# Global Constraint Catalog
## Volume II
## Time-Series Constraints

Ekaterina Arafailova, Nicolas Beldiceanu[1], Rémi Douence
Mines Nantes (CNRS/INRIA), Nantes, France

Mats Carlsson
SICS, Kista, Sweden

Pierre Flener, María Andreína Francisco Rodríguez, Justin Pearson
Uppsala University, Uppsala, Sweden

Helmut Simonis
Insight Centre for Data Analytics, University College Cork, Ireland

**Abstract:** First this report presents a restricted set of finite transducers used to synthesise structural time-series constraints described by means of a multi-layered function composition scheme. Second it provides the corresponding synthesised catalogue of structural time-series constraints where each constraint is explicitly described in terms of automata with accumulators.

September 24, 2016

[1]Corresponding author, Email: Nicolas.Beldiceanu@mines-nantes.fr

# Contents

# Preface

This second volume of the Global Constraint Catalogue [6] is devoted to time-series constraints. Within the context of Constraint Programming, time-series constraints go back to the work of Goldin and Kanellakis [10]. This volume contains 626 constraints, which are explicitly described in terms of automata with accumulators [3]. Checkers and propagators for all these constraints were synthesised [4] from 22 transducers [15, 8, 12, 14] described in the second chapter of this report.

As in the first volume, the global constraints described in this second volume are not only accessible to humans, who can read the catalogue when searching for some information. It is also available to machines, which can read and interpret it. This is why there also exists an electronic version of this catalogue where one can get, for all time-series constraints, a complete description in terms of meta-data used in the first volume. In fact, unlike the first volume, *all the meta-data* of the electronic version as well as *all text and figures* of this second volume were automatically generated. While this second volume is by no means supposed to contain all possible time-series constraints, it contributes in the context of time-series constraints to the *systematic reconstruction* of the Global Constraint Catalogue that we have previously advocated [7]. This reconstruction is based on the following methodology:

- First reuse, adapt or come up with abstractions, which allow to concisely represent structures and properties of time series as abstract combinatorial objects. In our context these abstractions essentially correspond to:

    1. Transducers where letters of the output alphabet are interpreted as semantic letters indicating how to recognise pattern occurrences.

    2. Transducers glue matrices expressing the relationship between the prefix, the suffix and the full sequence passed to a transducer.

    3. Properties associated to regular expressions corresponding to fragments of the input language of our transducers.

- Second, create from these abstract combinatorial objects a data base of concrete combinatorial objects.

i

- Third, synthesise concrete code for various technologies, languages, tasks from this data base of concrete combinatorial objects. In this context, correctness and efficiency [11] of the synthesised code are essentially side product of:

    - The correctness of the formulae of our data base which is itself based on the wellformedness of our abstractions.
    - The generality behind our abstract combinatorial objects.

The time-series catalogue is done in the following way:

- All time-series constraints are now defined in a *compositional way* from a few basic constituents, i.e., patterns, features, aggregators, and predicates, which completely define the meaning of a constraint, where patterns are defined using regular expressions.

- Constraint names are now constructed in a systematic way as the *concatenation* of pattern name, feature name, and aggregation or predicate name.

- Given a pattern $p$, checkers and constraints are now *systematically synthesised* from a transducer that, given an input sequence over the input alphabet $\{<, =, >\}$, compares two adjacent values of a time-series and determines an output sequence over a output semantic alphabet describing how to recognise the occurrences of $p$.

- For each time-series constraint associated with a pattern $p$, the generation of an automaton with accumulators is completely driven by the transducer associated with pattern $p$ as well as by *decoration tables* describing for each semantic letter of the output alphabet of the transducers how to generate accumulator updates. Code optimisation is ensured by using decoration tables that depend on properties of the pattern, of the feature, and of the aggregator associated with the time-series constraint.

- Lower and upper bounds of characteristics of time-series that appear in the restriction slot of a time-series constraint are synthesised from a *few parameterised formulae* that only depend on a restricted set of characteristics of the regular expression associated with the pattern.

- Parametrised glue matrices are provided for each transducer that corresponds to reversible time-series constraints. A concrete glue matrix is given for each reversible time-series constraint.

- Linear invariants are systematically obtained by applying the Farkas Lemma [9] to the automata with accumulators that were synthesised. They consist of *linear constraints typically linking consecutive accumulator values*, e.g., see the legend of the second automaton of the MAX_MAX_PEAK, MAX_RANGE_DECREASING, MAX_RANGE_INCREASING, MAX_WIDTH_STRICTLY_DECREASING_SEQUENCE, MAX_WIDTH_STRICTLY_INCREASING_SEQUENCE, MIN_MAX_PEAK, MIN_WIDTH_PLAIN, MIN_WIDTH_PLATEAU, NB_BUMP_ON_DECREASING_SEQUENCE, NB_DIP_ON_INCREASING_SEQUENCE, NB_GORGE, NB_PEAK, NB_SUMMIT, NB_VALLEY,

and ᴺᴮ_ᴢɪɢᴢᴀɢ constraints, which are generated even with non-linear accumulator updates. Missing linear invariants will be completed later on.

- Last but not least, time-series constraints were used for generating time-series verifying a conjunction of constraints both in the context of Constraint Programming [2] and in the context of Linear Programming. In this last case a linear reformulation is given in [1].

- In the context of sequential pattern mining, time-series constraint checkers can be used to identify and extract patterns from fixed sequences. While the time-series catalogue may need to be extended in order to capture more patterns, having a possibly large set of fixed time-series constraints is a natural safeguard to prevent overfitting when dealing with few sequences, at a price of not finding patterns that are not covered by the catalogue.

- Finally, both SICStus and MiniZinc code are synthesised. The later allows using time series constraints on many plate forms such as Choco, Gecode, ORtools, Cplex or Gurobi and is available from the Electronic Constraint Catalogue in Appendix A of this document.

The catalogue contains the following types of figures, which were all synthesised by computer programs producing Ti*k*Z [16] code:

- Figures representing the logo of each pattern.

- Figures for visualising the transducer associated with each pattern. The following scheme was consistently used over all transducers:

  – An arc labelled by the output symbol **out**, **maybe**, **found**, or **in** is respectively coloured in red, orange, blue, and violet.

  – If all input transitions of a state have the same colour, then the state uses this colour, otherwise it is coloured in black.

  – The name of a state is preceded by all input symbols of the transitions that enter this state. We use special characters for representing combinations of input symbols, e.g. $\leq$ for $<$ and $=$.

- Figures giving time series that respectively achieve lower and upper bounds of the `VALUE` argument of a time-series constraint.

- Figures illustrating each constraint on a relevant time series.

- Figures providing the synthesised automaton with accumulators associated with each constraint. Such automata use the same graphical scheme as the transducers from which they were generated.

In order to see PDF annotations and attached files, you are advised to use **Adobe Reader** with the pdf version of the catalogue.[1] If you do not see on your screen a small yellow bullet at the beginning of this paragraph, you are using a PDF viewer that does not fully support PDF annotations.

---

[1] Since we are using the LᴬTEX packages pdfcomment and attachfile and since most PDF viewers do not support PDF annotations or attachments.

To get started the reader should first read the introduction of the overview chapter 2, which presents the notions of pattern, seed transducer and glue matrix used throughout this document, as well as the conventions used for drawing time-series.

The initial work on synthesising automata with accumulators from transducers was done by N. Beldiceanu, R. Douence, and H. Simonis. The generation of the catalogue was done by H. Simonis except, for the examples and figures, which were handled by N. Beldiceanu. The code generation part dealing with constraint checkers was rewritten by M. Carlsson for getting optimised checkers that do not use any constraint. The creation of dedicated decoration tables that depend on properties of patterns, features, and aggregators was done by E. Arafailova under the supervision of N. Beldiceanu and R. Douence [2]. The bounds were done by E. Arafailova under the supervision of N. Beldiceanu; the generation of the corresponding code was done by H. Simonis. The generation of linear invariants [13, 2] and the parametrised glue matrices [5, 1] were done by M. A. Francisco Rodríguez under the supervision of P. Flener and J. Pearson.

The authors have benefit from the following support:

- E. Arafailova and R. Douence were supported by the European H2020 FETPROACT-2014 project "GRACeFUL" (project number 640954).

- From 2014 to 2016 N. Beldiceanu was partially supported by the Gaspard Monge Program for Optimization and Operations Research (PGMO), as well as by the European H2020 FETPROACT-2014 project "GRACeFUL".

- M. Carlsson was supported by SICS.

- P. Flener, J. Pearson and M. A. Francisco Rodríguez were supported by grants 2011-6133 and 2012-4908 of the Swedish Research Council (VR).

- During 2014 H. Simonis was partially supported by a senior researcher chair by the Région Pays de la Loire and by the EU FET grant ICON (project number 284715). From 2014 to 2016 H. Simonis was also supported by Science Foundation Ireland under Grant Number SFI/10/IN.1/I3032. The INSIGHT Centre for Data Analytics is supported by Science Foundation Ireland under Grant Number SFI/12/RC/2289.

Readers may send their suggestions via email to the corresponding author with `catalogue` as subject.

*Cork, Irland,    Nantes, France,    Uppsala, Sweden,    September 2016*
— EA, NB, RD, MC, PF, MAFR, JP, HS

# 1

# Describing a global constraint

## Contents

This chapter, taken from the first volume of the global constraint catalogue [6], recalls how to describe a global constraint and its arguments.

Since global constraints have to receive their arguments in some form, no matter whether we use the graph-based or automaton-based description, we start by describing the abstract data types that we use in order to specify the arguments of a global constraint. These abstract data types are not related to any specific programming language like Caml, C, C++, Java, or Prolog. If one wants to focus on a specific language, then one has to map these abstract data types to the data types that are available within the considered language. Second we describe all the restrictions that one can impose on the arguments of a global constraint. Finally, we show how to use these ingredients in order to declare the arguments of a global constraint.

## 1.1 Basic data types

We use the following *basic data types*:

- `atom` corresponds to an atom. Predefined atoms are `MININT` and `MAXINT`, which respectively correspond to the *smallest integer* and *largest integer*.

- `int` corresponds to an *integer value*.

- `dvar` corresponds to a *domain variable*. A *domain variable V* is a variable that will be assigned an *integer* value taken from an initial finite set of integer values denoted by $\text{dom}(V)$. $\underline{V}$ and $\overline{V}$ respectively denote the minimum and maximum values of $\text{dom}(V)$.

- `fdvar` corresponds to a *possibly unbounded domain variable*. A *possibly unbounded domain variable* is a variable that will be assigned an *integer* value from an initial finite set of integer values denoted by $\text{dom}(V)$ or from $]-\infty, +\infty[$.

- `sint` corresponds to a *finite set of integer values*.

- `svar` corresponds to a *set variable*. A *set variable* $V$ is a variable that will be assigned to a *finite set* of integer values. Its *lower bound* $\underline{V}$ denotes the set of integer values that for sure belong to $V$, while its *upper bound* $\overline{V}$ denotes the set of integer values that may belong to $V$. Let $\text{dom}(V) = \{\mathbf{v_1}, \ldots, \mathbf{v_n}, v_{n+1}, \ldots, v_m\}$ be a shortcut for combining the lower and upper bounds of $V$ in a single notation:

    - **Bold** values designate those values that only belong to $\underline{V}$.
    - *Italic* values indicate those values that belong only to $\overline{V}$.

- `mint` corresponds to a *multiset of integer values*.

- `mvar` corresponds to a *multiset variable*. A *multiset variable* is a variable that will be assigned to a *multiset of integer values*.

- `real` corresponds to a *real number*.

- `rvar` corresponds to a *real number variable*. A *real number variable* is a variable that will be assigned a *real number* taken from an initial finite set of intervals. A real number is usually represented by an interval of two floating-point numbers.

## 1.2  Compound data types

We use the following *compound data types*:

- `list(T)` corresponds to a list of elements of type $T$, where $T$ is a basic or compound data type.

- `collection(A_1, A_2, \ldots, A_n)` corresponds to a collection of ordered items, where each item consists of $n > 0$ attributes $A_1, A_2, \ldots, A_n$. Each attribute is an expression of the form $\mathbf{a} - T$, where $\mathbf{a}$ is the *name* of the attribute and $T$ the *type* of the attribute (a basic or compound data type). All names of the attributes of a given collection should be distinct and different from the keyword `key`, which corresponds to an implicit[1] attribute. The value of the `key` attribute is the position of an item within the collection. The first item of a collection is associated with position 1.

---

[1]This attribute is not explicitly defined.

The following notation is used for instantiated arguments:

- A *list of elements* $e_1, e_2, \ldots, e_n$ is denoted by $[e_1, e_2, \ldots, e_n]$.

- A *finite set of integers* $i_1, i_2, \ldots, i_n$ is denoted by $\{i_1, i_2, \ldots, i_n\}$.

- A *collection of $n$ items*, each item having $m$ attributes, is denoted by
  $\langle \mathtt{a}_1 - v_{11} \ldots \mathtt{a}_m - v_{1m}, \ \mathtt{a}_1 - v_{21} \ldots \mathtt{a}_m - v_{2m}, \ \ldots \ , \mathtt{a}_1 - v_{n1} \ldots \mathtt{a}_m - v_{nm} \rangle$.
  Each item is separated from the previous item by a comma. If the items of the
  collection involve a single attribute $\mathtt{a}_1$, then $\langle v_{11}, v_{21}, \ldots, v_{n1} \rangle$ can be used as a
  shortcut for $\langle \mathtt{a}_1 - v_{11}, \ \mathtt{a}_1 - v_{21}, \ \ldots \ , \mathtt{a}_1 - v_{n1} \rangle$.

- The $i^{\text{th}}$ *item of a collection* $\mathtt{c}$ is denoted by $\mathtt{c}[i]$.

- The *value of the attribute* $\mathtt{a}$ *of the $i^{\text{th}}$ item of a collection* $\mathtt{c}$ is denoted by $\mathtt{c}[i].\mathtt{a}$.
  Note that, within an arithmetic expression, we can use the shortcut $\mathtt{c}[i]$ when
  the collection $\mathtt{c}$ involves a single attribute.

- The *number of items* of a collection $\mathtt{c}$ is denoted by $|\mathtt{c}|$.

## 1.3 Restrictions

When defining the arguments of a global constraint, it is often the case that one needs
to express additional conditions that refine the type declarations of its arguments. For
this purpose we provide *restrictions* that allow the specification of these additional
conditions.
Currently the list of restrictions is:

- `in_list(Arg, ListAtoms)`

    - `Arg` is an argument of type `atom`,
    - `ListAtoms` is a non-empty list of distinct atoms.

  This restriction enforces `Arg` to be one of the atoms specified in `ListAtoms`.

- `in_list(Arg, Attr, ListIntOrAtom)`

    - `Arg` is an argument of type `collection`,
    - `Attr` is an attribute of type `int` or `atom` of `Arg`,
    - If `Attr` is an attribute of type `int`, then `ListIntOrAtom` is a non-empty
      list of distinct integers; if `Attr` is an attribute of type `atom`, then
      `ListIntOrAtom` is a non-empty list of distinct atoms.

  This restriction enforces, for all items of `Arg`, the attribute `Attr` to take its value
  within `ListIntOrAtom`.

- `in_attr(Arg1, Attr1, Arg2, Attr2)`

    - `Arg1` is an argument of type `collection`,

- **Attr1** is an attribute of type `dvar` or `int` of `Arg1`,
- **Arg2** is an argument of type `collection`,
- **Attr2** is an attribute of type `int` of `Arg2`.

Let $\mathcal{S}_2$ denote the set of values assigned to the `Attr2` attributes of the items of `Arg2`. This restriction enforces the following condition: for all items of `Arg1`, the attribute `Attr1` takes its value in the set $\mathcal{S}_2$.

- **`distinct(Arg, Attrs)`**

  - **Arg** is an argument of type `collection`,
  - **Attrs** is an attribute of type `int` or `dvar`, or a (possibly empty) list of distinct attributes of type `int` or `dvar` of `Arg`.

  For all pairs of distinct items of `Arg` this restriction enforces that there be at least one attribute specified by `Attrs` with two distinct values. If `Attrs` is the empty list, then all items of `Arg` should be distinct.

- **`increasing_seq(Arg, Attrs)`**

  - **Arg** is an argument of type `collection`,
  - **Attrs** is an attribute of type `int` or a list of distinct attributes of type `int` of `Arg`.

  Let $n$ and $m$ respectively denote the number of items of `Arg`, and the number of attributes of `Attrs`. For item $i$ of `Arg` let $t_i$ denote the tuple of values $\langle v_{i,1}, v_{i,2}, \ldots, v_{i,m} \rangle$ where $v_{i,j}$ is the value of attribute $j$ of `Attrs` of item $i$ of `Arg`. The restriction enforces a strict lexicographical ordering on the tuples $t_1, t_2, \ldots, t_n$.

- **`non_increasing_size(Arg, Attr)`**

  - **Arg** is an argument of type `collection`,
  - **Attr** is an attribute of type `collection` of `Arg`.

  This restriction enforces for each pair of consecutive items `Arg[i]`, `Arg[i + 1]` that the number of items of `Arg[i].Attr` is greater than or equal to the number of items of the collection `Arg[i + 1].Attr`.

- **`required(Arg, Attrs)`**

  - **Arg** is an argument of type `collection`,
  - **Attrs** is an attribute or a list of distinct attributes of `Arg`.

  This restriction enforces that all attributes denoted by `Attrs` be explicitly used within all items of `Arg`.

  This restriction is usually systematically used for every attribute of a collection. It is not used when some attributes may be implicitly defined according to other attributes. In this context, we use the `require_at_least` restriction, which we now introduce.

- `require_at_least(Atleast, Arg, Attrs)`

    - `Atleast` is a positive integer,

    - `Arg` is an argument of type `collection`,

    - `Attrs` is a non-empty list of distinct attributes of `Arg`. The length of this list should be strictly greater than `Atleast`.

    This restriction enforces that at least `Atleast` attributes of the list `Attrs` be explicitly used within all items of `Arg`.

- `same_size(Arg, Attr)`

    - `Arg` is an argument of type `collection`,

    - `Attr` is an attribute of type collection of `Arg`.

    This restriction enforces that all collections of `Attr` have the same number of items.

- `Term₁ Comparison Term₂`

    - $\texttt{Term}_1$ is a *term*. A *term* is an expression that can be evaluated to one or possibly several integer values.

    - `Comparison` is one of the comparison operators $\leq, \geq, <, >, =, \neq$.

    - $\texttt{Term}_2$ is a *term*.

    Let $v_{1,1}, v_{1,2}, \ldots, v_{1,n_1}$ and $v_{2,1}, v_{2,2}, \ldots, v_{2,n_2}$ be the values respectively associated with $\texttt{Term}_1$ and $\texttt{Term}_2$. The restriction $\texttt{Term}_1$ `Comparison` $\texttt{Term}_2$ enforces $v_{1,i}$ `Comparison` $v_{2,j}$ to hold for every $i \in [1, n_1]$ and every $j \in [1, n_2]$.

    A *term* is one of the following expressions:

    - $e$ , where $e$ is an integer. The corresponding value is $e$.

    - $|c|$ , where $c$ is an argument of type `collection`. The value of $|c|$ is the number of items of the collection $c$.

    - $\textbf{first}(c.a)$ : $\texttt{first}(c.a)$ denotes the value assigned to the attribute $a$ of the first item of the collection $c$. It is equal to $0$ if the collection is empty.

    - $\textbf{last}(c.a)$ : $\texttt{last}(c.a)$ denotes the value assigned to the attribute $a$ of the last item of the collection $c$. It is equal to $0$ if the collection is empty.

    - $\textbf{sum}(c.a)$ : $\texttt{sum}(c.a)$ denotes the sum of the values assigned to the attribute $a$ of the collection $c$; it is equal to $0$ if the collection is empty.

    - $\textbf{sum}(\ell)$ : $\texttt{sum}(\ell)$, where $\ell$ is a list of collection attributes, each of the form $c_i.a_i$ (with $i \in [1, n]$), is the sum of the values assigned to the attributes $a_i$ of the collections $c_i$ (with $i \in [1, n]$).

– `range(c.a)` : $\mathtt{range}(c.a)$ denotes the difference between the maximum value and minimum value plus one of the values assigned to the attribute $a$ of $c$; it is equal to 0 if the collection is empty.

– `range(ℓ)` : $\mathtt{range}(\ell)$, where $\ell$ is a list of collection attributes, each of the form $c_i.a_i$ (with $i \in [1, n]$), is the difference between the maximum value and minimum value plus one of the values assigned to the attributes $a_i$ of the $c_i$ (with $i \in [1, n]$).

– `minval(c.a)` : $\mathtt{minval}(c.a)$ denotes the minimum of the values assigned to the attribute $a$ of the collection denoted by $c$: it is equal to 0 if the collection is empty.

– `minval(ℓ)` : $\mathtt{minval}(\ell)$, where $\ell$ is a list of collection attributes, each of the form $c_i.a_i$ (with $i \in [1, n]$), is the minimum of the values assigned to the attributes $a_i$ of the $c_i$ (with $i \in [1, n]$).

– `maxval(c.a)` : $\mathtt{maxval}(c.a)$ denotes the maximum of the values assigned to the attribute $a$ of $c$; it is equal to 0 if the collection is empty.

– `maxval(ℓ)` : $\mathtt{maxval}(\ell)$, where $\ell$ is a list of collection attributes, each of the form $c_i.a_i$ (with $i \in [1, n]$), is the maximum of the values assigned to the attributes $a_i$ of the $c_i$ (with $i \in [1, n]$).

– `nval(c.a)` : $\mathtt{nval}(c.a)$ denotes the number of distinct values over the values assigned to the attribute $a$ of $c$, it is equal to 0 if the collection is empty.

– `nval(ℓ)` : $\mathtt{nval}(\ell)$, where $\ell$ is a list of collection attributes, each of the form $c_i.a_i$ (with $i \in [1, n]$), is the number of distinct values over the values assigned to the attributes $a_i$ of the $c_i$ (with $i \in [1, n]$).

– `prod(c.a)` : $\mathtt{prod}(c.a)$ denotes the product of the values assigned to the attribute $a$ of $c$, it is equal to 1 if the collection is empty.

– `prod(ℓ)` : $\mathtt{prod}(\ell)$, where $\ell$ is a list of collection attributes, each of the form $c_i.a_i$ (with $i \in [1, n]$), is the product of the values assigned to the attributes $a_i$ of the $c_i$ (with $i \in [1, n]$).

– `pfdc(c.a)` : $\mathtt{pfdc}(c.a)$, where $\mathtt{pfdc}$ is a pure functional dependency constraint of the form $\mathtt{pfdc}(v, col)$ that computes a value $v$ from a collection of variables $col$, and where $c.a$ is a collection with attribute $a$ denotes the $\mathtt{pfdc}$ of the values assigned to the attribute $a$ of $c$; it is equal to 0 if the collection $c$ is empty.

– `pfdc(ℓ)` : $\mathtt{pfdc}(\ell)$, where $\mathtt{pfdc}$ is a pure functional dependency constraint of the form $\mathtt{pfdc}(v, col)$ that computes a value $v$ from a collection of variables $col$, and where $\ell$ is a list of collection attributes, each of the form $c_i.a_i$ (with $i \in [1, n]$), is the $\mathtt{pfdc}$ of the values assigned to the attributes $a_i$ of the $c_i$ (with $i \in [1, n]$).

- $t$ , where $t$ is an argument of type `int`. The value of $t$ is the value of the corresponding argument.

- $v$ , where $v$ is an argument of type `dvar`. The value of $v$ is the value assigned to variable $v$. Note that restrictions are defined on the ground instance of a global constraint.

- $s$ , where $s$ is an argument of type `sint` or `svar`. The values denoted by $s$ are all the values of the corresponding set.

- $c.a$ , where $c$ is an argument of type `collection` and $a$ an attribute of $c$ of type `int` or `dvar`. The values denoted by $c.a$ are all the values corresponding to attribute $a$ for the items of $c$. When $c.a$ designates a domain variable we consider the value assigned to that variable.

- $c.a$ , where $c$ is an argument of type `collection` and $a$ an attribute of $c$ of type `sint` or `svar`. The values denoted by $c.a$ are all the values belonging to the sets corresponding to attribute $a$ for the items of $c$. When $c.a$ designates a set variable we consider the values that belong to that set.

- $t_1 \ \mathtt{op} \ t_2$ , where $t_1$ and $t_2$ are *terms* and op one of the operators $+$, $-$, $*$, $/$ or $\mathrm{mod}$.[2] Let $\mathcal{V}_1$ and $\mathcal{V}_2$ denote the sets of values respectively associated with the terms $t_1$ and $t_2$. The set of values associated with $t_1 \ \mathtt{op} \ t_2$ is $\mathcal{V}_{12} = \{v : v = v_1 \ \mathtt{op} \ v_2, v_1 \in \mathcal{V}_1, v_2 \in \mathcal{V}_2\}$.

- $|t|$ , where $t$ is a *term*. Let $\mathcal{V}$ denote the set of values associated with the term $t$. The set of values associated with $|t|$ is $\mathcal{V} = \{v : v = |val|, val \in \mathcal{V}\}$.

- $\min_{v \in [\ell, u]}(t)$ , where $v$ is a variable that occurs in the term $t$, and $\ell$, $u$ are two terms. The value of $\min_{v \in [\ell, u]}(t)$ is the smallest value among all possible values of the term $t$ where $v$ is assigned a value in the interval $[\ell, u]$.

- $\max_{v \in [\ell, u]}(t)$ , where $v$ is a variable that occurs in the term $t$, and $\ell$, $u$ are two terms. The value of $\max_{v \in [\ell, u]}(t)$ is the largest value among all possible values of the term $t$ where $v$ is assigned a value in the interval $[\ell, u]$.

- Using the $\vee$ connector we can express a disjunction between two restrictions .

- Using the $\Rightarrow$ connector we can express an implication between a restriction (or a conjunction of restrictions) and a restriction.

- We can also use a constraint $C$ of the catalogue for expressing a restriction as long as that constraint is not defined according to the constraint under consideration. The constraint $C$ should have a graph-based or automaton-based description so that its meaning is explicitly defined.

Within the restriction slot it is possible to define local variables to designate intermediate expressions. This is done by using the `where` statement after the set of restrictions.

---

[2] $/$ denotes a *floor division*, that is a division in which the result is rounded to the nearest integer that is smaller or equal.

## 1.4   Declaring a global constraint

Declaring a global constraint consists of providing the following information:

- A *term* `constraint`$(A_1, A_2, \ldots, A_n)$ , where `constraint` corresponds to the *name* of the global constraint and $A_1, A_2, \ldots, A_n$ to its *arguments*.

- A possibly empty *list of type declarations* , where each declaration has the form `type:type_declaration`; `type` is the *name* of the new type we define and `type_declaration` is a basic data type, a compound data type, or a previously defined type.

- An *argument declaration* $A_1{:}T_1, A_2{:}T_2, \ldots, A_n{:}T_n$ giving for each argument $A_1, A_2, \ldots, A_n$ of the global constraint `constraint` its type. Each type is a basic data type, a compound data type, or a type that was declared in the list of type declarations.

- A possibly empty *list of restrictions* , where each restriction is one of the restrictions described in Section 1.3.

## 1.5   Keywords

This section lists the keywords used in the time-series catalogue.

- **added**: Denotes the fact that, even if the same constant is added to all variables of the `VARIABLES` collection, the corresponding constraint still holds.

- **functional dependency**: A constraint that allows for representing a functional dependency between possibly several domain variables and a single domain variable. A sequence of variables $X_1, X_2, \ldots, X_n$ is said to functionally determine another variable $Y$ if and only if each potential tuple of values of $X_1, X_2, \ldots, X_n$ is associated with exactly one potential value of $Y$ (i.e., $Y$ is a function of $X_1, X_2, \ldots, X_n$).

- **reversed**: Denotes the fact that, even if the variables of the `VARIABLES` collection are reversed, the corresponding constraint still holds.

- **reverse of a constraint**: A constraint which has a reverse constraint, where the reverse is defined in the following way. Consider two constraints $c(col, r_1, \ldots, r_n)$ and $c'(col, r_1, \ldots, r_n)$ for which, in both cases, the argument *col* is a collection of items that functionally determines all the other arguments $r_1, \ldots, r_n$.

  The constraint $c'$ is the *reverse constraint* of constraint $c$ if, for any collection of items *col*, we have the equivalence $c(col, r_1, \ldots, r_n) \Leftrightarrow c'(col^{rev}, r_1, \ldots, r_n)$, where $col^{rev}$ denotes the collection *col* where the items of the collection are reversed. When constraints $c$ and $c'$ are identical we say that constraint $c$ is its own reverse.

# 2

# Patterns, Seed Transducers, Glue matrices and Decoration Tables

## Contents

Patterns focus on the *topological* aspect of subsequences of a time series. They are defined by two components:

- First a regular expression over the alphabet $\{<, =, >\}$.

- Second two non-negative integers $b$ and $a$ that are intended to respectively delete a prefix and a suffix of the pattern that should discarded for computing a characteristic of an occurrence of pattern.

Given a pattern and a time series $x_0, x_1, \ldots, x_{n-1}$ of integer constants, called the *input values* and forming the *input sequence*, a single integer is computed as follows:

I. Compare each pair of adjacent input values in order to build a sequence $s_0, s_1, \ldots, s_{n-2}$ of *signature values* over the alphabet $\{<, =, >\}$, as follows: $(x_i < x_{i+1} \Leftrightarrow s_i = \ '<') \land (x_i = x_{i+1} \Leftrightarrow s_i = \ '=') \land (x_i > x_{i+1} \Leftrightarrow s_i = \ '>')$. The signature values form the *signature sequence*.

II. Within the signature sequence, find *all* maximal words matching the regular expression associated with the pattern.

III. For each found pattern occurrence discard its prefix and suffix of length $b$ and $a$ to obtain an integer sequence $e$ for which we compute an integer *feature value*, so that we obtain a *feature sequence*. The features we currently consider are `one`, `width`, `surface`, `min`, `max`, and `range`, and correspond respectively to the value 1, to the number of elements of $e$, to $\sum_{i \in e} x_i$, to $\min_{i \in e} x_i$, to $\max_{i \in e} x_i$, and to $\max_{i \in e} x_i - \min_{i \in e} x_i$.

IV. Aggregate the values of the feature sequence into a single integer value. The aggregators we currently consider are taking the sum (`Sum`), taking the minimum (`Min`), and taking the maximum (`Max`). The feature `one` only makes sense with the `Sum` aggregator.

**Definition 1** (s-occurrence, i-occurrence, e-occurrence)**.** *Given an input sequence* $x_0, x_1, \ldots, x_{n-1}$, *its signature sequence* $S = s_0, s_1, \ldots, s_{n-2}$, *a pattern* $\langle r, b, a \rangle$, *and a non-empty signature subsequence* $s_i, s_{i+1}, \ldots, s_j$, *with* $0 \leq i \leq j \leq n-2$, *forming a maximal word that matches* $r$, *the* s-occurrence $(i..j)$ *is the index sequence* $i, i+1, \ldots, j$; *the* i-occurrence $[(i+b)..j]$ *is the index sequence* $i+b, \ldots, j$; *and the* e-occurrence $[[(i+b)..(j+1-a)]]$ *is the index sequence* $i+b, \ldots, j+1-a$.

Figures representing time-series use the following convention for denoting i-occurrences, s-occurrences and e-occurrences:

- Positions belonging to an i-occurrence are represented by a red circle •.

- Positions belonging to a s-occurrence, but not to an i-occurrence, are represented by a circle ○.

- Positions of an occurrence of pattern corresponding to an e-occurrence have a background coloured in yellow.

When an i-occurrence and an e-occurrence of two consecutive occurrences of a pattern coincide a red circle ● is used.

The name of a constraint is defined by the concatenation of the aggregator function, the feature name and the pattern. For each pattern a so-called *seed transducer* recognises all pattern occurrences.

**Definition 2** (seed transducer). *A seed transducer is a transducer for which all states are accepting with input alphabet* $\{<, =, >\}$, *output alphabet, also-called semantic alphabet,* $\{\mathbf{out}, \mathbf{maybe_b}, \mathbf{out_r}, \mathbf{found}, \mathbf{maybe_b}, \mathbf{in}, \mathbf{out_a}, \mathbf{found_e}\}$.

**Definition 3** (t-occurrence). *Given an input sequence $s$ over the input alphabet* $\{<, =, >\}$ *the* t-occurrence *of $s$ wrt a given seed transducer corresponds to the indices of the semantic letters of a maximum word within the generated output sequence that matches* '$\mathbf{maybe_b^*found_e} \mid \mathbf{maybe_b^*found(maybe_a^*in^+)^*}$'.

**Definition 4** (wellformedness). *A seed transducer is* well formed *if (1) all output sequences it produces are recognised by the finite deterministic automaton depicted by Figure 2.1, and if (2) the t-occurrence for any input sequence $s$ coincides with the i-occurrence of the pattern for $s$.*

Wellformedness of seed transducers guarantees that the automata with accumulators obtained by applying the decoration tables for replacing semantic letters of the output alphabet by accumulator updates compute the expected result [4].

Figure 2.1: Automaton describing the output language of a well-formed transducer; states $\notin$, ?, $\in$ respectively mean that (1) we are outside an occurrence of pattern, (2) we are potentially inside an occurrence of pattern, (3) we are inside an occurrence of pattern that is not yet finished.

Following the presentation in [5, 1], we introduce the notion of glue matrix. A *glue matrix* for a reversible time-series constraint $c$ specified by a seed transducer $\sigma$, an aggregator $g$ and a feature $f$ is a matrix indexed by the states of $\sigma$ and $\sigma^{\mathrm{rev}}$ and parametrised by $g$ and $f$, where $\sigma^{\mathrm{rev}}$ is a seed transducer of the reverse of constraint *c.* Consider the automaton $\mathcal{A}_\sigma$ (respectively $\mathcal{A}_\sigma^{\mathrm{rev}}$) with accumulators $R$, $C$ and $D$ obtained by applying the decoration table 2.35 to the seed transducer $\sigma$ (respectively $\sigma^{\mathrm{rev}}$). Let $\mathcal{A}_\sigma$ reach state $\overrightarrow{Q}$ and accumulator values $\langle \overrightarrow{C}, \overrightarrow{D}, \overrightarrow{R} \rangle$ on a prefix of a word $w$. Similarly let $\mathcal{A}_\sigma^{\mathrm{rev}}$ reach state $\overleftarrow{Q}$ and accumulator values $\langle \overleftarrow{C}, \overleftarrow{D}, \overleftarrow{R} \rangle$ on the reverse of the corresponding suffix of $w$. The value returned by $\mathcal{A}_\sigma$ on the entire word $w$ is $\phi_g(\overrightarrow{R}, \overleftarrow{R}, \Gamma)$, where $\Gamma$ is the entry of the glue matrix corresponding to row $\overrightarrow{Q}$ and column $\overleftarrow{Q}$. That entry contains an expression parametrised by $g$ and $f$ involving a subset of the accumulators $\overrightarrow{C}, \overrightarrow{D}, \overleftarrow{C}, \overleftarrow{D}$.

## 2.1 Patterns, their Seed Transducers and their Parametrised Glue Matrices

| Feature $f$ | $\mathrm{id}_f$ | $\min_f$ | $\max_f$ | $\phi_f$ | $\delta_f^i$ |
|---|---|---|---|---|---|
| `one` | 1 | 1 | 1 | 1 | 1 |
| `width` | 0 | $n+1$ | $+\infty$ | $+$ | 1 |
| `surface` | 0 | $-\infty$ | $+\infty$ | $+$ | $x_i$ |
| `max` | $-\infty$ | $-\infty$ | $+\infty$ | max | $x_i$ |
| `min` | $+\infty$ | $-\infty$ | $+\infty$ | min | $x_i$ |
| `range` | 0 | 0 | $+\infty$ | n/a | $x_i$ |

| Aggregator $g$ | $\phi_g$ | $\mathrm{default}_{gf}$ |
|---|---|---|
| `Max` | max | $\min_f$ |
| `Min` | min | $\max_f$ |
| `Sum` | $+$ | 0 |

Table 2.1: (Left) Features: identity, minimum, and maximum values; the operators $\phi_f$ and $\delta_f^i$ recursively define the feature value $v_u$ of a time series $x_\ell, \ldots, x_u$ by $v_\ell = \phi_f(\mathrm{id}_f, \delta_f^\ell)$ and $v_i = \phi_f(v_{i-1}, \delta_f^i)$ for $i > \ell$, where $\delta_f^i$ is the contribution of $x_i$ to $v_u$; $n$ stands for the length of the time-series. (Right) Aggregators: operators and identity values relative a feature $f$.

| Predicate $\theta$ | $\texttt{default}_\theta$ |
|---|---|
| $\leq$ | $-\infty$ |
| $=$ | $x$ (free variable) |
| $\geq$ | $+\infty$ |

Table 2.2: Default value for a predicate $\theta$ used in the decoration table 2.36

### 2.1.1 BUMP_ON_DECREASING_SEQUENCE

Given a sequence $S$ over the alphabet $\{<, =, >\}$, an occurrence of the pattern BUMP_ON_DECREASING_SEQUENCE is the subsequence of $S$ which matches the regular expression $>><>>$. Part (A) and parts (B,C) of Figure 2.2 respectively depict the seed transducer associated with the BUMP_ON_DECREASING_SEQUENCE pattern as well as one example of its execution on a time series.

Figure 2.2: (A) Seed transducer for the BUMP_ON_DECREASING_SEQUENCE pattern: $>><>$ with $b = 2$ and $a = 1$; (B) Illustrating the execution of the seed transducer on a time series (within the output $o$, $o_r$, $m_b$, $f_e$ are shortcut for **out**, **out**$_r$, **maybe**$_b$, **found**$_e$); (C) Illustrating the relation between the $s$-occurrence, the $i$-occurrence and the $e$-occurrence.

Figure 2.3: Parametrised automaton for any functional time-series constraints of the BUMP_ON_DECREASING_SEQUENCE pattern obtained by applying the decoration table 2.35 to the corresponding transducer

Figure 2.4: Parametrised automaton for any predicate time-series constraints of the BUMP_ON_DECREASING_SEQUENCE pattern obtained by applying the decoration table 2.36 to the corresponding transducer, where $\theta$ is the predicate (i.e., $\leq$, $=$, $\geq$) used for comparing two consecutive feature values

### 2.1.2 DECREASING

Given a sequence $S$ over the alphabet $\{<, =, >\}$, an occurrence of the pattern DE-CREASING is the subsequence of $S$ which matches the regular expression $>$. Part (A) and parts (B,C) of Figure 2.5 respectively depict the seed transducer associated with the DECREASING pattern as well as one example of its execution on a time series.

Figure 2.5: (A) Seed transducer for the DECREASING pattern: $>$ with $b = 0$ and $a = 0$; (B) Illustrating the execution of the seed transducer on a time series (within the output $\mathbf{o}$, $\mathbf{f_e}$ are shortcut for $\mathbf{out}$, $\mathbf{found_e}$); (C) Illustrating the relation between the $s$-occurrence, the $i$-occurrence and the $e$-occurrence.

| | $s$ |
|---|---|
| $s$ | $\phi_g(\overrightarrow{C}, \overleftarrow{C})$ |

Table 2.3: Parametrised glue matrix for any $g\_f\_$DECREASING constraint

Figure 2.6: Parametrised automaton for any functional time-series constraints of the DECREASING pattern obtained by applying the decoration table 2.35 to the corresponding transducer

Figure 2.7: Parametrised automaton for any predicate time-series constraints of the DECREASING pattern obtained by applying the decoration table 2.36 to the corresponding transducer, where $\theta$ is the predicate (i.e., $\leq$, $=$, $\geq$) used for comparing two consecutive feature values

### 2.1.3 DECREASING_SEQUENCE

Given a sequence $S$ over the alphabet $\{<, =, >\}$, an occurrence of the pattern DE-CREASING_SEQUENCE is the *maximal* subsequence of $S$ which matches the regular expression $> (> | =)^* > | >$. Part (A) and parts (B,C) of Figure 2.8 respectively depict the seed transducer associated with the DECREASING_SEQUENCE pattern as well as one example of its execution on a time series.

Figure 2.8: (A) Seed transducer for the DECREASING_SEQUENCE pattern: $> (> | =)^* > | >$ with $b = 0$ and $a = 0$; (B) Illustrating the execution of the seed transducer on a time series (within the output $\mathbf{o}$, $\mathbf{f}\ \mathbf{m}_a$, $\mathbf{o}_a$ are shortcut for $\mathbf{out}$, $\mathbf{found}\ \mathbf{maybe}_a$, $\mathbf{out}_a$); (C) Illustrating the relation between the *s*-occurrence, the *i*-occurrence and the *e*-occurrence.

Figure 2.9: Parametrised automaton for any functional time-series constraints of the DECREASING_SEQUENCE pattern obtained by applying the decoration table 2.35 to the corresponding transducer

Figure 2.10: Parametrised automaton for any predicate time-series constraints of the DECREASING_SEQUENCE pattern obtained by applying the decoration table 2.36 to the corresponding transducer, where $\theta$ is the predicate (i.e., $\leq$, $=$, $\geq$) used for comparing two consecutive feature values

|     | $s$ | $t$ |
|-----|-----|-----|
| $s$ | $\phi_g(\overrightarrow{C}, \overleftarrow{C})$ | $\phi_g(\overrightarrow{C}, \overleftarrow{C})$ |
| $t$ | $\phi_g(\overrightarrow{C}, \overleftarrow{C})$ | $\phi_f(\overrightarrow{C}, \overleftarrow{C}, \overrightarrow{D}, \overleftarrow{D}, \delta_f^i)$ |

Table 2.4: Parametrised glue matrix for any $g\_f\_$DECREASING_SEQUENCE constraint

### 2.1.4 DECREASING_TERRACE

Given a sequence $S$ over the alphabet $\{<, =, >\}$, an occurrence of the pattern DE-CREASING_TERRACE is the *maximal* subsequence of $S$ which matches the regular expression $> =^+ >$. Part (A) and parts (B,C) of Figure 2.11 respectively depict the seed transducer associated with the DECREASING_TERRACE pattern as well as one example of its execution on a time series.

Figure 2.11: (A) Seed transducer for the DECREASING_TERRACE pattern: $> =^+ >$ with $b = 1$ and $a = 1$; (B) Illustrating the execution of the seed transducer on a time series (within the output **o**, **m**$_b$, **o**$_r$, **f**$_e$ are shortcut for **out**, **maybe**$_b$, **out**$_r$, **found**$_e$); (C) Illustrating the relation between the *s*-occurrence, the *i*-occurrence and the *e*-occurrence.

Figure 2.12: Parametrised automaton for any functional time-series constraints of the DECREASING_TERRACE pattern obtained by applying the decoration table 2.35 to the corresponding transducer

Figure 2.13: Parametrised automaton for any predicate time-series constraints of the DECREASING_TERRACE pattern obtained by applying the decoration table 2.36 to the corresponding transducer, where $\theta$ is the predicate (i.e., $\leq$, $=$, $\geq$) used for comparing two consecutive feature values

|   | $s$ | $r$ | $t$ |
|---|---|---|---|
| $s$ | $\phi_g(\overrightarrow{C}, \overleftarrow{C})$ | $\phi_g(\overrightarrow{C}, \overleftarrow{C})$ | $\phi_g(\overrightarrow{C}, \overleftarrow{C})$ |
| $r$ | $\phi_g(\overrightarrow{C}, \overleftarrow{C})$ | $\phi_g(\overrightarrow{C}, \overleftarrow{C})$ | $\phi_f(\overrightarrow{D}, \overleftarrow{D}, \delta_f^i)$ |
| $t$ | $\phi_g(\overrightarrow{C}, \overleftarrow{C})$ | $\phi_f(\overrightarrow{D}, \overleftarrow{D}, \delta_f^i)$ | $\phi_f(\overrightarrow{D}, \overleftarrow{D}, \delta_f^i)$ |

Table 2.5: Parametrised glue matrix for any $g\_f\_$DECREASING_TERRACE constraint

### 2.1.5 DIP_ON_INCREASING_SEQUENCE

Given a sequence $S$ over the alphabet $\{<, =, >\}$, an occurrence of the pattern DIP_ON_INCREASING_SEQUENCE is the subsequence of $S$ which matches the regular expression $<<><<$. Part (A) and parts (B,C) of Figure 2.14 respectively depict the seed transducer associated with the DIP_ON_INCREASING_SEQUENCE pattern as well as one example of its execution on a time series.

Figure 2.14: (A) Seed transducer for the DIP_ON_INCREASING_SEQUENCE pattern: $<<><<$ with $b = 2$ and $a = 1$; (B) Illustrating the execution of the seed transducer on a time series (within the output $o$, $o_r$, $m_b$, $f_e$ are shortcut for **out**, **out**$_r$, **maybe**$_b$, **found**$_e$); (C) Illustrating the relation between the $s$-occurrence, the $i$-occurrence and the $e$-occurrence.

Figure 2.15: Parametrised automaton for any functional time-series constraints of the DIP_ON_INCREASING_SEQUENCE pattern obtained by applying the decoration table 2.35 to the corresponding transducer

Figure 2.16: Parametrised automaton for any predicate time-series constraints of the DIP_ON_INCREASING_SEQUENCE pattern obtained by applying the decoration table 2.36 to the corresponding transducer, where $\theta$ is the predicate (i.e., $\leq$, $=$, $\geq$) used for comparing two consecutive feature values

### 2.1.6 GORGE

Given a sequence $S$ over the alphabet $\{<, =, >\}$, an occurrence of the pattern GORGE is the *maximal* subsequence of $S$ which matches the regular expression $(> | (> (= | >)^* >))(< | (< (= | <)^* <))$. Part (A) and parts (B,C) of Figure 2.17 respectively depict the seed transducer associated with the GORGE pattern as well as one example of its execution on a time series.

Figure 2.17: (A) Seed transducer for the GORGE pattern: $(> | (> (= | >)^* >))(< | (< (= | <)^* <))$ with $b = 1$ and $a = 1$; (B) Illustrating the execution of the seed transducer on a time series (within the output $\mathbf{o}$, $\mathbf{f}$, $\mathbf{m_a}$, $\mathbf{o_a}$, $\mathbf{m_b}$, $\mathbf{o_r}$ are shortcut for **out**, **found**, **maybe$_a$**, **out$_a$**, **maybe$_b$**, **out$_r$**); (C) Illustrating the relation between the $s$-occurrence, the $i$-occurrence and the $e$-occurrence.

Figure 2.18: Parametrised automaton for any functional time-series constraints of the GORGE pattern obtained by applying the decoration table 2.35 to the corresponding transducer (transition $u \to r$ has the same accumulator update as transition $r \to u$)

Figure 2.19: Parametrised automaton for any predicate time-series constraints of the GORGE pattern obtained by applying the decoration table 2.36 to the corresponding transducer, where $\theta$ is the predicate (i.e., $\leq, =, \geq$) used for comparing two consecutive feature values; transition $u \to r$ has the same accumulator update as transition $r \to u$.

|   | $s$ | $r$ | $t$ | $u$ |
|---|---|---|---|---|
| $s$ | $\phi_g(\overrightarrow{C}, \overleftarrow{C})$ | $\phi_g(\overrightarrow{C}, \overleftarrow{C})$ | $\phi_g(\overrightarrow{C}, \overleftarrow{C})$ | $\phi_g(\overrightarrow{C}, \overleftarrow{C})$ |
| $r$ | $\phi_g(\overrightarrow{C}, \overleftarrow{C})$ | $\phi_f(\overrightarrow{D}, \overleftarrow{D}, \delta_f^i)$ | $\phi_f(\overleftarrow{C}, \overrightarrow{D}, \overleftarrow{D}, \delta_f^i)$ | $\phi_g(\overrightarrow{C}, \overleftarrow{C})$ |
| $t$ | $\phi_g(\overrightarrow{C}, \overleftarrow{C})$ | $\phi_f(\overrightarrow{C}, \overrightarrow{D}, \overleftarrow{D}, \delta_f^i)$ | $\phi_g(\overrightarrow{C}, \overleftarrow{C})$ | $\phi_f(\overrightarrow{C}, \overrightarrow{D}, \overleftarrow{D}, \delta_f^i)$ |
| $u$ | $\phi_g(\overrightarrow{C}, \overleftarrow{C})$ | $\phi_g(\overrightarrow{C}, \overleftarrow{C})$ | $\phi_f(\overleftarrow{C}, \overrightarrow{D}, \overleftarrow{D}, \delta_f^i)$ | $\phi_g(\overrightarrow{C}, \overleftarrow{C})$ |

Table 2.6: Parametrised glue matrix for any $g\_f\_\textsc{gorge}$ constraint

### 2.1.7 INCREASING

Given a sequence $S$ over the alphabet $\{<, =, >\}$, an occurrence of the pattern IN-CREASING is the subsequence of $S$ which matches the regular expression $<$. Part (A) and parts (B,C) of Figure 2.20 respectively depict the seed transducer associated with the INCREASING pattern as well as one example of its execution on a time series.

Figure 2.20: (A) Seed transducer for the INCREASING pattern: $<$ with $b = 0$ and $a = 0$; (B) Illustrating the execution of the seed transducer on a time series (within the output $\mathbf{o}$, $\mathbf{f_e}$ are shortcut for $\mathbf{out}$, $\mathbf{found_e}$); (C) Illustrating the relation between the $s$-occurrence, the $i$-occurrence and the $e$-occurrence.

$$
\begin{array}{c|c}
 & s \\
\hline
s & \phi_g(\overrightarrow{C}, \overleftarrow{C}) \\
\end{array}
$$

Table 2.7: Parametrised glue matrix for any $g\_f\_$INCREASING constraint

Figure 2.21: Parametrised automaton for any functional time-series constraints of the INCREASING pattern obtained by applying the decoration table 2.35 to the corresponding transducer

Figure 2.22: Parametrised automaton for any predicate time-series constraints of the INCREASING pattern obtained by applying the decoration table 2.36 to the corresponding transducer, where $\theta$ is the predicate (i.e., $\leq$, $=$, $\geq$) used for comparing two consecutive feature values

### 2.1.8  INCREASING_SEQUENCE

Given a sequence $S$ over the alphabet $\{<, =, >\}$, an occurrence of the pattern INCREASING_SEQUENCE is the *maximal* subsequence of $S$ which matches the regular expression $< (< | =)^* < | <$. Part (A) and parts (B,C) of Figure 2.23 respectively depict the seed transducer associated with the INCREASING_SEQUENCE pattern as well as one example of its execution on a time series.

Figure 2.23: (A) Seed transducer for the INCREASING_SEQUENCE pattern: $< (< | = )^* < | <$ with $b = 0$ and $a = 0$; (B) Illustrating the execution of the seed transducer on a time series (within the output **o**, **f**, **m**$_a$, **o**$_a$ are shortcut for **out**, **found**, **maybe**$_a$, **out**$_a$); (C) Illustrating the relation between the *s*-occurrence, the *i*-occurrence and the *e*-occurrence.

Figure 2.24: Parametrised automaton for any functional time-series constraints of the INCREASING_SEQUENCE pattern obtained by applying the decoration table 2.35 to the corresponding transducer

Figure 2.25: Parametrised automaton for any predicate time-series constraints of the INCREASING_SEQUENCE pattern obtained by applying the decoration table 2.36 to the corresponding transducer, where $\theta$ is the predicate (i.e., $\leq$, $=$, $\geq$) used for comparing two consecutive feature values

|   | $s$ | $t$ |
|---|---|---|
| $s$ | $\phi_g(\overrightarrow{C}, \overleftarrow{C})$ | $\phi_g(\overrightarrow{C}, \overleftarrow{C})$ |
| $t$ | $\phi_g(\overrightarrow{C}, \overleftarrow{C})$ | $\phi_f(\overrightarrow{C}, \overleftarrow{C}, \overrightarrow{D}, \overleftarrow{D}, \delta_f^i)$ |

Table 2.8: Parametrised glue matrix for any $g\_f\_\textsc{increasing\_sequence}$ constraint

### 2.1.9 INCREASING_TERRACE

Given a sequence $S$ over the alphabet $\{<, =, >\}$, an occurrence of the pattern IN-CREASING_TERRACE is the *maximal* subsequence of $S$ which matches the regular expression $<=^+<$. Part (A) and parts (B,C) of Figure 2.26 respectively depict the seed transducer associated with the INCREASING_TERRACE pattern as well as one example of its execution on a time series.

Figure 2.26: (A) Seed transducer for the INCREASING_TERRACE pattern: $<=^+<$ with $b = 1$ and $a = 1$; (B) Illustrating the execution of the seed transducer on a time series (within the output **o**, $\mathbf{m_b}$, $\mathbf{o_r}$, $\mathbf{f_e}$ are shortcut for **out**, $\mathbf{maybe_b}$, $\mathbf{out_r}$, $\mathbf{found_e}$); (C) Illustrating the relation between the $s$-occurrence, the $i$-occurrence and the $e$-occurrence.

Figure 2.27: Parametrised automaton for any functional time-series constraints of the INCREASING_TERRACE pattern obtained by applying the decoration table 2.35 to the corresponding transducer

Figure 2.28: Parametrised automaton for any predicate time-series constraints of the INCREASING_TERRACE pattern obtained by applying the decoration table 2.36 to the corresponding transducer, where $\theta$ is the predicate (i.e., $\leq$, $=$, $\geq$) used for comparing two consecutive feature values

|   | $s$ | $r$ | $t$ |
|---|---|---|---|
| $s$ | $\phi_g(\overrightarrow{C}, \overleftarrow{C})$ | $\phi_g(\overrightarrow{C}, \overleftarrow{C})$ | $\phi_g(\overrightarrow{C}, \overleftarrow{C})$ |
| $r$ | $\phi_g(\overrightarrow{C}, \overleftarrow{C})$ | $\phi_g(\overrightarrow{C}, \overleftarrow{C})$ | $\phi_f(\overrightarrow{D}, \overleftarrow{D}, \delta_f^i)$ |
| $t$ | $\phi_g(\overrightarrow{C}, \overleftarrow{C})$ | $\phi_f(\overrightarrow{D}, \overleftarrow{D}, \delta_f^i)$ | $\phi_f(\overrightarrow{D}, \overleftarrow{D}, \delta_f^i)$ |

Table 2.9: Parametrised glue matrix for any $g\_f\_$INCREASING_TERRACE constraint

### 2.1.10  INFLEXION

Given a sequence $S$ over the alphabet $\{<, =, >\}$, an occurrence of the pattern INFLEXION is the *maximal* subsequence of $S$ which matches the regular expression $< (< | =)^* > | > (> | =)^* <$. Part (A) and parts (B,C) of Figure 2.29 respectively depict the seed transducer associated with the INFLEXION pattern as well as one example of its execution on a time series.

**Figure 2.29:** (A) Seed transducer for the INFLEXION pattern: $< (< | =)^* > | > (> | =)^* <$ with $b = 1$ and $a = 1$; (B) Illustrating the execution of the seed transducer on a time series (within the output **o**, $\mathbf{m_b}$, $\mathbf{f_e}$ are shortcut for **out**, **maybe$_b$**, **found$_e$**); (C) Illustrating the relation between the $s$-occurrence, the $i$-occurrence and the $e$-occurrence.

Figure 2.30: Parametrised automaton for any functional time-series constraints of the INFLEXION pattern obtained by applying the decoration table 2.35 to the corresponding transducer (transition $r \to t$ has the same accumulators updates as transition $t \to r$)

Figure 2.31: Parametrised automaton for any predicate time-series constraints of the INFLEXION pattern obtained by applying the decoration table 2.36 to the corresponding transducer, where $\theta$ is the predicate (i.e., $\leq, =, \geq$) used for comparing two consecutive feature values; transition $r \to t$ has the same accumulators updates as transition $t \to r$.

### 2.1.11 PEAK

Given a sequence $S$ over the alphabet $\{<, =, >\}$, an occurrence of the pattern PEAK is the *maximal* subsequence of $S$ which matches the regular expression $< (= | < )^* (> | =)^* >$. Part (A) and parts (B,C) of Figure 2.32 respectively depict the seed transducer associated with the PEAK pattern as well as one example of its execution on a time series.

Figure 2.32: (A) Seed transducer for the PEAK pattern: $< (= | <)^* (> | =)^* >$ with $b = 1$ and $a = 1$; (B) Illustrating the execution of the seed transducer on a time series (within the output $\mathbf{o}$, $\mathbf{m_b}$, $\mathbf{f}$, $\mathbf{m_a}$, $\mathbf{o_a}$ are shortcut for $\mathbf{out}$, $\mathbf{maybe_b}$, $\mathbf{found}$, $\mathbf{maybe_a}$, $\mathbf{out_a}$); (C) Illustrating the relation between the $s$-occurrence, the $i$-occurrence and the $e$-occurrence.

Figure 2.33: Parametrised automaton for any functional time-series constraints of the PEAK pattern obtained by applying the decoration table 2.35 to the corresponding transducer

Figure 2.34: Parametrised automaton for any predicate time-series constraints of the PEAK pattern obtained by applying the decoration table 2.36 to the corresponding transducer, where $\theta$ is the predicate (i.e., $\leq$, $=$, $\geq$) used for comparing two consecutive feature values

|   | $s$ | $r$ | $t$ |
|---|---|---|---|
| $s$ | $\phi_g(\overrightarrow{C},\overleftarrow{C})$ | $\phi_g(\overrightarrow{C},\overleftarrow{C})$ | $\phi_g(\overrightarrow{C},\overleftarrow{C})$ |
| $r$ | $\phi_g(\overrightarrow{C},\overleftarrow{C})$ | $\phi_f(\overrightarrow{D},\overleftarrow{D},\delta_f^i)$ | $\phi_f(\overleftarrow{C},\overrightarrow{D},\overleftarrow{D},\delta_f^i)$ |
| $t$ | $\phi_g(\overrightarrow{C},\overleftarrow{C})$ | $\phi_f(\overrightarrow{C},\overrightarrow{D},\overleftarrow{D},\delta_f^i)$ | $\phi_g(\overrightarrow{C},\overleftarrow{C})$ |

Table 2.10: Parametrised glue matrix for any $g\_f\_\textsc{peak}$ constraint

### 2.1.12 PLAIN

Given a sequence $S$ over the alphabet $\{<, =, >\}$, an occurrence of the pattern PLAIN is the *maximal* subsequence of $S$ which matches the regular expression $>=^*<$. Part (A) and parts (B,C) of Figure 2.35 respectively depict the seed transducer associated with the PLAIN pattern as well as one example of its execution on a time series.

Figure 2.35: (A) Seed transducer for the PLAIN pattern: $>=^*<$ with $b = 1$ and $a = 1$; (B) Illustrating the execution of the seed transducer on a time series (within the output $\mathbf{o}$, $\mathbf{f_e}$, $\mathbf{m_b}$, $\mathbf{o_r}$ are shortcut for $\mathbf{out}$, $\mathbf{found_e}$, $\mathbf{maybe_b}$, $\mathbf{out_r}$); (C) Illustrating the relation between the $s$-occurrence, the $i$-occurrence and the $e$-occurrence.

Figure 2.36: Parametrised automaton for any functional time-series constraints of the PLAIN pattern obtained by applying the decoration table 2.35 to the corresponding transducer

Figure 2.37: Parametrised automaton for any predicate time-series constraints of the PLAIN pattern obtained by applying the decoration table 2.36 to the corresponding transducer, where $\theta$ is the predicate (i.e., $\leq, =, \geq$) used for comparing two consecutive feature values

|   | $s$ | $r$ | $t$ |
|---|---|---|---|
| $s$ | $\phi_g(\overrightarrow{C}, \overleftarrow{C})$ | $\phi_g(\overrightarrow{C}, \overleftarrow{C})$ | $\phi_g(\overrightarrow{C}, \overleftarrow{C})$ |
| $r$ | $\phi_g(\overrightarrow{C}, \overleftarrow{C})$ | $\phi_f(\overrightarrow{D}, \overleftarrow{D}, \delta_f^i)$ | $\phi_f(\overrightarrow{D}, \overleftarrow{D}, \delta_f^i)$ |
| $t$ | $\phi_g(\overrightarrow{C}, \overleftarrow{C})$ | $\phi_f(\overrightarrow{D}, \overleftarrow{D}, \delta_f^i)$ | $\phi_f(\overrightarrow{D}, \overleftarrow{D}, \delta_f^i)$ |

Table 2.11: Parametrised glue matrix for any $g\_f\_$PLAIN constraint

### 2.1.13 PLATEAU

Given a sequence $S$ over the alphabet $\{<, =, >\}$, an occurrence of the pattern PLATEAU is the *maximal* subsequence of $S$ which matches the regular expression $<=^*>$. Part (A) and parts (B,C) of Figure 2.38 respectively depict the seed transducer associated with the PLATEAU pattern as well as one example of its execution on a time series.

Figure 2.38: (A) Seed transducer for the PLATEAU pattern: $<=^*>$ with $b = 1$ and $a = 1$; (B) Illustrating the execution of the seed transducer on a time series (within the output $\mathbf{o}$, $\mathbf{f_e}$, $\mathbf{m_b}$, $\mathbf{o_r}$ are shortcut for $\mathbf{out}$, $\mathbf{found_e}$, $\mathbf{maybe_b}$, $\mathbf{out_r}$); (C) Illustrating the relation between the $s$-occurrence, the $i$-occurrence and the $e$-occurrence.

Figure 2.39: Parametrised automaton for any functional time-series constraints of the PLATEAU pattern obtained by applying the decoration table 2.35 to the corresponding transducer

Figure 2.40: Parametrised automaton for any predicate time-series constraints of the PLATEAU pattern obtained by applying the decoration table 2.36 to the corresponding transducer, where $\theta$ is the predicate (i.e., $\leq$, $=$, $\geq$) used for comparing two consecutive feature values

|   | $s$ | $r$ | $t$ |
|---|---|---|---|
| $s$ | $\phi_g(\overrightarrow{C}, \overleftarrow{C})$ | $\phi_g(\overrightarrow{C}, \overleftarrow{C})$ | $\phi_g(\overrightarrow{C}, \overleftarrow{C})$ |
| $r$ | $\phi_g(\overrightarrow{C}, \overleftarrow{C})$ | $\phi_f(\overrightarrow{D}, \overleftarrow{D}, \delta_f^i)$ | $\phi_f(\overrightarrow{D}, \overleftarrow{D}, \delta_f^i)$ |
| $t$ | $\phi_g(\overrightarrow{C}, \overleftarrow{C})$ | $\phi_f(\overrightarrow{D}, \overleftarrow{D}, \delta_f^i)$ | $\phi_f(\overrightarrow{D}, \overleftarrow{D}, \delta_f^i)$ |

Table 2.12: Parametrised glue matrix for any $g\_f$\_PLATEAU constraint

### 2.1.14 PROPER_PLAIN

Given a sequence $S$ over the alphabet $\{<, =, >\}$, an occurrence of the pattern PROPER_PLAIN is the *maximal* subsequence of $S$ which matches the regular expression $>=^+<$. Part (A) and parts (B,C) of Figure 2.41 respectively depict the seed transducer associated with the PROPER_PLAIN pattern as well as one example of its execution on a time series.

Figure 2.41: (A) Seed transducer for the PROPER_PLAIN pattern: $>=^+<$ with $b = 1$ and $a = 1$; (B) Illustrating the execution of the seed transducer on a time series (within the output $\mathbf{o}$, $\mathbf{m_b}$, $\mathbf{f_e}$ are shortcut for $\mathbf{out}$, $\mathbf{maybe_b}$, $\mathbf{found_e}$); (C) Illustrating the relation between the $s$-occurrence, the $i$-occurrence and the $e$-occurrence.

Figure 2.42: Parametrised automaton for any functional time-series constraints of the PROPER_PLAIN pattern obtained by applying the decoration table 2.35 to the corresponding transducer

Figure 2.43: Parametrised automaton for any predicate time-series constraints of the PROPER_PLAIN pattern obtained by applying the decoration table 2.36 to the corresponding transducer, where $\theta$ is the predicate (i.e., $\leq$, $=$, $\geq$) used for comparing two consecutive feature values

|   | $s$ | $r$ | $t$ |
|---|---|---|---|
| $s$ | $\phi_g(\overrightarrow{C}, \overleftarrow{C})$ | $\phi_g(\overrightarrow{C}, \overleftarrow{C})$ | $\phi_g(\overrightarrow{C}, \overleftarrow{C})$ |
| $r$ | $\phi_g(\overrightarrow{C}, \overleftarrow{C})$ | $\phi_g(\overrightarrow{C}, \overleftarrow{C})$ | $\phi_f(\overrightarrow{D}, \overleftarrow{D}, \delta_f^i)$ |
| $t$ | $\phi_g(\overrightarrow{C}, \overleftarrow{C})$ | $\phi_f(\overrightarrow{D}, \overleftarrow{D}, \delta_f^i)$ | $\phi_f(\overrightarrow{D}, \overleftarrow{D}, \delta_f^i)$ |

Table 2.13: Parametrised glue matrix for any $g\_f\_\text{PROPER\_PLAIN}$ constraint

### 2.1.15 PROPER_PLATEAU

Given a sequence $S$ over the alphabet $\{<, =, >\}$, an occurrence of the pattern PROPER_PLATEAU is the *maximal* subsequence of $S$ which matches the regular expression $<=^{+}>$. Part (A) and parts (B,C) of Figure 2.44 respectively depict the seed transducer associated with the PROPER_PLATEAU pattern as well as one example of its execution on a time series.

Figure 2.44: (A) Seed transducer for the PROPER_PLATEAU pattern: $<=^{+}>$ with $b = 1$ and $a = 1$; (B) Illustrating the execution of the seed transducer on a time series (within the output $\mathbf{o}$, $\mathbf{m_b}$, $\mathbf{f_e}$ are shortcut for $\mathbf{out}$, $\mathbf{maybe_b}$, $\mathbf{found_e}$); (C) Illustrating the relation between the $s$-occurrence, the $i$-occurrence and the $e$-occurrence.

Figure 2.45: Parametrised automaton for any functional time-series constraints of the PROPER_PLATEAU pattern obtained by applying the decoration table 2.35 to the corresponding transducer

Figure 2.46: Parametrised automaton for any predicate time-series constraints of the PROPER_PLATEAU pattern obtained by applying the decoration table 2.36 to the corresponding transducer, where $\theta$ is the predicate (i.e., $\leq$, $=$, $\geq$) used for comparing two consecutive feature values

|   | $s$ | $r$ | $t$ |
|---|---|---|---|
| $s$ | $\phi_g(\overrightarrow{C}, \overleftarrow{C})$ | $\phi_g(\overrightarrow{C}, \overleftarrow{C})$ | $\phi_g(\overrightarrow{C}, \overleftarrow{C})$ |
| $r$ | $\phi_g(\overrightarrow{C}, \overleftarrow{C})$ | $\phi_g(\overrightarrow{C}, \overleftarrow{C})$ | $\phi_f(\overrightarrow{D}, \overleftarrow{D}, \delta_f^i)$ |
| $t$ | $\phi_g(\overrightarrow{C}, \overleftarrow{C})$ | $\phi_f(\overrightarrow{D}, \overleftarrow{D}, \delta_f^i)$ | $\phi_f(\overrightarrow{D}, \overleftarrow{D}, \delta_f^i)$ |

Table 2.14: Parametrised glue matrix for any $g\_f$_PROPER_PLATEAU_ constraint

### 2.1.16 STEADY

Given a sequence $S$ over the alphabet $\{<, =, >\}$, an occurrence of the pattern STEADY is the subsequence of $S$ which matches the regular expression $=$. Part (A) and parts (B,C) of Figure 2.47 respectively depict the seed transducer associated with the STEADY pattern as well as one example of its execution on a time series.

Figure 2.47: (A) Seed transducer for the STEADY pattern: $=$ with $b = 0$ and $a = 0$; (B) Illustrating the execution of the seed transducer on a time series (within the output $\mathbf{f_e}$, $\mathbf{o}$ are shortcut for $\mathbf{found_e}$, $\mathbf{out}$); (C) Illustrating the relation between the $s$-occurrence, the $i$-occurrence and the $e$-occurrence.

Table 2.15: Parametrised glue matrix for any $g\_f\_$STEADY constraint

Figure 2.48: Parametrised automaton for any functional time-series constraints of the STEADY pattern obtained by applying the decoration table 2.35 to the corresponding transducer

Figure 2.49: Parametrised automaton for any predicate time-series constraints of the STEADY pattern obtained by applying the decoration table 2.36 to the corresponding transducer, where $\theta$ is the predicate (i.e., $\leq$, $=$, $\geq$) used for comparing two consecutive feature values

### 2.1.17 STEADY_SEQUENCE

Given a sequence $S$ over the alphabet $\{<, =, >\}$, an occurrence of the pattern STEADY_SEQUENCE is the *maximal* subsequence of $S$ which matches the regular expression $=^+$. Part (A) and parts (B,C) of Figure 2.50 respectively depict the seed transducer associated with the STEADY_SEQUENCE pattern as well as one example of its execution on a time series.

Figure 2.50: (A) Seed transducer for the STEADY_SEQUENCE pattern: $=^+$ with $b = 0$ and $a = 0$; (B) Illustrating the execution of the seed transducer on a time series (within the output $\mathbf{o}$, $\mathbf{f}$, $\mathbf{o_a}$ are shortcut for **out**, **found**, **out**$_a$); (C) Illustrating the relation between the $s$-occurrence, the $i$-occurrence and the $e$-occurrence.

|   | $s$ | $r$ |
|---|-----|-----|
| $s$ | $\phi_g(\overrightarrow{C}, \overleftarrow{C})$ | $\phi_g(\overrightarrow{C}, \overleftarrow{C})$ |
| $r$ | $\phi_g(\overrightarrow{C}, \overleftarrow{C})$ | $\phi_f(\overrightarrow{C}, \overleftarrow{C}, \overrightarrow{D}, \overleftarrow{D}, \delta_f^i)$ |

Table 2.16: Parametrised glue matrix for any $g\_f\_$STEADY_SEQUENCE constraint

Figure 2.51: Parametrised automaton for any functional time-series constraints of the `STEADY_SEQUENCE` pattern obtained by applying the decoration table 2.35 to the corresponding transducer

Figure 2.52: Parametrised automaton for any predicate time-series constraints of the `STEADY_SEQUENCE` pattern obtained by applying the decoration table 2.36 to the corresponding transducer, where $\theta$ is the predicate (i.e., $\leq$, $=$, $\geq$) used for comparing two consecutive feature values

## 2.1.18 **STRICTLY_DECREASING_SEQUENCE**

Given a sequence $S$ over the alphabet $\{<, =, >\}$, an occurrence of the pattern STRICTLY_DECREASING_SEQUENCE is the *maximal* subsequence of $S$ which matches the regular expression $>^+$. Part (A) and parts (B,C) of Figure 2.53 respectively depict the seed transducer associated with the STRICTLY_DECREASING_SEQUENCE pattern as well as one example of its execution on a time series.

Figure 2.53: (A) Seed transducer for the STRICTLY_DECREASING_SEQUENCE pattern: $>^+$ with $b = 0$ and $a = 0$; (B) Illustrating the execution of the seed transducer on a time series (within the output **o**, **f**, $\mathbf{o}_a$ are shortcut for **out**, **found**, **out**$_a$); (C) Illustrating the relation between the *s*-occurrence, the *i*-occurrence and the *e*-occurrence.

Figure 2.54: Parametrised automaton for any functional time-series constraints of the STRICTLY_DECREASING_SEQUENCE pattern obtained by applying the decoration table 2.35 to the corresponding transducer

Figure 2.55: Parametrised automaton for any predicate time-series constraints of the STRICTLY_DECREASING_SEQUENCE pattern obtained by applying the decoration table 2.36 to the corresponding transducer, where $\theta$ is the predicate (i.e., $\leq$, $=$, $\geq$) used for comparing two consecutive feature values

|   | $s$ | $r$ |
|---|---|---|
| $s$ | $\phi_g(\overrightarrow{C}, \overleftarrow{C})$ | $\phi_g(\overrightarrow{C}, \overleftarrow{C})$ |
| $r$ | $\phi_g(\overrightarrow{C}, \overleftarrow{C})$ | $\phi_f(\overrightarrow{C}, \overleftarrow{C}, \overrightarrow{D}, \overleftarrow{D}, \delta_f^i)$ |

Table 2.17: Parametrised glue matrix for any $g\_f\_$STRICTLY_DECREASING_SEQUENCE constraint

### 2.1.19 STRICTLY_INCREASING_SEQUENCE

Given a sequence $S$ over the alphabet $\{<, =, >\}$, an occurrence of the pattern STRICTLY_INCREASING_SEQUENCE is the *maximal* subsequence of $S$ which matches the regular expression $<^{+}$. Part (A) and parts (B,C) of Figure 2.56 respectively depict the seed transducer associated with the STRICTLY_INCREASING_SEQUENCE pattern as well as one example of its execution on a time series.

Figure 2.56: (A) Seed transducer for the STRICTLY_INCREASING_SEQUENCE pattern: $<^{+}$ with $b = 0$ and $a = 0$; (B) Illustrating the execution of the seed transducer on a time series (within the output $\mathbf{o}$, $\mathbf{f}$, $\mathbf{o_a}$ are shortcut for $\mathbf{out}$, $\mathbf{found}$, $\mathbf{out_a}$); (C) Illustrating the relation between the $s$-occurrence, the $i$-occurrence and the $e$-occurrence.

Figure 2.57: Parametrised automaton for any functional time-series constraints of the STRICTLY_INCREASING_SEQUENCE pattern obtained by applying the decoration table 2.35 to the corresponding transducer

Figure 2.58: Parametrised automaton for any predicate time-series constraints of the STRICTLY_INCREASING_SEQUENCE pattern obtained by applying the decoration table 2.36 to the corresponding transducer, where $\theta$ is the predicate (i.e., $\leq$, $=$, $\geq$) used for comparing two consecutive feature values

|   | $s$ | $r$ |
|---|---|---|
| $s$ | $\phi_g(\overrightarrow{C}, \overleftarrow{C})$ | $\phi_g(\overrightarrow{C}, \overleftarrow{C})$ |
| $r$ | $\phi_g(\overrightarrow{C}, \overleftarrow{C})$ | $\phi_f(\overrightarrow{C}, \overleftarrow{C}, \overrightarrow{D}, \overleftarrow{D}, \delta_f^i)$ |

Table 2.18: Parametrised glue matrix for any $g\_f\_\text{STRICTLY\_INCREASING\_SEQUENCE}$ constraint

## 2.1.20 SUMMIT

Given a sequence $S$ over the alphabet $\{<, =, >\}$, an occurrence of the pattern SUMMIT is the *maximal* subsequence of $S$ which matches the regular expression $(< \mid (< (= \mid <)^* <))(> \mid (> (= \mid >)^* >))$. Part (A) and parts (B,C) of Figure 2.59 respectively depict the seed transducer associated with the SUMMIT pattern as well as one example of its execution on a time series.

Figure 2.59: (A) Seed transducer for the SUMMIT pattern: $(< \mid (< (= \mid <)^* <))(> \mid (> (= \mid >)^* >))$ with $b = 1$ and $a = 1$; (B) Illustrating the execution of the seed transducer on a time series (within the output o, f, $m_a$, $o_a$, $m_b$, $o_r$ are shortcut for **out**, **found**, **maybe**$_a$, **out**$_a$, **maybe**$_b$, **out**$_r$); (C) Illustrating the relation between the $s$-occurrence, the $i$-occurrence and the $e$-occurrence.

Figure 2.60: Parametrised automaton for any functional time-series constraints of the SUMMIT pattern obtained by applying the decoration table 2.35 to the corresponding transducer (transition $u \to r$ has the same accumulator update as transition $r \to u$)

Figure 2.61: Parametrised automaton for any predicate time-series constraints of the SUMMIT pattern obtained by applying the decoration table 2.36 to the corresponding transducer, where $\theta$ is the predicate (i.e., $\leq$, $=$, $\geq$) used for comparing two consecutive feature values; transition $u \to r$ has the same accumulator update as transition $r \to u$.

|   | $s$ | $r$ | $t$ | $u$ |
|---|---|---|---|---|
| $s$ | $\phi_g(\overrightarrow{C}, \overleftarrow{C})$ | $\phi_g(\overrightarrow{C}, \overleftarrow{C})$ | $\phi_g(\overrightarrow{C}, \overleftarrow{C})$ | $\phi_g(\overrightarrow{C}, \overleftarrow{C})$ |
| $r$ | $\phi_g(\overrightarrow{C}, \overleftarrow{C})$ | $\phi_f(\overrightarrow{D}, \overleftarrow{D}, \delta_f^i)$ | $\phi_f(\overleftarrow{C}, \overrightarrow{D}, \overleftarrow{D}, \delta_f^i)$ | $\phi_g(\overrightarrow{C}, \overleftarrow{C})$ |
| $t$ | $\phi_g(\overrightarrow{C}, \overleftarrow{C})$ | $\phi_f(\overrightarrow{C}, \overrightarrow{D}, \overleftarrow{D}, \delta_f^i)$ | $\phi_g(\overrightarrow{C}, \overleftarrow{C})$ | $\phi_f(\overrightarrow{C}, \overrightarrow{D}, \overleftarrow{D}, \delta_f^i)$ |
| $u$ | $\phi_g(\overrightarrow{C}, \overleftarrow{C})$ | $\phi_g(\overrightarrow{C}, \overleftarrow{C})$ | $\phi_f(\overleftarrow{C}, \overrightarrow{D}, \overleftarrow{D}, \delta_f^i)$ | $\phi_g(\overrightarrow{C}, \overleftarrow{C})$ |

Table 2.19: Parametrised glue matrix for any $g\_f\_$SUMMIT constraint

## 2.1.21 VALLEY

Given a sequence $S$ over the alphabet $\{<, =, >\}$, an occurrence of the pattern VALLEY is the *maximal* subsequence of $S$ which matches the regular expression $> (= | >)^* (< | =)^* <$. Part (A) and parts (B,C) Figure 2.62 respectively depict the seed transducer associated with the VALLEY pattern as well as one example of its execution on a time series.

Figure 2.62: (A) Seed transducer for the VALLEY pattern: $> (= | >)^* (< | =)^* <$ with $b = 1$ and $a = 1$; (B) Illustrating the execution of the seed transducer on a time series (within the output $\mathbf{o}, \mathbf{m_b}, \mathbf{f}, \mathbf{m_a}, \mathbf{o_a}$ are shortcut for $\mathbf{out}, \mathbf{maybe_b}, \mathbf{found}, \mathbf{maybe_a}, \mathbf{out_a}$); (C) Illustrating the relation between the $s$-occurrence, the $i$-occurrence and the $e$-occurrence.

Figure 2.63: Parametrised automaton for any functional time-series constraints of the VALLEY pattern obtained by applying the decoration table 2.35 to the corresponding transducer

Figure 2.64: Parametrised automaton for any predicate time-series constraints of the VALLEY pattern obtained by applying the decoration table 2.36 to the corresponding transducer, where $\theta$ is the predicate (i.e., $\leq$, $=$, $\geq$) used for comparing two consecutive feature values

|   | $s$ | $r$ | $t$ |
|---|---|---|---|
| $s$ | $\phi_g(\overrightarrow{C}, \overleftarrow{C})$ | $\phi_g(\overrightarrow{C}, \overleftarrow{C})$ | $\phi_g(\overrightarrow{C}, \overleftarrow{C})$ |
| $r$ | $\phi_g(\overrightarrow{C}, \overleftarrow{C})$ | $\phi_f(\overrightarrow{D}, \overleftarrow{D}, \delta_f^i)$ | $\phi_f(\overleftarrow{C}, \overrightarrow{D}, \overleftarrow{D}, \delta_f^i)$ |
| $t$ | $\phi_g(\overrightarrow{C}, \overleftarrow{C})$ | $\phi_f(\overrightarrow{C}, \overrightarrow{D}, \overleftarrow{D}, \delta_f^i)$ | $\phi_g(\overrightarrow{C}, \overleftarrow{C})$ |

Table 2.20: Parametrised glue matrix for any $g\_f\_\texttt{VALLEY}$ constraint

## 2.1.22 ZIGZAG

Given a sequence $S$ over the alphabet $\{<, =, >\}$, an occurrence of the pattern ZIGZAG is the *maximal* subsequence of $S$ which matches the regular expression $(<>)^+(< \mid <>) \mid (><)^+(> \mid ><)$. Figures 2.65 and 2.66 respectively depict the seed transducer associated with the ZIGZAG pattern as well as one example of its execution on a time series.

Figure 2.65: Seed transducer for the ZIGZAG pattern: $(<>)^+(< \mid <>) \mid (><)^+(> \mid ><)$ with $b = 1$ and $a = 1$; missing transitions from $a$, $d$ to $s$ are labelled by $= :$ **out**, missing transitions from $b$, $e$ to $s$ are labelled by $= :$ **out**$_r$, and missing transitions from $c$, $f$ to $s$ are labelled by $= :$ **out**$_a$.

Figure 2.66: (A) Illustrating the execution of the seed transducer associated with the ZIGZAG pattern on a time series (within the output $\mathbf{o}$, $\mathbf{m}_b$, $\mathbf{f}$, $\mathbf{o}_a$ are shortcut for $\mathbf{out}$, $\mathbf{maybe}_b$, $\mathbf{found}$, $\mathbf{out}_a$); (B) Illustrating the relation between the $s$-occurrence, the $i$-occurrence and the $e$-occurrence.

|   | $s$ | $a$ | $b$ | $c$ | $d$ | $e$ | $f$ |
|---|---|---|---|---|---|---|---|
| $s$ | $\phi_g(\overrightarrow{C}, \overleftarrow{C})$ | $\phi_g(\overrightarrow{C}, \overleftarrow{C})$ | $\phi_g(\overrightarrow{C}, \overleftarrow{C})$ | $\phi_g(\overrightarrow{C}, \overleftarrow{C})$ | $\phi_g(\overrightarrow{C}, \overleftarrow{C})$ | $\phi_g(\overrightarrow{C}, \overleftarrow{C})$ | $\phi_g(\overrightarrow{C}, \overleftarrow{C})$ |
| $a$ | $\phi_g(\overrightarrow{C}, \overleftarrow{C})$ | $\phi_g(\overrightarrow{C}, \overleftarrow{C})$ | $\phi_g(\overrightarrow{C}, \overleftarrow{C})$ | $\phi_f(\overleftarrow{C}, \overleftarrow{D}, \overleftarrow{D}, \delta^i_f)$ | $\phi_g(\overrightarrow{C}, \overleftarrow{C})$ | $\phi_f(\overrightarrow{D}, \overleftarrow{D}, \delta^i_f)$ | $\phi_g(\overrightarrow{C}, \overleftarrow{C})$ |
| $b$ | $\phi_g(\overrightarrow{C}, \overleftarrow{C})$ | $\phi_g(\overrightarrow{C}, \overleftarrow{C})$ | $\phi_f(\overrightarrow{D}, \overleftarrow{D}, \delta^i_f)$ | $\phi_g(\overrightarrow{C}, \overleftarrow{C})$ | $\phi_f(\overrightarrow{D}, \overleftarrow{D}, \delta^i_f)$ | $\phi_g(\overrightarrow{C}, \overleftarrow{C})$ | $\phi_f(\overrightarrow{C}, \overleftarrow{D}, \overleftarrow{D}, \delta^i_f)$ |
| $c$ | $\phi_g(\overrightarrow{C}, \overleftarrow{C})$ | $\phi_f(\overrightarrow{C}, \overleftarrow{D}, \overleftarrow{D}, \delta^i_f)$ | $\phi_g(\overrightarrow{C}, \overleftarrow{C})$ | $\phi_f(\overrightarrow{C}, \overleftarrow{C}, \overleftarrow{D}, \overleftarrow{D}, \delta^i_f)$ | $\phi_g(\overrightarrow{C}, \overleftarrow{C})$ | $\phi_f(\overrightarrow{C}, \overleftarrow{D}, \overleftarrow{D}, \delta^i_f)$ | $\phi_g(\overrightarrow{C}, \overleftarrow{C})$ |
| $d$ | $\phi_g(\overrightarrow{C}, \overleftarrow{C})$ | $\phi_g(\overrightarrow{C}, \overleftarrow{C})$ | $\phi_f(\overrightarrow{D}, \overleftarrow{D}, \delta^i_f)$ | $\phi_g(\overrightarrow{C}, \overleftarrow{C})$ | $\phi_g(\overrightarrow{C}, \overleftarrow{C})$ | $\phi_g(\overrightarrow{C}, \overleftarrow{C})$ | $\phi_f(\overleftarrow{C}, \overleftarrow{D}, \overleftarrow{D}, \delta^i_f)$ |
| $e$ | $\phi_g(\overrightarrow{C}, \overleftarrow{C})$ | $\phi_f(\overrightarrow{D}, \overleftarrow{D}, \delta^i_f)$ | $\phi_g(\overrightarrow{C}, \overleftarrow{C})$ | $\phi_f(\overleftarrow{C}, \overleftarrow{D}, \overleftarrow{D}, \delta^i_f)$ | $\phi_g(\overrightarrow{C}, \overleftarrow{C})$ | $\phi_f(\overrightarrow{D}, \overleftarrow{D}, \delta^i_f)$ | $\phi_g(\overrightarrow{C}, \overleftarrow{C})$ |
| $f$ | $\phi_g(\overrightarrow{C}, \overleftarrow{C})$ | $\phi_g(\overrightarrow{C}, \overleftarrow{C})$ | $\phi_f(\overrightarrow{C}, \overleftarrow{D}, \overleftarrow{D}, \delta^i_f)$ | $\phi_g(\overrightarrow{C}, \overleftarrow{C})$ | $\phi_f(\overrightarrow{C}, \overleftarrow{D}, \overleftarrow{D}, \delta^i_f)$ | $\phi_g(\overrightarrow{C}, \overleftarrow{C})$ | $\phi_f(\overrightarrow{C}, \overleftarrow{C}, \overleftarrow{D}, \overleftarrow{D}, \delta^i_f)$ |

Table 2.21: Parametrised glue matrix for any $g\_f\_$ZIGZAG constraint

Figure 2.67: Parametrised automaton for any functional time-series constraints of the ZIGZAG pattern obtained by applying the decoration table 2.35 to the corresponding transducer; (1) missing transitions from $a, b, c, d, e, f$ to $s$ are labelled by $=$; (2) on transitions from $b, c, e, f$ to $s$ the accumulator $D$ is reset to its initial value; (3) on transitions from $c, f$ to $s$ the accumulator $R$ is updated wrt $C$ and the accumulator $C$ is reset to its initial value.

Figure 2.68: Parametrised automaton for any predicate time-series constraints of the ZIGZAG pattern obtained by applying the decoration table 2.36 to the corresponding transducer, where $\theta$ is the predicate (i.e., $\leq$, $=$, $\geq$) used for comparing two consecutive feature values; (1) missing transitions from $a, b, c, d, e, f$ to $s$ are labelled by $=$; (2) on transitions from $b, c, e, f$ to $s$ the accumulator $D$ is reset to its initial value; (3) on transitions from $c, f$ to $s$ the accumulator $F$ is reset to $C$, and the accumulator $R$ is updated wrt $C$ and $F$

## 2.2 Decoration Tables

This section provides the different decoration tables used for synthesising the automata with accumulators of this catalogue. While no automaton of this catalogue relies on tables 2.32 and 2.34, these two decoration tables may be used for generating distance constraints between consecutive occurrences of a pattern.

Table 2.22: Table for Type: cDRFoundEOut

| initialization | $C \leftarrow \textbf{default}_{\eta f}$ | | $D \leftarrow \textbf{id}_f$ | $R \leftarrow \textbf{default}_{\eta f}$ |
|---|---|---|---|---|
| return | $\phi_\eta(R,C)$ | | | |
| | | | Decoration | |
| Semantic Letter | After | Update of $C$ | Update of $D$ | Update of $R$ |
| **out** | | | $D \leftarrow \textbf{id}_f$ | |
| **out$_r$** | | | $D \leftarrow \textbf{id}_f$ | $R \leftarrow \phi_\eta(R,C)$ |
| **out$_a$** | | $C \leftarrow \textbf{default}_{\eta f}$ | $D \leftarrow \phi_f(D,\delta_f^i)$ | |
| **maybe$_b$** | 0 | | $D \leftarrow \phi_f(D,\delta_f^i)$ | |
| **maybe$_b$** | 1 | | $D \leftarrow \phi_f(D,\delta_f^{i+1})$ | |
| **maybe$_a$** | | | $D \leftarrow \phi_f(D,\delta_f^i)$ | |
| **found** | 0 | $C \leftarrow \phi_f(\phi_f(D,\delta_f^i),\delta_f^{i+1})$ | $D \leftarrow \textbf{id}_f$ | |
| **found** | 1 | $C \leftarrow \phi_f(D,\delta_f^i)$ | $D \leftarrow \textbf{id}_f$ | |
| **in** | 0 | $C \leftarrow \phi_f(C,\phi_f(D,\delta_f^{i+1}))$ | $D \leftarrow \textbf{id}_f$ | |
| **in** | 1 | $C \leftarrow \phi_f(C,\phi_f(D,\delta_f^i))$ | $D \leftarrow \textbf{id}_f$ | |
| **found$_e$** | 0 | | $D \leftarrow \textbf{id}_f$ | $R \leftarrow \phi_\eta(R,\phi_f(\phi_f(D,\delta_f^i),\delta_f^{i+1}))$ |
| **found$_e$** | 1 | | $D \leftarrow \textbf{id}_f$ | $R \leftarrow \phi_\eta(R,\phi_f(D,\delta_f^i))$ |

Table 2.23: Table for Type: cDRFoundEOutNoMA

| initialization | | $C \leftarrow \mathbf{default}_{\varphi f}$ | $D \leftarrow \mathrm{id}_f$ | $R \leftarrow \mathbf{default}_{\varphi f}$ |
|---|---|---|---|---|
| return | | $\varphi_g(R,C)$ | | |
| Semantic Letter | After | Update of $C$ | Update of $D$ | Update of $R$ |
| | | | Decoration | |
| **out** | | | $D \leftarrow \mathrm{id}_f$ | |
| **out**$_r$ | | | $D \leftarrow \mathrm{id}_f$ | $R \leftarrow \varphi_g(R,C)$ |
| **out**$_a$ | | $C \leftarrow \mathbf{default}_{\varphi f}$ | $D \leftarrow \mathrm{id}_f$ | |
| **maybe**$_b$ | 0 | | $D \leftarrow \varphi_f(D, \delta^i_f)$ | |
| **maybe**$_a$ | 1 | | $D \leftarrow \varphi_f(D, \delta^{i+1}_f)$ | |
| **maybe**$_a$ | | | $D \leftarrow \varphi_f(D, \delta^i_f)$ | |
| **found** | 0 | $C \leftarrow \varphi_f(\varphi_f(D, \delta^i_f), \delta^{i+1}_f)$ | $D \leftarrow \mathrm{id}_f$ | |
| **found** | 1 | $C \leftarrow \varphi_f(D, \delta^i_f)$ | $D \leftarrow \mathrm{id}_f$ | |
| **in** | 0 | $C \leftarrow \varphi_f(C, \delta^{i+1}_f)$ | $D \leftarrow \mathrm{id}_f$ | |
| **in** | 1 | $C \leftarrow \varphi_f(C, \delta^i_f)$ | $D \leftarrow \mathrm{id}_f$ | |
| **found**$_e$ | 0 | | $D \leftarrow \mathrm{id}_f$ | $R \leftarrow \varphi_g(R, \varphi_f(\varphi_f(D, \delta^i_f), \delta^{i+1}_f))$ |
| **found**$_a$ | 1 | | $D \leftarrow \mathrm{id}_f$ | $R \leftarrow \varphi_g(R, \varphi_f(\varphi_f(D, \delta^i_f)))$ |

Table 2.24: Table for Type: cRFoundEOutA

| initialization | | $C \leftarrow \mathtt{default}_{gf}$ | $R \leftarrow \mathtt{default}_{gf}$ |
|---|---|---|---|
| return | | $\phi_g(R, C)$ | |
| Semantic Letter | | Decoration | |
| | After | Update of $C$ | Update of $R$ |
| **out** | | | |
| **out**$_r$ | | $C \leftarrow \mathtt{default}_{gf}$ | |
| **out**$_a$ | | $C \leftarrow \mathtt{default}_{gf}$ | $R \leftarrow \phi_g(R, C)$ |
| **maybe**$_b$ | | | |
| **maybe**$_a$ | 0 | | |
| **maybe**$_a$ | 1 | | |
| **found** | 0 | $C \leftarrow \phi_f(\delta_f^i, \delta_f^{i+1})$ | |
| **found** | 1 | $C \leftarrow \delta_f^i$ | |
| **in** | 0 | $C \leftarrow \phi_f(C, \delta_f^{i+1})$ | |
| **in** | 1 | $C \leftarrow \phi_f(C, \delta_f^i)$ | |
| **found**$_e$ | 0 | $C \leftarrow \mathtt{default}_{gf}$ | $R \leftarrow \phi_g(R, \phi_f(\delta_f^i, \delta_f^{i+1}))$ |
| **found**$_e$ | 1 | $C \leftarrow \mathtt{default}_{gf}$ | $R \leftarrow \phi_g(R, \delta_f^i)$ |

Table 2.25: Table for Type: dRFoundFoundEInA

| initialization | | $D \leftarrow \mathtt{id}_f$ | $R \leftarrow \mathtt{default}_{gf}$ |
|---|---|---|---|
| **return** | | $R$ | |
| Semantic Letter | | | Decoration |
| | After | Update of $D$ | Update of $R$ |
| **out** | | | |
| **out**$_r$ | | $D \leftarrow \mathtt{id}_f$ | |
| **out**$_a$ | | $D \leftarrow \mathtt{id}_f$ | |
| **maybe**$_b$ | | $D \leftarrow \phi_f(D, \delta_f^i)$ | |
| **maybe**$_a$ | 0 | $D \leftarrow \phi_f(D, \delta_f^{i+1})$ | |
| **maybe**$_a$ | 1 | $D \leftarrow \phi_f(D, \delta_f^i)$ | |
| **found** | 0 | $D \leftarrow \mathtt{id}_f$ | $R \leftarrow \phi_g(R, \phi_f(\phi_f(D, \delta_f^i), \delta_f^{i+1}))$ |
| **found** | 1 | $D \leftarrow \mathtt{id}_f$ | $R \leftarrow \phi_g(R, \phi_f(D, \delta_f^i))$ |
| **in** | 0 | $D \leftarrow \mathtt{id}_f$ | $R \leftarrow \phi_g(R, \phi_f(D, \delta_f^{i+1}))$ |
| **in** | 1 | $D \leftarrow \mathtt{id}_f$ | $R \leftarrow \phi_g(R, \phi_f(D, \delta_f^i))$ |
| **found**$_e$ | 0 | $D \leftarrow \mathtt{id}_f$ | $R \leftarrow \phi_g(R, \phi_f(\phi_f(D, \delta_f^i), \delta_f^{i+1}))$ |
| **found**$_e$ | 1 | $D \leftarrow \mathtt{id}_f$ | $R \leftarrow \phi_g(R, \phi_f(D, \delta_f^i))$ |

Table 2.26: Table for Type: dRFoundFoundEInB

| initialization | | $D \leftarrow \mathtt{id}_f$ | $R \leftarrow \mathtt{default}_{gf}$ |
|---|---|---|---|
| **return** | | $R$ | |
| Semantic Letter | | | Decoration |
| | After | Update of $D$ | Update of $R$ |
| **out** | | | |
| **out**$_r$ | | $D \leftarrow \mathtt{id}_f$ | |
| **out**$_a$ | | $D \leftarrow \mathtt{id}_f$ | |
| **maybe**$_b$ | | $D \leftarrow \phi_f(D, \delta_f^i)$ | |
| **maybe**$_a$ | 0 | $D \leftarrow \phi_f(D, \delta_f^{i+1})$ | |
| **maybe**$_a$ | 1 | $D \leftarrow \phi_f(D, \delta_f^i)$ | |
| **found** | 0 | $D \leftarrow \phi_f(\phi_f(D, \delta_f^i), \delta_f^{i+1})$ | $R \leftarrow \phi_g(R, \phi_f(\phi_f(D, \delta_f^i), \delta_f^{i+1}))$ |
| **found** | 1 | $D \leftarrow \phi_f(D, \delta_f^i)$ | $R \leftarrow \phi_g(R, \phi_f(D, \delta_f^i))$ |
| **in** | 0 | $D \leftarrow \phi_f(D, \delta_f^i)$ | $R \leftarrow \phi_g(R, \phi_f(D, \delta_f^{i+1}))$ |
| **in** | 1 | $D \leftarrow \phi_f(D, \delta_f^i)$ | $R \leftarrow \phi_g(R, \phi_f(D, \delta_f^i))$ |
| **found**$_e$ | 0 | $D \leftarrow \mathtt{id}_f$ | $R \leftarrow \phi_g(R, \phi_f(\phi_f(D, \delta_f^i), \delta_f^{i+1}))$ |
| **found**$_e$ | 1 | $D \leftarrow \mathtt{id}_f$ | $R \leftarrow \phi_g(R, \phi_f(D, \delta_f^i))$ |

Table 2.27: Table for Type: dRFoundFoundEInMonMB

| | | |
|---|---|---|
| **initialization** | $D \leftarrow \mathtt{id}_f$ | $R \leftarrow \mathtt{default}_{gf}$ |
| **return** | $R$ | |
| Semantic Letter | | Decoration |
| | Update of $D$ | Update of $R$ |
| **out** | | |
| $\mathbf{out}_r$ | | |
| $\mathbf{out}_a$ | | |
| $\mathbf{maybe}_b$ | $D \leftarrow \delta_f^i$ | |
| $\mathbf{maybe}_a$ | | |
| **found** | | |
| **in** | | |
| $\mathbf{found_e}$ | | $R \leftarrow \phi_g(R, \phi_f(D, \delta_f^i))$ |

Table 2.28: Table for Type: dRFoundFoundEInNoMA

| initialization | | $D \leftarrow \mathtt{id}_f$ | $R \leftarrow \mathtt{default}_{gf}$ |
|---|---|---|---|
| **return** | | $R$ | |
| Semantic Letter | | | Decoration |
| | After | Update of $D$ | Update of $R$ |
| **out** | | | |
| **out**$_r$ | | $D \leftarrow \mathtt{id}_f$ | |
| **out**$_a$ | | $D \leftarrow \mathtt{id}_f$ | |
| **maybe**$_b$ | | $D \leftarrow \phi_f(D, \delta_f^i)$ | |
| **maybe**$_a$ | 0 | $D \leftarrow \phi_f(D, \delta_f^{i+1})$ | |
| **maybe**$_a$ | 1 | $D \leftarrow \phi_f(D, \delta_f^i)$ | |
| **found** | 0 | $D \leftarrow \mathtt{id}_f$ | $R \leftarrow \phi_g(R, \phi_f(\phi_f(D, \delta_f^i), \delta_f^{i+1}))$ |
| **found** | 1 | $D \leftarrow \mathtt{id}_f$ | $R \leftarrow \phi_g(R, \phi_f(D, \delta_f^i))$ |
| **in** | 0 | $D \leftarrow \mathtt{id}_f$ | $R \leftarrow \phi_g(R, \delta_f^{i+1})$ |
| **in** | 1 | $D \leftarrow \mathtt{id}_f$ | $R \leftarrow \phi_g(R, \delta_f^i)$ |
| **found**$_e$ | 0 | $D \leftarrow \mathtt{id}_f$ | $R \leftarrow \phi_g(R, \phi_f(\phi_f(D, \delta_f^i), \delta_f^{i+1}))$ |
| **found**$_e$ | 1 | $D \leftarrow \mathtt{id}_f$ | $R \leftarrow \phi_g(R, \phi_f(D, \delta_f^i))$ |

Table 2.29: Table for Type: dRFoundFoundEInNoMBA

| initialization | | $D \leftarrow \mathtt{id}_f$ | $R \leftarrow \mathtt{default}_{gf}$ |
|---|---|---|---|
| **return** | | $R$ | |
| Semantic Letter | | Decoration | |
| | After | Update of $D$ | Update of $R$ |
| **out** | | | |
| **out**$_r$ | | $D \leftarrow \mathtt{id}_f$ | |
| **out**$_a$ | | $D \leftarrow \mathtt{id}_f$ | |
| **maybe**$_b$ | | | |
| **maybe**$_a$ | 0 | $D \leftarrow \phi_f(D, \delta_f^{i+1})$ | |
| **maybe**$_a$ | 1 | $D \leftarrow \phi_f(D, \delta_f^i)$ | |
| **found** | 0 | $D \leftarrow \mathtt{id}_f$ | $R \leftarrow \phi_g(R, \phi_f(\delta_f^i, \delta_f^{i+1}))$ |
| **found** | 1 | $D \leftarrow \mathtt{id}_f$ | $R \leftarrow \phi_g(R, \delta_f^i)$ |
| **in** | 0 | $D \leftarrow \mathtt{id}_f$ | $R \leftarrow \phi_g(R, \phi_f(D, \delta_f^{i+1}))$ |
| **in** | 1 | $D \leftarrow \mathtt{id}_f$ | $R \leftarrow \phi_g(R, \phi_f(D, \delta_f^i))$ |
| **found**$_e$ | 0 | $D \leftarrow \mathtt{id}_f$ | $R \leftarrow \phi_g(R, \phi_f(\delta_f^i, \delta_f^{i+1}))$ |
| **found**$_e$ | 1 | $D \leftarrow \mathtt{id}_f$ | $R \leftarrow \phi_g(R, \delta_f^i)$ |

Table 2.30: Table for Type: dRFoundFoundEInNoMBB

| initialization | | $D \leftarrow \mathtt{id}_f$ | $R \leftarrow \mathtt{default}_{gf}$ |
|---|---|---|---|
| **return** | | $R$ | |
| Semantic Letter | | Decoration | |
| | After | Update of $D$ | Update of $R$ |
| **out** | | | |
| **out**$_r$ | | $D \leftarrow \mathtt{id}_f$ | |
| **out**$_a$ | | $D \leftarrow \mathtt{id}_f$ | |
| **maybe**$_b$ | | | |
| **maybe**$_a$ | 0 | $D \leftarrow \phi_f(D, \delta_f^{i+1})$ | |
| **maybe**$_a$ | 1 | $D \leftarrow \phi_f(D, \delta_f^i)$ | |
| **found** | 0 | $D \leftarrow \phi_f(\delta_f^i, \delta_f^{i+1})$ | $R \leftarrow \phi_g(R, \phi_f(\delta_f^i, \delta_f^{i+1}))$ |
| **found** | 1 | $D \leftarrow \delta_f^i$ | $R \leftarrow \phi_g(R, \delta_f^i)$ |
| **in** | 0 | $D \leftarrow \phi_f(D, \delta_f^i)$ | $R \leftarrow \phi_g(R, \phi_f(D, \delta_f^{i+1}))$ |
| **in** | 1 | $D \leftarrow \phi_f(D, \delta_f^i)$ | $R \leftarrow \phi_g(R, \phi_f(D, \delta_f^i))$ |
| **found**$_e$ | 0 | $D \leftarrow \mathtt{id}_f$ | $R \leftarrow \phi_g(R, \phi_f(\delta_f^i, \delta_f^{i+1}))$ |
| **found**$_e$ | 1 | $D \leftarrow \mathtt{id}_f$ | $R \leftarrow \phi_g(R, \delta_f^i)$ |

Table 2.31: Table for Type: footprint

| initialization | | $C \leftarrow 0$ |
|---|---|---|
| **return** | $p_n = 0$ | |
| Semantic Letter | Decoration | |
| | Guard | Update of $C$ |
| **out** | $p_i = 0$ | |
| **out**$_r$ | $p_i = 0$ | |
| **out**$_a$ | $p_i = 0$ | |
| **maybe**$_b$ | $p_i = p_{i+1}$ | |
| **maybe**$_a$ | $p_i = p_{i+1}$ | |
| **found**$_e$ | $p_i = C + 1$ | $C \leftarrow C + 1$ |
| **found** | $p_i = C + 1$ | $C \leftarrow C + 1$ |
| **in** | $p_i = C$ | |

Table 2.32: Table for Type: footprint_distance

| | Guard | Update of $C$ | Update of $D$ | Update of $F$ | Update of $M$ |
|---|---|---|---|---|---|
| **initialization** | | $C \leftarrow \mathtt{id}_g$ | $D \leftarrow 0$ | $F \leftarrow 0$ | $M \leftarrow 0$ |
| **return** | $C$ | | | | |
| Semantic Letter | | | Decoration | | |
| | Guard | Update of $C$ | Update of $D$ | Update of $F$ | Update of $M$ |
| **out** | | | $D \leftarrow D + 1$ | | |
| **out**$_r$ | | | $D \leftarrow D + M$ | | $M \leftarrow 0$ |
| **out**$_a$ | | | $D \leftarrow D + 1$ | | |
| **maybe**$_b$ | | | | | $M \leftarrow M + 1$ |
| **maybe**$_a$ | | | $D \leftarrow D + 1$ | | |
| **found**$_e$ | $F = 1$ | $C \leftarrow \phi_g(C, D)$ | $D \leftarrow 0$ | | |
| **found**$_e$ | $F = 0$ | | $D \leftarrow 0$ | $F \leftarrow 1$ | |
| **found** | $F = 1$ | $C \leftarrow \phi_g(C, D)$ | $D \leftarrow 0$ | | |
| **found** | $F = 0$ | | $D \leftarrow 0$ | $F \leftarrow 1$ | |
| **in** | | | $D \leftarrow 0$ | | |

Table 2.33: Table for Type: found ($<_g$ stands for $<$ when $g = \min$ and $>$ when $g = \max$)

| | |
|---|---|
| **initialization** | $f_n = 0$ <br> $C = R \Rightarrow ct_n = 1, at_n = 0$ <br> $C <_g R \Rightarrow ct_n = 0, at_n = 0$ |
| **return** | $C = R, R = \text{default}_{gf} \Rightarrow ct_n = 0, at_n = 0$ <br> $C = R, R \neq \text{default}_{gf} \Rightarrow ct_n = 1, at_n = 1$ <br> $R <_g C \Rightarrow ct_n = 0, at_n = 1$ <br> **return** $\phi_{gf}(R, C)$ |

| | | | $C \leftarrow \text{default}_{gf}$ | $D \leftarrow \text{id}_f$ | $R \leftarrow \text{default}_{gf}$ |
|---|---|---|---|---|---|
| | | | Decoration | | |
| Semantic Letter | After | Guard | Update of $C$ | Update of $D$ | Update of $R$ |
| **out** | | $f_i = 0, ct_{i+1} = ct_i, at_{i+1} = at_i$ | | $D \leftarrow \text{id}_f$ | |
| **out$_r$** | | $f_i = 0, ct_{i+1} = ct_i, at_{i+1} = at_i$ | | $D \leftarrow \text{id}_f$ | |
| **out$_a$** | | $c <_g r, f_i = 0, at_i = 0, at_{i+1} = ct_i$ | $C \leftarrow \text{default}_{gf}$ | $D \leftarrow \text{id}_f$ | $R \leftarrow \phi_g(R, C)$ |
| **out$_a$** | | $c = r, f_i = 0, at_{i+1} = ct_i, at_{i+1} = at_i$ | $C \leftarrow \text{default}_{gf}$ | $D \leftarrow \text{id}_f$ | $R \leftarrow \phi_g(R, C)$ |
| **out$_a$** | | $r <_g c, f_i = 0, ct_i = 0, at_{i+1} = at_i$ | $C \leftarrow \text{default}_{gf}$ | $D \leftarrow \text{id}_f$ | $R \leftarrow \phi_g(R, C)$ |
| **maybe$_b$** | 0 | $f_i = 0, ct_{i+1} = ct_i, at_{i+1} = at_i$ | | $D \leftarrow \phi_f(D, \delta_f^i)$ | |
| **maybe$_a$** | 0 | $f_i = 0, ct_{i+1} = ct_i, at_{i+1} = at_i$ | | $D \leftarrow \phi_f(D, \delta_f^{i+1})$ | |
| **maybe$_a$** | 1 | $f_i = 0, ct_{i+1} = ct_i, at_{i+1} = at_i$ | | $D \leftarrow \phi_f(D, \delta_f^i)$ | |
| **found** | 0 | $ct_i = 0, ct_{i+1} = f_i, at_{i+1} = at_i$ | $C \leftarrow \phi_f(D, \delta_f^i)$ | $D \leftarrow \text{id}_f$ | |
| **found** | 1 | $f_i = 0, ct_{i+1} = f_i, at_{i+1} = at_i$ | $C \leftarrow \phi_f(D, \delta_f^i)$ | $D \leftarrow \text{id}_f$ | |
| **in** | 0 | $f_i = 0, ct_{i+1} = ct_i, at_{i+1} = at_i$ | $C \leftarrow \phi_f(C, \phi_f(D, \delta_f^{i+1}))$ | $D \leftarrow \text{id}_f$ | |
| **in** | 1 | $f_i = 0, ct_{i+1} = ct_i, at_{i+1} = at_i$ | $C \leftarrow \phi_f(C, \phi_f(D, \delta_f^i))$ | $D \leftarrow \text{id}_f$ | |
| **found$_o$** | 0 | $\phi_f(\phi_f(d, \delta_f^i), \delta_f^{i+1}) <_g r, ct_i = f_i, at_i = 0, at_{i+1} = ct_i$ | | $D \leftarrow \text{id}_f$ | $R \leftarrow \phi_g(R, \phi_f(\phi_f(\phi_f(D, \delta_f^i), \delta_f^{i+1})))$ |
| **found$_o$** | 0 | $\phi_f(\phi_f(d, \delta_f^i), \delta_f^{i+1}) = r, at_{i+1} = f_i, at_{i+1} = ct_i, at_{i+1} = at_i$ | | $D \leftarrow \text{id}_f$ | $R \leftarrow \phi_g(R, \phi_f(\phi_f(\phi_f(D, \delta_f^i), \delta_f^{i+1})))$ |
| **found$_o$** | 0 | $r <_g \phi_f(\phi_f(d, \delta_f^i), \delta_f^{i+1}), f_i = 0, ct_i = 0, at_{i+1} = at_i$ | | $D \leftarrow \text{id}_f$ | $R \leftarrow \phi_g(R, \phi_f(\phi_f(\phi_f(D, \delta_f^i), \delta_f^{i+1})))$ |
| **found$_o$** | 1 | $\phi_f(d, \delta_f^i) <_g r, ct_i = f_i, at_{i+1} = 0, at_{i+1} = ct_i$ | | $D \leftarrow \text{id}_f$ | $R \leftarrow \phi_g(R, \phi_f(\phi_f(D, \delta_f^i)))$ |
| **found$_o$** | 1 | $\phi_f(d, \delta_f^i) = r, at_{i+1} = f_i, at_{i+1} = ct_i, at_{i+1} = at_i$ | | $D \leftarrow \text{id}_f$ | $R \leftarrow \phi_g(R, \phi_f(D, \delta_f^i))$ |
| **found$_o$** | 1 | $r <_g \phi_f(d, \delta_f^i), f_i = 0, ct_i = 0, at_{i+1} = at_i$ | | $D \leftarrow \text{id}_f$ | $R \leftarrow \phi_g(R, \phi_f(D, \delta_f^i))$ |

Table 2.34: Table for Type: found_distance

| initialization | | $C \leftarrow \mathtt{id}_g$ | $F \leftarrow 0$ |
|---|---|---|---|
| **return** | $C$ | | |
| Semantic Letter | | Decoration | |
| | Guard | Update of $C$ | Update of $F$ |
| **out** | | | |
| **out**$_r$ | | | |
| **out**$_a$ | | | |
| **maybe**$_b$ | | | |
| **maybe**$_a$ | | | |
| **found**$_e$ | $F > 0$ | $C \leftarrow \phi_g(C, i - F)$ | $F \leftarrow i$ |
| **found**$_e$ | $F = 0$ | | $F \leftarrow i$ |
| **found** | $F > 0$ | $C \leftarrow \phi_g(C, i - F)$ | $F \leftarrow i$ |
| **found** | $F = 0$ | | $F \leftarrow i$ |
| **in** | | | |

Table 2.35: Table for Type: function

| initialization | | $C \leftarrow \mathbf{default}_{\flat f}$ | $D \leftarrow \mathbf{id}_f$ | $R \leftarrow \mathbf{default}_{\flat f}$ |
| return | | $\phi_\flat(R,C)$ | | |
| | | | Decoration | |
| Semantic Letter | After | Update of $C$ | Update of $D$ | Update of $R$ |
|---|---|---|---|---|
| **out** | | | | |
| **out**$_r$ | | | $D \leftarrow \mathbf{id}_f$ | |
| **out**$_a$ | | $C \leftarrow \mathbf{default}_{\flat f}$ | $D \leftarrow \mathbf{id}_f$ | $R \leftarrow \phi_\flat(R,C)$ |
| **maybe**$_b$ | 0 | | $D \leftarrow \phi_f(D,\delta_f^i)$ | |
| **maybe**$_a$ | 0 | | $D \leftarrow \phi_f(D,\delta_f^{i+1})$ | |
| **maybe**$_a$ | 1 | | $D \leftarrow \phi_f(D,\delta_f^i)$ | |
| **found** | 0 | $C \leftarrow \phi_f(\phi_f(D,\delta_f^i),\delta_f^{i+1})$ | $D \leftarrow \mathbf{id}_f$ | |
| **found** | 1 | $C \leftarrow \phi_f(D,\delta_f^i)$ | $D \leftarrow \mathbf{id}_f$ | |
| **in** | 0 | $C \leftarrow \phi_f(C,\phi_f(D,\delta_f^{i+1}))$ | $D \leftarrow \mathbf{id}_f$ | |
| **in** | 1 | $C \leftarrow \phi_f(C,\phi_f(D,\delta_f^i))$ | $D \leftarrow \mathbf{id}_f$ | |
| **found**$_e$ | 0 | | $D \leftarrow \mathbf{id}_f$ | $R \leftarrow \phi_\flat(R,\phi_f(\phi_f(D,\delta_f^i),\delta_f^{i+1}))$ |
| **found**$_d$ | 1 | | $D \leftarrow \mathbf{id}_f$ | $R \leftarrow \phi_\flat(R,\phi_f(D,\delta_f^i))$ |

Table 2.36: Table for Type: predicate

| initialization | | $C \leftarrow \mathtt{default}_\theta$ | $F \leftarrow \mathtt{default}_\theta$ | $R \leftarrow 1$ |
| --- | --- | --- | --- | --- |
| return | | $R \wedge \theta(F,C)$ | | |

| | | | | Decoration | |
| Semantic Letter | After | Update of $C$ | Update of $D$ | Update of $F$ | Update of $R$ |
| --- | --- | --- | --- | --- | --- |
| **out** | | | $D \leftarrow \mathtt{id}_f$ | | |
| **out$_r$** | | | $D \leftarrow \mathtt{id}_f$ | | |
| **out$_a$** | | | $D \leftarrow \phi_f(D, \delta_f^i)$ | | |
| **maybe$_b$** | 0 | | $D \leftarrow \phi_f(D, \delta_f^{i+1})$ | $F \leftarrow C$ | $R \leftarrow R \wedge \theta(F,C)$ |
| **maybe$_a$** | 1 | | $D \leftarrow \phi_f(D, \delta_f^i)$ | | |
| **found** | 0 | $C \leftarrow \phi_f(\phi_f(D, \delta_f^i), \delta_f^{i+1})$ | $D \leftarrow \mathtt{id}_f$ | | |
| **found** | 1 | $C \leftarrow \phi_f(D, \delta_f^i)$ | $D \leftarrow \mathtt{id}_f$ | | |
| **in** | 0 | $C \leftarrow \phi_f(C, \phi_f(D, \delta_f^i))$ | $D \leftarrow \mathtt{id}_f$ | | |
| **in** | 1 | $C \leftarrow \phi_f(C, \phi_f(D, \delta_f^i))$ | $D \leftarrow \mathtt{id}_f$ | | |
| **found$_o$** | 0 | $C \leftarrow \phi_f(\phi_f(D, \delta_f^i), \delta_f^{i+1})$ | $D \leftarrow \mathtt{id}_f$ | $F \leftarrow \phi_f(\phi_f(D, \delta_f^i), \delta_f^{i+1})$ | $R \leftarrow R \wedge \theta(F, \phi_f(\phi_f(D, \delta_f^i), \delta_f^{i+1}))$ |
| **found$_o$** | 1 | $C \leftarrow \phi_f(D, \delta_f^i)$ | $D \leftarrow \mathtt{id}_f$ | $F \leftarrow \phi_f(D, \delta_f^i)$ | $R \leftarrow R \wedge \theta(F, \phi_f(D, \delta_f^i))$ |

Table 2.37: Table for Type: rFoundFoundE

| initialization | $R \leftarrow \texttt{default}_{gf}$ |
|---|---|
| **return** | $R$ |
| Semantic Letter | Decoration |
| | Update of $R$ |
| **out** | |
| **out**$_r$ | |
| **out**$_a$ | |
| **maybe**$_b$ | |
| **maybe**$_a$ | |
| **found** | $R \leftarrow \phi_g(R, \delta_f^i)$ |
| **in** | |
| **found**$_\mathbf{e}$ | $R \leftarrow \phi_g(R, \delta_f^i)$ |

Table 2.38: Table for Type: rFoundFoundEIn

| initialization | $R \leftarrow \text{default}_{gf}$ |
|---|---|
| **return** | $R$ |
| Semantic Letter | Decoration |
| | Update of $R$ |
| **out** | |
| **out**$_r$ | |
| **out**$_a$ | |
| **maybe**$_b$ | |
| **maybe**$_a$ | |
| **found** | $R \leftarrow \phi_g(R, \phi_f(\delta_f^i, \delta_f^{i+1}))$ |
| **in** | $R \leftarrow \phi_g(R, \delta_f^{i+1})$ |
| **found**$_e$ | $R \leftarrow \phi_g(R, \phi_f(\delta_f^i, \delta_f^{i+1}))$ |

Table 2.39: Table for Type: rFoundInSWZigzag

| | |
|---|---|
| **initialization** | $R \leftarrow \mathtt{default}_{gf}$ |
| **return** | $R$ |
| Semantic Letter | Decoration |
| | Update of $R$ |
| **out** | |
| **out**$_r$ | |
| **out**$_a$ | |
| **maybe**$_h$ | |
| **maybe**$_a$ | |
| **found** | $R \leftarrow \phi_g(\phi_g(R, \delta_f^i), \delta_f^{i+1})$ |
| **in** | $R \leftarrow \phi_g(R, \delta_f^i)$ |
| **found**$_e$ | |

Table 2.40: Table for Type: rangeCHRFoundEOutA

| initialization | $C \leftarrow \texttt{default}_{gf}$ | $H \leftarrow \texttt{VAR}_1$ | $R \leftarrow \texttt{default}_{gf}$ |
|---|---|---|---|
| **return** | $\phi_g(R, C)$ | | |
| Semantic Letter | | Decoration | |
| | Update of $C$ | Update of $H$ | Update of $R$ |
| **out** | | $H \leftarrow \delta_f^{i+1}$ | |
| **out**$_r$ | | $H \leftarrow \delta_f^{i+1}$ | |
| **out**$_a$ | $C \leftarrow \texttt{default}_{gf}$ | $H \leftarrow \delta_f^{i+1}$ | $R \leftarrow \phi_g(R, C)$ |
| **maybe**$_b$ | | | |
| **maybe**$_a$ | | | |
| **found** | $C \leftarrow H - \delta_f^{i+1}$ | | |
| **in** | $C \leftarrow H - \delta_f^{i+1}$ | | |
| **found**$_\mathbf{e}$ | | $H \leftarrow \delta_f^{i+1}$ | $R \leftarrow \phi_g(R, H - \delta_f^{i+1})$ |

Table 2.41: Table for Type: rangeCHRFoundEOutB

| | | | |
|---|---|---|---|
| **initialization** | $C \leftarrow \texttt{default}_{gf}$ | $H \leftarrow \texttt{VAR}_1$ | $R \leftarrow \texttt{default}_{gf}$ |
| **return** | $\phi_g(R, C)$ | | |
| Semantic Letter | Decoration | | |
| | Update of $C$ | Update of $H$ | Update of $R$ |
| **out** | | $H \leftarrow \delta_f^{i+1}$ | |
| **out**$_r$ | | $H \leftarrow \delta_f^{i+1}$ | |
| **out**$_a$ | $C \leftarrow \texttt{default}_{gf}$ | $H \leftarrow \delta_f^{i+1}$ | $R \leftarrow \phi_g(R, C)$ |
| **maybe**$_b$ | | | |
| **maybe**$_a$ | | | |
| **found** | $C \leftarrow \delta_f^{i+1} - H$ | | |
| **in** | $C \leftarrow \delta_f^{i+1} - H$ | | |
| **found**$_e$ | | $H \leftarrow \delta_f^{i+1}$ | $R \leftarrow \phi_g(R, \delta_f^{i+1} - H)$ |

Table 2.42: Table for Type: rangeHRFoundFoundEInA

| initialization | $H \leftarrow \text{VAR}_1$ | $R \leftarrow \text{default}_{gf}$ |
|---|---|---|
| **return** | $R$ | |
| Semantic Letter | | Decoration |
| | Update of $H$ | Update of $R$ |
| **out** | $H \leftarrow \delta_f^{i+1}$ | |
| **out**$_r$ | $H \leftarrow \delta_f^{i+1}$ | |
| **out**$_a$ | $H \leftarrow \delta_f^{i+1}$ | |
| **maybe**$_b$ | | |
| **maybe**$_a$ | | |
| **found** | | $R \leftarrow \phi_g(R, H - \delta_f^{i+1})$ |
| **in** | | $R \leftarrow \phi_g(R, H - \delta_f^{i+1})$ |
| **found**$_e$ | $H \leftarrow \delta_f^{i+1}$ | $R \leftarrow \phi_g(R, H - \delta_f^{i+1})$ |

Table 2.43: Table for Type: rangeHRFoundFoundEInB

| initialization | $H \leftarrow \mathtt{VAR}_1$ | $R \leftarrow \mathtt{default}_{gf}$ |
|---|---|---|
| **return** | $R$ | |
| Semantic Letter | | Decoration |
| | Update of $H$ | Update of $R$ |
| **out** | $H \leftarrow \delta_f^{i+1}$ | |
| **out**$_r$ | $H \leftarrow \delta_f^{i+1}$ | |
| **out**$_a$ | $H \leftarrow \delta_f^{i+1}$ | |
| **maybe**$_b$ | | |
| **maybe**$_a$ | | |
| **found** | | $R \leftarrow \phi_g(R, \delta_f^{i+1} - H)$ |
| **in** | | $R \leftarrow \phi_g(R, \delta_f^{i+1} - H)$ |
| **found**$_\mathbf{e}$ | $H \leftarrow \delta_f^{i+1}$ | $R \leftarrow \phi_g(R, \delta_f^{i+1} - H)$ |

Table 2.44: Table for Type: rangeRFoundFoundEInA

| | |
|---|---|
| **initialization** | $R \leftarrow \mathtt{default}_{gf}$ |
| **return** | $R$ |
| Semantic Letter | Decoration |
| | Update of $R$ |
| **out** | |
| **out**$_r$ | |
| **out**$_a$ | |
| **maybe**$_b$ | |
| **maybe**$_a$ | |
| **found** | $R \leftarrow \phi_g(R, \max(0, \delta_f^i - \delta_f^{i+1}))$ |
| **in** | $R \leftarrow \phi_g(R, \max(0, \delta_f^i - \delta_f^{i+1}))$ |
| **found**$_\mathbf{e}$ | $R \leftarrow \phi_g(R, \max(0, \delta_f^i - \delta_f^{i+1}))$ |

Table 2.45: Table for Type: rangeRFoundFoundEInB

| initialization | $R \leftarrow \texttt{default}_{gf}$ |
|---|---|
| **return** | $R$ |
| Semantic Letter | Decoration |
| | Update of $R$ |
| **out** | |
| **out**$_r$ | |
| **out**$_a$ | |
| **maybe**$_h$ | |
| **maybe**$_a$ | |
| **found** | $R \leftarrow \phi_g(R, \max(0, \delta_f^{i+1} - \delta_f^i))$ |
| **in** | $R \leftarrow \phi_g(R, \max(0, \delta_f^{i+1} - \delta_f^i))$ |
| **found**$_e$ | $R \leftarrow \phi_g(R, \max(0, \delta_f^{i+1} - \delta_f^i))$ |

Table 2.46: Table for Type: range_function

| initialization | $C \leftarrow \texttt{default}_{gf}$ | $H \leftarrow \texttt{VAR}_1$ | $R \leftarrow \texttt{default}_{gf}$ |
|---|---|---|---|
| **return** | $\phi_g(R, C)$ | | |
| Semantic Letter | | Decoration | |
| | Update of $C$ | Update of $H$ | Update of $R$ |
| **out** | | $H \leftarrow \delta_f^{i+1}$ | |
| **out**$_r$ | | $H \leftarrow \delta_f^{i+1}$ | |
| **out**$_a$ | $C \leftarrow \texttt{default}_{gf}$ | $H \leftarrow \delta_f^{i+1}$ | $R \leftarrow \phi_g(R, C)$ |
| **maybe**$_b$ | | | |
| **maybe**$_a$ | | | |
| **found**$_\mathbf{e}$ | | $H \leftarrow \delta_f^{i+1}$ | $R \leftarrow \phi_g(R, |H - \delta_f^{i+1}|)$ |
| **found** | $C \leftarrow |H - \delta_f^{i+1}|$ | | |
| **in** | $C \leftarrow |H - \delta_f^{i+1}|$ | | |

Table 2.47: Table for Type: range_predicate

| **initialization** | $C \leftarrow \texttt{default}_\theta$ | $F \leftarrow \texttt{default}_\theta$ | $H \leftarrow \texttt{VAR}_1$ | $R \leftarrow 1$ |
|---|---|---|---|---|
| **return** | $R \wedge \theta(F,C)$ | | | |
| Semantic Letter | Decoration | | | |
| | Update of $C$ | Update of $F$ | Update of $H$ | Update of $R$ |
| **out** | | | $H \leftarrow \delta_f^{i+1}$ | |
| **out**$_r$ | | | $H \leftarrow \delta_f^{i+1}$ | |
| **out**$_a$ | | $F \leftarrow |H - \delta_f^i|$ | $H \leftarrow \delta_f^{i+1}$ | $R \leftarrow R \wedge \theta(F, C)$ |
| **maybe**$_b$ | | | | |
| **maybe**$_a$ | | | | |
| **found**$_e$ | $C \leftarrow |H - \delta_f^{i+1}|$ | $F \leftarrow |H - \delta_f^{i+1}|$ | $H \leftarrow \delta_f^{i+1}$ | $R \leftarrow R \wedge \theta(F, |H - \delta_f^{i+1}|)$ |
| **found** | $C \leftarrow |H - \delta_f^{i+1}|$ | | | |
| **in** | $C \leftarrow |H - \delta_f^{i+1}|$ | | | |

# 3

# Global Constraint Catalogue

*"Mon problème, avec les classements, c'est qu'ils ne durent pas ; à peine ai-je fini de mettre de l'ordre que cet ordre est déjà caduc."*

– Georges Perec, *Penser/Classer*

## Contents

# 3.1 ALL_EQUAL_HEIGHT_DECREASING_TERRACE

▶ ▷　　　　DESCRIPTION　　　AUTOMATON

| | |
|---|---|
| **Origin** | Based on the DECREASING_TERRACE pattern. |
| **Constraint** | ALL_EQUAL_HEIGHT_DECREASING_TERRACE(VARIABLES) |
| **Argument** | VARIABLES : collection(var−dvar) |
| **Restriction** | required(VARIABLES, var) |

**Purpose**

Succeeds if the minima of the values in each occurrence of the DECREASING_TERRACE pattern in the time-series given by the VARIABLES collection are all the same.

An occurrence of the pattern DECREASING_TERRACE is the *maximal* subsequence which matches the regular expression $>=^+>$.

Assume that the occurrence of the pattern DECREASING_TERRACE starts at position $i$ and ends at position $j$. The feature MIN computes the minimum of the values from index $i + 1$ to index $j$.

**Example**

$(\langle 6, 3, 3, 1, 5, 4, 3, 3, 2, 4, 4, 5, 3, 3, 1, 1 \rangle)$

Figure 3.1 provides an example where the ALL_EQUAL_HEIGHT_DECREASING_TERRACE ($[6, 3, 3, 1, 5, 4, 3, 3, 2, 4, 4, 5, 3, 3, 1, 1]$) constraint holds.

Figure 3.1: Illustrating the ALL_EQUAL_HEIGHT_DECREASING_TERRACE constraint of the **Example** slot

**Typical**
$|\texttt{VARIABLES}| > 3$
$\texttt{range}(\texttt{VARIABLES.var}) > 2$

**Automaton**    Figure 3.2 depicts the automaton associated with the constraint ALL_EQUAL_HEIGHT_DECREASING_TERRACE.

Figure 3.2: Automaton for the ALL_EQUAL_HEIGHT_DECREASING_TERRACE constraint obtained by applying decoration Table 2.36 to the seed transducer of the DECREASING_TERRACE pattern

## 3.2 ALL_EQUAL_HEIGHT_INCREASING_TERRACE

► ▷ ◄        DESCRIPTION        AUTOMATON

**Origin**         Based on the INCREASING_TERRACE pattern.

**Constraint**     ALL_EQUAL_HEIGHT_INCREASING_TERRACE(VARIABLES)

**Argument**       VARIABLES : collection(var−dvar)

**Restriction**    required(VARIABLES, var)

**Purpose**        Succeeds if the minima of the values in each occurrence of the INCREASING_TERRACE pattern in the time-series given by the VARIABLES collection are all the same.
An occurrence of the pattern INCREASING_TERRACE is the *maximal* subsequence which matches the regular expression $< =^+ <$.
Assume that the occurrence of the pattern INCREASING_TERRACE starts at position $i$ and ends at position $j$. The feature MIN computes the minimum of the values from index $i + 1$ to index $j$.

**Example**        $(\langle 1, 4, 4, 6, 2, 3, 4, 4, 5, 3, 3, 2, 4, 4, 6, 6 \rangle)$

Figure 3.3 provides an example where the ALL_EQUAL_HEIGHT_INCREASING_TERRACE ($[1, 4, 4, 6, 2, 3, 4, 4, 5, 3, 3, 2, 4, 4, 6, 6]$) constraint holds.

Figure 3.3: Illustrating the ALL_EQUAL_HEIGHT_INCREASING_TERRACE constraint of the **Example** slot

**Typical**

$|\mathtt{VARIABLES}| > 3$
$\mathtt{range}(\mathtt{VARIABLES.var}) > 2$

**Automaton**    Figure 3.4 depicts the automaton associated with the constraint ALL_EQUAL_HEIGHT_INCREASING_TERRACE.

Figure 3.4: Automaton for the ALL_EQUAL_HEIGHT_INCREASING_TERRACE constraint obtained by applying decoration Table 2.36 to the seed transducer of the INCREASING_TERRACE pattern

## 3.3 ALL_EQUAL_HEIGHT_PLAIN

► ▷ ◄        DESCRIPTION        AUTOMATON

| | |
|---|---|
| **Origin** | Based on the PLAIN pattern. |
| **Constraint** | ALL_EQUAL_HEIGHT_PLAIN(VARIABLES) |
| **Argument** | VARIABLES : collection(var−dvar) |
| **Restriction** | required(VARIABLES, var) |
| **Purpose** | Succeeds if the minima of the values in each occurrence of the PLAIN pattern in the time-series given by the VARIABLES collection are all the same.<br><br>An occurrence of the pattern PLAIN is the *maximal* subsequence which matches the regular expression $> =^* <$.<br><br>Assume that the occurrence of the pattern PLAIN starts at position $i$ and ends at position $j$. The feature MIN computes the minimum of the values from index $i + 1$ to index $j$. |
| **Example** | $(\langle 1, 6, 3, 3, 7, 6, 6, 3, 3, 5, 5, 4, 3, 3, 6, 3 \rangle)$ |

Figure 3.5 provides an example where the ALL_EQUAL_HEIGHT_PLAIN ([1, 6, 3, 3, 7, 6, 6, 3, 3, 5, 5, 4, 3, 3, 6, 3]) constraint holds.

Figure 3.5: Illustrating the ALL_EQUAL_HEIGHT_PLAIN constraint of the **Example** slot

**Typical**          $|\texttt{VARIABLES}| > 2$
                     $\texttt{range}(\texttt{VARIABLES.var}) > 1$

**Automaton**       Figure 3.6 depicts the automaton associated with the constraint
ALL_EQUAL_HEIGHT_PLAIN.

Figure 3.6: Automaton for the ALL_EQUAL_HEIGHT_PLAIN constraint obtained by applying decoration Table 2.36 to the seed transducer of the PLAIN pattern

## 3.4  ALL_EQUAL_HEIGHT_PLATEAU

▶  ▷     ◀        DESCRIPTION          AUTOMATON

**Origin**          Based on the PLATEAU pattern.

**Constraint**      ALL_EQUAL_HEIGHT_PLATEAU(VARIABLES)

**Argument**        VARIABLES : collection(var−dvar)

**Restriction**     required(VARIABLES, var)

**Purpose**         Succeeds if the minima of the values in each occurrence of the PLATEAU pattern in the
                    time-series given by the VARIABLES collection are all the same.

                    An occurrence of the pattern PLATEAU is the *maximal* subsequence which matches the
                    regular expression $< =^* >$.

                    Assume that the occurrence of the pattern PLATEAU starts at position $i$ and ends at position $j$. The feature MIN computes the minimum of the values from index $i + 1$ to index $j$.

**Example**         $(\langle 7, 2, 5, 5, 1, 2, 2, 5, 5, 3, 3, 4, 5, 5, 2, 5 \rangle)$

Figure 3.7 provides an example where the ALL_EQUAL_HEIGHT_PLATEAU
$([7, 2, 5, 5, 1, 2, 2, 5, 5, 3, 3, 4, 5, 5, 2, 5])$ constraint holds.

**Typical**         $|\texttt{VARIABLES}| > 2$
                    range(VARIABLES.var) > 1

Figure 3.7: Illustrating the ALL_EQUAL_HEIGHT_PLATEAU constraint of the **Example** slot

**Automaton**   Figure 3.8 depicts the automaton associated with the constraint ALL_EQUAL_HEIGHT_PLATEAU.

Figure 3.8: Automaton for the ALL_EQUAL_HEIGHT_PLATEAU constraint obtained by applying decoration Table 2.36 to the seed transducer of the PLATEAU pattern

## 3.5 ALL_EQUAL_HEIGHT_PROPER_PLAIN

          DESCRIPTION          AUTOMATON

| | |
|---|---|
| **Origin** | Based on the PROPER_PLAIN pattern. |
| **Constraint** | ALL_EQUAL_HEIGHT_PROPER_PLAIN(VARIABLES) |
| **Argument** | VARIABLES : collection(var−dvar) |
| **Restriction** | required(VARIABLES, var) |

**Purpose**

Succeeds if the minima of the values in each occurrence of the PROPER_PLAIN pattern in the time-series given by the VARIABLES collection are all the same.

An occurrence of the pattern PROPER_PLAIN is the *maximal* subsequence which matches the regular expression $> =^+ <$.

Assume that the occurrence of the pattern PROPER_PLAIN starts at position $i$ and ends at position $j$. The feature MIN computes the minimum of the values from index $i + 1$ to index $j$.

**Example**

$(\langle 2, 7, 3, 3, 6, 6, 3, 7, 3, 3, 5, 6, 5, 3, 3, 5 \rangle)$

Figure 3.9 provides an example where the ALL_EQUAL_HEIGHT_PROPER_PLAIN ($[2, 7, 3, 3, 6, 6, 3, 7, 3, 3, 5, 6, 5, 3, 3, 5]$) constraint holds.

Figure 3.9: Illustrating the ALL_EQUAL_HEIGHT_PROPER_PLAIN constraint of the **Example** slot

**Typical**

$|\text{VARIABLES}| > 3$
$\text{range}(\text{VARIABLES.var}) > 1$

**Automaton**                Figure   3.10   depicts   the   automaton   associated   with   the   constraint
ALL_EQUAL_HEIGHT_PROPER_PLAIN.

Figure 3.10: Automaton for the ALL_EQUAL_HEIGHT_PROPER_PLAIN constraint obtained by applying decoration Table 2.36 to the seed transducer of the PROPER_PLAIN pattern

## 3.6   ALL_EQUAL_HEIGHT_PROPER_PLATEAU

▶ ▷   ◀          DESCRIPTION          AUTOMATON

| **Origin** | Based on the PROPER_PLATEAU pattern. |
| **Constraint** | ALL_EQUAL_HEIGHT_PROPER_PLATEAU(VARIABLES) |
| **Argument** | VARIABLES : collection(var−dvar) |
| **Restriction** | required(VARIABLES, var) |
| **Purpose** | Succeeds if the minima of the values in each occurrence of the PROPER_PLATEAU pattern in the time-series given by the VARIABLES collection are all the same.<br><br>An occurrence of the pattern PROPER_PLATEAU is the *maximal* subsequence which matches the regular expression $< =^+ >$.<br><br>Assume that the occurrence of the pattern PROPER_PLATEAU starts at position $i$ and ends at position $j$. The feature MIN computes the minimum of the values from index $i + 1$ to index $j$. |
| **Example** | $(\langle 7, 1, 5, 5, 2, 2, 5, 1, 5, 5, 3, 2, 3, 5, 5, 3 \rangle)$ |

Figure 3.11 provides an example where the ALL_EQUAL_HEIGHT_PROPER_PLATEAU
$([7, 1, 5, 5, 2, 2, 5, 1, 5, 5, 3, 2, 3, 5, 5, 3])$ constraint holds.

| **Typical** | $\lvert$VARIABLES$\rvert > 3$<br>range(VARIABLES.var) > 1 |

Figure 3.11: Illustrating the ALL_EQUAL_HEIGHT_PROPER_PLATEAU constraint of the **Example** slot

**Automaton**         Figure  3.12  depicts  the  automaton  associated  with  the  constraint
ALL_EQUAL_HEIGHT_PROPER_PLATEAU.

Figure 3.12:  Automaton for the ALL_EQUAL_HEIGHT_PROPER_PLATEAU con-
straint obtained by applying decoration Table 2.36 to the seed transducer of the
PROPER_PLATEAU pattern

## 3.7  ALL_EQUAL_HEIGHT_STEADY

► ▷ ◄          DESCRIPTION          AUTOMATON

| | |
|---|---|
| **Origin** | Based on the STEADY pattern. |
| **Constraint** | ALL_EQUAL_HEIGHT_STEADY(VARIABLES) |
| **Argument** | VARIABLES : collection(var−dvar) |
| **Restriction** | required(VARIABLES, var) |
| **Purpose** | Succeeds if the minima of the values in each occurrence of the STEADY pattern in the time-series given by the VARIABLES collection are all the same. <br> An occurrence of the pattern STEADY is the subsequence which matches the regular expression =. <br> Assume that the occurrence of the pattern STEADY starts at position $i$ and ends at position $j$. The feature MIN computes the minimum of the values from index $i$ to index $j + 1$. |
| **Example** | $(\langle 4, 4, 4, 2, 6, 4, 4, 4, 4, 3, 2, 1, 4, 4, 3, 4 \rangle)$ |

Figure 3.13 provides an example where the ALL_EQUAL_HEIGHT_STEADY ($[4, 4, 4, 2, 6, 4, 4, 4, 4, 3, 2, 1, 4, 4, 3, 4]$) constraint holds.

Figure 3.13: Illustrating the ALL_EQUAL_HEIGHT_STEADY constraint of the **Example** slot

**Typical**                   $|VARIABLES| > 1$

**Automaton**  Figure 3.14 depicts the automaton associated with the constraint ALL_EQUAL_HEIGHT_STEADY.

Figure 3.14: Automaton for the ALL_EQUAL_HEIGHT_STEADY constraint obtained by applying decoration Table 2.36 to the seed transducer of the STEADY pattern

## 3.8   ALL_EQUAL_HEIGHT_STEADY_SEQUENCE

► ▷ ◄         DESCRIPTION          AUTOMATON

**Origin**          Based on the STEADY_SEQUENCE pattern.

**Constraint**      ALL_EQUAL_HEIGHT_STEADY_SEQUENCE(VARIABLES)

**Argument**        VARIABLES : collection(var−dvar)

**Restriction**     required(VARIABLES, var)

**Purpose**         Succeeds if the minima of the values in each occurrence of the STEADY_SEQUENCE
                    pattern in the time-series given by the VARIABLES collection are all the same.
                    An occurrence of the pattern STEADY_SEQUENCE is the *maximal* subsequence which
                    matches the regular expression $=^+$.
                    Assume that the occurrence of the pattern STEADY_SEQUENCE starts at position $i$ and
                    ends at position $j$. The feature MIN computes the minimum of the values from index $i$
                    to index $j + 1$.

**Example**         $(\langle 4, 3, 5, 5, 5, 3, 1, 1, 5, 5, 6, 5, 5, 5, 5, 3, 5 \rangle)$

Figure 3.15 provides an example where the ALL_EQUAL_HEIGHT_STEADY_SEQUENCE
($[4, 3, 5, 5, 5, 3, 1, 1, 5, 5, 6, 5, 5, 5, 5, 3, 5]$) constraint holds.

Figure 3.15: Illustrating the ALL_EQUAL_HEIGHT_STEADY_SEQUENCE constraint of
the **Example** slot

**Typical**          $|\text{VARIABLES}| > 1$

**Automaton**          Figure 3.16 depicts the automaton associated with the constraint ALL_EQUAL_HEIGHT_STEADY_SEQUENCE.

Figure 3.16: Automaton for the ALL_EQUAL_HEIGHT_STEADY_SEQUENCE constraint obtained by applying decoration Table 2.36 to the seed transducer of the STEADY_SEQUENCE pattern

## 3.9  ALL_EQUAL_MAX_BUMP_ON_DECREASING_SEQUENCE

► ▷ ◄         DESCRIPTION          AUTOMATON

| | |
|---|---|
| **Origin** | Based on the BUMP_ON_DECREASING_SEQUENCE pattern. |
| **Constraint** | ALL_EQUAL_MAX_BUMP_ON_DECREASING_SEQUENCE(VARIABLES) |
| **Argument** | VARIABLES : collection(var−dvar) |
| **Restriction** | required(VARIABLES, var) |

**Purpose**

Succeeds if the maxima of the values in each occurrence of the BUMP_ON_DECREASING_SEQUENCE pattern in the time-series given by the VARIABLES collection are all the same.

An occurrence of the pattern BUMP_ON_DECREASING_SEQUENCE is the subsequence which matches the regular expression >><>>.

Assume that the occurrence of the pattern BUMP_ON_DECREASING_SEQUENCE starts at position $i$ and ends at position $j$. The feature MAX computes the maximum of the values from index $i + 2$ to index $j$.

**Example**

$$(\langle 7, 6, 5, 6, 5, 4, 1, 4, 7, 5, 4, 2, 6, 5, 4, 4 \rangle)$$

Figure 3.17 provides an example where the ALL_EQUAL_MAX_BUMP_ON_DECREASING_SEQUENCE ([7, 6, 5, 6, 5, 4, 1, 4, 7, 5, 4, 2, 6, 5, 4, 4]) constraint holds.

**Typical**

|VARIABLES| > 5
range(VARIABLES.var) > 2

Figure 3.17: Illustrating the ALL_EQUAL_MAX_BUMP_ON_DECREASING_SEQUENCE constraint of the **Example** slot

**Automaton**          Figure   3.18   depicts   the   automaton   associated   with   the   constraint
ALL_EQUAL_MAX_BUMP_ON_DECREASING_SEQUENCE.

Figure 3.18: Automaton for the ALL_EQUAL_MAX_BUMP_ON_DECREASING_SEQUENCE constraint obtained by applying decoration Table 2.36 to the seed transducer of the BUMP_ON_DECREASING_SEQUENCE pattern

## 3.10  ALL_EQUAL_MAX_DECREASING

► ▷ ◄          DESCRIPTION          AUTOMATON

| | |
|---|---|
| **Origin** | Based on the DECREASING pattern. |
| **Constraint** | ALL_EQUAL_MAX_DECREASING(VARIABLES) |
| **Argument** | VARIABLES : collection(var−dvar) |
| **Restriction** | required(VARIABLES, var) |
| **Purpose** | Succeeds if the maxima of the values in each occurrence of the DECREASING pattern in the time-series given by the VARIABLES collection are all the same.<br><br>An occurrence of the pattern DECREASING is the subsequence which matches the regular expression >.<br><br>Assume that the occurrence of the pattern DECREASING starts at position $i$ and ends at position $j$. The feature MAX computes the maximum of the values from index $i$ to index $j + 1$. |
| **Example** | $(\langle 1, 6, 6, 1, 1, 4, 6, 6, 4, 4, 4, 5, 5, 6, 1, 5 \rangle)$ |

Figure 3.19 provides an example where the ALL_EQUAL_MAX_DECREASING $([1, 6, 6, 1, 1, 4, 6, 6, 4, 4, 4, 5, 5, 6, 1, 5])$ constraint holds.

Figure 3.19: Illustrating the ALL_EQUAL_MAX_DECREASING constraint of the **Example** slot

**Typical**          $|\texttt{VARIABLES}| > 1$
                     $\texttt{range}(\texttt{VARIABLES.var}) > 1$

**Automaton**    Figure 3.20 depicts the automaton associated with the constraint ALL_EQUAL_MAX_DECREASING.

$$\left\{ \begin{array}{l} C \leftarrow X \\ D \leftarrow -\infty \\ F \leftarrow X \\ R \leftarrow 1 \end{array} \right\}$$

$$\left\{ \begin{array}{l} C \leftarrow \max(\max(D, \mathtt{VAR}_i), \mathtt{VAR}_{i+1}) \\ D \leftarrow -\infty \\ F \leftarrow \max(\max(D, \mathtt{VAR}_i), \mathtt{VAR}_{i+1}) \\ R \leftarrow R \wedge (F = \max(\max(D, \mathtt{VAR}_i), \mathtt{VAR}_{i+1})) \end{array} \right\}$$

$>$

$\geq \atop \leq$ $s$

$\leq$

$R \wedge (F = C)$

Figure 3.20: Automaton for the ALL_EQUAL_MAX_DECREASING constraint obtained by applying decoration Table 2.36 to the seed transducer of the DECREASING pattern

# 3.11 ALL_EQUAL_MAX_DECREASING_SEQUENCE

► ▷ ◄         **DESCRIPTION**          **AUTOMATON**

| | |
|---|---|
| **Origin** | Based on the DECREASING_SEQUENCE pattern. |
| **Constraint** | ALL_EQUAL_MAX_DECREASING_SEQUENCE(VARIABLES) |
| **Argument** | VARIABLES : collection(var−dvar) |
| **Restriction** | required(VARIABLES, var) |

**Purpose**

Succeeds if the maxima of the values in each occurrence of the DECREAS-ING_SEQUENCE pattern in the time-series given by the VARIABLES collection are all the same.

An occurrence of the pattern DECREASING_SEQUENCE is the *maximal* subsequence which matches the regular expression $> (> | =)^* > | >$.

Assume that the occurrence of the pattern DECREASING_SEQUENCE starts at position $i$ and ends at position $j$. The feature MAX computes the maximum of the values from index $i$ to index $j + 1$.

**Example**

$$(\langle 1, 6, 6, 1, 1, 4, 6, 6, 4, 4, 2, 1, 3, 6, 1, 5 \rangle)$$

Figure 3.21 provides an example where the ALL_EQUAL_MAX_DECREASING_SEQUENCE ($[1, 6, 6, 1, 1, 4, 6, 6, 4, 4, 2, 1, 3, 6, 1, 5]$) constraint holds.

**Typical**

$|\text{VARIABLES}| > 1$
$\text{range}(\text{VARIABLES.var}) > 1$

Figure 3.21: Illustrating the ALL_EQUAL_MAX_DECREASING_SEQUENCE constraint of the **Example** slot

**Automaton**    Figure 3.22 depicts the automaton associated with the constraint ALL_EQUAL_MAX_DECREASING_SEQUENCE.

Figure 3.22: Automaton for the ALL_EQUAL_MAX_DECREASING_SEQUENCE constraint obtained by applying decoration Table 2.36 to the seed transducer of the DECREASING_SEQUENCE pattern

## 3.12 ALL_EQUAL_MAX_DIP_ON_INCREASING_SEQUENCE

► ▷ ◄        DESCRIPTION        AUTOMATON

| | |
|---|---|
| **Origin** | Based on the DIP_ON_INCREASING_SEQUENCE pattern. |
| **Constraint** | ALL_EQUAL_MAX_DIP_ON_INCREASING_SEQUENCE(VARIABLES) |
| **Argument** | VARIABLES : collection(var−dvar) |
| **Restriction** | required(VARIABLES, var) |

**Purpose**

Succeeds if the maxima of the values in each occurrence of the DIP_ON_INCREASING_SEQUENCE pattern in the time-series given by the VARIABLES collection are all the same.

An occurrence of the pattern DIP_ON_INCREASING_SEQUENCE is the subsequence which matches the regular expression `<<><<`.

Assume that the occurrence of the pattern DIP_ON_INCREASING_SEQUENCE starts at position $i$ and ends at position $j$. The feature MAX computes the maximum of the values from index $i + 2$ to index $j$.

**Example**

$$(\langle 0, 1, 2, 1, 5, 6, 7, 3, 0, 2, 3, 5, 0, 2, 3, 3 \rangle)$$

Figure 3.23 provides an example where the ALL_EQUAL_MAX_DIP_ON_INCREASING_SEQUENCE ([0, 1, 2, 1, 5, 6, 7, 3, 0, 2, 3, 5, 0, 2, 3, 3]) constraint holds.

**Typical**

$|\text{VARIABLES}| > 5$
$\text{range}(\text{VARIABLES.var}) > 2$

Figure 3.23: Illustrating the ALL_EQUAL_MAX_DIP_ON_INCREASING_SEQUENCE constraint of the **Example** slot

**Automaton**          Figure 3.24 depicts the automaton associated with the constraint ALL_EQUAL_MAX_DIP_ON_INCREASING_SEQUENCE.

Figure 3.24: Automaton for the ALL_EQUAL_MAX_DIP_ON_INCREASING_SEQUENCE constraint obtained by applying decoration Table 2.36 to the seed transducer of the DIP_ON_INCREASING_SEQUENCE pattern

## 3.13   ALL_EQUAL_MAX_INCREASING

►  ▷  ◄          **DESCRIPTION**          **AUTOMATON**

| | |
|---|---|
| **Origin** | Based on the INCREASING pattern. |
| **Constraint** | ALL_EQUAL_MAX_INCREASING(VARIABLES) |
| **Argument** | VARIABLES : collection(var−dvar) |
| **Restriction** | required(VARIABLES, var) |
| **Purpose** | Succeeds if the maxima of the values in each occurrence of the INCREASING pattern in the time-series given by the VARIABLES collection are all the same. <br><br> An occurrence of the pattern INCREASING is the subsequence which matches the regular expression <. <br><br> Assume that the occurrence of the pattern INCREASING starts at position $i$ and ends at position $j$. The feature MAX computes the maximum of the values from index $i$ to index $j + 1$. |
| **Example** | $(\langle 6, 1, 1, 6, 6, 6, 5, 6, 5, 3, 3, 6, 4, 1, 6, 2 \rangle)$ |

Figure 3.25 provides an example where the ALL_EQUAL_MAX_INCREASING $([6, 1, 1, 6, 6, 6, 5, 6, 5, 3, 3, 6, 4, 1, 6, 2])$ constraint holds.

Figure 3.25: Illustrating the ALL_EQUAL_MAX_INCREASING constraint of the **Example** slot

**Typical**     $|\texttt{VARIABLES}| > 1$
                $\texttt{range}(\texttt{VARIABLES.var}) > 1$

**Automaton**    Figure 3.26 depicts the automaton associated with the constraint ALL_EQUAL_MAX_INCREASING.

Figure 3.26: Automaton for the ALL_EQUAL_MAX_INCREASING constraint obtained by applying decoration Table 2.36 to the seed transducer of the INCREASING pattern

## 3.14 ALL_EQUAL_MAX_INCREASING_SEQUENCE

          DESCRIPTION          AUTOMATON

| | |
|---|---|
| **Origin** | Based on the INCREASING_SEQUENCE pattern. |
| **Constraint** | ALL_EQUAL_MAX_INCREASING_SEQUENCE(VARIABLES) |
| **Argument** | VARIABLES : collection(var−dvar) |
| **Restriction** | required(VARIABLES, var) |

**Purpose**

Succeeds if the maxima of the values in each occurrence of the INCREASING_SEQUENCE pattern in the time-series given by the VARIABLES collection are all the same.

An occurrence of the pattern INCREASING_SEQUENCE is the *maximal* subsequence which matches the regular expression $< (< | =)^* < | <$.

Assume that the occurrence of the pattern INCREASING_SEQUENCE starts at position $i$ and ends at position $j$. The feature MAX computes the maximum of the values from index $i$ to index $j + 1$.

**Example**

$$(\langle 6, 1, 1, 6, 6, 3, 1, 1, 3, 3, 5, 6, 4, 1, 6, 2 \rangle)$$

Figure 3.27 provides an example where the ALL_EQUAL_MAX_INCREASING_SEQUENCE $([6, 1, 1, 6, 6, 3, 1, 1, 3, 3, 5, 6, 4, 1, 6, 2])$ constraint holds.

Figure 3.27: Illustrating the ALL_EQUAL_MAX_INCREASING_SEQUENCE constraint of the **Example** slot

**Typical**

$|\mathtt{VARIABLES}| > 1$
$\mathtt{range}(\mathtt{VARIABLES.var}) > 1$

**Automaton**    Figure   3.28   depicts   the   automaton   associated   with   the   constraint
ALL_EQUAL_MAX_INCREASING_SEQUENCE.

Figure 3.28:  Automaton for the ALL_EQUAL_MAX_INCREASING_SEQUENCE constraint obtained by applying decoration Table 2.36 to the seed transducer of the INCREASING_SEQUENCE pattern

# 3.15  ALL_EQUAL_MAX_INFLEXION

► ▷ ◄          DESCRIPTION          AUTOMATON

$< (< | =)^* > | > (> | =)^* <$

| | |
|---|---|
| **Origin** | Based on the INFLEXION pattern. |
| **Constraint** | ALL_EQUAL_MAX_INFLEXION(VARIABLES) |
| **Argument** | VARIABLES : collection(var−dvar) |
| **Restriction** | required(VARIABLES, var) |
| **Purpose** | Succeeds if the maxima of the values in each occurrence of the INFLEXION pattern in the time-series given by the VARIABLES collection are all the same. <br> An occurrence of the pattern INFLEXION is the *maximal* subsequence which matches the regular expression $< (< | =)^* > | > (> | =)^* <$. <br> Assume that the occurrence of the pattern INFLEXION starts at position $i$ and ends at position $j$. The feature MAX computes the maximum of the values from index $i + 1$ to index $j$. |
| **Example** | $(\langle 5, 5, 4, 4, 3, 2, 2, 3, 3, 3, 4, 4, 2, 2 \rangle)$ |

Figure 3.29 provides an example where the ALL_EQUAL_MAX_INFLEXION ([5, 5, 4, 4, 3, 2, 2, 3, 3, 3, 4, 4, 2, 2]) constraint holds.

Figure 3.29: Illustrating the ALL_EQUAL_MAX_INFLEXION constraint of the **Example** slot

| | |
|---|---|
| **Typical** | $\lvert$VARIABLES$\rvert > 2$ <br> range(VARIABLES.var) $> 1$ |

**Automaton**     Figure 3.30 depicts the automaton associated with the constraint ALL_EQUAL_MAX_INFLEXION.

Figure 3.30: Automaton for the ALL_EQUAL_MAX_INFLEXION constraint obtained by applying decoration Table 2.36 to the seed transducer of the INFLEXION pattern (transition $r \to t$ has the same accumulators updates as transition $t \to r$)

## 3.16 ALL_EQUAL_MAX_PEAK

► ▷ ◄          DESCRIPTION                    AUTOMATON

| | |
|---|---|
| **Origin** | Based on the PEAK pattern. |
| **Constraint** | ALL_EQUAL_MAX_PEAK(VARIABLES) |
| **Argument** | VARIABLES : collection(var−dvar) |
| **Restriction** | required(VARIABLES, var) |
| **Purpose** | Succeeds if the maxima of the values in each occurrence of the PEAK pattern in the time-series given by the VARIABLES collection are all the same. |
| | An occurrence of the pattern PEAK is the *maximal* subsequence which matches the regular expression $< (= \mid <)^* (> \mid =)^* >$. |
| | Assume that the occurrence of the pattern PEAK starts at position $i$ and ends at position $j$. The feature MAX computes the maximum of the values from index $i + 1$ to index $j$. |
| **Example** | $(\langle 1, 2, 2, 7, 7, 1, 2, 4, 5, 7, 1, 2, 3, 6, 7, 1 \rangle)$ |

Figure 3.31 provides an example where the ALL_EQUAL_MAX_PEAK $([1, 2, 2, 7, 7, 1, 2, 4, 5, 7, 1, 2, 3, 6, 7, 1])$ constraint holds.

Figure 3.31: Illustrating the ALL_EQUAL_MAX_PEAK constraint of the **Example** slot

**Typical**
$|\texttt{VARIABLES}| > 2$
$\texttt{range}(\texttt{VARIABLES.var}) > 1$

     Figure 3.32 depicts the automaton associated with the constraint ALL_EQUAL_MAX_PEAK.

Figure 3.32: Automaton for the ALL_EQUAL_MAX_PEAK constraint obtained by applying decoration Table 2.36 to the seed transducer of the PEAK pattern

## 3.17 ALL_EQUAL_MAX_STRICTLY_DECREASING_SEQUENCE

► ▷   ◄        DESCRIPTION          AUTOMATON

| | |
|---|---|
| **Origin** | Based on the STRICTLY_DECREASING_SEQUENCE pattern. |
| **Constraint** | ALL_EQUAL_MAX_STRICTLY_DECREASING_SEQUENCE(VARIABLES) |
| **Argument** | VARIABLES : collection(var−dvar) |
| **Restriction** | required(VARIABLES, var) |
| **Purpose** | Succeeds if the maxima of the values in each occurrence of the STRICTLY_DECREASING_SEQUENCE pattern in the time-series given by the VARIABLES collection are all the same. An occurrence of the pattern STRICTLY_DECREASING_SEQUENCE is the *maximal* sub-sequence which matches the regular expression $>^+$. Assume that the occurrence of the pattern STRICTLY_DECREASING_SEQUENCE starts at position $i$ and ends at position $j$. The feature MAX computes the maximum of the values from index $i$ to index $j + 1$. |
| **Example** | $(\langle 5, 5, 6, 3, 1, 2, 4, 4, 6, 6, 4, 1, 1, 6, 6, 1 \rangle)$ |

Figure 3.33 provides an example where the ALL_EQUAL_MAX_STRICTLY_DECREASING_SEQUENCE ($[5, 5, 6, 3, 1, 2, 4, 4, 6, 6, 4, 1, 1, 6, 6, 1]$) constraint holds.

| | |
|---|---|
| **Typical** | $\lvert \text{VARIABLES} \rvert > 1$ range(VARIABLES.var) $> 1$ |

Figure 3.33: Illustrating the ALL_EQUAL_MAX_STRICTLY_DECREASING_SEQUENCE constraint of the **Example** slot

**Automaton**          Figure 3.34 depicts the automaton associated with the constraint
ALL_EQUAL_MAX_STRICTLY_DECREASING_SEQUENCE.

Figure 3.34: Automaton for the ALL_EQUAL_MAX_STRICTLY_DECREASING_SEQUENCE
constraint obtained by applying decoration Table 2.36 to the seed transducer of the
STRICTLY_DECREASING_SEQUENCE pattern

## 3.18  ALL_EQUAL_MAX_STRICTLY_INCREASING_SEQUENCE

► ▷ ◄         **DESCRIPTION**          **AUTOMATON**

**Origin**          Based on the STRICTLY_INCREASING_SEQUENCE pattern.

**Constraint**      ALL_EQUAL_MAX_STRICTLY_INCREASING_SEQUENCE(VARIABLES)

**Argument**        VARIABLES : collection(var−dvar)

**Restriction**     required(VARIABLES, var)

**Purpose**         Succeeds if the maxima of the values in each occurrence of the
                    STRICTLY_INCREASING_SEQUENCE pattern in the time-series given by the VARIABLES
                    collection are all the same.
                    An occurrence of the pattern STRICTLY_INCREASING_SEQUENCE is the *maximal* sub-
                    sequence which matches the regular expression $<^+$.
                    Assume that the occurrence of the pattern STRICTLY_INCREASING_SEQUENCE starts at
                    position $i$ and ends at position $j$. The feature MAX computes the maximum of the values
                    from index $i$ to index $j + 1$.

**Example**         $(\langle 6, 1, 1, 6, 6, 3, 1, 1, 3, 4, 5, 6, 4, 1, 6, 2 \rangle)$

                    Figure 3.35 provides an example where the ALL_EQUAL_MAX_STRICTLY_INCREASING_SEQUENCE
                    ($[6, 1, 1, 6, 6, 3, 1, 1, 3, 4, 5, 6, 4, 1, 6, 2]$) constraint holds.

**Typical**         $|\text{VARIABLES}| > 1$
                    $\text{range}(\text{VARIABLES.var}) > 1$

Figure 3.35: Illustrating the ALL_EQUAL_MAX_STRICTLY_INCREASING_SEQUENCE constraint of the **Example** slot

**Automaton**        Figure   3.36   depicts   the   automaton   associated   with   the   constraint
ALL_EQUAL_MAX_STRICTLY_INCREASING_SEQUENCE.

Figure 3.36: Automaton for the ALL_EQUAL_MAX_STRICTLY_INCREASING_SEQUENCE
constraint obtained by applying decoration Table 2.36 to the seed transducer of the
STRICTLY_INCREASING_SEQUENCE pattern

## 3.19 ALL_EQUAL_MAX_SUMMIT

► ▷ ◄          DESCRIPTION          AUTOMATON

$$(< | < (= | <)^* <)(> | > (= | >)^* >)$$

| | |
|---|---|
| **Origin** | Based on the SUMMIT pattern. |
| **Constraint** | ALL_EQUAL_MAX_SUMMIT(VARIABLES) |
| **Argument** | VARIABLES : collection(var−dvar) |
| **Restriction** | required(VARIABLES, var) |
| **Purpose** | Succeeds if the maxima of the values in each occurrence of the SUMMIT pattern in the time-series given by the VARIABLES collection are all the same. <br> An occurrence of the pattern SUMMIT is the *maximal* subsequence which matches the regular expression $(< | < (= | <)^* <)(> | > (= | >)^* >)$. <br> Assume that the occurrence of the pattern SUMMIT starts at position $i$ and ends at position $j$. The feature MAX computes the maximum of the values from index $i+1$ to index $j$. |
| **Example** | $(\langle 1, 3, 5, 4, 2, 6, 6, 2, 3, 5, 4, 1, 4, 5, 3, 2 \rangle)$ |

Figure 3.37 provides an example where the ALL_EQUAL_MAX_SUMMIT $([1, 3, 5, 4, 2, 6, 6, 2, 3, 5, 4, 1, 4, 5, 3, 2])$ constraint holds.

Figure 3.37: Illustrating the ALL_EQUAL_MAX_SUMMIT constraint of the **Example** slot

**Typical**            $|\mathtt{VARIABLES}| > 2$
                       $\mathtt{range}(\mathtt{VARIABLES.var}) > 1$

**Automaton**   Figure 3.38 depicts the automaton associated with the constraint ALL_EQUAL_MAX_SUMMIT.

Figure 3.38: Automaton for the ALL_EQUAL_MAX_SUMMIT constraint obtained by applying decoration Table 2.36 to the seed transducer of the SUMMIT pattern (transition $u \to r$ has the same accumulator update as transition $r \to u$)

## 3.20 ALL_EQUAL_MAX_ZIGZAG

► ▷ ◄          DESCRIPTION          AUTOMATON

| | |
|---|---|
| **Origin** | Based on the ZIGZAG pattern. |
| **Constraint** | ALL_EQUAL_MAX_ZIGZAG(VARIABLES) |
| **Argument** | VARIABLES : collection(var−dvar) |
| **Restriction** | required(VARIABLES, var) |
| **Purpose** | Succeeds if the maxima of the values in each occurrence of the ZIGZAG pattern in the time-series given by the VARIABLES collection are all the same.<br><br>An occurrence of the pattern ZIGZAG is the *maximal* subsequence which matches the regular expression $(<>)^+(<\|<>)\|(><)^+(>\|><)$.<br><br>Assume that the occurrence of the pattern ZIGZAG starts at position $i$ and ends at position $j$. The feature MAX computes the maximum of the values from index $i+1$ to index $j$. |
| **Example** | $(\langle 1, 6, 1, 2, 7, 5, 1, 6, 3, 4, 2, 5, 6, 3, 4, 7 \rangle)$ |

Figure 3.39 provides an example where the ALL_EQUAL_MAX_ZIGZAG $([1, 6, 1, 2, 7, 5, 1, 6, 3, 4, 2, 5, 6, 3, 4, 7])$ constraint holds.

Figure 3.39: Illustrating the ALL_EQUAL_MAX_ZIGZAG constraint of the **Example** slot

**Typical**  $|\mathtt{VARIABLES}| > 3$
$\mathtt{range}(\mathtt{VARIABLES.var}) > 1$

**Automaton**     Figure 3.40 depicts the automaton associated with the constraint ALL_EQUAL_MAX_ZIGZAG.

Figure 3.40: Automaton for the ALL_EQUAL_MAX_ZIGZAG constraint obtained by applying decoration Table 2.36 to the seed transducer of the ZIGZAG pattern; (1) missing transitions from $a, b, c, d, e, f$ to $s$ are labelled by $=$; (2) on transitions from $b, c, e, f$ to $s$ the accumulator $D$ is reset to its initial value; (3) on transitions from $c, f$ to $s$ the accumulator $F$ is reset to $C$, and the accumulator $R$ is updated wrt $C$ and $F$

# 3.21  ALL_EQUAL_MIN_BUMP_ON_DECREASING_SEQUENCE

▶ ▷        ◀        **DESCRIPTION**        **AUTOMATON**

| | |
|---|---|
| **Origin** | Based on the BUMP_ON_DECREASING_SEQUENCE pattern. |
| **Constraint** | ALL_EQUAL_MIN_BUMP_ON_DECREASING_SEQUENCE(VARIABLES) |
| **Argument** | VARIABLES : collection(var−dvar) |
| **Restriction** | required(VARIABLES, var) |

**Purpose**

Succeeds if the minima of the values in each occurrence of the BUMP_ON_DECREASING_SEQUENCE pattern in the time-series given by the VARIABLES collection are all the same.

An occurrence of the pattern BUMP_ON_DECREASING_SEQUENCE is the subsequence which matches the regular expression >><>>.

Assume that the occurrence of the pattern BUMP_ON_DECREASING_SEQUENCE starts at position $i$ and ends at position $j$. The feature MIN computes the minimum of the values from index $i + 2$ to index $j$.

**Example**

$(\langle 7, 6, 5, 6, 2, 1, 1, 5, 4, 3, 2, 6, 5, 4, 4, 5 \rangle)$

Figure 3.41 provides an example where the ALL_EQUAL_MIN_BUMP_ON_DECREASING_SEQUENCE $([7, 6, 5, 6, 2, 1, 1, 5, 4, 3, 2, 6, 5, 4, 4, 5])$ constraint holds.

**Typical**

$|\mathtt{VARIABLES}| > 5$

$\mathtt{range}(\mathtt{VARIABLES.var}) > 2$

Figure 3.41: Illustrating the ALL_EQUAL_MIN_BUMP_ON_DECREASING_SEQUENCE constraint of the **Example** slot

**Automaton**        Figure 3.42 depicts the automaton associated with the constraint
                     ALL_EQUAL_MIN_BUMP_ON_DECREASING_SEQUENCE.

Figure 3.42: Automaton for the ALL_EQUAL_MIN_BUMP_ON_DECREASING_SEQUENCE
constraint obtained by applying decoration Table 2.36 to the seed transducer of the
BUMP_ON_DECREASING_SEQUENCE pattern

## 3.22 ALL_EQUAL_MIN_DECREASING

▶ ▷ ◀          DESCRIPTION          AUTOMATON

| | |
|---|---|
| **Origin** | Based on the DECREASING pattern. |
| **Constraint** | ALL_EQUAL_MIN_DECREASING(VARIABLES) |
| **Argument** | VARIABLES : collection(var−dvar) |
| **Restriction** | required(VARIABLES, var) |
| **Purpose** | Succeeds if the minima of the values in each occurrence of the DECREASING pattern in the time-series given by the VARIABLES collection are all the same. <br><br> An occurrence of the pattern DECREASING is the subsequence which matches the regular expression >. <br><br> Assume that the occurrence of the pattern DECREASING starts at position $i$ and ends at position $j$. The feature MIN computes the minimum of the values from index $i$ to index $j+1$. |
| **Example** | $(\langle 1, 6, 6, 1, 1, 4, 4, 5, 5, 5, 5, 1, 3, 6, 1, 5 \rangle)$ |

Figure 3.43 provides an example where the ALL_EQUAL_MIN_DECREASING $([1, 6, 6, 1, 1, 4, 4, 5, 5, 5, 5, 1, 3, 6, 1, 5])$ constraint holds.

Figure 3.43: Illustrating the ALL_EQUAL_MIN_DECREASING constraint of the **Example** slot

**Typical**
$$|\text{VARIABLES}| > 1$$
$$\text{range}(\text{VARIABLES.var}) > 1$$

**Automaton**        Figure 3.44 depicts the automaton associated with the constraint ALL_EQUAL_MIN_DECREASING.

Figure 3.44: Automaton for the ALL_EQUAL_MIN_DECREASING constraint obtained by applying decoration Table 2.36 to the seed transducer of the DECREASING pattern

## 3.23  ALL_EQUAL_MIN_DECREASING_SEQUENCE

► ▷ ◄          DESCRIPTION          AUTOMATON

| | |
|---|---|
| **Origin** | Based on the DECREASING_SEQUENCE pattern. |
| **Constraint** | ALL_EQUAL_MIN_DECREASING_SEQUENCE(VARIABLES) |
| **Argument** | VARIABLES : collection(var−dvar) |
| **Restriction** | required(VARIABLES, var) |

**Purpose**

Succeeds if the minima of the values in each occurrence of the DECREASING_SEQUENCE pattern in the time-series given by the VARIABLES collection are all the same.
An occurrence of the pattern DECREASING_SEQUENCE is the *maximal* subsequence which matches the regular expression $> (> | =)^* > | >$.
Assume that the occurrence of the pattern DECREASING_SEQUENCE starts at position $i$ and ends at position $j$. The feature MIN computes the minimum of the values from index $i$ to index $j + 1$.

**Example**

$(\langle 1, 6, 6, 1, 1, 4, 5, 5, 4, 4, 2, 1, 3, 4, 1, 5\rangle)$

Figure 3.45 provides an example where the ALL_EQUAL_MIN_DECREASING_SEQUENCE $([1, 6, 6, 1, 1, 4, 5, 5, 4, 4, 2, 1, 3, 4, 1, 5])$ constraint holds.

Figure 3.45: Illustrating the ALL_EQUAL_MIN_DECREASING_SEQUENCE constraint of the **Example** slot

**Typical**    $|\mathtt{VARIABLES}| > 1$
               $\mathrm{range}(\mathtt{VARIABLES.var}) > 1$

**Automaton**       Figure 3.46 depicts the automaton associated with the constraint
ALL_EQUAL_MIN_DECREASING_SEQUENCE.

Figure 3.46: Automaton for the ALL_EQUAL_MIN_DECREASING_SEQUENCE constraint obtained by applying decoration Table 2.36 to the seed transducer of the DECREASING_SEQUENCE pattern

## 3.24 ALL_EQUAL_MIN_DIP_ON_INCREASING_SEQUENCE

▶ ▷ ◀         DESCRIPTION         AUTOMATON

**Origin**         Based on the DIP_ON_INCREASING_SEQUENCE pattern.

**Constraint**      ALL_EQUAL_MIN_DIP_ON_INCREASING_SEQUENCE(VARIABLES)

**Argument**       VARIABLES : collection(var−dvar)

**Restriction**     required(VARIABLES, var)

**Purpose**        Succeeds if the minima of the values in each occurrence of the DIP_ON_INCREASING_SEQUENCE pattern in the time-series given by the VARIABLES collection are all the same.

An occurrence of the pattern DIP_ON_INCREASING_SEQUENCE is the subsequence which matches the regular expression <<><<.

Assume that the occurrence of the pattern DIP_ON_INCREASING_SEQUENCE starts at position $i$ and ends at position $j$. The feature MIN computes the minimum of the values from index $i + 2$ to index $j$.

**Example**        $(\langle 1, 2, 3, 2, 4, 7, 7, 3, 4, 5, 6, 2, 3, 4, 4, 3 \rangle)$

Figure 3.47 provides an example where the ALL_EQUAL_MIN_DIP_ON_INCREASING_SEQUENCE $([1, 2, 3, 2, 4, 7, 7, 3, 4, 5, 6, 2, 3, 4, 4, 3])$ constraint holds.

**Typical**        $|\text{VARIABLES}| > 5$
              $\text{range}(\text{VARIABLES.var}) > 2$

Figure 3.47: Illustrating the ALL_EQUAL_MIN_DIP_ON_INCREASING_SEQUENCE constraint of the **Example** slot

**Automaton**        Figure 3.48 depicts the automaton associated with the constraint
ALL_EQUAL_MIN_DIP_ON_INCREASING_SEQUENCE.

Figure 3.48: Automaton for the ALL_EQUAL_MIN_DIP_ON_INCREASING_SEQUENCE
constraint obtained by applying decoration Table 2.36 to the seed transducer of the
DIP_ON_INCREASING_SEQUENCE pattern

## 3.25 ALL_EQUAL_MIN_GORGE

| | |
|---|---|
| **Origin** | Based on the GORGE pattern. |
| **Constraint** | ALL_EQUAL_MIN_GORGE(VARIABLES) |
| **Argument** | VARIABLES : collection(var−dvar) |
| **Restriction** | required(VARIABLES, var) |
| **Purpose** | Succeeds if the minima of the values in each occurrence of the GORGE pattern in the time-series given by the VARIABLES collection are all the same. <br><br> An occurrence of the pattern GORGE is the *maximal* subsequence which matches the regular expression $(> \mid > (= \mid >)^* >)(< \mid < (= \mid <)^* <)$. <br><br> Assume that the occurrence of the pattern GORGE starts at position $i$ and ends at position $j$. The feature MIN computes the minimum of the values from index $i + 1$ to index $j$. |
| **Example** | $(\langle 6, 4, 2, 3, 5, 1, 1, 5, 4, 2, 3, 6, 3, 2, 4, 5 \rangle)$ |

Figure 3.49 provides an example where the ALL_EQUAL_MIN_GORGE ([6, 4, 2, 3, 5, 1, 1, 5, 4, 2, 3, 6, 3, 2, 4, 5]) constraint holds.

Figure 3.49: Illustrating the ALL_EQUAL_MIN_GORGE constraint of the **Example** slot

| | |
|---|---|
| **Typical** | $\mid$VARIABLES$\mid > 2$ <br> range(VARIABLES.var) $> 1$ |

**Automaton**          Figure 3.50 depicts the automaton associated with the constraint ALL_EQUAL_MIN_GORGE.

Figure 3.50: Automaton for the ALL_EQUAL_MIN_GORGE constraint obtained by applying decoration Table 2.36 to the seed transducer of the GORGE pattern (transition $u \rightarrow r$ has the same accumulator update as transition $r \rightarrow u$)

## 3.26  ALL_EQUAL_MIN_INCREASING

▶ ▷    ◀         **DESCRIPTION**         **AUTOMATON**

| | |
|---|---|
| **Origin** | Based on the INCREASING pattern. |
| **Constraint** | ALL_EQUAL_MIN_INCREASING(VARIABLES) |
| **Argument** | VARIABLES : collection(var−dvar) |
| **Restriction** | required(VARIABLES, var) |
| **Purpose** | Succeeds if the minima of the values in each occurrence of the INCREASING pattern in the time-series given by the VARIABLES collection are all the same.<br><br>An occurrence of the pattern INCREASING is the subsequence which matches the regular expression <.<br><br>Assume that the occurrence of the pattern INCREASING starts at position $i$ and ends at position $j$. The feature MIN computes the minimum of the values from index $i$ to index $j+1$. |
| **Example** | $(\langle 7, 2, 2, 6, 6, 3, 2, 2, 7, 4, 4, 3, 3, 2, 5, 1\rangle)$ |

Figure 3.51 provides an example where the ALL_EQUAL_MIN_INCREASING $([7, 2, 2, 6, 6, 3, 2, 2, 7, 4, 4, 3, 3, 2, 5, 1])$ constraint holds.

| | |
|---|---|
| **Typical** | $\lvert$VARIABLES$\rvert > 1$<br>range(VARIABLES.var) $> 1$ |

Figure 3.51: Illustrating the ALL_EQUAL_MIN_INCREASING constraint of the **Example** slot

**Automaton**     Figure   3.52   depicts   the   automaton   associated   with   the   constraint
ALL_EQUAL_MIN_INCREASING.

Figure 3.52: Automaton for the ALL_EQUAL_MIN_INCREASING constraint obtained by applying decoration Table 2.36 to the seed transducer of the INCREASING pattern

# 3.27 ALL_EQUAL_MIN_INCREASING_SEQUENCE

▶ ▷   ◀        **DESCRIPTION**        **AUTOMATON**

| | |
|---|---|
| **Origin** | Based on the INCREASING_SEQUENCE pattern. |
| **Constraint** | ALL_EQUAL_MIN_INCREASING_SEQUENCE(VARIABLES) |
| **Argument** | VARIABLES : collection(var−dvar) |
| **Restriction** | required(VARIABLES, var) |

**Purpose**

Succeeds if the minima of the values in each occurrence of the INCREASING_SEQUENCE pattern in the time-series given by the VARIABLES collection are all the same.

An occurrence of the pattern INCREASING_SEQUENCE is the *maximal* subsequence which matches the regular expression $< (< | =)^* < | <$.

Assume that the occurrence of the pattern INCREASING_SEQUENCE starts at position $i$ and ends at position $j$. The feature MIN computes the minimum of the values from index $i$ to index $j + 1$.

**Example**

$(\langle 7, 2, 2, 6, 6, 3, 2, 2, 3, 3, 5, 6, 4, 2, 5, 1 \rangle)$

Figure 3.53 provides an example where the ALL_EQUAL_MIN_INCREASING_SEQUENCE $([7, 2, 2, 6, 6, 3, 2, 2, 3, 3, 5, 6, 4, 2, 5, 1])$ constraint holds.

**Typical**

$|\text{VARIABLES}| > 1$
range(VARIABLES.var) $> 1$

Figure 3.53: Illustrating the ALL_EQUAL_MIN_INCREASING_SEQUENCE constraint of the **Example** slot

**Automaton**    Figure 3.54 depicts the automaton associated with the constraint ALL_EQUAL_MIN_INCREASING_SEQUENCE.

Figure 3.54: Automaton for the ALL_EQUAL_MIN_INCREASING_SEQUENCE constraint obtained by applying decoration Table 2.36 to the seed transducer of the INCREASING_SEQUENCE pattern

## 3.28  ALL_EQUAL_MIN_INFLEXION

► ▷ ◄          DESCRIPTION          AUTOMATON

| | |
|---|---|
| **Origin** | Based on the INFLEXION pattern. |
| **Constraint** | ALL_EQUAL_MIN_INFLEXION(VARIABLES) |
| **Argument** | VARIABLES : collection(var−dvar) |
| **Restriction** | required(VARIABLES, var) |
| **Purpose** | Succeeds if the minima of the values in each occurrence of the INFLEXION pattern in the time-series given by the VARIABLES collection are all the same. |
| | An occurrence of the pattern INFLEXION is the *maximal* subsequence which matches the regular expression $< (< \mid =)^* > \mid > (> \mid =)^* <$. |
| | Assume that the occurrence of the pattern INFLEXION starts at position $i$ and ends at position $j$. The feature MIN computes the minimum of the values from index $i + 1$ to index $j$. |
| **Example** | $(\langle 2, 2, 3, 6, 6, 5, 4, 3, 4, 4, 5 \rangle)$ |

Figure 3.55 provides an example where the ALL_EQUAL_MIN_INFLEXION ([2, 2, 3, 6, 6, 5, 4, 3, 4, 4, 5]) constraint holds.

Figure 3.55: Illustrating the ALL_EQUAL_MIN_INFLEXION constraint of the **Example** slot

**Typical**

$|\mathtt{VARIABLES}| > 2$

$\mathtt{range}(\mathtt{VARIABLES.var}) > 1$

**Automaton**    Figure 3.56 depicts the automaton associated with the constraint ALL_EQUAL_MIN_INFLEXION.

Figure 3.56: Automaton for the ALL_EQUAL_MIN_INFLEXION constraint obtained by applying decoration Table 2.36 to the seed transducer of the INFLEXION pattern (transition $r \rightarrow t$ has the same accumulators updates as transition $t \rightarrow r$)

## 3.29  ALL_EQUAL_MIN_STRICTLY_DECREASING_SEQUENCE

▶ ▷ ◀          DESCRIPTION          AUTOMATON

| | |
|---|---|
| **Origin** | Based on the STRICTLY_DECREASING_SEQUENCE pattern. |
| **Constraint** | ALL_EQUAL_MIN_STRICTLY_DECREASING_SEQUENCE(VARIABLES) |
| **Argument** | VARIABLES : collection(var−dvar) |
| **Restriction** | required(VARIABLES, var) |
| **Purpose** | Succeeds if the minima of the values in each occurrence of the STRICTLY_DECREASING_SEQUENCE pattern in the time-series given by the VARIABLES collection are all the same.<br>An occurrence of the pattern STRICTLY_DECREASING_SEQUENCE is the *maximal* subsequence which matches the regular expression $>^+$.<br>Assume that the occurrence of the pattern STRICTLY_DECREASING_SEQUENCE starts at position $i$ and ends at position $j$. The feature MIN computes the minimum of the values from index $i$ to index $j+1$. |
| **Example** | $(\langle 5, 1, 4, 3, 1, 2, 4, 4, 5, 5, 4, 1, 1, 6, 6, 1 \rangle)$ |

Figure 3.57 provides an example where the ALL_EQUAL_MIN_STRICTLY_DECREASING_SEQUENCE $([5, 1, 4, 3, 1, 2, 4, 4, 5, 5, 4, 1, 1, 6, 6, 1])$ constraint holds.

| | |
|---|---|
| **Typical** | $\|VARIABLES\| > 1$<br>range(VARIABLES.var) > 1 |

Figure 3.57: Illustrating the ALL_EQUAL_MIN_STRICTLY_DECREASING_SEQUENCE constraint of the **Example** slot

**Automaton**     Figure 3.58 depicts the automaton associated with the constraint
ALL␣EQUAL␣MIN␣STRICTLY␣DECREASING␣SEQUENCE.

Figure 3.58: Automaton for the ALL␣EQUAL␣MIN␣STRICTLY␣DECREASING␣SEQUENCE
constraint obtained by applying decoration Table 2.36 to the seed transducer of the
STRICTLY␣DECREASING␣SEQUENCE pattern

## 3.30 ALL_EQUAL_MIN_STRICTLY_INCREASING_SEQUENCE

▶ ▷ ◀　　DESCRIPTION　　AUTOMATON

**Origin**　　Based on the STRICTLY_INCREASING_SEQUENCE pattern.

**Constraint**　　ALL_EQUAL_MIN_STRICTLY_INCREASING_SEQUENCE(VARIABLES)

**Argument**　　VARIABLES : collection(var−dvar)

**Restriction**　　required(VARIABLES, var)

**Purpose**　　Succeeds if the minima of the values in each occurrence of the STRICTLY_INCREASING_SEQUENCE pattern in the time-series given by the VARIABLES collection are all the same.

An occurrence of the pattern STRICTLY_INCREASING_SEQUENCE is the *maximal* subsequence which matches the regular expression $<^+$.

Assume that the occurrence of the pattern STRICTLY_INCREASING_SEQUENCE starts at position $i$ and ends at position $j$. The feature MIN computes the minimum of the values from index $i$ to index $j + 1$.

**Example**　　$(\langle 7, 2, 2, 6, 6, 3, 2, 2, 3, 4, 5, 6, 4, 2, 5, 1 \rangle)$

Figure 3.59 provides an example where the ALL_EQUAL_MIN_STRICTLY_INCREASING_SEQUENCE $([7, 2, 2, 6, 6, 3, 2, 2, 3, 4, 5, 6, 4, 2, 5, 1])$ constraint holds.

**Typical**　　$|\text{VARIABLES}| > 1$
range(VARIABLES.var) $> 1$

Figure 3.59: Illustrating the ALL_EQUAL_MIN_STRICTLY_INCREASING_SEQUENCE constraint of the **Example** slot

**Automaton**    Figure 3.60 depicts the automaton associated with the constraint
ALL_EQUAL_MIN_STRICTLY_INCREASING_SEQUENCE.

Figure 3.60: Automaton for the ALL_EQUAL_MIN_STRICTLY_INCREASING_SEQUENCE
constraint obtained by applying decoration Table 2.36 to the seed transducer of the
STRICTLY_INCREASING_SEQUENCE pattern

# 3.31  ALL_EQUAL_MIN_VALLEY

          DESCRIPTION          AUTOMATON

**Origin**         Based on the VALLEY pattern.

**Constraint**     ALL_EQUAL_MIN_VALLEY(VARIABLES)

**Argument**       VARIABLES : collection(var−dvar)

**Restriction**    required(VARIABLES, var)

**Purpose**        Succeeds if the minima of the values in each occurrence of the VALLEY pattern in the time-series given by the VARIABLES collection are all the same.

An occurrence of the pattern VALLEY is the *maximal* subsequence which matches the regular expression $> (= | >)^* (< | =)^* <$.

Assume that the occurrence of the pattern VALLEY starts at position $i$ and ends at position $j$. The feature MIN computes the minimum of the values from index $i + 1$ to index $j$.

**Example**        $(\langle 7, 6, 6, 1, 1, 7, 6, 4, 3, 1, 7, 6, 5, 2, 1, 7 \rangle)$

Figure 3.61 provides an example where the ALL_EQUAL_MIN_VALLEY ($[7, 6, 6, 1, 1, 7, 6, 4, 3, 1, 7, 6, 5, 2, 1, 7]$) constraint holds.

Figure 3.61: Illustrating the ALL_EQUAL_MIN_VALLEY constraint of the **Example** slot

**Typical**                     $|\text{VARIABLES}| > 2$
                                $\text{range}(\text{VARIABLES.var}) > 1$

**Automaton**    Figure 3.62 depicts the automaton associated with the constraint
ALL_EQUAL_MIN_VALLEY.

Figure 3.62: Automaton for the ALL_EQUAL_MIN_VALLEY constraint obtained by applying decoration Table 2.36 to the seed transducer of the VALLEY pattern

# 3.32 ALL_EQUAL_MIN_ZIGZAG

► ▷ ◄          DESCRIPTION          AUTOMATON

$(<>)^{+}(<|<>)|(><)^{+}(>|><)$

| | |
|---|---|
| **Origin** | Based on the ZIGZAG pattern. |
| **Constraint** | ALL_EQUAL_MIN_ZIGZAG(VARIABLES) |
| **Argument** | VARIABLES : collection(var−dvar) |
| **Restriction** | required(VARIABLES, var) |
| **Purpose** | Succeeds if the minima of the values in each occurrence of the ZIGZAG pattern in the time-series given by the VARIABLES collection are all the same.<br><br>An occurrence of the pattern ZIGZAG is the *maximal* subsequence which matches the regular expression $(<>)^{+}(<|<>)|(><)^{+}(>|><)$.<br><br>Assume that the occurrence of the pattern ZIGZAG starts at position $i$ and ends at position $j$. The feature MIN computes the minimum of the values from index $i+1$ to index $j$. |
| **Example** | $(\langle 1, 3, 1, 2, 6, 5, 1, 4, 3, 4, 3, 5, 6, 1, 4, 7 \rangle)$ |

Figure 3.63 provides an example where the ALL_EQUAL_MIN_ZIGZAG ($[1, 3, 1, 2, 6, 5, 1, 4, 3, 4, 3, 5, 6, 1, 4, 7]$) constraint holds.

Figure 3.63: Illustrating the ALL_EQUAL_MIN_ZIGZAG constraint of the **Example** slot

**Typical**            $|\text{VARIABLES}| > 3$
                       $\text{range}(\text{VARIABLES.var}) > 1$

**Automaton**          Figure 3.64 depicts the automaton associated with the constraint
ALL_EQUAL_MIN_ZIGZAG.

Figure 3.64: Automaton for the ALL_EQUAL_MIN_ZIGZAG constraint obtained by applying decoration Table 2.36 to the seed transducer of the ZIGZAG pattern; (1) missing transitions from $a, b, c, d, e, f$ to $s$ are labelled by $=$; (2) on transitions from $b, c, e, f$ to $s$ the accumulator $D$ is reset to its initial value; (3) on transitions from $c, f$ to $s$ the accumulator $F$ is reset to $C$, and the accumulator $R$ is updated wrt $C$ and $F$

## 3.33   ALL_EQUAL_RANGE_DECREASING

| | |
|---|---|
| **Origin** | Based on the DECREASING pattern. |
| **Constraint** | ALL_EQUAL_RANGE_DECREASING(VARIABLES) |
| **Argument** | VARIABLES : collection(var−dvar) |
| **Restriction** | required(VARIABLES, var) |

**Purpose**

Succeeds if the differences between the largest and smallest value in each occurrence of the DECREASING pattern in the time-series given by the VARIABLES collection are the same.

An occurrence of the pattern DECREASING is the subsequence which matches the regular expression >.

Assume that the occurrence of the pattern DECREASING starts at position $i$ and ends at position $j$. The feature RANGE computes the range of the values from index $i$ to index $j + 1$.

**Example**

$(\langle 1, 4, 6, 2, 2, 4, 4, 5, 5, 5, 5, 1, 3, 6, 2, 5 \rangle)$

Figure 3.65 provides an example where the ALL_EQUAL_RANGE_DECREASING $([1, 4, 6, 2, 2, 4, 4, 5, 5, 5, 5, 1, 3, 6, 2, 5])$ constraint holds.

**Typical**

$|\text{VARIABLES}| > 1$
$\text{range}(\text{VARIABLES.var}) > 1$

Figure 3.65: Illustrating the ALL_EQUAL_RANGE_DECREASING constraint of the **Example** slot

**Automaton**     Figure  3.66  depicts  the  automaton  associated  with  the  constraint
ALL_EQUAL_RANGE_DECREASING.

Figure 3.66: Automaton for the ALL_EQUAL_RANGE_DECREASING constraint obtained by applying decoration Table 2.47 to the seed transducer of the DECREASING pattern

## 3.34 ALL_EQUAL_RANGE_DECREASING_SEQUENCE

► ▷ ◄          **DESCRIPTION**            **AUTOMATON**

| | |
|---|---|
| **Origin** | Based on the DECREASING_SEQUENCE pattern. |
| **Constraint** | ALL_EQUAL_RANGE_DECREASING_SEQUENCE(VARIABLES) |
| **Argument** | VARIABLES : collection(var−dvar) |
| **Restriction** | required(VARIABLES, var) |
| **Purpose** | Succeeds if the differences between the largest and smallest value in each occurrence of the DECREASING_SEQUENCE pattern in the time-series given by the VARIABLES collection are the same.<br>An occurrence of the pattern DECREASING_SEQUENCE is the *maximal* subsequence which matches the regular expression $> (> \mid =)^* > \mid >$.<br>Assume that the occurrence of the pattern DECREASING_SEQUENCE starts at position $i$ and ends at position $j$. The feature RANGE computes the range of the values from index $i$ to index $j + 1$. |
| **Example** | $(\langle 1, 6, 7, 2, 2, 4, 5, 5, 4, 4, 2, 0, 3, 6, 1, 5\rangle)$ |

Figure 3.67 provides an example where the ALL_EQUAL_RANGE_DECREASING_SEQUENCE $([1, 6, 7, 2, 2, 4, 5, 5, 4, 4, 2, 0, 3, 6, 1, 5])$ constraint holds.

| | |
|---|---|
| **Typical** | $\lvert\text{VARIABLES}\rvert > 1$<br>range(VARIABLES.var) > 1 |

Figure 3.67: Illustrating the ALL_EQUAL_RANGE_DECREASING_SEQUENCE constraint of the **Example** slot

**Automaton**　　　　Figure 3.68 depicts the automaton associated with the constraint ALL_EQUAL_RANGE_DECREASING_SEQUENCE.

Figure 3.68: Automaton for the ALL_EQUAL_RANGE_DECREASING_SEQUENCE constraint obtained by applying decoration Table 2.47 to the seed transducer of the DECREASING_SEQUENCE pattern

## 3.35   ALL_EQUAL_RANGE_INCREASING

► ▷   ◄        **DESCRIPTION**           **AUTOMATON**

| | |
|---|---|
| **Origin** | Based on the INCREASING pattern. |
| **Constraint** | ALL_EQUAL_RANGE_INCREASING(VARIABLES) |
| **Argument** | VARIABLES : collection(var−dvar) |
| **Restriction** | required(VARIABLES, var) |
| **Purpose** | Succeeds if the differences between the largest and smallest value in each occurrence of the INCREASING pattern in the time-series given by the VARIABLES collection are the same. <br> An occurrence of the pattern INCREASING is the subsequence which matches the regular expression <. <br> Assume that the occurrence of the pattern INCREASING starts at position $i$ and ends at position $j$. The feature RANGE computes the range of the values from index $i$ to index $j+1$. |
| **Example** | $(\langle 7, 3, 3, 6, 6, 3, 2, 2, 5, 4, 4, 3, 3, 1, 4, 1 \rangle)$ |

Figure 3.69 provides an example where the ALL_EQUAL_RANGE_INCREASING ([7, 3, 3, 6, 6, 3, 2, 2, 5, 4, 4, 3, 3, 1, 4, 1]) constraint holds.

| | |
|---|---|
| **Typical** | $\|$VARIABLES$\| > 1$ <br> range(VARIABLES.var) $> 1$ |

Figure 3.69: Illustrating the ALL_EQUAL_RANGE_INCREASING constraint of the **Example** slot

**Automaton**  Figure 3.70 depicts the automaton associated with the constraint ALL_EQUAL_RANGE_INCREASING.

Figure 3.70: Automaton for the ALL_EQUAL_RANGE_INCREASING constraint obtained by applying decoration Table 2.47 to the seed transducer of the INCREASING pattern

## 3.36 ALL_EQUAL_RANGE_INCREASING_SEQUENCE

► ▷ ◄          **DESCRIPTION**          **AUTOMATON**

| **Origin** | Based on the INCREASING_SEQUENCE pattern. |
|---|---|
| **Constraint** | ALL_EQUAL_RANGE_INCREASING_SEQUENCE(VARIABLES) |
| **Argument** | VARIABLES : collection(var−dvar) |
| **Restriction** | required(VARIABLES, var) |

**Purpose**

Succeeds if the differences between the largest and smallest value in each occurrence of the INCREASING_SEQUENCE pattern in the time-series given by the VARIABLES collection are the same.

An occurrence of the pattern INCREASING_SEQUENCE is the *maximal* subsequence which matches the regular expression $< (< | =)^* < | <$.

Assume that the occurrence of the pattern INCREASING_SEQUENCE starts at position $i$ and ends at position $j$. The feature RANGE computes the range of the values from index $i$ to index $j + 1$.

**Example**

$(\langle 7, 2, 2, 6, 6, 3, 1, 1, 2, 3, 3, 5, 4, 2, 6, 1 \rangle)$

Figure 3.71 provides an example where the ALL_EQUAL_RANGE_INCREASING_SEQUENCE $([7, 2, 2, 6, 6, 3, 1, 1, 2, 3, 3, 5, 4, 2, 6, 1])$ constraint holds.

**Typical**

$|VARIABLES| > 1$
range(VARIABLES.var) $> 1$

Figure 3.71: Illustrating the ALL_EQUAL_RANGE_INCREASING_SEQUENCE constraint of the **Example** slot

**Automaton**  Figure 3.72 depicts the automaton associated with the constraint ALL_EQUAL_RANGE_INCREASING_SEQUENCE.

Figure 3.72: Automaton for the ALL_EQUAL_RANGE_INCREASING_SEQUENCE constraint obtained by applying decoration Table 2.47 to the seed transducer of the INCREASING_SEQUENCE pattern

## 3.37 ALL_EQUAL_RANGE_STRICTLY_DECREASING_SEQUENCE

| | |
|---|---|
| **Origin** | Based on the STRICTLY_DECREASING_SEQUENCE pattern. |
| **Constraint** | ALL_EQUAL_RANGE_STRICTLY_DECREASING_SEQUENCE(VARIABLES) |
| **Argument** | VARIABLES : collection(var−dvar) |
| **Restriction** | required(VARIABLES, var) |

**Purpose**

Succeeds if the differences between the largest and smallest value in each occurrence of the STRICTLY_DECREASING_SEQUENCE pattern in the time-series given by the VARIABLES collection are the same.

An occurrence of the pattern STRICTLY_DECREASING_SEQUENCE is the *maximal* subsequence which matches the regular expression $>^+$.

Assume that the occurrence of the pattern STRICTLY_DECREASING_SEQUENCE starts at position $i$ and ends at position $j$. The feature RANGE computes the range of the values from index $i$ to index $j + 1$.

**Example**

$$(\langle 4, 1, 4, 3, 1, 2, 4, 4, 5, 5, 4, 2, 2, 6, 6, 3 \rangle)$$

Figure 3.73 provides an example where the ALL_EQUAL_RANGE_STRICTLY_DECREASING_SEQUENCE ([4, 1, 4, 3, 1, 2, 4, 4, 5, 5, 4, 2, 2, 6, 6, 3]) constraint holds.

**Typical**

$|\text{VARIABLES}| > 1$
$\text{range}(\text{VARIABLES.var}) > 1$

Figure 3.73: Illustrating the ALL_EQUAL_RANGE_STRICTLY_DECREASING_SEQUENCE constraint of the **Example** slot

**Automaton**     Figure 3.74 depicts the automaton associated with the constraint
ALL_EQUAL_RANGE_STRICTLY_DECREASING_SEQUENCE.

Figure 3.74: Automaton for the ALL_EQUAL_RANGE_STRICTLY_DECREASING_SEQUENCE constraint obtained by applying decoration Table 2.47 to the seed transducer of the STRICTLY_DECREASING_SEQUENCE pattern

# 3.38 ALL_EQUAL_RANGE_STRICTLY_INCREASING_SEQUENCE

► ▷ ◄        DESCRIPTION        AUTOMATON

| | |
|---|---|
| **Origin** | Based on the STRICTLY_INCREASING_SEQUENCE pattern. |
| **Constraint** | ALL_EQUAL_RANGE_STRICTLY_INCREASING_SEQUENCE(VARIABLES) |
| **Argument** | VARIABLES : collection(var−dvar) |
| **Restriction** | required(VARIABLES, var) |

**Purpose**

Succeeds if the differences between the largest and smallest value in each occurrence of the STRICTLY_INCREASING_SEQUENCE pattern in the time-series given by the VARIABLES collection are the same.

An occurrence of the pattern STRICTLY_INCREASING_SEQUENCE is the *maximal* subsequence which matches the regular expression $<^+$.

Assume that the occurrence of the pattern STRICTLY_INCREASING_SEQUENCE starts at position $i$ and ends at position $j$. The feature RANGE computes the range of the values from index $i$ to index $j + 1$.

**Example**

$$(\langle 7, 2, 2, 6, 6, 3, 1, 1, 2, 3, 4, 5, 4, 0, 4, 1 \rangle)$$

Figure 3.75 provides an example where the ALL_EQUAL_RANGE_STRICTLY_INCREASING_SEQUENCE ([7, 2, 2, 6, 6, 3, 1, 1, 2, 3, 4, 5, 4, 0, 4, 1]) constraint holds.

**Typical**

$|\text{VARIABLES}| > 1$

$\text{range}(\text{VARIABLES.var}) > 1$

Figure 3.75: Illustrating the ALL_EQUAL_RANGE_STRICTLY_INCREASING_SEQUENCE constraint of the **Example** slot

**Automaton**          Figure 3.76 depicts the automaton associated with the constraint
ALL_EQUAL_RANGE_STRICTLY_INCREASING_SEQUENCE.

$$\left\{ \begin{array}{l} C \leftarrow X \\ F \leftarrow X \\ H \leftarrow \texttt{VAR}_1 \\ R \leftarrow 1 \end{array} \right\}$$

$$\left\{ \begin{array}{l} F \leftarrow |H - \texttt{VAR}_i| \\ H \leftarrow \texttt{VAR}_{i+1} \\ R \leftarrow R \wedge (F = C) \end{array} \right\}$$

$\geq s$          $\{ H \overset{\geq}{\leftarrow} \texttt{VAR}_{i+1} \}$

$R \wedge (F = C)$          $\{ C \leftarrow |H \overset{<}{-} \texttt{VAR}_{i+1}| \}$

$< r$          $\{ C \leftarrow |H \overset{<}{-} \texttt{VAR}_{i+1}| \}$

Figure 3.76: Automaton for the ALL_EQUAL_RANGE_STRICTLY_INCREASING_SEQUENCE
constraint obtained by applying decoration Table 2.47 to the seed transducer of the
STRICTLY_INCREASING_SEQUENCE pattern

## 3.39 ALL_EQUAL_SURF_BUMP_ON_DECREASING_SEQUENCE

► ▷ ◄  **DESCRIPTION**  **AUTOMATON**

| | |
|---|---|
| **Origin** | Based on the BUMP_ON_DECREASING_SEQUENCE pattern. |
| **Constraint** | ALL_EQUAL_SURF_BUMP_ON_DECREASING_SEQUENCE(VARIABLES) |
| **Argument** | VARIABLES : collection(var−dvar) |
| **Restriction** | required(VARIABLES, var) |
| **Purpose** | Succeeds if the surface of all occurrences of the BUMP_ON_DECREASING_SEQUENCE pattern in the time-series given by the VARIABLES collection are the same. <br> An occurrence of the pattern BUMP_ON_DECREASING_SEQUENCE is the subsequence which matches the regular expression >><>>. <br> Assume that the occurrence of the pattern BUMP_ON_DECREASING_SEQUENCE starts at position $i$ and ends at position $j$. The feature SURF computes the sum of the values from index $i + 2$ to index $j$. |
| **Example** | $(\langle 7, 6, 5, 6, 5, 4, 2, 3, 7, 4, 3, 2, 8, 6, 4, 4 \rangle)$ |

Figure 3.77 provides an example where the ALL_EQUAL_SURF_BUMP_ON_DECREASING_SEQUENCE ([7, 6, 5, 6, 5, 4, 2, 3, 7, 4, 3, 2, 8, 6, 4, 4]) constraint holds.

| | |
|---|---|
| **Typical** | $|\texttt{VARIABLES}| > 5$ <br> range(VARIABLES.var) $> 2$ |

Figure 3.77: Illustrating the ALL_EQUAL_SURF_BUMP_ON_DECREASING_SEQUENCE constraint of the **Example** slot

**Automaton**  Figure 3.78 depicts the automaton associated with the constraint ALL_EQUAL_SURF_BUMP_ON_DECREASING_SEQUENCE.

Figure 3.78: Automaton for the ALL_EQUAL_SURF_BUMP_ON_DECREASING_SEQUENCE constraint obtained by applying decoration Table 2.36 to the seed transducer of the BUMP_ON_DECREASING_SEQUENCE pattern

# 3.40 ALL_EQUAL_SURF_DECREASING

► ▷ ◄          **DESCRIPTION**          **AUTOMATON**

**Origin**          Based on the DECREASING pattern.

**Constraint**          ALL_EQUAL_SURF_DECREASING(VARIABLES)

**Argument**          VARIABLES : collection(var−dvar)

**Restriction**          required(VARIABLES, var)

**Purpose**          Succeeds if the surface of all occurrences of the DECREASING pattern in the time-series given by the VARIABLES collection are the same.

An occurrence of the pattern DECREASING is the subsequence which matches the regular expression >.

Assume that the occurrence of the pattern DECREASING starts at position $i$ and ends at position $j$. The feature SURF computes the sum of the values from index $i$ to index $j+1$.

**Example**          $(\langle 1, 5, 5, 2, 2, 3, 4, 6, 1, 2, 3, 4, 5, 2, 2, 3 \rangle)$

Figure 3.79 provides an example where the ALL_EQUAL_SURF_DECREASING ($[1, 5, 5, 2, 2, 3, 4, 6, 1, 2, 3, 4, 5, 2, 2, 3]$) constraint holds.

Figure 3.79: Illustrating the ALL_EQUAL_SURF_DECREASING constraint of the **Example** slot

**Typical**    $|\texttt{VARIABLES}| > 1$
$\texttt{range}(\texttt{VARIABLES.var}) > 1$

**Automaton**    Figure 3.80 depicts the automaton associated with the constraint ALL_EQUAL_SURF_DECREASING.

Figure 3.80: Automaton for the ALL_EQUAL_SURF_DECREASING constraint obtained by applying decoration Table 2.36 to the seed transducer of the DECREASING pattern

## 3.41  ALL_EQUAL_SURF_DECREASING_SEQUENCE

► ▷ ◄          DESCRIPTION          AUTOMATON

$> (> | =)^* > |$
$>$

| | |
|---|---|
| **Origin** | Based on the DECREASING_SEQUENCE pattern. |
| **Constraint** | ALL_EQUAL_SURF_DECREASING_SEQUENCE(VARIABLES) |
| **Argument** | VARIABLES : collection(var−dvar) |
| **Restriction** | required(VARIABLES, var) |
| **Purpose** | Succeeds if the surface of all occurrences of the DECREASING_SEQUENCE pattern in the time-series given by the VARIABLES collection are the same. <br> An occurrence of the pattern DECREASING_SEQUENCE is the *maximal* subsequence which matches the regular expression $> (> \| =)^* > \| >$. <br> Assume that the occurrence of the pattern DECREASING_SEQUENCE starts at position $i$ and ends at position $j$. The feature SURF computes the sum of the values from index $i$ to index $j + 1$. |
| **Example** | $(\langle 1, 6, 6, 4, 3, 3, 2, 2, 5, 6, 6, 5, 5, 2, 3, 3 \rangle)$ |

Figure 3.81 provides an example where the ALL_EQUAL_SURF_DECREASING_SEQUENCE ($[1, 6, 6, 4, 3, 3, 2, 2, 5, 6, 6, 5, 5, 2, 3, 3]$) constraint holds.

Figure 3.81: Illustrating the ALL_EQUAL_SURF_DECREASING_SEQUENCE constraint of the **Example** slot

**Typical**          $|\texttt{VARIABLES}| > 1$
             $\texttt{range}(\texttt{VARIABLES.var}) > 1$

**Automaton**     Figure 3.82 depicts the automaton associated with the constraint ALL_EQUAL_SURF_DECREASING_SEQUENCE.

Figure 3.82: Automaton for the ALL_EQUAL_SURF_DECREASING_SEQUENCE constraint obtained by applying decoration Table 2.36 to the seed transducer of the DECREASING_SEQUENCE pattern

## 3.42  ALL_EQUAL_SURF_DECREASING_TERRACE

▶ ▷ ◀          DESCRIPTION          AUTOMATON

| | |
|---|---|
| **Origin** | Based on the DECREASING_TERRACE pattern. |
| **Constraint** | ALL_EQUAL_SURF_DECREASING_TERRACE(VARIABLES) |
| **Argument** | VARIABLES : collection(var−dvar) |
| **Restriction** | required(VARIABLES, var) |
| **Purpose** | Succeeds if the surface of all occurrences of the DECREASING_TERRACE pattern in the time-series given by the VARIABLES collection are the same. <br> An occurrence of the pattern DECREASING_TERRACE is the *maximal* subsequence which matches the regular expression $> =^+ >$. <br> Assume that the occurrence of the pattern DECREASING_TERRACE starts at position $i$ and ends at position $j$. The feature SURF computes the sum of the values from index $i + 1$ to index $j$. |
| **Example** | $(\langle 5, 4, 4, 1, 6, 6, 2, 2, 2, 2, 1, 6, 4, 4, 2, 2 \rangle)$ |

Figure 3.83 provides an example where the ALL_EQUAL_SURF_DECREASING_TERRACE $([5, 4, 4, 1, 6, 6, 2, 2, 2, 2, 1, 6, 4, 4, 2, 2])$ constraint holds.

Figure 3.83: Illustrating the ALL_EQUAL_SURF_DECREASING_TERRACE constraint of the **Example** slot

**Typical**
$|\texttt{VARIABLES}| > 3$
$\texttt{range}(\texttt{VARIABLES.var}) > 2$

**Automaton**    Figure 3.84 depicts the automaton associated with the constraint ALL_EQUAL_SURF_DECREASING_TERRACE.

Figure 3.84: Automaton for the ALL_EQUAL_SURF_DECREASING_TERRACE constraint obtained by applying decoration Table 2.36 to the seed transducer of the DECREAS-ING_TERRACE pattern

## 3.43 ALL_EQUAL_SURF_DIP_ON_INCREASING_SEQUENCE

► ▷ ◄        DESCRIPTION        AUTOMATON

**Origin**        Based on the DIP_ON_INCREASING_SEQUENCE pattern.

**Constraint**    ALL_EQUAL_SURF_DIP_ON_INCREASING_SEQUENCE(VARIABLES)

**Argument**      VARIABLES : collection(var−dvar)

**Restriction**   required(VARIABLES, var)

**Purpose**       Succeeds if the surface of all occurrences of the DIP_ON_INCREASING_SEQUENCE pattern in the time-series given by the VARIABLES collection are the same.

An occurrence of the pattern DIP_ON_INCREASING_SEQUENCE is the subsequence which matches the regular expression <<><<.

Assume that the occurrence of the pattern DIP_ON_INCREASING_SEQUENCE starts at position $i$ and ends at position $j$. The feature SURF computes the sum of the values from index $i + 2$ to index $j$.

**Example**       $(\langle 1, 2, 3, 2, 3, 4, 6, 5, 1, 4, 5, 6, 0, 2, 4, 4 \rangle)$

Figure 3.85 provides an example where the ALL_EQUAL_SURF_DIP_ON_INCREASING_SEQUENCE $([1, 2, 3, 2, 3, 4, 6, 5, 1, 4, 5, 6, 0, 2, 4, 4])$ constraint holds.

**Typical**       $|\text{VARIABLES}| > 5$
                  range(VARIABLES.var) > 2

Figure 3.85: Illustrating the ALL_EQUAL_SURF_DIP_ON_INCREASING_SEQUENCE constraint of the **Example** slot

**Automaton**    Figure 3.86 depicts the automaton associated with the constraint ALL_EQUAL_SURF_DIP_ON_INCREASING_SEQUENCE.

Figure 3.86: Automaton for the ALL_EQUAL_SURF_DIP_ON_INCREASING_SEQUENCE constraint obtained by applying decoration Table 2.36 to the seed transducer of the DIP_ON_INCREASING_SEQUENCE pattern

## 3.44 ALL_EQUAL_SURF_GORGE

► ▷ ◄            DESCRIPTION            AUTOMATON

$(> | > (= | >)^* >)(< | < (= | <)^* <)$

| | |
|---|---|
| **Origin** | Based on the GORGE pattern. |
| **Constraint** | ALL_EQUAL_SURF_GORGE(VARIABLES) |
| **Argument** | VARIABLES : collection(var−dvar) |
| **Restriction** | required(VARIABLES, var) |
| **Purpose** | Succeeds if the surface of all occurrences of the GORGE pattern in the time-series given by the VARIABLES collection are the same. <br> An occurrence of the pattern GORGE is the *maximal* subsequence which matches the regular expression $(> | > (= | >)^* >)(< | < (= | <)^* <)$. <br> Assume that the occurrence of the pattern GORGE starts at position $i$ and ends at position $j$. The feature SURF computes the sum of the values from index $i + 1$ to index $j$. |
| **Example** | $(\langle 6, 4, 2, 3, 5, 1, 1, 5, 4, 2, 3, 6, 3, 2, 4, 5 \rangle)$ |

Figure 3.87 provides an example where the ALL_EQUAL_SURF_GORGE $([6, 4, 2, 3, 5, 1, 1, 5, 4, 2, 3, 6, 3, 2, 4, 5])$ constraint holds.

Figure 3.87: Illustrating the ALL_EQUAL_SURF_GORGE constraint of the **Example** slot

| | |
|---|---|
| **Typical** | $|\texttt{VARIABLES}| > 2$ <br> range(VARIABLES.var) $> 1$ |

**Automaton**    Figure 3.88 depicts the automaton associated with the constraint
ALL_EQUAL_SURF_GORGE.

Figure 3.88: Automaton for the ALL_EQUAL_SURF_GORGE constraint obtained by applying decoration Table 2.36 to the seed transducer of the GORGE pattern (transition $u \rightarrow r$ has the same accumulator update as transition $r \rightarrow u$)

## 3.45   ALL_EQUAL_SURF_INCREASING

► ▷   ◄          **DESCRIPTION**          **AUTOMATON**

| | |
|---|---|
| **Origin** | Based on the INCREASING pattern. |
| **Constraint** | ALL_EQUAL_SURF_INCREASING(VARIABLES) |
| **Argument** | VARIABLES : collection(var−dvar) |
| **Restriction** | required(VARIABLES, var) |
| **Purpose** | Succeeds if the surface of all occurrences of the INCREASING pattern in the time-series given by the VARIABLES collection are the same.<br><br>An occurrence of the pattern INCREASING is the subsequence which matches the regular expression <.<br><br>Assume that the occurrence of the pattern INCREASING starts at position $i$ and ends at position $j$. The feature SURF computes the sum of the values from index $i$ to index $j+1$. |
| **Example** | $(\langle 6, 1, 1, 7, 4, 4, 2, 2, 6, 4, 2, 6, 5, 3, 5, 5\rangle)$ |

Figure 3.89 provides an example where the ALL_EQUAL_SURF_INCREASING $([6, 1, 1, 7, 4, 4, 2, 2, 6, 4, 2, 6, 5, 3, 5, 5])$ constraint holds.

| | |
|---|---|
| **Typical** | $\lvert \texttt{VARIABLES} \rvert > 1$<br>range(VARIABLES.var) > 1 |

Figure 3.89: Illustrating the ALL_EQUAL_SURF_INCREASING constraint of the **Example** slot

**Automaton**     Figure 3.90 depicts the automaton associated with the constraint
ALL_EQUAL_SURF_INCREASING.

Figure 3.90: Automaton for the ALL_EQUAL_SURF_INCREASING constraint obtained
by applying decoration Table 2.36 to the seed transducer of the INCREASING pattern

# 3.46   ALL_EQUAL_SURF_INCREASING_SEQUENCE

► ▷ ◄          **DESCRIPTION**          **AUTOMATON**

$< (< | =)^* < |$
$<$

**Origin**          Based on the INCREASING_SEQUENCE pattern.

**Constraint**      ALL_EQUAL_SURF_INCREASING_SEQUENCE(VARIABLES)

**Argument**        VARIABLES : collection(var−dvar)

**Restriction**     required(VARIABLES, var)

**Purpose**         Succeeds if the surface of all occurrences of the INCREASING_SEQUENCE pattern in the time-series given by the VARIABLES collection are the same.

An occurrence of the pattern INCREASING_SEQUENCE is the *maximal* subsequence which matches the regular expression $< (< | =)^* < | <$.

Assume that the occurrence of the pattern INCREASING_SEQUENCE starts at position $i$ and ends at position $j$. The feature SURF computes the sum of the values from index $i$ to index $j + 1$.

**Example**         $(\langle 6, 1, 1, 3, 4, 4, 5, 5, 2, 1, 1, 5, 5, 6, 4, 4 \rangle)$

Figure 3.91 provides an example where the ALL_EQUAL_SURF_INCREASING_SEQUENCE $([6, 1, 1, 3, 4, 4, 5, 5, 2, 1, 1, 5, 5, 6, 4, 4])$ constraint holds.

Figure 3.91: Illustrating the ALL_EQUAL_SURF_INCREASING_SEQUENCE constraint of the **Example** slot

**Typical**

$|\mathtt{VARIABLES}| > 1$

$\mathtt{range}(\mathtt{VARIABLES.var}) > 1$

**Automaton**       Figure 3.92 depicts the automaton associated with the constraint
ALL_EQUAL_SURF_INCREASING_SEQUENCE.

Figure 3.92: Automaton for the ALL_EQUAL_SURF_INCREASING_SEQUENCE constraint obtained by applying decoration Table 2.36 to the seed transducer of the INCREASING_SEQUENCE pattern

## 3.47  ALL_EQUAL_SURF_INCREASING_TERRACE

► ▷ ◄       DESCRIPTION        AUTOMATON

| | |
|---|---|
| **Origin** | Based on the INCREASING_TERRACE pattern. |
| **Constraint** | ALL_EQUAL_SURF_INCREASING_TERRACE(VARIABLES) |
| **Argument** | VARIABLES : collection(var−dvar) |
| **Restriction** | required(VARIABLES, var) |
| **Purpose** | Succeeds if the surface of all occurrences of the INCREASING_TERRACE pattern in the time-series given by the VARIABLES collection are the same.<br>An occurrence of the pattern INCREASING_TERRACE is the *maximal* subsequence which matches the regular expression $< =^+ <$.<br>Assume that the occurrence of the pattern INCREASING_TERRACE starts at position $i$ and ends at position $j$. The feature SURF computes the sum of the values from index $i + 1$ to index $j$. |
| **Example** | $(\langle 2, 3, 3, 6, 1, 1, 2, 2, 2, 4, 6, 1, 3, 3, 5, 5 \rangle)$ |

Figure 3.93 provides an example where the ALL_EQUAL_SURF_INCREASING_TERRACE ([2, 3, 3, 6, 1, 1, 2, 2, 2, 4, 6, 1, 3, 3, 5, 5]) constraint holds.

Figure 3.93: Illustrating the ALL_EQUAL_SURF_INCREASING_TERRACE constraint of the **Example** slot

**Typical**
$|\texttt{VARIABLES}| > 3$
$\texttt{range}(\texttt{VARIABLES.var}) > 2$

**Automaton**   Figure 3.94 depicts the automaton associated with the constraint ALL_EQUAL_SURF_INCREASING_TERRACE.

Figure 3.94: Automaton for the ALL_EQUAL_SURF_INCREASING_TERRACE constraint obtained by applying decoration Table 2.36 to the seed transducer of the INCREAS-ING_TERRACE pattern

## 3.48 ALL_EQUAL_SURF_INFLEXION

▶ ▷ ◀        DESCRIPTION        AUTOMATON

$< (< | =)^* > | > (> | =)^* <$

| | |
|---|---|
| **Origin** | Based on the INFLEXION pattern. |
| **Constraint** | ALL_EQUAL_SURF_INFLEXION(VARIABLES) |
| **Argument** | VARIABLES : collection(var−dvar) |
| **Restriction** | required(VARIABLES, var) |
| **Purpose** | Succeeds if the surface of all occurrences of the INFLEXION pattern in the time-series given by the VARIABLES collection are the same.<br><br>An occurrence of the pattern INFLEXION is the *maximal* subsequence which matches the regular expression $< (< | =)^* > | > (> | =)^* <$.<br><br>Assume that the occurrence of the pattern INFLEXION starts at position $i$ and ends at position $j$. The feature SURF computes the sum of the values from index $i + 1$ to index $j$. |
| **Example** | $(\langle 2, 2, 3, 6, 6, 5, 5, 3, 2, 4, 5, 6, 3, 3, 3, 3 \rangle)$ |

Figure 3.95 provides an example where the ALL_EQUAL_SURF_INFLEXION ([2, 2, 3, 6, 6, 5, 5, 3, 2, 4, 5, 6, 3, 3, 3, 3]) constraint holds.

Figure 3.95: Illustrating the ALL_EQUAL_SURF_INFLEXION constraint of the **Example** slot

**Typical**           $|\texttt{VARIABLES}| > 2$
                      $\texttt{range}(\texttt{VARIABLES.var}) > 1$

**Automaton**    Figure 3.96 depicts the automaton associated with the constraint ALL_EQUAL_SURF_INFLEXION.

Figure 3.96: Automaton for the ALL_EQUAL_SURF_INFLEXION constraint obtained by applying decoration Table 2.36 to the seed transducer of the INFLEXION pattern (transition $r \rightarrow t$ has the same accumulators updates as transition $t \rightarrow r$)

## 3.49 ALL_EQUAL_SURF_PEAK

► ▷ ◄          DESCRIPTION          AUTOMATON

$< (= | <)^* (> | =)^* >$

| | |
|---|---|
| **Origin** | Based on the PEAK pattern. |
| **Constraint** | ALL_EQUAL_SURF_PEAK(VARIABLES) |
| **Argument** | VARIABLES : collection(var−dvar) |
| **Restriction** | required(VARIABLES, var) |
| **Purpose** | Succeeds if the surface of all occurrences of the PEAK pattern in the time-series given by the VARIABLES collection are the same. An occurrence of the pattern PEAK is the *maximal* subsequence which matches the regular expression $< (= | <)^* (> | =)^* >$. Assume that the occurrence of the pattern PEAK starts at position $i$ and ends at position $j$. The feature SURF computes the sum of the values from index $i + 1$ to index $j$. |
| **Example** | $(\langle 1, 2, 2, 7, 7, 1, 2, 4, 5, 7, 1, 2, 3, 6, 7, 1 \rangle)$ |

Figure 3.97 provides an example where the ALL_EQUAL_SURF_PEAK ($[1, 2, 2, 7, 7, 1, 2, 4, 5, 7, 1, 2, 3, 6, 7, 1]$) constraint holds.

Figure 3.97: Illustrating the ALL_EQUAL_SURF_PEAK constraint of the **Example** slot

**Typical**

$|\mathtt{VARIABLES}| > 2$
$\mathtt{range}(\mathtt{VARIABLES.var}) > 1$

          Figure 3.98 depicts the automaton associated with the constraint ALL_EQUAL_SURF_PEAK.

Figure 3.98: Automaton for the ALL_EQUAL_SURF_PEAK constraint obtained by applying decoration Table 2.36 to the seed transducer of the PEAK pattern

## 3.50 ALL_EQUAL_SURF_PLAIN

► ▷ ◄            DESCRIPTION            AUTOMATON

| | |
|---|---|
| **Origin** | Based on the PLAIN pattern. |
| **Constraint** | ALL_EQUAL_SURF_PLAIN(VARIABLES) |
| **Argument** | VARIABLES : collection(var−dvar) |
| **Restriction** | required(VARIABLES, var) |
| **Purpose** | Succeeds if the surface of all occurrences of the PLAIN pattern in the time-series given by the VARIABLES collection are the same. An occurrence of the pattern PLAIN is the *maximal* subsequence which matches the regular expression >=* <. Assume that the occurrence of the pattern PLAIN starts at position $i$ and ends at position $j$. The feature SURF computes the sum of the values from index $i+1$ to index $j$. |
| **Example** | $(\langle 1, 6, 3, 3, 7, 6, 6, 3, 3, 5, 5, 4, 3, 3, 6, 3 \rangle)$ |

Figure 3.99 provides an example where the ALL_EQUAL_SURF_PLAIN $([1, 6, 3, 3, 7, 6, 6, 3, 3, 5, 5, 4, 3, 3, 6, 3])$ constraint holds.

Figure 3.99: Illustrating the ALL_EQUAL_SURF_PLAIN constraint of the **Example** slot

**Typical**    $|\texttt{VARIABLES}| > 2$
$\texttt{range}(\texttt{VARIABLES.var}) > 1$

**Automaton**          Figure 3.100 depicts the automaton associated with the constraint
ALL_EQUAL_SURF_PLAIN.

Figure 3.100: Automaton for the ALL_EQUAL_SURF_PLAIN constraint obtained by applying decoration Table 2.36 to the seed transducer of the PLAIN pattern

## 3.51 ALL_EQUAL_SURF_PLATEAU

► ▷ ◄        DESCRIPTION          AUTOMATON

| | |
|---|---|
| **Origin** | Based on the PLATEAU pattern. |
| **Constraint** | ALL_EQUAL_SURF_PLATEAU(VARIABLES) |
| **Argument** | VARIABLES : collection(var−dvar) |
| **Restriction** | required(VARIABLES, var) |
| **Purpose** | Succeeds if the surface of all occurrences of the PLATEAU pattern in the time-series given by the VARIABLES collection are the same. An occurrence of the pattern PLATEAU is the *maximal* subsequence which matches the regular expression $<=^*>$. Assume that the occurrence of the pattern PLATEAU starts at position $i$ and ends at position $j$. The feature SURF computes the sum of the values from index $i+1$ to index $j$. |
| **Example** | $(\langle 7, 2, 5, 5, 1, 2, 2, 5, 5, 3, 3, 4, 5, 5, 2, 5 \rangle)$ |

Figure 3.101 provides an example where the ALL_EQUAL_SURF_PLATEAU $([7, 2, 5, 5, 1, 2, 2, 5, 5, 3, 3, 4, 5, 5, 2, 5])$ constraint holds.

| | |
|---|---|
| **Typical** | $\lvert$VARIABLES$\rvert > 2$ <br> range(VARIABLES.var) $> 1$ |

Figure 3.101: Illustrating the ALL_EQUAL_SURF_PLATEAU constraint of the **Example** slot

**Automaton**                Figure   3.102   depicts   the   automaton   associated   with   the   constraint
ALL_EQUAL_SURF_PLATEAU.

Figure 3.102: Automaton for the ALL_EQUAL_SURF_PLATEAU constraint obtained by applying decoration Table 2.36 to the seed transducer of the PLATEAU pattern

## 3.52 ALL_EQUAL_SURF_PROPER_PLAIN

► ▷ ◄          DESCRIPTION          AUTOMATON

| | |
|---|---|
| **Origin** | Based on the PROPER_PLAIN pattern. |
| **Constraint** | ALL_EQUAL_SURF_PROPER_PLAIN(VARIABLES) |
| **Argument** | VARIABLES : collection(var−dvar) |
| **Restriction** | required(VARIABLES, var) |
| **Purpose** | Succeeds if the surface of all occurrences of the PROPER_PLAIN pattern in the time-series given by the VARIABLES collection are the same. An occurrence of the pattern PROPER_PLAIN is the *maximal* subsequence which matches the regular expression $> =^+ <$. Assume that the occurrence of the pattern PROPER_PLAIN starts at position $i$ and ends at position $j$. The feature SURF computes the sum of the values from index $i + 1$ to index $j$. |
| **Example** | $(\langle 2, 7, 3, 3, 6, 6, 3, 7, 3, 3, 5, 6, 5, 3, 3, 5 \rangle)$ |

Figure 3.103 provides an example where the ALL_EQUAL_SURF_PROPER_PLAIN $([2, 7, 3, 3, 6, 6, 3, 7, 3, 3, 5, 6, 5, 3, 3, 5])$ constraint holds.

Figure 3.103: Illustrating the ALL_EQUAL_SURF_PROPER_PLAIN constraint of the **Example** slot

**Typical**
$$|\mathtt{VARIABLES}| > 3$$
$$\mathtt{range}(\mathtt{VARIABLES.var}) > 1$$

**Automaton**  Figure 3.104 depicts the automaton associated with the constraint ALL_EQUAL_SURF_PROPER_PLAIN.

Figure 3.104: Automaton for the ALL_EQUAL_SURF_PROPER_PLAIN constraint obtained by applying decoration Table 2.36 to the seed transducer of the PROPER_PLAIN pattern

## 3.53 ALL_EQUAL_SURF_PROPER_PLATEAU

► ▷ ◄          DESCRIPTION          AUTOMATON

**Origin**          Based on the PROPER_PLATEAU pattern.

**Constraint**      ALL_EQUAL_SURF_PROPER_PLATEAU(VARIABLES)

**Argument**        VARIABLES : collection(var−dvar)

**Restriction**     required(VARIABLES, var)

**Purpose**         Succeeds if the surface of all occurrences of the PROPER_PLATEAU pattern in the time-series given by the VARIABLES collection are the same.

An occurrence of the pattern PROPER_PLATEAU is the *maximal* subsequence which matches the regular expression $< =^+ >$.

Assume that the occurrence of the pattern PROPER_PLATEAU starts at position $i$ and ends at position $j$. The feature SURF computes the sum of the values from index $i+1$ to index $j$.

**Example**         $(\langle 7,1,5,5,2,2,5,1,5,5,3,2,3,5,5,3\rangle)$

Figure 3.105 provides an example where the ALL_EQUAL_SURF_PROPER_PLATEAU ([7, 1, 5, 5, 2, 2, 5, 1, 5, 5, 3, 2, 3, 5, 5, 3]) constraint holds.

**Typical**         $|\text{VARIABLES}| > 3$
                    range(VARIABLES.var) > 1

Figure 3.105: Illustrating the ALL_EQUAL_SURF_PROPER_PLATEAU constraint of the **Example** slot

**Automaton**          Figure  3.106  depicts  the  automaton  associated  with  the  constraint
ALL_EQUAL_SURF_PROPER_PLATEAU.

Figure 3.106: Automaton for the ALL_EQUAL_SURF_PROPER_PLATEAU constraint obtained by applying decoration Table 2.36 to the seed transducer of the PROPER_PLATEAU pattern

# 3.54 ALL_EQUAL_SURF_STEADY

► ▷ ◄          DESCRIPTION          AUTOMATON

| | |
|---|---|
| **Origin** | Based on the STEADY pattern. |
| **Constraint** | ALL_EQUAL_SURF_STEADY(VARIABLES) |
| **Argument** | VARIABLES : collection(var−dvar) |
| **Restriction** | required(VARIABLES, var) |
| **Purpose** | Succeeds if the surface of all occurrences of the STEADY pattern in the time-series given by the VARIABLES collection are the same.<br><br>An occurrence of the pattern STEADY is the subsequence which matches the regular expression =.<br><br>Assume that the occurrence of the pattern STEADY starts at position $i$ and ends at position $j$. The feature SURF computes the sum of the values from index $i$ to index $j + 1$. |
| **Example** | $(\langle 4, 4, 4, 2, 6, 4, 4, 4, 4, 3, 2, 1, 4, 4, 3, 4 \rangle)$ |

Figure 3.107 provides an example where the ALL_EQUAL_SURF_STEADY ($[4, 4, 4, 2, 6, 4, 4, 4, 4, 3, 2, 1, 4, 4, 3, 4]$) constraint holds.

Figure 3.107: Illustrating the ALL_EQUAL_SURF_STEADY constraint of the **Example** slot

**Typical**                    $|\texttt{VARIABLES}| > 1$

**Automaton**          Figure 3.108 depicts the automaton associated with the constraint
                       ALL_EQUAL_SURF_STEADY.

Figure 3.108: Automaton for the ALL_EQUAL_SURF_STEADY constraint obtained by applying decoration Table 2.36 to the seed transducer of the STEADY pattern

## 3.55 ALL_EQUAL_SURF_STEADY_SEQUENCE

► ▷   ◄        DESCRIPTION          AUTOMATON

| | |
|---|---|
| **Origin** | Based on the STEADY_SEQUENCE pattern. |
| **Constraint** | ALL_EQUAL_SURF_STEADY_SEQUENCE(VARIABLES) |
| **Argument** | VARIABLES : collection(var−dvar) |
| **Restriction** | required(VARIABLES, var) |
| **Purpose** | Succeeds if the surface of all occurrences of the STEADY_SEQUENCE pattern in the time-series given by the VARIABLES collection are the same. <br> An occurrence of the pattern STEADY_SEQUENCE is the *maximal* subsequence which matches the regular expression $=^+$. <br> Assume that the occurrence of the pattern STEADY_SEQUENCE starts at position $i$ and ends at position $j$. The feature SURF computes the sum of the values from index $i$ to index $j + 1$. |
| **Example** | $(\langle 6, 6, 5, 3, 3, 3, 3, 1, 4, 4, 4, 5, 6, 6, 7, 2 \rangle)$ |

Figure 3.109 provides an example where the ALL_EQUAL_SURF_STEADY_SEQUENCE $([6, 6, 5, 3, 3, 3, 3, 1, 4, 4, 4, 5, 6, 6, 7, 2])$ constraint holds.

| | |
|---|---|
| **Typical** | $|\texttt{VARIABLES}| > 1$ |

Figure 3.109: Illustrating the ALL_EQUAL_SURF_STEADY_SEQUENCE constraint of the **Example** slot

**Automaton**  Figure 3.110 depicts the automaton associated with the constraint ALL_EQUAL_SURF_STEADY_SEQUENCE.

Figure 3.110: Automaton for the ALL_EQUAL_SURF_STEADY_SEQUENCE constraint obtained by applying decoration Table 2.36 to the seed transducer of the STEADY_SEQUENCE pattern

## 3.56  ALL_EQUAL_SURF_STRICTLY_DECREASING_SEQUENCE

▶ ▷ ◀        DESCRIPTION        AUTOMATON

| | |
|---|---|
| **Origin** | Based on the STRICTLY_DECREASING_SEQUENCE pattern. |
| **Constraint** | ALL_EQUAL_SURF_STRICTLY_DECREASING_SEQUENCE(VARIABLES) |
| **Argument** | VARIABLES : collection(var−dvar) |
| **Restriction** | required(VARIABLES, var) |
| **Purpose** | Succeeds if the surface of all occurrences of the STRICTLY_DECREASING_SEQUENCE pattern in the time-series given by the VARIABLES collection are the same. <br><br> An occurrence of the pattern STRICTLY_DECREASING_SEQUENCE is the *maximal* subsequence which matches the regular expression $>^+$. <br><br> Assume that the occurrence of the pattern STRICTLY_DECREASING_SEQUENCE starts at position $i$ and ends at position $j$. The feature SURF computes the sum of the values from index $i$ to index $j + 1$. |
| **Example** | $(\langle 3, 6, 5, 4, 5, 5, 6, 6, 5, 3, 1, 2, 6, 5, 3, 1 \rangle)$ |

Figure 3.111 provides an example where the ALL_EQUAL_SURF_STRICTLY_DECREASING_SEQUENCE $([3, 6, 5, 4, 5, 5, 6, 6, 5, 3, 1, 2, 6, 5, 3, 1])$ constraint holds.

| | |
|---|---|
| **Typical** | $\lvert$VARIABLES$\rvert > 1$ <br> range(VARIABLES.var) $> 1$ |

Figure 3.111: Illustrating the ALL_EQUAL_SURF_STRICTLY_DECREASING_SEQUENCE constraint of the **Example** slot

**Automaton**    Figure 3.112 depicts the automaton associated with the constraint ALL_EQUAL_SURF_STRICTLY_DECREASING_SEQUENCE.

Figure 3.112: Automaton for the ALL_EQUAL_SURF_STRICTLY_DECREASING_SEQUENCE constraint obtained by applying decoration Table 2.36 to the seed transducer of the STRICTLY_DECREASING_SEQUENCE pattern

## 3.57  ALL_EQUAL_SURF_STRICTLY_INCREASING_SEQUENCE

▶ ▷ ◀          **DESCRIPTION**          **AUTOMATON**

| | |
|---|---|
| **Origin** | Based on the STRICTLY_INCREASING_SEQUENCE pattern. |
| **Constraint** | ALL_EQUAL_SURF_STRICTLY_INCREASING_SEQUENCE(VARIABLES) |
| **Argument** | VARIABLES : collection(var−dvar) |
| **Restriction** | required(VARIABLES, var) |
| **Purpose** | Succeeds if the surface of all occurrences of the STRICTLY_INCREASING_SEQUENCE pattern in the time-series given by the VARIABLES collection are the same.<br><br>An occurrence of the pattern STRICTLY_INCREASING_SEQUENCE is the *maximal* subsequence which matches the regular expression $<^+$.<br><br>Assume that the occurrence of the pattern STRICTLY_INCREASING_SEQUENCE starts at position $i$ and ends at position $j$. The feature SURF computes the sum of the values from index $i$ to index $j + 1$. |
| **Example** | $(\langle 6, 1, 1, 2, 3, 4, 5, 5, 2, 1, 1, 3, 5, 6, 4, 4 \rangle)$ |

Figure 3.113 provides an example where the ALL_EQUAL_SURF_STRICTLY_INCREASING_SEQUENCE $([6, 1, 1, 2, 3, 4, 5, 5, 2, 1, 1, 3, 5, 6, 4, 4])$ constraint holds.

| | |
|---|---|
| **Typical** | $\|\text{VARIABLES}\| > 1$<br>range(VARIABLES.var) $> 1$ |

Figure 3.113: Illustrating the ALL_EQUAL_SURF_STRICTLY_INCREASING_SEQUENCE constraint of the **Example** slot

**Automaton**    Figure 3.114 depicts the automaton associated with the constraint ALL_EQUAL_SURF_STRICTLY_INCREASING_SEQUENCE.

Figure 3.114: Automaton for the ALL_EQUAL_SURF_STRICTLY_INCREASING_SEQUENCE constraint obtained by applying decoration Table 2.36 to the seed transducer of the STRICTLY_INCREASING_SEQUENCE pattern

## 3.58 ALL_EQUAL_SURF_SUMMIT

► ▷ ◄          DESCRIPTION          AUTOMATON

$(< | < (= | <)^* <)()( > | > (= | >)^* >)$

| | |
|---|---|
| **Origin** | Based on the SUMMIT pattern. |
| **Constraint** | ALL_EQUAL_SURF_SUMMIT(VARIABLES) |
| **Argument** | VARIABLES : collection(var−dvar) |
| **Restriction** | required(VARIABLES, var) |
| **Purpose** | Succeeds if the surface of all occurrences of the SUMMIT pattern in the time-series given by the VARIABLES collection are the same. |
| | An occurrence of the pattern SUMMIT is the *maximal* subsequence which matches the regular expression $(< | < (= | <)^* <)( > | > (= | >)^* >)$. |
| | Assume that the occurrence of the pattern SUMMIT starts at position $i$ and ends at position $j$. The feature SURF computes the sum of the values from index $i + 1$ to index $j$. |
| **Example** | $(\langle 1, 3, 5, 4, 2, 6, 6, 2, 3, 5, 4, 1, 4, 5, 3, 2 \rangle)$ |

Figure 3.115 provides an example where the ALL_EQUAL_SURF_SUMMIT $([1, 3, 5, 4, 2, 6, 6, 2, 3, 5, 4, 1, 4, 5, 3, 2])$ constraint holds.

Figure 3.115: Illustrating the ALL_EQUAL_SURF_SUMMIT constraint of the **Example** slot

**Typical**         $|\texttt{VARIABLES}| > 2$
                    $\texttt{range}(\texttt{VARIABLES.var}) > 1$

**Automaton**     Figure 3.116 depicts the automaton associated with the constraint ALL_EQUAL_SURF_SUMMIT.

Figure 3.116: Automaton for the ALL_EQUAL_SURF_SUMMIT constraint obtained by applying decoration Table 2.36 to the seed transducer of the SUMMIT pattern (transition $u \to r$ has the same accumulator update as transition $r \to u$)

# 3.59 ALL_EQUAL_SURF_VALLEY

▶ ▷          ◀          DESCRIPTION          AUTOMATON

| **Origin** | Based on the VALLEY pattern. |
| **Constraint** | ALL_EQUAL_SURF_VALLEY(VARIABLES) |
| **Argument** | VARIABLES : collection(var−dvar) |
| **Restriction** | required(VARIABLES, var) |

**Purpose**

Succeeds if the surface of all occurrences of the VALLEY pattern in the time-series given by the VARIABLES collection are the same.

An occurrence of the pattern VALLEY is the *maximal* subsequence which matches the regular expression $> (= | >)^* (< | =)^* <$.

Assume that the occurrence of the pattern VALLEY starts at position $i$ and ends at position $j$. The feature SURF computes the sum of the values from index $i + 1$ to index $j$.

**Example**

$(\langle 7, 6, 6, 1, 1, 7, 6, 4, 3, 1, 7, 6, 5, 2, 1, 7 \rangle)$

Figure 3.117 provides an example where the ALL_EQUAL_SURF_VALLEY ([7, 6, 6, 1, 1, 7, 6, 4, 3, 1, 7, 6, 5, 2, 1, 7]) constraint holds.

**Typical**

$|\texttt{VARIABLES}| > 2$
$\texttt{range}(\texttt{VARIABLES.var}) > 1$

Figure 3.117: Illustrating the ALL_EQUAL_SURF_VALLEY constraint of the **Example** slot

**Automaton**  Figure 3.118 depicts the automaton associated with the constraint ALL_EQUAL_SURF_VALLEY.

Figure 3.118: Automaton for the ALL_EQUAL_SURF_VALLEY constraint obtained by applying decoration Table 2.36 to the seed transducer of the VALLEY pattern

## 3.60 ALL_EQUAL_SURF_ZIGZAG

► ▷ ◄     DESCRIPTION     AUTOMATON

| | |
|---|---|
| **Origin** | Based on the ZIGZAG pattern. |
| **Constraint** | ALL_EQUAL_SURF_ZIGZAG(VARIABLES) |
| **Argument** | VARIABLES : collection(var−dvar) |
| **Restriction** | required(VARIABLES, var) |

**Purpose**

Succeeds if the surface of all occurrences of the ZIGZAG pattern in the time-series given by the VARIABLES collection are the same.

An occurrence of the pattern ZIGZAG is the *maximal* subsequence which matches the regular expression $(<>)^+(<\mid<>) \mid (><)^+(>\mid><)$.

Assume that the occurrence of the pattern ZIGZAG starts at position $i$ and ends at position $j$. The feature SURF computes the sum of the values from index $i + 1$ to index $j$.

**Example**

$(\langle 1, 6, 1, 2, 6, 5, 1, 2, 1, 2, 1, 4, 5, 2, 4, 7 \rangle)$

Figure 3.119 provides an example where the ALL_EQUAL_SURF_ZIGZAG ([1, 6, 1, 2, 6, 5, 1, 2, 1, 2, 1, 4, 5, 2, 4, 7]) constraint holds.

**Typical**

$|\texttt{VARIABLES}| > 3$

range(VARIABLES.var) $> 1$

Figure 3.119: Illustrating the ALL_EQUAL_SURF_ZIGZAG constraint of the **Example** slot

**Automaton**       Figure 3.120 depicts the automaton associated with the constraint ALL_EQUAL_SURF_ZIGZAG.

Figure 3.120: Automaton for the ALL_EQUAL_SURF_ZIGZAG constraint obtained by applying decoration Table 2.36 to the seed transducer of the ZIGZAG pattern; (1) missing transitions from $a, b, c, d, e, f$ to $s$ are labelled by $=$; (2) on transitions from $b, c, e, f$ to $s$ the accumulator $D$ is reset to its initial value; (3) on transitions from $c, f$ to $s$ the accumulator $F$ is reset to $C$, and the accumulator $R$ is updated wrt $C$ and $F$

## 3.61  ALL_EQUAL_WIDTH_DECREASING_SEQUENCE

► ▷ ◄          **DESCRIPTION**          **AUTOMATON**

| | |
|---|---|
| **Origin** | Based on the DECREASING_SEQUENCE pattern. |
| **Constraint** | ALL_EQUAL_WIDTH_DECREASING_SEQUENCE(VARIABLES) |
| **Argument** | VARIABLES : collection(var−dvar) |
| **Restriction** | required(VARIABLES, var) |
| **Purpose** | Succeeds if the width of all occurrences of the DECREASING_SEQUENCE pattern in the time-series given by the VARIABLES collection are the same.<br>An occurrence of the pattern DECREASING_SEQUENCE is the *maximal* subsequence which matches the regular expression $> (> | =)^* > | >$.<br>Assume that the occurrence of the pattern DECREASING_SEQUENCE starts at position $i$ and ends at position $j$. The feature WIDTH computes the value $j - i + 2$. |
| **Example** | $(\langle 1, 3, 3, 2, 2, 1, 1, 4, 6, 6, 4, 4, 2, 2, 3, 4 \rangle)$ |

Figure 3.121 provides an example where the ALL_EQUAL_WIDTH_DECREASING_SEQUENCE $([1, 3, 3, 2, 2, 1, 1, 4, 6, 6, 4, 4, 2, 2, 3, 4])$ constraint holds.

Figure 3.121: Illustrating the ALL_EQUAL_WIDTH_DECREASING_SEQUENCE constraint of the **Example** slot

**Typical**
$|\texttt{VARIABLES}| > 1$
$\texttt{range}(\texttt{VARIABLES.var}) > 1$

**Automaton**          Figure 3.122 depicts the automaton associated with the constraint
ALL_EQUAL_WIDTH_DECREASING_SEQUENCE.

Figure 3.122: Automaton for the ALL_EQUAL_WIDTH_DECREASING_SEQUENCE constraint obtained by applying decoration Table 2.36 to the seed transducer of the DECREASING_SEQUENCE pattern

## 3.62 ALL_EQUAL_WIDTH_DECREASING_TERRACE

▶  ▷  ◀        DESCRIPTION        AUTOMATON

**Origin**        Based on the DECREASING_TERRACE pattern.

**Constraint**    ALL_EQUAL_WIDTH_DECREASING_TERRACE(VARIABLES)

**Argument**      VARIABLES : collection(var−dvar)

**Restriction**   required(VARIABLES, var)

**Purpose**       Succeeds if the width of all occurrences of the DECREASING_TERRACE pattern in the time-series given by the VARIABLES collection are the same.

An occurrence of the pattern DECREASING_TERRACE is the *maximal* subsequence which matches the regular expression $> =^+ >$.

Assume that the occurrence of the pattern DECREASING_TERRACE starts at position $i$ and ends at position $j$. The feature WIDTH computes the value $j - i$.

**Example**       $(\langle 6, 5, 5, 4, 3, 3, 1, 1, 3, 3, 5, 2, 2, 1, 1, 1\rangle)$

Figure 3.123 provides an example where the ALL_EQUAL_WIDTH_DECREASING_TERRACE $([6, 5, 5, 4, 3, 3, 1, 1, 3, 3, 5, 2, 2, 1, 1, 1])$ constraint holds.

Figure 3.123: Illustrating the ALL_EQUAL_WIDTH_DECREASING_TERRACE constraint of the **Example** slot

**Typical**     $|\texttt{VARIABLES}| > 3$
$\texttt{range}(\texttt{VARIABLES.var}) > 2$

**Automaton**     Figure 3.124 depicts the automaton associated with the constraint ALL_EQUAL_WIDTH_DECREASING_TERRACE.

Figure 3.124: Automaton for the ALL_EQUAL_WIDTH_DECREASING_TERRACE constraint obtained by applying decoration Table 2.36 to the seed transducer of the DECREASING_TERRACE pattern

## 3.63 ALL_EQUAL_WIDTH_GORGE

► ▷ ◄        DESCRIPTION        AUTOMATON

| | |
|---|---|
| **Origin** | Based on the GORGE pattern. |
| **Constraint** | ALL_EQUAL_WIDTH_GORGE(VARIABLES) |
| **Argument** | VARIABLES : collection(var−dvar) |
| **Restriction** | required(VARIABLES, var) |
| **Purpose** | Succeeds if the width of all occurrences of the GORGE pattern in the time-series given by the VARIABLES collection are the same. <br> An occurrence of the pattern GORGE is the *maximal* subsequence which matches the regular expression $(> | > (= | >)^* )(< | < (= | <)^* <)$. <br> Assume that the occurrence of the pattern GORGE starts at position $i$ and ends at position $j$. The feature WIDTH computes the value $j - i$. |
| **Example** | $(\langle 6, 4, 2, 3, 5, 1, 1, 5, 4, 2, 3, 6, 3, 2, 4, 5 \rangle)$ |

Figure 3.125 provides an example where the ALL_EQUAL_WIDTH_GORGE ([6, 4, 2, 3, 5, 1, 1, 5, 4, 2, 3, 6, 3, 2, 4, 5]) constraint holds.

Figure 3.125: Illustrating the ALL_EQUAL_WIDTH_GORGE constraint of the **Example** slot

**Typical**
$$|\texttt{VARIABLES}| > 2$$
$$\texttt{range}(\texttt{VARIABLES.var}) > 1$$

**Automaton**    Figure 3.126 depicts the automaton associated with the constraint ALL_EQUAL_WIDTH_GORGE.

Figure 3.126: Automaton for the ALL_EQUAL_WIDTH_GORGE constraint obtained by applying decoration Table 2.36 to the seed transducer of the GORGE pattern (transition $u \rightarrow r$ has the same accumulator update as transition $r \rightarrow u$)

# 3.64 ALL_EQUAL_WIDTH_INCREASING_SEQUENCE

▶ ▷ ◀    DESCRIPTION    AUTOMATON

$< (< | =)^* < |$
$<$

| | |
|---|---|
| **Origin** | Based on the INCREASING_SEQUENCE pattern. |
| **Constraint** | ALL_EQUAL_WIDTH_INCREASING_SEQUENCE(VARIABLES) |
| **Argument** | VARIABLES : collection(var−dvar) |
| **Restriction** | required(VARIABLES, var) |
| **Purpose** | Succeeds if the width of all occurrences of the INCREASING_SEQUENCE pattern in the time-series given by the VARIABLES collection are the same. An occurrence of the pattern INCREASING_SEQUENCE is the *maximal* subsequence which matches the regular expression $< (< | =)^* < | <$. Assume that the occurrence of the pattern INCREASING_SEQUENCE starts at position $i$ and ends at position $j$. The feature WIDTH computes the value $j − i + 2$. |
| **Example** | $(\langle 6, 4, 4, 5, 5, 6, 6, 3, 1, 1, 3, 3, 5, 5, 4, 3 \rangle)$ |

Figure 3.127 provides an example where the ALL_EQUAL_WIDTH_INCREASING_SEQUENCE ([6, 4, 4, 5, 5, 6, 6, 3, 1, 1, 3, 3, 5, 5, 4, 3]) constraint holds.

Figure 3.127: Illustrating the ALL_EQUAL_WIDTH_INCREASING_SEQUENCE constraint of the **Example** slot

**Typical**         $|\mathtt{VARIABLES}| > 1$
$\mathtt{range}(\mathtt{VARIABLES.var}) > 1$

**Automaton**          Figure    3.128    depicts    the    automaton    associated    with    the    constraint
ALL_EQUAL_WIDTH_INCREASING_SEQUENCE.

Figure 3.128: Automaton for the ALL_EQUAL_WIDTH_INCREASING_SEQUENCE constraint obtained by applying decoration Table 2.36 to the seed transducer of the INCREASING_SEQUENCE pattern

# 3.65 ALL_EQUAL_WIDTH_INCREASING_TERRACE

▶ ▷ ◀        DESCRIPTION        AUTOMATON

| | |
|---|---|
| **Origin** | Based on the INCREASING_TERRACE pattern. |
| **Constraint** | ALL_EQUAL_WIDTH_INCREASING_TERRACE(VARIABLES) |
| **Argument** | VARIABLES : collection(var−dvar) |
| **Restriction** | required(VARIABLES, var) |
| **Purpose** | Succeeds if the width of all occurrences of the INCREASING_TERRACE pattern in the time-series given by the VARIABLES collection are the same.<br>An occurrence of the pattern INCREASING_TERRACE is the *maximal* subsequence which matches the regular expression $<=^+<$.<br>Assume that the occurrence of the pattern INCREASING_TERRACE starts at position $i$ and ends at position $j$. The feature WIDTH computes the value $j - i$. |
| **Example** | $(\langle 1, 2, 2, 3, 4, 4, 6, 6, 4, 4, 2, 5, 5, 6, 6, 6 \rangle)$ |

Figure 3.129 provides an example where the ALL_EQUAL_WIDTH_INCREASING_TERRACE $([1, 2, 2, 3, 4, 4, 6, 6, 4, 4, 2, 5, 5, 6, 6, 6])$ constraint holds.

Figure 3.129: Illustrating the ALL_EQUAL_WIDTH_INCREASING_TERRACE constraint of the **Example** slot

**Typical**          $|\mathtt{VARIABLES}| > 3$
$\mathtt{range}(\mathtt{VARIABLES.var}) > 2$

**Automaton**    Figure 3.130 depicts the automaton associated with the constraint ALL_EQUAL_WIDTH_INCREASING_TERRACE.

Figure 3.130: Automaton for the ALL_EQUAL_WIDTH_INCREASING_TERRACE constraint obtained by applying decoration Table 2.36 to the seed transducer of the IN-CREASING_TERRACE pattern

## 3.66 ALL_EQUAL_WIDTH_INFLEXION

► ▷ ◄          DESCRIPTION          AUTOMATON

$< (< | =)^* > | > (> | =)^* <$

| | |
|---|---|
| **Origin** | Based on the INFLEXION pattern. |
| **Constraint** | ALL_EQUAL_WIDTH_INFLEXION(VARIABLES) |
| **Argument** | VARIABLES : collection(var−dvar) |
| **Restriction** | required(VARIABLES, var) |
| **Purpose** | Succeeds if the width of all occurrences of the INFLEXION pattern in the time-series given by the VARIABLES collection are the same. <br> An occurrence of the pattern INFLEXION is the *maximal* subsequence which matches the regular expression $< (< | =)^* > | > (> | =)^* <$. <br> Assume that the occurrence of the pattern INFLEXION starts at position $i$ and ends at position $j$. The feature WIDTH computes the value $j - i$. |
| **Example** | $(\langle 2, 2, 3, 6, 6, 5, 3, 3, 4, 4, 5, 3, 3, 3, 4, 6 \rangle)$ |

Figure 3.131 provides an example where the ALL_EQUAL_WIDTH_INFLEXION $([2, 2, 3, 6, 6, 5, 3, 3, 4, 4, 5, 3, 3, 3, 4, 6])$ constraint holds.

Figure 3.131: Illustrating the ALL_EQUAL_WIDTH_INFLEXION constraint of the **Example** slot

| | |
|---|---|
| **Typical** | $\|\text{VARIABLES}\| > 2$ <br> range(VARIABLES.var) > 1 |

**Automaton** Figure 3.132 depicts the automaton associated with the constraint ALL_EQUAL_WIDTH_INFLEXION.

Figure 3.132: Automaton for the ALL_EQUAL_WIDTH_INFLEXION constraint obtained by applying decoration Table 2.36 to the seed transducer of the INFLEXION pattern (transition $r \to t$ has the same accumulators updates as transition $t \to r$)

## 3.67  ALL_EQUAL_WIDTH_PEAK

►  ▷      ◄        DESCRIPTION            AUTOMATON

| | |
|---|---|
| **Origin** | Based on the PEAK pattern. |
| **Constraint** | ALL_EQUAL_WIDTH_PEAK(VARIABLES) |
| **Argument** | VARIABLES : collection(var−dvar) |
| **Restriction** | required(VARIABLES, var) |
| **Purpose** | Succeeds if the width of all occurrences of the PEAK pattern in the time-series given by the VARIABLES collection are the same. <br><br> An occurrence of the pattern PEAK is the *maximal* subsequence which matches the regular expression $< (= \mid <)^* (> \mid =)^* >$. <br><br> Assume that the occurrence of the pattern PEAK starts at position $i$ and ends at position $j$. The feature WIDTH computes the value $j - i$. |
| **Example** | $(\langle 1, 2, 2, 7, 7, 1, 2, 4, 5, 7, 1, 2, 3, 6, 7, 1 \rangle)$ |

Figure 3.133 provides an example where the ALL_EQUAL_WIDTH_PEAK ($[1, 2, 2, 7, 7, 1, 2, 4, 5, 7, 1, 2, 3, 6, 7, 1]$) constraint holds.

| | |
|---|---|
| **Typical** | $\lvert \text{VARIABLES} \rvert > 2$ <br> range(VARIABLES.var) $> 1$ |

Figure 3.133: Illustrating the ALL_EQUAL_WIDTH_PEAK constraint of the **Example** slot

**Automaton**     Figure 3.134 depicts the automaton associated with the constraint ALL_EQUAL_WIDTH_PEAK.

Figure 3.134: Automaton for the ALL_EQUAL_WIDTH_PEAK constraint obtained by applying decoration Table 2.36 to the seed transducer of the PEAK pattern

## 3.68 ALL_EQUAL_WIDTH_PLAIN

| | |
|---|---|
| **Origin** | Based on the PLAIN pattern. |
| **Constraint** | ALL_EQUAL_WIDTH_PLAIN(VARIABLES) |
| **Argument** | VARIABLES : collection(var−dvar) |
| **Restriction** | required(VARIABLES, var) |

**Purpose**

Succeeds if the width of all occurrences of the PLAIN pattern in the time-series given by the VARIABLES collection are the same.

An occurrence of the pattern PLAIN is the *maximal* subsequence which matches the regular expression $> =^* <$.

Assume that the occurrence of the pattern PLAIN starts at position $i$ and ends at position $j$. The feature WIDTH computes the value $j - i$.

**Example**

$(\langle 1, 6, 3, 3, 7, 6, 6, 3, 3, 5, 5, 4, 3, 3, 6, 3 \rangle)$

Figure 3.135 provides an example where the ALL_EQUAL_WIDTH_PLAIN $([1, 6, 3, 3, 7, 6, 6, 3, 3, 5, 5, 4, 3, 3, 6, 3])$ constraint holds.

**Typical**

$|\texttt{VARIABLES}| > 2$
range(VARIABLES.var) $> 1$

Figure 3.135: Illustrating the ALL_EQUAL_WIDTH_PLAIN constraint of the **Example** slot

**Automaton**                 Figure     3.136    depicts    the    automaton    associated    with    the    constraint
ALL_EQUAL_WIDTH_PLAIN.

Figure 3.136: Automaton for the ALL_EQUAL_WIDTH_PLAIN constraint obtained by applying decoration Table 2.36 to the seed transducer of the PLAIN pattern

## 3.69 ALL_EQUAL_WIDTH_PLATEAU

► ▷ ◄          DESCRIPTION                AUTOMATON

| | |
|---|---|
| **Origin** | Based on the PLATEAU pattern. |
| **Constraint** | ALL_EQUAL_WIDTH_PLATEAU(VARIABLES) |
| **Argument** | VARIABLES : collection(var−dvar) |
| **Restriction** | required(VARIABLES, var) |
| **Purpose** | Succeeds if the width of all occurrences of the PLATEAU pattern in the time-series given by the VARIABLES collection are the same.<br>An occurrence of the pattern PLATEAU is the *maximal* subsequence which matches the regular expression $< = ^{*} >$.<br>Assume that the occurrence of the pattern PLATEAU starts at position $i$ and ends at position $j$. The feature WIDTH computes the value $j - i$. |
| **Example** | $(\langle 7, 2, 5, 5, 1, 2, 2, 5, 5, 3, 3, 4, 5, 5, 2, 5 \rangle)$ |

Figure 3.137 provides an example where the ALL_EQUAL_WIDTH_PLATEAU ([7, 2, 5, 5, 1, 2, 2, 5, 5, 3, 3, 4, 5, 5, 2, 5]) constraint holds.

| | |
|---|---|
| **Typical** | $|$VARIABLES$| > 2$<br>range(VARIABLES.var) $> 1$ |

Figure 3.137: Illustrating the ALL_EQUAL_WIDTH_PLATEAU constraint of the **Example** slot

**Automaton**     Figure   3.138   depicts   the   automaton   associated   with   the   constraint
ALL_EQUAL_WIDTH_PLATEAU.

Figure 3.138: Automaton for the ALL_EQUAL_WIDTH_PLATEAU constraint obtained by applying decoration Table 2.36 to the seed transducer of the PLATEAU pattern

## 3.70  ALL_EQUAL_WIDTH_PROPER_PLAIN

        DESCRIPTION        AUTOMATON

| | |
|---|---|
| **Origin** | Based on the PROPER_PLAIN pattern. |
| **Constraint** | ALL_EQUAL_WIDTH_PROPER_PLAIN(VARIABLES) |
| **Argument** | VARIABLES : collection(var−dvar) |
| **Restriction** | required(VARIABLES, var) |
| **Purpose** | Succeeds if the width of all occurrences of the PROPER_PLAIN pattern in the time-series given by the VARIABLES collection are the same. An occurrence of the pattern PROPER_PLAIN is the *maximal* subsequence which matches the regular expression $> =^+ <$. Assume that the occurrence of the pattern PROPER_PLAIN starts at position $i$ and ends at position $j$. The feature WIDTH computes the value $j − i$. |
| **Example** | $(\langle 2, 7, 3, 3, 6, 6, 3, 7, 3, 3, 5, 6, 5, 3, 3, 5\rangle)$ |

Figure 3.139 provides an example where the ALL_EQUAL_WIDTH_PROPER_PLAIN ([2, 7, 3, 3, 6, 6, 3, 7, 3, 3, 5, 6, 5, 3, 3, 5]) constraint holds.

Figure 3.139: Illustrating the ALL_EQUAL_WIDTH_PROPER_PLAIN constraint of the **Example** slot

**Typical**  $|\texttt{VARIABLES}| > 3$
$\texttt{range}(\texttt{VARIABLES.var}) > 1$

**Automaton**          Figure 3.140 depicts the automaton associated with the constraint
ALL_EQUAL_WIDTH_PROPER_PLAIN.

Figure 3.140: Automaton for the ALL_EQUAL_WIDTH_PROPER_PLAIN constraint obtained by applying decoration Table 2.36 to the seed transducer of the PROPER_PLAIN pattern

## 3.71  ALL_EQUAL_WIDTH_PROPER_PLATEAU

▶ ▷ ◀        DESCRIPTION        AUTOMATON

| | |
|---|---|
| **Origin** | Based on the PROPER_PLATEAU pattern. |
| **Constraint** | ALL_EQUAL_WIDTH_PROPER_PLATEAU(VARIABLES) |
| **Argument** | VARIABLES : collection(var−dvar) |
| **Restriction** | required(VARIABLES, var) |
| **Purpose** | Succeeds if the width of all occurrences of the PROPER_PLATEAU pattern in the time-series given by the VARIABLES collection are the same.<br><br>An occurrence of the pattern PROPER_PLATEAU is the *maximal* subsequence which matches the regular expression < =⁺ >.<br><br>Assume that the occurrence of the pattern PROPER_PLATEAU starts at position $i$ and ends at position $j$. The feature WIDTH computes the value $j - i$. |
| **Example** | $(\langle 7, 1, 5, 5, 2, 2, 5, 1, 5, 5, 3, 2, 3, 5, 5, 3 \rangle)$ |

Figure 3.141 provides an example where the ALL_EQUAL_WIDTH_PROPER_PLATEAU $([7, 1, 5, 5, 2, 2, 5, 1, 5, 5, 3, 2, 3, 5, 5, 3])$ constraint holds.

| | |
|---|---|
| **Typical** | $\lvert$VARIABLES$\rvert > 3$<br>range(VARIABLES.var) $> 1$ |

Figure 3.141: Illustrating the ALL_EQUAL_WIDTH_PROPER_PLATEAU constraint of the **Example** slot

**Automaton**       Figure 3.142 depicts the automaton associated with the constraint
ALL_EQUAL_WIDTH_PROPER_PLATEAU.

Figure 3.142: Automaton for the ALL_EQUAL_WIDTH_PROPER_PLATEAU con-
straint obtained by applying decoration Table 2.36 to the seed transducer of the
PROPER_PLATEAU pattern

# 3.72 ALL_EQUAL_WIDTH_STEADY_SEQUENCE

►  ▷  ◄          DESCRIPTION          AUTOMATON

**Origin**          Based on the STEADY_SEQUENCE pattern.

**Constraint**          ALL_EQUAL_WIDTH_STEADY_SEQUENCE(VARIABLES)

**Argument**          VARIABLES  :  collection(var−dvar)

**Restriction**          required(VARIABLES, var)

**Purpose**          Succeeds if the width of all occurrences of the STEADY_SEQUENCE pattern in the time-series given by the VARIABLES collection are the same.

An occurrence of the pattern STEADY_SEQUENCE is the *maximal* subsequence which matches the regular expression =$^+$.

Assume that the occurrence of the pattern STEADY_SEQUENCE starts at position $i$ and ends at position $j$. The feature WIDTH computes the value $j − i + 2$.

**Example**          $(\langle 3, 5, 5, 5, 3, 1, 4, 4, 4, 6, 5, 3, 2, 2, 2, 5 \rangle)$

Figure 3.143 provides an example where the ALL_EQUAL_WIDTH_STEADY_SEQUENCE ([3, 5, 5, 5, 3, 1, 4, 4, 4, 6, 5, 3, 2, 2, 2, 5]) constraint holds.

Figure 3.143: Illustrating the ALL_EQUAL_WIDTH_STEADY_SEQUENCE constraint of the **Example** slot

**Typical**    $|VARIABLES| > 1$

**Automaton**          Figure 3.144 depicts the automaton associated with the constraint
ALL_EQUAL_WIDTH_STEADY_SEQUENCE.

$$\left\{ \begin{array}{l} C \leftarrow X \\ D \leftarrow 0 \\ F \leftarrow X \\ R \leftarrow 1 \end{array} \right\}$$

Figure 3.144: Automaton for the ALL_EQUAL_WIDTH_STEADY_SEQUENCE constraint obtained by applying decoration Table 2.36 to the seed transducer of the STEADY_SEQUENCE pattern

# 3.73  ALL_EQUAL_WIDTH_STRICTLY_DECREASING_SEQUENCE

▶ ▷ ◀        **DESCRIPTION**        **AUTOMATON**

**Origin**            Based on the STRICTLY_DECREASING_SEQUENCE pattern.

**Constraint**        ALL_EQUAL_WIDTH_STRICTLY_DECREASING_SEQUENCE(VARIABLES)

**Argument**          VARIABLES : collection(var−dvar)

**Restriction**       required(VARIABLES, var)

**Purpose**           Succeeds if the width of all occurrences of the STRICTLY_DECREASING_SEQUENCE pattern in the time-series given by the VARIABLES collection are the same.
                      An occurrence of the pattern STRICTLY_DECREASING_SEQUENCE is the *maximal* subsequence which matches the regular expression $>^+$.
                      Assume that the occurrence of the pattern STRICTLY_DECREASING_SEQUENCE starts at position $i$ and ends at position $j$. The feature WIDTH computes the value $j - i + 2$.

**Example**           $(\langle 4, 3, 2, 2, 4, 4, 6, 6, 4, 1, 1, 2, 2, 4, 3, 1 \rangle)$

Figure 3.145 provides an example where the ALL_EQUAL_WIDTH_STRICTLY_DECREASING_SEQUENCE ([4, 3, 2, 2, 4, 4, 6, 6, 4, 1, 1, 2, 2, 4, 3, 1]) constraint holds.

Figure 3.145: Illustrating the ALL_EQUAL_WIDTH_STRICTLY_DECREASING_SEQUENCE constraint of the **Example** slot

**Typical**

$|\texttt{VARIABLES}| > 1$
$\texttt{range}(\texttt{VARIABLES.var}) > 1$

**Automaton**    Figure 3.146 depicts the automaton associated with the constraint ALL_EQUAL_WIDTH_STRICTLY_DECREASING_SEQUENCE.

Figure 3.146: Automaton for the ALL_EQUAL_WIDTH_STRICTLY_DECREASING_SEQUENCE constraint obtained by applying decoration Table 2.36 to the seed transducer of the STRICTLY_DECREASING_SEQUENCE pattern

## 3.74 ALL_EQUAL_WIDTH_STRICTLY_INCREASING_SEQUENCE

▶ ▷ ◀        DESCRIPTION        AUTOMATON

| | |
|---|---|
| **Origin** | Based on the STRICTLY_INCREASING_SEQUENCE pattern. |
| **Constraint** | ALL_EQUAL_WIDTH_STRICTLY_INCREASING_SEQUENCE(VARIABLES) |
| **Argument** | VARIABLES : collection(var−dvar) |
| **Restriction** | required(VARIABLES, var) |
| **Purpose** | Succeeds if the width of all occurrences of the STRICTLY_INCREASING_SEQUENCE pattern in the time-series given by the VARIABLES collection are the same. An occurrence of the pattern STRICTLY_INCREASING_SEQUENCE is the *maximal* subsequence which matches the regular expression $<^+$. Assume that the occurrence of the pattern STRICTLY_INCREASING_SEQUENCE starts at position $i$ and ends at position $j$. The feature WIDTH computes the value $j - i + 2$. |
| **Example** | $(\langle 6, 3, 3, 4, 5, 6, 6, 3, 1, 1, 2, 3, 5, 5, 4, 3 \rangle)$ |

Figure 3.147 provides an example where the ALL_EQUAL_WIDTH_STRICTLY_INCREASING_SEQUENCE
([6, 3, 3, 4, 5, 6, 6, 3, 1, 1, 2, 3, 5, 5, 4, 3]) constraint holds.

Figure 3.147: Illustrating the ALL_EQUAL_WIDTH_STRICTLY_INCREASING_SEQUENCE
constraint of the **Example** slot

**Typical**
$|\texttt{VARIABLES}| > 1$
$\texttt{range}(\texttt{VARIABLES.var}) > 1$

**Automaton**            Figure   3.148   depicts   the   automaton   associated   with   the   constraint
ALL_EQUAL_WIDTH_STRICTLY_INCREASING_SEQUENCE.

Figure 3.148: Automaton for the ALL_EQUAL_WIDTH_STRICTLY_INCREASING_SEQUENCE constraint obtained by applying decoration Table 2.36 to the seed transducer of the STRICTLY_INCREASING_SEQUENCE pattern

## 3.75  ALL_EQUAL_WIDTH_SUMMIT

► ▷ ◄          DESCRIPTION          AUTOMATON

$$(< | < (= | <)^* <)(>|> (= | >)^* >)$$

| **Origin** | Based on the SUMMIT pattern. |
| **Constraint** | ALL_EQUAL_WIDTH_SUMMIT(VARIABLES) |
| **Argument** | VARIABLES : collection(var−dvar) |
| **Restriction** | required(VARIABLES, var) |
| **Purpose** | Succeeds if the width of all occurrences of the SUMMIT pattern in the time-series given by the VARIABLES collection are the same. <br> An occurrence of the pattern SUMMIT is the *maximal* subsequence which matches the regular expression $(< | < (= | <)^* <)(> | > (= | >)^* >)$. <br> Assume that the occurrence of the pattern SUMMIT starts at position $i$ and ends at position $j$. The feature WIDTH computes the value $j - i$. |
| **Example** | $(\langle 1, 3, 5, 4, 2, 6, 6, 2, 3, 5, 4, 1, 4, 5, 3, 2\rangle)$ |

Figure 3.149 provides an example where the ALL_EQUAL_WIDTH_SUMMIT ($[1, 3, 5, 4, 2, 6, 6, 2, 3, 5, 4, 1, 4, 5, 3, 2]$) constraint holds.

Figure 3.149: Illustrating the ALL_EQUAL_WIDTH_SUMMIT constraint of the **Example** slot

**Typical**          $|\texttt{VARIABLES}| > 2$
                     $\texttt{range}(\texttt{VARIABLES.var}) > 1$

**Automaton**    Figure 3.150 depicts the automaton associated with the constraint ALL_EQUAL_WIDTH_SUMMIT.

Figure 3.150: Automaton for the ALL_EQUAL_WIDTH_SUMMIT constraint obtained by applying decoration Table 2.36 to the seed transducer of the SUMMIT pattern (transition $u \to r$ has the same accumulator update as transition $r \to u$)

## 3.76 ALL_EQUAL_WIDTH_VALLEY

▶ ▷    ◀         DESCRIPTION            AUTOMATON

$> (= | >)^* (< | =)^* <$

| | |
|---|---|
| **Origin** | Based on the VALLEY pattern. |
| **Constraint** | ALL_EQUAL_WIDTH_VALLEY(VARIABLES) |
| **Argument** | VARIABLES : collection(var−dvar) |
| **Restriction** | required(VARIABLES, var) |
| **Purpose** | Succeeds if the width of all occurrences of the VALLEY pattern in the time-series given by the VARIABLES collection are the same. |
| | An occurrence of the pattern VALLEY is the *maximal* subsequence which matches the regular expression $> (= | >)^* (< | =)^* <$. |
| | Assume that the occurrence of the pattern VALLEY starts at position $i$ and ends at position $j$. The feature WIDTH computes the value $j - i$. |
| **Example** | $(\langle 7, 6, 6, 1, 1, 7, 6, 4, 3, 1, 7, 6, 5, 2, 1, 7 \rangle)$ |

Figure 3.151 provides an example where the ALL_EQUAL_WIDTH_VALLEY ([7, 6, 6, 1, 1, 7, 6, 4, 3, 1, 7, 6, 5, 2, 1, 7]) constraint holds.

| | |
|---|---|
| **Typical** | $|\texttt{VARIABLES}| > 2$ |
| | range(VARIABLES.var) $> 1$ |

Figure 3.151: Illustrating the ALL_EQUAL_WIDTH_VALLEY constraint of the **Example** slot

**Automaton**    Figure    3.152    depicts    the    automaton    associated    with    the    constraint
ALL_EQUAL_WIDTH_VALLEY.

Figure 3.152: Automaton for the ALL_EQUAL_WIDTH_VALLEY constraint obtained by applying decoration Table 2.36 to the seed transducer of the VALLEY pattern

## 3.77  ALL_EQUAL_WIDTH_ZIGZAG

▶ ▷ ◀          DESCRIPTION          AUTOMATON

$(<>)^+(<|<>)\,|\,(><)^+(>|><)$

| **Origin** | Based on the ZIGZAG pattern. |
|---|---|

| **Constraint** | ALL_EQUAL_WIDTH_ZIGZAG(VARIABLES) |
|---|---|

| **Argument** | VARIABLES : collection(var−dvar) |
|---|---|

| **Restriction** | required(VARIABLES, var) |
|---|---|

| **Purpose** | Succeeds if the width of all occurrences of the ZIGZAG pattern in the time-series given by the VARIABLES collection are the same. |
|---|---|
| | An occurrence of the pattern ZIGZAG is the *maximal* subsequence which matches the regular expression $(<>)^+(<|<>)\,|\,(><)^+(>|><)$. |
| | Assume that the occurrence of the pattern ZIGZAG starts at position $i$ and ends at position $j$. The feature WIDTH computes the value $j - i$. |

| **Example** | $(\langle 3, 4, 1, 4, 3, 3, 4, 2, 6, 3, 2, 0, 3, 1, 2, 3\rangle)$ |
|---|---|

Figure 3.153 provides an example where the ALL_EQUAL_WIDTH_ZIGZAG ([3, 4, 1, 4, 3, 3, 4, 2, 6, 3, 2, 0, 3, 1, 2, 3]) constraint holds.

| **Typical** | $|\mathtt{VARIABLES}| > 3$ |
|---|---|
| | range(VARIABLES.var) $> 1$ |

Figure 3.153: Illustrating the ALL_EQUAL_WIDTH_ZIGZAG constraint of the **Example** slot

**Automaton**          Figure 3.154 depicts the automaton associated with the constraint
                       ALL_EQUAL_WIDTH_ZIGZAG.

Figure 3.154: Automaton for the ALL_EQUAL_WIDTH_ZIGZAG constraint obtained by applying decoration Table 2.36 to the seed transducer of the ZIGZAG pattern; (1) missing transitions from $a, b, c, d, e, f$ to $s$ are labelled by $=$; (2) on transitions from $b, c, e, f$ to $s$ the accumulator $D$ is reset to its initial value; (3) on transitions from $c, f$ to $s$ the accumulator $F$ is reset to $C$, and the accumulator $R$ is updated wrt $C$ and $F$

## 3.78　DECREASING_HEIGHT_DECREASING_TERRACE

▶ ▷ ◁ ◀　　**DESCRIPTION**　　　**AUTOMATON**

| | |
|---|---|
| **Origin** | Based on the DECREASING_TERRACE pattern. |
| **Constraint** | DECREASING_HEIGHT_DECREASING_TERRACE(VARIABLES) |
| **Argument** | VARIABLES : collection(var−dvar) |
| **Restriction** | required(VARIABLES, var) |

**Purpose**

> Succeeds if the minima of the values in each occurrence of the DECREASING_TERRACE pattern in the time-series given by the VARIABLES collection are decreasing.
> An occurrence of the pattern DECREASING_TERRACE is the *maximal* subsequence which matches the regular expression $> =^+ >$.
> Assume that the occurrence of the pattern DECREASING_TERRACE starts at position $i$ and ends at position $j$. The feature MIN computes the minimum of the values from index $i + 1$ to index $j$.

**Example**

> $(\langle 7, 6, 6, 6, 4, 3, 3, 2, 2, 4, 4, 6, 3, 3, 1, 1 \rangle)$

Figure 3.155 provides an example where the DECREAS­ING_HEIGHT_DECREASING_TERRACE $([7, 6, 6, 6, 4, 3, 3, 2, 2, 4, 4, 6, 3, 3, 1, 1])$ con­straint holds.

**Typical**

> $|\texttt{VARIABLES}| > 3$
> range(VARIABLES.var) > 2

Figure 3.155: Illustrating the DECREASING_HEIGHT_DECREASING_TERRACE constraint of the **Example** slot

**Automaton**          Figure 3.156 depicts the automaton associated with the constraint DECREAS-
                       ING_HEIGHT_DECREASING_TERRACE.

Figure 3.156: Automaton for the DECREASING_HEIGHT_DECREASING_TERRACE con-
straint obtained by applying decoration Table 2.36 to the seed transducer of the DE-
CREASING_TERRACE pattern

## 3.79  DECREASING_HEIGHT_INCREASING_TERRACE

▶ ▷ ◁ ◀        **DESCRIPTION**          **AUTOMATON**

| **Origin** | Based on the INCREASING_TERRACE pattern. |
|---|---|

| **Constraint** | DECREASING_HEIGHT_INCREASING_TERRACE(VARIABLES) |
|---|---|

| **Argument** | VARIABLES : collection(var−dvar) |
|---|---|

| **Restriction** | required(VARIABLES, var) |
|---|---|

| **Purpose** | Succeeds if the minima of the values in each occurrence of the INCREASING_TERRACE pattern in the time-series given by the VARIABLES collection are decreasing.<br>An occurrence of the pattern INCREASING_TERRACE is the *maximal* subsequence which matches the regular expression $<=^+<$.<br>Assume that the occurrence of the pattern INCREASING_TERRACE starts at position $i$ and ends at position $j$. The feature MIN computes the minimum of the values from index $i+1$ to index $j$. |
|---|---|

| **Example** | $(\langle 2, 5, 5, 5, 6, 2, 3, 4, 4, 5, 3, 3, 1, 2, 2, 3\rangle)$ |
|---|---|

Figure 3.157 provides an example where the DECREASING_HEIGHT_INCREASING_TERRACE ([2, 5, 5, 5, 6, 2, 3, 4, 4, 5, 3, 3, 1, 2, 2, 3]) constraint holds.

| **Typical** | $\lvert$VARIABLES$\rvert > 3$<br>range(VARIABLES.var) $> 2$ |
|---|---|

Figure 3.157: Illustrating the DECREASING_HEIGHT_INCREASING_TERRACE constraint of the **Example** slot

**Automaton**         Figure 3.158 depicts the automaton associated with the constraint DECREAS-ING_HEIGHT_INCREASING_TERRACE.

Figure 3.158: Automaton for the DECREASING_HEIGHT_INCREASING_TERRACE constraint obtained by applying decoration Table 2.36 to the seed transducer of the IN-CREASING_TERRACE pattern

## 3.80  DECREASING_HEIGHT_PLAIN

▶ ▷ ◁ ◀          DESCRIPTION          AUTOMATON

| | |
|---|---|
| **Origin** | Based on the PLAIN pattern. |
| **Constraint** | DECREASING_HEIGHT_PLAIN(VARIABLES) |
| **Argument** | VARIABLES : collection(var−dvar) |
| **Restriction** | required(VARIABLES, var) |
| **Purpose** | Succeeds if the minima of the values in each occurrence of the PLAIN pattern in the time-series given by the VARIABLES collection are decreasing. An occurrence of the pattern PLAIN is the *maximal* subsequence which matches the regular expression >=*<. Assume that the occurrence of the pattern PLAIN starts at position $i$ and ends at position $j$. The feature MIN computes the minimum of the values from index $i + 1$ to index $j$. |
| **Example** | $(\langle 2, 3, 6, 5, 5, 7, 6, 6, 4, 5, 5, 4, 3, 6, 6, 3 \rangle)$ |

Figure 3.159 provides an example where the DECREASING_HEIGHT_PLAIN $([2, 3, 6, 5, 5, 7, 6, 6, 4, 5, 5, 4, 3, 6, 6, 3])$ constraint holds.

Figure 3.159: Illustrating the DECREASING_HEIGHT_PLAIN constraint of the **Example** slot

**Typical**

$$|\mathtt{VARIABLES}| > 2$$
$$\mathtt{range}(\mathtt{VARIABLES.var}) > 1$$

**Automaton**           Figure 3.160 depicts the automaton associated with the constraint DECREAS-
                        ING_HEIGHT_PLAIN.

Figure 3.160: Automaton for the DECREASING_HEIGHT_PLAIN constraint obtained by
applying decoration Table 2.36 to the seed transducer of the PLAIN pattern

## 3.81   DECREASING_HEIGHT_PLATEAU

▶ ▷ ◁ ◀          DESCRIPTION          AUTOMATON

| | |
|---|---|
| **Origin** | Based on the PLATEAU pattern. |
| **Constraint** | DECREASING_HEIGHT_PLATEAU(VARIABLES) |
| **Argument** | VARIABLES : collection(var−dvar) |
| **Restriction** | required(VARIABLES, var) |
| **Purpose** | Succeeds if the minima of the values in each occurrence of the PLATEAU pattern in the time-series given by the VARIABLES collection are decreasing. An occurrence of the pattern PLATEAU is the *maximal* subsequence which matches the regular expression $<=^*>$. Assume that the occurrence of the pattern PLATEAU starts at position $i$ and ends at position $j$. The feature MIN computes the minimum of the values from index $i+1$ to index $j$. |
| **Example** | $(\langle 5, 2, 2, 5, 5, 4, 3, 3, 4, 2, 2, 1, 3, 2, 5, 7 \rangle)$ |

Figure 3.161 provides an example where the DECREASING_HEIGHT_PLATEAU ($[5, 2, 2, 5, 5, 4, 3, 3, 4, 2, 2, 1, 3, 2, 5, 7]$) constraint holds.

| | |
|---|---|
| **Typical** | $\lvert$VARIABLES$\rvert > 2$ <br> range(VARIABLES.var) $> 1$ |

Figure 3.161: Illustrating the DECREASING_HEIGHT_PLATEAU constraint of the **Example** slot

**Automaton**          Figure 3.162 depicts the automaton associated with the constraint DECREAS-
ING_HEIGHT_PLATEAU.

Figure 3.162: Automaton for the DECREASING_HEIGHT_PLATEAU constraint obtained
by applying decoration Table 2.36 to the seed transducer of the PLATEAU pattern

## 3.82　DECREASING_HEIGHT_PROPER_PLAIN

▶ ▷ ◁ ◀　　　　**DESCRIPTION**　　　　**AUTOMATON**

**Origin**　　　Based on the PROPER_PLAIN pattern.

**Constraint**　　DECREASING_HEIGHT_PROPER_PLAIN(VARIABLES)

**Argument**　　VARIABLES : collection(var−dvar)

**Restriction**　　required(VARIABLES, var)

**Purpose**　　Succeeds if the minima of the values in each occurrence of the PROPER_PLAIN pattern in the time-series given by the VARIABLES collection are decreasing.

An occurrence of the pattern PROPER_PLAIN is the *maximal* subsequence which matches the regular expression $> =^+ <$.

Assume that the occurrence of the pattern PROPER_PLAIN starts at position $i$ and ends at position $j$. The feature MIN computes the minimum of the values from index $i + 1$ to index $j$.

**Example**　　$(\langle 2, 7, 5, 5, 5, 6, 3, 7, 4, 4, 5, 6, 5, 3, 3, 5\rangle)$

Figure 3.163 provides an example where the DECREASING_HEIGHT_PROPER_PLAIN $([2, 7, 5, 5, 5, 6, 3, 7, 4, 4, 5, 6, 5, 3, 3, 5])$ constraint holds.

**Typical**　　$|\texttt{VARIABLES}| > 3$
　　　　　　　range(VARIABLES.var) > 1

Figure 3.163: Illustrating the DECREASING_HEIGHT_PROPER_PLAIN constraint of the **Example** slot

**Automaton**          Figure 3.164 depicts the automaton associated with the constraint DECREAS-
ING_HEIGHT_PROPER_PLAIN.

Figure 3.164: Automaton for the DECREASING_HEIGHT_PROPER_PLAIN con-
straint obtained by applying decoration Table 2.36 to the seed transducer of the
PROPER_PLAIN pattern

## 3.83  DECREASING_HEIGHT_PROPER_PLATEAU

▶ ▷ ◁ ◀        DESCRIPTION        AUTOMATON

**Origin**          Based on the PROPER_PLATEAU pattern.

**Constraint**      DECREASING_HEIGHT_PROPER_PLATEAU(VARIABLES)

**Argument**        VARIABLES : collection(var−dvar)

**Restriction**     required(VARIABLES, var)

**Purpose**         Succeeds if the minima of the values in each occurrence of the PROPER_PLATEAU pattern
                    in the time-series given by the VARIABLES collection are decreasing.
                    An occurrence of the pattern PROPER_PLATEAU is the *maximal* subsequence which
                    matches the regular expression $< =^+ >$.
                    Assume that the occurrence of the pattern PROPER_PLATEAU starts at position $i$ and ends
                    at position $j$. The feature MIN computes the minimum of the values from index $i + 1$ to
                    index $j$.

**Example**         $(\langle 3, 5, 5, 5, 3, 2, 3, 4, 4, 1, 5, 2, 3, 3, 1, 7 \rangle)$

                    Figure 3.165 provides an example where the DECREASING_HEIGHT_PROPER_PLATEAU
                    ([3, 5, 5, 5, 3, 2, 3, 4, 4, 1, 5, 2, 3, 3, 1, 7]) constraint holds.

**Typical**         $|\texttt{VARIABLES}| > 3$
                    range(VARIABLES.var) $> 1$

Figure 3.165: Illustrating the DECREASING_HEIGHT_PROPER_PLATEAU constraint of the **Example** slot

**Automaton**            Figure 3.166 depicts the automaton associated with the constraint DECREAS-
                         ING_HEIGHT_PROPER_PLATEAU.

Figure 3.166: Automaton for the DECREASING_HEIGHT_PROPER_PLATEAU con-
straint obtained by applying decoration Table 2.36 to the seed transducer of the
PROPER_PLATEAU pattern

## 3.84 DECREASING_HEIGHT_STEADY

▶ ▷ ◁ ◀          **DESCRIPTION**          **AUTOMATON**

| | |
|---|---|
| **Origin** | Based on the STEADY pattern. |
| **Constraint** | DECREASING_HEIGHT_STEADY(VARIABLES) |
| **Argument** | VARIABLES : collection(var−dvar) |
| **Restriction** | required(VARIABLES, var) |
| **Purpose** | Succeeds if the minima of the values in each occurrence of the STEADY pattern in the time-series given by the VARIABLES collection are decreasing.<br><br>An occurrence of the pattern STEADY is the subsequence which matches the regular expression =.<br><br>Assume that the occurrence of the pattern STEADY starts at position $i$ and ends at position $j$. The feature MIN computes the minimum of the values from index $i$ to index $j + 1$. |
| **Example** | $(\langle 3, 6, 6, 6, 2, 7, 5, 5, 5, 6, 5, 5, 5, 3, 3, 7 \rangle)$ |

Figure 3.167 provides an example where the DECREASING_HEIGHT_STEADY ([3, 6, 6, 6, 2, 7, 5, 5, 5, 6, 5, 5, 5, 3, 3, 7]) constraint holds.

Figure 3.167: Illustrating the DECREASING_HEIGHT_STEADY constraint of the **Example** slot

Typical          $|VARIABLES| > 1$

**Automaton**     Figure 3.168 depicts the automaton associated with the constraint DECREAS-
                  ING_HEIGHT_STEADY.

Figure 3.168: Automaton for the DECREASING_HEIGHT_STEADY constraint obtained
by applying decoration Table 2.36 to the seed transducer of the STEADY pattern

## 3.85  DECREASING_HEIGHT_STEADY_SEQUENCE

► ▷ ◁ ◄          **DESCRIPTION**          **AUTOMATON**

| | |
|---|---|
| **Origin** | Based on the STEADY_SEQUENCE pattern. |
| **Constraint** | DECREASING_HEIGHT_STEADY_SEQUENCE(VARIABLES) |
| **Argument** | VARIABLES : collection(var−dvar) |
| **Restriction** | required(VARIABLES, var) |
| **Purpose** | Succeeds if the minima of the values in each occurrence of the STEADY_SEQUENCE pattern in the time-series given by the VARIABLES collection are decreasing. An occurrence of the pattern STEADY_SEQUENCE is the *maximal* subsequence which matches the regular expression $=^+$. Assume that the occurrence of the pattern STEADY_SEQUENCE starts at position $i$ and ends at position $j$. The feature MIN computes the minimum of the values from index $i$ to index $j + 1$. |
| **Example** | $(\langle 4, 3, 2, 5, 5, 5, 3, 2, 4, 4, 6, 5, 3, 2, 2, 6 \rangle)$ |

Figure 3.169 provides an example where the DECREASING_HEIGHT_STEADY_SEQUENCE $([4, 3, 2, 5, 5, 5, 3, 2, 4, 4, 6, 5, 3, 2, 2, 6])$ constraint holds.

Figure 3.169: Illustrating the DECREASING_HEIGHT_STEADY_SEQUENCE constraint of the **Example** slot

**Typical**                $|\texttt{VARIABLES}| > 1$

**Automaton**    Figure 3.170 depicts the automaton associated with the constraint DECREAS-
ING_HEIGHT_STEADY_SEQUENCE.

Figure 3.170: Automaton for the DECREASING_HEIGHT_STEADY_SEQUENCE con-
straint obtained by applying decoration Table 2.36 to the seed transducer of the
STEADY_SEQUENCE pattern

## 3.86   DECREASING_MAX_BUMP_ON_DECREASING_SEQUENCE

▶ ▷ ◁ ◀          **DESCRIPTION**                **AUTOMATON**

**Origin**          Based on the BUMP_ON_DECREASING_SEQUENCE pattern.

**Constraint**      DECREASING_MAX_BUMP_ON_DECREASING_SEQUENCE(VARIABLES)

**Argument**        VARIABLES : collection(var−dvar)

**Restriction**     required(VARIABLES, var)

**Purpose**         Succeeds if the maxima of the values in each occurrence of the BUMP_ON_DECREASING_SEQUENCE pattern in the time-series given by the VARIABLES collection are decreasing.

An occurrence of the pattern BUMP_ON_DECREASING_SEQUENCE is the subsequence which matches the regular expression >><>>.

Assume that the occurrence of the pattern BUMP_ON_DECREASING_SEQUENCE starts at position $i$ and ends at position $j$. The feature MAX computes the maximum of the values from index $i + 2$ to index $j$.

**Example**         $(\langle 7, 6, 5, 6, 5, 4, 1, 2, 7, 5, 4, 2, 5, 4, 3, 3 \rangle)$

Figure 3.171 provides an example where the DECREAS-ING_MAX_BUMP_ON_DECREASING_SEQUENCE $([7, 6, 5, 6, 5, 4, 1, 2, 7, 5, 4, 2, 5, 4, 3, 3])$ constraint holds.

**Typical**         $|\mathtt{VARIABLES}| > 5$
                    range(VARIABLES.var) > 2

Figure 3.171: Illustrating the DECREASING_MAX_BUMP_ON_DECREASING_SEQUENCE constraint of the **Example** slot

**Automaton**      Figure 3.172 depicts the automaton associated with the constraint DECREAS-ING_MAX_BUMP_ON_DECREASING_SEQUENCE.

Figure 3.172: Automaton for the DECREAS-ING_MAX_BUMP_ON_DECREASING_SEQUENCE constraint obtained by applying decoration Table 2.36 to the seed transducer of the BUMP_ON_DECREASING_SEQUENCE pattern

## 3.87 DECREASING_MAX_DECREASING

▶ ▷ ◁ ◀          **DESCRIPTION**          **AUTOMATON**

**Origin**          Based on the DECREASING pattern.

**Constraint**     DECREASING_MAX_DECREASING(VARIABLES)

**Argument**       VARIABLES : collection(var−dvar)

**Restriction**    required(VARIABLES, var)

**Purpose**        Succeeds if the maxima of the values in each occurrence of the DECREASING pattern in
                   the time-series given by the VARIABLES collection are decreasing.
                   An occurrence of the pattern DECREASING is the subsequence which matches the regular
                   expression >.
                   Assume that the occurrence of the pattern DECREASING starts at position $i$ and ends at
                   position $j$. The feature MAX computes the maximum of the values from index $i$ to index
                   $j + 1$.

**Example**        $(\langle 1, 1, 3, 4, 4, 5, 7, 7, 5, 4, 4, 2, 2, 2, 5, 5 \rangle)$

                   Figure 3.173 provides an example where the DECREASING_MAX_DECREASING
                   $([1, 1, 3, 4, 4, 5, 7, 7, 5, 4, 4, 2, 2, 2, 5, 5])$ constraint holds.

**Typical**        $|\texttt{VARIABLES}| > 1$
                   range(VARIABLES.var) > 1

Figure 3.173: Illustrating the DECREASING_MAX_DECREASING constraint of the **Example** slot

**Automaton**     Figure 3.174 depicts the automaton associated with the constraint DECREAS-ING_MAX_DECREASING.

$$\left\{ \begin{array}{l} C \leftarrow +\infty \\ D \leftarrow -\infty \\ F \leftarrow +\infty \\ R \leftarrow 1 \end{array} \right\}$$

$$\left\{ \begin{array}{l} C \leftarrow \max(\max(D, \mathtt{VAR}_i), \mathtt{VAR}_{i+1}) \\ D \leftarrow -\infty \\ F \leftarrow \max(\max(D, \mathtt{VAR}_i), \mathtt{VAR}_{i+1}) \\ R \leftarrow R \wedge (F \geq \max(\max(D, \mathtt{VAR}_i), \mathtt{VAR}_{i+1})) \end{array} \right\}$$

$R \wedge (F \geq C)$

Figure 3.174: Automaton for the DECREASING_MAX_DECREASING constraint obtained by applying decoration Table 2.36 to the seed transducer of the DECREASING pattern

## 3.88 DECREASING_MAX_DECREASING_SEQUENCE

▶ ▷ ◁ ◀        **DESCRIPTION**          **AUTOMATON**

| **Origin** | Based on the DECREASING_SEQUENCE pattern. |
|---|---|

| **Constraint** | DECREASING_MAX_DECREASING_SEQUENCE(VARIABLES) |
|---|---|

| **Argument** | VARIABLES : collection(var−dvar) |
|---|---|

| **Restriction** | required(VARIABLES, var) |
|---|---|

**Purpose**

Succeeds if the maxima of the values in each occurrence of the DECREAS-ING_SEQUENCE pattern in the time-series given by the VARIABLES collection are de-creasing.

An occurrence of the pattern DECREASING_SEQUENCE is the *maximal* subsequence which matches the regular expression $> (> | =)^* > | >$.

Assume that the occurrence of the pattern DECREASING_SEQUENCE starts at position $i$ and ends at position $j$. The feature MAX computes the maximum of the values from index $i$ to index $j + 1$.

**Example**

$(\langle 4, 7, 4, 2, 3, 4, 4, 6, 5, 5, 4, 3, 2, 2, 3, 2 \rangle)$

Figure 3.175 provides an example where the DECREAS-ING_MAX_DECREASING_SEQUENCE $([4, 7, 4, 2, 3, 4, 4, 6, 5, 5, 4, 3, 2, 2, 3, 2])$ constraint holds.

**Typical**

$|VARIABLES| > 1$
range(VARIABLES.var) $> 1$

Figure 3.175: Illustrating the DECREASING_MAX_DECREASING_SEQUENCE constraint of the **Example** slot

**Automaton**        Figure 3.176 depicts the automaton associated with the constraint DECREAS-
                     ING_MAX_DECREASING_SEQUENCE.

Figure 3.176: Automaton for the DECREASING_MAX_DECREASING_SEQUENCE con-
straint obtained by applying decoration Table 2.36 to the seed transducer of the DE-
CREASING_SEQUENCE pattern

## 3.89   DECREASING_MAX_DIP_ON_INCREASING_SEQUENCE

▶ ▷ ◁ ◀         **DESCRIPTION**          **AUTOMATON**

**Origin**          Based on the DIP_ON_INCREASING_SEQUENCE pattern.

**Constraint**      DECREASING_MAX_DIP_ON_INCREASING_SEQUENCE(VARIABLES)

**Argument**        VARIABLES : collection(var−dvar)

**Restriction**     required(VARIABLES, var)

**Purpose**         Succeeds if the maxima of the values in each occurrence of the DIP_ON_INCREASING_SEQUENCE pattern in the time-series given by the VARIABLES collection are decreasing.

An occurrence of the pattern DIP_ON_INCREASING_SEQUENCE is the subsequence which matches the regular expression $<<><<$.

Assume that the occurrence of the pattern DIP_ON_INCREASING_SEQUENCE starts at position $i$ and ends at position $j$. The feature MAX computes the maximum of the values from index $i + 2$ to index $j$.

**Example**         $(\langle 4, 5, 6, 0, 2, 4, 4, 1, 2, 3, 2, 3, 4, 6, 5, 1 \rangle)$

Figure 3.177 provides an example where the DECREAS-ING_MAX_DIP_ON_INCREASING_SEQUENCE $([4, 5, 6, 0, 2, 4, 4, 1, 2, 3, 2, 3, 4, 6, 5, 1])$ constraint holds.

**Typical**         $|\text{VARIABLES}| > 5$
                    range(VARIABLES.var) $> 2$

Figure 3.177: Illustrating the DECREASING_MAX_DIP_ON_INCREASING_SEQUENCE constraint of the **Example** slot

**Automaton**            Figure 3.178 depicts the automaton associated with the constraint DECREAS-
ING_MAX_DIP_ON_INCREASING_SEQUENCE.

Figure 3.178: Automaton for the DECREAS-
ING_MAX_DIP_ON_INCREASING_SEQUENCE constraint obtained by applying
decoration Table 2.36 to the seed transducer of the DIP_ON_INCREASING_SEQUENCE
pattern

## 3.90 DECREASING_MAX_INCREASING

▶ ▷ ◁ ◀          DESCRIPTION          AUTOMATON

| | |
|---|---|
| **Origin** | Based on the INCREASING pattern. |
| **Constraint** | DECREASING_MAX_INCREASING(VARIABLES) |
| **Argument** | VARIABLES : collection(var−dvar) |
| **Restriction** | required(VARIABLES, var) |
| **Purpose** | Succeeds if the maxima of the values in each occurrence of the INCREASING pattern in the time-series given by the VARIABLES collection are decreasing. |
| | An occurrence of the pattern INCREASING is the subsequence which matches the regular expression <. |
| | Assume that the occurrence of the pattern INCREASING starts at position $i$ and ends at position $j$. The feature MAX computes the maximum of the values from index $i$ to index $j + 1$. |
| **Example** | $(\langle 4, 4, 5, 5, 4, 3, 3, 1, 4, 2, 4, 3, 3, 1, 2, 1\rangle)$ |

Figure 3.179 provides an example where the DECREASING_MAX_INCREASING $([4, 4, 5, 5, 4, 3, 3, 1, 4, 2, 4, 3, 3, 1, 2, 1])$ constraint holds.

Figure 3.179: Illustrating the DECREASING_MAX_INCREASING constraint of the **Example** slot

**Typical**          $|\mathtt{VARIABLES}| > 1$
$\mathtt{range}(\mathtt{VARIABLES.var}) > 1$

**Automaton**   Figure 3.180 depicts the automaton associated with the constraint DECREAS-
ING_MAX_INCREASING.

Figure 3.180: Automaton for the DECREASING_MAX_INCREASING constraint obtained
by applying decoration Table 2.36 to the seed transducer of the INCREASING pattern

## 3.91 DECREASING_MAX_INCREASING_SEQUENCE

► ▷ ◁ ◄            **DESCRIPTION**          **AUTOMATON**

| | |
|---|---|
| **Origin** | Based on the INCREASING_SEQUENCE pattern. |
| **Constraint** | DECREASING_MAX_INCREASING_SEQUENCE(VARIABLES) |
| **Argument** | VARIABLES : collection(var−dvar) |
| **Restriction** | required(VARIABLES, var) |

**Purpose**

> Succeeds if the maxima of the values in each occurrence of the INCREASING_SEQUENCE pattern in the time-series given by the VARIABLES collection are decreasing.
> An occurrence of the pattern INCREASING_SEQUENCE is the *maximal* subsequence which matches the regular expression $< (< | =)^* < | <$.
> Assume that the occurrence of the pattern INCREASING_SEQUENCE starts at position $i$ and ends at position $j$. The feature MAX computes the maximum of the values from index $i$ to index $j + 1$.

**Example**

> $(\langle 3, 4, 5, 5, 4, 3, 3, 1, 2, 2, 4, 3, 3, 1, 3, 1 \rangle)$

Figure 3.181 provides an example where the DECREASING_MAX_INCREASING_SEQUENCE $([3, 4, 5, 5, 4, 3, 3, 1, 2, 2, 4, 3, 3, 1, 3, 1])$ constraint holds.

Figure 3.181: Illustrating the DECREASING_MAX_INCREASING_SEQUENCE constraint of the **Example** slot

**Typical**      $|\mathtt{VARIABLES}| > 1$
$\mathtt{range}(\mathtt{VARIABLES.var}) > 1$

**Automaton**       Figure 3.182 depicts the automaton associated with the constraint DECREAS-
ING_MAX_INCREASING_SEQUENCE.

Figure 3.182: Automaton for the DECREASING_MAX_INCREASING_SEQUENCE con-
straint obtained by applying decoration Table 2.36 to the seed transducer of the IN-
CREASING_SEQUENCE pattern

## 3.92 DECREASING_MAX_INFLEXION

▶ ▷ ◁ ◀        **DESCRIPTION**        **AUTOMATON**

| | |
|---|---|
| **Origin** | Based on the INFLEXION pattern. |
| **Constraint** | DECREASING_MAX_INFLEXION(VARIABLES) |
| **Argument** | VARIABLES : collection(var−dvar) |
| **Restriction** | required(VARIABLES, var) |
| **Purpose** | Succeeds if the maxima of the values in each occurrence of the INFLEXION pattern in the time-series given by the VARIABLES collection are decreasing. <br><br> An occurrence of the pattern INFLEXION is the *maximal* subsequence which matches the regular expression $< (< | =)^* > | > (> | =)^* <$. <br><br> Assume that the occurrence of the pattern INFLEXION starts at position $i$ and ends at position $j$. The feature MAX computes the maximum of the values from index $i + 1$ to index $j$. |
| **Example** | $(\langle 3, 4, 3, 2, 2, 1, 1, 2, 2, 5, 5 \rangle)$ |

Figure 3.183 provides an example where the DECREASING_MAX_INFLEXION $([3, 4, 3, 2, 2, 1, 1, 2, 2, 5, 5])$ constraint holds.

Figure 3.183: Illustrating the DECREASING_MAX_INFLEXION constraint of the **Example** slot

**Typical**

$|\mathtt{VARIABLES}| > 2$

$\mathtt{range}(\mathtt{VARIABLES.var}) > 1$

**Automaton**        Figure 3.184 depicts the automaton associated with the constraint DECREAS-
                     ING_MAX_INFLEXION.

Figure 3.184: Automaton for the DECREASING_MAX_INFLEXION constraint obtained
by applying decoration Table 2.36 to the seed transducer of the INFLEXION pattern
(transition $r \rightarrow t$ has the same accumulators updates as transition $t \rightarrow r$)

## 3.93 DECREASING_MAX_PEAK

► ▷ ◁ ◀          **DESCRIPTION**          **AUTOMATON**

| | |
|---|---|
| **Origin** | Based on the PEAK pattern. |
| **Constraint** | DECREASING_MAX_PEAK(VARIABLES) |
| **Argument** | VARIABLES : collection(var−dvar) |
| **Restriction** | required(VARIABLES, var) |
| **Purpose** | Succeeds if the maxima of the values in each occurrence of the PEAK pattern in the time-series given by the VARIABLES collection are decreasing.<br><br>An occurrence of the pattern PEAK is the *maximal* subsequence which matches the regular expression $< (= | <)^* (> | =)^* >$.<br><br>Assume that the occurrence of the pattern PEAK starts at position $i$ and ends at position $j$. The feature MAX computes the maximum of the values from index $i + 1$ to index $j$. |
| **Example** | $(\langle 1, 6, 6, 6, 2, 5, 4, 3, 2, 2, 3, 2, 1, 5, 5, 7 \rangle)$ |

Figure 3.185 provides an example where the DECREASING_MAX_PEAK
$([1, 6, 6, 6, 2, 5, 4, 3, 2, 2, 3, 2, 1, 5, 5, 7])$ constraint holds.

Figure 3.185: Illustrating the DECREASING_MAX_PEAK constraint of the **Example** slot

**Typical**

$|\mathtt{VARIABLES}| > 2$
$\mathtt{range}(\mathtt{VARIABLES.var}) > 1$

**Automaton**     Figure 3.186 depicts the automaton associated with the constraint DECREAS-
ING_MAX_PEAK.

Figure 3.186: Automaton for the DECREASING_MAX_PEAK constraint obtained by ap-
plying decoration Table 2.36 to the seed transducer of the PEAK pattern

## 3.94   DECREASING_MAX_STRICTLY_DECREASING_SEQUENCE

▶ ▷ ◁ ◀      **DESCRIPTION**      **AUTOMATON**

**Origin**      Based on the STRICTLY_DECREASING_SEQUENCE pattern.

**Constraint**      DECREASING_MAX_STRICTLY_DECREASING_SEQUENCE(VARIABLES)

**Argument**      VARIABLES : collection(var−dvar)

**Restriction**      required(VARIABLES, var)

**Purpose**      Succeeds if the maxima of the values in each occurrence of the STRICTLY_DECREASING_SEQUENCE pattern in the time-series given by the VARIABLES collection are decreasing.

An occurrence of the pattern STRICTLY_DECREASING_SEQUENCE is the *maximal* subsequence which matches the regular expression $>^+$.

Assume that the occurrence of the pattern STRICTLY_DECREASING_SEQUENCE starts at position $i$ and ends at position $j$. The feature MAX computes the maximum of the values from index $i$ to index $j + 1$.

**Example**      $(\langle 6, 4, 6, 5, 5, 3, 5, 5, 4, 4, 4, 3, 2, 2, 3, 4 \rangle)$

Figure 3.187 provides an example where the DECREASING_MAX_STRICTLY_DECREASING_SEQUENCE $([6, 4, 6, 5, 5, 3, 5, 5, 4, 4, 4, 3, 2, 2, 3, 4])$ constraint holds.

**Typical**      $|\text{VARIABLES}| > 1$
                   range(VARIABLES.var) > 1

Figure 3.187: Illustrating the DECREASING_MAX_STRICTLY_DECREASING_SEQUENCE constraint of the **Example** slot

**Automaton**          Figure 3.188 depicts the automaton associated with the constraint DECREAS-ING_MAX_STRICTLY_DECREASING_SEQUENCE.

Figure 3.188: Automaton for the DECREAS-ING_MAX_STRICTLY_DECREASING_SEQUENCE constraint obtained by applying decoration Table 2.36 to the seed transducer of the STRICTLY_DECREASING_SEQUENCE pattern

## 3.95 DECREASING_MAX_STRICTLY_INCREASING_SEQUENCE

► ▷ ◁ ◀    **DESCRIPTION**        **AUTOMATON**

| | |
|---|---|
| **Origin** | Based on the STRICTLY_INCREASING_SEQUENCE pattern. |
| **Constraint** | DECREASING_MAX_STRICTLY_INCREASING_SEQUENCE(VARIABLES) |
| **Argument** | VARIABLES : collection(var−dvar) |
| **Restriction** | required(VARIABLES, var) |

**Purpose**

Succeeds if the maxima of the values in each occurrence of the STRICTLY_INCREASING_SEQUENCE pattern in the time-series given by the VARIABLES collection are decreasing.

An occurrence of the pattern STRICTLY_INCREASING_SEQUENCE is the *maximal* subsequence which matches the regular expression $<^+$.

Assume that the occurrence of the pattern STRICTLY_INCREASING_SEQUENCE starts at position $i$ and ends at position $j$. The feature MAX computes the maximum of the values from index $i$ to index $j + 1$.

**Example**

$(\langle 3, 4, 5, 5, 4, 3, 3, 1, 2, 3, 4, 3, 3, 1, 3, 1 \rangle)$

Figure 3.189 provides an example where the DECREASING_MAX_STRICTLY_INCREASING_SEQUENCE $([3, 4, 5, 5, 4, 3, 3, 1, 2, 3, 4, 3, 3, 1, 3, 1])$ constraint holds.

**Typical**

$|\text{VARIABLES}| > 1$
$\text{range}(\text{VARIABLES.var}) > 1$

Figure 3.189: Illustrating the DECREASING_MAX_STRICTLY_INCREASING_SEQUENCE constraint of the **Example** slot

**Automaton**            Figure 3.190 depicts the automaton associated with the constraint DECREAS-
                         ING_MAX_STRICTLY_INCREASING_SEQUENCE.

Figure       3.190:                 Automaton       for       the       DECREAS-
ING_MAX_STRICTLY_INCREASING_SEQUENCE constraint obtained by applying dec-
oration Table 2.36 to the seed transducer of the STRICTLY_INCREASING_SEQUENCE
pattern

## 3.96 DECREASING_MAX_SUMMIT

► ▷ ◁ ◀    **DESCRIPTION**    **AUTOMATON**

$(< | < (= | <)^* <)(> | > (= | >)^* >)$

| **Origin** | Based on the SUMMIT pattern. |
|---|---|

| **Constraint** | DECREASING_MAX_SUMMIT(VARIABLES) |
|---|---|

| **Argument** | VARIABLES : collection(var−dvar) |
|---|---|

| **Restriction** | required(VARIABLES, var) |
|---|---|

| **Purpose** | Succeeds if the maxima of the values in each occurrence of the SUMMIT pattern in the time-series given by the VARIABLES collection are decreasing. An occurrence of the pattern SUMMIT is the *maximal* subsequence which matches the regular expression $(< | < (= | <)^* <)(> | > (= | >)^* >)$. Assume that the occurrence of the pattern SUMMIT starts at position $i$ and ends at position $j$. The feature MAX computes the maximum of the values from index $i + 1$ to index $j$. |
|---|---|

| **Example** | $(\langle 1, 4, 4, 5, 4, 2, 6, 6, 2, 3, 5, 4, 1, 2, 4, 2 \rangle)$ |
|---|---|

Figure 3.191 provides an example where the DECREASING_MAX_SUMMIT $([1, 4, 4, 5, 4, 2, 6, 6, 2, 3, 5, 4, 1, 2, 4, 2])$ constraint holds.

| **Typical** | $|\text{VARIABLES}| > 2$ |
|---|---|
| | range(VARIABLES.var) $> 1$ |

Figure 3.191: Illustrating the DECREASING_MAX_SUMMIT constraint of the **Example** slot

**Automaton**    Figure 3.192 depicts the automaton associated with the constraint DECREAS-
ING_MAX_SUMMIT.

Figure 3.192: Automaton for the DECREASING_MAX_SUMMIT constraint obtained by
applying decoration Table 2.36 to the seed transducer of the SUMMIT pattern (transition
$u \rightarrow r$ has the same accumulator update as transition $r \rightarrow u$)

## 3.97 DECREASING_MAX_ZIGZAG

▶ ▷ ◁ ◀          **DESCRIPTION**          **AUTOMATON**

$(<>)^+(<\,|\,<>)\,|\,(><)^+(>\,|\,><)$

| **Origin** | Based on the ZIGZAG pattern. |
| **Constraint** | DECREASING_MAX_ZIGZAG(VARIABLES) |
| **Argument** | VARIABLES : collection(var−dvar) |
| **Restriction** | required(VARIABLES, var) |
| **Purpose** | Succeeds if the maxima of the values in each occurrence of the ZIGZAG pattern in the time-series given by the VARIABLES collection are decreasing. |
| | An occurrence of the pattern ZIGZAG is the *maximal* subsequence which matches the regular expression $(<>)^+(<\,|\,<>)\,|\,(><)^+(>\,|\,><)$. |
| | Assume that the occurrence of the pattern ZIGZAG starts at position $i$ and ends at position $j$. The feature MAX computes the maximum of the values from index $i+1$ to index $j$. |
| **Example** | $(\langle 6, 5, 7, 4, 5, 1, 1, 6, 3, 6, 4, 3, 5, 2, 7, 7\rangle)$ |

Figure 3.193 provides an example where the DECREASING_MAX_ZIGZAG $([6, 5, 7, 4, 5, 1, 1, 6, 3, 6, 4, 3, 5, 2, 7, 7])$ constraint holds.

| **Typical** | $|\texttt{VARIABLES}| > 3$ |
| | range(VARIABLES.var) $> 1$ |

Figure 3.193: Illustrating the DECREASING_MAX_ZIGZAG constraint of the **Example** slot

**Automaton**          Figure 3.194 depicts the automaton associated with the constraint DECREAS-
                       ING_MAX_ZIGZAG.

Figure 3.194: Automaton for the DECREASING_MAX_ZIGZAG constraint obtained by applying decoration Table 2.36 to the seed transducer of the ZIGZAG pattern; (1) missing transitions from $a, b, c, d, e, f$ to $s$ are labelled by $=$; (2) on transitions from $b, c, e, f$ to $s$ the accumulator $D$ is reset to its initial value; (3) on transitions from $c, f$ to $s$ the accumulator $F$ is reset to $C$, and the accumulator $R$ is updated wrt $C$ and $F$

## 3.98 DECREASING_MIN_BUMP_ON_DECREASING_SEQUENCE

▶ ▷ ◁ ◀  **DESCRIPTION**   **AUTOMATON**

| | |
|---|---|
| **Origin** | Based on the BUMP_ON_DECREASING_SEQUENCE pattern. |
| **Constraint** | DECREASING_MIN_BUMP_ON_DECREASING_SEQUENCE(VARIABLES) |
| **Argument** | VARIABLES : collection(var−dvar) |
| **Restriction** | required(VARIABLES, var) |

**Purpose**

Succeeds if the minima of the values in each occurrence of the BUMP_ON_DECREASING_SEQUENCE pattern in the time-series given by the VARIABLES collection are decreasing.

An occurrence of the pattern BUMP_ON_DECREASING_SEQUENCE is the subsequence which matches the regular expression '>><>>'.

Assume that the occurrence of the pattern BUMP_ON_DECREASING_SEQUENCE starts at position $i$ and ends at position $j$. The feature MIN computes the minimum of the values from index $i + 2$ to index $j$.

**Example**

$(\langle 7, 6, 5, 6, 4, 1, 1, 5, 5, 4, 2, 6, 5, 4, 4, 5\rangle)$

Figure 3.195 provides an example where the DECREASING_MIN_BUMP_ON_DECREASING_SEQUENCE ([7, 6, 5, 6, 4, 1, 1, 5, 5, 4, 2, 6, 5, 4, 4, 5]) constraint holds.

**Typical**

$|\mathtt{VARIABLES}| > 5$

range(VARIABLES.var) > 2

Figure 3.195: Illustrating the DECREASING_MIN_BUMP_ON_DECREASING_SEQUENCE constraint of the **Example** slot

**Automaton**     Figure 3.196 depicts the automaton associated with the constraint DECREAS-
ING_MIN_BUMP_ON_DECREASING_SEQUENCE.

Figure 3.196: Automaton for the DECREAS-
ING_MIN_BUMP_ON_DECREASING_SEQUENCE constraint obtained by applying deco-
ration Table 2.36 to the seed transducer of the BUMP_ON_DECREASING_SEQUENCE
pattern

## 3.99  DECREASING_MIN_DECREASING

        **DESCRIPTION**                  **AUTOMATON**

| | |
|---|---|
| **Origin** | Based on the DECREASING pattern. |
| **Constraint** | DECREASING_MIN_DECREASING(VARIABLES) |
| **Argument** | VARIABLES : collection(var−dvar) |
| **Restriction** | required(VARIABLES, var) |

**Purpose**

Succeeds if the minima of the values in each occurrence of the DECREASING pattern in the time-series given by the VARIABLES collection are decreasing.

An occurrence of the pattern DECREASING is the subsequence which matches the regular expression >.

Assume that the occurrence of the pattern DECREASING starts at position $i$ and ends at position $j$. The feature MIN computes the minimum of the values from index $i$ to index $j + 1$.

**Example**

$$(\langle 4, 7, 6, 4, 5, 5, 6, 4, 4, 3, 3, 2, 2, 3, 1\rangle)$$

Figure 3.197 provides an example where the DECREASING_MIN_DECREASING ([4, 7, 6, 4, 5, 5, 6, 4, 4, 3, 3, 2, 2, 3, 1]) constraint holds.

**Typical**

$|\texttt{VARIABLES}| > 1$

$\texttt{range}(\texttt{VARIABLES.var}) > 1$

Figure 3.197: Illustrating the DECREASING_MIN_DECREASING constraint of the **Example** slot

**Automaton**     Figure 3.198 depicts the automaton associated with the constraint DECREAS-
ING_MIN_DECREASING.

Figure 3.198: Automaton for the DECREASING_MIN_DECREASING constraint obtained
by applying decoration Table 2.36 to the seed transducer of the DECREASING pattern

## 3.100 DECREASING_MIN_DECREASING_SEQUENCE

▶ ▷ ◁ ◀          **DESCRIPTION**          **AUTOMATON**

**Origin**            Based on the DECREASING_SEQUENCE pattern.

**Constraint**        DECREASING_MIN_DECREASING_SEQUENCE(VARIABLES)

**Argument**          VARIABLES : collection(var−dvar)

**Restriction**       required(VARIABLES, var)

**Purpose**           Succeeds if the minima of the values in each occurrence of the DECREASING_SEQUENCE
                      pattern in the time-series given by the VARIABLES collection are decreasing.
                      An occurrence of the pattern DECREASING_SEQUENCE is the *maximal* subsequence
                      which matches the regular expression $> (> | =)^* > | >$.
                      Assume that the occurrence of the pattern DECREASING_SEQUENCE starts at position $i$
                      and ends at position $j$. The feature MIN computes the minimum of the values from index
                      $i$ to index $j + 1$.

**Example**           $(\langle 4, 7, 6, 4, 5, 5, 5, 6, 5, 5, 4, 4, 2, 2, 3, 1 \rangle)$

                      Figure 3.199 provides an example where the DECREASING_MIN_DECREASING_SEQUENCE
                      ([4, 7, 6, 4, 5, 5, 5, 6, 5, 5, 4, 4, 2, 2, 3, 1]) constraint holds.

**Typical**           $|\text{VARIABLES}| > 1$
                      range(VARIABLES.var) > 1

Figure 3.199: Illustrating the DECREASING_MIN_DECREASING_SEQUENCE constraint of the **Example** slot

**Automaton**  Figure 3.200 depicts the automaton associated with the constraint DECREAS-
ING_MIN_DECREASING_SEQUENCE.

Figure 3.200: Automaton for the DECREASING_MIN_DECREASING_SEQUENCE con-
straint obtained by applying decoration Table 2.36 to the seed transducer of the DE-
CREASING_SEQUENCE pattern

# 3.101   DECREASING_MIN_DIP_ON_INCREASING_SEQUENCE

▶ ▷ ◁ ◀          **DESCRIPTION**          **AUTOMATON**

**Origin**          Based on the DIP_ON_INCREASING_SEQUENCE pattern.

**Constraint**      DECREASING_MIN_DIP_ON_INCREASING_SEQUENCE(VARIABLES)

**Argument**        VARIABLES : collection(var−dvar)

**Restriction**     required(VARIABLES, var)

**Purpose**         Succeeds if the minima of the values in each occurrence of the DIP_ON_INCREASING_SEQUENCE pattern in the time-series given by the VARIABLES collection are decreasing.

An occurrence of the pattern DIP_ON_INCREASING_SEQUENCE is the subsequence which matches the regular expression `<<><<`.

Assume that the occurrence of the pattern DIP_ON_INCREASING_SEQUENCE starts at position $i$ and ends at position $j$. The feature MIN computes the minimum of the values from index $i + 2$ to index $j$.

**Example**         $(\langle 4, 5, 6, 1, 2, 4, 4, 1, 2, 3, 0, 3, 4, 6, 5, 1 \rangle)$

Figure 3.201 provides an example where the DECREAS-ING_MIN_DIP_ON_INCREASING_SEQUENCE $([4, 5, 6, 1, 2, 4, 4, 1, 2, 3, 0, 3, 4, 6, 5, 1])$ constraint holds.

**Typical**         $|\text{VARIABLES}| > 5$
                    range(VARIABLES.var) > 2

Figure 3.201: Illustrating the DECREASING_MIN_DIP_ON_INCREASING_SEQUENCE constraint of the **Example** slot

**Automaton**     Figure 3.202 depicts the automaton associated with the constraint DECREAS-
ING_MIN_DIP_ON_INCREASING_SEQUENCE.

Figure 3.202:     Automaton     for     the     DECREAS-
ING_MIN_DIP_ON_INCREASING_SEQUENCE constraint obtained by applying dec-
oration Table 2.36 to the seed transducer of the DIP_ON_INCREASING_SEQUENCE
pattern

## 3.102  DECREASING_MIN_GORGE

▶ ▷ ◁ ◀          **DESCRIPTION**          **AUTOMATON**

$(> | > (= | >)^*) (< | < (= | <)^* <)$

**Origin**          Based on the GORGE pattern.

**Constraint**          DECREASING_MIN_GORGE(VARIABLES)

**Argument**          VARIABLES : collection(var−dvar)

**Restriction**          required(VARIABLES, var)

**Purpose**          Succeeds if the minima of the values in each occurrence of the GORGE pattern in the time-series given by the VARIABLES collection are decreasing.

An occurrence of the pattern GORGE is the *maximal* subsequence which matches the regular expression $(> | > (= | >)^*) (< | < (= | <)^* <)$.

Assume that the occurrence of the pattern GORGE starts at position $i$ and ends at position $j$. The feature MIN computes the minimum of the values from index $i + 1$ to index $j$.

**Example**          $(\langle 6, 2, 3, 4, 5, 6, 1, 1, 5, 4, 2, 3, 6, 1, 4, 5 \rangle)$

Figure 3.203 provides an example where the DECREASING_MIN_GORGE ($[6, 2, 3, 4, 5, 6, 1, 1, 5, 4, 2, 3, 6, 1, 4, 5]$) constraint holds.

Figure 3.203: Illustrating the DECREASING_MIN_GORGE constraint of the **Example** slot

**Typical**

$|\texttt{VARIABLES}| > 2$
$\texttt{range}(\texttt{VARIABLES.var}) > 1$

**Automaton**  Figure 3.204 depicts the automaton associated with the constraint DECREAS-ING_MIN_GORGE.

Figure 3.204: Automaton for the DECREASING_MIN_GORGE constraint obtained by applying decoration Table 2.36 to the seed transducer of the GORGE pattern (transition $u \rightarrow r$ has the same accumulator update as transition $r \rightarrow u$)

## 3.103 DECREASING_MIN_INCREASING

▶ ▷ ◁ ◀          DESCRIPTION          AUTOMATON

| | |
|---|---|
| **Origin** | Based on the INCREASING pattern. |
| **Constraint** | DECREASING_MIN_INCREASING(VARIABLES) |
| **Argument** | VARIABLES : collection(var−dvar) |
| **Restriction** | required(VARIABLES, var) |
| **Purpose** | Succeeds if the minima of the values in each occurrence of the INCREASING pattern in the time-series given by the VARIABLES collection are decreasing. An occurrence of the pattern INCREASING is the subsequence which matches the regular expression <. Assume that the occurrence of the pattern INCREASING starts at position $i$ and ends at position $j$. The feature MIN computes the minimum of the values from index $i$ to index $j+1$. |
| **Example** | $(\langle 4, 4, 5, 5, 4, 3, 3, 2, 4, 3, 3, 2, 2, 1, 3, 1 \rangle)$ |

Figure 3.205 provides an example where the DECREASING_MIN_INCREASING $([4, 4, 5, 5, 4, 3, 3, 2, 4, 3, 3, 2, 2, 1, 3, 1])$ constraint holds.

Figure 3.205: Illustrating the DECREASING_MIN_INCREASING constraint of the **Example** slot

**Typical**

$|\mathtt{VARIABLES}| > 1$
$\mathtt{range}(\mathtt{VARIABLES.var}) > 1$

**Automaton**   Figure 3.206 depicts the automaton associated with the constraint DECREAS-
ING_MIN_INCREASING.

Figure 3.206: Automaton for the DECREASING_MIN_INCREASING constraint obtained
by applying decoration Table 2.36 to the seed transducer of the INCREASING pattern

## 3.104 DECREASING_MIN_INCREASING_SEQUENCE

▶ ▷ ◁ ◀          **DESCRIPTION**          **AUTOMATON**

**Origin**          Based on the INCREASING_SEQUENCE pattern.

**Constraint**      DECREASING_MIN_INCREASING_SEQUENCE(VARIABLES)

**Argument**        VARIABLES : collection(var−dvar)

**Restriction**     required(VARIABLES, var)

**Purpose**
Succeeds if the minima of the values in each occurrence of the INCREASING_SEQUENCE pattern in the time-series given by the VARIABLES collection are decreasing.

An occurrence of the pattern INCREASING_SEQUENCE is the *maximal* subsequence which matches the regular expression $< (< | =)^* < | <$.

Assume that the occurrence of the pattern INCREASING_SEQUENCE starts at position $i$ and ends at position $j$. The feature MIN computes the minimum of the values from index $i$ to index $j + 1$.

**Example**
$(\langle 3, 4, 5, 5, 4, 3, 3, 1, 4, 3, 3, 2, 2, 1, 3, 1 \rangle)$

Figure 3.207 provides an example where the DECREASING_MIN_INCREASING_SEQUENCE $([3, 4, 5, 5, 4, 3, 3, 1, 4, 3, 3, 2, 2, 1, 3, 1])$ constraint holds.

Figure 3.207: Illustrating the DECREASING_MIN_INCREASING_SEQUENCE constraint of the **Example** slot

**Typical**        $|\mathtt{VARIABLES}| > 1$
$\mathtt{range}(\mathtt{VARIABLES.var}) > 1$

**Automaton**    Figure 3.208 depicts the automaton associated with the constraint DECREAS-
ING_MIN_INCREASING_SEQUENCE.

Figure 3.208: Automaton for the DECREASING_MIN_INCREASING_SEQUENCE con-
straint obtained by applying decoration Table 2.36 to the seed transducer of the IN-
CREASING_SEQUENCE pattern

## 3.105  DECREASING_MIN_INFLEXION

▶ ▷ ◁ ◀              **DESCRIPTION**        **AUTOMATON**

$< (< | =)^* > | > (> | =)^* <$

| | |
|---|---|
| **Origin** | Based on the INFLEXION pattern. |
| **Constraint** | DECREASING_MIN_INFLEXION(VARIABLES) |
| **Argument** | VARIABLES : collection(var−dvar) |
| **Restriction** | required(VARIABLES, var) |
| **Purpose** | Succeeds if the minima of the values in each occurrence of the INFLEXION pattern in the time-series given by the VARIABLES collection are decreasing.<br><br>An occurrence of the pattern INFLEXION is the *maximal* subsequence which matches the regular expression $< (< | =)^* > | > (> | =)^* <$.<br><br>Assume that the occurrence of the pattern INFLEXION starts at position $i$ and ends at position $j$. The feature MIN computes the minimum of the values from index $i + 1$ to index $j$. |
| **Example** | $(\langle 3, 4, 3, 2, 2, 1, 1, 2, 2, 5, 5 \rangle)$ |

Figure 3.209 provides an example where the DECREASING_MIN_INFLEXION ($[3, 4, 3, 2, 2, 1, 1, 2, 2, 5, 5]$) constraint holds.

Figure 3.209: Illustrating the DECREASING_MIN_INFLEXION constraint of the **Example** slot

**Typical**

$|\text{VARIABLES}| > 2$
$\text{range}(\text{VARIABLES.var}) > 1$

    Figure 3.210 depicts the automaton associated with the constraint DECREAS-
ING_MIN_INFLEXION.

Figure 3.210: Automaton for the DECREASING_MIN_INFLEXION constraint obtained
by applying decoration Table 2.36 to the seed transducer of the INFLEXION pattern
(transition $r \to t$ has the same accumulators updates as transition $t \to r$)

# 3.106 DECREASING_MIN_STRICTLY_DECREASING_SEQUENCE

▶ ▷ ◁ ◀      **DESCRIPTION**         **AUTOMATON**

**Origin**          Based on the STRICTLY_DECREASING_SEQUENCE pattern.

**Constraint**      DECREASING_MIN_STRICTLY_DECREASING_SEQUENCE(VARIABLES)

**Argument**       VARIABLES : collection(var−dvar)

**Restriction**     required(VARIABLES, var)

**Purpose**

Succeeds if the minima of the values in each occurrence of the STRICTLY_DECREASING_SEQUENCE pattern in the time-series given by the VARIABLES collection are decreasing.

An occurrence of the pattern STRICTLY_DECREASING_SEQUENCE is the *maximal* sub-sequence which matches the regular expression $>^{+}$.

Assume that the occurrence of the pattern STRICTLY_DECREASING_SEQUENCE starts at position $i$ and ends at position $j$. The feature MIN computes the minimum of the values from index $i$ to index $j + 1$.

**Example**       $(\langle 6, 5, 6, 5, 5, 4, 4, 3, 6, 2, 4, 3, 2, 2, 3, 4 \rangle)$

Figure 3.211 provides an example where the DECREAS-ING_MIN_STRICTLY_DECREASING_SEQUENCE $([6, 5, 6, 5, 5, 4, 4, 3, 6, 2, 4, 3, 2, 2, 3, 4])$ constraint holds.

**Typical**       |VARIABLES| > 1
               range(VARIABLES.var) > 1

Figure 3.211: Illustrating the DECREASING_MIN_STRICTLY_DECREASING_SEQUENCE constraint of the **Example** slot

**Automaton** Figure 3.212 depicts the automaton associated with the constraint DECREAS-ING_MIN_STRICTLY_DECREASING_SEQUENCE.

Figure 3.212: Automaton for the DECREAS-ING_MIN_STRICTLY_DECREASING_SEQUENCE constraint obtained by applying decoration Table 2.36 to the seed transducer of the STRICTLY_DECREASING_SEQUENCE pattern

# 3.107 DECREASING_MIN_STRICTLY_INCREASING_SEQUENCE

▶ ▷ ◁ ◀          **DESCRIPTION**          **AUTOMATON**

**Origin**           Based on the STRICTLY_INCREASING_SEQUENCE pattern.

**Constraint**       DECREASING_MIN_STRICTLY_INCREASING_SEQUENCE(VARIABLES)

**Argument**         VARIABLES : collection(var−dvar)

**Restriction**      required(VARIABLES, var)

**Purpose**          Succeeds if the minima of the values in each occurrence of the STRICTLY_INCREASING_SEQUENCE pattern in the time-series given by the VARIABLES collection are decreasing.
An occurrence of the pattern STRICTLY_INCREASING_SEQUENCE is the *maximal* subsequence which matches the regular expression $<^+$.
Assume that the occurrence of the pattern STRICTLY_INCREASING_SEQUENCE starts at position $i$ and ends at position $j$. The feature MIN computes the minimum of the values from index $i$ to index $j + 1$.

**Example**          $(\langle 3, 4, 5, 5, 4, 3, 3, 1, 4, 3, 3, 2, 2, 1, 3, 1 \rangle)$

Figure 3.213 provides an example where the DECREASING_MIN_STRICTLY_INCREASING_SEQUENCE $([3, 4, 5, 5, 4, 3, 3, 1, 4, 3, 3, 2, 2, 1, 3, 1])$ constraint holds.

**Typical**          $|\text{VARIABLES}| > 1$
range(VARIABLES.var) > 1

Figure 3.213: Illustrating the DECREASING_MIN_STRICTLY_INCREASING_SEQUENCE constraint of the **Example** slot

**Automaton**    Figure 3.214 depicts the automaton associated with the constraint DECREAS-ING_MIN_STRICTLY_INCREASING_SEQUENCE.

Figure 3.214: Automaton for the DECREAS-ING_MIN_STRICTLY_INCREASING_SEQUENCE constraint obtained by applying dec-oration Table 2.36 to the seed transducer of the STRICTLY_INCREASING_SEQUENCE pattern

## 3.108   DECREASING_MIN_VALLEY

► ▷ ◁ ◀        **DESCRIPTION**              **AUTOMATON**

| **Origin** | Based on the VALLEY pattern. |

| **Constraint** | DECREASING_MIN_VALLEY(VARIABLES) |

| **Argument** | VARIABLES : collection(var−dvar) |

| **Restriction** | required(VARIABLES, var) |

**Purpose**

Succeeds if the minima of the values in each occurrence of the VALLEY pattern in the time-series given by the VARIABLES collection are decreasing.

An occurrence of the pattern VALLEY is the *maximal* subsequence which matches the regular expression $> (= | >)^* (< | =)^* <$.

Assume that the occurrence of the pattern VALLEY starts at position $i$ and ends at position $j$. The feature MIN computes the minimum of the values from index $i + 1$ to index $j$.

**Example**

$(\langle 1, 3, 3, 7, 6, 5, 4, 6, 6, 5, 4, 3, 6, 2, 2, 7 \rangle)$

Figure 3.215 provides an example where the DECREASING_MIN_VALLEY $([1, 3, 3, 7, 6, 5, 4, 6, 6, 5, 4, 3, 6, 2, 2, 7])$ constraint holds.

**Typical**

$|\texttt{VARIABLES}| > 2$
range(VARIABLES.var) $> 1$

Figure 3.215: Illustrating the DECREASING_MIN_VALLEY constraint of the **Example** slot

**Automaton**     Figure 3.216 depicts the automaton associated with the constraint DECREAS-
ING_MIN_VALLEY.

Figure 3.216: Automaton for the DECREASING_MIN_VALLEY constraint obtained by
applying decoration Table 2.36 to the seed transducer of the VALLEY pattern

## 3.109  DECREASING_MIN_ZIGZAG

▶ ▷ ◁ ◀          **DESCRIPTION**          **AUTOMATON**

$(<>)^+(<\,|\,<>)\,|\,(><)^+(>\,|\,><)$

| | |
|---|---|
| **Origin** | Based on the ZIGZAG pattern. |
| **Constraint** | DECREASING_MIN_ZIGZAG(VARIABLES) |
| **Argument** | VARIABLES : collection(var−dvar) |
| **Restriction** | required(VARIABLES, var) |
| **Purpose** | Succeeds if the minima of the values in each occurrence of the ZIGZAG pattern in the time-series given by the VARIABLES collection are decreasing. An occurrence of the pattern ZIGZAG is the *maximal* subsequence which matches the regular expression $(<>)^+(<\,|\,<>)\,|\,(><)^+(>\,|\,><)$. Assume that the occurrence of the pattern ZIGZAG starts at position $i$ and ends at position $j$. The feature MIN computes the minimum of the values from index $i+1$ to index $j$. |
| **Example** | $(\langle 6, 5, 7, 4, 5, 1, 1, 6, 3, 6, 4, 3, 5, 2, 7, 7 \rangle)$ |

Figure 3.217 provides an example where the DECREASING_MIN_ZIGZAG $([6, 5, 7, 4, 5, 1, 1, 6, 3, 6, 4, 3, 5, 2, 7, 7])$ constraint holds.

| | |
|---|---|
| **Typical** | $|\texttt{VARIABLES}| > 3$ <br> range(VARIABLES.var) $> 1$ |

Figure 3.217: Illustrating the DECREASING_MIN_ZIGZAG constraint of the **Example** slot

**Automaton**         Figure 3.218 depicts the automaton associated with the constraint DECREAS-
ING_MIN_ZIGZAG.

Figure 3.218: Automaton for the DECREASING_MIN_ZIGZAG constraint obtained by applying decoration Table 2.36 to the seed transducer of the ZIGZAG pattern; (1) missing transitions from $a, b, c, d, e, f$ to $s$ are labelled by $=$; (2) on transitions from $b, c, e, f$ to $s$ the accumulator $D$ is reset to its initial value; (3) on transitions from $c, f$ to $s$ the accumulator $F$ is reset to $C$, and the accumulator $R$ is updated wrt $C$ and $F$

# 3.110  DECREASING_RANGE_DECREASING

▶ ▷ ◁ ◀        **DESCRIPTION**            **AUTOMATON**

| | |
|---|---|
| **Origin** | Based on the DECREASING pattern. |
| **Constraint** | DECREASING_RANGE_DECREASING(VARIABLES) |
| **Argument** | VARIABLES : collection(var−dvar) |
| **Restriction** | required(VARIABLES, var) |
| **Purpose** | Succeeds if the differences between the largest and smallest value in each occurrence of the DECREASING pattern in the time-series given by the VARIABLES collection are decreasing.<br>An occurrence of the pattern DECREASING is the subsequence which matches the regular expression >.<br>Assume that the occurrence of the pattern DECREASING starts at position $i$ and ends at position $j$. The feature RANGE computes the range of the values from index $i$ to index $j + 1$. |
| **Example** | $(\langle 4, 8, 6, 4, 5, 5, 5, 8, 6, 6, 5, 5, 4, 4, 6, 5 \rangle)$ |

Figure 3.219 provides an example where the DECREASING_RANGE_DECREASING ($[4, 8, 6, 4, 5, 5, 5, 8, 6, 6, 5, 5, 4, 4, 6, 5]$) constraint holds.

| | |
|---|---|
| **Typical** | $\lvert \texttt{VARIABLES} \rvert > 1$<br>range(VARIABLES.var) > 1 |

Figure 3.219: Illustrating the DECREASING_RANGE_DECREASING constraint of the **Example** slot

**Automaton**    Figure 3.220 depicts the automaton associated with the constraint DECREAS-
ING_RANGE_DECREASING.

$$
\left\{
\begin{array}{l}
C \leftarrow +\infty \\
F \leftarrow +\infty \\
H \leftarrow \mathtt{VAR}_1 \\
R \leftarrow 1
\end{array}
\right\}
$$

$$
\left\{
\begin{array}{l}
C \leftarrow |H - \mathtt{VAR}_{i+1}| \\
F \leftarrow |H - \mathtt{VAR}_{i+1}| \\
H \leftarrow \mathtt{VAR}_{i+1} \\
R \leftarrow R \wedge (F \geq |H - \mathtt{VAR}_{i+1}|)
\end{array}
\right\}
$$

$$\{H \leftarrow \mathtt{VAR}_{i+1}\}$$

$$\boxed{R \wedge (F \geq C)}$$

Figure 3.220: Automaton for the DECREASING_RANGE_DECREASING constraint ob-
tained by applying decoration Table 2.47 to the seed transducer of the DECREASING
pattern

# 3.111    DECREASING_RANGE_DECREASING_SEQUENCE

▶ ▷ ◁ ◀    **DESCRIPTION**    **AUTOMATON**

**Origin**    Based on the DECREASING_SEQUENCE pattern.

**Constraint**    DECREASING_RANGE_DECREASING_SEQUENCE(VARIABLES)

**Argument**    VARIABLES : collection(var−dvar)

**Restriction**    required(VARIABLES, var)

**Purpose**    Succeeds if the differences between the largest and smallest value in each occurrence of the DECREASING_SEQUENCE pattern in the time-series given by the VARIABLES collection are decreasing.

An occurrence of the pattern DECREASING_SEQUENCE is the *maximal* subsequence which matches the regular expression $> (> | =)^* > | >$.

Assume that the occurrence of the pattern DECREASING_SEQUENCE starts at position $i$ and ends at position $j$. The feature RANGE computes the range of the values from index $i$ to index $j + 1$.

**Example**    $(\langle 4, 7, 6, 1, 5, 5, 5, 6, 5, 5, 4, 4, 2, 2, 3, 1 \rangle)$

Figure 3.221 provides an example where the DECREASING_RANGE_DECREASING_SEQUENCE $([4, 7, 6, 1, 5, 5, 5, 6, 5, 5, 4, 4, 2, 2, 3, 1])$ constraint holds.

**Typical**    $|\text{VARIABLES}| > 1$
range(VARIABLES.var) > 1

Figure 3.221: Illustrating the DECREASING_RANGE_DECREASING_SEQUENCE constraint of the **Example** slot

**Automaton**    Figure 3.222 depicts the automaton associated with the constraint DECREAS-
ING_RANGE_DECREASING_SEQUENCE.

Figure 3.222: Automaton for the DECREASING_RANGE_DECREASING_SEQUENCE
constraint obtained by applying decoration Table 2.47 to the seed transducer of the
DECREASING_SEQUENCE pattern

## 3.112  DECREASING_RANGE_INCREASING

▶ ▷ ◁ ◀          **DESCRIPTION**          **AUTOMATON**

| | |
|---|---|
| **Origin** | Based on the INCREASING pattern. |
| **Constraint** | DECREASING_RANGE_INCREASING(VARIABLES) |
| **Argument** | VARIABLES : collection(var−dvar) |
| **Restriction** | required(VARIABLES, var) |

**Purpose**

Succeeds if the differences between the largest and smallest value in each occurrence of the INCREASING pattern in the time-series given by the VARIABLES collection are decreasing.

An occurrence of the pattern INCREASING is the subsequence which matches the regular expression $<$.

Assume that the occurrence of the pattern INCREASING starts at position $i$ and ends at position $j$. The feature RANGE computes the range of the values from index $i$ to index $j+1$.

**Example**

$$(\langle 4, 6, 8, 4, 4, 5, 5, 6, 6, 7, 5, 5, 5, 5, 6, 4 \rangle)$$

Figure 3.223 provides an example where the DECREASING_RANGE_INCREASING ($[4, 6, 8, 4, 4, 5, 5, 6, 6, 7, 5, 5, 5, 5, 6, 4]$) constraint holds.

**Typical**

$|\text{VARIABLES}| > 1$
$\text{range}(\text{VARIABLES.var}) > 1$

Figure 3.223: Illustrating the DECREASING_RANGE_INCREASING constraint of the **Example** slot

**Automaton**        Figure 3.224 depicts the automaton associated with the constraint DECREAS-
ING_RANGE_INCREASING.

Figure 3.224: Automaton for the DECREASING_RANGE_INCREASING constraint ob-
tained by applying decoration Table 2.47 to the seed transducer of the INCREASING
pattern

# 3.113 DECREASING_RANGE_INCREASING_SEQUENCE

▶ ▷ ◁ ◀         **DESCRIPTION**         **AUTOMATON**

**Origin**          Based on the INCREASING_SEQUENCE pattern.

**Constraint**      DECREASING_RANGE_INCREASING_SEQUENCE(VARIABLES)

**Argument**        VARIABLES : collection(var−dvar)

**Restriction**     required(VARIABLES, var)

**Purpose**         Succeeds if the differences between the largest and smallest value in each occurrence of
                    the INCREASING_SEQUENCE pattern in the time-series given by the VARIABLES collec-
                    tion are decreasing.
                    An occurrence of the pattern INCREASING_SEQUENCE is the *maximal* subsequence
                    which matches the regular expression $< (< | =)^* < | <$.
                    Assume that the occurrence of the pattern INCREASING_SEQUENCE starts at position $i$
                    and ends at position $j$. The feature RANGE computes the range of the values from index
                    $i$ to index $j + 1$.

**Example**         $(\langle 1, 4, 5, 5, 4, 3, 3, 1, 4, 3, 3, 2, 2, 1, 3, 1 \rangle)$

                    Figure 3.225 provides an example where the DECREAS-
                    ING_RANGE_INCREASING_SEQUENCE $([1, 4, 5, 5, 4, 3, 3, 1, 4, 3, 3, 2, 2, 1, 3, 1])$ con-
                    straint holds.

**Typical**         $|\text{VARIABLES}| > 1$
                    $\text{range}(\text{VARIABLES.var}) > 1$

Figure 3.225: Illustrating the DECREASING_RANGE_INCREASING_SEQUENCE constraint of the **Example** slot

**Automaton**            Figure 3.226 depicts the automaton associated with the constraint DECREAS-
ING_RANGE_INCREASING_SEQUENCE.

$$\left\{ \begin{array}{l} C \leftarrow +\infty \\ F \leftarrow +\infty \\ H \leftarrow \text{VAR}_1 \\ R \leftarrow 1 \end{array} \right\}$$

Figure 3.226: Automaton for the DECREASING_RANGE_INCREASING_SEQUENCE
constraint obtained by applying decoration Table 2.47 to the seed transducer of the
INCREASING_SEQUENCE pattern

# 3.114 DECREASING_RANGE_STRICTLY_DECREASING_SEQUENCE

▶ ▷ ◁ ◀    **DESCRIPTION**         **AUTOMATON**

| | |
|---|---|
| **Origin** | Based on the STRICTLY_DECREASING_SEQUENCE pattern. |
| **Constraint** | DECREASING_RANGE_STRICTLY_DECREASING_SEQUENCE(VARIABLES) |
| **Argument** | VARIABLES : collection(var−dvar) |
| **Restriction** | required(VARIABLES, var) |

**Purpose**

Succeeds if the differences between the largest and smallest value in each occurrence of the STRICTLY_DECREASING_SEQUENCE pattern in the time-series given by the VARIABLES collection are decreasing.

An occurrence of the pattern STRICTLY_DECREASING_SEQUENCE is the *maximal* subsequence which matches the regular expression $>^+$.

Assume that the occurrence of the pattern STRICTLY_DECREASING_SEQUENCE starts at position $i$ and ends at position $j$. The feature RANGE computes the range of the values from index $i$ to index $j + 1$.

**Example**    $(\langle 6, 3, 3, 5, 4, 2, 7, 6, 5, 6, 4, 4, 3, 3, 2, 6 \rangle)$

Figure 3.227 provides an example where the DECREASING_RANGE_STRICTLY_DECREASING_SEQUENCE ([6, 3, 3, 5, 4, 2, 7, 6, 5, 6, 4, 4, 3, 3, 2, 6]) constraint holds.

**Typical**    $|\text{VARIABLES}| > 1$
               $\text{range}(\text{VARIABLES.var}) > 1$

Figure 3.227: Illustrating the DECREASING_RANGE_STRICTLY_DECREASING_SEQUENCE constraint of the **Example** slot

**Automaton**      Figure 3.228 depicts the automaton associated with the constraint DECREAS-
ING_RANGE_STRICTLY_DECREASING_SEQUENCE.

Figure 3.228: Automaton for the DECREAS-
ING_RANGE_STRICTLY_DECREASING_SEQUENCE constraint obtained
by applying decoration Table 2.47 to the seed transducer of the
STRICTLY_DECREASING_SEQUENCE pattern

# 3.115 DECREASING_RANGE_STRICTLY_INCREASING_SEQUENCE

▶ ▷ ◁ ◀          **DESCRIPTION**          **AUTOMATON**

| | |
|---|---|
| **Origin** | Based on the STRICTLY_INCREASING_SEQUENCE pattern. |
| **Constraint** | DECREASING_RANGE_STRICTLY_INCREASING_SEQUENCE(VARIABLES) |
| **Argument** | VARIABLES : collection(var−dvar) |
| **Restriction** | required(VARIABLES, var) |

**Purpose**

Succeeds if the differences between the largest and smallest value in each occurrence of the STRICTLY_INCREASING_SEQUENCE pattern in the time-series given by the VARIABLES collection are decreasing.

An occurrence of the pattern STRICTLY_INCREASING_SEQUENCE is the *maximal* subsequence which matches the regular expression $<^+$.

Assume that the occurrence of the pattern STRICTLY_INCREASING_SEQUENCE starts at position $i$ and ends at position $j$. The feature RANGE computes the range of the values from index $i$ to index $j + 1$.

**Example**

$$(\langle 1, 4, 5, 5, 4, 1, 0, 2, 3, 3, 6, 2, 3, 4, 4, 6 \rangle)$$

Figure 3.229 provides an example where the DECREASING_RANGE_STRICTLY_INCREASING_SEQUENCE ([1, 4, 5, 5, 4, 1, 0, 2, 3, 3, 6, 2, 3, 4, 4, 6]) constraint holds.

**Typical**

$|\text{VARIABLES}| > 1$

$\text{range}(\text{VARIABLES.var}) > 1$

Figure 3.229: Illustrating the DECREASING_RANGE_STRICTLY_INCREASING_SEQUENCE constraint of the **Example** slot

**Automaton**    Figure 3.230 depicts the automaton associated with the constraint DECREAS-ING_RANGE_STRICTLY_INCREASING_SEQUENCE.

Figure 3.230: Automaton for the DECREAS-ING_RANGE_STRICTLY_INCREASING_SEQUENCE constraint obtained by applying decoration Table 2.47 to the seed transducer of the STRICTLY_INCREASING_SEQUENCE pattern

# 3.116 DECREASING_SURF_BUMP_ON_DECREASING_SEQUENCE

► ▷ ◁ ◀      **DESCRIPTION**          **AUTOMATON**

**Origin**           Based on the BUMP_ON_DECREASING_SEQUENCE pattern.

**Constraint**       DECREASING_SURF_BUMP_ON_DECREASING_SEQUENCE(VARIABLES)

**Argument**         VARIABLES : collection(var−dvar)

**Restriction**      required(VARIABLES, var)

**Purpose**          Succeeds if the values denoting the surface of each occurrence of the
                     BUMP_ON_DECREASING_SEQUENCE pattern in the time-series given by the VARIABLES
                     collection are decreasing.
                     An occurrence of the pattern BUMP_ON_DECREASING_SEQUENCE is the subsequence
                     which matches the regular expression `>><>>`.
                     Assume that the occurrence of the pattern BUMP_ON_DECREASING_SEQUENCE starts at
                     position $i$ and ends at position $j$. The feature SURF computes the sum of the values from
                     index $i + 2$ to index $j$.

**Example**          $(\langle 7, 6, 5, 6, 5, 4, 1, 4, 7, 5, 4, 2, 5, 4, 3, 3 \rangle)$

                     Figure 3.231 provides an example where the DECREAS-
                     ING_SURF_BUMP_ON_DECREASING_SEQUENCE $([7, 6, 5, 6, 5, 4, 1, 4, 7, 5, 4, 2, 5, 4, 3, 3])$
                     constraint holds.

**Typical**          $|\text{VARIABLES}| > 5$
                     range(VARIABLES.var) > 2

Figure 3.231: Illustrating the DECREASING_SURF_BUMP_ON_DECREASING_SEQUENCE constraint of the **Example** slot

**Automaton**        Figure 3.232 depicts the automaton associated with the constraint DECREAS-
ING_SURF_BUMP_ON_DECREASING_SEQUENCE.

Figure 3.232: Automaton for the DECREAS-
ING_SURF_BUMP_ON_DECREASING_SEQUENCE constraint obtained by applying dec-
oration Table 2.36 to the seed transducer of the BUMP_ON_DECREASING_SEQUENCE
pattern

## 3.117 DECREASING_SURF_DECREASING

▶ ▷ ◁ ◀          DESCRIPTION          AUTOMATON

| | |
|---|---|
| **Origin** | Based on the DECREASING pattern. |
| **Constraint** | DECREASING_SURF_DECREASING(VARIABLES) |
| **Argument** | VARIABLES : collection(var−dvar) |
| **Restriction** | required(VARIABLES, var) |
| **Purpose** | Succeeds if the values denoting the surface of each occurrence of the DECREASING pattern in the time-series given by the VARIABLES collection are decreasing. An occurrence of the pattern DECREASING is the subsequence which matches the regular expression >. Assume that the occurrence of the pattern DECREASING starts at position $i$ and ends at position $j$. The feature SURF computes the sum of the values from index $i$ to index $j+1$. |
| **Example** | $(\langle 2, 6, 5, 2, 3, 3, 3, 4, 3, 3, 2, 4, 4, 5, 7, 7 \rangle)$ |

Figure 3.233 provides an example where the DECREASING_SURF_DECREASING ($[2, 6, 5, 2, 3, 3, 3, 4, 3, 3, 2, 4, 4, 5, 7, 7]$) constraint holds.

Figure 3.233: Illustrating the DECREASING_SURF_DECREASING constraint of the **Example** slot

**Typical**

$|\mathtt{VARIABLES}| > 1$
$\mathtt{range}(\mathtt{VARIABLES.var}) > 1$

**Automaton**        Figure 3.234 depicts the automaton associated with the constraint DECREAS-
                     ING_SURF_DECREASING.

$$
\left\{
\begin{array}{l}
C \leftarrow +\infty \\
D \leftarrow 0 \\
F \leftarrow +\infty \\
R \leftarrow 1
\end{array}
\right\}
$$

$$
\left\{
\begin{array}{l}
C \leftarrow D + \mathrm{VAR}_i + \mathrm{VAR}_{i+1} \\
D \leftarrow 0 \\
F \leftarrow D + \mathrm{VAR}_i + \mathrm{VAR}_{i+1} \\
R \leftarrow R \wedge (F \geq D + \mathrm{VAR}_i + \mathrm{VAR}_{i+1})
\end{array}
\right\}
$$

$$ > $$

$$ \geq s \quad \leq $$

$$ \boxed{R \wedge (F \geq C)} $$

Figure 3.234: Automaton for the DECREASING_SURF_DECREASING constraint ob-
tained by applying decoration Table 2.36 to the seed transducer of the DECREASING
pattern

# 3.118 DECREASING_SURF_DECREASING_SEQUENCE

▶ ▷ ◁ ◀          **DESCRIPTION**          **AUTOMATON**

| | |
|---|---|
| **Origin** | Based on the DECREASING_SEQUENCE pattern. |
| **Constraint** | DECREASING_SURF_DECREASING_SEQUENCE(VARIABLES) |
| **Argument** | VARIABLES : collection(var−dvar) |
| **Restriction** | required(VARIABLES, var) |

**Purpose**

Succeeds if the values denoting the surface of each occurrence of the DECREAS-ING_SEQUENCE pattern in the time-series given by the VARIABLES collection are decreasing.

An occurrence of the pattern DECREASING_SEQUENCE is the *maximal* subsequence which matches the regular expression $> (> | =)^* > | >$.

Assume that the occurrence of the pattern DECREASING_SEQUENCE starts at position $i$ and ends at position $j$. The feature SURF computes the sum of the values from index $i$ to index $j + 1$.

**Example**

$(\langle 2, 6, 5, 2, 3, 3, 3, 4, 3, 3, 2, 4, 4, 5, 7, 5 \rangle)$

Figure 3.235 provides an example where the DECREAS-ING_SURF_DECREASING_SEQUENCE ([2, 6, 5, 2, 3, 3, 3, 4, 3, 3, 2, 4, 4, 5, 7, 5]) constraint holds.

**Typical**

$|\text{VARIABLES}| > 1$
range(VARIABLES.var) > 1

Figure 3.235: Illustrating the DECREASING_SURF_DECREASING_SEQUENCE constraint of the **Example** slot

**Automaton**          Figure 3.236 depicts the automaton associated with the constraint DECREAS-
                       ING_SURF_DECREASING_SEQUENCE.

$$\left\{ \begin{array}{l} C \leftarrow +\infty \\ D \leftarrow 0 \\ F \leftarrow +\infty \\ R \leftarrow 1 \end{array} \right\}$$

$$\downarrow$$

$$\boxed{\leq s} \qquad \leq$$

$$\left\{ \begin{array}{l} D \leftarrow 0 \\ F \leftarrow C \\ R \leftarrow R \wedge (F \geq C') \end{array} \right\} \quad \boxed{R \wedge (F \geq t)} \quad \left\{ \begin{array}{l} \overset{>}{C \leftarrow D + \text{VAR}_i + \text{VAR}_{i+1}} \\ D \leftarrow 0 \end{array} \right\}$$

$$\overset{=}{\{D \leftarrow D + \text{VAR}_{i+1}\}} \quad \boxed{\geq t} \quad \left\{ \begin{array}{l} \overset{>}{C \leftarrow C + D + \text{VAR}_{i+1}} \\ D \leftarrow 0 \end{array} \right\}$$

Figure 3.236: Automaton for the DECREASING_SURF_DECREASING_SEQUENCE con-
straint obtained by applying decoration Table 2.36 to the seed transducer of the DE-
CREASING_SEQUENCE pattern

# 3.119 DECREASING_SURF_DECREASING_TERRACE

▶ ▷ ◁ ◀        **DESCRIPTION**              **AUTOMATON**

**Origin**        Based on the DECREASING_TERRACE pattern.

**Constraint**    DECREASING_SURF_DECREASING_TERRACE(VARIABLES)

**Argument**      VARIABLES : collection(var−dvar)

**Restriction**   required(VARIABLES, var)

**Purpose**

Succeeds if the values denoting the surface of each occurrence of the DECREAS-ING_TERRACE pattern in the time-series given by the VARIABLES collection are decreasing.

An occurrence of the pattern DECREASING_TERRACE is the *maximal* subsequence which matches the regular expression $>=^+>$.

Assume that the occurrence of the pattern DECREASING_TERRACE starts at position $i$ and ends at position $j$. The feature SURF computes the sum of the values from index $i + 1$ to index $j$.

**Example**       $(\langle 7, 6, 6, 6, 4, 3, 3, 2, 2, 4, 4, 6, 3, 3, 1, 1 \rangle)$

Figure 3.237 provides an example where the DECREASING_SURF_DECREASING_TERRACE ([7, 6, 6, 6, 4, 3, 3, 2, 2, 4, 4, 6, 3, 3, 1, 1]) constraint holds.

**Typical**       $|\text{VARIABLES}| > 3$
                  range(VARIABLES.var) > 2

Figure 3.237: Illustrating the DECREASING_SURF_DECREASING_TERRACE constraint of the **Example** slot

**Automaton**    Figure 3.238 depicts the automaton associated with the constraint DECREAS-
ING_SURF_DECREASING_TERRACE.

Figure 3.238: Automaton for the DECREASING_SURF_DECREASING_TERRACE con-
straint obtained by applying decoration Table 2.36 to the seed transducer of the DE-
CREASING_TERRACE pattern

## 3.120 DECREASING_SURF_DIP_ON_INCREASING_SEQUENCE

      **DESCRIPTION**      **AUTOMATON**

| | |
|---|---|
| **Origin** | Based on the DIP_ON_INCREASING_SEQUENCE pattern. |
| **Constraint** | DECREASING_SURF_DIP_ON_INCREASING_SEQUENCE(VARIABLES) |
| **Argument** | VARIABLES : collection(var−dvar) |
| **Restriction** | required(VARIABLES, var) |

**Purpose**

Succeeds if the values denoting the surface of each occurrence of the DIP_ON_INCREASING_SEQUENCE pattern in the time-series given by the VARIABLES collection are decreasing.

An occurrence of the pattern DIP_ON_INCREASING_SEQUENCE is the subsequence which matches the regular expression `<<><<`.

Assume that the occurrence of the pattern DIP_ON_INCREASING_SEQUENCE starts at position $i$ and ends at position $j$. The feature SURF computes the sum of the values from index $i + 2$ to index $j$.

**Example**

$(\langle 4, 5, 6, 0, 3, 4, 4, 1, 2, 3, 2, 3, 4, 6, 5, 1 \rangle)$

Figure 3.239 provides an example where the DECREASING_SURF_DIP_ON_INCREASING_SEQUENCE $([5, 6, 0, 3, 4, 4, 1, 2, 3, 2, 3, 4, 6, 5, 1])$ constraint holds.

**Typical**

$|\text{VARIABLES}| > 5$

$\text{range}(\text{VARIABLES.var}) > 2$

Figure 3.239: Illustrating the DECREASING_SURF_DIP_ON_INCREASING_SEQUENCE constraint of the **Example** slot

**Automaton**          Figure 3.240 depicts the automaton associated with the constraint DECREAS-
ING_SURF_DIP_ON_INCREASING_SEQUENCE.

Figure 3.240: Automaton for the DECREAS-
ING_SURF_DIP_ON_INCREASING_SEQUENCE constraint obtained by applying
decoration Table 2.36 to the seed transducer of the DIP_ON_INCREASING_SEQUENCE
pattern

## 3.121 DECREASING_SURF_GORGE

► ▷ ◁ ◀            DESCRIPTION            AUTOMATON

$(> | > (= | >)^*>)(< | < (= | <)^* <)$

| | |
|---|---|
| **Origin** | Based on the GORGE pattern. |
| **Constraint** | DECREASING_SURF_GORGE(VARIABLES) |
| **Argument** | VARIABLES : collection(var−dvar) |
| **Restriction** | required(VARIABLES, var) |
| **Purpose** | Succeeds if the values denoting the surface of each occurrence of the GORGE pattern in the time-series given by the VARIABLES collection are decreasing.<br><br>An occurrence of the pattern GORGE is the *maximal* subsequence which matches the regular expression $(> | > (= | >)^*>)(< | < (= | <)^* <)$.<br><br>Assume that the occurrence of the pattern GORGE starts at position $i$ and ends at position $j$. The feature SURF computes the sum of the values from index $i + 1$ to index $j$. |
| **Example** | $(\langle 6, 2, 3, 4, 5, 6, 1, 1, 5, 4, 2, 3, 6, 1, 4, 5 \rangle)$ |

Figure 3.241 provides an example where the DECREASING_SURF_GORGE
$([6, 2, 3, 4, 5, 6, 1, 1, 5, 4, 2, 3, 6, 1, 4, 5])$ constraint holds.

Figure 3.241: Illustrating the DECREASING_SURF_GORGE constraint of the **Example** slot

**Typical**
$|\mathtt{VARIABLES}| > 2$
$\mathtt{range}(\mathtt{VARIABLES.var}) > 1$

**Automaton**       Figure 3.242 depicts the automaton associated with the constraint DECREAS-
                    ING_SURF_GORGE.

Figure 3.242: Automaton for the DECREASING_SURF_GORGE constraint obtained by
applying decoration Table 2.36 to the seed transducer of the GORGE pattern (transition
$u \rightarrow r$ has the same accumulator update as transition $r \rightarrow u$)

## 3.122 DECREASING_SURF_INCREASING

| | |
|---|---|
| **Origin** | Based on the INCREASING pattern. |
| **Constraint** | DECREASING_SURF_INCREASING(VARIABLES) |
| **Argument** | VARIABLES : collection(var−dvar) |
| **Restriction** | required(VARIABLES, var) |
| **Purpose** | Succeeds if the values denoting the surface of each occurrence of the INCREASING pattern in the time-series given by the VARIABLES collection are decreasing.<br><br>An occurrence of the pattern INCREASING is the subsequence which matches the regular expression <.<br><br>Assume that the occurrence of the pattern INCREASING starts at position $i$ and ends at position $j$. The feature SURF computes the sum of the values from index $i$ to index $j+1$. |
| **Example** | $(\langle 4, 4, 5, 5, 4, 3, 3, 1, 4, 3, 3, 2, 2, 1, 3, 1 \rangle)$ |

Figure 3.243 provides an example where the DECREASING_SURF_INCREASING ($[4, 4, 5, 5, 4, 3, 3, 1, 4, 3, 3, 2, 2, 1, 3, 1]$) constraint holds.

Figure 3.243: Illustrating the DECREASING_SURF_INCREASING constraint of the **Example** slot

| | |
|---|---|
| **Typical** | $\lvert$VARIABLES$\rvert > 1$<br>range(VARIABLES.var) > 1 |

**Automaton** Figure 3.244 depicts the automaton associated with the constraint DECREAS-ING_SURF_INCREASING.

Figure 3.244: Automaton for the DECREASING_SURF_INCREASING constraint obtained by applying decoration Table 2.36 to the seed transducer of the INCREASING pattern

# 3.123 DECREASING_SURF_INCREASING_SEQUENCE

▶ ▷ ◁ ◀     **DESCRIPTION**     **AUTOMATON**

<div align="right">

$< (< | =)^* < |$
$<$

</div>

| | |
|---|---|
| **Origin** | Based on the INCREASING_SEQUENCE pattern. |
| **Constraint** | DECREASING_SURF_INCREASING_SEQUENCE(VARIABLES) |
| **Argument** | VARIABLES : collection(var−dvar) |
| **Restriction** | required(VARIABLES, var) |
| **Purpose** | Succeeds if the values denoting the surface of each occurrence of the INCREASING_SEQUENCE pattern in the time-series given by the VARIABLES collection are decreasing.<br><br>An occurrence of the pattern INCREASING_SEQUENCE is the *maximal* subsequence which matches the regular expression $< (< | =)^* < | <$.<br><br>Assume that the occurrence of the pattern INCREASING_SEQUENCE starts at position $i$ and ends at position $j$. The feature SURF computes the sum of the values from index $i$ to index $j + 1$. |
| **Example** | $(\langle 3, 4, 5, 5, 4, 3, 3, 1, 4, 3, 3, 2, 2, 1, 3, 1 \rangle)$ |

Figure 3.245 provides an example where the DECREASING_SURF_INCREASING_SEQUENCE $([3, 4, 5, 5, 4, 3, 3, 1, 4, 3, 3, 2, 2, 1, 3, 1])$ constraint holds.

| | |
|---|---|
| **Typical** | $|\text{VARIABLES}| > 1$<br>range(VARIABLES.var) $> 1$ |

Figure 3.245: Illustrating the DECREASING_SURF_INCREASING_SEQUENCE constraint of the **Example** slot

**Automaton**     Figure 3.246 depicts the automaton associated with the constraint DECREAS-
ING_SURF_INCREASING_SEQUENCE.

$$\left\{ \begin{array}{l} C \leftarrow +\infty \\ D \leftarrow 0 \\ F \leftarrow +\infty \\ R \leftarrow 1 \end{array} \right\}$$

$\geq s$      $\geq$

$R \wedge (F \geq C)$

$\left\{ \begin{array}{l} D \leftarrow 0 \\ F \leftarrow C \\ R \leftarrow R \wedge (F \geq C) \end{array} \right\}$  $>$

$\left\{ \begin{array}{l} C \leftarrow D + \mathtt{VAR}_i + \mathtt{VAR}_{i+1} \\ D \leftarrow 0 \end{array} \right\}$  $<$

$\{ D \leftarrow D + \mathtt{VAR}_{i+1} \}$  $=$      $< t$      $\left\{ \begin{array}{l} C \leftarrow C + D + \mathtt{VAR}_{i+1} \\ D \leftarrow 0 \end{array} \right\}$  $<$

Figure 3.246: Automaton for the DECREASING_SURF_INCREASING_SEQUENCE con-
straint obtained by applying decoration Table 2.36 to the seed transducer of the IN-
CREASING_SEQUENCE pattern

# 3.124 DECREASING_SURF_INCREASING_TERRACE

▶ ▷ ◁ ◀                    **DESCRIPTION**              **AUTOMATON**

| | |
|---|---|
| **Origin** | Based on the INCREASING_TERRACE pattern. |
| **Constraint** | DECREASING_SURF_INCREASING_TERRACE(VARIABLES) |
| **Argument** | VARIABLES : collection(var−dvar) |
| **Restriction** | required(VARIABLES, var) |

**Purpose**

Succeeds if the values denoting the surface of each occurrence of the INCREAS-ING_TERRACE pattern in the time-series given by the VARIABLES collection are decreasing.

An occurrence of the pattern INCREASING_TERRACE is the *maximal* subsequence which matches the regular expression $<=^+<$.

Assume that the occurrence of the pattern INCREASING_TERRACE starts at position $i$ and ends at position $j$. The feature SURF computes the sum of the values from index $i + 1$ to index $j$.

**Example**

$(\langle 2, 5, 5, 5, 6, 2, 3, 4, 4, 5, 3, 3, 1, 2, 2, 3\rangle)$

Figure 3.247 provides an example where the DECREASING_SURF_INCREASING_TERRACE $([2, 5, 5, 5, 6, 2, 3, 4, 4, 5, 3, 3, 1, 2, 2, 3])$ constraint holds.

**Typical**

$|\text{VARIABLES}| > 3$
$\text{range}(\text{VARIABLES.var}) > 2$

Figure 3.247: Illustrating the DECREASING_SURF_INCREASING_TERRACE constraint of the **Example** slot

**Automaton**        Figure 3.248 depicts the automaton associated with the constraint DECREAS-
                     ING_SURF_INCREASING_TERRACE.

Figure 3.248: Automaton for the DECREASING_SURF_INCREASING_TERRACE con-
straint obtained by applying decoration Table 2.36 to the seed transducer of the IN-
CREASING_TERRACE pattern

# 3.125   DECREASING_SURF_INFLEXION

▶ ▷ ◁ ◀        **DESCRIPTION**              **AUTOMATON**

$< (< | =)^* > | > (> | =)^* <$

| | |
|---|---|
| **Origin** | Based on the INFLEXION pattern. |
| **Constraint** | DECREASING_SURF_INFLEXION(VARIABLES) |
| **Argument** | VARIABLES : collection(var−dvar) |
| **Restriction** | required(VARIABLES, var) |
| **Purpose** | Succeeds if the values denoting the surface of each occurrence of the INFLEXION pattern in the time-series given by the VARIABLES collection are decreasing. <br><br> An occurrence of the pattern INFLEXION is the *maximal* subsequence which matches the regular expression $< (< | =)^* > | > (> | =)^* <$. <br><br> Assume that the occurrence of the pattern INFLEXION starts at position $i$ and ends at position $j$. The feature SURF computes the sum of the values from index $i + 1$ to index $j$. |
| **Example** | $(\langle 6, 5, 5, 4, 3, 2, 3, 3, 4, 6, 3, 3, 1, 2, 1, 1 \rangle)$ |

Figure 3.249 provides an example where the DECREASING_SURF_INFLEXION ([6, 5, 5, 4, 3, 2, 3, 3, 4, 6, 3, 3, 1, 2, 1, 1]) constraint holds.

| | |
|---|---|
| **Typical** | $\|\text{VARIABLES}\| > 2$ <br> range(VARIABLES.var) $> 1$ |

Figure 3.249: Illustrating the DECREASING_SURF_INFLEXION constraint of the **Example** slot

**Automaton**     Figure 3.250 depicts the automaton associated with the constraint DECREAS-
ING_SURF_INFLEXION.

Figure 3.250: Automaton for the DECREASING_SURF_INFLEXION constraint obtained
by applying decoration Table 2.36 to the seed transducer of the INFLEXION pattern
(transition $r \rightarrow t$ has the same accumulators updates as transition $t \rightarrow r$)

# 3.126   DECREASING_SURF_PEAK

▶ ▷ ◁ ◀        **DESCRIPTION**            **AUTOMATON**

$< (= | <)^* (> | =)^* >$

| | |
|---|---|
| **Origin** | Based on the PEAK pattern. |
| **Constraint** | DECREASING_SURF_PEAK(VARIABLES) |
| **Argument** | VARIABLES : collection(var−dvar) |
| **Restriction** | required(VARIABLES, var) |

**Purpose**

Succeeds if the values denoting the surface of each occurrence of the PEAK pattern in the time-series given by the VARIABLES collection are decreasing.

An occurrence of the pattern PEAK is the *maximal* subsequence which matches the regular expression $< (= | <)^* (> | =)^* >$.

Assume that the occurrence of the pattern PEAK starts at position $i$ and ends at position $j$. The feature SURF computes the sum of the values from index $i + 1$ to index $j$.

**Example**

$(\langle 1, 6, 6, 6, 2, 5, 4, 3, 2, 2, 3, 2, 1, 5, 5, 7 \rangle)$

Figure 3.251 provides an example where the DECREASING_SURF_PEAK $([1, 6, 6, 6, 2, 5, 4, 3, 2, 2, 3, 2, 1, 5, 5, 7])$ constraint holds.

**Typical**

$|\mathtt{VARIABLES}| > 2$
$\mathrm{range}(\mathtt{VARIABLES.var}) > 1$

Figure 3.251: Illustrating the DECREASING_SURF_PEAK constraint of the **Example** slot

**Automaton**  Figure 3.252 depicts the automaton associated with the constraint DECREAS-ING_SURF_PEAK.

Figure 3.252: Automaton for the DECREASING_SURF_PEAK constraint obtained by applying decoration Table 2.36 to the seed transducer of the PEAK pattern

# 3.127 DECREASING_SURF_PLAIN

► ▷ ◁ ◄          **DESCRIPTION**          **AUTOMATON**

| | |
|---|---|
| **Origin** | Based on the PLAIN pattern. |
| **Constraint** | DECREASING_SURF_PLAIN(VARIABLES) |
| **Argument** | VARIABLES : collection(var−dvar) |
| **Restriction** | required(VARIABLES, var) |
| **Purpose** | Succeeds if the values denoting the surface of each occurrence of the PLAIN pattern in the time-series given by the VARIABLES collection are decreasing. An occurrence of the pattern PLAIN is the *maximal* subsequence which matches the regular expression >=*<. Assume that the occurrence of the pattern PLAIN starts at position $i$ and ends at position $j$. The feature SURF computes the sum of the values from index $i + 1$ to index $j$. |
| **Example** | $(\langle 2, 3, 6, 5, 5, 7, 6, 6, 4, 5, 5, 4, 3, 6, 6, 3 \rangle)$ |

Figure 3.253 provides an example where the DECREASING_SURF_PLAIN $([2, 3, 6, 5, 5, 7, 6, 6, 4, 5, 5, 4, 3, 6, 6, 3])$ constraint holds.

Figure 3.253: Illustrating the DECREASING_SURF_PLAIN constraint of the **Example** slot

**Typical**         $|\texttt{VARIABLES}| > 2$
                    $\texttt{range}(\texttt{VARIABLES.var}) > 1$

**Automaton**    Figure 3.254 depicts the automaton associated with the constraint DECREAS-ING_SURF_PLAIN.

Figure 3.254: Automaton for the DECREASING_SURF_PLAIN constraint obtained by applying decoration Table 2.36 to the seed transducer of the PLAIN pattern

## 3.128   DECREASING_SURF_PLATEAU

▶ ▷ ◁ ◀            DESCRIPTION            AUTOMATON

| | |
|---|---|
| **Origin** | Based on the PLATEAU pattern. |
| **Constraint** | DECREASING_SURF_PLATEAU(VARIABLES) |
| **Argument** | VARIABLES : collection(var−dvar) |
| **Restriction** | required(VARIABLES, var) |
| **Purpose** | Succeeds if the values denoting the surface of each occurrence of the PLATEAU pattern in the time-series given by the VARIABLES collection are decreasing. An occurrence of the pattern PLATEAU is the *maximal* subsequence which matches the regular expression $<=^*>$. Assume that the occurrence of the pattern PLATEAU starts at position $i$ and ends at position $j$. The feature SURF computes the sum of the values from index $i+1$ to index $j$. |
| **Example** | $(\langle 5, 2, 2, 5, 5, 4, 3, 3, 4, 2, 2, 1, 3, 2, 5, 7\rangle)$ |

Figure 3.255 provides an example where the DECREASING_SURF_PLATEAU ([5, 2, 2, 5, 5, 4, 3, 3, 4, 2, 2, 1, 3, 2, 5, 7]) constraint holds.

| | |
|---|---|
| **Typical** | $\|$VARIABLES$\| > 2$ <br> range(VARIABLES.var) $> 1$ |

Figure 3.255: Illustrating the DECREASING_SURF_PLATEAU constraint of the **Example** slot

**Automaton**     Figure 3.256 depicts the automaton associated with the constraint DECREAS-ING_SURF_PLATEAU.

Figure 3.256: Automaton for the DECREASING_SURF_PLATEAU constraint obtained by applying decoration Table 2.36 to the seed transducer of the PLATEAU pattern

## 3.129 DECREASING_SURF_PROPER_PLAIN

▶ ▷ ◁ ◀        DESCRIPTION          AUTOMATON

**Origin**         Based on the PROPER_PLAIN pattern.

**Constraint**     DECREASING_SURF_PROPER_PLAIN(VARIABLES)

**Argument**       VARIABLES : collection(var−dvar)

**Restriction**    required(VARIABLES, var)

**Purpose**        Succeeds if the values denoting the surface of each occurrence of the PROPER_PLAIN pattern in the time-series given by the VARIABLES collection are decreasing.
An occurrence of the pattern PROPER_PLAIN is the *maximal* subsequence which matches the regular expression $> =^+ <$.
Assume that the occurrence of the pattern PROPER_PLAIN starts at position $i$ and ends at position $j$. The feature SURF computes the sum of the values from index $i + 1$ to index $j$.

**Example**        $(\langle 2, 7, 5, 5, 5, 6, 3, 7, 4, 4, 5, 6, 5, 3, 3, 5 \rangle)$

Figure 3.257 provides an example where the DECREASING_SURF_PROPER_PLAIN ($[2, 7, 5, 5, 5, 6, 3, 7, 4, 4, 5, 6, 5, 3, 3, 5]$) constraint holds.

**Typical**        $|\text{VARIABLES}| > 3$
range(VARIABLES.var) > 1

Figure 3.257: Illustrating the DECREASING_SURF_PROPER_PLAIN constraint of the **Example** slot

                 Figure 3.258 depicts the automaton associated with the constraint DECREAS-
ING_SURF_PROPER_PLAIN.

Figure 3.258: Automaton for the DECREASING_SURF_PROPER_PLAIN constraint obtained by applying decoration Table 2.36 to the seed transducer of the PROPER_PLAIN pattern

## 3.130   DECREASING_SURF_PROPER_PLATEAU

► ▷ ◁ ◀          DESCRIPTION          AUTOMATON

| | |
|---|---|
| **Origin** | Based on the PROPER_PLATEAU pattern. |
| **Constraint** | DECREASING_SURF_PROPER_PLATEAU(VARIABLES) |
| **Argument** | VARIABLES : collection(var−dvar) |
| **Restriction** | required(VARIABLES, var) |

**Purpose**

Succeeds if the values denoting the surface of each occurrence of the PROPER_PLATEAU pattern in the time-series given by the VARIABLES collection are decreasing.

An occurrence of the pattern PROPER_PLATEAU is the *maximal* subsequence which matches the regular expression $< =^+ >$.

Assume that the occurrence of the pattern PROPER_PLATEAU starts at position $i$ and ends at position $j$. The feature SURF computes the sum of the values from index $i+1$ to index $j$.

**Example**

$(\langle 3, 5, 5, 5, 3, 2, 3, 4, 4, 1, 5, 2, 3, 3, 1, 7\rangle)$

Figure 3.259 provides an example where the DECREASING_SURF_PROPER_PLATEAU ([3, 5, 5, 5, 3, 2, 3, 4, 4, 1, 5, 2, 3, 3, 1, 7]) constraint holds.

**Typical**

$|\texttt{VARIABLES}| > 3$

range(VARIABLES.var) $> 1$

Figure 3.259: Illustrating the DECREASING_SURF_PROPER_PLATEAU constraint of the **Example** slot

**Automaton**        Figure 3.260 depicts the automaton associated with the constraint DECREAS-
ING_SURF_PROPER_PLATEAU.

Figure 3.260: Automaton for the DECREASING_SURF_PROPER_PLATEAU con-
straint obtained by applying decoration Table 2.36 to the seed transducer of the
PROPER_PLATEAU pattern

## 3.131  DECREASING_SURF_STEADY

▶ ▷ ◁ ◀          **DESCRIPTION**          **AUTOMATON**

| | |
|---|---|
| **Origin** | Based on the STEADY pattern. |
| **Constraint** | DECREASING_SURF_STEADY(VARIABLES) |
| **Argument** | VARIABLES : collection(var−dvar) |
| **Restriction** | required(VARIABLES, var) |
| **Purpose** | Succeeds if the values denoting the surface of each occurrence of the STEADY pattern in the time-series given by the VARIABLES collection are decreasing. An occurrence of the pattern STEADY is the subsequence which matches the regular expression =. Assume that the occurrence of the pattern STEADY starts at position $i$ and ends at index $j$. The feature SURF computes the sum of the values from index $i$ to index $j+1$. |
| **Example** | $(\langle 3, 6, 6, 6, 2, 7, 5, 5, 5, 6, 5, 5, 5, 3, 3, 7 \rangle)$ |

Figure 3.261 provides an example where the DECREASING_SURF_STEADY ([3, 6, 6, 6, 2, 7, 5, 5, 5, 6, 5, 5, 5, 3, 3, 7]) constraint holds.

Figure 3.261: Illustrating the DECREASING_SURF_STEADY constraint of the **Example** slot

**Typical** $|VARIABLES| > 1$

**Automaton**          Figure 3.262 depicts the automaton associated with the constraint DECREAS-ING_SURF_STEADY.

Figure 3.262: Automaton for the DECREASING_SURF_STEADY constraint obtained by applying decoration Table 2.36 to the seed transducer of the STEADY pattern

# 3.132 DECREASING_SURF_STEADY_SEQUENCE

DESCRIPTION            AUTOMATON

| | |
|---|---|
| **Origin** | Based on the STEADY_SEQUENCE pattern. |
| **Constraint** | DECREASING_SURF_STEADY_SEQUENCE(VARIABLES) |
| **Argument** | VARIABLES : collection(var−dvar) |
| **Restriction** | required(VARIABLES, var) |
| **Purpose** | Succeeds if the values denoting the surface of each occurrence of the STEADY_SEQUENCE pattern in the time-series given by the VARIABLES collection are decreasing. An occurrence of the pattern STEADY_SEQUENCE is the *maximal* subsequence which matches the regular expression $=^{+}$. Assume that the occurrence of the pattern STEADY_SEQUENCE starts at position $i$ and ends at position $j$. The feature SURF computes the sum of the values from index $i$ to index $j + 1$. |
| **Example** | $(\langle 4, 3, 2, 5, 5, 5, 3, 2, 4, 4, 6, 5, 3, 2, 2, 6 \rangle)$ |

Figure 3.263 provides an example where the DECREASING_SURF_STEADY_SEQUENCE $([4, 3, 2, 5, 5, 5, 3, 2, 4, 4, 6, 5, 3, 2, 2, 6])$ constraint holds.

Figure 3.263: Illustrating the DECREASING_SURF_STEADY_SEQUENCE constraint of the **Example** slot

**Typical**    $|\mathtt{VARIABLES}| > 1$

**Automaton**        Figure 3.264 depicts the automaton associated with the constraint DECREAS-
ING_SURF_STEADY_SEQUENCE.

Figure 3.264: Automaton for the DECREASING_SURF_STEADY_SEQUENCE con-
straint obtained by applying decoration Table 2.36 to the seed transducer of the
STEADY_SEQUENCE pattern

# 3.133   DECREASING_SURF_STRICTLY_DECREASING_SEQUENCE

▶ ▷ ◁ ◀        **DESCRIPTION**              **AUTOMATON**

| | |
|---|---|
| **Origin** | Based on the STRICTLY_DECREASING_SEQUENCE pattern. |
| **Constraint** | DECREASING_SURF_STRICTLY_DECREASING_SEQUENCE(VARIABLES) |
| **Argument** | VARIABLES : collection(var−dvar) |
| **Restriction** | required(VARIABLES, var) |

**Purpose**

Succeeds if the values denoting the surface of each occurrence of the STRICTLY_DECREASING_SEQUENCE pattern in the time-series given by the VARIABLES collection are decreasing.

An occurrence of the pattern STRICTLY_DECREASING_SEQUENCE is the *maximal* subsequence which matches the regular expression $>^+$.

Assume that the occurrence of the pattern STRICTLY_DECREASING_SEQUENCE starts at position $i$ and ends at position $j$. The feature SURF computes the sum of the values from index $i$ to index $j + 1$.

**Example**

$(\langle 7, 4, 6, 5, 5, 4, 4, 3, 6, 1, 4, 2, 1, 2, 3, 4 \rangle)$

Figure 3.265 provides an example where the DECREASING_SURF_STRICTLY_DECREASING_SEQUENCE $([7, 4, 6, 5, 5, 4, 4, 3, 6, 1, 4, 2, 1, 2, 3, 4])$ constraint holds.

**Typical**

$|\texttt{VARIABLES}| > 1$

range(VARIABLES.var) $> 1$

Figure 3.265: Illustrating the DECREASING_SURF_STRICTLY_DECREASING_SEQUENCE constraint of the **Example** slot

**Automaton**        Figure 3.266 depicts the automaton associated with the constraint DECREAS-
ING_SURF_STRICTLY_DECREASING_SEQUENCE.

Figure        3.266:          Automaton       for      the       DECREAS-
ING_SURF_STRICTLY_DECREASING_SEQUENCE constraint obtained by applying dec-
oration Table 2.36 to the seed transducer of the STRICTLY_DECREASING_SEQUENCE
pattern

## 3.134 DECREASING_SURF_STRICTLY_INCREASING_SEQUENCE

▶ ▷ ◁ ◀        DESCRIPTION        AUTOMATON

| | |
|---|---|
| **Origin** | Based on the STRICTLY_INCREASING_SEQUENCE pattern. |
| **Constraint** | DECREASING_SURF_STRICTLY_INCREASING_SEQUENCE(VARIABLES) |
| **Argument** | VARIABLES : collection(var−dvar) |
| **Restriction** | required(VARIABLES, var) |

**Purpose**

Succeeds if the values denoting the surface of each occurrence of the STRICTLY_INCREASING_SEQUENCE pattern in the time-series given by the VARIABLES collection are decreasing.

An occurrence of the pattern STRICTLY_INCREASING_SEQUENCE is the *maximal* subsequence which matches the regular expression $<^+$.

Assume that the occurrence of the pattern STRICTLY_INCREASING_SEQUENCE starts at position $i$ and ends at position $j$. The feature SURF computes the sum of the values from index $i$ to index $j + 1$.

**Example**

$(\langle 3, 4, 5, 5, 4, 3, 3, 1, 4, 3, 3, 2, 2, 1, 3, 1 \rangle)$

Figure 3.267 provides an example where the DECREASING_SURF_STRICTLY_INCREASING_SEQUENCE $([3, 4, 5, 5, 4, 3, 3, 1, 4, 3, 3, 2, 2, 1, 3, 1])$ constraint holds.

**Typical**

$|\texttt{VARIABLES}| > 1$
range(VARIABLES.var) > 1

Figure 3.267: Illustrating the DECREASING_SURF_STRICTLY_INCREASING_SEQUENCE constraint of the **Example** slot

**Automaton**    Figure 3.268 depicts the automaton associated with the constraint DECREAS-
ING_SURF_STRICTLY_INCREASING_SEQUENCE.

$$\left\{ \begin{array}{l} C \leftarrow +\infty \\ D \leftarrow 0 \\ F \leftarrow +\infty \\ R \leftarrow 1 \end{array} \right\}$$

$\geq s$    $\geq$

$R \wedge (F \geq C)$

$$\left\{ \begin{array}{l} D \leftarrow 0 \\ F \leftarrow C \\ R \leftarrow R \wedge (F \geq C) \end{array} \right\} \qquad \overset{<}{\left\{ \begin{array}{l} C \leftarrow D + \texttt{VAR}_i + \texttt{VAR}_{i+1} \\ D \leftarrow 0 \end{array} \right\}}$$

$< r$    $\overset{<}{\left\{ \begin{array}{l} C \leftarrow C + D + \texttt{VAR}_{i+1} \\ D \leftarrow 0 \end{array} \right\}}$

Figure        3.268:           Automaton        for       the       DECREAS-
ING_SURF_STRICTLY_INCREASING_SEQUENCE constraint obtained by applying dec-
oration Table 2.36 to the seed transducer of the STRICTLY_INCREASING_SEQUENCE
pattern

# 3.135 DECREASING_SURF_SUMMIT

▶ ▷ ◁ ◀         **DESCRIPTION**         **AUTOMATON**

$(< | < (= | <)^* <)(> | > (= | >)^* >)$

| **Origin** | Based on the SUMMIT pattern. |
|---|---|
| **Constraint** | DECREASING_SURF_SUMMIT(VARIABLES) |
| **Argument** | VARIABLES : collection(var−dvar) |
| **Restriction** | required(VARIABLES, var) |

**Purpose**

Succeeds if the values denoting the surface of each occurrence of the SUMMIT pattern in the time-series given by the VARIABLES collection are decreasing.

An occurrence of the pattern SUMMIT is the *maximal* subsequence which matches the regular expression $(< | < (= | <)^* <)(> | > (= | >)^* >)$.

Assume that the occurrence of the pattern SUMMIT starts at position $i$ and ends at position $j$. The feature SURF computes the sum of the values from index $i + 1$ to index $j$.

**Example**

$(\langle 1, 4, 4, 5, 4, 2, 6, 6, 2, 3, 5, 4, 1, 2, 4, 2 \rangle)$

Figure 3.269 provides an example where the DECREASING_SURF_SUMMIT $([1, 4, 4, 5, 4, 2, 6, 6, 2, 3, 5, 4, 1, 2, 4, 2])$ constraint holds.

**Typical**

$|\texttt{VARIABLES}| > 2$
range(VARIABLES.var) $> 1$

Figure 3.269: Illustrating the DECREASING_SURF_SUMMIT constraint of the **Example** slot

**Automaton**    Figure 3.270 depicts the automaton associated with the constraint DECREAS-
ING_SURF_SUMMIT.

Figure 3.270: Automaton for the DECREASING_SURF_SUMMIT constraint obtained by
applying decoration Table 2.36 to the seed transducer of the SUMMIT pattern (transition
$u \rightarrow r$ has the same accumulator update as transition $r \rightarrow u$)

# 3.136   DECREASING_SURF_VALLEY

▶ ▷ ◁ ◀        **DESCRIPTION**        **AUTOMATON**

| | |
|---|---|
| **Origin** | Based on the VALLEY pattern. |
| **Constraint** | DECREASING_SURF_VALLEY(VARIABLES) |
| **Argument** | VARIABLES : collection(var−dvar) |
| **Restriction** | required(VARIABLES, var) |
| **Purpose** | Succeeds if the values denoting the surface of each occurrence of the VALLEY pattern in the time-series given by the VARIABLES collection are decreasing. An occurrence of the pattern VALLEY is the *maximal* subsequence which matches the regular expression $> (= \mid >)^* (< \mid =)^* <$. Assume that the occurrence of the pattern VALLEY starts at position $i$ and ends at position $j$. The feature SURF computes the sum of the values from index $i + 1$ to index $j$. |
| **Example** | $(\langle 1, 3, 3, 7, 6, 5, 4, 6, 6, 5, 4, 3, 6, 2, 2, 7 \rangle)$ |

Figure 3.271 provides an example where the DECREASING_SURF_VALLEY $([1, 3, 3, 7, 6, 5, 4, 6, 6, 5, 4, 3, 6, 2, 2, 7])$ constraint holds.

| | |
|---|---|
| **Typical** | $\lvert \texttt{VARIABLES} \rvert > 2$ |
| | range(VARIABLES.var) > 1 |

Figure 3.271: Illustrating the DECREASING_SURF_VALLEY constraint of the **Example** slot

**Automaton**   Figure 3.272 depicts the automaton associated with the constraint DECREAS-
ING_SURF_VALLEY.

Figure 3.272: Automaton for the DECREASING_SURF_VALLEY constraint obtained by
applying decoration Table 2.36 to the seed transducer of the VALLEY pattern

# 3.137 DECREASING_SURF_ZIGZAG

▶ ▷ ◁ ◀        **DESCRIPTION**         **AUTOMATON**

$(<>)^+(< | <>) | (><)^+(> | ><)$

| | |
|---|---|
| **Origin** | Based on the ZIGZAG pattern. |
| **Constraint** | DECREASING_SURF_ZIGZAG(VARIABLES) |
| **Argument** | VARIABLES : collection(var−dvar) |
| **Restriction** | required(VARIABLES, var) |
| **Purpose** | Succeeds if the values denoting the surface of each occurrence of the ZIGZAG pattern in the time-series given by the VARIABLES collection are decreasing. <br> An occurrence of the pattern ZIGZAG is the *maximal* subsequence which matches the regular expression $(<>)^+(< | <>) | (><)^+(> | ><)$. <br> Assume that the occurrence of the pattern ZIGZAG starts at position $i$ and ends at position $j$. The feature SURF computes the sum of the values from index $i + 1$ to index $j$. |
| **Example** | $(\langle 6, 5, 7, 4, 5, 1, 1, 6, 3, 6, 4, 3, 5, 2, 7, 7 \rangle)$ |

Figure 3.273 provides an example where the DECREASING_SURF_ZIGZAG $([6, 5, 7, 4, 5, 1, 1, 6, 3, 6, 4, 3, 5, 2, 7, 7])$ constraint holds.

| | |
|---|---|
| **Typical** | $|\text{VARIABLES}| > 3$ <br> range(VARIABLES.var) $> 1$ |

Figure 3.273: Illustrating the DECREASING_SURF_ZIGZAG constraint of the **Example** slot

**Automaton**          Figure 3.274 depicts the automaton associated with the constraint DECREAS-
                       ING_SURF_ZIGZAG.

Figure 3.274: Automaton for the DECREASING_SURF_ZIGZAG constraint obtained by applying decoration Table 2.36 to the seed transducer of the ZIGZAG pattern; (1) missing transitions from $a, b, c, d, e, f$ to $s$ are labelled by $=$; (2) on transitions from $b, c, e, f$ to $s$ the accumulator $D$ is reset to its initial value; (3) on transitions from $c, f$ to $s$ the accumulator $F$ is reset to $C$, and the accumulator $R$ is updated wrt $C$ and $F$

# 3.138   DECREASING_WIDTH_DECREASING_SEQUENCE

▶ ▷ ◁ ◀            **DESCRIPTION**            **AUTOMATON**

$> (> | =)^* > |$
$>$

**Origin**            Based on the DECREASING_SEQUENCE pattern.

**Constraint**       DECREASING_WIDTH_DECREASING_SEQUENCE(VARIABLES)

**Argument**         VARIABLES : collection(var−dvar)

**Restriction**      required(VARIABLES, var)

**Purpose**          Succeeds if the values denoting the width of each occurrence of the DECREAS-
                     ING_SEQUENCE pattern in the time-series given by the VARIABLES collection are de-
                     creasing.
                     An occurrence of the pattern DECREASING_SEQUENCE is the *maximal* subsequence
                     which matches the regular expression $> (> | =)^* > | >$.
                     Assume that the occurrence of the pattern DECREASING_SEQUENCE starts at position $i$
                     and ends at position $j$. The feature WIDTH computes the value $j - i + 2$.

**Example**          $(\langle 2, 6, 5, 2, 3, 3, 3, 4, 3, 2, 3, 4, 4, 5, 7, 5 \rangle)$

                     Figure    3.275    provides    an    example    where    the    DECREAS-
                     ING_WIDTH_DECREASING_SEQUENCE    ($[2, 6, 5, 2, 3, 3, 3, 4, 3, 2, 3, 4, 4, 5, 7, 5]$)    con-
                     straint holds.

**Typical**          $|\text{VARIABLES}| > 1$
                     range(VARIABLES.var) > 1

Figure 3.275: Illustrating the DECREASING_WIDTH_DECREASING_SEQUENCE constraint of the **Example** slot

**Automaton**        Figure 3.276 depicts the automaton associated with the constraint DECREAS-
                     ING_WIDTH_DECREASING_SEQUENCE.

Figure 3.276: Automaton for the DECREASING_WIDTH_DECREASING_SEQUENCE constraint obtained by applying decoration Table 2.36 to the seed transducer of the DECREASING_SEQUENCE pattern

# 3.139 DECREASING_WIDTH_DECREASING_TERRACE

          **DESCRIPTION**          **AUTOMATON**

| | |
|---|---|
| **Origin** | Based on the DECREASING_TERRACE pattern. |
| **Constraint** | DECREASING_WIDTH_DECREASING_TERRACE(VARIABLES) |
| **Argument** | VARIABLES : collection(var−dvar) |
| **Restriction** | required(VARIABLES, var) |

**Purpose**

Succeeds if the values denoting the width of each occurrence of the DECREAS-ING_TERRACE pattern in the time-series given by the VARIABLES collection are decreasing.

An occurrence of the pattern DECREASING_TERRACE is the *maximal* subsequence which matches the regular expression $>=^+>$.

Assume that the occurrence of the pattern DECREASING_TERRACE starts at position $i$ and ends at position $j$. The feature WIDTH computes the value $j - i$.

**Example**

$$(\langle 7, 6, 6, 6, 4, 3, 3, 2, 2, 4, 4, 6, 3, 3, 1, 1 \rangle)$$

Figure 3.277 provides an example where the DECREAS-ING_WIDTH_DECREASING_TERRACE ($[7, 6, 6, 6, 4, 3, 3, 2, 2, 4, 4, 6, 3, 3, 1, 1]$) constraint holds.

**Typical**

$|\text{VARIABLES}| > 3$

range(VARIABLES.var) $> 2$

Figure 3.277: Illustrating the DECREASING_WIDTH_DECREASING_TERRACE constraint of the **Example** slot

**Automaton**       Figure 3.278 depicts the automaton associated with the constraint DECREAS-
ING_WIDTH_DECREASING_TERRACE.

Figure 3.278: Automaton for the DECREASING_WIDTH_DECREASING_TERRACE con-
straint obtained by applying decoration Table 2.36 to the seed transducer of the DE-
CREASING_TERRACE pattern

## 3.140 DECREASING_WIDTH_GORGE

          **DESCRIPTION**          **AUTOMATON**

$(> | > (= | >)^* >)(< | < (= | <)^* <)$

| | |
|---|---|
| **Origin** | Based on the GORGE pattern. |
| **Constraint** | DECREASING_WIDTH_GORGE(VARIABLES) |
| **Argument** | VARIABLES : collection(var−dvar) |
| **Restriction** | required(VARIABLES, var) |
| **Purpose** | Succeeds if the values denoting the width of each occurrence of the GORGE pattern in the time-series given by the VARIABLES collection are decreasing. An occurrence of the pattern GORGE is the *maximal* subsequence which matches the regular expression $(> | > (= | >)^* >)(< | < (= | <)^* <)$. Assume that the occurrence of the pattern GORGE starts at position $i$ and ends at position $j$. The feature WIDTH computes the value $j - i$. |
| **Example** | $(\langle 6, 2, 3, 4, 5, 6, 1, 1, 5, 4, 2, 3, 6, 1, 4, 5 \rangle)$ |

Figure 3.279 provides an example where the DECREASING_WIDTH_GORGE ([6, 2, 3, 4, 5, 6, 1, 1, 5, 4, 2, 3, 6, 1, 4, 5]) constraint holds.

Figure 3.279: Illustrating the DECREASING_WIDTH_GORGE constraint of the **Example** slot

**Typical**

$|\texttt{VARIABLES}| > 2$
$\texttt{range}(\texttt{VARIABLES.var}) > 1$

**Automaton** Figure 3.280 depicts the automaton associated with the constraint DECREASING_WIDTH_GORGE.

Figure 3.280: Automaton for the DECREASING_WIDTH_GORGE constraint obtained by applying decoration Table 2.36 to the seed transducer of the GORGE pattern (transition $u \rightarrow r$ has the same accumulator update as transition $r \rightarrow u$)

## 3.141 DECREASING_WIDTH_INCREASING_SEQUENCE

▶ ▷ ◁ ◀            **DESCRIPTION**            **AUTOMATON**

| **Origin** | Based on the INCREASING_SEQUENCE pattern. |

| **Constraint** | DECREASING_WIDTH_INCREASING_SEQUENCE(VARIABLES) |

| **Argument** | VARIABLES : collection(var−dvar) |

| **Restriction** | required(VARIABLES, var) |

| **Purpose** | Succeeds if the values denoting the width of each occurrence of the INCREASING_SEQUENCE pattern in the time-series given by the VARIABLES collection are decreasing. An occurrence of the pattern INCREASING_SEQUENCE is the *maximal* subsequence which matches the regular expression $< (< | =)^* < | <$. Assume that the occurrence of the pattern INCREASING_SEQUENCE starts at position $i$ and ends at position $j$. The feature WIDTH computes the value $j - i + 2$. |

| **Example** | $(\langle 3, 4, 5, 5, 4, 3, 3, 1, 4, 3, 3, 2, 2, 1, 3, 1 \rangle)$ |

Figure 3.281 provides an example where the DECREASING_WIDTH_INCREASING_SEQUENCE $([3, 4, 5, 5, 4, 3, 3, 1, 4, 3, 3, 2, 2, 1, 3, 1])$ constraint holds.

Figure 3.281: Illustrating the DECREASING_WIDTH_INCREASING_SEQUENCE constraint of the **Example** slot

**Typical**
$$|\texttt{VARIABLES}| > 1$$
$$\texttt{range}(\texttt{VARIABLES.var}) > 1$$

**Automaton** Figure 3.282 depicts the automaton associated with the constraint DECREAS-ING_WIDTH_INCREASING_SEQUENCE.

Figure 3.282: Automaton for the DECREASING_WIDTH_INCREASING_SEQUENCE constraint obtained by applying decoration Table 2.36 to the seed transducer of the IN-CREASING_SEQUENCE pattern

## 3.142 DECREASING_WIDTH_INCREASING_TERRACE

▶ ▷ ◁ ◀        DESCRIPTION         AUTOMATON

**Origin**        Based on the INCREASING_TERRACE pattern.

**Constraint**    DECREASING_WIDTH_INCREASING_TERRACE(VARIABLES)

**Argument**      VARIABLES : collection(var−dvar)

**Restriction**   required(VARIABLES, var)

**Purpose**       Succeeds if the values denoting the width of each occurrence of the INCREAS-ING_TERRACE pattern in the time-series given by the VARIABLES collection are decreas-ing.
An occurrence of the pattern INCREASING_TERRACE is the *maximal* subsequence which matches the regular expression $<=^+<$.
Assume that the occurrence of the pattern INCREASING_TERRACE starts at position $i$ and ends at position $j$. The feature WIDTH computes the value $j - i$.

**Example**       $(\langle 2, 5, 5, 5, 6, 2, 3, 4, 4, 5, 3, 3, 1, 2, 2, 3 \rangle)$

Figure 3.283 provides an example where the DECREAS-ING_WIDTH_INCREASING_TERRACE ([2, 5, 5, 5, 6, 2, 3, 4, 4, 5, 3, 3, 1, 2, 2, 3]) constraint holds.

**Typical**       |VARIABLES| > 3
                  range(VARIABLES.var) > 2

Figure 3.283: Illustrating the DECREASING_WIDTH_INCREASING_TERRACE constraint of the **Example** slot

    Figure 3.284 depicts the automaton associated with the constraint DECREAS-
ING_WIDTH_INCREASING_TERRACE.

Figure 3.284: Automaton for the DECREASING_WIDTH_INCREASING_TERRACE con-
straint obtained by applying decoration Table 2.36 to the seed transducer of the IN-
CREASING_TERRACE pattern

## 3.143  DECREASING_WIDTH_INFLEXION

► ▷ ◁ ◄          **DESCRIPTION**          **AUTOMATON**

$< (< | =)^* > | > (> | =)^* <$

**Origin**          Based on the INFLEXION pattern.

**Constraint**          DECREASING_WIDTH_INFLEXION(VARIABLES)

**Argument**          VARIABLES : collection(var−dvar)

**Restriction**          required(VARIABLES, var)

**Purpose**          Succeeds if the values denoting the width of each occurrence of the INFLEXION pattern in the time-series given by the VARIABLES collection are decreasing.

An occurrence of the pattern INFLEXION is the *maximal* subsequence which matches the regular expression $< (< | =)^* > | > (> | =)^* <$.

Assume that the occurrence of the pattern INFLEXION starts at position $i$ and ends at position $j$. The feature WIDTH computes the value $j - i$.

**Example**          $(\langle 6, 5, 5, 4, 3, 2, 3, 3, 4, 6, 3, 3, 1, 2, 1, 1 \rangle)$

Figure 3.285 provides an example where the DECREASING_WIDTH_INFLEXION ([6, 5, 5, 4, 3, 2, 3, 3, 4, 6, 3, 3, 1, 2, 1, 1]) constraint holds.

Figure 3.285: Illustrating the DECREASING_WIDTH_INFLEXION constraint of the **Example** slot

**Typical**
$$|\texttt{VARIABLES}| > 2$$
$$\texttt{range}(\texttt{VARIABLES.var}) > 1$$

**Automaton**   Figure 3.286 depicts the automaton associated with the constraint DECREAS-ING_WIDTH_INFLEXION.

Figure 3.286: Automaton for the DECREASING_WIDTH_INFLEXION constraint obtained by applying decoration Table 2.36 to the seed transducer of the INFLEXION pattern (transition $r \rightarrow t$ has the same accumulators updates as transition $t \rightarrow r$)

# 3.144 DECREASING_WIDTH_PEAK

▶ ▷ ◁ ◀          **DESCRIPTION**          **AUTOMATON**

| | |
|---|---|
| **Origin** | Based on the PEAK pattern. |
| **Constraint** | DECREASING_WIDTH_PEAK(VARIABLES) |
| **Argument** | VARIABLES : collection(var−dvar) |
| **Restriction** | required(VARIABLES, var) |

**Purpose**

Succeeds if the values denoting the width of each occurrence of the PEAK pattern in the time-series given by the VARIABLES collection are decreasing.

An occurrence of the pattern PEAK is the *maximal* subsequence which matches the regular expression $< (= | <)^* (> | =)^* >$.

Assume that the occurrence of the pattern PEAK starts at position $i$ and ends at position $j$. The feature WIDTH computes the value $j - i$.

**Example**

$(\langle 1, 6, 6, 6, 2, 5, 4, 3, 2, 2, 3, 2, 1, 5, 5, 7 \rangle)$

Figure 3.287 provides an example where the DECREASING_WIDTH_PEAK $([1, 6, 6, 6, 2, 5, 4, 3, 2, 2, 3, 2, 1, 5, 5, 7])$ constraint holds.

**Typical**

$|\text{VARIABLES}| > 2$

range(VARIABLES.var) $> 1$

Figure 3.287: Illustrating the DECREASING_WIDTH_PEAK constraint of the **Example** slot

**Automaton**                 Figure 3.288 depicts the automaton associated with the constraint DECREAS-
ING_WIDTH_PEAK.

Figure 3.288: Automaton for the DECREASING_WIDTH_PEAK constraint obtained by
applying decoration Table 2.36 to the seed transducer of the PEAK pattern

# 3.145 DECREASING_WIDTH_PLAIN

► ▷ ◁ ◀          DESCRIPTION          AUTOMATON

**Origin**         Based on the PLAIN pattern.

**Constraint**     DECREASING_WIDTH_PLAIN(VARIABLES)

**Argument**       VARIABLES : collection(var−dvar)

**Restriction**    required(VARIABLES, var)

**Purpose**        Succeeds if the values denoting the width of each occurrence of the PLAIN pattern in the time-series given by the VARIABLES collection are decreasing.

An occurrence of the pattern PLAIN is the *maximal* subsequence which matches the regular expression >=*<.

Assume that the occurrence of the pattern PLAIN starts at position $i$ and ends at position $j$. The feature WIDTH computes the value $j − i$.

**Example**        $(\langle 2, 3, 6, 5, 5, 7, 6, 6, 4, 5, 5, 4, 3, 6, 6, 3 \rangle)$

Figure 3.289 provides an example where the DECREASING_WIDTH_PLAIN ([2, 3, 6, 5, 5, 7, 6, 6, 4, 5, 5, 4, 3, 6, 6, 3]) constraint holds.

Figure 3.289: Illustrating the DECREASING_WIDTH_PLAIN constraint of the **Example** slot

**Typical**

$|\mathtt{VARIABLES}| > 2$

$\mathrm{range}(\mathtt{VARIABLES.var}) > 1$

**Automaton**          Figure 3.290 depicts the automaton associated with the constraint DECREAS-
                       ING_WIDTH_PLAIN.

Figure 3.290: Automaton for the DECREASING_WIDTH_PLAIN constraint obtained by
applying decoration Table 2.36 to the seed transducer of the PLAIN pattern

# 3.146 DECREASING_WIDTH_PLATEAU

▶ ▷ ◁ ◀        **DESCRIPTION**          **AUTOMATON**

| | |
|---|---|
| **Origin** | Based on the PLATEAU pattern. |
| **Constraint** | DECREASING_WIDTH_PLATEAU(VARIABLES) |
| **Argument** | VARIABLES : collection(var−dvar) |
| **Restriction** | required(VARIABLES, var) |
| **Purpose** | Succeeds if the values denoting the width of each occurrence of the PLATEAU pattern in the time-series given by the VARIABLES collection are decreasing. An occurrence of the pattern PLATEAU is the *maximal* subsequence which matches the regular expression $<=^*>$. Assume that the occurrence of the pattern PLATEAU starts at position $i$ and ends at position $j$. The feature WIDTH computes the value $j - i$. |
| **Example** | $(\langle 5, 2, 2, 5, 5, 4, 3, 3, 4, 2, 2, 1, 3, 2, 5, 7 \rangle)$ |

Figure 3.291 provides an example where the DECREASING_WIDTH_PLATEAU $([5, 2, 2, 5, 5, 4, 3, 3, 4, 2, 2, 1, 3, 2, 5, 7])$ constraint holds.

| | |
|---|---|
| **Typical** | $\lvert$VARIABLES$\rvert > 2$ <br> range(VARIABLES.var) $> 1$ |

Figure 3.291: Illustrating the DECREASING_WIDTH_PLATEAU constraint of the **Example** slot

**Automaton**         Figure 3.292 depicts the automaton associated with the constraint DECREAS-
                      ING_WIDTH_PLATEAU.

Figure 3.292: Automaton for the DECREASING_WIDTH_PLATEAU constraint obtained
by applying decoration Table 2.36 to the seed transducer of the PLATEAU pattern

## 3.147   DECREASING_WIDTH_PROPER_PLAIN

▶ ▷ ◁ ◀              **DESCRIPTION**          **AUTOMATON**

**Origin**              Based on the PROPER_PLAIN pattern.

**Constraint**          DECREASING_WIDTH_PROPER_PLAIN(VARIABLES)

**Argument**            VARIABLES : collection(var−dvar)

**Restriction**         required(VARIABLES, var)

**Purpose**             Succeeds if the values denoting the width of each occurrence of the PROPER_PLAIN
                        pattern in the time-series given by the VARIABLES collection are decreasing.
                        An occurrence of the pattern PROPER_PLAIN is the *maximal* subsequence which matches
                        the regular expression $> =^+ <$.
                        Assume that the occurrence of the pattern PROPER_PLAIN starts at position $i$ and ends at
                        position $j$. The feature WIDTH computes the value $j - i$.

**Example**             $(\langle 2, 7, 5, 5, 5, 6, 3, 7, 4, 4, 5, 6, 5, 3, 3, 5 \rangle)$

Figure 3.293 provides an example where the DECREASING_WIDTH_PROPER_PLAIN
([2, 7, 5, 5, 5, 6, 3, 7, 4, 4, 5, 6, 5, 3, 3, 5]) constraint holds.

Figure 3.293: Illustrating the DECREASING_WIDTH_PROPER_PLAIN constraint of the
**Example** slot

**Typical**

$|\text{VARIABLES}| > 3$

$\text{range}(\text{VARIABLES.var}) > 1$

**Automaton**          Figure 3.294 depicts the automaton associated with the constraint DECREAS-
ING_WIDTH_PROPER_PLAIN.

Figure 3.294: Automaton for the DECREASING_WIDTH_PROPER_PLAIN constraint ob-
tained by applying decoration Table 2.36 to the seed transducer of the PROPER_PLAIN
pattern

## 3.148 DECREASING_WIDTH_PROPER_PLATEAU

▶ ▷ ◁ ◀    **DESCRIPTION**        **AUTOMATON**

| | |
|---|---|
| **Origin** | Based on the PROPER_PLATEAU pattern. |
| **Constraint** | DECREASING_WIDTH_PROPER_PLATEAU(VARIABLES) |
| **Argument** | VARIABLES : collection(var−dvar) |
| **Restriction** | required(VARIABLES, var) |

**Purpose**

Succeeds if the values denoting the width of each occurrence of the PROPER_PLATEAU pattern in the time-series given by the VARIABLES collection are decreasing.

An occurrence of the pattern PROPER_PLATEAU is the *maximal* subsequence which matches the regular expression $< =^+ >$.

Assume that the occurrence of the pattern PROPER_PLATEAU starts at position $i$ and ends at position $j$. The feature WIDTH computes the value $j − i$.

**Example**

$(\langle 3, 5, 5, 5, 3, 2, 3, 4, 4, 1, 5, 2, 3, 3, 1, 7 \rangle)$

Figure 3.295 provides an example where the DECREASING_WIDTH_PROPER_PLATEAU ([3, 5, 5, 5, 3, 2, 3, 4, 4, 1, 5, 2, 3, 3, 1, 7]) constraint holds.

**Typical**

$|\text{VARIABLES}| > 3$
range(VARIABLES.var) $> 1$

Figure 3.295: Illustrating the DECREASING_WIDTH_PROPER_PLATEAU constraint of the **Example** slot

**Automaton**          Figure 3.296 depicts the automaton associated with the constraint DECREAS-
ING_WIDTH_PROPER_PLATEAU.

Figure 3.296: Automaton for the DECREASING_WIDTH_PROPER_PLATEAU con-
straint obtained by applying decoration Table 2.36 to the seed transducer of the
PROPER_PLATEAU pattern

## 3.149   DECREASING_WIDTH_STEADY_SEQUENCE

▶ ▷ ◁ ◀          **DESCRIPTION**          **AUTOMATON**

| | |
|---|---|
| **Origin** | Based on the STEADY_SEQUENCE pattern. |
| **Constraint** | DECREASING_WIDTH_STEADY_SEQUENCE(VARIABLES) |
| **Argument** | VARIABLES : collection(var−dvar) |
| **Restriction** | required(VARIABLES, var) |

**Purpose**

Succeeds if the values denoting the width of each occurrence of the STEADY_SEQUENCE pattern in the time-series given by the VARIABLES collection are decreasing.

An occurrence of the pattern STEADY_SEQUENCE is the *maximal* subsequence which matches the regular expression $=^+$.

Assume that the occurrence of the pattern STEADY_SEQUENCE starts at position $i$ and ends at position $j$. The feature WIDTH computes the value $j - i + 2$.

**Example**

$(\langle 4, 3, 2, 5, 5, 5, 3, 2, 4, 4, 6, 5, 3, 2, 2, 6 \rangle)$

Figure 3.297 provides an example where the DECREASING_WIDTH_STEADY_SEQUENCE $([4, 3, 2, 5, 5, 5, 3, 2, 4, 4, 6, 5, 3, 2, 2, 6])$ constraint holds.

Figure 3.297: Illustrating the DECREASING_WIDTH_STEADY_SEQUENCE constraint of the **Example** slot

| | |
|---|---|
| **Typical** | \|VARIABLES\| > 1 |

**Automaton**   Figure 3.298 depicts the automaton associated with the constraint DECREAS-ING_WIDTH_STEADY_SEQUENCE.

Figure 3.298: Automaton for the DECREASING_WIDTH_STEADY_SEQUENCE constraint obtained by applying decoration Table 2.36 to the seed transducer of the STEADY_SEQUENCE pattern

### 3.150 DECREASING_WIDTH_STRICTLY_DECREASING_SEQUENCE

▶ ▷ ◁ ◀          **DESCRIPTION**          **AUTOMATON**

| | |
|---|---|
| **Origin** | Based on the STRICTLY_DECREASING_SEQUENCE pattern. |
| **Constraint** | DECREASING_WIDTH_STRICTLY_DECREASING_SEQUENCE(VARIABLES) |
| **Argument** | VARIABLES : collection(var−dvar) |
| **Restriction** | required(VARIABLES, var) |

**Purpose**

Succeeds if the values denoting the width of each occurrence of the STRICTLY_DECREASING_SEQUENCE pattern in the time-series given by the VARIABLES collection are decreasing.

An occurrence of the pattern STRICTLY_DECREASING_SEQUENCE is the *maximal* sub-sequence which matches the regular expression $>^+$.

Assume that the occurrence of the pattern STRICTLY_DECREASING_SEQUENCE starts at position $i$ and ends at position $j$. The feature WIDTH computes the value $j - i + 2$.

**Example**

$(\langle 1, 1, 2, 2, 3, 2, 1, 5, 6, 6, 5, 4, 4, 6, 6, 1 \rangle)$

Figure 3.299 provides an example where the DECREAS-ING_WIDTH_STRICTLY_DECREASING_SEQUENCE ([1, 1, 2, 2, 3, 2, 1, 5, 6, 6, 5, 4, 4, 6, 6, 1]) constraint holds.

**Typical**

$|\texttt{VARIABLES}| > 1$
range(VARIABLES.var) $> 1$

Figure 3.299: Illustrating the DECREASING_WIDTH_STRICTLY_DECREASING_SEQUENCE constraint of the **Example** slot

**Automaton**          Figure 3.300 depicts the automaton associated with the constraint DECREAS-
ING_WIDTH_STRICTLY_DECREASING_SEQUENCE.

Figure 3.300: Automaton for the DECREAS-
ING_WIDTH_STRICTLY_DECREASING_SEQUENCE constraint obtained
by applying decoration Table 2.36 to the seed transducer of the
STRICTLY_DECREASING_SEQUENCE pattern

# 3.151 DECREASING_WIDTH_STRICTLY_INCREASING_SEQUENCE

▶ ▷ ◁ ◀          **DESCRIPTION**          **AUTOMATON**

**Origin**          Based on the STRICTLY_INCREASING_SEQUENCE pattern.

**Constraint**      DECREASING_WIDTH_STRICTLY_INCREASING_SEQUENCE(VARIABLES)

**Argument**        VARIABLES : collection(var−dvar)

**Restriction**     required(VARIABLES, var)

**Purpose**         Succeeds if the values denoting the width of each occurrence of the
                    STRICTLY_INCREASING_SEQUENCE pattern in the time-series given by the VARIABLES
                    collection are decreasing.
                    An occurrence of the pattern STRICTLY_INCREASING_SEQUENCE is the *maximal* sub-
                    sequence which matches the regular expression $<^+$.
                    Assume that the occurrence of the pattern STRICTLY_INCREASING_SEQUENCE starts at
                    position $i$ and ends at position $j$. The feature WIDTH computes the value $j - i + 2$.

**Example**         $(\langle 3, 4, 5, 5, 4, 3, 3, 1, 4, 3, 3, 2, 2, 1, 3, 1 \rangle)$

                    Figure 3.301 provides an example where the DECREAS-
                    ING_WIDTH_STRICTLY_INCREASING_SEQUENCE ([3, 4, 5, 5, 4, 3, 3, 1, 4, 3, 3, 2, 2, 1, 3, 1])
                    constraint holds.

**Typical**         $|\text{VARIABLES}| > 1$
                    range(VARIABLES.var) > 1

Figure 3.301: Illustrating the DECREASING_WIDTH_STRICTLY_INCREASING_SEQUENCE constraint of the **Example** slot

**Automaton**    Figure 3.302 depicts the automaton associated with the constraint DECREAS-
ING_WIDTH_STRICTLY_INCREASING_SEQUENCE.

Figure 3.302: Automaton for the DECREAS-
ING_WIDTH_STRICTLY_INCREASING_SEQUENCE constraint obtained
by applying decoration Table 2.36 to the seed transducer of the
STRICTLY_INCREASING_SEQUENCE pattern

## 3.152 DECREASING_WIDTH_SUMMIT

          **DESCRIPTION**          **AUTOMATON**

$(< | < (= | <)^* <) (>) | > (= | >)^* >)$

| | |
|---|---|
| **Origin** | Based on the SUMMIT pattern. |
| **Constraint** | DECREASING_WIDTH_SUMMIT(VARIABLES) |
| **Argument** | VARIABLES : collection(var−dvar) |
| **Restriction** | required(VARIABLES, var) |
| **Purpose** | Succeeds if the values denoting the width of each occurrence of the SUMMIT pattern in the time-series given by the VARIABLES collection are decreasing. An occurrence of the pattern SUMMIT is the *maximal* subsequence which matches the regular expression $(< | < (= | <)^* <) (>) | > (= | >)^* >)$. Assume that the occurrence of the pattern SUMMIT starts at position $i$ and ends at position $j$. The feature WIDTH computes the value $j - i$. |
| **Example** | $(\langle 1, 4, 4, 5, 4, 2, 6, 6, 2, 3, 5, 4, 1, 2, 4, 2 \rangle)$ |

Figure 3.303 provides an example where the DECREASING_WIDTH_SUMMIT $([1, 4, 4, 5, 4, 2, 6, 6, 2, 3, 5, 4, 1, 2, 4, 2])$ constraint holds.

Figure 3.303: Illustrating the DECREASING_WIDTH_SUMMIT constraint of the **Example** slot

**Typical**

$|\texttt{VARIABLES}| > 2$

$\texttt{range}(\texttt{VARIABLES.var}) > 1$

**Automaton**    Figure 3.304 depicts the automaton associated with the constraint DECREAS-
ING_WIDTH_SUMMIT.

Figure 3.304: Automaton for the DECREASING_WIDTH_SUMMIT constraint obtained
by applying decoration Table 2.36 to the seed transducer of the SUMMIT pattern (tran-
sition $u \rightarrow r$ has the same accumulator update as transition $r \rightarrow u$)

# 3.153   DECREASING_WIDTH_VALLEY

          **DESCRIPTION**              **AUTOMATON**

$$> (= | >)^* (< | =)^* <$$

| | |
|---|---|
| **Origin** | Based on the VALLEY pattern. |
| **Constraint** | DECREASING_WIDTH_VALLEY(VARIABLES) |
| **Argument** | VARIABLES : collection(var−dvar) |
| **Restriction** | required(VARIABLES, var) |
| **Purpose** | Succeeds if the values denoting the width of each occurrence of the VALLEY pattern in the time-series given by the VARIABLES collection are decreasing. <br><br> An occurrence of the pattern VALLEY is the *maximal* subsequence which matches the regular expression $> (= | >)^* (< | =)^* <$. <br><br> Assume that the occurrence of the pattern VALLEY starts at position $i$ and ends at position $j$. The feature WIDTH computes the value $j - i$. |
| **Example** | $(\langle 1, 3, 3, 7, 6, 5, 4, 6, 6, 5, 4, 3, 6, 2, 2, 7 \rangle)$ |

Figure 3.305 provides an example where the DECREASING_WIDTH_VALLEY ($[1, 3, 3, 7, 6, 5, 4, 6, 6, 5, 4, 3, 6, 2, 2, 7]$) constraint holds.

| | |
|---|---|
| **Typical** | $|\mathtt{VARIABLES}| > 2$ <br> range(VARIABLES.var) $> 1$ |

Figure 3.305: Illustrating the DECREASING_WIDTH_VALLEY constraint of the **Example** slot

**Automaton**         Figure 3.306 depicts the automaton associated with the constraint DECREAS-
                      ING_WIDTH_VALLEY.

Figure 3.306: Automaton for the DECREASING_WIDTH_VALLEY constraint obtained
by applying decoration Table 2.36 to the seed transducer of the VALLEY pattern

## 3.154  DECREASING_WIDTH_ZIGZAG

▶ ▷ ◁ ◀         **DESCRIPTION**          **AUTOMATON**

| | |
|---|---|
| **Origin** | Based on the ZIGZAG pattern. |
| **Constraint** | DECREASING_WIDTH_ZIGZAG(VARIABLES) |
| **Argument** | VARIABLES : collection(var−dvar) |
| **Restriction** | required(VARIABLES, var) |
| **Purpose** | Succeeds if the values denoting the width of each occurrence of the ZIGZAG pattern in the time-series given by the VARIABLES collection are decreasing. <br> An occurrence of the pattern ZIGZAG is the *maximal* subsequence which matches the regular expression $(<>)^+(< \mid <>) \mid (><)^+(> \mid ><)$. <br> Assume that the occurrence of the pattern ZIGZAG starts at position $i$ and ends at position $j$. The feature WIDTH computes the value $j - i$. |
| **Example** | $(\langle 6, 5, 7, 4, 5, 1, 1, 6, 3, 6, 4, 3, 5, 2, 7, 7 \rangle)$ |

Figure 3.307 provides an example where the DECREASING_WIDTH_ZIGZAG ([6, 5, 7, 4, 5, 1, 1, 6, 3, 6, 4, 3, 5, 2, 7, 7]) constraint holds.

| | |
|---|---|
| **Typical** | $\mid$VARIABLES$\mid > 3$ <br> range(VARIABLES.var) $> 1$ |

Figure 3.307: Illustrating the DECREASING_WIDTH_ZIGZAG constraint of the **Example** slot

**Automaton**  Figure 3.308 depicts the automaton associated with the constraint DECREAS-ING_WIDTH_ZIGZAG.

Figure 3.308: Automaton for the DECREASING_WIDTH_ZIGZAG constraint obtained by applying decoration Table 2.36 to the seed transducer of the ZIGZAG pattern; (1) missing transitions from $a, b, c, d, e, f$ to $s$ are labelled by $=$; (2) on transitions from $b, c, e, f$ to $s$ the accumulator $D$ is reset to its initial value; (3) on transitions from $c, f$ to $s$ the accumulator $F$ is reset to $C$, and the accumulator $R$ is updated wrt $C$ and $F$

# 3.155 INCREASING_HEIGHT_DECREASING_TERRACE

► ▷ ◁ ◄        **DESCRIPTION**              **AUTOMATON**

| | |
|---|---|
| **Origin** | Based on the DECREASING_TERRACE pattern. |
| **Constraint** | INCREASING_HEIGHT_DECREASING_TERRACE(VARIABLES) |
| **Argument** | VARIABLES : collection(var−dvar) |
| **Restriction** | required(VARIABLES, var) |
| **Purpose** | Succeeds if the minima of the values in each occurrence of the DECREASING_TERRACE pattern in the time-series given by the VARIABLES collection are increasing. An occurrence of the pattern DECREASING_TERRACE is the *maximal* subsequence which matches the regular expression $>=^+>$. Assume that the occurrence of the pattern DECREASING_TERRACE starts at position $i$ and ends at position $j$. The feature MIN computes the minimum of the values from index $i+1$ to index $j$. |
| **Example** | $(\langle 5, 2, 2, 1, 5, 4, 3, 3, 2, 4, 4, 6, 5, 5, 5, 4\rangle)$ |

Figure 3.309 provides an example where the INCREAS-ING_HEIGHT_DECREASING_TERRACE ([5, 2, 2, 1, 5, 4, 3, 3, 2, 4, 4, 6, 5, 5, 5, 4]) constraint holds.

| | |
|---|---|
| **Typical** | $|\texttt{VARIABLES}| > 3$ |
| | range(VARIABLES.var) > 2 |

Figure 3.309: Illustrating the INCREASING_HEIGHT_DECREASING_TERRACE constraint of the **Example** slot

**Automaton**    Figure 3.310 depicts the automaton associated with the constraint INCREAS-ING_HEIGHT_DECREASING_TERRACE.

Figure 3.310: Automaton for the INCREASING_HEIGHT_DECREASING_TERRACE constraint obtained by applying decoration Table 2.36 to the seed transducer of the DECREASING_TERRACE pattern

## 3.156   INCREASING_HEIGHT_INCREASING_TERRACE

**Origin**            Based on the INCREASING_TERRACE pattern.

**Constraint**        INCREASING_HEIGHT_INCREASING_TERRACE(VARIABLES)

**Argument**          VARIABLES : collection(var−dvar)

**Restriction**       required(VARIABLES, var)

**Purpose**           Succeeds if the minima of the values in each occurrence of the INCREASING_TERRACE
                      pattern in the time-series given by the VARIABLES collection are increasing.
                      An occurrence of the pattern INCREASING_TERRACE is the *maximal* subsequence which
                      matches the regular expression $<=^+<$.
                      Assume that the occurrence of the pattern INCREASING_TERRACE starts at position $i$
                      and ends at position $j$. The feature MIN computes the minimum of the values from index
                      $i + 1$ to index $j$.

**Example**           $(\langle 1, 2, 2, 4, 5, 5, 6, 6, 4, 4, 2, 5, 5, 5, 7, 7 \rangle)$

                      Figure    3.311    provides    an    example    where    the    INCREAS-
                      ING_HEIGHT_INCREASING_TERRACE $([1, 2, 2, 4, 5, 5, 6, 6, 4, 4, 2, 5, 5, 5, 7, 7])$ constraint
                      holds.

**Typical**           $|\texttt{VARIABLES}| > 3$
                      range(VARIABLES.var) $> 2$

Figure 3.311: Illustrating the INCREASING_HEIGHT_INCREASING_TERRACE constraint of the **Example** slot

**Automaton**     Figure 3.312 depicts the automaton associated with the constraint INCREAS-
ING_HEIGHT_INCREASING_TERRACE.

Figure 3.312: Automaton for the INCREASING_HEIGHT_INCREASING_TERRACE con-
straint obtained by applying decoration Table 2.36 to the seed transducer of the IN-
CREASING_TERRACE pattern

## 3.157 INCREASING_HEIGHT_PLAIN

▶ ▷ ◁ ◀        DESCRIPTION          AUTOMATON

| | |
|---|---|
| **Origin** | Based on the PLAIN pattern. |
| **Constraint** | INCREASING_HEIGHT_PLAIN(VARIABLES) |
| **Argument** | VARIABLES : collection(var−dvar) |
| **Restriction** | required(VARIABLES, var) |
| **Purpose** | Succeeds if the minima of the values in each occurrence of the PLAIN pattern in the time-series given by the VARIABLES collection are increasing.<br>An occurrence of the pattern PLAIN is the *maximal* subsequence which matches the regular expression $> =^* <$.<br>Assume that the occurrence of the pattern PLAIN starts at position $i$ and ends at position $j$. The feature MIN computes the minimum of the values from index $i+1$ to index $j$. |
| **Example** | $(\langle 3, 6, 6, 3, 4, 5, 5, 4, 6, 6, 7, 5, 5, 6, 3, 2 \rangle)$ |

Figure 3.313 provides an example where the INCREASING_HEIGHT_PLAIN $([3, 6, 6, 3, 4, 5, 5, 4, 6, 6, 7, 5, 5, 6, 3, 2])$ constraint holds.

Figure 3.313: Illustrating the INCREASING_HEIGHT_PLAIN constraint of the **Example** slot

**Typical**     $|\mathtt{VARIABLES}| > 2$
                $\mathtt{range}(\mathtt{VARIABLES.var}) > 1$

**Automaton**        Figure 3.314 depicts the automaton associated with the constraint INCREAS-
                     ING_HEIGHT_PLAIN.

Figure 3.314: Automaton for the INCREASING_HEIGHT_PLAIN constraint obtained by
applying decoration Table 2.36 to the seed transducer of the PLAIN pattern

# 3.158 INCREASING_HEIGHT_PLATEAU

► ▷ ◁ ◄          **DESCRIPTION**          **AUTOMATON**

| | |
|---|---|
| **Origin** | Based on the PLATEAU pattern. |
| **Constraint** | INCREASING_HEIGHT_PLATEAU(VARIABLES) |
| **Argument** | VARIABLES : collection(var−dvar) |
| **Restriction** | required(VARIABLES, var) |
| **Purpose** | Succeeds if the minima of the values in each occurrence of the PLATEAU pattern in the time-series given by the VARIABLES collection are increasing.<br><br>An occurrence of the pattern PLATEAU is the *maximal* subsequence which matches the regular expression $< =^* >$.<br><br>Assume that the occurrence of the pattern PLATEAU starts at position $i$ and ends at position $j$. The feature MIN computes the minimum of the values from index $i + 1$ to index $j$. |
| **Example** | $(\langle 7, 5, 2, 3, 1, 2, 2, 4, 3, 3, 4, 5, 5, 2, 2, 5 \rangle)$ |

Figure 3.315 provides an example where the INCREASING_HEIGHT_PLATEAU $([7, 5, 2, 3, 1, 2, 2, 4, 3, 3, 4, 5, 5, 2, 2, 5])$ constraint holds.

| | |
|---|---|
| **Typical** | $|\texttt{VARIABLES}| > 2$<br>range(VARIABLES.var) $> 1$ |

Figure 3.315: Illustrating the INCREASING_HEIGHT_PLATEAU constraint of the **Example** slot

**Automaton**        Figure 3.316 depicts the automaton associated with the constraint INCREAS-
ING_HEIGHT_PLATEAU.

Figure 3.316: Automaton for the INCREASING_HEIGHT_PLATEAU constraint obtained
by applying decoration Table 2.36 to the seed transducer of the PLATEAU pattern

# 3.159   INCREASING_HEIGHT_PROPER_PLAIN

▶ ▷ ◁ ◀     **DESCRIPTION**          **AUTOMATON**

**Origin**          Based on the PROPER_PLAIN pattern.

**Constraint**      INCREASING_HEIGHT_PROPER_PLAIN(VARIABLES)

**Argument**        VARIABLES : collection(var−dvar)

**Restriction**     required(VARIABLES, var)

**Purpose**         Succeeds if the minima of the values in each occurrence of the PROPER_PLAIN pattern
                    in the time-series given by the VARIABLES collection are increasing.
                    An occurrence of the pattern PROPER_PLAIN is the *maximal* subsequence which matches
                    the regular expression $> =^+ <$.
                    Assume that the occurrence of the pattern PROPER_PLAIN starts at position $i$ and ends
                    at position $j$. The feature MIN computes the minimum of the values from index $i + 1$ to
                    index $j$.

**Example**         $(\langle 5, 3, 3, 5, 6, 5, 4, 4, 7, 3, 6, 5, 5, 5, 7, 2 \rangle)$

                    Figure 3.317 provides an example where the INCREASING_HEIGHT_PROPER_PLAIN
                    $([5, 3, 3, 5, 6, 5, 4, 4, 7, 3, 6, 5, 5, 5, 7, 2])$ constraint holds.

**Typical**         $|\texttt{VARIABLES}| > 3$
                    range(VARIABLES.var) $> 1$

Figure 3.317: Illustrating the INCREASING_HEIGHT_PROPER_PLAIN constraint of the **Example** slot

**Automaton**          Figure 3.318 depicts the automaton associated with the constraint INCREAS-
ING_HEIGHT_PROPER_PLAIN.

Figure 3.318: Automaton for the INCREASING_HEIGHT_PROPER_PLAIN constraint ob-
tained by applying decoration Table 2.36 to the seed transducer of the PROPER_PLAIN
pattern

# 3.160 INCREASING_HEIGHT_PROPER_PLATEAU

▶ ▷ ◁ ◀        **DESCRIPTION**            **AUTOMATON**

**Origin**          Based on the PROPER_PLATEAU pattern.

**Constraint**      INCREASING_HEIGHT_PROPER_PLATEAU(VARIABLES)

**Argument**        VARIABLES : collection(var−dvar)

**Restriction**     required(VARIABLES, var)

**Purpose**         Succeeds if the minima of the values in each occurrence of the PROPER_PLATEAU pattern
                    in the time-series given by the VARIABLES collection are increasing.

                    An occurrence of the pattern PROPER_PLATEAU is the *maximal* subsequence which
                    matches the regular expression $< =^+ >$.

                    Assume that the occurrence of the pattern PROPER_PLATEAU starts at position $i$ and ends
                    at position $j$. The feature MIN computes the minimum of the values from index $i + 1$ to
                    index $j$.

**Example**         $(\langle 7, 1, 3, 3, 2, 5, 1, 4, 4, 3, 2, 3, 5, 5, 5, 3 \rangle)$

                    Figure 3.319 provides an example where the INCREASING_HEIGHT_PROPER_PLATEAU
                    $([7, 1, 3, 3, 2, 5, 1, 4, 4, 3, 2, 3, 5, 5, 5, 3])$ constraint holds.

**Typical**         $|\mathtt{VARIABLES}| > 3$
                    range(VARIABLES.var) $> 1$

Figure 3.319: Illustrating the INCREASING_HEIGHT_PROPER_PLATEAU constraint of the **Example** slot

**Automaton**       Figure 3.320 depicts the automaton associated with the constraint INCREAS-
ING_HEIGHT_PROPER_PLATEAU.

Figure 3.320: Automaton for the INCREASING_HEIGHT_PROPER_PLATEAU con-
straint obtained by applying decoration Table 2.36 to the seed transducer of the
PROPER_PLATEAU pattern

# 3.161 INCREASING_HEIGHT_STEADY

▶ ▷ ◁ ◀          **DESCRIPTION**          **AUTOMATON**

| | |
|---|---|
| **Origin** | Based on the STEADY pattern. |
| **Constraint** | INCREASING_HEIGHT_STEADY(VARIABLES) |
| **Argument** | VARIABLES : collection(var−dvar) |
| **Restriction** | required(VARIABLES, var) |
| **Purpose** | Succeeds if the minima of the values in each occurrence of the STEADY pattern in the time-series given by the VARIABLES collection are increasing.<br>An occurrence of the pattern STEADY is the subsequence which matches the regular expression =.<br>Assume that the occurrence of the pattern STEADY starts at position $i$ and ends at position $j$. The feature MIN computes the minimum of the values from index $i$ to index $j + 1$. |
| **Example** | $(\langle 1, 1, 7, 3, 3, 5, 5, 5, 6, 5, 5, 5, 7, 2, 6, 6 \rangle)$ |

Figure 3.321 provides an example where the INCREASING_HEIGHT_STEADY $([1, 1, 7, 3, 3, 5, 5, 5, 6, 5, 5, 5, 7, 2, 6, 6])$ constraint holds.

| | |
|---|---|
| **Typical** | $|\texttt{VARIABLES}| > 1$ |

Figure 3.321: Illustrating the INCREASING_HEIGHT_STEADY constraint of the **Example** slot

**Automaton**    Figure 3.322 depicts the automaton associated with the constraint INCREAS-ING_HEIGHT_STEADY.

$$\left\{ \begin{array}{l} C \leftarrow -\infty \\ D \leftarrow +\infty \\ F \leftarrow -\infty \\ R \leftarrow 1 \end{array} \right\}$$

$$\overset{=}{\left\{ \begin{array}{l} C \leftarrow \min(\min(D, \mathtt{VAR}_i), \mathtt{VAR}_{i+1}) \\ D \leftarrow +\infty \\ F \leftarrow \min(\min(D, \mathtt{VAR}_i), \mathtt{VAR}_{i+1}) \\ R \leftarrow R \wedge (F \leq \min(\min(D, \mathtt{VAR}_i), \mathtt{VAR}_{i+1})) \end{array} \right\}}$$

$$\boxed{R \wedge (F \leq C)}$$

Figure 3.322: Automaton for the INCREASING_HEIGHT_STEADY constraint obtained by applying decoration Table 2.36 to the seed transducer of the STEADY pattern

# 3.162 INCREASING_HEIGHT_STEADY_SEQUENCE

▶ ▷ ◁ ◀          **DESCRIPTION**          **AUTOMATON**

| | |
|---|---|
| **Origin** | Based on the STEADY_SEQUENCE pattern. |
| **Constraint** | INCREASING_HEIGHT_STEADY_SEQUENCE(VARIABLES) |
| **Argument** | VARIABLES : collection(var−dvar) |
| **Restriction** | required(VARIABLES, var) |
| **Purpose** | Succeeds if the minima of the values in each occurrence of the STEADY_SEQUENCE pattern in the time-series given by the VARIABLES collection are increasing. An occurrence of the pattern STEADY_SEQUENCE is the *maximal* subsequence which matches the regular expression $=^+$. Assume that the occurrence of the pattern STEADY_SEQUENCE starts at position $i$ and ends at position $j$. The feature MIN computes the minimum of the values from index $i$ to index $j + 1$. |
| **Example** | $(\langle 6, 2, 2, 3, 5, 6, 4, 4, 2, 3, 5, 5, 5, 2, 3, 4 \rangle)$ |

Figure 3.323 provides an example where the INCREASING_HEIGHT_STEADY_SEQUENCE $([6, 2, 2, 3, 5, 6, 4, 4, 2, 3, 5, 5, 5, 2, 3, 4])$ constraint holds.

Figure 3.323: Illustrating the INCREASING_HEIGHT_STEADY_SEQUENCE constraint of the **Example** slot

**Typical**    $|VARIABLES| > 1$

**Automaton**          Figure 3.324 depicts the automaton associated with the constraint INCREAS-
ING_HEIGHT_STEADY_SEQUENCE.

Figure 3.324:  Automaton for the INCREASING_HEIGHT_STEADY_SEQUENCE con-
straint obtained by applying decoration Table 2.36 to the seed transducer of the
STEADY_SEQUENCE pattern

# 3.163   INCREASING_MAX_BUMP_ON_DECREASING_SEQUENCE

▶ ▷ ◁ ◀        **DESCRIPTION**        **AUTOMATON**

| | |
|---|---|
| **Origin** | Based on the BUMP_ON_DECREASING_SEQUENCE pattern. |
| **Constraint** | INCREASING_MAX_BUMP_ON_DECREASING_SEQUENCE(VARIABLES) |
| **Argument** | VARIABLES : collection(var−dvar) |
| **Restriction** | required(VARIABLES, var) |

**Purpose**

Succeeds if the maxima of the values in each occurrence of the BUMP_ON_DECREASING_SEQUENCE pattern in the time-series given by the VARIABLES collection are increasing.

An occurrence of the pattern BUMP_ON_DECREASING_SEQUENCE is the subsequence which matches the regular expression '>><>>'.

Assume that the occurrence of the pattern BUMP_ON_DECREASING_SEQUENCE starts at position $i$ and ends at position $j$. The feature MAX computes the maximum of the values from index $i + 2$ to index $j$.

**Example**

$$(\langle 7, 5, 4, 2, 5, 4, 3, 3, 5, 7, 6, 5, 6, 5, 4, 1 \rangle)$$

Figure 3.325 provides an example where the INCREAS­ING_MAX_BUMP_ON_DECREASING_SEQUENCE ([7, 5, 4, 2, 5, 4, 3, 3, 5, 7, 6, 5, 6, 5, 4, 1]) constraint holds.

**Typical**

$|VARIABLES| > 5$
range(VARIABLES.var) > 2

Figure 3.325: Illustrating the INCREASING_MAX_BUMP_ON_DECREASING_SEQUENCE constraint of the **Example** slot

**Automaton**　　Figure 3.326 depicts the automaton associated with the constraint INCREAS-ING_MAX_BUMP_ON_DECREASING_SEQUENCE.

Figure 3.326: Automaton for the INCREASING_MAX_BUMP_ON_DECREASING_SEQUENCE constraint obtained by applying decoration Table 2.36 to the seed transducer of the BUMP_ON_DECREASING_SEQUENCE pattern

## 3.164   INCREASING_MAX_DECREASING

▶ ▷ ◁ ◀        **DESCRIPTION**                **AUTOMATON**

| | |
|---|---|
| **Origin** | Based on the DECREASING pattern. |
| **Constraint** | INCREASING_MAX_DECREASING(VARIABLES) |
| **Argument** | VARIABLES : collection(var−dvar) |
| **Restriction** | required(VARIABLES, var) |
| **Purpose** | Succeeds if the maxima of the values in each occurrence of the DECREASING pattern in the time-series given by the VARIABLES collection are increasing. An occurrence of the pattern DECREASING is the subsequence which matches the regular expression >. Assume that the occurrence of the pattern DECREASING starts at position $i$ and ends at position $j$. The feature MAX computes the maximum of the values from index $i$ to index $j + 1$. |
| **Example** | $(\langle 2, 1, 1, 1, 3, 2, 2, 3, 1, 1, 5, 3, 4, 5, 6, 5 \rangle)$ |

Figure 3.327 provides an example where the INCREASING_MAX_DECREASING $([2, 1, 1, 1, 3, 2, 2, 3, 1, 1, 5, 3, 4, 5, 6, 5])$ constraint holds.

| | |
|---|---|
| **Typical** | $\lvert$VARIABLES$\rvert > 1$ |
| | range(VARIABLES.var) $> 1$ |

Figure 3.327: Illustrating the INCREASING_MAX_DECREASING constraint of the **Example** slot

**Automaton** Figure 3.328 depicts the automaton associated with the constraint INCREAS-ING_MAX_DECREASING.

$$\left\{ \begin{array}{l} C \leftarrow -\infty \\ D \leftarrow -\infty \\ F \leftarrow -\infty \\ R \leftarrow 1 \end{array} \right\}$$

$$\left\{ \begin{array}{l} C \leftarrow \max(\max(D, \text{VAR}_i), \text{VAR}_{i+1}) \\ D \leftarrow -\infty \\ F \leftarrow \max(\max(D, \text{VAR}_i), \text{VAR}_{i+1}) \\ R \leftarrow R \wedge (F \leq \max(\max(D, \text{VAR}_i), \text{VAR}_{i+1})) \end{array} \right\}$$

$>$

$\geq \wedge \leq$  $s$  $\leq$

$R \wedge (F \leq C)$

Figure 3.328: Automaton for the INCREASING_MAX_DECREASING constraint obtained by applying decoration Table 2.36 to the seed transducer of the DECREASING pattern

# 3.165   INCREASING_MAX_DECREASING_SEQUENCE

▶ ▷ ◁ ◀          **DESCRIPTION**          **AUTOMATON**

**Origin**          Based on the DECREASING_SEQUENCE pattern.

**Constraint**      INCREASING_MAX_DECREASING_SEQUENCE(VARIABLES)

**Argument**        VARIABLES : collection(var−dvar)

**Restriction**     required(VARIABLES, var)

**Purpose**         Succeeds if the maxima of the values in each occurrence of the DECREAS-
                    ING_SEQUENCE pattern in the time-series given by the VARIABLES collection are in-
                    creasing.
                    An occurrence of the pattern DECREASING_SEQUENCE is the *maximal* subsequence
                    which matches the regular expression $> (> | =)^* > | >$.
                    Assume that the occurrence of the pattern DECREASING_SEQUENCE starts at position
                    $i$ and ends at position $j$. The feature MAX computes the maximum of the values from
                    index $i$ to index $j + 1$.

**Example**         $(\langle 4, 3, 2, 2, 3, 4, 4, 6, 5, 5, 3, 4, 4, 6, 4, 6 \rangle)$

                    Figure 3.329 provides an example where the INCREAS-
                    ING_MAX_DECREASING_SEQUENCE ([4, 3, 2, 2, 3, 4, 4, 6, 5, 5, 3, 4, 4, 6, 4, 6]) constraint
                    holds.

**Typical**         $|\text{VARIABLES}| > 1$
                    range(VARIABLES.var) > 1

Figure 3.329: Illustrating the INCREASING_MAX_DECREASING_SEQUENCE constraint of the **Example** slot

**Automaton**     Figure 3.330 depicts the automaton associated with the constraint INCREAS-
                  ING_MAX_DECREASING_SEQUENCE.

Figure 3.330: Automaton for the INCREASING_MAX_DECREASING_SEQUENCE con-
straint obtained by applying decoration Table 2.36 to the seed transducer of the DE-
CREASING_SEQUENCE pattern

# 3.166   INCREASING_MAX_DIP_ON_INCREASING_SEQUENCE

▶ ▷ ◁ ◀        **DESCRIPTION**          **AUTOMATON**

| | |
|---|---|
| **Origin** | Based on the DIP_ON_INCREASING_SEQUENCE pattern. |
| **Constraint** | INCREASING_MAX_DIP_ON_INCREASING_SEQUENCE(VARIABLES) |
| **Argument** | VARIABLES : collection(var−dvar) |
| **Restriction** | required(VARIABLES, var) |

**Purpose**

Succeeds if the maxima of the values in each occurrence of the DIP_ON_INCREASING_SEQUENCE pattern in the time-series given by the VARIABLES collection are increasing.

An occurrence of the pattern DIP_ON_INCREASING_SEQUENCE is the subsequence which matches the regular expression $<<><<$.

Assume that the occurrence of the pattern DIP_ON_INCREASING_SEQUENCE starts at position $i$ and ends at position $j$. The feature MAX computes the maximum of the values from index $i + 2$ to index $j$.

**Example**

$$(\langle 1, 2, 3, 2, 3, 4, 6, 5, 1, 4, 5, 6, 0, 2, 4, 4 \rangle)$$

Figure 3.331 provides an example where the INCREASING_MAX_DIP_ON_INCREASING_SEQUENCE ($[1, 2, 3, 2, 3, 4, 6, 5, 1, 4, 5, 6, 0, 2, 4, 4]$) constraint holds.

**Typical**

$|\text{VARIABLES}| > 5$

$\text{range}(\text{VARIABLES.var}) > 2$

Figure 3.331: Illustrating the INCREASING_MAX_DIP_ON_INCREASING_SEQUENCE constraint of the **Example** slot

**Automaton**     Figure 3.332 depicts the automaton associated with the constraint INCREAS-
ING_MAX_DIP_ON_INCREASING_SEQUENCE.

Figure 3.332: Automaton for the INCREASING_MAX_DIP_ON_INCREASING_SEQUENCE
constraint obtained by applying decoration Table 2.36 to the seed transducer of the
DIP_ON_INCREASING_SEQUENCE pattern

## 3.167  INCREASING_MAX_INCREASING

          **DESCRIPTION**          **AUTOMATON**

| | |
|---|---|
| **Origin** | Based on the INCREASING pattern. |
| **Constraint** | INCREASING_MAX_INCREASING(VARIABLES) |
| **Argument** | VARIABLES : collection(var−dvar) |
| **Restriction** | required(VARIABLES, var) |
| **Purpose** | Succeeds if the maxima of the values in each occurrence of the INCREASING pattern in the time-series given by the VARIABLES collection are increasing. An occurrence of the pattern INCREASING is the subsequence which matches the regular expression $<$. Assume that the occurrence of the pattern INCREASING starts at position $i$ and ends at position $j$. The feature MAX computes the maximum of the values from index $i$ to index $j+1$. |
| **Example** | $(\langle 2, 1, 2, 0, 2, 3, 1, 1, 3, 4, 2, 1, 1, 0, 6, 7 \rangle)$ |

Figure 3.333 provides an example where the INCREASING_MAX_INCREASING $([2, 1, 2, 0, 2, 3, 1, 1, 3, 4, 2, 1, 1, 0, 6, 7])$ constraint holds.

| | |
|---|---|
| **Typical** | $|\texttt{VARIABLES}| > 1$ <br> range(VARIABLES.var) $> 1$ |

Figure 3.333: Illustrating the INCREASING_MAX_INCREASING constraint of the **Example** slot

**Automaton**    Figure 3.334 depicts the automaton associated with the constraint INCREAS-
ING_MAX_INCREASING.

$$\left\{ \begin{array}{l} C \leftarrow -\infty \\ D \leftarrow -\infty \\ F \leftarrow -\infty \\ R \leftarrow 1 \end{array} \right\}$$

$$\left\{ \begin{array}{l} C \leftarrow \max(\max(D, \mathtt{VAR}_i), \mathtt{VAR}_{i+1}) \\ D \leftarrow -\infty \\ F \leftarrow \max(\max(D, \mathtt{VAR}_i), \mathtt{VAR}_{i+1}) \\ R \leftarrow R \wedge (F \leq \max(\max(D, \mathtt{VAR}_i), \mathtt{VAR}_{i+1})) \end{array} \right\} \quad <$$

$$\geq \qquad s \qquad \geq$$

$$\boxed{R \wedge (F \leq C)}$$

Figure 3.334: Automaton for the INCREASING_MAX_INCREASING constraint obtained
by applying decoration Table 2.36 to the seed transducer of the INCREASING pattern

# 3.168   INCREASING_MAX_INCREASING_SEQUENCE

▶ ▷ ◁ ◀        **DESCRIPTION**          **AUTOMATON**

**Origin**              Based on the INCREASING_SEQUENCE pattern.

**Constraint**          INCREASING_MAX_INCREASING_SEQUENCE(VARIABLES)

**Argument**            VARIABLES   :   collection(var−dvar)

**Restriction**         required(VARIABLES, var)

**Purpose**             Succeeds if the maxima of the values in each occurrence of the INCREASING_SEQUENCE
                        pattern in the time-series given by the VARIABLES collection are increasing.
                        An occurrence of the pattern INCREASING_SEQUENCE is the *maximal* subsequence
                        which matches the regular expression $<\ (<\ |\ =)^*\ <\ |\ <$.
                        Assume that the occurrence of the pattern INCREASING_SEQUENCE starts at position $i$
                        and ends at position $j$. The feature MAX computes the maximum of the values from
                        index $i$ to index $j + 1$.

**Example**             $(\langle 4, 1, 4, 5, 5, 4, 4, 2, 3, 3, 4, 5, 6, 6, 5, 6 \rangle)$

                        Figure 3.335 provides an example where the INCREASING_MAX_INCREASING_SEQUENCE
                        $([4, 1, 4, 5, 5, 4, 4, 2, 3, 3, 4, 5, 6, 6, 5, 6])$ constraint holds.

**Typical**             $|\text{VARIABLES}| > 1$
                        range(VARIABLES.var) > 1

Figure 3.335: Illustrating the INCREASING_MAX_INCREASING_SEQUENCE constraint of the **Example** slot

**Automaton**  Figure 3.336 depicts the automaton associated with the constraint INCREAS-
ING_MAX_INCREASING_SEQUENCE.

Figure 3.336: Automaton for the INCREASING_MAX_INCREASING_SEQUENCE con-
straint obtained by applying decoration Table 2.36 to the seed transducer of the IN-
CREASING_SEQUENCE pattern

## 3.169 INCREASING_MAX_INFLEXION

▶ ▷ ◁ ◀          **DESCRIPTION**          **AUTOMATON**

$< (< | =)^* > | > (> | =)^* <$

| | |
|---|---|
| **Origin** | Based on the INFLEXION pattern. |
| **Constraint** | INCREASING_MAX_INFLEXION(VARIABLES) |
| **Argument** | VARIABLES : collection(var−dvar) |
| **Restriction** | required(VARIABLES, var) |
| **Purpose** | Succeeds if the maxima of the values in each occurrence of the INFLEXION pattern in the time-series given by the VARIABLES collection are increasing. |
| | An occurrence of the pattern INFLEXION is the *maximal* subsequence which matches the regular expression $< (< | =)^* > | > (> | =)^* <$. |
| | Assume that the occurrence of the pattern INFLEXION starts at position $i$ and ends at position $j$. The feature MAX computes the maximum of the values from index $i + 1$ to index $j$. |
| **Example** | $(\langle 5, 5, 2, 2, 1, 1, 2, 2, 3, 4, 3 \rangle)$ |

Figure 3.337 provides an example where the INCREASING_MAX_INFLEXION
$([5, 5, 2, 2, 1, 1, 2, 2, 3, 4, 3])$ constraint holds.

Figure 3.337: Illustrating the INCREASING_MAX_INFLEXION constraint of the **Example** slot

**Typical**  $|\mathtt{VARIABLES}| > 2$
$\mathtt{range}(\mathtt{VARIABLES.var}) > 1$

**Automaton**    Figure 3.338 depicts the automaton associated with the constraint INCREAS-
ING_MAX_INFLEXION.

Figure 3.338: Automaton for the INCREASING_MAX_INFLEXION constraint obtained
by applying decoration Table 2.36 to the seed transducer of the INFLEXION pattern
(transition $r \rightarrow t$ has the same accumulators updates as transition $t \rightarrow r$)

# 3.170 INCREASING_MAX_PEAK

▶ ▷ ◁ ◀          DESCRIPTION          AUTOMATON

| | |
|---|---|
| **Origin** | Based on the PEAK pattern. |
| **Constraint** | INCREASING_MAX_PEAK(VARIABLES) |
| **Argument** | VARIABLES : collection(var−dvar) |
| **Restriction** | required(VARIABLES, var) |
| **Purpose** | Succeeds if the maxima of the values in each occurrence of the PEAK pattern in the time-series given by the VARIABLES collection are increasing. |
| | An occurrence of the pattern PEAK is the *maximal* subsequence which matches the regular expression $< (= | <)^* (>|=)^* >$. |
| | Assume that the occurrence of the pattern PEAK starts at position $i$ and ends at position $j$. The feature MAX computes the maximum of the values from index $i + 1$ to index $j$. |
| **Example** | $(\langle 7, 5, 5, 1, 2, 3, 2, 2, 3, 4, 5, 2, 6, 6, 6, 1 \rangle)$ |

Figure 3.339 provides an example where the INCREASING_MAX_PEAK $([7, 5, 5, 1, 2, 3, 2, 2, 3, 4, 5, 2, 6, 6, 6, 1])$ constraint holds.

Figure 3.339: Illustrating the INCREASING_MAX_PEAK constraint of the **Example** slot

**Typical**    $|\texttt{VARIABLES}| > 2$
$\texttt{range}(\texttt{VARIABLES.var}) > 1$

**Automaton**     Figure 3.340 depicts the automaton associated with the constraint INCREAS-ING_MAX_PEAK.

Figure 3.340: Automaton for the INCREASING_MAX_PEAK constraint obtained by applying decoration Table 2.36 to the seed transducer of the PEAK pattern

# 3.171 INCREASING_MAX_STRICTLY_DECREASING_SEQUENCE

▶ ▷ ◁ ◀        **DESCRIPTION**              **AUTOMATON**

**Origin**          Based on the STRICTLY_DECREASING_SEQUENCE pattern.

**Constraint**      INCREASING_MAX_STRICTLY_DECREASING_SEQUENCE(VARIABLES)

**Argument**        VARIABLES : collection(var−dvar)

**Restriction**     required(VARIABLES, var)

**Purpose**         Succeeds if the maxima of the values in each occurrence of the
                    STRICTLY_DECREASING_SEQUENCE pattern in the time-series given by the VARIABLES
                    collection are increasing.
                    An occurrence of the pattern STRICTLY_DECREASING_SEQUENCE is the *maximal* sub-
                    sequence which matches the regular expression $>^+$.
                    Assume that the occurrence of the pattern STRICTLY_DECREASING_SEQUENCE starts at
                    position $i$ and ends at position $j$. The feature MAX computes the maximum of the values
                    from index $i$ to index $j + 1$.

**Example**         $(\langle 2, 3, 2, 2, 3, 4, 5, 5, 6, 5, 4, 3, 2, 4, 7, 4 \rangle)$

                    Figure 3.341 provides an example where the INCREAS-
                    ING_MAX_STRICTLY_DECREASING_SEQUENCE $([2, 3, 2, 2, 3, 4, 5, 5, 6, 5, 4, 3, 2, 4, 7, 4])$
                    constraint holds.

**Typical**         $|\texttt{VARIABLES}| > 1$
                    range(VARIABLES.var) $> 1$

Figure 3.341: Illustrating the INCREASING_MAX_STRICTLY_DECREASING_SEQUENCE constraint of the **Example** slot

**Automaton**      Figure 3.342 depicts the automaton associated with the constraint INCREAS-
ING_MAX_STRICTLY_DECREASING_SEQUENCE.

$$\left\{ \begin{array}{l} C \leftarrow -\infty \\ D \leftarrow -\infty \\ F \leftarrow -\infty \\ R \leftarrow 1 \end{array} \right\}$$

$\leq s$    $\leq$

$$\left\{ \begin{array}{l} D \leftarrow -\infty \\ F \leftarrow C \\ R \leftarrow R \wedge (F \leq C) \end{array} \right\}$$

$R \wedge (F \leq C)$

$$\left\{ \begin{array}{l} C \leftarrow \max(\max(D, \mathtt{VAR}_i), \mathtt{VAR}_{i+1}) \\ D \leftarrow -\infty \end{array} \right\}$$

$> r$

$$\left\{ \begin{array}{l} C \leftarrow \max(C, \max(D, \mathtt{VAR}_{i+1})) \\ D \leftarrow -\infty \end{array} \right\}$$

Figure 3.342: Automaton for the INCREASING_MAX_STRICTLY_DECREASING_SEQUENCE
constraint obtained by applying decoration Table 2.36 to the seed transducer of the
STRICTLY_DECREASING_SEQUENCE pattern

## 3.172  INCREASING_MAX_STRICTLY_INCREASING_SEQUENCE

▶ ▷ ◁ ◀          **DESCRIPTION**          **AUTOMATON**

| | |
|---|---|
| **Origin** | Based on the STRICTLY_INCREASING_SEQUENCE pattern. |
| **Constraint** | INCREASING_MAX_STRICTLY_INCREASING_SEQUENCE(VARIABLES) |
| **Argument** | VARIABLES : collection(var−dvar) |
| **Restriction** | required(VARIABLES, var) |

**Purpose**

Succeeds if the maxima of the values in each occurrence of the STRICTLY_INCREASING_SEQUENCE pattern in the time-series given by the VARIABLES collection are increasing.

An occurrence of the pattern STRICTLY_INCREASING_SEQUENCE is the *maximal* sub-sequence which matches the regular expression $<^{+}$.

Assume that the occurrence of the pattern STRICTLY_INCREASING_SEQUENCE starts at position $i$ and ends at position $j$. The feature MAX computes the maximum of the values from index $i$ to index $j + 1$.

**Example**

$(\langle 4, 1, 4, 5, 5, 4, 4, 2, 2, 3, 4, 5, 6, 6, 5, 6 \rangle)$

Figure 3.343 provides an example where the INCREASING_MAX_STRICTLY_INCREASING_SEQUENCE $([4, 1, 4, 5, 5, 4, 4, 2, 2, 3, 4, 5, 6, 6, 5, 6])$ constraint holds.

**Typical**

$|VARIABLES| > 1$
range(VARIABLES.var) $> 1$

Figure 3.343: Illustrating the INCREASING_MAX_STRICTLY_INCREASING_SEQUENCE constraint of the **Example** slot

**Automaton**    Figure 3.344 depicts the automaton associated with the constraint INCREAS-ING_MAX_STRICTLY_INCREASING_SEQUENCE.

Figure 3.344: Automaton for the INCREASING_MAX_STRICTLY_INCREASING_SEQUENCE constraint obtained by applying decoration Table 2.36 to the seed transducer of the STRICTLY_INCREASING_SEQUENCE pattern

# 3.173  INCREASING_MAX_SUMMIT

$(< | < (= | <)^* <)(> | > (= | >)^* >)$

| | |
|---|---|
| **Origin** | Based on the SUMMIT pattern. |
| **Constraint** | INCREASING_MAX_SUMMIT(VARIABLES) |
| **Argument** | VARIABLES : collection(var−dvar) |
| **Restriction** | required(VARIABLES, var) |
| **Purpose** | Succeeds if the maxima of the values in each occurrence of the SUMMIT pattern in the time-series given by the VARIABLES collection are increasing.<br>An occurrence of the pattern SUMMIT is the *maximal* subsequence which matches the regular expression $(< | < (= | <)^* <)(> | > (= | >)^* >)$.<br>Assume that the occurrence of the pattern SUMMIT starts at position $i$ and ends at position $j$. The feature MAX computes the maximum of the values from index $i+1$ to index $j$. |
| **Example** | $(\langle 1, 5, 2, 1, 6, 6, 2, 3, 5, 4, 1, 4, 6, 4, 3, 2 \rangle)$ |

Figure 3.345 provides an example where the INCREASING_MAX_SUMMIT $([1, 5, 2, 1, 6, 6, 2, 3, 5, 4, 1, 4, 6, 4, 3, 2])$ constraint holds.

| | |
|---|---|
| **Typical** | $\lvert$VARIABLES$\rvert > 2$<br>range(VARIABLES.var) $> 1$ |

Figure 3.345: Illustrating the INCREASING_MAX_SUMMIT constraint of the **Example** slot

**Automaton**     Figure 3.346 depicts the automaton associated with the constraint INCREAS-
ING_MAX_SUMMIT.

Figure 3.346: Automaton for the INCREASING_MAX_SUMMIT constraint obtained by
applying decoration Table 2.36 to the seed transducer of the SUMMIT pattern (transition
$u \rightarrow r$ has the same accumulator update as transition $r \rightarrow u$)

# 3.174 INCREASING_MAX_ZIGZAG

▶ ▷ ◁ ◀          **DESCRIPTION**          **AUTOMATON**

| | |
|---|---|
| **Origin** | Based on the ZIGZAG pattern. |
| **Constraint** | INCREASING_MAX_ZIGZAG(VARIABLES) |
| **Argument** | VARIABLES : collection(var−dvar) |
| **Restriction** | required(VARIABLES, var) |
| **Purpose** | Succeeds if the maxima of the values in each occurrence of the ZIGZAG pattern in the time-series given by the VARIABLES collection are increasing.<br>An occurrence of the pattern ZIGZAG is the *maximal* subsequence which matches the regular expression $(<>)^+(< \mid <>) \mid (><)^+(> \mid ><)$.<br>Assume that the occurrence of the pattern ZIGZAG starts at position $i$ and ends at position $j$. The feature MAX computes the maximum of the values from index $i + 1$ to index $j$. |
| **Example** | $(\langle 7, 7, 2, 5, 3, 4, 6, 3, 6, 1, 1, 5, 4, 7, 5, 6 \rangle)$ |

Figure 3.347 provides an example where the INCREASING_MAX_ZIGZAG $([7, 7, 2, 5, 3, 4, 6, 3, 6, 1, 1, 5, 4, 7, 5, 6])$ constraint holds.

| | |
|---|---|
| **Typical** | $\mid$VARIABLES$\mid > 3$<br>range(VARIABLES.var) $> 1$ |

Figure 3.347: Illustrating the INCREASING_MAX_ZIGZAG constraint of the **Example** slot

**Automaton**            Figure 3.348 depicts the automaton associated with the constraint INCREAS-
                         ING_MAX_ZIGZAG.

Figure 3.348: Automaton for the INCREASING_MAX_ZIGZAG constraint obtained by applying decoration Table 2.36 to the seed transducer of the ZIGZAG pattern; (1) missing transitions from $a, b, c, d, e, f$ to $s$ are labelled by =; (2) on transitions from $b, c, e, f$ to $s$ the accumulator $D$ is reset to its initial value; (3) on transitions from $c, f$ to $s$ the accumulator $F$ is reset to $C$, and the accumulator $R$ is updated wrt $C$ and $F$

## 3.175 INCREASING_MIN_BUMP_ON_DECREASING_SEQUENCE

▶ ▷ ◁ ◀    **DESCRIPTION**    **AUTOMATON**

| | |
|---|---|
| **Origin** | Based on the BUMP_ON_DECREASING_SEQUENCE pattern. |
| **Constraint** | INCREASING_MIN_BUMP_ON_DECREASING_SEQUENCE(VARIABLES) |
| **Argument** | VARIABLES : collection(var−dvar) |
| **Restriction** | required(VARIABLES, var) |

**Purpose**

Succeeds if the minima of the values in each occurrence of the BUMP_ON_DECREASING_SEQUENCE pattern in the time-series given by the VARIABLES collection are increasing.

An occurrence of the pattern BUMP_ON_DECREASING_SEQUENCE is the subsequence which matches the regular expression `>><>`.

Assume that the occurrence of the pattern BUMP_ON_DECREASING_SEQUENCE starts at position $i$ and ends at position $j$. The feature MIN computes the minimum of the values from index $i + 2$ to index $j$.

**Example**

$$(\langle 5, 5, 4, 2, 6, 5, 4, 4, 5, 7, 6, 5, 6, 4, 1, 1 \rangle)$$

Figure 3.349 provides an example where the INCREASING_MIN_BUMP_ON_DECREASING_SEQUENCE ([5, 5, 4, 2, 6, 5, 4, 4, 5, 7, 6, 5, 6, 4, 1, 1]) constraint holds.

**Typical**

$|\texttt{VARIABLES}| > 5$

$\texttt{range}(\texttt{VARIABLES.var}) > 2$

Figure 3.349: Illustrating the INCREASING_MIN_BUMP_ON_DECREASING_SEQUENCE constraint of the **Example** slot

**Automaton**      Figure 3.350 depicts the automaton associated with the constraint INCREAS-ING_MIN_BUMP_ON_DECREASING_SEQUENCE.

Figure 3.350: Automaton for the INCREASING_MIN_BUMP_ON_DECREASING_SEQUENCE constraint obtained by applying decoration Table 2.36 to the seed transducer of the BUMP_ON_DECREASING_SEQUENCE pattern

## 3.176 INCREASING_MIN_DECREASING

► ▷ ◁ ◀        **DESCRIPTION**              **AUTOMATON**

| | |
|---|---|
| **Origin** | Based on the DECREASING pattern. |
| **Constraint** | INCREASING_MIN_DECREASING(VARIABLES) |
| **Argument** | VARIABLES : collection(var−dvar) |
| **Restriction** | required(VARIABLES, var) |
| **Purpose** | Succeeds if the minima of the values in each occurrence of the DECREASING pattern in the time-series given by the VARIABLES collection are increasing.<br><br>An occurrence of the pattern DECREASING is the subsequence which matches the regular expression >.<br><br>Assume that the occurrence of the pattern DECREASING starts at position $i$ and ends at position $j$. The feature MIN computes the minimum of the values from index $i$ to index $j+1$. |
| **Example** | $(\langle 3, 6, 6, 1, 1, 3, 3, 4, 5, 6, 1, 5, 3, 3, 6, 5\rangle)$ |

Figure 3.351 provides an example where the INCREASING_MIN_DECREASING $([3, 6, 6, 1, 1, 3, 3, 4, 5, 6, 1, 5, 3, 3, 6, 5])$ constraint holds.

| | |
|---|---|
| **Typical** | $\|$VARIABLES$\| > 1$<br>range(VARIABLES.var) $> 1$ |

Figure 3.351: Illustrating the INCREASING_MIN_DECREASING constraint of the **Example** slot

**Automaton**          Figure 3.352 depicts the automaton associated with the constraint INCREAS-
                       ING_MIN_DECREASING.

Figure 3.352: Automaton for the INCREASING_MIN_DECREASING constraint obtained
by applying decoration Table 2.36 to the seed transducer of the DECREASING pattern

## 3.177 INCREASING_MIN_DECREASING_SEQUENCE

DESCRIPTION            AUTOMATON

| Origin | Based on the DECREASING_SEQUENCE pattern. |
|---|---|

| Constraint | INCREASING_MIN_DECREASING_SEQUENCE(VARIABLES) |
|---|---|

| Argument | VARIABLES : collection(var−dvar) |
|---|---|

| Restriction | required(VARIABLES, var) |
|---|---|

| Purpose | Succeeds if the minima of the values in each occurrence of the DECREASING_SEQUENCE pattern in the time-series given by the VARIABLES collection are increasing. |
|---|---|
| | An occurrence of the pattern DECREASING_SEQUENCE is the *maximal* subsequence which matches the regular expression $> (> \mid =)^* > \mid >$. |
| | Assume that the occurrence of the pattern DECREASING_SEQUENCE starts at position $i$ and ends at position $j$. The feature MIN computes the minimum of the values from index $i$ to index $j + 1$. |

| Example | $(\langle 4, 3, 2, 2, 3, 4, 4, 6, 3, 4, 4, 5, 5, 6, 4, 6 \rangle)$ |
|---|---|

Figure 3.353 provides an example where the INCREASING_MIN_DECREASING_SEQUENCE $([4, 3, 2, 2, 3, 4, 4, 6, 3, 4, 4, 5, 5, 6, 4, 6])$ constraint holds.

Figure 3.353: Illustrating the INCREASING_MIN_DECREASING_SEQUENCE constraint of the **Example** slot

**Typical**

$|\text{VARIABLES}| > 1$

$\text{range}(\text{VARIABLES.var}) > 1$

**Automaton**        Figure 3.354 depicts the automaton associated with the constraint INCREAS-ING_MIN_DECREASING_SEQUENCE.

Figure 3.354: Automaton for the INCREASING_MIN_DECREASING_SEQUENCE constraint obtained by applying decoration Table 2.36 to the seed transducer of the DE-CREASING_SEQUENCE pattern

# 3.178 INCREASING_MIN_DIP_ON_INCREASING_SEQUENCE

▶ ▷ ◁ ◀      **DESCRIPTION**      **AUTOMATON**

**Origin**      Based on the DIP_ON_INCREASING_SEQUENCE pattern.

**Constraint**      INCREASING_MIN_DIP_ON_INCREASING_SEQUENCE(VARIABLES)

**Argument**      VARIABLES : collection(var−dvar)

**Restriction**      required(VARIABLES, var)

**Purpose**      Succeeds if the minima of the values in each occurrence of the DIP_ON_INCREASING_SEQUENCE pattern in the time-series given by the VARIABLES collection are increasing.

An occurrence of the pattern DIP_ON_INCREASING_SEQUENCE is the subsequence which matches the regular expression `<<><<`.

Assume that the occurrence of the pattern DIP_ON_INCREASING_SEQUENCE starts at position $i$ and ends at position $j$. The feature MIN computes the minimum of the values from index $i + 2$ to index $j$.

**Example**      $(\langle 1, 2, 3, 0, 3, 4, 6, 5, 1, 4, 5, 6, 1, 2, 4, 4 \rangle)$

Figure 3.355 provides an example where the INCREASING_MIN_DIP_ON_INCREASING_SEQUENCE $([1, 2, 3, 0, 3, 4, 6, 5, 1, 4, 5, 6, 1, 2, 4, 4])$ constraint holds.

**Typical**      $|\text{VARIABLES}| > 5$

range(VARIABLES.var) > 2

Figure 3.355: Illustrating the INCREASING_MIN_DIP_ON_INCREASING_SEQUENCE constraint of the **Example** slot

**Automaton**        Figure 3.356 depicts the automaton associated with the constraint INCREAS-
ING_MIN_DIP_ON_INCREASING_SEQUENCE.

Figure 3.356: Automaton for the INCREASING_MIN_DIP_ON_INCREASING_SEQUENCE
constraint obtained by applying decoration Table 2.36 to the seed transducer of the
DIP_ON_INCREASING_SEQUENCE pattern

## 3.179 INCREASING_MIN_GORGE

► ▷ ◁ ◄          **DESCRIPTION**          **AUTOMATON**

$(> | > (= | >)^*)(< | < (= | <)^* <)$

**Origin**              Based on the GORGE pattern.

**Constraint**          INCREASING_MIN_GORGE(VARIABLES)

**Argument**            VARIABLES : collection(var−dvar)

**Restriction**         required(VARIABLES, var)

**Purpose**             Succeeds if the minima of the values in each occurrence of the GORGE pattern in the time-series given by the VARIABLES collection are increasing.

An occurrence of the pattern GORGE is the *maximal* subsequence which matches the regular expression $(> | > (= | >)^*)(< | < (= | <)^* <)$.

Assume that the occurrence of the pattern GORGE starts at position $i$ and ends at position $j$. The feature MIN computes the minimum of the values from index $i + 1$ to index $j$.

**Example**             $(\langle 6, 2, 5, 1, 1, 5, 4, 2, 3, 6, 5, 4, 4, 3, 5, 5 \rangle)$

Figure 3.357 provides an example where the INCREASING_MIN_GORGE $([6, 2, 5, 1, 1, 5, 4, 2, 3, 6, 5, 4, 4, 3, 5, 5])$ constraint holds.

Figure 3.357: Illustrating the INCREASING_MIN_GORGE constraint of the **Example** slot

**Typical**
$$|\mathtt{VARIABLES}| > 2$$
$$\mathtt{range}(\mathtt{VARIABLES.var}) > 1$$

**Automaton**  Figure 3.358 depicts the automaton associated with the constraint INCREAS-ING_MIN_GORGE.

Figure 3.358: Automaton for the INCREASING_MIN_GORGE constraint obtained by applying decoration Table 2.36 to the seed transducer of the GORGE pattern (transition $u \rightarrow r$ has the same accumulator update as transition $r \rightarrow u$)

## 3.180 INCREASING_MIN_INCREASING

▶ ▷ ◁ ◀          DESCRIPTION          AUTOMATON

**Origin**          Based on the INCREASING pattern.

**Constraint**     INCREASING_MIN_INCREASING(VARIABLES)

**Argument**       VARIABLES : collection(var−dvar)

**Restriction**    required(VARIABLES, var)

**Purpose**        Succeeds if the minima of the values in each occurrence of the INCREASING pattern in the time-series given by the VARIABLES collection are increasing.

An occurrence of the pattern INCREASING is the subsequence which matches the regular expression $<$.

Assume that the occurrence of the pattern INCREASING starts at position $i$ and ends at position $j$. The feature MIN computes the minimum of the values from index $i$ to index $j + 1$.

**Example**        $(\langle 4, 1, 2, 4, 3, 3, 3, 2, 3, 3, 4, 4, 6, 6, 5, 7 \rangle)$

Figure 3.359 provides an example where the INCREASING_MIN_INCREASING $([4, 1, 2, 4, 3, 3, 3, 2, 3, 3, 4, 4, 6, 6, 5, 7])$ constraint holds.

**Typical**        $|\text{VARIABLES}| > 1$
                   range(VARIABLES.var) > 1

Figure 3.359: Illustrating the INCREASING_MIN_INCREASING constraint of the **Example** slot

**Automaton**        Figure 3.360 depicts the automaton associated with the constraint INCREAS-ING_MIN_INCREASING.

Figure 3.360: Automaton for the INCREASING_MIN_INCREASING constraint obtained by applying decoration Table 2.36 to the seed transducer of the INCREASING pattern

# 3.181 INCREASING_MIN_INCREASING_SEQUENCE

▶ ▷ ◁ ◀        **DESCRIPTION**            **AUTOMATON**

**Origin**           Based on the INCREASING_SEQUENCE pattern.

**Constraint**       INCREASING_MIN_INCREASING_SEQUENCE(VARIABLES)

**Argument**         VARIABLES : collection(var−dvar)

**Restriction**      required(VARIABLES, var)

**Purpose**          Succeeds if the minima of the values in each occurrence of the INCREASING_SEQUENCE
                     pattern in the time-series given by the VARIABLES collection are increasing.
                     An occurrence of the pattern INCREASING_SEQUENCE is the *maximal* subsequence
                     which matches the regular expression $< (< | =)^* < | <$.
                     Assume that the occurrence of the pattern INCREASING_SEQUENCE starts at position $i$
                     and ends at position $j$. The feature MIN computes the minimum of the values from index
                     $i$ to index $j + 1$.

**Example**          $(\langle 4, 1, 2, 4, 3, 3, 3, 2, 3, 3, 4, 4, 6, 6, 5, 7 \rangle)$

                     Figure 3.361 provides an example where the INCREASING_MIN_INCREASING_SEQUENCE
                     $([4, 1, 2, 4, 3, 3, 3, 2, 3, 3, 4, 4, 6, 6, 5, 7])$ constraint holds.

**Typical**          $|\text{VARIABLES}| > 1$
                     range(VARIABLES.var) > 1

Figure 3.361: Illustrating the INCREASING_MIN_INCREASING_SEQUENCE constraint of the **Example** slot

**Automaton**    Figure 3.362 depicts the automaton associated with the constraint INCREAS-ING_MIN_INCREASING_SEQUENCE.

Figure 3.362: Automaton for the INCREASING_MIN_INCREASING_SEQUENCE constraint obtained by applying decoration Table 2.36 to the seed transducer of the INCREASING_SEQUENCE pattern

# 3.182  INCREASING_MIN_INFLEXION

        **DESCRIPTION**          **AUTOMATON**

$< (< | =)^* > | > (> | =)^* <$

| | |
|---|---|
| **Origin** | Based on the INFLEXION pattern. |
| **Constraint** | INCREASING_MIN_INFLEXION(VARIABLES) |
| **Argument** | VARIABLES : collection(var−dvar) |
| **Restriction** | required(VARIABLES, var) |
| **Purpose** | Succeeds if the minima of the values in each occurrence of the INFLEXION pattern in the time-series given by the VARIABLES collection are increasing. <br> An occurrence of the pattern INFLEXION is the *maximal* subsequence which matches the regular expression $< (< | =)^* > | > (> | =)^* <$. <br> Assume that the occurrence of the pattern INFLEXION starts at position $i$ and ends at position $j$. The feature MIN computes the minimum of the values from index $i + 1$ to index $j$. |
| **Example** | $(\langle 1, 1, 2, 4, 4, 5, 3, 3, 3, 4, 5, 6 \rangle)$ |

Figure 3.363 provides an example where the INCREASING_MIN_INFLEXION ([1, 1, 2, 4, 4, 5, 3, 3, 3, 4, 5, 6]) constraint holds.

| | |
|---|---|
| **Typical** | $|\text{VARIABLES}| > 2$ <br> range(VARIABLES.var) $> 1$ |

Figure 3.363: Illustrating the INCREASING_MIN_INFLEXION constraint of the **Example** slot

**Automaton**    Figure 3.364 depicts the automaton associated with the constraint INCREAS-
ING_MIN_INFLEXION.

Figure 3.364: Automaton for the INCREASING_MIN_INFLEXION constraint obtained
by applying decoration Table 2.36 to the seed transducer of the INFLEXION pattern
(transition $r \rightarrow t$ has the same accumulators updates as transition $t \rightarrow r$)

# 3.183 INCREASING_MIN_STRICTLY_DECREASING_SEQUENCE

▶ ▷ ◁ ◀    DESCRIPTION         AUTOMATON

| | |
|---|---|
| **Origin** | Based on the STRICTLY_DECREASING_SEQUENCE pattern. |
| **Constraint** | INCREASING_MIN_STRICTLY_DECREASING_SEQUENCE(VARIABLES) |
| **Argument** | VARIABLES : collection(var−dvar) |
| **Restriction** | required(VARIABLES, var) |

**Purpose**

Succeeds if the minima of the values in each occurrence of the STRICTLY_DECREASING_SEQUENCE pattern in the time-series given by the VARIABLES collection are increasing.

An occurrence of the pattern STRICTLY_DECREASING_SEQUENCE is the *maximal* subsequence which matches the regular expression $>^+$.

Assume that the occurrence of the pattern STRICTLY_DECREASING_SEQUENCE starts at position $i$ and ends at position $j$. The feature MIN computes the minimum of the values from index $i$ to index $j + 1$.

**Example**

$$(\langle 1, 3, 2, 2, 4, 4, 5, 5, 6, 3, 5, 5, 4, 6, 7, 4 \rangle)$$

Figure 3.365 provides an example where the INCREASING_MIN_STRICTLY_DECREASING_SEQUENCE $([1, 3, 2, 2, 4, 4, 5, 5, 6, 3, 5, 5, 4, 6, 7, 4])$ constraint holds.

**Typical**

$|VARIABLES| > 1$
range(VARIABLES.var) $> 1$

Figure 3.365: Illustrating the INCREASING_MIN_STRICTLY_DECREASING_SEQUENCE constraint of the **Example** slot

**Automaton**    Figure 3.366 depicts the automaton associated with the constraint INCREAS-
ING_MIN_STRICTLY_DECREASING_SEQUENCE.

Figure 3.366: Automaton for the INCREASING_MIN_STRICTLY_DECREASING_SEQUENCE
constraint obtained by applying decoration Table 2.36 to the seed transducer of the
STRICTLY_DECREASING_SEQUENCE pattern

## 3.184 INCREASING_MIN_STRICTLY_INCREASING_SEQUENCE

▶ ▷ ◁ ◀      **DESCRIPTION**      **AUTOMATON**

| | |
|---|---|
| **Origin** | Based on the STRICTLY_INCREASING_SEQUENCE pattern. |
| **Constraint** | INCREASING_MIN_STRICTLY_INCREASING_SEQUENCE(VARIABLES) |
| **Argument** | VARIABLES : collection(var−dvar) |
| **Restriction** | required(VARIABLES, var) |

**Purpose**

Succeeds if the minima of the values in each occurrence of the STRICTLY_INCREASING_SEQUENCE pattern in the time-series given by the VARIABLES collection are increasing.

An occurrence of the pattern STRICTLY_INCREASING_SEQUENCE is the *maximal* subsequence which matches the regular expression $<^+$.

Assume that the occurrence of the pattern STRICTLY_INCREASING_SEQUENCE starts at position $i$ and ends at position $j$. The feature MIN computes the minimum of the values from index $i$ to index $j + 1$.

**Example**

$$(\langle 4, 1, 2, 4, 3, 3, 3, 2, 3, 3, 4, 5, 6, 6, 5, 7 \rangle)$$

Figure 3.367 provides an example where the INCREASING_MIN_STRICTLY_INCREASING_SEQUENCE $([4, 1, 2, 4, 3, 3, 3, 2, 3, 3, 4, 5, 6, 6, 5, 7])$ constraint holds.

**Typical**

$|VARIABLES| > 1$
$range(VARIABLES.var) > 1$

Figure 3.367: Illustrating the INCREASING_MIN_STRICTLY_INCREASING_SEQUENCE constraint of the **Example** slot

**Automaton**     Figure 3.368 depicts the automaton associated with the constraint INCREAS-
ING_MIN_STRICTLY_INCREASING_SEQUENCE.

Figure 3.368: Automaton for the INCREASING_MIN_STRICTLY_INCREASING_SEQUENCE
constraint obtained by applying decoration Table 2.36 to the seed transducer of the
STRICTLY_INCREASING_SEQUENCE pattern

## 3.185   INCREASING_MIN_VALLEY

▶ ▷ ◁ ◀          **DESCRIPTION**          **AUTOMATON**

| | |
|---|---|
| **Origin** | Based on the VALLEY pattern. |
| **Constraint** | INCREASING_MIN_VALLEY(VARIABLES) |
| **Argument** | VARIABLES : collection(var−dvar) |
| **Restriction** | required(VARIABLES, var) |
| **Purpose** | Succeeds if the minima of the values in each occurrence of the VALLEY pattern in the time-series given by the VARIABLES collection are increasing. An occurrence of the pattern VALLEY is the *maximal* subsequence which matches the regular expression $> (= | >)^* (< | =)^* <$. Assume that the occurrence of the pattern VALLEY starts at position $i$ and ends at position $j$. The feature MIN computes the minimum of the values from index $i + 1$ to index $j$. |
| **Example** | $(\langle 7, 2, 2, 6, 3, 4, 5, 6, 6, 4, 4, 6, 7, 3, 3, 1 \rangle)$ |

Figure 3.369 provides an example where the INCREASING_MIN_VALLEY $([7, 2, 2, 6, 3, 4, 5, 6, 6, 4, 4, 6, 7, 3, 3, 1])$ constraint holds.

| | |
|---|---|
| **Typical** | $\|$VARIABLES$\| > 2$ |
| | range(VARIABLES.var) $> 1$ |

Figure 3.369: Illustrating the INCREASING_MIN_VALLEY constraint of the **Example** slot

**Automaton**          Figure 3.370 depicts the automaton associated with the constraint INCREAS-
ING_MIN_VALLEY.

Figure 3.370: Automaton for the INCREASING_MIN_VALLEY constraint obtained by
applying decoration Table 2.36 to the seed transducer of the VALLEY pattern

## 3.186 INCREASING_MIN_ZIGZAG

► ▷ ◁ ◀          **DESCRIPTION**          **AUTOMATON**

**Origin**          Based on the ZIGZAG pattern.

**Constraint**          INCREASING_MIN_ZIGZAG(VARIABLES)

**Argument**          VARIABLES : collection(var−dvar)

**Restriction**          required(VARIABLES, var)

**Purpose**          Succeeds if the minima of the values in each occurrence of the ZIGZAG pattern in the time-series given by the VARIABLES collection are increasing.

An occurrence of the pattern ZIGZAG is the *maximal* subsequence which matches the regular expression $(<>)^+(< \mid <>) \mid (><)^+(> \mid ><)$.

Assume that the occurrence of the pattern ZIGZAG starts at position $i$ and ends at position $j$. The feature MIN computes the minimum of the values from index $i + 1$ to index $j$.

**Example**          $(\langle 7, 7, 2, 5, 3, 4, 6, 3, 6, 1, 1, 5, 4, 7, 5, 6 \rangle)$

Figure 3.371 provides an example where the INCREASING_MIN_ZIGZAG $([7, 7, 2, 5, 3, 4, 6, 3, 6, 1, 1, 5, 4, 7, 5, 6])$ constraint holds.

**Typical**          $|\texttt{VARIABLES}| > 3$
range(VARIABLES.var) > 1

Figure 3.371: Illustrating the INCREASING_MIN_ZIGZAG constraint of the **Example** slot

**Automaton**          Figure 3.372 depicts the automaton associated with the constraint INCREAS-
                       ING_MIN_ZIGZAG.

Figure 3.372: Automaton for the INCREASING_MIN_ZIGZAG constraint obtained by applying decoration Table 2.36 to the seed transducer of the ZIGZAG pattern; (1) missing transitions from $a, b, c, d, e, f$ to $s$ are labelled by $=$; (2) on transitions from $b, c, e, f$ to $s$ the accumulator $D$ is reset to its initial value; (3) on transitions from $c, f$ to $s$ the accumulator $F$ is reset to $C$, and the accumulator $R$ is updated wrt $C$ and $F$

## 3.187  INCREASING_RANGE_DECREASING

        **DESCRIPTION**                    **AUTOMATON**

| | |
|---|---|
| **Origin** | Based on the DECREASING pattern. |
| **Constraint** | INCREASING_RANGE_DECREASING(VARIABLES) |
| **Argument** | VARIABLES : collection(var−dvar) |
| **Restriction** | required(VARIABLES, var) |
| **Purpose** | Succeeds if the differences between the largest and smallest value in each occurrence of the DECREASING pattern in the time-series given by the VARIABLES collection are increasing. <br> An occurrence of the pattern DECREASING is the subsequence which matches the regular expression >. <br> Assume that the occurrence of the pattern DECREASING starts at position $i$ and ends at position $j$. The feature RANGE computes the range of the values from index $i$ to index $j + 1$. |
| **Example** | $(\langle 3, 6, 6, 5, 4, 4, 6, 4, 2, 2, 0, 5, 2, 6, 3, 3\rangle)$ |

Figure 3.373 provides an example where the INCREASING_RANGE_DECREASING
($[3, 6, 6, 5, 4, 4, 6, 4, 2, 2, 0, 5, 2, 6, 3, 3]$) constraint holds.

| | |
|---|---|
| **Typical** | $|\texttt{VARIABLES}| > 1$ <br> range(VARIABLES.var) $> 1$ |

Figure 3.373: Illustrating the INCREASING_RANGE_DECREASING constraint of the **Example** slot

**Automaton**        Figure 3.374 depicts the automaton associated with the constraint INCREAS-
ING_RANGE_DECREASING.

$$\left\{ \begin{array}{l} C \leftarrow -\infty \\ F \leftarrow -\infty \\ H \leftarrow \text{VAR}_1 \\ R \leftarrow 1 \end{array} \right\}$$

$$\left\{ \begin{array}{l} C \leftarrow |H - \text{VAR}_{i+1}| \\ F \leftarrow |H - \text{VAR}_{i+1}| \\ H \leftarrow \text{VAR}_{i+1} \\ R \leftarrow R \wedge (F \leq |H - \text{VAR}_{i+1}|) \end{array} \right\}$$

$$\{H \leftarrow \text{VAR}_{i+1}\}$$

$$R \wedge (F \leq C)$$

Figure 3.374: Automaton for the INCREASING_RANGE_DECREASING constraint ob-
tained by applying decoration Table 2.47 to the seed transducer of the DECREASING
pattern

# 3.188   INCREASING_RANGE_DECREASING_SEQUENCE

▶ ▷ ◁ ◀          **DESCRIPTION**          **AUTOMATON**

**Origin**          Based on the DECREASING_SEQUENCE pattern.

**Constraint**      INCREASING_RANGE_DECREASING_SEQUENCE(VARIABLES)

**Argument**        VARIABLES : collection(var−dvar)

**Restriction**     required(VARIABLES, var)

**Purpose**         Succeeds if the differences between the largest and smallest value in each occurrence of
                    the DECREASING_SEQUENCE pattern in the time-series given by the VARIABLES collec-
                    tion are increasing.
                    An occurrence of the pattern DECREASING_SEQUENCE is the *maximal* subsequence
                    which matches the regular expression $> (> | =)^* > | >$.
                    Assume that the occurrence of the pattern DECREASING_SEQUENCE starts at position $i$
                    and ends at position $j$. The feature RANGE computes the range of the values from index
                    $i$ to index $j + 1$.

**Example**         $(\langle 4, 3, 2, 2, 0, 0, 3, 4, 4, 6, 2, 4, 4, 6, 5, 1 \rangle)$

                    Figure     3.375     provides     an     example     where     the     INCREAS-
                    ING_RANGE_DECREASING_SEQUENCE    $([4, 3, 2, 2, 0, 0, 3, 4, 4, 6, 2, 4, 4, 6, 5, 1])$    con-
                    straint holds.

**Typical**         $|\texttt{VARIABLES}| > 1$
                    $\texttt{range}(\texttt{VARIABLES.var}) > 1$

Figure 3.375: Illustrating the INCREASING_RANGE_DECREASING_SEQUENCE constraint of the **Example** slot

**Automaton**       Figure 3.376 depicts the automaton associated with the constraint INCREAS-
                    ING_RANGE_DECREASING_SEQUENCE.

Figure 3.376: Automaton for the INCREASING_RANGE_DECREASING_SEQUENCE
constraint obtained by applying decoration Table 2.47 to the seed transducer of the
DECREASING_SEQUENCE pattern

# 3.189   INCREASING_RANGE_INCREASING

► ▷ ◁ ◀         **DESCRIPTION**          **AUTOMATON**

| | |
|---|---|
| **Origin** | Based on the INCREASING pattern. |
| **Constraint** | INCREASING_RANGE_INCREASING(VARIABLES) |
| **Argument** | VARIABLES : collection(var−dvar) |
| **Restriction** | required(VARIABLES, var) |
| **Purpose** | Succeeds if the differences between the largest and smallest value in each occurrence of the INCREASING pattern in the time-series given by the VARIABLES collection are increasing. <br><br> An occurrence of the pattern INCREASING is the subsequence which matches the regular expression <. <br><br> Assume that the occurrence of the pattern INCREASING starts at position $i$ and ends at position $j$. The feature RANGE computes the range of the values from index $i$ to index $j + 1$. |
| **Example** | $(\langle 4, 5, 6, 6, 3, 3, 4, 6, 2, 2, 1, 4, 7, 0, 5, 5\rangle)$ |

Figure 3.377 provides an example where the INCREASING_RANGE_INCREASING
($[4, 5, 6, 6, 3, 3, 4, 6, 2, 2, 1, 4, 7, 0, 5, 5]$) constraint holds.

| | |
|---|---|
| **Typical** | $\lvert\text{VARIABLES}\rvert > 1$ <br> range(VARIABLES.var) $> 1$ |

Figure 3.377: Illustrating the INCREASING_RANGE_INCREASING constraint of the **Example** slot

**Automaton**    Figure 3.378 depicts the automaton associated with the constraint INCREAS-ING_RANGE_INCREASING.

$$\left\{ \begin{array}{l} C \leftarrow -\infty \\ F \leftarrow -\infty \\ H \leftarrow \mathtt{VAR}_1 \\ R \leftarrow 1 \end{array} \right\}$$

$$\left\{ \begin{array}{l} C \leftarrow |H - \mathtt{VAR}_{i+1}| \\ F \leftarrow |H - \mathtt{VAR}_{i+1}| \\ H \leftarrow \mathtt{VAR}_{i+1} \\ R \leftarrow R \wedge (F \leq |H - \mathtt{VAR}_{i+1}|) \end{array} \right\}$$

$$\{H \leftarrow \mathtt{VAR}_{i+1}\}$$

$$\boxed{R \wedge (F \leq C)}$$

Figure 3.378: Automaton for the INCREASING_RANGE_INCREASING constraint obtained by applying decoration Table 2.47 to the seed transducer of the INCREASING pattern

# 3.190 INCREASING_RANGE_INCREASING_SEQUENCE

▶ ▷ ◁ ◀ **DESCRIPTION** **AUTOMATON**

| | |
|---|---|
| **Origin** | Based on the INCREASING_SEQUENCE pattern. |
| **Constraint** | INCREASING_RANGE_INCREASING_SEQUENCE(VARIABLES) |
| **Argument** | VARIABLES : collection(var−dvar) |
| **Restriction** | required(VARIABLES, var) |

**Purpose**

Succeeds if the differences between the largest and smallest value in each occurrence of the INCREASING_SEQUENCE pattern in the time-series given by the VARIABLES collection are increasing.

An occurrence of the pattern INCREASING_SEQUENCE is the *maximal* subsequence which matches the regular expression $< (< | =)^* < | <$.

Assume that the occurrence of the pattern INCREASING_SEQUENCE starts at position $i$ and ends at position $j$. The feature RANGE computes the range of the values from index $i$ to index $j + 1$.

**Example**

$(\langle 4, 1, 2, 4, 3, 3, 3, 2, 3, 3, 4, 4, 6, 6, 3, 7 \rangle)$

Figure 3.379 provides an example where the INCREASING_RANGE_INCREASING_SEQUENCE ([4, 1, 2, 4, 3, 3, 3, 2, 3, 3, 4, 4, 6, 6, 3, 7]) constraint holds.

**Typical**

$|\text{VARIABLES}| > 1$
$\text{range}(\text{VARIABLES.var}) > 1$

Figure 3.379: Illustrating the INCREASING_RANGE_INCREASING_SEQUENCE constraint of the **Example** slot

**Automaton**    Figure 3.380 depicts the automaton associated with the constraint INCREAS-
ING_RANGE_INCREASING_SEQUENCE.

Figure 3.380: Automaton for the INCREASING_RANGE_INCREASING_SEQUENCE con-
straint obtained by applying decoration Table 2.47 to the seed transducer of the IN-
CREASING_SEQUENCE pattern

## 3.191   INCREASING_RANGE_STRICTLY_DECREASING_SEQUENCE

▶ ▷ ◁ ◀      **DESCRIPTION**      **AUTOMATON**

| | |
|---|---|
| **Origin** | Based on the STRICTLY_DECREASING_SEQUENCE pattern. |
| **Constraint** | INCREASING_RANGE_STRICTLY_DECREASING_SEQUENCE(VARIABLES) |
| **Argument** | VARIABLES : collection(var−dvar) |
| **Restriction** | required(VARIABLES, var) |

**Purpose**

Succeeds if the differences between the largest and smallest value in each occurrence of the STRICTLY_DECREASING_SEQUENCE pattern in the time-series given by the VARIABLES collection are increasing.

An occurrence of the pattern STRICTLY_DECREASING_SEQUENCE is the *maximal* subsequence which matches the regular expression $>^+$.

Assume that the occurrence of the pattern STRICTLY_DECREASING_SEQUENCE starts at position $i$ and ends at position $j$. The feature RANGE computes the range of the values from index $i$ to index $j + 1$.

**Example**

$$(\langle 1, 3, 2, 2, 5, 6, 5, 3, 3, 0, 1, 3, 5, 1, 1, 2 \rangle)$$

Figure 3.381 provides an example where the IN-CREASING_RANGE_STRICTLY_DECREASING_SEQUENCE $([1, 3, 2, 2, 5, 6, 5, 3, 3, 0, 1, 3, 5, 1, 1, 2])$ constraint holds.

**Typical**

$|VARIABLES| > 1$

$range(VARIABLES.var) > 1$

Figure 3.381: Illustrating the INCREASING_RANGE_STRICTLY_DECREASING_SEQUENCE constraint of the **Example** slot

**Automaton**     Figure 3.382 depicts the automaton associated with the constraint INCREAS-
ING_RANGE_STRICTLY_DECREASING_SEQUENCE.

Figure 3.382: Automaton for the INCREASING_RANGE_STRICTLY_DECREASING_SEQUENCE
constraint obtained by applying decoration Table 2.47 to the seed transducer of the
STRICTLY_DECREASING_SEQUENCE pattern

## 3.192  INCREASING_RANGE_STRICTLY_INCREASING_SEQUENCE

▶ ▷ ◁ ◀          **DESCRIPTION**          **AUTOMATON**

| | |
|---|---|
| **Origin** | Based on the STRICTLY_INCREASING_SEQUENCE pattern. |
| **Constraint** | INCREASING_RANGE_STRICTLY_INCREASING_SEQUENCE(VARIABLES) |
| **Argument** | VARIABLES : collection(var−dvar) |
| **Restriction** | required(VARIABLES, var) |
| **Purpose** | Succeeds if the differences between the largest and smallest value in each occurrence of the STRICTLY_INCREASING_SEQUENCE pattern in the time-series given by the VARIABLES collection are increasing. <br><br> An occurrence of the pattern STRICTLY_INCREASING_SEQUENCE is the *maximal* subsequence which matches the regular expression $<^{+}$. <br><br> Assume that the occurrence of the pattern STRICTLY_INCREASING_SEQUENCE starts at position $i$ and ends at position $j$. The feature RANGE computes the range of the values from index $i$ to index $j + 1$. |
| **Example** | $(\langle 4, 1, 2, 2, 3, 0, 1, 1, 4, 1, 1, 2, 3, 4, 4, 8 \rangle)$ |

Figure 3.383 provides an example where the INCREASING_RANGE_STRICTLY_INCREASING_SEQUENCE $([4, 1, 2, 2, 3, 0, 1, 1, 4, 1, 1, 2, 3, 4, 4, 8])$ constraint holds.

| | |
|---|---|
| **Typical** | $|\texttt{VARIABLES}| > 1$ <br> range(VARIABLES.var) > 1 |

Figure 3.383: Illustrating the INCREASING_RANGE_STRICTLY_INCREASING_SEQUENCE constraint of the **Example** slot

**Automaton**          Figure 3.384 depicts the automaton associated with the constraint INCREAS-
                       ING_RANGE_STRICTLY_INCREASING_SEQUENCE.

Figure 3.384: Automaton for the INCREASING_RANGE_STRICTLY_INCREASING_SEQUENCE constraint obtained by applying decoration Table 2.47 to the seed transducer of the STRICTLY_INCREASING_SEQUENCE pattern

# 3.193   INCREASING_SURF_BUMP_ON_DECREASING_SEQUENCE

▶ ▷ ◁ ◀       **DESCRIPTION**          **AUTOMATON**

| | |
|---|---|
| **Origin** | Based on the BUMP_ON_DECREASING_SEQUENCE pattern. |
| **Constraint** | INCREASING_SURF_BUMP_ON_DECREASING_SEQUENCE(VARIABLES) |
| **Argument** | VARIABLES : collection(var−dvar) |
| **Restriction** | required(VARIABLES, var) |

**Purpose**

Succeeds if the values denoting the surface of each occurrence of the BUMP_ON_DECREASING_SEQUENCE pattern in the time-series given by the VARIABLES collection are increasing.

An occurrence of the pattern BUMP_ON_DECREASING_SEQUENCE is the subsequence which matches the regular expression `>><>>`.

Assume that the occurrence of the pattern BUMP_ON_DECREASING_SEQUENCE starts at position $i$ and ends at position $j$. The feature SURF computes the sum of the values from index $i + 2$ to index $j$.

**Example**

$(\langle 7, 6, 4, 2, 5, 4, 3, 3, 5, 7, 6, 5, 6, 5, 4, 1 \rangle)$

Figure 3.385 provides an example where the INCREAS-ING_SURF_BUMP_ON_DECREASING_SEQUENCE $([7, 6, 4, 2, 5, 4, 3, 3, 5, 7, 6, 5, 6, 5, 4, 1])$ constraint holds.

**Typical**

$|\text{VARIABLES}| > 5$

$\text{range}(\text{VARIABLES.var}) > 2$

Figure 3.385: Illustrating the INCREASING_SURF_BUMP_ON_DECREASING_SEQUENCE constraint of the **Example** slot

**Automaton**    Figure 3.386 depicts the automaton associated with the constraint INCREAS-ING_SURF_BUMP_ON_DECREASING_SEQUENCE.

Figure 3.386: Automaton for the INCREASING_SURF_BUMP_ON_DECREASING_SEQUENCE constraint obtained by applying decoration Table 2.36 to the seed transducer of the BUMP_ON_DECREASING_SEQUENCE pattern

## 3.194 INCREASING_SURF_DECREASING

► ▷ ◁ ◄          DESCRIPTION          AUTOMATON

| | |
|---|---|
| **Origin** | Based on the DECREASING pattern. |
| **Constraint** | INCREASING_SURF_DECREASING(VARIABLES) |
| **Argument** | VARIABLES : collection(var−dvar) |
| **Restriction** | required(VARIABLES, var) |
| **Purpose** | Succeeds if the values denoting the surface of each occurrence of the DECREASING pattern in the time-series given by the VARIABLES collection are increasing. An occurrence of the pattern DECREASING is the subsequence which matches the regular expression >. Assume that the occurrence of the pattern DECREASING starts at position $i$ and ends at position $j$. The feature SURF computes the sum of the values from index $i$ to index $j+1$. |
| **Example** | $(\langle 3, 3, 2, 2, 3, 4, 4, 6, 3, 4, 4, 5, 5, 6, 4, 6\rangle)$ |

Figure 3.387 provides an example where the INCREASING_SURF_DECREASING ([3, 3, 2, 2, 3, 4, 4, 6, 3, 4, 4, 5, 5, 6, 4, 6]) constraint holds.

Figure 3.387: Illustrating the INCREASING_SURF_DECREASING constraint of the **Example** slot

| | |
|---|---|
| **Typical** | $|\text{VARIABLES}| > 1$<br>range(VARIABLES.var) > 1 |

**Automaton**  Figure 3.388 depicts the automaton associated with the constraint INCREAS-ING_SURF_DECREASING.

Figure 3.388: Automaton for the INCREASING_SURF_DECREASING constraint obtained by applying decoration Table 2.36 to the seed transducer of the DECREASING pattern

# 3.195 INCREASING_SURF_DECREASING_SEQUENCE

▶ ▷ ◁ ◀        **DESCRIPTION**        **AUTOMATON**

**Origin**          Based on the DECREASING_SEQUENCE pattern.

**Constraint**      INCREASING_SURF_DECREASING_SEQUENCE(VARIABLES)

**Argument**        VARIABLES : collection(var−dvar)

**Restriction**     required(VARIABLES, var)

**Purpose**         Succeeds if the values denoting the surface of each occurrence of the DECREAS-
                    ING_SEQUENCE pattern in the time-series given by the VARIABLES collection are in-
                    creasing.
                    An occurrence of the pattern DECREASING_SEQUENCE is the *maximal* subsequence
                    which matches the regular expression $> (> | =)^* > | >$.
                    Assume that the occurrence of the pattern DECREASING_SEQUENCE starts at position $i$
                    and ends at position $j$. The feature SURF computes the sum of the values from index $i$ to
                    index $j + 1$.

**Example**         $(\langle 4, 3, 2, 2, 3, 4, 4, 6, 3, 4, 4, 5, 5, 6, 4, 6 \rangle)$

                    Figure 3.389 provides an example where the INCREAS-
                    ING_SURF_DECREASING_SEQUENCE $([4, 3, 2, 2, 3, 4, 4, 6, 3, 4, 4, 5, 5, 6, 4, 6])$ constraint
                    holds.

**Typical**         $|\text{VARIABLES}| > 1$
                    range(VARIABLES.var) $> 1$

Figure 3.389: Illustrating the INCREASING_SURF_DECREASING_SEQUENCE constraint of the **Example** slot

**Automaton**    Figure 3.390 depicts the automaton associated with the constraint INCREAS-ING_SURF_DECREASING_SEQUENCE.

Figure 3.390: Automaton for the INCREASING_SURF_DECREASING_SEQUENCE constraint obtained by applying decoration Table 2.36 to the seed transducer of the DE-CREASING_SEQUENCE pattern

## 3.196   INCREASING_SURF_DECREASING_TERRACE

▶ ▷ ◁ ◀        **DESCRIPTION**              **AUTOMATON**

| | |
|---|---|
| **Origin** | Based on the DECREASING_TERRACE pattern. |
| **Constraint** | INCREASING_SURF_DECREASING_TERRACE(VARIABLES) |
| **Argument** | VARIABLES : collection(var−dvar) |
| **Restriction** | required(VARIABLES, var) |

**Purpose**

Succeeds if the values denoting the surface of each occurrence of the DECREAS-ING_TERRACE pattern in the time-series given by the VARIABLES collection are increasing.

An occurrence of the pattern DECREASING_TERRACE is the *maximal* subsequence which matches the regular expression $>=^+>$.

Assume that the occurrence of the pattern DECREASING_TERRACE starts at position $i$ and ends at position $j$. The feature SURF computes the sum of the values from index $i+1$ to index $j$.

**Example**

$$(\langle 5, 2, 2, 1, 5, 4, 3, 3, 2, 4, 4, 6, 5, 5, 5, 4 \rangle)$$

Figure 3.391 provides an example where the INCREASING_SURF_DECREASING_TERRACE ([5, 2, 2, 1, 5, 4, 3, 3, 2, 4, 4, 6, 5, 5, 5, 4]) constraint holds.

**Typical**

$|\text{VARIABLES}| > 3$

$\text{range}(\text{VARIABLES.var}) > 2$

Figure 3.391: Illustrating the INCREASING_SURF_DECREASING_TERRACE constraint of the **Example** slot

**Automaton**     Figure 3.392 depicts the automaton associated with the constraint INCREAS-
ING_SURF_DECREASING_TERRACE.

Figure 3.392: Automaton for the INCREASING_SURF_DECREASING_TERRACE con-
straint obtained by applying decoration Table 2.36 to the seed transducer of the DE-
CREASING_TERRACE pattern

# 3.197  INCREASING_SURF_DIP_ON_INCREASING_SEQUENCE

▶ ▷ ◁ ◀      **DESCRIPTION**            **AUTOMATON**

| | |
|---|---|
| **Origin** | Based on the DIP_ON_INCREASING_SEQUENCE pattern. |
| **Constraint** | INCREASING_SURF_DIP_ON_INCREASING_SEQUENCE(VARIABLES) |
| **Argument** | VARIABLES : collection(var−dvar) |
| **Restriction** | required(VARIABLES, var) |

**Purpose**

Succeeds if the values denoting the surface of each occurrence of the DIP_ON_INCREASING_SEQUENCE pattern in the time-series given by the VARIABLES collection are increasing.

An occurrence of the pattern DIP_ON_INCREASING_SEQUENCE is the subsequence which matches the regular expression `<<><<`.

Assume that the occurrence of the pattern DIP_ON_INCREASING_SEQUENCE starts at position $i$ and ends at position $j$. The feature SURF computes the sum of the values from index $i + 2$ to index $j$.

**Example**

$$(\langle 1, 2, 3, 0, 3, 4, 6, 5, 1, 4, 5, 6, 1, 2, 4, 4 \rangle)$$

Figure 3.393 provides an example where the INCREASING_SURF_DIP_ON_INCREASING_SEQUENCE $([1, 2, 3, 0, 3, 4, 6, 5, 1, 4, 5, 6, 1, 2, 4, 4])$ constraint holds.

**Typical**

$|\texttt{VARIABLES}| > 5$

$\texttt{range}(\texttt{VARIABLES.var}) > 2$

Figure 3.393: Illustrating the INCREASING_SURF_DIP_ON_INCREASING_SEQUENCE constraint of the **Example** slot

**Automaton**        Figure 3.394 depicts the automaton associated with the constraint INCREAS-
ING_SURF_DIP_ON_INCREASING_SEQUENCE.

Figure 3.394: Automaton for the INCREASING_SURF_DIP_ON_INCREASING_SEQUENCE
constraint obtained by applying decoration Table 2.36 to the seed transducer of the
DIP_ON_INCREASING_SEQUENCE pattern

## 3.198   INCREASING_SURF_GORGE

▶ ▷ ◁ ◀          **DESCRIPTION**          **AUTOMATON**

$(>|>(=|>)^*>)(<|<(=|<)^*<)$

**Origin**          Based on the GORGE pattern.

**Constraint**      INCREASING_SURF_GORGE(VARIABLES)

**Argument**        VARIABLES  :  collection(var−dvar)

**Restriction**     required(VARIABLES, var)

**Purpose**         Succeeds if the values denoting the surface of each occurrence of the GORGE pattern in
                    the time-series given by the VARIABLES collection are increasing.
                    An occurrence of the pattern GORGE is the *maximal* subsequence which matches the
                    regular expression $(>|>(=|>)^*>)(<|<(=|<)^*<)$.
                    Assume that the occurrence of the pattern GORGE starts at position $i$ and ends at position
                    $j$. The feature SURF computes the sum of the values from index $i + 1$ to index $j$.

**Example**         $(\langle 6, 2, 5, 1, 1, 5, 4, 2, 3, 6, 5, 4, 4, 3, 5, 5 \rangle)$

Figure 3.395 provides an example where the INCREASING_SURF_GORGE
$([6, 2, 5, 1, 1, 5, 4, 2, 3, 6, 5, 4, 4, 3, 5, 5])$ constraint holds.

Figure 3.395: Illustrating the INCREASING_SURF_GORGE constraint of the **Example**
slot

**Typical**          $|\texttt{VARIABLES}| > 2$
             $\texttt{range}(\texttt{VARIABLES.var}) > 1$

**Automaton**    Figure 3.396 depicts the automaton associated with the constraint INCREAS-
ING_SURF_GORGE.

Figure 3.396: Automaton for the INCREASING_SURF_GORGE constraint obtained by
applying decoration Table 2.36 to the seed transducer of the GORGE pattern (transition
$u \rightarrow r$ has the same accumulator update as transition $r \rightarrow u$)

## 3.199 INCREASING_SURF_INCREASING

▶ ▷ ◁ ◀          **DESCRIPTION**          **AUTOMATON**

| | |
|---|---|
| **Origin** | Based on the INCREASING pattern. |
| **Constraint** | INCREASING_SURF_INCREASING(VARIABLES) |
| **Argument** | VARIABLES : collection(var−dvar) |
| **Restriction** | required(VARIABLES, var) |
| **Purpose** | Succeeds if the values denoting the surface of each occurrence of the INCREASING pattern in the time-series given by the VARIABLES collection are increasing. An occurrence of the pattern INCREASING is the subsequence which matches the regular expression <. Assume that the occurrence of the pattern INCREASING starts at position $i$ and ends at position $j$. The feature SURF computes the sum of the values from index $i$ to index $j+1$. |
| **Example** | $(\langle 6, 1, 2, 3, 5, 5, 5, 4, 5, 6, 4, 3, 3, 2, 1, 1 \rangle)$ |

Figure 3.397 provides an example where the INCREASING_SURF_INCREASING $([6, 1, 2, 3, 5, 5, 5, 4, 5, 6, 4, 3, 3, 2, 1, 1])$ constraint holds.

Figure 3.397: Illustrating the INCREASING_SURF_INCREASING constraint of the **Example** slot

**Typical**

$|\text{VARIABLES}| > 1$

$\text{range}(\text{VARIABLES.var}) > 1$

**Automaton**    Figure 3.398 depicts the automaton associated with the constraint INCREAS-ING_SURF_INCREASING.

Figure 3.398: Automaton for the INCREASING_SURF_INCREASING constraint obtained by applying decoration Table 2.36 to the seed transducer of the INCREASING pattern

## 3.200   INCREASING_SURF_INCREASING_SEQUENCE

| **Origin** | Based on the INCREASING_SEQUENCE pattern. |
|---|---|

| **Constraint** | INCREASING_SURF_INCREASING_SEQUENCE(VARIABLES) |
|---|---|

| **Argument** | VARIABLES : collection(var−dvar) |
|---|---|

| **Restriction** | required(VARIABLES, var) |
|---|---|

**Purpose**

Succeeds if the values denoting the surface of each occurrence of the INCREASING_SEQUENCE pattern in the time-series given by the VARIABLES collection are increasing.

An occurrence of the pattern INCREASING_SEQUENCE is the *maximal* subsequence which matches the regular expression $< (< | =)^* < | <$.

Assume that the occurrence of the pattern INCREASING_SEQUENCE starts at position $i$ and ends at position $j$. The feature SURF computes the sum of the values from index $i$ to index $j + 1$.

**Example**

$$(\langle 6, 2, 3, 6, 5, 5, 5, 4, 5, 6, 1, 3, 3, 4, 5, 5 \rangle)$$

Figure 3.399 provides an example where the INCREASING_SURF_INCREASING_SEQUENCE $([6, 2, 3, 6, 5, 5, 5, 4, 5, 6, 1, 3, 3, 4, 5, 5])$ constraint holds.

**Typical**

$|\texttt{VARIABLES}| > 1$

$\texttt{range}(\texttt{VARIABLES.var}) > 1$

Figure 3.399: Illustrating the INCREASING_SURF_INCREASING_SEQUENCE constraint of the **Example** slot

**Automaton**         Figure 3.400 depicts the automaton associated with the constraint INCREAS-
ING_SURF_INCREASING_SEQUENCE.

Figure 3.400: Automaton for the INCREASING_SURF_INCREASING_SEQUENCE con-
straint obtained by applying decoration Table 2.36 to the seed transducer of the IN-
CREASING_SEQUENCE pattern

## 3.201 INCREASING_SURF_INCREASING_TERRACE

▶ ▷ ◁ ◀　　　DESCRIPTION　　　AUTOMATON

**Origin**　　Based on the INCREASING_TERRACE pattern.

**Constraint**　　INCREASING_SURF_INCREASING_TERRACE(VARIABLES)

**Argument**　　VARIABLES : collection(var−dvar)

**Restriction**　　required(VARIABLES, var)

**Purpose**　　Succeeds if the values denoting the surface of each occurrence of the INCREAS-ING_TERRACE pattern in the time-series given by the VARIABLES collection are increasing.

An occurrence of the pattern INCREASING_TERRACE is the *maximal* subsequence which matches the regular expression $<=^+<$.

Assume that the occurrence of the pattern INCREASING_TERRACE starts at position $i$ and ends at position $j$. The feature SURF computes the sum of the values from index $i + 1$ to index $j$.

**Example**　　$(\langle 1, 2, 2, 4, 5, 5, 6, 6, 4, 4, 2, 5, 5, 5, 7, 7 \rangle)$

Figure 3.401 provides an example where the INCREASING_SURF_INCREASING_TERRACE $([1, 2, 2, 4, 5, 5, 6, 6, 4, 4, 2, 5, 5, 5, 7, 7])$ constraint holds.

**Typical**　　$|\text{VARIABLES}| > 3$
range(VARIABLES.var) $> 2$

Figure 3.401: Illustrating the INCREASING_SURF_INCREASING_TERRACE constraint of the **Example** slot

**Automaton**    Figure 3.402 depicts the automaton associated with the constraint INCREAS-
ING_SURF_INCREASING_TERRACE.

Figure 3.402: Automaton for the INCREASING_SURF_INCREASING_TERRACE con-
straint obtained by applying decoration Table 2.36 to the seed transducer of the IN-
CREASING_TERRACE pattern

## 3.202   INCREASING_SURF_INFLEXION

► ▷ ◁ ◄        **DESCRIPTION**          **AUTOMATON**

$< (< | =)^* > | > (> | =)^* <$

| | |
|---|---|
| **Origin** | Based on the INFLEXION pattern. |
| **Constraint** | INCREASING_SURF_INFLEXION(VARIABLES) |
| **Argument** | VARIABLES : collection(var−dvar) |
| **Restriction** | required(VARIABLES, var) |
| **Purpose** | Succeeds if the values denoting the surface of each occurrence of the INFLEXION pattern in the time-series given by the VARIABLES collection are increasing. <br><br> An occurrence of the pattern INFLEXION is the *maximal* subsequence which matches the regular expression $< (< | =)^* > | > (> | =)^* <$. <br><br> Assume that the occurrence of the pattern INFLEXION starts at position $i$ and ends at position $j$. The feature SURF computes the sum of the values from index $i + 1$ to index $j$. |
| **Example** | $(\langle 1, 1, 2, 3, 3, 2, 2, 2, 2, 1, 6, 6, 5, 5, 5 \rangle)$ |

Figure 3.403 provides an example where the INCREASING_SURF_INFLEXION $([1, 1, 2, 3, 3, 2, 2, 2, 2, 1, 6, 6, 5, 5, 5])$ constraint holds.

| | |
|---|---|
| **Typical** | $|\text{VARIABLES}| > 2$ <br> range(VARIABLES.var) $> 1$ |

Figure 3.403: Illustrating the INCREASING_SURF_INFLEXION constraint of the **Example** slot

**Automaton**     Figure 3.404 depicts the automaton associated with the constraint INCREAS-
ING_SURF_INFLEXION.

Figure 3.404: Automaton for the INCREASING_SURF_INFLEXION constraint obtained
by applying decoration Table 2.36 to the seed transducer of the INFLEXION pattern
(transition $r \rightarrow t$ has the same accumulators updates as transition $t \rightarrow r$)

## 3.203 INCREASING_SURF_PEAK

▶ ▷ ◁ ◀          **DESCRIPTION**          **AUTOMATON**

$< (= | <)^* (> | =)^* >$

| | |
|---|---|
| **Origin** | Based on the PEAK pattern. |
| **Constraint** | INCREASING_SURF_PEAK(VARIABLES) |
| **Argument** | VARIABLES : collection(var−dvar) |
| **Restriction** | required(VARIABLES, var) |
| **Purpose** | Succeeds if the values denoting the surface of each occurrence of the PEAK pattern in the time-series given by the VARIABLES collection are increasing.<br><br>An occurrence of the pattern PEAK is the *maximal* subsequence which matches the regular expression $< (= | <)^* (> | =)^* >$.<br><br>Assume that the occurrence of the pattern PEAK starts at position $i$ and ends at position $j$. The feature SURF computes the sum of the values from index $i + 1$ to index $j$. |
| **Example** | $(\langle 7, 5, 5, 1, 2, 3, 2, 2, 3, 4, 5, 2, 6, 6, 6, 1 \rangle)$ |

Figure 3.405 provides an example where the INCREASING_SURF_PEAK $([7, 5, 5, 1, 2, 3, 2, 2, 3, 4, 5, 2, 6, 6, 6, 1])$ constraint holds.

Figure 3.405: Illustrating the INCREASING_SURF_PEAK constraint of the **Example** slot

**Typical**

$|\mathtt{VARIABLES}| > 2$
$\mathtt{range}(\mathtt{VARIABLES.var}) > 1$

**Automaton**        Figure 3.406 depicts the automaton associated with the constraint INCREAS-
ING_SURF_PEAK.

Figure 3.406: Automaton for the INCREASING_SURF_PEAK constraint obtained by ap-
plying decoration Table 2.36 to the seed transducer of the PEAK pattern

## 3.204 INCREASING_SURF_PLAIN

► ▷ ◁ ◄          DESCRIPTION          AUTOMATON

| | |
|---|---|
| **Origin** | Based on the PLAIN pattern. |
| **Constraint** | INCREASING_SURF_PLAIN(VARIABLES) |
| **Argument** | VARIABLES : collection(var−dvar) |
| **Restriction** | required(VARIABLES, var) |
| **Purpose** | Succeeds if the values denoting the surface of each occurrence of the PLAIN pattern in the time-series given by the VARIABLES collection are increasing. An occurrence of the pattern PLAIN is the *maximal* subsequence which matches the regular expression > =* <. Assume that the occurrence of the pattern PLAIN starts at position $i$ and ends at position $j$. The feature SURF computes the sum of the values from index $i+1$ to index $j$. |
| **Example** | $(\langle 3, 6, 6, 3, 4, 5, 5, 4, 6, 6, 7, 5, 5, 6, 3, 2 \rangle)$ |

Figure 3.407 provides an example where the INCREASING_SURF_PLAIN $([3, 6, 6, 3, 4, 5, 5, 4, 6, 6, 7, 5, 5, 6, 3, 2])$ constraint holds.

Figure 3.407: Illustrating the INCREASING_SURF_PLAIN constraint of the **Example** slot

**Typical**

$|\text{VARIABLES}| > 2$

$\text{range}(\text{VARIABLES.var}) > 1$

**Automaton**          Figure 3.408 depicts the automaton associated with the constraint INCREAS-
ING_SURF_PLAIN.

Figure 3.408: Automaton for the INCREASING_SURF_PLAIN constraint obtained by
applying decoration Table 2.36 to the seed transducer of the PLAIN pattern

## 3.205   INCREASING_SURF_PLATEAU

► ▷ ◁ ◀        **DESCRIPTION**          **AUTOMATON**

| | |
|---|---|
| **Origin** | Based on the PLATEAU pattern. |
| **Constraint** | INCREASING_SURF_PLATEAU(VARIABLES) |
| **Argument** | VARIABLES : collection(var−dvar) |
| **Restriction** | required(VARIABLES, var) |
| **Purpose** | Succeeds if the values denoting the surface of each occurrence of the PLATEAU pattern in the time-series given by the VARIABLES collection are increasing. <br><br> An occurrence of the pattern PLATEAU is the *maximal* subsequence which matches the regular expression $<=^*>$. <br><br> Assume that the occurrence of the pattern PLATEAU starts at position $i$ and ends at position $j$. The feature SURF computes the sum of the values from index $i+1$ to index $j$. |
| **Example** | $(\langle 7, 5, 2, 3, 1, 2, 2, 4, 3, 3, 4, 5, 5, 2, 2, 5\rangle)$ |

Figure 3.409 provides an example where the INCREASING_SURF_PLATEAU $([7, 5, 2, 3, 1, 2, 2, 4, 3, 3, 4, 5, 5, 2, 2, 5])$ constraint holds.

| | |
|---|---|
| **Typical** | $\|\text{VARIABLES}\| > 2$ <br> range(VARIABLES.var) $> 1$ |

Figure 3.409: Illustrating the INCREASING_SURF_PLATEAU constraint of the **Example** slot

**Automaton**     Figure 3.410 depicts the automaton associated with the constraint INCREAS-
ING_SURF_PLATEAU.

Figure 3.410: Automaton for the INCREASING_SURF_PLATEAU constraint obtained by
applying decoration Table 2.36 to the seed transducer of the PLATEAU pattern

## 3.206 INCREASING_SURF_PROPER_PLAIN

▶  ▷  ◁  ◀        DESCRIPTION        AUTOMATON

**Origin**          Based on the PROPER_PLAIN pattern.

**Constraint**      INCREASING_SURF_PROPER_PLAIN(VARIABLES)

**Argument**        VARIABLES  :  collection(var−dvar)

**Restriction**     required(VARIABLES, var)

**Purpose**         Succeeds if the values denoting the surface of each occurrence of the PROPER_PLAIN pattern in the time-series given by the VARIABLES collection are increasing.
An occurrence of the pattern PROPER_PLAIN is the *maximal* subsequence which matches the regular expression $> =^+ <$.
Assume that the occurrence of the pattern PROPER_PLAIN starts at position $i$ and ends at position $j$. The feature SURF computes the sum of the values from index $i + 1$ to index $j$.

**Example**         $(\langle 5, 3, 3, 5, 6, 5, 4, 4, 7, 3, 6, 5, 5, 5, 7, 2 \rangle)$

Figure 3.411 provides an example where the INCREASING_SURF_PROPER_PLAIN $([5, 3, 3, 5, 6, 5, 4, 4, 7, 3, 6, 5, 5, 5, 7, 2])$ constraint holds.

**Typical**         $|VARIABLES| > 3$
range(VARIABLES.var) > 1

Figure 3.411: Illustrating the INCREASING_SURF_PROPER_PLAIN constraint of the **Example** slot

**Automaton**   Figure 3.412 depicts the automaton associated with the constraint INCREAS-
ING_SURF_PROPER_PLAIN.

Figure 3.412: Automaton for the INCREASING_SURF_PROPER_PLAIN constraint ob-
tained by applying decoration Table 2.36 to the seed transducer of the PROPER_PLAIN
pattern

## 3.207 INCREASING_SURF_PROPER_PLATEAU

▶ ▷ ◁ ◀   **DESCRIPTION**   **AUTOMATON**

| | |
|---|---|
| **Origin** | Based on the PROPER_PLATEAU pattern. |
| **Constraint** | INCREASING_SURF_PROPER_PLATEAU(VARIABLES) |
| **Argument** | VARIABLES : collection(var−dvar) |
| **Restriction** | required(VARIABLES, var) |

**Purpose**

> Succeeds if the values denoting the surface of each occurrence of the PROPER_PLATEAU pattern in the time-series given by the VARIABLES collection are increasing.
>
> An occurrence of the pattern PROPER_PLATEAU is the *maximal* subsequence which matches the regular expression $< =^+ >$.
>
> Assume that the occurrence of the pattern PROPER_PLATEAU starts at position $i$ and ends at position $j$. The feature SURF computes the sum of the values from index $i+1$ to index $j$.

**Example**

$$(\langle 7, 1, 3, 3, 2, 5, 1, 4, 4, 3, 2, 3, 5, 5, 5, 3 \rangle)$$

Figure 3.413 provides an example where the INCREASING_SURF_PROPER_PLATEAU ([7, 1, 3, 3, 2, 5, 1, 4, 4, 3, 2, 3, 5, 5, 5, 3]) constraint holds.

**Typical**

$|VARIABLES| > 3$
range(VARIABLES.var) > 1

Figure 3.413: Illustrating the INCREASING_SURF_PROPER_PLATEAU constraint of the **Example** slot

**Automaton**          Figure 3.414 depicts the automaton associated with the constraint INCREAS-ING_SURF_PROPER_PLATEAU.

Figure 3.414: Automaton for the INCREASING_SURF_PROPER_PLATEAU constraint obtained by applying decoration Table 2.36 to the seed transducer of the PROPER_PLATEAU pattern

## 3.208  INCREASING_SURF_STEADY

► ▷ ◁ ◄        **DESCRIPTION**        **AUTOMATON**

| | |
|---|---|
| **Origin** | Based on the STEADY pattern. |
| **Constraint** | INCREASING_SURF_STEADY(VARIABLES) |
| **Argument** | VARIABLES : collection(var−dvar) |
| **Restriction** | required(VARIABLES, var) |
| **Purpose** | Succeeds if the values denoting the surface of each occurrence of the STEADY pattern in the time-series given by the VARIABLES collection are increasing.<br>An occurrence of the pattern STEADY is the subsequence which matches the regular expression =.<br>Assume that the occurrence of the pattern STEADY starts at position $i$ and ends at position $j$. The feature SURF computes the sum of the values from index $i$ to index $j + 1$. |
| **Example** | $(\langle 1, 1, 7, 3, 3, 5, 5, 5, 6, 5, 5, 5, 7, 2, 6, 6 \rangle)$ |

Figure 3.415 provides an example where the INCREASING_SURF_STEADY ($[1, 1, 7, 3, 3, 5, 5, 5, 6, 5, 5, 5, 7, 2, 6, 6]$) constraint holds.

| | |
|---|---|
| **Typical** | $|\texttt{VARIABLES}| > 1$ |

Figure 3.415: Illustrating the INCREASING_SURF_STEADY constraint of the **Example** slot

**Automaton**    Figure 3.416 depicts the automaton associated with the constraint INCREAS-
ING_SURF_STEADY.

$$\left\{ \begin{array}{l} C \leftarrow -\infty \\ D \leftarrow 0 \\ F \leftarrow -\infty \\ R \leftarrow 1 \end{array} \right\}$$

$$\left\{ \begin{array}{l} C \leftarrow D + \text{VAR}_i + \text{VAR}_{i+1} \\ D \leftarrow 0 \\ F \leftarrow D + \text{VAR}_i + \text{VAR}_{i+1} \\ R \leftarrow R \wedge (F \leq D + \text{VAR}_i + \text{VAR}_{i+1}) \end{array} \right\}$$

$$R \wedge (F \leq C)$$

Figure 3.416: Automaton for the INCREASING_SURF_STEADY constraint obtained by
applying decoration Table 2.36 to the seed transducer of the STEADY pattern

# 3.209 INCREASING_SURF_STEADY_SEQUENCE

       **DESCRIPTION**          **AUTOMATON**

**Origin**          Based on the STEADY_SEQUENCE pattern.

**Constraint**      INCREASING_SURF_STEADY_SEQUENCE(VARIABLES)

**Argument**        VARIABLES : collection(var−dvar)

**Restriction**     required(VARIABLES, var)

**Purpose**         Succeeds if the values denoting the surface of each occurrence of the
                    STEADY_SEQUENCE pattern in the time-series given by the VARIABLES collection
                    are increasing.
                    An occurrence of the pattern STEADY_SEQUENCE is the *maximal* subsequence which
                    matches the regular expression $=^+$.
                    Assume that the occurrence of the pattern STEADY_SEQUENCE starts at position $i$ and
                    ends at position $j$. The feature SURF computes the sum of the values from index $i$ to
                    index $j + 1$.

**Example**         $(\langle 6, 2, 2, 3, 5, 6, 4, 4, 2, 3, 5, 5, 5, 2, 3, 4 \rangle)$

Figure 3.417 provides an example where the INCREASING_SURF_STEADY_SEQUENCE
($[6, 2, 2, 3, 5, 6, 4, 4, 2, 3, 5, 5, 5, 2, 3, 4]$) constraint holds.

Figure 3.417: Illustrating the INCREASING_SURF_STEADY_SEQUENCE constraint of
the **Example** slot

**Typical**            $|\text{VARIABLES}| > 1$

**Automaton**            Figure 3.418 depicts the automaton associated with the constraint INCREAS-ING_SURF_STEADY_SEQUENCE.

Figure 3.418: Automaton for the INCREASING_SURF_STEADY_SEQUENCE constraint obtained by applying decoration Table 2.36 to the seed transducer of the STEADY_SEQUENCE pattern

# 3.210   INCREASING_SURF_STRICTLY_DECREASING_SEQUENCE

▶ ▷ ◁ ◀      **DESCRIPTION**      **AUTOMATON**

| | |
|---|---|
| **Origin** | Based on the STRICTLY_DECREASING_SEQUENCE pattern. |
| **Constraint** | INCREASING_SURF_STRICTLY_DECREASING_SEQUENCE(VARIABLES) |
| **Argument** | VARIABLES : collection(var−dvar) |
| **Restriction** | required(VARIABLES, var) |

**Purpose**

Succeeds if the values denoting the surface of each occurrence of the STRICTLY_DECREASING_SEQUENCE pattern in the time-series given by the VARIABLES collection are increasing.

An occurrence of the pattern STRICTLY_DECREASING_SEQUENCE is the *maximal* sub-sequence which matches the regular expression $>^+$.

Assume that the occurrence of the pattern STRICTLY_DECREASING_SEQUENCE starts at position $i$ and ends at position $j$. The feature SURF computes the sum of the values from index $i$ to index $j + 1$.

**Example**

$(\langle 1, 4, 3, 2, 7, 3, 3, 7, 3, 3, 3, 4, 4, 5, 7, 5 \rangle)$

Figure 3.419 provides an example where the INCREASING_SURF_STRICTLY_DECREASING_SEQUENCE $([1, 4, 3, 2, 7, 3, 3, 7, 3, 3, 3, 4, 4, 5, 7, 5])$ constraint holds.

**Typical**

$|\text{VARIABLES}| > 1$

$\text{range}(\text{VARIABLES.var}) > 1$

Figure 3.419: Illustrating the INCREASING_SURF_STRICTLY_DECREASING_SEQUENCE constraint of the **Example** slot

**Automaton**    Figure 3.420 depicts the automaton associated with the constraint INCREAS-
ING_SURF_STRICTLY_DECREASING_SEQUENCE.

Figure 3.420: Automaton for the INCREASING_SURF_STRICTLY_DECREASING_SEQUENCE
constraint obtained by applying decoration Table 2.36 to the seed transducer of the
STRICTLY_DECREASING_SEQUENCE pattern

## 3.211 INCREASING_SURF_STRICTLY_INCREASING_SEQUENCE

▶ ▷ ◁ ◀  **DESCRIPTION**    **AUTOMATON**

| | |
|---|---|
| **Origin** | Based on the STRICTLY_INCREASING_SEQUENCE pattern. |
| **Constraint** | INCREASING_SURF_STRICTLY_INCREASING_SEQUENCE(VARIABLES) |
| **Argument** | VARIABLES : collection(var−dvar) |
| **Restriction** | required(VARIABLES, var) |

**Purpose**

Succeeds if the values denoting the surface of each occurrence of the STRICTLY_INCREASING_SEQUENCE pattern in the time-series given by the VARIABLES collection are increasing.

An occurrence of the pattern STRICTLY_INCREASING_SEQUENCE is the *maximal* subsequence which matches the regular expression $<^+$.

Assume that the occurrence of the pattern STRICTLY_INCREASING_SEQUENCE starts at position $i$ and ends at position $j$. The feature SURF computes the sum of the values from index $i$ to index $j + 1$.

**Example**

$$(\langle 6, 2, 3, 6, 5, 5, 4, 5, 6, 1, 2, 3, 4, 5, 5 \rangle)$$

Figure 3.421 provides an example where the INCREASING_SURF_STRICTLY_INCREASING_SEQUENCE $([6, 2, 3, 6, 5, 5, 4, 5, 6, 1, 2, 3, 4, 5, 5])$ constraint holds.

**Typical**

$|\mathtt{VARIABLES}| > 1$

$\mathtt{range}(\mathtt{VARIABLES.var}) > 1$

Figure 3.421: Illustrating the INCREASING_SURF_STRICTLY_INCREASING_SEQUENCE constraint of the **Example** slot

**Automaton**  Figure 3.422 depicts the automaton associated with the constraint INCREAS-ING_SURF_STRICTLY_INCREASING_SEQUENCE.

Figure 3.422: Automaton for the INCREASING_SURF_STRICTLY_INCREASING_SEQUENCE constraint obtained by applying decoration Table 2.36 to the seed transducer of the STRICTLY_INCREASING_SEQUENCE pattern

## 3.212  INCREASING_SURF_SUMMIT

► ▷ ◁ ◀          DESCRIPTION          AUTOMATON

$(< | < (= | <)^* <)(> | > (= | >)^* >)$

| | |
|---|---|
| **Origin** | Based on the SUMMIT pattern. |
| **Constraint** | INCREASING_SURF_SUMMIT(VARIABLES) |
| **Argument** | VARIABLES : collection(var−dvar) |
| **Restriction** | required(VARIABLES, var) |
| **Purpose** | Succeeds if the values denoting the surface of each occurrence of the SUMMIT pattern in the time-series given by the VARIABLES collection are increasing. <br> An occurrence of the pattern SUMMIT is the *maximal* subsequence which matches the regular expression $(< | < (= | <)^* <)(> | > (= | >)^* >)$. <br> Assume that the occurrence of the pattern SUMMIT starts at position $i$ and ends at position $j$. The feature SURF computes the sum of the values from index $i + 1$ to index $j$. |
| **Example** | $(\langle 1, 5, 2, 1, 6, 6, 2, 3, 5, 4, 1, 4, 6, 4, 3, 2 \rangle)$ |

Figure 3.423 provides an example where the INCREASING_SURF_SUMMIT ($[1, 5, 2, 1, 6, 6, 2, 3, 5, 4, 1, 4, 6, 4, 3, 2]$) constraint holds.

| | |
|---|---|
| **Typical** | $|\texttt{VARIABLES}| > 2$ <br> range(VARIABLES.var) $> 1$ |

Figure 3.423: Illustrating the INCREASING_SURF_SUMMIT constraint of the **Example** slot

**Automaton**     Figure 3.424 depicts the automaton associated with the constraint INCREAS-
ING_SURF_SUMMIT.

Figure 3.424: Automaton for the INCREASING_SURF_SUMMIT constraint obtained by
applying decoration Table 2.36 to the seed transducer of the SUMMIT pattern (transition
$u \rightarrow r$ has the same accumulator update as transition $r \rightarrow u$)

## 3.213 INCREASING_SURF_VALLEY

▶ ▷ ◁ ◀        **DESCRIPTION**              **AUTOMATON**

$> (= | >)^* (< | =)^* <$

| | |
|---|---|
| **Origin** | Based on the VALLEY pattern. |
| **Constraint** | INCREASING_SURF_VALLEY(VARIABLES) |
| **Argument** | VARIABLES : collection(var−dvar) |
| **Restriction** | required(VARIABLES, var) |
| **Purpose** | Succeeds if the values denoting the surface of each occurrence of the VALLEY pattern in the time-series given by the VARIABLES collection are increasing.<br><br>An occurrence of the pattern VALLEY is the *maximal* subsequence which matches the regular expression $> (= | >)^* (< | =)^* <$.<br><br>Assume that the occurrence of the pattern VALLEY starts at position $i$ and ends at position $j$. The feature SURF computes the sum of the values from index $i + 1$ to index $j$. |

| | |
|---|---|
| **Example** | $(\langle 7, 2, 2, 6, 3, 4, 5, 6, 6, 4, 4, 6, 7, 3, 3, 1 \rangle)$ |

Figure 3.425 provides an example where the INCREASING_SURF_VALLEY ([7, 2, 2, 6, 3, 4, 5, 6, 6, 4, 4, 6, 7, 3, 3, 1]) constraint holds.

| | |
|---|---|
| **Typical** | $|$VARIABLES$| > 2$<br>range(VARIABLES.var) $> 1$ |

Figure 3.425: Illustrating the INCREASING_SURF_VALLEY constraint of the **Example** slot

**Automaton**     Figure 3.426 depicts the automaton associated with the constraint INCREAS-ING_SURF_VALLEY.

Figure 3.426: Automaton for the INCREASING_SURF_VALLEY constraint obtained by applying decoration Table 2.36 to the seed transducer of the VALLEY pattern

## 3.214  INCREASING_SURF_ZIGZAG

        **DESCRIPTION**              **AUTOMATON**

| | |
|---|---|
| **Origin** | Based on the ZIGZAG pattern. |
| **Constraint** | INCREASING_SURF_ZIGZAG(VARIABLES) |
| **Argument** | VARIABLES : collection(var−dvar) |
| **Restriction** | required(VARIABLES, var) |

**Purpose**

Succeeds if the values denoting the surface of each occurrence of the ZIGZAG pattern in the time-series given by the VARIABLES collection are increasing.

An occurrence of the pattern ZIGZAG is the *maximal* subsequence which matches the regular expression $(\texttt{<>})^+(\texttt{<} \mid \texttt{<>}) \mid (\texttt{><})^+(\texttt{>} \mid \texttt{><})$.

Assume that the occurrence of the pattern ZIGZAG starts at position $i$ and ends at position $j$. The feature SURF computes the sum of the values from index $i + 1$ to index $j$.

**Example**

$(\langle 7, 7, 2, 5, 3, 4, 6, 3, 6, 1, 1, 5, 4, 7, 5, 6 \rangle)$

Figure 3.427 provides an example where the INCREASING_SURF_ZIGZAG $([7, 7, 2, 5, 3, 4, 6, 3, 6, 1, 1, 5, 4, 7, 5, 6])$ constraint holds.

**Typical**

$|\texttt{VARIABLES}| > 3$

$\texttt{range}(\texttt{VARIABLES.var}) > 1$

Figure 3.427: Illustrating the INCREASING_SURF_ZIGZAG constraint of the **Example** slot

**Automaton**          Figure 3.428 depicts the automaton associated with the constraint INCREAS-
                       ING_SURF_ZIGZAG.

Figure 3.428: Automaton for the INCREASING_SURF_ZIGZAG constraint obtained by applying decoration Table 2.36 to the seed transducer of the ZIGZAG pattern; (1) missing transitions from $a, b, c, d, e, f$ to $s$ are labelled by $=$; (2) on transitions from $b, c, e, f$ to $s$ the accumulator $D$ is reset to its initial value; (3) on transitions from $c, f$ to $s$ the accumulator $F$ is reset to $C$, and the accumulator $R$ is updated wrt $C$ and $F$

## 3.215 INCREASING_WIDTH_DECREASING_SEQUENCE

▶ ▷ ◁ ◀          DESCRIPTION          AUTOMATON

| **Origin** | Based on the DECREASING_SEQUENCE pattern. |
|---|---|

| **Constraint** | INCREASING_WIDTH_DECREASING_SEQUENCE(VARIABLES) |
|---|---|

| **Argument** | VARIABLES : collection(var−dvar) |
|---|---|

| **Restriction** | required(VARIABLES, var) |
|---|---|

| **Purpose** | Succeeds if the values denoting the width of each occurrence of the DECREASING_SEQUENCE pattern in the time-series given by the VARIABLES collection are increasing.<br>An occurrence of the pattern DECREASING_SEQUENCE is the *maximal* subsequence which matches the regular expression $> (> \mid =)^* > \mid >$.<br>Assume that the occurrence of the pattern DECREASING_SEQUENCE starts at position $i$ and ends at position $j$. The feature WIDTH computes the value $j - i + 2$. |
|---|---|

| **Example** | $(\langle 1, 6, 6, 4, 4, 5, 6, 6, 5, 2, 3, 3, 2, 2, 1, 1 \rangle)$ |
|---|---|

Figure   3.429   provides   an   example   where   the   INCREAS-
ING_WIDTH_DECREASING_SEQUENCE   ([1, 6, 6, 4, 4, 5, 6, 6, 5, 2, 3, 3, 2, 2, 1, 1])   con-
straint holds.

| **Typical** | $\lvert$VARIABLES$\rvert > 1$<br>range(VARIABLES.var) $> 1$ |
|---|---|

Figure 3.429: Illustrating the INCREASING_WIDTH_DECREASING_SEQUENCE constraint of the **Example** slot

**Automaton**     Figure 3.430 depicts the automaton associated with the constraint INCREAS-
ING_WIDTH_DECREASING_SEQUENCE.

Figure 3.430: Automaton for the INCREASING_WIDTH_DECREASING_SEQUENCE con-
straint obtained by applying decoration Table 2.36 to the seed transducer of the DE-
CREASING_SEQUENCE pattern

## 3.216 INCREASING_WIDTH_DECREASING_TERRACE

          DESCRIPTION          AUTOMATON

| | |
|---|---|
| **Origin** | Based on the DECREASING_TERRACE pattern. |
| **Constraint** | INCREASING_WIDTH_DECREASING_TERRACE(VARIABLES) |
| **Argument** | VARIABLES : collection(var−dvar) |
| **Restriction** | required(VARIABLES, var) |

**Purpose**

Succeeds if the values denoting the width of each occurrence of the DECREAS-ING_TERRACE pattern in the time-series given by the VARIABLES collection are increas-ing.

An occurrence of the pattern DECREASING_TERRACE is the *maximal* subsequence which matches the regular expression $>=^+>$.

Assume that the occurrence of the pattern DECREASING_TERRACE starts at position $i$ and ends at position $j$. The feature WIDTH computes the value $j - i$.

**Example**

$$(\langle 5, 2, 2, 1, 5, 4, 3, 3, 2, 4, 4, 6, 5, 5, 5, 4 \rangle)$$

Figure 3.431 provides an example where the INCREAS-ING_WIDTH_DECREASING_TERRACE $([5, 2, 2, 1, 5, 4, 3, 3, 2, 4, 4, 6, 5, 5, 5, 4])$ constraint holds.

**Typical**

$|\text{VARIABLES}| > 3$

range(VARIABLES.var) $> 2$

Figure 3.431: Illustrating the INCREASING_WIDTH_DECREASING_TERRACE constraint of the **Example** slot

**Automaton**          Figure 3.432 depicts the automaton associated with the constraint INCREAS-
ING_WIDTH_DECREASING_TERRACE.

Figure 3.432: Automaton for the INCREASING_WIDTH_DECREASING_TERRACE con-
straint obtained by applying decoration Table 2.36 to the seed transducer of the DE-
CREASING_TERRACE pattern

## 3.217 INCREASING_WIDTH_GORGE

► ▷ ◁ ◄          DESCRIPTION          AUTOMATON

**Origin**          Based on the GORGE pattern.

**Constraint**      INCREASING_WIDTH_GORGE(VARIABLES)

**Argument**        VARIABLES : collection(var−dvar)

**Restriction**     required(VARIABLES, var)

**Purpose**

Succeeds if the values denoting the width of each occurrence of the GORGE pattern in the time-series given by the VARIABLES collection are increasing.

An occurrence of the pattern GORGE is the *maximal* subsequence which matches the regular expression $(> | > (= | >)^* >)(< | < (= | <)^* <)$.

Assume that the occurrence of the pattern GORGE starts at position $i$ and ends at position $j$. The feature WIDTH computes the value $j - i$.

**Example**         $(\langle 6, 2, 5, 1, 1, 5, 4, 2, 3, 6, 5, 4, 4, 3, 5, 5 \rangle)$

Figure 3.433 provides an example where the INCREASING_WIDTH_GORGE $([6, 2, 5, 1, 1, 5, 4, 2, 3, 6, 5, 4, 4, 3, 5, 5])$ constraint holds.

Figure 3.433: Illustrating the INCREASING_WIDTH_GORGE constraint of the **Example** slot

**Typical**

$|\texttt{VARIABLES}| > 2$
$\texttt{range}(\texttt{VARIABLES.var}) > 1$

**Automaton**     Figure 3.434 depicts the automaton associated with the constraint INCREAS-
ING_WIDTH_GORGE.

Figure 3.434: Automaton for the INCREASING_WIDTH_GORGE constraint obtained by
applying decoration Table 2.36 to the seed transducer of the GORGE pattern (transition
$u \to r$ has the same accumulator update as transition $r \to u$)

# 3.218 INCREASING_WIDTH_INCREASING_SEQUENCE

▶ ▷ ◁ ◀        **DESCRIPTION**        **AUTOMATON**

| **Origin** | Based on the INCREASING_SEQUENCE pattern. |

| **Constraint** | INCREASING_WIDTH_INCREASING_SEQUENCE(VARIABLES) |

| **Argument** | VARIABLES : collection(var−dvar) |

| **Restriction** | required(VARIABLES, var) |

| **Purpose** | Succeeds if the values denoting the width of each occurrence of the INCREASING_SEQUENCE pattern in the time-series given by the VARIABLES collection are increasing.<br>An occurrence of the pattern INCREASING_SEQUENCE is the *maximal* subsequence which matches the regular expression $< (< | =)^* < | <$.<br>Assume that the occurrence of the pattern INCREASING_SEQUENCE starts at position $i$ and ends at position $j$. The feature WIDTH computes the value $j - i + 2$. |

| **Example** | $(\langle 6, 2, 3, 6, 5, 5, 4, 5, 6, 1, 3, 3, 4, 5, 5 \rangle)$ |

Figure 3.435 provides an example where the INCREASING_WIDTH_INCREASING_SEQUENCE ([6, 2, 3, 6, 5, 5, 4, 5, 6, 1, 3, 3, 4, 5, 5]) constraint holds.

| **Typical** | $|\text{VARIABLES}| > 1$<br>range(VARIABLES.var) $> 1$ |

Figure 3.435: Illustrating the INCREASING_WIDTH_INCREASING_SEQUENCE constraint of the **Example** slot

**Automaton**    Figure 3.436 depicts the automaton associated with the constraint INCREAS-
ING_WIDTH_INCREASING_SEQUENCE.

Figure 3.436: Automaton for the INCREASING_WIDTH_INCREASING_SEQUENCE con-
straint obtained by applying decoration Table 2.36 to the seed transducer of the IN-
CREASING_SEQUENCE pattern

## 3.219 INCREASING_WIDTH_INCREASING_TERRACE

▶ ▷ ◁ ◀          **DESCRIPTION**          **AUTOMATON**

| | |
|---|---|
| **Origin** | Based on the INCREASING_TERRACE pattern. |
| **Constraint** | INCREASING_WIDTH_INCREASING_TERRACE(VARIABLES) |
| **Argument** | VARIABLES : collection(var−dvar) |
| **Restriction** | required(VARIABLES, var) |
| **Purpose** | Succeeds if the values denoting the width of each occurrence of the INCREASING_TERRACE pattern in the time-series given by the VARIABLES collection are increasing. |

An occurrence of the pattern INCREASING_TERRACE is the *maximal* subsequence which matches the regular expression $<=^+<$.

Assume that the occurrence of the pattern INCREASING_TERRACE starts at position $i$ and ends at position $j$. The feature WIDTH computes the value $j - i$.

| | |
|---|---|
| **Example** | $(\langle 1, 2, 2, 4, 5, 5, 6, 6, 4, 4, 2, 5, 5, 5, 7, 7 \rangle)$ |

Figure 3.437 provides an example where the INCREASING_WIDTH_INCREASING_TERRACE $([1, 2, 2, 4, 5, 5, 6, 6, 4, 4, 2, 5, 5, 5, 7, 7])$ constraint holds.

| | |
|---|---|
| **Typical** | $\|\text{VARIABLES}\| > 3$ |
| | range(VARIABLES.var) $> 2$ |

Figure 3.437: Illustrating the INCREASING_WIDTH_INCREASING_TERRACE constraint of the **Example** slot

**Automaton**       Figure 3.438 depicts the automaton associated with the constraint INCREAS-
ING_WIDTH_INCREASING_TERRACE.

Figure 3.438: Automaton for the INCREASING_WIDTH_INCREASING_TERRACE constraint obtained by applying decoration Table 2.36 to the seed transducer of the INCREASING_TERRACE pattern

# 3.220 INCREASING_WIDTH_INFLEXION

► ▷ ◁ ◄       **DESCRIPTION**        **AUTOMATON**

$< (< | =)^* > | > (> | =)^* <$

| | |
|---|---|
| **Origin** | Based on the INFLEXION pattern. |
| **Constraint** | INCREASING_WIDTH_INFLEXION(VARIABLES) |
| **Argument** | VARIABLES : collection(var−dvar) |
| **Restriction** | required(VARIABLES, var) |
| **Purpose** | Succeeds if the values denoting the width of each occurrence of the INFLEXION pattern in the time-series given by the VARIABLES collection are increasing. <br><br> An occurrence of the pattern INFLEXION is the *maximal* subsequence which matches the regular expression $< (< | =)^* > | > (> | =)^* <$. <br><br> Assume that the occurrence of the pattern INFLEXION starts at position $i$ and ends at position $j$. The feature WIDTH computes the value $j − i$. |
| **Example** | $(\langle 1, 1, 2, 1, 3, 3, 6, 4, 3, 3, 2, 3, 4, 5, 5, 6 \rangle)$ |

Figure 3.439 provides an example where the INCREASING_WIDTH_INFLEXION ($[1, 1, 2, 1, 3, 3, 6, 4, 3, 3, 2, 3, 4, 5, 5, 6]$) constraint holds.

Figure 3.439: Illustrating the INCREASING_WIDTH_INFLEXION constraint of the **Example** slot

**Typical**

$|\text{VARIABLES}| > 2$
$\text{range}(\text{VARIABLES.var}) > 1$

**Automaton**   Figure 3.440 depicts the automaton associated with the constraint INCREAS-ING_WIDTH_INFLEXION.

Figure 3.440: Automaton for the INCREASING_WIDTH_INFLEXION constraint obtained by applying decoration Table 2.36 to the seed transducer of the INFLEXION pattern (transition $r \rightarrow t$ has the same accumulators updates as transition $t \rightarrow r$)

## 3.221  INCREASING_WIDTH_PEAK

▶ ▷ ◁ ◀          **DESCRIPTION**          **AUTOMATON**

| | |
|---|---|
| **Origin** | Based on the PEAK pattern. |
| **Constraint** | INCREASING_WIDTH_PEAK(VARIABLES) |
| **Argument** | VARIABLES : collection(var−dvar) |
| **Restriction** | required(VARIABLES, var) |
| **Purpose** | Succeeds if the values denoting the width of each occurrence of the PEAK pattern in the time-series given by the VARIABLES collection are increasing. <br> An occurrence of the pattern PEAK is the *maximal* subsequence which matches the regular expression $< (= \mid <)^* (> \mid =)^* >$. <br> Assume that the occurrence of the pattern PEAK starts at position $i$ and ends at position $j$. The feature WIDTH computes the value $j - i$. |
| **Example** | $(\langle 7, 5, 5, 1, 2, 3, 2, 2, 3, 4, 5, 2, 6, 6, 6, 1 \rangle)$ |

Figure 3.441 provides an example where the INCREASING_WIDTH_PEAK ([7, 5, 5, 1, 2, 3, 2, 2, 3, 4, 5, 2, 6, 6, 6, 1]) constraint holds.

| | |
|---|---|
| **Typical** | $\lvert \texttt{VARIABLES} \rvert > 2$ <br> range(VARIABLES.var) > 1 |

Figure 3.441: Illustrating the INCREASING_WIDTH_PEAK constraint of the **Example** slot

**Automaton**       Figure 3.442 depicts the automaton associated with the constraint INCREAS-
ING_WIDTH_PEAK.

Figure 3.442: Automaton for the INCREASING_WIDTH_PEAK constraint obtained by
applying decoration Table 2.36 to the seed transducer of the PEAK pattern

## 3.222 INCREASING_WIDTH_PLAIN

► ▷ ◁ ◄     DESCRIPTION          AUTOMATON

**Origin**          Based on the PLAIN pattern.

**Constraint**      INCREASING_WIDTH_PLAIN(VARIABLES)

**Argument**        VARIABLES : collection(var−dvar)

**Restriction**     required(VARIABLES, var)

**Purpose**         Succeeds if the values denoting the width of each occurrence of the PLAIN pattern in the time-series given by the VARIABLES collection are increasing.

An occurrence of the pattern PLAIN is the *maximal* subsequence which matches the regular expression >=* <.

Assume that the occurrence of the pattern PLAIN starts at position $i$ and ends at position $j$. The feature WIDTH computes the value $j - i$.

**Example**         $(\langle 3, 6, 6, 3, 4, 5, 5, 4, 6, 6, 7, 5, 5, 6, 3, 2 \rangle)$

Figure 3.443 provides an example where the INCREASING_WIDTH_PLAIN ([3, 6, 6, 3, 4, 5, 5, 4, 6, 6, 7, 5, 5, 6, 3, 2]) constraint holds.

Figure 3.443: Illustrating the INCREASING_WIDTH_PLAIN constraint of the **Example** slot

**Typical**

$|\text{VARIABLES}| > 2$
$\text{range}(\text{VARIABLES.var}) > 1$

**Automaton**          Figure 3.444 depicts the automaton associated with the constraint INCREAS-
                       ING_WIDTH_PLAIN.

Figure 3.444: Automaton for the INCREASING_WIDTH_PLAIN constraint obtained by
applying decoration Table 2.36 to the seed transducer of the PLAIN pattern

# 3.223  INCREASING_WIDTH_PLATEAU

► ▷ ◁ ◄        DESCRIPTION        AUTOMATON

| | |
|---|---|
| **Origin** | Based on the PLATEAU pattern. |
| **Constraint** | INCREASING_WIDTH_PLATEAU(VARIABLES) |
| **Argument** | VARIABLES : collection(var−dvar) |
| **Restriction** | required(VARIABLES, var) |
| **Purpose** | Succeeds if the values denoting the width of each occurrence of the PLATEAU pattern in the time-series given by the VARIABLES collection are increasing.<br><br>An occurrence of the pattern PLATEAU is the *maximal* subsequence which matches the regular expression $<=^* >$.<br><br>Assume that the occurrence of the pattern PLATEAU starts at position $i$ and ends at position $j$. The feature WIDTH computes the value $j − i$. |
| **Example** | $(\langle 7, 5, 2, 3, 1, 2, 2, 4, 3, 3, 4, 5, 5, 2, 2, 5 \rangle)$ |

Figure 3.445 provides an example where the INCREASING_WIDTH_PLATEAU $([7, 5, 2, 3, 1, 2, 2, 4, 3, 3, 4, 5, 5, 2, 2, 5])$ constraint holds.

| | |
|---|---|
| **Typical** | $\lvert\texttt{VARIABLES}\rvert > 2$<br>range(VARIABLES.var) $> 1$ |

Figure 3.445: Illustrating the INCREASING_WIDTH_PLATEAU constraint of the **Example** slot

**Automaton**      Figure 3.446 depicts the automaton associated with the constraint INCREAS-
                   ING_WIDTH_PLATEAU.

Figure 3.446: Automaton for the INCREASING_WIDTH_PLATEAU constraint obtained
by applying decoration Table 2.36 to the seed transducer of the PLATEAU pattern

## 3.224  INCREASING_WIDTH_PROPER_PLAIN

▶ ▷ ◁ ◀          **DESCRIPTION**          **AUTOMATON**

**Origin**          Based on the PROPER_PLAIN pattern.

**Constraint**      INCREASING_WIDTH_PROPER_PLAIN(VARIABLES)

**Argument**        VARIABLES : collection(var−dvar)

**Restriction**     required(VARIABLES, var)

**Purpose**         Succeeds if the values denoting the width of each occurrence of the PROPER_PLAIN pattern in the time-series given by the VARIABLES collection are increasing.
An occurrence of the pattern PROPER_PLAIN is the *maximal* subsequence which matches the regular expression $> =^+ <$.
Assume that the occurrence of the pattern PROPER_PLAIN starts at position $i$ and ends at position $j$. The feature WIDTH computes the value $j - i$.

**Example**         $(\langle 5, 3, 3, 5, 6, 5, 4, 4, 7, 3, 6, 5, 5, 5, 7, 2 \rangle)$

Figure 3.447 provides an example where the INCREASING_WIDTH_PROPER_PLAIN ([5, 3, 3, 5, 6, 5, 4, 4, 7, 3, 6, 5, 5, 5, 7, 2]) constraint holds.

Figure 3.447: Illustrating the INCREASING_WIDTH_PROPER_PLAIN constraint of the **Example** slot

**Typical**          $|\texttt{VARIABLES}| > 3$
                     $\texttt{range}(\texttt{VARIABLES.var}) > 1$

**Automaton**        Figure 3.448 depicts the automaton associated with the constraint INCREAS-
ING_WIDTH_PROPER_PLAIN.

Figure 3.448: Automaton for the INCREASING_WIDTH_PROPER_PLAIN constraint ob-
tained by applying decoration Table 2.36 to the seed transducer of the PROPER_PLAIN
pattern

## 3.225 INCREASING_WIDTH_PROPER_PLATEAU

    **DESCRIPTION**    **AUTOMATON**

| **Origin** | Based on the PROPER_PLATEAU pattern. |

| **Constraint** | INCREASING_WIDTH_PROPER_PLATEAU(VARIABLES) |

| **Argument** | VARIABLES : collection(var−dvar) |

| **Restriction** | required(VARIABLES, var) |

| **Purpose** | Succeeds if the values denoting the width of each occurrence of the PROPER_PLATEAU pattern in the time-series given by the VARIABLES collection are increasing. An occurrence of the pattern PROPER_PLATEAU is the *maximal* subsequence which matches the regular expression $< =^+ >$. Assume that the occurrence of the pattern PROPER_PLATEAU starts at position $i$ and ends at position $j$. The feature WIDTH computes the value $j − i$. |

| **Example** | $(\langle 7, 1, 3, 3, 2, 5, 1, 4, 4, 3, 2, 3, 5, 5, 5, 3 \rangle)$ |

Figure 3.449 provides an example where the INCREASING_WIDTH_PROPER_PLATEAU $([7, 1, 3, 3, 2, 5, 1, 4, 4, 3, 2, 3, 5, 5, 5, 3])$ constraint holds.

| **Typical** | $\|$VARIABLES$\| > 3$ |
|            | range(VARIABLES.var) $> 1$ |

Figure 3.449: Illustrating the INCREASING_WIDTH_PROPER_PLATEAU constraint of the **Example** slot

**Automaton**      Figure 3.450 depicts the automaton associated with the constraint INCREAS-ING_WIDTH_PROPER_PLATEAU.

Figure 3.450: Automaton for the INCREASING_WIDTH_PROPER_PLATEAU constraint obtained by applying decoration Table 2.36 to the seed transducer of the PROPER_PLATEAU pattern

## 3.226  INCREASING_WIDTH_STEADY_SEQUENCE

▶ ▷ ◁ ◀        DESCRIPTION          AUTOMATON

| | |
|---|---|
| **Origin** | Based on the STEADY_SEQUENCE pattern. |
| **Constraint** | INCREASING_WIDTH_STEADY_SEQUENCE(VARIABLES) |
| **Argument** | VARIABLES : collection(var−dvar) |
| **Restriction** | required(VARIABLES, var) |
| **Purpose** | Succeeds if the values denoting the width of each occurrence of the STEADY_SEQUENCE pattern in the time-series given by the VARIABLES collection are increasing. An occurrence of the pattern STEADY_SEQUENCE is the *maximal* subsequence which matches the regular expression $=^{+}$. Assume that the occurrence of the pattern STEADY_SEQUENCE starts at position $i$ and ends at position $j$. The feature WIDTH computes the value $j − i + 2$. |
| **Example** | $(\langle 6, 2, 2, 3, 5, 6, 4, 4, 2, 3, 5, 5, 5, 2, 3, 4\rangle)$ |

Figure 3.451 provides an example where the INCREASING_WIDTH_STEADY_SEQUENCE ([6, 2, 2, 3, 5, 6, 4, 4, 2, 3, 5, 5, 5, 2, 3, 4]) constraint holds.

Figure 3.451: Illustrating the INCREASING_WIDTH_STEADY_SEQUENCE constraint of the **Example** slot

| | |
|---|---|
| **Typical** | $|\text{VARIABLES}| > 1$ |

**Automaton**   Figure 3.452 depicts the automaton associated with the constraint INCREAS-ING_WIDTH_STEADY_SEQUENCE.

Figure 3.452: Automaton for the INCREASING_WIDTH_STEADY_SEQUENCE constraint obtained by applying decoration Table 2.36 to the seed transducer of the STEADY_SEQUENCE pattern

# 3.227 INCREASING_WIDTH_STRICTLY_DECREASING_SEQUENCE

▶ ▷ ◁ ◀        DESCRIPTION          AUTOMATON

| **Origin** | Based on the STRICTLY_DECREASING_SEQUENCE pattern. |
| **Constraint** | INCREASING_WIDTH_STRICTLY_DECREASING_SEQUENCE(VARIABLES) |
| **Argument** | VARIABLES : collection(var−dvar) |
| **Restriction** | required(VARIABLES, var) |

**Purpose**

Succeeds if the values denoting the width of each occurrence of the STRICTLY_DECREASING_SEQUENCE pattern in the time-series given by the VARIABLES collection are increasing.

An occurrence of the pattern STRICTLY_DECREASING_SEQUENCE is the *maximal* subsequence which matches the regular expression $>^+$.

Assume that the occurrence of the pattern STRICTLY_DECREASING_SEQUENCE starts at position $i$ and ends at position $j$. The feature WIDTH computes the value $j - i + 2$.

**Example**

$(\langle 5, 7, 4, 4, 2, 3, 4, 3, 2, 2, 4, 5, 6, 3, 2 \rangle)$

Figure 3.453 provides an example where the INCREASING_WIDTH_STRICTLY_DECREASING_SEQUENCE ($[5, 7, 4, 4, 2, 3, 4, 3, 2, 2, 4, 5, 6, 3, 2]$) constraint holds.

**Typical**

$|\text{VARIABLES}| > 1$

range(VARIABLES.var) > 1

Figure 3.453: Illustrating the INCREASING_WIDTH_STRICTLY_DECREASING_SEQUENCE constraint of the **Example** slot

**Automaton**          Figure 3.454 depicts the automaton associated with the constraint INCREAS-
                       ING_WIDTH_STRICTLY_DECREASING_SEQUENCE.

Figure 3.454: Automaton for the INCREASING_WIDTH_STRICTLY_DECREASING_SEQUENCE
constraint obtained by applying decoration Table 2.36 to the seed transducer of the
STRICTLY_DECREASING_SEQUENCE pattern

## 3.228 INCREASING_WIDTH_STRICTLY_INCREASING_SEQUENCE

▶ ▷ ◁ ◀          **DESCRIPTION**          **AUTOMATON**

| | |
|---|---|
| **Origin** | Based on the STRICTLY_INCREASING_SEQUENCE pattern. |
| **Constraint** | INCREASING_WIDTH_STRICTLY_INCREASING_SEQUENCE(VARIABLES) |
| **Argument** | VARIABLES : collection(var−dvar) |
| **Restriction** | required(VARIABLES, var) |

**Purpose**

Succeeds if the values denoting the width of each occurrence of the STRICTLY_INCREASING_SEQUENCE pattern in the time-series given by the VARIABLES collection are increasing.

An occurrence of the pattern STRICTLY_INCREASING_SEQUENCE is the *maximal* subsequence which matches the regular expression $<^+$.

Assume that the occurrence of the pattern STRICTLY_INCREASING_SEQUENCE starts at position $i$ and ends at position $j$. The feature WIDTH computes the value $j - i + 2$.

**Example**

$$(\langle 6, 2, 3, 6, 5, 5, 5, 4, 5, 6, 1, 2, 3, 4, 5, 5 \rangle)$$

Figure 3.455 provides an example where the INCREASING_WIDTH_STRICTLY_INCREASING_SEQUENCE ($[6, 2, 3, 6, 5, 5, 5, 4, 5, 6, 1, 2, 3, 4, 5, 5]$) constraint holds.

**Typical**

$|\text{VARIABLES}| > 1$
$\text{range}(\text{VARIABLES.var}) > 1$

Figure 3.455: Illustrating the INCREASING_WIDTH_STRICTLY_INCREASING_SEQUENCE constraint of the **Example** slot

**Automaton**    Figure 3.456 depicts the automaton associated with the constraint INCREAS-
ING_WIDTH_STRICTLY_INCREASING_SEQUENCE.

Figure 3.456: Automaton for the INCREASING_WIDTH_STRICTLY_INCREASING_SEQUENCE
constraint obtained by applying decoration Table 2.36 to the seed transducer of the
STRICTLY_INCREASING_SEQUENCE pattern

## 3.229 INCREASING_WIDTH_SUMMIT

▶ ▷ ◁ ◀          DESCRIPTION          AUTOMATON

$$(< | < (= | <)^* <)(>|> (= | >)^* >)$$

| | |
|---|---|
| **Origin** | Based on the SUMMIT pattern. |
| **Constraint** | INCREASING_WIDTH_SUMMIT(VARIABLES) |
| **Argument** | VARIABLES : collection(var−dvar) |
| **Restriction** | required(VARIABLES, var) |
| **Purpose** | Succeeds if the values denoting the width of each occurrence of the SUMMIT pattern in the time-series given by the VARIABLES collection are increasing. |
| | An occurrence of the pattern SUMMIT is the *maximal* subsequence which matches the regular expression $(< | < (= | <)^* <)(>|> (= | >)^* >)$. |
| | Assume that the occurrence of the pattern SUMMIT starts at position $i$ and ends at position $j$. The feature WIDTH computes the value $j - i$. |
| **Example** | $(\langle 1, 5, 2, 1, 6, 6, 2, 3, 5, 4, 1, 4, 6, 4, 3, 2 \rangle)$ |

Figure 3.457 provides an example where the INCREASING_WIDTH_SUMMIT $([1, 5, 2, 1, 6, 6, 2, 3, 5, 4, 1, 4, 6, 4, 3, 2])$ constraint holds.

Figure 3.457: Illustrating the INCREASING_WIDTH_SUMMIT constraint of the **Example** slot

**Typical**
$|\mathtt{VARIABLES}| > 2$
$\mathtt{range}(\mathtt{VARIABLES.var}) > 1$

**Automaton**     Figure 3.458 depicts the automaton associated with the constraint INCREAS-
ING_WIDTH_SUMMIT.

Figure 3.458: Automaton for the INCREASING_WIDTH_SUMMIT constraint obtained by
applying decoration Table 2.36 to the seed transducer of the SUMMIT pattern (transition
$u \rightarrow r$ has the same accumulator update as transition $r \rightarrow u$)

## 3.230  INCREASING_WIDTH_VALLEY

        **DESCRIPTION**          **AUTOMATON**

$> (= | >)^* (< | =)^* <$

| **Origin** | Based on the VALLEY pattern. |
|---|---|

| **Constraint** | INCREASING_WIDTH_VALLEY(VARIABLES) |
|---|---|

| **Argument** | VARIABLES : collection(var−dvar) |
|---|---|

| **Restriction** | required(VARIABLES, var) |
|---|---|

**Purpose**

Succeeds if the values denoting the width of each occurrence of the VALLEY pattern in the time-series given by the VARIABLES collection are increasing.

An occurrence of the pattern VALLEY is the *maximal* subsequence which matches the regular expression $> (= | >)^* (< | =)^* <$.

Assume that the occurrence of the pattern VALLEY starts at position $i$ and ends at position $j$. The feature WIDTH computes the value $j - i$.

**Example**

$(\langle 7, 2, 2, 6, 3, 4, 5, 6, 6, 4, 4, 6, 7, 3, 3, 1 \rangle)$

Figure 3.459 provides an example where the INCREASING_WIDTH_VALLEY $([7, 2, 2, 6, 3, 4, 5, 6, 6, 4, 4, 6, 7, 3, 3, 1])$ constraint holds.

**Typical**

$|\text{VARIABLES}| > 2$
range(VARIABLES.var) > 1

Figure 3.459: Illustrating the INCREASING_WIDTH_VALLEY constraint of the **Example** slot

**Automaton**    Figure 3.460 depicts the automaton associated with the constraint INCREAS-
ING_WIDTH_VALLEY.

Figure 3.460: Automaton for the INCREASING_WIDTH_VALLEY constraint obtained by
applying decoration Table 2.36 to the seed transducer of the VALLEY pattern

## 3.231 INCREASING_WIDTH_ZIGZAG

▶ ▷ ◁ ◀          **DESCRIPTION**          **AUTOMATON**

| | |
|---|---|
| **Origin** | Based on the ZIGZAG pattern. |
| **Constraint** | INCREASING_WIDTH_ZIGZAG(VARIABLES) |
| **Argument** | VARIABLES : collection(var−dvar) |
| **Restriction** | required(VARIABLES, var) |
| **Purpose** | Succeeds if the values denoting the width of each occurrence of the ZIGZAG pattern in the time-series given by the VARIABLES collection are increasing.<br><br>An occurrence of the pattern ZIGZAG is the *maximal* subsequence which matches the regular expression $(<>)^+(< | <>) | (><)^+(> | ><)$.<br><br>Assume that the occurrence of the pattern ZIGZAG starts at position $i$ and ends at position $j$. The feature WIDTH computes the value $j - i$. |
| **Example** | $(\langle 7, 7, 2, 5, 3, 4, 6, 3, 6, 1, 1, 5, 4, 7, 5, 6 \rangle)$ |

Figure 3.461 provides an example where the INCREASING_WIDTH_ZIGZAG $([7, 7, 2, 5, 3, 4, 6, 3, 6, 1, 1, 5, 4, 7, 5, 6])$ constraint holds.

| | |
|---|---|
| **Typical** | $|\texttt{VARIABLES}| > 3$<br>range(VARIABLES.var) $> 1$ |

Figure 3.461: Illustrating the INCREASING_WIDTH_ZIGZAG constraint of the **Example** slot

**Automaton**          Figure 3.462 depicts the automaton associated with the constraint INCREAS-
                       ING_WIDTH_ZIGZAG.

Figure 3.462: Automaton for the INCREASING_WIDTH_ZIGZAG constraint obtained by applying decoration Table 2.36 to the seed transducer of the ZIGZAG pattern; (1) missing transitions from $a, b, c, d, e, f$ to $s$ are labelled by $=$; (2) on transitions from $b, c, e, f$ to $s$ the accumulator $D$ is reset to its initial value; (3) on transitions from $c, f$ to $s$ the accumulator $F$ is reset to $C$, and the accumulator $R$ is updated wrt $C$ and $F$

## 3.232   MAX_HEIGHT_DECREASING_TERRACE

▶ ▷ ◁ ◀        DESCRIPTION          AUTOMATON

| | |
|---|---|
| **Origin** | Based on the DECREASING_TERRACE pattern. |
| **Constraint** | MAX_HEIGHT_DECREASING_TERRACE(VALUE, VARIABLES) |
| **Arguments** | VALUE       : dvar<br>VARIABLES : collection(var−dvar) |
| **Restrictions** | $\mathtt{sv} \leq 3 \vee \mathtt{rv} \leq 2 \Rightarrow \mathtt{VALUE} = -\infty$ |
| | $\mathtt{VALUE} = -\infty \vee \mathtt{VALUE} \geq \mathtt{minv} + 1$ |
| | $\mathtt{VALUE} \leq \mathtt{maxv} - 1$ ① |
| | required(VARIABLES, var) |
| | where |
| | maxv = maxval(VARIABLES.var) |
| | minv = minval(VARIABLES.var) |
| | sv = \|VARIABLES\| |
| | rv = range(VARIABLES.var) |
| **Purpose** | VALUE is the maximum of all minimum values in each occurrence of the DECREAS-ING_TERRACE pattern in the time-series given by the VARIABLES collection. If the pattern does not occur, VALUE takes the default value $-\infty$.<br>An occurrence of the pattern DECREASING_TERRACE is the *maximal* subsequence which matches the regular expression $> =^{+} >$.<br>Assume that the occurrence of the pattern DECREASING_TERRACE starts at position $i$ and ends at position $j$. The feature MIN computes the minimum of the values from index $i + 1$ to index $j$. |
| **Example** | $(4, \langle 6, 4, 4, 4, 5, 2, 2, 1, 3, 3, 5, 4, 4, 3, 3, 3 \rangle)$ |

Figure 3.463 provides an example where the MAX_HEIGHT_DECREASING_TERRACE $(4, [6, 4, 4, 4, 5, 2, 2, 1, 3, 3, 5, 4, 4, 3, 3, 3])$ constraint holds.

| | |
|---|---|
| **Typical** | \|VARIABLES\| > 3<br>range(VARIABLES.var) > 2 |
| **Arg. properties** | Functional dependency: VALUE determined by VARIABLES. |

Figure 3.463: Illustrating the MAX_HEIGHT_DECREASING_TERRACE constraint of the **Example** slot

 Figures 3.464 and 3.465 respectively depict the automaton associated with the constraint MAX_HEIGHT_DECREASING_TERRACE and its simplified form.

Figure 3.464: Automaton for the MAX_HEIGHT_DECREASING_TERRACE constraint obtained by applying decoration Table 2.35 to the seed transducer of the DECREAS-ING_TERRACE pattern where default is $-\infty$

Figure 3.465: Automaton for the MAX_HEIGHT_DECREASING_TERRACE constraint obtained by applying decoration Table 2.37 to the seed transducer of the DECREAS-ING_TERRACE pattern where default is $-\infty$; $R_i - R_{i-1} \geq 0$ is a linear invariant.

| | $s$ | $r$ | $t$ |
|---|---|---|---|
| $s$ | $\max(\overrightarrow{C}, \overleftarrow{C})$ | $\max(\overrightarrow{C}, \overleftarrow{C})$ | $\max(\overrightarrow{C}, \overleftarrow{C})$ |
| $r$ | $\max(\overrightarrow{C}, \overleftarrow{C})$ | $\max(\overrightarrow{C}, \overleftarrow{C})$ | $\min(\overrightarrow{D}, \overleftarrow{D}, \mathtt{VAR}_i)$ |
| $t$ | $\max(\overrightarrow{C}, \overleftarrow{C})$ | $\min(\overrightarrow{D}, \overleftarrow{D}, \mathtt{VAR}_i)$ | $\min(\overrightarrow{D}, \overleftarrow{D}, \mathtt{VAR}_i)$ |

Table 3.1: Glue matrix for the MAX_HEIGHT_DECREASING_TERRACE constraint defined as the composition of the DECREASING_TERRACE pattern , the feature MIN , and the aggregator max ; cells of the glue matrix are coloured with the colour of the constituent to which they are related.

## 3.233   MAX_HEIGHT_INCREASING_TERRACE

▶ ▷ ◁ ◀          DESCRIPTION          AUTOMATON

| | |
|---|---|
| **Origin** | Based on the INCREASING_TERRACE pattern. |
| **Constraint** | MAX_HEIGHT_INCREASING_TERRACE(VALUE, VARIABLES) |
| **Arguments** | VALUE      : dvar<br>VARIABLES  : collection(var−dvar) |

**Restrictions**

$$\mathtt{sv} \leq 3 \vee \mathtt{rv} \leq 2 \Rightarrow \mathtt{VALUE} = -\infty$$
$$\mathtt{VALUE} = -\infty \vee \mathtt{VALUE} \geq \mathtt{minv} + 1$$
$$\mathtt{VALUE} \leq \mathtt{maxv} - 1 \textcircled{1}$$
required(VARIABLES, var)

where

$$\mathtt{maxv} = \mathsf{maxval}(\mathtt{VARIABLES.var})$$
$$\mathtt{minv} = \mathsf{minval}(\mathtt{VARIABLES.var})$$
$$\mathtt{sv} = |\mathtt{VARIABLES}|$$
$$\mathtt{rv} = \mathsf{range}(\mathtt{VARIABLES.var})$$

**Purpose**

VALUE is the maximum of all minimum values in each occurrence of the INCREAS­ING_TERRACE pattern in the time-series given by the VARIABLES collection. If the pat­tern does not occur, VALUE takes the default value $-\infty$.

An occurrence of the pattern INCREASING_TERRACE is the *maximal* subsequence which matches the regular expression $<=^+<$.

Assume that the occurrence of the pattern INCREASING_TERRACE starts at position $i$ and ends at position $j$. The feature MIN computes the minimum of the values from index $i + 1$ to index $j$.

**Example**

$$(5, \langle 1, 3, 3, 3, 2, 5, 5, 6, 4, 4, 2, 3, 3, 3, 4, 4 \rangle)$$

Figure 3.466 provides an example where the MAX_HEIGHT_INCREASING_TERRACE $(5, [1, 3, 3, 3, 2, 5, 5, 6, 4, 4, 2, 3, 3, 3, 4, 4])$ constraint holds.

| | |
|---|---|
| **Typical** | $\|\mathtt{VARIABLES}\| > 3$<br>range(VARIABLES.var) > 2 |
| **Arg. properties** | Functional dependency: VALUE determined by VARIABLES. |

Figure 3.466: Illustrating the MAX_HEIGHT_INCREASING_TERRACE constraint of the **Example** slot

**Automaton**     Figures 3.467 and 3.468 respectively depict the automaton associated with the constraint
MAX_HEIGHT_INCREASING_TERRACE and its simplified form.

Figure 3.467: Automaton for the MAX_HEIGHT_INCREASING_TERRACE constraint
obtained by applying decoration Table 2.35 to the seed transducer of the INCREAS-
ING_TERRACE pattern where default is $-\infty$

Figure 3.468: Automaton for the MAX_HEIGHT_INCREASING_TERRACE constraint
obtained by applying decoration Table 2.37 to the seed transducer of the INCREAS-
ING_TERRACE pattern where default is $-\infty$; $R_i - R_{i-1} \geq 0$ is a linear invariant.

|   | $s$ | $r$ | $t$ |
|---|---|---|---|
| $s$ | $\max(\overrightarrow{C}, \overleftarrow{C})$ | $\max(\overrightarrow{C}, \overleftarrow{C})$ | $\max(\overrightarrow{C}, \overleftarrow{C})$ |
| $r$ | $\max(\overrightarrow{C}, \overleftarrow{C})$ | $\max(\overrightarrow{C}, \overleftarrow{C})$ | $\min(\overrightarrow{D}, \overleftarrow{D}, \mathtt{VAR}_i)$ |
| $t$ | $\max(\overrightarrow{C}, \overleftarrow{C})$ | $\min(\overrightarrow{D}, \overleftarrow{D}, \mathtt{VAR}_i)$ | $\min(\overrightarrow{D}, \overleftarrow{D}, \mathtt{VAR}_i)$ |

Table 3.2: Glue matrix for the MAX_HEIGHT_INCREASING_TERRACE constraint defined as the composition of the INCREASING_TERRACE pattern, the feature MIN, and the aggregator max; cells of the glue matrix are coloured with the colour of the constituent to which they are related.

## 3.234   MAX_HEIGHT_PLAIN

► ▷ ◁ ◄        DESCRIPTION        AUTOMATON

| | |
|---|---|
| **Origin** | Based on the PLAIN pattern. |
| **Constraint** | MAX_HEIGHT_PLAIN(VALUE, VARIABLES) |
| **Arguments** | VALUE      : dvar<br>VARIABLES : collection(var−dvar) |
| **Restrictions** | $\mathtt{sv} \leq 2 \vee \mathtt{rv} \leq 1 \Rightarrow \mathtt{VALUE} = -\infty$<br>$\mathtt{VALUE} = -\infty \vee \mathtt{VALUE} \geq \mathtt{minv}$<br>$\mathtt{VALUE} \leq \mathtt{maxv} - 1$ ①<br>required(VARIABLES, var)<br>where<br>maxv = maxval(VARIABLES.var)<br>minv = minval(VARIABLES.var)<br>sv = \|VARIABLES\|<br>rv = range(VARIABLES.var) |

| **Purpose** | VALUE is the maximum of all minimum values in each occurrence of the PLAIN pattern in the time-series given by the VARIABLES collection. If the pattern does not occur, VALUE takes the default value $-\infty$.<br>An occurrence of the pattern PLAIN is the *maximal* subsequence which matches the regular expression $> =^* <$.<br>Assume that the occurrence of the pattern PLAIN starts at position $i$ and ends at position $j$. The feature MIN computes the minimum of the values from index $i + 1$ to index $j$. |

| **Example** | $(5, \langle 2, 3, 6, 5, 7, 6, 6, 4, 5, 5, 4, 3, 3, 6, 6, 3 \rangle)$ |

Figure   3.469   provides   an   example   where   the   MAX_HEIGHT_PLAIN $(5, [2, 3, 6, 5, 7, 6, 6, 4, 5, 5, 4, 3, 3, 6, 6, 3])$ constraint holds.

| **Typical** | $\|\mathtt{VARIABLES}\| > 2$<br>range(VARIABLES.var) $> 1$ |
| **Symmetry** | Items of VARIABLES can be reversed. |
| **Arg. properties** | Functional dependency: VALUE determined by VARIABLES. |

Figure 3.469: Illustrating the MAX_HEIGHT_PLAIN constraint of the **Example** slot

**Automaton**   Figures 3.470 and 3.471 respectively depict the automaton associated with the constraint MAX_HEIGHT_PLAIN and its simplified form.

Figure 3.470: Automaton for the MAX_HEIGHT_PLAIN constraint obtained by applying decoration Table 2.35 to the seed transducer of the PLAIN pattern where default is $-\infty$

Figure 3.471: Automaton for the MAX_HEIGHT_PLAIN constraint obtained by applying decoration Table 2.37 to the seed transducer of the PLAIN pattern where default is $-\infty$; $R_i - R_{i-1} \geq 0$ is a linear invariant.

|     | $s$ | $r$ | $t$ |
|-----|-----|-----|-----|
| $s$ | $\max(\overrightarrow{C}, \overleftarrow{C})$ | $\max(\overrightarrow{C}, \overleftarrow{C})$ | $\max(\overrightarrow{C}, \overleftarrow{C})$ |
| $r$ | $\max(\overrightarrow{C}, \overleftarrow{C})$ | $\min(\overrightarrow{D}, \overleftarrow{D}, \mathtt{VAR}_i)$ | $\min(\overrightarrow{D}, \overleftarrow{D}, \mathtt{VAR}_i)$ |
| $t$ | $\max(\overrightarrow{C}, \overleftarrow{C})$ | $\min(\overrightarrow{D}, \overleftarrow{D}, \mathtt{VAR}_i)$ | $\min(\overrightarrow{D}, \overleftarrow{D}, \mathtt{VAR}_i)$ |

Table 3.3: Glue matrix for the MAX_HEIGHT_PLAIN constraint defined as the composition of the PLAIN pattern , the feature MIN , and the aggregator max ; cells of the glue matrix are coloured with the colour of the constituent to which they are related.

## 3.235   MAX_HEIGHT_PLATEAU

▶ ▷ ◁ ◀          DESCRIPTION          AUTOMATON

| | |
|---|---|
| **Origin** | Based on the PLATEAU pattern. |
| **Constraint** | MAX_HEIGHT_PLATEAU(VALUE, VARIABLES) |
| **Arguments** | VALUE       : dvar<br>VARIABLES : collection(var−dvar) |
| **Restrictions** | $\mathtt{sv} \leq 2 \vee \mathtt{rv} \leq 1 \Rightarrow \mathtt{VALUE} = -\infty$<br>$\mathtt{VALUE} = -\infty \vee \mathtt{VALUE} \geq \mathtt{minv} + 1$<br>$\mathtt{VALUE} \leq \mathtt{maxv}①$<br>required(VARIABLES, var)<br>where<br>maxv =maxval(VARIABLES.var)<br>minv =minval(VARIABLES.var)<br>sv = \|VARIABLES\|<br>rv =range(VARIABLES.var) |
| **Purpose** | VALUE is the maximum of all minimum values in each occurrence of the PLATEAU pattern in the time-series given by the VARIABLES collection. If the pattern does not occur, VALUE takes the default value $-\infty$.<br>An occurrence of the pattern PLATEAU is the *maximal* subsequence which matches the regular expression $<=^* >$.<br>Assume that the occurrence of the pattern PLATEAU starts at position $i$ and ends at position $j$. The feature MIN computes the minimum of the values from index $i + 1$ to index $j$. |
| **Example** | $(5, \langle 7, 5, 2, 3, 1, 2, 2, 4, 3, 3, 4, 5, 5, 2, 2, 5 \rangle)$ |

Figure 3.472 provides an example where the MAX_HEIGHT_PLATEAU $(5, [7, 5, 2, 3, 1, 2, 2, 4, 3, 3, 4, 5, 5, 2, 2, 5])$ constraint holds.

| | |
|---|---|
| **Typical** | $\|\mathtt{VARIABLES}\| > 2$<br>range(VARIABLES.var) > 1 |
| **Symmetry** | Items of VARIABLES can be reversed. |
| **Arg. properties** | Functional dependency: VALUE determined by VARIABLES. |

Figure 3.472: Illustrating the MAX_HEIGHT_PLATEAU constraint of the **Example** slot

 Figures 3.473 and 3.474 respectively depict the automaton associated with the constraint MAX_HEIGHT_PLATEAU and its simplified form.

Figure 3.473: Automaton for the MAX_HEIGHT_PLATEAU constraint obtained by applying decoration Table 2.35 to the seed transducer of the PLATEAU pattern where `default` is $-\infty$

Figure 3.474: Automaton for the MAX_HEIGHT_PLATEAU constraint obtained by applying decoration Table 2.37 to the seed transducer of the PLATEAU pattern where `default` is $-\infty$; $R_i - R_{i-1} \geq 0$ is a linear invariant.

|   | $s$ | $r$ | $t$ |
|---|---|---|---|
| $s$ | $\max(\overrightarrow{C}, \overleftarrow{C})$ | $\max(\overrightarrow{C}, \overleftarrow{C})$ | $\max(\overrightarrow{C}, \overleftarrow{C})$ |
| $r$ | $\max(\overrightarrow{C}, \overleftarrow{C})$ | $\min(\overrightarrow{D}, \overleftarrow{D}, \mathtt{VAR}_i)$ | $\min(\overrightarrow{D}, \overleftarrow{D}, \mathtt{VAR}_i)$ |
| $t$ | $\max(\overrightarrow{C}, \overleftarrow{C})$ | $\min(\overrightarrow{D}, \overleftarrow{D}, \mathtt{VAR}_i)$ | $\min(\overrightarrow{D}, \overleftarrow{D}, \mathtt{VAR}_i)$ |

Table 3.4: Glue matrix for the MAX_HEIGHT_PLATEAU constraint defined as the composition of the PLATEAU pattern , the feature MIN , and the aggregator max ; cells of the glue matrix are coloured with the colour of the constituent to which they are related.

## 3.236 MAX_HEIGHT_PROPER_PLAIN

            DESCRIPTION            AUTOMATON

**Origin**          Based on the PROPER_PLAIN pattern.

**Constraint**      MAX_HEIGHT_PROPER_PLAIN(VALUE, VARIABLES)

**Arguments**       VALUE      : dvar
                    VARIABLES  : collection(var−dvar)

**Restrictions**    $\mathtt{sv} \leq 3 \vee \mathtt{rv} \leq 1 \Rightarrow \mathtt{VALUE} = -\infty$
                    $\mathtt{VALUE} = -\infty \vee \mathtt{VALUE} \geq \mathtt{minv}$
                    $\mathtt{VALUE} \leq \mathtt{maxv} - 1$①
                    required(VARIABLES, var)
                    where
                    $\mathtt{maxv} =$ maxval(VARIABLES.var)
                    $\mathtt{minv} =$ minval(VARIABLES.var)
                    $\mathtt{sv} = |\mathtt{VARIABLES}|$
                    $\mathtt{rv} =$ range(VARIABLES.var)

**Purpose**         VALUE is the maximum of all minimum values in each occurrence of the PROPER_PLAIN
                    pattern in the time-series given by the VARIABLES collection. If the pattern does not
                    occur, VALUE takes the default value $-\infty$.
                    An occurrence of the pattern PROPER_PLAIN is the *maximal* subsequence which matches
                    the regular expression $>=^+<$.
                    Assume that the occurrence of the pattern PROPER_PLAIN starts at position $i$ and ends
                    at position $j$. The feature MIN computes the minimum of the values from index $i + 1$ to
                    index $j$.

**Example**         $(5, \langle 2, 7, 5, 5, 6, 3, 7, 4, 4, 5, 6, 5, 3, 3, 3, 5 \rangle)$

                    Figure 3.475 provides an example where the MAX_HEIGHT_PROPER_PLAIN
                    $(5, [2, 7, 5, 5, 6, 3, 7, 4, 4, 5, 6, 5, 3, 3, 3, 5])$ constraint holds.

**Typical**         $|\mathtt{VARIABLES}| > 3$
                    range(VARIABLES.var) > 1

**Symmetry**        Items of VARIABLES can be reversed.

**Arg. properties** Functional dependency: VALUE determined by VARIABLES.

Figure 3.475: Illustrating the MAX_HEIGHT_PROPER_PLAIN constraint of the **Example** slot

**Automaton**          Figures 3.476 and 3.477 respectively depict the automaton associated with the constraint
MAX_HEIGHT_PROPER_PLAIN and its simplified form.

Figure 3.476: Automaton for the MAX_HEIGHT_PROPER_PLAIN constraint obtained
by applying decoration Table 2.35 to the seed transducer of the PROPER_PLAIN pattern
where default is $-\infty$

Figure 3.477: Automaton for the MAX_HEIGHT_PROPER_PLAIN constraint obtained
by applying decoration Table 2.37 to the seed transducer of the PROPER_PLAIN pattern
where default is $-\infty$; $R_i - R_{i-1} \geq 0$ is a linear invariant.

|   | $s$ | $r$ | $t$ |
|---|---|---|---|
| $s$ | $\max(\overrightarrow{C}, \overleftarrow{C})$ | $\max(\overrightarrow{C}, \overleftarrow{C})$ | $\max(\overrightarrow{C}, \overleftarrow{C})$ |
| $r$ | $\max(\overrightarrow{C}, \overleftarrow{C})$ | $\max(\overrightarrow{C}, \overleftarrow{C})$ | $\min(\overrightarrow{D}, \overleftarrow{D}, \text{VAR}_i)$ |
| $t$ | $\max(\overrightarrow{C}, \overleftarrow{C})$ | $\min(\overrightarrow{D}, \overleftarrow{D}, \text{VAR}_i)$ | $\min(\overrightarrow{D}, \overleftarrow{D}, \text{VAR}_i)$ |

Table 3.5: Glue matrix for the MAX_HEIGHT_PROPER_PLAIN constraint defined as the composition of the PROPER_PLAIN pattern, the feature MIN, and the aggregator max; cells of the glue matrix are coloured with the colour of the constituent to which they are related.

## 3.237   MAX_HEIGHT_PROPER_PLATEAU

► ▷ ◁ ◀        **DESCRIPTION**              **AUTOMATON**

| **Origin** | Based on the PROPER_PLATEAU pattern. |
|---|---|

| **Constraint** | MAX_HEIGHT_PROPER_PLATEAU(VALUE, VARIABLES) |
|---|---|

| **Arguments** | VALUE      : dvar |
|---|---|
| | VARIABLES : collection(var−dvar) |

| **Restrictions** | $\mathtt{sv} \leq 3 \vee \mathtt{rv} \leq 1 \Rightarrow \mathtt{VALUE} = -\infty$ |
|---|---|
| | $\mathtt{VALUE} = -\infty \vee \mathtt{VALUE} \geq \mathtt{minv} + 1$ |
| | $\mathtt{VALUE} \leq \mathtt{maxv}$① |
| | required(VARIABLES, var) |
| | where |
| | maxv = maxval(VARIABLES.var) |
| | minv = minval(VARIABLES.var) |
| | sv = \|VARIABLES\| |
| | rv = range(VARIABLES.var) |

**Purpose**

VALUE is the maximum of all minimum values in each occurrence of the PROPER_PLATEAU pattern in the time-series given by the VARIABLES collection. If the pattern does not occur, VALUE takes the default value $-\infty$.

An occurrence of the pattern PROPER_PLATEAU is the *maximal* subsequence which matches the regular expression $<=^+>$.

Assume that the occurrence of the pattern PROPER_PLATEAU starts at position $i$ and ends at position $j$. The feature MIN computes the minimum of the values from index $i+1$ to index $j$.

| **Example** | $(5, \langle 7, 1, 3, 3, 2, 5, 1, 4, 4, 3, 2, 3, 5, 5, 5, 3 \rangle)$ |
|---|---|

Figure 3.478 provides an example where the MAX_HEIGHT_PROPER_PLATEAU $(5, [7, 1, 3, 3, 2, 5, 1, 4, 4, 3, 2, 3, 5, 5, 5, 3])$ constraint holds.

| **Typical** | $\|\mathtt{VARIABLES}\| > 3$ |
|---|---|
| | range(VARIABLES.var) > 1 |

| **Symmetry** | Items of VARIABLES can be reversed. |
|---|---|

| **Arg. properties** | Functional dependency: VALUE determined by VARIABLES. |
|---|---|

Figure 3.478: Illustrating the MAX_HEIGHT_PROPER_PLATEAU constraint of the **Example** slot

     Figures 3.479 and 3.480 respectively depict the automaton associated with the constraint
MAX_HEIGHT_PROPER_PLATEAU and its simplified form.

Figure 3.479: Automaton for the MAX_HEIGHT_PROPER_PLATEAU constraint obtained
by applying decoration Table 2.35 to the seed transducer of the PROPER_PLATEAU
pattern where default is $-\infty$

Figure 3.480: Automaton for the MAX_HEIGHT_PROPER_PLATEAU constraint obtained
by applying decoration Table 2.37 to the seed transducer of the PROPER_PLATEAU
pattern where default is $-\infty$; $R_i - R_{i-1} \geq 0$ is a linear invariant.

|   | $s$ | $r$ | $t$ |
|---|-----|-----|-----|
| $s$ | $\max(\overrightarrow{C}, \overleftarrow{C})$ | $\max(\overrightarrow{C}, \overleftarrow{C})$ | $\max(\overrightarrow{C}, \overleftarrow{C})$ |
| $r$ | $\max(\overrightarrow{C}, \overleftarrow{C})$ | $\max(\overrightarrow{C}, \overleftarrow{C})$ | $\min(\overrightarrow{D}, \overleftarrow{D}, \mathtt{VAR}_i)$ |
| $t$ | $\max(\overrightarrow{C}, \overleftarrow{C})$ | $\min(\overrightarrow{D}, \overleftarrow{D}, \mathtt{VAR}_i)$ | $\min(\overrightarrow{D}, \overleftarrow{D}, \mathtt{VAR}_i)$ |

Table 3.6: Glue matrix for the MAX_HEIGHT_PROPER_PLATEAU constraint defined as the composition of the PROPER_PLATEAU pattern, the feature MIN, and the aggregator max; cells of the glue matrix are coloured with the colour of the constituent to which they are related.

## 3.238   MAX_HEIGHT_STEADY

► ▷ ◁ ◀        DESCRIPTION        AUTOMATON

| | |
|---|---|
| **Origin** | Based on the STEADY pattern. |
| **Constraint** | MAX_HEIGHT_STEADY(VALUE, VARIABLES) |
| **Arguments** | VALUE : dvar<br>VARIABLES : collection(var−dvar) |

**Restrictions**

$\text{sv} \leq 1 \Rightarrow \text{VALUE} = -\infty$
$\text{VALUE} = -\infty \lor \text{VALUE} \geq \text{minv}$
$\text{VALUE} \leq \text{maxv}①$
required(VARIABLES, var)
where
maxv = maxval(VARIABLES.var)
minv = minval(VARIABLES.var)
$\text{sv} = |\text{VARIABLES}|$

**Purpose**

VALUE is the maximum of all minimum values in each occurrence of the STEADY pattern in the time-series given by the VARIABLES collection. If the pattern does not occur, VALUE takes the default value $-\infty$.
An occurrence of the pattern STEADY is the subsequence which matches the regular expression =.
Assume that the occurrence of the pattern STEADY starts at position $i$ and ends at position $j$. The feature MIN computes the minimum of the values from index $i$ to index $j + 1$.

**Example**

$(6, \langle 1, 1, 7, 3, 3, 5, 5, 5, 6, 5, 5, 5, 7, 2, 6, 6 \rangle)$

Figure 3.481 provides an example where the MAX_HEIGHT_STEADY $(6, [1, 1, 7, 3, 3, 5, 5, 5, 6, 5, 5, 5, 7, 2, 6, 6])$ constraint holds.

| | |
|---|---|
| **Typical** | $|\text{VARIABLES}| > 1$ |

| | |
|---|---|
| **Symmetry** | Items of VARIABLES can be reversed. |
| **Arg. properties** | Functional dependency: VALUE determined by VARIABLES. |

Figure 3.481: Illustrating the MAX_HEIGHT_STEADY constraint of the **Example** slot

**Automaton**     Figures 3.482 and 3.483 respectively depict the automaton associated with the constraint MAX_HEIGHT_STEADY and its simplified form.

Figure 3.482: Automaton for the MAX_HEIGHT_STEADY constraint obtained by applying decoration Table 2.35 to the seed transducer of the STEADY pattern where default is $-\infty$

Figure 3.483: Automaton for the MAX_HEIGHT_STEADY constraint obtained by applying decoration Table 2.37 to the seed transducer of the STEADY pattern where default is $-\infty$; $R_i - R_{i-1} \geq 0$ is a linear invariant.

Table 3.7: Glue matrix for the MAX_HEIGHT_STEADY constraint defined as the composition of the STEADY pattern , the feature MIN , and the aggregator max ; cells of the glue matrix are coloured with the colour of the constituent to which they are related.

## 3.239  MAX_HEIGHT_STEADY_SEQUENCE

▶ ▷ ◁ ◀        **DESCRIPTION**            **AUTOMATON**

| | |
|---|---|
| **Origin** | Based on the STEADY_SEQUENCE pattern. |
| **Constraint** | MAX_HEIGHT_STEADY_SEQUENCE(VALUE, VARIABLES) |
| **Arguments** | VALUE     : dvar<br>VARIABLES : collection(var−dvar) |

**Restrictions**

$\text{sv} \leq 1 \Rightarrow \text{VALUE} = -\infty$
$\text{VALUE} = -\infty \vee \text{VALUE} \geq \text{minv}$
$\text{VALUE} \leq \text{maxv}①$
required(VARIABLES, var)
where
maxv = maxval(VARIABLES.var)
minv = minval(VARIABLES.var)
$\text{sv} = |\text{VARIABLES}|$

**Purpose**

VALUE is the maximum of all minimum values in each occurrence of the STEADY_SEQUENCE pattern in the time-series given by the VARIABLES collection. If the pattern does not occur, VALUE takes the default value $-\infty$.

An occurrence of the pattern STEADY_SEQUENCE is the *maximal* subsequence which matches the regular expression $=^+$.

Assume that the occurrence of the pattern STEADY_SEQUENCE starts at position $i$ and ends at position $j$. The feature MIN computes the minimum of the values from index $i$ to index $j + 1$.

**Example**

$(5, \langle 3, 1, 1, 4, 5, 5, 5, 6, 2, 2, 4, 4, 3, 2, 1, 1 \rangle)$

Figure 3.484 provides an example where the MAX_HEIGHT_STEADY_SEQUENCE $(5, [3, 1, 1, 4, 5, 5, 5, 6, 2, 2, 4, 4, 3, 2, 1, 1])$ constraint holds.

| | |
|---|---|
| **Typical** | $|\text{VARIABLES}| > 1$ |
| **Symmetry** | Items of VARIABLES can be reversed. |
| **Arg. properties** | Functional dependency: VALUE determined by VARIABLES. |

Figure 3.484: Illustrating the MAX_HEIGHT_STEADY_SEQUENCE constraint of the **Example** slot

**Automaton**       Figures 3.485 and 3.486 respectively depict the automaton associated with the constraint
MAX_HEIGHT_STEADY_SEQUENCE and its simplified form.

Figure 3.485: Automaton for the MAX_HEIGHT_STEADY_SEQUENCE constraint
obtained by applying decoration Table 2.35 to the seed transducer of the
STEADY_SEQUENCE pattern where default is $-\infty$

Figure 3.486: Automaton for the MAX_HEIGHT_STEADY_SEQUENCE constraint
obtained by applying decoration Table 2.37 to the seed transducer of the
STEADY_SEQUENCE pattern where default is $-\infty$; $R_i - R_{i-1} \geq 0$ is a linear invariant.

|   | $s$ | $r$ |
|---|-----|-----|
| $s$ | $\max(\overrightarrow{C}, \overleftarrow{C})$ | $\max(\overrightarrow{C}, \overleftarrow{C})$ |
| $r$ | $\max(\overrightarrow{C}, \overleftarrow{C})$ | $\min(\overrightarrow{C}, \overleftarrow{C}, \overrightarrow{D}, \overleftarrow{D}, \mathtt{VAR}_i)$ |

Table 3.8: Glue matrix for the MAX_HEIGHT_STEADY_SEQUENCE constraint defined as the composition of the STEADY_SEQUENCE pattern , the feature MIN , and the aggregator max ; cells of the glue matrix are coloured with the colour of the constituent to which they are related.

# 3.240   MAX_MAX_BUMP_ON_DECREASING_SEQUENCE

▶ ▷ ◁ ◀     **DESCRIPTION**          **AUTOMATON**

| **Origin** | Based on the BUMP_ON_DECREASING_SEQUENCE pattern. |
| --- | --- |

| **Constraint** | MAX_MAX_BUMP_ON_DECREASING_SEQUENCE(VALUE, VARIABLES) |
| --- | --- |

| **Arguments** | VALUE      :  dvar |
| --- | --- |
| | VARIABLES  :  collection(var−dvar) |

**Restrictions**

$\mathtt{sv} \le 5 \lor \mathtt{rv} \le 2 \Rightarrow \mathtt{VALUE} = -\infty$
$\mathtt{VALUE} = -\infty \lor \mathtt{VALUE} \ge \mathtt{minv} + 2$
$\mathtt{VALUE} \le \mathtt{maxv}$①
required(VARIABLES, var)
where
maxv =maxval(VARIABLES.var)
minv =minval(VARIABLES.var)
sv = |VARIABLES|
rv =range(VARIABLES.var)

**Purpose**

VALUE is the maximum of all maximum values in each occurrence of the BUMP_ON_DECREASING_SEQUENCE pattern in the time-series given by the VARIABLES collection. If the pattern does not occur, VALUE takes the default value $-\infty$.
An occurrence of the pattern BUMP_ON_DECREASING_SEQUENCE is the subsequence which matches the regular expression >><>>.
Assume that the occurrence of the pattern BUMP_ON_DECREASING_SEQUENCE starts at position $i$ and ends at position $j$. The feature MAX computes the maximum of the values from index $i + 2$ to index $j$.

**Example**

$(6, \langle 7, 6, 5, 6, 5, 4, 1, 4, 7, 5, 4, 2, 5, 4, 3, 3 \rangle)$

Figure 3.487 provides an example where the MAX_MAX_BUMP_ON_DECREASING_SEQUENCE $(6, [7, 6, 5, 6, 5, 4, 1, 4, 7, 5, 4, 2, 5, 4, 3, 3])$ constraint holds.

| **Typical** | $|\mathtt{VARIABLES}| > 5$ |
| --- | --- |
| | range(VARIABLES.var) > 2 |

| **Arg. properties** | Functional dependency: VALUE determined by VARIABLES. |
| --- | --- |

Figure 3.487: Illustrating the MAX_MAX_BUMP_ON_DECREASING_SEQUENCE constraint of the **Example** slot

**Automaton**    Figures 3.488 and 3.489 respectively depict the automaton associated with the constraint
MAX_MAX_BUMP_ON_DECREASING_SEQUENCE and its simplified form.

Figure 3.488: Automaton for the MAX_MAX_BUMP_ON_DECREASING_SEQUENCE
constraint obtained by applying decoration Table 2.35 to the seed transducer of the
BUMP_ON_DECREASING_SEQUENCE pattern where default is $-\infty$

Figure 3.489: Automaton for the MAX_MAX_BUMP_ON_DECREASING_SEQUENCE constraint obtained by applying decoration Table 2.27 to the seed transducer of the BUMP_ON_DECREASING_SEQUENCE pattern where default is $-\infty$; $R_i - R_{i-1} \geq 0$ is a linear invariant.

# 3.241 MAX_MAX_DECREASING

| | |
|---|---|
| **Origin** | Based on the DECREASING pattern. |
| **Constraint** | MAX_MAX_DECREASING(VALUE, VARIABLES) |
| **Arguments** | VALUE       :  dvar |
| | VARIABLES  :  collection(var−dvar) |

**Restrictions**

$$\text{sv} \leq 1 \vee \text{rv} \leq 1 \Rightarrow \text{VALUE} = -\infty$$
$$\text{VALUE} = -\infty \vee \text{VALUE} \geq \text{minv} + 1$$
$$\text{VALUE} \leq \text{maxv}①$$
$$\text{required}(\text{VARIABLES}, \text{var})$$
where
$$\text{maxv} = \text{maxval}(\text{VARIABLES.var})$$
$$\text{minv} = \text{minval}(\text{VARIABLES.var})$$
$$\text{sv} = |\text{VARIABLES}|$$
$$\text{rv} = \text{range}(\text{VARIABLES.var})$$

**Purpose**

VALUE is the maximum of all maximum values in each occurrence of the DECREASING pattern in the time-series given by the VARIABLES collection. If the pattern does not occur, VALUE takes the default value $-\infty$.

An occurrence of the pattern DECREASING is the subsequence which matches the regular expression >.

Assume that the occurrence of the pattern DECREASING starts at position $i$ and ends at position $j$. The feature MAX computes the maximum of the values from index $i$ to index $j + 1$.

**Example**

$$(6, \langle 3, 4, 2, 2, 5, 6, 6, 4, 4, 3, 1, 1, 4, 6, 4, 4 \rangle)$$

Figure 3.490 provides an example where the MAX_MAX_DECREASING $(6, [3, 4, 2, 2, 5, 6, 6, 4, 4, 3, 1, 1, 4, 6, 4, 4])$ constraint holds.

**Typical**

$$|\text{VARIABLES}| > 1$$
$$\text{range}(\text{VARIABLES.var}) > 1$$

**Arg. properties**

Functional dependency: VALUE determined by VARIABLES.

Figure 3.490: Illustrating the MAX_MAX_DECREASING constraint of the **Example** slot

**Automaton**     Figures 3.491 and 3.492 respectively depict the automaton associated with the constraint MAX_MAX_DECREASING and its simplified form.

$$\left\{\begin{array}{l} C \leftarrow \texttt{default} \\ D \leftarrow -\infty \\ R \leftarrow \texttt{default} \end{array}\right\}$$

$$\left\{\begin{array}{l} D \leftarrow -\infty \\ R \leftarrow \max(R, \max(\max(D, \texttt{VAR}_i), \texttt{VAR}_{i+1})) \end{array}\right\}$$

$\max(R, C)$

Figure 3.491: Automaton for the MAX_MAX_DECREASING constraint obtained by applying decoration Table 2.35 to the seed transducer of the DECREASING pattern where `default` is $-\infty$

$\{R \leftarrow \texttt{default}\}$

$\{R \leftarrow \max(R, \texttt{VAR}_i)\}$

$R$

Figure 3.492: Automaton for the MAX_MAX_DECREASING constraint obtained by applying decoration Table 2.37 to the seed transducer of the DECREASING pattern where `default` is $-\infty$; $R_i - R_{i-1} \geq 0$ is a linear invariant.

| | $s$ | |
|---|---|---|
| $s$ | $\max(\overrightarrow{C}, \overleftarrow{C})$ | |

Table 3.9: Glue matrix for the MAX_MAX_DECREASING constraint defined as the composition of the DECREASING pattern, the feature MAX, and the aggregator max; cells of the glue matrix are coloured with the colour of the constituent to which they are related.

# 3.242　MAX_MAX_DECREASING_SEQUENCE

▶ ▷ ◁ ◀　　**DESCRIPTION**　　**AUTOMATON**

> (> | =)* > |
> >

| **Origin** | Based on the DECREASING_SEQUENCE pattern. |
|---|---|

| **Constraint** | MAX_MAX_DECREASING_SEQUENCE(VALUE, VARIABLES) |
|---|---|

**Arguments**

```
VALUE      : dvar
VARIABLES  : collection(var−dvar)
```

**Restrictions**

$\mathtt{sv} \leq 1 \vee \mathtt{rv} \leq 1 \Rightarrow \mathtt{VALUE} = -\infty$

$\mathtt{VALUE} = -\infty \vee \mathtt{VALUE} \geq \mathtt{minv} + 1$

$\mathtt{VALUE} \leq \mathtt{maxv}①$

required(VARIABLES, var)

where

maxv = maxval(VARIABLES.var)

minv = minval(VARIABLES.var)

sv = |VARIABLES|

rv = range(VARIABLES.var)

VALUE ≤ **2**

**Purpose**

VALUE is the maximum of all maximum values in each occurrence of the DECREAS­ING_SEQUENCE pattern in the time-series given by the VARIABLES collection. If the pattern does not occur, VALUE takes the default value $-\infty$.

An occurrence of the pattern DECREASING_SEQUENCE is the *maximal* subsequence which matches the regular expression > (> | =)* > | >.

Assume that the occurrence of the pattern DECREASING_SEQUENCE starts at position $i$ and ends at position $j$. The feature MAX computes the maximum of the values from index $i$ to index $j + 1$.

**Example**

$(6, \langle 3, 4, 2, 2, 5, 6, 6, 4, 4, 3, 1, 1, 4, 6, 4, 4 \rangle)$

Figure 3.493 provides an example where the MAX_MAX_DECREASING_SEQUENCE $(6, [3, 4, 2, 2, 5, 6, 6, 4, 4, 3, 1, 1, 4, 6, 4, 4])$ constraint holds.

**Typical**

$|\mathtt{VARIABLES}| > 1$

range(VARIABLES.var) > 1

**Arg. properties**

Functional dependency: VALUE determined by VARIABLES.

Figure 3.493: Illustrating the MAX_MAX_DECREASING_SEQUENCE constraint of the **Example** slot

 Figures 3.494 and 3.495 respectively depict the automaton associated with the constraint MAX_MAX_DECREASING_SEQUENCE and its simplified form.

Figure 3.494: Automaton for the MAX_MAX_DECREASING_SEQUENCE constraint obtained by applying decoration Table 2.35 to the seed transducer of the DECREASING_SEQUENCE pattern where default is $-\infty$

Figure 3.495: Automaton for the MAX_MAX_DECREASING_SEQUENCE constraint obtained by applying decoration Table 2.37 to the seed transducer of the DECREASING_SEQUENCE pattern where default is $-\infty$; $R_i - R_{i-1} \geq 0$ is a linear invariant.

|     | $s$ | $t$ |
| --- | --- | --- |
| $s$ | $\max(\overrightarrow{C}, \overleftarrow{C})$ | $\max(\overrightarrow{C}, \overleftarrow{C})$ |
| $t$ | $\max(\overrightarrow{C}, \overleftarrow{C})$ | $\max(\overrightarrow{C}, \overleftarrow{C}, \overrightarrow{D}, \overleftarrow{D}, \mathtt{VAR}_i)$ |

Table 3.10: Glue matrix for the MAX_MAX_DECREASING_SEQUENCE constraint defined as the composition of the DECREASING_SEQUENCE pattern , the feature MAX , and the aggregator max ; cells of the glue matrix are coloured with the colour of the constituent to which they are related.

## 3.243 MAX_MAX_DIP_ON_INCREASING_SEQUENCE

▶ ▷ ◁ ◀          DESCRIPTION          AUTOMATON

**Origin**         Based on the DIP_ON_INCREASING_SEQUENCE pattern.

**Constraint**     MAX_MAX_DIP_ON_INCREASING_SEQUENCE(VALUE, VARIABLES)

**Arguments**      VALUE     : dvar
                   VARIABLES : collection(var−dvar)

**Restrictions**   $\mathtt{sv} \leq 5 \vee \mathtt{rv} \leq 2 \Rightarrow \mathtt{VALUE} = -\infty$
                   $\mathtt{VALUE} = -\infty \vee \mathtt{VALUE} \geq \mathtt{minv} + 2$
                   $\mathtt{VALUE} \leq \mathtt{maxv}①$
                   required(VARIABLES, var)
                   where
                   maxv =maxval(VARIABLES.var)
                   minv =minval(VARIABLES.var)
                   sv = |VARIABLES|
                   rv =range(VARIABLES.var)

**Purpose**        VALUE is the maximum of all maximum values in each occurrence of the
                   DIP_ON_INCREASING_SEQUENCE pattern in the time-series given by the VARIABLES
                   collection. If the pattern does not occur, VALUE takes the default value $-\infty$.
                   An occurrence of the pattern DIP_ON_INCREASING_SEQUENCE is the subsequence
                   which matches the regular expression <<><<.
                   Assume that the occurrence of the pattern DIP_ON_INCREASING_SEQUENCE starts at
                   position $i$ and ends at position $j$. The feature MAX computes the maximum of the values
                   from index $i + 2$ to index $j$.

**Example**        $(6, \langle 1, 2, 3, 2, 5, 6, 7, 4, 1, 3, 4, 6, 1, 2, 4, 4 \rangle)$

                   Figure 3.496 provides an example where the MAX_MAX_DIP_ON_INCREASING_SEQUENCE
                   $(6, [1, 2, 3, 2, 5, 6, 7, 4, 1, 3, 4, 6, 1, 2, 4, 4])$ constraint holds.

**Typical**        $|\mathtt{VARIABLES}| > 5$
                   range(VARIABLES.var) > 2

**Arg. properties** Functional dependency: VALUE determined by VARIABLES.

Figure 3.496: Illustrating the MAX_MAX_DIP_ON_INCREASING_SEQUENCE constraint of the **Example** slot

**Automaton**   Figures 3.497 and 3.498 respectively depict the automaton associated with the constraint
MAX_MAX_DIP_ON_INCREASING_SEQUENCE and its simplified form.

Figure 3.497: Automaton for the MAX_MAX_DIP_ON_INCREASING_SEQUENCE constraint obtained by applying decoration Table 2.35 to the seed transducer of the DIP_ON_INCREASING_SEQUENCE pattern where default is $-\infty$

Figure 3.498: Automaton for the MAX_MAX_DIP_ON_INCREASING_SEQUENCE constraint obtained by applying decoration Table 2.25 to the seed transducer of the DIP_ON_INCREASING_SEQUENCE pattern where default is $-\infty$; $R_i - R_{i-1} \geq 0$ is a linear invariant.

# 3.244   MAX_MAX_INCREASING

        **DESCRIPTION**        **AUTOMATON**

| | |
|---|---|
| **Origin** | Based on the INCREASING pattern. |
| **Constraint** | MAX_MAX_INCREASING(VALUE, VARIABLES) |
| **Arguments** | VALUE      : dvar |
| | VARIABLES  : collection(var−dvar) |

**Restrictions**

$\text{sv} \leq 1 \vee \text{rv} \leq 1 \Rightarrow \text{VALUE} = -\infty$

$\text{VALUE} = -\infty \vee \text{VALUE} \geq \text{minv} + 1$

$\text{VALUE} \leq \text{maxv}①$

required(VARIABLES, var)

where

maxv =maxval(VARIABLES.var)

minv =minval(VARIABLES.var)

sv = |VARIABLES|

rv =range(VARIABLES.var)

**Purpose**

VALUE is the maximum of all maximum values in each occurrence of the INCREASING pattern in the time-series given by the VARIABLES collection. If the pattern does not occur, VALUE takes the default value $-\infty$.

An occurrence of the pattern INCREASING is the subsequence which matches the regular expression $<$.

Assume that the occurrence of the pattern INCREASING starts at position $i$ and ends at position $j$. The feature MAX computes the maximum of the values from index $i$ to index $j + 1$.

| | |
|---|---|
| **Example** | $(6, \langle 4, 3, 5, 5, 2, 1, 1, 3, 3, 4, 6, 6, 3, 1, 3, 3 \rangle)$ |

Figure 3.499 provides an example where the MAX_MAX_INCREASING $(6, [4, 3, 5, 5, 2, 1, 1, 3, 3, 4, 6, 6, 3, 1, 3, 3])$ constraint holds.

| | |
|---|---|
| **Typical** | $|\text{VARIABLES}| > 1$ |
| | range(VARIABLES.var) > 1 |
| **Arg. properties** | Functional dependency: VALUE determined by VARIABLES. |

Figure 3.499: Illustrating the MAX_MAX_INCREASING constraint of the **Example** slot

**Automaton** Figures 3.500 and 3.501 respectively depict the automaton associated with the constraint MAX_MAX_INCREASING and its simplified form.

Figure 3.500: Automaton for the MAX_MAX_INCREASING constraint obtained by applying decoration Table 2.35 to the seed transducer of the INCREASING pattern where `default` is $-\infty$

Figure 3.501: Automaton for the MAX_MAX_INCREASING constraint obtained by applying decoration Table 2.38 to the seed transducer of the INCREASING pattern where `default` is $-\infty$; $R_i - R_{i-1} \geq 0$ is a linear invariant.

Table 3.11: Glue matrix for the MAX_MAX_INCREASING constraint defined as the composition of the INCREASING pattern, the feature MAX, and the aggregator max; cells of the glue matrix are coloured with the colour of the constituent to which they are related.

# 3.245   MAX_MAX_INCREASING_SEQUENCE

▶ ▷ ◁ ◀        **DESCRIPTION**          **AUTOMATON**

**Origin**           Based on the INCREASING_SEQUENCE pattern.

**Constraint**       MAX_MAX_INCREASING_SEQUENCE(VALUE, VARIABLES)

**Arguments**        VALUE      : dvar
                     VARIABLES  : collection(var−dvar)

**Restrictions**     $\text{sv} \le 1 \vee \text{rv} \le 1 \Rightarrow \text{VALUE} = -\infty$

                     $\text{VALUE} = -\infty \vee \text{VALUE} \ge \text{minv} + 1$
                     $\text{VALUE} \le \text{maxv①}$
                     required(VARIABLES, var)
                     where
                     maxv = maxval(VARIABLES.var)
                     minv = minval(VARIABLES.var)
                     sv = |VARIABLES|
                     rv = range(VARIABLES.var)

**Purpose**          VALUE is the maximum of all maximum values in each occurrence of the INCREAS-
                     ING_SEQUENCE pattern in the time-series given by the VARIABLES collection. If the
                     pattern does not occur, VALUE takes the default value $-\infty$.
                     An occurrence of the pattern INCREASING_SEQUENCE is the *maximal* subsequence
                     which matches the regular expression $< (< | =)^* < | <$.
                     Assume that the occurrence of the pattern INCREASING_SEQUENCE starts at position $i$
                     and ends at position $j$. The feature MAX computes the maximum of the values from
                     index $i$ to index $j + 1$.

**Example**          $(6, \langle 4, 3, 5, 5, 2, 1, 1, 3, 3, 4, 6, 6, 3, 1, 1, 3, 3 \rangle)$

                     Figure 3.502 provides an example where the MAX_MAX_INCREASING_SEQUENCE
                     $(6, [4, 3, 5, 5, 2, 1, 1, 3, 3, 4, 6, 6, 3, 1, 1, 3, 3])$ constraint holds.

**Typical**          $|\text{VARIABLES}| > 1$
                     range(VARIABLES.var) > 1

**Arg. properties**  Functional dependency: VALUE determined by VARIABLES.

Figure 3.502: Illustrating the MAX_MAX_INCREASING_SEQUENCE constraint of the **Example** slot

**Automaton**     Figures 3.503 and 3.504 respectively depict the automaton associated with the constraint MAX_MAX_INCREASING_SEQUENCE and its simplified form.

Figure 3.503: Automaton for the MAX_MAX_INCREASING_SEQUENCE constraint obtained by applying decoration Table 2.35 to the seed transducer of the INCREASING_SEQUENCE pattern where default is $-\infty$

Figure 3.504: Automaton for the MAX_MAX_INCREASING_SEQUENCE constraint obtained by applying decoration Table 2.38 to the seed transducer of the INCREASING_SEQUENCE pattern where default is $-\infty$; $R_i - R_{i-1} \geq 0$ is a linear invariant.

|   | $s$ | $t$ |
|---|---|---|
| $s$ | $\max(\overrightarrow{C}, \overleftarrow{C})$ | $\max(\overrightarrow{C}, \overleftarrow{C})$ |
| $t$ | $\max(\overrightarrow{C}, \overleftarrow{C})$ | $\max(\overrightarrow{C}, \overleftarrow{C}, \overrightarrow{D}, \overleftarrow{D}, \texttt{VAR}_i)$ |

Table 3.12: Glue matrix for the MAX_MAX_INCREASING_SEQUENCE constraint defined as the composition of the INCREASING_SEQUENCE pattern, the feature MAX, and the aggregator max; cells of the glue matrix are coloured with the colour of the constituent to which they are related.

# 3.246   MAX_MAX_INFLEXION

▶ ▷ ◁ ◀        DESCRIPTION              AUTOMATON

$< (< | =)^* > | > (> | =)^* <$

**Origin**          Based on the INFLEXION pattern.

**Constraint**      MAX_MAX_INFLEXION(VALUE, VARIABLES)

**Arguments**       VALUE     : dvar
                    VARIABLES : collection(var−dvar)

**Restrictions**    $\mathtt{sv} \leq 2 \vee \mathtt{rv} \leq 1 \Rightarrow \mathtt{VALUE} = -\infty$
                    $\mathtt{VALUE} = -\infty \vee \mathtt{VALUE} \geq \mathtt{minv}$
                    $\mathtt{VALUE} \leq \mathtt{maxv}①$
                    required(VARIABLES, var)
                    where
                    maxv = maxval(VARIABLES.var)
                    minv = minval(VARIABLES.var)
                    $\mathtt{sv} = |\mathtt{VARIABLES}|$
                    rv = range(VARIABLES.var)

**Purpose**         VALUE is the maximum of all maximum values in each occurrence of the INFLEXION
                    pattern in the time-series given by the VARIABLES collection. If the pattern does not
                    occur, VALUE takes the default value $-\infty$.
                    An occurrence of the pattern INFLEXION is the *maximal* subsequence which matches the
                    regular expression $< (< | =)^* > | > (> | =)^* <$.
                    Assume that the occurrence of the pattern INFLEXION starts at position $i$ and ends at
                    position $j$. The feature MAX computes the maximum of the values from index $i + 1$ to
                    index $j$.

**Example**         $(6, \langle 1, 2, 6, 6, 4, 4, 3, 5, 2, 5, 1, 5, 3, 3, 4, 4 \rangle)$

                    Figure   3.505   provides   an   example   where   the   MAX_MAX_INFLEXION
                    $(6, [1, 2, 6, 6, 4, 4, 3, 5, 2, 5, 1, 5, 3, 3, 4, 4])$ constraint holds.

**Typical**         $|\mathtt{VARIABLES}| > 2$
                    range(VARIABLES.var) $> 1$

**Symmetry**        Items of VARIABLES can be reversed.

**Arg. properties** Functional dependency: VALUE determined by VARIABLES.

Figure 3.505: Illustrating the MAX_MAX_INFLEXION constraint of the **Example** slot

**Automaton**    Figures 3.506 and 3.507 respectively depict the automaton associated with the constraint MAX_MAX_INFLEXION and its simplified form.

Figure 3.506: Automaton for the MAX_MAX_INFLEXION constraint obtained by applying decoration Table 2.35 to the seed transducer of the INFLEXION pattern where `default` is $-\infty$ (transition $r \rightarrow t$ has the same accumulators updates as transition $t \rightarrow r$)

Figure 3.507: Automaton for the MAX_MAX_INFLEXION constraint obtained by applying decoration Table 2.25 to the seed transducer of the INFLEXION pattern where `default` is $-\infty$ (transition $r \rightarrow t$ has the same accumulators updates as transition $t \rightarrow r$); $R_i - R_{i-1} \geq 0$ is a linear invariant.

## 3.247 MAX_MAX_PEAK

▶ ▷ ◁ ◀          **DESCRIPTION**          **AUTOMATON**

$< (= | <)^* (> | =)^* >$

| | |
|---|---|
| **Origin** | Based on the PEAK pattern. |
| **Constraint** | MAX_MAX_PEAK(VALUE, VARIABLES) |
| **Arguments** | VALUE      :   dvar<br>VARIABLES : collection(var−dvar) |

**Restrictions**

$\mathtt{sv} \leq 2 \vee \mathtt{rv} \leq 1 \Rightarrow \mathtt{VALUE} = -\infty$
$\mathtt{VALUE} = -\infty \vee \mathtt{VALUE} \geq \mathtt{minv} + 1$
$\mathtt{VALUE} \leq \mathtt{maxv}①$
required(VARIABLES, var)
where
$\mathtt{maxv} = \mathtt{maxval}(\mathtt{VARIABLES.var})$
$\mathtt{minv} = \mathtt{minval}(\mathtt{VARIABLES.var})$
$\mathtt{sv} = |\mathtt{VARIABLES}|$
$\mathtt{rv} = \mathtt{range}(\mathtt{VARIABLES.var})$

VALUE ≤ **2**

**Purpose**

VALUE is the maximum of all maximum values in each occurrence of the PEAK pattern in the time-series given by the VARIABLES collection. If the pattern does not occur, VALUE takes the default value $-\infty$.

An occurrence of the pattern PEAK is the *maximal* subsequence which matches the regular expression $< (= | <)^* (> | =)^* >$.

Assume that the occurrence of the pattern PEAK starts at position $i$ and ends at position $j$. The feature MAX computes the maximum of the values from index $i + 1$ to index $j$.

**Example**

$(6, \langle 7, 5, 5, 1, 4, 5, 2, 2, 3, 5, 6, 2, 3, 3, 3, 1 \rangle)$

Figure 3.508 provides an example where the MAX_MAX_PEAK $(6, [7, 5, 5, 1, 4, 5, 2, 2, 3, 5, 6, 2, 3, 3, 3, 1])$ constraint holds.

| | |
|---|---|
| **Typical** | $|\mathtt{VARIABLES}| > 2$<br>range(VARIABLES.var) > 1 |
| **Symmetry** | Items of VARIABLES can be reversed. |
| **Arg. properties** | Functional dependency: VALUE determined by VARIABLES. |

Figure 3.508: Illustrating the MAX_MAX_PEAK constraint of the **Example** slot

**Automaton**  Figures 3.509 and 3.510 respectively depict the automaton associated with the constraint MAX_MAX_PEAK and its simplified form.

Figure 3.509: Automaton for the MAX_MAX_PEAK constraint obtained by applying decoration Table 2.35 to the seed transducer of the PEAK pattern where default is $-\infty$

Figure 3.510: Automaton for the MAX_MAX_PEAK constraint obtained by applying decoration Table 2.37 to the seed transducer of the PEAK pattern where default is $-\infty$; $R_i - R_{i-1} \geq 0$ is a linear invariant.

|     | $s$ | $r$ | $t$ |
| --- | --- | --- | --- |
| $s$ | $\max(\overrightarrow{C}, \overleftarrow{C})$ | $\max(\overrightarrow{C}, \overleftarrow{C})$ | $\max(\overrightarrow{C}, \overleftarrow{C})$ |
| $r$ | $\max(\overrightarrow{C}, \overleftarrow{C})$ | $\max(\overrightarrow{D}, \overleftarrow{D}, \mathtt{VAR}_i)$ | $\max(\overleftarrow{C}, \overrightarrow{D}, \overleftarrow{D}, \mathtt{VAR}_i)$ |
| $t$ | $\max(\overrightarrow{C}, \overleftarrow{C})$ | $\max(\overrightarrow{C}, \overrightarrow{D}, \overleftarrow{D}, \mathtt{VAR}_i)$ | $\max(\overrightarrow{C}, \overleftarrow{C})$ |

Table 3.13: Glue matrix for the MAX_MAX_PEAK constraint defined as the composition of the `PEAK pattern`, the feature MAX, and the `aggregator max`; cells of the glue matrix are coloured with the colour of the constituent to which they are related.

# 3.248   MAX_MAX_STRICTLY_DECREASING_SEQUENCE

▶ ▷ ◁ ◀        **DESCRIPTION**          **AUTOMATON**

| | |
|---|---|
| **Origin** | Based on the STRICTLY_DECREASING_SEQUENCE pattern. |
| **Constraint** | MAX_MAX_STRICTLY_DECREASING_SEQUENCE(VALUE, VARIABLES) |
| **Arguments** | VALUE      : dvar <br> VARIABLES : collection(var−dvar) |
| **Restrictions** | $sv \leq 1 \vee rv \leq 1 \Rightarrow VALUE = -\infty$ <br> $VALUE = -\infty \vee VALUE \geq minv + 1$ <br> $VALUE \leq maxv①$ <br> required(VARIABLES, var) <br> where <br> maxv = maxval(VARIABLES.var) <br> minv = minval(VARIABLES.var) <br> $sv = |VARIABLES|$ <br> rv = range(VARIABLES.var) |

**Purpose**

VALUE is the maximum of all maximum values in each occurrence of the STRICTLY_DECREASING_SEQUENCE pattern in the time-series given by the VARIABLES collection. If the pattern does not occur, VALUE takes the default value $-\infty$.

An occurrence of the pattern STRICTLY_DECREASING_SEQUENCE is the *maximal* subsequence which matches the regular expression $>^+$.

Assume that the occurrence of the pattern STRICTLY_DECREASING_SEQUENCE starts at position $i$ and ends at position $j$. The feature MAX computes the maximum of the values from index $i$ to index $j + 1$.

| | |
|---|---|
| **Example** | $(6, \langle 4, 4, 6, 4, 1, 1, 3, 4, 4, 6, 6, 5, 2, 2, 4, 3 \rangle)$ |

Figure 3.511 provides an example where the MAX_MAX_STRICTLY_DECREASING_SEQUENCE $(6, [4, 4, 6, 4, 1, 1, 3, 4, 4, 6, 6, 5, 2, 2, 4, 3])$ constraint holds.

| | |
|---|---|
| **Typical** | $|VARIABLES| > 1$ <br> range(VARIABLES.var) > 1 |
| **Arg. properties** | Functional dependency: VALUE determined by VARIABLES. |

Figure 3.511: Illustrating the MAX_MAX_STRICTLY_DECREASING_SEQUENCE constraint of the **Example** slot

**Automaton**    Figures 3.512 and 3.513 respectively depict the automaton associated with the constraint
MAX_MAX_STRICTLY_DECREASING_SEQUENCE and its simplified form.

Figure 3.512: Automaton for the MAX_MAX_STRICTLY_DECREASING_SEQUENCE constraint obtained by applying decoration Table 2.35 to the seed transducer of the STRICTLY_DECREASING_SEQUENCE pattern where default is $-\infty$

Figure 3.513: Automaton for the MAX_MAX_STRICTLY_DECREASING_SEQUENCE constraint obtained by applying decoration Table 2.37 to the seed transducer of the STRICTLY_DECREASING_SEQUENCE pattern where default is $-\infty$; $R_i - R_{i-1} \geq 0$ is a linear invariant.

|     | $s$ | $r$ |
| --- | --- | --- |
| $s$ | $\max(\overrightarrow{C}, \overleftarrow{C})$ | $\max(\overrightarrow{C}, \overleftarrow{C})$ |
| $r$ | $\max(\overrightarrow{C}, \overleftarrow{C})$ | $\max(\overrightarrow{C}, \overleftarrow{C}, \overrightarrow{D}, \overleftarrow{D}, \mathtt{VAR}_i)$ |

Table 3.14: Glue matrix for the MAX_MAX_STRICTLY_DECREASING_SEQUENCE constraint defined as the composition of the STRICTLY_DECREASING_SEQUENCE pattern, the feature MAX, and the aggregator max; cells of the glue matrix are coloured with the colour of the constituent to which they are related.

## 3.249   MAX_MAX_STRICTLY_INCREASING_SEQUENCE

▶ ▷ ◁ ◀        **DESCRIPTION**          **AUTOMATON**

| | |
|---|---|
| **Origin** | Based on the STRICTLY_INCREASING_SEQUENCE pattern. |
| **Constraint** | MAX_MAX_STRICTLY_INCREASING_SEQUENCE(VALUE, VARIABLES) |
| **Arguments** | VALUE      :  dvar <br> VARIABLES  :  collection(var−dvar) |
| **Restrictions** | $\text{sv} \leq 1 \lor \text{rv} \leq 1 \Rightarrow \text{VALUE} = -\infty$ <br> $\text{VALUE} = -\infty \lor \text{VALUE} \geq \text{minv} + 1$ <br> $\text{VALUE} \leq \text{maxv}$① <br> required(VARIABLES, var) <br> where <br> maxv =maxval(VARIABLES.var) <br> minv =minval(VARIABLES.var) <br> sv = \|VARIABLES\| <br> rv =range(VARIABLES.var) |
| **Purpose** | VALUE is the maximum of all maximum values in each occurrence of the STRICTLY_INCREASING_SEQUENCE pattern in the time-series given by the VARIABLES collection. If the pattern does not occur, VALUE takes the default value $-\infty$. <br> An occurrence of the pattern STRICTLY_INCREASING_SEQUENCE is the *maximal* sub-sequence which matches the regular expression $<^+$. <br> Assume that the occurrence of the pattern STRICTLY_INCREASING_SEQUENCE starts at position $i$ and ends at position $j$. The feature MAX computes the maximum of the values from index $i$ to index $j + 1$. |
| **Example** | $(6, \langle 4, 3, 5, 5, 2, 1, 1, 2, 3, 4, 6, 6, 3, 1, 1, 2, 3 \rangle)$ |
| | Figure 3.514 provides an example where the MAX_MAX_STRICTLY_INCREASING_SEQUENCE $(6, [4, 3, 5, 5, 2, 1, 1, 2, 3, 4, 6, 6, 3, 1, 1, 2, 3])$ constraint holds. |
| **Typical** | $\|\text{VARIABLES}\| > 1$ <br> range(VARIABLES.var) > 1 |
| **Arg. properties** | Functional dependency: VALUE determined by VARIABLES. |

Figure 3.514: Illustrating the MAX_MAX_STRICTLY_INCREASING_SEQUENCE constraint of the **Example** slot

 Figures 3.515 and 3.516 respectively depict the automaton associated with the constraint MAX_MAX_STRICTLY_INCREASING_SEQUENCE and its simplified form.

Figure 3.515: Automaton for the MAX_MAX_STRICTLY_INCREASING_SEQUENCE constraint obtained by applying decoration Table 2.35 to the seed transducer of the STRICTLY_INCREASING_SEQUENCE pattern where default is $-\infty$

Figure 3.516: Automaton for the MAX_MAX_STRICTLY_INCREASING_SEQUENCE constraint obtained by applying decoration Table 2.38 to the seed transducer of the STRICTLY_INCREASING_SEQUENCE pattern where default is $-\infty$; $R_i - R_{i-1} \geq 0$ is a linear invariant.

|   | $s$ | $r$ |
|---|---|---|
| $s$ | $\max(\overrightarrow{C}, \overleftarrow{C})$ | $\max(\overrightarrow{C}, \overleftarrow{C})$ |
| $r$ | $\max(\overrightarrow{C}, \overleftarrow{C})$ | $\max(\overrightarrow{C}, \overleftarrow{C}, \overrightarrow{D}, \overleftarrow{D}, \text{VAR}_i)$ |

Table 3.15: Glue matrix for the MAX_MAX_STRICTLY_INCREASING_SEQUENCE constraint defined as the composition of the STRICTLY_INCREASING_SEQUENCE pattern, the feature MAX, and the aggregator max; cells of the glue matrix are coloured with the colour of the constituent to which they are related.

## 3.250   MAX_MAX_SUMMIT

▶ ▷ ◁ ◀        **DESCRIPTION**        **AUTOMATON**

$(< | < (= | <)^* <)(> | > (= | >)^* >)$

| | |
|---|---|
| **Origin** | Based on the SUMMIT pattern. |
| **Constraint** | MAX_MAX_SUMMIT(VALUE, VARIABLES) |
| **Arguments** | VALUE       : dvar <br> VARIABLES : collection(var−dvar) |

**Restrictions**

$\mathtt{sv} \leq 2 \vee \mathtt{rv} \leq 1 \Rightarrow \mathtt{VALUE} = -\infty$
$\mathtt{VALUE} = -\infty \vee \mathtt{VALUE} \geq \mathtt{minv} + 1$
$\mathtt{VALUE} \leq \mathtt{maxv}①$
required(VARIABLES, var)
where
maxv = maxval(VARIABLES.var)
minv = minval(VARIABLES.var)
$\mathtt{sv} = |\mathtt{VARIABLES}|$
rv = range(VARIABLES.var)

VALUE ≤ **2**

**Purpose**

VALUE is the maximum of all maximum values in each occurrence of the SUMMIT pattern in the time-series given by the VARIABLES collection. If the pattern does not occur, VALUE takes the default value $-\infty$.

An occurrence of the pattern SUMMIT is the *maximal* subsequence which matches the regular expression $(< | < (= | <)^* <)(> | > (= | >)^* >)$.

Assume that the occurrence of the pattern SUMMIT starts at position $i$ and ends at position $j$. The feature MAX computes the maximum of the values from index $i + 1$ to index $j$.

**Example**

$(5, \langle 7, 1, 5, 4, 4, 3, 3, 4, 6, 6, 2, 3, 4, 2, 3, 1 \rangle)$

Figure 3.517 provides an example where the MAX_MAX_SUMMIT $(5, [7, 1, 5, 4, 4, 3, 3, 4, 6, 6, 2, 3, 4, 2, 3, 1])$ constraint holds.

| | |
|---|---|
| **Typical** | $|\mathtt{VARIABLES}| > 2$ <br> range(VARIABLES.var) > 1 |
| **Symmetry** | Items of VARIABLES can be reversed. |
| **Arg. properties** | Functional dependency: VALUE determined by VARIABLES. |

Figure 3.517: Illustrating the MAX_MAX_SUMMIT constraint of the **Example** slot

**Automaton**     Figures 3.518 and 3.519 respectively depict the automaton associated with the constraint MAX_MAX_SUMMIT and its simplified form.

Figure 3.518: Automaton for the MAX_MAX_SUMMIT constraint obtained by applying decoration Table 2.35 to the seed transducer of the SUMMIT pattern where default is $-\infty$ (transition $u \to r$ has the same accumulator update as transition $r \to u$)

| | $s$ | $r$ | $t$ | $u$ |
|---|---|---|---|---|
| $s$ | $\max(\overrightarrow{C}, \overleftarrow{C})$ | $\max(\overrightarrow{C}, \overleftarrow{C})$ | $\max(\overrightarrow{C}, \overleftarrow{C})$ | $\max(\overrightarrow{C}, \overleftarrow{C})$ |
| $r$ | $\max(\overrightarrow{C}, \overleftarrow{C})$ | $\max(\overrightarrow{D}, \overleftarrow{D}, \text{VAR}_i)$ | $\max(\overleftarrow{C}, \overrightarrow{D}, \overleftarrow{D}, \text{VAR}_i)$ | $\max(\overrightarrow{C}, \overleftarrow{C})$ |
| $t$ | $\max(\overrightarrow{C}, \overleftarrow{C})$ | $\max(\overrightarrow{C}, \overrightarrow{D}, \overleftarrow{D}, \text{VAR}_i)$ | $\max(\overrightarrow{C}, \overleftarrow{C})$ | $\max(\overrightarrow{C}, \overrightarrow{D}, \overleftarrow{D}, \text{VAR}_i)$ |
| $u$ | $\max(\overrightarrow{C}, \overleftarrow{C})$ | $\max(\overrightarrow{C}, \overleftarrow{C})$ | $\max(\overleftarrow{C}, \overrightarrow{D}, \overleftarrow{D}, \text{VAR}_i)$ | $\max(\overrightarrow{C}, \overleftarrow{C})$ |

Table 3.16: Glue matrix for the MAX_MAX_SUMMIT constraint defined as the composition of the SUMMIT pattern , the feature MAX , and the aggregator max ; cells of the glue matrix are coloured with the colour of the constituent to which they are related.

Figure 3.519: Automaton for the MAX\_MAX\_SUMMIT constraint obtained by applying decoration Table 2.37 to the seed transducer of the SUMMIT pattern where default is $-\infty$ (transition $u \to r$ has the same accumulator update as transition $r \to u$); $R_i - R_{i-1} \geq 0$ is a linear invariant.

## 3.251  MAX_MAX_ZIGZAG

► ▷ ◁ ◀         **DESCRIPTION**         **AUTOMATON**

$(<>)^+(< | <>) | (><)^+(> | ><)$

| | |
|---|---|
| **Origin** | Based on the ZIGZAG pattern. |
| **Constraint** | MAX_MAX_ZIGZAG(VALUE, VARIABLES) |
| **Arguments** | VALUE      : dvar |
| | VARIABLES : collection(var−dvar) |

**Restrictions**

$\text{sv} \leq 3 \vee \text{rv} \leq 1 \Rightarrow \text{VALUE} = -\infty$

$\text{VALUE} = -\infty \vee \text{VALUE} \geq \text{minv} + 1$

$\text{VALUE} \leq \text{maxv}①$

required(VARIABLES, var)

where

maxv = maxval(VARIABLES.var)

minv = minval(VARIABLES.var)

sv = |VARIABLES|

rv = range(VARIABLES.var)

VALUE ≤ 2

**Purpose**

VALUE is the maximum of all maximum values in each occurrence of the ZIGZAG pattern in the time-series given by the VARIABLES collection. If the pattern does not occur, VALUE takes the default value $-\infty$.

An occurrence of the pattern ZIGZAG is the *maximal* subsequence which matches the regular expression $(<>)^+(< | <>) | (><)^+(> | ><)$.

Assume that the occurrence of the pattern ZIGZAG starts at position $i$ and ends at position $j$. The feature MAX computes the maximum of the values from index $i + 1$ to index $j$.

**Example**

$(7, \langle 4, 1, 3, 1, 4, 6, 1, 5, 5, 2, 7, 2, 3, 1, 6, 1 \rangle)$

Figure 3.520 provides an example where the MAX_MAX_ZIGZAG $(7, [4, 1, 3, 1, 4, 6, 1, 5, 5, 2, 7, 2, 3, 1, 6, 1])$ constraint holds.

| | |
|---|---|
| **Typical** | $|\text{VARIABLES}| > 3$ |
| | range(VARIABLES.var) > 1 |
| **Symmetry** | Items of VARIABLES can be reversed. |
| **Arg. properties** | Functional dependency: VALUE determined by VARIABLES. |

Figure 3.520: Illustrating the MAX_MAX_ZIGZAG constraint of the **Example** slot

**Automaton**          Figures 3.521 and 3.522 respectively depict the automaton associated with the constraint
                       MAX_MAX_ZIGZAG and its simplified form.

Figure 3.521: Automaton for the MAX_MAX_ZIGZAG constraint obtained by applying decoration Table 2.35 to the seed transducer of the ZIGZAG pattern where default is $-\infty$; (1) missing transitions from $a, b, c, d, e, f$ to $s$ are labelled by $=$; (2) on transitions from $b, c, e, f$ to $s$ the accumulator $D$ is reset to its initial value; (3) on transitions from $c, f$ to $s$ the accumulator $R$ is updated wrt $C$ and the accumulator $C$ is reset to its initial value

Figure 3.522: Automaton for the MAX_MAX_ZIGZAG constraint obtained by applying decoration Table 2.28 to the seed transducer of the ZIGZAG pattern where default is $-\infty$; (1) missing transitions from $a, b, c, d, e, f$ to $s$ are labelled by $=$; (2) on transitions from $b, c, e, f$ to $s$ the accumulator $D$ is reset to its initial value; $R_i - R_{i-1} \geq 0$ is a linear invariant.

| | $s$ | $a$ | $b$ | $c$ | $d$ | $e$ | $f$ |
|---|---|---|---|---|---|---|---|
| $s$ | $\max(\overrightarrow{C},\overleftarrow{C})$ | $\max(\overrightarrow{C},\overleftarrow{C})$ | $\max(\overrightarrow{C},\overleftarrow{C})$ | $\max(\overrightarrow{C},\overleftarrow{C})$ | $\max(\overrightarrow{C},\overleftarrow{C})$ | $\max(\overrightarrow{C},\overleftarrow{C})$ | $\max(\overrightarrow{C},\overleftarrow{C})$ |
| $a$ | $\max(\overrightarrow{C},\overleftarrow{C})$ | $\max(\overrightarrow{C},\overleftarrow{C})$ | $\max(\overrightarrow{C},\overleftarrow{C})$ | $\max(\overleftarrow{C},\overrightarrow{D},\overleftarrow{D},\text{VAR}_i)$ | $\max(\overrightarrow{C},\overleftarrow{C})$ | $\max(\overrightarrow{D},\overleftarrow{D},\text{VAR}_i)$ | $\max(\overrightarrow{C},\overleftarrow{C})$ |
| $b$ | $\max(\overrightarrow{C},\overleftarrow{C})$ | $\max(\overrightarrow{C},\overleftarrow{C})$ | $\max(\overrightarrow{D},\overleftarrow{D},\text{VAR}_i)$ | $\max(\overrightarrow{C},\overleftarrow{C})$ | $\max(\overrightarrow{D},\overleftarrow{D},\text{VAR}_i)$ | $\max(\overrightarrow{C},\overleftarrow{C})$ | $\max(\overleftarrow{C},\overrightarrow{D},\overleftarrow{D},\text{VAR}_i)$ |
| $c$ | $\max(\overrightarrow{C},\overleftarrow{C})$ | $\max(\overrightarrow{C},\overrightarrow{D},\overleftarrow{D},\text{VAR}_i)$ | $\max(\overrightarrow{C},\overleftarrow{C})$ | $\max(\overrightarrow{C},\overleftarrow{C},\overrightarrow{D},\overleftarrow{D},\text{VAR}_i)$ | $\max(\overrightarrow{C},\overleftarrow{C})$ | $\max(\overrightarrow{C},\overrightarrow{D},\overleftarrow{D},\text{VAR}_i)$ | $\max(\overrightarrow{C},\overleftarrow{C})$ |
| $d$ | $\max(\overrightarrow{C},\overleftarrow{C})$ | $\max(\overrightarrow{C},\overleftarrow{C})$ | $\max(\overrightarrow{D},\overleftarrow{D},\text{VAR}_i)$ | $\max(\overrightarrow{C},\overleftarrow{C})$ | $\max(\overrightarrow{C},\overleftarrow{C})$ | $\max(\overrightarrow{C},\overleftarrow{C})$ | $\max(\overleftarrow{C},\overrightarrow{D},\overleftarrow{D},\text{VAR}_i)$ |
| $e$ | $\max(\overrightarrow{C},\overleftarrow{C})$ | $\max(\overrightarrow{D},\overleftarrow{D},\text{VAR}_i)$ | $\max(\overrightarrow{C},\overleftarrow{C})$ | $\max(\overleftarrow{C},\overrightarrow{D},\overleftarrow{D},\text{VAR}_i)$ | $\max(\overrightarrow{C},\overleftarrow{C})$ | $\max(\overrightarrow{D},\overleftarrow{D},\text{VAR}_i)$ | $\max(\overrightarrow{C},\overleftarrow{C})$ |
| $f$ | $\max(\overrightarrow{C},\overleftarrow{C})$ | $\max(\overrightarrow{C},\overleftarrow{C})$ | $\max(\overrightarrow{C},\overrightarrow{D},\overleftarrow{D},\text{VAR}_i)$ | $\max(\overrightarrow{C},\overleftarrow{C})$ | $\max(\overrightarrow{C},\overrightarrow{D},\overleftarrow{D},\text{VAR}_i)$ | $\max(\overrightarrow{C},\overleftarrow{C})$ | $\max(\overrightarrow{C},\overleftarrow{C},\overrightarrow{D},\overleftarrow{D},\text{VAR}_i)$ |

Table 3.17: Glue matrix for the MAX_MAX_ZIGZAG constraint defined as the composition of the ZIGZAG pattern, the feature MAX, and the aggregator max; cells of the glue matrix are coloured with the colour of the constituent to which they are related.

# 3.252　MAX_MIN_BUMP_ON_DECREASING_SEQUENCE

▶ ▷ ◁ ◀　　**DESCRIPTION**　　**AUTOMATON**

| | |
|---|---|
| **Origin** | Based on the BUMP_ON_DECREASING_SEQUENCE pattern. |
| **Constraint** | MAX_MIN_BUMP_ON_DECREASING_SEQUENCE(VALUE, VARIABLES) |
| **Arguments** | VALUE　　　　: dvar <br> VARIABLES : collection(var−dvar) |

**Restrictions**

$\mathtt{sv} \leq 5 \lor \mathtt{rv} \leq 2 \Rightarrow \mathtt{VALUE} = -\infty$
$\mathtt{VALUE} = -\infty \lor \mathtt{VALUE} \geq \mathtt{minv}$
$\mathtt{VALUE} \leq \mathtt{maxv} - 2$ ①
required(VARIABLES, var)
where
maxv = maxval(VARIABLES.var)
minv = minval(VARIABLES.var)
sv = |VARIABLES|
rv = range(VARIABLES.var)

**Purpose**

VALUE is the maximum of all minimum values in each occurrence of the BUMP_ON_DECREASING_SEQUENCE pattern in the time-series given by the VARIABLES collection. If the pattern does not occur, VALUE takes the default value $-\infty$.

An occurrence of the pattern BUMP_ON_DECREASING_SEQUENCE is the subsequence which matches the regular expression >><>>.

Assume that the occurrence of the pattern BUMP_ON_DECREASING_SEQUENCE starts at position $i$ and ends at position $j$. The feature MIN computes the minimum of the values from index $i + 2$ to index $j$.

**Example**

$(5, \langle 7, 6, 5, 6, 5, 4, 1, 4, 7, 5, 4, 2, 5, 4, 3, 3 \rangle)$

Figure 3.523 provides an example where the MAX_MIN_BUMP_ON_DECREASING_SEQUENCE $(5, [7, 6, 5, 6, 5, 4, 1, 4, 7, 5, 4, 2, 5, 4, 3, 3])$ constraint holds.

| | |
|---|---|
| **Typical** | $\|\mathtt{VARIABLES}\| > 5$ <br> range(VARIABLES.var) > 2 |
| **Arg. properties** | Functional dependency: VALUE determined by VARIABLES. |

Figure 3.523: Illustrating the MAX_MIN_BUMP_ON_DECREASING_SEQUENCE constraint of the **Example** slot

**Automaton**     Figures 3.524 and 3.525 respectively depict the automaton associated with the constraint
MAX_MIN_BUMP_ON_DECREASING_SEQUENCE and its simplified form.

Figure 3.524: Automaton for the MAX_MIN_BUMP_ON_DECREASING_SEQUENCE constraint obtained by applying decoration Table 2.35 to the seed transducer of the BUMP_ON_DECREASING_SEQUENCE pattern where default is $-\infty$

Figure 3.525: Automaton for the MAX_MIN_BUMP_ON_DECREASING_SEQUENCE constraint obtained by applying decoration Table 2.25 to the seed transducer of the BUMP_ON_DECREASING_SEQUENCE pattern where default is $-\infty$; $R_i - R_{i-1} \geq 0$ is a linear invariant.

# 3.253    MAX_MIN_DECREASING

    **DESCRIPTION**        **AUTOMATON**

| | |
|---|---|
| **Origin** | Based on the DECREASING pattern. |
| **Constraint** | MAX_MIN_DECREASING(VALUE, VARIABLES) |
| **Arguments** | VALUE      : dvar<br>VARIABLES : collection(var−dvar) |
| **Restrictions** | $\text{sv} \leq 1 \lor \text{rv} \leq 1 \Rightarrow \text{VALUE} = -\infty$<br>$\text{VALUE} = -\infty \lor \text{VALUE} \geq \text{minv}$<br>$\text{VALUE} \leq \text{maxv} - 1①$<br>required(VARIABLES, var)<br>where<br>maxv =maxval(VARIABLES.var)<br>minv =minval(VARIABLES.var)<br>sv = \|VARIABLES\|<br>rv =range(VARIABLES.var) |
| **Purpose** | VALUE is the maximum of all minimum values in each occurrence of the DECREASING pattern in the time-series given by the VARIABLES collection. If the pattern does not occur, VALUE takes the default value $-\infty$.<br>An occurrence of the pattern DECREASING is the subsequence which matches the regular expression >.<br>Assume that the occurrence of the pattern DECREASING starts at position $i$ and ends at position $j$. The feature MIN computes the minimum of the values from index $i$ to index $j + 1$. |
| **Example** | $(4, \langle 3, 4, 2, 2, 5, 6, 6, 4, 4, 3, 1, 1, 4, 6, 4, 4 \rangle)$ |

Figure 3.526 provides an example where the MAX_MIN_DECREASING $(4, [3, 4, 2, 2, 5, 6, 6, 4, 4, 3, 1, 1, 4, 6, 4, 4])$ constraint holds.

| | |
|---|---|
| **Typical** | $\|\text{VARIABLES}\| > 1$<br>range(VARIABLES.var) > 1 |
| **Arg. properties** | Functional dependency: VALUE determined by VARIABLES. |

Figure 3.526: Illustrating the MAX_MIN_DECREASING constraint of the **Example** slot

**Automaton**        Figures 3.527 and 3.528 respectively depict the automaton associated with the constraint
                    MAX_MIN_DECREASING and its simplified form.

Figure 3.527: Automaton for the MAX_MIN_DECREASING constraint obtained by applying decoration Table 2.35 to the seed transducer of the DECREASING pattern where `default` is $-\infty$

Figure 3.528: Automaton for the MAX_MIN_DECREASING constraint obtained by applying decoration Table 2.38 to the seed transducer of the DECREASING pattern where `default` is $-\infty$; $R_i - R_{i-1} \geq 0$ is a linear invariant.

Table 3.18: Glue matrix for the MAX_MIN_DECREASING constraint defined as the composition of the DECREASING pattern , the feature MIN , and the aggregator max ; cells of the glue matrix are coloured with the colour of the constituent to which they are related.

# 3.254   MAX_MIN_DECREASING_SEQUENCE

▶ ▷ ◁ ◀        **DESCRIPTION**        **AUTOMATON**

> (> | =)* > |
> >

| **Origin** | Based on the DECREASING_SEQUENCE pattern. |

| **Constraint** | MAX_MIN_DECREASING_SEQUENCE(VALUE, VARIABLES) |

| **Arguments** | VALUE      : dvar |
| | VARIABLES  : collection(var−dvar) |

**Restrictions**

$$\text{sv} \leq 1 \vee \text{rv} \leq 1 \Rightarrow \text{VALUE} = -\infty$$
$$\text{VALUE} = -\infty \vee \text{VALUE} \geq \text{minv}$$
$$\text{VALUE} \leq \text{maxv} - 1 ①$$
required(VARIABLES, var)
where
maxv = maxval(VARIABLES.var)
minv = minval(VARIABLES.var)
sv = |VARIABLES|
rv = range(VARIABLES.var)

VALUE ≤ 1

2
2 − 1
0
①

**Purpose**

VALUE is the maximum of all minimum values in each occurrence of the DECREAS-ING_SEQUENCE pattern in the time-series given by the VARIABLES collection. If the pattern does not occur, VALUE takes the default value $-\infty$.

An occurrence of the pattern DECREASING_SEQUENCE is the *maximal* subsequence which matches the regular expression > (> | =)* > | >.

Assume that the occurrence of the pattern DECREASING_SEQUENCE starts at position $i$ and ends at position $j$. The feature MIN computes the minimum of the values from index $i$ to index $j + 1$.

**Example**

$$(4, \langle 3, 4, 2, 2, 5, 6, 6, 4, 4, 3, 1, 1, 4, 6, 4, 4\rangle)$$

Figure 3.529 provides an example where the MAX_MIN_DECREASING_SEQUENCE $(4, [3, 4, 2, 2, 5, 6, 6, 4, 4, 3, 1, 1, 4, 6, 4, 4])$ constraint holds.

**Typical**

|VARIABLES| > 1
range(VARIABLES.var) > 1

**Arg. properties**

Functional dependency: VALUE determined by VARIABLES.

Figure 3.529: Illustrating the MAX_MIN_DECREASING_SEQUENCE constraint of the **Example** slot

**Automaton**   Figures 3.530 and 3.531 respectively depict the automaton associated with the constraint MAX_MIN_DECREASING_SEQUENCE and its simplified form.

Figure 3.530: Automaton for the MAX_MIN_DECREASING_SEQUENCE constraint obtained by applying decoration Table 2.35 to the seed transducer of the DECREASING_SEQUENCE pattern where default is $-\infty$

Figure 3.531: Automaton for the MAX_MIN_DECREASING_SEQUENCE constraint obtained by applying decoration Table 2.24 to the seed transducer of the DECREASING_SEQUENCE pattern where default is $-\infty$; $R_i - R_{i-1} \geq 0$ is a linear invariant.

|   | $s$ | $t$ |
|---|---|---|
| $s$ | $\max(\overrightarrow{C}, \overleftarrow{C})$ | $\max(\overrightarrow{C}, \overleftarrow{C})$ |
| $t$ | $\max(\overrightarrow{C}, \overleftarrow{C})$ | $\min(\overrightarrow{C}, \overleftarrow{C}, \overrightarrow{D}, \overleftarrow{D}, \texttt{VAR}_i)$ |

Table 3.19: Glue matrix for the MAX_MIN_DECREASING_SEQUENCE constraint defined as the composition of the DECREASING_SEQUENCE pattern, the feature MIN, and the aggregator max; cells of the glue matrix are coloured with the colour of the constituent to which they are related.

# 3.255 MAX_MIN_DIP_ON_INCREASING_SEQUENCE

▶ ▷ ◁ ◀      **DESCRIPTION**      **AUTOMATON**

**Origin**          Based on the DIP_ON_INCREASING_SEQUENCE pattern.

**Constraint**      MAX_MIN_DIP_ON_INCREASING_SEQUENCE(VALUE, VARIABLES)

**Arguments**       VALUE     : dvar
                    VARIABLES : collection(var−dvar)

**Restrictions**    $\text{sv} \leq 5 \vee \text{rv} \leq 2 \Rightarrow \text{VALUE} = -\infty$
                    $\text{VALUE} = -\infty \vee \text{VALUE} \geq \text{minv}$
                    $\text{VALUE} \leq \text{maxv} - 2①$
                    required(VARIABLES, var)
                    where
                    maxv =maxval(VARIABLES.var)
                    minv =minval(VARIABLES.var)
                    sv =|VARIABLES|
                    rv =range(VARIABLES.var)

**Purpose**         VALUE is the maximum of all minimum values in each occurrence of the
                    DIP_ON_INCREASING_SEQUENCE pattern in the time-series given by the VARIABLES
                    collection. If the pattern does not occur, VALUE takes the default value $-\infty$.
                    An occurrence of the pattern DIP_ON_INCREASING_SEQUENCE is the subsequence
                    which matches the regular expression <<><<.
                    Assume that the occurrence of the pattern DIP_ON_INCREASING_SEQUENCE starts at
                    position $i$ and ends at position $j$. The feature MIN computes the minimum of the values
                    from index $i + 2$ to index $j$.

**Example**         $(2, \langle 1, 2, 3, 2, 5, 6, 7, 4, 1, 3, 4, 6, 1, 2, 4, 4 \rangle)$

                    Figure 3.532 provides an example where the MAX_MIN_DIP_ON_INCREASING_SEQUENCE
                    $(2, [1, 2, 3, 2, 5, 6, 7, 4, 1, 3, 4, 6, 1, 2, 4, 4])$ constraint holds.

**Typical**         $|\text{VARIABLES}| > 5$
                    range(VARIABLES.var) > 2

**Arg. properties** Functional dependency: VALUE determined by VARIABLES.

Figure 3.532: Illustrating the MAX_MIN_DIP_ON_INCREASING_SEQUENCE constraint of the **Example** slot

**Automaton**     Figures 3.533 and 3.534 respectively depict the automaton associated with the constraint
MAX_MIN_DIP_ON_INCREASING_SEQUENCE and its simplified form.

Figure 3.533: Automaton for the MAX_MIN_DIP_ON_INCREASING_SEQUENCE constraint obtained by applying decoration Table 2.35 to the seed transducer of the DIP_ON_INCREASING_SEQUENCE pattern where default is $-\infty$

Figure 3.534: Automaton for the MAX_MIN_DIP_ON_INCREASING_SEQUENCE constraint obtained by applying decoration Table 2.27 to the seed transducer of the DIP_ON_INCREASING_SEQUENCE pattern where default is $-\infty$; $R_i - R_{i-1} \geq 0$ is a linear invariant.

## 3.256 MAX_MIN_GORGE

▶ ▷ ◁ ◀        **DESCRIPTION**        **AUTOMATON**

$(> | > (= | >)^* >)(< | < (= | <)^* <)$

| | |
|---|---|
| **Origin** | Based on the GORGE pattern. |
| **Constraint** | MAX_MIN_GORGE(VALUE, VARIABLES) |
| **Arguments** | VALUE       : dvar<br>VARIABLES  : collection(var−dvar) |

**Restrictions**

$\mathtt{sv} \leq 2 \vee \mathtt{rv} \leq 1 \Rightarrow \mathtt{VALUE} = -\infty$
$\mathtt{VALUE} = -\infty \vee \mathtt{VALUE} \geq \mathtt{minv}$
$\mathtt{VALUE} \leq \mathtt{maxv} - 1$ ①
required(VARIABLES, var)
where
$\mathtt{maxv} =$maxval(VARIABLES.var)
$\mathtt{minv} =$minval(VARIABLES.var)
$\mathtt{sv} = |\mathtt{VARIABLES}|$
$\mathtt{rv} =$range(VARIABLES.var)

**Purpose**

VALUE is the maximum of all minimum values in each occurrence of the GORGE pattern in the time-series given by the VARIABLES collection. If the pattern does not occur, VALUE takes the default value $-\infty$.
An occurrence of the pattern GORGE is the *maximal* subsequence which matches the regular expression $(> | > (= | >)^* >)(< | < (= | <)^* <)$.
Assume that the occurrence of the pattern GORGE starts at position $i$ and ends at position $j$. The feature MIN computes the minimum of the values from index $i + 1$ to index $j$.

**Example**

$(5, \langle 1, 7, 3, 4, 4, 5, 5, 4, 2, 2, 6, 5, 4, 6, 5, 7 \rangle)$

Figure 3.535 provides an example where the MAX_MIN_GORGE $(5, [1, 7, 3, 4, 4, 5, 5, 4, 2, 2, 6, 5, 4, 6, 5, 7])$ constraint holds.

| | |
|---|---|
| **Typical** | $|\mathtt{VARIABLES}| > 2$<br>range(VARIABLES.var) $> 1$ |
| **Symmetry** | Items of VARIABLES can be reversed. |
| **Arg. properties** | Functional dependency: VALUE determined by VARIABLES. |

Figure 3.535: Illustrating the MAX_MIN_GORGE constraint of the **Example** slot

     Figures 3.536 and 3.537 respectively depict the automaton associated with the constraint
MAX_MIN_GORGE and its simplified form.

Figure 3.536: Automaton for the MAX_MIN_GORGE constraint obtained by applying decoration Table 2.35 to the seed transducer of the GORGE pattern where default is $-\infty$ (transition $u \rightarrow r$ has the same accumulator update as transition $r \rightarrow u$)

|   | $s$ | $r$ | $t$ | $u$ |
|---|---|---|---|---|
| $s$ | $\max(\overrightarrow{C}, \overleftarrow{C})$ | $\max(\overrightarrow{C}, \overleftarrow{C})$ | $\max(\overrightarrow{C}, \overleftarrow{C})$ | $\max(\overrightarrow{C}, \overleftarrow{C})$ |
| $r$ | $\max(\overrightarrow{C}, \overleftarrow{C})$ | $\min(\overrightarrow{D}, \overleftarrow{D}, \mathtt{VAR}_i)$ | $\min(\overleftarrow{C}, \overrightarrow{D}, \overleftarrow{D}, \mathtt{VAR}_i)$ | $\max(\overrightarrow{C}, \overleftarrow{C})$ |
| $t$ | $\max(\overrightarrow{C}, \overleftarrow{C})$ | $\min(\overrightarrow{C}, \overrightarrow{D}, \overleftarrow{D}, \mathtt{VAR}_i)$ | $\max(\overrightarrow{C}, \overleftarrow{C})$ | $\min(\overrightarrow{C}, \overrightarrow{D}, \overleftarrow{D}, \mathtt{VAR}_i)$ |
| $u$ | $\max(\overrightarrow{C}, \overleftarrow{C})$ | $\max(\overrightarrow{C}, \overleftarrow{C})$ | $\min(\overleftarrow{C}, \overrightarrow{D}, \overleftarrow{D}, \mathtt{VAR}_i)$ | $\max(\overrightarrow{C}, \overleftarrow{C})$ |

Table 3.20: Glue matrix for the MAX_MIN_GORGE constraint defined as the composition of the GORGE pattern , the feature MIN , and the aggregator max ; cells of the glue matrix are coloured with the colour of the constituent to which they are related.

Figure 3.537: Automaton for the MAX_MIN_GORGE constraint obtained by applying decoration Table 2.37 to the seed transducer of the GORGE pattern where `default` is $-\infty$; $R_i - R_{i-1} \geq 0$ is a linear invariant.

# 3.257   MAX_MIN_INCREASING

▶ ▷ ◁ ◀        **DESCRIPTION**        **AUTOMATON**

| | |
|---|---|
| **Origin** | Based on the INCREASING pattern. |
| **Constraint** | MAX_MIN_INCREASING(VALUE, VARIABLES) |
| **Arguments** | VALUE      : dvar<br>VARIABLES : collection(var−dvar) |

**Restrictions**

$$\mathtt{sv} \le 1 \lor \mathtt{rv} \le 1 \Rightarrow \mathtt{VALUE} = -\infty$$
$$\mathtt{VALUE} = -\infty \lor \mathtt{VALUE} \ge \mathtt{minv}$$
$$\mathtt{VALUE} \le \mathtt{maxv} - 1 ①$$
required(VARIABLES, var)
where
$$\mathtt{maxv} = \mathtt{maxval}(\mathtt{VARIABLES.var})$$
$$\mathtt{minv} = \mathtt{minval}(\mathtt{VARIABLES.var})$$
$$\mathtt{sv} = |\mathtt{VARIABLES}|$$
$$\mathtt{rv} = \mathtt{range}(\mathtt{VARIABLES.var})$$

**Purpose**

VALUE is the maximum of all minimum values in each occurrence of the INCREASING pattern in the time-series given by the VARIABLES collection. If the pattern does not occur, VALUE takes the default value $-\infty$.

An occurrence of the pattern INCREASING is the subsequence which matches the regular expression $<$.

Assume that the occurrence of the pattern INCREASING starts at position $i$ and ends at position $j$. The feature MIN computes the minimum of the values from index $i$ to index $j + 1$.

**Example**

$$(4, \langle 4, 3, 5, 5, 2, 1, 1, 3, 3, 4, 6, 6, 3, 1, 3, 3 \rangle)$$

Figure 3.538 provides an example where the MAX_MIN_INCREASING $(4, [4, 3, 5, 5, 2, 1, 1, 3, 3, 4, 6, 6, 3, 1, 3, 3])$ constraint holds.

| | |
|---|---|
| **Typical** | $|\mathtt{VARIABLES}| > 1$<br>range(VARIABLES.var) $> 1$ |
| **Arg. properties** | Functional dependency: VALUE determined by VARIABLES. |

Figure 3.538: Illustrating the MAX_MIN_INCREASING constraint of the **Example** slot

   Figures 3.539 and 3.540 respectively depict the automaton associated with the constraint MAX_MIN_INCREASING and its simplified form.

Figure 3.539: Automaton for the MAX_MIN_INCREASING constraint obtained by applying decoration Table 2.35 to the seed transducer of the INCREASING pattern where `default` is $-\infty$

Figure 3.540: Automaton for the MAX_MIN_INCREASING constraint obtained by applying decoration Table 2.37 to the seed transducer of the INCREASING pattern where `default` is $-\infty$; $R_i - R_{i-1} \geq 0$ is a linear invariant.

Table 3.21: Glue matrix for the MAX_MIN_INCREASING constraint defined as the composition of the INCREASING pattern , the feature MIN , and the aggregator max ; cells of the glue matrix are coloured with the colour of the constituent to which they are related.

## 3.258    MAX_MIN_INCREASING_SEQUENCE

▶ ▷ ◁ ◀        **DESCRIPTION**        **AUTOMATON**

| **Origin** | Based on the INCREASING_SEQUENCE pattern. |

| **Constraint** | MAX_MIN_INCREASING_SEQUENCE(VALUE, VARIABLES) |

| **Arguments** | VALUE      : dvar<br>VARIABLES  : collection(var−dvar) |

**Restrictions**

$$\mathtt{sv} \leq 1 \vee \mathtt{rv} \leq 1 \Rightarrow \mathtt{VALUE} = -\infty$$
$$\mathtt{VALUE} = -\infty \vee \mathtt{VALUE} \geq \mathtt{minv}$$
$$\mathtt{VALUE} \leq \mathtt{maxv} - 1①$$
$$\mathtt{required}(\mathtt{VARIABLES}, \mathtt{var})$$
where
$$\mathtt{maxv} = \mathtt{maxval}(\mathtt{VARIABLES.var})$$
$$\mathtt{minv} = \mathtt{minval}(\mathtt{VARIABLES.var})$$
$$\mathtt{sv} = |\mathtt{VARIABLES}|$$
$$\mathtt{rv} = \mathtt{range}(\mathtt{VARIABLES.var})$$

**Purpose**

VALUE is the maximum of all minimum values in each occurrence of the INCREAS-ING_SEQUENCE pattern in the time-series given by the VARIABLES collection. If the pattern does not occur, VALUE takes the default value $-\infty$.

An occurrence of the pattern INCREASING_SEQUENCE is the *maximal* subsequence which matches the regular expression $< (< | =)^* < |<$.

Assume that the occurrence of the pattern INCREASING_SEQUENCE starts at position $i$ and ends at position $j$. The feature MIN computes the minimum of the values from index $i$ to index $j + 1$.

**Example**

$$(3, \langle 4, 3, 5, 5, 2, 1, 1, 3, 3, 4, 6, 6, 3, 1, 3, 3 \rangle)$$

Figure 3.541 provides an example where the MAX_MIN_INCREASING_SEQUENCE $(3, [4, 3, 5, 5, 2, 1, 1, 3, 3, 4, 6, 6, 3, 1, 3, 3])$ constraint holds.

| **Typical** | $\|\mathtt{VARIABLES}\| > 1$<br>$\mathtt{range}(\mathtt{VARIABLES.var}) > 1$ |

| **Arg. properties** | Functional dependency: VALUE determined by VARIABLES. |

Figure 3.541: Illustrating the MAX_MIN_INCREASING_SEQUENCE constraint of the **Example** slot

Figures 3.542 and 3.543 respectively depict the automaton associated with the constraint MAX_MIN_INCREASING_SEQUENCE and its simplified form.

Figure 3.542: Automaton for the MAX_MIN_INCREASING_SEQUENCE constraint obtained by applying decoration Table 2.35 to the seed transducer of the INCREASING_SEQUENCE pattern where default is $-\infty$

Figure 3.543: Automaton for the MAX_MIN_INCREASING_SEQUENCE constraint obtained by applying decoration Table 2.37 to the seed transducer of the INCREASING_SEQUENCE pattern where default is $-\infty$; $R_i - R_{i-1} \geq 0$ is a linear invariant.

|   | $s$ | $t$ |
|---|---|---|
| $s$ | $\max(\overrightarrow{C}, \overleftarrow{C})$ | $\max(\overrightarrow{C}, \overleftarrow{C})$ |
| $t$ | $\max(\overrightarrow{C}, \overleftarrow{C})$ | $\min(\overrightarrow{C}, \overleftarrow{C}, \overrightarrow{D}, \overleftarrow{D}, \text{VAR}_i)$ |

Table 3.22: Glue matrix for the MAX_MIN_INCREASING_SEQUENCE constraint defined as the composition of the INCREASING_SEQUENCE pattern, the feature MIN, and the aggregator max; cells of the glue matrix are coloured with the colour of the constituent to which they are related.

## 3.259   MAX_MIN_INFLEXION

$< (< | =)^* > | > (> | =)^* <$

**Origin**            Based on the INFLEXION pattern.

**Constraint**        MAX_MIN_INFLEXION(VALUE, VARIABLES)

**Arguments**         VALUE      : dvar
                      VARIABLES  : collection(var−dvar)

**Restrictions**      $\text{sv} \leq 2 \vee \text{rv} \leq 1 \Rightarrow \text{VALUE} = -\infty$
                      $\text{VALUE} = -\infty \vee \text{VALUE} \geq \text{minv}$
                      $\text{VALUE} \leq \text{maxv}①$
                      required(VARIABLES, var)
                      where
                      maxv =maxval(VARIABLES.var)
                      minv =minval(VARIABLES.var)
                      sv = |VARIABLES|
                      rv =range(VARIABLES.var)

                      VALUE ≤ **2**

**Purpose**           VALUE is the maximum of all minimum values in each occurrence of the INFLEXION
                      pattern in the time-series given by the VARIABLES collection. If the pattern does not
                      occur, VALUE takes the default value $-\infty$.
                      An occurrence of the pattern INFLEXION is the *maximal* subsequence which matches the
                      regular expression $< (< | =)^* > | > (> | =)^* <$.
                      Assume that the occurrence of the pattern INFLEXION starts at position $i$ and ends at
                      position $j$. The feature MIN computes the minimum of the values from index $i + 1$ to
                      index $j$.

**Example**           $(5, \langle 1, 2, 6, 6, 4, 4, 3, 5, 2, 5, 1, 5, 3, 3, 4, 4 \rangle)$

                      Figure   3.544   provides   an   example   where   the   MAX_MIN_INFLEXION
                      $(5, [1, 2, 6, 6, 4, 4, 3, 5, 2, 5, 1, 5, 3, 3, 4, 4])$ constraint holds.

**Typical**           $|\text{VARIABLES}| > 2$
                      range(VARIABLES.var) > 1

**Symmetry**          Items of VARIABLES can be reversed.

**Arg. properties**   Functional dependency: VALUE determined by VARIABLES.

Figure 3.544: Illustrating the MAX_MIN_INFLEXION constraint of the **Example** slot

**Automaton**    Figures 3.545 and 3.546 respectively depict the automaton associated with the constraint MAX_MIN_INFLEXION and its simplified form.

Figure 3.545: Automaton for the MAX_MIN_INFLEXION constraint obtained by applying decoration Table 2.35 to the seed transducer of the INFLEXION pattern where default is $-\infty$ (transition $r \rightarrow t$ has the same accumulators updates as transition $t \rightarrow r$)

Figure 3.546: Automaton for the MAX_MIN_INFLEXION constraint obtained by applying decoration Table 2.25 to the seed transducer of the INFLEXION pattern where default is $-\infty$ (transition $r \rightarrow t$ has the same accumulators updates as transition $t \rightarrow r$); $R_i - R_{i-1} \geq 0$ is a linear invariant.

## 3.260 MAX_MIN_STRICTLY_DECREASING_SEQUENCE

▶ ▷ ◁ ◀        **DESCRIPTION**              **AUTOMATON**

| | |
|---|---|
| **Origin** | Based on the STRICTLY_DECREASING_SEQUENCE pattern. |
| **Constraint** | MAX_MIN_STRICTLY_DECREASING_SEQUENCE(VALUE, VARIABLES) |
| **Arguments** | VALUE      : dvar<br>VARIABLES : collection(var−dvar) |
| **Restrictions** | $\mathtt{sv} \leq 1 \vee \mathtt{rv} \leq 1 \Rightarrow \mathtt{VALUE} = -\infty$<br>$\mathtt{VALUE} = -\infty \vee \mathtt{VALUE} \geq \mathtt{minv}$<br>$\mathtt{VALUE} \leq \mathtt{maxv} - 1 ①$<br>required(VARIABLES, var)<br>where<br>maxv = maxval(VARIABLES.var)<br>minv = minval(VARIABLES.var)<br>$\mathtt{sv} = \lvert\mathtt{VARIABLES}\rvert$<br>rv = range(VARIABLES.var) |
| **Purpose** | VALUE is the maximum of all minimum values in each occurrence of the STRICTLY_DECREASING_SEQUENCE pattern in the time-series given by the VARIABLES collection. If the pattern does not occur, VALUE takes the default value $-\infty$.<br>An occurrence of the pattern STRICTLY_DECREASING_SEQUENCE is the *maximal* subsequence which matches the regular expression $>^{+}$.<br>Assume that the occurrence of the pattern STRICTLY_DECREASING_SEQUENCE starts at position $i$ and ends at position $j$. The feature MIN computes the minimum of the values from index $i$ to index $j + 1$. |
| **Example** | $(3, \langle 4, 4, 6, 4, 1, 1, 3, 4, 4, 6, 6, 5, 2, 2, 4, 3 \rangle)$ |

Figure 3.547 provides an example where the MAX_MIN_STRICTLY_DECREASING_SEQUENCE $(3, [4, 4, 6, 4, 1, 1, 3, 4, 4, 6, 6, 5, 2, 2, 4, 3])$ constraint holds.

| | |
|---|---|
| **Typical** | $\lvert\mathtt{VARIABLES}\rvert > 1$<br>range(VARIABLES.var) > 1 |
| **Arg. properties** | Functional dependency: VALUE determined by VARIABLES. |

Figure 3.547: Illustrating the MAX_MIN_STRICTLY_DECREASING_SEQUENCE constraint of the **Example** slot

**Automaton**     Figures 3.548 and 3.549 respectively depict the automaton associated with the constraint
MAX_MIN_STRICTLY_DECREASING_SEQUENCE and its simplified form.

Figure 3.548: Automaton for the MAX_MIN_STRICTLY_DECREASING_SEQUENCE constraint obtained by applying decoration Table 2.35 to the seed transducer of the STRICTLY_DECREASING_SEQUENCE pattern where default is $-\infty$

Figure 3.549: Automaton for the MAX_MIN_STRICTLY_DECREASING_SEQUENCE constraint obtained by applying decoration Table 2.24 to the seed transducer of the STRICTLY_DECREASING_SEQUENCE pattern where default is $-\infty$; $R_i - R_{i-1} \geq 0$ is a linear invariant.

|   | $s$ | $r$ |
|---|---|---|
| $s$ | $\max(\overrightarrow{C}, \overleftarrow{C})$ | $\max(\overrightarrow{C}, \overleftarrow{C})$ |
| $r$ | $\max(\overrightarrow{C}, \overleftarrow{C})$ | $\min(\overrightarrow{C}, \overleftarrow{C}, \overrightarrow{D}, \overleftarrow{D}, \texttt{VAR}_i)$ |

Table 3.23: Glue matrix for the MAX_MIN_STRICTLY_DECREASING_SEQUENCE constraint defined as the composition of the STRICTLY_DECREASING_SEQUENCE pattern, the feature MIN, and the aggregator max; cells of the glue matrix are coloured with the colour of the constituent to which they are related.

# 3.261 MAX_MIN_STRICTLY_INCREASING_SEQUENCE

▶ ▷ ◁ ◀     DESCRIPTION          AUTOMATON

| | |
|---|---|
| **Origin** | Based on the STRICTLY_INCREASING_SEQUENCE pattern. |
| **Constraint** | MAX_MIN_STRICTLY_INCREASING_SEQUENCE(VALUE, VARIABLES) |
| **Arguments** | VALUE     : dvar<br>VARIABLES : collection(var−dvar) |

**Restrictions**

$\mathtt{sv} \leq 1 \vee \mathtt{rv} \leq 1 \Rightarrow \mathtt{VALUE} = -\infty$
$\mathtt{VALUE} = -\infty \vee \mathtt{VALUE} \geq \mathtt{minv}$
$\mathtt{VALUE} \leq \mathtt{maxv} - 1 \text{①}$
required(VARIABLES, var)
where
$\mathtt{maxv} = \mathtt{maxval}(\mathtt{VARIABLES.var})$
$\mathtt{minv} = \mathtt{minval}(\mathtt{VARIABLES.var})$
$\mathtt{sv} = |\mathtt{VARIABLES}|$
$\mathtt{rv} = \mathtt{range}(\mathtt{VARIABLES.var})$

**Purpose**

VALUE is the maximum of all minimum values in each occurrence of the STRICTLY_INCREASING_SEQUENCE pattern in the time-series given by the VARIABLES collection. If the pattern does not occur, VALUE takes the default value $-\infty$.

An occurrence of the pattern STRICTLY_INCREASING_SEQUENCE is the *maximal* sub-sequence which matches the regular expression $<^+$.

Assume that the occurrence of the pattern STRICTLY_INCREASING_SEQUENCE starts at position $i$ and ends at position $j$. The feature MIN computes the minimum of the values from index $i$ to index $j + 1$.

**Example**

$(3, \langle 4, 3, 5, 5, 2, 1, 1, 2, 3, 4, 6, 6, 3, 1, 2, 3 \rangle)$

Figure 3.550 provides an example where the MAX_MIN_STRICTLY_INCREASING_SEQUENCE $(3, [4, 3, 5, 5, 2, 1, 1, 2, 3, 4, 6, 6, 3, 1, 2, 3])$ constraint holds.

| | |
|---|---|
| **Typical** | $\|\mathtt{VARIABLES}\| > 1$<br>range(VARIABLES.var) > 1 |
| **Arg. properties** | Functional dependency: VALUE determined by VARIABLES. |

Figure 3.550: Illustrating the MAX_MIN_STRICTLY_INCREASING_SEQUENCE constraint of the **Example** slot

 Figures 3.551 and 3.552 respectively depict the automaton associated with the constraint MAX_MIN_STRICTLY_INCREASING_SEQUENCE and its simplified form.

Figure 3.551: Automaton for the MAX_MIN_STRICTLY_INCREASING_SEQUENCE constraint obtained by applying decoration Table 2.35 to the seed transducer of the STRICTLY_INCREASING_SEQUENCE pattern where default is $-\infty$

Figure 3.552: Automaton for the MAX_MIN_STRICTLY_INCREASING_SEQUENCE constraint obtained by applying decoration Table 2.37 to the seed transducer of the STRICTLY_INCREASING_SEQUENCE pattern where default is $-\infty$; $R_i - R_{i-1} \geq 0$ is a linear invariant.

|   | $s$ | $r$ |
|---|---|---|
| $s$ | $\max(\overrightarrow{C}, \overleftarrow{C})$ | $\max(\overrightarrow{C}, \overleftarrow{C})$ |
| $r$ | $\max(\overrightarrow{C}, \overleftarrow{C})$ | $\min(\overrightarrow{C}, \overleftarrow{C}, \overrightarrow{D}, \overleftarrow{D}, \mathtt{VAR}_i)$ |

Table 3.24: Glue matrix for the MAX_MIN_STRICTLY_INCREASING_SEQUENCE constraint defined as the composition of the STRICTLY_INCREASING_SEQUENCE pattern , the feature MIN , and the aggregator max ; cells of the glue matrix are coloured with the colour of the constituent to which they are related.

## 3.262 MAX_MIN_VALLEY

► ▷ ◁ ◄     **DESCRIPTION**          **AUTOMATON**

| **Origin** | Based on the VALLEY pattern. |
| **Constraint** | MAX_MIN_VALLEY(VALUE, VARIABLES) |
| **Arguments** | VALUE      : dvar |
| | VARIABLES  : collection(var−dvar) |

**Restrictions**

$\mathtt{sv} \leq 2 \vee \mathtt{rv} \leq 1 \Rightarrow \mathtt{VALUE} = -\infty$
$\mathtt{VALUE} = -\infty \vee \mathtt{VALUE} \geq \mathtt{minv}$
$\mathtt{VALUE} \leq \mathtt{maxv} - 1$①
required(VARIABLES, var)
where
$\mathtt{maxv} =$ maxval(VARIABLES.var)
$\mathtt{minv} =$ minval(VARIABLES.var)
$\mathtt{sv} = |\mathtt{VARIABLES}|$
$\mathtt{rv} =$ range(VARIABLES.var)

**Purpose**

VALUE is the maximum of all minimum values in each occurrence of the VALLEY pattern in the time-series given by the VARIABLES collection. If the pattern does not occur, VALUE takes the default value $-\infty$.

An occurrence of the pattern VALLEY is the *maximal* subsequence which matches the regular expression $> (= | >)^* (< | =)^* <$.

Assume that the occurrence of the pattern VALLEY starts at position $i$ and ends at position $j$. The feature MIN computes the minimum of the values from index $i + 1$ to index $j$.

**Example**

$(5, \langle 1, 3, 7, 4, 3, 6, 6, 5, 3, 3, 2, 6, 5, 5, 5, 7 \rangle)$

Figure 3.553 provides an example where the MAX_MIN_VALLEY $(5, [1, 3, 7, 4, 3, 6, 6, 5, 3, 3, 2, 6, 5, 5, 5, 7])$ constraint holds.

**Typical**

$|\mathtt{VARIABLES}| > 2$
range(VARIABLES.var) $> 1$

**Symmetry**

Items of VARIABLES can be reversed.

**Arg. properties**

Functional dependency: VALUE determined by VARIABLES.

Figure 3.553: Illustrating the max_min_valley constraint of the **Example** slot

**Automaton**    Figures 3.554 and 3.555 respectively depict the automaton associated with the constraint MAX_MIN_VALLEY and its simplified form.

Figure 3.554: Automaton for the MAX_MIN_VALLEY constraint obtained by applying decoration Table 2.35 to the seed transducer of the VALLEY pattern where default is $-\infty$

Figure 3.555: Automaton for the MAX_MIN_VALLEY constraint obtained by applying decoration Table 2.37 to the seed transducer of the VALLEY pattern where default is $-\infty$; $R_i - R_{i-1} \geq 0$ is a linear invariant.

|   | $s$ | $r$ | $t$ |
|---|---|---|---|
| $s$ | $\max(\overrightarrow{C}, \overleftarrow{C})$ | $\max(\overrightarrow{C}, \overleftarrow{C})$ | $\max(\overrightarrow{C}, \overleftarrow{C})$ |
| $r$ | $\max(\overrightarrow{C}, \overleftarrow{C})$ | $\min(\overrightarrow{D}, \overleftarrow{D}, \mathtt{VAR}_i)$ | $\min(\overleftarrow{C}, \overrightarrow{D}, \overleftarrow{D}, \mathtt{VAR}_i)$ |
| $t$ | $\max(\overrightarrow{C}, \overleftarrow{C})$ | $\min(\overrightarrow{C}, \overrightarrow{D}, \overleftarrow{D}, \mathtt{VAR}_i)$ | $\max(\overrightarrow{C}, \overleftarrow{C})$ |

Table 3.25: Glue matrix for the MAX_MIN_VALLEY constraint defined as the composition of the VALLEY pattern , the feature MIN , and the aggregator max ; cells of the glue matrix are coloured with the colour of the constituent to which they are related.

## 3.263 MAX_MIN_ZIGZAG

  **DESCRIPTION**    **AUTOMATON**

| | | |
|---|---|---|
| **Origin** | Based on the ZIGZAG pattern. | $(<>)^+(<\|<>)\|(><)^+(>\|><)$ |

**Constraint**    MAX_MIN_ZIGZAG(VALUE, VARIABLES)

**Arguments**
```
VALUE      : dvar
VARIABLES  : collection(var−dvar)
```

**Restrictions**

$\mathtt{sv} \leq 3 \vee \mathtt{rv} \leq 1 \Rightarrow \mathtt{VALUE} = -\infty$

$\mathtt{VALUE} = -\infty \vee \mathtt{VALUE} \geq \mathtt{minv}$

$\mathtt{VALUE} \leq \mathtt{maxv} - 1$①

required(VARIABLES, var)

where

maxv =maxval(VARIABLES.var)

minv =minval(VARIABLES.var)

sv = |VARIABLES|

rv =range(VARIABLES.var)

**Purpose**

VALUE is the maximum of all minimum values in each occurrence of the ZIGZAG pattern in the time-series given by the VARIABLES collection. If the pattern does not occur, VALUE takes the default value $-\infty$.

An occurrence of the pattern ZIGZAG is the *maximal* subsequence which matches the regular expression $(<>)^+(<\|<>)\|(><)^+(>\|><)$.

Assume that the occurrence of the pattern ZIGZAG starts at position $i$ and ends at position $j$. The feature MIN computes the minimum of the values from index $i + 1$ to index $j$.

**Example**    $(1, \langle 4, 1, 3, 1, 4, 6, 1, 5, 5, 2, 7, 2, 3, 1, 6, 1 \rangle)$

Figure 3.556 provides an example where the MAX_MIN_ZIGZAG $(1, [4, 1, 3, 1, 4, 6, 1, 5, 5, 2, 7, 2, 3, 1, 6, 1])$ constraint holds.

**Typical**

$|\mathtt{VARIABLES}| > 3$

range(VARIABLES.var) > 1

**Symmetry**    Items of VARIABLES can be reversed.

**Arg. properties**    Functional dependency: VALUE determined by VARIABLES.

Figure 3.556: Illustrating the MAX_MIN_ZIGZAG constraint of the **Example** slot

**Automaton** Figures 3.557 and 3.558 respectively depict the automaton associated with the constraint MAX_MIN_ZIGZAG and its simplified form.

Figure 3.557: Automaton for the MAX_MIN_ZIGZAG constraint obtained by applying decoration Table 2.35 to the seed transducer of the ZIGZAG pattern where default is $-\infty$; (1) missing transitions from $a, b, c, d, e, f$ to $s$ are labelled by $=$; (2) on transitions from $b, c, e, f$ to $s$ the accumulator $D$ is reset to its initial value; (3) on transitions from $c, f$ to $s$ the accumulator $R$ is updated wrt $C$ and the accumulator $C$ is reset to its initial value

Figure 3.558: Automaton for the MAX_MIN_ZIGZAG constraint obtained by applying decoration Table 2.23 to the seed transducer of the ZIGZAG pattern where default is $-\infty$; missing transitions from $a, b, c, d, e, f$ to $s$ are labelled by $=$; (2) on transitions from $b, c, e, f$ to $s$ the accumulator $D$ is reset to its initial value; (3) on transitions from $c, f$ to $s$ the accumulator $R$ is updated wrt $C$ and the accumulator $C$ is reset to its initial value.; $R_i - R_{i-1} \geq 0$ is a linear invariant.

|   | $s$ | $a$ | $b$ | $c$ | $d$ | $e$ | $f$ |
|---|-----|-----|-----|-----|-----|-----|-----|
| $s$ | $\max(\overrightarrow{C}, \overleftarrow{C})$ | $\max(\overrightarrow{C}, \overleftarrow{C})$ | $\max(\overrightarrow{C}, \overleftarrow{C})$ | $\max(\overrightarrow{C}, \overleftarrow{C})$ | $\max(\overrightarrow{C}, \overleftarrow{C})$ | $\max(\overrightarrow{C}, \overleftarrow{C})$ | $\max(\overrightarrow{C}, \overleftarrow{C})$ |
| $a$ | $\max(\overrightarrow{C}, \overleftarrow{C})$ | $\max(\overrightarrow{C}, \overleftarrow{C})$ | $\max(\overrightarrow{C}, \overleftarrow{C})$ | $\min(\overrightarrow{C}, \overrightarrow{D}, \overleftarrow{D}, \text{VAR}_i)$ | $\max(\overrightarrow{C}, \overleftarrow{C})$ | $\min(\overrightarrow{D}, \overleftarrow{D}, \text{VAR}_i)$ | $\max(\overrightarrow{C}, \overleftarrow{C})$ |
| $b$ | $\max(\overrightarrow{C}, \overleftarrow{C})$ | $\min(\overrightarrow{C}, \overrightarrow{D}, \overleftarrow{D}, \text{VAR}_i)$ | $\min(\overrightarrow{D}, \overleftarrow{D}, \text{VAR}_i)$ | $\max(\overrightarrow{C}, \overleftarrow{C})$ | $\min(\overrightarrow{D}, \overleftarrow{D}, \text{VAR}_i)$ | $\max(\overrightarrow{C}, \overrightarrow{D}, \overleftarrow{D}, \text{VAR}_i)$ | $\min(\overleftarrow{C}, \overleftarrow{D}, \overrightarrow{D}, \text{VAR}_i)$ |
| $c$ | $\max(\overrightarrow{C}, \overleftarrow{C})$ | $\max(\overrightarrow{C}, \overleftarrow{C})$ | $\max(\overrightarrow{C}, \overleftarrow{C})$ | $\min(\overrightarrow{C}, \overleftarrow{C}, \overrightarrow{D}, \overleftarrow{D}, \text{VAR}_i)$ | $\max(\overrightarrow{C}, \overleftarrow{C})$ | $\min(\overrightarrow{C}, \overrightarrow{D}, \overleftarrow{D}, \text{VAR}_i)$ | $\max(\overrightarrow{C}, \overleftarrow{C})$ |
| $d$ | $\max(\overrightarrow{C}, \overleftarrow{C})$ | $\max(\overrightarrow{C}, \overleftarrow{C})$ | $\min(\overrightarrow{D}, \overleftarrow{D}, \text{VAR}_i)$ | $\max(\overrightarrow{C}, \overleftarrow{C})$ | $\max(\overrightarrow{C}, \overleftarrow{C})$ | $\max(\overrightarrow{C}, \overleftarrow{C})$ | $\min(\overleftarrow{C}, \overleftarrow{D}, \overrightarrow{D}, \text{VAR}_i)$ |
| $e$ | $\max(\overrightarrow{C}, \overleftarrow{C})$ | $\min(\overrightarrow{D}, \overleftarrow{D}, \text{VAR}_i)$ | $\max(\overrightarrow{C}, \overleftarrow{C})$ | $\min(\overrightarrow{C}, \overrightarrow{D}, \overleftarrow{D}, \text{VAR}_i)$ | $\max(\overrightarrow{C}, \overleftarrow{C})$ | $\min(\overrightarrow{D}, \overleftarrow{D}, \text{VAR}_i)$ | $\max(\overrightarrow{C}, \overleftarrow{C})$ |
| $f$ | $\max(\overrightarrow{C}, \overleftarrow{C})$ | $\max(\overrightarrow{C}, \overleftarrow{C})$ | $\min(\overrightarrow{C}, \overrightarrow{D}, \overleftarrow{D}, \text{VAR}_i)$ | $\max(\overrightarrow{C}, \overleftarrow{C})$ | $\min(\overrightarrow{C}, \overrightarrow{D}, \overleftarrow{D}, \text{VAR}_i)$ | $\max(\overrightarrow{C}, \overleftarrow{C})$ | $\min(\overrightarrow{C}, \overleftarrow{C}, \overrightarrow{D}, \overleftarrow{D}, \text{VAR}_i)$ |

Table 3.26: Glue matrix for the MAX_MIN_ZIGZAG constraint defined as the composition of the ZIGZAG pattern, the feature MIN, and the aggregator max; cells of the glue matrix are coloured with the colour of the constituent to which they are related.

## 3.264   MAX_RANGE_DECREASING

DESCRIPTION          AUTOMATON

| | |
|---|---|
| **Origin** | Based on the DECREASING pattern. |
| **Constraint** | MAX_RANGE_DECREASING(VALUE, VARIABLES) |
| **Arguments** | VALUE       : dvar<br>VARIABLES : collection(var−dvar) |
| **Restrictions** | $\mathtt{sv} \leq 1 \vee \mathtt{rv} \leq 1 \Rightarrow \mathtt{VALUE} = 0$<br>$\mathtt{VALUE} \geq 0$<br>$\mathtt{VALUE} \leq \mathtt{rv} - 1$ ①<br>required(VARIABLES, var)<br>where<br>$\mathtt{rv} =$ range(VARIABLES.var)<br>$\mathtt{sv} = |$VARIABLES$|$ |
| **Purpose** | VALUE is the maximum value of the differences between the largest and smallest value in each occurrence of the DECREASING pattern in the time-series given by the VARIABLES collection. If the pattern does not occur, VALUE takes the default value 0.<br>An occurrence of the pattern DECREASING is the subsequence which matches the regular expression >.<br>Assume that the occurrence of the pattern DECREASING starts at position $i$ and ends at position $j$. The feature RANGE computes the range of the values from index $i$ to index $j + 1$. |
| **Example** | $(2, \langle 3, 4, 2, 2, 5, 6, 6, 4, 4, 3, 1, 1, 4, 6, 4, 4 \rangle)$ |

Figure 3.559 provides an example where the MAX_RANGE_DECREASING $(2, [3, 4, 2, 2, 5, 6, 6, 4, 4, 3, 1, 1, 4, 6, 4, 4])$ constraint holds.

| | |
|---|---|
| **Typical** | $|$VARIABLES$| > 1$<br>range(VARIABLES.var) > 1 |
| **Symmetry** | One and the same constant can be added to the var attribute of all items of VARIABLES. |
| **Arg. properties** | Functional dependency: VALUE determined by VARIABLES. |

Figure 3.559: Illustrating the MAX_RANGE_DECREASING constraint of the **Example** slot

**Automaton**     Figures 3.560 and 3.561 respectively depict the automaton associated with the constraint MAX_RANGE_DECREASING and its simplified form.

$$\left\{ \begin{array}{l} C \leftarrow \texttt{default} \\ H \leftarrow \texttt{VAR}_1 \\ R \leftarrow \texttt{default} \end{array} \right\}$$

$$\left\{ \begin{array}{l} H \leftarrow \texttt{VAR}_{i+1} \\ R \leftarrow \max(R, |H - \texttt{VAR}_{i+1}|) \end{array} \right\} \qquad \overset{>}{\underset{\geq}{\lessgtr}} s \qquad \{H \overset{\leq}{\leftarrow} \texttt{VAR}_{i+1}\}$$

$$\boxed{\max(R, C)}$$

Figure 3.560: Automaton for the MAX_RANGE_DECREASING constraint obtained by applying decoration Table 2.46 to the seed transducer of the DECREASING pattern where default is 0

$$\{R \leftarrow \texttt{default}\}$$

$$\{R \leftarrow \max(R, \max(0, \texttt{VAR}_i - \texttt{VAR}_{i+1}))\} \qquad \overset{>}{\underset{\geq}{\lessgtr}} s \qquad \leq$$

$$\boxed{R}$$

Figure 3.561: Automaton for the MAX_RANGE_DECREASING constraint obtained by applying decoration Table 2.44 to the seed transducer of the DECREASING pattern where default is 0; $R_i - R_{i-1} \geq 0$ and $R_i + \texttt{VAR}_{i-1} - \texttt{VAR}_{i-2} \geq 0$ are linear invariants.

## 3.265  MAX_RANGE_DECREASING_SEQUENCE

► ▷ ◁ ◄        DESCRIPTION            AUTOMATON

**Origin**          Based on the DECREASING_SEQUENCE pattern.

**Constraint**      MAX_RANGE_DECREASING_SEQUENCE(VALUE, VARIABLES)

**Arguments**       VALUE      : dvar
                    VARIABLES  : collection(var−dvar)

**Restrictions**    $\mathtt{sv} \leq 1 \vee \mathtt{rv} \leq 1 \Rightarrow \mathtt{VALUE} = 0$
                    $\mathtt{VALUE} \geq 0$
                    $\mathtt{VALUE} \leq \mathtt{rv} - 1$ ①
                    required(VARIABLES, var)
                    where
                    $\mathtt{rv} =$ range(VARIABLES.var)
                    $\mathtt{sv} = |\mathtt{VARIABLES}|$

**Purpose**         VALUE is the maximum value of the differences between the largest and smallest value
                    in each occurrence of the DECREASING_SEQUENCE pattern in the time-series given by
                    the VARIABLES collection. If the pattern does not occur, VALUE takes the default value
                    0.
                    An occurrence of the pattern DECREASING_SEQUENCE is the *maximal* subsequence
                    which matches the regular expression $> (> | =)^* > | >$.
                    Assume that the occurrence of the pattern DECREASING_SEQUENCE starts at position $i$
                    and ends at position $j$. The feature RANGE computes the range of the values from index
                    $i$ to index $j + 1$.

**Example**         $(5, \langle 3, 4, 2, 2, 5, 6, 6, 4, 4, 3, 1, 1, 4, 6, 4, 4 \rangle)$

                    Figure 3.562 provides an example where the MAX_RANGE_DECREASING_SEQUENCE
                    $(5, [3, 4, 2, 2, 5, 6, 6, 4, 4, 3, 1, 1, 4, 6, 4, 4])$ constraint holds.

**Typical**         $|\mathtt{VARIABLES}| > 1$
                    range(VARIABLES.var) $> 1$

**Symmetry**        One and the same constant can be added to the var attribute of all items of VARIABLES.

**Arg. properties** Functional dependency: VALUE determined by VARIABLES.

Figure 3.562: Illustrating the MAX_RANGE_DECREASING_SEQUENCE constraint of the **Example** slot

 Figures 3.563 and 3.564 respectively depict the automaton associated with the constraint MAX_RANGE_DECREASING_SEQUENCE and its simplified form.

Figure 3.563: Automaton for the MAX_RANGE_DECREASING_SEQUENCE constraint obtained by applying decoration Table 2.46 to the seed transducer of the DECREAS-ING_SEQUENCE pattern where default is $0$

Figure 3.564: Automaton for the MAX_RANGE_DECREASING_SEQUENCE constraint obtained by applying decoration Table 2.42 to the seed transducer of the DECREAS-ING_SEQUENCE pattern where default is $0$; $R_i - R_{i-1} \geq 0$ is a linear invariant.

## 3.266   MAX_RANGE_INCREASING

► ▷ ◁ ◄        **DESCRIPTION**        **AUTOMATON**

| | |
|---|---|
| **Origin** | Based on the INCREASING pattern. |
| **Constraint** | MAX_RANGE_INCREASING(VALUE, VARIABLES) |
| **Arguments** | VALUE      : dvar |
| | VARIABLES  : collection(var−dvar) |

**Restrictions**

$\text{sv} \leq 1 \vee \text{rv} \leq 1 \Rightarrow \text{VALUE} = 0$
$\text{VALUE} \geq 0$
$\text{VALUE} \leq \text{rv} - 1$ ⓘ
required(VARIABLES, var)
where
$\text{rv} =$ range(VARIABLES.var)
$\text{sv} = |\text{VARIABLES}|$

**Purpose**

VALUE is the maximum value of the differences between the largest and smallest value in each occurrence of the INCREASING pattern in the time-series given by the VARIABLES collection. If the pattern does not occur, VALUE takes the default value 0.
An occurrence of the pattern INCREASING is the subsequence which matches the regular expression <.
Assume that the occurrence of the pattern INCREASING starts at position $i$ and ends at position $j$. The feature RANGE computes the range of the values from index $i$ to index $j + 1$.

**Example**

$(2, \langle 4, 3, 5, 5, 2, 1, 1, 3, 3, 4, 6, 6, 3, 1, 3, 3 \rangle)$

Figure 3.565 provides an example where the MAX_RANGE_INCREASING $(2, [4, 3, 5, 5, 2, 1, 1, 3, 3, 4, 6, 6, 3, 1, 3, 3])$ constraint holds.

| | |
|---|---|
| **Typical** | $|\text{VARIABLES}| > 1$ |
| | range(VARIABLES.var) $> 1$ |
| **Symmetry** | One and the same constant can be added to the var attribute of all items of VARIABLES. |
| **Arg. properties** | Functional dependency: VALUE determined by VARIABLES. |

Figure 3.565: Illustrating the MAX_RANGE_INCREASING constraint of the **Example** slot

**Automaton**     Figures 3.566 and 3.567 respectively depict the automaton associated with the constraint MAX_RANGE_INCREASING and its simplified form.

Figure 3.566: Automaton for the MAX_RANGE_INCREASING constraint obtained by applying decoration Table 2.46 to the seed transducer of the INCREASING pattern where default is 0

Figure 3.567: Automaton for the MAX_RANGE_INCREASING constraint obtained by applying decoration Table 2.45 to the seed transducer of the INCREASING pattern where default is 0; $R_i - R_{i-1} \geq 0$ and $R_i - \text{VAR}_{i-1} + \text{VAR}_{i-2} \geq 0$ are linear invariants.

## 3.267  MAX_RANGE_INCREASING_SEQUENCE

▶ ▷ ◁ ◀          **DESCRIPTION**          **AUTOMATON**

$< (< | =)^* < |$
$<$

| **Origin** | Based on the INCREASING_SEQUENCE pattern. |

| **Constraint** | MAX_RANGE_INCREASING_SEQUENCE(VALUE, VARIABLES) |

| **Arguments** | VALUE      : dvar |
|  | VARIABLES  : collection(var−dvar) |

**Restrictions**

$\mathtt{sv} \leq 1 \vee \mathtt{rv} \leq 1 \Rightarrow \mathtt{VALUE} = 0$
$\mathtt{VALUE} \geq 0$
$\mathtt{VALUE} \leq \mathtt{rv} - 1$ ①
required(VARIABLES, var)
where
$\mathtt{rv} =$ range(VARIABLES.var)
$\mathtt{sv} = |\mathtt{VARIABLES}|$

VALUE ≤ **3**

**Purpose**

VALUE is the maximum value of the differences between the largest and smallest value in each occurrence of the INCREASING_SEQUENCE pattern in the time-series given by the VARIABLES collection. If the pattern does not occur, VALUE takes the default value 0.

An occurrence of the pattern INCREASING_SEQUENCE is the *maximal* subsequence which matches the regular expression $< (< | =)^* < | <$.

Assume that the occurrence of the pattern INCREASING_SEQUENCE starts at position $i$ and ends at position $j$. The feature RANGE computes the range of the values from index $i$ to index $j + 1$.

**Example**

$(5, \langle 4, 3, 5, 5, 2, 1, 1, 3, 3, 4, 6, 6, 3, 1, 3, 3 \rangle)$

Figure 3.568 provides an example where the MAX_RANGE_INCREASING_SEQUENCE $(5, [4, 3, 5, 5, 2, 1, 1, 3, 3, 4, 6, 6, 3, 1, 3, 3])$ constraint holds.

**Typical**

$|\mathtt{VARIABLES}| > 1$
range(VARIABLES.var) $> 1$

**Symmetry**

One and the same constant can be added to the var attribute of all items of VARIABLES.

**Arg. properties**

Functional dependency: VALUE determined by VARIABLES.

Figure 3.568: Illustrating the MAX_RANGE_INCREASING_SEQUENCE constraint of the **Example** slot

 Figures 3.569 and 3.570 respectively depict the automaton associated with the constraint MAX_RANGE_INCREASING_SEQUENCE and its simplified form.

Figure 3.569: Automaton for the MAX_RANGE_INCREASING_SEQUENCE constraint obtained by applying decoration Table 2.46 to the seed transducer of the INCREASING_SEQUENCE pattern where default is $0$

Figure 3.570: Automaton for the MAX_RANGE_INCREASING_SEQUENCE constraint obtained by applying decoration Table 2.43 to the seed transducer of the INCREASING_SEQUENCE pattern where default is $0$; $R_i - R_{i-1} \geq 0$ is a linear invariant.

# 3.268   MAX_RANGE_STRICTLY_DECREASING_SEQUENCE

▶ ▷ ◁ ◀          DESCRIPTION          AUTOMATON

| | |
|---|---|
| **Origin** | Based on the STRICTLY_DECREASING_SEQUENCE pattern. |
| **Constraint** | MAX_RANGE_STRICTLY_DECREASING_SEQUENCE(VALUE, VARIABLES) |
| **Arguments** | VALUE     : dvar<br>VARIABLES : collection(var−dvar) |
| **Restrictions** | $\text{sv} \leq 1 \vee \text{rv} \leq 1 \Rightarrow \text{VALUE} = 0$<br>$\text{VALUE} \geq 0$<br>$\text{VALUE} \leq \text{rv} - 1$ ⓘ<br>required(VARIABLES, var)<br>where<br>rv = range(VARIABLES.var)<br>sv = \|VARIABLES\| |
| **Purpose** | VALUE is the maximum value of the differences between the largest and smallest value in each occurrence of the STRICTLY_DECREASING_SEQUENCE pattern in the time-series given by the VARIABLES collection. If the pattern does not occur, VALUE takes the default value 0.<br>An occurrence of the pattern STRICTLY_DECREASING_SEQUENCE is the *maximal* sub-sequence which matches the regular expression $>^+$.<br>Assume that the occurrence of the pattern STRICTLY_DECREASING_SEQUENCE starts at position $i$ and ends at position $j$. The feature RANGE computes the range of the values from index $i$ to index $j + 1$. |
| **Example** | $(5, \langle 4, 4, 6, 4, 1, 1, 3, 4, 4, 6, 6, 5, 2, 2, 4, 3 \rangle)$ |

Figure 3.571 provides an example where the MAX_RANGE_STRICTLY_DECREASING_SEQUENCE $(5, [4, 4, 6, 4, 1, 1, 3, 4, 4, 6, 6, 5, 2, 2, 4, 3])$ constraint holds.

| | |
|---|---|
| **Typical** | $\|\text{VARIABLES}\| > 1$<br>range(VARIABLES.var) > 1 |
| **Symmetry** | One and the same constant can be added to the var attribute of all items of VARIABLES. |
| **Arg. properties** | Functional dependency: VALUE determined by VARIABLES. |

Figure 3.571: Illustrating the MAX_RANGE_STRICTLY_DECREASING_SEQUENCE constraint of the **Example** slot

**Automaton**      Figures 3.572 and 3.573 respectively depict the automaton associated with the constraint
MAX_RANGE_STRICTLY_DECREASING_SEQUENCE and its simplified form.

Figure 3.572: Automaton for the MAX_RANGE_STRICTLY_DECREASING_SEQUENCE
constraint obtained by applying decoration Table 2.46 to the seed transducer of the
STRICTLY_DECREASING_SEQUENCE pattern where default is 0

Figure 3.573: Automaton for the MAX_RANGE_STRICTLY_DECREASING_SEQUENCE
constraint obtained by applying decoration Table 2.42 to the seed transducer of the
STRICTLY_DECREASING_SEQUENCE pattern where default is 0; $R_i - R_{i-1} \geq 0$ is a
linear invariant.

## 3.269 MAX_RANGE_STRICTLY_INCREASING_SEQUENCE

　　　DESCRIPTION　　　AUTOMATON

| | |
|---|---|
| **Origin** | Based on the STRICTLY_INCREASING_SEQUENCE pattern. |
| **Constraint** | MAX_RANGE_STRICTLY_INCREASING_SEQUENCE(VALUE, VARIABLES) |
| **Arguments** | VALUE : dvar<br>VARIABLES : collection(var−dvar) |

**Restrictions**

$$\mathtt{sv} \leq 1 \vee \mathtt{rv} \leq 1 \Rightarrow \mathtt{VALUE} = 0$$
$$\mathtt{VALUE} \geq 0$$
$$\mathtt{VALUE} \leq \mathtt{rv} - 1 ①$$
required(VARIABLES, var)
where
$$\mathtt{rv} = \mathtt{range}(\mathtt{VARIABLES.var})$$
$$\mathtt{sv} = |\mathtt{VARIABLES}|$$

**Purpose**

VALUE is the maximum value of the differences between the largest and smallest value in each occurrence of the STRICTLY_INCREASING_SEQUENCE pattern in the time-series given by the VARIABLES collection. If the pattern does not occur, VALUE takes the default value 0.

An occurrence of the pattern STRICTLY_INCREASING_SEQUENCE is the *maximal* sub-sequence which matches the regular expression $<^+$.

Assume that the occurrence of the pattern STRICTLY_INCREASING_SEQUENCE starts at position $i$ and ends at position $j$. The feature RANGE computes the range of the values from index $i$ to index $j + 1$.

**Example**

$$(5, \langle 4, 3, 5, 5, 2, 1, 1, 2, 3, 4, 6, 6, 3, 1, 2, 3 \rangle)$$

Figure 3.574 provides an example where the MAX_RANGE_STRICTLY_INCREASING_SEQUENCE $(5, [4, 3, 5, 5, 2, 1, 1, 2, 3, 4, 6, 6, 3, 1, 2, 3])$ constraint holds.

| | |
|---|---|
| **Typical** | $\|\mathtt{VARIABLES}\| > 1$<br>range(VARIABLES.var) > 1 |
| **Symmetry** | One and the same constant can be added to the var attribute of all items of VARIABLES. |
| **Arg. properties** | Functional dependency: VALUE determined by VARIABLES. |

Figure 3.574: Illustrating the MAX_RANGE_STRICTLY_INCREASING_SEQUENCE constraint of the **Example** slot

Figures 3.575 and 3.576 respectively depict the automaton associated with the constraint MAX_RANGE_STRICTLY_INCREASING_SEQUENCE and its simplified form.

Figure 3.575: Automaton for the MAX_RANGE_STRICTLY_INCREASING_SEQUENCE constraint obtained by applying decoration Table 2.46 to the seed transducer of the STRICTLY_INCREASING_SEQUENCE pattern where default is 0

Figure 3.576: Automaton for the MAX_RANGE_STRICTLY_INCREASING_SEQUENCE constraint obtained by applying decoration Table 2.43 to the seed transducer of the STRICTLY_INCREASING_SEQUENCE pattern where default is 0; $R_i - R_{i-1} \geq 0$ is a linear invariant.

## 3.270  MAX_SURF_BUMP_ON_DECREASING_SEQUENCE

► ▷ ◁ ◀        **DESCRIPTION**            **AUTOMATON**

| | |
|---|---|
| **Origin** | Based on the BUMP_ON_DECREASING_SEQUENCE pattern. |
| **Constraint** | MAX_SURF_BUMP_ON_DECREASING_SEQUENCE(VALUE, VARIABLES) |
| **Arguments** | VALUE     : dvar<br>VARIABLES : collection(var−dvar) |

**Restrictions**

$\mathtt{sv} \leq 5 \vee \mathtt{rv} \leq 2 \Rightarrow \mathtt{VALUE} = -\infty$
$\mathtt{VALUE} = -\infty \vee \mathtt{VALUE} \geq 3 * \mathtt{minv} + 3$
$\mathtt{VALUE} \leq 3 * \mathtt{maxv} - 3$①
required(VARIABLES, var)
where
maxv = maxval(VARIABLES.var)
minv = minval(VARIABLES.var)
sv = |VARIABLES|
rv = range(VARIABLES.var)

**Purpose**

VALUE is the maximal surface of occurrences of the BUMP_ON_DECREASING_SEQUENCE pattern in the time-series given by the VARIABLES collection. If the pattern does not occur, VALUE takes the default value $-\infty$.

An occurrence of the pattern BUMP_ON_DECREASING_SEQUENCE is the subsequence which matches the regular expression >><>>.

Assume that the occurrence of the pattern BUMP_ON_DECREASING_SEQUENCE starts at position $i$ and ends at position $j$. The feature SURF computes the sum of the values from index $i + 2$ to index $j$.

**Example**

$(16, \langle 7, 6, 5, 6, 5, 4, 1, 4, 7, 5, 4, 2, 5, 4, 3, 3 \rangle)$

Figure 3.577 provides an example where the MAX_SURF_BUMP_ON_DECREASING_SEQUENCE $(16, [7, 6, 5, 6, 5, 4, 1, 4, 7, 5, 4, 2, 5, 4, 3, 3])$ constraint holds.

**Typical**

$|\mathtt{VARIABLES}| > 5$
range(VARIABLES.var) > 2

**Arg. properties**

Functional dependency: VALUE determined by VARIABLES.

Figure 3.577: Illustrating the max_surf_bump_on_decreasing_sequence constraint of the **Example** slot

**Automaton**      Figures 3.578 and 3.579 respectively depict the automaton associated with the constraint
MAX_SURF_BUMP_ON_DECREASING_SEQUENCE and its simplified form.

Figure 3.578: Automaton for the MAX_SURF_BUMP_ON_DECREASING_SEQUENCE
constraint obtained by applying decoration Table 2.35 to the seed transducer of the
BUMP_ON_DECREASING_SEQUENCE pattern where default is $-\infty$

Figure 3.579: Automaton for the MAX_SURF_BUMP_ON_DECREASING_SEQUENCE constraint obtained by applying decoration Table 2.25 to the seed transducer of the BUMP_ON_DECREASING_SEQUENCE pattern where default is $-\infty$; $R_i - R_{i-1} \geq 0$ is a linear invariant.

## 3.271 MAX_SURF_DECREASING

        DESCRIPTION        AUTOMATON

| | |
|---|---|
| **Origin** | Based on the DECREASING pattern. |
| **Constraint** | MAX_SURF_DECREASING(VALUE, VARIABLES) |
| **Arguments** | VALUE      : dvar<br>VARIABLES  : collection(var−dvar) |

**Restrictions**

$\text{sv} \leq 1 \vee \text{rv} \leq 1 \Rightarrow \text{VALUE} = -\infty$

$\text{VALUE} = -\infty \vee \text{VALUE} \geq 2 * \text{minv} + 1$

$\text{VALUE} \leq 2 * \text{maxv} - 1$①

required(VARIABLES, var)

where

maxv =maxval(VARIABLES.var)

minv =minval(VARIABLES.var)

$\text{sv} = |\text{VARIABLES}|$

rv =range(VARIABLES.var)

**Purpose**

VALUE is the maximal surface of occurrences of the DECREASING pattern in the time-series given by the VARIABLES collection. If the pattern does not occur, VALUE takes the default value $-\infty$.

An occurrence of the pattern DECREASING is the subsequence which matches the regular expression >.

Assume that the occurrence of the pattern DECREASING starts at position $i$ and ends at position $j$. The feature SURF computes the sum of the values from index $i$ to index $j+1$.

**Example**

$(10, \langle 3, 4, 2, 2, 5, 6, 6, 4, 4, 3, 1, 1, 4, 6, 4, 4 \rangle)$

Figure 3.580 provides an example where the MAX_SURF_DECREASING $(10, [3, 4, 2, 2, 5, 6, 6, 4, 4, 3, 1, 1, 4, 6, 4, 4])$ constraint holds.

**Typical**

$|\text{VARIABLES}| > 1$

range(VARIABLES.var) > 1

**Arg. properties**

Functional dependency: VALUE determined by VARIABLES.

Figure 3.580: Illustrating the MAX_SURF_DECREASING constraint of the **Example** slot

**Automaton**   Figures 3.581 and 3.582 respectively depict the automaton associated with the constraint MAX_SURF_DECREASING and its simplified form.

Figure 3.581: Automaton for the MAX_SURF_DECREASING constraint obtained by applying decoration Table 2.35 to the seed transducer of the DECREASING pattern where `default` is $-\infty$

Figure 3.582: Automaton for the MAX_SURF_DECREASING constraint obtained by applying decoration Table 2.38 to the seed transducer of the DECREASING pattern where `default` is $-\infty$; $R_i - R_{i-1} \geq 0$ is a linear invariant.

Table 3.27: Glue matrix for the MAX_SURF_DECREASING constraint defined as the composition of the DECREASING pattern , the feature SURF , and the aggregator max ; cells of the glue matrix are coloured with the colour of the constituent to which they are related.

# 3.272 MAX_SURF_DECREASING_SEQUENCE

▶ ▷ ◁ ◀          **DESCRIPTION**          **AUTOMATON**

**Origin**          Based on the DECREASING_SEQUENCE pattern.

**Constraint**      MAX_SURF_DECREASING_SEQUENCE(VALUE, VARIABLES)

**Arguments**       VALUE      : dvar
                    VARIABLES  : collection(var−dvar)

**Restrictions**    $\mathtt{sv} \leq 1 \vee \mathtt{rv} \leq 1 \Rightarrow \mathtt{VALUE} = -\infty$
                    $\mathtt{rv} = 2 \Rightarrow \mathtt{VALUE} = -\infty \vee \mathtt{VALUE} \geq 2 * \mathtt{minv} + 1$
                    $\mathtt{rv} \geq 3 \Rightarrow \mathtt{VALUE} = -\infty \vee \mathtt{VALUE} \geq \min(2 * \mathtt{minv} + 1, \mathtt{sv} * (\mathtt{minv} + 1))$
                    $\mathtt{rv} = 2 \Rightarrow \mathtt{VALUE} \leq 2 * \mathtt{maxv} - 1$①
                    $\mathtt{rv} \geq 3 \Rightarrow \mathtt{VALUE} \leq \max(2 * \mathtt{maxv} - 1$②$, \mathtt{sv} * (\mathtt{maxv} - 1)$③$)$
                    required(VARIABLES, var)
                    where
                    $\mathtt{maxv} = $ maxval(VARIABLES.var)
                    $\mathtt{rv} = $ range(VARIABLES.var)
                    $\mathtt{sv} = |\mathtt{VARIABLES}|$
                    $\mathtt{minv} = $ minval(VARIABLES.var)

**Purpose**         VALUE is the maximal surface of occurrences of the DECREASING_SEQUENCE pattern in
                    the time-series given by the VARIABLES collection. If the pattern does not occur, VALUE
                    takes the default value $-\infty$.
                    An occurrence of the pattern DECREASING_SEQUENCE is the *maximal* subsequence
                    which matches the regular expression $> (> | =)^* > | >$.
                    Assume that the occurrence of the pattern DECREASING_SEQUENCE starts at position $i$
                    and ends at position $j$. The feature SURF computes the sum of the values from index $i$ to
                    index $j + 1$.

**Example**         $(18, \langle 3, 4, 2, 2, 5, 6, 6, 4, 4, 3, 1, 1, 4, 6, 4, 4 \rangle)$

                    Figure 3.583 provides an example where the MAX_SURF_DECREASING_SEQUENCE
                    $(18, [3, 4, 2, 2, 5, 6, 6, 4, 4, 3, 1, 1, 4, 6, 4, 4])$ constraint holds.

**Typical**         $|\mathtt{VARIABLES}| > 1$
                    range(VARIABLES.var) $> 1$

**Arg. properties** Functional dependency: VALUE determined by VARIABLES.

Figure 3.583: Illustrating the MAX_SURF_DECREASING_SEQUENCE constraint of the **Example** slot

**Automaton** Figure 3.584 depicts the automaton associated with the constraint MAX_SURF_DECREASING_SEQUENCE.

Figure 3.584: Automaton for the MAX_SURF_DECREASING_SEQUENCE constraint obtained by applying decoration Table 2.35 to the seed transducer of the DECREASING_SEQUENCE pattern where `default` is $-\infty$; $R_i - R_{i-1} \geq 0$ is a linear invariant.

|   | $s$ | $t$ |
|---|---|---|
| $s$ | $\max(\overrightarrow{C}, \overleftarrow{C})$ | $\max(\overrightarrow{C}, \overleftarrow{C})$ |
| $t$ | $\max(\overrightarrow{C}, \overleftarrow{C})$ | $\overrightarrow{C} + \overleftarrow{C} + \overrightarrow{D} + \overleftarrow{D} + \mathtt{VAR}_i$ |

Table 3.28: Glue matrix for the MAX_SURF_DECREASING_SEQUENCE constraint defined as the composition of the DECREASING_SEQUENCE pattern, the feature SURF, and the aggregator max; cells of the glue matrix are coloured with the colour of the constituent to which they are related.

# 3.273   MAX_SURF_DECREASING_TERRACE

   **DESCRIPTION**      **AUTOMATON**

**Origin**            Based on the DECREASING_TERRACE pattern.

**Constraint**        MAX_SURF_DECREASING_TERRACE(VALUE, VARIABLES)

**Arguments**         VALUE      : dvar
                      VARIABLES  : collection(var−dvar)

**Restrictions**      $\mathtt{sv} \leq 3 \vee \mathtt{rv} \leq 2 \Rightarrow \mathtt{VALUE} = -\infty$
                      $\mathtt{VALUE} = -\infty \vee \mathtt{VALUE} \geq \min(2*(\mathtt{minv}+1),(\mathtt{sv}-2)*(\mathtt{minv}+1))$
                      $\mathtt{VALUE} \leq \max(2*(\mathtt{maxv}-1)①,(\mathtt{sv}-2)*(\mathtt{maxv}-1)②)$
                      required(VARIABLES, var)
                      where
                      maxv =maxval(VARIABLES.var)
                      sv = |VARIABLES|
                      minv =minval(VARIABLES.var)
                      rv =range(VARIABLES.var)

**Purpose**           VALUE is the maximal surface of occurrences of the DECREASING_TERRACE pattern in
                      the time-series given by the VARIABLES collection. If the pattern does not occur, VALUE
                      takes the default value $-\infty$.
                      An occurrence of the pattern DECREASING_TERRACE is the *maximal* subsequence which
                      matches the regular expression $>=^+>$.
                      Assume that the occurrence of the pattern DECREASING_TERRACE starts at position $i$
                      and ends at position $j$. The feature SURF computes the sum of the values from index
                      $i+1$ to index $j$.

**Example**           $(8, \langle 6, 4, 4, 4, 5, 2, 2, 1, 3, 3, 5, 4, 4, 3, 3, 3 \rangle)$

                      Figure 3.585 provides an example where the MAX_SURF_DECREASING_TERRACE
                      $(8, [6, 4, 4, 4, 5, 2, 2, 1, 3, 3, 5, 4, 4, 3, 3, 3])$ constraint holds.

**Typical**           $|\mathtt{VARIABLES}| > 3$
                      range(VARIABLES.var) $> 2$

**Arg. properties**   Functional dependency: VALUE determined by VARIABLES.

Figure 3.585: Illustrating the MAX_SURF_DECREASING_TERRACE constraint of the **Example** slot

 Figures 3.586 and 3.587 respectively depict the automaton associated with the constraint MAX_SURF_DECREASING_TERRACE and its simplified form.

Figure 3.586: Automaton for the MAX_SURF_DECREASING_TERRACE constraint obtained by applying decoration Table 2.35 to the seed transducer of the DECREAS­ING_TERRACE pattern where default is $-\infty$

Figure 3.587: Automaton for the MAX_SURF_DECREASING_TERRACE constraint obtained by applying decoration Table 2.25 to the seed transducer of the DECREAS­ING_TERRACE pattern where default is $-\infty$; $R_i - R_{i-1} \geq 0$ is a linear invariant.

|   | $s$ | $r$ | $t$ |
|---|---|---|---|
| $s$ | $\max(\overrightarrow{C}, \overleftarrow{C})$ | $\max(\overrightarrow{C}, \overleftarrow{C})$ | $\max(\overrightarrow{C}, \overleftarrow{C})$ |
| $r$ | $\max(\overrightarrow{C}, \overleftarrow{C})$ | $\max(\overrightarrow{C}, \overleftarrow{C})$ | $\overrightarrow{D} + \overleftarrow{D} + \text{VAR}_i$ |
| $t$ | $\max(\overrightarrow{C}, \overleftarrow{C})$ | $\overrightarrow{D} + \overleftarrow{D} + \text{VAR}_i$ | $\overrightarrow{D} + \overleftarrow{D} + \text{VAR}_i$ |

Table 3.29: Glue matrix for the MAX_SURF_DECREASING_TERRACE constraint defined as the composition of the DECREASING_TERRACE pattern , the feature SURF , and the aggregator max ; cells of the glue matrix are coloured with the colour of the constituent to which they are related.

# 3.274   MAX_SURF_DIP_ON_INCREASING_SEQUENCE

▶ ▷ ◁ ◀    **DESCRIPTION**    **AUTOMATON**

| | |
|---|---|
| **Origin** | Based on the DIP_ON_INCREASING_SEQUENCE pattern. |
| **Constraint** | MAX_SURF_DIP_ON_INCREASING_SEQUENCE(VALUE, VARIABLES) |
| **Arguments** | VALUE     : dvar |
| | VARIABLES : collection(var−dvar) |

**Restrictions**

$\text{sv} \leq 5 \vee \text{rv} \leq 2 \Rightarrow \text{VALUE} = -\infty$

$\text{VALUE} = -\infty \vee \text{VALUE} \geq 3 * \text{minv} + 3$

$\text{VALUE} \leq 3 * \text{maxv} - 3 ①$

required(VARIABLES, var)

where

maxv = maxval(VARIABLES.var)

minv = minval(VARIABLES.var)

sv = |VARIABLES|

rv = range(VARIABLES.var)

**Purpose**

VALUE is the maximal surface of occurrences of the DIP_ON_INCREASING_SEQUENCE pattern in the time-series given by the VARIABLES collection. If the pattern does not occur, VALUE takes the default value $-\infty$.

An occurrence of the pattern DIP_ON_INCREASING_SEQUENCE is the subsequence which matches the regular expression <<><<.

Assume that the occurrence of the pattern DIP_ON_INCREASING_SEQUENCE starts at position $i$ and ends at position $j$. The feature SURF computes the sum of the values from index $i + 2$ to index $j$.

**Example**

$(10, \langle 1, 2, 3, 2, 5, 6, 7, 4, 1, 3, 4, 6, 1, 2, 4, 4 \rangle)$

Figure 3.588 provides an example where the MAX_SURF_DIP_ON_INCREASING_SEQUENCE $(10, [1, 2, 3, 2, 5, 6, 7, 4, 1, 3, 4, 6, 1, 2, 4, 4])$ constraint holds.

| | |
|---|---|
| **Typical** | $|\text{VARIABLES}| > 5$ |
| | range(VARIABLES.var) $> 2$ |
| **Arg. properties** | Functional dependency: VALUE determined by VARIABLES. |

Figure 3.588: Illustrating the MAX_SURF_DIP_ON_INCREASING_SEQUENCE constraint of the **Example** slot

**Automaton**     Figures 3.589 and 3.590 respectively depict the automaton associated with the constraint MAX_SURF_DIP_ON_INCREASING_SEQUENCE and its simplified form.

Figure 3.589: Automaton for the MAX_SURF_DIP_ON_INCREASING_SEQUENCE constraint obtained by applying decoration Table 2.35 to the seed transducer of the DIP_ON_INCREASING_SEQUENCE pattern where default is $-\infty$

Figure 3.590: Automaton for the MAX_SURF_DIP_ON_INCREASING_SEQUENCE constraint obtained by applying decoration Table 2.25 to the seed transducer of the DIP_ON_INCREASING_SEQUENCE pattern where default is $-\infty$; $R_i - R_{i-1} \geq 0$ is a linear invariant.

## 3.275 MAX_SURF_GORGE

▶ ▷ ◁ ◀        DESCRIPTION        AUTOMATON

$(> \mid > (= \mid >)^* >)(< \mid < (= \mid <)^* <)$

**Origin**          Based on the GORGE pattern.

**Constraint**      MAX_SURF_GORGE(VALUE, VARIABLES)

**Arguments**       VALUE      : dvar
                    VARIABLES  : collection(var−dvar)

**Restrictions**    $\mathtt{sv} \le 2 \vee \mathtt{rv} \le 1 \Rightarrow \mathtt{VALUE} = -\infty$
                    $\mathtt{rv} = 2 \Rightarrow \mathtt{VALUE} = -\infty \vee \mathtt{VALUE} \ge \mathtt{minv}$
                    $\mathtt{rv} \ge 3 \Rightarrow \mathtt{VALUE} = -\infty \vee \mathtt{VALUE} \ge \min(\mathtt{minv}, (\mathtt{sv} - 2) * (\mathtt{minv} + 1) - 1)$
                    $\mathtt{rv} = 2 \Rightarrow \mathtt{VALUE} \le \mathtt{maxv} - 1$①
                    $\mathtt{rv} \ge 3 \Rightarrow \mathtt{VALUE} \le \max(\mathtt{maxv} - 1$②$, (\mathtt{sv} - 2) * (\mathtt{maxv} - 1) - 1$③$)$
                    required(VARIABLES, var)
                    where
                    $\mathtt{maxv} = \mathtt{maxval}(\mathtt{VARIABLES.var})$
                    $\mathtt{rv} = \mathtt{range}(\mathtt{VARIABLES.var})$
                    $\mathtt{sv} = |\mathtt{VARIABLES}|$
                    $\mathtt{minv} = \mathtt{minval}(\mathtt{VARIABLES.var})$

**Purpose**         VALUE is the maximal surface of occurrences of the GORGE pattern in the time-series given by the VARIABLES collection. If the pattern does not occur, VALUE takes the default value $-\infty$.
                    An occurrence of the pattern GORGE is the *maximal* subsequence which matches the regular expression $(> \mid > (= \mid >)^* >)(< \mid < (= \mid <)^* <)$.
                    Assume that the occurrence of the pattern GORGE starts at position $i$ and ends at position $j$. The feature SURF computes the sum of the values from index $i + 1$ to index $j$.

**Example**         $(11, \langle 1, 7, 3, 4, 4, 5, 5, 4, 2, 2, 6, 5, 4, 6, 5, 7 \rangle)$

Figure  3.591  provides an example where the MAX_SURF_GORGE $(11, [1, 7, 3, 4, 4, 5, 5, 4, 2, 2, 6, 5, 4, 6, 5, 7])$ constraint holds.

**Typical**         $|\mathtt{VARIABLES}| > 2$
                    $\mathtt{range}(\mathtt{VARIABLES.var}) > 1$

**Symmetry**        Items of VARIABLES can be reversed.

**Arg. properties** Functional dependency: VALUE determined by VARIABLES.

Figure 3.591: Illustrating the MAX_SURF_GORGE constraint of the **Example** slot

**Automaton**          Figure 3.592 depicts the automaton associated with the constraint MAX_SURF_GORGE.

Figure 3.592: Automaton for the MAX_SURF_GORGE constraint obtained by applying decoration Table 2.35 to the seed transducer of the GORGE pattern where default is $-\infty$ (transition $u \to r$ has the same accumulator update as transition $r \to u$); $R_i - R_{i-1} \geq 0$ is a linear invariant.

| | $s$ | $r$ | $t$ | $u$ |
|---|---|---|---|---|
| $s$ | $\max(\overrightarrow{C}, \overleftarrow{C})$ | $\max(\overrightarrow{C}, \overleftarrow{C})$ | $\max(\overrightarrow{C}, \overleftarrow{C})$ | $\max(\overrightarrow{C}, \overleftarrow{C})$ |
| $r$ | $\max(\overrightarrow{C}, \overleftarrow{C})$ | $\overrightarrow{D} + \overleftarrow{D} + \mathtt{VAR}_i$ | $\overleftarrow{C} + \overrightarrow{D} + \overleftarrow{D} + \mathtt{VAR}_i$ | $\max(\overrightarrow{C}, \overleftarrow{C})$ |
| $t$ | $\max(\overrightarrow{C}, \overleftarrow{C})$ | $\overrightarrow{C} + \overrightarrow{D} + \overleftarrow{D} + \mathtt{VAR}_i$ | $\max(\overrightarrow{C}, \overleftarrow{C})$ | $\overrightarrow{C} + \overrightarrow{D} + \overleftarrow{D} + \mathtt{VAR}_i$ |
| $u$ | $\max(\overrightarrow{C}, \overleftarrow{C})$ | $\max(\overrightarrow{C}, \overleftarrow{C})$ | $\overleftarrow{C} + \overrightarrow{D} + \overleftarrow{D} + \mathtt{VAR}_i$ | $\max(\overrightarrow{C}, \overleftarrow{C})$ |

Table 3.30: Glue matrix for the MAX_SURF_GORGE constraint defined as the composition of the GORGE pattern , the feature SURF , and the aggregator max ; cells of the glue matrix are coloured with the colour of the constituent to which they are related.

# 3.276   MAX_SURF_INCREASING

     **DESCRIPTION**          **AUTOMATON**

| | |
|---|---|
| **Origin** | Based on the INCREASING pattern. |
| **Constraint** | MAX_SURF_INCREASING(VALUE, VARIABLES) |
| **Arguments** | VALUE      : dvar<br>VARIABLES : collection(var−dvar) |

**Restrictions**

$$\text{sv} \leq 1 \lor \text{rv} \leq 1 \Rightarrow \text{VALUE} = -\infty$$
$$\text{VALUE} = -\infty \lor \text{VALUE} \geq 2 * \text{minv} + 1$$
$$\text{VALUE} \leq 2 * \text{maxv} - 1 \text{①}$$
required(VARIABLES, var)
where
$$\text{maxv} = \text{maxval}(\text{VARIABLES.var})$$
$$\text{minv} = \text{minval}(\text{VARIABLES.var})$$
$$\text{sv} = |\text{VARIABLES}|$$
$$\text{rv} = \text{range}(\text{VARIABLES.var})$$

**Purpose**

VALUE is the maximal surface of occurrences of the INCREASING pattern in the time-series given by the VARIABLES collection. If the pattern does not occur, VALUE takes the default value $-\infty$.

An occurrence of the pattern INCREASING is the subsequence which matches the regular expression <.

Assume that the occurrence of the pattern INCREASING starts at position $i$ and ends at position $j$. The feature SURF computes the sum of the values from index $i$ to index $j+1$.

**Example**

$$(10, \langle 4, 3, 5, 5, 2, 1, 1, 3, 3, 4, 6, 6, 3, 1, 3, 3 \rangle)$$

Figure 3.593 provides an example where the MAX_SURF_INCREASING $(10, [4, 3, 5, 5, 2, 1, 1, 3, 3, 4, 6, 6, 3, 1, 3, 3])$ constraint holds.

**Typical**

$$|\text{VARIABLES}| > 1$$
$$\text{range}(\text{VARIABLES.var}) > 1$$

**Arg. properties**

Functional dependency: VALUE determined by VARIABLES.

Figure 3.593: Illustrating the MAX_SURF_INCREASING constraint of the **Example** slot

**Automaton**     Figures 3.594 and 3.595 respectively depict the automaton associated with the constraint MAX_SURF_INCREASING and its simplified form.

Figure 3.594: Automaton for the MAX_SURF_INCREASING constraint obtained by applying decoration Table 2.35 to the seed transducer of the INCREASING pattern where default is $-\infty$

Figure 3.595: Automaton for the MAX_SURF_INCREASING constraint obtained by applying decoration Table 2.38 to the seed transducer of the INCREASING pattern where default is $-\infty$; $R_i - R_{i-1} \geq 0$ is a linear invariant.

Table 3.31: Glue matrix for the MAX_SURF_INCREASING constraint defined as the composition of the INCREASING pattern, the feature SURF, and the aggregator max; cells of the glue matrix are coloured with the colour of the constituent to which they are related.

# 3.277   MAX_SURF_INCREASING_SEQUENCE

▶ ▷ ◁ ◀        **DESCRIPTION**      **AUTOMATON**

$< (< | =)^* < |$
$<$

**Origin**            Based on the INCREASING_SEQUENCE pattern.

**Constraint**        MAX_SURF_INCREASING_SEQUENCE(VALUE, VARIABLES)

**Arguments**         VALUE      : dvar
                      VARIABLES  : collection(var−dvar)

**Restrictions**      $\text{sv} \leq 1 \vee \text{rv} \leq 1 \Rightarrow \text{VALUE} = -\infty$
                      $\text{rv} = 2 \Rightarrow \text{VALUE} = -\infty \vee \text{VALUE} \geq 2 * \text{minv} + 1$
                      $\text{rv} \geq 3 \Rightarrow \text{VALUE} = -\infty \vee \text{VALUE} \geq \min(2 * \text{minv} + 1, \text{sv} * (\text{minv} + 1))$
                      $\text{rv} = 2 \Rightarrow \text{VALUE} \leq 2 * \text{maxv} - 1$ ①
                      $\text{rv} \geq 3 \Rightarrow \text{VALUE} \leq \max(2 * \text{maxv} - 1$ ②$, \text{sv} * (\text{maxv} - 1)$ ③$)$
                      required(VARIABLES, var)
                      where
                      $\text{maxv} = $maxval(VARIABLES.var)
                      $\text{rv} = $range(VARIABLES.var)
                      $\text{sv} = |$VARIABLES$|$
                      $\text{minv} = $minval(VARIABLES.var)

**Purpose**           VALUE is the maximal surface of occurrences of the INCREASING_SEQUENCE pattern in
                      the time-series given by the VARIABLES collection. If the pattern does not occur, VALUE
                      takes the default value $-\infty$.
                      An occurrence of the pattern INCREASING_SEQUENCE is the *maximal* subsequence
                      which matches the regular expression $< (< | =)^* < | <$.
                      Assume that the occurrence of the pattern INCREASING_SEQUENCE starts at position $i$
                      and ends at position $j$. The feature SURF computes the sum of the values from index $i$ to
                      index $j + 1$.

**Example**           $(17, \langle 4, 3, 5, 5, 2, 1, 1, 3, 3, 4, 6, 6, 3, 1, 3, 3 \rangle)$

                      Figure 3.596 provides an example where the MAX_SURF_INCREASING_SEQUENCE
                      $(17, [4, 3, 5, 5, 2, 1, 1, 3, 3, 4, 6, 6, 3, 1, 3, 3])$ constraint holds.

**Typical**           $|$VARIABLES$| > 1$
                      range(VARIABLES.var) $> 1$

**Arg. properties**   Functional dependency: VALUE determined by VARIABLES.

Figure 3.596: Illustrating the MAX_SURF_INCREASING_SEQUENCE constraint of the **Example** slot

**Automaton**    Figure 3.597 depicts the automaton associated with the constraint MAX_SURF_INCREASING_SEQUENCE.

Figure 3.597: Automaton for the MAX_SURF_INCREASING_SEQUENCE constraint obtained by applying decoration Table 2.35 to the seed transducer of the INCREASING_SEQUENCE pattern where `default` is $-\infty$; $R_i - R_{i-1} \geq 0$ is a linear invariant.

|   | $s$ | $t$ |
|---|---|---|
| $s$ | $\max(\overrightarrow{C}, \overleftarrow{C})$ | $\max(\overrightarrow{C}, \overleftarrow{C})$ |
| $t$ | $\max(\overrightarrow{C}, \overleftarrow{C})$ | $\overrightarrow{C} + \overleftarrow{C} + \overrightarrow{D} + \overleftarrow{D} + \mathtt{VAR}_i$ |

Table 3.32: Glue matrix for the MAX_SURF_INCREASING_SEQUENCE constraint defined as the composition of the INCREASING_SEQUENCE pattern, the feature SURF, and the aggregator max; cells of the glue matrix are coloured with the colour of the constituent to which they are related.

## 3.278   MAX_SURF_INCREASING_TERRACE

▶ ▷ ◁ ◀        **DESCRIPTION**        **AUTOMATON**

**Origin**         Based on the INCREASING_TERRACE pattern.

**Constraint**     MAX_SURF_INCREASING_TERRACE(VALUE, VARIABLES)

**Arguments**      VALUE      : dvar
                   VARIABLES  : collection(var−dvar)

**Restrictions**   $\mathtt{sv} \leq 3 \vee \mathtt{rv} \leq 2 \Rightarrow \mathtt{VALUE} = -\infty$
                   $\mathtt{VALUE} = -\infty \vee \mathtt{VALUE} \geq \min(2 * (\mathtt{minv} + 1), (\mathtt{sv} - 2) * (\mathtt{minv} + 1))$
                   $\mathtt{VALUE} \leq \max(2 * (\mathtt{maxv} - 1)①, (\mathtt{sv} - 2) * (\mathtt{maxv} - 1)②)$
                   required(VARIABLES, var)
                   where
                   maxv = maxval(VARIABLES.var)
                   sv = |VARIABLES|
                   minv = minval(VARIABLES.var)
                   rv = range(VARIABLES.var)

**Purpose**        VALUE is the maximal surface of occurrences of the INCREASING_TERRACE pattern in
                   the time-series given by the VARIABLES collection. If the pattern does not occur, VALUE
                   takes the default value $-\infty$.
                   An occurrence of the pattern INCREASING_TERRACE is the *maximal* subsequence which
                   matches the regular expression $<=^{+}<$.
                   Assume that the occurrence of the pattern INCREASING_TERRACE starts at position $i$
                   and ends at position $j$. The feature SURF computes the sum of the values from index
                   $i + 1$ to index $j$.

**Example**        $(10, \langle 1, 3, 3, 3, 2, 5, 5, 6, 4, 4, 2, 3, 3, 3, 4, 4 \rangle)$

                   Figure 3.598 provides an example where the MAX_SURF_INCREASING_TERRACE
                   $(10, [1, 3, 3, 3, 2, 5, 5, 6, 4, 4, 2, 3, 3, 3, 4, 4])$ constraint holds.

**Typical**        $|\mathtt{VARIABLES}| > 3$
                   range(VARIABLES.var) > 2

**Arg. properties**  Functional dependency: VALUE determined by VARIABLES.

Figure 3.598: Illustrating the MAX_SURF_INCREASING_TERRACE constraint of the **Example** slot

  Figures 3.599 and 3.600 respectively depict the automaton associated with the constraint
MAX_SURF_INCREASING_TERRACE and its simplified form.

Figure 3.599: Automaton for the MAX_SURF_INCREASING_TERRACE constraint obtained by applying decoration Table 2.35 to the seed transducer of the INCREASING_TERRACE pattern where default is $-\infty$

Figure 3.600: Automaton for the MAX_SURF_INCREASING_TERRACE constraint obtained by applying decoration Table 2.25 to the seed transducer of the INCREASING_TERRACE pattern where default is $-\infty$; $R_i - R_{i-1} \geq 0$ is a linear invariant.

|   | $s$ | $r$ | $t$ |
|---|---|---|---|
| $s$ | $\max(\overrightarrow{C}, \overleftarrow{C})$ | $\max(\overrightarrow{C}, \overleftarrow{C})$ | $\max(\overrightarrow{C}, \overleftarrow{C})$ |
| $r$ | $\max(\overrightarrow{C}, \overleftarrow{C})$ | $\max(\overrightarrow{C}, \overleftarrow{C})$ | $\overrightarrow{D} + \overleftarrow{D} + \mathtt{VAR}_i$ |
| $t$ | $\max(\overrightarrow{C}, \overleftarrow{C})$ | $\overrightarrow{D} + \overleftarrow{D} + \mathtt{VAR}_i$ | $\overrightarrow{D} + \overleftarrow{D} + \mathtt{VAR}_i$ |

Table 3.33: Glue matrix for the MAX_SURF_INCREASING_TERRACE constraint defined as the composition of the INCREASING_TERRACE pattern , the feature SURF , and the aggregator max ; cells of the glue matrix are coloured with the colour of the constituent to which they are related.

# 3.279  MAX_SURF_INFLEXION

▶ ▷ ◁ ◀          **DESCRIPTION**          **AUTOMATON**

$< (< | =)^* > | > (> | =)^* <$

| | |
|---|---|
| **Origin** | Based on the INFLEXION pattern. |
| **Constraint** | MAX_SURF_INFLEXION(VALUE, VARIABLES) |
| **Arguments** | VALUE       :  dvar<br>VARIABLES  :  collection(var−dvar) |

**Restrictions**

$\mathtt{sv} \leq 2 \vee \mathtt{rv} \leq 1 \Rightarrow \mathtt{VALUE} = -\infty$
$\mathtt{VALUE} = -\infty \vee \mathtt{VALUE} \geq \min(\mathtt{minv}, (\mathtt{sv} - 2) * \mathtt{minv})$
$\mathtt{VALUE} \leq \max(\mathtt{maxv}①, (\mathtt{sv} - 2) * \mathtt{maxv}②)$
required(VARIABLES, var)
where
$\mathtt{maxv} = $maxval$(\mathtt{VARIABLES.var})$
$\mathtt{sv} = |\mathtt{VARIABLES}|$
$\mathtt{minv} = $minval$(\mathtt{VARIABLES.var})$
$\mathtt{rv} = $range$(\mathtt{VARIABLES.var})$

**Purpose**

VALUE is the maximal surface of occurrences of the INFLEXION pattern in the time-series given by the VARIABLES collection. If the pattern does not occur, VALUE takes the default value $-\infty$.
An occurrence of the pattern INFLEXION is the *maximal* subsequence which matches the regular expression $< (< | =)^* > | > (> | =)^* <$.
Assume that the occurrence of the pattern INFLEXION starts at position $i$ and ends at position $j$. The feature SURF computes the sum of the values from index $i + 1$ to index $j$.

**Example**          $(14, \langle 1, 2, 6, 6, 4, 4, 3, 5, 2, 5, 1, 5, 3, 3, 4, 4 \rangle)$

Figure 3.601 provides an example where the MAX_SURF_INFLEXION $(14, [1, 2, 6, 6, 4, 4, 3, 5, 2, 5, 1, 5, 3, 3, 4, 4])$ constraint holds.

| | |
|---|---|
| **Typical** | $|\mathtt{VARIABLES}| > 2$<br>range$(\mathtt{VARIABLES.var}) > 1$ |
| **Symmetry** | Items of VARIABLES can be reversed. |
| **Arg. properties** | Functional dependency: VALUE determined by VARIABLES. |

Figure 3.601: Illustrating the MAX_SURF_INFLEXION constraint of the **Example** slot

**Automaton**   Figures 3.602 and 3.603 respectively depict the automaton associated with the constraint MAX_SURF_INFLEXION and its simplified form.

Figure 3.602: Automaton for the MAX_SURF_INFLEXION constraint obtained by applying decoration Table 2.35 to the seed transducer of the INFLEXION pattern where default is $-\infty$ (transition $r \rightarrow t$ has the same accumulators updates as transition $t \rightarrow r$)

Figure 3.603: Automaton for the MAX_SURF_INFLEXION constraint obtained by applying decoration Table 2.25 to the seed transducer of the INFLEXION pattern where default is $-\infty$ (transition $r \rightarrow t$ has the same accumulators updates as transition $t \rightarrow r$); $R_i - R_{i-1} \geq 0$ is a linear invariant.

## 3.280   MAX_SURF_PEAK

► ▷ ◁ ◄        **DESCRIPTION**         **AUTOMATON**

$< (= | <)^* (> | =)^* >$

**Origin**          Based on the PEAK pattern.

**Constraint**      MAX_SURF_PEAK(VALUE, VARIABLES)

**Arguments**       VARIABLE   : dvar
                    VARIABLES  : collection(var−dvar)

**Restrictions**    $\mathtt{sv} \leq 2 \vee \mathtt{rv} \leq 1 \Rightarrow \mathtt{VALUE} = -\infty$
                    $\mathtt{VALUE} = -\infty \vee \mathtt{VALUE} \geq \min(\mathtt{minv} + 1, (\mathtt{sv} - 2) * (\mathtt{minv} + 1))$
                    $\mathtt{VALUE} \leq \max(\mathtt{maxv}❶, (\mathtt{sv} - 2) * \mathtt{maxv}❷)$
                    required(VARIABLES, var)
                    where
                    $\mathtt{minv} = \mathtt{minval}(\mathtt{VARIABLES.var})$
                    $\mathtt{maxv} = \mathtt{maxval}(\mathtt{VARIABLES.var})$
                    $\mathtt{sv} = |\mathtt{VARIABLES}|$
                    $\mathtt{rv} = \mathtt{range}(\mathtt{VARIABLES.var})$

**Purpose**         VALUE is the maximal surface of occurrences of the PEAK pattern in the time-series given by the VARIABLES collection. If the pattern does not occur, VALUE takes the default value $-\infty$.
                    An occurrence of the pattern PEAK is the *maximal* subsequence which matches the regular expression $< (= | <)^* (> | =)^* >$.
                    Assume that the occurrence of the pattern PEAK starts at position $i$ and ends at position $j$. The feature SURF computes the sum of the values from index $i + 1$ to index $j$.

**Example**         $(14, \langle 7, 5, 5, 1, 4, 5, 2, 2, 3, 5, 6, 2, 3, 3, 3, 1 \rangle)$

Figure 3.604 provides an example where the MAX_SURF_PEAK $(14, [7, 5, 5, 1, 4, 5, 2, 2, 3, 5, 6, 2, 3, 3, 3, 1])$ constraint holds.

**Typical**         $|\mathtt{VARIABLES}| > 2$
                    range(VARIABLES.var) > 1

**Symmetry**        Items of VARIABLES can be reversed.

**Arg. properties** Functional dependency: VALUE determined by VARIABLES.

Figure 3.604: Illustrating the MAX_SURF_PEAK constraint of the **Example** slot

**Automaton**        Figure 3.605 depicts the automaton associated with the constraint MAX_SURF_PEAK.

Figure 3.605: Automaton for the MAX_SURF_PEAK constraint obtained by applying decoration Table 2.35 to the seed transducer of the PEAK pattern where default is $-\infty$; $R_i - R_{i-1} \geq 0$ is a linear invariant.

|   | $s$ | $r$ | $t$ |
|---|-----|-----|-----|
| $s$ | $\max(\overrightarrow{C}, \overleftarrow{C})$ | $\max(\overrightarrow{C}, \overleftarrow{C})$ | $\max(\overrightarrow{C}, \overleftarrow{C})$ |
| $r$ | $\max(\overrightarrow{C}, \overleftarrow{C})$ | $\overrightarrow{D} + \overleftarrow{D} + \mathtt{VAR}_i$ | $\overleftarrow{C} + \overrightarrow{D} + \overleftarrow{D} + \mathtt{VAR}_i$ |
| $t$ | $\max(\overrightarrow{C}, \overleftarrow{C})$ | $\overrightarrow{C} + \overrightarrow{D} + \overleftarrow{D} + \mathtt{VAR}_i$ | $\max(\overrightarrow{C}, \overleftarrow{C})$ |

Table 3.34: Glue matrix for the MAX_SURF_PEAK constraint defined as the composition of the PEAK pattern , the feature SURF , and the aggregator max ; cells of the glue matrix are coloured with the colour of the constituent to which they are related.

## 3.281 MAX_SURF_PLAIN

                    DESCRIPTION                    AUTOMATON

| | |
|---|---|
| **Origin** | Based on the PLAIN pattern. |
| **Constraint** | MAX_SURF_PLAIN(VALUE, VARIABLES) |
| **Arguments** | VALUE      : dvar<br>VARIABLES  : collection(var−dvar) |

**Restrictions**

$$\text{sv} \leq 2 \vee \text{rv} \leq 1 \Rightarrow \text{VALUE} = -\infty$$
$$\text{VALUE} = -\infty \vee \text{VALUE} \geq \min(\text{minv}, (\text{sv} - 2) * \text{minv})$$
$$\text{VALUE} \leq \max(\text{maxv} - 1①, (\text{sv} - 2) * (\text{maxv} - 1)②)$$
required(VARIABLES, var)
where
maxv = maxval(VARIABLES.var)
sv = |VARIABLES|
minv = minval(VARIABLES.var)
rv = range(VARIABLES.var)

**Purpose**

VALUE is the maximal surface of occurrences of the PLAIN pattern in the time-series given by the VARIABLES collection. If the pattern does not occur, VALUE takes the default value $-\infty$.

An occurrence of the pattern PLAIN is the *maximal* subsequence which matches the regular expression $> =^* <$.

Assume that the occurrence of the pattern PLAIN starts at position $i$ and ends at position $j$. The feature SURF computes the sum of the values from index $i + 1$ to index $j$.

**Example**

$(6, \langle 2, 3, 6, 5, 7, 6, 6, 4, 5, 5, 4, 3, 3, 6, 6, 3 \rangle)$

Figure 3.606 provides an example where the MAX_SURF_PLAIN $(6, [2, 3, 6, 5, 7, 6, 6, 4, 5, 5, 4, 3, 3, 6, 6, 3])$ constraint holds.

| | |
|---|---|
| **Typical** | $|\text{VARIABLES}| > 2$<br>range(VARIABLES.var) $> 1$ |
| **Symmetry** | Items of VARIABLES can be reversed. |
| **Arg. properties** | Functional dependency: VALUE determined by VARIABLES. |

Figure 3.606: Illustrating the MAX_SURF_PLAIN constraint of the **Example** slot

**Automaton**     Figures 3.607 and 3.608 respectively depict the automaton associated with the constraint MAX_SURF_PLAIN and its simplified form.

Figure 3.607: Automaton for the MAX_SURF_PLAIN constraint obtained by applying decoration Table 2.35 to the seed transducer of the PLAIN pattern where default is $-\infty$

Figure 3.608: Automaton for the MAX_SURF_PLAIN constraint obtained by applying decoration Table 2.25 to the seed transducer of the PLAIN pattern where default is $-\infty$; $R_i - R_{i-1} \geq 0$ is a linear invariant.

|   | $s$ | $r$ | $t$ |
|---|---|---|---|
| $s$ | $\max(\overrightarrow{C}, \overleftarrow{C})$ | $\max(\overrightarrow{C}, \overleftarrow{C})$ | $\max(\overrightarrow{C}, \overleftarrow{C})$ |
| $r$ | $\max(\overrightarrow{C}, \overleftarrow{C})$ | $\overrightarrow{D} + \overleftarrow{D} + \mathtt{VAR}_i$ | $\overrightarrow{D} + \overleftarrow{D} + \mathtt{VAR}_i$ |
| $t$ | $\max(\overrightarrow{C}, \overleftarrow{C})$ | $\overrightarrow{D} + \overleftarrow{D} + \mathtt{VAR}_i$ | $\overrightarrow{D} + \overleftarrow{D} + \mathtt{VAR}_i$ |

Table 3.35: Glue matrix for the MAX_SURF_PLAIN constraint defined as the composition of the PLAIN pattern , the feature SURF , and the aggregator max ; cells of the glue matrix are coloured with the colour of the constituent to which they are related.

## 3.282 MAX_SURF_PLATEAU

▶ ▷ ◁ ◀       DESCRIPTION            AUTOMATON

| | |
|---|---|
| **Origin** | Based on the PLATEAU pattern. |
| **Constraint** | MAX_SURF_PLATEAU(VALUE, VARIABLES) |
| **Arguments** | VALUE       : dvar<br>VARIABLES : collection(var−dvar) |

**Restrictions**

$$\mathtt{sv} \leq 2 \lor \mathtt{rv} \leq 1 \Rightarrow \mathtt{VALUE} = -\infty$$
$$\mathtt{VALUE} = -\infty \lor \mathtt{VALUE} \geq \min(\mathtt{minv} + 1, (\mathtt{sv} - 2) * (\mathtt{minv} + 1))$$
$$\mathtt{VALUE} \leq \max(\mathtt{maxv}①, (\mathtt{sv} - 2) * \mathtt{maxv}②)$$
required(VARIABLES, var)

where

maxv =maxval(VARIABLES.var)
sv = |VARIABLES|
minv =minval(VARIABLES.var)
rv =range(VARIABLES.var)

**Purpose**

VALUE is the maximal surface of occurrences of the PLATEAU pattern in the time-series given by the VARIABLES collection. If the pattern does not occur, VALUE takes the default value $-\infty$.

An occurrence of the pattern PLATEAU is the *maximal* subsequence which matches the regular expression $<=^*>$.

Assume that the occurrence of the pattern PLATEAU starts at position $i$ and ends at position $j$. The feature SURF computes the sum of the values from index $i + 1$ to index $j$.

**Example**

$$(10, \langle 7, 5, 2, 3, 1, 2, 2, 4, 3, 3, 4, 5, 5, 2, 2, 5 \rangle)$$

Figure 3.609 provides an example where the MAX_SURF_PLATEAU $(10, [7, 5, 2, 3, 1, 2, 2, 4, 3, 3, 4, 5, 5, 2, 2, 5])$ constraint holds.

| | |
|---|---|
| **Typical** | $|\mathtt{VARIABLES}| > 2$<br>range(VARIABLES.var) > 1 |
| **Symmetry** | Items of VARIABLES can be reversed. |
| **Arg. properties** | Functional dependency: VALUE determined by VARIABLES. |

Figure 3.609: Illustrating the MAX_SURF_PLATEAU constraint of the **Example** slot

    Figures 3.610 and 3.611 respectively depict the automaton associated with the constraint MAX_SURF_PLATEAU and its simplified form.

Figure 3.610: Automaton for the MAX_SURF_PLATEAU constraint obtained by applying decoration Table 2.35 to the seed transducer of the PLATEAU pattern where default is $-\infty$

Figure 3.611: Automaton for the MAX_SURF_PLATEAU constraint obtained by applying decoration Table 2.25 to the seed transducer of the PLATEAU pattern where default is $-\infty$; $R_i - R_{i-1} \geq 0$ is a linear invariant.

|     | $s$ | $r$ | $t$ |
|-----|-----|-----|-----|
| $s$ | $\max(\overrightarrow{C}, \overleftarrow{C})$ | $\max(\overrightarrow{C}, \overleftarrow{C})$ | $\max(\overrightarrow{C}, \overleftarrow{C})$ |
| $r$ | $\max(\overrightarrow{C}, \overleftarrow{C})$ | $\overrightarrow{D} + \overleftarrow{D} + \text{VAR}_i$ | $\overrightarrow{D} + \overleftarrow{D} + \text{VAR}_i$ |
| $t$ | $\max(\overrightarrow{C}, \overleftarrow{C})$ | $\overrightarrow{D} + \overleftarrow{D} + \text{VAR}_i$ | $\overrightarrow{D} + \overleftarrow{D} + \text{VAR}_i$ |

Table 3.36: Glue matrix for the MAX_SURF_PLATEAU constraint defined as the composition of the PLATEAU pattern , the feature SURF , and the aggregator max ; cells of the glue matrix are coloured with the colour of the constituent to which they are related.

## 3.283  MAX_SURF_PROPER_PLAIN

▶ ▷ ◁ ◀        **DESCRIPTION**        **AUTOMATON**

| | |
|---|---|
| **Origin** | Based on the PROPER_PLAIN pattern. |
| **Constraint** | MAX_SURF_PROPER_PLAIN(VALUE, VARIABLES) |
| **Arguments** | VALUE  : dvar<br>VARIABLES : collection(var−dvar) |

**Restrictions**

$\mathtt{sv} \leq 3 \vee \mathtt{rv} \leq 1 \Rightarrow \mathtt{VALUE} = -\infty$
$\mathtt{VALUE} = -\infty \vee \mathtt{VALUE} \geq \min(2 * \mathtt{minv}, (\mathtt{sv} - 2) * \mathtt{minv})$
$\mathtt{VALUE} \leq \max(2 * (\mathtt{maxv} - 1)①, (\mathtt{sv} - 2) * (\mathtt{maxv} - 1)②)$
required(VARIABLES, var)
where
$\quad \mathtt{maxv} = \mathsf{maxval}(\mathtt{VARIABLES.var})$
$\quad \mathtt{sv} = |\mathtt{VARIABLES}|$
$\quad \mathtt{minv} = \mathsf{minval}(\mathtt{VARIABLES.var})$
$\quad \mathtt{rv} = \mathsf{range}(\mathtt{VARIABLES.var})$

**Purpose**

VALUE is the maximal surface of occurrences of the PROPER_PLAIN pattern in the time-series given by the VARIABLES collection. If the pattern does not occur, VALUE takes the default value $-\infty$.

An occurrence of the pattern PROPER_PLAIN is the *maximal* subsequence which matches the regular expression $> =^+ <$.

Assume that the occurrence of the pattern PROPER_PLAIN starts at position $i$ and ends at position $j$. The feature SURF computes the sum of the values from index $i + 1$ to index $j$.

**Example**

$(10, \langle 2, 7, 5, 5, 6, 3, 7, 4, 4, 5, 6, 5, 3, 3, 3, 5 \rangle)$

Figure 3.612 provides an example where the MAX_SURF_PROPER_PLAIN $(10, [2, 7, 5, 5, 6, 3, 7, 4, 4, 5, 6, 5, 3, 3, 3, 5])$ constraint holds.

| | |
|---|---|
| **Typical** | $\|\mathtt{VARIABLES}\| > 3$<br>range(VARIABLES.var) $> 1$ |
| **Symmetry** | Items of VARIABLES can be reversed. |
| **Arg. properties** | Functional dependency: VALUE determined by VARIABLES. |

Figure 3.612: Illustrating the MAX_SURF_PROPER_PLAIN constraint of the **Example** slot

**Automaton**     Figures 3.613 and 3.614 respectively depict the automaton associated with the constraint
MAX_SURF_PROPER_PLAIN and its simplified form.

Figure 3.613: Automaton for the MAX_SURF_PROPER_PLAIN constraint obtained by
applying decoration Table 2.35 to the seed transducer of the PROPER_PLAIN pattern
where default is $-\infty$

Figure 3.614: Automaton for the MAX_SURF_PROPER_PLAIN constraint obtained by
applying decoration Table 2.25 to the seed transducer of the PROPER_PLAIN pattern
where default is $-\infty$; $R_i - R_{i-1} \geq 0$ is a linear invariant.

| | $s$ | $r$ | $t$ |
|---|---|---|---|
| $s$ | $\max(\overrightarrow{C}, \overleftarrow{C})$ | $\max(\overrightarrow{C}, \overleftarrow{C})$ | $\max(\overrightarrow{C}, \overleftarrow{C})$ |
| $r$ | $\max(\overrightarrow{C}, \overleftarrow{C})$ | $\max(\overrightarrow{C}, \overleftarrow{C})$ | $\overrightarrow{D} + \overleftarrow{D} + \mathtt{VAR}_i$ |
| $t$ | $\max(\overrightarrow{C}, \overleftarrow{C})$ | $\overrightarrow{D} + \overleftarrow{D} + \mathtt{VAR}_i$ | $\overrightarrow{D} + \overleftarrow{D} + \mathtt{VAR}_i$ |

Table 3.37: Glue matrix for the MAX_SURF_PROPER_PLAIN constraint defined as the composition of the PROPER_PLAIN pattern , the feature SURF , and the aggregator max ; cells of the glue matrix are coloured with the colour of the constituent to which they are related.

## 3.284  MAX_SURF_PROPER_PLATEAU

▶ ▷ ◁ ◀          DESCRIPTION          AUTOMATON

| | |
|---|---|
| **Origin** | Based on the PROPER_PLATEAU pattern. |
| **Constraint** | MAX_SURF_PROPER_PLATEAU(VALUE, VARIABLES) |
| **Arguments** | VALUE : dvar<br>VARIABLES : collection(var−dvar) |

**Restrictions**

$\mathtt{sv} \leq 3 \vee \mathtt{rv} \leq 1 \Rightarrow \mathtt{VALUE} = -\infty$

$\mathtt{VALUE} = -\infty \vee \mathtt{VALUE} \geq \min(2 * (\mathtt{minv} + 1), (\mathtt{sv} - 2) * (\mathtt{minv} + 1))$

$\mathtt{VALUE} \leq \max(2 * \mathtt{maxv}①, (\mathtt{sv} - 2) * \mathtt{maxv}②)$

required(VARIABLES, var)

where

maxv =maxval(VARIABLES.var)

sv = |VARIABLES|

minv =minval(VARIABLES.var)

rv =range(VARIABLES.var)

**Purpose**

VALUE is the maximal surface of occurrences of the PROPER_PLATEAU pattern in the time-series given by the VARIABLES collection. If the pattern does not occur, VALUE takes the default value $-\infty$.

An occurrence of the pattern PROPER_PLATEAU is the *maximal* subsequence which matches the regular expression $<=^+>$.

Assume that the occurrence of the pattern PROPER_PLATEAU starts at position $i$ and ends at position $j$. The feature SURF computes the sum of the values from index $i+1$ to index $j$.

**Example**

$(15, \langle 7, 1, 3, 3, 2, 5, 1, 4, 4, 3, 2, 3, 5, 5, 5, 3 \rangle)$

Figure 3.615 provides an example where the MAX_SURF_PROPER_PLATEAU $(15, [7, 1, 3, 3, 2, 5, 1, 4, 4, 3, 2, 3, 5, 5, 5, 3])$ constraint holds.

| | |
|---|---|
| **Typical** | $|\mathtt{VARIABLES}| > 3$<br>range(VARIABLES.var) $> 1$ |
| **Symmetry** | Items of VARIABLES can be reversed. |
| **Arg. properties** | Functional dependency: VALUE determined by VARIABLES. |

Figure 3.615: Illustrating the MAX_SURF_PROPER_PLATEAU constraint of the **Example** slot

**Automaton**   Figures 3.616 and 3.617 respectively depict the automaton associated with the constraint
MAX_SURF_PROPER_PLATEAU and its simplified form.

Figure 3.616: Automaton for the MAX_SURF_PROPER_PLATEAU constraint obtained by
applying decoration Table 2.35 to the seed transducer of the PROPER_PLATEAU pattern
where default is $-\infty$

Figure 3.617: Automaton for the MAX_SURF_PROPER_PLATEAU constraint obtained by
applying decoration Table 2.25 to the seed transducer of the PROPER_PLATEAU pattern
where default is $-\infty$; $R_i - R_{i-1} \geq 0$ is a linear invariant.

|     | $s$ | $r$ | $t$ |
| --- | --- | --- | --- |
| $s$ | $\max(\overrightarrow{C}, \overleftarrow{C})$ | $\max(\overrightarrow{C}, \overleftarrow{C})$ | $\max(\overrightarrow{C}, \overleftarrow{C})$ |
| $r$ | $\max(\overrightarrow{C}, \overleftarrow{C})$ | $\max(\overrightarrow{C}, \overleftarrow{C})$ | $\overrightarrow{D} + \overleftarrow{D} + \text{VAR}_i$ |
| $t$ | $\max(\overrightarrow{C}, \overleftarrow{C})$ | $\overrightarrow{D} + \overleftarrow{D} + \text{VAR}_i$ | $\overrightarrow{D} + \overleftarrow{D} + \text{VAR}_i$ |

Table 3.38: Glue matrix for the MAX_SURF_PROPER_PLATEAU constraint defined as the composition of the PROPER_PLATEAU pattern , the feature SURF , and the aggregator max ; cells of the glue matrix are coloured with the colour of the constituent to which they are related.

## 3.285 MAX_SURF_STEADY

DESCRIPTION            AUTOMATON

| | |
|---|---|
| **Origin** | Based on the STEADY pattern. |
| **Constraint** | MAX_SURF_STEADY(VALUE, VARIABLES) |
| **Arguments** | VALUE      : dvar<br>VARIABLES  : collection(var−dvar) |

**Restrictions**

$\mathtt{sv} \leq 1 \Rightarrow \mathtt{VALUE} = -\infty$

$\mathtt{VALUE} = -\infty \lor \mathtt{VALUE} \geq 2 * \mathtt{minv}$

$\mathtt{VALUE} \leq 2 * \mathtt{maxv} ①$

required(VARIABLES, var)

where

maxv = maxval(VARIABLES.var)

minv = minval(VARIABLES.var)

$\mathtt{sv} = |\mathtt{VARIABLES}|$

**Purpose**

VALUE is the maximal surface of occurrences of the STEADY pattern in the time-series given by the VARIABLES collection. If the pattern does not occur, VALUE takes the default value $-\infty$.

An occurrence of the pattern STEADY is the subsequence which matches the regular expression =.

Assume that the occurrence of the pattern STEADY starts at position $i$ and ends at position $j$. The feature SURF computes the sum of the values from index $i$ to index $j + 1$.

**Example**

$(12, \langle 1, 1, 7, 3, 3, 5, 5, 5, 6, 5, 5, 5, 7, 2, 6, 6 \rangle)$

Figure 3.618 provides an example where the MAX_SURF_STEADY $(12, [1, 1, 7, 3, 3, 5, 5, 5, 6, 5, 5, 5, 7, 2, 6, 6])$ constraint holds.

| | |
|---|---|
| **Typical** | $|\mathtt{VARIABLES}| > 1$ |
| **Symmetry** | Items of VARIABLES can be reversed. |
| **Arg. properties** | Functional dependency: VALUE determined by VARIABLES. |

Figure 3.618: Illustrating the MAX_SURF_STEADY constraint of the **Example** slot

**Automaton**     Figures 3.619 and 3.620 respectively depict the automaton associated with the constraint MAX_SURF_STEADY and its simplified form.

$$\left\{\begin{matrix} D \leftarrow 0 \\ R \leftarrow \max(R, D + \mathtt{VAR}_i + \mathtt{VAR}_{i+1}) \end{matrix}\right\} \stackrel{=}{}$$

$$\left\{\begin{matrix} C \leftarrow \mathtt{default} \\ D \leftarrow 0 \\ R \leftarrow \mathtt{default} \end{matrix}\right\}$$

$$\max(R, C)$$

Figure 3.619: Automaton for the MAX_SURF_STEADY constraint obtained by applying decoration Table 2.35 to the seed transducer of the STEADY pattern where `default` is $-\infty$

$$\{R \leftarrow \max(R, \mathtt{VAR}_i + \mathtt{VAR}_{i+1})\} \stackrel{=}{}$$

$$\{R \leftarrow \mathtt{default}\}$$

$$R$$

Figure 3.620: Automaton for the MAX_SURF_STEADY constraint obtained by applying decoration Table 2.38 to the seed transducer of the STEADY pattern where `default` is $-\infty$; $R_i - R_{i-1} \geq 0$ is a linear invariant.

Table 3.39: Glue matrix for the MAX_SURF_STEADY constraint defined as the composition of the STEADY pattern , the feature SURF , and the aggregator max ; cells of the glue matrix are coloured with the colour of the constituent to which they are related.

## 3.286   MAX_SURF_STEADY_SEQUENCE

► ▷ ◁ ◀        DESCRIPTION              AUTOMATON

**Origin**        Based on the STEADY_SEQUENCE pattern.

**Constraint**    MAX_SURF_STEADY_SEQUENCE(VALUE, VARIABLES)

**Arguments**     VALUE      : dvar
                  VARIABLES  : collection(var−dvar)

**Restrictions**  $\mathtt{sv} \leq 1 \Rightarrow \mathtt{VALUE} = -\infty$
                  $\mathtt{rv} = 1 \Rightarrow \mathtt{VALUE} = -\infty \vee \mathtt{VALUE} \geq \mathtt{sv} * \mathtt{minv}$
                  $\mathtt{rv} \geq 2 \Rightarrow \mathtt{VALUE} = -\infty \vee \mathtt{VALUE} \geq \min(2 * \mathtt{minv}, \mathtt{sv} * \mathtt{minv})$
                  $\mathtt{rv} = 1 \Rightarrow \mathtt{VALUE} \leq \mathtt{sv} * \mathtt{maxv}①$
                  $\mathtt{rv} \geq 2 \Rightarrow \mathtt{VALUE} \leq \max(2 * \mathtt{maxv}②, \mathtt{sv} * \mathtt{maxv}③)$
                  required(VARIABLES, var)
                  where
                  maxv =maxval(VARIABLES.var)
                  rv =range(VARIABLES.var)
                  sv = |VARIABLES|
                  minv =minval(VARIABLES.var)

**Purpose**       VALUE is the maximal surface of occurrences of the STEADY_SEQUENCE pattern in the
                  time-series given by the VARIABLES collection. If the pattern does not occur, VALUE
                  takes the default value $-\infty$.
                  An occurrence of the pattern STEADY_SEQUENCE is the *maximal* subsequence which
                  matches the regular expression $=^+$.
                  Assume that the occurrence of the pattern STEADY_SEQUENCE starts at position $i$ and
                  ends at position $j$. The feature SURF computes the sum of the values from index $i$ to
                  index $j + 1$.

**Example**       $(15, \langle 3, 1, 1, 4, 5, 5, 5, 6, 2, 2, 4, 4, 3, 2, 1, 1 \rangle)$

                  Figure 3.621 provides an example where the MAX_SURF_STEADY_SEQUENCE
                  $(15, [3, 1, 1, 4, 5, 5, 5, 6, 2, 2, 4, 4, 3, 2, 1, 1])$ constraint holds.

**Typical**       $|\mathtt{VARIABLES}| > 1$

**Symmetry**      Items of VARIABLES can be reversed.

**Arg. properties**  Functional dependency: VALUE determined by VARIABLES.

Figure 3.621: Illustrating the MAX_SURF_STEADY_SEQUENCE constraint of the **Example** slot

**Automaton**     Figures 3.622 and 3.623 respectively depict the automaton associated with the constraint
MAX_SURF_STEADY_SEQUENCE and its simplified form.

Figure 3.622: Automaton for the MAX_SURF_STEADY_SEQUENCE constraint obtained
by applying decoration Table 2.35 to the seed transducer of the STEADY_SEQUENCE
pattern where default is $-\infty$

Figure 3.623: Automaton for the MAX_SURF_STEADY_SEQUENCE constraint obtained
by applying decoration Table 2.24 to the seed transducer of the STEADY_SEQUENCE
pattern where default is $-\infty$; $R_i - R_{i-1} \geq 0$ is a linear invariant.

|   | $s$ | $r$ |
|---|---|---|
| $s$ | $\max(\overrightarrow{C}, \overleftarrow{C})$ | $\max(\overrightarrow{C}, \overleftarrow{C})$ |
| $r$ | $\max(\overrightarrow{C}, \overleftarrow{C})$ | $\overrightarrow{C} + \overleftarrow{C} + \overrightarrow{D} + \overleftarrow{D} + \mathtt{VAR}_i$ |

Table 3.40: Glue matrix for the MAX_SURF_STEADY_SEQUENCE constraint defined as the composition of the STEADY_SEQUENCE pattern , the feature SURF , and the aggregator max ; cells of the glue matrix are coloured with the colour of the constituent to which they are related.

# 3.287  MAX_SURF_STRICTLY_DECREASING_SEQUENCE

► ▷ ◁ ◄          **DESCRIPTION**          **AUTOMATON**

**Origin**          Based on the STRICTLY_DECREASING_SEQUENCE pattern.

**Constraint**      MAX_SURF_STRICTLY_DECREASING_SEQUENCE(VALUE, VARIABLES)

**Arguments**       VALUE     : dvar
                    VARIABLES : collection(var−dvar)

**Restrictions**    $\mathtt{sv} \leq 1 \vee \mathtt{rv} \leq 1 \Rightarrow \mathtt{VALUE} = -\infty$

$\mathtt{minv} < 0 \Rightarrow \mathtt{VALUE} = -\infty \vee \mathtt{VALUE} \geq \ell_1 * \mathtt{minv} + \lfloor \ell_1 * (\ell_1 - 1)/2 \rfloor$

$\mathtt{minv} \geq 0 \Rightarrow \mathtt{VALUE} = -\infty \vee \mathtt{VALUE} \geq 2 * \mathtt{minv} + 1$

$\mathtt{maxv} > 0 \Rightarrow \mathtt{VALUE} \leq \ell_2 * \mathtt{maxv} - \lfloor \ell_2 * (\ell_2 - 1)/2 \rfloor$ ①

$\mathtt{maxv} \leq 0 \Rightarrow \mathtt{VALUE} \leq 2 * \mathtt{maxv} - 1$ ②

required(VARIABLES, var)

where

$\mathtt{maxv} = \mathtt{maxval}(\mathtt{VARIABLES.var})$
$\mathtt{rv} = \mathtt{range}(\mathtt{VARIABLES.var})$
$\mathtt{sv} = |\mathtt{VARIABLES}|$
$\ell_1 = \min(\min(\mathtt{sv}, \mathtt{rv}), |\mathtt{minv}|)$
$\ell_2 = \min(\min(\mathtt{sv}, \mathtt{rv}), |\mathtt{maxv}|)$
$\mathtt{minv} = \mathtt{minval}(\mathtt{VARIABLES.var})$

**Purpose**         VALUE is the maximal surface of occurrences of the STRICTLY_DECREASING_SEQUENCE pattern in the time-series given by the VARIABLES collection. If the pattern does not occur, VALUE takes the default value $-\infty$.

An occurrence of the pattern STRICTLY_DECREASING_SEQUENCE is the *maximal* subsequence which matches the regular expression $>^{+}$.

Assume that the occurrence of the pattern STRICTLY_DECREASING_SEQUENCE starts at position $i$ and ends at position $j$. The feature SURF computes the sum of the values from index $i$ to index $j + 1$.

**Example**         $(13, \langle 4, 4, 6, 4, 1, 1, 3, 4, 4, 6, 6, 5, 2, 2, 4, 3 \rangle)$

Figure 3.624 provides an example where the MAX_SURF_STRICTLY_DECREASING_SEQUENCE $(13, [4, 4, 6, 4, 1, 1, 3, 4, 4, 6, 6, 5, 2, 2, 4, 3])$ constraint holds.

**Typical**         $|\mathtt{VARIABLES}| > 1$
                    range(VARIABLES.var) > 1

**Arg. properties** Functional dependency: VALUE determined by VARIABLES.

Figure 3.624: Illustrating the MAX_SURF_STRICTLY_DECREASING_SEQUENCE constraint of the **Example** slot

      Figures 3.625 and 3.626 respectively depict the automaton associated with the constraint MAX_SURF_STRICTLY_DECREASING_SEQUENCE and its simplified form.

Figure 3.625: Automaton for the MAX_SURF_STRICTLY_DECREASING_SEQUENCE constraint obtained by applying decoration Table 2.35 to the seed transducer of the STRICTLY_DECREASING_SEQUENCE pattern where default is $-\infty$

Figure 3.626: Automaton for the MAX_SURF_STRICTLY_DECREASING_SEQUENCE constraint obtained by applying decoration Table 2.24 to the seed transducer of the STRICTLY_DECREASING_SEQUENCE pattern where default is $-\infty$; $R_i - R_{i-1} \geq 0$ is a linear invariant.

|     | $s$ | $r$ |
| --- | --- | --- |
| $s$ | $\max(\overrightarrow{C}, \overleftarrow{C})$ | $\max(\overrightarrow{C}, \overleftarrow{C})$ |
| $r$ | $\max(\overrightarrow{C}, \overleftarrow{C})$ | $\overrightarrow{C} + \overleftarrow{C} + \overrightarrow{D} + \overleftarrow{D} + \texttt{VAR}_i$ |

Table 3.41: Glue matrix for the MAX_SURF_STRICTLY_DECREASING_SEQUENCE constraint defined as the composition of the STRICTLY_DECREASING_SEQUENCE pattern, the feature SURF, and the aggregator max; cells of the glue matrix are coloured with the colour of the constituent to which they are related.

## 3.288  MAX_SURF_STRICTLY_INCREASING_SEQUENCE

▶ ▷ ◁ ◀        **DESCRIPTION**              **AUTOMATON**

**Origin**          Based on the STRICTLY_INCREASING_SEQUENCE pattern.

**Constraint**      MAX_SURF_STRICTLY_INCREASING_SEQUENCE(VALUE, VARIABLES)

**Arguments**       VALUE      : dvar
                    VARIABLES  : collection(var−dvar)

**Restrictions**    $\mathtt{sv} \leq 1 \vee \mathtt{rv} \leq 1 \Rightarrow \mathtt{VALUE} = -\infty$
                    $\mathtt{minv} < 0 \Rightarrow \mathtt{VALUE} = -\infty \vee \mathtt{VALUE} \geq \ell_1 * \mathtt{minv} + \lfloor \ell_1 * (\ell_1 - 1)/2 \rfloor$
                    $\mathtt{minv} \geq 0 \Rightarrow \mathtt{VALUE} = -\infty \vee \mathtt{VALUE} \geq 2 * \mathtt{minv} + 1$
                    $\mathtt{maxv} > 0 \Rightarrow \mathtt{VALUE} \leq \ell_2 * \mathtt{maxv} - \lfloor \ell_2 * (\ell_2 - 1)/2 \rfloor$ ①
                    $\mathtt{maxv} \leq 0 \Rightarrow \mathtt{VALUE} \leq 2 * \mathtt{maxv} - 1$ ②
                    required(VARIABLES, var)
                    where
                    $\mathtt{maxv} = \mathtt{maxval}(\mathtt{VARIABLES.var})$
                    $\mathtt{rv} = \mathtt{range}(\mathtt{VARIABLES.var})$
                    $\mathtt{sv} = |\mathtt{VARIABLES}|$
                    $\ell_1 = \min(\min(\mathtt{sv}, \mathtt{rv}), |\mathtt{minv}|)$
                    $\ell_2 = \min(\min(\mathtt{sv}, \mathtt{rv}), |\mathtt{maxv}|)$
                    $\mathtt{minv} = \mathtt{minval}(\mathtt{VARIABLES.var})$

**Purpose**         VALUE is the maximal surface of occurrences of the STRICTLY_INCREASING_SEQUENCE pattern in the time-series given by the VARIABLES collection. If the pattern does not occur, VALUE takes the default value $-\infty$.
                    An occurrence of the pattern STRICTLY_INCREASING_SEQUENCE is the *maximal* subsequence which matches the regular expression $<^+$.
                    Assume that the occurrence of the pattern STRICTLY_INCREASING_SEQUENCE starts at position $i$ and ends at position $j$. The feature SURF computes the sum of the values from index $i$ to index $j + 1$.

**Example**         $(16, \langle 4, 3, 5, 5, 2, 1, 1, 2, 3, 4, 6, 6, 3, 1, 2, 3 \rangle)$

                    Figure 3.627 provides an example where the MAX_SURF_STRICTLY_INCREASING_SEQUENCE $(16, [4, 3, 5, 5, 2, 1, 1, 2, 3, 4, 6, 6, 3, 1, 2, 3])$ constraint holds.

**Typical**         $|\mathtt{VARIABLES}| > 1$
                    $\mathtt{range}(\mathtt{VARIABLES.var}) > 1$

**Arg. properties** Functional dependency: VALUE determined by VARIABLES.

Figure 3.627: Illustrating the MAX_SURF_STRICTLY_INCREASING_SEQUENCE constraint of the **Example** slot

 Figures 3.628 and 3.629 respectively depict the automaton associated with the constraint MAX_SURF_STRICTLY_INCREASING_SEQUENCE and its simplified form.

Figure 3.628: Automaton for the MAX_SURF_STRICTLY_INCREASING_SEQUENCE constraint obtained by applying decoration Table 2.35 to the seed transducer of the STRICTLY_INCREASING_SEQUENCE pattern where default is $-\infty$

Figure 3.629: Automaton for the MAX_SURF_STRICTLY_INCREASING_SEQUENCE constraint obtained by applying decoration Table 2.24 to the seed transducer of the STRICTLY_INCREASING_SEQUENCE pattern where default is $-\infty$; $R_i - R_{i-1} \geq 0$ is a linear invariant.

|   | $s$ | $r$ |
|---|-----|-----|
| $s$ | $\max(\overrightarrow{C}, \overleftarrow{C})$ | $\max(\overrightarrow{C}, \overleftarrow{C})$ |
| $r$ | $\max(\overrightarrow{C}, \overleftarrow{C})$ | $\overrightarrow{C} + \overleftarrow{C} + \overrightarrow{D} + \overleftarrow{D} + \text{VAR}_i$ |

Table 3.42: Glue matrix for the MAX_SURF_STRICTLY_INCREASING_SEQUENCE constraint defined as the composition of the STRICTLY_INCREASING_SEQUENCE pattern, the feature SURF, and the aggregator max; cells of the glue matrix are coloured with the colour of the constituent to which they are related.

## 3.289  MAX_SURF_SUMMIT

        DESCRIPTION              AUTOMATON

$(< | < (= | <)^* <)(> | > (= | >)^* >)$

| | |
|---|---|
| **Origin** | Based on the SUMMIT pattern. |
| **Constraint** | MAX_SURF_SUMMIT(VALUE, VARIABLES) |
| **Arguments** | VALUE      : dvar<br>VARIABLES : collection(var−dvar) |
| **Restrictions** | $\text{sv} \leq 2 \vee \text{rv} \leq 1 \Rightarrow \text{VALUE} = -\infty$<br>$\text{rv} = 2 \Rightarrow \text{VALUE} = -\infty \vee \text{VALUE} \geq \text{minv} + 1$<br>$\text{rv} \geq 3 \Rightarrow$<br>$\text{VALUE} = -\infty \vee \text{VALUE} \geq \min(\text{minv} + 1, (\text{sv} - 2) * (\text{minv} + 1) + 1)$<br>$\text{rv} = 2 \Rightarrow \text{VALUE} \leq \text{maxv}①$<br>$\text{rv} \geq 3 \Rightarrow \text{VALUE} \leq \max(\text{maxv}②, (\text{sv} - 2) * (\text{maxv} - 1) + 1③)$<br>required(VARIABLES, var)<br>where<br>maxv =maxval(VARIABLES.var)<br>rv =range(VARIABLES.var)<br>sv =\|VARIABLES\|<br>minv =minval(VARIABLES.var) |
| **Purpose** | VALUE is the maximal surface of occurrences of the SUMMIT pattern in the time-series given by the VARIABLES collection. If the pattern does not occur, VALUE takes the default value $-\infty$.<br>An occurrence of the pattern SUMMIT is the *maximal* subsequence which matches the regular expression $(< | < (= | <)^* <)(> | > (= | >)^* >)$.<br>Assume that the occurrence of the pattern SUMMIT starts at position $i$ and ends at position $j$. The feature SURF computes the sum of the values from index $i + 1$ to index $j$. |
| **Example** | $(13, \langle 7, 1, 5, 4, 4, 3, 3, 4, 6, 6, 2, 3, 4, 2, 3, 1 \rangle)$ |

Figure 3.630 provides an example where the MAX_SURF_SUMMIT $(13, [7, 1, 5, 4, 4, 3, 3, 4, 6, 6, 2, 3, 4, 2, 3, 1])$ constraint holds.

| | |
|---|---|
| **Typical** | $\|\text{VARIABLES}\| > 2$<br>range(VARIABLES.var) > 1 |
| **Symmetry** | Items of VARIABLES can be reversed. |
| **Arg. properties** | Functional dependency: VALUE determined by VARIABLES. |

Figure 3.630: Illustrating the MAX_SURF_SUMMIT constraint of the **Example** slot

**Automaton**        Figure 3.631 depicts the automaton associated with the constraint MAX_SURF_SUMMIT.

Figure 3.631: Automaton for the MAX_SURF_SUMMIT constraint obtained by applying decoration Table 2.35 to the seed transducer of the SUMMIT pattern where default is $-\infty$ (transition $u \rightarrow r$ has the same accumulator update as transition $r \rightarrow u$); $R_i - R_{i-1} \geq 0$ is a linear invariant.

| | $s$ | $r$ | $t$ | $u$ |
|---|---|---|---|---|
| $s$ | $\max(\overrightarrow{C}, \overleftarrow{C})$ | $\max(\overrightarrow{C}, \overleftarrow{C})$ | $\max(\overrightarrow{C}, \overleftarrow{C})$ | $\max(\overrightarrow{C}, \overleftarrow{C})$ |
| $r$ | $\max(\overrightarrow{C}, \overleftarrow{C})$ | $\overrightarrow{D} + \overleftarrow{D} + \text{VAR}_i$ | $\overleftarrow{C} + \overrightarrow{D} + \overleftarrow{D} + \text{VAR}_i$ | $\max(\overrightarrow{C}, \overleftarrow{C})$ |
| $t$ | $\max(\overrightarrow{C}, \overleftarrow{C})$ | $\overrightarrow{C} + \overrightarrow{D} + \overleftarrow{D} + \text{VAR}_i$ | $\max(\overrightarrow{C}, \overleftarrow{C})$ | $\overrightarrow{C} + \overrightarrow{D} + \overleftarrow{D} + \text{VAR}_i$ |
| $u$ | $\max(\overrightarrow{C}, \overleftarrow{C})$ | $\max(\overrightarrow{C}, \overleftarrow{C})$ | $\overleftarrow{C} + \overrightarrow{D} + \overleftarrow{D} + \text{VAR}_i$ | $\max(\overrightarrow{C}, \overleftarrow{C})$ |

Table 3.43: Glue matrix for the MAX_SURF_SUMMIT constraint defined as the composition of the SUMMIT pattern , the feature SURF , and the aggregator max ; cells of the glue matrix are coloured with the colour of the constituent to which they are related.

## 3.290   MAX_SURF_VALLEY

▶ ▷ ◁ ◀        DESCRIPTION          AUTOMATON

**Origin**        Based on the VALLEY pattern.

**Constraint**    MAX_SURF_VALLEY(VALUE, VARIABLES)

**Arguments**     VALUE      : dvar
                  VARIABLES  : collection(var−dvar)

**Restrictions**  $\mathtt{sv} \leq 2 \vee \mathtt{rv} \leq 1 \Rightarrow \mathtt{VALUE} = -\infty$
                  $\mathtt{VALUE} = -\infty \vee \mathtt{VALUE} \geq \min(\mathtt{minv}, (\mathtt{sv}-2) * \mathtt{minv})$
                  $\mathtt{VALUE} \leq \max(\mathtt{maxv} - 1①, (\mathtt{sv}-2) * (\mathtt{maxv}-1)②)$
                  required(VARIABLES, var)
                  where
                  $\mathtt{maxv} = \mathtt{maxval}(\mathtt{VARIABLES.var})$
                  $\mathtt{sv} = |\mathtt{VARIABLES}|$
                  $\mathtt{minv} = \mathtt{minval}(\mathtt{VARIABLES.var})$
                  $\mathtt{rv} = \mathtt{range}(\mathtt{VARIABLES.var})$

**Purpose**       VALUE is the maximal surface of occurrences of the VALLEY pattern in the time-series given by the VARIABLES collection. If the pattern does not occur, VALUE takes the default value $-\infty$.
                  An occurrence of the pattern VALLEY is the *maximal* subsequence which matches the regular expression $> (= | >)^* (< | =)^* <$.
                  Assume that the occurrence of the pattern VALLEY starts at position $i$ and ends at position $j$. The feature SURF computes the sum of the values from index $i + 1$ to index $j$.

**Example**       $(15, \langle 1, 3, 7, 4, 3, 6, 6, 5, 3, 3, 2, 6, 5, 5, 5, 7 \rangle)$

                  Figure 3.632 provides an example where the MAX_SURF_VALLEY $(15, [1, 3, 7, 4, 3, 6, 6, 5, 3, 3, 2, 6, 5, 5, 5, 7])$ constraint holds.

**Typical**       $|\mathtt{VARIABLES}| > 2$
                  $\mathtt{range}(\mathtt{VARIABLES.var}) > 1$

**Symmetry**      Items of VARIABLES can be reversed.

**Arg. properties**  Functional dependency: VALUE determined by VARIABLES.

Figure 3.632: Illustrating the MAX_SURF_VALLEY constraint of the **Example** slot

**Automaton**    Figure 3.633 depicts the automaton associated with the constraint MAX_SURF_VALLEY.

Figure 3.633: Automaton for the MAX_SURF_VALLEY constraint obtained by applying decoration Table 2.35 to the seed transducer of the VALLEY pattern where default is $-\infty$; $R_i - R_{i-1} \geq 0$ is a linear invariant.

|   | $s$ | $r$ | $t$ |
|---|---|---|---|
| $s$ | $\max(\overrightarrow{C}, \overleftarrow{C})$ | $\max(\overrightarrow{C}, \overleftarrow{C})$ | $\max(\overrightarrow{C}, \overleftarrow{C})$ |
| $r$ | $\max(\overrightarrow{C}, \overleftarrow{C})$ | $\overrightarrow{D} + \overleftarrow{D} + \text{VAR}_i$ | $\overleftarrow{C} + \overrightarrow{D} + \overleftarrow{D} + \text{VAR}_i$ |
| $t$ | $\max(\overrightarrow{C}, \overleftarrow{C})$ | $\overrightarrow{C} + \overrightarrow{D} + \overleftarrow{D} + \text{VAR}_i$ | $\max(\overrightarrow{C}, \overleftarrow{C})$ |

Table 3.44: Glue matrix for the MAX_SURF_VALLEY constraint defined as the composition of the VALLEY pattern , the feature SURF , and the aggregator max ; cells of the glue matrix are coloured with the colour of the constituent to which they are related.

## 3.291 MAX_SURF_ZIGZAG

▶ ▷ ◁ ◀      DESCRIPTION          AUTOMATON

$(<>)^+(<|<>)|(><)^+(>|><)$

| | | |
|---|---|---|
| **Origin** | Based on the ZIGZAG pattern. | |
| **Constraint** | MAX_SURF_ZIGZAG(VALUE, VARIABLES) | |
| **Arguments** | VALUE       : dvar<br>VARIABLES : collection(var−dvar) | |

**Restrictions**

$\mathtt{sv} \leq 3 \vee \mathtt{rv} \leq 1 \Rightarrow \mathtt{VALUE} = -\infty$

$\bigvee \left( \begin{array}{l} \mathtt{VALUE} = -\infty, \\ \mathtt{VALUE} \geq \min \left( \begin{array}{l} 2 * \mathtt{minv} + 1, \\ \lfloor (\mathtt{sv}-1)/2 \rfloor * \mathtt{minv} + \lfloor (\mathtt{sv}-2)/2 \rfloor * (\mathtt{minv}+1) \end{array} \right) \end{array} \right)$

$\mathtt{VALUE} \leq \max \left( \begin{array}{l} 2 * \mathtt{maxv} - 1①, \\ \lfloor (\mathtt{sv}-1)/2 \rfloor * \mathtt{maxv} + \lfloor (\mathtt{sv}-2)/2 \rfloor * (\mathtt{maxv}-1)② \end{array} \right)$

required(VARIABLES, var)

where

minv = minval(VARIABLES.var)
maxv = maxval(VARIABLES.var)
sv = |VARIABLES|
rv = range(VARIABLES.var)

**Purpose**

VALUE is the maximal surface of occurrences of the ZIGZAG pattern in the time-series given by the VARIABLES collection. If the pattern does not occur, VALUE takes the default value $-\infty$.

An occurrence of the pattern ZIGZAG is the *maximal* subsequence which matches the regular expression $(<>)^+(<|<>)|(><)^+(>|><)$.

Assume that the occurrence of the pattern ZIGZAG starts at position $i$ and ends at position $j$. The feature SURF computes the sum of the values from index $i+1$ to index $j$.

**Example**

$(21, \langle 4, 1, 3, 1, 4, 6, 1, 5, 5, 2, 7, 2, 3, 1, 6, 1 \rangle)$

Figure 3.634 provides an example where the MAX_SURF_ZIGZAG $(21, [4, 1, 3, 1, 4, 6, 1, 5, 5, 2, 7, 2, 3, 1, 6, 1])$ constraint holds.

| | |
|---|---|
| **Typical** | $|\mathtt{VARIABLES}| > 3$<br>range(VARIABLES.var) $> 1$ |
| **Symmetry** | Items of VARIABLES can be reversed. |
| **Arg. properties** | Functional dependency: VALUE determined by VARIABLES. |

Figure 3.634: Illustrating the MAX_SURF_ZIGZAG constraint of the **Example** slot

**Automaton**          Figures 3.635 and 3.636 respectively depict the automaton associated with the constraint
                       MAX_SURF_ZIGZAG and its simplified form.

Figure 3.635: Automaton for the MAX_SURF_ZIGZAG constraint obtained by applying decoration Table 2.35 to the seed transducer of the ZIGZAG pattern where default is $-\infty$; (1) missing transitions from $a, b, c, d, e, f$ to $s$ are labelled by $=$; (2) on transitions from $b, c, e, f$ to $s$ the accumulator $D$ is reset to its initial value; (3) on transitions from $c, f$ to $s$ the accumulator $R$ is updated wrt $C$ and the accumulator $C$ is reset to its initial value

Figure 3.636: Automaton for the MAX_SURF_ZIGZAG constraint obtained by applying decoration Table 2.23 to the seed transducer of the ZIGZAG pattern where default is $-\infty$; missing transitions from $a, b, c, d, e, f$ to $s$ are labelled by $=$; (2) on transitions from $b, c, e, f$ to $s$ the accumulator $D$ is reset to its initial value; (3) on transitions from $c, f$ to $s$ the accumulator $R$ is updated wrt $C$ and the accumulator $C$ is reset to its initial value.; $R_i - R_{i-1} \geq 0$ is a linear invariant.

|   | $s$ | $a$ | $b$ | $c$ | $d$ | $e$ | $f$ |
|---|---|---|---|---|---|---|---|
| $s$ | $\max(\overrightarrow{\mathcal{C}},\overleftarrow{\mathcal{C}})$ | $\max(\overrightarrow{\mathcal{C}},\overleftarrow{\mathcal{C}})$ | $\max(\overrightarrow{\mathcal{C}},\overleftarrow{\mathcal{C}})$ | $\max(\overrightarrow{\mathcal{C}},\overleftarrow{\mathcal{C}})$ | $\max(\overrightarrow{\mathcal{C}},\overleftarrow{\mathcal{C}})$ | $\max(\overrightarrow{\mathcal{C}},\overleftarrow{\mathcal{C}})$ | $\max(\overrightarrow{\mathcal{C}},\overleftarrow{\mathcal{C}})$ |
| $a$ | $\max(\overrightarrow{\mathcal{C}},\overleftarrow{\mathcal{C}})$ | $\max(\overrightarrow{\mathcal{C}},\overleftarrow{\mathcal{C}})$ | $\max(\overrightarrow{\mathcal{C}},\overleftarrow{\mathcal{C}})$ | $\overleftarrow{\mathcal{C}}+\overrightarrow{\mathcal{D}}+\overleftarrow{\mathcal{D}}+\mathtt{VAR}_i$ | $\max(\overrightarrow{\mathcal{C}},\overleftarrow{\mathcal{C}})$ | $\overrightarrow{\mathcal{D}}+\overleftarrow{\mathcal{D}}+\mathtt{VAR}_i$ | $\max(\overrightarrow{\mathcal{C}},\overleftarrow{\mathcal{C}})$ |
| $b$ | $\max(\overrightarrow{\mathcal{C}},\overleftarrow{\mathcal{C}})$ | $\max(\overrightarrow{\mathcal{C}},\overleftarrow{\mathcal{C}})$ | $\overrightarrow{\mathcal{D}}+\overleftarrow{\mathcal{D}}+\mathtt{VAR}_i$ | $\max(\overrightarrow{\mathcal{C}},\overleftarrow{\mathcal{C}})$ | $\overrightarrow{\mathcal{D}}+\overleftarrow{\mathcal{D}}+\mathtt{VAR}_i$ | $\max(\overrightarrow{\mathcal{C}},\overleftarrow{\mathcal{C}})$ | $\overleftarrow{\mathcal{C}}+\overrightarrow{\mathcal{D}}+\overleftarrow{\mathcal{D}}+\mathtt{VAR}_i$ |
| $c$ | $\max(\overrightarrow{\mathcal{C}},\overleftarrow{\mathcal{C}})$ | $\overrightarrow{\mathcal{C}}+\overrightarrow{\mathcal{D}}+\overleftarrow{\mathcal{D}}+\mathtt{VAR}_i$ | $\max(\overrightarrow{\mathcal{C}},\overleftarrow{\mathcal{C}})$ | $\overrightarrow{\mathcal{C}}+\overleftarrow{\mathcal{C}}+\overrightarrow{\mathcal{D}}+\overleftarrow{\mathcal{D}}+\mathtt{VAR}_i$ | $\max(\overrightarrow{\mathcal{C}},\overleftarrow{\mathcal{C}})$ | $\overleftarrow{\mathcal{C}}+\overrightarrow{\mathcal{D}}+\overleftarrow{\mathcal{D}}+\mathtt{VAR}_i$ | $\max(\overrightarrow{\mathcal{C}},\overleftarrow{\mathcal{C}})$ |
| $d$ | $\max(\overrightarrow{\mathcal{C}},\overleftarrow{\mathcal{C}})$ | $\max(\overrightarrow{\mathcal{C}},\overleftarrow{\mathcal{C}})$ | $\overrightarrow{\mathcal{D}}+\overleftarrow{\mathcal{D}}+\mathtt{VAR}_i$ | $\max(\overrightarrow{\mathcal{C}},\overleftarrow{\mathcal{C}})$ | $\max(\overrightarrow{\mathcal{C}},\overleftarrow{\mathcal{C}})$ | $\max(\overrightarrow{\mathcal{C}},\overleftarrow{\mathcal{C}})$ | $\overleftarrow{\mathcal{C}}+\overrightarrow{\mathcal{D}}+\overleftarrow{\mathcal{D}}+\mathtt{VAR}_i$ |
| $e$ | $\max(\overrightarrow{\mathcal{C}},\overleftarrow{\mathcal{C}})$ | $\overrightarrow{\mathcal{D}}+\overleftarrow{\mathcal{D}}+\mathtt{VAR}_i$ | $\max(\overrightarrow{\mathcal{C}},\overleftarrow{\mathcal{C}})$ | $\overleftarrow{\mathcal{C}}+\overrightarrow{\mathcal{D}}+\overleftarrow{\mathcal{D}}+\mathtt{VAR}_i$ | $\max(\overrightarrow{\mathcal{C}},\overleftarrow{\mathcal{C}})$ | $\overrightarrow{\mathcal{D}}+\overleftarrow{\mathcal{D}}+\mathtt{VAR}_i$ | $\max(\overrightarrow{\mathcal{C}},\overleftarrow{\mathcal{C}})$ |
| $f$ | $\max(\overrightarrow{\mathcal{C}},\overleftarrow{\mathcal{C}})$ | $\max(\overrightarrow{\mathcal{C}},\overleftarrow{\mathcal{C}})$ | $\overrightarrow{\mathcal{C}}+\overrightarrow{\mathcal{D}}+\overleftarrow{\mathcal{D}}+\mathtt{VAR}_i$ | $\max(\overrightarrow{\mathcal{C}},\overleftarrow{\mathcal{C}})$ | $\overrightarrow{\mathcal{C}}+\overrightarrow{\mathcal{D}}+\overleftarrow{\mathcal{D}}+\mathtt{VAR}_i$ | $\max(\overrightarrow{\mathcal{C}},\overleftarrow{\mathcal{C}})$ | $\overrightarrow{\mathcal{C}}+\overleftarrow{\mathcal{C}}+\overrightarrow{\mathcal{D}}+\overleftarrow{\mathcal{D}}+\mathtt{VAR}_i$ |

Table 3.45: Glue matrix for the MAX_SURF_ZIGZAG constraint defined as the composition of the ZIGZAG pattern, the feature SURF, and the aggregator max; cells of the glue matrix are coloured with the colour of the constituent to which they are related.

# 3.292   MAX_WIDTH_DECREASING_SEQUENCE

▶ ▷ ◁ ◀        **DESCRIPTION**        **AUTOMATON**

| | |
|---|---|
| **Origin** | Based on the DECREASING_SEQUENCE pattern. |
| **Constraint** | MAX_WIDTH_DECREASING_SEQUENCE(VALUE, VARIABLES) |
| **Arguments** | VALUE      : dvar |
| | VARIABLES  : collection(var−dvar) |

**Restrictions**

$\mathtt{sv} \leq 1 \vee \mathtt{rv} \leq 1 \Rightarrow \mathtt{VALUE} = 0$
$\mathtt{VALUE} = 0 \vee \mathtt{VALUE} \geq 2$
$\mathtt{rv} = 2 \Rightarrow \mathtt{VALUE} \leq 2①$
$\mathtt{rv} \geq 3 \Rightarrow \mathtt{VALUE} \leq \mathtt{sv}②$
required(VARIABLES, var)
where
  $\mathtt{sv} = |\mathtt{VARIABLES}|$
  $\mathtt{rv} = $ range(VARIABLES.var)

**Purpose**

VALUE is the maximal width of occurrences of the DECREASING_SEQUENCE pattern in the time-series given by the VARIABLES collection. If the pattern does not occur, VALUE takes the default value 0.
An occurrence of the pattern DECREASING_SEQUENCE is the *maximal* subsequence which matches the regular expression $> (> | =)^* > | >$.
Assume that the occurrence of the pattern DECREASING_SEQUENCE starts at position $i$ and ends at position $j$. The feature WIDTH computes the value $j - i + 2$.

**Example**

$(5, \langle 3, 4, 2, 2, 5, 6, 6, 4, 4, 3, 1, 1, 4, 6, 4, 4 \rangle)$

Figure 3.637 provides an example where the MAX_WIDTH_DECREASING_SEQUENCE $(5, [3, 4, 2, 2, 5, 6, 6, 4, 4, 3, 1, 1, 4, 6, 4, 4])$ constraint holds.

**Typical**

$|\mathtt{VARIABLES}| > 1$
range(VARIABLES.var) $> 1$

**Symmetry**

One and the same constant can be added to the var attribute of all items of VARIABLES.

**Arg. properties**

Functional dependency: VALUE determined by VARIABLES.

Figure 3.637: Illustrating the MAX_WIDTH_DECREASING_SEQUENCE constraint of the **Example** slot

 Figures 3.638 and 3.639 respectively depict the automaton associated with the constraint MAX_WIDTH_DECREASING_SEQUENCE and its simplified form.

Figure 3.638: Automaton for the MAX_WIDTH_DECREASING_SEQUENCE constraint obtained by applying decoration Table 2.35 to the seed transducer of the DECREAS­ING_SEQUENCE pattern where default is $0$

| | $s$ | $t$ |
|---|---|---|
| $s$ | $\max(\overrightarrow{C}, \overleftarrow{C})$ | $\max(\overrightarrow{C}, \overleftarrow{C})$ |
| $t$ | $\max(\overrightarrow{C}, \overleftarrow{C})$ | $\overrightarrow{C} + \overleftarrow{C} + \overrightarrow{D} + \overleftarrow{D} + 1$ |

Table 3.46: Glue matrix for the MAX_WIDTH_DECREASING_SEQUENCE con­straint defined as the composition of the DECREASING_SEQUENCE pattern, the feature WIDTH, and the aggregator max; cells of the glue matrix are coloured with the colour of the constituent to which they are related.

Figure 3.639: Automaton for the MAX_WIDTH_DECREASING_SEQUENCE constraint obtained by applying decoration Table 2.30 to the seed transducer of the DECREAS-ING_SEQUENCE pattern where default is 0; $R_i - R_{i-1} \geq 0$ and $-R_i + R_{i-1} + D_i \geq 0$ are linear invariants.

## 3.293   MAX_WIDTH_DECREASING_TERRACE

   **DESCRIPTION**        **AUTOMATON**

| | |
|---|---|
| **Origin** | Based on the DECREASING_TERRACE pattern. |
| **Constraint** | MAX_WIDTH_DECREASING_TERRACE(VALUE, VARIABLES) |
| **Arguments** | VALUE      : dvar<br>VARIABLES : collection(var−dvar) |

**Restrictions**

$\mathtt{sv} \leq 3 \vee \mathtt{rv} \leq 2 \Rightarrow \mathtt{VALUE} = 0$

$\mathtt{VALUE} = 0 \vee \mathtt{VALUE} \geq 2$

$\mathtt{VALUE} \leq \max(0, \mathtt{sv} - 2 ①)$

required(VARIABLES, var)

where

$\mathtt{sv} = |\mathtt{VARIABLES}|$

$\mathtt{rv} = \mathtt{range}(\mathtt{VARIABLES.var})$

**Purpose**

VALUE is the maximal width of occurrences of the DECREASING_TERRACE pattern in the time-series given by the VARIABLES collection. If the pattern does not occur, VALUE takes the default value 0.

An occurrence of the pattern DECREASING_TERRACE is the *maximal* subsequence which matches the regular expression $> =^+ >$.

Assume that the occurrence of the pattern DECREASING_TERRACE starts at position $i$ and ends at position $j$. The feature WIDTH computes the value $j - i$.

**Example**

$(2, \langle 6, 4, 4, 4, 5, 2, 2, 1, 3, 3, 5, 4, 4, 3, 3, 3 \rangle)$

Figure 3.640 provides an example where the MAX_WIDTH_DECREASING_TERRACE $(2, [6, 4, 4, 4, 5, 2, 2, 1, 3, 3, 5, 4, 4, 3, 3, 3])$ constraint holds.

**Typical**

$|\mathtt{VARIABLES}| > 3$

range(VARIABLES.var) $> 2$

**Symmetry**

One and the same constant can be added to the var attribute of all items of VARIABLES.

**Arg. properties**

Functional dependency: VALUE determined by VARIABLES.

Figure 3.640: Illustrating the MAX_WIDTH_DECREASING_TERRACE constraint of the **Example** slot

**Automaton**    Figures 3.641 and 3.642 respectively depict the automaton associated with the constraint MAX_WIDTH_DECREASING_TERRACE and its simplified form.

Figure 3.641: Automaton for the MAX_WIDTH_DECREASING_TERRACE constraint obtained by applying decoration Table 2.35 to the seed transducer of the DECREASING_TERRACE pattern where default is 0

| | $s$ | $r$ | $t$ |
|---|---|---|---|
| $s$ | $\max(\overrightarrow{C}, \overleftarrow{C})$ | $\max(\overrightarrow{C}, \overleftarrow{C})$ | $\max(\overrightarrow{C}, \overleftarrow{C})$ |
| $r$ | $\max(\overrightarrow{C}, \overleftarrow{C})$ | $\max(\overrightarrow{C}, \overleftarrow{C})$ | $\overrightarrow{D} + \overleftarrow{D} + 1$ |
| $t$ | $\max(\overrightarrow{C}, \overleftarrow{C})$ | $\overrightarrow{D} + \overleftarrow{D} + 1$ | $\overrightarrow{D} + \overleftarrow{D} + 1$ |

Table 3.47: Glue matrix for the MAX_WIDTH_DECREASING_TERRACE constraint defined as the composition of the DECREASING_TERRACE pattern , the feature WIDTH , and the aggregator max ; cells of the glue matrix are coloured with the colour of the constituent to which they are related.

Figure 3.642: Automaton for the MAX_WIDTH_DECREASING_TERRACE constraint obtained by applying decoration Table 2.25 to the seed transducer of the DECREASING_TERRACE pattern where default is 0; $R_i - R_{i-1} \geq 0$ and $-R_i + R_{i-1} + D_i \geq 0$ are linear invariants.

## 3.294  MAX_WIDTH_GORGE

▶ ▷ ◁ ◀        **DESCRIPTION**        **AUTOMATON**

$(> | > (= | >)^* >)(< | < (= | <)^* <)$

| | |
|---|---|
| **Origin** | Based on the GORGE pattern. |
| **Constraint** | MAX_WIDTH_GORGE(VALUE, VARIABLES) |
| **Arguments** | VALUE      : dvar |
| | VARIABLES : collection(var−dvar) |

**Restrictions**

$\mathtt{sv} \leq 2 \vee \mathtt{rv} \leq 1 \Rightarrow \mathtt{VALUE} = 0$
$\mathtt{VALUE} = 0 \vee \mathtt{VALUE} \geq 1$
$\mathtt{rv} = 2 \Rightarrow \mathtt{VALUE} \leq 1 ①$
$\mathtt{rv} \geq 3 \Rightarrow \mathtt{VALUE} \leq \max(0, \mathtt{sv} - 2②)$
required(VARIABLES, var)
where
$\mathtt{sv} = |\mathtt{VARIABLES}|$
$\mathtt{rv} = \text{range}(\mathtt{VARIABLES.var})$

**Purpose**

VALUE is the maximal width of occurrences of the GORGE pattern in the time-series given by the VARIABLES collection. If the pattern does not occur, VALUE takes the default value 0.

An occurrence of the pattern GORGE is the *maximal* subsequence which matches the regular expression $(> | > (= | >)^* >)(< | < (= | <)^* <)$.

Assume that the occurrence of the pattern GORGE starts at position $i$ and ends at position $j$. The feature WIDTH computes the value $j - i$.

| | |
|---|---|
| **Example** | $(3, \langle 1, 7, 3, 4, 4, 5, 5, 4, 2, 2, 6, 5, 4, 6, 5, 7 \rangle)$ |

Figure 3.643 provides an example where the MAX_WIDTH_GORGE $(3, [1, 7, 3, 4, 4, 5, 5, 4, 2, 2, 6, 5, 4, 6, 5, 7])$ constraint holds.

| | |
|---|---|
| **Typical** | $|\mathtt{VARIABLES}| > 2$ |
| | range(VARIABLES.var) > 1 |

**Symmetries**

- Items of VARIABLES can be reversed.
- One and the same constant can be added to the var attribute of all items of VARIABLES.

| | |
|---|---|
| **Arg. properties** | Functional dependency: VALUE determined by VARIABLES. |

Figure 3.643: Illustrating the MAX_WIDTH_GORGE constraint of the **Example** slot

**Automaton**     Figures 3.644 and 3.645 respectively depict the automaton associated with the constraint
MAX_WIDTH_GORGE and its simplified form.

Figure 3.644: Automaton for the MAX_WIDTH_GORGE constraint obtained by applying
decoration Table 2.35 to the seed transducer of the GORGE pattern where default is 0
(transition $u \rightarrow r$ has the same accumulator update as transition $r \rightarrow u$)

Figure 3.645: Automaton for the MAX_WIDTH_GORGE constraint obtained by applying decoration Table 2.26 to the seed transducer of the GORGE pattern where `default` is $0$ (transition $u \rightarrow r$ has the same accumulator update as transition $r \rightarrow u$); $R_i - R_{i-1} \geq 0$ and $-R_i + R_{i-1} + D_i \geq 0$ are linear invariants.

|   | $s$ | $r$ | $t$ | $u$ |
|---|---|---|---|---|
| $s$ | $\max(\overrightarrow{C}, \overleftarrow{C})$ | $\max(\overrightarrow{C}, \overleftarrow{C})$ | $\max(\overrightarrow{C}, \overleftarrow{C})$ | $\max(\overrightarrow{C}, \overleftarrow{C})$ |
| $r$ | $\max(\overrightarrow{C}, \overleftarrow{C})$ | $\overrightarrow{D} + \overleftarrow{D} + 1$ | $\overleftarrow{C} + \overrightarrow{D} + \overleftarrow{D} + 1$ | $\max(\overrightarrow{C}, \overleftarrow{C})$ |
| $t$ | $\max(\overrightarrow{C}, \overleftarrow{C})$ | $\overrightarrow{C} + \overrightarrow{D} + \overleftarrow{D} + 1$ | $\max(\overrightarrow{C}, \overleftarrow{C})$ | $\overrightarrow{C} + \overrightarrow{D} + \overleftarrow{D} + 1$ |
| $u$ | $\max(\overrightarrow{C}, \overleftarrow{C})$ | $\max(\overrightarrow{C}, \overleftarrow{C})$ | $\overleftarrow{C} + \overrightarrow{D} + \overleftarrow{D} + 1$ | $\max(\overrightarrow{C}, \overleftarrow{C})$ |

Table 3.48: Glue matrix for the MAX_WIDTH_GORGE constraint defined as the composition of the GORGE pattern , the feature WIDTH , and the aggregator max ; cells of the glue matrix are coloured with the colour of the constituent to which they are related.

## 3.295   MAX_WIDTH_INCREASING_SEQUENCE

► ▷ ◁ ◀        DESCRIPTION              AUTOMATON

| | |
|---|---|
| **Origin** | Based on the INCREASING_SEQUENCE pattern. |
| **Constraint** | MAX_WIDTH_INCREASING_SEQUENCE(VALUE, VARIABLES) |
| **Arguments** | VALUE     : dvar |
| | VARIABLES : collection(var−dvar) |
| **Restrictions** | $\mathtt{sv} \leq 1 \vee \mathtt{rv} \leq 1 \Rightarrow \mathtt{VALUE} = 0$ |
| | $\mathtt{VALUE} = 0 \vee \mathtt{VALUE} \geq 2$ |
| | $\mathtt{rv} = 2 \Rightarrow \mathtt{VALUE} \leq 2①$ |
| | $\mathtt{rv} \geq 3 \Rightarrow \mathtt{VALUE} \leq \mathtt{sv}②$ |
| | required(VARIABLES, var) |
| | where |
| | $\mathtt{sv} = \|\mathtt{VARIABLES}\|$ |
| | $\mathtt{rv} =$ range(VARIABLES.var) |
| **Purpose** | VALUE is the maximal width of occurrences of the INCREASING_SEQUENCE pattern in the time-series given by the VARIABLES collection. If the pattern does not occur, VALUE takes the default value 0. |
| | An occurrence of the pattern INCREASING_SEQUENCE is the *maximal* subsequence which matches the regular expression $< (< | =)^* < | <$. |
| | Assume that the occurrence of the pattern INCREASING_SEQUENCE starts at position $i$ and ends at position $j$. The feature WIDTH computes the value $j - i + 2$. |
| **Example** | $(5, \langle 4, 3, 5, 5, 2, 1, 1, 3, 3, 4, 6, 6, 3, 1, 3, 3 \rangle)$ |

Figure 3.646 provides an example where the MAX_WIDTH_INCREASING_SEQUENCE $(5, [4, 3, 5, 5, 2, 1, 1, 3, 3, 4, 6, 6, 3, 1, 3, 3])$ constraint holds.

| | |
|---|---|
| **Typical** | $\|\mathtt{VARIABLES}\| > 1$ |
| | range(VARIABLES.var) $> 1$ |
| **Symmetry** | One and the same constant can be added to the var attribute of all items of VARIABLES. |
| **Arg. properties** | Functional dependency: VALUE determined by VARIABLES. |

Figure 3.646: Illustrating the MAX_WIDTH_INCREASING_SEQUENCE constraint of the **Example** slot

**Automaton**  Figures 3.647 and 3.648 respectively depict the automaton associated with the constraint MAX_WIDTH_INCREASING_SEQUENCE and its simplified form.

Figure 3.647: Automaton for the MAX_WIDTH_INCREASING_SEQUENCE constraint obtained by applying decoration Table 2.35 to the seed transducer of the INCREASING_SEQUENCE pattern where default is 0

| | $s$ | $t$ |
|---|---|---|
| $s$ | $\max(\overrightarrow{C}, \overleftarrow{C})$ | $\max(\overrightarrow{C}, \overleftarrow{C})$ |
| $t$ | $\max(\overrightarrow{C}, \overleftarrow{C})$ | $\overrightarrow{C} + \overleftarrow{C} + \overrightarrow{D} + \overleftarrow{D} + 1$ |

Table 3.49: Glue matrix for the MAX_WIDTH_INCREASING_SEQUENCE constraint defined as the composition of the INCREASING_SEQUENCE pattern , the feature WIDTH , and the aggregator max ; cells of the glue matrix are coloured with the colour of the constituent to which they are related.

Figure 3.648: Automaton for the MAX_WIDTH_INCREASING_SEQUENCE constraint obtained by applying decoration Table 2.30 to the seed transducer of the INCREAS­ING_SEQUENCE pattern where default is 0; $R_i - R_{i-1} \geq 0$ and $-R_i + R_{i-1} + D_i \geq 0$ are linear invariants.

# 3.296 MAX_WIDTH_INCREASING_TERRACE

▶ ▷ ◁ ◀     DESCRIPTION          AUTOMATON

**Origin**          Based on the INCREASING_TERRACE pattern.

**Constraint**      MAX_WIDTH_INCREASING_TERRACE(VALUE, VARIABLES)

**Arguments**       VALUE      :  dvar
                    VARIABLES  :  collection(var−dvar)

**Restrictions**    $\mathtt{sv} \leq 3 \vee \mathtt{rv} \leq 2 \Rightarrow \mathtt{VALUE} = 0$
                    $\mathtt{VALUE} = 0 \vee \mathtt{VALUE} \geq 2$
                    $\mathtt{VALUE} \leq \max(0, \mathtt{sv} - 2①)$
                    required(VARIABLES, var)
                    where
                    $\mathtt{sv} = |\mathtt{VARIABLES}|$
                    $\mathtt{rv} = \mathtt{range}(\mathtt{VARIABLES.var})$

**Purpose**         VALUE is the maximal width of occurrences of the INCREASING_TERRACE pattern in the
                    time-series given by the VARIABLES collection. If the pattern does not occur, VALUE
                    takes the default value 0.
                    An occurrence of the pattern INCREASING_TERRACE is the *maximal* subsequence which
                    matches the regular expression $< =^+ <$.
                    Assume that the occurrence of the pattern INCREASING_TERRACE starts at position $i$
                    and ends at position $j$. The feature WIDTH computes the value $j - i$.

**Example**         $(3, \langle 1, 3, 3, 3, 2, 5, 5, 6, 4, 4, 2, 3, 3, 3, 4, 4 \rangle)$

                    Figure 3.649 provides an example where the MAX_WIDTH_INCREASING_TERRACE
                    $(3, [1, 3, 3, 3, 2, 5, 5, 6, 4, 4, 2, 3, 3, 3, 4, 4])$ constraint holds.

**Typical**         $|\mathtt{VARIABLES}| > 3$
                    $\mathtt{range}(\mathtt{VARIABLES.var}) > 2$

**Symmetry**        One and the same constant can be added to the var attribute of all items of VARIABLES.

**Arg. properties** Functional dependency: VALUE determined by VARIABLES.

Figure 3.649: Illustrating the MAX_WIDTH_INCREASING_TERRACE constraint of the **Example** slot

**Automaton**    Figures 3.650 and 3.651 respectively depict the automaton associated with the constraint
MAX_WIDTH_INCREASING_TERRACE and its simplified form.

Figure 3.650: Automaton for the MAX_WIDTH_INCREASING_TERRACE constraint obtained by applying decoration Table 2.35 to the seed transducer of the INCREASING_TERRACE pattern where default is 0

|   | $s$ | $r$ | $t$ |
|---|-----|-----|-----|
| $s$ | $\max(\overrightarrow{C}, \overleftarrow{C})$ | $\max(\overrightarrow{C}, \overleftarrow{C})$ | $\max(\overrightarrow{C}, \overleftarrow{C})$ |
| $r$ | $\max(\overrightarrow{C}, \overleftarrow{C})$ | $\max(\overrightarrow{C}, \overleftarrow{C})$ | $\overrightarrow{D} + \overleftarrow{D} + 1$ |
| $t$ | $\max(\overrightarrow{C}, \overleftarrow{C})$ | $\overrightarrow{D} + \overleftarrow{D} + 1$ | $\overrightarrow{D} + \overleftarrow{D} + 1$ |

Table 3.50: Glue matrix for the MAX_WIDTH_INCREASING_TERRACE constraint defined as the composition of the INCREASING_TERRACE pattern , the feature WIDTH , and the aggregator max ; cells of the glue matrix are coloured with the colour of the constituent to which they are related.

Figure 3.651: Automaton for the MAX_WIDTH_INCREASING_TERRACE constraint obtained by applying decoration Table 2.25 to the seed transducer of the INCREASING_TERRACE pattern where default is $0$; $R_i - R_{i-1} \geq 0$ and $-R_i + R_{i-1} + D_i \geq 0$ are linear invariants.

PL

## 3.297  MAX_WIDTH_INFLEXION

▶ ▷ ◁ ◀          DESCRIPTION          AUTOMATON

$< (< | =)^* > | > (> | =)^* <$

| | |
|---|---|
| **Origin** | Based on the INFLEXION pattern. |
| **Constraint** | MAX_WIDTH_INFLEXION(VALUE, VARIABLES) |

| **Arguments** | |
|---|---|
| | VALUE      : dvar |
| | VARIABLES : collection(var−dvar) |

**Restrictions**

$\mathtt{sv} \leq 2 \lor \mathtt{rv} \leq 1 \Rightarrow \mathtt{VALUE} = 0$
$\mathtt{VALUE} \geq 0$
$\mathtt{VALUE} \leq \max(0, \mathtt{sv} - 2①)$
required(VARIABLES, var)
where
$\mathtt{sv} = |\mathtt{VARIABLES}|$
$\mathtt{rv} = \mathtt{range}(\mathtt{VARIABLES.var})$

VALUE ≤ **4** = 6 − 2

VALUE ≤ **4** = 6 − 2

**Purpose**

VALUE is the maximal width of occurrences of the INFLEXION pattern in the time-series given by the VARIABLES collection. If the pattern does not occur, VALUE takes the default value 0.

An occurrence of the pattern INFLEXION is the *maximal* subsequence which matches the regular expression $< (< | =)^* > | > (> | =)^* <$.

Assume that the occurrence of the pattern INFLEXION starts at position $i$ and ends at position $j$. The feature WIDTH computes the value $j - i$.

**Example**

$(3, \langle 1, 2, 6, 6, 4, 4, 3, 5, 2, 5, 1, 5, 3, 3, 4, 4 \rangle)$

Figure 3.652 provides an example where the MAX_WIDTH_INFLEXION $(3, [1, 2, 6, 6, 4, 4, 3, 5, 2, 5, 1, 5, 3, 3, 4, 4])$ constraint holds.

| **Typical** | |
|---|---|
| | $|\mathtt{VARIABLES}| > 2$ |
| | range(VARIABLES.var) > 1 |

**Symmetries**

- Items of VARIABLES can be reversed.
- One and the same constant can be added to the var attribute of all items of VARIABLES.

| **Arg. properties** | Functional dependency: VALUE determined by VARIABLES. |
|---|---|

Figure 3.652: Illustrating the MAX_WIDTH_INFLEXION constraint of the **Example** slot

**Automaton**  Figures 3.653 and 3.654 respectively depict the automaton associated with the constraint MAX_WIDTH_INFLEXION and its simplified form.

Figure 3.653: Automaton for the MAX_WIDTH_INFLEXION constraint obtained by applying decoration Table 2.35 to the seed transducer of the INFLEXION pattern where default is 0 (transition $r \to t$ has the same accumulators updates as transition $t \to r$)

Figure 3.654: Automaton for the MAX_WIDTH_INFLEXION constraint obtained by applying decoration Table 2.25 to the seed transducer of the INFLEXION pattern where default is 0 (transition $r \to t$ has the same accumulators updates as transition $t \to r$); $R_i - R_{i-1} \geq 0$ and $-R_i + R_{i-1} + D_i \geq 0$ are linear invariants.

## 3.298 MAX_WIDTH_PEAK

▶ ▷ ◁ ◀        DESCRIPTION              AUTOMATON

$< (= | <)^* (> | =)^* >$

**Origin**            Based on the PEAK pattern.

**Constraint**        MAX_WIDTH_PEAK(VALUE, VARIABLES)

**Arguments**         VALUE       :  dvar
                      VARIABLES   :  collection(var−dvar)

**Restrictions**      $\mathtt{sv} \leq 2 \vee \mathtt{rv} \leq 1 \Rightarrow \mathtt{VALUE} = 0$
                      $\mathtt{VALUE} \geq 0$
                      $\mathtt{VALUE} \leq \max(0, \mathtt{sv} - 2①)$
                      required(VARIABLES, var)
                      where
                         $\mathtt{sv} = |\mathtt{VARIABLES}|$
                         $\mathtt{rv} = $range(VARIABLES.var)

VALUE ≤ **4** = 6 − 2

**Purpose**           VALUE is the maximal width of occurrences of the PEAK pattern in the time-series given
                      by the VARIABLES collection. If the pattern does not occur, VALUE takes the default
                      value 0.
                      An occurrence of the pattern PEAK is the *maximal* subsequence which matches the reg-
                      ular expression $< (= | <)^* (> | =)^* >$.
                      Assume that the occurrence of the pattern PEAK starts at position $i$ and ends at position
                      $j$. The feature WIDTH computes the value $j - i$.

**Example**           $(3, \langle 7, 5, 5, 1, 4, 5, 2, 2, 3, 5, 6, 2, 3, 3, 3, 1 \rangle)$

                      Figure   3.655   provides   an   example   where   the   MAX_WIDTH_PEAK
                      $(3, [7, 5, 5, 1, 4, 5, 2, 2, 3, 5, 6, 2, 3, 3, 3, 1])$ constraint holds.

**Typical**           $|\mathtt{VARIABLES}| > 2$
                      range(VARIABLES.var) > 1

**Symmetries**        • Items of VARIABLES can be reversed.

                      • One and the same constant can be added to the var attribute of all items of
                        VARIABLES.

**Arg. properties**   Functional dependency: VALUE determined by VARIABLES.

Figure 3.655: Illustrating the MAX_WIDTH_PEAK constraint of the **Example** slot

**Automaton**     Figures 3.656 and 3.657 respectively depict the automaton associated with the constraint MAX_WIDTH_PEAK and its simplified form.

Figure 3.656: Automaton for the MAX_WIDTH_PEAK constraint obtained by applying decoration Table 2.35 to the seed transducer of the PEAK pattern where default is 0

Figure 3.657: Automaton for the MAX_WIDTH_PEAK constraint obtained by applying decoration Table 2.26 to the seed transducer of the PEAK pattern where default is 0; $R_i - R_{i-1} \geq 0$ and $-R_i + R_{i-1} + D_i \geq 0$ are linear invariants.

|   | $s$ | $r$ | $t$ |
|---|---|---|---|
| $s$ | $\max(\overrightarrow{C}, \overleftarrow{C})$ | $\max(\overrightarrow{C}, \overleftarrow{C})$ | $\max(\overrightarrow{C}, \overleftarrow{C})$ |
| $r$ | $\max(\overrightarrow{C}, \overleftarrow{C})$ | $\overrightarrow{D} + \overleftarrow{D} + 1$ | $\overleftarrow{C} + \overrightarrow{D} + \overleftarrow{D} + 1$ |
| $t$ | $\max(\overrightarrow{C}, \overleftarrow{C})$ | $\overrightarrow{C} + \overrightarrow{D} + \overleftarrow{D} + 1$ | $\max(\overrightarrow{C}, \overleftarrow{C})$ |

Table 3.51: Glue matrix for the MAX_WIDTH_PEAK constraint defined as the composition of the PEAK pattern , the feature WIDTH , and the aggregator max ; cells of the glue matrix are coloured with the colour of the constituent to which they are related.

## 3.299   MAX_WIDTH_PLAIN

▶ ▷ ◁ ◀        DESCRIPTION            AUTOMATON

| | |
|---|---|
| **Origin** | Based on the PLAIN pattern. |
| **Constraint** | MAX_WIDTH_PLAIN(VALUE, VARIABLES) |
| **Arguments** | VALUE      : dvar<br>VARIABLES  : collection(var−dvar) |
| **Restrictions** | $\mathtt{sv} \leq 2 \vee \mathtt{rv} \leq 1 \Rightarrow \mathtt{VALUE} = 0$<br>$\mathtt{VALUE} \geq 0$<br>$\mathtt{VALUE} \leq \max(0, \mathtt{sv} - 2 ①)$<br>required(VARIABLES, var)<br>where<br>$\mathtt{sv} = \|\mathtt{VARIABLES}\|$<br>$\mathtt{rv} = \mathtt{range}(\mathtt{VARIABLES.var})$ |
| **Purpose** | VALUE is the maximal width of occurrences of the PLAIN pattern in the time-series given by the VARIABLES collection. If the pattern does not occur, VALUE takes the default value 0.<br>An occurrence of the pattern PLAIN is the *maximal* subsequence which matches the regular expression $> =^* <$.<br>Assume that the occurrence of the pattern PLAIN starts at position $i$ and ends at position $j$. The feature WIDTH computes the value $j - i$. |
| **Example** | $(2, \langle 2, 3, 6, 5, 7, 6, 6, 4, 5, 5, 4, 3, 3, 6, 6, 3 \rangle)$ |

Figure 3.658 provides an example where the MAX_WIDTH_PLAIN $(2, [2, 3, 6, 5, 7, 6, 6, 4, 5, 5, 4, 3, 3, 6, 6, 3])$ constraint holds.

| | |
|---|---|
| **Typical** | $\|\mathtt{VARIABLES}\| > 2$<br>$\mathtt{range}(\mathtt{VARIABLES.var}) > 1$ |
| **Symmetries** | • Items of VARIABLES can be reversed.<br>• One and the same constant can be added to the var attribute of all items of VARIABLES. |
| **Arg. properties** | Functional dependency: VALUE determined by VARIABLES. |

Figure 3.658: Illustrating the MAX_WIDTH_PLAIN constraint of the **Example** slot

**Automaton**    Figures 3.659 and 3.660 respectively depict the automaton associated with the constraint
MAX_WIDTH_PLAIN and its simplified form.

Figure 3.659: Automaton for the MAX_WIDTH_PLAIN constraint obtained by applying
decoration Table 2.35 to the seed transducer of the PLAIN pattern where default is 0

Figure 3.660: Automaton for the MAX_WIDTH_PLAIN constraint obtained by applying
decoration Table 2.25 to the seed transducer of the PLAIN pattern where default is 0;
$R_i - R_{i-1} \geq 0$ and $-R_i + R_{i-1} + D_i \geq 0$ are linear invariants.

|     | $s$ | $r$ | $t$ |
|-----|-----|-----|-----|
| $s$ | $\max(\overrightarrow{C}, \overleftarrow{C})$ | $\max(\overrightarrow{C}, \overleftarrow{C})$ | $\max(\overrightarrow{C}, \overleftarrow{C})$ |
| $r$ | $\max(\overrightarrow{C}, \overleftarrow{C})$ | $\overrightarrow{D} + \overleftarrow{D} + 1$ | $\overrightarrow{D} + \overleftarrow{D} + 1$ |
| $t$ | $\max(\overrightarrow{C}, \overleftarrow{C})$ | $\overrightarrow{D} + \overleftarrow{D} + 1$ | $\overrightarrow{D} + \overleftarrow{D} + 1$ |

Table 3.52: Glue matrix for the MAX_WIDTH_PLAIN constraint defined as the composition of the PLAIN pattern, the feature WIDTH, and the aggregator max; cells of the glue matrix are coloured with the colour of the constituent to which they are related.

## 3.300 MAX_WIDTH_PLATEAU

▶ ▷ ◁ ◀     DESCRIPTION          AUTOMATON

| | |
|---|---|
| **Origin** | Based on the PLATEAU pattern. |
| **Constraint** | MAX_WIDTH_PLATEAU(VALUE, VARIABLES) |
| **Arguments** | VALUE      : dvar<br>VARIABLES : collection(var−dvar) |
| **Restrictions** | $\mathtt{sv} \le 2 \vee \mathtt{rv} \le 1 \Rightarrow \mathtt{VALUE} = 0$<br>$\mathtt{VALUE} \ge 0$<br>$\mathtt{VALUE} \le \max(0, \mathtt{sv} - 2①)$<br>required(VARIABLES, var)<br>where<br>$\mathtt{sv} = |\mathtt{VARIABLES}|$<br>$\mathtt{rv} = \mathtt{range}(\mathtt{VARIABLES.var})$ |

| | |
|---|---|
| **Purpose** | VALUE is the maximal width of occurrences of the PLATEAU pattern in the time-series given by the VARIABLES collection. If the pattern does not occur, VALUE takes the default value 0.<br>An occurrence of the pattern PLATEAU is the *maximal* subsequence which matches the regular expression $< =^* >$.<br>Assume that the occurrence of the pattern PLATEAU starts at position $i$ and ends at position $j$. The feature WIDTH computes the value $j - i$. |
| **Example** | $(4, \langle 1, 3, 3, 5, 5, 5, 5, 2, 4, 4, 4, 3, 3, 1, 5, 5 \rangle)$ |

Figure 3.661 provides an example where the MAX_WIDTH_PLATEAU $(4, [1, 3, 3, 5, 5, 5, 5, 2, 4, 4, 4, 3, 3, 1, 5, 5])$ constraint holds.

| | |
|---|---|
| **Typical** | $|\mathtt{VARIABLES}| > 2$<br>$\mathtt{range}(\mathtt{VARIABLES.var}) > 1$ |

| | |
|---|---|
| **Symmetries** | • Items of VARIABLES can be reversed.<br>• One and the same constant can be added to the var attribute of all items of VARIABLES. |
| **Arg. properties** | Functional dependency: VALUE determined by VARIABLES. |

Figure 3.661: Illustrating the MAX_WIDTH_PLATEAU constraint of the **Example** slot

**Automaton**   Figures 3.662 and 3.663 respectively depict the automaton associated with the constraint
MAX_WIDTH_PLATEAU and its simplified form.

Figure 3.662: Automaton for the MAX_WIDTH_PLATEAU constraint obtained by applying decoration Table 2.35 to the seed transducer of the PLATEAU pattern where default is 0

Figure 3.663: Automaton for the MAX_WIDTH_PLATEAU constraint obtained by applying decoration Table 2.25 to the seed transducer of the PLATEAU pattern where default is 0; $R_i - R_{i-1} \geq 0$ and $-R_i + R_{i-1} + D_i \geq 0$ are linear invariants.

|   | $s$ | $r$ | $t$ |
|---|---|---|---|
| $s$ | $\max(\overrightarrow{C}, \overleftarrow{C})$ | $\max(\overrightarrow{C}, \overleftarrow{C})$ | $\max(\overrightarrow{C}, \overleftarrow{C})$ |
| $r$ | $\max(\overrightarrow{C}, \overleftarrow{C})$ | $\overrightarrow{D} + \overleftarrow{D} + 1$ | $\overrightarrow{D} + \overleftarrow{D} + 1$ |
| $t$ | $\max(\overrightarrow{C}, \overleftarrow{C})$ | $\overrightarrow{D} + \overleftarrow{D} + 1$ | $\overrightarrow{D} + \overleftarrow{D} + 1$ |

Table 3.53: Glue matrix for the MAX_WIDTH_PLATEAU constraint defined as the composition of the PLATEAU pattern, the feature WIDTH, and the aggregator max; cells of the glue matrix are coloured with the colour of the constituent to which they are related.

## 3.301 MAX_WIDTH_PROPER_PLAIN

   DESCRIPTION          AUTOMATON

| **Origin** | Based on the PROPER_PLAIN pattern. |
|---|---|

| **Constraint** | MAX_WIDTH_PROPER_PLAIN(VALUE, VARIABLES) |
|---|---|

**Arguments**

```
VALUE      :  dvar
VARIABLES  :  collection(var−dvar)
```

**Restrictions**

$\text{sv} \leq 3 \vee \text{rv} \leq 1 \Rightarrow \text{VALUE} = 0$
$\text{VALUE} = 0 \vee \text{VALUE} \geq 2$
$\text{VALUE} \leq \max(0, \text{sv} - 2①)$
required(VARIABLES, var)
where
$\text{sv} = |\text{VARIABLES}|$
$\text{rv} = \text{range}(\text{VARIABLES.var})$

**Purpose**

VALUE is the maximal width of occurrences of the PROPER_PLAIN pattern in the time-series given by the VARIABLES collection. If the pattern does not occur, VALUE takes the default value 0.
An occurrence of the pattern PROPER_PLAIN is the *maximal* subsequence which matches the regular expression $> =^+ <$.
Assume that the occurrence of the pattern PROPER_PLAIN starts at position $i$ and ends at position $j$. The feature WIDTH computes the value $j - i$.

**Example**

$(3, \langle 2, 7, 5, 5, 6, 3, 7, 4, 4, 5, 6, 5, 3, 3, 3, 5 \rangle)$

Figure 3.664 provides an example where the MAX_WIDTH_PROPER_PLAIN $(3, [2, 7, 5, 5, 6, 3, 7, 4, 4, 5, 6, 5, 3, 3, 3, 5])$ constraint holds.

**Typical**

$|\text{VARIABLES}| > 3$
$\text{range}(\text{VARIABLES.var}) > 1$

**Symmetries**

- Items of VARIABLES can be reversed.
- One and the same constant can be added to the var attribute of all items of VARIABLES.

**Arg. properties**

Functional dependency: VALUE determined by VARIABLES.

Figure 3.664: Illustrating the MAX_WIDTH_PROPER_PLAIN constraint of the **Example** slot

  Figures 3.665 and 3.666 respectively depict the automaton associated with the constraint MAX_WIDTH_PROPER_PLAIN and its simplified form.

Figure 3.665: Automaton for the MAX_WIDTH_PROPER_PLAIN constraint obtained by applying decoration Table 2.35 to the seed transducer of the PROPER_PLAIN pattern where default is 0

Figure 3.666: Automaton for the MAX_WIDTH_PROPER_PLAIN constraint obtained by applying decoration Table 2.25 to the seed transducer of the PROPER_PLAIN pattern where default is 0; $R_i - R_{i-1} \geq 0$ and $-R_i + R_{i-1} + D_i \geq 0$ are linear invariants.

|   | $s$ | $r$ | $t$ |
|---|---|---|---|
| $s$ | $\max(\overrightarrow{C}, \overleftarrow{C})$ | $\max(\overrightarrow{C}, \overleftarrow{C})$ | $\max(\overrightarrow{C}, \overleftarrow{C})$ |
| $r$ | $\max(\overrightarrow{C}, \overleftarrow{C})$ | $\max(\overrightarrow{C}, \overleftarrow{C})$ | $\overrightarrow{D} + \overleftarrow{D} + 1$ |
| $t$ | $\max(\overrightarrow{C}, \overleftarrow{C})$ | $\overrightarrow{D} + \overleftarrow{D} + 1$ | $\overrightarrow{D} + \overleftarrow{D} + 1$ |

Table 3.54: Glue matrix for the MAX_WIDTH_PROPER_PLAIN constraint defined as the composition of the PROPER_PLAIN pattern , the feature WIDTH , and the aggregator max ; cells of the glue matrix are coloured with the colour of the constituent to which they are related.

# 3.302 MAX_WIDTH_PROPER_PLATEAU

► ▷ ◁ ◀        **DESCRIPTION**                **AUTOMATON**

| | |
|---|---|
| **Origin** | Based on the PROPER_PLATEAU pattern. |
| **Constraint** | MAX_WIDTH_PROPER_PLATEAU(VALUE, VARIABLES) |
| **Arguments** | VALUE      : dvar |
| | VARIABLES  : collection(var−dvar) |

**Restrictions**

$\text{sv} \leq 3 \vee \text{rv} \leq 1 \Rightarrow \text{VALUE} = 0$
$\text{VALUE} = 0 \vee \text{VALUE} \geq 2$
$\text{VALUE} \leq \max(0, \text{sv} - 2\textcircled{2})$
required(VARIABLES, var)
where
$\text{sv} = |\text{VARIABLES}|$
$\text{rv} = \text{range}(\text{VARIABLES.var})$

**Purpose**

VALUE is the maximal width of occurrences of the PROPER_PLATEAU pattern in the time-series given by the VARIABLES collection. If the pattern does not occur, VALUE takes the default value 0.

An occurrence of the pattern PROPER_PLATEAU is the *maximal* subsequence which matches the regular expression $<=^+>$.

Assume that the occurrence of the pattern PROPER_PLATEAU starts at position $i$ and ends at position $j$. The feature WIDTH computes the value $j - i$.

**Example**

$(3, \langle 7, 1, 3, 3, 2, 5, 1, 4, 4, 3, 2, 3, 5, 5, 5, 3 \rangle)$

Figure 3.667 provides an example where the MAX_WIDTH_PROPER_PLATEAU $(3, [7, 1, 3, 3, 2, 5, 1, 4, 4, 3, 2, 3, 5, 5, 5, 3])$ constraint holds.

**Typical**

$|\text{VARIABLES}| > 3$
$\text{range}(\text{VARIABLES.var}) > 1$

**Symmetries**

- Items of VARIABLES can be reversed.
- One and the same constant can be added to the var attribute of all items of VARIABLES.

**Arg. properties**

Functional dependency: VALUE determined by VARIABLES.

Figure 3.667: Illustrating the MAX_WIDTH_PROPER_PLATEAU constraint of the **Example** slot

 Figures 3.668 and 3.669 respectively depict the automaton associated with the constraint MAX_WIDTH_PROPER_PLATEAU and its simplified form.

Figure 3.668: Automaton for the MAX_WIDTH_PROPER_PLATEAU constraint obtained by applying decoration Table 2.35 to the seed transducer of the PROPER_PLATEAU pattern where default is 0

Figure 3.669: Automaton for the MAX_WIDTH_PROPER_PLATEAU constraint obtained by applying decoration Table 2.25 to the seed transducer of the PROPER_PLATEAU pattern where default is 0; $R_i - R_{i-1} \geq 0$ and $-R_i + R_{i-1} + D_i \geq 0$ are linear invariants.

|   | $s$ | $r$ | $t$ |
|---|-----|-----|-----|
| $s$ | $\max(\overrightarrow{C}, \overleftarrow{C})$ | $\max(\overrightarrow{C}, \overleftarrow{C})$ | $\max(\overrightarrow{C}, \overleftarrow{C})$ |
| $r$ | $\max(\overrightarrow{C}, \overleftarrow{C})$ | $\max(\overrightarrow{C}, \overleftarrow{C})$ | $\overrightarrow{D} + \overleftarrow{D} + 1$ |
| $t$ | $\max(\overrightarrow{C}, \overleftarrow{C})$ | $\overrightarrow{D} + \overleftarrow{D} + 1$ | $\overrightarrow{D} + \overleftarrow{D} + 1$ |

Table 3.55: Glue matrix for the MAX_WIDTH_PROPER_PLATEAU constraint defined as the composition of the PROPER_PLATEAU pattern, the feature WIDTH, and the aggregator max; cells of the glue matrix are coloured with the colour of the constituent to which they are related.

## 3.303   MAX_WIDTH_STEADY_SEQUENCE

► ▷ ◁ ◀          **DESCRIPTION**          **AUTOMATON**

| | |
|---|---|
| **Origin** | Based on the STEADY_SEQUENCE pattern. |
| **Constraint** | MAX_WIDTH_STEADY_SEQUENCE(VALUE, VARIABLES) |
| **Arguments** | VALUE     : dvar |
| | VARIABLES : collection(var−dvar) |
| **Restrictions** | $\mathtt{sv} \leq 1 \Rightarrow \mathtt{VALUE} = 0$ |
| | $\mathtt{VALUE} = 0 \vee \mathtt{VALUE} \geq 2$ |
| | $\mathtt{rv} = 1 \wedge \mathtt{sv} \geq 2 \Rightarrow \mathtt{VALUE} \geq \mathtt{sv}$ |
| | $\mathtt{rv} \geq 2 \wedge \mathtt{sv} \geq 2 \Rightarrow \mathtt{VALUE} \geq 0$ |
| | $\mathtt{VALUE} \leq \mathtt{sv}①$ |
| | required(VARIABLES, var) |
| | where |
| | $\mathtt{sv} = |\mathtt{VARIABLES}|$ |
| | $\mathtt{rv} = \text{range}(\mathtt{VARIABLES.var})$ |
| **Purpose** | VALUE is the maximal width of occurrences of the STEADY_SEQUENCE pattern in the time-series given by the VARIABLES collection. If the pattern does not occur, VALUE takes the default value 0. |
| | An occurrence of the pattern STEADY_SEQUENCE is the *maximal* subsequence which matches the regular expression $=^{+}$. |
| | Assume that the occurrence of the pattern STEADY_SEQUENCE starts at position $i$ and ends at position $j$. The feature WIDTH computes the value $j - i + 2$. |
| **Example** | $(3, \langle 3, 1, 1, 4, 5, 5, 5, 6, 2, 2, 4, 4, 3, 2, 1, 1 \rangle)$ |

Figure 3.670 provides an example where the MAX_WIDTH_STEADY_SEQUENCE $(3, [3, 1, 1, 4, 5, 5, 5, 6, 2, 2, 4, 4, 3, 2, 1, 1])$ constraint holds.

| | |
|---|---|
| **Typical** | $|\mathtt{VARIABLES}| > 1$ |

| | |
|---|---|
| **Symmetries** | • Items of VARIABLES can be reversed. |
| | • One and the same constant can be added to the var attribute of all items of VARIABLES. |
| **Arg. properties** | Functional dependency: VALUE determined by VARIABLES. |

Figure 3.670: Illustrating the MAX_WIDTH_STEADY_SEQUENCE constraint of the **Example** slot

**Automaton**     Figures 3.671 and 3.672 respectively depict the automaton associated with the constraint
MAX_WIDTH_STEADY_SEQUENCE and its simplified form.

Figure 3.671: Automaton for the MAX_WIDTH_STEADY_SEQUENCE constraint
obtained by applying decoration Table 2.35 to the seed transducer of the
STEADY_SEQUENCE pattern where default is 0

|     | $s$ | $r$ |
|-----|-----|-----|
| $s$ | $\max(\overrightarrow{C}, \overleftarrow{C})$ | $\max(\overrightarrow{C}, \overleftarrow{C})$ |
| $r$ | $\max(\overrightarrow{C}, \overleftarrow{C})$ | $\overrightarrow{C} + \overleftarrow{C} + \overrightarrow{D} + \overleftarrow{D} + 1$ |

Table 3.56: Glue matrix for the MAX_WIDTH_STEADY_SEQUENCE constraint defined
as the composition of the STEADY_SEQUENCE pattern , the feature WIDTH , and the
aggregator max ; cells of the glue matrix are coloured with the colour of the con-
stituent to which they are related.

Figure 3.672: Automaton for the MAX_WIDTH_STEADY_SEQUENCE constraint obtained by applying decoration Table 2.30 to the seed transducer of the STEADY_SEQUENCE pattern where default is $0$; $R_i - R_{i-1} \geq 0$ and $-R_i + R_{i-1} + D_i \geq 0$ are linear invariants.

## 3.304  MAX_WIDTH_STRICTLY_DECREASING_SEQUENCE

▶ ▷ ◁ ◀      **DESCRIPTION**          **AUTOMATON**

| | |
|---|---|
| **Origin** | Based on the STRICTLY_DECREASING_SEQUENCE pattern. |
| **Constraint** | MAX_WIDTH_STRICTLY_DECREASING_SEQUENCE(VALUE, VARIABLES) |
| **Arguments** | VALUE     :  dvar<br>VARIABLES  :  collection(var−dvar) |

**Restrictions**

$\text{sv} \leq 1 \lor \text{rv} \leq 1 \Rightarrow \text{VALUE} = 0$
$\text{VALUE} = 0 \lor \text{VALUE} \geq 2$
$\text{VALUE} \leq \min(\text{sv}, \text{rv}①)$
required(VARIABLES, var)
where
$\text{sv} = |\text{VARIABLES}|$
$\text{rv} = \text{range}(\text{VARIABLES.var})$

**Purpose**

VALUE is the maximal width of occurrences of the STRICTLY_DECREASING_SEQUENCE pattern in the time-series given by the VARIABLES collection. If the pattern does not occur, VALUE takes the default value 0.

An occurrence of the pattern STRICTLY_DECREASING_SEQUENCE is the *maximal* sub-sequence which matches the regular expression $>^+$.

Assume that the occurrence of the pattern STRICTLY_DECREASING_SEQUENCE starts at position $i$ and ends at position $j$. The feature WIDTH computes the value $j - i + 2$.

**Example**

$(3, \langle 4, 4, 6, 4, 1, 1, 3, 4, 4, 6, 6, 5, 2, 2, 4, 3 \rangle)$

Figure 3.673 provides an example where the MAX_WIDTH_STRICTLY_DECREASING_SEQUENCE $(3, [4, 4, 6, 4, 1, 1, 3, 4, 4, 6, 6, 5, 2, 2, 4, 3])$ constraint holds.

**Typical**

$|\text{VARIABLES}| > 1$
range(VARIABLES.var) $> 1$

**Symmetry**

One and the same constant can be added to the var attribute of all items of VARIABLES.

**Arg. properties**

Functional dependency: VALUE determined by VARIABLES.

Figure 3.673: Illustrating the MAX_WIDTH_STRICTLY_DECREASING_SEQUENCE constraint of the **Example** slot

**Automaton**    Figures 3.674 and 3.675 respectively depict the automaton associated with the constraint MAX_WIDTH_STRICTLY_DECREASING_SEQUENCE and its simplified form.

$$\left\{ \begin{array}{l} C \leftarrow \texttt{default} \\ D \leftarrow 0 \\ R \leftarrow \texttt{default} \end{array} \right\}$$

$\leq s$    $\leq$

$$\left\{ \begin{array}{l} \leq \\ C \leftarrow \texttt{default} \\ D \leftarrow 0 \\ R \leftarrow \max(R, C) \end{array} \right\}$$

$\max(R, C)$

$$\left\{ \begin{array}{l} > \\ C \leftarrow D + 2 \\ D \leftarrow 0 \end{array} \right\}$$

$> r$    $$\left\{ \begin{array}{l} > \\ C \leftarrow C + D + 1 \\ D \leftarrow 0 \end{array} \right\}$$

Figure 3.674: Automaton for the MAX_WIDTH_STRICTLY_DECREASING_SEQUENCE constraint obtained by applying decoration Table 2.35 to the seed transducer of the STRICTLY_DECREASING_SEQUENCE pattern where `default` is 0

|     | $s$ | $r$ |
| --- | --- | --- |
| $s$ | $\max(\overrightarrow{C}, \overleftarrow{C})$ | $\max(\overrightarrow{C}, \overleftarrow{C})$ |
| $r$ | $\max(\overrightarrow{C}, \overleftarrow{C})$ | $\overrightarrow{C} + \overleftarrow{C} + \overrightarrow{D} + \overleftarrow{D} + 1$ |

Table 3.57: Glue matrix for the MAX_WIDTH_STRICTLY_DECREASING_SEQUENCE constraint defined as the composition of the STRICTLY_DECREASING_SEQUENCE pattern , the feature WIDTH , and the aggregator max ; cells of the glue matrix are coloured with the colour of the constituent to which they are related.

Figure 3.675: Automaton for the MAX_WIDTH_STRICTLY_DECREASING_SEQUENCE constraint obtained by applying decoration Table 2.30 to the seed transducer of the STRICTLY_DECREASING_SEQUENCE pattern where default is 0; $R_i - R_{i-1} \geq 0$ and $-R_i + R_{i-1} + 2 \geq 0$ are linear invariants.

# 3.305 MAX_WIDTH_STRICTLY_INCREASING_SEQUENCE

▶ ▷ ◁ ◀        **DESCRIPTION**        **AUTOMATON**

| | |
|---|---|
| **Origin** | Based on the STRICTLY_INCREASING_SEQUENCE pattern. |
| **Constraint** | MAX_WIDTH_STRICTLY_INCREASING_SEQUENCE(VALUE, VARIABLES) |
| **Arguments** | VALUE : dvar |
| | VARIABLES : collection(var−dvar) |
| **Restrictions** | $\mathtt{sv} \leq 1 \vee \mathtt{rv} \leq 1 \Rightarrow \mathtt{VALUE} = 0$ |
| | $\mathtt{VALUE} = 0 \vee \mathtt{VALUE} \geq 2$ |
| | $\mathtt{VALUE} \leq \min(\mathtt{sv}, \mathtt{rv}①)$ |
| | required(VARIABLES, var) |
| | where |
| | $\mathtt{sv} = |\mathtt{VARIABLES}|$ |
| | $\mathtt{rv} = $ range(VARIABLES.var) |

**Purpose**

VALUE is the maximal width of occurrences of the STRICTLY_INCREASING_SEQUENCE pattern in the time-series given by the VARIABLES collection. If the pattern does not occur, VALUE takes the default value 0.

An occurrence of the pattern STRICTLY_INCREASING_SEQUENCE is the *maximal* subsequence which matches the regular expression $<^{+}$.

Assume that the occurrence of the pattern STRICTLY_INCREASING_SEQUENCE starts at position $i$ and ends at position $j$. The feature WIDTH computes the value $j - i + 2$.

**Example**

$(5, \langle 4, 3, 5, 5, 2, 1, 1, 2, 3, 4, 6, 6, 3, 1, 2, 3 \rangle)$

Figure 3.676 provides an example where the MAX_WIDTH_STRICTLY_INCREASING_SEQUENCE $(5, [4, 3, 5, 5, 2, 1, 1, 2, 3, 4, 6, 6, 3, 1, 2, 3])$ constraint holds.

| | |
|---|---|
| **Typical** | $|\mathtt{VARIABLES}| > 1$ |
| | range(VARIABLES.var) > 1 |
| **Symmetry** | One and the same constant can be added to the var attribute of all items of VARIABLES. |
| **Arg. properties** | Functional dependency: VALUE determined by VARIABLES. |

Figure 3.676: Illustrating the MAX_WIDTH_STRICTLY_INCREASING_SEQUENCE constraint of the **Example** slot

**Automaton**    Figures 3.677 and 3.678 respectively depict the automaton associated with the constraint
MAX_WIDTH_STRICTLY_INCREASING_SEQUENCE and its simplified form.

Figure 3.677: Automaton for the MAX_WIDTH_STRICTLY_INCREASING_SEQUENCE
constraint obtained by applying decoration Table 2.35 to the seed transducer of the
STRICTLY_INCREASING_SEQUENCE pattern where default is $0$

|     | $s$ | $r$ |
|-----|-----|-----|
| $s$ | $\max(\overrightarrow{C}, \overleftarrow{C})$ | $\max(\overrightarrow{C}, \overleftarrow{C})$ |
| $r$ | $\max(\overrightarrow{C}, \overleftarrow{C})$ | $\overrightarrow{C} + \overleftarrow{C} + \overrightarrow{D} + \overleftarrow{D} + 1$ |

Table 3.58: Glue matrix for the MAX_WIDTH_STRICTLY_INCREASING_SEQUENCE
constraint defined as the composition of the STRICTLY_INCREASING_SEQUENCE pattern,
the feature WIDTH, and the aggregator max; cells of the glue matrix are coloured
with the colour of the constituent to which they are related.

Figure 3.678: Automaton for the MAX_WIDTH_STRICTLY_INCREASING_SEQUENCE constraint obtained by applying decoration Table 2.30 to the seed transducer of the STRICTLY_INCREASING_SEQUENCE pattern where default is $0$; $R_i - R_{i-1} \geq 0$ and $-R_i + R_{i-1} + 2 \geq 0$ are linear invariants.

## 3.306    MAX_WIDTH_SUMMIT

$(< | < (= | <)^* <)(> | > (= | >)^* >)$

| | |
|---|---|
| **Origin** | Based on the SUMMIT pattern. |
| **Constraint** | MAX_WIDTH_SUMMIT(VALUE, VARIABLES) |
| **Arguments** | VALUE      : dvar |
| | VARIABLES  : collection(var−dvar) |
| **Restrictions** | $\mathtt{sv} \leq 2 \vee \mathtt{rv} \leq 1 \Rightarrow \mathtt{VALUE} = 0$ |
| | $\mathtt{VALUE} \geq 0$ |
| | $\mathtt{rv} = 2 \Rightarrow \mathtt{VALUE} \leq 1$① |
| | $\mathtt{rv} \geq 3 \Rightarrow \mathtt{VALUE} \leq \max(0, \mathtt{sv} - 2$②$)$ |
| | required(VARIABLES, var) |
| | where |
| | $\mathtt{sv} = \lvert \mathtt{VARIABLES} \rvert$ |
| | $\mathtt{rv} = $ range(VARIABLES.var) |

VALUE ≤ **1**

VALUE ≤ **5** = 7 − 2

| | |
|---|---|
| **Purpose** | VALUE is the maximal width of occurrences of the SUMMIT pattern in the time-series given by the VARIABLES collection. If the pattern does not occur, VALUE takes the default value 0. |
| | An occurrence of the pattern SUMMIT is the *maximal* subsequence which matches the regular expression $(< | < (= | <)^* <)(> | > (= | >)^* >)$. |
| | Assume that the occurrence of the pattern SUMMIT starts at position $i$ and ends at position $j$. The feature WIDTH computes the value $j - i$. |
| **Example** | $(3, \langle 7, 1, 5, 4, 4, 3, 3, 4, 6, 6, 2, 3, 4, 2, 3, 1 \rangle)$ |

Figure    3.679    provides    an    example    where    the    MAX_WIDTH_SUMMIT $(3, [7, 1, 5, 4, 4, 3, 3, 4, 6, 6, 2, 3, 4, 2, 3, 1])$ constraint holds.

| | |
|---|---|
| **Typical** | $\lvert \mathtt{VARIABLES} \rvert > 2$ |
| | range(VARIABLES.var) $> 1$ |
| **Symmetries** | • Items of VARIABLES can be reversed. |
| | • One and the same constant can be added to the var attribute of all items of VARIABLES. |
| **Arg. properties** | Functional dependency: VALUE determined by VARIABLES. |

Figure 3.679: Illustrating the MAX_WIDTH_SUMMIT constraint of the **Example** slot

**Automaton**   Figures 3.680 and 3.681 respectively depict the automaton associated with the constraint
MAX_WIDTH_SUMMIT and its simplified form.

Figure 3.680: Automaton for the MAX_WIDTH_SUMMIT constraint obtained by apply-
ing decoration Table 2.35 to the seed transducer of the SUMMIT pattern where default
is 0 (transition $u \rightarrow r$ has the same accumulator update as transition $r \rightarrow u$)

Figure 3.681: Automaton for the MAX_WIDTH_SUMMIT constraint obtained by applying decoration Table 2.26 to the seed transducer of the SUMMIT pattern where default is 0 (transition $u \to r$ has the same accumulator update as transition $r \to u$); $R_i - R_{i-1} \geq 0$ and $-R_i + R_{i-1} + D_i \geq 0$ are linear invariants.

|     | $s$ | $r$ | $t$ | $u$ |
|-----|-----|-----|-----|-----|
| $s$ | $\max(\overrightarrow{C},\overleftarrow{C})$ | $\max(\overrightarrow{C},\overleftarrow{C})$ | $\max(\overrightarrow{C},\overleftarrow{C})$ | $\max(\overrightarrow{C},\overleftarrow{C})$ |
| $r$ | $\max(\overrightarrow{C},\overleftarrow{C})$ | $\overrightarrow{D}+\overleftarrow{D}+1$ | $\overleftarrow{C}+\overrightarrow{D}+\overleftarrow{D}+1$ | $\max(\overrightarrow{C},\overleftarrow{C})$ |
| $t$ | $\max(\overrightarrow{C},\overleftarrow{C})$ | $\overrightarrow{C}+\overrightarrow{D}+\overleftarrow{D}+1$ | $\max(\overrightarrow{C},\overleftarrow{C})$ | $\overrightarrow{C}+\overrightarrow{D}+\overleftarrow{D}+1$ |
| $u$ | $\max(\overrightarrow{C},\overleftarrow{C})$ | $\max(\overrightarrow{C},\overleftarrow{C})$ | $\overleftarrow{C}+\overrightarrow{D}+\overleftarrow{D}+1$ | $\max(\overrightarrow{C},\overleftarrow{C})$ |

Table 3.59: Glue matrix for the MAX_WIDTH_SUMMIT constraint defined as the composition of the SUMMIT pattern, the feature WIDTH, and the aggregator max; cells of the glue matrix are coloured with the colour of the constituent to which they are related.

## 3.307  MAX_WIDTH_VALLEY

► ▷ ◁ ◀        DESCRIPTION          AUTOMATON

**Origin**          Based on the VALLEY pattern.

**Constraint**      MAX_WIDTH_VALLEY(VALUE, VARIABLES)

**Arguments**       VALUE      : dvar
                    VARIABLES  : collection(var−dvar)

**Restrictions**    $\mathtt{sv} \leq 2 \vee \mathtt{rv} \leq 1 \Rightarrow \mathtt{VALUE} = 0$
                    $\mathtt{VALUE} \geq 0$
                    $\mathtt{VALUE} \leq \max(0, \mathtt{sv} - 2①)$
                    required(VARIABLES, var)
                    where
                    $\mathtt{sv} = |\mathtt{VARIABLES}|$
                    $\mathtt{rv} = \mathtt{range}(\mathtt{VARIABLES.var})$

**Purpose**         VALUE is the maximal width of occurrences of the VALLEY pattern in the time-series given by the VARIABLES collection. If the pattern does not occur, VALUE takes the default value 0.
                    An occurrence of the pattern VALLEY is the *maximal* subsequence which matches the regular expression > (= | >)* (< | =)* <.
                    Assume that the occurrence of the pattern VALLEY starts at position $i$ and ends at position $j$. The feature WIDTH computes the value $j - i$.

**Example**         $(4, \langle 1, 3, 7, 4, 3, 6, 6, 5, 3, 3, 2, 6, 5, 5, 5, 7 \rangle)$

                    Figure 3.682 provides an example where the MAX_WIDTH_VALLEY $(4, [1, 3, 7, 4, 3, 6, 6, 5, 3, 3, 2, 6, 5, 5, 5, 7])$ constraint holds.

**Typical**         $|\mathtt{VARIABLES}| > 2$
                    $\mathtt{range}(\mathtt{VARIABLES.var}) > 1$

**Symmetries**      • Items of VARIABLES can be reversed.
                    • One and the same constant can be added to the var attribute of all items of VARIABLES.

**Arg. properties** Functional dependency: VALUE determined by VARIABLES.

Figure 3.682: Illustrating the MAX_WIDTH_VALLEY constraint of the **Example** slot

**Automaton**     Figures 3.683 and 3.684 respectively depict the automaton associated with the constraint
MAX_WIDTH_VALLEY and its simplified form.

Figure 3.683: Automaton for the MAX_WIDTH_VALLEY constraint obtained by applying decoration Table 2.35 to the seed transducer of the VALLEY pattern where default is 0

Figure 3.684: Automaton for the MAX_WIDTH_VALLEY constraint obtained by applying decoration Table 2.26 to the seed transducer of the VALLEY pattern where default is 0; $R_i - R_{i-1} \geq 0$ and $-R_i + R_{i-1} + D_i \geq 0$ are linear invariants.

| | $s$ | $r$ | $t$ |
|---|---|---|---|
| $s$ | $\max(\overrightarrow{C}, \overleftarrow{C})$ | $\max(\overrightarrow{C}, \overleftarrow{C})$ | $\max(\overrightarrow{C}, \overleftarrow{C})$ |
| $r$ | $\max(\overrightarrow{C}, \overleftarrow{C})$ | $\overrightarrow{D} + \overleftarrow{D} + 1$ | $\overleftarrow{C} + \overrightarrow{D} + \overleftarrow{D} + 1$ |
| $t$ | $\max(\overrightarrow{C}, \overleftarrow{C})$ | $\overrightarrow{C} + \overrightarrow{D} + \overleftarrow{D} + 1$ | $\max(\overrightarrow{C}, \overleftarrow{C})$ |

Table 3.60: Glue matrix for the MAX_WIDTH_VALLEY constraint defined as the composition of the VALLEY pattern , the feature WIDTH , and the aggregator max ; cells of the glue matrix are coloured with the colour of the constituent to which they are related.

## 3.308 MAX_WIDTH_ZIGZAG

     DESCRIPTION     AUTOMATON

$(<>)^+(<\,|\,<>)\,|\,(><)^+(>\,|\,><)$

| | |
|---|---|
| **Origin** | Based on the ZIGZAG pattern. |
| **Constraint** | MAX_WIDTH_ZIGZAG(VALUE, VARIABLES) |
| **Arguments** | VALUE : dvar<br>VARIABLES : collection(var−dvar) |

**Restrictions**

$\mathrm{sv} \leq 3 \vee \mathrm{rv} \leq 1 \Rightarrow \mathtt{VALUE} = 0$

$\mathtt{VALUE} = 0 \vee \mathtt{VALUE} \geq 2$

$\mathtt{VALUE} \leq \max(0, \mathrm{sv} - 2\text{①})$

required(VARIABLES, var)

where

$\mathrm{sv} = |\mathtt{VARIABLES}|$

$\mathrm{rv} =$ range(VARIABLES.var)

**Purpose**

VALUE is the maximal width of occurrences of the ZIGZAG pattern in the time-series given by the VARIABLES collection. If the pattern does not occur, VALUE takes the default value 0.

An occurrence of the pattern ZIGZAG is the *maximal* subsequence which matches the regular expression $(<>)^+(<\,|\,<>)\,|\,(><)^+(>\,|\,><)$.

Assume that the occurrence of the pattern ZIGZAG starts at position $i$ and ends at position $j$. The feature WIDTH computes the value $j - i$.

**Example**

$(6, \langle 4, 1, 3, 1, 4, 6, 1, 5, 5, 2, 7, 2, 3, 1, 6, 1 \rangle)$

Figure 3.685 provides an example where the MAX_WIDTH_ZIGZAG $(6, [4, 1, 3, 1, 4, 6, 1, 5, 5, 2, 7, 2, 3, 1, 6, 1])$ constraint holds.

| | |
|---|---|
| **Typical** | $|\mathtt{VARIABLES}| > 3$<br>range(VARIABLES.var) $> 1$ |
| **Symmetries** | • Items of VARIABLES can be reversed.<br>• One and the same constant can be added to the var attribute of all items of VARIABLES. |
| **Arg. properties** | Functional dependency: VALUE determined by VARIABLES. |

Figure 3.685: Illustrating the MAX_WIDTH_ZIGZAG constraint of the **Example** slot

**Automaton**       Figures 3.686 and 3.687 respectively depict the automaton associated with the constraint
                    MAX_WIDTH_ZIGZAG and its simplified form.

Figure 3.686: Automaton for the MAX_WIDTH_ZIGZAG constraint obtained by applying decoration Table 2.35 to the seed transducer of the ZIGZAG pattern where default is 0; (1) missing transitions from $a, b, c, d, e, f$ to $s$ are labelled by $=$; (2) on transitions from $b, c, e, f$ to $s$ the accumulator $D$ is reset to its initial value; (3) on transitions from $c, f$ to $s$ the accumulator $R$ is updated wrt $C$ and the accumulator $C$ is reset to its initial value

Figure 3.687: Automaton for the MAX_WIDTH_ZIGZAG constraint obtained by applying decoration Table 2.26 to the seed transducer of the ZIGZAG pattern where default is 0; (1) missing transitions from $a, b, c, d, e, f$ to $s$ are labelled by $=$; (2) on transitions from $b, c, e, f$ to $s$ the accumulator $D$ is reset to its initial value; $R_i - R_{i-1} \geq 0$ and $-R_i + R_{i-1} + 2 \geq 0$ are linear invariants.

|   | s | a | b | c | d | e | f |
|---|---|---|---|---|---|---|---|
| **s** | $\max(\overrightarrow{C},\overleftarrow{C})$ | $\max(\overrightarrow{C},\overleftarrow{C})$ | $\max(\overrightarrow{C},\overleftarrow{C})$ | $\max(\overrightarrow{C},\overleftarrow{C})$ | $\max(\overrightarrow{C},\overleftarrow{C})$ | $\max(\overrightarrow{C},\overleftarrow{C})$ | $\max(\overrightarrow{C},\overleftarrow{C})$ |
| **a** | $\max(\overrightarrow{C},\overleftarrow{C})$ | $\max(\overrightarrow{C},\overleftarrow{C})$ | $\max(\overrightarrow{C},\overleftarrow{C})$ | $\overleftarrow{C}+\overrightarrow{D}+\overleftarrow{D}+1$ | $\max(\overrightarrow{C},\overleftarrow{C})$ | $\overrightarrow{D}+\overleftarrow{D}+1$ | $\max(\overrightarrow{C},\overleftarrow{C})$ |
| **b** | $\max(\overrightarrow{C},\overleftarrow{C})$ | $\max(\overrightarrow{C},\overleftarrow{C})$ | $\overrightarrow{D}+\overleftarrow{D}+1$ | $\max(\overrightarrow{C},\overleftarrow{C})$ | $\overrightarrow{D}+\overleftarrow{D}+1$ | $\max(\overrightarrow{C},\overleftarrow{C})$ | $\overleftarrow{C}+\overrightarrow{D}+\overleftarrow{D}+1$ |
| **c** | $\max(\overrightarrow{C},\overleftarrow{C})$ | $\overrightarrow{C}+\overrightarrow{D}+\overleftarrow{D}+1$ | $\max(\overrightarrow{C},\overleftarrow{C})$ | $\overrightarrow{C}+\overleftarrow{C}+\overrightarrow{D}+\overleftarrow{D}+1$ | $\max(\overrightarrow{C},\overleftarrow{C})$ | $\overrightarrow{C}+\overrightarrow{D}+\overleftarrow{D}+1$ | $\max(\overrightarrow{C},\overleftarrow{C})$ |
| **d** | $\max(\overrightarrow{C},\overleftarrow{C})$ | $\max(\overrightarrow{C},\overleftarrow{C})$ | $\overrightarrow{D}+\overleftarrow{D}+1$ | $\max(\overrightarrow{C},\overleftarrow{C})$ | $\max(\overrightarrow{C},\overleftarrow{C})$ | $\max(\overrightarrow{C},\overleftarrow{C})$ | $\overleftarrow{C}+\overrightarrow{D}+\overleftarrow{D}+1$ |
| **e** | $\max(\overrightarrow{C},\overleftarrow{C})$ | $\overrightarrow{D}+\overleftarrow{D}+1$ | $\max(\overrightarrow{C},\overleftarrow{C})$ | $\overleftarrow{C}+\overrightarrow{D}+\overleftarrow{D}+1$ | $\max(\overrightarrow{C},\overleftarrow{C})$ | $\overrightarrow{D}+\overleftarrow{D}+1$ | $\max(\overrightarrow{C},\overleftarrow{C})$ |
| **f** | $\max(\overrightarrow{C},\overleftarrow{C})$ | $\max(\overrightarrow{C},\overleftarrow{C})$ | $\overrightarrow{C}+\overrightarrow{D}+\overleftarrow{D}+1$ | $\max(\overrightarrow{C},\overleftarrow{C})$ | $\overrightarrow{C}+\overrightarrow{D}+\overleftarrow{D}+1$ | $\max(\overrightarrow{C},\overleftarrow{C})$ | $\overrightarrow{C}+\overleftarrow{C}+\overrightarrow{D}+\overleftarrow{D}+1$ |

Table 3.61: Glue matrix for the MAX_WIDTH_ZIGZAG constraint defined as the composition of the `ZIGZAG pattern`, the feature `WIDTH`, and the `aggregator max`; cells of the glue matrix are coloured with the colour of the constituent to which they are related.

## 3.309   MIN_HEIGHT_DECREASING_TERRACE

▶ ▷ ◁ ◀          DESCRIPTION            AUTOMATON

**Origin**          Based on the DECREASING_TERRACE pattern.

**Constraint**      MIN_HEIGHT_DECREASING_TERRACE(VALUE, VARIABLES)

**Arguments**       VALUE       : dvar
                    VARIABLES   : collection(var−dvar)

**Restrictions**    $\text{sv} \leq 3 \vee \text{rv} \leq 2 \Rightarrow \text{VALUE} = +\infty$

                    $\text{VALUE} \geq \text{minv} + 1$①

                    $\text{VALUE} = +\infty \vee \text{VALUE} \leq \text{maxv} - 1$

                    required(VARIABLES, var)

                    where

                    minv =minval(VARIABLES.var)

                    maxv =maxval(VARIABLES.var)

                    sv = |VARIABLES|

                    rv =range(VARIABLES.var)

**Purpose**         VALUE is the minimum of all minimum values in each occurrence of the DECREAS-
                    ING_TERRACE pattern in the time-series given by the VARIABLES collection. If the pat-
                    tern does not occur, VALUE takes the default value $+\infty$.

                    An occurrence of the pattern DECREASING_TERRACE is the *maximal* subsequence which
                    matches the regular expression $>=^+>$.

                    Assume that the occurrence of the pattern DECREASING_TERRACE starts at position $i$
                    and ends at position $j$. The feature MIN computes the minimum of the values from index
                    $i + 1$ to index $j$.

**Example**         $(2, \langle 6, 4, 4, 4, 5, 2, 2, 1, 3, 3, 5, 4, 4, 3, 3, 3 \rangle)$

                    Figure 3.688 provides an example where the MIN_HEIGHT_DECREASING_TERRACE
                    $(2, [6, 4, 4, 4, 5, 2, 2, 1, 3, 3, 5, 4, 4, 3, 3, 3])$ constraint holds.

**Typical**         $|\text{VARIABLES}| > 3$

                    range(VARIABLES.var) > 2

**Arg. properties** Functional dependency: VALUE determined by VARIABLES.

Figure 3.688: Illustrating the MIN_HEIGHT_DECREASING_TERRACE constraint of the **Example** slot

 Figures 3.689 and 3.690 respectively depict the automaton associated with the constraint MIN_HEIGHT_DECREASING_TERRACE and its simplified form.

Figure 3.689: Automaton for the MIN_HEIGHT_DECREASING_TERRACE constraint obtained by applying decoration Table 2.35 to the seed transducer of the DECREASING_TERRACE pattern where default is $+\infty$

Figure 3.690: Automaton for the MIN_HEIGHT_DECREASING_TERRACE constraint obtained by applying decoration Table 2.37 to the seed transducer of the DECREASING_TERRACE pattern where default is $+\infty$; $-R_i + R_{i-1} \geq 0$ is a linear invariant.

|   | $s$ | $r$ | $t$ |
|---|---|---|---|
| $s$ | $\min(\overrightarrow{C}, \overleftarrow{C})$ | $\min(\overrightarrow{C}, \overleftarrow{C})$ | $\min(\overrightarrow{C}, \overleftarrow{C})$ |
| $r$ | $\min(\overrightarrow{C}, \overleftarrow{C})$ | $\min(\overrightarrow{C}, \overleftarrow{C})$ | $\min(\overrightarrow{D}, \overleftarrow{D}, \texttt{VAR}_i)$ |
| $t$ | $\min(\overrightarrow{C}, \overleftarrow{C})$ | $\min(\overrightarrow{D}, \overleftarrow{D}, \texttt{VAR}_i)$ | $\min(\overrightarrow{D}, \overleftarrow{D}, \texttt{VAR}_i)$ |

Table 3.62: Glue matrix for the MIN_HEIGHT_DECREASING_TERRACE constraint defined as the composition of the DECREASING_TERRACE pattern , the feature MIN , and the aggregator min ; cells of the glue matrix are coloured with the colour of the constituent to which they are related.

# 3.310 MIN_HEIGHT_INCREASING_TERRACE

▶ ▷ ◁ ◀        **DESCRIPTION**          **AUTOMATON**

| | |
|---|---|
| **Origin** | Based on the INCREASING_TERRACE pattern. |
| **Constraint** | MIN_HEIGHT_INCREASING_TERRACE(VALUE, VARIABLES) |
| **Arguments** | VALUE : dvar<br>VARIABLES : collection(var−dvar) |
| **Restrictions** | $\mathtt{sv} \leq 3 \vee \mathtt{rv} \leq 2 \Rightarrow \mathtt{VALUE} = +\infty$<br>$\mathtt{VALUE} \geq \mathtt{minv} + 1$①<br>$\mathtt{VALUE} = +\infty \vee \mathtt{VALUE} \leq \mathtt{maxv} - 1$<br>required(VARIABLES, var)<br>where<br>minv =minval(VARIABLES.var)<br>maxv =maxval(VARIABLES.var)<br>sv = \|VARIABLES\|<br>rv =range(VARIABLES.var) |

**Purpose**

VALUE is the minimum of all minimum values in each occurrence of the INCREAS-ING_TERRACE pattern in the time-series given by the VARIABLES collection. If the pattern does not occur, VALUE takes the default value $+\infty$.

An occurrence of the pattern INCREASING_TERRACE is the *maximal* subsequence which matches the regular expression $<=^+<$.

Assume that the occurrence of the pattern INCREASING_TERRACE starts at position $i$ and ends at position $j$. The feature MIN computes the minimum of the values from index $i + 1$ to index $j$.

**Example**

$(3, \langle 1, 3, 3, 3, 2, 5, 5, 6, 4, 4, 2, 3, 3, 3, 4, 4\rangle)$

Figure 3.691 provides an example where the MIN_HEIGHT_INCREASING_TERRACE $(3, [1, 3, 3, 3, 2, 5, 5, 6, 4, 4, 2, 3, 3, 3, 4, 4])$ constraint holds.

| | |
|---|---|
| **Typical** | \|VARIABLES\| > 3<br>range(VARIABLES.var) > 2 |
| **Arg. properties** | Functional dependency: VALUE determined by VARIABLES. |

Figure 3.691: Illustrating the MIN_HEIGHT_INCREASING_TERRACE constraint of the **Example** slot

**Automaton**   Figures 3.692 and 3.693 respectively depict the automaton associated with the constraint MIN_HEIGHT_INCREASING_TERRACE and its simplified form.

Figure 3.692: Automaton for the MIN_HEIGHT_INCREASING_TERRACE constraint obtained by applying decoration Table 2.35 to the seed transducer of the INCREASING_TERRACE pattern where default is $+\infty$

Figure 3.693: Automaton for the MIN_HEIGHT_INCREASING_TERRACE constraint obtained by applying decoration Table 2.37 to the seed transducer of the INCREASING_TERRACE pattern where default is $+\infty$; $-R_i + R_{i-1} \geq 0$ is a linear invariant.

|   | $s$ | $r$ | $t$ |
|---|-----|-----|-----|
| $s$ | $\min(\overrightarrow{C}, \overleftarrow{C})$ | $\min(\overrightarrow{C}, \overleftarrow{C})$ | $\min(\overrightarrow{C}, \overleftarrow{C})$ |
| $r$ | $\min(\overrightarrow{C}, \overleftarrow{C})$ | $\min(\overrightarrow{C}, \overleftarrow{C})$ | $\min(\overrightarrow{D}, \overleftarrow{D}, \texttt{VAR}_i)$ |
| $t$ | $\min(\overrightarrow{C}, \overleftarrow{C})$ | $\min(\overrightarrow{D}, \overleftarrow{D}, \texttt{VAR}_i)$ | $\min(\overrightarrow{D}, \overleftarrow{D}, \texttt{VAR}_i)$ |

Table 3.63: Glue matrix for the MIN_HEIGHT_INCREASING_TERRACE constraint defined as the composition of the INCREASING_TERRACE pattern , the feature MIN , and the aggregator min ; cells of the glue matrix are coloured with the colour of the constituent to which they are related.

# 3.311 MIN_HEIGHT_PLAIN

        **DESCRIPTION**              **AUTOMATON**

| | |
|---|---|
| **Origin** | Based on the PLAIN pattern. |
| **Constraint** | MIN_HEIGHT_PLAIN(VALUE, VARIABLES) |
| **Arguments** | VALUE      : dvar<br>VARIABLES  : collection(var−dvar) |

**Restrictions**

$\text{sv} \leq 2 \vee \text{rv} \leq 1 \Rightarrow \text{VALUE} = +\infty$

$\text{VALUE} \geq \text{minv}①$

$\text{VALUE} = +\infty \vee \text{VALUE} \leq \text{maxv} - 1$

required(VARIABLES, var)

where

minv = minval(VARIABLES.var)

maxv = maxval(VARIABLES.var)

sv = |VARIABLES|

rv = range(VARIABLES.var)

**Purpose**

VALUE is the minimum of all minimum values in each occurrence of the PLAIN pattern in the time-series given by the VARIABLES collection. If the pattern does not occur, VALUE takes the default value $+\infty$.

An occurrence of the pattern PLAIN is the *maximal* subsequence which matches the regular expression $> = ^{*} <$.

Assume that the occurrence of the pattern PLAIN starts at position $i$ and ends at position $j$. The feature MIN computes the minimum of the values from index $i + 1$ to index $j$.

**Example**

$(3, \langle 2, 3, 6, 5, 7, 6, 6, 4, 5, 5, 4, 3, 3, 6, 6, 3 \rangle)$

Figure 3.694 provides an example where the MIN_HEIGHT_PLAIN $(3, [2, 3, 6, 5, 7, 6, 6, 4, 5, 5, 4, 3, 3, 6, 6, 3])$ constraint holds.

| | |
|---|---|
| **Typical** | $\|\text{VARIABLES}\| > 2$<br>range(VARIABLES.var) $> 1$ |
| **Symmetry** | Items of VARIABLES can be reversed. |
| **Arg. properties** | Functional dependency: VALUE determined by VARIABLES. |

Figure 3.694: Illustrating the MIN_HEIGHT_PLAIN constraint of the **Example** slot

**Automaton**     Figures 3.695 and 3.696 respectively depict the automaton associated with the constraint MIN_HEIGHT_PLAIN and its simplified form.

Figure 3.695: Automaton for the MIN_HEIGHT_PLAIN constraint obtained by applying decoration Table 2.35 to the seed transducer of the PLAIN pattern where default is $+\infty$

Figure 3.696: Automaton for the MIN_HEIGHT_PLAIN constraint obtained by applying decoration Table 2.37 to the seed transducer of the PLAIN pattern where default is $+\infty$; $-R_i + R_{i-1} \geq 0$ is a linear invariant.

|   | $s$ | $r$ | $t$ |
|---|---|---|---|
| $s$ | $\min(\overrightarrow{C}, \overleftarrow{C})$ | $\min(\overrightarrow{C}, \overleftarrow{C})$ | $\min(\overrightarrow{C}, \overleftarrow{C})$ |
| $r$ | $\min(\overrightarrow{C}, \overleftarrow{C})$ | $\min(\overrightarrow{D}, \overleftarrow{D}, \mathtt{VAR}_i)$ | $\min(\overrightarrow{D}, \overleftarrow{D}, \mathtt{VAR}_i)$ |
| $t$ | $\min(\overrightarrow{C}, \overleftarrow{C})$ | $\min(\overrightarrow{D}, \overleftarrow{D}, \mathtt{VAR}_i)$ | $\min(\overrightarrow{D}, \overleftarrow{D}, \mathtt{VAR}_i)$ |

Table 3.64: Glue matrix for the MIN_HEIGHT_PLAIN constraint defined as the composition of the PLAIN pattern, the feature MIN, and the aggregator min; cells of the glue matrix are coloured with the colour of the constituent to which they are related.

## 3.312  MIN_HEIGHT_PLATEAU

▶ ▷ ◁ ◀        DESCRIPTION        AUTOMATON

**Origin**            Based on the PLATEAU pattern.

**Constraint**        MIN_HEIGHT_PLATEAU(VALUE, VARIABLES)

**Arguments**         VALUE     : dvar
                      VARIABLES : collection(var−dvar)

**Restrictions**      $\mathtt{sv} \leq 2 \vee \mathtt{rv} \leq 1 \Rightarrow \mathtt{VALUE} = +\infty$
                      $\mathtt{VALUE} \geq \mathtt{minv} + 1①$
                      $\mathtt{VALUE} = +\infty \vee \mathtt{VALUE} \leq \mathtt{maxv}$
                      required(VARIABLES, var)
                      where
                      minv =minval(VARIABLES.var)
                      maxv =maxval(VARIABLES.var)
                      sv = |VARIABLES|
                      rv =range(VARIABLES.var)

**Purpose**           VALUE is the minimum of all minimum values in each occurrence of the PLATEAU pattern
                      in the time-series given by the VARIABLES collection. If the pattern does not occur,
                      VALUE takes the default value $+\infty$.
                      An occurrence of the pattern PLATEAU is the *maximal* subsequence which matches the
                      regular expression $<=^*>$.
                      Assume that the occurrence of the pattern PLATEAU starts at position $i$ and ends at posi-
                      tion $j$. The feature MIN computes the minimum of the values from index $i + 1$ to index
                      $j$.

**Example**           $(3, \langle 7, 5, 2, 3, 1, 2, 2, 4, 3, 3, 4, 5, 5, 2, 2, 5 \rangle)$

                      Figure   3.697   provides   an   example   where   the   MIN_HEIGHT_PLATEAU
                      $(3, [7, 5, 2, 3, 1, 2, 2, 4, 3, 3, 4, 5, 5, 2, 2, 5])$ constraint holds.

**Typical**           $|\mathtt{VARIABLES}| > 2$
                      range(VARIABLES.var) > 1

**Symmetry**          Items of VARIABLES can be reversed.

**Arg. properties**   Functional dependency: VALUE determined by VARIABLES.

Figure 3.697: Illustrating the MIN_HEIGHT_PLATEAU constraint of the **Example** slot

     Figures 3.698 and 3.699 respectively depict the automaton associated with the constraint
MIN_HEIGHT_PLATEAU and its simplified form.

Figure 3.698: Automaton for the MIN_HEIGHT_PLATEAU constraint obtained by applying decoration Table 2.35 to the seed transducer of the PLATEAU pattern where default is $+\infty$

Figure 3.699: Automaton for the MIN_HEIGHT_PLATEAU constraint obtained by applying decoration Table 2.37 to the seed transducer of the PLATEAU pattern where default is $+\infty$; $-R_i + R_{i-1} \geq 0$ is a linear invariant.

|  | $s$ | $r$ | $t$ |
|---|---|---|---|
| $s$ | $\min(\overrightarrow{C}, \overleftarrow{C})$ | $\min(\overrightarrow{C}, \overleftarrow{C})$ | $\min(\overrightarrow{C}, \overleftarrow{C})$ |
| $r$ | $\min(\overrightarrow{C}, \overleftarrow{C})$ | $\min(\overrightarrow{D}, \overleftarrow{D}, \texttt{VAR}_i)$ | $\min(\overrightarrow{D}, \overleftarrow{D}, \texttt{VAR}_i)$ |
| $t$ | $\min(\overrightarrow{C}, \overleftarrow{C})$ | $\min(\overrightarrow{D}, \overleftarrow{D}, \texttt{VAR}_i)$ | $\min(\overrightarrow{D}, \overleftarrow{D}, \texttt{VAR}_i)$ |

Table 3.65: Glue matrix for the MIN_HEIGHT_PLATEAU constraint defined as the composition of the PLATEAU pattern , the feature MIN , and the aggregator min ; cells of the glue matrix are coloured with the colour of the constituent to which they are related.

## 3.313   MIN_HEIGHT_PROPER_PLAIN

▶ ▷ ◁ ◀          **DESCRIPTION**          **AUTOMATON**

| | |
|---|---|
| **Origin** | Based on the PROPER_PLAIN pattern. |
| **Constraint** | MIN_HEIGHT_PROPER_PLAIN(VALUE, VARIABLES) |
| **Arguments** | VALUE      : dvar<br>VARIABLES : collection(var−dvar) |

**Restrictions**

$\mathtt{sv} \leq 3 \vee \mathtt{rv} \leq 1 \Rightarrow \mathtt{VALUE} = +\infty$

$\mathtt{VALUE} \geq \mathtt{minv}$①

$\mathtt{VALUE} = +\infty \vee \mathtt{VALUE} \leq \mathtt{maxv} - 1$

required(VARIABLES, var)

where

$\mathtt{minv} = \mathtt{minval}(\mathtt{VARIABLES.var})$

$\mathtt{maxv} = \mathtt{maxval}(\mathtt{VARIABLES.var})$

$\mathtt{sv} = |\mathtt{VARIABLES}|$

$\mathtt{rv} = \mathtt{range}(\mathtt{VARIABLES.var})$

**Purpose**

VALUE is the minimum of all minimum values in each occurrence of the PROPER_PLAIN pattern in the time-series given by the VARIABLES collection. If the pattern does not occur, VALUE takes the default value $+\infty$.

An occurrence of the pattern PROPER_PLAIN is the *maximal* subsequence which matches the regular expression $> =^+ <$.

Assume that the occurrence of the pattern PROPER_PLAIN starts at position $i$ and ends at position $j$. The feature MIN computes the minimum of the values from index $i + 1$ to index $j$.

| | |
|---|---|
| **Example** | $(3, \langle 2, 7, 5, 5, 6, 3, 7, 4, 4, 5, 6, 5, 3, 3, 3, 5 \rangle)$ |

Figure 3.700 provides an example where the MIN_HEIGHT_PROPER_PLAIN $(3, [2, 7, 5, 5, 6, 3, 7, 4, 4, 5, 6, 5, 3, 3, 3, 5])$ constraint holds.

| | |
|---|---|
| **Typical** | $\|\mathtt{VARIABLES}\| > 3$<br>range(VARIABLES.var) $> 1$ |
| **Symmetry** | Items of VARIABLES can be reversed. |
| **Arg. properties** | Functional dependency: VALUE determined by VARIABLES. |

Figure 3.700: Illustrating the MIN_HEIGHT_PROPER_PLAIN constraint of the **Example** slot

                  Figures 3.701 and 3.702 respectively depict the automaton associated with the constraint
MIN_HEIGHT_PROPER_PLAIN and its simplified form.

Figure 3.701: Automaton for the MIN_HEIGHT_PROPER_PLAIN constraint obtained by
applying decoration Table 2.35 to the seed transducer of the PROPER_PLAIN pattern
where default is $+\infty$

Figure 3.702: Automaton for the MIN_HEIGHT_PROPER_PLAIN constraint obtained by
applying decoration Table 2.37 to the seed transducer of the PROPER_PLAIN pattern
where default is $+\infty$; $-R_i + R_{i-1} \geq 0$ is a linear invariant.

|     | $s$ | $r$ | $t$ |
| --- | --- | --- | --- |
| $s$ | $\min(\overrightarrow{C}, \overleftarrow{C})$ | $\min(\overrightarrow{C}, \overleftarrow{C})$ | $\min(\overrightarrow{C}, \overleftarrow{C})$ |
| $r$ | $\min(\overrightarrow{C}, \overleftarrow{C})$ | $\min(\overrightarrow{C}, \overleftarrow{C})$ | $\min(\overrightarrow{D}, \overleftarrow{D}, \mathtt{VAR}_i)$ |
| $t$ | $\min(\overrightarrow{C}, \overleftarrow{C})$ | $\min(\overrightarrow{D}, \overleftarrow{D}, \mathtt{VAR}_i)$ | $\min(\overrightarrow{D}, \overleftarrow{D}, \mathtt{VAR}_i)$ |

Table 3.66: Glue matrix for the MIN_HEIGHT_PROPER_PLAIN constraint defined as the composition of the PROPER_PLAIN pattern, the feature MIN, and the aggregator min; cells of the glue matrix are coloured with the colour of the constituent to which they are related.

## 3.314 MIN_HEIGHT_PROPER_PLATEAU

► ▷ ◁ ◄        DESCRIPTION              AUTOMATON

| **Origin** | Based on the PROPER_PLATEAU pattern. |
|---|---|

| **Constraint** | MIN_HEIGHT_PROPER_PLATEAU(VALUE, VARIABLES) |
|---|---|

| **Arguments** | VALUE      : dvar |
|---|---|
| | VARIABLES : collection(var−dvar) |

**Restrictions**

$\text{sv} \leq 3 \vee \text{rv} \leq 1 \Rightarrow \text{VALUE} = +\infty$
$\text{VALUE} \geq \text{minv} + 1 \textcircled{1}$
$\text{VALUE} = +\infty \vee \text{VALUE} \leq \text{maxv}$
required(VARIABLES, var)
where
minv = minval(VARIABLES.var)
maxv = maxval(VARIABLES.var)
sv = |VARIABLES|
rv = range(VARIABLES.var)

**Purpose**

VALUE is the minimum of all minimum values in each occurrence of the PROPER_PLATEAU pattern in the time-series given by the VARIABLES collection. If the pattern does not occur, VALUE takes the default value $+\infty$.

An occurrence of the pattern PROPER_PLATEAU is the *maximal* subsequence which matches the regular expression $<=^+>$.

Assume that the occurrence of the pattern PROPER_PLATEAU starts at position $i$ and ends at position $j$. The feature MIN computes the minimum of the values from index $i + 1$ to index $j$.

| **Example** | $(3, \langle 7, 1, 3, 3, 2, 5, 1, 4, 4, 3, 2, 3, 5, 5, 5, 3 \rangle)$ |
|---|---|

Figure 3.703 provides an example where the MIN_HEIGHT_PROPER_PLATEAU $(3, [7, 1, 3, 3, 2, 5, 1, 4, 4, 3, 2, 3, 5, 5, 5, 3])$ constraint holds.

| **Typical** | $|\text{VARIABLES}| > 3$ |
|---|---|
| | range(VARIABLES.var) > 1 |

| **Symmetry** | Items of VARIABLES can be reversed. |
|---|---|

| **Arg. properties** | Functional dependency: VALUE determined by VARIABLES. |
|---|---|

Figure 3.703: Illustrating the MIN_HEIGHT_PROPER_PLATEAU constraint of the **Example** slot

 Figures 3.704 and 3.705 respectively depict the automaton associated with the constraint MIN_HEIGHT_PROPER_PLATEAU and its simplified form.

Figure 3.704: Automaton for the MIN_HEIGHT_PROPER_PLATEAU constraint obtained by applying decoration Table 2.35 to the seed transducer of the PROPER_PLATEAU pattern where default is $+\infty$

Figure 3.705: Automaton for the MIN_HEIGHT_PROPER_PLATEAU constraint obtained by applying decoration Table 2.37 to the seed transducer of the PROPER_PLATEAU pattern where default is $+\infty$; $-R_i + R_{i-1} \geq 0$ is a linear invariant.

|     | $s$ | $r$ | $t$ |
|-----|-----|-----|-----|
| $s$ | $\min(\overrightarrow{C}, \overleftarrow{C})$ | $\min(\overrightarrow{C}, \overleftarrow{C})$ | $\min(\overrightarrow{C}, \overleftarrow{C})$ |
| $r$ | $\min(\overrightarrow{C}, \overleftarrow{C})$ | $\min(\overrightarrow{C}, \overleftarrow{C})$ | $\min(\overrightarrow{D}, \overleftarrow{D}, \text{VAR}_i)$ |
| $t$ | $\min(\overrightarrow{C}, \overleftarrow{C})$ | $\min(\overrightarrow{D}, \overleftarrow{D}, \text{VAR}_i)$ | $\min(\overrightarrow{D}, \overleftarrow{D}, \text{VAR}_i)$ |

Table 3.67: Glue matrix for the MIN_HEIGHT_PROPER_PLATEAU constraint defined as the composition of the PROPER_PLATEAU pattern, the feature MIN, and the aggregator min; cells of the glue matrix are coloured with the colour of the constituent to which they are related.

## 3.315   MIN_HEIGHT_STEADY

► ▷ ◁ ◀     DESCRIPTION     AUTOMATON

| | |
|---|---|
| **Origin** | Based on the STEADY pattern. |
| **Constraint** | MIN_HEIGHT_STEADY(VALUE, VARIABLES) |
| **Arguments** | VALUE     : dvar |
| | VARIABLES : collection(var−dvar) |

**Restrictions**

$\text{sv} \leq 1 \Rightarrow \text{VALUE} = +\infty$
$\text{VALUE} \geq \text{minv}①$
$\text{VALUE} = +\infty \vee \text{VALUE} \leq \text{maxv}$
required(VARIABLES, var)
where
minv = minval(VARIABLES.var)
maxv = maxval(VARIABLES.var)
sv = |VARIABLES|

**Purpose**

VALUE is the minimum of all minimum values in each occurrence of the STEADY pattern in the time-series given by the VARIABLES collection. If the pattern does not occur, VALUE takes the default value $+\infty$.

An occurrence of the pattern STEADY is the subsequence which matches the regular expression =.

Assume that the occurrence of the pattern STEADY starts at position $i$ and ends at position $j$. The feature MIN computes the minimum of the values from index $i$ to index $j + 1$.

**Example**

$(1, \langle 1, 1, 7, 3, 3, 5, 5, 5, 6, 5, 5, 5, 7, 2, 6, 6 \rangle)$

Figure 3.706 provides an example where the MIN_HEIGHT_STEADY $(1, [1, 1, 7, 3, 3, 5, 5, 5, 6, 5, 5, 5, 7, 2, 6, 6])$ constraint holds.

| | |
|---|---|
| **Typical** | |VARIABLES| > 1 |

| | |
|---|---|
| **Symmetry** | Items of VARIABLES can be reversed. |
| **Arg. properties** | Functional dependency: VALUE determined by VARIABLES. |

Figure 3.706: Illustrating the MIN_HEIGHT_STEADY constraint of the **Example** slot

**Automaton**    Figures 3.707 and 3.708 respectively depict the automaton associated with the constraint
MIN_HEIGHT_STEADY and its simplified form.

Figure 3.707: Automaton for the MIN_HEIGHT_STEADY constraint obtained by applying decoration Table 2.35 to the seed transducer of the STEADY pattern where default is $+\infty$

Figure 3.708: Automaton for the MIN_HEIGHT_STEADY constraint obtained by applying decoration Table 2.37 to the seed transducer of the STEADY pattern where default is $+\infty$; $-R_i + R_{i-1} \geq 0$ is a linear invariant.

Table 3.68: Glue matrix for the MIN_HEIGHT_STEADY constraint defined as the composition of the STEADY pattern , the feature MIN , and the aggregator min ; cells of the glue matrix are coloured with the colour of the constituent to which they are related.

## 3.316  MIN_HEIGHT_STEADY_SEQUENCE

► ▷ ◁ ◄          DESCRIPTION          AUTOMATON

| | |
|---|---|
| **Origin** | Based on the STEADY_SEQUENCE pattern. |
| **Constraint** | MIN_HEIGHT_STEADY_SEQUENCE(VALUE, VARIABLES) |
| **Arguments** | VALUE      : dvar |
| | VARIABLES  : collection(var−dvar) |

**Restrictions**

$\text{sv} \leq 1 \Rightarrow \text{VALUE} = +\infty$

$\text{VALUE} \geq \text{minv}①$

$\text{VALUE} = +\infty \vee \text{VALUE} \leq \text{maxv}$

required(VARIABLES, var)

where

minv = minval(VARIABLES.var)

maxv = maxval(VARIABLES.var)

$\text{sv} = |\text{VARIABLES}|$

**Purpose**

VALUE is the minimum of all minimum values in each occurrence of the STEADY_SEQUENCE pattern in the time-series given by the VARIABLES collection. If the pattern does not occur, VALUE takes the default value $+\infty$.

An occurrence of the pattern STEADY_SEQUENCE is the *maximal* subsequence which matches the regular expression $=^+$.

Assume that the occurrence of the pattern STEADY_SEQUENCE starts at position $i$ and ends at position $j$. The feature MIN computes the minimum of the values from index $i$ to index $j + 1$.

| | |
|---|---|
| **Example** | $(1, \langle 3, 1, 1, 4, 5, 5, 5, 6, 2, 2, 4, 4, 3, 2, 1, 1 \rangle)$ |

Figure 3.709 provides an example where the MIN_HEIGHT_STEADY_SEQUENCE $(1, [3, 1, 1, 4, 5, 5, 5, 6, 2, 2, 4, 4, 3, 2, 1, 1])$ constraint holds.

| | |
|---|---|
| **Typical** | $|\text{VARIABLES}| > 1$ |

| | |
|---|---|
| **Symmetry** | Items of VARIABLES can be reversed. |
| **Arg. properties** | Functional dependency: VALUE determined by VARIABLES. |

Figure 3.709: Illustrating the MIN_HEIGHT_STEADY_SEQUENCE constraint of the **Example** slot

**Automaton**      Figures 3.710 and 3.711 respectively depict the automaton associated with the constraint
MIN_HEIGHT_STEADY_SEQUENCE and its simplified form.

$$\left\{ \begin{array}{l} C \leftarrow \texttt{default} \\ D \leftarrow +\infty \\ R \leftarrow \texttt{default} \end{array} \right\}$$

Figure 3.710: Automaton for the MIN_HEIGHT_STEADY_SEQUENCE constraint
obtained by applying decoration Table 2.35 to the seed transducer of the
STEADY_SEQUENCE pattern where default is $+\infty$

Figure 3.711: Automaton for the MIN_HEIGHT_STEADY_SEQUENCE constraint
obtained by applying decoration Table 2.37 to the seed transducer of the
STEADY_SEQUENCE pattern where default is $+\infty$; $-R_i + R_{i-1} \geq 0$ is a linear invariant.

|   | $s$ | $r$ |
|---|---|---|
| $s$ | $\min(\overrightarrow{C}, \overleftarrow{C})$ | $\min(\overrightarrow{C}, \overleftarrow{C})$ |
| $r$ | $\min(\overrightarrow{C}, \overleftarrow{C})$ | $\min(\overrightarrow{C}, \overleftarrow{C}, \overrightarrow{D}, \overleftarrow{D}, \mathtt{VAR}_i)$ |

Table 3.69: Glue matrix for the MIN_HEIGHT_STEADY_SEQUENCE constraint defined as the composition of the STEADY_SEQUENCE pattern, the feature MIN, and the aggregator min; cells of the glue matrix are coloured with the colour of the constituent to which they are related.

# 3.317   MIN_MAX_BUMP_ON_DECREASING_SEQUENCE

▶ ▷ ◁ ◀          **DESCRIPTION**          **AUTOMATON**

| | |
|---|---|
| **Origin** | Based on the BUMP_ON_DECREASING_SEQUENCE pattern. |
| **Constraint** | MIN_MAX_BUMP_ON_DECREASING_SEQUENCE(VALUE, VARIABLES) |
| **Arguments** | VALUE       :  dvar<br>VARIABLES  :  collection(var−dvar) |

**Restrictions**

$\text{sv} \leq 5 \vee \text{rv} \leq 2 \Rightarrow \text{VALUE} = +\infty$

$\text{VALUE} \geq \text{minv} + 2 ①$

$\text{VALUE} = +\infty \vee \text{VALUE} \leq \text{maxv}$

required(VARIABLES, var)

where

minv = minval(VARIABLES.var)

maxv = maxval(VARIABLES.var)

sv = |VARIABLES|

rv = range(VARIABLES.var)

**Purpose**

VALUE is the minimum of all maximum values in each occurrence of the BUMP_ON_DECREASING_SEQUENCE pattern in the time-series given by the VARIABLES collection. If the pattern does not occur, VALUE takes the default value $+\infty$.

An occurrence of the pattern BUMP_ON_DECREASING_SEQUENCE is the subsequence which matches the regular expression >><>>.

Assume that the occurrence of the pattern BUMP_ON_DECREASING_SEQUENCE starts at position $i$ and ends at position $j$. The feature MAX computes the maximum of the values from index $i + 2$ to index $j$.

**Example**

$(5, \langle 7, 6, 5, 6, 5, 4, 1, 4, 7, 5, 4, 2, 5, 4, 3, 3 \rangle)$

Figure 3.712 provides an example where the MIN_MAX_BUMP_ON_DECREASING_SEQUENCE $(5, [7, 6, 5, 6, 5, 4, 1, 4, 7, 5, 4, 2, 5, 4, 3, 3])$ constraint holds.

| | |
|---|---|
| **Typical** | $|\text{VARIABLES}| > 5$<br>range(VARIABLES.var) $> 2$ |
| **Arg. properties** | Functional dependency: VALUE determined by VARIABLES. |

Figure 3.712: Illustrating the MIN_MAX_BUMP_ON_DECREASING_SEQUENCE constraint of the **Example** slot

**Automaton**     Figures 3.713 and 3.714 respectively depict the automaton associated with the constraint
MIN_MAX_BUMP_ON_DECREASING_SEQUENCE and its simplified form.

Figure 3.713: Automaton for the MIN_MAX_BUMP_ON_DECREASING_SEQUENCE constraint obtained by applying decoration Table 2.35 to the seed transducer of the BUMP_ON_DECREASING_SEQUENCE pattern where default is $+\infty$

Figure 3.714: Automaton for the MIN_MAX_BUMP_ON_DECREASING_SEQUENCE constraint obtained by applying decoration Table 2.27 to the seed transducer of the BUMP_ON_DECREASING_SEQUENCE pattern where default is $+\infty$; $-R_i + R_{i-1} \geq 0$ is a linear invariant.

## 3.318   MIN_MAX_DECREASING

► ▷ ◁ ◄          DESCRIPTION          AUTOMATON

| | |
|---|---|
| **Origin** | Based on the DECREASING pattern. |
| **Constraint** | MIN_MAX_DECREASING(VALUE, VARIABLES) |
| **Arguments** | VALUE       : dvar<br>VARIABLES  : collection(var−dvar) |
| **Restrictions** | $\text{sv} \leq 1 \vee \text{rv} \leq 1 \Rightarrow \text{VALUE} = +\infty$<br>$\text{VALUE} \geq \text{minv} + 1 ①$<br>$\text{VALUE} = +\infty \vee \text{VALUE} \leq \text{maxv}$<br>required(VARIABLES, var)<br>where<br>minv = minval(VARIABLES.var)<br>maxv = maxval(VARIABLES.var)<br>sv = \|VARIABLES\|<br>rv = range(VARIABLES.var) |
| **Purpose** | VALUE is the minimum of all maximum values in each occurrence of the DECREASING pattern in the time-series given by the VARIABLES collection. If the pattern does not occur, VALUE takes the default value $+\infty$.<br>An occurrence of the pattern DECREASING is the subsequence which matches the regular expression >.<br>Assume that the occurrence of the pattern DECREASING starts at position $i$ and ends at position $j$. The feature MAX computes the maximum of the values from index $i$ to index $j + 1$. |
| **Example** | $(3, \langle 3, 4, 2, 2, 5, 6, 6, 4, 4, 3, 1, 1, 4, 6, 4, 4 \rangle)$ |

Figure 3.715 provides an example where the MIN_MAX_DECREASING $(3, [3, 4, 2, 2, 5, 6, 6, 4, 4, 3, 1, 1, 4, 6, 4, 4])$ constraint holds.

| | |
|---|---|
| **Typical** | $\|\text{VARIABLES}\| > 1$<br>range(VARIABLES.var) > 1 |
| **Arg. properties** | Functional dependency: VALUE determined by VARIABLES. |

Figure 3.715: Illustrating the MIN_MAX_DECREASING constraint of the **Example** slot

**Automaton**     Figures 3.716 and 3.717 respectively depict the automaton associated with the constraint MIN_MAX_DECREASING and its simplified form.

Figure 3.716: Automaton for the MIN_MAX_DECREASING constraint obtained by applying decoration Table 2.35 to the seed transducer of the DECREASING pattern where default is $+\infty$

Figure 3.717: Automaton for the MIN_MAX_DECREASING constraint obtained by applying decoration Table 2.37 to the seed transducer of the DECREASING pattern where default is $+\infty$; $-R_i + R_{i-1} \geq 0$ is a linear invariant.

Table 3.70: Glue matrix for the MIN_MAX_DECREASING constraint defined as the composition of the DECREASING pattern, the feature MAX, and the aggregator min; cells of the glue matrix are coloured with the colour of the constituent to which they are related.

# 3.319 MIN_MAX_DECREASING_SEQUENCE

| **Origin** | Based on the DECREASING_SEQUENCE pattern. |
|---|---|

| **Constraint** | MIN_MAX_DECREASING_SEQUENCE(VALUE, VARIABLES) |
|---|---|

| **Arguments** | VALUE : dvar |
|---|---|
| | VARIABLES : collection(var−dvar) |

**Restrictions**

$sv \leq 1 \lor rv \leq 1 \Rightarrow$ VALUE $= +\infty$

VALUE $\geq minv + 1$ ①

VALUE $= +\infty \lor$ VALUE $\leq maxv$

required(VARIABLES, var)

where

$minv =$ minval(VARIABLES.var)

$maxv =$ maxval(VARIABLES.var)

$sv = |$VARIABLES$|$

$rv =$ range(VARIABLES.var)

**Purpose**

VALUE is the minimum of all maximum values in each occurrence of the DECREAS-ING_SEQUENCE pattern in the time-series given by the VARIABLES collection. If the pattern does not occur, VALUE takes the default value $+\infty$.

An occurrence of the pattern DECREASING_SEQUENCE is the *maximal* subsequence which matches the regular expression $> (> | =)^* > |>$.

Assume that the occurrence of the pattern DECREASING_SEQUENCE starts at position $i$ and ends at position $j$. The feature MAX computes the maximum of the values from index $i$ to index $j + 1$.

**Example**

$(4, \langle 3, 4, 2, 2, 5, 6, 6, 4, 4, 3, 1, 1, 4, 6, 4, 4 \rangle)$

Figure 3.718 provides an example where the MIN_MAX_DECREASING_SEQUENCE $(4, [3, 4, 2, 2, 5, 6, 6, 4, 4, 3, 1, 1, 4, 6, 4, 4])$ constraint holds.

| **Typical** | $|$VARIABLES$| > 1$ |
|---|---|
| | range(VARIABLES.var) $> 1$ |

| **Arg. properties** | Functional dependency: VALUE determined by VARIABLES. |
|---|---|

Figure 3.718: Illustrating the MIN_MAX_DECREASING_SEQUENCE constraint of the **Example** slot

 Figures 3.719 and 3.720 respectively depict the automaton associated with the constraint MIN_MAX_DECREASING_SEQUENCE and its simplified form.

Figure 3.719: Automaton for the MIN_MAX_DECREASING_SEQUENCE constraint obtained by applying decoration Table 2.35 to the seed transducer of the DECREASING_SEQUENCE pattern where default is $+\infty$

Figure 3.720: Automaton for the MIN_MAX_DECREASING_SEQUENCE constraint obtained by applying decoration Table 2.37 to the seed transducer of the DECREASING_SEQUENCE pattern where default is $+\infty$; $-R_i + R_{i-1} \geq 0$ is a linear invariant.

|   | $s$ | $t$ |
|---|---|---|
| $s$ | $\min(\overrightarrow{C}, \overleftarrow{C})$ | $\min(\overrightarrow{C}, \overleftarrow{C})$ |
| $t$ | $\min(\overrightarrow{C}, \overleftarrow{C})$ | $\max(\overrightarrow{C}, \overleftarrow{C}, \overrightarrow{D}, \overleftarrow{D}, \text{VAR}_i)$ |

Table 3.71: Glue matrix for the MIN_MAX_DECREASING_SEQUENCE constraint defined as the composition of the DECREASING_SEQUENCE pattern, the feature MAX, and the aggregator min; cells of the glue matrix are coloured with the colour of the constituent to which they are related.

## 3.320 MIN_MAX_DIP_ON_INCREASING_SEQUENCE

► ▷ ◁ ◄     **DESCRIPTION**          **AUTOMATON**

| | |
|---|---|
| **Origin** | Based on the DIP_ON_INCREASING_SEQUENCE pattern. |
| **Constraint** | MIN_MAX_DIP_ON_INCREASING_SEQUENCE(VALUE, VARIABLES) |
| **Arguments** | VALUE      : dvar<br>VARIABLES : collection(var−dvar) |

**Restrictions**

$\mathtt{sv} \leq 5 \lor \mathtt{rv} \leq 2 \Rightarrow \mathtt{VALUE} = +\infty$
$\mathtt{VALUE} \geq \mathtt{minv} + 2$①
$\mathtt{VALUE} = +\infty \lor \mathtt{VALUE} \leq \mathtt{maxv}$
required(VARIABLES, var)
where
$\mathtt{minv} =$ minval(VARIABLES.var)
$\mathtt{maxv} =$ maxval(VARIABLES.var)
$\mathtt{sv} = |\mathtt{VARIABLES}|$
$\mathtt{rv} =$ range(VARIABLES.var)

**Purpose**

VALUE is the minimum of all maximum values in each occurrence of the DIP_ON_INCREASING_SEQUENCE pattern in the time-series given by the VARIABLES collection. If the pattern does not occur, VALUE takes the default value $+\infty$.

An occurrence of the pattern DIP_ON_INCREASING_SEQUENCE is the subsequence which matches the regular expression <<><<.

Assume that the occurrence of the pattern DIP_ON_INCREASING_SEQUENCE starts at position $i$ and ends at position $j$. The feature MAX computes the maximum of the values from index $i + 2$ to index $j$.

**Example**

$(5, \langle 1, 2, 3, 2, 5, 6, 7, 4, 1, 3, 4, 6, 1, 2, 4, 4 \rangle)$

Figure 3.721 provides an example where the MIN_MAX_DIP_ON_INCREASING_SEQUENCE $(5, [1, 2, 3, 2, 5, 6, 7, 4, 1, 3, 4, 6, 1, 2, 4, 4])$ constraint holds.

| | |
|---|---|
| **Typical** | $|\mathtt{VARIABLES}| > 5$<br>range(VARIABLES.var) $> 2$ |
| **Arg. properties** | Functional dependency: VALUE determined by VARIABLES. |

Figure 3.721: Illustrating the MIN_MAX_DIP_ON_INCREASING_SEQUENCE constraint of the **Example** slot

**Automaton**     Figures 3.722 and 3.723 respectively depict the automaton associated with the constraint
MIN_MAX_DIP_ON_INCREASING_SEQUENCE and its simplified form.

Figure 3.722: Automaton for the MIN_MAX_DIP_ON_INCREASING_SEQUENCE constraint obtained by applying decoration Table 2.35 to the seed transducer of the DIP_ON_INCREASING_SEQUENCE pattern where default is $+\infty$

Figure 3.723: Automaton for the MIN_MAX_DIP_ON_INCREASING_SEQUENCE constraint obtained by applying decoration Table 2.25 to the seed transducer of the DIP_ON_INCREASING_SEQUENCE pattern where default is $+\infty$; $-R_i + R_{i-1} \geq 0$ is a linear invariant.

## 3.321   MIN_MAX_INCREASING

► ▷ ◁ ◄        **DESCRIPTION**        **AUTOMATON**

**Origin**          Based on the INCREASING pattern.

**Constraint**      MIN_MAX_INCREASING(VALUE, VARIABLES)

**Arguments**       VALUE     : dvar
                    VARIABLES : collection(var−dvar)

**Restrictions**    $\text{sv} \leq 1 \vee \text{rv} \leq 1 \Rightarrow \text{VALUE} = +\infty$
                    $\text{VALUE} \geq \text{minv} + 1 ①$
                    $\text{VALUE} = +\infty \vee \text{VALUE} \leq \text{maxv}$
                    required(VARIABLES, var)
                    where
                    minv = minval(VARIABLES.var)
                    maxv = maxval(VARIABLES.var)
                    sv = |VARIABLES|
                    rv = range(VARIABLES.var)

**Purpose**         VALUE is the minimum of all maximum values in each occurrence of the INCREASING
                    pattern in the time-series given by the VARIABLES collection. If the pattern does not
                    occur, VALUE takes the default value $+\infty$.
                    An occurrence of the pattern INCREASING is the subsequence which matches the regular
                    expression $<$.
                    Assume that the occurrence of the pattern INCREASING starts at position $i$ and ends at
                    position $j$. The feature MAX computes the maximum of the values from index $i$ to index
                    $j + 1$.

**Example**         $(3, \langle 4, 3, 5, 5, 2, 1, 1, 3, 3, 4, 6, 6, 3, 1, 3, 3 \rangle)$

                    Figure 3.724 provides an example where the MIN_MAX_INCREASING
                    $(3, [4, 3, 5, 5, 2, 1, 1, 3, 3, 4, 6, 6, 3, 1, 3, 3])$ constraint holds.

**Typical**         $|\text{VARIABLES}| > 1$
                    range(VARIABLES.var) > 1

**Arg. properties** Functional dependency: VALUE determined by VARIABLES.

Figure 3.724: Illustrating the MIN_MAX_INCREASING constraint of the **Example** slot

**Automaton**    Figures 3.725 and 3.726 respectively depict the automaton associated with the constraint MIN_MAX_INCREASING and its simplified form.

Figure 3.725: Automaton for the MIN_MAX_INCREASING constraint obtained by applying decoration Table 2.35 to the seed transducer of the INCREASING pattern where `default` is $+\infty$

Figure 3.726: Automaton for the MIN_MAX_INCREASING constraint obtained by applying decoration Table 2.38 to the seed transducer of the INCREASING pattern where `default` is $+\infty$; $-R_i + R_{i-1} \geq 0$ is a linear invariant.

Table 3.72: Glue matrix for the MIN_MAX_INCREASING constraint defined as the composition of the INCREASING pattern, the feature MAX, and the aggregator min; cells of the glue matrix are coloured with the colour of the constituent to which they are related.

# 3.322   MIN_MAX_INCREASING_SEQUENCE

▶ ▷ ◁ ◀        **DESCRIPTION**        **AUTOMATON**

| **Origin** | Based on the INCREASING_SEQUENCE pattern. |
|---|---|

| **Constraint** | MIN_MAX_INCREASING_SEQUENCE(VALUE, VARIABLES) |
|---|---|

| **Arguments** | VALUE     : dvar |
|---|---|
| | VARIABLES : collection(var−dvar) |

**Restrictions**

$\mathtt{sv} \leq 1 \vee \mathtt{rv} \leq 1 \Rightarrow \mathtt{VALUE} = +\infty$

$\mathtt{VALUE} \geq \mathtt{minv} + 1①$

$\mathtt{VALUE} = +\infty \vee \mathtt{VALUE} \leq \mathtt{maxv}$

required(VARIABLES, var)

where

minv =minval(VARIABLES.var)

maxv =maxval(VARIABLES.var)

sv = |VARIABLES|

rv =range(VARIABLES.var)

**Purpose**

VALUE is the minimum of all maximum values in each occurrence of the INCREAS-ING_SEQUENCE pattern in the time-series given by the VARIABLES collection. If the pattern does not occur, VALUE takes the default value $+\infty$.

An occurrence of the pattern INCREASING_SEQUENCE is the *maximal* subsequence which matches the regular expression $< (< | =)^* < |$.

Assume that the occurrence of the pattern INCREASING_SEQUENCE starts at position $i$ and ends at position $j$. The feature MAX computes the maximum of the values from index $i$ to index $j + 1$.

**Example**

$(3, \langle 4, 3, 5, 5, 2, 1, 1, 3, 3, 4, 6, 6, 3, 1, 3, 3 \rangle)$

Figure 3.727 provides an example where the MIN_MAX_INCREASING_SEQUENCE $(3, [4, 3, 5, 5, 2, 1, 1, 3, 3, 4, 6, 6, 3, 1, 3, 3])$ constraint holds.

**Typical**

$|\mathtt{VARIABLES}| > 1$

range(VARIABLES.var) > 1

**Arg. properties**

Functional dependency: VALUE determined by VARIABLES.

Figure 3.727: Illustrating the MIN_MAX_INCREASING_SEQUENCE constraint of the **Example** slot

Figures 3.728 and 3.729 respectively depict the automaton associated with the constraint MIN_MAX_INCREASING_SEQUENCE and its simplified form.

Figure 3.728: Automaton for the MIN_MAX_INCREASING_SEQUENCE constraint obtained by applying decoration Table 2.35 to the seed transducer of the INCREASING_SEQUENCE pattern where default is $+\infty$

Figure 3.729: Automaton for the MIN_MAX_INCREASING_SEQUENCE constraint obtained by applying decoration Table 2.24 to the seed transducer of the INCREASING_SEQUENCE pattern where default is $+\infty$; $-R_i + R_{i-1} \geq 0$ is a linear invariant.

|   | $s$ | $t$ |
|---|---|---|
| $s$ | $\min(\overrightarrow{C}, \overleftarrow{C})$ | $\min(\overrightarrow{C}, \overleftarrow{C})$ |
| $t$ | $\min(\overrightarrow{C}, \overleftarrow{C})$ | $\max(\overrightarrow{C}, \overleftarrow{C}, \overrightarrow{D}, \overleftarrow{D}, \text{VAR}_i)$ |

Table 3.73: Glue matrix for the MIN_MAX_INCREASING_SEQUENCE constraint defined as the composition of the INCREASING_SEQUENCE pattern , the feature MAX , and the aggregator min ; cells of the glue matrix are coloured with the colour of the constituent to which they are related.

## 3.323 MIN_MAX_INFLEXION

► ▷ ◁ ◀        **DESCRIPTION**        **AUTOMATON**

$< (< | =)^* > | > (> | =)^* <$

**Origin**          Based on the INFLEXION pattern.

**Constraint**      MIN_MAX_INFLEXION(VALUE, VARIABLES)

**Arguments**       VALUE      : dvar
                    VARIABLES  : collection(var−dvar)

**Restrictions**    $\mathtt{sv} \leq 2 \lor \mathtt{rv} \leq 1 \Rightarrow \mathtt{VALUE} = +\infty$
                    $\mathtt{VALUE} \geq \mathtt{minv}①$
                    $\mathtt{VALUE} = +\infty \lor \mathtt{VALUE} \leq \mathtt{maxv}$
                    required(VARIABLES, var)
                    where
                    $\mathtt{minv} =$ minval(VARIABLES.var)
                    $\mathtt{maxv} =$ maxval(VARIABLES.var)
                    $\mathtt{sv} = |\mathtt{VARIABLES}|$
                    $\mathtt{rv} =$ range(VARIABLES.var)

**Purpose**         VALUE is the minimum of all maximum values in each occurrence of the INFLEXION
                    pattern in the time-series given by the VARIABLES collection. If the pattern does not
                    occur, VALUE takes the default value $+\infty$.
                    An occurrence of the pattern INFLEXION is the *maximal* subsequence which matches the
                    regular expression $< (< | =)^* > | > (> | =)^* <$.
                    Assume that the occurrence of the pattern INFLEXION starts at position $i$ and ends at
                    position $j$. The feature MAX computes the maximum of the values from index $i + 1$ to
                    index $j$.

**Example**         $(1, \langle 1, 2, 6, 6, 4, 4, 3, 5, 2, 5, 1, 5, 3, 3, 4, 4\rangle)$

                    Figure 3.730 provides an example where the MIN_MAX_INFLEXION
                    $(1, [1, 2, 6, 6, 4, 4, 3, 5, 2, 5, 1, 5, 3, 3, 4, 4])$ constraint holds.

**Typical**         $|\mathtt{VARIABLES}| > 2$
                    range(VARIABLES.var) $> 1$

**Symmetry**        Items of VARIABLES can be reversed.

**Arg. properties** Functional dependency: VALUE determined by VARIABLES.

Figure 3.730: Illustrating the MIN_MAX_INFLEXION constraint of the **Example** slot

  Figures 3.731 and 3.732 respectively depict the automaton associated with the constraint MIN_MAX_INFLEXION and its simplified form.

Figure 3.731: Automaton for the MIN_MAX_INFLEXION constraint obtained by applying decoration Table 2.35 to the seed transducer of the INFLEXION pattern where `default` is $+\infty$ (transition $r \to t$ has the same accumulators updates as transition $t \to r$)

Figure 3.732: Automaton for the MIN_MAX_INFLEXION constraint obtained by applying decoration Table 2.25 to the seed transducer of the INFLEXION pattern where `default` is $+\infty$ (transition $r \to t$ has the same accumulators updates as transition $t \to r$); $-R_i + R_{i-1} \geq 0$ is a linear invariant.

## 3.324 MIN_MAX_PEAK

▶ ▷ ◁ ◀    **DESCRIPTION**    **AUTOMATON**

**Origin**      Based on the PEAK pattern.

**Constraint**   MIN_MAX_PEAK(VALUE, VARIABLES)

**Arguments**   
```
VALUE      : dvar
VARIABLES  : collection(var−dvar)
```

**Restrictions**

$\text{sv} \leq 2 \vee \text{rv} \leq 1 \Rightarrow \text{VALUE} = +\infty$

$\text{VALUE} \geq \text{minv} + 1\text{①}$

$\text{VALUE} = +\infty \vee \text{VALUE} \leq \text{maxv}$

required(VARIABLES, var)

where

minv = minval(VARIABLES.var)

maxv = maxval(VARIABLES.var)

sv = |VARIABLES|

rv = range(VARIABLES.var)

**Purpose**

VALUE is the minimum of all maximum values in each occurrence of the PEAK pattern in the time-series given by the VARIABLES collection. If the pattern does not occur, VALUE takes the default value $+\infty$.

An occurrence of the pattern PEAK is the *maximal* subsequence which matches the regular expression $< (= | <)^* (> | =)^* >$.

Assume that the occurrence of the pattern PEAK starts at position $i$ and ends at position $j$. The feature MAX computes the maximum of the values from index $i + 1$ to index $j$.

**Example**   $(3, \langle 7, 5, 5, 1, 4, 5, 2, 2, 3, 5, 6, 2, 3, 3, 3, 1 \rangle)$

Figure 3.733 provides an example where the MIN_MAX_PEAK $(3, [7, 5, 5, 1, 4, 5, 2, 2, 3, 5, 6, 2, 3, 3, 3, 1])$ constraint holds.

**Typical**   
$|\text{VARIABLES}| > 2$
range(VARIABLES.var) > 1

**Symmetry**   Items of VARIABLES can be reversed.

**Arg. properties**   Functional dependency: VALUE determined by VARIABLES.

Figure 3.733: Illustrating the MIN_MAX_PEAK constraint of the **Example** slot

**Automaton**     Figures 3.734 and 3.735 respectively depict the automaton associated with the constraint MIN_MAX_PEAK and its simplified form.

Figure 3.734: Automaton for the MIN_MAX_PEAK constraint obtained by applying decoration Table 2.35 to the seed transducer of the PEAK pattern where default is $+\infty$

Figure 3.735: Automaton for the MIN_MAX_PEAK constraint obtained by applying decoration Table 2.37 to the seed transducer of the PEAK pattern where default is $+\infty$; $-R_i + R_{i-1} \geq 0$ is a linear invariant.

| | $s$ | $r$ | $t$ |
|---|---|---|---|
| $s$ | $\min(\overrightarrow{C}, \overleftarrow{C})$ | $\min(\overrightarrow{C}, \overleftarrow{C})$ | $\min(\overrightarrow{C}, \overleftarrow{C})$ |
| $r$ | $\min(\overrightarrow{C}, \overleftarrow{C})$ | $\max(\overrightarrow{D}, \overleftarrow{D}, \mathtt{VAR}_i)$ | $\max(\overleftarrow{C}, \overrightarrow{D}, \overleftarrow{D}, \mathtt{VAR}_i)$ |
| $t$ | $\min(\overrightarrow{C}, \overleftarrow{C})$ | $\max(\overrightarrow{C}, \overrightarrow{D}, \overleftarrow{D}, \mathtt{VAR}_i)$ | $\min(\overrightarrow{C}, \overleftarrow{C})$ |

Table 3.74: Glue matrix for the MIN_MAX_PEAK constraint defined as the composition of the PEAK pattern , the feature MAX , and the aggregator min ; cells of the glue matrix are coloured with the colour of the constituent to which they are related.

## 3.325   MIN_MAX_STRICTLY_DECREASING_SEQUENCE

▶ ▷ ◁ ◀        **DESCRIPTION**            **AUTOMATON**

| | |
|---|---|
| **Origin** | Based on the STRICTLY_DECREASING_SEQUENCE pattern. |
| **Constraint** | MIN_MAX_STRICTLY_DECREASING_SEQUENCE(VALUE, VARIABLES) |
| **Arguments** | VALUE       : dvar<br>VARIABLES : collection(var−dvar) |

**Restrictions**

$sv \leq 1 \lor rv \leq 1 \Rightarrow VALUE = +\infty$

$VALUE \geq minv + 1$ ①

$VALUE = +\infty \lor VALUE \leq maxv$

required(VARIABLES, var)

where

minv = minval(VARIABLES.var)

maxv = maxval(VARIABLES.var)

sv = |VARIABLES|

rv = range(VARIABLES.var)

**Purpose**

VALUE is the minimum of all maximum values in each occurrence of the STRICTLY_DECREASING_SEQUENCE pattern in the time-series given by the VARIABLES collection. If the pattern does not occur, VALUE takes the default value $+\infty$.

An occurrence of the pattern STRICTLY_DECREASING_SEQUENCE is the *maximal* sub-sequence which matches the regular expression $>^{+}$.

Assume that the occurrence of the pattern STRICTLY_DECREASING_SEQUENCE starts at position $i$ and ends at position $j$. The feature MAX computes the maximum of the values from index $i$ to index $j + 1$.

**Example**

$(4, \langle 4, 4, 6, 4, 1, 1, 3, 4, 4, 6, 6, 5, 2, 2, 4, 3 \rangle)$

Figure 3.736 provides an example where the MIN_MAX_STRICTLY_DECREASING_SEQUENCE $(4, [4, 4, 6, 4, 1, 1, 3, 4, 4, 6, 6, 5, 2, 2, 4, 3])$ constraint holds.

| | |
|---|---|
| **Typical** | \|VARIABLES\| > 1<br>range(VARIABLES.var) > 1 |
| **Arg. properties** | Functional dependency: VALUE determined by VARIABLES. |

Figure 3.736: Illustrating the MIN_MAX_STRICTLY_DECREASING_SEQUENCE constraint of the **Example** slot

**Automaton**     Figures 3.737 and 3.738 respectively depict the automaton associated with the constraint MIN_MAX_STRICTLY_DECREASING_SEQUENCE and its simplified form.

Figure 3.737: Automaton for the MIN_MAX_STRICTLY_DECREASING_SEQUENCE constraint obtained by applying decoration Table 2.35 to the seed transducer of the STRICTLY_DECREASING_SEQUENCE pattern where default is $+\infty$

Figure 3.738: Automaton for the MIN_MAX_STRICTLY_DECREASING_SEQUENCE constraint obtained by applying decoration Table 2.37 to the seed transducer of the STRICTLY_DECREASING_SEQUENCE pattern where default is $+\infty$; $-R_i + R_{i-1} \geq 0$ is a linear invariant.

|   | $s$ | $r$ |
|---|---|---|
| $s$ | $\min(\overrightarrow{C}, \overleftarrow{C})$ | $\min(\overrightarrow{C}, \overleftarrow{C})$ |
| $r$ | $\min(\overrightarrow{C}, \overleftarrow{C})$ | $\max(\overrightarrow{C}, \overleftarrow{C}, \overrightarrow{D}, \overleftarrow{D}, \mathtt{VAR}_i)$ |

Table 3.75: Glue matrix for the MIN_MAX_STRICTLY_DECREASING_SEQUENCE constraint defined as the composition of the STRICTLY_DECREASING_SEQUENCE pattern, the feature MAX, and the aggregator min; cells of the glue matrix are coloured with the colour of the constituent to which they are related.

# 3.326 MIN_MAX_STRICTLY_INCREASING_SEQUENCE

▶ ▷ ◁ ◀　　**DESCRIPTION**　　　**AUTOMATON**

| | |
|---|---|
| **Origin** | Based on the STRICTLY_INCREASING_SEQUENCE pattern. |
| **Constraint** | MIN_MAX_STRICTLY_INCREASING_SEQUENCE(VALUE, VARIABLES) |

**Arguments**

```
VALUE      : dvar
VARIABLES  : collection(var−dvar)
```

**Restrictions**

$\mathtt{sv} \leq 1 \lor \mathtt{rv} \leq 1 \Rightarrow \mathtt{VALUE} = +\infty$

$\mathtt{VALUE} \geq \mathtt{minv} + 1$ ①

$\mathtt{VALUE} = +\infty \lor \mathtt{VALUE} \leq \mathtt{maxv}$

required(VARIABLES, var)

where

minv = minval(VARIABLES.var)

maxv = maxval(VARIABLES.var)

sv = |VARIABLES|

rv = range(VARIABLES.var)

**Purpose**

VALUE is the minimum of all maximum values in each occurrence of the STRICTLY_INCREASING_SEQUENCE pattern in the time-series given by the VARIABLES collection. If the pattern does not occur, VALUE takes the default value $+\infty$.

An occurrence of the pattern STRICTLY_INCREASING_SEQUENCE is the *maximal* subsequence which matches the regular expression $<^{+}$.

Assume that the occurrence of the pattern STRICTLY_INCREASING_SEQUENCE starts at position $i$ and ends at position $j$. The feature MAX computes the maximum of the values from index $i$ to index $j + 1$.

**Example**

$$(3, \langle 4, 3, 5, 5, 2, 1, 1, 2, 3, 4, 6, 6, 3, 1, 2, 3 \rangle)$$

Figure 3.739 provides an example where the MIN_MAX_STRICTLY_INCREASING_SEQUENCE $(3, [4, 3, 5, 5, 2, 1, 1, 2, 3, 4, 6, 6, 3, 1, 2, 3])$ constraint holds.

**Typical**

$|\mathtt{VARIABLES}| > 1$

range(VARIABLES.var) > 1

**Arg. properties**

Functional dependency: VALUE determined by VARIABLES.

Figure 3.739: Illustrating the MIN_MAX_STRICTLY_INCREASING_SEQUENCE constraint of the **Example** slot

**Automaton**  Figures 3.740 and 3.741 respectively depict the automaton associated with the constraint MIN_MAX_STRICTLY_INCREASING_SEQUENCE and its simplified form.

Figure 3.740: Automaton for the MIN_MAX_STRICTLY_INCREASING_SEQUENCE constraint obtained by applying decoration Table 2.35 to the seed transducer of the STRICTLY_INCREASING_SEQUENCE pattern where default is $+\infty$

Figure 3.741: Automaton for the MIN_MAX_STRICTLY_INCREASING_SEQUENCE constraint obtained by applying decoration Table 2.24 to the seed transducer of the STRICTLY_INCREASING_SEQUENCE pattern where default is $+\infty$; $-R_i + R_{i-1} \geq 0$ is a linear invariant.

|   | $s$ | $r$ |
|---|-----|-----|
| $s$ | $\min(\overrightarrow{C}, \overleftarrow{C})$ | $\min(\overrightarrow{C}, \overleftarrow{C})$ |
| $r$ | $\min(\overrightarrow{C}, \overleftarrow{C})$ | $\max(\overrightarrow{C}, \overleftarrow{C}, \overrightarrow{D}, \overleftarrow{D}, \text{VAR}_i)$ |

Table 3.76: Glue matrix for the MIN_MAX_STRICTLY_INCREASING_SEQUENCE constraint defined as the composition of the STRICTLY_INCREASING_SEQUENCE pattern, the feature MAX, and the aggregator min; cells of the glue matrix are coloured with the colour of the constituent to which they are related.

# 3.327   MIN_MAX_SUMMIT

► ▷ ◁ ◀        **DESCRIPTION**        **AUTOMATON**

$(< | < (= | <)^* <)(> | > (= | >)^* >)$

| | |
|---|---|
| **Origin** | Based on the SUMMIT pattern. |
| **Constraint** | MIN_MAX_SUMMIT(VALUE, VARIABLES) |
| **Arguments** | VALUE      : dvar <br> VARIABLES : collection(var−dvar) |
| **Restrictions** | $sv \le 2 \vee rv \le 1 \Rightarrow$ VALUE $= +\infty$ <br> VALUE $\ge$ minv $+ 1$① <br> VALUE $= +\infty \vee$ VALUE $\le$ maxv <br> required(VARIABLES, var) <br> where <br> minv =minval(VARIABLES.var) <br> maxv =maxval(VARIABLES.var) <br> sv = \|VARIABLES\| <br> rv =range(VARIABLES.var) |

VALUE $\ge$ 1

| | |
|---|---|
| **Purpose** | VALUE is the minimum of all maximum values in each occurrence of the SUMMIT pattern in the time-series given by the VARIABLES collection. If the pattern does not occur, VALUE takes the default value $+\infty$. <br> An occurrence of the pattern SUMMIT is the *maximal* subsequence which matches the regular expression $(< | < (= | <)^* <)(> | > (= | >)^* >)$. <br> Assume that the occurrence of the pattern SUMMIT starts at position $i$ and ends at position $j$. The feature MAX computes the maximum of the values from index $i + 1$ to index $j$. |

| | |
|---|---|
| **Example** | $(3, \langle 7, 1, 5, 4, 4, 3, 3, 4, 6, 6, 2, 3, 4, 2, 3, 1 \rangle)$ |

Figure 3.742 provides an example where the MIN_MAX_SUMMIT $(3, [7, 1, 5, 4, 4, 3, 3, 4, 6, 6, 2, 3, 4, 2, 3, 1])$ constraint holds.

| | |
|---|---|
| **Typical** | \|VARIABLES\| $> 2$ <br> range(VARIABLES.var) $> 1$ |
| **Symmetry** | Items of VARIABLES can be reversed. |
| **Arg. properties** | Functional dependency: VALUE determined by VARIABLES. |

Figure 3.742: Illustrating the MIN_MAX_SUMMIT constraint of the **Example** slot

**Automaton**  Figures 3.743 and 3.744 respectively depict the automaton associated with the constraint MIN_MAX_SUMMIT and its simplified form.

Figure 3.743: Automaton for the MIN_MAX_SUMMIT constraint obtained by applying decoration Table 2.35 to the seed transducer of the SUMMIT pattern where default is $+\infty$ (transition $u \rightarrow r$ has the same accumulator update as transition $r \rightarrow u$)

| | $s$ | $r$ | $t$ | $u$ |
|---|---|---|---|---|
| $s$ | $\min(\overrightarrow{C}, \overleftarrow{C})$ | $\min(\overrightarrow{C}, \overleftarrow{C})$ | $\min(\overrightarrow{C}, \overleftarrow{C})$ | $\min(\overrightarrow{C}, \overleftarrow{C})$ |
| $r$ | $\min(\overrightarrow{C}, \overleftarrow{C})$ | $\max(\overrightarrow{D}, \overleftarrow{D}, \text{VAR}_i)$ | $\max(\overleftarrow{C}, \overrightarrow{D}, \overleftarrow{D}, \text{VAR}_i)$ | $\min(\overrightarrow{C}, \overleftarrow{C})$ |
| $t$ | $\min(\overrightarrow{C}, \overleftarrow{C})$ | $\max(\overrightarrow{C}, \overrightarrow{D}, \overleftarrow{D}, \text{VAR}_i)$ | $\min(\overrightarrow{C}, \overleftarrow{C})$ | $\max(\overrightarrow{C}, \overrightarrow{D}, \overleftarrow{D}, \text{VAR}_i)$ |
| $u$ | $\min(\overrightarrow{C}, \overleftarrow{C})$ | $\min(\overrightarrow{C}, \overleftarrow{C})$ | $\max(\overleftarrow{C}, \overrightarrow{D}, \overleftarrow{D}, \text{VAR}_i)$ | $\min(\overrightarrow{C}, \overleftarrow{C})$ |

Table 3.77: Glue matrix for the MIN_MAX_SUMMIT constraint defined as the composition of the SUMMIT pattern , the feature MAX , and the aggregator min ; cells of the glue matrix are coloured with the colour of the constituent to which they are related.

Figure 3.744: Automaton for the MIN_MAX_SUMMIT constraint obtained by applying decoration Table 2.37 to the seed transducer of the SUMMIT pattern where default is $+\infty$ (transition $u \to r$ has the same accumulator update as transition $r \to u$); $-R_i + R_{i-1} \geq 0$ is a linear invariant.

## 3.328  MIN_MAX_ZIGZAG

▶ ▷ ◁ ◀          **DESCRIPTION**          **AUTOMATON**

| | | |
|---|---|---|
| **Origin** | Based on the ZIGZAG pattern. | |

| | |
|---|---|
| **Constraint** | MIN_MAX_ZIGZAG(VALUE, VARIABLES) |
| **Arguments** | VALUE      : dvar |
| | VARIABLES  : collection(var−dvar) |
| **Restrictions** | $\texttt{sv} \leq 3 \vee \texttt{rv} \leq 1 \Rightarrow \texttt{VALUE} = +\infty$ |
| | $\texttt{VALUE} \geq \texttt{minv} + 1 \,①$ |
| | $\texttt{VALUE} = +\infty \vee \texttt{VALUE} \leq \texttt{maxv}$ |
| | required(VARIABLES, var) |
| | where |
| | $\texttt{minv} = $ minval(VARIABLES.var) |
| | $\texttt{maxv} = $ maxval(VARIABLES.var) |
| | $\texttt{sv} = |\texttt{VARIABLES}|$ |
| | $\texttt{rv} = $ range(VARIABLES.var) |
| **Purpose** | VALUE is the minimum of all maximum values in each occurrence of the ZIGZAG pattern in the time-series given by the VARIABLES collection. If the pattern does not occur, VALUE takes the default value $+\infty$. |
| | An occurrence of the pattern ZIGZAG is the *maximal* subsequence which matches the regular expression $(<>)^+(< | <>) | (><)^+(> | ><)$. |
| | Assume that the occurrence of the pattern ZIGZAG starts at position $i$ and ends at position $j$. The feature MAX computes the maximum of the values from index $i + 1$ to index $j$. |
| **Example** | $(3, \langle 4, 1, 3, 1, 4, 6, 1, 5, 5, 2, 7, 2, 3, 1, 6, 1 \rangle)$ |

Figure   3.745   provides   an   example   where   the   MIN_MAX_ZIGZAG $(3, [4, 1, 3, 1, 4, 6, 1, 5, 5, 2, 7, 2, 3, 1, 6, 1])$ constraint holds.

| | |
|---|---|
| **Typical** | $|\texttt{VARIABLES}| > 3$ |
| | range(VARIABLES.var) $> 1$ |
| **Symmetry** | Items of VARIABLES can be reversed. |
| **Arg. properties** | Functional dependency: VALUE determined by VARIABLES. |

Figure 3.745: Illustrating the MIN_MAX_ZIGZAG constraint of the **Example** slot

**Automaton**          Figures 3.746 and 3.747 respectively depict the automaton associated with the constraint
                       MIN_MAX_ZIGZAG and its simplified form.

Figure 3.746: Automaton for the MIN_MAX_ZIGZAG constraint obtained by applying decoration Table 2.35 to the seed transducer of the ZIGZAG pattern where default is $+\infty$; (1) missing transitions from $a, b, c, d, e, f$ to $s$ are labelled by $=$; (2) on transitions from $b, c, e, f$ to $s$ the accumulator $D$ is reset to its initial value; (3) on transitions from $c, f$ to $s$ the accumulator $R$ is updated wrt $C$ and the accumulator $C$ is reset to its initial value

Figure 3.747: Automaton for the MIN_MAX_ZIGZAG constraint obtained by applying decoration Table 2.23 to the seed transducer of the ZIGZAG pattern where default is $+\infty$; missing transitions from $a, b, c, d, e, f$ to $s$ are labelled by $=$; (2) on transitions from $b, c, e, f$ to $s$ the accumulator $D$ is reset to its initial value; (3) on transitions from $c, f$ to $s$ the accumulator $R$ is updated wrt $C$ and the accumulator $C$ is reset to its initial value.; $-R_i + R_{i-1} \geq 0$ is a linear invariant.

| | $s$ | $a$ | $b$ | $c$ | $d$ | $e$ | $f$ |
|---|---|---|---|---|---|---|---|
| $s$ | $\min(\overrightarrow{C}, \overleftarrow{C})$ | $\min(\overrightarrow{C}, \overleftarrow{C})$ | $\min(\overrightarrow{C}, \overleftarrow{C})$ | $\min(\overrightarrow{C}, \overleftarrow{C})$ | $\min(\overrightarrow{C}, \overleftarrow{C})$ | $\min(\overrightarrow{C}, \overleftarrow{C})$ | $\min(\overrightarrow{C}, \overleftarrow{C})$ |
| $a$ | $\min(\overrightarrow{C}, \overleftarrow{C})$ | $\min(\overrightarrow{C}, \overleftarrow{C})$ | $\min(\overrightarrow{C}, \overleftarrow{C})$ | $\max(\overrightarrow{C}, \overrightarrow{D}, \overleftarrow{D}, \mathrm{VAR}_i)$ | $\min(\overrightarrow{C}, \overleftarrow{C})$ | $\max(\overrightarrow{D}, \overleftarrow{D}, \mathrm{VAR}_i)$ | $\min(\overrightarrow{C}, \overleftarrow{C})$ |
| $b$ | $\min(\overrightarrow{C}, \overleftarrow{C})$ | $\min(\overrightarrow{C}, \overleftarrow{C})$ | $\max(\overrightarrow{D}, \overleftarrow{D}, \mathrm{VAR}_i)$ | $\min(\overrightarrow{C}, \overleftarrow{C})$ | $\max(\overrightarrow{D}, \overleftarrow{D}, \mathrm{VAR}_i)$ | $\min(\overrightarrow{C}, \overleftarrow{C})$ | $\max(\overleftarrow{C}, \overrightarrow{D}, \overleftarrow{D}, \mathrm{VAR}_i)$ |
| $c$ | $\min(\overrightarrow{C}, \overleftarrow{C})$ | $\max(\overleftarrow{C}, \overrightarrow{D}, \overleftarrow{D}, \mathrm{VAR}_i)$ | $\min(\overrightarrow{C}, \overleftarrow{C})$ | $\max(\overrightarrow{C}, \overleftarrow{C}, \overrightarrow{D}, \overleftarrow{D}, \mathrm{VAR}_i)$ | $\min(\overrightarrow{C}, \overleftarrow{C})$ | $\max(\overrightarrow{C}, \overrightarrow{D}, \overleftarrow{D}, \mathrm{VAR}_i)$ | $\min(\overrightarrow{C}, \overleftarrow{C})$ |
| $d$ | $\min(\overrightarrow{C}, \overleftarrow{C})$ | $\min(\overrightarrow{C}, \overleftarrow{C})$ | $\max(\overrightarrow{D}, \overleftarrow{D}, \mathrm{VAR}_i)$ | $\min(\overrightarrow{C}, \overleftarrow{C})$ | $\min(\overrightarrow{C}, \overleftarrow{C})$ | $\min(\overrightarrow{C}, \overleftarrow{C})$ | $\max(\overleftarrow{C}, \overrightarrow{D}, \overleftarrow{D}, \mathrm{VAR}_i)$ |
| $e$ | $\min(\overrightarrow{C}, \overleftarrow{C})$ | $\max(\overrightarrow{D}, \overleftarrow{D}, \mathrm{VAR}_i)$ | $\min(\overrightarrow{C}, \overleftarrow{C})$ | $\max(\overleftarrow{C}, \overrightarrow{D}, \overleftarrow{D}, \mathrm{VAR}_i)$ | $\min(\overrightarrow{C}, \overleftarrow{C})$ | $\max(\overrightarrow{D}, \overleftarrow{D}, \mathrm{VAR}_i)$ | $\min(\overrightarrow{C}, \overleftarrow{C})$ |
| $f$ | $\min(\overrightarrow{C}, \overleftarrow{C})$ | $\min(\overrightarrow{C}, \overleftarrow{C})$ | $\max(\overrightarrow{C}, \overrightarrow{D}, \overleftarrow{D}, \mathrm{VAR}_i)$ | $\min(\overrightarrow{C}, \overleftarrow{C})$ | $\max(\overrightarrow{C}, \overleftarrow{C}, \overrightarrow{D}, \overleftarrow{D}, \mathrm{VAR}_i)$ | $\min(\overrightarrow{C}, \overleftarrow{C})$ | $\max(\overrightarrow{C}, \overleftarrow{C}, \overrightarrow{D}, \overleftarrow{D}, \mathrm{VAR}_i)$ |

Table 3.78: Glue matrix for the MIN_MAX_ZIGZAG constraint defined as the composition of the ZIGZAG pattern , the feature MAX , and the aggregator min ; cells of the glue matrix are coloured with the colour of the constituent to which they are related.

# 3.329   MIN_MIN_BUMP_ON_DECREASING_SEQUENCE

▶ ▷ ◁ ◀      **DESCRIPTION**      **AUTOMATON**

| | |
|---|---|
| **Origin** | Based on the BUMP_ON_DECREASING_SEQUENCE pattern. |
| **Constraint** | MIN_MIN_BUMP_ON_DECREASING_SEQUENCE(VALUE, VARIABLES) |
| **Arguments** | VALUE      :   dvar<br>VARIABLES : collection(var−dvar) |

**Restrictions**

$\mathtt{sv} \leq 5 \lor \mathtt{rv} \leq 2 \Rightarrow \mathtt{VALUE} = +\infty$

$\mathtt{VALUE} \geq \mathtt{minv}$ ⓘ

$\mathtt{VALUE} = +\infty \lor \mathtt{VALUE} \leq \mathtt{maxv} - 2$

required(VARIABLES, var)

where

minv = minval(VARIABLES.var)

maxv = maxval(VARIABLES.var)

sv = |VARIABLES|

rv = range(VARIABLES.var)

**Purpose**

VALUE is the minimum of all minimum values in each occurrence of the BUMP_ON_DECREASING_SEQUENCE pattern in the time-series given by the VARIABLES collection. If the pattern does not occur, VALUE takes the default value $+\infty$.

An occurrence of the pattern BUMP_ON_DECREASING_SEQUENCE is the subsequence which matches the regular expression >><>>.

Assume that the occurrence of the pattern BUMP_ON_DECREASING_SEQUENCE starts at position $i$ and ends at position $j$. The feature MIN computes the minimum of the values from index $i + 2$ to index $j$.

**Example**

$(2, \langle 7, 6, 5, 6, 5, 4, 1, 4, 7, 5, 4, 2, 5, 4, 3, 3 \rangle)$

Figure 3.748 provides an example where the MIN_MIN_BUMP_ON_DECREASING_SEQUENCE $(2, [7, 6, 5, 6, 5, 4, 1, 4, 7, 5, 4, 2, 5, 4, 3, 3])$ constraint holds.

**Typical**

$|\mathtt{VARIABLES}| > 5$

range(VARIABLES.var) > 2

**Arg. properties**

Functional dependency: VALUE determined by VARIABLES.

Figure 3.748: Illustrating the MIN_MIN_BUMP_ON_DECREASING_SEQUENCE constraint of the **Example** slot

**Automaton**       Figures 3.749 and 3.750 respectively depict the automaton associated with the constraint MIN_MIN_BUMP_ON_DECREASING_SEQUENCE and its simplified form.

Figure 3.749: Automaton for the MIN_MIN_BUMP_ON_DECREASING_SEQUENCE constraint obtained by applying decoration Table 2.35 to the seed transducer of the BUMP_ON_DECREASING_SEQUENCE pattern where default is $+\infty$

Figure 3.750: Automaton for the MIN_MIN_BUMP_ON_DECREASING_SEQUENCE constraint obtained by applying decoration Table 2.25 to the seed transducer of the BUMP_ON_DECREASING_SEQUENCE pattern where default is $+\infty$; $-R_i + R_{i-1} \geq 0$ is a linear invariant.

# 3.330   MIN_MIN_DECREASING

► ▷ ◁ ◀        **DESCRIPTION**            **AUTOMATON**

**Origin**              Based on the DECREASING pattern.

**Constraint**          MIN_MIN_DECREASING(VALUE, VARIABLES)

**Arguments**           VALUE      : dvar
                        VARIABLES  : collection(var−dvar)

**Restrictions**        $sv \leq 1 \lor rv \leq 1 \Rightarrow$ VALUE $= +\infty$
                        VALUE $\geq$ minv①
                        VALUE $= +\infty \lor$ VALUE $\leq$ maxv $- 1$
                        required(VARIABLES, var)
                        where
                        minv =minval(VARIABLES.var)
                        maxv =maxval(VARIABLES.var)
                        $sv = |$VARIABLES$|$
                        $rv =$range(VARIABLES.var)

**Purpose**             VALUE is the minimum of all minimum values in each occurrence of the DECREASING
                        pattern in the time-series given by the VARIABLES collection. If the pattern does not
                        occur, VALUE takes the default value $+\infty$.
                        An occurrence of the pattern DECREASING is the subsequence which matches the regular
                        expression >.
                        Assume that the occurrence of the pattern DECREASING starts at position $i$ and ends at
                        position $j$. The feature MIN computes the minimum of the values from index $i$ to index
                        $j + 1$.

**Example**             $(1, \langle 3, 4, 2, 2, 5, 6, 6, 4, 4, 3, 1, 1, 4, 6, 4, 4 \rangle)$

                        Figure 3.751 provides an example where the MIN_MIN_DECREASING
                        $(1, [3, 4, 2, 2, 5, 6, 6, 4, 4, 3, 1, 1, 4, 6, 4, 4])$ constraint holds.

**Typical**             $|$VARIABLES$| > 1$
                        range(VARIABLES.var) $> 1$

**Arg. properties**     Functional dependency: VALUE determined by VARIABLES.

Figure 3.751: Illustrating the MIN_MIN_DECREASING constraint of the **Example** slot

**Automaton**  Figures 3.752 and 3.753 respectively depict the automaton associated with the constraint MIN_MIN_DECREASING and its simplified form.

Figure 3.752: Automaton for the MIN_MIN_DECREASING constraint obtained by applying decoration Table 2.35 to the seed transducer of the DECREASING pattern where `default` is $+\infty$

Figure 3.753: Automaton for the MIN_MIN_DECREASING constraint obtained by applying decoration Table 2.38 to the seed transducer of the DECREASING pattern where `default` is $+\infty$; $-R_i + R_{i-1} \geq 0$ is a linear invariant.

Table 3.79: Glue matrix for the MIN_MIN_DECREASING constraint defined as the composition of the DECREASING pattern , the feature MIN , and the aggregator min ; cells of the glue matrix are coloured with the colour of the constituent to which they are related.

# 3.331 MIN_MIN_DECREASING_SEQUENCE

▶ ▷ ◁ ◀          **DESCRIPTION**          **AUTOMATON**

| **Origin** | Based on the DECREASING_SEQUENCE pattern. |

| **Constraint** | MIN_MIN_DECREASING_SEQUENCE(VALUE, VARIABLES) |

| **Arguments** | VALUE : dvar |
| | VARIABLES : collection(var−dvar) |

**Restrictions**

$\mathtt{sv} \leq 1 \vee \mathtt{rv} \leq 1 \Rightarrow \mathtt{VALUE} = +\infty$

$\mathtt{VALUE} \geq \mathtt{minv}①$

$\mathtt{VALUE} = +\infty \vee \mathtt{VALUE} \leq \mathtt{maxv} - 1$

required(VARIABLES, var)

where

minv =minval(VARIABLES.var)

maxv =maxval(VARIABLES.var)

sv = |VARIABLES|

rv =range(VARIABLES.var)

**Purpose**

VALUE is the minimum of all minimum values in each occurrence of the DECREAS-ING_SEQUENCE pattern in the time-series given by the VARIABLES collection. If the pattern does not occur, VALUE takes the default value $+\infty$.

An occurrence of the pattern DECREASING_SEQUENCE is the *maximal* subsequence which matches the regular expression > (> | =)* > | >.

Assume that the occurrence of the pattern DECREASING_SEQUENCE starts at position $i$ and ends at position $j$. The feature MIN computes the minimum of the values from index $i$ to index $j + 1$.

**Example**

$(1, \langle 3, 4, 2, 2, 5, 6, 6, 4, 4, 3, 1, 1, 4, 6, 4, 4 \rangle)$

Figure 3.754 provides an example where the MIN_MIN_DECREASING_SEQUENCE $(1, [3, 4, 2, 2, 5, 6, 6, 4, 4, 3, 1, 1, 4, 6, 4, 4])$ constraint holds.

| **Typical** | |VARIABLES| > 1 |
| | range(VARIABLES.var) > 1 |

| **Arg. properties** | Functional dependency: VALUE determined by VARIABLES. |

Figure 3.754: Illustrating the MIN_MIN_DECREASING_SEQUENCE constraint of the **Example** slot

  Figures 3.755 and 3.756 respectively depict the automaton associated with the constraint MIN_MIN_DECREASING_SEQUENCE and its simplified form.

Figure 3.755: Automaton for the MIN_MIN_DECREASING_SEQUENCE constraint obtained by applying decoration Table 2.35 to the seed transducer of the DECREASING_SEQUENCE pattern where default is $+\infty$

Figure 3.756: Automaton for the MIN_MIN_DECREASING_SEQUENCE constraint obtained by applying decoration Table 2.38 to the seed transducer of the DECREASING_SEQUENCE pattern where default is $+\infty$; $-R_i + R_{i-1} \geq 0$ is a linear invariant.

|   | $s$ | $t$ |
|---|---|---|
| $s$ | $\min(\overrightarrow{C}, \overleftarrow{C})$ | $\min(\overrightarrow{C}, \overleftarrow{C})$ |
| $t$ | $\min(\overrightarrow{C}, \overleftarrow{C})$ | $\min(\overrightarrow{C}, \overleftarrow{C}, \overrightarrow{D}, \overleftarrow{D}, \mathtt{VAR}_i)$ |

Table 3.80: Glue matrix for the MIN_MIN_DECREASING_SEQUENCE constraint defined as the composition of the DECREASING_SEQUENCE pattern , the feature MIN , and the aggregator min ; cells of the glue matrix are coloured with the colour of the constituent to which they are related.

# 3.332   MIN_MIN_DIP_ON_INCREASING_SEQUENCE

▶ ▷ ◁ ◀        **DESCRIPTION**          **AUTOMATON**

| | |
|---|---|
| **Origin** | Based on the DIP_ON_INCREASING_SEQUENCE pattern. |
| **Constraint** | MIN_MIN_DIP_ON_INCREASING_SEQUENCE(VALUE, VARIABLES) |
| **Arguments** | VALUE      : dvar<br>VARIABLES  : collection(var−dvar) |

**Restrictions**

$\text{sv} \leq 5 \vee \text{rv} \leq 2 \Rightarrow \text{VALUE} = +\infty$

$\text{VALUE} \geq \text{minv}①$

$\text{VALUE} = +\infty \vee \text{VALUE} \leq \text{maxv} - 2$

required(VARIABLES, var)

where

minv = minval(VARIABLES.var)

maxv = maxval(VARIABLES.var)

sv = |VARIABLES|

rv = range(VARIABLES.var)

**Purpose**

VALUE is the minimum of all minimum values in each occurrence of the DIP_ON_INCREASING_SEQUENCE pattern in the time-series given by the VARIABLES collection. If the pattern does not occur, VALUE takes the default value $+\infty$.

An occurrence of the pattern DIP_ON_INCREASING_SEQUENCE is the subsequence which matches the regular expression <<><<.

Assume that the occurrence of the pattern DIP_ON_INCREASING_SEQUENCE starts at position $i$ and ends at position $j$. The feature MIN computes the minimum of the values from index $i + 2$ to index $j$.

**Example**

$(1, \langle 1, 2, 3, 2, 5, 6, 7, 4, 1, 3, 4, 6, 1, 2, 4, 4 \rangle)$

Figure 3.757 provides an example where the MIN_MIN_DIP_ON_INCREASING_SEQUENCE $(1, [1, 2, 3, 2, 5, 6, 7, 4, 1, 3, 4, 6, 1, 2, 4, 4])$ constraint holds.

| | |
|---|---|
| **Typical** | \|VARIABLES\| > 5<br>range(VARIABLES.var) > 2 |
| **Arg. properties** | Functional dependency: VALUE determined by VARIABLES. |

Figure 3.757: Illustrating the MIN_MIN_DIP_ON_INCREASING_SEQUENCE constraint of the **Example** slot

**Automaton**    Figures 3.758 and 3.759 respectively depict the automaton associated with the constraint MIN_MIN_DIP_ON_INCREASING_SEQUENCE and its simplified form.

Figure 3.758: Automaton for the MIN_MIN_DIP_ON_INCREASING_SEQUENCE constraint obtained by applying decoration Table 2.35 to the seed transducer of the DIP_ON_INCREASING_SEQUENCE pattern where default is $+\infty$

Figure 3.759: Automaton for the MIN_MIN_DIP_ON_INCREASING_SEQUENCE constraint obtained by applying decoration Table 2.27 to the seed transducer of the DIP_ON_INCREASING_SEQUENCE pattern where `default` is $+\infty$; $-R_i + R_{i-1} \geq 0$ is a linear invariant.

## 3.333  MIN_MIN_GORGE

$(> | > (= | >)^*>)(< | < (= | <)^* <)$

| | |
|---|---|
| **Origin** | Based on the GORGE pattern. |
| **Constraint** | MIN_MIN_GORGE(VALUE, VARIABLES) |
| **Arguments** | VALUE      : dvar<br>VARIABLES : collection(var−dvar) |
| **Restrictions** | $\text{sv} \leq 2 \vee \text{rv} \leq 1 \Rightarrow \text{VALUE} = +\infty$<br>$\text{VALUE} \geq \text{minv}①$<br>$\text{VALUE} = +\infty \vee \text{VALUE} \leq \text{maxv} - 1$<br>required(VARIABLES, var)<br>where<br>$\text{minv} = \text{minval}(\text{VARIABLES.var})$<br>$\text{maxv} = \text{maxval}(\text{VARIABLES.var})$<br>$\text{sv} = |\text{VARIABLES}|$<br>$\text{rv} = \text{range}(\text{VARIABLES.var})$ |

VALUE ≥ 0

| | |
|---|---|
| **Purpose** | VALUE is the minimum of all minimum values in each occurrence of the GORGE pattern in the time-series given by the VARIABLES collection. If the pattern does not occur, VALUE takes the default value $+\infty$.<br>An occurrence of the pattern GORGE is the *maximal* subsequence which matches the regular expression $(> | > (= | >)^* >)(< | < (= | <)^* <)$.<br>Assume that the occurrence of the pattern GORGE starts at position $i$ and ends at position $j$. The feature MIN computes the minimum of the values from index $i + 1$ to index $j$. |

| | |
|---|---|
| **Example** | $(3, \langle 1, 7, 3, 4, 4, 5, 5, 4, 2, 2, 6, 5, 4, 6, 5, 7 \rangle)$ |

Figure 3.760 provides an example where the MIN_MIN_GORGE $(3, [1, 7, 3, 4, 4, 5, 5, 4, 2, 2, 6, 5, 4, 6, 5, 7])$ constraint holds.

| | |
|---|---|
| **Typical** | $|\text{VARIABLES}| > 2$<br>range(VARIABLES.var) > 1 |
| **Symmetry** | Items of VARIABLES can be reversed. |
| **Arg. properties** | Functional dependency: VALUE determined by VARIABLES. |

Figure 3.760: Illustrating the MIN_MIN_GORGE constraint of the **Example** slot

**Automaton**     Figures 3.761 and 3.762 respectively depict the automaton associated with the constraint
MIN_MIN_GORGE and its simplified form.

Figure 3.761: Automaton for the MIN_MIN_GORGE constraint obtained by applying
decoration Table 2.35 to the seed transducer of the GORGE pattern where default is
$+\infty$ (transition $u \rightarrow r$ has the same accumulator update as transition $r \rightarrow u$)

|   | $s$ | $r$ | $t$ | $u$ |
|---|---|---|---|---|
| $s$ | $\min(\overrightarrow{C}, \overleftarrow{C})$ | $\min(\overrightarrow{C}, \overleftarrow{C})$ | $\min(\overrightarrow{C}, \overleftarrow{C})$ | $\min(\overrightarrow{C}, \overleftarrow{C})$ |
| $r$ | $\min(\overrightarrow{C}, \overleftarrow{C})$ | $\min(\overrightarrow{D}, \overleftarrow{D}, \text{VAR}_i)$ | $\min(\overleftarrow{C}, \overrightarrow{D}, \overleftarrow{D}, \text{VAR}_i)$ | $\min(\overrightarrow{C}, \overleftarrow{C})$ |
| $t$ | $\min(\overrightarrow{C}, \overleftarrow{C})$ | $\min(\overrightarrow{C}, \overrightarrow{D}, \overleftarrow{D}, \text{VAR}_i)$ | $\min(\overrightarrow{C}, \overleftarrow{C})$ | $\min(\overrightarrow{C}, \overrightarrow{D}, \overleftarrow{D}, \text{VAR}_i)$ |
| $u$ | $\min(\overrightarrow{C}, \overleftarrow{C})$ | $\min(\overrightarrow{C}, \overleftarrow{C})$ | $\min(\overleftarrow{C}, \overrightarrow{D}, \overleftarrow{D}, \text{VAR}_i)$ | $\min(\overrightarrow{C}, \overleftarrow{C})$ |

Table 3.81: Glue matrix for the MIN_MIN_GORGE constraint defined as the composition
of the GORGE pattern , the feature MIN , and the aggregator min ; cells of the glue
matrix are coloured with the colour of the constituent to which they are related.

Figure 3.762: Automaton for the MIN_MIN_GORGE constraint obtained by applying decoration Table 2.37 to the seed transducer of the GORGE pattern where default is $+\infty$; $-R_i + R_{i-1} \geq 0$ is a linear invariant.

# 3.334   MIN_MIN_INCREASING

   **DESCRIPTION**          **AUTOMATON**

| | |
|---|---|
| **Origin** | Based on the INCREASING pattern. |
| **Constraint** | MIN_MIN_INCREASING(VALUE, VARIABLES) |
| **Arguments** | VALUE      : dvar<br>VARIABLES  : collection(var−dvar) |

**Restrictions**

$\text{sv} \leq 1 \lor \text{rv} \leq 1 \Rightarrow \text{VALUE} = +\infty$

$\text{VALUE} \geq \text{minv}①$

$\text{VALUE} = +\infty \lor \text{VALUE} \leq \text{maxv} - 1$

required(VARIABLES, var)

where

$\text{minv} = \text{minval}(\text{VARIABLES.var})$

$\text{maxv} = \text{maxval}(\text{VARIABLES.var})$

$\text{sv} = |\text{VARIABLES}|$

$\text{rv} = \text{range}(\text{VARIABLES.var})$

**Purpose**

VALUE is the minimum of all minimum values in each occurrence of the INCREASING pattern in the time-series given by the VARIABLES collection. If the pattern does not occur, VALUE takes the default value $+\infty$.

An occurrence of the pattern INCREASING is the subsequence which matches the regular expression $<$.

Assume that the occurrence of the pattern INCREASING starts at position $i$ and ends at position $j$. The feature MIN computes the minimum of the values from index $i$ to index $j + 1$.

**Example**

$$(1, \langle 4, 3, 5, 5, 2, 1, 1, 3, 3, 4, 6, 6, 3, 1, 3, 3 \rangle)$$

Figure 3.763 provides an example where the MIN_MIN_INCREASING $(1, [4, 3, 5, 5, 2, 1, 1, 3, 3, 4, 6, 6, 3, 1, 3, 3])$ constraint holds.

**Typical**

$|\text{VARIABLES}| > 1$

$\text{range}(\text{VARIABLES.var}) > 1$

**Arg. properties**

Functional dependency: VALUE determined by VARIABLES.

Figure 3.763: Illustrating the MIN_MIN_INCREASING constraint of the **Example** slot

**Automaton**     Figures 3.764 and 3.765 respectively depict the automaton associated with the constraint MIN_MIN_INCREASING and its simplified form.

Figure 3.764: Automaton for the MIN_MIN_INCREASING constraint obtained by applying decoration Table 2.35 to the seed transducer of the INCREASING pattern where `default` is $+\infty$

Figure 3.765: Automaton for the MIN_MIN_INCREASING constraint obtained by applying decoration Table 2.37 to the seed transducer of the INCREASING pattern where `default` is $+\infty$; $-R_i + R_{i-1} \geq 0$ is a linear invariant.

Table 3.82: Glue matrix for the MIN_MIN_INCREASING constraint defined as the composition of the INCREASING pattern , the feature MIN , and the aggregator min ; cells of the glue matrix are coloured with the colour of the constituent to which they are related.

# 3.335    MIN_MIN_INCREASING_SEQUENCE

▶ ▷ ◁ ◀    **DESCRIPTION**        **AUTOMATON**

| **Origin** | Based on the INCREASING_SEQUENCE pattern. |
|---|---|

| **Constraint** | MIN_MIN_INCREASING_SEQUENCE(VALUE, VARIABLES) |
|---|---|

| **Arguments** | VALUE      : dvar |
|---|---|
| | VARIABLES : collection(var−dvar) |

**Restrictions**

$\text{sv} \leq 1 \vee \text{rv} \leq 1 \Rightarrow \text{VALUE} = +\infty$

$\text{VALUE} \geq \text{minv}①$

$\text{VALUE} = +\infty \vee \text{VALUE} \leq \text{maxv} - 1$

required(VARIABLES, var)

where

minv = minval(VARIABLES.var)

maxv = maxval(VARIABLES.var)

sv = |VARIABLES|

rv = range(VARIABLES.var)

**Purpose**

VALUE is the minimum of all minimum values in each occurrence of the INCREAS-ING_SEQUENCE pattern in the time-series given by the VARIABLES collection. If the pattern does not occur, VALUE takes the default value $+\infty$.

An occurrence of the pattern INCREASING_SEQUENCE is the *maximal* subsequence which matches the regular expression $< (< | =)^* < | <$.

Assume that the occurrence of the pattern INCREASING_SEQUENCE starts at position $i$ and ends at position $j$. The feature MIN computes the minimum of the values from index $i$ to index $j + 1$.

**Example**

$(1, \langle 4, 3, 5, 5, 2, 1, 1, 3, 3, 4, 6, 6, 3, 1, 3, 3 \rangle)$

Figure 3.766 provides an example where the MIN_MIN_INCREASING_SEQUENCE $(1, [4, 3, 5, 5, 2, 1, 1, 3, 3, 4, 6, 6, 3, 1, 3, 3])$ constraint holds.

**Typical**

|VARIABLES| > 1

range(VARIABLES.var) > 1

**Arg. properties**

Functional dependency: VALUE determined by VARIABLES.

Figure 3.766: Illustrating the MIN_MIN_INCREASING_SEQUENCE constraint of the **Example** slot

 Figures 3.767 and 3.768 respectively depict the automaton associated with the constraint MIN_MIN_INCREASING_SEQUENCE and its simplified form.

Figure 3.767: Automaton for the MIN_MIN_INCREASING_SEQUENCE constraint obtained by applying decoration Table 2.35 to the seed transducer of the INCREASING_SEQUENCE pattern where default is $+\infty$

Figure 3.768: Automaton for the MIN_MIN_INCREASING_SEQUENCE constraint obtained by applying decoration Table 2.37 to the seed transducer of the INCREASING_SEQUENCE pattern where default is $+\infty$; $-R_i + R_{i-1} \geq 0$ is a linear invariant.

|     | $s$ | $t$ |
| --- | --- | --- |
| $s$ | $\min(\overrightarrow{C}, \overleftarrow{C})$ | $\min(\overrightarrow{C}, \overleftarrow{C})$ |
| $t$ | $\min(\overrightarrow{C}, \overleftarrow{C})$ | $\min(\overrightarrow{C}, \overleftarrow{C}, \overrightarrow{D}, \overleftarrow{D}, \texttt{VAR}_i)$ |

Table 3.83: Glue matrix for the MIN_MIN_INCREASING_SEQUENCE constraint defined as the composition of the  INCREASING_SEQUENCE pattern , the  feature MIN , and the  aggregator min ; cells of the glue matrix are coloured with the colour of the constituent to which they are related.

# 3.336   MIN_MIN_INFLEXION

   **DESCRIPTION**          **AUTOMATON**

$< (< | =)^* > | > (> | =)^* <$

| | |
|---|---|
| **Origin** | Based on the INFLEXION pattern. |
| **Constraint** | MIN_MIN_INFLEXION(VALUE, VARIABLES) |
| **Arguments** | VALUE       : dvar <br> VARIABLES  : collection(var−dvar) |
| **Restrictions** | $\mathtt{sv} \leq 2 \vee \mathtt{rv} \leq 1 \Rightarrow \mathtt{VALUE} = +\infty$ <br> $\mathtt{VALUE} \geq \mathtt{minv}①$ <br> $\mathtt{VALUE} = +\infty \vee \mathtt{VALUE} \leq \mathtt{maxv}$ <br> required(VARIABLES, var) <br> where <br> minv =minval(VARIABLES.var) <br> maxv =maxval(VARIABLES.var) <br> sv = \|VARIABLES\| <br> rv =range(VARIABLES.var) |

VALUE ≥ 0

| | |
|---|---|
| **Purpose** | VALUE is the minimum of all minimum values in each occurrence of the INFLEXION pattern in the time-series given by the VARIABLES collection. If the pattern does not occur, VALUE takes the default value $+\infty$. <br> An occurrence of the pattern INFLEXION is the *maximal* subsequence which matches the regular expression $< (< | =)^* > | > (> | =)^* <$. <br> Assume that the occurrence of the pattern INFLEXION starts at position $i$ and ends at position $j$. The feature MIN computes the minimum of the values from index $i + 1$ to index $j$. |

| | |
|---|---|
| **Example** | $(1, \langle 1, 2, 6, 6, 4, 4, 3, 5, 2, 5, 1, 5, 3, 3, 4, 4 \rangle)$ |

Figure 3.769 provides an example where the MIN_MIN_INFLEXION $(1, [1, 2, 6, 6, 4, 4, 3, 5, 2, 5, 1, 5, 3, 3, 4, 4])$ constraint holds.

| | |
|---|---|
| **Typical** | $\|\mathtt{VARIABLES}\| > 2$ <br> range(VARIABLES.var) $> 1$ |
| **Symmetry** | Items of VARIABLES can be reversed. |
| **Arg. properties** | Functional dependency: VALUE determined by VARIABLES. |

Figure 3.769: Illustrating the MIN_MIN_INFLEXION constraint of the **Example** slot

 Figures 3.770 and 3.771 respectively depict the automaton associated with the constraint MIN_MIN_INFLEXION and its simplified form.

Figure 3.770: Automaton for the MIN_MIN_INFLEXION constraint obtained by applying decoration Table 2.35 to the seed transducer of the INFLEXION pattern where `default` is $+\infty$ (transition $r \rightarrow t$ has the same accumulators updates as transition $t \rightarrow r$)

Figure 3.771: Automaton for the MIN_MIN_INFLEXION constraint obtained by applying decoration Table 2.25 to the seed transducer of the INFLEXION pattern where `default` is $+\infty$ (transition $r \rightarrow t$ has the same accumulators updates as transition $t \rightarrow r$); $-R_i + R_{i-1} \geq 0$ is a linear invariant.

## 3.337 MIN_MIN_STRICTLY_DECREASING_SEQUENCE

► ▷ ◁ ◀     DESCRIPTION          AUTOMATON

| | |
|---|---|
| **Origin** | Based on the STRICTLY_DECREASING_SEQUENCE pattern. |
| **Constraint** | MIN_MIN_STRICTLY_DECREASING_SEQUENCE(VALUE, VARIABLES) |
| **Arguments** | VALUE      : dvar <br> VARIABLES  : collection(var−dvar) |

**Restrictions**

$\mathtt{sv} \leq 1 \lor \mathtt{rv} \leq 1 \Rightarrow \mathtt{VALUE} = +\infty$
$\mathtt{VALUE} \geq \mathtt{minv}①$
$\mathtt{VALUE} = +\infty \lor \mathtt{VALUE} \leq \mathtt{maxv} - 1$
required(VARIABLES, var)
where
$\mathtt{minv} = \mathtt{minval}(\mathtt{VARIABLES.var})$
$\mathtt{maxv} = \mathtt{maxval}(\mathtt{VARIABLES.var})$
$\mathtt{sv} = |\mathtt{VARIABLES}|$
$\mathtt{rv} = \mathtt{range}(\mathtt{VARIABLES.var})$

**Purpose**

VALUE is the minimum of all minimum values in each occurrence of the STRICTLY_DECREASING_SEQUENCE pattern in the time-series given by the VARIABLES collection. If the pattern does not occur, VALUE takes the default value $+\infty$.
An occurrence of the pattern STRICTLY_DECREASING_SEQUENCE is the *maximal* subsequence which matches the regular expression $>^+$.
Assume that the occurrence of the pattern STRICTLY_DECREASING_SEQUENCE starts at position $i$ and ends at position $j$. The feature MIN computes the minimum of the values from index $i$ to index $j + 1$.

**Example**

$(1, \langle 4, 4, 6, 4, 1, 1, 3, 4, 4, 6, 6, 5, 2, 2, 4, 3 \rangle)$

Figure 3.772 provides an example where the MIN_MIN_STRICTLY_DECREASING_SEQUENCE $(1, [4, 4, 6, 4, 1, 1, 3, 4, 4, 6, 6, 5, 2, 2, 4, 3])$ constraint holds.

**Typical**

$|\mathtt{VARIABLES}| > 1$
range(VARIABLES.var) $> 1$

**Arg. properties**

Functional dependency: VALUE determined by VARIABLES.

Figure 3.772: Illustrating the MIN_MIN_STRICTLY_DECREASING_SEQUENCE constraint of the **Example** slot

**Automaton**     Figures 3.773 and 3.774 respectively depict the automaton associated with the constraint MIN_MIN_STRICTLY_DECREASING_SEQUENCE and its simplified form.

Figure 3.773: Automaton for the MIN_MIN_STRICTLY_DECREASING_SEQUENCE constraint obtained by applying decoration Table 2.35 to the seed transducer of the STRICTLY_DECREASING_SEQUENCE pattern where default is $+\infty$

Figure 3.774: Automaton for the MIN_MIN_STRICTLY_DECREASING_SEQUENCE constraint obtained by applying decoration Table 2.38 to the seed transducer of the STRICTLY_DECREASING_SEQUENCE pattern where default is $+\infty$; $-R_i + R_{i-1} \geq 0$ is a linear invariant.

|     | $s$ | $r$ |
| --- | --- | --- |
| $s$ | $\min(\overrightarrow{C}, \overleftarrow{C})$ | $\min(\overrightarrow{C}, \overleftarrow{C})$ |
| $r$ | $\min(\overrightarrow{C}, \overleftarrow{C})$ | $\min(\overrightarrow{C}, \overleftarrow{C}, \overrightarrow{D}, \overleftarrow{D}, \mathtt{VAR}_i)$ |

Table 3.84: Glue matrix for the MIN_MIN_STRICTLY_DECREASING_SEQUENCE constraint defined as the composition of the STRICTLY_DECREASING_SEQUENCE pattern, the feature MIN, and the aggregator min; cells of the glue matrix are coloured with the colour of the constituent to which they are related.

## 3.338 MIN_MIN_STRICTLY_INCREASING_SEQUENCE

▶ ▷ ◁ ◀    **DESCRIPTION**    **AUTOMATON**

| | |
|---|---|
| **Origin** | Based on the STRICTLY_INCREASING_SEQUENCE pattern. |
| **Constraint** | MIN_MIN_STRICTLY_INCREASING_SEQUENCE(VALUE, VARIABLES) |
| **Arguments** | VALUE : dvar <br> VARIABLES : collection(var−dvar) |

**Restrictions**

$sv \leq 1 \lor rv \leq 1 \Rightarrow VALUE = +\infty$

$VALUE \geq minv①$

$VALUE = +\infty \lor VALUE \leq maxv − 1$

required(VARIABLES, var)

where

minv =minval(VARIABLES.var)

maxv =maxval(VARIABLES.var)

sv = |VARIABLES|

rv =range(VARIABLES.var)

**Purpose**

VALUE is the minimum of all minimum values in each occurrence of the STRICTLY_INCREASING_SEQUENCE pattern in the time-series given by the VARIABLES collection. If the pattern does not occur, VALUE takes the default value $+\infty$.

An occurrence of the pattern STRICTLY_INCREASING_SEQUENCE is the *maximal* sub-sequence which matches the regular expression $<^+$.

Assume that the occurrence of the pattern STRICTLY_INCREASING_SEQUENCE starts at position $i$ and ends at position $j$. The feature MIN computes the minimum of the values from index $i$ to index $j + 1$.

**Example**

$(1, \langle 4, 3, 5, 5, 2, 1, 1, 2, 3, 4, 6, 6, 3, 1, 2, 3 \rangle)$

Figure 3.775 provides an example where the MIN_MIN_STRICTLY_INCREASING_SEQUENCE $(1, [4, 3, 5, 5, 2, 1, 1, 2, 3, 4, 6, 6, 3, 1, 2, 3])$ constraint holds.

**Typical**

|VARIABLES| > 1

range(VARIABLES.var) > 1

**Arg. properties**

Functional dependency: VALUE determined by VARIABLES.

Figure 3.775: Illustrating the MIN_MIN_STRICTLY_INCREASING_SEQUENCE constraint of the **Example** slot

**Automaton**    Figures 3.776 and 3.777 respectively depict the automaton associated with the constraint MIN_MIN_STRICTLY_INCREASING_SEQUENCE and its simplified form.

Figure 3.776: Automaton for the MIN_MIN_STRICTLY_INCREASING_SEQUENCE constraint obtained by applying decoration Table 2.35 to the seed transducer of the STRICTLY_INCREASING_SEQUENCE pattern where default is $+\infty$

Figure 3.777: Automaton for the MIN_MIN_STRICTLY_INCREASING_SEQUENCE constraint obtained by applying decoration Table 2.37 to the seed transducer of the STRICTLY_INCREASING_SEQUENCE pattern where default is $+\infty$; $-R_i + R_{i-1} \geq 0$ is a linear invariant.

|   | $s$ | $r$ |
|---|-----|-----|
| $s$ | $\min(\overrightarrow{C}, \overleftarrow{C})$ | $\min(\overrightarrow{C}, \overleftarrow{C})$ |
| $r$ | $\min(\overrightarrow{C}, \overleftarrow{C})$ | $\min(\overrightarrow{C}, \overleftarrow{C}, \overrightarrow{D}, \overleftarrow{D}, \text{VAR}_i)$ |

Table 3.85: Glue matrix for the MIN_MIN_STRICTLY_INCREASING_SEQUENCE constraint defined as the composition of the STRICTLY_INCREASING_SEQUENCE pattern, the feature MIN, and the aggregator min; cells of the glue matrix are coloured with the colour of the constituent to which they are related.

## 3.339 MIN_MIN_VALLEY

▶ ▷ ◁ ◀     **DESCRIPTION**        **AUTOMATON**

$> (= | >)^* (< | =)^* <$

| **Origin** | Based on the VALLEY pattern. |

| **Constraint** | MIN_MIN_VALLEY(VALUE, VARIABLES) |

| **Arguments** | VALUE      : dvar |
| | VARIABLES : collection(var−dvar) |

**Restrictions**

$\mathtt{sv} \leq 2 \lor \mathtt{rv} \leq 1 \Rightarrow \mathtt{VALUE} = +\infty$

$\mathtt{VALUE} \geq \mathtt{minv}①$

$\mathtt{VALUE} = +\infty \lor \mathtt{VALUE} \leq \mathtt{maxv} - 1$

required(VARIABLES, var)

where

$\mathtt{minv} = \mathtt{minval}(\mathtt{VARIABLES.var})$

$\mathtt{maxv} = \mathtt{maxval}(\mathtt{VARIABLES.var})$

$\mathtt{sv} = |\mathtt{VARIABLES}|$

$\mathtt{rv} = \mathtt{range}(\mathtt{VARIABLES.var})$

VALUE ≥ 0

**Purpose**

VALUE is the minimum of all minimum values in each occurrence of the VALLEY pattern in the time-series given by the VARIABLES collection. If the pattern does not occur, VALUE takes the default value $+\infty$.

An occurrence of the pattern VALLEY is the *maximal* subsequence which matches the regular expression $> (= | >)^* (< | =)^* <$.

Assume that the occurrence of the pattern VALLEY starts at position $i$ and ends at position $j$. The feature MIN computes the minimum of the values from index $i + 1$ to index $j$.

**Example**

$(2, \langle 1, 3, 7, 4, 3, 6, 6, 5, 3, 3, 2, 6, 5, 5, 5, 7 \rangle)$

Figure 3.778 provides an example where the MIN_MIN_VALLEY $(2, [1, 3, 7, 4, 3, 6, 6, 5, 3, 3, 2, 6, 5, 5, 5, 7])$ constraint holds.

| **Typical** | $|\mathtt{VARIABLES}| > 2$ |
| | range(VARIABLES.var) $> 1$ |

| **Symmetry** | Items of VARIABLES can be reversed. |

| **Arg. properties** | Functional dependency: VALUE determined by VARIABLES. |

Figure 3.778: Illustrating the MIN_MIN_VALLEY constraint of the **Example** slot

**Automaton**     Figures 3.779 and 3.780 respectively depict the automaton associated with the constraint MIN_MIN_VALLEY and its simplified form.

Figure 3.779: Automaton for the MIN_MIN_VALLEY constraint obtained by applying decoration Table 2.35 to the seed transducer of the VALLEY pattern where default is $+\infty$

Figure 3.780: Automaton for the MIN_MIN_VALLEY constraint obtained by applying decoration Table 2.37 to the seed transducer of the VALLEY pattern where default is $+\infty$; $-R_i + R_{i-1} \geq 0$ is a linear invariant.

|   | $s$ | $r$ | $t$ |
|---|---|---|---|
| $s$ | $\min(\overrightarrow{C}, \overleftarrow{C})$ | $\min(\overrightarrow{C}, \overleftarrow{C})$ | $\min(\overrightarrow{C}, \overleftarrow{C})$ |
| $r$ | $\min(\overrightarrow{C}, \overleftarrow{C})$ | $\min(\overrightarrow{D}, \overleftarrow{D}, \text{VAR}_i)$ | $\min(\overleftarrow{C}, \overrightarrow{D}, \overleftarrow{D}, \text{VAR}_i)$ |
| $t$ | $\min(\overrightarrow{C}, \overleftarrow{C})$ | $\min(\overrightarrow{C}, \overrightarrow{D}, \overleftarrow{D}, \text{VAR}_i)$ | $\min(\overrightarrow{C}, \overleftarrow{C})$ |

Table 3.86: Glue matrix for the MIN_MIN_VALLEY constraint defined as the composition of the VALLEY pattern , the feature MIN , and the aggregator min ; cells of the glue matrix are coloured with the colour of the constituent to which they are related.

## 3.340 MIN_MIN_ZIGZAG

▶ ▷ ◁ ◀        **DESCRIPTION**        **AUTOMATON**

**Origin**              Based on the ZIGZAG pattern.

**Constraint**          MIN_MIN_ZIGZAG(VALUE, VARIABLES)

**Arguments**           VALUE      : dvar
                        VARIABLES  : collection(var−dvar)

**Restrictions**        $\mathtt{sv} \leq 3 \vee \mathtt{rv} \leq 1 \Rightarrow \mathtt{VALUE} = +\infty$
                        $\mathtt{VALUE} \geq \mathtt{minv}①$
                        $\mathtt{VALUE} = +\infty \vee \mathtt{VALUE} \leq \mathtt{maxv} - 1$
                        required(VARIABLES, var)
                        where
                        minv = minval(VARIABLES.var)
                        maxv = maxval(VARIABLES.var)
                        sv = |VARIABLES|
                        rv = range(VARIABLES.var)

**Purpose**             VALUE is the minimum of all minimum values in each occurrence of the ZIGZAG pattern
                        in the time-series given by the VARIABLES collection. If the pattern does not occur,
                        VALUE takes the default value $+\infty$.
                        An occurrence of the pattern ZIGZAG is the *maximal* subsequence which matches the
                        regular expression $(<>)^+ (< | <>) | (><)^+ (> | ><)$.
                        Assume that the occurrence of the pattern ZIGZAG starts at position $i$ and ends at position
                        $j$. The feature MIN computes the minimum of the values from index $i + 1$ to index $j$.

**Example**             $(1, \langle 4, 1, 3, 1, 4, 6, 1, 5, 5, 2, 7, 2, 3, 1, 6, 1 \rangle)$

                        Figure 3.781 provides an example where the MIN_MIN_ZIGZAG
                        $(1, [4, 1, 3, 1, 4, 6, 1, 5, 5, 2, 7, 2, 3, 1, 6, 1])$ constraint holds.

**Typical**             $|\mathtt{VARIABLES}| > 3$
                        range(VARIABLES.var) > 1

**Symmetry**            Items of VARIABLES can be reversed.

**Arg. properties**     Functional dependency: VALUE determined by VARIABLES.

Figure 3.781: Illustrating the MIN_MIN_ZIGZAG constraint of the **Example** slot

**Automaton**          Figures 3.782 and 3.783 respectively depict the automaton associated with the constraint
                       MIN_MIN_ZIGZAG and its simplified form.

Figure 3.782: Automaton for the MIN_MIN_ZIGZAG constraint obtained by applying decoration Table 2.35 to the seed transducer of the ZIGZAG pattern where default is $+\infty$; (1) missing transitions from $a, b, c, d, e, f$ to $s$ are labelled by $=$; (2) on transitions from $b, c, e, f$ to $s$ the accumulator $D$ is reset to its initial value; (3) on transitions from $c, f$ to $s$ the accumulator $R$ is updated wrt $C$ and the accumulator $C$ is reset to its initial value

Figure 3.783: Automaton for the MIN_MIN_ZIGZAG constraint obtained by applying decoration Table 2.28 to the seed transducer of the ZIGZAG pattern where default is $+\infty$; (1) missing transitions from $a, b, c, d, e, f$ to $s$ are labelled by $=$; (2) on transitions from $b, c, e, f$ to $s$ the accumulator $D$ is reset to its initial value; $-R_i + R_{i-1} \geq 0$ is a linear invariant.

| | $s$ | $a$ | $b$ | $c$ | $d$ | $e$ | $f$ |
|---|---|---|---|---|---|---|---|
| $s$ | $\min(\overrightarrow{C},\overleftarrow{C})$ | $\min(\overrightarrow{C},\overleftarrow{C})$ | $\min(\overrightarrow{C},\overleftarrow{C})$ | $\min(\overrightarrow{C},\overleftarrow{C})$ | $\min(\overrightarrow{C},\overleftarrow{C})$ | $\min(\overrightarrow{C},\overleftarrow{C})$ | $\min(\overrightarrow{C},\overleftarrow{C})$ |
| $a$ | $\min(\overrightarrow{C},\overleftarrow{C})$ | $\min(\overrightarrow{C},\overleftarrow{C})$ | $\min(\overrightarrow{C},\overleftarrow{C})$ | $\min(\overrightarrow{C},\overrightarrow{D},\overleftarrow{D},\mathbf{VAR}_i)$ | $\min(\overrightarrow{C},\overleftarrow{C})$ | $\min(\overrightarrow{D},\overleftarrow{D},\mathbf{VAR}_i)$ | $\min(\overrightarrow{C},\overleftarrow{C})$ |
| $b$ | $\min(\overrightarrow{C},\overleftarrow{C})$ | $\min(\overrightarrow{C},\overleftarrow{C})$ | $\min(\overrightarrow{D},\overleftarrow{D},\mathbf{VAR}_i)$ | $\min(\overrightarrow{C},\overleftarrow{C})$ | $\min(\overrightarrow{D},\overleftarrow{D},\mathbf{VAR}_i)$ | $\min(\overrightarrow{C},\overleftarrow{C})$ | $\min(\overrightarrow{C},\overrightarrow{D},\overleftarrow{D},\mathbf{VAR}_i)$ |
| $c$ | $\min(\overrightarrow{C},\overleftarrow{C})$ | $\min(\overrightarrow{C},\overrightarrow{D},\overleftarrow{D},\mathbf{VAR}_i)$ | $\min(\overrightarrow{C},\overleftarrow{C})$ | $\min(\overrightarrow{C},\overleftarrow{C},\overrightarrow{D},\overleftarrow{D},\mathbf{VAR}_i)$ | $\min(\overrightarrow{C},\overleftarrow{C})$ | $\min(\overrightarrow{C},\overrightarrow{D},\overleftarrow{D},\mathbf{VAR}_i)$ | $\min(\overrightarrow{C},\overleftarrow{C})$ |
| $d$ | $\min(\overrightarrow{C},\overleftarrow{C})$ | $\min(\overrightarrow{C},\overleftarrow{C})$ | $\min(\overrightarrow{D},\overleftarrow{D},\mathbf{VAR}_i)$ | $\min(\overrightarrow{C},\overleftarrow{C})$ | $\min(\overrightarrow{C},\overleftarrow{C})$ | $\min(\overrightarrow{C},\overleftarrow{C})$ | $\min(\overrightarrow{C},\overrightarrow{D},\overleftarrow{D},\mathbf{VAR}_i)$ |
| $e$ | $\min(\overrightarrow{C},\overleftarrow{C})$ | $\min(\overrightarrow{D},\overleftarrow{D},\mathbf{VAR}_i)$ | $\min(\overrightarrow{C},\overleftarrow{C})$ | $\min(\overrightarrow{C},\overrightarrow{D},\overleftarrow{D},\mathbf{VAR}_i)$ | $\min(\overrightarrow{C},\overleftarrow{C})$ | $\min(\overrightarrow{D},\overleftarrow{D},\mathbf{VAR}_i)$ | $\min(\overrightarrow{C},\overleftarrow{C})$ |
| $f$ | $\min(\overrightarrow{C},\overleftarrow{C})$ | $\min(\overrightarrow{C},\overleftarrow{C})$ | $\min(\overrightarrow{C},\overrightarrow{D},\overleftarrow{D},\mathbf{VAR}_i)$ | $\min(\overrightarrow{C},\overleftarrow{C})$ | $\min(\overrightarrow{C},\overrightarrow{D},\overleftarrow{D},\mathbf{VAR}_i)$ | $\min(\overrightarrow{C},\overleftarrow{C})$ | $\min(\overrightarrow{C},\overleftarrow{C},\overrightarrow{D},\overleftarrow{D},\mathbf{VAR}_i)$ |

Table 3.87: Glue matrix for the MIN_MIN_ZIGZAG constraint defined as the composition of the ZIGZAG pattern , the feature MIN , and the aggregator min ; cells of the glue matrix are coloured with the colour of the constituent to which they are related.

# 3.341   MIN_RANGE_DECREASING

| | |
|---|---|
| **Origin** | Based on the DECREASING pattern. |
| **Constraint** | MIN_RANGE_DECREASING(VALUE, VARIABLES) |
| **Arguments** | VALUE      : dvar<br>VARIABLES : collection(var−dvar) |

**Restrictions**

$$\texttt{sv} \leq 1 \vee \texttt{rv} \leq 1 \Rightarrow \texttt{VALUE} = +\infty$$
$$\texttt{VALUE} \geq 1①$$
required(VARIABLES, var)
where
$$\texttt{sv} = |\texttt{VARIABLES}|$$
$$\texttt{rv} = \texttt{range}(\texttt{VARIABLES.var})$$

**Purpose**

VALUE is the minimum value of the differences between the largest and smallest value in each occurrence of the DECREASING pattern in the time-series given by the VARIABLES collection. If the pattern does not occur, VALUE takes the default value $+\infty$.

An occurrence of the pattern DECREASING is the subsequence which matches the regular expression >.

Assume that the occurrence of the pattern DECREASING starts at position $i$ and ends at position $j$. The feature RANGE computes the range of the values from index $i$ to index $j + 1$.

**Example**

$$(1, \langle 3, 4, 2, 2, 5, 6, 6, 4, 4, 3, 1, 1, 4, 6, 4, 4 \rangle)$$

Figure 3.784 provides an example where the MIN_RANGE_DECREASING $(1, [3, 4, 2, 2, 5, 6, 6, 4, 4, 3, 1, 1, 4, 6, 4, 4])$ constraint holds.

**Typical**

$$|\texttt{VARIABLES}| > 1$$
$$\texttt{range}(\texttt{VARIABLES.var}) > 1$$

**Symmetry**

One and the same constant can be added to the var attribute of all items of VARIABLES.

**Arg. properties**

Functional dependency: VALUE determined by VARIABLES.

Figure 3.784: Illustrating the MIN_RANGE_DECREASING constraint of the **Example** slot

**Automaton**     Figures 3.785 and 3.786 respectively depict the automaton associated with the constraint MIN_RANGE_DECREASING and its simplified form.

Figure 3.785: Automaton for the MIN_RANGE_DECREASING constraint obtained by applying decoration Table 2.46 to the seed transducer of the DECREASING pattern where `default` is $+\infty$

Figure 3.786: Automaton for the MIN_RANGE_DECREASING constraint obtained by applying decoration Table 2.44 to the seed transducer of the DECREASING pattern where `default` is $+\infty$; $-R_i + R_{i-1} \geq 0$ is a linear invariant.

# 3.342   MIN_RANGE_DECREASING_SEQUENCE

► ▷ ◁ ◀        DESCRIPTION            AUTOMATON

**Origin**            Based on the DECREASING_SEQUENCE pattern.

**Constraint**        MIN_RANGE_DECREASING_SEQUENCE(VALUE, VARIABLES)

**Arguments**         VALUE      : dvar
                      VARIABLES  : collection(var−dvar)

**Restrictions**      $\mathtt{sv} \leq 1 \lor \mathtt{rv} \leq 1 \Rightarrow \mathtt{VALUE} = +\infty$
                      $\mathtt{VALUE} \geq 1$①
                      required(VARIABLES, var)
                      where
                      $\mathtt{sv} = |\mathtt{VARIABLES}|$
                      $\mathtt{rv} = \mathrm{range}(\mathtt{VARIABLES.var})$

**Purpose**           VALUE is the minimum value of the differences between the largest and smallest value in
                      each occurrence of the DECREASING_SEQUENCE pattern in the time-series given by the
                      VARIABLES collection. If the pattern does not occur, VALUE takes the default value $+\infty$.
                      An occurrence of the pattern DECREASING_SEQUENCE is the *maximal* subsequence
                      which matches the regular expression $> (> | =)^* | >$.
                      Assume that the occurrence of the pattern DECREASING_SEQUENCE starts at position $i$
                      and ends at position $j$. The feature RANGE computes the range of the values from index
                      $i$ to index $j + 1$.

**Example**           $(2, \langle 3, 4, 2, 2, 5, 6, 6, 4, 4, 3, 1, 1, 4, 6, 4, 4 \rangle)$

                      Figure 3.787 provides an example where the MIN_RANGE_DECREASING_SEQUENCE
                      $(2, [3, 4, 2, 2, 5, 6, 6, 4, 4, 3, 1, 1, 4, 6, 4, 4])$ constraint holds.

**Typical**           $|\mathtt{VARIABLES}| > 1$
                      $\mathrm{range}(\mathtt{VARIABLES.var}) > 1$

**Symmetry**          One and the same constant can be added to the var attribute of all items of VARIABLES.

**Arg. properties**   Functional dependency: VALUE determined by VARIABLES.

Figure 3.787: Illustrating the MIN_RANGE_DECREASING_SEQUENCE constraint of the **Example** slot

**Automaton**  Figures 3.788 and 3.789 respectively depict the automaton associated with the constraint MIN_RANGE_DECREASING_SEQUENCE and its simplified form.

Figure 3.788:  Automaton for the MIN_RANGE_DECREASING_SEQUENCE constraint obtained by applying decoration Table 2.46 to the seed transducer of the DECREAS-ING_SEQUENCE pattern where `default` is $+\infty$

Figure 3.789:  Automaton for the MIN_RANGE_DECREASING_SEQUENCE constraint obtained by applying decoration Table 2.40 to the seed transducer of the DECREAS-ING_SEQUENCE pattern where `default` is $+\infty$; $-R_i + R_{i-1} \geq 0$ is a linear invariant.

## 3.343   MIN_RANGE_INCREASING

► ▷ ◁ ◀        **DESCRIPTION**        **AUTOMATON**

**Origin**            Based on the INCREASING pattern.

**Constraint**        MIN_RANGE_INCREASING(VALUE, VARIABLES)

**Arguments**         VALUE      : dvar
                      VARIABLES  : collection(var−dvar)

**Restrictions**      $sv \leq 1 \vee rv \leq 1 \Rightarrow VALUE = +\infty$
                      $VALUE \geq 1$①
                      required(VARIABLES, var)
                      where
                      $sv = |VARIABLES|$
                      $rv = $range(VARIABLES.var)

**Purpose**           VALUE is the minimum value of the differences between the largest and smallest value in
                      each occurrence of the INCREASING pattern in the time-series given by the VARIABLES
                      collection. If the pattern does not occur, VALUE takes the default value $+\infty$.
                      An occurrence of the pattern INCREASING is the subsequence which matches the regular
                      expression $<$.
                      Assume that the occurrence of the pattern INCREASING starts at position $i$ and ends at
                      position $j$. The feature RANGE computes the range of the values from index $i$ to index
                      $j + 1$.

**Example**           $(1, \langle 4, 3, 5, 5, 2, 1, 1, 3, 3, 4, 6, 6, 3, 1, 3, 3 \rangle)$

                      Figure 3.790 provides an example where the MIN_RANGE_INCREASING
                      $(1, [4, 3, 5, 5, 2, 1, 1, 3, 3, 4, 6, 6, 3, 1, 3, 3])$ constraint holds.

**Typical**           $|VARIABLES| > 1$
                      range(VARIABLES.var) $> 1$

**Symmetry**          One and the same constant can be added to the var attribute of all items of VARIABLES.

**Arg. properties**   Functional dependency: VALUE determined by VARIABLES.

Figure 3.790: Illustrating the MIN_RANGE_INCREASING constraint of the **Example** slot

**Automaton**      Figures 3.791 and 3.792 respectively depict the automaton associated with the constraint MIN_RANGE_INCREASING and its simplified form.

Figure 3.791: Automaton for the MIN_RANGE_INCREASING constraint obtained by applying decoration Table 2.46 to the seed transducer of the INCREASING pattern where default is $+\infty$

Figure 3.792: Automaton for the MIN_RANGE_INCREASING constraint obtained by applying decoration Table 2.45 to the seed transducer of the INCREASING pattern where default is $+\infty$; $-R_i + R_{i-1} \geq 0$ is a linear invariant.

# 3.344 MIN_RANGE_INCREASING_SEQUENCE

► ▷ ◁ ◄      DESCRIPTION          AUTOMATON

$< (< | =)^* < |$
$<$

| **Origin** | Based on the INCREASING_SEQUENCE pattern. |
|---|---|

| **Constraint** | MIN_RANGE_INCREASING_SEQUENCE(VALUE, VARIABLES) |
|---|---|

| **Arguments** | VALUE     : dvar |
|---|---|
| | VARIABLES : collection(var−dvar) |

**Restrictions**

$\mathtt{sv} \leq 1 \vee \mathtt{rv} \leq 1 \Rightarrow \mathtt{VALUE} = +\infty$

$\mathtt{VALUE} \geq 1$①

required(VARIABLES, var)

where

$\mathtt{sv} = |\mathtt{VARIABLES}|$

$\mathtt{rv} = \mathtt{range}(\mathtt{VARIABLES.var})$

VALUE ≥ 1

**Purpose**

VALUE is the minimum value of the differences between the largest and smallest value in each occurrence of the INCREASING_SEQUENCE pattern in the time-series given by the VARIABLES collection. If the pattern does not occur, VALUE takes the default value $+\infty$.
An occurrence of the pattern INCREASING_SEQUENCE is the *maximal* subsequence which matches the regular expression $< (< | =)^* < | <$.
Assume that the occurrence of the pattern INCREASING_SEQUENCE starts at position $i$ and ends at position $j$. The feature RANGE computes the range of the values from index $i$ to index $j + 1$.

**Example**

$(2, \langle 4, 3, 5, 5, 2, 1, 1, 3, 3, 4, 6, 6, 3, 1, 3, 3 \rangle)$

Figure 3.793 provides an example where the MIN_RANGE_INCREASING_SEQUENCE $(2, [4, 3, 5, 5, 2, 1, 1, 3, 3, 4, 6, 6, 3, 1, 3, 3])$ constraint holds.

| **Typical** | $|\mathtt{VARIABLES}| > 1$ |
|---|---|
| | range(VARIABLES.var) > 1 |

| **Symmetry** | One and the same constant can be added to the var attribute of all items of VARIABLES. |
|---|---|

| **Arg. properties** | Functional dependency: VALUE determined by VARIABLES. |
|---|---|

Figure 3.793: Illustrating the MIN_RANGE_INCREASING_SEQUENCE constraint of the **Example** slot

  Figures 3.794 and 3.795 respectively depict the automaton associated with the constraint MIN_RANGE_INCREASING_SEQUENCE and its simplified form.

Figure 3.794:  Automaton for the MIN_RANGE_INCREASING_SEQUENCE constraint obtained by applying decoration Table 2.46 to the seed transducer of the INCREASING_SEQUENCE pattern where default is $+\infty$

Figure 3.795:  Automaton for the MIN_RANGE_INCREASING_SEQUENCE constraint obtained by applying decoration Table 2.41 to the seed transducer of the INCREASING_SEQUENCE pattern where default is $+\infty$; $-R_i + R_{i-1} \geq 0$ is a linear invariant.

# 3.345   MIN_RANGE_STRICTLY_DECREASING_SEQUENCE

► ▷ ◁ ◀     **DESCRIPTION**          **AUTOMATON**

| | |
|---|---|
| **Origin** | Based on the STRICTLY_DECREASING_SEQUENCE pattern. |
| **Constraint** | MIN_RANGE_STRICTLY_DECREASING_SEQUENCE(VALUE, VARIABLES) |
| **Arguments** | VALUE : dvar<br>VARIABLES : collection(var−dvar) |
| **Restrictions** | $sv \leq 1 \vee rv \leq 1 \Rightarrow VALUE = +\infty$<br>$VALUE \geq 1$①<br>required(VARIABLES, var)<br>where<br>$sv = \|VARIABLES\|$<br>$rv = range(VARIABLES.var)$ |
| **Purpose** | VALUE is the minimum value of the differences between the largest and smallest value in each occurrence of the STRICTLY_DECREASING_SEQUENCE pattern in the time-series given by the VARIABLES collection. If the pattern does not occur, VALUE takes the default value $+\infty$.<br>An occurrence of the pattern STRICTLY_DECREASING_SEQUENCE is the *maximal* subsequence which matches the regular expression $>^+$.<br>Assume that the occurrence of the pattern STRICTLY_DECREASING_SEQUENCE starts at position $i$ and ends at position $j$. The feature RANGE computes the range of the values from index $i$ to index $j + 1$. |
| **Example** | $(1, \langle 4, 4, 6, 4, 1, 1, 3, 4, 4, 6, 6, 5, 2, 2, 4, 3 \rangle)$ |

Figure 3.796 provides an example where the MIN_RANGE_STRICTLY_DECREASING_SEQUENCE $(1, [4, 6, 4, 1, 1, 3, 4, 4, 6, 6, 5, 2, 2, 4, 3])$ constraint holds.

| | |
|---|---|
| **Typical** | $\|VARIABLES\| > 1$<br>range(VARIABLES.var) > 1 |
| **Symmetry** | One and the same constant can be added to the var attribute of all items of VARIABLES. |
| **Arg. properties** | Functional dependency: VALUE determined by VARIABLES. |

Figure 3.796: Illustrating the MIN_RANGE_STRICTLY_DECREASING_SEQUENCE constraint of the **Example** slot

**Automaton**    Figures 3.797 and 3.798 respectively depict the automaton associated with the constraint MIN_RANGE_STRICTLY_DECREASING_SEQUENCE and its simplified form.

Figure 3.797: Automaton for the MIN_RANGE_STRICTLY_DECREASING_SEQUENCE constraint obtained by applying decoration Table 2.46 to the seed transducer of the STRICTLY_DECREASING_SEQUENCE pattern where default is $+\infty$

Figure 3.798: Automaton for the MIN_RANGE_STRICTLY_DECREASING_SEQUENCE constraint obtained by applying decoration Table 2.40 to the seed transducer of the STRICTLY_DECREASING_SEQUENCE pattern where default is $+\infty$; $-R_i + R_{i-1} \geq 0$ is a linear invariant.

# 3.346   MIN_RANGE_STRICTLY_INCREASING_SEQUENCE

**DESCRIPTION**          **AUTOMATON**

| | |
|---|---|
| **Origin** | Based on the STRICTLY_INCREASING_SEQUENCE pattern. |
| **Constraint** | MIN_RANGE_STRICTLY_INCREASING_SEQUENCE(VALUE, VARIABLES) |
| **Arguments** | VALUE      : dvar <br> VARIABLES  : collection(var−dvar) |
| **Restrictions** | $\mathtt{sv} \leq 1 \vee \mathtt{rv} \leq 1 \Rightarrow \mathtt{VALUE} = +\infty$ <br> $\mathtt{VALUE} \geq 1$① <br> required(VARIABLES, var) <br> where <br> $\mathtt{sv} = |\mathtt{VARIABLES}|$ <br> $\mathtt{rv} = $ range(VARIABLES.var) |

| | |
|---|---|
| **Purpose** | VALUE is the minimum value of the differences between the largest and smallest value in each occurrence of the STRICTLY_INCREASING_SEQUENCE pattern in the time-series given by the VARIABLES collection. If the pattern does not occur, VALUE takes the default value $+\infty$. <br><br> An occurrence of the pattern STRICTLY_INCREASING_SEQUENCE is the *maximal* sub-sequence which matches the regular expression $<^+$. <br><br> Assume that the occurrence of the pattern STRICTLY_INCREASING_SEQUENCE starts at position $i$ and ends at position $j$. The feature RANGE computes the range of the values from index $i$ to index $j + 1$. |
| **Example** | $(2, \langle 4, 3, 5, 5, 2, 1, 1, 2, 3, 4, 6, 6, 3, 1, 2, 3 \rangle)$ |

Figure 3.799 provides an example where the MIN_RANGE_STRICTLY_INCREASING_SEQUENCE $(2, [4, 3, 5, 5, 2, 1, 1, 2, 3, 4, 6, 6, 3, 1, 2, 3])$ constraint holds.

| | |
|---|---|
| **Typical** | $|\mathtt{VARIABLES}| > 1$ <br> range(VARIABLES.var) $> 1$ |

| | |
|---|---|
| **Symmetry** | One and the same constant can be added to the var attribute of all items of VARIABLES. |
| **Arg. properties** | Functional dependency: VALUE determined by VARIABLES. |

Figure 3.799: Illustrating the MIN_RANGE_STRICTLY_INCREASING_SEQUENCE constraint of the **Example** slot

**Automaton**    Figures 3.800 and 3.801 respectively depict the automaton associated with the constraint MIN_RANGE_STRICTLY_INCREASING_SEQUENCE and its simplified form.

Figure 3.800: Automaton for the MIN_RANGE_STRICTLY_INCREASING_SEQUENCE constraint obtained by applying decoration Table 2.46 to the seed transducer of the STRICTLY_INCREASING_SEQUENCE pattern where default is $+\infty$

Figure 3.801: Automaton for the MIN_RANGE_STRICTLY_INCREASING_SEQUENCE constraint obtained by applying decoration Table 2.41 to the seed transducer of the STRICTLY_INCREASING_SEQUENCE pattern where default is $+\infty$; $-R_i + R_{i-1} \geq 0$ is a linear invariant.

## 3.347　MIN_SURF_BUMP_ON_DECREASING_SEQUENCE

▶ ▷ ◁ ◀　　**DESCRIPTION**　　**AUTOMATON**

| | |
|---|---|
| **Origin** | Based on the BUMP_ON_DECREASING_SEQUENCE pattern. |
| **Constraint** | MIN_SURF_BUMP_ON_DECREASING_SEQUENCE(VALUE, VARIABLES) |
| **Arguments** | VALUE　　　: dvar<br>VARIABLES : collection(var−dvar) |

**Restrictions**

$\text{sv} \leq 5 \vee \text{rv} \leq 2 \Rightarrow \text{VALUE} = +\infty$

$\text{VALUE} \geq 3 * \text{minv} + 3$①

$\text{VALUE} = +\infty \vee \text{VALUE} \leq 3 * \text{maxv} - 3$

required(VARIABLES, var)

where

minv =minval(VARIABLES.var)

rv =range(VARIABLES.var)

sv = |VARIABLES|

maxv =maxval(VARIABLES.var)

**Purpose**

VALUE is the minimal surface of occurrences of the BUMP_ON_DECREASING_SEQUENCE pattern in the time-series given by the VARIABLES collection. If the pattern does not occur, VALUE takes the default value $+\infty$.

An occurrence of the pattern BUMP_ON_DECREASING_SEQUENCE is the subsequence which matches the regular expression >><>>.

Assume that the occurrence of the pattern BUMP_ON_DECREASING_SEQUENCE starts at position $i$ and ends at position $j$. The feature SURF computes the sum of the values from index $i + 2$ to index $j$.

**Example**

$(11, \langle 7, 6, 5, 6, 5, 4, 1, 4, 7, 5, 4, 2, 5, 4, 3, 3 \rangle)$

Figure 3.802 provides an example where the MIN_SURF_BUMP_ON_DECREASING_SEQUENCE $(11, [7, 6, 5, 6, 5, 4, 1, 4, 7, 5, 4, 2, 5, 4, 3, 3])$ constraint holds.

| | |
|---|---|
| **Typical** | $\|\text{VARIABLES}\| > 5$<br>range(VARIABLES.var) > 2 |
| **Arg. properties** | Functional dependency: VALUE determined by VARIABLES. |

Figure 3.802: Illustrating the MIN_SURF_BUMP_ON_DECREASING_SEQUENCE constraint of the **Example** slot

**Automaton**   Figures 3.803 and 3.804 respectively depict the automaton associated with the constraint
MIN_SURF_BUMP_ON_DECREASING_SEQUENCE and its simplified form.

Figure 3.803: Automaton for the MIN_SURF_BUMP_ON_DECREASING_SEQUENCE
constraint obtained by applying decoration Table 2.35 to the seed transducer of the
BUMP_ON_DECREASING_SEQUENCE pattern where default is $+\infty$

Figure 3.804: Automaton for the MIN_SURF_BUMP_ON_DECREASING_SEQUENCE constraint obtained by applying decoration Table 2.25 to the seed transducer of the BUMP_ON_DECREASING_SEQUENCE pattern where default is $+\infty$; $-R_i + R_{i-1} \geq 0$ is a linear invariant.

# 3.348   MIN_SURF_DECREASING

      **DESCRIPTION**          **AUTOMATON**

| | |
|---|---|
| **Origin** | Based on the DECREASING pattern. |
| **Constraint** | MIN_SURF_DECREASING(VALUE, VARIABLES) |
| **Arguments** | VALUE      : dvar<br>VARIABLES  : collection(var−dvar) |
| **Restrictions** | $\mathtt{sv} \leq 1 \vee \mathtt{rv} \leq 1 \Rightarrow \mathtt{VALUE} = +\infty$<br>$\mathtt{VALUE} \geq 2 * \mathtt{minv} + 1$ ①<br>$\mathtt{VALUE} = +\infty \vee \mathtt{VALUE} \leq 2 * \mathtt{maxv} - 1$<br>required(VARIABLES, var)<br>where<br>minv =minval(VARIABLES.var)<br>maxv =maxval(VARIABLES.var)<br>sv = \|VARIABLES\|<br>rv =range(VARIABLES.var) |
| **Purpose** | VALUE is the minimal surface of occurrences of the DECREASING pattern in the time-series given by the VARIABLES collection. If the pattern does not occur, VALUE takes the default value $+\infty$.<br>An occurrence of the pattern DECREASING is the subsequence which matches the regular expression >.<br>Assume that the occurrence of the pattern DECREASING starts at position $i$ and ends at position $j$. The feature SURF computes the sum of the values from index $i$ to index $j+1$. |
| **Example** | $(4, \langle 3, 4, 2, 2, 5, 6, 6, 4, 4, 3, 1, 1, 4, 6, 4, 4 \rangle)$ |

Figure 3.805 provides an example where the MIN_SURF_DECREASING $(4, [3, 4, 2, 2, 5, 6, 6, 4, 4, 3, 1, 1, 4, 6, 4, 4])$ constraint holds.

| | |
|---|---|
| **Typical** | $\|\mathtt{VARIABLES}\| > 1$<br>range(VARIABLES.var) > 1 |
| **Arg. properties** | Functional dependency: VALUE determined by VARIABLES. |

Figure 3.805: Illustrating the MIN_SURF_DECREASING constraint of the **Example** slot

**Automaton**    Figures 3.806 and 3.807 respectively depict the automaton associated with the constraint MIN_SURF_DECREASING and its simplified form.

Figure 3.806: Automaton for the MIN_SURF_DECREASING constraint obtained by applying decoration Table 2.35 to the seed transducer of the DECREASING pattern where `default` is $+\infty$

Figure 3.807: Automaton for the MIN_SURF_DECREASING constraint obtained by applying decoration Table 2.38 to the seed transducer of the DECREASING pattern where `default` is $+\infty$; $-R_i + R_{i-1} \geq 0$ is a linear invariant.

Table 3.88: Glue matrix for the MIN_SURF_DECREASING constraint defined as the composition of the DECREASING pattern , the feature SURF , and the aggregator min ; cells of the glue matrix are coloured with the colour of the constituent to which they are related.

## 3.349   MIN_SURF_DECREASING_SEQUENCE

▶ ▷ ◁ ◀       **DESCRIPTION**         **AUTOMATON**

| **Origin** | Based on the DECREASING_SEQUENCE pattern. |
|---|---|

| **Constraint** | MIN_SURF_DECREASING_SEQUENCE(VALUE, VARIABLES) |
|---|---|

| **Arguments** | VALUE      : dvar |
|---|---|
|  | VARIABLES : collection(var−dvar) |

**Restrictions**

$\text{sv} \leq 1 \vee \text{rv} \leq 1 \Rightarrow \text{VALUE} = +\infty$

$\text{rv} = 2 \Rightarrow \text{VALUE} \geq 2 * \text{minv} + 1$ ①

$\text{rv} \geq 3 \Rightarrow \text{VALUE} \geq \min(2 * \text{minv} + 1$②$, \text{sv} * (\text{minv} + 1)$③$)$

$\text{rv} = 2 \Rightarrow \text{VALUE} = +\infty \vee \text{VALUE} \leq 2 * \text{maxv} - 1$

$\text{rv} \geq 3 \Rightarrow \text{VALUE} = +\infty \vee \text{VALUE} \leq \max(2 * \text{maxv} - 1, \text{sv} * (\text{maxv} - 1))$

required(VARIABLES, var)

where

minv = minval(VARIABLES.var)

rv = range(VARIABLES.var)

sv = |VARIABLES|

maxv = maxval(VARIABLES.var)

**Purpose**

VALUE is the minimal surface of occurrences of the DECREASING_SEQUENCE pattern in the time-series given by the VARIABLES collection. If the pattern does not occur, VALUE takes the default value $+\infty$.

An occurrence of the pattern DECREASING_SEQUENCE is the *maximal* subsequence which matches the regular expression $> (> | =)^* > | >$.

Assume that the occurrence of the pattern DECREASING_SEQUENCE starts at position $i$ and ends at position $j$. The feature SURF computes the sum of the values from index $i$ to index $j + 1$.

**Example**

$(6, \langle 3, 4, 2, 2, 5, 6, 6, 4, 4, 3, 1, 1, 4, 6, 4, 4 \rangle)$

Figure 3.808 provides an example where the MIN_SURF_DECREASING_SEQUENCE $(6, [3, 4, 2, 2, 5, 6, 6, 4, 4, 3, 1, 1, 4, 6, 4, 4])$ constraint holds.

| **Typical** | $|\text{VARIABLES}| > 1$ |
|---|---|
|  | range(VARIABLES.var) > 1 |

| **Arg. properties** | Functional dependency: VALUE determined by VARIABLES. |
|---|---|

Figure 3.808: Illustrating the MIN_SURF_DECREASING_SEQUENCE constraint of the **Example** slot

**Automaton**  Figure 3.809 depicts the automaton associated with the constraint MIN_SURF_DECREASING_SEQUENCE.

Figure 3.809: Automaton for the MIN_SURF_DECREASING_SEQUENCE constraint obtained by applying decoration Table 2.35 to the seed transducer of the DECREAS-ING_SEQUENCE pattern where `default` is $+\infty$; $-R_i + R_{i-1} \geq 0$ is a linear invariant.

|     | $s$ | $t$ |
|-----|-----|-----|
| $s$ | $\min(\overrightarrow{C}, \overleftarrow{C})$ | $\min(\overrightarrow{C}, \overleftarrow{C})$ |
| $t$ | $\min(\overrightarrow{C}, \overleftarrow{C})$ | $\overrightarrow{C} + \overleftarrow{C} + \overrightarrow{D} + \overleftarrow{D} + \texttt{VAR}_i$ |

Table 3.89: Glue matrix for the MIN_SURF_DECREASING_SEQUENCE constraint defined as the composition of the DECREASING_SEQUENCE pattern , the feature SURF , and the aggregator min ; cells of the glue matrix are coloured with the colour of the constituent to which they are related.

# 3.350  MIN_SURF_DECREASING_TERRACE

▶ ▷ ◁ ◀        **DESCRIPTION**              **AUTOMATON**

| **Origin** | Based on the DECREASING_TERRACE pattern. |
| --- | --- |

| **Constraint** | MIN_SURF_DECREASING_TERRACE(VALUE, VARIABLES) |
| --- | --- |

| **Arguments** | VALUE      : dvar |
| --- | --- |
|  | VARIABLES : collection(var−dvar) |

**Restrictions**

$$\mathtt{sv} \leq 3 \vee \mathtt{rv} \leq 2 \Rightarrow \mathtt{VALUE} = +\infty$$
$$\mathtt{VALUE} \geq \min(2 * (\mathtt{minv} + 1) ①, (\mathtt{sv} - 2) * (\mathtt{minv} + 1) ②)$$
$$\mathtt{VALUE} = +\infty \vee \mathtt{VALUE} \leq \max(2 * (\mathtt{maxv} - 1), (\mathtt{sv} - 2) * (\mathtt{maxv} - 1))$$
required(VARIABLES, var)
where
minv = minval(VARIABLES.var)
sv = |VARIABLES|
maxv = maxval(VARIABLES.var)
rv = range(VARIABLES.var)

**Purpose**

VALUE is the minimal surface of occurrences of the DECREASING_TERRACE pattern in the time-series given by the VARIABLES collection. If the pattern does not occur, VALUE takes the default value $+\infty$.

An occurrence of the pattern DECREASING_TERRACE is the *maximal* subsequence which matches the regular expression $>=^+>$.

Assume that the occurrence of the pattern DECREASING_TERRACE starts at position $i$ and ends at position $j$. The feature SURF computes the sum of the values from index $i + 1$ to index $j$.

**Example**

$$(4, \langle 6, 4, 4, 4, 5, 2, 2, 1, 3, 3, 5, 4, 4, 3, 3, 3 \rangle)$$

Figure 3.810 provides an example where the MIN_SURF_DECREASING_TERRACE $(4, [6, 4, 4, 4, 5, 2, 2, 1, 3, 3, 5, 4, 4, 3, 3, 3])$ constraint holds.

**Typical**

$$|\mathtt{VARIABLES}| > 3$$
$$\mathtt{range}(\mathtt{VARIABLES.var}) > 2$$

**Arg. properties**

Functional dependency: VALUE determined by VARIABLES.

Figure 3.810: Illustrating the MIN_SURF_DECREASING_TERRACE constraint of the **Example** slot

 Figures 3.811 and 3.812 respectively depict the automaton associated with the constraint MIN_SURF_DECREASING_TERRACE and its simplified form.

Figure 3.811: Automaton for the MIN_SURF_DECREASING_TERRACE constraint obtained by applying decoration Table 2.35 to the seed transducer of the DECREAS-ING_TERRACE pattern where default is $+\infty$

Figure 3.812: Automaton for the MIN_SURF_DECREASING_TERRACE constraint obtained by applying decoration Table 2.25 to the seed transducer of the DECREAS-ING_TERRACE pattern where default is $+\infty$; $-R_i + R_{i-1} \geq 0$ is a linear invariant.

|   | $s$ | $r$ | $t$ |
|---|---|---|---|
| $s$ | $\min(\overrightarrow{C}, \overleftarrow{C})$ | $\min(\overrightarrow{C}, \overleftarrow{C})$ | $\min(\overrightarrow{C}, \overleftarrow{C})$ |
| $r$ | $\min(\overrightarrow{C}, \overleftarrow{C})$ | $\min(\overrightarrow{C}, \overleftarrow{C})$ | $\overrightarrow{D} + \overleftarrow{D} + \texttt{VAR}_i$ |
| $t$ | $\min(\overrightarrow{C}, \overleftarrow{C})$ | $\overrightarrow{D} + \overleftarrow{D} + \texttt{VAR}_i$ | $\overrightarrow{D} + \overleftarrow{D} + \texttt{VAR}_i$ |

Table 3.90: Glue matrix for the MIN_SURF_DECREASING_TERRACE constraint defined as the composition of the DECREASING_TERRACE pattern, the feature SURF, and the aggregator min; cells of the glue matrix are coloured with the colour of the constituent to which they are related.

# 3.351   MIN_SURF_DIP_ON_INCREASING_SEQUENCE

▶ ▷ ◁ ◀     **DESCRIPTION**          **AUTOMATON**

| **Origin** | Based on the DIP_ON_INCREASING_SEQUENCE pattern. |

| **Constraint** | MIN_SURF_DIP_ON_INCREASING_SEQUENCE(VALUE, VARIABLES) |

| **Arguments** | VALUE      : dvar |
|               | VARIABLES  : collection(var−dvar) |

**Restrictions**

$sv \leq 5 \lor rv \leq 2 \Rightarrow$ VALUE $= +\infty$

VALUE $\geq 3 * $ minv $+ 3$ ①

VALUE $= +\infty \lor$ VALUE $\leq 3 * $ maxv $- 3$

required(VARIABLES, var)

where

minv = minval(VARIABLES.var)

maxv = maxval(VARIABLES.var)

sv = |VARIABLES|

rv = range(VARIABLES.var)

**Purpose**

VALUE is the minimal surface of occurrences of the DIP_ON_INCREASING_SEQUENCE pattern in the time-series given by the VARIABLES collection. If the pattern does not occur, VALUE takes the default value $+\infty$.

An occurrence of the pattern DIP_ON_INCREASING_SEQUENCE is the subsequence which matches the regular expression <<><<.

Assume that the occurrence of the pattern DIP_ON_INCREASING_SEQUENCE starts at position $i$ and ends at position $j$. The feature SURF computes the sum of the values from index $i + 2$ to index $j$.

**Example**

$(9, \langle 1, 2, 3, 2, 5, 6, 7, 4, 1, 3, 4, 6, 1, 2, 4, 4 \rangle)$

Figure 3.813 provides an example where the MIN_SURF_DIP_ON_INCREASING_SEQUENCE $(9, [1, 2, 3, 2, 5, 6, 7, 4, 1, 3, 4, 6, 1, 2, 4, 4])$ constraint holds.

**Typical**

|VARIABLES| $> 5$

range(VARIABLES.var) $> 2$

**Arg. properties**

Functional dependency: VALUE determined by VARIABLES.

Figure 3.813: Illustrating the MIN_SURF_DIP_ON_INCREASING_SEQUENCE constraint of the **Example** slot

**Automaton**    Figures 3.814 and 3.815 respectively depict the automaton associated with the constraint
                 MIN_SURF_DIP_ON_INCREASING_SEQUENCE and its simplified form.

Figure 3.814: Automaton for the MIN_SURF_DIP_ON_INCREASING_SEQUENCE constraint obtained by applying decoration Table 2.35 to the seed transducer of the DIP_ON_INCREASING_SEQUENCE pattern where default is $+\infty$

Figure 3.815: Automaton for the MIN_SURF_DIP_ON_INCREASING_SEQUENCE constraint obtained by applying decoration Table 2.25 to the seed transducer of the DIP_ON_INCREASING_SEQUENCE pattern where default is $+\infty$; $-R_i + R_{i-1} \geq 0$ is a linear invariant.

## 3.352 MIN_SURF_GORGE

   DESCRIPTION          AUTOMATON

$(> | > (= | >)^* >)(< | < (= | <)^* <)$

| | |
|---|---|
| **Origin** | Based on the GORGE pattern. |
| **Constraint** | MIN_SURF_GORGE(VALUE, VARIABLES) |
| **Arguments** | VALUE      : dvar<br>VARIABLES : collection(var−dvar) |

**Restrictions**

$\mathtt{sv} \leq 2 \vee \mathtt{rv} \leq 1 \Rightarrow \mathtt{VALUE} = +\infty$

$\mathtt{rv} = 2 \Rightarrow \mathtt{VALUE} \geq \mathtt{minv}$①

$\mathtt{rv} \geq 3 \Rightarrow \mathtt{VALUE} \geq \min(\mathtt{minv}②, (\mathtt{sv} - 2) * (\mathtt{minv} + 1) - 1③)$

$\mathtt{rv} = 2 \Rightarrow \mathtt{VALUE} = +\infty \vee \mathtt{VALUE} \leq \mathtt{maxv} - 1$

$\mathtt{rv} \geq 3 \Rightarrow$

$\quad \mathtt{VALUE} = +\infty \vee \mathtt{VALUE} \leq \max(\mathtt{maxv} - 1, (\mathtt{sv} - 2) * (\mathtt{maxv} - 1) - 1)$

required(VARIABLES, var)

where

minv =minval(VARIABLES.var)

rv =range(VARIABLES.var)

sv =|VARIABLES|

maxv =maxval(VARIABLES.var)

**Purpose**

VALUE is the minimal surface of occurrences of the GORGE pattern in the time-series given by the VARIABLES collection. If the pattern does not occur, VALUE takes the default value $+\infty$.

An occurrence of the pattern GORGE is the *maximal* subsequence which matches the regular expression $(> | > (= | >)^* >)(< | < (= | <)^* <)$.

Assume that the occurrence of the pattern GORGE starts at position $i$ and ends at position $j$. The feature SURF computes the sum of the values from index $i + 1$ to index $j$.

**Example**

$(5, \langle 1, 7, 3, 4, 4, 5, 5, 4, 2, 2, 6, 5, 4, 6, 5, 7 \rangle)$

Figure 3.816 provides an example where the MIN_SURF_GORGE $(5, [1, 7, 3, 4, 4, 5, 5, 4, 2, 2, 6, 5, 4, 6, 5, 7])$ constraint holds.

**Typical**

$|\mathtt{VARIABLES}| > 2$

$\mathtt{range}(\mathtt{VARIABLES.var}) > 1$

**Symmetry** | Items of VARIABLES can be reversed.

**Arg. properties** | Functional dependency: VALUE determined by VARIABLES.

Figure 3.816: Illustrating the MIN_SURF_GORGE constraint of the **Example** slot

**Automaton**    Figure 3.817 depicts the automaton associated with the constraint MIN_SURF_GORGE.

Figure 3.817: Automaton for the MIN_SURF_GORGE constraint obtained by applying decoration Table 2.35 to the seed transducer of the GORGE pattern where default is $+\infty$ (transition $u \to r$ has the same accumulator update as transition $r \to u$); $-R_i + R_{i-1} \geq 0$ is a linear invariant.

|   | $s$ | $r$ | $t$ | $u$ |
|---|---|---|---|---|
| $s$ | $\min(\overrightarrow{C}, \overleftarrow{C})$ | $\min(\overrightarrow{C}, \overleftarrow{C})$ | $\min(\overrightarrow{C}, \overleftarrow{C})$ | $\min(\overrightarrow{C}, \overleftarrow{C})$ |
| $r$ | $\min(\overrightarrow{C}, \overleftarrow{C})$ | $\overrightarrow{D} + \overleftarrow{D} + \mathtt{VAR}_i$ | $\overleftarrow{C} + \overrightarrow{D} + \overleftarrow{D} + \mathtt{VAR}_i$ | $\min(\overrightarrow{C}, \overleftarrow{C})$ |
| $t$ | $\min(\overrightarrow{C}, \overleftarrow{C})$ | $\overrightarrow{C} + \overrightarrow{D} + \overleftarrow{D} + \mathtt{VAR}_i$ | $\min(\overrightarrow{C}, \overleftarrow{C})$ | $\overrightarrow{C} + \overrightarrow{D} + \overleftarrow{D} + \mathtt{VAR}_i$ |
| $u$ | $\min(\overrightarrow{C}, \overleftarrow{C})$ | $\min(\overrightarrow{C}, \overleftarrow{C})$ | $\overleftarrow{C} + \overrightarrow{D} + \overleftarrow{D} + \mathtt{VAR}_i$ | $\min(\overrightarrow{C}, \overleftarrow{C})$ |

Table 3.91: Glue matrix for the MIN_SURF_GORGE constraint defined as the composition of the GORGE pattern , the feature SURF , and the aggregator min ; cells of the glue matrix are coloured with the colour of the constituent to which they are related.

## 3.353  MIN_SURF_INCREASING

► ▷ ◁ ◀        **DESCRIPTION**        **AUTOMATON**

**Origin**         Based on the INCREASING pattern.

**Constraint**     MIN_SURF_INCREASING(VALUE, VARIABLES)

**Arguments**      VALUE      : dvar
                   VARIABLES  : collection(var−dvar)

**Restrictions**   $sv \leq 1 \lor rv \leq 1 \Rightarrow$ VALUE $= +\infty$
                   VALUE $\geq 2 * minv + 1$①
                   VALUE $= +\infty \lor$ VALUE $\leq 2 * maxv - 1$
                   required(VARIABLES, var)
                   where
                   minv =minval(VARIABLES.var)
                   maxv =maxval(VARIABLES.var)
                   sv = |VARIABLES|
                   rv =range(VARIABLES.var)

**Purpose**        VALUE is the minimal surface of occurrences of the INCREASING pattern in the time-
                   series given by the VARIABLES collection. If the pattern does not occur, VALUE takes the
                   default value $+\infty$.
                   An occurrence of the pattern INCREASING is the subsequence which matches the regular
                   expression $<$.
                   Assume that the occurrence of the pattern INCREASING starts at position $i$ and ends at
                   position $j$. The feature SURF computes the sum of the values from index $i$ to index $j+1$.

**Example**        $(4, \langle 4, 3, 5, 5, 2, 1, 1, 3, 3, 4, 6, 6, 3, 1, 3, 3\rangle)$

                   Figure 3.818 provides an example where the MIN_SURF_INCREASING
                   $(4, [4, 3, 5, 5, 2, 1, 1, 3, 3, 4, 6, 6, 3, 1, 3, 3])$ constraint holds.

**Typical**        |VARIABLES| $> 1$
                   range(VARIABLES.var) $> 1$

**Arg. properties** Functional dependency: VALUE determined by VARIABLES.

Figure 3.818: Illustrating the MIN_SURF_INCREASING constraint of the **Example** slot

**Automaton**     Figures 3.819 and 3.820 respectively depict the automaton associated with the constraint MIN_SURF_INCREASING and its simplified form.

Figure 3.819: Automaton for the MIN_SURF_INCREASING constraint obtained by applying decoration Table 2.35 to the seed transducer of the INCREASING pattern where `default` is $+\infty$

Figure 3.820: Automaton for the MIN_SURF_INCREASING constraint obtained by applying decoration Table 2.38 to the seed transducer of the INCREASING pattern where `default` is $+\infty$; $-R_i + R_{i-1} \geq 0$ is a linear invariant.

Table 3.92: Glue matrix for the MIN_SURF_INCREASING constraint defined as the composition of the INCREASING pattern, the feature SURF, and the aggregator min; cells of the glue matrix are coloured with the colour of the constituent to which they are related.

## 3.354 MIN_SURF_INCREASING_SEQUENCE

► ▷ ◁ ◀      **DESCRIPTION**      **AUTOMATON**

| **Origin** | Based on the INCREASING_SEQUENCE pattern. |
|---|---|

| **Constraint** | MIN_SURF_INCREASING_SEQUENCE(VALUE, VARIABLES) |
|---|---|

| **Arguments** | VALUE      : dvar |
|---|---|
| | VARIABLES : collection(var−dvar) |

**Restrictions**

$\mathtt{sv} \leq 1 \vee \mathtt{rv} \leq 1 \Rightarrow \mathtt{VALUE} = +\infty$
$\mathtt{rv} = 2 \Rightarrow \mathtt{VALUE} \geq 2 * \mathtt{minv} + 1$①
$\mathtt{rv} \geq 3 \Rightarrow \mathtt{VALUE} \geq \min(2 * \mathtt{minv} + 1$②$, \mathtt{sv} * (\mathtt{minv} + 1))$③
$\mathtt{rv} = 2 \Rightarrow \mathtt{VALUE} = +\infty \vee \mathtt{VALUE} \leq 2 * \mathtt{maxv} - 1$
$\mathtt{rv} \geq 3 \Rightarrow \mathtt{VALUE} = +\infty \vee \mathtt{VALUE} \leq \max(2 * \mathtt{maxv} - 1, \mathtt{sv} * (\mathtt{maxv} - 1))$
required(VARIABLES, var)
where

$\mathtt{minv} = \mathtt{minval}(\mathtt{VARIABLES.var})$
$\mathtt{rv} = \mathtt{range}(\mathtt{VARIABLES.var})$
$\mathtt{sv} = |\mathtt{VARIABLES}|$
$\mathtt{maxv} = \mathtt{maxval}(\mathtt{VARIABLES.var})$

**Purpose**

VALUE is the minimal surface of occurrences of the INCREASING_SEQUENCE pattern in the time-series given by the VARIABLES collection. If the pattern does not occur, VALUE takes the default value $+\infty$.

An occurrence of the pattern INCREASING_SEQUENCE is the *maximal* subsequence which matches the regular expression $< (< | =)^* < | <$.

Assume that the occurrence of the pattern INCREASING_SEQUENCE starts at position $i$ and ends at position $j$. The feature SURF computes the sum of the values from index $i$ to index $j + 1$.

**Example**

$(4, \langle 4, 3, 5, 5, 2, 1, 1, 3, 3, 4, 6, 6, 3, 1, 3, 3 \rangle)$

Figure 3.821 provides an example where the MIN_SURF_INCREASING_SEQUENCE $(4, [4, 3, 5, 5, 2, 1, 1, 3, 3, 4, 6, 6, 3, 1, 3, 3])$ constraint holds.

| **Typical** | $|\mathtt{VARIABLES}| > 1$ |
|---|---|
| | $\mathtt{range}(\mathtt{VARIABLES.var}) > 1$ |

| **Arg. properties** | Functional dependency: VALUE determined by VARIABLES. |
|---|---|

Figure 3.821: Illustrating the MIN_SURF_INCREASING_SEQUENCE constraint of the **Example** slot

**Automaton**    Figure 3.822 depicts the automaton associated with the constraint MIN_SURF_INCREASING_SEQUENCE.

Figure 3.822: Automaton for the MIN_SURF_INCREASING_SEQUENCE constraint obtained by applying decoration Table 2.35 to the seed transducer of the INCREASING_SEQUENCE pattern where `default` is $+\infty$; $-R_i + R_{i-1} \geq 0$ is a linear invariant.

|   | $s$ | $t$ |
|---|---|---|
| $s$ | $\min(\overrightarrow{C}, \overleftarrow{C})$ | $\min(\overrightarrow{C}, \overleftarrow{C})$ |
| $t$ | $\min(\overrightarrow{C}, \overleftarrow{C})$ | $\overrightarrow{C} + \overleftarrow{C} + \overrightarrow{D} + \overleftarrow{D} + \mathtt{VAR}_i$ |

Table 3.93: Glue matrix for the MIN_SURF_INCREASING_SEQUENCE constraint defined as the composition of the INCREASING_SEQUENCE pattern, the feature SURF, and the aggregator min; cells of the glue matrix are coloured with the colour of the constituent to which they are related.

# 3.355　MIN_SURF_INCREASING_TERRACE

▶ ▷ ◁ ◀　　　**DESCRIPTION**　　　**AUTOMATON**

| | |
|---|---|
| **Origin** | Based on the INCREASING_TERRACE pattern. |
| **Constraint** | MIN_SURF_INCREASING_TERRACE(VALUE, VARIABLES) |
| **Arguments** | VALUE　　　　: dvar |
| | VARIABLES : collection(var−dvar) |

**Restrictions**

$$\mathtt{sv} \le 3 \vee \mathtt{rv} \le 2 \Rightarrow \mathtt{VALUE} = +\infty$$
$$\mathtt{VALUE} \ge \min(2 * (\mathtt{minv} + 1)\textcircled{\scriptsize 1}, (\mathtt{sv} - 2) * (\mathtt{minv} + 1)\textcircled{\scriptsize 2})$$
$$\mathtt{VALUE} = +\infty \vee \mathtt{VALUE} \le \max(2 * (\mathtt{maxv} - 1), (\mathtt{sv} - 2) * (\mathtt{maxv} - 1))$$
required(VARIABLES, var)
where
$$\mathtt{minv} = \mathtt{minval}(\mathtt{VARIABLES.var})$$
$$\mathtt{sv} = |\mathtt{VARIABLES}|$$
$$\mathtt{maxv} = \mathtt{maxval}(\mathtt{VARIABLES.var})$$
$$\mathtt{rv} = \mathtt{range}(\mathtt{VARIABLES.var})$$

**Purpose**

VALUE is the minimal surface of occurrences of the INCREASING_TERRACE pattern in the time-series given by the VARIABLES collection. If the pattern does not occur, VALUE takes the default value $+\infty$.

An occurrence of the pattern INCREASING_TERRACE is the *maximal* subsequence which matches the regular expression $<=^+<$.

Assume that the occurrence of the pattern INCREASING_TERRACE starts at position $i$ and ends at position $j$. The feature SURF computes the sum of the values from index $i + 1$ to index $j$.

**Example**

$$(9, \langle 1, 3, 3, 3, 2, 5, 5, 6, 4, 4, 2, 3, 3, 3, 4, 4 \rangle)$$

Figure 3.823 provides an example where the MIN_SURF_INCREASING_TERRACE $(9, [1, 3, 3, 3, 2, 5, 5, 6, 4, 4, 2, 3, 3, 3, 4, 4])$ constraint holds.

| | |
|---|---|
| **Typical** | $|\mathtt{VARIABLES}| > 3$ |
| | range(VARIABLES.var) $> 2$ |
| **Arg. properties** | Functional dependency: VALUE determined by VARIABLES. |

Figure 3.823: Illustrating the MIN_SURF_INCREASING_TERRACE constraint of the **Example** slot

 Figures 3.824 and 3.825 respectively depict the automaton associated with the constraint MIN_SURF_INCREASING_TERRACE and its simplified form.

Figure 3.824: Automaton for the MIN_SURF_INCREASING_TERRACE constraint obtained by applying decoration Table 2.35 to the seed transducer of the INCREASING_TERRACE pattern where default is $+\infty$

Figure 3.825: Automaton for the MIN_SURF_INCREASING_TERRACE constraint obtained by applying decoration Table 2.25 to the seed transducer of the INCREASING_TERRACE pattern where default is $+\infty$; $-R_i + R_{i-1} \geq 0$ is a linear invariant.

|   | $s$ | $r$ | $t$ |
|---|---|---|---|
| $s$ | $\min(\overrightarrow{C}, \overleftarrow{C})$ | $\min(\overrightarrow{C}, \overleftarrow{C})$ | $\min(\overrightarrow{C}, \overleftarrow{C})$ |
| $r$ | $\min(\overrightarrow{C}, \overleftarrow{C})$ | $\min(\overrightarrow{C}, \overleftarrow{C})$ | $\overrightarrow{D} + \overleftarrow{D} + \mathtt{VAR}_i$ |
| $t$ | $\min(\overrightarrow{C}, \overleftarrow{C})$ | $\overrightarrow{D} + \overleftarrow{D} + \mathtt{VAR}_i$ | $\overrightarrow{D} + \overleftarrow{D} + \mathtt{VAR}_i$ |

Table 3.94: Glue matrix for the MIN_SURF_INCREASING_TERRACE constraint defined as the composition of the INCREASING_TERRACE pattern , the feature SURF , and the aggregator min ; cells of the glue matrix are coloured with the colour of the constituent to which they are related.

# 3.356   MIN_SURF_INFLEXION

▶ ▷ ◁ ◀          **DESCRIPTION**          **AUTOMATON**

| **Origin** | Based on the INFLEXION pattern. |
|---|---|

| **Constraint** | MIN_SURF_INFLEXION(VALUE, VARIABLES) |
|---|---|

| **Arguments** | VALUE      :   dvar |
|---|---|
| | VARIABLES : collection(var−dvar) |

**Restrictions**

$sv \leq 2 \vee rv \leq 1 \Rightarrow$ VALUE $= +\infty$
VALUE $\geq \min(\text{minv}①, (sv - 2) * \text{minv}②)$
VALUE $= +\infty \vee$ VALUE $\leq \max(\text{maxv}, (sv - 2) * \text{maxv})$
required(VARIABLES, var)
where
minv = minval(VARIABLES.var)
sv = |VARIABLES|
maxv = maxval(VARIABLES.var)
rv = range(VARIABLES.var)

**Purpose**

VALUE is the minimal surface of occurrences of the INFLEXION pattern in the time-series given by the VARIABLES collection. If the pattern does not occur, VALUE takes the default value $+\infty$.

An occurrence of the pattern INFLEXION is the *maximal* subsequence which matches the regular expression $< (< | =)^* > | > (> | =)^* <$.

Assume that the occurrence of the pattern INFLEXION starts at position $i$ and ends at position $j$. The feature SURF computes the sum of the values from index $i + 1$ to index $j$.

**Example**

$(1, \langle 1, 2, 6, 6, 4, 4, 3, 5, 2, 5, 1, 5, 3, 3, 4, 4 \rangle)$

Figure 3.826 provides an example where the MIN_SURF_INFLEXION $(1, [1, 2, 6, 6, 4, 4, 3, 5, 2, 5, 1, 5, 3, 3, 4, 4])$ constraint holds.

| **Typical** | $|$VARIABLES$| > 2$ |
|---|---|
| | range(VARIABLES.var) $> 1$ |

| **Symmetry** | Items of VARIABLES can be reversed. |
|---|---|

| **Arg. properties** | Functional dependency: VALUE determined by VARIABLES. |
|---|---|

Figure 3.826: Illustrating the MIN_SURF_INFLEXION constraint of the **Example** slot

**Automaton**       Figures 3.827 and 3.828 respectively depict the automaton associated with the constraint
                    MIN_SURF_INFLEXION and its simplified form.

Figure 3.827: Automaton for the MIN_SURF_INFLEXION constraint obtained by applying decoration Table 2.35 to the seed transducer of the INFLEXION pattern where `default` is $+\infty$ (transition $r \rightarrow t$ has the same accumulators updates as transition $t \rightarrow r$)

Figure 3.828: Automaton for the MIN_SURF_INFLEXION constraint obtained by applying decoration Table 2.25 to the seed transducer of the INFLEXION pattern where `default` is $+\infty$ (transition $r \rightarrow t$ has the same accumulators updates as transition $t \rightarrow r$); $-R_i + R_{i-1} \geq 0$ is a linear invariant.

## 3.357 MIN_SURF_PEAK

► ▷ ◁ ◄          **DESCRIPTION**          **AUTOMATON**

$< (= | <)^* (> | =)^* >$

| **Origin** | Based on the PEAK pattern. |
|---|---|

| **Constraint** | MIN_SURF_PEAK(VALUE, VARIABLES) |
|---|---|

| **Arguments** | VALUE      : dvar |
|---|---|
| | VARIABLES : collection(var−dvar) |

| **Restrictions** | $\mathtt{sv} \leq 2 \vee \mathtt{rv} \leq 1 \Rightarrow \mathtt{VALUE} = +\infty$ |
|---|---|
| | $\mathtt{VALUE} \geq \min(\mathtt{minv} + 1①, (\mathtt{sv} - 2) * (\mathtt{minv} + 1)②)$ |
| | $\mathtt{VALUE} = +\infty \vee \mathtt{VALUE} \leq \max(\mathtt{maxv}, (\mathtt{sv} - 2) * \mathtt{maxv})$ |
| | required(VARIABLES, var) |
| | where |
| | $\mathtt{minv} = \mathtt{minval}(\mathtt{VARIABLES.var})$ |
| | $\mathtt{sv} = |\mathtt{VARIABLES}|$ |
| | $\mathtt{maxv} = \mathtt{maxval}(\mathtt{VARIABLES.var})$ |
| | $\mathtt{rv} = \mathtt{range}(\mathtt{VARIABLES.var})$ |

VALUE ≥ **1**        VALUE ≥ −**18** = (11 − 2) * (−3 + 1)

| **Purpose** | VALUE is the minimal surface of occurrences of the PEAK pattern in the time-series given by the VARIABLES collection. If the pattern does not occur, VALUE takes the default value $+\infty$. |
|---|---|
| | An occurrence of the pattern PEAK is the *maximal* subsequence which matches the regular expression $< (= | <)^* (> | =)^* >$. |
| | Assume that the occurrence of the pattern PEAK starts at position $i$ and ends at position $j$. The feature SURF computes the sum of the values from index $i + 1$ to index $j$. |

| **Example** | $(9, \langle 7, 5, 5, 1, 4, 5, 2, 2, 3, 5, 6, 2, 3, 3, 3, 1 \rangle)$ |
|---|---|

Figure 3.829 provides an example where the MIN_SURF_PEAK $(9, [7, 5, 5, 1, 4, 5, 2, 2, 3, 5, 6, 2, 3, 3, 3, 1])$ constraint holds.

| **Typical** | $|\mathtt{VARIABLES}| > 2$ |
|---|---|
| | range(VARIABLES.var) $> 1$ |

| **Symmetry** | Items of VARIABLES can be reversed. |
|---|---|

| **Arg. properties** | Functional dependency: VALUE determined by VARIABLES. |
|---|---|

Figure 3.829: Illustrating the MIN_SURF_PEAK constraint of the **Example** slot

**Automaton**  Figure 3.830 depicts the automaton associated with the constraint MIN_SURF_PEAK.

Figure 3.830: Automaton for the MIN_SURF_PEAK constraint obtained by applying decoration Table 2.35 to the seed transducer of the PEAK pattern where default is $+\infty$; $-R_i + R_{i-1} \geq 0$ is a linear invariant.

| | $s$ | $r$ | $t$ |
|---|---|---|---|
| $s$ | $\min(\overrightarrow{C}, \overleftarrow{C})$ | $\min(\overrightarrow{C}, \overleftarrow{C})$ | $\min(\overrightarrow{C}, \overleftarrow{C})$ |
| $r$ | $\min(\overrightarrow{C}, \overleftarrow{C})$ | $\overrightarrow{D} + \overleftarrow{D} + \mathtt{VAR}_i$ | $\overleftarrow{C} + \overrightarrow{D} + \overleftarrow{D} + \mathtt{VAR}_i$ |
| $t$ | $\min(\overrightarrow{C}, \overleftarrow{C})$ | $\overrightarrow{C} + \overrightarrow{D} + \overleftarrow{D} + \mathtt{VAR}_i$ | $\min(\overrightarrow{C}, \overleftarrow{C})$ |

Table 3.95: Glue matrix for the MIN_SURF_PEAK constraint defined as the composition of the PEAK pattern, the feature SURF, and the aggregator min; cells of the glue matrix are coloured with the colour of the constituent to which they are related.

## 3.358   MIN_SURF_PLAIN

► ▷ ◁ ◀        DESCRIPTION        AUTOMATON

**Origin**        Based on the PLAIN pattern.

**Constraint**        MIN_SURF_PLAIN(VALUE, VARIABLES)

**Arguments**        VALUE      : dvar
                     VARIABLES  : collection(var−dvar)

**Restrictions**     $\text{sv} \leq 2 \vee \text{rv} \leq 1 \Rightarrow \text{VALUE} = +\infty$
                     $\text{VALUE} \geq \min(\text{minv}①, (\text{sv} - 2) * \text{minv}②)$
                     $\text{VALUE} = +\infty \vee \text{VALUE} \leq \max(\text{maxv} - 1, (\text{sv} - 2) * (\text{maxv} - 1))$
                     required(VARIABLES, var)
                     where
                     $\text{minv} = \text{minval}(\text{VARIABLES.var})$
                     $\text{sv} = |\text{VARIABLES}|$
                     $\text{maxv} = \text{maxval}(\text{VARIABLES.var})$
                     $\text{rv} = \text{range}(\text{VARIABLES.var})$

**Purpose**        VALUE is the minimal surface of occurrences of the PLAIN pattern in the time-series given by the VARIABLES collection. If the pattern does not occur, VALUE takes the default value $+\infty$.
                   An occurrence of the pattern PLAIN is the *maximal* subsequence which matches the regular expression $> =^* <$.
                   Assume that the occurrence of the pattern PLAIN starts at position $i$ and ends at position $j$. The feature SURF computes the sum of the values from index $i + 1$ to index $j$.

**Example**        $(4, \langle 2, 3, 6, 5, 7, 6, 6, 4, 5, 5, 4, 3, 3, 6, 6, 3\rangle)$

Figure 3.831 provides an example where the MIN_SURF_PLAIN $(4, [2, 3, 6, 5, 7, 6, 6, 4, 5, 5, 4, 3, 3, 6, 6, 3])$ constraint holds.

**Typical**        $|\text{VARIABLES}| > 2$
                   $\text{range}(\text{VARIABLES.var}) > 1$

**Symmetry**       Items of VARIABLES can be reversed.

**Arg. properties**   Functional dependency: VALUE determined by VARIABLES.

Figure 3.831: Illustrating the MIN_SURF_PLAIN constraint of the **Example** slot

**Automaton**        Figures 3.832 and 3.833 respectively depict the automaton associated with the constraint
                     MIN_SURF_PLAIN and its simplified form.

Figure 3.832: Automaton for the MIN_SURF_PLAIN constraint obtained by applying decoration Table 2.35 to the seed transducer of the PLAIN pattern where default is $+\infty$

Figure 3.833: Automaton for the MIN_SURF_PLAIN constraint obtained by applying decoration Table 2.25 to the seed transducer of the PLAIN pattern where default is $+\infty$; $-R_i + R_{i-1} \geq 0$ is a linear invariant.

|     | $s$ | $r$ | $t$ |
| --- | --- | --- | --- |
| $s$ | $\min(\overrightarrow{C}, \overleftarrow{C})$ | $\min(\overrightarrow{C}, \overleftarrow{C})$ | $\min(\overrightarrow{C}, \overleftarrow{C})$ |
| $r$ | $\min(\overrightarrow{C}, \overleftarrow{C})$ | $\overrightarrow{D} + \overleftarrow{D} + \mathtt{VAR}_i$ | $\overrightarrow{D} + \overleftarrow{D} + \mathtt{VAR}_i$ |
| $t$ | $\min(\overrightarrow{C}, \overleftarrow{C})$ | $\overrightarrow{D} + \overleftarrow{D} + \mathtt{VAR}_i$ | $\overrightarrow{D} + \overleftarrow{D} + \mathtt{VAR}_i$ |

Table 3.96: Glue matrix for the MIN_SURF_PLAIN constraint defined as the composition of the PLAIN pattern, the feature SURF, and the aggregator min; cells of the glue matrix are coloured with the colour of the constituent to which they are related.

## 3.359   MIN_SURF_PLATEAU

▶ ▷ ◁ ◀        DESCRIPTION          AUTOMATON

| | |
|---|---|
| **Origin** | Based on the PLATEAU pattern. |
| **Constraint** | MIN_SURF_PLATEAU(VALUE, VARIABLES) |
| **Arguments** | VALUE       : dvar<br>VARIABLES : collection(var−dvar) |

**Restrictions**

$$\text{sv} \leq 2 \vee \text{rv} \leq 1 \Rightarrow \text{VALUE} = +\infty$$
$$\text{VALUE} \geq \min(\text{minv} + 1 ①, (\text{sv} - 2) * (\text{minv} + 1) ②)$$
$$\text{VALUE} = +\infty \vee \text{VALUE} \leq \max(\text{maxv}, (\text{sv} - 2) * \text{maxv})$$
required(VARIABLES, var)

where

minv = minval(VARIABLES.var)
sv = |VARIABLES|
maxv = maxval(VARIABLES.var)
rv = range(VARIABLES.var)

**Purpose**

VALUE is the minimal surface of occurrences of the PLATEAU pattern in the time-series given by the VARIABLES collection. If the pattern does not occur, VALUE takes the default value $+\infty$.

An occurrence of the pattern PLATEAU is the *maximal* subsequence which matches the regular expression $< =^* >$.

Assume that the occurrence of the pattern PLATEAU starts at position $i$ and ends at position $j$. The feature SURF computes the sum of the values from index $i + 1$ to index $j$.

**Example**

$$(3, \langle 7, 5, 2, 3, 1, 2, 2, 4, 3, 3, 4, 5, 5, 2, 2, 5 \rangle)$$

Figure 3.834 provides an example where the MIN_SURF_PLATEAU $(3, [7, 5, 2, 3, 1, 2, 2, 4, 3, 3, 4, 5, 5, 2, 2, 5])$ constraint holds.

| | |
|---|---|
| **Typical** | \|VARIABLES\| > 2<br>range(VARIABLES.var) > 1 |
| **Symmetry** | Items of VARIABLES can be reversed. |
| **Arg. properties** | Functional dependency: VALUE determined by VARIABLES. |

Figure 3.834: Illustrating the MIN_SURF_PLATEAU constraint of the **Example** slot

 Figures 3.835 and 3.836 respectively depict the automaton associated with the constraint
MIN_SURF_PLATEAU and its simplified form.

Figure 3.835: Automaton for the MIN_SURF_PLATEAU constraint obtained by applying
decoration Table 2.35 to the seed transducer of the PLATEAU pattern where default is
$+\infty$

Figure 3.836: Automaton for the MIN_SURF_PLATEAU constraint obtained by applying
decoration Table 2.25 to the seed transducer of the PLATEAU pattern where default
is $+\infty$; $-R_i + R_{i-1} \geq 0$ is a linear invariant.

|       | $s$                                      | $r$                                                  | $t$                                                  |
|-------|------------------------------------------|------------------------------------------------------|------------------------------------------------------|
| $s$   | $\min(\overrightarrow{C},\overleftarrow{C})$ | $\min(\overrightarrow{C},\overleftarrow{C})$ | $\min(\overrightarrow{C},\overleftarrow{C})$ |
| $r$   | $\min(\overrightarrow{C},\overleftarrow{C})$ | $\overrightarrow{D}+\overleftarrow{D}+\mathtt{VAR}_i$ | $\overrightarrow{D}+\overleftarrow{D}+\mathtt{VAR}_i$ |
| $t$   | $\min(\overrightarrow{C},\overleftarrow{C})$ | $\overrightarrow{D}+\overleftarrow{D}+\mathtt{VAR}_i$ | $\overrightarrow{D}+\overleftarrow{D}+\mathtt{VAR}_i$ |

Table 3.97: Glue matrix for the MIN_SURF_PLATEAU constraint defined as the composition of the PLATEAU pattern , the feature SURF , and the aggregator min ; cells of the glue matrix are coloured with the colour of the constituent to which they are related.

## 3.360   MIN_SURF_PROPER_PLAIN

► ▷ ◁ ◀          DESCRIPTION          AUTOMATON

| **Origin** | Based on the PROPER_PLAIN pattern. |
| **Constraint** | MIN_SURF_PROPER_PLAIN(VALUE, VARIABLES) |
| **Arguments** | VALUE      : dvar<br>VARIABLES  : collection(var−dvar) |

**Restrictions**

$\mathtt{sv} \leq 3 \vee \mathtt{rv} \leq 1 \Rightarrow \mathtt{VALUE} = +\infty$

$\mathtt{VALUE} \geq \min(2 * \mathtt{minv}①, (\mathtt{sv} - 2) * \mathtt{minv}②)$

$\mathtt{VALUE} = +\infty \vee \mathtt{VALUE} \leq \max(2 * (\mathtt{maxv} - 1), (\mathtt{sv} - 2) * (\mathtt{maxv} - 1))$

required(VARIABLES, var)

where

$\mathtt{minv} = \mathtt{minval}(\mathtt{VARIABLES.var})$

$\mathtt{sv} = |\mathtt{VARIABLES}|$

$\mathtt{maxv} = \mathtt{maxval}(\mathtt{VARIABLES.var})$

$\mathtt{rv} = \mathtt{range}(\mathtt{VARIABLES.var})$

**Purpose**

VALUE is the minimal surface of occurrences of the PROPER_PLAIN pattern in the time-series given by the VARIABLES collection. If the pattern does not occur, VALUE takes the default value $+\infty$.

An occurrence of the pattern PROPER_PLAIN is the *maximal* subsequence which matches the regular expression $> =^+ <$.

Assume that the occurrence of the pattern PROPER_PLAIN starts at position $i$ and ends at position $j$. The feature SURF computes the sum of the values from index $i + 1$ to index $j$.

**Example**

$$(8, \langle 2, 7, 5, 5, 6, 3, 7, 4, 4, 5, 6, 5, 3, 3, 3, 5 \rangle)$$

Figure 3.837 provides an example where the MIN_SURF_PROPER_PLAIN $(8, [2, 7, 5, 5, 6, 3, 7, 4, 4, 5, 6, 5, 3, 3, 3, 5])$ constraint holds.

| **Typical** | $|\mathtt{VARIABLES}| > 3$<br>$\mathtt{range}(\mathtt{VARIABLES.var}) > 1$ |
| **Symmetry** | Items of VARIABLES can be reversed. |
| **Arg. properties** | Functional dependency: VALUE determined by VARIABLES. |

Figure 3.837: Illustrating the MIN_SURF_PROPER_PLAIN constraint of the **Example** slot

 Figures 3.838 and 3.839 respectively depict the automaton associated with the constraint MIN_SURF_PROPER_PLAIN and its simplified form.

Figure 3.838: Automaton for the MIN_SURF_PROPER_PLAIN constraint obtained by applying decoration Table 2.35 to the seed transducer of the PROPER_PLAIN pattern where `default` is $+\infty$

Figure 3.839: Automaton for the MIN_SURF_PROPER_PLAIN constraint obtained by applying decoration Table 2.25 to the seed transducer of the PROPER_PLAIN pattern where `default` is $+\infty$; $-R_i + R_{i-1} \geq 0$ is a linear invariant.

|   | $s$ | $r$ | $t$ |
|---|---|---|---|
| $s$ | $\min(\overrightarrow{C}, \overleftarrow{C})$ | $\min(\overrightarrow{C}, \overleftarrow{C})$ | $\min(\overrightarrow{C}, \overleftarrow{C})$ |
| $r$ | $\min(\overrightarrow{C}, \overleftarrow{C})$ | $\min(\overrightarrow{C}, \overleftarrow{C})$ | $\overrightarrow{D} + \overleftarrow{D} + \texttt{VAR}_i$ |
| $t$ | $\min(\overrightarrow{C}, \overleftarrow{C})$ | $\overrightarrow{D} + \overleftarrow{D} + \texttt{VAR}_i$ | $\overrightarrow{D} + \overleftarrow{D} + \texttt{VAR}_i$ |

Table 3.98: Glue matrix for the MIN_SURF_PROPER_PLAIN constraint defined as the composition of the PROPER_PLAIN pattern , the feature SURF , and the aggregator min ; cells of the glue matrix are coloured with the colour of the constituent to which they are related.

# 3.361 MIN_SURF_PROPER_PLATEAU

▶ ▷ ◁ ◀          DESCRIPTION               AUTOMATON

**Origin**          Based on the PROPER_PLATEAU pattern.

**Constraint**      MIN_SURF_PROPER_PLATEAU(VALUE, VARIABLES)

**Arguments**       VALUE      : dvar
                    VARIABLES  : collection(var−dvar)

**Restrictions**    $sv \leq 3 \vee rv \leq 1 \Rightarrow VALUE = +\infty$
                    $VALUE \geq \min(2*(minv+1)①, (sv-2)*(minv+1)②)$
                    $VALUE = +\infty \vee VALUE \leq \max(2*maxv, (sv-2)*maxv)$
                    required(VARIABLES, var)
                    where
                    minv =minval(VARIABLES.var)
                    $sv = |VARIABLES|$
                    maxv =maxval(VARIABLES.var)
                    rv =range(VARIABLES.var)

**Purpose**         VALUE is the minimal surface of occurrences of the PROPER_PLATEAU pattern in the
                    time-series given by the VARIABLES collection. If the pattern does not occur, VALUE
                    takes the default value $+\infty$.
                    An occurrence of the pattern PROPER_PLATEAU is the *maximal* subsequence which
                    matches the regular expression $<=^+>$.
                    Assume that the occurrence of the pattern PROPER_PLATEAU starts at position $i$ and ends
                    at position $j$. The feature SURF computes the sum of the values from index $i+1$ to index
                    $j$.

**Example**         $(6, \langle 7, 1, 3, 3, 2, 5, 1, 4, 4, 3, 2, 3, 5, 5, 5, 3 \rangle)$

                    Figure 3.840 provides an example where the MIN_SURF_PROPER_PLATEAU
                    $(6, [7, 1, 3, 3, 2, 5, 1, 4, 4, 3, 2, 3, 5, 5, 5, 3])$ constraint holds.

**Typical**         $|VARIABLES| > 3$
                    range(VARIABLES.var) $> 1$

**Symmetry**        Items of VARIABLES can be reversed.

**Arg. properties** Functional dependency: VALUE determined by VARIABLES.

Figure 3.840: Illustrating the MIN_SURF_PROPER_PLATEAU constraint of the **Example** slot

**Automaton**  Figures 3.841 and 3.842 respectively depict the automaton associated with the constraint MIN_SURF_PROPER_PLATEAU and its simplified form.

Figure 3.841: Automaton for the MIN_SURF_PROPER_PLATEAU constraint obtained by applying decoration Table 2.35 to the seed transducer of the PROPER_PLATEAU pattern where default is $+\infty$

Figure 3.842: Automaton for the MIN_SURF_PROPER_PLATEAU constraint obtained by applying decoration Table 2.25 to the seed transducer of the PROPER_PLATEAU pattern where default is $+\infty$; $-R_i + R_{i-1} \geq 0$ is a linear invariant.

|     | $s$ | $r$ | $t$ |
| --- | --- | --- | --- |
| $s$ | $\min(\overrightarrow{C}, \overleftarrow{C})$ | $\min(\overrightarrow{C}, \overleftarrow{C})$ | $\min(\overrightarrow{C}, \overleftarrow{C})$ |
| $r$ | $\min(\overrightarrow{C}, \overleftarrow{C})$ | $\min(\overrightarrow{C}, \overleftarrow{C})$ | $\overrightarrow{D} + \overleftarrow{D} + \mathtt{VAR}_i$ |
| $t$ | $\min(\overrightarrow{C}, \overleftarrow{C})$ | $\overrightarrow{D} + \overleftarrow{D} + \mathtt{VAR}_i$ | $\overrightarrow{D} + \overleftarrow{D} + \mathtt{VAR}_i$ |

Table 3.99: Glue matrix for the MIN_SURF_PROPER_PLATEAU constraint defined as the composition of the PROPER_PLATEAU pattern , the feature SURF , and the aggregator min ; cells of the glue matrix are coloured with the colour of the constituent to which they are related.

## 3.362   MIN_SURF_STEADY

       DESCRIPTION          AUTOMATON

| | |
|---|---|
| **Origin** | Based on the STEADY pattern. |
| **Constraint** | MIN_SURF_STEADY(VALUE, VARIABLES) |
| **Arguments** | VALUE      : dvar <br> VARIABLES  : collection(var−dvar) |

**Restrictions**

$\text{sv} \leq 1 \Rightarrow \text{VALUE} = +\infty$

$\text{VALUE} \geq 2 * \text{minv}①$

$\text{VALUE} = +\infty \vee \text{VALUE} \leq 2 * \text{maxv}$

required(VARIABLES, var)

where

minv = minval(VARIABLES.var)

maxv = maxval(VARIABLES.var)

sv = |VARIABLES|

VALUE $\geq$ **2** = 2 * 1

**Purpose**

VALUE is the minimal surface of occurrences of the STEADY pattern in the time-series given by the VARIABLES collection. If the pattern does not occur, VALUE takes the default value $+\infty$.

An occurrence of the pattern STEADY is the subsequence which matches the regular expression =.

Assume that the occurrence of the pattern STEADY starts at position $i$ and ends at position $j$. The feature SURF computes the sum of the values from index $i$ to index $j + 1$.

**Example**

$(2, \langle 1, 1, 7, 3, 3, 5, 5, 5, 6, 5, 5, 5, 7, 2, 6, 6 \rangle)$

Figure 3.843 provides an example where the MIN_SURF_STEADY $(2, [1, 1, 7, 3, 3, 5, 5, 5, 6, 5, 5, 5, 7, 2, 6, 6])$ constraint holds.

| | |
|---|---|
| **Typical** | $\|\text{VARIABLES}\| > 1$ |
| **Symmetry** | Items of VARIABLES can be reversed. |
| **Arg. properties** | Functional dependency: VALUE determined by VARIABLES. |

Figure 3.843: Illustrating the MIN_SURF_STEADY constraint of the **Example** slot

**Automaton**  Figures 3.844 and 3.845 respectively depict the automaton associated with the constraint MIN_SURF_STEADY and its simplified form.

Figure 3.844: Automaton for the MIN_SURF_STEADY constraint obtained by applying decoration Table 2.35 to the seed transducer of the STEADY pattern where default is $+\infty$

Figure 3.845: Automaton for the MIN_SURF_STEADY constraint obtained by applying decoration Table 2.38 to the seed transducer of the STEADY pattern where default is $+\infty$; $-R_i + R_{i-1} \geq 0$ is a linear invariant.

Table 3.100: Glue matrix for the MIN_SURF_STEADY constraint defined as the composition of the STEADY pattern , the feature SURF , and the aggregator min ; cells of the glue matrix are coloured with the colour of the constituent to which they are related.

## 3.363   MIN_SURF_STEADY_SEQUENCE

| | |
|---|---|
| **Origin** | Based on the STEADY_SEQUENCE pattern. |
| **Constraint** | MIN_SURF_STEADY_SEQUENCE(VALUE, VARIABLES) |
| **Arguments** | VALUE      : dvar |
| | VARIABLES  : collection(var−dvar) |

**Restrictions**

$\text{sv} \leq 1 \Rightarrow \text{VALUE} = +\infty$
$\text{rv} = 1 \Rightarrow \text{VALUE} \geq \text{sv} * \text{minv}①$
$\text{rv} \geq 2 \Rightarrow \text{VALUE} \geq \min(2 * \text{minv}②, \text{sv} * \text{minv}③)$
$\text{rv} = 1 \Rightarrow \text{VALUE} = +\infty \vee \text{VALUE} \leq \text{sv} * \text{maxv}$
$\text{rv} \geq 2 \Rightarrow \text{VALUE} = +\infty \vee \text{VALUE} \leq \max(2 * \text{maxv}, \text{sv} * \text{maxv})$
required(VARIABLES, var)
where
  minv =minval(VARIABLES.var)
  rv =range(VARIABLES.var)
  sv =|VARIABLES|
  maxv =maxval(VARIABLES.var)

**Purpose**

VALUE is the minimal surface of occurrences of the STEADY_SEQUENCE pattern in the time-series given by the VARIABLES collection. If the pattern does not occur, VALUE takes the default value $+\infty$.

An occurrence of the pattern STEADY_SEQUENCE is the *maximal* subsequence which matches the regular expression $=^+$.

Assume that the occurrence of the pattern STEADY_SEQUENCE starts at position $i$ and ends at position $j$. The feature SURF computes the sum of the values from index $i$ to index $j + 1$.

**Example**

$(2, \langle 3, 1, 1, 4, 5, 5, 5, 6, 2, 2, 4, 4, 3, 2, 1, 1 \rangle)$

Figure 3.846 provides an example where the MIN_SURF_STEADY_SEQUENCE $(2, [3, 1, 1, 4, 5, 5, 5, 6, 2, 2, 4, 4, 3, 2, 1, 1])$ constraint holds.

| | |
|---|---|
| **Typical** | $|\text{VARIABLES}| > 1$ |

| | |
|---|---|
| **Symmetry** | Items of VARIABLES can be reversed. |
| **Arg. properties** | Functional dependency: VALUE determined by VARIABLES. |

Figure 3.846: Illustrating the MIN_SURF_STEADY_SEQUENCE constraint of the **Example** slot

 Figures 3.847 and 3.848 respectively depict the automaton associated with the constraint MIN_SURF_STEADY_SEQUENCE and its simplified form.

Figure 3.847: Automaton for the MIN_SURF_STEADY_SEQUENCE constraint obtained by applying decoration Table 2.35 to the seed transducer of the STEADY_SEQUENCE pattern where default is $+\infty$

Figure 3.848: Automaton for the MIN_SURF_STEADY_SEQUENCE constraint obtained by applying decoration Table 2.24 to the seed transducer of the STEADY_SEQUENCE pattern where default is $+\infty$; $-R_i + R_{i-1} \geq 0$ is a linear invariant.

|   | $s$ | $r$ |
|---|---|---|
| $s$ | $\min(\overrightarrow{C}, \overleftarrow{C})$ | $\min(\overrightarrow{C}, \overleftarrow{C})$ |
| $r$ | $\min(\overrightarrow{C}, \overleftarrow{C})$ | $\overrightarrow{C} + \overleftarrow{C} + \overrightarrow{D} + \overleftarrow{D} + \mathtt{VAR}_i$ |

Table 3.101: Glue matrix for the MIN_SURF_STEADY_SEQUENCE constraint defined as the composition of the STEADY_SEQUENCE pattern , the feature SURF , and the aggregator min ; cells of the glue matrix are coloured with the colour of the constituent to which they are related.

# 3.364   MIN_SURF_STRICTLY_DECREASING_SEQUENCE

▶ ▷ ◁ ◀      **DESCRIPTION**      **AUTOMATON**

| | |
|---|---|
| **Origin** | Based on the STRICTLY_DECREASING_SEQUENCE pattern. |
| **Constraint** | MIN_SURF_STRICTLY_DECREASING_SEQUENCE(VALUE, VARIABLES) |
| **Arguments** | VALUE : dvar <br> VARIABLES : collection(var−dvar) |

**Restrictions**

$sv \leq 1 \lor rv \leq 1 \Rightarrow \text{VALUE} = +\infty$

$minv < 0 \Rightarrow \text{VALUE} \geq \ell_1 * minv + \lfloor \ell_1 * (\ell_1 - 1)/2 \rfloor$ ①

$minv \geq 0 \Rightarrow \text{VALUE} \geq 2 * minv + 1$ ②

$maxv > 0 \Rightarrow \text{VALUE} = +\infty \lor \text{VALUE} \leq \ell_2 * maxv - \lfloor \ell_2 * (\ell_2 - 1)/2 \rfloor$

$maxv \leq 0 \Rightarrow \text{VALUE} = +\infty \lor \text{VALUE} \leq 2 * maxv - 1$

required(VARIABLES, var)

where

$minv = $ minval(VARIABLES.var)

$rv = $ range(VARIABLES.var)

$sv = |\text{VARIABLES}|$

$\ell_1 = \min(\min(sv, rv), |minv|)$

$\ell_2 = \min(\min(sv, rv), |maxv|)$

$maxv = $ maxval(VARIABLES.var)

VALUE ≥ −6    ①    $\ell_1 = 3$   $\ell_2 = 1$

VALUE ≥ −1    ①    $\ell_1 = 1$   $\ell_2 = 2$

VALUE ≥ 3    ②

**Purpose**

VALUE is the minimal surface of occurrences of the STRICTLY_DECREASING_SEQUENCE pattern in the time-series given by the VARIABLES collection. If the pattern does not occur, VALUE takes the default value $+\infty$.

An occurrence of the pattern STRICTLY_DECREASING_SEQUENCE is the *maximal* subsequence which matches the regular expression $>^+$.

Assume that the occurrence of the pattern STRICTLY_DECREASING_SEQUENCE starts at position $i$ and ends at position $j$. The feature SURF computes the sum of the values from index $i$ to index $j + 1$.

**Example**

$(7, \langle 4, 4, 6, 4, 1, 1, 3, 4, 4, 6, 6, 5, 2, 2, 4, 3 \rangle)$

Figure 3.849 provides an example where the MIN_SURF_STRICTLY_DECREASING_SEQUENCE $(7, [4, 4, 6, 4, 1, 1, 3, 4, 4, 6, 6, 5, 2, 2, 4, 3])$ constraint holds.

| | |
|---|---|
| **Typical** | $|\text{VARIABLES}| > 1$ <br> range(VARIABLES.var) $> 1$ |
| **Arg. properties** | Functional dependency: VALUE determined by VARIABLES. |

Figure 3.849: Illustrating the MIN_SURF_STRICTLY_DECREASING_SEQUENCE constraint of the **Example** slot

 Figures 3.850 and 3.851 respectively depict the automaton associated with the constraint MIN_SURF_STRICTLY_DECREASING_SEQUENCE and its simplified form.

Figure 3.850: Automaton for the MIN_SURF_STRICTLY_DECREASING_SEQUENCE constraint obtained by applying decoration Table 2.35 to the seed transducer of the STRICTLY_DECREASING_SEQUENCE pattern where default is $+\infty$

Figure 3.851: Automaton for the MIN_SURF_STRICTLY_DECREASING_SEQUENCE constraint obtained by applying decoration Table 2.24 to the seed transducer of the STRICTLY_DECREASING_SEQUENCE pattern where default is $+\infty$; $-R_i + R_{i-1} \geq 0$ is a linear invariant.

|   | $s$ | $r$ |
|---|-----|-----|
| $s$ | $\min(\overrightarrow{C}, \overleftarrow{C})$ | $\min(\overrightarrow{C}, \overleftarrow{C})$ |
| $r$ | $\min(\overrightarrow{C}, \overleftarrow{C})$ | $\overrightarrow{C} + \overleftarrow{C} + \overrightarrow{D} + \overleftarrow{D} + \mathtt{VAR}_i$ |

Table 3.102: Glue matrix for the MIN_SURF_STRICTLY_DECREASING_SEQUENCE constraint defined as the composition of the STRICTLY_DECREASING_SEQUENCE pattern, the feature SURF, and the aggregator min; cells of the glue matrix are coloured with the colour of the constituent to which they are related.

# 3.365  MIN_SURF_STRICTLY_INCREASING_SEQUENCE

▶ ▷ ◁ ◀        **DESCRIPTION**              **AUTOMATON**

| | |
|---|---|
| **Origin** | Based on the STRICTLY_INCREASING_SEQUENCE pattern. |
| **Constraint** | MIN_SURF_STRICTLY_INCREASING_SEQUENCE(VALUE, VARIABLES) |
| **Arguments** | VALUE       : dvar<br>VARIABLES : collection(var−dvar) |

**Restrictions**

$sv \leq 1 \lor rv \leq 1 \Rightarrow VALUE = +\infty$

$minv < 0 \Rightarrow VALUE \geq \ell_1 * minv + \lfloor \ell_1 * (\ell_1 - 1)/2 \rfloor$ ①

$minv \geq 0 \Rightarrow VALUE \geq 2 * minv + 1$ ②

$maxv > 0 \Rightarrow VALUE = +\infty \lor VALUE \leq \ell_2 * maxv - \lfloor \ell_2 * (\ell_2 - 1)/2 \rfloor$

$maxv \leq 0 \Rightarrow VALUE = +\infty \lor VALUE \leq 2 * maxv - 1$

required(VARIABLES, var)

where

$minv =$ minval(VARIABLES.var)

$rv =$ range(VARIABLES.var)

$sv = |$VARIABLES$|$

$\ell_1 = \min(\min(sv, rv), |minv|)$

$\ell_2 = \min(\min(sv, rv), |maxv|)$

$maxv =$ maxval(VARIABLES.var)

**Purpose**

VALUE is the minimal surface of occurrences of the STRICTLY_INCREASING_SEQUENCE pattern in the time-series given by the VARIABLES collection. If the pattern does not occur, VALUE takes the default value $+\infty$.

An occurrence of the pattern STRICTLY_INCREASING_SEQUENCE is the *maximal* sub-sequence which matches the regular expression $<^+$.

Assume that the occurrence of the pattern STRICTLY_INCREASING_SEQUENCE starts at position $i$ and ends at position $j$. The feature SURF computes the sum of the values from index $i$ to index $j + 1$.

**Example**

$(6, \langle 4, 3, 5, 5, 2, 1, 1, 2, 3, 4, 6, 6, 3, 1, 2, 3 \rangle)$

Figure 3.852 provides an example where the MIN_SURF_STRICTLY_INCREASING_SEQUENCE
$(6, [4, 3, 5, 5, 2, 1, 1, 2, 3, 4, 6, 6, 3, 1, 2, 3])$ constraint holds.

| | |
|---|---|
| **Typical** | $|$VARIABLES$| > 1$<br>range(VARIABLES.var) $> 1$ |
| **Arg. properties** | Functional dependency: VALUE determined by VARIABLES. |

Figure 3.852: Illustrating the MIN_SURF_STRICTLY_INCREASING_SEQUENCE constraint of the **Example** slot

 Figures 3.853 and 3.854 respectively depict the automaton associated with the constraint MIN_SURF_STRICTLY_INCREASING_SEQUENCE and its simplified form.

Figure 3.853: Automaton for the MIN_SURF_STRICTLY_INCREASING_SEQUENCE constraint obtained by applying decoration Table 2.35 to the seed transducer of the STRICTLY_INCREASING_SEQUENCE pattern where default is $+\infty$

Figure 3.854: Automaton for the MIN_SURF_STRICTLY_INCREASING_SEQUENCE constraint obtained by applying decoration Table 2.24 to the seed transducer of the STRICTLY_INCREASING_SEQUENCE pattern where default is $+\infty$; $-R_i + R_{i-1} \geq 0$ is a linear invariant.

| | $s$ | $r$ |
|---|---|---|
| $s$ | $\min(\overrightarrow{C}, \overleftarrow{C})$ | $\min(\overrightarrow{C}, \overleftarrow{C})$ |
| $r$ | $\min(\overrightarrow{C}, \overleftarrow{C})$ | $\overrightarrow{C} + \overleftarrow{C} + \overrightarrow{D} + \overleftarrow{D} + \mathtt{VAR}_i$ |

Table 3.103: Glue matrix for the MIN_SURF_STRICTLY_INCREASING_SEQUENCE constraint defined as the composition of the STRICTLY_INCREASING_SEQUENCE pattern, the feature SURF, and the aggregator min; cells of the glue matrix are coloured with the colour of the constituent to which they are related.

# 3.366 MIN_SURF_SUMMIT

▶ ▷ ◁ ◀        DESCRIPTION        AUTOMATON

$(< | < (= | <)^* <)(> | > (= | >)^* >)$

| | |
|---|---|
| **Origin** | Based on the SUMMIT pattern. |
| **Constraint** | MIN_SURF_SUMMIT(VALUE, VARIABLES) |
| **Arguments** | VALUE      : dvar |
| | VARIABLES : collection(var−dvar) |

**Restrictions**

$\text{sv} \leq 2 \vee \text{rv} \leq 1 \Rightarrow \text{VALUE} = +\infty$

$\text{rv} = 2 \Rightarrow \text{VALUE} \geq \text{minv} + 1$ ①

$\text{rv} \geq 3 \Rightarrow \text{VALUE} \geq \min(\text{minv} + 1$①$, (\text{sv} - 2) * (\text{minv} + 1) + 1$②$)$

$\text{rv} = 2 \Rightarrow \text{VALUE} = +\infty \vee \text{VALUE} \leq \text{maxv}$

$\text{rv} \geq 3 \Rightarrow \text{VALUE} = +\infty \vee \text{VALUE} \leq \max(\text{maxv}, (\text{sv} - 2) * (\text{maxv} - 1) + 1)$

required(VARIABLES, var)

where

minv = minval(VARIABLES.var)

rv = range(VARIABLES.var)

sv = |VARIABLES|

maxv = maxval(VARIABLES.var)

**Purpose**

VALUE is the minimal surface of occurrences of the SUMMIT pattern in the time-series given by the VARIABLES collection. If the pattern does not occur, VALUE takes the default value $+\infty$.

An occurrence of the pattern SUMMIT is the *maximal* subsequence which matches the regular expression $(< | < (= | <)^* <)(> | > (= | >)^* >)$.

Assume that the occurrence of the pattern SUMMIT starts at position $i$ and ends at position $j$. The feature SURF computes the sum of the values from index $i + 1$ to index $j$.

**Example**

$(3, \langle 7, 1, 5, 4, 4, 3, 3, 4, 6, 6, 2, 3, 4, 2, 3, 1 \rangle)$

Figure 3.855 provides an example where the MIN_SURF_SUMMIT $(3, [7, 1, 5, 4, 4, 3, 3, 4, 6, 6, 2, 3, 4, 2, 3, 1])$ constraint holds.

| | |
|---|---|
| **Typical** | $\lvert\text{VARIABLES}\rvert > 2$ |
| | range(VARIABLES.var) $> 1$ |
| **Symmetry** | Items of VARIABLES can be reversed. |
| **Arg. properties** | Functional dependency: VALUE determined by VARIABLES. |

Figure 3.855: Illustrating the MIN_SURF_SUMMIT constraint of the **Example** slot

**Automaton**          Figure 3.856 depicts the automaton associated with the constraint MIN_SURF_SUMMIT.

Figure 3.856: Automaton for the MIN_SURF_SUMMIT constraint obtained by applying decoration Table 2.35 to the seed transducer of the SUMMIT pattern where default is $+\infty$ (transition $u \to r$ has the same accumulator update as transition $r \to u$); $-R_i + R_{i-1} \geq 0$ is a linear invariant.

|   | $s$ | $r$ | $t$ | $u$ |
|---|-----|-----|-----|-----|
| $s$ | $\min(\overrightarrow{C}, \overleftarrow{C})$ | $\min(\overrightarrow{C}, \overleftarrow{C})$ | $\min(\overrightarrow{C}, \overleftarrow{C})$ | $\min(\overrightarrow{C}, \overleftarrow{C})$ |
| $r$ | $\min(\overrightarrow{C}, \overleftarrow{C})$ | $\overrightarrow{D} + \overleftarrow{D} + \mathtt{VAR}_i$ | $\overleftarrow{C} + \overrightarrow{D} + \overleftarrow{D} + \mathtt{VAR}_i$ | $\min(\overrightarrow{C}, \overleftarrow{C})$ |
| $t$ | $\min(\overrightarrow{C}, \overleftarrow{C})$ | $\overrightarrow{C} + \overrightarrow{D} + \overleftarrow{D} + \mathtt{VAR}_i$ | $\min(\overrightarrow{C}, \overleftarrow{C})$ | $\overrightarrow{C} + \overrightarrow{D} + \overleftarrow{D} + \mathtt{VAR}_i$ |
| $u$ | $\min(\overrightarrow{C}, \overleftarrow{C})$ | $\min(\overrightarrow{C}, \overleftarrow{C})$ | $\overleftarrow{C} + \overrightarrow{D} + \overleftarrow{D} + \mathtt{VAR}_i$ | $\min(\overrightarrow{C}, \overleftarrow{C})$ |

Table 3.104: Glue matrix for the MIN_SURF_SUMMIT constraint defined as the composition of the SUMMIT pattern , the feature SURF , and the aggregator min ; cells of the glue matrix are coloured with the colour of the constituent to which they are related.

# 3.367   MIN_SURF_VALLEY

     DESCRIPTION          AUTOMATON

| | |
|---|---|
| **Origin** | Based on the VALLEY pattern. |
| **Constraint** | MIN_SURF_VALLEY(VALUE, VARIABLES) |
| **Arguments** | VALUE      : dvar<br>VARIABLES  : collection(var−dvar) |

**Restrictions**

$\text{sv} \leq 2 \vee \text{rv} \leq 1 \Rightarrow \text{VALUE} = +\infty$
$\text{VALUE} \geq \min(\text{minv}①, (\text{sv} - 2) * \text{minv}②)$
$\text{VALUE} = +\infty \vee \text{VALUE} \leq \max(\text{maxv} - 1, (\text{sv} - 2) * (\text{maxv} - 1))$
required(VARIABLES, var)
where
$\text{minv} = \text{minval}(\text{VARIABLES.var})$
$\text{sv} = |\text{VARIABLES}|$
$\text{maxv} = \text{maxval}(\text{VARIABLES.var})$
$\text{rv} = \text{range}(\text{VARIABLES.var})$

**Purpose**

VALUE is the minimal surface of occurrences of the VALLEY pattern in the time-series given by the VARIABLES collection. If the pattern does not occur, VALUE takes the default value $+\infty$.

An occurrence of the pattern VALLEY is the *maximal* subsequence which matches the regular expression $> (= | >)^* (< | =)^* <$.

Assume that the occurrence of the pattern VALLEY starts at position $i$ and ends at position $j$. The feature SURF computes the sum of the values from index $i + 1$ to index $j$.

**Example**

$(7, \langle 1, 3, 7, 4, 3, 6, 6, 5, 3, 3, 2, 6, 5, 5, 5, 7 \rangle)$

Figure 3.857 provides an example where the MIN_SURF_VALLEY $(7, [1, 3, 7, 4, 3, 6, 6, 5, 3, 3, 2, 6, 5, 5, 5, 7])$ constraint holds.

| | |
|---|---|
| **Typical** | $|\text{VARIABLES}| > 2$<br>range(VARIABLES.var) $> 1$ |
| **Symmetry** | Items of VARIABLES can be reversed. |
| **Arg. properties** | Functional dependency: VALUE determined by VARIABLES. |

Figure 3.857: Illustrating the MIN_SURF_VALLEY constraint of the **Example** slot

**Automaton**     Figure 3.858 depicts the automaton associated with the constraint MIN_SURF_VALLEY.

Figure 3.858: Automaton for the MIN_SURF_VALLEY constraint obtained by applying decoration Table 2.35 to the seed transducer of the VALLEY pattern where default is $+\infty$; $-R_i + R_{i-1} \geq 0$ is a linear invariant.

|   | $s$ | $r$ | $t$ |
|---|---|---|---|
| $s$ | $\min(\overrightarrow{C}, \overleftarrow{C})$ | $\min(\overrightarrow{C}, \overleftarrow{C})$ | $\min(\overrightarrow{C}, \overleftarrow{C})$ |
| $r$ | $\min(\overrightarrow{C}, \overleftarrow{C})$ | $\overrightarrow{D} + \overleftarrow{D} + \mathtt{VAR}_i$ | $\overleftarrow{C} + \overrightarrow{D} + \overleftarrow{D} + \mathtt{VAR}_i$ |
| $t$ | $\min(\overrightarrow{C}, \overleftarrow{C})$ | $\overrightarrow{C} + \overrightarrow{D} + \overleftarrow{D} + \mathtt{VAR}_i$ | $\min(\overrightarrow{C}, \overleftarrow{C})$ |

Table 3.105: Glue matrix for the MIN_SURF_VALLEY constraint defined as the composition of the VALLEY pattern , the feature SURF , and the aggregator min ; cells of the glue matrix are coloured with the colour of the constituent to which they are related.

## 3.368 MIN_SURF_ZIGZAG

        DESCRIPTION        AUTOMATON

**Origin**          Based on the ZIGZAG pattern.

**Constraint**      MIN_SURF_ZIGZAG(VALUE, VARIABLES)

**Arguments**       VALUE      : dvar
                    VARIABLES  : collection(var−dvar)

**Restrictions**    $sv \leq 3 \vee rv \leq 1 \Rightarrow VALUE = +\infty$

$VALUE \geq \min \left( \begin{array}{l} 2 * \mathtt{minv} + 1 \textcircled{1}, \\ \lfloor (sv-1)/2 \rfloor * \mathtt{minv} + \lfloor (sv-2)/2 \rfloor * (\mathtt{minv}+1) \textcircled{2} \end{array} \right)$

$\vee \left( \begin{array}{l} VALUE = +\infty, \\ VALUE \leq \max \left( \begin{array}{l} 2 * \mathtt{maxv} - 1, \\ \lfloor (sv-1)/2 \rfloor * \mathtt{maxv} + \lfloor (sv-2)/2 \rfloor * (\mathtt{maxv}-1) \end{array} \right) \end{array} \right)$

required(VARIABLES, var)

where

minv =minval(VARIABLES.var)
sv =|VARIABLES|
maxv =maxval(VARIABLES.var)
rv =range(VARIABLES.var)

**Purpose**         VALUE is the minimal surface of occurrences of the ZIGZAG pattern in the time-series
                    given by the VARIABLES collection. If the pattern does not occur, VALUE takes the default
                    value $+\infty$.

                    An occurrence of the pattern ZIGZAG is the *maximal* subsequence which matches the
                    regular expression $(<>)^+(< | <>) | (><)^+(> | ><)$.

                    Assume that the occurrence of the pattern ZIGZAG starts at position $i$ and ends at position
                    $j$. The feature SURF computes the sum of the values from index $i+1$ to index $j$.

**Example**         $(5, \langle 4, 1, 3, 1, 4, 6, 1, 5, 5, 2, 7, 2, 3, 1, 6, 1 \rangle)$

                    Figure 3.859 provides an example where the MIN_SURF_ZIGZAG
                    $(5, [4, 1, 3, 1, 4, 6, 1, 5, 5, 2, 7, 2, 3, 1, 6, 1])$ constraint holds.

**Typical**         $|\mathtt{VARIABLES}| > 3$
                    $range(\mathtt{VARIABLES.var}) > 1$

**Symmetry**        Items of VARIABLES can be reversed.

**Arg. properties** Functional dependency: VALUE determined by VARIABLES.

Figure 3.859: Illustrating the MIN_SURF_ZIGZAG constraint of the **Example** slot

**Automaton**　　　Figures 3.860 and 3.861 respectively depict the automaton associated with the constraint MIN_SURF_ZIGZAG and its simplified form.

Figure 3.860: Automaton for the MIN_SURF_ZIGZAG constraint obtained by applying decoration Table 2.35 to the seed transducer of the ZIGZAG pattern where default is $+\infty$; (1) missing transitions from $a, b, c, d, e, f$ to $s$ are labelled by $=$; (2) on transitions from $b, c, e, f$ to $s$ the accumulator $D$ is reset to its initial value; (3) on transitions from $c, f$ to $s$ the accumulator $R$ is updated wrt $C$ and the accumulator $C$ is reset to its initial value

Figure 3.861: Automaton for the MIN_SURF_ZIGZAG constraint obtained by applying decoration Table 2.23 to the seed transducer of the ZIGZAG pattern where default is $+\infty$; missing transitions from $a, b, c, d, e, f$ to $s$ are labelled by $=$; (2) on transitions from $b, c, e, f$ to $s$ the accumulator $D$ is reset to its initial value; (3) on transitions from $c, f$ to $s$ the accumulator $R$ is updated wrt $C$ and the accumulator $C$ is reset to its initial value.; $-R_i + R_{i-1} \geq 0$ is a linear invariant.

|  | $s$ | $a$ | $b$ | $c$ | $d$ | $e$ | $f$ |
|---|---|---|---|---|---|---|---|
| **$s$** | $\min(\overrightarrow{C},\overleftarrow{C})$ | $\min(\overrightarrow{C},\overleftarrow{C})$ | $\min(\overrightarrow{C},\overleftarrow{C})$ | $\min(\overrightarrow{C},\overleftarrow{C})$ | $\min(\overrightarrow{C},\overleftarrow{C})$ | $\min(\overrightarrow{C},\overleftarrow{C})$ | $\min(\overrightarrow{C},\overleftarrow{C})$ |
| **$a$** | $\min(\overrightarrow{C},\overleftarrow{C})$ | $\min(\overrightarrow{C},\overleftarrow{C})$ | $\min(\overrightarrow{C},\overleftarrow{C})$ | $\overrightarrow{C}+\overrightarrow{D}+\overleftarrow{D}+\text{VAR}_i$ | $\min(\overrightarrow{C},\overleftarrow{C})$ | $\overrightarrow{D}+\overleftarrow{D}+\text{VAR}_i$ | $\min(\overrightarrow{C},\overleftarrow{C})$ |
| **$b$** | $\min(\overrightarrow{C},\overleftarrow{C})$ | $\overrightarrow{C}+\overrightarrow{D}+\overleftarrow{D}+\text{VAR}_i$ | $\overrightarrow{D}+\overleftarrow{D}+\text{VAR}_i$ | $\min(\overrightarrow{C},\overleftarrow{C})$ | $\overrightarrow{D}+\overleftarrow{D}+\text{VAR}_i$ | $\min(\overrightarrow{C},\overleftarrow{C})$ | $\overrightarrow{C}+\overrightarrow{D}+\overleftarrow{D}+\text{VAR}_i$ |
| **$c$** | $\min(\overrightarrow{C},\overleftarrow{C})$ | $\min(\overrightarrow{C},\overleftarrow{C})$ | $\min(\overrightarrow{C},\overleftarrow{C})$ | $\overrightarrow{C}+\overrightarrow{C}+\overrightarrow{D}+\overleftarrow{D}+\text{VAR}_i$ | $\min(\overrightarrow{C},\overleftarrow{C})$ | $\overrightarrow{C}+\overrightarrow{D}+\overleftarrow{D}+\text{VAR}_i$ | $\min(\overrightarrow{C},\overleftarrow{C})$ |
| **$d$** | $\min(\overrightarrow{C},\overleftarrow{C})$ | $\overrightarrow{D}+\overleftarrow{D}+\text{VAR}_i$ | $\overrightarrow{D}+\overleftarrow{D}+\text{VAR}_i$ | $\min(\overrightarrow{C},\overleftarrow{C})$ | $\min(\overrightarrow{C},\overleftarrow{C})$ | $\min(\overrightarrow{C},\overleftarrow{C})$ | $\overrightarrow{C}+\overrightarrow{D}+\overleftarrow{D}+\text{VAR}_i$ |
| **$e$** | $\min(\overrightarrow{C},\overleftarrow{C})$ | $\min(\overrightarrow{C},\overleftarrow{C})$ | $\min(\overrightarrow{C},\overleftarrow{C})$ | $\overrightarrow{C}+\overrightarrow{D}+\overleftarrow{D}+\text{VAR}_i$ | $\min(\overrightarrow{C},\overleftarrow{C})$ | $\overrightarrow{D}+\overleftarrow{D}+\text{VAR}_i$ | $\min(\overrightarrow{C},\overleftarrow{C})$ |
| **$f$** | $\min(\overrightarrow{C},\overleftarrow{C})$ | $\min(\overrightarrow{C},\overleftarrow{C})$ | $\overrightarrow{C}+\overrightarrow{D}+\overleftarrow{D}+\text{VAR}_i$ | $\min(\overrightarrow{C},\overleftarrow{C})$ | $\overrightarrow{C}+\overrightarrow{D}+\overleftarrow{D}+\text{VAR}_i$ | $\min(\overrightarrow{C},\overleftarrow{C})$ | $\overrightarrow{C}+\overleftarrow{C}+\overrightarrow{D}+\overleftarrow{D}+\text{VAR}_i$ |

Table 3.106: Glue matrix for the MIN_SURF_ZIGZAG constraint defined as the composition of the ZIGZAG pattern , the feature SURF , and the aggregator min ; cells of the glue matrix are coloured with the colour of the constituent to which they are related.

# 3.369 MIN_WIDTH_DECREASING_SEQUENCE

▶ ▷ ◁ ◀        **DESCRIPTION**        **AUTOMATON**

> (> | =)* > |
> 

| **Origin** | Based on the DECREASING_SEQUENCE pattern. |

| **Constraint** | MIN_WIDTH_DECREASING_SEQUENCE(VALUE, VARIABLES) |

| **Arguments** | VALUE       : dvar |
|                | VARIABLES : collection(var−dvar) |

**Restrictions**

$$\mathtt{sv} \leq 1 \vee \mathtt{rv} \leq 1 \Rightarrow \mathtt{VALUE} = \mathtt{sv} + 1$$
$$\mathtt{VALUE} \geq 2$$
$$\mathtt{rv} = 2 \Rightarrow \mathtt{VALUE} = \mathtt{sv} + 1 \vee \mathtt{VALUE} \leq 2 ①$$
$$\mathtt{rv} \geq 3 \Rightarrow \mathtt{VALUE} = \mathtt{sv} + 1 \vee \mathtt{VALUE} \leq \mathtt{sv} ②$$
required(VARIABLES, var)
where
$$\mathtt{sv} = |\mathtt{VARIABLES}|$$
$$\mathtt{rv} = \mathtt{range}(\mathtt{VARIABLES.var})$$

VALUE ≤ 2          VALUE ≤ 6
①                      ②

**Purpose**

VALUE is the minimal width of occurrences of the DECREASING_SEQUENCE pattern in the time-series given by the VARIABLES collection. If the pattern does not occur, VALUE takes the default value |VARIABLES| + 1.

An occurrence of the pattern DECREASING_SEQUENCE is the *maximal* subsequence which matches the regular expression > (> | =)* > | >.

Assume that the occurrence of the pattern DECREASING_SEQUENCE starts at position $i$ and ends at position $j$. The feature WIDTH computes the value $j - i + 2$.

**Example**

$$(2, \langle 3, 4, 2, 2, 5, 6, 6, 4, 4, 3, 1, 1, 4, 6, 4, 4 \rangle)$$

Figure 3.862 provides an example where the MIN_WIDTH_DECREASING_SEQUENCE $(2, [3, 4, 2, 2, 5, 6, 6, 4, 4, 3, 1, 1, 4, 6, 4, 4])$ constraint holds.

**Typical**

$$|\mathtt{VARIABLES}| > 1$$
$$\mathtt{range}(\mathtt{VARIABLES.var}) > 1$$

**Symmetry**       One and the same constant can be added to the var attribute of all items of VARIABLES.

**Arg. properties**       Functional dependency: VALUE determined by VARIABLES.

Figure 3.862: Illustrating the MIN_WIDTH_DECREASING_SEQUENCE constraint of the **Example** slot

**Automaton**    Figure 3.863 depicts the automaton associated with the constraint MIN_WIDTH_DECREASING_SEQUENCE.

Figure 3.863: Automaton for the MIN_WIDTH_DECREASING_SEQUENCE constraint obtained by applying decoration Table 2.35 to the seed transducer of the DECREASING_SEQUENCE pattern where default is $|\text{VARIABLES}| + 1$; $-R_i + R_{i-1} \geq 0$ is a linear invariant.

|       | $s$                                      | $t$                                                                  |
| ----- | ---------------------------------------- | -------------------------------------------------------------------- |
| $s$   | $\min(\overrightarrow{C}, \overleftarrow{C})$ | $\min(\overrightarrow{C}, \overleftarrow{C})$                         |
| $t$   | $\min(\overrightarrow{C}, \overleftarrow{C})$ | $\overrightarrow{C} + \overleftarrow{C} + \overrightarrow{D} + \overleftarrow{D} + 1$ |

Table 3.107: Glue matrix for the MIN_WIDTH_DECREASING_SEQUENCE constraint defined as the composition of the DECREASING_SEQUENCE pattern , the feature WIDTH , and the aggregator min ; cells of the glue matrix are coloured with the colour of the constituent to which they are related.

## 3.370 MIN_WIDTH_DECREASING_TERRACE

▶ ▷ ◁ ◀    DESCRIPTION      AUTOMATON

| | |
|---|---|
| **Origin** | Based on the DECREASING_TERRACE pattern. |
| **Constraint** | MIN_WIDTH_DECREASING_TERRACE(VALUE, VARIABLES) |
| **Arguments** | VALUE : dvar<br>VARIABLES : collection(var−dvar) |

**Restrictions**

$sv \leq 3 \lor rv \leq 2 \Rightarrow \text{VALUE} = sv + 1$

$\text{VALUE} \geq 2$

$\text{VALUE} = sv + 1 \lor \text{VALUE} \leq sv - 2$ ①

required(VARIABLES, var)

where

$sv = |\text{VARIABLES}|$

$rv = \text{range}(\text{VARIABLES.var})$

**Purpose**

VALUE is the minimal width of occurrences of the DECREASING_TERRACE pattern in the time-series given by the VARIABLES collection. If the pattern does not occur, VALUE takes the default value $|\text{VARIABLES}| + 1$.

An occurrence of the pattern DECREASING_TERRACE is the *maximal* subsequence which matches the regular expression $>=^+ >$.

Assume that the occurrence of the pattern DECREASING_TERRACE starts at position $i$ and ends at position $j$. The feature WIDTH computes the value $j - i$.

**Example**

$(2, \langle 6, 4, 4, 4, 5, 2, 2, 1, 3, 3, 5, 4, 4, 3, 3, 3 \rangle)$

Figure 3.864 provides an example where the MIN_WIDTH_DECREASING_TERRACE $(2, [6, 4, 4, 4, 5, 2, 2, 1, 3, 3, 5, 4, 4, 3, 3, 3])$ constraint holds.

**Typical**

$|\text{VARIABLES}| > 3$

$\text{range}(\text{VARIABLES.var}) > 2$

**Symmetry**

One and the same constant can be added to the var attribute of all items of VARIABLES.

**Arg. properties**

Functional dependency: VALUE determined by VARIABLES.

Figure 3.864: Illustrating the MIN_WIDTH_DECREASING_TERRACE constraint of the **Example** slot

**Automaton**  Figures 3.865 and 3.866 respectively depict the automaton associated with the constraint MIN_WIDTH_DECREASING_TERRACE and its simplified form.

Figure 3.865: Automaton for the MIN_WIDTH_DECREASING_TERRACE constraint obtained by applying decoration Table 2.35 to the seed transducer of the DECREASING_TERRACE pattern where default is $|\texttt{VARIABLES}| + 1$

|   | $s$ | $r$ | $t$ |
|---|---|---|---|
| $s$ | $\min(\overrightarrow{C}, \overleftarrow{C})$ | $\min(\overrightarrow{C}, \overleftarrow{C})$ | $\min(\overrightarrow{C}, \overleftarrow{C})$ |
| $r$ | $\min(\overrightarrow{C}, \overleftarrow{C})$ | $\min(\overrightarrow{C}, \overleftarrow{C})$ | $\overrightarrow{D} + \overleftarrow{D} + 1$ |
| $t$ | $\min(\overrightarrow{C}, \overleftarrow{C})$ | $\overrightarrow{D} + \overleftarrow{D} + 1$ | $\overrightarrow{D} + \overleftarrow{D} + 1$ |

Table 3.108: Glue matrix for the MIN_WIDTH_DECREASING_TERRACE constraint defined as the composition of the DECREASING_TERRACE pattern , the feature WIDTH , and the aggregator min ; cells of the glue matrix are coloured with the colour of the constituent to which they are related.

Figure 3.866: Automaton for the MIN_WIDTH_DECREASING_TERRACE constraint obtained by applying decoration Table 2.25 to the seed transducer of the DECREASING_TERRACE pattern where default is $|\text{VARIABLES}| + 1$; $-R_i + R_{i-1} \geq 0$ is a linear invariant.

# 3.371   MIN_WIDTH_GORGE

$(> | > (= | >)^* >)(< | < (= | <)^* <)$

| | |
|---|---|
| **Origin** | Based on the GORGE pattern. |
| **Constraint** | MIN_WIDTH_GORGE(VALUE, VARIABLES) |
| **Arguments** | VALUE      : dvar<br>VARIABLES  : collection(var−dvar) |

**Restrictions**

$sv \leq 2 \vee rv \leq 1 \Rightarrow \text{VALUE} = sv + 1$

$\text{VALUE} \geq 1$

$rv = 2 \Rightarrow \text{VALUE} = sv + 1 \vee \text{VALUE} \leq 1$①

$rv \geq 3 \Rightarrow \text{VALUE} = sv + 1 \vee \text{VALUE} \leq sv - 2$②

required(VARIABLES, var)

where

$sv = |\text{VARIABLES}|$

$rv = \text{range}(\text{VARIABLES.var})$

VALUE ≤ 1

$\text{VALUE} \leq 5 = 7 - 2$

**Purpose**

VALUE is the minimal width of occurrences of the GORGE pattern in the time-series given by the VARIABLES collection. If the pattern does not occur, VALUE takes the default value $|\text{VARIABLES}| + 1$.

An occurrence of the pattern GORGE is the *maximal* subsequence which matches the regular expression $(> | > (= | >)^* >)(< | < (= | <)^* <)$.

Assume that the occurrence of the pattern GORGE starts at position $i$ and ends at position $j$. The feature WIDTH computes the value $j - i$.

**Example**

$(1, \langle 1, 7, 3, 4, 4, 5, 5, 4, 2, 2, 6, 5, 4, 6, 5, 7 \rangle)$

Figure 3.867 provides an example where the MIN_WIDTH_GORGE $(1, [1, 7, 3, 4, 4, 5, 5, 4, 2, 2, 6, 5, 4, 6, 5, 7])$ constraint holds.

**Typical**

$|\text{VARIABLES}| > 2$

range(VARIABLES.var) > 1

**Symmetries**

- Items of VARIABLES can be reversed.
- One and the same constant can be added to the var attribute of all items of VARIABLES.

**Arg. properties**

Functional dependency: VALUE determined by VARIABLES.

Figure 3.867: Illustrating the MIN_WIDTH_GORGE constraint of the **Example** slot

**Automaton**          Figure 3.868 depicts the automaton associated with the constraint MIN_WIDTH_GORGE.

Figure 3.868: Automaton for the MIN_WIDTH_GORGE constraint obtained by applying decoration Table 2.35 to the seed transducer of the GORGE pattern where default is |VARIABLES| + 1 (transition $u \rightarrow r$ has the same accumulator update as transition $r \rightarrow u$); $-R_i + R_{i-1} \geq 0$ is a linear invariant.

|   | $s$ | $r$ | $t$ | $u$ |
|---|---|---|---|---|
| $s$ | $\min(\overrightarrow{C}, \overleftarrow{C})$ | $\min(\overrightarrow{C}, \overleftarrow{C})$ | $\min(\overrightarrow{C}, \overleftarrow{C})$ | $\min(\overrightarrow{C}, \overleftarrow{C})$ |
| $r$ | $\min(\overrightarrow{C}, \overleftarrow{C})$ | $\overrightarrow{D} + \overleftarrow{D} + 1$ | $\overleftarrow{C} + \overrightarrow{D} + \overleftarrow{D} + 1$ | $\min(\overrightarrow{C}, \overleftarrow{C})$ |
| $t$ | $\min(\overrightarrow{C}, \overleftarrow{C})$ | $\overrightarrow{C} + \overrightarrow{D} + \overleftarrow{D} + 1$ | $\min(\overrightarrow{C}, \overleftarrow{C})$ | $\overrightarrow{C} + \overrightarrow{D} + \overleftarrow{D} + 1$ |
| $u$ | $\min(\overrightarrow{C}, \overleftarrow{C})$ | $\min(\overrightarrow{C}, \overleftarrow{C})$ | $\overleftarrow{C} + \overrightarrow{D} + \overleftarrow{D} + 1$ | $\min(\overrightarrow{C}, \overleftarrow{C})$ |

Table 3.109: Glue matrix for the MIN_WIDTH_GORGE constraint defined as the composition of the GORGE pattern , the feature WIDTH , and the aggregator min ; cells of the glue matrix are coloured with the colour of the constituent to which they are related.

# 3.372　MIN_WIDTH_INCREASING_SEQUENCE

▶ ▷ ◁ ◀　　**DESCRIPTION**　　**AUTOMATON**

| **Origin** | Based on the INCREASING_SEQUENCE pattern. |
|---|---|

| **Constraint** | MIN_WIDTH_INCREASING_SEQUENCE(VALUE, VARIABLES) |
|---|---|

| **Arguments** | VALUE　　　: dvar |
|---|---|
| | VARIABLES : collection(var−dvar) |

**Restrictions**

$\mathtt{sv} \leq 1 \vee \mathtt{rv} \leq 1 \Rightarrow \mathtt{VALUE} = \mathtt{sv} + 1$

$\mathtt{VALUE} \geq 2$

$\mathtt{rv} = 2 \Rightarrow \mathtt{VALUE} = \mathtt{sv} + 1 \vee \mathtt{VALUE} \leq 2$ ①

$\mathtt{rv} \geq 3 \Rightarrow \mathtt{VALUE} = \mathtt{sv} + 1 \vee \mathtt{VALUE} \leq \mathtt{sv}$ ②

required(VARIABLES, var)

where

$\quad \mathtt{sv} = |\mathtt{VARIABLES}|$

$\quad \mathtt{rv} = $ range(VARIABLES.var)

**Purpose**

VALUE is the minimal width of occurrences of the INCREASING_SEQUENCE pattern in the time-series given by the VARIABLES collection. If the pattern does not occur, VALUE takes the default value $|\mathtt{VARIABLES}| + 1$.

An occurrence of the pattern INCREASING_SEQUENCE is the *maximal* subsequence which matches the regular expression $< (< | =)^* < | <$.

Assume that the occurrence of the pattern INCREASING_SEQUENCE starts at position $i$ and ends at position $j$. The feature WIDTH computes the value $j - i + 2$.

**Example**

$(2, \langle 4, 3, 5, 5, 2, 1, 1, 3, 3, 4, 6, 6, 3, 1, 3, 3 \rangle)$

Figure 3.869 provides an example where the MIN_WIDTH_INCREASING_SEQUENCE $(2, [4, 3, 5, 5, 2, 1, 1, 3, 3, 4, 6, 6, 3, 1, 3, 3])$ constraint holds.

| **Typical** | $|\mathtt{VARIABLES}| > 1$ |
|---|---|
| | range(VARIABLES.var) > 1 |

| **Symmetry** | One and the same constant can be added to the var attribute of all items of VARIABLES. |
|---|---|

| **Arg. properties** | Functional dependency: VALUE determined by VARIABLES. |
|---|---|

Figure 3.869: Illustrating the MIN_WIDTH_INCREASING_SEQUENCE constraint of the **Example** slot

**Automaton**     Figure 3.870 depicts the automaton associated with the constraint MIN_WIDTH_INCREASING_SEQUENCE.

Figure 3.870: Automaton for the MIN_WIDTH_INCREASING_SEQUENCE constraint obtained by applying decoration Table 2.35 to the seed transducer of the INCREASING_SEQUENCE pattern where default is $|\texttt{VARIABLES}| + 1$; $-R_i + R_{i-1} \geq 0$ is a linear invariant.

|   | $s$ | $t$ |
|---|---|---|
| $s$ | $\min(\overrightarrow{C}, \overleftarrow{C})$ | $\min(\overrightarrow{C}, \overleftarrow{C})$ |
| $t$ | $\min(\overrightarrow{C}, \overleftarrow{C})$ | $\overrightarrow{C} + \overleftarrow{C} + \overrightarrow{D} + \overleftarrow{D} + 1$ |

Table 3.110: Glue matrix for the MIN_WIDTH_INCREASING_SEQUENCE constraint defined as the composition of the INCREASING_SEQUENCE pattern , the feature WIDTH , and the aggregator min ; cells of the glue matrix are coloured with the colour of the constituent to which they are related.

# 3.373   MIN_WIDTH_INCREASING_TERRACE

▶ ▷ ◁ ◀          DESCRIPTION          AUTOMATON

**Origin**         Based on the INCREASING_TERRACE pattern.

**Constraint**     MIN_WIDTH_INCREASING_TERRACE(VALUE, VARIABLES)

**Arguments**      VALUE      :  dvar
                   VARIABLES  :  collection(var−dvar)

**Restrictions**   $\text{sv} \leq 3 \lor \text{rv} \leq 2 \Rightarrow \text{VALUE} = \text{sv} + 1$
                   $\text{VALUE} \geq 2$
                   $\text{VALUE} = \text{sv} + 1 \lor \text{VALUE} \leq \text{sv} - 2$①
                   required(VARIABLES, var)
                   where
                   $\text{sv} = |\text{VARIABLES}|$
                   $\text{rv} = \text{range}(\text{VARIABLES.var})$

**Purpose**        VALUE is the minimal width of occurrences of the INCREASING_TERRACE pattern in the time-series given by the VARIABLES collection. If the pattern does not occur, VALUE takes the default value $|\text{VARIABLES}| + 1$.
                   An occurrence of the pattern INCREASING_TERRACE is the *maximal* subsequence which matches the regular expression $< =^+ <$.
                   Assume that the occurrence of the pattern INCREASING_TERRACE starts at position $i$ and ends at position $j$. The feature WIDTH computes the value $j - i$.

**Example**        $(2, \langle 1, 3, 3, 3, 2, 5, 5, 6, 4, 4, 2, 3, 3, 3, 4, 4 \rangle)$

                   Figure 3.871 provides an example where the MIN_WIDTH_INCREASING_TERRACE $(2, [1, 3, 3, 3, 2, 5, 5, 6, 4, 4, 2, 3, 3, 3, 4, 4])$ constraint holds.

**Typical**        $|\text{VARIABLES}| > 3$
                   $\text{range}(\text{VARIABLES.var}) > 2$

**Symmetry**       One and the same constant can be added to the var attribute of all items of VARIABLES.

**Arg. properties** Functional dependency: VALUE determined by VARIABLES.

Figure 3.871: Illustrating the MIN_WIDTH_INCREASING_TERRACE constraint of the **Example** slot

 Figures 3.872 and 3.873 respectively depict the automaton associated with the constraint MIN_WIDTH_INCREASING_TERRACE and its simplified form.

Figure 3.872: Automaton for the MIN_WIDTH_INCREASING_TERRACE constraint obtained by applying decoration Table 2.35 to the seed transducer of the INCREASING_TERRACE pattern where default is $|\mathtt{VARIABLES}| + 1$

|   | $s$ | $r$ | $t$ |
|---|---|---|---|
| $s$ | $\min(\overrightarrow{C}, \overleftarrow{C})$ | $\min(\overrightarrow{C}, \overleftarrow{C})$ | $\min(\overrightarrow{C}, \overleftarrow{C})$ |
| $r$ | $\min(\overrightarrow{C}, \overleftarrow{C})$ | $\min(\overrightarrow{C}, \overleftarrow{C})$ | $\overrightarrow{D} + \overleftarrow{D} + 1$ |
| $t$ | $\min(\overrightarrow{C}, \overleftarrow{C})$ | $\overrightarrow{D} + \overleftarrow{D} + 1$ | $\overrightarrow{D} + \overleftarrow{D} + 1$ |

Table 3.111: Glue matrix for the MIN_WIDTH_INCREASING_TERRACE constraint defined as the composition of the INCREASING_TERRACE pattern , the feature WIDTH , and the aggregator min ; cells of the glue matrix are coloured with the colour of the constituent to which they are related.

Figure 3.873: Automaton for the MIN_WIDTH_INCREASING_TERRACE constraint obtained by applying decoration Table 2.25 to the seed transducer of the INCREASING_TERRACE pattern where default is $|\text{VARIABLES}| + 1$; $-R_i + R_{i-1} \geq 0$ is a linear invariant.

PL

## 3.374  MIN_WIDTH_INFLEXION

▶ ▷ ◁ ◀         DESCRIPTION            AUTOMATON

$< (< | =)^* > | > (> | =)^* <$

**Origin**           Based on the INFLEXION pattern.

**Constraint**       MIN_WIDTH_INFLEXION(VALUE, VARIABLES)

**Arguments**        VALUE      : dvar
                     VARIABLES  : collection(var−dvar)

**Restrictions**     $\mathtt{sv} \leq 2 \vee \mathtt{rv} \leq 1 \Rightarrow \mathtt{VALUE} = \mathtt{sv} + 1$
                     $\mathtt{VALUE} \geq 1$
                     $\mathtt{VALUE} = \mathtt{sv} + 1 \vee \mathtt{VALUE} \leq \mathtt{sv} - 2$ ①
                     required(VARIABLES, var)
                     where
                     $\mathtt{sv} = |\mathtt{VARIABLES}|$
                     $\mathtt{rv} = \mathtt{range}(\mathtt{VARIABLES.var})$

VALUE ≤ **4** = 6 − 2

VALUE ≤ **4** = 6 − 2

**Purpose**          VALUE is the minimal width of occurrences of the INFLEXION pattern in the time-series given by the VARIABLES collection. If the pattern does not occur, VALUE takes the default value $|\mathtt{VARIABLES}| + 1$.
                     An occurrence of the pattern INFLEXION is the *maximal* subsequence which matches the regular expression $< (< | =)^* > | > (> | =)^* <$.
                     Assume that the occurrence of the pattern INFLEXION starts at position $i$ and ends at position $j$. The feature WIDTH computes the value $j − i$.

**Example**          $(1, \langle 1, 2, 6, 6, 4, 4, 3, 5, 2, 5, 1, 5, 3, 3, 4, 4 \rangle)$

Figure 3.874 provides an example where the MIN_WIDTH_INFLEXION $(1, [1, 2, 6, 6, 4, 4, 3, 5, 2, 5, 1, 5, 3, 3, 4, 4])$ constraint holds.

**Typical**          $|\mathtt{VARIABLES}| > 2$
                     $\mathtt{range}(\mathtt{VARIABLES.var}) > 1$

**Symmetries**       • Items of VARIABLES can be reversed.
                     • One and the same constant can be added to the var attribute of all items of VARIABLES.

**Arg. properties**  Functional dependency: VALUE determined by VARIABLES.

Figure 3.874: Illustrating the MIN_WIDTH_INFLEXION constraint of the **Example** slot

**Automaton**  Figures 3.875 and 3.876 respectively depict the automaton associated with the constraint MIN_WIDTH_INFLEXION and its simplified form.

Figure 3.875: Automaton for the MIN_WIDTH_INFLEXION constraint obtained by applying decoration Table 2.35 to the seed transducer of the INFLEXION pattern where `default` is |VARIABLES| + 1 (transition $r \rightarrow t$ has the same accumulators updates as transition $t \rightarrow r$)

Figure 3.876: Automaton for the MIN_WIDTH_INFLEXION constraint obtained by applying decoration Table 2.25 to the seed transducer of the INFLEXION pattern where `default` is |VARIABLES| + 1 (transition $r \rightarrow t$ has the same accumulators updates as transition $t \rightarrow r$); $-R_i + R_{i-1} \geq 0$ is a linear invariant.

# 3.375   MIN_WIDTH_PEAK

▶ ▷ ◁ ◀        **DESCRIPTION**        **AUTOMATON**

$< (= | <)^* (> | =)^* >$

| **Origin** | Based on the PEAK pattern. |
|---|---|

| **Constraint** | MIN_WIDTH_PEAK(VALUE, VARIABLES) |
|---|---|

**Arguments**

```
VALUE      : dvar
VARIABLES  : collection(var−dvar)
```

**Restrictions**

$sv \leq 2 \vee rv \leq 1 \Rightarrow VALUE = sv + 1$
$VALUE \geq 1$
$VALUE = sv + 1 \vee VALUE \leq sv - 2 ①$
required(VARIABLES, var)
where
$sv = |VARIABLES|$
$rv = range(VARIABLES.var)$

VALUE $\leq 4 = 6 - 2$

**Purpose**

VALUE is the minimal width of occurrences of the PEAK pattern in the time-series given by the VARIABLES collection. If the pattern does not occur, VALUE takes the default value $|VARIABLES| + 1$.
An occurrence of the pattern PEAK is the *maximal* subsequence which matches the regular expression $< (= | <)^* (> | =)^* >$.
Assume that the occurrence of the pattern PEAK starts at position $i$ and ends at position $j$. The feature WIDTH computes the value $j - i$.

**Example**

$(2, \langle 7, 5, 5, 1, 4, 5, 2, 2, 3, 5, 6, 2, 3, 3, 3, 1 \rangle)$

Figure 3.877 provides an example where the MIN_WIDTH_PEAK $(2, [7, 5, 5, 1, 4, 5, 2, 2, 3, 5, 6, 2, 3, 3, 3, 1])$ constraint holds.

**Typical**

$|VARIABLES| > 2$
$range(VARIABLES.var) > 1$

**Symmetries**

- Items of VARIABLES can be reversed.
- One and the same constant can be added to the var attribute of all items of VARIABLES.

**Arg. properties**

Functional dependency: VALUE determined by VARIABLES.

Figure 3.877: Illustrating the MIN_WIDTH_PEAK constraint of the **Example** slot

        Figure 3.878 depicts the automaton associated with the constraint MIN_WIDTH_PEAK.

Figure 3.878: Automaton for the MIN_WIDTH_PEAK constraint obtained by applying decoration Table 2.35 to the seed transducer of the PEAK pattern where default is $|\mathtt{VARIABLES}| + 1$; $-R_i + R_{i-1} \geq 0$ is a linear invariant.

|   | $s$ | $r$ | $t$ |
|---|---|---|---|
| $s$ | $\min(\overrightarrow{C}, \overleftarrow{C})$ | $\min(\overrightarrow{C}, \overleftarrow{C})$ | $\min(\overrightarrow{C}, \overleftarrow{C})$ |
| $r$ | $\min(\overrightarrow{C}, \overleftarrow{C})$ | $\overrightarrow{D} + \overleftarrow{D} + 1$ | $\overleftarrow{C} + \overrightarrow{D} + \overleftarrow{D} + 1$ |
| $t$ | $\min(\overrightarrow{C}, \overleftarrow{C})$ | $\overrightarrow{C} + \overrightarrow{D} + \overleftarrow{D} + 1$ | $\min(\overrightarrow{C}, \overleftarrow{C})$ |

Table 3.112: Glue matrix for the MIN_WIDTH_PEAK constraint defined as the composition of the PEAK pattern , the feature WIDTH , and the aggregator min ; cells of the glue matrix are coloured with the colour of the constituent to which they are related.

# 3.376  MIN_WIDTH_PLAIN

▶ ▷ ◁ ◀        **DESCRIPTION**          **AUTOMATON**

| | |
|---|---|
| **Origin** | Based on the PLAIN pattern. |
| **Constraint** | MIN_WIDTH_PLAIN(VALUE, VARIABLES) |
| **Arguments** | VALUE      : dvar<br>VARIABLES  : collection(var−dvar) |
| **Restrictions** | $sv \leq 2 \vee rv \leq 1 \Rightarrow VALUE = sv + 1$<br>$VALUE \geq 1$<br>$VALUE = sv + 1 \vee VALUE \leq sv - 2$①<br>required(VARIABLES, var)<br>where<br>$sv = \lvert VARIABLES \rvert$<br>$rv = range(VARIABLES.var)$ |
| **Purpose** | VALUE is the minimal width of occurrences of the PLAIN pattern in the time-series given by the VARIABLES collection. If the pattern does not occur, VALUE takes the default value $\lvert VARIABLES \rvert + 1$.<br>An occurrence of the pattern PLAIN is the *maximal* subsequence which matches the regular expression $> =^* <$.<br>Assume that the occurrence of the pattern PLAIN starts at position $i$ and ends at position $j$. The feature WIDTH computes the value $j - i$. |
| **Example** | $(1, \langle 2, 3, 6, 5, 7, 6, 6, 4, 5, 5, 4, 3, 3, 6, 6, 3 \rangle)$ |

Figure 3.879 provides an example where the MIN_WIDTH_PLAIN $(1, [2, 3, 6, 5, 7, 6, 6, 4, 5, 5, 4, 3, 3, 6, 6, 3])$ constraint holds.

| | |
|---|---|
| **Typical** | $\lvert VARIABLES \rvert > 2$<br>$range(VARIABLES.var) > 1$ |
| **Symmetries** | • Items of VARIABLES can be reversed.<br>• One and the same constant can be added to the var attribute of all items of VARIABLES. |
| **Arg. properties** | Functional dependency: VALUE determined by VARIABLES. |

Figure 3.879: Illustrating the MIN_WIDTH_PLAIN constraint of the **Example** slot

 Figures 3.880 and 3.881 respectively depict the automaton associated with the constraint MIN_WIDTH_PLAIN and its simplified form.

Figure 3.880: Automaton for the MIN_WIDTH_PLAIN constraint obtained by applying decoration Table 2.35 to the seed transducer of the PLAIN pattern where default is $|\mathtt{VARIABLES}| + 1$

Figure 3.881: Automaton for the MIN_WIDTH_PLAIN constraint obtained by applying decoration Table 2.25 to the seed transducer of the PLAIN pattern where default is $|\mathtt{VARIABLES}| + 1$; $-R_i + R_{i-1} \geq 0$ is a linear invariant.

|   | $s$ | $r$ | $t$ |
|---|---|---|---|
| $s$ | $\min(\overrightarrow{C}, \overleftarrow{C})$ | $\min(\overrightarrow{C}, \overleftarrow{C})$ | $\min(\overrightarrow{C}, \overleftarrow{C})$ |
| $r$ | $\min(\overrightarrow{C}, \overleftarrow{C})$ | $\overrightarrow{D} + \overleftarrow{D} + 1$ | $\overrightarrow{D} + \overleftarrow{D} + 1$ |
| $t$ | $\min(\overrightarrow{C}, \overleftarrow{C})$ | $\overrightarrow{D} + \overleftarrow{D} + 1$ | $\overrightarrow{D} + \overleftarrow{D} + 1$ |

Table 3.113: Glue matrix for the MIN_WIDTH_PLAIN constraint defined as the composition of the PLAIN pattern, the feature WIDTH, and the aggregator min; cells of the glue matrix are coloured with the colour of the constituent to which they are related.

## 3.377 MIN_WIDTH_PLATEAU

| | |
|---|---|
| **Origin** | Based on the PLATEAU pattern. |
| **Constraint** | MIN_WIDTH_PLATEAU(VALUE, VARIABLES) |
| **Arguments** | VALUE      : dvar<br>VARIABLES : collection(var−dvar) |

**Restrictions**

$$\text{sv} \leq 2 \vee \text{rv} \leq 1 \Rightarrow \text{VALUE} = \text{sv} + 1$$
$$\text{VALUE} \geq 1$$
$$\text{VALUE} = \text{sv} + 1 \vee \text{VALUE} \leq \text{sv} - 2 \,①$$
required(VARIABLES, var)
where
$$\text{sv} = |\text{VARIABLES}|$$
$$\text{rv} = \text{range}(\text{VARIABLES.var})$$

**Purpose**

VALUE is the minimal width of occurrences of the PLATEAU pattern in the time-series given by the VARIABLES collection. If the pattern does not occur, VALUE takes the default value $|\text{VARIABLES}| + 1$.

An occurrence of the pattern PLATEAU is the *maximal* subsequence which matches the regular expression $<=^*>$.

Assume that the occurrence of the pattern PLATEAU starts at position $i$ and ends at position $j$. The feature WIDTH computes the value $j - i$.

**Example**

$$(3, \langle 1, 3, 3, 5, 5, 5, 5, 2, 4, 4, 4, 3, 3, 1, 5, 5 \rangle)$$

Figure 3.882 provides an example where the MIN_WIDTH_PLATEAU $(3, [1, 3, 3, 5, 5, 5, 5, 2, 4, 4, 4, 3, 3, 1, 5, 5])$ constraint holds.

| | |
|---|---|
| **Typical** | $\|\text{VARIABLES}\| > 2$<br>range(VARIABLES.var) $> 1$ |
| **Symmetries** | • Items of VARIABLES can be reversed.<br>• One and the same constant can be added to the var attribute of all items of VARIABLES. |
| **Arg. properties** | Functional dependency: VALUE determined by VARIABLES. |

Figure 3.882: Illustrating the MIN_WIDTH_PLATEAU constraint of the **Example** slot

**Automaton**     Figures 3.883 and 3.884 respectively depict the automaton associated with the constraint
MIN_WIDTH_PLATEAU and its simplified form.

Figure 3.883: Automaton for the MIN_WIDTH_PLATEAU constraint obtained by applying decoration Table 2.35 to the seed transducer of the PLATEAU pattern where `default` is $|\texttt{VARIABLES}| + 1$

Figure 3.884: Automaton for the MIN_WIDTH_PLATEAU constraint obtained by applying decoration Table 2.25 to the seed transducer of the PLATEAU pattern where `default` is $|\texttt{VARIABLES}| + 1$; $-R_i + R_{i-1} \geq 0$ is a linear invariant.

|   | $s$ | $r$ | $t$ |
|---|---|---|---|
| $s$ | $\min(\overrightarrow{C}, \overleftarrow{C})$ | $\min(\overrightarrow{C}, \overleftarrow{C})$ | $\min(\overrightarrow{C}, \overleftarrow{C})$ |
| $r$ | $\min(\overrightarrow{C}, \overleftarrow{C})$ | $\overrightarrow{D} + \overleftarrow{D} + 1$ | $\overrightarrow{D} + \overleftarrow{D} + 1$ |
| $t$ | $\min(\overrightarrow{C}, \overleftarrow{C})$ | $\overrightarrow{D} + \overleftarrow{D} + 1$ | $\overrightarrow{D} + \overleftarrow{D} + 1$ |

Table 3.114: Glue matrix for the MIN_WIDTH_PLATEAU constraint defined as the composition of the PLATEAU pattern, the feature WIDTH, and the aggregator min; cells of the glue matrix are coloured with the colour of the constituent to which they are related.

# 3.378   MIN_WIDTH_PROPER_PLAIN

▶ ▷ ◁ ◀        DESCRIPTION              AUTOMATON

| | |
|---|---|
| **Origin** | Based on the PROPER_PLAIN pattern. |
| **Constraint** | MIN_WIDTH_PROPER_PLAIN(VALUE, VARIABLES) |
| **Arguments** | VALUE      : dvar |
| | VARIABLES  : collection(var−dvar) |
| **Restrictions** | $\text{sv} \leq 3 \vee \text{rv} \leq 1 \Rightarrow \text{VALUE} = \text{sv} + 1$ |
| | $\text{VALUE} \geq 2$ |
| | $\text{VALUE} = \text{sv} + 1 \vee \text{VALUE} \leq \text{sv} - 2$ ① |
| | required(VARIABLES, var) |
| | where |
| | $\text{sv} = |\text{VARIABLES}|$ |
| | $\text{rv} = \text{range}(\text{VARIABLES.var})$ |
| **Purpose** | VALUE is the minimal width of occurrences of the PROPER_PLAIN pattern in the time-series given by the VARIABLES collection. If the pattern does not occur, VALUE takes the default value $|\text{VARIABLES}| + 1$.<br>An occurrence of the pattern PROPER_PLAIN is the *maximal* subsequence which matches the regular expression $> =^+ <$.<br>Assume that the occurrence of the pattern PROPER_PLAIN starts at position $i$ and ends at position $j$. The feature WIDTH computes the value $j - i$. |

| | |
|---|---|
| **Example** | $(2, \langle 2, 7, 5, 5, 6, 3, 7, 4, 4, 5, 6, 5, 3, 3, 3, 5 \rangle)$ |

Figure 3.885 provides an example where the MIN_WIDTH_PROPER_PLAIN $(2, [2, 7, 5, 5, 6, 3, 7, 4, 4, 5, 6, 5, 3, 3, 3, 5])$ constraint holds.

| | |
|---|---|
| **Typical** | $|\text{VARIABLES}| > 3$ |
| | $\text{range}(\text{VARIABLES.var}) > 1$ |
| **Symmetries** | • Items of VARIABLES can be reversed. |
| | • One and the same constant can be added to the var attribute of all items of VARIABLES. |
| **Arg. properties** | Functional dependency: VALUE determined by VARIABLES. |

Figure 3.885: Illustrating the MIN_WIDTH_PROPER_PLAIN constraint of the **Example** slot

 Figures 3.886 and 3.887 respectively depict the automaton associated with the constraint
MIN_WIDTH_PROPER_PLAIN and its simplified form.

Figure 3.886: Automaton for the MIN_WIDTH_PROPER_PLAIN constraint obtained by
applying decoration Table 2.35 to the seed transducer of the PROPER_PLAIN pattern
where default is $|\text{VARIABLES}| + 1$

Figure 3.887: Automaton for the MIN_WIDTH_PROPER_PLAIN constraint obtained by
applying decoration Table 2.25 to the seed transducer of the PROPER_PLAIN pattern
where default is $|\text{VARIABLES}| + 1$; $-R_i + R_{i-1} \geq 0$ is a linear invariant.

|   | $s$ | $r$ | $t$ |
|---|---|---|---|
| $s$ | $\min(\overrightarrow{C}, \overleftarrow{C})$ | $\min(\overrightarrow{C}, \overleftarrow{C})$ | $\min(\overrightarrow{C}, \overleftarrow{C})$ |
| $r$ | $\min(\overrightarrow{C}, \overleftarrow{C})$ | $\min(\overrightarrow{C}, \overleftarrow{C})$ | $\overrightarrow{D} + \overleftarrow{D} + 1$ |
| $t$ | $\min(\overrightarrow{C}, \overleftarrow{C})$ | $\overrightarrow{D} + \overleftarrow{D} + 1$ | $\overrightarrow{D} + \overleftarrow{D} + 1$ |

Table 3.115: Glue matrix for the MIN_WIDTH_PROPER_PLAIN constraint defined as the composition of the PROPER_PLAIN pattern , the feature WIDTH , and the aggregator min ; cells of the glue matrix are coloured with the colour of the constituent to which they are related.

## 3.379　MIN_WIDTH_PROPER_PLATEAU

　　　　DESCRIPTION　　　　　AUTOMATON

**Origin**　　　　Based on the PROPER_PLATEAU pattern.

**Constraint**　　MIN_WIDTH_PROPER_PLATEAU(VALUE, VARIABLES)

**Arguments**
```
VALUE      : dvar
VARIABLES  : collection(var−dvar)
```

**Restrictions**
$\text{sv} \leq 3 \lor \text{rv} \leq 1 \Rightarrow \text{VALUE} = \text{sv} + 1$
$\text{VALUE} \geq 2$
$\text{VALUE} = \text{sv} + 1 \lor \text{VALUE} \leq \text{sv} - 2①$
required(VARIABLES, var)
where
$\text{sv} = |\text{VARIABLES}|$
$\text{rv} = \text{range}(\text{VARIABLES.var})$

**Purpose**
VALUE is the minimal width of occurrences of the PROPER_PLATEAU pattern in the time-series given by the VARIABLES collection. If the pattern does not occur, VALUE takes the default value $|\text{VARIABLES}| + 1$.

An occurrence of the pattern PROPER_PLATEAU is the *maximal* subsequence which matches the regular expression $< =^+ >$.

Assume that the occurrence of the pattern PROPER_PLATEAU starts at position $i$ and ends at position $j$. The feature WIDTH computes the value $j - i$.

**Example**
$(2, \langle 7, 1, 3, 3, 2, 5, 1, 4, 4, 3, 2, 3, 5, 5, 5, 3 \rangle)$

Figure 3.888 provides an example where the MIN_WIDTH_PROPER_PLATEAU $(2, [7, 1, 3, 3, 2, 5, 1, 4, 4, 3, 2, 3, 5, 5, 5, 3])$ constraint holds.

**Typical**
$|\text{VARIABLES}| > 3$
$\text{range}(\text{VARIABLES.var}) > 1$

**Symmetries**
- Items of VARIABLES can be reversed.
- One and the same constant can be added to the var attribute of all items of VARIABLES.

**Arg. properties**
Functional dependency: VALUE determined by VARIABLES.

Figure 3.888: Illustrating the MIN_WIDTH_PROPER_PLATEAU constraint of the **Example** slot

**Automaton**  Figures 3.889 and 3.890 respectively depict the automaton associated with the constraint MIN_WIDTH_PROPER_PLATEAU and its simplified form.

Figure 3.889: Automaton for the MIN_WIDTH_PROPER_PLATEAU constraint obtained by applying decoration Table 2.35 to the seed transducer of the PROPER_PLATEAU pattern where default is $|\text{VARIABLES}| + 1$

Figure 3.890: Automaton for the MIN_WIDTH_PROPER_PLATEAU constraint obtained by applying decoration Table 2.25 to the seed transducer of the PROPER_PLATEAU pattern where default is $|\text{VARIABLES}| + 1$; $-R_i + R_{i-1} \geq 0$ is a linear invariant.

|   | $s$ | $r$ | $t$ |
|---|---|---|---|
| $s$ | $\min(\overrightarrow{C}, \overleftarrow{C})$ | $\min(\overrightarrow{C}, \overleftarrow{C})$ | $\min(\overrightarrow{C}, \overleftarrow{C})$ |
| $r$ | $\min(\overrightarrow{C}, \overleftarrow{C})$ | $\min(\overrightarrow{C}, \overleftarrow{C})$ | $\overrightarrow{D} + \overleftarrow{D} + 1$ |
| $t$ | $\min(\overrightarrow{C}, \overleftarrow{C})$ | $\overrightarrow{D} + \overleftarrow{D} + 1$ | $\overrightarrow{D} + \overleftarrow{D} + 1$ |

Table 3.116: Glue matrix for the MIN_WIDTH_PROPER_PLATEAU constraint defined as the composition of the PROPER_PLATEAU pattern, the feature WIDTH, and the aggregator min; cells of the glue matrix are coloured with the colour of the constituent to which they are related.

## 3.380   MIN_WIDTH_STEADY_SEQUENCE

▶ ▷ ◁ ◀        DESCRIPTION              AUTOMATON

**Origin**          Based on the STEADY_SEQUENCE pattern.

**Constraint**      MIN_WIDTH_STEADY_SEQUENCE(VALUE, VARIABLES)

**Arguments**       VALUE     : dvar
                    VARIABLES : collection(var−dvar)

**Restrictions**    $\text{sv} \leq 1 \Rightarrow \text{VALUE} = \text{sv} + 1$
                    $\text{rv} = 1 \Rightarrow \text{VALUE} \geq \text{sv}$
                    $\text{rv} \geq 2 \Rightarrow \text{VALUE} \geq 2$
                    $\text{VALUE} = \text{sv} + 1 \lor \text{VALUE} \leq \text{sv}①$
                    required(VARIABLES, var)
                    where
                    $\text{rv} = $ range(VARIABLES.var)
                    $\text{sv} = |\text{VARIABLES}|$

**Purpose**         VALUE is the minimal width of occurrences of the STEADY_SEQUENCE pattern in the
                    time-series given by the VARIABLES collection. If the pattern does not occur, VALUE
                    takes the default value $|\text{VARIABLES}| + 1$.
                    An occurrence of the pattern STEADY_SEQUENCE is the *maximal* subsequence which
                    matches the regular expression $=^+$.
                    Assume that the occurrence of the pattern STEADY_SEQUENCE starts at position $i$ and
                    ends at position $j$. The feature WIDTH computes the value $j - i + 2$.

**Example**         $(2, \langle 3, 1, 1, 4, 5, 5, 5, 6, 2, 2, 4, 4, 3, 2, 1, 1 \rangle)$

                    Figure 3.891 provides an example where the MIN_WIDTH_STEADY_SEQUENCE
                    $(2, [3, 1, 1, 4, 5, 5, 5, 6, 2, 2, 4, 4, 3, 2, 1, 1])$ constraint holds.

**Typical**         $|\text{VARIABLES}| > 1$

**Symmetries**      • Items of VARIABLES can be reversed.

                    • One and the same constant can be added to the var attribute of all items of
                      VARIABLES.

**Arg. properties** Functional dependency: VALUE determined by VARIABLES.

Figure 3.891: Illustrating the MIN_WIDTH_STEADY_SEQUENCE constraint of the **Example** slot

Automaton        Figures 3.892 and 3.893 respectively depict the automaton associated with the constraint
                 MIN_WIDTH_STEADY_SEQUENCE and its simplified form.

Figure 3.892: Automaton for the MIN_WIDTH_STEADY_SEQUENCE constraint obtained by applying decoration Table 2.35 to the seed transducer of the STEADY_SEQUENCE pattern where default is $|\texttt{VARIABLES}| + 1$

Figure 3.893: Automaton for the MIN_WIDTH_STEADY_SEQUENCE constraint obtained by applying decoration Table 2.24 to the seed transducer of the STEADY_SEQUENCE pattern where default is $|\texttt{VARIABLES}| + 1$; $-R_i + R_{i-1} \geq 0$ is a linear invariant.

|   | $s$ | $r$ |
|---|---|---|
| $s$ | $\min(\overrightarrow{C}, \overleftarrow{C})$ | $\min(\overrightarrow{C}, \overleftarrow{C})$ |
| $r$ | $\min(\overrightarrow{C}, \overleftarrow{C})$ | $\overrightarrow{C} + \overleftarrow{C} + \overrightarrow{D} + \overleftarrow{D} + 1$ |

Table 3.117: Glue matrix for the MIN_WIDTH_STEADY_SEQUENCE constraint defined as the composition of the STEADY_SEQUENCE pattern , the feature WIDTH , and the aggregator min ; cells of the glue matrix are coloured with the colour of the constituent to which they are related.

# 3.381　MIN_WIDTH_STRICTLY_DECREASING_SEQUENCE

▶ ▷ ◁ ◀　　　DESCRIPTION　　　AUTOMATON

| **Origin** | Based on the STRICTLY_DECREASING_SEQUENCE pattern. |
|---|---|
| **Constraint** | MIN_WIDTH_STRICTLY_DECREASING_SEQUENCE(VALUE, VARIABLES) |

| **Arguments** | VALUE　　　　: dvar<br>VARIABLES　: collection(var−dvar) |
|---|---|

**Restrictions**

$\mathtt{sv} \leq 1 \vee \mathtt{rv} \leq 1 \Rightarrow \mathtt{VALUE} = \mathtt{sv} + 1$

$\mathtt{VALUE} \geq 2$

$\mathtt{VALUE} = \mathtt{sv} + 1 \vee \mathtt{VALUE} \leq \min(\mathtt{sv}, \mathtt{rv}①)$

required(VARIABLES, var)

where

$\mathtt{sv} = |\mathtt{VARIABLES}|$

$\mathtt{rv} = \mathtt{range}(\mathtt{VARIABLES.var})$

**Purpose**

VALUE is the minimal width of occurrences of the STRICTLY_DECREASING_SEQUENCE pattern in the time-series given by the VARIABLES collection. If the pattern does not occur, VALUE takes the default value $|\mathtt{VARIABLES}| + 1$.

An occurrence of the pattern STRICTLY_DECREASING_SEQUENCE is the *maximal* subsequence which matches the regular expression $>^+$.

Assume that the occurrence of the pattern STRICTLY_DECREASING_SEQUENCE starts at position $i$ and ends at position $j$. The feature WIDTH computes the value $j - i + 2$.

**Example**

$(2, \langle 4, 4, 6, 4, 1, 1, 3, 4, 4, 6, 6, 5, 2, 2, 4, 3 \rangle)$

Figure 3.894 provides an example where the MIN_WIDTH_STRICTLY_DECREASING_SEQUENCE $(2, [4, 4, 6, 4, 1, 1, 3, 4, 4, 6, 6, 5, 2, 2, 4, 3])$ constraint holds.

**Typical**

$|\mathtt{VARIABLES}| > 1$

range(VARIABLES.var) > 1

**Symmetry**

One and the same constant can be added to the var attribute of all items of VARIABLES.

**Arg. properties**

Functional dependency: VALUE determined by VARIABLES.

Figure 3.894: Illustrating the MIN_WIDTH_STRICTLY_DECREASING_SEQUENCE constraint of the **Example** slot

 Figures 3.895 and 3.896 respectively depict the automaton associated with the constraint MIN_WIDTH_STRICTLY_DECREASING_SEQUENCE and its simplified form.

Figure 3.895: Automaton for the MIN_WIDTH_STRICTLY_DECREASING_SEQUENCE constraint obtained by applying decoration Table 2.35 to the seed transducer of the STRICTLY_DECREASING_SEQUENCE pattern where default is $|\text{VARIABLES}| + 1$

Figure 3.896: Automaton for the MIN_WIDTH_STRICTLY_DECREASING_SEQUENCE constraint obtained by applying decoration Table 2.24 to the seed transducer of the STRICTLY_DECREASING_SEQUENCE pattern where default is $|\text{VARIABLES}|+1$; $-R_i + R_{i-1} \geq 0$ is a linear invariant.

|   | $s$ | $r$ |
|---|---|---|
| $s$ | $\min(\overrightarrow{C}, \overleftarrow{C})$ | $\min(\overrightarrow{C}, \overleftarrow{C})$ |
| $r$ | $\min(\overrightarrow{C}, \overleftarrow{C})$ | $\overrightarrow{C} + \overleftarrow{C} + \overrightarrow{D} + \overleftarrow{D} + 1$ |

Table 3.118: Glue matrix for the MIN_WIDTH_STRICTLY_DECREASING_SEQUENCE constraint defined as the composition of the STRICTLY_DECREASING_SEQUENCE pattern , the feature WIDTH , and the aggregator min ; cells of the glue matrix are coloured with the colour of the constituent to which they are related.

# 3.382 MIN_WIDTH_STRICTLY_INCREASING_SEQUENCE

    **DESCRIPTION**    **AUTOMATON**

| | |
|---|---|
| **Origin** | Based on the STRICTLY_INCREASING_SEQUENCE pattern. |
| **Constraint** | MIN_WIDTH_STRICTLY_INCREASING_SEQUENCE(VALUE, VARIABLES) |
| **Arguments** | VALUE    : dvar<br>VARIABLES : collection(var−dvar) |

**Restrictions**

$$sv \leq 1 \vee rv \leq 1 \Rightarrow VALUE = sv + 1$$
$$VALUE \geq 2$$
$$VALUE = sv + 1 \vee VALUE \leq \min(sv, rv ①)$$
required(VARIABLES, var)

where

$$sv = |VARIABLES|$$
$$rv = range(VARIABLES.var)$$

**Purpose**

VALUE is the minimal width of occurrences of the STRICTLY_INCREASING_SEQUENCE pattern in the time-series given by the VARIABLES collection. If the pattern does not occur, VALUE takes the default value $|VARIABLES| + 1$.

An occurrence of the pattern STRICTLY_INCREASING_SEQUENCE is the *maximal* sub-sequence which matches the regular expression $<^+$.

Assume that the occurrence of the pattern STRICTLY_INCREASING_SEQUENCE starts at position $i$ and ends at position $j$. The feature WIDTH computes the value $j - i + 2$.

**Example**

$$(2, \langle 4, 3, 5, 5, 2, 1, 1, 2, 3, 4, 6, 6, 3, 1, 2, 3 \rangle)$$

Figure 3.897 provides an example where the MIN_WIDTH_STRICTLY_INCREASING_SEQUENCE $(2, [4, 3, 5, 5, 2, 1, 1, 2, 3, 4, 6, 6, 3, 1, 2, 3])$ constraint holds.

**Typical**

$$|VARIABLES| > 1$$
$$range(VARIABLES.var) > 1$$

**Symmetry**

One and the same constant can be added to the var attribute of all items of VARIABLES.

**Arg. properties**

Functional dependency: VALUE determined by VARIABLES.

Figure 3.897: Illustrating the MIN_WIDTH_STRICTLY_INCREASING_SEQUENCE constraint of the **Example** slot

Figures 3.898 and 3.899 respectively depict the automaton associated with the constraint MIN_WIDTH_STRICTLY_INCREASING_SEQUENCE and its simplified form.

Figure 3.898: Automaton for the MIN_WIDTH_STRICTLY_INCREASING_SEQUENCE constraint obtained by applying decoration Table 2.35 to the seed transducer of the STRICTLY_INCREASING_SEQUENCE pattern where default is $|\texttt{VARIABLES}| + 1$

Figure 3.899: Automaton for the MIN_WIDTH_STRICTLY_INCREASING_SEQUENCE constraint obtained by applying decoration Table 2.24 to the seed transducer of the STRICTLY_INCREASING_SEQUENCE pattern where default is $|\texttt{VARIABLES}| + 1$; $-R_i + R_{i-1} \geq 0$ is a linear invariant.

|   | $s$ | $r$ |
|---|---|---|
| $s$ | $\min(\overrightarrow{C}, \overleftarrow{C})$ | $\min(\overrightarrow{C}, \overleftarrow{C})$ |
| $r$ | $\min(\overrightarrow{C}, \overleftarrow{C})$ | $\overrightarrow{C} + \overleftarrow{C} + \overrightarrow{D} + \overleftarrow{D} + 1$ |

Table 3.119: Glue matrix for the MIN_WIDTH_STRICTLY_INCREASING_SEQUENCE constraint defined as the composition of the STRICTLY_INCREASING_SEQUENCE pattern, the feature WIDTH, and the aggregator min; cells of the glue matrix are coloured with the colour of the constituent to which they are related.

# 3.383   MIN_WIDTH_SUMMIT

▶ ▷ ◁ ◀        **DESCRIPTION**          **AUTOMATON**

$(< | < (= | <)^* <)(> | > (= | >)^* >)$

| **Origin** | Based on the SUMMIT pattern. |
|---|---|

| **Constraint** | MIN_WIDTH_SUMMIT(VALUE, VARIABLES) |
|---|---|

| **Arguments** | VALUE     : dvar |
|---|---|
| | VARIABLES : collection(var−dvar) |

**Restrictions**

$\mathtt{sv} \leq 2 \vee \mathtt{rv} \leq 1 \Rightarrow \mathtt{VALUE} = \mathtt{sv} + 1$

$\mathtt{VALUE} \geq 1$

$\mathtt{rv} = 2 \Rightarrow \mathtt{VALUE} = \mathtt{sv} + 1 \vee \mathtt{VALUE} \leq 1$①

$\mathtt{rv} \geq 3 \Rightarrow \mathtt{VALUE} = \mathtt{sv} + 1 \vee \mathtt{VALUE} \leq \mathtt{sv} - 2$②

required(VARIABLES, var)

where

$\mathtt{sv} = |\mathtt{VARIABLES}|$

$\mathtt{rv} = \mathtt{range}(\mathtt{VARIABLES.var})$

**Purpose**

VALUE is the minimal width of occurrences of the SUMMIT pattern in the time-series given by the VARIABLES collection. If the pattern does not occur, VALUE takes the default value $|\mathtt{VARIABLES}| + 1$.

An occurrence of the pattern SUMMIT is the *maximal* subsequence which matches the regular expression $(< | < (= | <)^* <)(> | > (= | >)^* >)$.

Assume that the occurrence of the pattern SUMMIT starts at position $i$ and ends at position $j$. The feature WIDTH computes the value $j - i$.

**Example**

$(1, \langle 7, 1, 5, 4, 4, 3, 3, 4, 6, 6, 2, 3, 4, 2, 3, 1 \rangle)$

Figure 3.900 provides an example where the MIN_WIDTH_SUMMIT $(1, [7, 1, 5, 4, 4, 3, 3, 4, 6, 6, 2, 3, 4, 2, 3, 1])$ constraint holds.

**Typical**

$|\mathtt{VARIABLES}| > 2$

$\mathtt{range}(\mathtt{VARIABLES.var}) > 1$

**Symmetries**

- Items of VARIABLES can be reversed.
- One and the same constant can be added to the var attribute of all items of VARIABLES.

**Arg. properties**

Functional dependency: VALUE determined by VARIABLES.

Figure 3.900: Illustrating the MIN_WIDTH_SUMMIT constraint of the **Example** slot

                   Figure 3.901 depicts the automaton associated with the constraint MIN_WIDTH_SUMMIT.

Figure 3.901: Automaton for the MIN_WIDTH_SUMMIT constraint obtained by applying decoration Table 2.35 to the seed transducer of the SUMMIT pattern where default is $|\texttt{VARIABLES}| + 1$ (transition $u \rightarrow r$ has the same accumulator update as transition $r \rightarrow u$); $-R_i + R_{i-1} \geq 0$ is a linear invariant.

|   | $s$ | $r$ | $t$ | $u$ |
|---|---|---|---|---|
| $s$ | $\min(\overrightarrow{C}, \overleftarrow{C})$ | $\min(\overrightarrow{C}, \overleftarrow{C})$ | $\min(\overrightarrow{C}, \overleftarrow{C})$ | $\min(\overrightarrow{C}, \overleftarrow{C})$ |
| $r$ | $\min(\overrightarrow{C}, \overleftarrow{C})$ | $\overrightarrow{D} + \overleftarrow{D} + 1$ | $\overleftarrow{C} + \overrightarrow{D} + \overleftarrow{D} + 1$ | $\min(\overrightarrow{C}, \overleftarrow{C})$ |
| $t$ | $\min(\overrightarrow{C}, \overleftarrow{C})$ | $\overrightarrow{C} + \overrightarrow{D} + \overleftarrow{D} + 1$ | $\min(\overrightarrow{C}, \overleftarrow{C})$ | $\overrightarrow{C} + \overrightarrow{D} + \overleftarrow{D} + 1$ |
| $u$ | $\min(\overrightarrow{C}, \overleftarrow{C})$ | $\min(\overrightarrow{C}, \overleftarrow{C})$ | $\overleftarrow{C} + \overrightarrow{D} + \overleftarrow{D} + 1$ | $\min(\overrightarrow{C}, \overleftarrow{C})$ |

Table 3.120: Glue matrix for the MIN_WIDTH_SUMMIT constraint defined as the composition of the SUMMIT pattern, the feature WIDTH, and the aggregator min; cells of the glue matrix are coloured with the colour of the constituent to which they are related.

## 3.384  MIN_WIDTH_VALLEY

▶ ▷ ◁ ◀         **DESCRIPTION**          **AUTOMATON**

$> (= | >)^* (< | =)^* <$

| | |
|---|---|
| **Origin** | Based on the VALLEY pattern. |
| **Constraint** | MIN_WIDTH_VALLEY(VALUE, VARIABLES) |
| **Arguments** | VALUE      : dvar |
| | VARIABLES  : collection(var−dvar) |
| **Restrictions** | $\mathtt{sv} \leq 2 \vee \mathtt{rv} \leq 1 \Rightarrow \mathtt{VALUE} = \mathtt{sv} + 1$ |
| | $\mathtt{VALUE} \geq 1$ |
| | $\mathtt{VALUE} = \mathtt{sv} + 1 \vee \mathtt{VALUE} \leq \mathtt{sv} - 2$ ① |
| | required(VARIABLES, var) |
| | where |
| | $\mathtt{sv} = |\mathtt{VARIABLES}|$ |
| | $\mathtt{rv} = \text{range}(\mathtt{VARIABLES.var})$ |

VALUE ≤ **4** = 6 − 2

| | |
|---|---|
| **Purpose** | VALUE is the minimal width of occurrences of the VALLEY pattern in the time-series given by the VARIABLES collection. If the pattern does not occur, VALUE takes the default value $|\mathtt{VARIABLES}| + 1$. |
| | An occurrence of the pattern VALLEY is the *maximal* subsequence which matches the regular expression $> (= | >)^* (< | =)^* <$. |
| | Assume that the occurrence of the pattern VALLEY starts at position $i$ and ends at position $j$. The feature WIDTH computes the value $j - i$. |
| **Example** | $(2, \langle 1, 3, 7, 4, 3, 6, 6, 5, 3, 3, 2, 6, 5, 5, 5, 7 \rangle)$ |

Figure 3.902 provides an example where the MIN_WIDTH_VALLEY $(2, [1, 3, 7, 4, 3, 6, 6, 5, 3, 3, 2, 6, 5, 5, 5, 7])$ constraint holds.

| | |
|---|---|
| **Typical** | $|\mathtt{VARIABLES}| > 2$ |
| | range(VARIABLES.var) > 1 |
| **Symmetries** | • Items of VARIABLES can be reversed. |
| | • One and the same constant can be added to the var attribute of all items of VARIABLES. |
| **Arg. properties** | Functional dependency: VALUE determined by VARIABLES. |

Figure 3.902: Illustrating the MIN_WIDTH_VALLEY constraint of the **Example** slot

**Automaton**      Figure 3.903 depicts the automaton associated with the constraint MIN_WIDTH_VALLEY.

Figure 3.903: Automaton for the MIN_WIDTH_VALLEY constraint obtained by applying decoration Table 2.35 to the seed transducer of the VALLEY pattern where default is $|\texttt{VARIABLES}| + 1$; $-R_i + R_{i-1} \geq 0$ is a linear invariant.

|   | $s$ | $r$ | $t$ |
|---|---|---|---|
| $s$ | $\min(\overrightarrow{C}, \overleftarrow{C})$ | $\min(\overrightarrow{C}, \overleftarrow{C})$ | $\min(\overrightarrow{C}, \overleftarrow{C})$ |
| $r$ | $\min(\overrightarrow{C}, \overleftarrow{C})$ | $\overrightarrow{D} + \overleftarrow{D} + 1$ | $\overleftarrow{C} + \overrightarrow{D} + \overleftarrow{D} + 1$ |
| $t$ | $\min(\overrightarrow{C}, \overleftarrow{C})$ | $\overrightarrow{C} + \overrightarrow{D} + \overleftarrow{D} + 1$ | $\min(\overrightarrow{C}, \overleftarrow{C})$ |

Table 3.121: Glue matrix for the MIN_WIDTH_VALLEY constraint defined as the composition of the VALLEY pattern , the feature WIDTH , and the aggregator min ; cells of the glue matrix are coloured with the colour of the constituent to which they are related.

## 3.385 MIN_WIDTH_ZIGZAG

     DESCRIPTION          AUTOMATON

| | |
|---|---|
| **Origin** | Based on the ZIGZAG pattern. |
| **Constraint** | MIN_WIDTH_ZIGZAG(VALUE, VARIABLES) |

| | | |
|---|---|---|
| **Arguments** | VALUE | : dvar |
| | VARIABLES | : collection(var−dvar) |

**Restrictions**

$$\text{sv} \leq 3 \vee \text{rv} \leq 1 \Rightarrow \text{VALUE} = \text{sv} + 1$$
$$\text{VALUE} \geq 2$$
$$\text{VALUE} = \text{sv} + 1 \vee \text{VALUE} \leq \text{sv} - 2 \text{①}$$
required(VARIABLES, var)
where
$$\text{sv} = |\text{VARIABLES}|$$
$$\text{rv} = \text{range}(\text{VARIABLES.var})$$

**Purpose**

VALUE is the minimal width of occurrences of the ZIGZAG pattern in the time-series given by the VARIABLES collection. If the pattern does not occur, VALUE takes the default value $|\text{VARIABLES}| + 1$.

An occurrence of the pattern ZIGZAG is the *maximal* subsequence which matches the regular expression $(<>)^+(< | <>) | (><)^+(> | ><)$.

Assume that the occurrence of the pattern ZIGZAG starts at position $i$ and ends at position $j$. The feature WIDTH computes the value $j - i$.

**Example**

$$(2, \langle 4, 1, 3, 1, 4, 6, 1, 5, 5, 2, 7, 2, 3, 1, 6, 1 \rangle)$$

Figure 3.904 provides an example where the MIN_WIDTH_ZIGZAG $(2, [4, 1, 3, 1, 4, 6, 1, 5, 5, 2, 7, 2, 3, 1, 6, 1])$ constraint holds.

**Typical**

$$|\text{VARIABLES}| > 3$$
$$\text{range}(\text{VARIABLES.var}) > 1$$

**Symmetries**

- Items of VARIABLES can be reversed.
- One and the same constant can be added to the var attribute of all items of VARIABLES.

**Arg. properties**

Functional dependency: VALUE determined by VARIABLES.

Figure 3.904: Illustrating the MIN_WIDTH_ZIGZAG constraint of the **Example** slot

**Automaton**          Figures 3.905 and 3.906 respectively depict the automaton associated with the constraint
                       MIN_WIDTH_ZIGZAG and its simplified form.

Figure 3.905: Automaton for the MIN_WIDTH_ZIGZAG constraint obtained by applying decoration Table 2.35 to the seed transducer of the ZIGZAG pattern where default is |VARIABLES| + 1; (1) missing transitions from $a, b, c, d, e, f$ to $s$ are labelled by =; (2) on transitions from $b, c, e, f$ to $s$ the accumulator $D$ is reset to its initial value; (3) on transitions from $c, f$ to $s$ the accumulator $R$ is updated wrt $C$ and the accumulator $C$ is reset to its initial value

Figure 3.906: Automaton for the MIN_WIDTH_ZIGZAG constraint obtained by applying decoration Table 2.23 to the seed transducer of the ZIGZAG pattern where default is $|\mathtt{VARIABLES}| + 1$; missing transitions from $a, b, c, d, e, f$ to $s$ are labelled by $=$; (2) on transitions from $b, c, e, f$ to $s$ the accumulator $D$ is reset to its initial value; (3) on transitions from $c, f$ to $s$ the accumulator $R$ is updated wrt $C$ and the accumulator $C$ is reset to its initial value.; $-R_i + R_{i-1} \geq 0$ is a linear invariant.

| | $s$ | $a$ | $b$ | $c$ | $d$ | $e$ | $f$ |
|---|---|---|---|---|---|---|---|
| $s$ | $\min(\overleftarrow{C},\overrightarrow{C})$ | $\min(\overleftarrow{C},\overrightarrow{C})$ | $\min(\overleftarrow{C},\overrightarrow{C})$ | $\min(\overleftarrow{C},\overrightarrow{C})$ | $\min(\overleftarrow{C},\overrightarrow{C})$ | $\min(\overleftarrow{C},\overrightarrow{C})$ | $\min(\overleftarrow{C},\overrightarrow{C})$ |
| $a$ | $\min(\overleftarrow{C},\overrightarrow{C})$ | $\min(\overleftarrow{C},\overrightarrow{C})$ | $\overrightarrow{D}+\overleftarrow{D}+1$ | $\overleftarrow{C}+\overrightarrow{D}+\overleftarrow{D}+1$ | $\overrightarrow{D}+\overleftarrow{D}+1$ | $\overrightarrow{D}+\overleftarrow{D}+1$ | $\min(\overleftarrow{C},\overrightarrow{C})$ |
| $b$ | $\min(\overleftarrow{C},\overrightarrow{C})$ | $\min(\overleftarrow{C},\overrightarrow{C})$ | $\min(\overleftarrow{C},\overrightarrow{C})$ | $\min(\overleftarrow{C},\overrightarrow{C})$ | $\min(\overleftarrow{C},\overrightarrow{C})$ | $\min(\overleftarrow{C},\overrightarrow{C})$ | $\overleftarrow{C}+\overrightarrow{D}+\overleftarrow{D}+1$ |
| $c$ | $\min(\overleftarrow{C},\overrightarrow{C})$ | $\overleftarrow{C}+\overrightarrow{D}+\overleftarrow{D}+1$ | $\overrightarrow{D}+\overleftarrow{D}+1$ | $\overleftarrow{C}+\overrightarrow{C}+\overrightarrow{D}+\overleftarrow{D}+1$ | $\min(\overleftarrow{C},\overrightarrow{C})$ | $\overleftarrow{C}+\overrightarrow{D}+\overleftarrow{D}+1$ | $\min(\overleftarrow{C},\overrightarrow{C})$ |
| $d$ | $\min(\overleftarrow{C},\overrightarrow{C})$ | $\min(\overleftarrow{C},\overrightarrow{C})$ | $\overrightarrow{D}+\overleftarrow{D}+1$ | $\min(\overleftarrow{C},\overrightarrow{C})$ | $\overrightarrow{D}+\overleftarrow{D}+1$ | $\min(\overleftarrow{C},\overrightarrow{C})$ | $\overleftarrow{C}+\overrightarrow{D}+\overleftarrow{D}+1$ |
| $e$ | $\min(\overleftarrow{C},\overrightarrow{C})$ | $\overrightarrow{D}+\overleftarrow{D}+1$ | $\min(\overleftarrow{C},\overrightarrow{C})$ | $\overleftarrow{C}+\overrightarrow{D}+\overleftarrow{D}+1$ | $\min(\overleftarrow{C},\overrightarrow{C})$ | $\overrightarrow{D}+\overleftarrow{D}+1$ | $\min(\overleftarrow{C},\overrightarrow{C})$ |
| $f$ | $\min(\overleftarrow{C},\overrightarrow{C})$ | $\min(\overleftarrow{C},\overrightarrow{C})$ | $\overleftarrow{C}+\overrightarrow{D}+\overleftarrow{D}+1$ | $\min(\overleftarrow{C},\overrightarrow{C})$ | $\overleftarrow{C}+\overrightarrow{D}+\overleftarrow{D}+1$ | $\min(\overleftarrow{C},\overrightarrow{C})$ | $\overleftarrow{C}+\overrightarrow{C}+\overrightarrow{D}+\overleftarrow{D}+1$ |

Table 3.122: Glue matrix for the MIN_WIDTH_ZIGZAG constraint defined as the composition of the ZIGZAG pattern , the feature WIDTH , and the aggregator min ; cells of the glue matrix are coloured with the colour of the constituent to which they are related.

## 3.386 NB_BUMP_ON_DECREASING_SEQUENCE

    **DESCRIPTION**          **AUTOMATON**

**Origin**         Based on the BUMP_ON_DECREASING_SEQUENCE pattern.

**Constraint**     NB_BUMP_ON_DECREASING_SEQUENCE(VALUE, VARIABLES)

**Arguments**      VALUE      : dvar
                   VARIABLES  : collection(var−dvar)

**Restrictions**   $\mathtt{sv} \leq 5 \vee \mathtt{rv} \leq 2 \Rightarrow \mathtt{VALUE} = 0$
                   $\mathtt{VALUE} \geq 0$
                   $\mathtt{VALUE} \leq \max(0, \lfloor(\mathtt{sv}-3)/3\rfloor①)$
                   required(VARIABLES, var)
                   where
                   $\mathtt{sv} = |\mathtt{VARIABLES}|$
                   $\mathtt{rv} = $ range(VARIABLES.var)

**Purpose**        VALUE is the number of occurrences of the BUMP_ON_DECREASING_SEQUENCE pattern
                   in the time-series given by the VARIABLES collection. If the pattern does not occur,
                   VALUE takes the default value 0.
                   An occurrence of the pattern BUMP_ON_DECREASING_SEQUENCE is the subsequence
                   which matches the regular expression $> > < > >$.

**Example**        $(2, \langle 7, 6, 5, 6, 5, 4, 1, 4, 7, 5, 4, 2, 5, 4, 3, 3 \rangle)$

                   Figure 3.907 provides an example where the NB_BUMP_ON_DECREASING_SEQUENCE
                   $(2, [7, 6, 5, 6, 5, 4, 1, 4, 7, 5, 4, 2, 5, 4, 3, 3])$ constraint holds.

**Typical**        $|\mathtt{VARIABLES}| > 5$
                   range(VARIABLES.var) $> 2$

**Symmetry**       One and the same constant can be added to the var attribute of all items of VARIABLES.

**Arg. properties** Functional dependency: VALUE determined by VARIABLES.

Figure 3.907: Illustrating the nb_bump_on_decreasing_sequence constraint of the **Example** slot

**Automaton**     Figures 3.908 and 3.909 respectively depict the automaton associated with the constraint NB_BUMP_ON_DECREASING_SEQUENCE and its simplified form.

Figure 3.908: Automaton for the NB_BUMP_ON_DECREASING_SEQUENCE constraint obtained by applying decoration Table 2.35 to the seed transducer of the BUMP_ON_DECREASING_SEQUENCE pattern where default is 0

Figure 3.909: Automaton for the NB_BUMP_ON_DECREASING_SEQUENCE constraint obtained by applying decoration Table 2.37 to the seed transducer of the BUMP_ON_DECREASING_SEQUENCE pattern where default is $0$; $R_i - R_{i-1} \geq 0$ and $-R_i + R_{i-3} + 1 \geq 0$ are linear invariants.

# 3.387 NB_DECREASING

**DESCRIPTION**          **AUTOMATON**

| | |
|---|---|
| **Origin** | Based on the DECREASING pattern. |
| **Constraint** | NB_DECREASING(VALUE, VARIABLES) |
| **Arguments** | VALUE      : dvar |
| | VARIABLES  : collection(var−dvar) |

**Restrictions**

$$\mathtt{sv} \leq 1 \vee \mathtt{rv} \leq 1 \Rightarrow \mathtt{VALUE} = 0$$
$$\mathtt{VALUE} \geq 0$$
$$\mathtt{VALUE} \leq \max(0, (\ell - 1) * \lfloor \mathtt{sv}/\ell \rfloor + \max(0, \mathtt{sv} \bmod \ell - 1)) ①$$
required(VARIABLES, var)
where
$$\mathtt{sv} = |\mathtt{VARIABLES}|$$
$$\mathtt{rv} = \mathrm{range}(\mathtt{VARIABLES.var})$$
$$\ell = \min(\mathtt{sv}, \mathtt{rv})$$

**Purpose**

VALUE is the number of occurrences of the DECREASING pattern in the time-series given by the VARIABLES collection. If the pattern does not occur, VALUE takes the default value 0.

An occurrence of the pattern DECREASING is the subsequence which matches the regular expression >.

**Example**

$$(5, \langle 3, 4, 2, 2, 5, 6, 6, 4, 4, 3, 1, 1, 4, 6, 4, 4 \rangle)$$

Figure 3.910 provides an example where the NB_DECREASING $(5, [3, 4, 2, 2, 5, 6, 6, 4, 4, 3, 1, 1, 4, 6, 4, 4])$ constraint holds.

| | |
|---|---|
| **Typical** | $|\mathtt{VARIABLES}| > 1$ |
| | $\mathrm{range}(\mathtt{VARIABLES.var}) > 1$ |
| **Symmetry** | One and the same constant can be added to the var attribute of all items of VARIABLES. |
| **Arg. properties** | Functional dependency: VALUE determined by VARIABLES. |

Figure 3.910: Illustrating the NB_DECREASING constraint of the **Example** slot

**Automaton**     Figures 3.911 and 3.912 respectively depict the automaton associated with the constraint NB_DECREASING and its simplified form.

Figure 3.911: Automaton for the NB_DECREASING constraint obtained by applying decoration Table 2.35 to the seed transducer of the DECREASING pattern where default is 0

Figure 3.912: Automaton for the NB_DECREASING constraint obtained by applying decoration Table 2.37 to the seed transducer of the DECREASING pattern where default is 0; $R_i - R_{i-1} \geq 0$ and $-R_i + R_{i-1} + 1 \geq 0$ are linear invariants.

Table 3.123: Glue matrix for the NB_DECREASING constraint defined as the composition of the DECREASING pattern, the feature ONE, and the aggregator sum; cells of the glue matrix are coloured with the colour of the constituent to which they are related.

## 3.388   NB_DECREASING_SEQUENCE

▶ ▷ ◁ ◀          **DESCRIPTION**          **AUTOMATON**

$> (> | =)^* > |$
$>$

| **Origin** | Based on the DECREASING_SEQUENCE pattern. |

| **Constraint** | NB_DECREASING_SEQUENCE(VALUE, VARIABLES) |

| **Arguments** | VALUE      : dvar |
| | VARIABLES : collection(var−dvar) |

**Restrictions**

$\mathtt{sv} \leq 1 \vee \mathtt{rv} \leq 1 \Rightarrow \mathtt{VALUE} = 0$
$\mathtt{VALUE} \geq 0$
$\mathtt{VALUE} \leq \lfloor \mathtt{sv}/2 \rfloor$ ①
required(VARIABLES, var)
where
$\quad \mathtt{sv} = |\mathtt{VARIABLES}|$
$\quad \mathtt{rv} = \mathtt{range}(\mathtt{VARIABLES.var})$

VALUE ≤ **6**

**Purpose**

VALUE is the number of occurrences of the DECREASING_SEQUENCE pattern in the time-series given by the VARIABLES collection. If the pattern does not occur, VALUE takes the default value 0.
An occurrence of the pattern DECREASING_SEQUENCE is the *maximal* subsequence which matches the regular expression $> (> | =)^* > | >$.

**Example**

$(3, \langle 3, 4, 2, 2, 5, 6, 6, 4, 4, 3, 1, 1, 4, 6, 4, 4 \rangle)$

Figure 3.913 provides an example where the NB_DECREASING_SEQUENCE $(3, [3, 4, 2, 2, 5, 6, 6, 4, 4, 3, 1, 1, 4, 6, 4, 4])$ constraint holds.

| **Typical** | $|\mathtt{VARIABLES}| > 1$ |
| | $\mathtt{range}(\mathtt{VARIABLES.var}) > 1$ |

| **Symmetry** | One and the same constant can be added to the var attribute of all items of VARIABLES. |

| **Arg. properties** | Functional dependency: VALUE determined by VARIABLES. |

Figure 3.913: Illustrating the NB_DECREASING_SEQUENCE constraint of the **Example** slot

**Automaton**      Figures 3.914 and 3.915 respectively depict the automaton associated with the constraint
                   NB_DECREASING_SEQUENCE and its simplified form.

Figure 3.914: Automaton for the NB_DECREASING_SEQUENCE constraint obtained by
applying decoration Table 2.35 to the seed transducer of the DECREASING_SEQUENCE
pattern where default is 0

Figure 3.915: Automaton for the NB_DECREASING_SEQUENCE constraint obtained by
applying decoration Table 2.37 to the seed transducer of the DECREASING_SEQUENCE
pattern where default is 0; $R_i - R_{i-1} \geq 0$ and $-R_i + R_{i-2} + 1 \geq 0$ are linear invariants.

|   | $s$ | $t$ |
|---|---|---|
| $s$ | $\overrightarrow{C} + \overleftarrow{C}$ | $\overrightarrow{C} + \overleftarrow{C}$ |
| $t$ | $\overrightarrow{C} + \overleftarrow{C}$ | $1$ |

Table 3.124: Glue matrix for the NB_DECREASING_SEQUENCE constraint defined as the composition of the DECREASING_SEQUENCE pattern , the feature ONE , and the aggregator sum ; cells of the glue matrix are coloured with the colour of the constituent to which they are related.

## 3.389  NB_DECREASING_TERRACE

▶ ▷ ◁ ◀       DESCRIPTION        AUTOMATON

| | |
|---|---|
| **Origin** | Based on the DECREASING_TERRACE pattern. |
| **Constraint** | NB_DECREASING_TERRACE(VALUE, VARIABLES) |
| **Arguments** | VALUE     : dvar<br>VARIABLES : collection(var−dvar) |

**Restrictions**

$\mathtt{sv} \leq 3 \lor \mathtt{rv} \leq 2 \Rightarrow \mathtt{VALUE} = 0$

$\mathtt{VALUE} \geq 0$

$\mathtt{VALUE} \leq \max(0, \lfloor (\ell - 2)/2 \rfloor * \lfloor \mathtt{sv}/\ell \rfloor + \max(0, \lfloor (\mathtt{sv} \bmod \ell - 2)/2 \rfloor))$ ①

required(VARIABLES, var)

where

$\mathtt{sv} = |\mathtt{VARIABLES}|$

$\mathtt{rv} = \mathrm{range}(\mathtt{VARIABLES.var})$

$\ell = \max(1, \min(\mathtt{sv}, 2 * \mathtt{rv} - 2))$

**Purpose**

VALUE is the number of occurrences of the DECREASING_TERRACE pattern in the time-series given by the VARIABLES collection. If the pattern does not occur, VALUE takes the default value 0.

An occurrence of the pattern DECREASING_TERRACE is the *maximal* subsequence which matches the regular expression $> =^+ >$.

**Example**

$(2, \langle 6, 4, 4, 4, 5, 2, 2, 1, 3, 3, 5, 4, 4, 3, 3, 3 \rangle)$

Figure 3.916 provides an example where the NB_DECREASING_TERRACE $(2, [6, 4, 4, 4, 5, 2, 2, 1, 3, 3, 5, 4, 4, 3, 3, 3])$ constraint holds.

| | |
|---|---|
| **Typical** | $|\mathtt{VARIABLES}| > 3$<br>$\mathrm{range}(\mathtt{VARIABLES.var}) > 2$ |
| **Symmetry** | One and the same constant can be added to the var attribute of all items of VARIABLES. |
| **Arg. properties** | Functional dependency: VALUE determined by VARIABLES. |

Figure 3.916: Illustrating the NB_DECREASING_TERRACE constraint of the **Example** slot

**Automaton**    Figures 3.917 and 3.918 respectively depict the automaton associated with the constraint NB_DECREASING_TERRACE and its simplified form.

Figure 3.917: Automaton for the NB_DECREASING_TERRACE constraint obtained by applying decoration Table 2.35 to the seed transducer of the DECREASING_TERRACE pattern where default is 0

Figure 3.918: Automaton for the NB_DECREASING_TERRACE constraint obtained by applying decoration Table 2.37 to the seed transducer of the DECREASING_TERRACE pattern where default is 0; $R_i - R_{i-1} \geq 0$ and $-R_i + R_{i-2} + 1 \geq 0$ are linear invariants.

|   | $s$ | $r$ | $t$ |
|---|---|---|---|
| $s$ | $\overrightarrow{C} + \overleftarrow{C}$ | $\overrightarrow{C} + \overleftarrow{C}$ | $\overrightarrow{C} + \overleftarrow{C}$ |
| $r$ | $\overrightarrow{C} + \overleftarrow{C}$ | $\overrightarrow{C} + \overleftarrow{C}$ | 1 |
| $t$ | $\overrightarrow{C} + \overleftarrow{C}$ | 1 | 1 |

Table 3.125: Glue matrix for the NB_DECREASING_TERRACE constraint defined as the composition of the DECREASING_TERRACE pattern , the feature ONE , and the aggregator sum ; cells of the glue matrix are coloured with the colour of the constituent to which they are related.

## 3.390　NB_DIP_ON_INCREASING_SEQUENCE

► ▷ ◁ ◀　　　**DESCRIPTION**　　**AUTOMATON**

**Origin**　　Based on the DIP_ON_INCREASING_SEQUENCE pattern.

**Constraint**　　NB_DIP_ON_INCREASING_SEQUENCE(VALUE, VARIABLES)

**Arguments**　　VALUE　　　　: dvar
　　　　　　　　　VARIABLES : collection(var−dvar)

**Restrictions**　　$\text{sv} \leq 5 \vee \text{rv} \leq 2 \Rightarrow \text{VALUE} = 0$
　　　　　　　　$\text{VALUE} \geq 0$
　　　　　　　　$\text{VALUE} \leq \max(0, \lfloor (\text{sv} - 3)/3 \rfloor ①)$
　　　　　　　　required(VARIABLES, var)
　　　　　　　　where
　　　　　　　　$\text{sv} = |\text{VARIABLES}|$
　　　　　　　　$\text{rv} = \text{range}(\text{VARIABLES.var})$

**Purpose**　　VALUE is the number of occurrences of the DIP_ON_INCREASING_SEQUENCE pattern in the time-series given by the VARIABLES collection. If the pattern does not occur, VALUE takes the default value 0.
An occurrence of the pattern DIP_ON_INCREASING_SEQUENCE is the subsequence which matches the regular expression $<<><<$.

**Example**　　$(2, \langle 1, 2, 3, 2, 5, 6, 7, 4, 1, 3, 4, 6, 1, 2, 4, 4 \rangle)$

Figure 3.919 provides an example where the NB_DIP_ON_INCREASING_SEQUENCE $(2, [1, 2, 3, 2, 5, 6, 7, 4, 1, 3, 4, 6, 1, 2, 4, 4])$ constraint holds.

**Typical**　　$|\text{VARIABLES}| > 5$
　　　　　　　$\text{range}(\text{VARIABLES.var}) > 2$

**Symmetry**　　One and the same constant can be added to the var attribute of all items of VARIABLES.

**Arg. properties**　　Functional dependency: VALUE determined by VARIABLES.

Figure 3.919: Illustrating the NB_DIP_ON_INCREASING_SEQUENCE constraint of the **Example** slot

**Automaton**    Figures 3.920 and 3.921 respectively depict the automaton associated with the constraint NB_DIP_ON_INCREASING_SEQUENCE and its simplified form.

Figure 3.920: Automaton for the NB_DIP_ON_INCREASING_SEQUENCE constraint obtained by applying decoration Table 2.35 to the seed transducer of the DIP_ON_INCREASING_SEQUENCE pattern where default is 0

Figure 3.921: Automaton for the NB_DIP_ON_INCREASING_SEQUENCE constraint obtained by applying decoration Table 2.37 to the seed transducer of the DIP_ON_INCREASING_SEQUENCE pattern where default is 0; $R_i - R_{i-1} \geq 0$ and $-R_i + R_{i-3} + 1 \geq 0$ are linear invariants.

# 3.391 NB_GORGE

◄ ▷ ◁ ◄        **DESCRIPTION**        **AUTOMATON**

$(> | > (= | >)^* >)(< | < (= | <)^* <)$

**Origin**        Based on the GORGE pattern.

**Constraint**        NB_GORGE(VALUE, VARIABLES)

**Arguments**        VALUE      : dvar
                     VARIABLES  : collection(var−dvar)

**Restrictions**     $\mathtt{sv} \leq 2 \vee \mathtt{rv} \leq 1 \Rightarrow \mathtt{VALUE} = 0$
                     $\mathtt{VALUE} \geq 0$
                     $\mathtt{VALUE} \leq \max(0, \lfloor (\mathtt{sv} - 1)/2 \rfloor ①)$
                     required(VARIABLES, var)
                     where
                     $\mathtt{sv} = |\mathtt{VARIABLES}|$
                     $\mathtt{rv} = \mathsf{range}(\mathtt{VARIABLES.var})$

**Purpose**          VALUE is the number of occurrences of the GORGE pattern in the time-series given by the
                     VARIABLES collection. If the pattern does not occur, VALUE takes the default value 0.
                     An occurrence of the pattern GORGE is the *maximal* subsequence which matches the
                     regular expression $(> | > (= | >)^* >)(< | < (= | <)^* <)$.

**Example**          $(3, \langle 1, 7, 3, 4, 4, 5, 5, 4, 2, 2, 6, 5, 4, 6, 5, 7 \rangle)$

                     Figure    3.922    provides    an    example    where    the    NB_GORGE
                     $(3, [1, 7, 3, 4, 4, 5, 5, 4, 2, 2, 6, 5, 4, 6, 5, 7])$ constraint holds.

**Typical**          $|\mathtt{VARIABLES}| > 2$
                     $\mathsf{range}(\mathtt{VARIABLES.var}) > 1$

**Symmetries**       • Items of VARIABLES can be reversed.

                     • One and the same constant can be added to the var attribute of all items of
                       VARIABLES.

**Arg. properties**  Functional dependency: VALUE determined by VARIABLES.

Figure 3.922: Illustrating the NB_GORGE constraint of the **Example** slot

**Automaton**     Figures 3.923 and 3.924 respectively depict the automaton associated with the constraint NB_GORGE and its simplified form.

Figure 3.923: Automaton for the NB_GORGE constraint obtained by applying decoration Table 2.35 to the seed transducer of the GORGE pattern where default is 0 (transition $u \rightarrow r$ has the same accumulator update as transition $r \rightarrow u$)

Figure 3.924: Automaton for the NB_GORGE constraint obtained by applying decoration Table 2.37 to the seed transducer of the GORGE pattern where default is 0; $R_i - R_{i-1} \geq 0$ and $-R_i + R_{i-2} + 1 \geq 0$ are linear invariants.

|   | $s$ | $r$ | $t$ | $u$ |
|---|---|---|---|---|
| $s$ | $\overrightarrow{C} + \overleftarrow{C}$ | $\overrightarrow{C} + \overleftarrow{C}$ | $\overrightarrow{C} + \overleftarrow{C}$ | $\overrightarrow{C} + \overleftarrow{C}$ |
| $r$ | $\overrightarrow{C} + \overleftarrow{C}$ | 1 | 1 | $\overrightarrow{C} + \overleftarrow{C}$ |
| $t$ | $\overrightarrow{C} + \overleftarrow{C}$ | 1 | $\overrightarrow{C} + \overleftarrow{C}$ | 1 |
| $u$ | $\overrightarrow{C} + \overleftarrow{C}$ | $\overrightarrow{C} + \overleftarrow{C}$ | 1 | $\overrightarrow{C} + \overleftarrow{C}$ |

Table 3.126: Glue matrix for the NB_GORGE constraint defined as the composition of the GORGE pattern , the feature ONE , and the aggregator sum ; cells of the glue matrix are coloured with the colour of the constituent to which they are related.

## 3.392  NB_INCREASING

▶ ▷ ◁ ◀     **DESCRIPTION**          **AUTOMATON**

| | |
|---|---|
| **Origin** | Based on the INCREASING pattern. |
| **Constraint** | NB_INCREASING(VALUE, VARIABLES) |
| **Arguments** | VALUE      : dvar |
| | VARIABLES : collection(var−dvar) |

**Restrictions**

$$\text{sv} \leq 1 \lor \text{rv} \leq 1 \Rightarrow \text{VALUE} = 0$$
$$\text{VALUE} \geq 0$$
$$\text{VALUE} \leq \max(0, (\ell - 1) * \lfloor \text{sv}/\ell \rfloor + \max(0, \text{sv} \bmod \ell - 1))①$$
required(VARIABLES, var)
where
$$\text{sv} = |\text{VARIABLES}|$$
$$\text{rv} = \text{range}(\text{VARIABLES.var})$$
$$\ell = \min(\text{sv}, \text{rv})$$

**Purpose**

VALUE is the number of occurrences of the INCREASING pattern in the time-series given by the VARIABLES collection. If the pattern does not occur, VALUE takes the default value 0.

An occurrence of the pattern INCREASING is the subsequence which matches the regular expression $<$.

**Example**

$$(5, \langle 4, 3, 5, 5, 2, 1, 1, 3, 3, 4, 6, 6, 3, 1, 3, 3 \rangle)$$

Figure 3.925 provides an example where the NB_INCREASING $(5, [4, 3, 5, 5, 2, 1, 1, 3, 3, 4, 6, 6, 3, 1, 3, 3])$ constraint holds.

**Typical**

$$|\text{VARIABLES}| > 1$$
$$\text{range}(\text{VARIABLES.var}) > 1$$

**Symmetry**     One and the same constant can be added to the var attribute of all items of VARIABLES.

**Arg. properties**     Functional dependency: VALUE determined by VARIABLES.

Figure 3.925: Illustrating the NB_INCREASING constraint of the **Example** slot

**Automaton**    Figures 3.926 and 3.927 respectively depict the automaton associated with the constraint
NB_INCREASING and its simplified form.

Figure 3.926: Automaton for the NB_INCREASING constraint obtained by applying
decoration Table 2.35 to the seed transducer of the INCREASING pattern where default
is 0

Figure 3.927: Automaton for the NB_INCREASING constraint obtained by applying
decoration Table 2.37 to the seed transducer of the INCREASING pattern where default
is 0; $R_i - R_{i-1} \geq 0$ and $-R_i + R_{i-1} + 1 \geq 0$ are linear invariants.

Table 3.127: Glue matrix for the NB_INCREASING constraint defined as the composi-
tion of the INCREASING pattern, the feature ONE, and the aggregator sum; cells of
the glue matrix are coloured with the colour of the constituent to which they are related.

# 3.393  NB_INCREASING_SEQUENCE

► ▷ ◁ ◄          **DESCRIPTION**          **AUTOMATON**

**Origin**            Based on the INCREASING_SEQUENCE pattern.

**Constraint**        NB_INCREASING_SEQUENCE(VALUE, VARIABLES)

**Arguments**         VALUE     : dvar
                      VARIABLES : collection(var−dvar)

**Restrictions**      $sv \leq 1 \lor rv \leq 1 \Rightarrow VALUE = 0$
                      $VALUE \geq 0$
                      $VALUE \leq \lfloor sv/2 \rfloor$ ①
                      required(VARIABLES, var)
                      where
                          $sv = |VARIABLES|$
                          $rv = range(VARIABLES.var)$

**Purpose**           VALUE is the number of occurrences of the INCREASING_SEQUENCE pattern in the time-
                      series given by the VARIABLES collection. If the pattern does not occur, VALUE takes the
                      default value 0.
                      An occurrence of the pattern INCREASING_SEQUENCE is the *maximal* subsequence
                      which matches the regular expression $< (< | =)^* < | <$.

**Example**           $(3, \langle 4, 3, 5, 5, 2, 1, 1, 3, 3, 4, 6, 6, 3, 1, 3, 3 \rangle)$

                      Figure 3.928 provides an example where the NB_INCREASING_SEQUENCE
                      $(3, [4, 3, 5, 5, 2, 1, 1, 3, 3, 4, 6, 6, 3, 1, 3, 3])$ constraint holds.

**Typical**           $|VARIABLES| > 1$
                      $range(VARIABLES.var) > 1$

**Symmetry**          One and the same constant can be added to the var attribute of all items of VARIABLES.

**Arg. properties**   Functional dependency: VALUE determined by VARIABLES.

Figure 3.928: Illustrating the NB_INCREASING_SEQUENCE constraint of the **Example** slot

**Automaton**   Figures 3.929 and 3.930 respectively depict the automaton associated with the constraint NB_INCREASING_SEQUENCE and its simplified form.

Figure 3.929: Automaton for the NB_INCREASING_SEQUENCE constraint obtained by applying decoration Table 2.35 to the seed transducer of the INCREASING_SEQUENCE pattern where default is 0

Figure 3.930: Automaton for the NB_INCREASING_SEQUENCE constraint obtained by applying decoration Table 2.37 to the seed transducer of the INCREASING_SEQUENCE pattern where default is 0; $R_i - R_{i-1} \geq 0$ and $-R_i + R_{i-2} + 1 \geq 0$ are linear invariants.

|       | $s$                                      | $t$                                      |
|-------|------------------------------------------|------------------------------------------|
| $s$   | $\overrightarrow{C} + \overleftarrow{C}$ | $\overrightarrow{C} + \overleftarrow{C}$ |
| $t$   | $\overrightarrow{C} + \overleftarrow{C}$ | $1$                                      |

Table 3.128: Glue matrix for the NB_INCREASING_SEQUENCE constraint defined as the composition of the INCREASING_SEQUENCE pattern , the feature ONE , and the aggregator sum ; cells of the glue matrix are coloured with the colour of the constituent to which they are related.

## 3.394  NB_INCREASING_TERRACE

▶ ▷ ◁ ◀     DESCRIPTION          AUTOMATON

| **Origin** | Based on the INCREASING_TERRACE pattern. |
|---|---|

| **Constraint** | NB_INCREASING_TERRACE(VALUE, VARIABLES) |
|---|---|

| **Arguments** | VALUE       : dvar |
|---|---|
|  | VARIABLES  : collection(var−dvar) |

**Restrictions**

$sv \leq 3 \vee rv \leq 2 \Rightarrow$ VALUE $= 0$

VALUE $\geq 0$

VALUE $\leq \max(0, \lfloor(\ell-2)/2\rfloor * \lfloor sv/\ell \rfloor + \max(0, \lfloor (sv \bmod \ell - 2)/2 \rfloor))$①

required(VARIABLES, var)

where

$sv = |\text{VARIABLES}|$

$rv = \text{range}(\text{VARIABLES.var})$

$\ell = \max(1, \min(sv, 2 * rv - 2))$

**Purpose**

VALUE is the number of occurrences of the INCREASING_TERRACE pattern in the time-series given by the VARIABLES collection. If the pattern does not occur, VALUE takes the default value 0.

An occurrence of the pattern INCREASING_TERRACE is the *maximal* subsequence which matches the regular expression $< =^+ <$.

**Example**

$(2, \langle 1, 3, 3, 3, 2, 5, 5, 6, 4, 4, 2, 3, 3, 3, 4, 4 \rangle)$

Figure 3.931 provides an example where the NB_INCREASING_TERRACE $(2, [1, 3, 3, 3, 2, 5, 5, 6, 4, 4, 2, 3, 3, 3, 4, 4])$ constraint holds.

| **Typical** | $|\text{VARIABLES}| > 3$ |
|---|---|
|  | range(VARIABLES.var) $> 2$ |

| **Symmetry** | One and the same constant can be added to the var attribute of all items of VARIABLES. |
|---|---|

| **Arg. properties** | Functional dependency: VALUE determined by VARIABLES. |
|---|---|

Figure 3.931: Illustrating the NB_INCREASING_TERRACE constraint of the **Example** slot

**Automaton**     Figures 3.932 and 3.933 respectively depict the automaton associated with the constraint
                  NB_INCREASING_TERRACE and its simplified form.

Figure 3.932: Automaton for the NB_INCREASING_TERRACE constraint obtained by
applying decoration Table 2.35 to the seed transducer of the INCREASING_TERRACE
pattern where default is 0

Figure 3.933: Automaton for the NB_INCREASING_TERRACE constraint obtained by
applying decoration Table 2.37 to the seed transducer of the INCREASING_TERRACE
pattern where default is 0; $R_i - R_{i-1} \geq 0$ and $-R_i + R_{i-2} + 1 \geq 0$ are linear invariants.

|     | $s$ | $r$ | $t$ |
| --- | --- | --- | --- |
| $s$ | $\overrightarrow{C} + \overleftarrow{C}$ | $\overrightarrow{C} + \overleftarrow{C}$ | $\overrightarrow{C} + \overleftarrow{C}$ |
| $r$ | $\overrightarrow{C} + \overleftarrow{C}$ | $\overrightarrow{C} + \overleftarrow{C}$ | 1 |
| $t$ | $\overrightarrow{C} + \overleftarrow{C}$ | 1 | 1 |

Table 3.129: Glue matrix for the NB_INCREASING_TERRACE constraint defined as the composition of the INCREASING_TERRACE pattern, the feature ONE, and the aggregator sum; cells of the glue matrix are coloured with the colour of the constituent to which they are related.

# 3.395 NB_INFLEXION

▶ ▷ ◁ ◀          DESCRIPTION          AUTOMATON

$< (< | =)^* > | > (> | =)^* <$

**Origin**            Based on the INFLEXION pattern.

**Constraint**        NB_INFLEXION(VALUE, VARIABLES)

**Arguments**         VALUE     : dvar
                      VARIABLES : collection(var−dvar)

**Restrictions**      $\mathtt{sv} \leq 2 \vee \mathtt{rv} \leq 1 \Rightarrow \mathtt{VALUE} = 0$
                      $\mathtt{VALUE} \geq 0$
                      $\mathtt{VALUE} \leq \max(0, \mathtt{sv} - 2①)$
                      required(VARIABLES, var)
                      where
                      $\mathtt{sv} = |\mathtt{VARIABLES}|$
                      $\mathtt{rv} =$ range(VARIABLES.var)

VALUE ≤ **6** = 8 − 2

VALUE ≤ **6** = 8 − 2

**Purpose**           VALUE is the number of occurrences of the INFLEXION pattern in the time-series given by
                      the VARIABLES collection. If the pattern does not occur, VALUE takes the default value
                      0.
                      An occurrence of the pattern INFLEXION is the *maximal* subsequence which matches the
                      regular expression $< (< | =)^* > | > (> | =)^* <$.

**Example**           $(8, \langle 1, 2, 6, 6, 4, 4, 3, 5, 2, 5, 1, 5, 3, 3, 4, 4 \rangle)$

Figure   3.934   provides   an   example   where   the   NB_INFLEXION
$(8, [1, 2, 6, 6, 4, 4, 3, 5, 2, 5, 1, 5, 3, 3, 4, 4])$ constraint holds.

**Typical**           $|\mathtt{VARIABLES}| > 2$
                      range(VARIABLES.var) > 1

**Symmetries**        • Items of VARIABLES can be reversed.
                      • One and the same constant can be added to the var attribute of all items of
                        VARIABLES.

**Arg. properties**   Functional dependency: VALUE determined by VARIABLES.

Figure 3.934: Illustrating the NB_INFLEXION constraint of the **Example** slot

**Automaton**    Figures 3.935 and 3.936 respectively depict the automaton associated with the constraint NB_INFLEXION and its simplified form.

Figure 3.935: Automaton for the NB_INFLEXION constraint obtained by applying decoration Table 2.35 to the seed transducer of the INFLEXION pattern where default is 0 (transition $r \rightarrow t$ has the same accumulators updates as transition $t \rightarrow r$)

Figure 3.936: Automaton for the NB_INFLEXION constraint obtained by applying decoration Table 2.37 to the seed transducer of the INFLEXION pattern where default is 0 (transition $r \rightarrow t$ has the same accumulator update as transition $t \rightarrow r$); $R_i - R_{i-1} \geq 0$ and $-R_i + R_{i-1} + 1 \geq 0$ are linear invariants.

## 3.396  NB_PEAK

     **DESCRIPTION**         **AUTOMATON**

$< (= | <)^* (> | =)^* >$

| | |
|---|---|
| **Origin** | Based on the PEAK pattern. |
| **Constraint** | NB_PEAK(VALUE, VARIABLES) |
| **Arguments** | VALUE      : dvar |
| | VARIABLES  : collection(var−dvar) |
| **Restrictions** | $\mathtt{sv} \leq 2 \lor \mathtt{rv} \leq 1 \Rightarrow \mathtt{VALUE} = 0$ |
| | $\mathtt{VALUE} \geq 0$ |
| | $\mathtt{VALUE} \leq \max(0, \lfloor (\mathtt{sv} - 1)/2 \rfloor)$ ① |
| | required(VARIABLES, var) |
| | where |
| | $\mathtt{sv} = |\mathtt{VARIABLES}|$ |
| | $\mathtt{rv} = \mathtt{range}(\mathtt{VARIABLES.var})$ |

VALUE ≤ **6**

| | |
|---|---|
| **Purpose** | VALUE is the number of occurrences of the PEAK pattern in the time-series given by the VARIABLES collection. If the pattern does not occur, VALUE takes the default value 0. |
| | An occurrence of the pattern PEAK is the *maximal* subsequence which matches the regular expression $< (= | <)^* (> | =)^* >$. |

| | |
|---|---|
| **Example** | $(3, \langle 7, 5, 5, 1, 4, 5, 2, 2, 3, 5, 6, 2, 3, 3, 3, 1 \rangle)$ |

Figure    3.937    provides    an    example    where    the    NB_PEAK
$(3, [7, 5, 5, 1, 4, 5, 2, 2, 3, 5, 6, 2, 3, 3, 3, 1])$ constraint holds.

| | |
|---|---|
| **Typical** | $|\mathtt{VARIABLES}| > 2$ |
| | range(VARIABLES.var) > 1 |
| **Symmetries** | • Items of VARIABLES can be reversed. |
| | • One and the same constant can be added to the var attribute of all items of VARIABLES. |
| **Arg. properties** | Functional dependency: VALUE determined by VARIABLES. |

Figure 3.937: Illustrating the NB_PEAK constraint of the **Example** slot

**Automaton**    Figures 3.938 and 3.939 respectively depict the automaton associated with the constraint NB_PEAK and its simplified form.

Figure 3.938: Automaton for the NB_PEAK constraint obtained by applying decoration Table 2.35 to the seed transducer of the PEAK pattern where default is 0

Figure 3.939: Automaton for the NB_PEAK constraint obtained by applying decoration Table 2.37 to the seed transducer of the PEAK pattern where default is 0; $R_i - R_{i-1} \geq 0$ and $-R_i + R_{i-2} + 1 \geq 0$ are linear invariants.

|   | $s$ | $r$ | $t$ |
|---|---|---|---|
| $s$ | $\overrightarrow{C} + \overleftarrow{C}$ | $\overrightarrow{C} + \overleftarrow{C}$ | $\overrightarrow{C} + \overleftarrow{C}$ |
| $r$ | $\overrightarrow{C} + \overleftarrow{C}$ | 1 | 1 |
| $t$ | $\overrightarrow{C} + \overleftarrow{C}$ | 1 | $\overrightarrow{C} + \overleftarrow{C}$ |

Table 3.130: Glue matrix for the NB_PEAK constraint defined as the composition of the PEAK pattern , the feature ONE , and the aggregator sum ; cells of the glue matrix are coloured with the colour of the constituent to which they are related.

# 3.397  NB_PLAIN

► ▷ ◁ ◄        DESCRIPTION          AUTOMATON

**Origin**        Based on the PLAIN pattern.

**Constraint**    NB_PLAIN(VALUE, VARIABLES)

**Arguments**     VALUE       : dvar
                  VARIABLES   : collection(var−dvar)

**Restrictions**  $sv \leq 2 \lor rv \leq 1 \Rightarrow VALUE = 0$
                  $VALUE \geq 0$
                  $VALUE \leq \max(0, \lfloor (sv - 1)/2 \rfloor)$ ①
                  required(VARIABLES, var)
                  where
                  $sv = |VARIABLES|$
                  $rv = $ range(VARIABLES.var)

**Purpose**       VALUE is the number of occurrences of the PLAIN pattern in the time-series given by the
                  VARIABLES collection. If the pattern does not occur, VALUE takes the default value 0.
                  An occurrence of the pattern PLAIN is the *maximal* subsequence which matches the
                  regular expression > =* <.

**Example**       $(3, \langle 2, 3, 6, 5, 7, 6, 6, 4, 5, 5, 4, 3, 3, 6, 6, 3 \rangle)$

                  Figure   3.940   provides   an   example   where   the   NB_PLAIN
                  $(3, [2, 3, 6, 5, 7, 6, 6, 4, 5, 5, 4, 3, 3, 6, 6, 3])$ constraint holds.

**Typical**       $|VARIABLES| > 2$
                  range(VARIABLES.var) $> 1$

**Symmetries**    • Items of VARIABLES can be reversed.
                  • One and the same constant can be added to the var attribute of all items of
                    VARIABLES.

**Arg. properties**  Functional dependency: VALUE determined by VARIABLES.

Figure 3.940: Illustrating the NB_PLAIN constraint of the **Example** slot

**Automaton**       Figures 3.941 and 3.942 respectively depict the automaton associated with the constraint
                    NB_PLAIN and its simplified form.

Figure 3.941: Automaton for the NB_PLAIN constraint obtained by applying decoration
Table 2.35 to the seed transducer of the PLAIN pattern where default is 0

Figure 3.942: Automaton for the NB_PLAIN constraint obtained by applying decoration
Table 2.37 to the seed transducer of the PLAIN pattern where default is 0; $R_i - R_{i-1} \geq$
$0$ and $-R_i + R_{i-2} + 1 \geq 0$ are linear invariants.

|     | $s$ | $r$ | $t$ |
| --- | --- | --- | --- |
| $s$ | $\overrightarrow{C} + \overleftarrow{C}$ | $\overrightarrow{C} + \overleftarrow{C}$ | $\overrightarrow{C} + \overleftarrow{C}$ |
| $r$ | $\overrightarrow{C} + \overleftarrow{C}$ | 1 | 1 |
| $t$ | $\overrightarrow{C} + \overleftarrow{C}$ | 1 | 1 |

Table 3.131: Glue matrix for the NB_PLAIN constraint defined as the composition of the PLAIN pattern , the feature ONE , and the aggregator sum ; cells of the glue matrix are coloured with the colour of the constituent to which they are related.

## 3.398 NB_PLATEAU

▶ ▷ ◁ ◀          **DESCRIPTION**          **AUTOMATON**

**Origin**            Based on the PLATEAU pattern.

**Constraint**        NB_PLATEAU(VALUE, VARIABLES)

**Arguments**         VALUE      : dvar
                      VARIABLES  : collection(var−dvar)

**Restrictions**      $sv \leq 2 \vee rv \leq 1 \Rightarrow VALUE = 0$
                      $VALUE \geq 0$
                      $VALUE \leq \max(0, \lfloor (sv - 1)/2 \rfloor)$ ①
                      required(VARIABLES, var)
                      where
                      $sv = |VARIABLES|$
                      $rv = $ range(VARIABLES.var)

**Purpose**           VALUE is the number of occurrences of the PLATEAU pattern in the time-series given by
                      the VARIABLES collection. If the pattern does not occur, VALUE takes the default value
                      0.
                      An occurrence of the pattern PLATEAU is the *maximal* subsequence which matches the
                      regular expression $< =^* >$.

**Example**           $(3, \langle 7, 5, 2, 3, 1, 2, 2, 4, 3, 3, 4, 5, 5, 2, 2, 5 \rangle)$

                      Figure    3.943    provides    an    example    where    the    NB_PLATEAU
                      $(3, [7, 5, 2, 3, 1, 2, 2, 4, 3, 3, 4, 5, 5, 2, 2, 5])$ constraint holds.

**Typical**           $|VARIABLES| > 2$
                      range(VARIABLES.var) > 1

**Symmetries**        • Items of VARIABLES can be reversed.

                      • One and the same constant can be added to the var attribute of all items of
                        VARIABLES.

**Arg. properties**   Functional dependency: VALUE determined by VARIABLES.

Figure 3.943: Illustrating the NB_PLATEAU constraint of the **Example** slot

**Automaton**    Figures 3.944 and 3.945 respectively depict the automaton associated with the constraint
NB_PLATEAU and its simplified form.

Figure 3.944: Automaton for the NB_PLATEAU constraint obtained by applying decoration Table 2.35 to the seed transducer of the PLATEAU pattern where default is 0

Figure 3.945: Automaton for the NB_PLATEAU constraint obtained by applying decoration Table 2.37 to the seed transducer of the PLATEAU pattern where default is 0; $R_i - R_{i-1} \geq 0$ and $-R_i + R_{i-2} + 1 \geq 0$ are linear invariants.

|   | $s$ | $r$ | $t$ |
|---|---|---|---|
| $s$ | $\overrightarrow{C} + \overleftarrow{C}$ | $\overrightarrow{C} + \overleftarrow{C}$ | $\overrightarrow{C} + \overleftarrow{C}$ |
| $r$ | $\overrightarrow{C} + \overleftarrow{C}$ | 1 | 1 |
| $t$ | $\overrightarrow{C} + \overleftarrow{C}$ | 1 | 1 |

Table 3.132: Glue matrix for the NB_PLATEAU constraint defined as the composition of the PLATEAU pattern , the feature ONE , and the aggregator sum ; cells of the glue matrix are coloured with the colour of the constituent to which they are related.

## 3.399  NB_PROPER_PLAIN

**DESCRIPTION**          **AUTOMATON**

$> =^+ <$

**Origin**          Based on the PROPER_PLAIN pattern.

**Constraint**      NB_PROPER_PLAIN(VALUE, VARIABLES)

**Arguments**       VALUE     : dvar
                    VARIABLES : collection(var−dvar)

**Restrictions**    $\mathtt{sv} \leq 3 \vee \mathtt{rv} \leq 1 \Rightarrow \mathtt{VALUE} = 0$
                    $\mathtt{VALUE} \geq 0$
                    $\mathtt{VALUE} \leq \max(0, \lfloor (\mathtt{sv} - 1)/3 \rfloor)$①
                    required(VARIABLES, var)
                    where
                    $\mathtt{sv} = |\mathtt{VARIABLES}|$
                    $\mathtt{rv} = \text{range}(\mathtt{VARIABLES.var})$

VALUE ≤ **4**

**Purpose**         VALUE is the number of occurrences of the PROPER_PLAIN pattern in the time-series
                    given by the VARIABLES collection. If the pattern does not occur, VALUE takes the default
                    value 0.
                    An occurrence of the pattern PROPER_PLAIN is the *maximal* subsequence which matches
                    the regular expression $> =^+ <$.

**Example**         $(3, \langle 2, 7, 5, 5, 6, 3, 7, 4, 4, 5, 6, 5, 3, 3, 3, 5 \rangle)$

Figure 3.946 provides an example where the NB_PROPER_PLAIN
$(3, [2, 7, 5, 5, 6, 3, 7, 4, 4, 5, 6, 5, 3, 3, 3, 5])$ constraint holds.

**Typical**         $|\mathtt{VARIABLES}| > 3$
                    $\text{range}(\mathtt{VARIABLES.var}) > 1$

**Symmetries**      • Items of VARIABLES can be reversed.
                    • One and the same constant can be added to the var attribute of all items of
                      VARIABLES.

**Arg. properties** Functional dependency: VALUE determined by VARIABLES.

Figure 3.946: Illustrating the NB_PROPER_PLAIN constraint of the **Example** slot

**Automaton**      Figures 3.947 and 3.948 respectively depict the automaton associated with the constraint
NB_PROPER_PLAIN and its simplified form.

Figure 3.947: Automaton for the NB_PROPER_PLAIN constraint obtained by apply-
ing decoration Table 2.35 to the seed transducer of the PROPER_PLAIN pattern where
default is 0

Figure 3.948: Automaton for the NB_PROPER_PLAIN constraint obtained by apply-
ing decoration Table 2.37 to the seed transducer of the PROPER_PLAIN pattern where
default is 0; $R_i - R_{i-1} \geq 0$ and $-R_i + R_{i-3} + 1 \geq 0$ are linear invariants.

|   | $s$ | $r$ | $t$ |
|---|---|---|---|
| $s$ | $\overrightarrow{C}+\overleftarrow{C}$ | $\overrightarrow{C}+\overleftarrow{C}$ | $\overrightarrow{C}+\overleftarrow{C}$ |
| $r$ | $\overrightarrow{C}+\overleftarrow{C}$ | $\overrightarrow{C}+\overleftarrow{C}$ | 1 |
| $t$ | $\overrightarrow{C}+\overleftarrow{C}$ | 1 | 1 |

Table 3.133: Glue matrix for the NB_PROPER_PLAIN constraint defined as the composition of the PROPER_PLAIN pattern , the feature ONE , and the aggregator sum ; cells of the glue matrix are coloured with the colour of the constituent to which they are related.

## 3.400  NB_PROPER_PLATEAU

▶ ▷ ◁ ◀          DESCRIPTION          AUTOMATON

**Origin**          Based on the PROPER_PLATEAU pattern.

**Constraint**      NB_PROPER_PLATEAU(VALUE, VARIABLES)

**Arguments**       VALUE      : dvar
                    VARIABLES  : collection(var−dvar)

**Restrictions**    $\mathtt{sv} \leq 3 \lor \mathtt{rv} \leq 1 \Rightarrow \mathtt{VALUE} = 0$
                    $\mathtt{VALUE} \geq 0$
                    $\mathtt{VALUE} \leq \max(0, \lfloor (\mathtt{sv} - 1)/3 \rfloor)$ ①
                    required(VARIABLES, var)
                    where
                    $\mathtt{sv} = |\mathtt{VARIABLES}|$
                    $\mathtt{rv} = \text{range}(\mathtt{VARIABLES.var})$

**Purpose**         VALUE is the number of occurrences of the PROPER_PLATEAU pattern in the time-series
                    given by the VARIABLES collection. If the pattern does not occur, VALUE takes the default
                    value 0.
                    An occurrence of the pattern PROPER_PLATEAU is the *maximal* subsequence which
                    matches the regular expression $< =^+ >$.

**Example**         $(3, \langle 7, 1, 3, 3, 2, 5, 1, 4, 4, 3, 2, 3, 5, 5, 5, 3 \rangle)$

                    Figure 3.949 provides an example where the NB_PROPER_PLATEAU
                    $(3, [7, 1, 3, 3, 2, 5, 1, 4, 4, 3, 2, 3, 5, 5, 5, 3])$ constraint holds.

**Typical**         $|\mathtt{VARIABLES}| > 3$
                    $\text{range}(\mathtt{VARIABLES.var}) > 1$

**Symmetries**      • Items of VARIABLES can be reversed.
                    • One and the same constant can be added to the var attribute of all items of
                      VARIABLES.

**Arg. properties** Functional dependency: VALUE determined by VARIABLES.

Figure 3.949: Illustrating the NB_PROPER_PLATEAU constraint of the **Example** slot

**Automaton**     Figures 3.950 and 3.951 respectively depict the automaton associated with the constraint NB_PROPER_PLATEAU and its simplified form.

Figure 3.950: Automaton for the NB_PROPER_PLATEAU constraint obtained by applying decoration Table 2.35 to the seed transducer of the PROPER_PLATEAU pattern where default is 0

Figure 3.951: Automaton for the NB_PROPER_PLATEAU constraint obtained by applying decoration Table 2.37 to the seed transducer of the PROPER_PLATEAU pattern where default is 0; $R_i - R_{i-1} \geq 0$ and $-R_i + R_{i-3} + 1 \geq 0$ are linear invariants.

| | $s$ | $r$ | $t$ |
|---|---|---|---|
| $s$ | $\overrightarrow{C} + \overleftarrow{C}$ | $\overrightarrow{C} + \overleftarrow{C}$ | $\overrightarrow{C} + \overleftarrow{C}$ |
| $r$ | $\overrightarrow{C} + \overleftarrow{C}$ | $\overrightarrow{C} + \overleftarrow{C}$ | 1 |
| $t$ | $\overrightarrow{C} + \overleftarrow{C}$ | 1 | 1 |

Table 3.134: Glue matrix for the NB_PROPER_PLATEAU constraint defined as the composition of the PROPER_PLATEAU pattern , the feature ONE , and the aggregator sum ; cells of the glue matrix are coloured with the colour of the constituent to which they are related.

## 3.401  NB_STEADY

**DESCRIPTION**          **AUTOMATON**

| | |
|---|---|
| **Origin** | Based on the STEADY pattern. |
| **Constraint** | NB_STEADY(VALUE, VARIABLES) |
| **Arguments** | VALUE     : dvar <br> VARIABLES : collection(var−dvar) |

**Restrictions**

$\mathtt{sv} \leq 1 \Rightarrow \mathtt{VALUE} = 0$
$\mathtt{rv} = 1 \Rightarrow \mathtt{VALUE} \geq \mathtt{sv} - 1$
$\mathtt{rv} \geq 2 \Rightarrow \mathtt{VALUE} \geq 0$
$\mathtt{VALUE} \leq \max(0, \mathtt{sv} - 1)$①
required(VARIABLES, var)
where
$\mathtt{sv} = |\mathtt{VARIABLES}|$
$\mathtt{rv} = $range(VARIABLES.var)

**Purpose**

VALUE is the number of occurrences of the STEADY pattern in the time-series given by the VARIABLES collection. If the pattern does not occur, VALUE takes the default value 0.
An occurrence of the pattern STEADY is the subsequence which matches the regular expression =.

**Example**

$(7, \langle 1, 1, 7, 3, 3, 5, 5, 5, 6, 5, 5, 5, 7, 2, 6, 6 \rangle)$

Figure 3.952 provides an example where the NB_STEADY $(7, [1, 1, 7, 3, 3, 5, 5, 5, 6, 5, 5, 5, 7, 2, 6, 6])$ constraint holds.

| | |
|---|---|
| **Typical** | $|\mathtt{VARIABLES}| > 1$ |
| **Symmetries** | • Items of VARIABLES can be reversed. <br> • One and the same constant can be added to the var attribute of all items of VARIABLES. |
| **Arg. properties** | Functional dependency: VALUE determined by VARIABLES. |

Figure 3.952: Illustrating the NB_STEADY constraint of the **Example** slot

**Automaton**  Figures 3.953 and 3.954 respectively depict the automaton associated with the constraint NB_STEADY and its simplified form.

Figure 3.953: Automaton for the NB_STEADY constraint obtained by applying decoration Table 2.35 to the seed transducer of the STEADY pattern where default is 0

Figure 3.954: Automaton for the NB_STEADY constraint obtained by applying decoration Table 2.37 to the seed transducer of the STEADY pattern where default is 0; $R_i - R_{i-1} \geq 0$ and $-R_i + R_{i-1} + 1 \geq 0$ are linear invariants.

Table 3.135: Glue matrix for the NB_STEADY constraint defined as the composition of the STEADY pattern , the feature ONE , and the aggregator sum ; cells of the glue matrix are coloured with the colour of the constituent to which they are related.

## 3.402   NB_STEADY_SEQUENCE

► ▷ ◁ ◀          **DESCRIPTION**          **AUTOMATON**

| | |
|---|---|
| **Origin** | Based on the STEADY_SEQUENCE pattern. |
| **Constraint** | NB_STEADY_SEQUENCE(VALUE, VARIABLES) |
| **Arguments** | VALUE     : dvar<br>VARIABLES : collection(var−dvar) |

**Restrictions**

$sv \leq 1 \Rightarrow VALUE = 0$

$rv = 1 \land sv \geq 2 \Rightarrow VALUE \geq 1$

$rv \geq 2 \Rightarrow VALUE \geq 0$

$rv = 1 \land sv \geq 2 \Rightarrow VALUE \leq 1$

$rv \geq 2 \land sv \geq 2 \Rightarrow VALUE \leq \lfloor sv/2 \rfloor ①$

required(VARIABLES, var)

where

$sv = |VARIABLES|$

$rv = range(VARIABLES.var)$

**Purpose**

VALUE is the number of occurrences of the STEADY_SEQUENCE pattern in the time-series given by the VARIABLES collection. If the pattern does not occur, VALUE takes the default value 0.

An occurrence of the pattern STEADY_SEQUENCE is the *maximal* subsequence which matches the regular expression $=^+$.

**Example**

$$(5, \langle 3, 1, 1, 4, 5, 5, 5, 6, 2, 2, 4, 4, 3, 2, 1, 1 \rangle)$$

Figure 3.955 provides an example where the NB_STEADY_SEQUENCE $(5, [3, 1, 1, 4, 5, 5, 5, 6, 2, 2, 4, 4, 3, 2, 1, 1])$ constraint holds.

| | |
|---|---|
| **Typical** | $|VARIABLES| > 1$ |

**Symmetries**

- Items of VARIABLES can be reversed.
- One and the same constant can be added to the var attribute of all items of VARIABLES.

| | |
|---|---|
| **Arg. properties** | Functional dependency: VALUE determined by VARIABLES. |

Figure 3.955: Illustrating the NB_STEADY_SEQUENCE constraint of the **Example** slot

**Automaton**     Figures 3.956 and 3.957 respectively depict the automaton associated with the constraint
NB_STEADY_SEQUENCE and its simplified form.

Figure 3.956: Automaton for the NB_STEADY_SEQUENCE constraint obtained by applying decoration Table 2.35 to the seed transducer of the STEADY_SEQUENCE pattern where default is 0

Figure 3.957: Automaton for the NB_STEADY_SEQUENCE constraint obtained by applying decoration Table 2.37 to the seed transducer of the STEADY_SEQUENCE pattern where default is 0; $R_i - R_{i-1} \geq 0$ and $-R_i + R_{i-2} + 1 \geq 0$ are linear invariants.

|   | $s$ | $r$ |
|---|---|---|
| $s$ | $\overrightarrow{C} + \overleftarrow{C}$ | $\overrightarrow{C} + \overleftarrow{C}$ |
| $r$ | $\overrightarrow{C} + \overleftarrow{C}$ | $1$ |

Table 3.136: Glue matrix for the NB_STEADY_SEQUENCE constraint defined as the composition of the STEADY_SEQUENCE pattern, the feature ONE, and the aggregator sum; cells of the glue matrix are coloured with the colour of the constituent to which they are related.

# 3.403 NB_STRICTLY_DECREASING_SEQUENCE

     **DESCRIPTION**          **AUTOMATON**

**Origin**          Based on the STRICTLY_DECREASING_SEQUENCE pattern.

**Constraint**      NB_STRICTLY_DECREASING_SEQUENCE(VALUE, VARIABLES)

**Arguments**       VALUE     : dvar
                    VARIABLES : collection(var−dvar)

**Restrictions**    $\mathtt{sv} \leq 1 \vee \mathtt{rv} \leq 1 \Rightarrow \mathtt{VALUE} = 0$
                    $\mathtt{VALUE} \geq 0$
                    $\mathtt{VALUE} \leq \lfloor \mathtt{sv}/2 \rfloor$ ①
                    required(VARIABLES, var)
                    where
                      $\mathtt{sv} = |\mathtt{VARIABLES}|$
                      $\mathtt{rv} = \mathtt{range}(\mathtt{VARIABLES.var})$

**Purpose**         VALUE is the number of occurrences of the STRICTLY_DECREASING_SEQUENCE pattern
                    in the time-series given by the VARIABLES collection. If the pattern does not occur,
                    VALUE takes the default value 0.
                    An occurrence of the pattern STRICTLY_DECREASING_SEQUENCE is the *maximal* sub-
                    sequence which matches the regular expression $>^+$.

**Example**         $(3, \langle 4, 4, 6, 4, 1, 1, 3, 4, 4, 6, 6, 5, 2, 2, 4, 3 \rangle)$

                    Figure 3.958 provides an example where the NB_STRICTLY_DECREASING_SEQUENCE
                    $(3, [4, 4, 6, 4, 1, 1, 3, 4, 4, 6, 6, 5, 2, 2, 4, 3])$ constraint holds.

**Typical**         $|\mathtt{VARIABLES}| > 1$
                    $\mathtt{range}(\mathtt{VARIABLES.var}) > 1$

**Symmetry**        One and the same constant can be added to the var attribute of all items of VARIABLES.

**Arg. properties** Functional dependency: VALUE determined by VARIABLES.

Figure 3.958: Illustrating the NB_STRICTLY_DECREASING_SEQUENCE constraint of the **Example** slot

**Automaton**     Figures 3.959 and 3.960 respectively depict the automaton associated with the constraint NB_STRICTLY_DECREASING_SEQUENCE and its simplified form.

Figure 3.959: Automaton for the NB_STRICTLY_DECREASING_SEQUENCE constraint obtained by applying decoration Table 2.35 to the seed transducer of the STRICTLY_DECREASING_SEQUENCE pattern where default is 0

Figure 3.960: Automaton for the NB_STRICTLY_DECREASING_SEQUENCE constraint obtained by applying decoration Table 2.37 to the seed transducer of the STRICTLY_DECREASING_SEQUENCE pattern where default is 0; $R_i - R_{i-1} \geq 0$ and $-R_i + R_{i-2} + 1 \geq 0$ are linear invariants.

|       | $s$ | $r$ |
|-------|------------------------------------|------------------------------------|
| $s$   | $\overrightarrow{C} + \overleftarrow{C}$ | $\overrightarrow{C} + \overleftarrow{C}$ |
| $r$   | $\overrightarrow{C} + \overleftarrow{C}$ | $1$ |

Table 3.137: Glue matrix for the NB_STRICTLY_DECREASING_SEQUENCE constraint defined as the composition of the STRICTLY_DECREASING_SEQUENCE pattern , the feature ONE , and the aggregator sum ; cells of the glue matrix are coloured with the colour of the constituent to which they are related.

## 3.404   NB_STRICTLY_INCREASING_SEQUENCE

       DESCRIPTION          AUTOMATON

| | |
|---|---|
| **Origin** | Based on the STRICTLY_INCREASING_SEQUENCE pattern. |
| **Constraint** | NB_STRICTLY_INCREASING_SEQUENCE(VALUE, VARIABLES) |
| **Arguments** | VALUE      :  dvar <br> VARIABLES  :  collection(var−dvar) |
| **Restrictions** | $\mathtt{sv} \leq 1 \lor \mathtt{rv} \leq 1 \Rightarrow \mathtt{VALUE} = 0$ <br> $\mathtt{VALUE} \geq 0$ <br> $\mathtt{VALUE} \leq \lfloor \mathtt{sv}/2 \rfloor$ ① <br> required(VARIABLES, var) <br> where <br> $\mathtt{sv} = \mid\mathtt{VARIABLES}\mid$ <br> $\mathtt{rv} = \mathtt{range}(\mathtt{VARIABLES.var})$ |
| **Purpose** | VALUE is the number of occurrences of the STRICTLY_INCREASING_SEQUENCE pattern in the time-series given by the VARIABLES collection. If the pattern does not occur, VALUE takes the default value 0. <br> An occurrence of the pattern STRICTLY_INCREASING_SEQUENCE is the *maximal* sub-sequence which matches the regular expression $<^+$. |
| **Example** | $(3, \langle 4, 3, 5, 5, 2, 1, 1, 2, 3, 4, 6, 6, 3, 1, 2, 3 \rangle)$ |
| | Figure 3.961 provides an example where the NB_STRICTLY_INCREASING_SEQUENCE $(3, [4, 3, 5, 5, 2, 1, 1, 2, 3, 4, 6, 6, 3, 1, 2, 3])$ constraint holds. |
| **Typical** | $\mid\mathtt{VARIABLES}\mid > 1$ <br> $\mathtt{range}(\mathtt{VARIABLES.var}) > 1$ |
| **Symmetry** | One and the same constant can be added to the var attribute of all items of VARIABLES. |
| **Arg. properties** | Functional dependency: VALUE determined by VARIABLES. |

Figure 3.961: Illustrating the NB_STRICTLY_INCREASING_SEQUENCE constraint of the **Example** slot

**Automaton**         Figures 3.962 and 3.963 respectively depict the automaton associated with the constraint
NB_STRICTLY_INCREASING_SEQUENCE and its simplified form.

Figure 3.962: Automaton for the NB_STRICTLY_INCREASING_SEQUENCE constraint obtained by applying decoration Table 2.35 to the seed transducer of the STRICTLY_INCREASING_SEQUENCE pattern where default is 0

Figure 3.963: Automaton for the NB_STRICTLY_INCREASING_SEQUENCE constraint obtained by applying decoration Table 2.37 to the seed transducer of the STRICTLY_INCREASING_SEQUENCE pattern where default is 0; $R_i - R_{i-1} \geq 0$ and $-R_i + R_{i-2} + 1 \geq 0$ are linear invariants.

|   | $s$ | $r$ |
|---|---|---|
| $s$ | $\overrightarrow{C} + \overleftarrow{C}$ | $\overrightarrow{C} + \overleftarrow{C}$ |
| $r$ | $\overrightarrow{C} + \overleftarrow{C}$ | $1$ |

Table 3.138: Glue matrix for the NB_STRICTLY_INCREASING_SEQUENCE constraint defined as the composition of the STRICTLY_INCREASING_SEQUENCE pattern , the feature ONE , and the aggregator sum ; cells of the glue matrix are coloured with the colour of the constituent to which they are related.

# 3.405 NB_SUMMIT

► ▷ ◁ ◀   **DESCRIPTION**   **AUTOMATON**

$(< | < (= | <)^* <)(> | > (= | >)^* >)$

| Origin | Based on the SUMMIT pattern. |
|---|---|

| Constraint | NB_SUMMIT(VALUE, VARIABLES) |
|---|---|

| Arguments | VALUE      : dvar |
|---|---|
| | VARIABLES  : collection(var−dvar) |

**Restrictions**

$sv \leq 2 \lor rv \leq 1 \Rightarrow VALUE = 0$

$VALUE \geq 0$

$VALUE \leq \max(0, \lfloor (sv-1)/2 \rfloor)$ ①

required(VARIABLES, var)

where

$sv = |VARIABLES|$

$rv = range(VARIABLES.var)$

VALUE ≤ **6**

**Purpose**

VALUE is the number of occurrences of the SUMMIT pattern in the time-series given by the VARIABLES collection. If the pattern does not occur, VALUE takes the default value 0.

An occurrence of the pattern SUMMIT is the *maximal* subsequence which matches the regular expression $(< | < (= | <)^* <)(> | > (= | >)^* >)$.

**Example**

$(3, \langle 7, 1, 5, 4, 4, 3, 3, 4, 6, 6, 2, 3, 4, 2, 3, 1 \rangle)$

Figure 3.964 provides an example where the NB_SUMMIT $(3, [7, 1, 5, 4, 4, 3, 3, 4, 6, 6, 2, 3, 4, 2, 3, 1])$ constraint holds.

**Typical**

$|VARIABLES| > 2$

$range(VARIABLES.var) > 1$

**Symmetries**

- Items of VARIABLES can be reversed.
- One and the same constant can be added to the var attribute of all items of VARIABLES.

**Arg. properties**

Functional dependency: VALUE determined by VARIABLES.

Figure 3.964: Illustrating the NB_SUMMIT constraint of the **Example** slot

**Automaton**     Figures 3.965 and 3.966 respectively depict the automaton associated with the constraint NB_SUMMIT and its simplified form.

Figure 3.965: Automaton for the NB_SUMMIT constraint obtained by applying decoration Table 2.35 to the seed transducer of the SUMMIT pattern where default is 0 (transition $u \rightarrow r$ has the same accumulator update as transition $r \rightarrow u$)

Figure 3.966: Automaton for the NB_SUMMIT constraint obtained by applying decoration Table 2.37 to the seed transducer of the SUMMIT pattern where `default` is $0$; $R_i - R_{i-1} \geq 0$ and $-R_i + R_{i-2} + 1 \geq 0$ are linear invariants.

|   | $s$ | $r$ | $t$ | $u$ |
|---|---|---|---|---|
| $s$ | $\overrightarrow{C} + \overleftarrow{C}$ | $\overrightarrow{C} + \overleftarrow{C}$ | $\overrightarrow{C} + \overleftarrow{C}$ | $\overrightarrow{C} + \overleftarrow{C}$ |
| $r$ | $\overrightarrow{C} + \overleftarrow{C}$ | $1$ | $1$ | $\overrightarrow{C} + \overleftarrow{C}$ |
| $t$ | $\overrightarrow{C} + \overleftarrow{C}$ | $1$ | $\overrightarrow{C} + \overleftarrow{C}$ | $1$ |
| $u$ | $\overrightarrow{C} + \overleftarrow{C}$ | $\overrightarrow{C} + \overleftarrow{C}$ | $1$ | $\overrightarrow{C} + \overleftarrow{C}$ |

Table 3.139: Glue matrix for the NB_SUMMIT constraint defined as the composition of the SUMMIT pattern, the feature ONE, and the aggregator sum; cells of the glue matrix are coloured with the colour of the constituent to which they are related.

# 3.406 NB_VALLEY

     **DESCRIPTION**          **AUTOMATON**

**Origin**          Based on the VALLEY pattern.

**Constraint**      NB_VALLEY(VALUE, VARIABLES)

**Arguments**       VALUE      : dvar
                    VARIABLES  : collection(var−dvar)

**Restrictions**    $\mathtt{sv} \leq 2 \vee \mathtt{rv} \leq 1 \Rightarrow \mathtt{VALUE} = 0$
                    $\mathtt{VALUE} \geq 0$
                    $\mathtt{VALUE} \leq \max(0, \lfloor (\mathtt{sv} - 1)/2 \rfloor)$ ①
                    required(VARIABLES, var)
                    where
                    $\mathtt{sv} = |\mathtt{VARIABLES}|$
                    $\mathtt{rv} = \text{range}(\mathtt{VARIABLES.var})$

**Purpose**         VALUE is the number of occurrences of the VALLEY pattern in the time-series given by
                    the VARIABLES collection. If the pattern does not occur, VALUE takes the default value
                    0.
                    An occurrence of the pattern VALLEY is the *maximal* subsequence which matches the
                    regular expression $> (= | >)^* (< | =)^* <$.

**Example**         $(3, \langle 1, 3, 7, 4, 3, 6, 6, 5, 3, 3, 2, 6, 5, 5, 5, 7 \rangle)$

                    Figure    3.967    provides    an    example    where    the    NB_VALLEY
                    $(3, [1, 3, 7, 4, 3, 6, 6, 5, 3, 3, 2, 6, 5, 5, 5, 7])$ constraint holds.

**Typical**         $|\mathtt{VARIABLES}| > 2$
                    range(VARIABLES.var) > 1

**Symmetries**      • Items of VARIABLES can be reversed.

                    • One and the same constant can be added to the var attribute of all items of
                      VARIABLES.

**Arg. properties** Functional dependency: VALUE determined by VARIABLES.

Figure 3.967: Illustrating the NB_VALLEY constraint of the **Example** slot

**Automaton**  Figures 3.968 and 3.969 respectively depict the automaton associated with the constraint NB_VALLEY and its simplified form.

Figure 3.968: Automaton for the NB_VALLEY constraint obtained by applying decoration Table 2.35 to the seed transducer of the VALLEY pattern where default is 0

Figure 3.969: Automaton for the NB_VALLEY constraint obtained by applying decoration Table 2.37 to the seed transducer of the VALLEY pattern where default is 0; $R_i - R_{i-1} \geq 0$ and $-R_i + R_{i-2} + 1 \geq 0$ are linear invariants.

|   | $s$ | $r$ | $t$ |
|---|---|---|---|
| $s$ | $\overrightarrow{C} + \overleftarrow{C}$ | $\overrightarrow{C} + \overleftarrow{C}$ | $\overrightarrow{C} + \overleftarrow{C}$ |
| $r$ | $\overrightarrow{C} + \overleftarrow{C}$ | 1 | 1 |
| $t$ | $\overrightarrow{C} + \overleftarrow{C}$ | 1 | $\overrightarrow{C} + \overleftarrow{C}$ |

Table 3.140: Glue matrix for the NB_VALLEY constraint defined as the composition of the VALLEY pattern, the feature ONE, and the aggregator sum; cells of the glue matrix are coloured with the colour of the constituent to which they are related.

# 3.407 NB_ZIGZAG

    DESCRIPTION    AUTOMATON

| | |
|---|---|
| **Origin** | Based on the ZIGZAG pattern. |
| **Constraint** | NB_ZIGZAG(VALUE, VARIABLES) |

| | | |
|---|---|---|
| **Arguments** | VALUE | : dvar |
| | VARIABLES | : collection(var−dvar) |

**Restrictions**

$$sv \leq 3 \vee rv \leq 1 \Rightarrow VALUE = 0$$
$$VALUE \geq 0$$
$$rv = 2 \Rightarrow VALUE \leq \lfloor sv/4 \rfloor \ ①$$
$$rv \geq 3 \Rightarrow VALUE \leq \max(0, \lfloor (sv-1)/3 \rfloor) \ ②$$
required(VARIABLES, var)
where
$$sv = |VARIABLES|$$
$$rv = range(VARIABLES.var)$$

**Purpose**

VALUE is the number of occurrences of the ZIGZAG pattern in the time-series given by the VARIABLES collection. If the pattern does not occur, VALUE takes the default value 0.

An occurrence of the pattern ZIGZAG is the *maximal* subsequence which matches the regular expression $(<>)^+(< | <>) | (><)^+(> | ><)$.

**Example**

$$(3, \langle 4, 1, 3, 1, 4, 6, 1, 5, 5, 2, 7, 2, 3, 1, 6, 1 \rangle)$$

Figure 3.970 provides an example where the NB_ZIGZAG $(3, [4, 1, 3, 1, 4, 6, 1, 5, 5, 2, 7, 2, 3, 1, 6, 1])$ constraint holds.

| | |
|---|---|
| **Typical** | $|VARIABLES| > 3$ |
| | $range(VARIABLES.var) > 1$ |

**Symmetries**

- Items of VARIABLES can be reversed.
- One and the same constant can be added to the var attribute of all items of VARIABLES.

| | |
|---|---|
| **Arg. properties** | Functional dependency: VALUE determined by VARIABLES. |

Figure 3.970: Illustrating the NB_ZIGZAG constraint of the **Example** slot

**Automaton**     Figures 3.971 and 3.972 respectively depict the automaton associated with the constraint NB_ZIGZAG and its simplified form.

|   | $s$ | $a$ | $b$ | $c$ | $d$ | $e$ | $f$ |
|---|---|---|---|---|---|---|---|
| $s$ | $\overrightarrow{C} + \overleftarrow{C}$ | $\overrightarrow{C} + \overleftarrow{C}$ | $\overrightarrow{C} + \overleftarrow{C}$ | $\overrightarrow{C} + \overleftarrow{C}$ | $\overrightarrow{C} + \overleftarrow{C}$ | $\overrightarrow{C} + \overleftarrow{C}$ | $\overrightarrow{C} + \overleftarrow{C}$ |
| $a$ | $\overrightarrow{C} + \overleftarrow{C}$ | $\overrightarrow{C} + \overleftarrow{C}$ | $\overrightarrow{C} + \overleftarrow{C}$ | $1$ | $\overrightarrow{C} + \overleftarrow{C}$ | $1$ | $\overrightarrow{C} + \overleftarrow{C}$ |
| $b$ | $\overrightarrow{C} + \overleftarrow{C}$ | $\overrightarrow{C} + \overleftarrow{C}$ | $1$ | $\overrightarrow{C} + \overleftarrow{C}$ | $1$ | $\overrightarrow{C} + \overleftarrow{C}$ | $1$ |
| $c$ | $\overrightarrow{C} + \overleftarrow{C}$ | $1$ | $\overrightarrow{C} + \overleftarrow{C}$ | $1$ | $\overrightarrow{C} + \overleftarrow{C}$ | $1$ | $\overrightarrow{C} + \overleftarrow{C}$ |
| $d$ | $\overrightarrow{C} + \overleftarrow{C}$ | $\overrightarrow{C} + \overleftarrow{C}$ | $1$ | $\overrightarrow{C} + \overleftarrow{C}$ | $\overrightarrow{C} + \overleftarrow{C}$ | $\overrightarrow{C} + \overleftarrow{C}$ | $1$ |
| $e$ | $\overrightarrow{C} + \overleftarrow{C}$ | $1$ | $\overrightarrow{C} + \overleftarrow{C}$ | $1$ | $\overrightarrow{C} + \overleftarrow{C}$ | $\overrightarrow{C} + \overleftarrow{C}$ | $\overrightarrow{C} + \overleftarrow{C}$ |
| $f$ | $\overrightarrow{C} + \overleftarrow{C}$ | $\overrightarrow{C} + \overleftarrow{C}$ | $1$ | $\overrightarrow{C} + \overleftarrow{C}$ | $1$ | $\overrightarrow{C} + \overleftarrow{C}$ | $1$ |

Table 3.141: Glue matrix for the NB_ZIGZAG constraint defined as the composition of the ZIGZAG pattern , the feature ONE , and the aggregator sum ; cells of the glue matrix are coloured with the colour of the constituent to which they are related.

Figure 3.971: Automaton for the NB_ZIGZAG constraint obtained by applying decoration Table 2.35 to the seed transducer of the ZIGZAG pattern where default is 0; (1) missing transitions from $a, b, c, d, e, f$ to $s$ are labelled by $=$; (2) on transitions from $b, c, e, f$ to $s$ the accumulator $D$ is reset to its initial value; (3) on transitions from $c, f$ to $s$ the accumulator $R$ is updated wrt $C$ and the accumulator $C$ is reset to its initial value

Figure 3.972: Automaton for the NB_ZIGZAG constraint obtained by applying decoration Table 2.37 to the seed transducer of the ZIGZAG pattern where default is $0$; missing transitions from $a, b, c, d, e, f$ to $s$ are labelled by $=$; $R_i - R_{i-1} \geq 0$ and $-R_i + R_{i-3} + 1 \geq 0$ are linear invariants.

# 3.408   POS_MAX_HEIGHT_DECREASING_TERRACE

► ▷ ◁ ◄        DESCRIPTION            AUTOMATON

| | |
|---|---|
| **Origin** | Based on constraint MAX_HEIGHT_DECREASING_TERRACE. |
| **Constraint** | POS_MAX_HEIGHT_DECREASING_TERRACE(VALUE, VARIABLES, FOUND) |
| **Arguments** | VALUE     : dvar<br>VARIABLES : collection(var−dvar)<br>FOUND     : collection(var−dvar) |
| **Restrictions** | $sv \leq 3 \vee rv \leq 2 \Rightarrow VALUE = -\infty$<br>$VALUE = -\infty \vee VALUE \geq minv + 1$<br>$VALUE \leq maxv - 1$①<br>required(VARIABLES, var)<br>required(FOUND, var)<br>where<br>$maxv = $maxval(VARIABLES.var)<br>$minv = $minval(VARIABLES.var)<br>$sv = |VARIABLES|$<br>$rv = $range(VARIABLES.var) |

| | |
|---|---|
| **Purpose** | The constraint MAX_HEIGHT_DECREASING_TERRACE(VALUE, VARIABLES) holds. In addition, FOUND is a collection of 0/1 variables where the value 1 indicates the position of the **found** letter in those occurrences of the pattern DECREASING_TERRACE for which the feature value is VALUE.<br><br>The position of the **found** letter in an occurrence of a pattern is the first position where the occurrence of pattern is identified, even if the pattern is not complete.<br><br>An occurrence of the pattern DECREASING_TERRACE is the *maximal* subsequence which matches the regular expression $> =^+ >$.<br><br>Assume that the occurrence of the pattern DECREASING_TERRACE starts at position $i$ and ends at position $j$. The feature MIN computes the minimum of the values from index $i + 1$ to index $j$. |
| **Example** | $\left( \begin{array}{l} 4, \langle 6, 4, 4, 4, 5, 2, 2, 1, 3, 3, 5, 4, 4, 3, 3, 3 \rangle, \\ \langle 0, 0, 0, 0, 0, 0, 0, 0, 0, 0, 0, 0, 0, 1, 0, 0, 0 \rangle \end{array} \right)$ |

Figure 3.973 provides an example where the POS_MAX_HEIGHT_DECREASING_TERRACE (4, [6, 4, 4, 4, 5, 2, 2, 1, 3, 3, 5, 4, 4, 3, 3, 3], [0, 0, 0, 0, 0, 0, 0, 0, 0, 0, 0, 0, 0, 1, 0, 0, 0]) constraint holds.

| | |
|---|---|
| **Typical** | $|VARIABLES| > 3$<br>range(VARIABLES.var) $> 2$ |

Figure 3.973: Illustrating the POS_MAX_HEIGHT_DECREASING_TERRACE constraint of the **Example** slot

**Arg. properties**

- Functional dependency: VALUE determined by VARIABLES.
- Functional dependency: FOUND determined by VARIABLES.

**Automaton**    Similar to the automaton of the MAX_HEIGHT_DECREASING_TERRACE constraint but use the decoration table 2.33.

# 3.409 POS_MAX_HEIGHT_INCREASING_TERRACE

▶ ▷ ◁ ◀    **DESCRIPTION**    **AUTOMATON**

| | |
|---|---|
| **Origin** | Based on constraint MAX_HEIGHT_INCREASING_TERRACE. |
| **Constraint** | POS_MAX_HEIGHT_INCREASING_TERRACE(VALUE, VARIABLES, FOUND) |

**Arguments**

```
VALUE     : dvar
VARIABLES : collection(var−dvar)
FOUND     : collection(var−dvar)
```

**Restrictions**

$\mathtt{sv} \leq 3 \lor \mathtt{rv} \leq 2 \Rightarrow \mathtt{VALUE} = -\infty$

$\mathtt{VALUE} = -\infty \lor \mathtt{VALUE} \geq \mathtt{minv} + 1$

$\mathtt{VALUE} \leq \mathtt{maxv} - 1$①

required(VARIABLES, var)

required(FOUND, var)

where

$\mathtt{maxv} = \mathtt{maxval}(\mathtt{VARIABLES.var})$

$\mathtt{minv} = \mathtt{minval}(\mathtt{VARIABLES.var})$

$\mathtt{sv} = |\mathtt{VARIABLES}|$

$\mathtt{rv} = \mathtt{range}(\mathtt{VARIABLES.var})$

**Purpose**

The constraint MAX_HEIGHT_INCREASING_TERRACE(VALUE, VARIABLES) holds. In addition, FOUND is a collection of 0/1 variables where the value 1 indicates the position of the **found** letter in those occurrences of the pattern INCREASING_TERRACE for which the feature value is VALUE.

The position of the **found** letter in an occurrence of a pattern is the first position where the occurrence of pattern is identified, even if the pattern is not complete.

An occurrence of the pattern INCREASING_TERRACE is the *maximal* subsequence which matches the regular expression $<=^+<$.

Assume that the occurrence of the pattern INCREASING_TERRACE starts at position $i$ and ends at position $j$. The feature MIN computes the minimum of the values from index $i+1$ to index $j$.

**Example**

$$\left( \begin{array}{l} 5, \langle 1, 3, 3, 3, 2, 5, 5, 6, 4, 4, 2, 3, 3, 3, 4, 4 \rangle, \\ \langle 0, 0, 0, 0, 0, 0, 1, 0, 0, 0, 0, 0, 0, 0, 0, 0 \rangle \end{array} \right)$$

Figure 3.974 provides an example where the POS_MAX_HEIGHT_INCREASING_TERRACE (5, [1, 3, 3, 3, 2, 5, 5, 6, 4, 4, 2, 3, 3, 3, 4, 4], [0, 0, 0, 0, 0, 0, 1, 0, 0, 0, 0, 0, 0, 0, 0, 0]) constraint holds.

**Typical**

$|\mathtt{VARIABLES}| > 3$

$\mathtt{range}(\mathtt{VARIABLES.var}) > 2$

Figure 3.974: Illustrating the POS_MAX_HEIGHT_INCREASING_TERRACE constraint of the **Example** slot

**Arg. properties**
- Functional dependency: VALUE determined by VARIABLES.
- Functional dependency: FOUND determined by VARIABLES.

**Automaton**          Similar to the automaton of the MAX_HEIGHT_INCREASING_TERRACE constraint but use
                       the decoration table 2.33.

## 3.410 POS_MAX_HEIGHT_PLAIN

▶ ▷ ◁ ◀        **DESCRIPTION**        **AUTOMATON**

| | |
|---|---|
| **Origin** | Based on constraint MAX_HEIGHT_PLAIN. |
| **Constraint** | POS_MAX_HEIGHT_PLAIN(VALUE, VARIABLES, FOUND) |
| **Arguments** | VALUE      : dvar<br>VARIABLES : collection(var−dvar)<br>FOUND      : collection(var−dvar) |
| **Restrictions** | $\mathtt{sv} \leq 2 \vee \mathtt{rv} \leq 1 \Rightarrow \mathtt{VALUE} = -\infty$<br>$\mathtt{VALUE} = -\infty \vee \mathtt{VALUE} \geq \mathtt{minv}$<br>$\mathtt{VALUE} \leq \mathtt{maxv} - 1$①<br>required(VARIABLES, var)<br>required(FOUND, var)<br>where<br>    maxv = maxval(VARIABLES.var)<br>    minv = minval(VARIABLES.var)<br>    sv = \|VARIABLES\|<br>    rv = range(VARIABLES.var) |
| **Purpose** | The constraint MAX_HEIGHT_PLAIN(VALUE, VARIABLES) holds. In addition, FOUND is a collection of 0/1 variables where the value 1 indicates the position of the **found** letter in those occurrences of the pattern PLAIN for which the feature value is VALUE.<br>The position of the **found** letter in an occurrence of a pattern is the first position where the occurrence of pattern is identified, even if the pattern is not complete.<br>An occurrence of the pattern PLAIN is the *maximal* subsequence which matches the regular expression $>=^* <$.<br>Assume that the occurrence of the pattern PLAIN starts at position $i$ and ends at position $j$. The feature MIN computes the minimum of the values from index $i+1$ to index $j$. |
| **Example** | $\left( \begin{array}{l} 5, \langle 2, 3, 6, 5, 7, 6, 6, 4, 5, 5, 4, 3, 3, 6, 6, 3 \rangle, \\ \langle 0, 0, 0, 1, 0, 0, 0, 0, 0, 0, 0, 0, 0, 0, 0, 0 \rangle \end{array} \right)$ |

Figure 3.975 provides an example where the POS_MAX_HEIGHT_PLAIN $(5, [2, 3, 6, 5, 7, 6, 6, 4, 5, 5, 4, 3, 3, 6, 6, 3], [0, 0, 0, 1, 0, 0, 0, 0, 0, 0, 0, 0, 0, 0, 0, 0])$ constraint holds.

| | |
|---|---|
| **Typical** | $\|VARIABLES\| > 2$<br>range(VARIABLES.var) > 1 |
| **Arg. properties** | • Functional dependency: VALUE determined by VARIABLES.<br>• Functional dependency: FOUND determined by VARIABLES. |

Figure 3.975: Illustrating the POS_MAX_HEIGHT_PLAIN constraint of the **Example** slot

**Automaton**     Similar to the automaton of the MAX_HEIGHT_PLAIN constraint but use the decoration table 2.33.

## 3.411 POS_MAX_HEIGHT_PLATEAU

      DESCRIPTION          AUTOMATON

| | |
|---|---|
| **Origin** | Based on constraint MAX_HEIGHT_PLATEAU. |
| **Constraint** | POS_MAX_HEIGHT_PLATEAU(VALUE, VARIABLES, FOUND) |

**Arguments**

```
VALUE      : dvar
VARIABLES  : collection(var−dvar)
FOUND      : collection(var−dvar)
```

**Restrictions**

$\text{sv} \leq 2 \lor \text{rv} \leq 1 \Rightarrow \text{VALUE} = -\infty$

$\text{VALUE} = -\infty \lor \text{VALUE} \geq \text{minv} + 1$

$\text{VALUE} \leq \text{maxv}①$

required(VARIABLES, var)

required(FOUND, var)

where

maxv = maxval(VARIABLES.var)

minv = minval(VARIABLES.var)

$\text{sv} = |\text{VARIABLES}|$

rv = range(VARIABLES.var)

**Purpose**

The constraint MAX_HEIGHT_PLATEAU(VALUE, VARIABLES) holds. In addition, FOUND is a collection of 0/1 variables where the value 1 indicates the position of the **found** letter in those occurrences of the pattern PLATEAU for which the feature value is VALUE. The position of the **found** letter in an occurrence of a pattern is the first position where the occurrence of pattern is identified, even if the pattern is not complete.

An occurrence of the pattern PLATEAU is the *maximal* subsequence which matches the regular expression $<=^*>$.

Assume that the occurrence of the pattern PLATEAU starts at position $i$ and ends at position $j$. The feature MIN computes the minimum of the values from index $i + 1$ to index $j$.

**Example**

$$\left( \begin{array}{l} 5, \langle 7, 5, 2, 3, 1, 2, 2, 4, 3, 3, 4, 5, 5, 2, 2, 5 \rangle, \\ \langle 0, 0, 0, 0, 0, 0, 0, 0, 0, 0, 0, 0, 0, 1, 0, 0, 0 \rangle \end{array} \right)$$

Figure 3.976 provides an example where the POS_MAX_HEIGHT_PLATEAU $(5, [7, 5, 2, 3, 1, 2, 2, 4, 3, 3, 4, 5, 5, 2, 2, 5], [0, 0, 0, 0, 0, 0, 0, 0, 0, 0, 0, 0, 0, 1, 0, 0, 0])$ constraint holds.

**Typical**

$|\text{VARIABLES}| > 2$

range(VARIABLES.var) > 1

Figure 3.976: Illustrating the POS_MAX_HEIGHT_PLATEAU constraint of the **Example** slot

**Arg. properties**

- Functional dependency: VALUE determined by VARIABLES.
- Functional dependency: FOUND determined by VARIABLES.

**Automaton**      Similar to the automaton of the MAX_HEIGHT_PLATEAU constraint but use the decoration table 2.33.

## 3.412  POS_MAX_HEIGHT_PROPER_PLAIN

▶ ▷ ◁ ◀     **DESCRIPTION**          **AUTOMATON**

| | |
|---|---|
| **Origin** | Based on constraint MAX_HEIGHT_PROPER_PLAIN. |
| **Constraint** | POS_MAX_HEIGHT_PROPER_PLAIN(VALUE, VARIABLES, FOUND) |

**Arguments**

| | | |
|---|---|---|
| VALUE | : | dvar |
| VARIABLES | : | collection(var−dvar) |
| FOUND | : | collection(var−dvar) |

**Restrictions**

$sv \le 3 \vee rv \le 1 \Rightarrow \text{VALUE} = -\infty$

$\text{VALUE} = -\infty \vee \text{VALUE} \ge minv$

$\text{VALUE} \le maxv - 1 \text{①}$

required(VARIABLES, var)

required(FOUND, var)

where

$maxv = \text{maxval}(\text{VARIABLES.var})$

$minv = \text{minval}(\text{VARIABLES.var})$

$sv = |\text{VARIABLES}|$

$rv = \text{range}(\text{VARIABLES.var})$

**Purpose**

The constraint MAX_HEIGHT_PROPER_PLAIN(VALUE, VARIABLES) holds. In addition, FOUND is a collection of 0/1 variables where the value 1 indicates the position of the **found** letter in those occurrences of the pattern PROPER_PLAIN for which the feature value is VALUE.

The position of the **found** letter in an occurrence of a pattern is the first position where the occurrence of pattern is identified, even if the pattern is not complete.

An occurrence of the pattern PROPER_PLAIN is the *maximal* subsequence which matches the regular expression $> =^+ <$.

Assume that the occurrence of the pattern PROPER_PLAIN starts at position $i$ and ends at position $j$. The feature MIN computes the minimum of the values from index $i + 1$ to index $j$.

**Example**

$$\left( \begin{array}{l} 5, \langle 2, 7, 5, 5, 6, 3, 7, 4, 4, 5, 6, 5, 3, 3, 3, 5 \rangle, \\ \langle 0, 0, 0, 1, 0, 0, 0, 0, 0, 0, 0, 0, 0, 0, 0, 0 \rangle \end{array} \right)$$

Figure 3.977 provides an example where the POS_MAX_HEIGHT_PROPER_PLAIN (5, [2, 7, 5, 5, 6, 3, 7, 4, 4, 5, 6, 5, 3, 3, 3, 5], [0, 0, 0, 1, 0, 0, 0, 0, 0, 0, 0, 0, 0, 0, 0, 0]) constraint holds.

**Typical**

|VARIABLES| > 3

range(VARIABLES.var) > 1

Figure 3.977: Illustrating the POS_MAX_HEIGHT_PROPER_PLAIN constraint of the **Example** slot

**Arg. properties**
- Functional dependency: VALUE determined by VARIABLES.
- Functional dependency: FOUND determined by VARIABLES.

**Automaton**     Similar to the automaton of the MAX_HEIGHT_PROPER_PLAIN constraint but use the decoration table 2.33.

# 3.413 POS_MAX_HEIGHT_PROPER_PLATEAU

▶ ▷ ◁ ◀        DESCRIPTION              AUTOMATON

| | |
|---|---|
| **Origin** | Based on constraint MAX_HEIGHT_PROPER_PLATEAU. |
| **Constraint** | POS_MAX_HEIGHT_PROPER_PLATEAU(VALUE, VARIABLES, FOUND) |

**Arguments**

```
VALUE     : dvar
VARIABLES : collection(var-dvar)
FOUND     : collection(var-dvar)
```

**Restrictions**

$sv \leq 3 \lor rv \leq 1 \Rightarrow$ VALUE $= -\infty$
VALUE $= -\infty \lor$ VALUE $\geq$ minv $+ 1$
VALUE $\leq$ maxv①
required(VARIABLES, var)
required(FOUND, var)
where
maxv $=$ maxval(VARIABLES.var)
minv $=$ minval(VARIABLES.var)
sv $= |$VARIABLES$|$
rv $=$ range(VARIABLES.var)

**Purpose**

The constraint MAX_HEIGHT_PROPER_PLATEAU(VALUE, VARIABLES) holds. In addition, FOUND is a collection of 0/1 variables where the value 1 indicates the position of the **found** letter in those occurrences of the pattern PROPER_PLATEAU for which the feature value is VALUE.

The position of the **found** letter in an occurrence of a pattern is the first position where the occurrence of pattern is identified, even if the pattern is not complete.

An occurrence of the pattern PROPER_PLATEAU is the *maximal* subsequence which matches the regular expression $<=^+ >$.

Assume that the occurrence of the pattern PROPER_PLATEAU starts at position $i$ and ends at position $j$. The feature MIN computes the minimum of the values from index $i + 1$ to index $j$.

**Example**

$$\left( \begin{array}{l} 5, \langle 7, 1, 3, 3, 2, 5, 1, 4, 4, 3, 2, 3, 5, 5, 5, 3 \rangle, \\ \langle 0, 0, 0, 0, 0, 0, 0, 0, 0, 0, 0, 0, 0, 0, 1, 0 \rangle \end{array} \right)$$

Figure 3.978 provides an example where the POS_MAX_HEIGHT_PROPER_PLATEAU $(5, [7, 1, 3, 3, 2, 5, 1, 4, 4, 3, 2, 3, 5, 5, 5, 3], [0, 0, 0, 0, 0, 0, 0, 0, 0, 0, 0, 0, 0, 0, 1, 0])$ constraint holds.

**Typical**

$|$VARIABLES$| > 3$
range(VARIABLES.var) $> 1$

Figure 3.978: Illustrating the POS_MAX_HEIGHT_PROPER_PLATEAU constraint of the **Example** slot

| Arg. properties | • Functional dependency: VALUE determined by VARIABLES. |
|---|---|
|  | • Functional dependency: FOUND determined by VARIABLES. |

**Automaton**        Similar to the automaton of the MAX_HEIGHT_PROPER_PLATEAU constraint but use the
                     decoration table 2.33.

## 3.414 POS_MAX_HEIGHT_STEADY

          **DESCRIPTION**                **AUTOMATON**

| | |
|---|---|
| **Origin** | Based on constraint MAX_HEIGHT_STEADY. |
| **Constraint** | POS_MAX_HEIGHT_STEADY(VALUE, VARIABLES, FOUND) |
| **Arguments** | VALUE     : dvar |
| | VARIABLES : collection(var−dvar) |
| | FOUND     : collection(var−dvar) |

**Restrictions**

$\mathtt{sv} \leq 1 \Rightarrow \mathtt{VALUE} = -\infty$

$\mathtt{VALUE} = -\infty \vee \mathtt{VALUE} \geq \mathtt{minv}$

$\mathtt{VALUE} \leq \mathtt{maxv}①$

required(VARIABLES, var)

required(FOUND, var)

where

maxv = maxval(VARIABLES.var)

minv = minval(VARIABLES.var)

sv = |VARIABLES|

**Purpose**

The constraint MAX_HEIGHT_STEADY(VALUE, VARIABLES) holds. In addition, FOUND is a collection of 0/1 variables where the value 1 indicates the position of the **found** letter in those occurrences of the pattern STEADY for which the feature value is VALUE.

The position of the **found** letter in an occurrence of a pattern is the first position where the occurrence of pattern is identified, even if the pattern is not complete.

An occurrence of the pattern STEADY is the subsequence which matches the regular expression =.

Assume that the occurrence of the pattern STEADY starts at position $i$ and ends at position $j$. The feature MIN computes the minimum of the values from index $i$ to index $j + 1$.

**Example**

$$\left( \begin{array}{l} 6, \langle 1, 1, 7, 3, 3, 5, 5, 5, 6, 5, 5, 5, 7, 2, 6, 6 \rangle, \\ \langle 0, 0, 0, 0, 0, 0, 0, 0, 0, 0, 0, 0, 0, 0, 1, 0 \rangle \end{array} \right)$$

Figure 3.979 provides an example where the POS_MAX_HEIGHT_STEADY (6, [1, 1, 7, 3, 3, 5, 5, 5, 6, 5, 5, 5, 7, 2, 6, 6], [0, 0, 0, 0, 0, 0, 0, 0, 0, 0, 0, 0, 0, 0, 1, 0]) constraint holds.

| | |
|---|---|
| **Typical** | |VARIABLES| > 1 |

| | |
|---|---|
| **Arg. properties** | • Functional dependency: VALUE determined by VARIABLES. |
| | • Functional dependency: FOUND determined by VARIABLES. |

Figure 3.979: Illustrating the POS_MAX_HEIGHT_STEADY constraint of the **Example** slot

**Automaton**        Similar to the automaton of the MAX_HEIGHT_STEADY constraint but use the decoration table 2.33.

## 3.415 POS_MAX_HEIGHT_STEADY_SEQUENCE

▶ ▷ ◁ ◀      **DESCRIPTION**          **AUTOMATON**

| | |
|---|---|
| **Origin** | Based on constraint MAX_HEIGHT_STEADY_SEQUENCE. |
| **Constraint** | POS_MAX_HEIGHT_STEADY_SEQUENCE(VALUE, VARIABLES, FOUND) |
| **Arguments** | VALUE      : dvar |
| | VARIABLES  : collection(var−dvar) |
| | FOUND      : collection(var−dvar) |

**Restrictions**

$$sv \leq 1 \Rightarrow \text{VALUE} = -\infty$$
$$\text{VALUE} = -\infty \lor \text{VALUE} \geq minv$$
$$\text{VALUE} \leq maxv①$$
required(VARIABLES, var)
required(FOUND, var)
where
$$maxv = \text{maxval}(\text{VARIABLES.var})$$
$$minv = \text{minval}(\text{VARIABLES.var})$$
$$sv = |\text{VARIABLES}|$$

**Purpose**

The constraint MAX_HEIGHT_STEADY_SEQUENCE(VALUE, VARIABLES) holds. In addition, FOUND is a collection of 0/1 variables where the value 1 indicates the position of the **found** letter in those occurrences of the pattern STEADY_SEQUENCE for which the feature value is VALUE.

The position of the **found** letter in an occurrence of a pattern is the first position where the occurrence of pattern is identified, even if the pattern is not complete.

An occurrence of the pattern STEADY_SEQUENCE is the *maximal* subsequence which matches the regular expression $=^+$.

Assume that the occurrence of the pattern STEADY_SEQUENCE starts at position $i$ and ends at position $j$. The feature MIN computes the minimum of the values from index $i$ to index $j + 1$.

**Example**

$$\left( \begin{array}{l} 5, \langle 3, 1, 1, 4, 5, 5, 5, 6, 2, 2, 4, 4, 3, 2, 1, 1 \rangle, \\ \langle 0, 0, 0, 0, 1, 0, 0, 0, 0, 0, 0, 0, 0, 0, 0, 0 \rangle \end{array} \right)$$

Figure 3.980 provides an example where the POS_MAX_HEIGHT_STEADY_SEQUENCE $(5, [3, 1, 1, 4, 5, 5, 5, 6, 2, 2, 4, 4, 3, 2, 1, 1], [0, 0, 0, 0, 1, 0, 0, 0, 0, 0, 0, 0, 0, 0, 0, 0])$ constraint holds.

**Typical**

$$|\text{VARIABLES}| > 1$$

**Arg. properties**

- Functional dependency: VALUE determined by VARIABLES.
- Functional dependency: FOUND determined by VARIABLES.

Figure 3.980: Illustrating the POS_MAX_HEIGHT_STEADY_SEQUENCE constraint of the **Example** slot

**Automaton**    Similar to the automaton of the MAX_HEIGHT_STEADY_SEQUENCE constraint but use the decoration table 2.33.

# 3.416 POS_MAX_MAX_BUMP_ON_DECREASING_SEQUENCE

► ▷ ◁ ◀    DESCRIPTION    AUTOMATON

**Origin**  Based on constraint MAX_MAX_BUMP_ON_DECREASING_SEQUENCE.

**Constraint**  POS_MAX_MAX_BUMP_ON_DECREASING_SEQUENCE(VALUE, VARIABLES, FOUND)

**Arguments**

```
VALUE     : dvar
VARIABLES : collection(var-dvar)
FOUND     : collection(var-dvar)
```

**Restrictions**

$sv \leq 5 \vee rv \leq 2 \Rightarrow VALUE = -\infty$

$VALUE = -\infty \vee VALUE \geq minv + 2$

$VALUE \leq maxv①$

required(VARIABLES, var)

required(FOUND, var)

where

maxv = maxval(VARIABLES.var)

minv = minval(VARIABLES.var)

sv = |VARIABLES|

rv = range(VARIABLES.var)

**Purpose**

The constraint MAX_MAX_BUMP_ON_DECREASING_SEQUENCE(VALUE, VARIABLES) holds. In addition, FOUND is a collection of 0/1 variables where the value 1 indicates the position of the **found** letter in those occurrences of the pattern BUMP_ON_DECREASING_SEQUENCE for which the feature value is VALUE.

The position of the **found** letter in an occurrence of a pattern is the first position where the occurrence of pattern is identified, even if the pattern is not complete.

An occurrence of the pattern BUMP_ON_DECREASING_SEQUENCE is the subsequence which matches the regular expression `>><>>`.

Assume that the occurrence of the pattern BUMP_ON_DECREASING_SEQUENCE starts at position $i$ and ends at position $j$. The feature MAX computes the maximum of the values from index $i + 2$ to index $j$.

**Example**

$$\left( \begin{array}{l} 6, \langle 7, 6, 5, 6, 5, 4, 1, 4, 7, 5, 4, 2, 5, 4, 3, 3 \rangle, \\ \langle 0, 0, 0, 0, 1, 0, 0, 0, 0, 0, 0, 0, 0, 0, 0, 0 \rangle \end{array} \right)$$

Figure 3.981 provides an example where the POS_MAX_MAX_BUMP_ON_DECREASING_SEQUENCE $(6, [7, 6, 5, 6, 5, 4, 1, 4, 7, 5, 4, 2, 5, 4, 3, 3], [0, 0, 0, 0, 1, 0, 0, 0, 0, 0, 0, 0, 0, 0, 0, 0])$ constraint holds.

**Typical**

|VARIABLES| > 5

range(VARIABLES.var) > 2

Figure 3.981: Illustrating the POS_MAX_MAX_BUMP_ON_DECREASING_SEQUENCE constraint of the **Example** slot

**Arg. properties**
- Functional dependency: VALUE determined by VARIABLES.
- Functional dependency: FOUND determined by VARIABLES.

**Automaton**     Similar to the automaton of the MAX_MAX_BUMP_ON_DECREASING_SEQUENCE constraint but use the decoration table 2.33.

## 3.417 POS_MAX_MAX_DECREASING

DESCRIPTION    AUTOMATON

**Origin**     Based on constraint MAX_MAX_DECREASING.

**Constraint**     POS_MAX_MAX_DECREASING(VALUE, VARIABLES, FOUND)

**Arguments**
```
VALUE     : dvar
VARIABLES : collection(var−dvar)
FOUND     : collection(var−dvar)
```

**Restrictions**

$sv \leq 1 \lor rv \leq 1 \Rightarrow VALUE = -\infty$
$VALUE = -\infty \lor VALUE \geq minv + 1$
$VALUE \leq maxv①$
required(VARIABLES, var)
required(FOUND, var)
where
maxv =maxval(VARIABLES.var)
minv =minval(VARIABLES.var)
sv = |VARIABLES|
rv =range(VARIABLES.var)

**Purpose**

The constraint MAX_MAX_DECREASING(VALUE, VARIABLES) holds. In addition, FOUND is a collection of 0/1 variables where the value 1 indicates the position of the **found** letter in those occurrences of the pattern DECREASING for which the feature value is VALUE.

The position of the **found** letter in an occurrence of a pattern is the first position where the occurrence of pattern is identified, even if the pattern is not complete.

An occurrence of the pattern DECREASING is the subsequence which matches the regular expression >.

Assume that the occurrence of the pattern DECREASING starts at position $i$ and ends at position $j$. The feature MAX computes the maximum of the values from index $i$ to index $j + 1$.

**Example**

$$\left( \begin{array}{l} 6, \langle 3, 4, 2, 2, 5, 6, 6, 4, 4, 3, 1, 1, 4, 6, 4, 4 \rangle, \\ \langle 0, 0, 0, 0, 0, 0, 1, 0, 0, 0, 0, 0, 0, 1, 0, 0 \rangle \end{array} \right)$$

Figure 3.982 provides an example where the POS_MAX_MAX_DECREASING (6, [3, 4, 2, 2, 5, 6, 6, 4, 4, 3, 1, 1, 4, 6, 4, 4], [0, 0, 0, 0, 0, 0, 1, 0, 0, 0, 0, 0, 0, 1, 0, 0]) constraint holds.

**Typical**

|VARIABLES| > 1
range(VARIABLES.var) > 1

Figure 3.982: Illustrating the POS_MAX_MAX_DECREASING constraint of the **Example** slot

| Arg. properties | • Functional dependency: VALUE determined by VARIABLES. |
|---|---|
| | • Functional dependency: FOUND determined by VARIABLES. |

**Automaton**    Similar to the automaton of the MAX_MAX_DECREASING constraint but use the decoration
table 2.33.

# 3.418   POS_MAX_MAX_DECREASING_SEQUENCE

▶ ▷ ◁ ◀        **DESCRIPTION**          **AUTOMATON**

**Origin**        Based on constraint MAX_MAX_DECREASING_SEQUENCE.

**Constraint**    POS_MAX_MAX_DECREASING_SEQUENCE(VALUE, VARIABLES, FOUND)

**Arguments**     VALUE     : dvar
                  VARIABLES : collection(var−dvar)
                  FOUND     : collection(var−dvar)

**Restrictions**  $sv \leq 1 \lor rv \leq 1 \Rightarrow VALUE = -\infty$
                  $VALUE = -\infty \lor VALUE \geq minv + 1$
                  $VALUE \leq maxv$①
                  required(VARIABLES, var)
                  required(FOUND, var)
                  where
                  maxv = maxval(VARIABLES.var)
                  minv = minval(VARIABLES.var)
                  sv = |VARIABLES|
                  rv = range(VARIABLES.var)

**Purpose**       The constraint MAX_MAX_DECREASING_SEQUENCE(VALUE, VARIABLES) holds. In addition, FOUND is a collection of 0/1 variables where the value 1 indicates the position of the **found** letter in those occurrences of the pattern DECREASING_SEQUENCE for which the feature value is VALUE.

                  The position of the **found** letter in an occurrence of a pattern is the first position where the occurrence of pattern is identified, even if the pattern is not complete.

                  An occurrence of the pattern DECREASING_SEQUENCE is the *maximal* subsequence which matches the regular expression $> (> | =)^* > | >$.

                  Assume that the occurrence of the pattern DECREASING_SEQUENCE starts at position $i$ and ends at position $j$. The feature MAX computes the maximum of the values from index $i$ to index $j + 1$.

**Example**       $$\left( \begin{array}{l} 6, \langle 3, 4, 2, 2, 5, 6, 6, 4, 4, 3, 1, 1, 4, 6, 4, 4 \rangle, \\ \langle 0, 0, 0, 0, 0, 0, 1, 0, 0, 0, 0, 0, 0, 1, 0, 0 \rangle \end{array} \right)$$

                  Figure 3.983 provides an example where the POS_MAX_MAX_DECREASING_SEQUENCE $(6, [3, 4, 2, 2, 5, 6, 6, 4, 4, 3, 1, 1, 4, 6, 4, 4], [0, 0, 0, 0, 0, 0, 1, 0, 0, 0, 0, 0, 0, 1, 0, 0])$ constraint holds.

**Typical**       $|VARIABLES| > 1$
                  range(VARIABLES.var) > 1

Figure 3.983: Illustrating the POS_MAX_MAX_DECREASING_SEQUENCE constraint of the **Example** slot

**Arg. properties**
- Functional dependency: VALUE determined by VARIABLES.
- Functional dependency: FOUND determined by VARIABLES.

**Automaton**    Similar to the automaton of the MAX_MAX_DECREASING_SEQUENCE constraint but use the decoration table 2.33.

## 3.419 POS_MAX_MAX_DIP_ON_INCREASING_SEQUENCE

► ▷ ◁ ◄          **DESCRIPTION**          **AUTOMATON**

| | |
|---|---|
| **Origin** | Based on constraint MAX_MAX_DIP_ON_INCREASING_SEQUENCE. |
| **Constraint** | POS_MAX_MAX_DIP_ON_INCREASING_SEQUENCE(VALUE, VARIABLES, FOUND) |

**Arguments**

```
VALUE      : dvar
VARIABLES  : collection(var−dvar)
FOUND      : collection(var−dvar)
```

**Restrictions**

$sv \leq 5 \vee rv \leq 2 \Rightarrow VALUE = -\infty$

$VALUE = -\infty \vee VALUE \geq minv + 2$

$VALUE \leq maxv$①

required(VARIABLES, var)

required(FOUND, var)

where

maxv = maxval(VARIABLES.var)

minv = minval(VARIABLES.var)

sv = |VARIABLES|

rv = range(VARIABLES.var)

**Purpose**

The constraint MAX_MAX_DIP_ON_INCREASING_SEQUENCE(VALUE, VARIABLES) holds. In addition, FOUND is a collection of 0/1 variables where the value 1 indicates the position of the **found** letter in those occurrences of the pattern DIP_ON_INCREASING_SEQUENCE for which the feature value is VALUE.

The position of the **found** letter in an occurrence of a pattern is the first position where the occurrence of pattern is identified, even if the pattern is not complete.

An occurrence of the pattern DIP_ON_INCREASING_SEQUENCE is the subsequence which matches the regular expression `<<><<`.

Assume that the occurrence of the pattern DIP_ON_INCREASING_SEQUENCE starts at position $i$ and ends at position $j$. The feature MAX computes the maximum of the values from index $i + 2$ to index $j$.

**Example**

$$\left( \begin{array}{l} 6, \langle 1, 2, 3, 2, 5, 6, 7, 4, 1, 3, 4, 6, 1, 2, 4, 4 \rangle, \\ \langle 0, 0, 0, 0, 0, 0, 0, 0, 0, 0, 0, 0, 0, 1, 0, 0 \rangle \end{array} \right)$$

Figure 3.984 provides an example where the POS_MAX_MAX_DIP_ON_INCREASING_SEQUENCE $(6, [1, 2, 3, 2, 5, 6, 7, 4, 1, 3, 4, 6, 1, 2, 4, 4], [0, 0, 0, 0, 0, 0, 0, 0, 0, 0, 0, 0, 0, 1, 0, 0])$ constraint holds.

**Typical**

|VARIABLES| > 5

range(VARIABLES.var) > 2

Figure 3.984: Illustrating the POS_MAX_MAX_DIP_ON_INCREASING_SEQUENCE constraint of the **Example** slot

**Arg. properties**
- Functional dependency: VALUE determined by VARIABLES.
- Functional dependency: FOUND determined by VARIABLES.

**Automaton**      Similar to the automaton of the MAX_MAX_DIP_ON_INCREASING_SEQUENCE constraint but use the decoration table 2.33.

## 3.420  POS_MAX_MAX_INCREASING

▶ ▷ ◁ ◀          DESCRIPTION              AUTOMATON

**Origin**           Based on constraint MAX_MAX_INCREASING.

**Constraint**       POS_MAX_MAX_INCREASING(VALUE, VARIABLES, FOUND)

**Arguments**        VALUE      : dvar
                     VARIABLES  : collection(var−dvar)
                     FOUND      : collection(var−dvar)

**Restrictions**     $\text{sv} \leq 1 \lor \text{rv} \leq 1 \Rightarrow \text{VALUE} = -\infty$
                     $\text{VALUE} = -\infty \lor \text{VALUE} \geq \text{minv} + 1$
                     $\text{VALUE} \leq \text{maxv}①$
                     required(VARIABLES, var)
                     required(FOUND, var)
                     where
                     $\text{maxv} = \text{maxval}(\text{VARIABLES.var})$
                     $\text{minv} = \text{minval}(\text{VARIABLES.var})$
                     $\text{sv} = |\text{VARIABLES}|$
                     $\text{rv} = \text{range}(\text{VARIABLES.var})$

**Purpose**          The constraint MAX_MAX_INCREASING(VALUE, VARIABLES) holds. In addition, FOUND
                     is a collection of 0/1 variables where the value 1 indicates the position of the **found**
                     letter in those occurrences of the pattern INCREASING for which the feature value is
                     VALUE.
                     The position of the **found** letter in an occurrence of a pattern is the first position where
                     the occurrence of pattern is identified, even if the pattern is not complete.
                     An occurrence of the pattern INCREASING is the subsequence which matches the regular
                     expression <.
                     Assume that the occurrence of the pattern INCREASING starts at position $i$ and ends at
                     position $j$. The feature MAX computes the maximum of the values from index $i$ to index
                     $j + 1$.

**Example**          $\left( \begin{array}{l} 6, \langle 4, 3, 5, 5, 2, 1, 1, 3, 3, 4, 6, 6, 3, 1, 1, 3, 3 \rangle, \\ \langle 0, 0, 0, 0, 0, 0, 0, 0, 0, 0, 1, 0, 0, 0, 0, 0, 0 \rangle \end{array} \right)$

                     Figure 3.985 provides an example where the POS_MAX_MAX_INCREASING
                     $(6, [4, 3, 5, 5, 2, 1, 1, 3, 3, 4, 6, 6, 3, 1, 1, 3, 3], [0, 0, 0, 0, 0, 0, 0, 0, 0, 0, 1, 0, 0, 0, 0, 0, 0])$
                     constraint holds.

**Typical**          $|\text{VARIABLES}| > 1$
                     $\text{range}(\text{VARIABLES.var}) > 1$

Figure 3.985: Illustrating the POS_MAX_MAX_INCREASING constraint of the **Example** slot

**Arg. properties**

- Functional dependency: VALUE determined by VARIABLES.
- Functional dependency: FOUND determined by VARIABLES.

**Automaton**     Similar to the automaton of the MAX_MAX_INCREASING constraint but use the decoration table 2.33.

# 3.421 POS_MAX_MAX_INCREASING_SEQUENCE

      DESCRIPTION        AUTOMATON

| | |
|---|---|
| **Origin** | Based on constraint MAX_MAX_INCREASING_SEQUENCE. |
| **Constraint** | POS_MAX_MAX_INCREASING_SEQUENCE(VALUE, VARIABLES, FOUND) |

**Arguments**

| | | |
|---|---|---|
| VALUE | : | dvar |
| VARIABLES | : | collection(var−dvar) |
| FOUND | : | collection(var−dvar) |

**Restrictions**

$\mathtt{sv} \leq 1 \vee \mathtt{rv} \leq 1 \Rightarrow \mathtt{VALUE} = -\infty$

$\mathtt{VALUE} = -\infty \vee \mathtt{VALUE} \geq \mathtt{minv} + 1$

$\mathtt{VALUE} \leq \mathtt{maxv}①$

required(VARIABLES, var)

required(FOUND, var)

where

$\mathtt{maxv} = \mathtt{maxval}(\mathtt{VARIABLES.var})$

$\mathtt{minv} = \mathtt{minval}(\mathtt{VARIABLES.var})$

$\mathtt{sv} = |\mathtt{VARIABLES}|$

$\mathtt{rv} = \mathtt{range}(\mathtt{VARIABLES.var})$

**Purpose**

The constraint MAX_MAX_INCREASING_SEQUENCE(VALUE, VARIABLES) holds. In addition, FOUND is a collection of 0/1 variables where the value 1 indicates the position of the **found** letter in those occurrences of the pattern INCREASING_SEQUENCE for which the feature value is VALUE.

The position of the **found** letter in an occurrence of a pattern is the first position where the occurrence of pattern is identified, even if the pattern is not complete.

An occurrence of the pattern INCREASING_SEQUENCE is the *maximal* subsequence which matches the regular expression $< (< | =)^* < | <$.

Assume that the occurrence of the pattern INCREASING_SEQUENCE starts at position $i$ and ends at position $j$. The feature MAX computes the maximum of the values from index $i$ to index $j + 1$.

**Example**

$$\left( \begin{array}{l} 6, \langle 4, 3, 5, 5, 2, 1, 1, 3, 3, 4, 6, 6, 3, 1, 1, 3, 3 \rangle, \\ \langle 0, 0, 0, 0, 0, 0, 1, 0, 0, 0, 0, 0, 0, 0, 0, 0, 0 \rangle \end{array} \right)$$

Figure 3.986 provides an example where the POS_MAX_MAX_INCREASING_SEQUENCE $(6, [4, 3, 5, 5, 2, 1, 1, 3, 3, 4, 6, 6, 3, 1, 1, 3, 3], [0, 0, 0, 0, 0, 0, 1, 0, 0, 0, 0, 0, 0, 0, 0, 0, 0])$ constraint holds.

**Typical**

$|\mathtt{VARIABLES}| > 1$

$\mathtt{range}(\mathtt{VARIABLES.var}) > 1$

Figure 3.986: Illustrating the POS_MAX_MAX_INCREASING_SEQUENCE constraint of the **Example** slot

**Arg. properties**

- Functional dependency: VALUE determined by VARIABLES.
- Functional dependency: FOUND determined by VARIABLES.

**Automaton**       Similar to the automaton of the MAX_MAX_INCREASING_SEQUENCE constraint but use the
                    decoration table 2.33.

## 3.422 POS_MAX_MAX_INFLEXION

► ▷ ◁ ◀          **DESCRIPTION**          **AUTOMATON**

$$< (< \mid =)^* > \mid > (> \mid =)^* <$$

| | |
|---|---|
| **Origin** | Based on constraint MAX_MAX_INFLEXION. |
| **Constraint** | POS_MAX_MAX_INFLEXION(VALUE, VARIABLES, FOUND) |
| **Arguments** | VALUE     : dvar<br>VARIABLES : collection(var−dvar)<br>FOUND     : collection(var−dvar) |

**Restrictions**

$\mathtt{sv} \leq 2 \vee \mathtt{rv} \leq 1 \Rightarrow \mathtt{VALUE} = -\infty$

$\mathtt{VALUE} = -\infty \vee \mathtt{VALUE} \geq \mathtt{minv}$

$\mathtt{VALUE} \leq \mathtt{maxv}$ ①

required(VARIABLES, var)

required(FOUND, var)

where

$\mathtt{maxv} = \mathtt{maxval}(\mathtt{VARIABLES.var})$

$\mathtt{minv} = \mathtt{minval}(\mathtt{VARIABLES.var})$

$\mathtt{sv} = |\mathtt{VARIABLES}|$

$\mathtt{rv} = \mathtt{range}(\mathtt{VARIABLES.var})$

**Purpose**

The constraint MAX_MAX_INFLEXION(VALUE, VARIABLES) holds. In addition, FOUND is a collection of 0/1 variables where the value 1 indicates the position of the **found** letter in those occurrences of the pattern INFLEXION for which the feature value is VALUE.

The position of the **found** letter in an occurrence of a pattern is the first position where the occurrence of pattern is identified, even if the pattern is not complete.

An occurrence of the pattern INFLEXION is the *maximal* subsequence which matches the regular expression $< (< \mid =)^* > \mid > (> \mid =)^* <$.

Assume that the occurrence of the pattern INFLEXION starts at position $i$ and ends at position $j$. The feature MAX computes the maximum of the values from index $i + 1$ to index $j$.

**Example**

$$\left( \begin{array}{l} 6, \langle 1, 2, 6, 6, 4, 4, 3, 5, 2, 5, 1, 5, 3, 3, 4, 4 \rangle, \\ \langle 0, 0, 0, 1, 0, 0, 0, 0, 0, 0, 0, 0, 0, 0, 0, 0 \rangle \end{array} \right)$$

Figure 3.987 provides an example where the POS_MAX_MAX_INFLEXION $(6, [1, 2, 6, 6, 4, 4, 3, 5, 2, 5, 1, 5, 3, 3, 4, 4], [0, 0, 0, 1, 0, 0, 0, 0, 0, 0, 0, 0, 0, 0, 0, 0])$ constraint holds.

**Typical**

$|\mathtt{VARIABLES}| > 2$

$\mathtt{range}(\mathtt{VARIABLES.var}) > 1$

Figure 3.987: Illustrating the POS_MAX_MAX_INFLEXION constraint of the **Example** slot

**Arg. properties**

- Functional dependency: VALUE determined by VARIABLES.
- Functional dependency: FOUND determined by VARIABLES.

**Automaton**    Similar to the automaton of the MAX_MAX_INFLEXION constraint but use the decoration table 2.33.

## 3.423 POS_MAX_MAX_PEAK

| | |
|---|---|
| **Origin** | Based on constraint MAX_MAX_PEAK. |
| **Constraint** | POS_MAX_MAX_PEAK(VALUE, VARIABLES, FOUND) |
| **Arguments** | VALUE : dvar |
| | VARIABLES : collection(var−dvar) |
| | FOUND : collection(var−dvar) |

**Restrictions**

$$\mathtt{sv} \leq 2 \lor \mathtt{rv} \leq 1 \Rightarrow \mathtt{VALUE} = -\infty$$
$$\mathtt{VALUE} = -\infty \lor \mathtt{VALUE} \geq \mathtt{minv} + 1$$
$$\mathtt{VALUE} \leq \mathtt{maxv}①$$
required(VARIABLES, var)
required(FOUND, var)
where
$$\mathtt{maxv} = \mathtt{maxval}(\mathtt{VARIABLES.var})$$
$$\mathtt{minv} = \mathtt{minval}(\mathtt{VARIABLES.var})$$
$$\mathtt{sv} = |\mathtt{VARIABLES}|$$
$$\mathtt{rv} = \mathtt{range}(\mathtt{VARIABLES.var})$$

**Purpose**

The constraint MAX_MAX_PEAK(VALUE, VARIABLES) holds. In addition, FOUND is a collection of 0/1 variables where the value 1 indicates the position of the **found** letter in those occurrences of the pattern PEAK for which the feature value is VALUE.

The position of the **found** letter in an occurrence of a pattern is the first position where the occurrence of pattern is identified, even if the pattern is not complete.

An occurrence of the pattern PEAK is the *maximal* subsequence which matches the regular expression $< (= | <)^* (> | =)^* >$.

Assume that the occurrence of the pattern PEAK starts at position $i$ and ends at position $j$. The feature MAX computes the maximum of the values from index $i + 1$ to index $j$.

**Example**

$$\left( \begin{array}{l} 6, \langle 7, 5, 5, 1, 4, 5, 2, 2, 3, 5, 6, 2, 3, 3, 3, 1 \rangle, \\ \langle 0, 0, 0, 0, 0, 0, 0, 0, 0, 0, 0, 1, 0, 0, 0, 0 \rangle \end{array} \right)$$

Figure 3.988 provides an example where the POS_MAX_MAX_PEAK $(6, [7, 5, 5, 1, 4, 5, 2, 2, 3, 5, 6, 2, 3, 3, 3, 1], [0, 0, 0, 0, 0, 0, 0, 0, 0, 0, 0, 1, 0, 0, 0, 0])$ constraint holds.

| | |
|---|---|
| **Typical** | $\lvert \mathtt{VARIABLES} \rvert > 2$ |
| | range(VARIABLES.var) $> 1$ |
| **Arg. properties** | • Functional dependency: VALUE determined by VARIABLES. |
| | • Functional dependency: FOUND determined by VARIABLES. |

Figure 3.988: Illustrating the POS_MAX_MAX_PEAK constraint of the **Example** slot

**Automaton**          Similar to the automaton of the MAX_MAX_PEAK constraint but use the decoration table 2.33.

# 3.424   POS_MAX_MAX_STRICTLY_DECREASING_SEQUENCE

▶ ▷ ◁ ◀        **DESCRIPTION**              **AUTOMATON**

| | |
|---|---|
| **Origin** | Based on constraint MAX_MAX_STRICTLY_DECREASING_SEQUENCE. |
| **Constraint** | POS_MAX_MAX_STRICTLY_DECREASING_SEQUENCE(VALUE, VARIABLES, FOUND) |
| **Arguments** | VALUE      : dvar<br>VARIABLES  : collection(var−dvar)<br>FOUND      : collection(var−dvar) |

**Restrictions**

$$sv \leq 1 \lor rv \leq 1 \Rightarrow VALUE = -\infty$$
$$VALUE = -\infty \lor VALUE \geq minv + 1$$
$$VALUE \leq maxv①$$
required(VARIABLES, var)
required(FOUND, var)
where
maxv = maxval(VARIABLES.var)
minv = minval(VARIABLES.var)
sv = |VARIABLES|
rv = range(VARIABLES.var)

**Purpose**

The constraint MAX_MAX_STRICTLY_DECREASING_SEQUENCE(VALUE, VARIABLES) holds. In addition, FOUND is a collection of 0/1 variables where the value 1 indicates the position of the **found** letter in those occurrences of the pattern STRICTLY_DECREASING_SEQUENCE for which the feature value is VALUE.

The position of the **found** letter in an occurrence of a pattern is the first position where the occurrence of pattern is identified, even if the pattern is not complete.

An occurrence of the pattern STRICTLY_DECREASING_SEQUENCE is the *maximal* subsequence which matches the regular expression $>^+$.

Assume that the occurrence of the pattern STRICTLY_DECREASING_SEQUENCE starts at position $i$ and ends at position $j$. The feature MAX computes the maximum of the values from index $i$ to index $j + 1$.

**Example**

$$\left( \begin{array}{l} 6, \langle 4, 4, 6, 4, 1, 1, 3, 4, 4, 6, 6, 5, 2, 2, 4, 3 \rangle , \\ \langle 0, 0, 1, 0, 0, 0, 0, 0, 0, 0, 0, 1, 0, 0, 0, 0, 0 \rangle \end{array} \right)$$

Figure 3.989 provides an example where the POS_MAX_MAX_STRICTLY_DECREASING_SEQUENCE (6, [4, 4, 6, 4, 1, 1, 3, 4, 4, 6, 6, 5, 2, 2, 4, 3], [0, 0, 1, 0, 0, 0, 0, 0, 0, 0, 0, 1, 0, 0, 0, 0, 0]) constraint holds.

**Typical**

|VARIABLES| > 1
range(VARIABLES.var) > 1

Figure 3.989: Illustrating the `POS_MAX_MAX_STRICTLY_DECREASING_SEQUENCE` constraint of the **Example** slot

**Arg. properties**

- Functional dependency: `VALUE` determined by `VARIABLES`.
- Functional dependency: `FOUND` determined by `VARIABLES`.

**Automaton**    Similar to the automaton of the MAX_MAX_STRICTLY_DECREASING_SEQUENCE constraint but use the decoration table 2.33.

# 3.425　POS_MAX_MAX_STRICTLY_INCREASING_SEQUENCE

▶ ▷ ◁ ◀　　DESCRIPTION　　　　AUTOMATON

**Origin**　　Based on constraint MAX_MAX_STRICTLY_INCREASING_SEQUENCE.

**Constraint**　　POS_MAX_MAX_STRICTLY_INCREASING_SEQUENCE(VALUE, VARIABLES, FOUND)

**Arguments**
```
VALUE     : dvar
VARIABLES : collection(var−dvar)
FOUND     : collection(var−dvar)
```

**Restrictions**

$\mathtt{sv} \le 1 \lor \mathtt{rv} \le 1 \Rightarrow \mathtt{VALUE} = -\infty$

$\mathtt{VALUE} = -\infty \lor \mathtt{VALUE} \ge \mathtt{minv} + 1$

$\mathtt{VALUE} \le \mathtt{maxv}①$

required(VARIABLES, var)

required(FOUND, var)

where

maxv = maxval(VARIABLES.var)

minv = minval(VARIABLES.var)

sv = |VARIABLES|

rv = range(VARIABLES.var)

The constraint MAX_MAX_STRICTLY_INCREASING_SEQUENCE(VALUE, VARIABLES) holds. In addition, FOUND is a collection of 0/1 variables where the value 1 indicates the position of the **found** letter in those occurrences of the pattern STRICTLY_INCREASING_SEQUENCE for which the feature value is VALUE.

The position of the **found** letter in an occurrence of a pattern is the first position where the occurrence of pattern is identified, even if the pattern is not complete.

An occurrence of the pattern STRICTLY_INCREASING_SEQUENCE is the *maximal* subsequence which matches the regular expression $<^+$.

Assume that the occurrence of the pattern STRICTLY_INCREASING_SEQUENCE starts at position $i$ and ends at position $j$. The feature MAX computes the maximum of the values from index $i$ to index $j + 1$.

**Purpose**

**Example**

$$\left( \begin{array}{l} 6, \langle 4, 3, 5, 5, 2, 1, 1, 2, 3, 4, 6, 6, 3, 1, 2, 3 \rangle, \\ \langle 0, 0, 0, 0, 0, 0, 1, 0, 0, 0, 0, 0, 0, 0, 0, 0 \rangle \end{array} \right)$$

Figure 3.990 provides an example where the POS_MAX_MAX_STRICTLY_INCREASING_SEQUENCE (6, [4, 3, 5, 5, 2, 1, 1, 2, 3, 4, 6, 6, 3, 1, 2, 3], [0, 0, 0, 0, 0, 0, 1, 0, 0, 0, 0, 0, 0, 0, 0, 0]) constraint holds.

**Typical**

|VARIABLES| > 1

range(VARIABLES.var) > 1

Figure 3.990: Illustrating the POS_MAX_MAX_STRICTLY_INCREASING_SEQUENCE constraint of the **Example** slot

**Arg. properties**

- Functional dependency: VALUE determined by VARIABLES.
- Functional dependency: FOUND determined by VARIABLES.

**Automaton**     Similar to the automaton of the MAX_MAX_STRICTLY_INCREASING_SEQUENCE constraint but use the decoration table 2.33.

## 3.426 POS_MAX_MAX_SUMMIT

$(< | < (= | <)^* <) (> | > (= | >)^* >)$

| | |
|---|---|
| **Origin** | Based on constraint MAX_MAX_SUMMIT. |
| **Constraint** | POS_MAX_MAX_SUMMIT(VALUE, VARIABLES, FOUND) |
| **Arguments** | VALUE : dvar<br>VARIABLES : collection(var−dvar)<br>FOUND : collection(var−dvar) |

**Restrictions**

$\mathtt{sv} \leq 2 \vee \mathtt{rv} \leq 1 \Rightarrow \mathtt{VALUE} = -\infty$

$\mathtt{VALUE} = -\infty \vee \mathtt{VALUE} \geq \mathtt{minv} + 1$

$\mathtt{VALUE} \leq \mathtt{maxv}①$

required(VARIABLES, var)

required(FOUND, var)

where

$\mathtt{maxv} = \mathtt{maxval}(\mathtt{VARIABLES.var})$

$\mathtt{minv} = \mathtt{minval}(\mathtt{VARIABLES.var})$

$\mathtt{sv} = |\mathtt{VARIABLES}|$

$\mathtt{rv} = \mathtt{range}(\mathtt{VARIABLES.var})$

**Purpose**

The constraint MAX_MAX_SUMMIT(VALUE, VARIABLES) holds. In addition, FOUND is a collection of 0/1 variables where the value 1 indicates the position of the **found** letter in those occurrences of the pattern SUMMIT for which the feature value is VALUE.

The position of the **found** letter in an occurrence of a pattern is the first position where the occurrence of pattern is identified, even if the pattern is not complete.

An occurrence of the pattern SUMMIT is the *maximal* subsequence which matches the regular expression $(< | < (= | <)^* <) (> | > (= | >)^* >)$.

Assume that the occurrence of the pattern SUMMIT starts at position $i$ and ends at position $j$. The feature MAX computes the maximum of the values from index $i+1$ to index $j$.

**Example**

$$\left( \begin{array}{l} 5, \langle 7, 1, 5, 4, 4, 3, 3, 4, 6, 6, 2, 3, 4, 2, 3, 1 \rangle, \\ \langle 0, 0, 1, 0, 0, 0, 0, 0, 0, 0, 0, 0, 0, 0, 0, 0 \rangle \end{array} \right)$$

Figure 3.991 provides an example where the POS_MAX_MAX_SUMMIT $(5, [7, 1, 5, 4, 4, 3, 3, 4, 6, 6, 2, 3, 4, 2, 3, 1], [0, 0, 1, 0, 0, 0, 0, 0, 0, 0, 0, 0, 0, 0, 0, 0])$ constraint holds.

**Typical**

$|\mathtt{VARIABLES}| > 2$

$\mathtt{range}(\mathtt{VARIABLES.var}) > 1$

Figure 3.991: Illustrating the POS_MAX_MAX_SUMMIT constraint of the **Example** slot

**Arg. properties**
- Functional dependency: VALUE determined by VARIABLES.
- Functional dependency: FOUND determined by VARIABLES.

**Automaton**     Similar to the automaton of the MAX_MAX_SUMMIT constraint but use the decoration table 2.33.

## 3.427 POS_MAX_MAX_ZIGZAG

| | |
|---|---|
| **Origin** | Based on constraint MAX_MAX_ZIGZAG. |
| **Constraint** | POS_MAX_MAX_ZIGZAG(VALUE, VARIABLES, FOUND) |
| **Arguments** | VALUE : dvar <br> VARIABLES : collection(var−dvar) <br> FOUND : collection(var−dvar) |

**Restrictions**

$sv \leq 3 \vee rv \leq 1 \Rightarrow \text{VALUE} = -\infty$

$\text{VALUE} = -\infty \vee \text{VALUE} \geq \text{minv} + 1$

$\text{VALUE} \leq \text{maxv}①$

required(VARIABLES, var)

required(FOUND, var)

where

maxv = maxval(VARIABLES.var)

minv = minval(VARIABLES.var)

$sv = |\text{VARIABLES}|$

rv = range(VARIABLES.var)

**Purpose**

The constraint MAX_MAX_ZIGZAG(VALUE, VARIABLES) holds. In addition, FOUND is a collection of 0/1 variables where the value 1 indicates the position of the **found** letter in those occurrences of the pattern ZIGZAG for which the feature value is VALUE.

The position of the **found** letter in an occurrence of a pattern is the first position where the occurrence of pattern is identified, even if the pattern is not complete.

An occurrence of the pattern ZIGZAG is the *maximal* subsequence which matches the regular expression $(<>)^+ (< | <>) | (><)^+ (> | ><)$.

Assume that the occurrence of the pattern ZIGZAG starts at position $i$ and ends at position $j$. The feature MAX computes the maximum of the values from index $i + 1$ to index $j$.

**Example**

$$\left( \begin{array}{l} 7, \langle 4,1,3,1,4,6,1,5,5,2,7,2,3,1,6,1 \rangle , \\ \langle 0,0,0,0,0,0,0,0,0,0,1,0,0,0,0,0 \rangle \end{array} \right)$$

Figure 3.992 provides an example where the POS_MAX_MAX_ZIGZAG $(7, [4,1,3,1,4,6,1,5,5,2,7,2,3,1,6,1], [0,0,0,0,0,0,0,0,0,0,1,0,0,0,0,0])$ constraint holds.

**Typical**

$|\text{VARIABLES}| > 3$

range(VARIABLES.var) > 1

**Arg. properties**

- Functional dependency: VALUE determined by VARIABLES.
- Functional dependency: FOUND determined by VARIABLES.

Figure 3.992: Illustrating the POS_MAX_MAX_ZIGZAG constraint of the **Example** slot

**Automaton**          Similar to the automaton of the MAX_MAX_ZIGZAG constraint but use the decoration table 2.33.

# 3.428 POS_MAX_MIN_BUMP_ON_DECREASING_SEQUENCE

▶ ▷ ◁ ◀        DESCRIPTION        AUTOMATON

**Origin**  Based on constraint MAX_MIN_BUMP_ON_DECREASING_SEQUENCE.

**Constraint**  POS_MAX_MIN_BUMP_ON_DECREASING_SEQUENCE(VALUE, VARIABLES, FOUND)

**Arguments**
```
VALUE      : dvar
VARIABLES  : collection(var−dvar)
FOUND      : collection(var−dvar)
```

**Restrictions**

$sv \leq 5 \vee rv \leq 2 \Rightarrow VALUE = -\infty$
$VALUE = -\infty \vee VALUE \geq minv$
$VALUE \leq maxv - 2 ①$
required(VARIABLES, var)
required(FOUND, var)
where
maxv = maxval(VARIABLES.var)
minv = minval(VARIABLES.var)
$sv = |VARIABLES|$
$rv = range(VARIABLES.var)$

**Purpose**

The constraint MAX_MIN_BUMP_ON_DECREASING_SEQUENCE(VALUE, VARIABLES) holds. In addition, FOUND is a collection of 0/1 variables where the value 1 indicates the position of the **found** letter in those occurrences of the pattern BUMP_ON_DECREASING_SEQUENCE for which the feature value is VALUE.

The position of the **found** letter in an occurrence of a pattern is the first position where the occurrence of pattern is identified, even if the pattern is not complete.

An occurrence of the pattern BUMP_ON_DECREASING_SEQUENCE is the subsequence which matches the regular expression >><>.

Assume that the occurrence of the pattern BUMP_ON_DECREASING_SEQUENCE starts at position $i$ and ends at position $j$. The feature MIN computes the minimum of the values from index $i + 2$ to index $j$.

**Example**

$$\left( \begin{array}{l} 5, \langle 7, 6, 5, 6, 5, 4, 1, 4, 7, 5, 4, 2, 5, 4, 3, 3 \rangle , \\ \langle 0, 0, 0, 0, 1, 0, 0, 0, 0, 0, 0, 0, 0, 0, 0, 0 \rangle \end{array} \right)$$

Figure 3.993 provides an example where the POS_MAX_MIN_BUMP_ON_DECREASING_SEQUENCE (5, [7, 6, 5, 6, 5, 4, 1, 4, 7, 5, 4, 2, 5, 4, 3, 3], [0, 0, 0, 0, 1, 0, 0, 0, 0, 0, 0, 0, 0, 0, 0, 0]) constraint holds.

**Typical**

$|VARIABLES| > 5$
$range(VARIABLES.var) > 2$

Figure 3.993: Illustrating the POS_MAX_MIN_BUMP_ON_DECREASING_SEQUENCE constraint of the **Example** slot

**Arg. properties**
- Functional dependency: VALUE determined by VARIABLES.
- Functional dependency: FOUND determined by VARIABLES.

**Automaton**        Similar to the automaton of the MAX_MIN_BUMP_ON_DECREASING_SEQUENCE constraint but use the decoration table 2.33.

## 3.429   POS_MAX_MIN_DECREASING

▶ ▷ ◁ ◀          **DESCRIPTION**                **AUTOMATON**

| | |
|---|---|
| **Origin** | Based on constraint MAX_MIN_DECREASING. |
| **Constraint** | POS_MAX_MIN_DECREASING(VALUE, VARIABLES, FOUND) |

| | | |
|---|---|---|
| **Arguments** | VALUE | : dvar |
| | VARIABLES | : collection(var−dvar) |
| | FOUND | : collection(var−dvar) |

**Restrictions**

$\mathtt{sv} \leq 1 \vee \mathtt{rv} \leq 1 \Rightarrow \mathtt{VALUE} = -\infty$

$\mathtt{VALUE} = -\infty \vee \mathtt{VALUE} \geq \mathtt{minv}$

$\mathtt{VALUE} \leq \mathtt{maxv} - 1$①

required(VARIABLES, var)

required(FOUND, var)

where

$\mathtt{maxv} = \mathtt{maxval}(\mathtt{VARIABLES.var})$

$\mathtt{minv} = \mathtt{minval}(\mathtt{VARIABLES.var})$

$\mathtt{sv} = |\mathtt{VARIABLES}|$

$\mathtt{rv} = \mathtt{range}(\mathtt{VARIABLES.var})$

**Purpose**

The constraint MAX_MIN_DECREASING(VALUE, VARIABLES) holds. In addition, FOUND is a collection of 0/1 variables where the value 1 indicates the position of the **found** letter in those occurrences of the pattern DECREASING for which the feature value is VALUE.

The position of the **found** letter in an occurrence of a pattern is the first position where the occurrence of pattern is identified, even if the pattern is not complete.

An occurrence of the pattern DECREASING is the subsequence which matches the regular expression >.

Assume that the occurrence of the pattern DECREASING starts at position $i$ and ends at position $j$. The feature MIN computes the minimum of the values from index $i$ to index $j + 1$.

**Example**

$$\left( \begin{array}{l} 4, \langle 3, 4, 2, 2, 5, 6, 6, 4, 4, 3, 1, 1, 4, 6, 4, 4 \rangle\,, \\ \langle 0, 0, 0, 0, 0, 0, 1, 0, 0, 0, 0, 0, 0, 1, 0, 0 \rangle \end{array} \right)$$

Figure 3.994 provides an example where the POS_MAX_MIN_DECREASING $(4, [3, 4, 2, 2, 5, 6, 6, 4, 4, 3, 1, 1, 4, 6, 4, 4], [0, 0, 0, 0, 0, 0, 1, 0, 0, 0, 0, 0, 0, 1, 0, 0])$ constraint holds.

**Typical**

$|\mathtt{VARIABLES}| > 1$

$\mathtt{range}(\mathtt{VARIABLES.var}) > 1$

Figure 3.994: Illustrating the POS_MAX_MIN_DECREASING constraint of the **Example** slot

**Arg. properties**
- Functional dependency: VALUE determined by VARIABLES.
- Functional dependency: FOUND determined by VARIABLES.

**Automaton**     Similar to the automaton of the MAX_MIN_DECREASING constraint but use the decoration table 2.33.

# 3.430 POS_MAX_MIN_DECREASING_SEQUENCE

► ▷ ◁ ◀        **DESCRIPTION**              **AUTOMATON**

**Origin**          Based on constraint MAX_MIN_DECREASING_SEQUENCE.

**Constraint**      POS_MAX_MIN_DECREASING_SEQUENCE(VALUE, VARIABLES, FOUND)

**Arguments**
```
VALUE     : dvar
VARIABLES : collection(var−dvar)
FOUND     : collection(var−dvar)
```

**Restrictions**
$$sv \leq 1 \lor rv \leq 1 \Rightarrow VALUE = -\infty$$
$$VALUE = -\infty \lor VALUE \geq minv$$
$$VALUE \leq maxv - 1①$$
required(VARIABLES, var)
required(FOUND, var)
where
$$maxv = maxval(VARIABLES.var)$$
$$minv = minval(VARIABLES.var)$$
$$sv = |VARIABLES|$$
$$rv = range(VARIABLES.var)$$

**Purpose**

The constraint MAX_MIN_DECREASING_SEQUENCE(VALUE, VARIABLES) holds. In addition, FOUND is a collection of 0/1 variables where the value 1 indicates the position of the **found** letter in those occurrences of the pattern DECREASING_SEQUENCE for which the feature value is VALUE.

The position of the **found** letter in an occurrence of a pattern is the first position where the occurrence of pattern is identified, even if the pattern is not complete.

An occurrence of the pattern DECREASING_SEQUENCE is the *maximal* subsequence which matches the regular expression $> (> | =)^* > | >$.

Assume that the occurrence of the pattern DECREASING_SEQUENCE starts at position $i$ and ends at position $j$. The feature MIN computes the minimum of the values from index $i$ to index $j + 1$.

**Example**

$$\left( \begin{array}{l} 4, \langle 3, 4, 2, 2, 5, 6, 6, 4, 4, 3, 1, 1, 4, 6, 4, 4 \rangle, \\ \langle 0, 0, 0, 0, 0, 0, 0, 0, 0, 0, 0, 0, 0, 1, 0, 0 \rangle \end{array} \right)$$

Figure 3.995 provides an example where the POS_MAX_MIN_DECREASING_SEQUENCE (4, [3, 4, 2, 2, 5, 6, 6, 4, 4, 3, 1, 1, 4, 6, 4, 4], [0, 0, 0, 0, 0, 0, 0, 0, 0, 0, 0, 0, 0, 1, 0, 0]) constraint holds.

**Typical**
$$|VARIABLES| > 1$$
$$range(VARIABLES.var) > 1$$

Figure 3.995: Illustrating the POS_MAX_MIN_DECREASING_SEQUENCE constraint of the **Example** slot

**Arg. properties**
- Functional dependency: VALUE determined by VARIABLES.
- Functional dependency: FOUND determined by VARIABLES.

**Automaton**       Similar to the automaton of the MAX_MIN_DECREASING_SEQUENCE constraint but use the
                    decoration table 2.33.

# 3.431   POS_MAX_MIN_DIP_ON_INCREASING_SEQUENCE

▶ ▷ ◁ ◀     **DESCRIPTION**     **AUTOMATON**

**Origin**     Based on constraint MAX_MIN_DIP_ON_INCREASING_SEQUENCE.

**Constraint**     POS_MAX_MIN_DIP_ON_INCREASING_SEQUENCE(VALUE, VARIABLES, FOUND)

**Arguments**

| | | |
|---|---|---|
| VALUE | : | dvar |
| VARIABLES | : | collection(var−dvar) |
| FOUND | : | collection(var−dvar) |

**Restrictions**

$\mathtt{sv} \leq 5 \lor \mathtt{rv} \leq 2 \Rightarrow \mathtt{VALUE} = -\infty$

$\mathtt{VALUE} = -\infty \lor \mathtt{VALUE} \geq \mathtt{minv}$

$\mathtt{VALUE} \leq \mathtt{maxv} - 2①$

required(VARIABLES, var)

required(FOUND, var)

where

$\mathtt{maxv} = \mathtt{maxval}(\mathtt{VARIABLES.var})$

$\mathtt{minv} = \mathtt{minval}(\mathtt{VARIABLES.var})$

$\mathtt{sv} = |\mathtt{VARIABLES}|$

$\mathtt{rv} = \mathtt{range}(\mathtt{VARIABLES.var})$

**Purpose**

The constraint MAX_MIN_DIP_ON_INCREASING_SEQUENCE(VALUE, VARIABLES) holds. In addition, FOUND is a collection of 0/1 variables where the value 1 indicates the position of the **found** letter in those occurrences of the pattern DIP_ON_INCREASING_SEQUENCE for which the feature value is VALUE.

The position of the **found** letter in an occurrence of a pattern is the first position where the occurrence of pattern is identified, even if the pattern is not complete.

An occurrence of the pattern DIP_ON_INCREASING_SEQUENCE is the subsequence which matches the regular expression <<><<.

Assume that the occurrence of the pattern DIP_ON_INCREASING_SEQUENCE starts at position $i$ and ends at position $j$. The feature MIN computes the minimum of the values from index $i + 2$ to index $j$.

**Example**

$$\left( \begin{array}{l} 2, \langle 1, 2, 3, 2, 5, 6, 7, 4, 1, 3, 4, 6, 1, 2, 4, 4 \rangle, \\ \langle 0, 0, 0, 0, 1, 0, 0, 0, 0, 0, 0, 0, 0, 0, 0, 0 \rangle \end{array} \right)$$

Figure 3.996 provides an example where the POS_MAX_MIN_DIP_ON_INCREASING_SEQUENCE $(2, [1, 2, 3, 2, 5, 6, 7, 4, 1, 3, 4, 6, 1, 2, 4, 4], [0, 0, 0, 0, 1, 0, 0, 0, 0, 0, 0, 0, 0, 0, 0, 0])$ constraint holds.

**Typical**

$|\mathtt{VARIABLES}| > 5$

$\mathtt{range}(\mathtt{VARIABLES.var}) > 2$

Figure 3.996: Illustrating the POS_MAX_MIN_DIP_ON_INCREASING_SEQUENCE constraint of the **Example** slot

**Arg. properties**

- Functional dependency: VALUE determined by VARIABLES.
- Functional dependency: FOUND determined by VARIABLES.

**Automaton**     Similar to the automaton of the MAX_MIN_DIP_ON_INCREASING_SEQUENCE constraint but
use the decoration table 2.33.

## 3.432  POS_MAX_MIN_GORGE

   **DESCRIPTION**        **AUTOMATON**

$(> \mid > (= \mid >)^* >)(< \mid < (= \mid <)^* <)$

| | |
|---|---|
| **Origin** | Based on constraint MAX_MIN_GORGE. |
| **Constraint** | POS_MAX_MIN_GORGE(VALUE, VARIABLES, FOUND) |
| **Arguments** | VALUE       : dvar<br>VARIABLES : collection(var−dvar)<br>FOUND     : collection(var−dvar) |
| **Restrictions** | $\mathtt{sv} \leq 2 \vee \mathtt{rv} \leq 1 \Rightarrow \mathtt{VALUE} = -\infty$<br>$\mathtt{VALUE} = -\infty \vee \mathtt{VALUE} \geq \mathtt{minv}$<br>$\mathtt{VALUE} \leq \mathtt{maxv} - 1$①<br>required(VARIABLES, var)<br>required(FOUND, var)<br>where<br>maxv =maxval(VARIABLES.var)<br>minv =minval(VARIABLES.var)<br>$\mathtt{sv} = \|\mathtt{VARIABLES}\|$<br>rv =range(VARIABLES.var) |

**Purpose**

The constraint MAX_MIN_GORGE(VALUE, VARIABLES) holds. In addition, FOUND is a collection of 0/1 variables where the value 1 indicates the position of the **found** letter in those occurrences of the pattern GORGE for which the feature value is VALUE.

The position of the **found** letter in an occurrence of a pattern is the first position where the occurrence of pattern is identified, even if the pattern is not complete.

An occurrence of the pattern GORGE is the *maximal* subsequence which matches the regular expression $(> \mid > (= \mid >)^* >)(< \mid < (= \mid <)^* <)$.

Assume that the occurrence of the pattern GORGE starts at position $i$ and ends at position $j$. The feature MIN computes the minimum of the values from index $i + 1$ to index $j$.

**Example**

$$\left( \begin{array}{l} 5, \langle 1, 7, 3, 4, 4, 5, 5, 4, 2, 2, 6, 5, 4, 6, 5, 7 \rangle, \\ \langle 0, 0, 0, 0, 0, 0, 0, 0, 0, 0, 0, 0, 0, 0, 1, 0 \rangle \end{array} \right)$$

Figure 3.997 provides an example where the POS_MAX_MIN_GORGE $(5, [1, 7, 3, 4, 4, 5, 5, 4, 2, 2, 6, 5, 4, 6, 5, 7], [0, 0, 0, 0, 0, 0, 0, 0, 0, 0, 0, 0, 0, 0, 1, 0])$ constraint holds.

| | |
|---|---|
| **Typical** | $\|\mathtt{VARIABLES}\| > 2$<br>range(VARIABLES.var) $> 1$ |
| **Arg. properties** | • Functional dependency: VALUE determined by VARIABLES.<br>• Functional dependency: FOUND determined by VARIABLES. |

Figure 3.997: Illustrating the POS_MAX_MIN_GORGE constraint of the **Example** slot

**Automaton**      Similar to the automaton of the MAX_MIN_GORGE constraint but use the decoration table 2.33.

# 3.433 POS_MAX_MIN_INCREASING

► ▷ ◁ ◀         DESCRIPTION         AUTOMATON

**Origin**        Based on constraint MAX_MIN_INCREASING.

**Constraint**    POS_MAX_MIN_INCREASING(VALUE, VARIABLES, FOUND)

**Arguments**
```
VALUE     : dvar
VARIABLES : collection(var−dvar)
FOUND     : collection(var−dvar)
```

**Restrictions**
$$\texttt{sv} \leq 1 \vee \texttt{rv} \leq 1 \Rightarrow \texttt{VALUE} = -\infty$$
$$\texttt{VALUE} = -\infty \vee \texttt{VALUE} \geq \texttt{minv}$$
$$\texttt{VALUE} \leq \texttt{maxv} - 1 ①$$
required(VARIABLES, var)
required(FOUND, var)
where
$$\texttt{maxv} = \texttt{maxval}(\texttt{VARIABLES.var})$$
$$\texttt{minv} = \texttt{minval}(\texttt{VARIABLES.var})$$
$$\texttt{sv} = |\texttt{VARIABLES}|$$
$$\texttt{rv} = \texttt{range}(\texttt{VARIABLES.var})$$

**Purpose**

The constraint MAX_MIN_INCREASING(VALUE, VARIABLES) holds. In addition, FOUND is a collection of 0/1 variables where the value 1 indicates the position of the **found** letter in those occurrences of the pattern INCREASING for which the feature value is VALUE.

The position of the **found** letter in an occurrence of a pattern is the first position where the occurrence of pattern is identified, even if the pattern is not complete.

An occurrence of the pattern INCREASING is the subsequence which matches the regular expression <.

Assume that the occurrence of the pattern INCREASING starts at position $i$ and ends at position $j$. The feature MIN computes the minimum of the values from index $i$ to index $j + 1$.

**Example**
$$\left( \begin{array}{l} 4, \langle 4, 3, 5, 5, 2, 1, 1, 3, 3, 4, 6, 6, 3, 1, 1, 3, 3 \rangle, \\ \langle 0, 0, 0, 0, 0, 0, 0, 0, 0, 0, 1, 0, 0, 0, 0, 0, 0 \rangle \end{array} \right)$$

Figure 3.998 provides an example where the POS_MAX_MIN_INCREASING (4, [4, 3, 5, 5, 2, 1, 1, 3, 3, 4, 6, 6, 3, 1, 1, 3, 3], [0, 0, 0, 0, 0, 0, 0, 0, 0, 0, 1, 0, 0, 0, 0, 0, 0]) constraint holds.

**Typical**
$$|\texttt{VARIABLES}| > 1$$
$$\texttt{range}(\texttt{VARIABLES.var}) > 1$$

Figure 3.998: Illustrating the POS_MAX_MIN_INCREASING constraint of the **Example** slot

**Arg. properties**
- Functional dependency: VALUE determined by VARIABLES.
- Functional dependency: FOUND determined by VARIABLES.

**Automaton**     Similar to the automaton of the MAX_MIN_INCREASING constraint but use the decoration table 2.33.

# 3.434　POS_MAX_MIN_INCREASING_SEQUENCE

▶ ▷ ◁ ◀　　**DESCRIPTION**　　**AUTOMATON**

| | |
|---|---|
| **Origin** | Based on constraint MAX_MIN_INCREASING_SEQUENCE. |

**Constraint**　　POS_MAX_MIN_INCREASING_SEQUENCE(VALUE, VARIABLES, FOUND)

**Arguments**

```
VALUE     : dvar
VARIABLES : collection(var-dvar)
FOUND     : collection(var-dvar)
```

**Restrictions**

$sv \leq 1 \vee rv \leq 1 \Rightarrow \text{VALUE} = -\infty$

$\text{VALUE} = -\infty \vee \text{VALUE} \geq minv$

$\text{VALUE} \leq maxv - 1$①

required(VARIABLES, var)

required(FOUND, var)

where

$maxv = \text{maxval}(\text{VARIABLES.var})$

$minv = \text{minval}(\text{VARIABLES.var})$

$sv = |\text{VARIABLES}|$

$rv = \text{range}(\text{VARIABLES.var})$

**Purpose**

The constraint MAX_MIN_INCREASING_SEQUENCE(VALUE, VARIABLES) holds. In addition, FOUND is a collection of 0/1 variables where the value 1 indicates the position of the **found** letter in those occurrences of the pattern INCREASING_SEQUENCE for which the feature value is VALUE.

The position of the **found** letter in an occurrence of a pattern is the first position where the occurrence of pattern is identified, even if the pattern is not complete.

An occurrence of the pattern INCREASING_SEQUENCE is the *maximal* subsequence which matches the regular expression $< (< | =)^* < | <$.

Assume that the occurrence of the pattern INCREASING_SEQUENCE starts at position $i$ and ends at position $j$. The feature MIN computes the minimum of the values from index $i$ to index $j + 1$.

**Example**

$$\left( \begin{array}{l} 3, \langle 4, 3, 5, 5, 2, 1, 1, 3, 3, 4, 6, 6, 3, 1, 1, 3, 3 \rangle, \\ \langle 0, 1, 0, 0, 0, 0, 0, 0, 0, 0, 0, 0, 0, 0, 0, 0, 0 \rangle \end{array} \right)$$

Figure 3.999 provides an example where the POS_MAX_MIN_INCREASING_SEQUENCE $(3, [4, 3, 5, 5, 2, 1, 1, 3, 3, 4, 6, 6, 3, 1, 1, 3, 3], [0, 1, 0, 0, 0, 0, 0, 0, 0, 0, 0, 0, 0, 0, 0, 0, 0])$ constraint holds.

**Typical**

$|\text{VARIABLES}| > 1$

$\text{range}(\text{VARIABLES.var}) > 1$

Figure 3.999: Illustrating the POS_MAX_MIN_INCREASING_SEQUENCE constraint of the **Example** slot

**Arg. properties**
- Functional dependency: VALUE determined by VARIABLES.
- Functional dependency: FOUND determined by VARIABLES.

**Automaton**       Similar to the automaton of the MAX_MIN_INCREASING_SEQUENCE constraint but use the
                    decoration table 2.33.

# 3.435 POS_MAX_MIN_INFLEXION

▶ ▷ ◁ ◀     **DESCRIPTION**          **AUTOMATON**

$< (< | =)^* > | > (> | =)^* <$

| | |
|---|---|
| **Origin** | Based on constraint MAX_MIN_INFLEXION. |
| **Constraint** | POS_MAX_MIN_INFLEXION(VALUE, VARIABLES, FOUND) |

**Arguments**

```
VALUE     : dvar
VARIABLES : collection(var−dvar)
FOUND     : collection(var−dvar)
```

**Restrictions**

$\text{sv} \leq 2 \vee \text{rv} \leq 1 \Rightarrow \text{VALUE} = -\infty$

$\text{VALUE} = -\infty \vee \text{VALUE} \geq \text{minv}$

$\text{VALUE} \leq \text{maxv}①$

required(VARIABLES, var)

required(FOUND, var)

where

$\text{maxv} = \text{maxval}(\text{VARIABLES.var})$

$\text{minv} = \text{minval}(\text{VARIABLES.var})$

$\text{sv} = |\text{VARIABLES}|$

$\text{rv} = \text{range}(\text{VARIABLES.var})$

VALUE ≤ **2**

**Purpose**

The constraint MAX_MIN_INFLEXION(VALUE, VARIABLES) holds. In addition, FOUND is a collection of 0/1 variables where the value 1 indicates the position of the **found** letter in those occurrences of the pattern INFLEXION for which the feature value is VALUE.

The position of the **found** letter in an occurrence of a pattern is the first position where the occurrence of pattern is identified, even if the pattern is not complete.

An occurrence of the pattern INFLEXION is the *maximal* subsequence which matches the regular expression $< (< | =)^* > | > (> | =)^* <$.

Assume that the occurrence of the pattern INFLEXION starts at position $i$ and ends at position $j$. The feature MIN computes the minimum of the values from index $i + 1$ to index $j$.

**Example**

$$\left( \begin{array}{l} 5, \langle 1, 2, 6, 6, 4, 4, 3, 5, 2, 5, 1, 5, 3, 3, 4, 4 \rangle , \\ \langle 0, 0, 0, 0, 0, 0, 0, 1, 0, 1, 0, 1, 0, 0, 0, 0 \rangle \end{array} \right)$$

Figure 3.1000 provides an example where the POS_MAX_MIN_INFLEXION $(5, [1, 2, 6, 6, 4, 4, 3, 5, 2, 5, 1, 5, 3, 3, 4, 4], [0, 0, 0, 0, 0, 0, 0, 1, 0, 1, 0, 1, 0, 0, 0, 0])$ constraint holds.

**Typical**

$|\text{VARIABLES}| > 2$

$\text{range}(\text{VARIABLES.var}) > 1$

Figure 3.1000: Illustrating the POS_MAX_MIN_INFLEXION constraint of the **Example** slot

**Arg. properties**    • Functional dependency: VALUE determined by VARIABLES.

                       • Functional dependency: FOUND determined by VARIABLES.

**Automaton**        Similar to the automaton of the MAX_MIN_INFLEXION constraint but use the decoration
                     table 2.33.

# 3.436   POS_MAX_MIN_STRICTLY_DECREASING_SEQUENCE

▶ ▷ ◁ ◀          **DESCRIPTION**          **AUTOMATON**

| | |
|---|---|
| **Origin** | Based on constraint MAX_MIN_STRICTLY_DECREASING_SEQUENCE. |
| **Constraint** | POS_MAX_MIN_STRICTLY_DECREASING_SEQUENCE(VALUE, VARIABLES, FOUND) |

**Arguments**

```
VALUE     : dvar
VARIABLES : collection(var−dvar)
FOUND     : collection(var−dvar)
```

**Restrictions**

$\mathtt{sv} \leq 1 \vee \mathtt{rv} \leq 1 \Rightarrow \mathtt{VALUE} = -\infty$

$\mathtt{VALUE} = -\infty \vee \mathtt{VALUE} \geq \mathtt{minv}$

$\mathtt{VALUE} \leq \mathtt{maxv} - 1$ ①

required(VARIABLES, var)

required(FOUND, var)

where

$\mathtt{maxv} = \mathtt{maxval}(\mathtt{VARIABLES.var})$

$\mathtt{minv} = \mathtt{minval}(\mathtt{VARIABLES.var})$

$\mathtt{sv} = |\mathtt{VARIABLES}|$

$\mathtt{rv} = \mathtt{range}(\mathtt{VARIABLES.var})$

**Purpose**

The constraint MAX_MIN_STRICTLY_DECREASING_SEQUENCE(VALUE, VARIABLES) holds. In addition, FOUND is a collection of 0/1 variables where the value 1 indicates the position of the **found** letter in those occurrences of the pattern STRICTLY_DECREASING_SEQUENCE for which the feature value is VALUE.

The position of the **found** letter in an occurrence of a pattern is the first position where the occurrence of pattern is identified, even if the pattern is not complete.

An occurrence of the pattern STRICTLY_DECREASING_SEQUENCE is the *maximal* sub-sequence which matches the regular expression $>^+$.

Assume that the occurrence of the pattern STRICTLY_DECREASING_SEQUENCE starts at position $i$ and ends at position $j$. The feature MIN computes the minimum of the values from index $i$ to index $j + 1$.

**Example**

$$\left( \begin{array}{l} 3, \langle 4, 4, 6, 4, 1, 1, 3, 4, 4, 6, 6, 5, 2, 2, 4, 3 \rangle, \\ \langle 0, 0, 0, 0, 0, 0, 0, 0, 0, 0, 0, 0, 0, 0, 1, 0 \rangle \end{array} \right)$$

Figure 3.1001 provides an example where the POS_MAX_MIN_STRICTLY_DECREASING_SEQUENCE $(3, [4, 4, 6, 4, 1, 1, 3, 4, 4, 6, 6, 5, 2, 2, 4, 3], [0, 0, 0, 0, 0, 0, 0, 0, 0, 0, 0, 0, 0, 0, 1, 0])$ constraint holds.

**Typical**

$|\mathtt{VARIABLES}| > 1$

$\mathtt{range}(\mathtt{VARIABLES.var}) > 1$

Figure 3.1001: Illustrating the POS_MAX_MIN_STRICTLY_DECREASING_SEQUENCE constraint of the **Example** slot

**Arg. properties**

- Functional dependency: VALUE determined by VARIABLES.
- Functional dependency: FOUND determined by VARIABLES.

**Automaton**     Similar to the automaton of the MAX_MIN_STRICTLY_DECREASING_SEQUENCE constraint but use the decoration table 2.33.

## 3.437   POS_MAX_MIN_STRICTLY_INCREASING_SEQUENCE

▶ ▷ ◁ ◀        **DESCRIPTION**              **AUTOMATON**

| | |
|---|---|
| **Origin** | Based on constraint MAX_MIN_STRICTLY_INCREASING_SEQUENCE. |
| **Constraint** | POS_MAX_MIN_STRICTLY_INCREASING_SEQUENCE(VALUE, VARIABLES, FOUND) |

**Arguments**

```
VALUE     : dvar
VARIABLES : collection(var−dvar)
FOUND     : collection(var−dvar)
```

**Restrictions**

$\mathrm{sv} \leq 1 \vee \mathrm{rv} \leq 1 \Rightarrow \mathtt{VALUE} = -\infty$

$\mathtt{VALUE} = -\infty \vee \mathtt{VALUE} \geq \mathrm{minv}$

$\mathtt{VALUE} \leq \mathrm{maxv} - 1$①

required(VARIABLES, var)

required(FOUND, var)

where

$\mathrm{maxv} = \mathrm{maxval}(\mathtt{VARIABLES.var})$

$\mathrm{minv} = \mathrm{minval}(\mathtt{VARIABLES.var})$

$\mathrm{sv} = |\mathtt{VARIABLES}|$

$\mathrm{rv} = \mathrm{range}(\mathtt{VARIABLES.var})$

**Purpose**

The constraint MAX_MIN_STRICTLY_INCREASING_SEQUENCE(VALUE, VARIABLES) holds. In addition, FOUND is a collection of 0/1 variables where the value 1 indicates the position of the **found** letter in those occurrences of the pattern STRICTLY_INCREASING_SEQUENCE for which the feature value is VALUE.

The position of the **found** letter in an occurrence of a pattern is the first position where the occurrence of pattern is identified, even if the pattern is not complete.

An occurrence of the pattern STRICTLY_INCREASING_SEQUENCE is the *maximal* subsequence which matches the regular expression $<^+$.

Assume that the occurrence of the pattern STRICTLY_INCREASING_SEQUENCE starts at position $i$ and ends at position $j$. The feature MIN computes the minimum of the values from index $i$ to index $j + 1$.

**Example**

$$\left( \begin{array}{l} 3, \langle 4, 3, 5, 5, 2, 1, 1, 2, 3, 4, 6, 6, 3, 1, 2, 3 \rangle , \\ \langle 0, 1, 0, 0, 0, 0, 0, 0, 0, 0, 0, 0, 0, 0, 0, 0 \rangle \end{array} \right)$$

Figure 3.1002 provides an example where the POS_MAX_MIN_STRICTLY_INCREASING_SEQUENCE $(3, [4, 3, 5, 5, 2, 1, 1, 2, 3, 4, 6, 6, 3, 1, 2, 3], [0, 1, 0, 0, 0, 0, 0, 0, 0, 0, 0, 0, 0, 0, 0, 0])$ constraint holds.

**Typical**

$|\mathtt{VARIABLES}| > 1$

$\mathrm{range}(\mathtt{VARIABLES.var}) > 1$

Figure 3.1002: Illustrating the POS_MAX_MIN_STRICTLY_INCREASING_SEQUENCE constraint of the **Example** slot

| Arg. properties | • Functional dependency: VALUE determined by VARIABLES. |
|---|---|
| | • Functional dependency: FOUND determined by VARIABLES. |

**Automaton**        Similar to the automaton of the MAX_MIN_STRICTLY_INCREASING_SEQUENCE constraint but use the decoration table 2.33.

# 3.438 POS_MAX_MIN_VALLEY

► ▷ ◁ ◀        DESCRIPTION            AUTOMATON

| Origin | Based on constraint MAX_MIN_VALLEY. |
|---|---|

| Constraint | POS_MAX_MIN_VALLEY(VALUE, VARIABLES, FOUND) |
|---|---|

| Arguments | VALUE      :  dvar |
|---|---|
|  | VARIABLES :  collection(var−dvar) |
|  | FOUND     :  collection(var−dvar) |

**Restrictions**

$\text{sv} \leq 2 \vee \text{rv} \leq 1 \Rightarrow \text{VALUE} = -\infty$

$\text{VALUE} = -\infty \vee \text{VALUE} \geq \text{minv}$

$\text{VALUE} \leq \text{maxv} - 1$①

required(VARIABLES, var)

required(FOUND, var)

where

maxv = maxval(VARIABLES.var)

minv = minval(VARIABLES.var)

sv = |VARIABLES|

rv = range(VARIABLES.var)

**Purpose**

The constraint MAX_MIN_VALLEY(VALUE, VARIABLES) holds. In addition, FOUND is a collection of 0/1 variables where the value 1 indicates the position of the **found** letter in those occurrences of the pattern VALLEY for which the feature value is VALUE.

The position of the **found** letter in an occurrence of a pattern is the first position where the occurrence of pattern is identified, even if the pattern is not complete.

An occurrence of the pattern VALLEY is the *maximal* subsequence which matches the regular expression $> (= | >)^* (< | =)^* <$.

Assume that the occurrence of the pattern VALLEY starts at position $i$ and ends at position $j$. The feature MIN computes the minimum of the values from index $i + 1$ to index $j$.

**Example**

$$\left( \begin{array}{l} 5, \langle 1, 3, 7, 4, 3, 6, 6, 5, 3, 3, 2, 6, 5, 5, 5, 7 \rangle, \\ \langle 0, 0, 0, 0, 0, 0, 0, 0, 0, 0, 0, 0, 0, 0, 1, 0 \rangle \end{array} \right)$$

Figure 3.1003 provides an example where the POS_MAX_MIN_VALLEY $(5, [1, 3, 7, 4, 3, 6, 6, 5, 3, 3, 2, 6, 5, 5, 5, 7], [0, 0, 0, 0, 0, 0, 0, 0, 0, 0, 0, 0, 0, 0, 1, 0])$ constraint holds.

| Typical | $|\text{VARIABLES}| > 2$ |
|---|---|
|  | range(VARIABLES.var) > 1 |

| Arg. properties | • Functional dependency: VALUE determined by VARIABLES. |
|---|---|
|  | • Functional dependency: FOUND determined by VARIABLES. |

Figure 3.1003: Illustrating the POS_MAX_MIN_VALLEY constraint of the **Example** slot

**Automaton**         Similar to the automaton of the MAX_MIN_VALLEY constraint but use the decoration table 2.33.

## 3.439  POS_MAX_MIN_ZIGZAG

► ▷ ◁ ◄          DESCRIPTION          AUTOMATON

**Origin**         Based on constraint MAX_MIN_ZIGZAG.

**Constraint**     POS_MAX_MIN_ZIGZAG(VALUE, VARIABLES, FOUND)

**Arguments**      VALUE      : dvar
                   VARIABLES  : collection(var−dvar)
                   FOUND      : collection(var−dvar)

**Restrictions**   $sv \leq 3 \vee rv \leq 1 \Rightarrow \text{VALUE} = -\infty$

                   $\text{VALUE} = -\infty \vee \text{VALUE} \geq minv$

                   $\text{VALUE} \leq maxv - 1①$

                   required(VARIABLES, var)

                   required(FOUND, var)

                   where

                   $maxv = \text{maxval}(\text{VARIABLES.var})$

                   $minv = \text{minval}(\text{VARIABLES.var})$

                   $sv = |\text{VARIABLES}|$

                   $rv = \text{range}(\text{VARIABLES.var})$

**Purpose**        The constraint MAX_MIN_ZIGZAG(VALUE, VARIABLES) holds. In addition, FOUND is a collection of 0/1 variables where the value 1 indicates the position of the **found** letter in those occurrences of the pattern ZIGZAG for which the feature value is VALUE.

                   The position of the **found** letter in an occurrence of a pattern is the first position where the occurrence of pattern is identified, even if the pattern is not complete.

                   An occurrence of the pattern ZIGZAG is the *maximal* subsequence which matches the regular expression $(<>)^+(< | <>) | (><)^+(> | ><)$.

                   Assume that the occurrence of the pattern ZIGZAG starts at position $i$ and ends at position $j$. The feature MIN computes the minimum of the values from index $i + 1$ to index $j$.

**Example**        $\left( \begin{array}{l} 1, \langle 4, 1, 3, 1, 4, 6, 1, 5, 5, 2, 7, 2, 3, 1, 6, 1 \rangle, \\ \langle 0, 0, 1, 0, 0, 0, 1, 0, 0, 0, 1, 0, 0, 0, 0, 0 \rangle \end{array} \right)$

                   Figure 3.1004 provides an example where the POS_MAX_MIN_ZIGZAG $(1, [4, 1, 3, 1, 4, 6, 1, 5, 5, 2, 7, 2, 3, 1, 6, 1], [0, 0, 1, 0, 0, 0, 1, 0, 0, 0, 1, 0, 0, 0, 0, 0])$ constraint holds.

**Typical**        $|\text{VARIABLES}| > 3$

                   $\text{range}(\text{VARIABLES.var}) > 1$

**Arg. properties**  
                   • Functional dependency: VALUE determined by VARIABLES.

                   • Functional dependency: FOUND determined by VARIABLES.

Figure 3.1004: Illustrating the POS_MAX_MIN_ZIGZAG constraint of the **Example** slot

**Automaton**            Similar to the automaton of the MAX_MIN_ZIGZAG constraint but use the decoration table 2.33.

## 3.440   POS_MAX_SURF_BUMP_ON_DECREASING_SEQUENCE

▶ ▷ ◁ ◀    **DESCRIPTION**         **AUTOMATON**

| | |
|---|---|
| **Origin** | Based on constraint MAX_SURF_BUMP_ON_DECREASING_SEQUENCE. |
| **Constraint** | POS_MAX_SURF_BUMP_ON_DECREASING_SEQUENCE(VALUE, VARIABLES, FOUND) |

**Arguments**

```
VALUE     : dvar
VARIABLES : collection(var−dvar)
FOUND     : collection(var−dvar)
```

**Restrictions**

$sv \leq 5 \vee rv \leq 2 \Rightarrow \text{VALUE} = -\infty$

$\text{VALUE} = -\infty \vee \text{VALUE} \geq 3 * \text{minv} + 3$

$\text{VALUE} \leq 3 * \text{maxv} - 3①$

required(VARIABLES, var)

required(FOUND, var)

where

maxv =maxval(VARIABLES.var)

minv =minval(VARIABLES.var)

sv = |VARIABLES|

rv =range(VARIABLES.var)

**Purpose**

The constraint MAX_SURF_BUMP_ON_DECREASING_SEQUENCE(VALUE, VARIABLES) holds. In addition, FOUND is a collection of 0/1 variables where the value 1 indicates the position of the **found** letter in those occurrences of the pattern BUMP_ON_DECREASING_SEQUENCE for which the feature value is VALUE.

The position of the **found** letter in an occurrence of a pattern is the first position where the occurrence of pattern is identified, even if the pattern is not complete.

An occurrence of the pattern BUMP_ON_DECREASING_SEQUENCE is the subsequence which matches the regular expression '>><>>'.

Assume that the occurrence of the pattern BUMP_ON_DECREASING_SEQUENCE starts at position $i$ and ends at position $j$. The feature SURF computes the sum of the values from index $i + 2$ to index $j$.

**Example**

$$\left( \begin{array}{l} 16, \langle 7, 6, 5, 6, 5, 4, 1, 4, 7, 5, 4, 2, 5, 4, 3, 3 \rangle, \\ \langle 0, 0, 0, 0, 1, 0, 0, 0, 0, 0, 0, 0, 0, 0, 0, 0 \rangle \end{array} \right)$$

Figure 3.1005 provides an example where the POS_MAX_SURF_BUMP_ON_DECREASING_SEQUENCE $(16, [7, 6, 5, 6, 5, 4, 1, 4, 7, 5, 4, 2, 5, 4, 3, 3], [0, 0, 0, 0, 1, 0, 0, 0, 0, 0, 0, 0, 0, 0, 0, 0])$ constraint holds.

**Typical**

|VARIABLES| > 5

range(VARIABLES.var) > 2

Figure 3.1005: Illustrating the POS_MAX_SURF_BUMP_ON_DECREASING_SEQUENCE constraint of the **Example** slot

| Arg. properties | |
|---|---|
| | • Functional dependency: VALUE determined by VARIABLES. |
| | • Functional dependency: FOUND determined by VARIABLES. |

**Automaton**      Similar to the automaton of the MAX_SURF_BUMP_ON_DECREASING_SEQUENCE constraint but use the decoration table 2.33.

# 3.441 POS_MAX_SURF_DECREASING

► ▷ ◁ ◀        **DESCRIPTION**        **AUTOMATON**

| | |
|---|---|
| **Origin** | Based on constraint MAX_SURF_DECREASING. |
| **Constraint** | POS_MAX_SURF_DECREASING(VALUE, VARIABLES, FOUND) |

| | | |
|---|---|---|
| **Arguments** | VALUE | : dvar |
| | VARIABLES | : collection(var−dvar) |
| | FOUND | : collection(var−dvar) |

**Restrictions**

$\mathtt{sv} \leq 1 \vee \mathtt{rv} \leq 1 \Rightarrow \mathtt{VALUE} = -\infty$

$\mathtt{VALUE} = -\infty \vee \mathtt{VALUE} \geq 2 * \mathtt{minv} + 1$

$\mathtt{VALUE} \leq 2 * \mathtt{maxv} - 1$①

required(VARIABLES, var)

required(FOUND, var)

where

$\mathtt{maxv} = \mathtt{maxval}(\mathtt{VARIABLES.var})$

$\mathtt{minv} = \mathtt{minval}(\mathtt{VARIABLES.var})$

$\mathtt{sv} = |\mathtt{VARIABLES}|$

$\mathtt{rv} = \mathtt{range}(\mathtt{VARIABLES.var})$

**Purpose**

The constraint MAX_SURF_DECREASING(VALUE, VARIABLES) holds. In addition, FOUND is a collection of 0/1 variables where the value 1 indicates the position of the **found** letter in those occurrences of the pattern DECREASING for which the feature value is VALUE.

The position of the **found** letter in an occurrence of a pattern is the first position where the occurrence of pattern is identified, even if the pattern is not complete.

An occurrence of the pattern DECREASING is the subsequence which matches the regular expression >.

Assume that the occurrence of the pattern DECREASING starts at position $i$ and ends at position $j$. The feature SURF computes the sum of the values from index $i$ to index $j+1$.

**Example**

$$\left( \begin{array}{l} 10, \langle 3, 4, 2, 2, 5, 6, 6, 4, 4, 3, 1, 1, 4, 6, 4, 4 \rangle, \\ \langle 0, 0, 0, 0, 0, 0, 1, 0, 0, 0, 0, 0, 0, 1, 0, 0 \rangle \end{array} \right)$$

Figure 3.1006 provides an example where the POS_MAX_SURF_DECREASING $(10, [3, 4, 2, 2, 5, 6, 6, 4, 4, 3, 1, 1, 4, 6, 4, 4], [0, 0, 0, 0, 0, 0, 1, 0, 0, 0, 0, 0, 0, 1, 0, 0])$ constraint holds.

**Typical**

$|\mathtt{VARIABLES}| > 1$

$\mathtt{range}(\mathtt{VARIABLES.var}) > 1$

Figure 3.1006: Illustrating the POS_MAX_SURF_DECREASING constraint of the **Example** slot

| Arg. properties | • Functional dependency: VALUE determined by VARIABLES. |
| | • Functional dependency: FOUND determined by VARIABLES. |

**Automaton**          Similar to the automaton of the MAX_SURF_DECREASING constraint but use the decoration table 2.33.

# 3.442   POS_MAX_SURF_DECREASING_SEQUENCE

► ▷ ◁ ◄         **DESCRIPTION**            **AUTOMATON**

| | |
|---|---|
| **Origin** | Based on constraint MAX_SURF_DECREASING_SEQUENCE. |
| **Constraint** | POS_MAX_SURF_DECREASING_SEQUENCE(VALUE, VARIABLES, FOUND) |

**Arguments**

```
VALUE     : dvar
VARIABLES : collection(var−dvar)
FOUND     : collection(var−dvar)
```

**Restrictions**

$sv \leq 1 \vee rv \leq 1 \Rightarrow \mathtt{VALUE} = -\infty$

$rv = 2 \Rightarrow \mathtt{VALUE} = -\infty \vee \mathtt{VALUE} \geq 2 * \mathtt{minv} + 1$

$rv \geq 3 \Rightarrow \mathtt{VALUE} = -\infty \vee \mathtt{VALUE} \geq \min(2 * \mathtt{minv} + 1, sv * (\mathtt{minv} + 1))$

$rv = 2 \Rightarrow \mathtt{VALUE} \leq 2 * \mathtt{maxv} - 1①$

$rv \geq 3 \Rightarrow \mathtt{VALUE} \leq \max(2 * \mathtt{maxv} - 1②, sv * (\mathtt{maxv} - 1)③)$

required(VARIABLES, var)

required(FOUND, var)

where

$\mathtt{maxv} = \mathtt{maxval}(\mathtt{VARIABLES.var})$

$rv = \mathtt{range}(\mathtt{VARIABLES.var})$

$sv = |\mathtt{VARIABLES}|$

$\mathtt{minv} = \mathtt{minval}(\mathtt{VARIABLES.var})$

**Purpose**

The constraint MAX_SURF_DECREASING_SEQUENCE(VALUE, VARIABLES) holds. In addition, FOUND is a collection of 0/1 variables where the value 1 indicates the position of the **found** letter in those occurrences of the pattern DECREASING_SEQUENCE for which the feature value is VALUE.

The position of the **found** letter in an occurrence of a pattern is the first position where the occurrence of pattern is identified, even if the pattern is not complete.

An occurrence of the pattern DECREASING_SEQUENCE is the *maximal* subsequence which matches the regular expression $> (> | =)^* > | >$.

Assume that the occurrence of the pattern DECREASING_SEQUENCE starts at position $i$ and ends at position $j$. The feature SURF computes the sum of the values from index $i$ to index $j + 1$.

**Example**

$$\left( \begin{array}{l} 18, \langle 3, 4, 2, 2, 5, 6, 6, 4, 4, 3, 1, 1, 4, 6, 4, 4 \rangle, \\ \langle 0, 0, 0, 0, 0, 0, 1, 0, 0, 0, 0, 0, 0, 0, 0, 0 \rangle \end{array} \right)$$

Figure 3.1007 provides an example where the POS_MAX_SURF_DECREASING_SEQUENCE $(18, [3, 4, 2, 2, 5, 6, 6, 4, 4, 3, 1, 1, 4, 6, 4, 4], [0, 0, 0, 0, 0, 0, 1, 0, 0, 0, 0, 0, 0, 0, 0, 0])$ constraint holds.

Figure 3.1007: Illustrating the POS_MAX_SURF_DECREASING_SEQUENCE constraint of the **Example** slot

| **Typical** | $|\texttt{VARIABLES}| > 1$ |
| | $\text{range}(\texttt{VARIABLES.var}) > 1$ |

| **Arg. properties** | • Functional dependency: VALUE determined by VARIABLES. |
| | • Functional dependency: FOUND determined by VARIABLES. |

**Automaton**       Similar to the automaton of the MAX_SURF_DECREASING_SEQUENCE constraint but use
                    the decoration table 2.33.

# 3.443   POS_MAX_SURF_DECREASING_TERRACE

▶ ▷ ◁ ◀        **DESCRIPTION**        **AUTOMATON**

**Origin**        Based on constraint MAX_SURF_DECREASING_TERRACE.

**Constraint**        POS_MAX_SURF_DECREASING_TERRACE(VALUE, VARIABLES, FOUND)

**Arguments**

    VALUE       : dvar
    VARIABLES   : collection(var−dvar)
    FOUND       : collection(var−dvar)

**Restrictions**

$\mathtt{sv} \leq 3 \vee \mathtt{rv} \leq 2 \Rightarrow \mathtt{VALUE} = -\infty$

$\mathtt{VALUE} = -\infty \vee \mathtt{VALUE} \geq \min(2 * (\mathtt{minv} + 1), (\mathtt{sv} - 2) * (\mathtt{minv} + 1))$

$\mathtt{VALUE} \leq \max(2 * (\mathtt{maxv} - 1)①, (\mathtt{sv} - 2) * (\mathtt{maxv} - 1)②)$

required(VARIABLES, var)

required(FOUND, var)

where

$\mathtt{maxv} = \mathtt{maxval}(\mathtt{VARIABLES.var})$

$\mathtt{sv} = |\mathtt{VARIABLES}|$

$\mathtt{minv} = \mathtt{minval}(\mathtt{VARIABLES.var})$

$\mathtt{rv} = \mathtt{range}(\mathtt{VARIABLES.var})$

**Purpose**

The constraint MAX_SURF_DECREASING_TERRACE(VALUE, VARIABLES) holds. In addition, FOUND is a collection of 0/1 variables where the value 1 indicates the position of the **found** letter in those occurrences of the pattern DECREASING_TERRACE for which the feature value is VALUE.

The position of the **found** letter in an occurrence of a pattern is the first position where the occurrence of pattern is identified, even if the pattern is not complete.

An occurrence of the pattern DECREASING_TERRACE is the *maximal* subsequence which matches the regular expression $>=^+>$.

Assume that the occurrence of the pattern DECREASING_TERRACE starts at position $i$ and ends at position $j$. The feature SURF computes the sum of the values from index $i + 1$ to index $j$.

**Example**

$$\left( \begin{array}{l} 8, \langle 6, 4, 4, 4, 5, 2, 2, 1, 3, 3, 5, 4, 4, 3, 3, 3 \rangle, \\ \langle 0, 0, 0, 0, 0, 0, 0, 0, 0, 0, 0, 0, 0, 1, 0, 0, 0 \rangle \end{array} \right)$$

Figure 3.1008 provides an example where the POS_MAX_SURF_DECREASING_TERRACE $(8, [6, 4, 4, 4, 5, 2, 2, 1, 3, 3, 5, 4, 4, 3, 3, 3], [0, 0, 0, 0, 0, 0, 0, 0, 0, 0, 0, 0, 0, 1, 0, 0, 0])$ constraint holds.

**Typical**

$|\mathtt{VARIABLES}| > 3$

range(VARIABLES.var) $> 2$

Figure 3.1008: Illustrating the POS_MAX_SURF_DECREASING_TERRACE constraint of the **Example** slot

**Arg. properties**
- Functional dependency: VALUE determined by VARIABLES.
- Functional dependency: FOUND determined by VARIABLES.

**Automaton**        Similar to the automaton of the MAX_SURF_DECREASING_TERRACE constraint but use the
                     decoration table 2.33.

# 3.444 POS_MAX_SURF_DIP_ON_INCREASING_SEQUENCE

▶ ▷ ◁ ◀    **DESCRIPTION**    **AUTOMATON**

| | |
|---|---|
| **Origin** | Based on constraint MAX_SURF_DIP_ON_INCREASING_SEQUENCE. |
| **Constraint** | POS_MAX_SURF_DIP_ON_INCREASING_SEQUENCE(VALUE, VARIABLES, FOUND) |

**Arguments**

VALUE      : dvar
VARIABLES  : collection(var−dvar)
FOUND      : collection(var−dvar)

**Restrictions**

$sv \leq 5 \vee rv \leq 2 \Rightarrow$ VALUE $= -\infty$
VALUE $= -\infty \vee$ VALUE $\geq 3 * minv + 3$
VALUE $\leq 3 * maxv - 3$ ①
required(VARIABLES, var)
required(FOUND, var)
where
maxv $=$ maxval(VARIABLES.var)
minv $=$ minval(VARIABLES.var)
sv $= |$VARIABLES$|$
rv $=$ range(VARIABLES.var)

**Purpose**

The constraint MAX_SURF_DIP_ON_INCREASING_SEQUENCE(VALUE, VARIABLES) holds. In addition, FOUND is a collection of 0/1 variables where the value 1 indicates the position of the **found** letter in those occurrences of the pattern DIP_ON_INCREASING_SEQUENCE for which the feature value is VALUE.

The position of the **found** letter in an occurrence of a pattern is the first position where the occurrence of pattern is identified, even if the pattern is not complete.

An occurrence of the pattern DIP_ON_INCREASING_SEQUENCE is the subsequence which matches the regular expression `<<><<`.

Assume that the occurrence of the pattern DIP_ON_INCREASING_SEQUENCE starts at position $i$ and ends at position $j$. The feature SURF computes the sum of the values from index $i + 2$ to index $j$.

**Example**

$$\left( \begin{array}{l} 10, \langle 1, 2, 3, 2, 5, 6, 7, 4, 1, 3, 4, 6, 1, 2, 4, 4 \rangle, \\ \langle 0, 0, 0, 0, 1, 0, 0, 0, 0, 0, 0, 0, 0, 0, 0, 0 \rangle \end{array} \right)$$

Figure 3.1009 provides an example where the POS_MAX_SURF_DIP_ON_INCREASING_SEQUENCE $(10, [1, 2, 3, 2, 5, 6, 7, 4, 1, 3, 4, 6, 1, 2, 4, 4], [0, 0, 0, 0, 1, 0, 0, 0, 0, 0, 0, 0, 0, 0, 0, 0])$ constraint holds.

**Typical**

$|$VARIABLES$| > 5$
range(VARIABLES.var) $> 2$

Figure 3.1009: Illustrating the POS_MAX_SURF_DIP_ON_INCREASING_SEQUENCE constraint of the **Example** slot

| Arg. properties | • Functional dependency: VALUE determined by VARIABLES. |
| | • Functional dependency: FOUND determined by VARIABLES. |

**Automaton**    Similar to the automaton of the MAX_SURF_DIP_ON_INCREASING_SEQUENCE constraint but use the decoration table 2.33.

# 3.445 POS_MAX_SURF_GORGE

► ▷ ◁ ◄        DESCRIPTION        AUTOMATON

$(> | > (= | >)^*>)(< | < (= | <)^* <)$

**Origin**      Based on constraint MAX_SURF_GORGE.

**Constraint**  POS_MAX_SURF_GORGE(VALUE, VARIABLES, FOUND)

**Arguments**
```
VALUE      : dvar
VARIABLES  : collection(var−dvar)
FOUND      : collection(var−dvar)
```

**Restrictions**
$\mathtt{sv} \leq 2 \vee \mathtt{rv} \leq 1 \Rightarrow \mathtt{VALUE} = -\infty$
$\mathtt{rv} = 2 \Rightarrow \mathtt{VALUE} = -\infty \vee \mathtt{VALUE} \geq \mathtt{minv}$
$\mathtt{rv} \geq 3 \Rightarrow \mathtt{VALUE} = -\infty \vee \mathtt{VALUE} \geq \min(\mathtt{minv}, (\mathtt{sv} - 2) * (\mathtt{minv} + 1) - 1)$
$\mathtt{rv} = 2 \Rightarrow \mathtt{VALUE} \leq \mathtt{maxv} - 1 ①$
$\mathtt{rv} \geq 3 \Rightarrow \mathtt{VALUE} \leq \max(\mathtt{maxv} - 1②, (\mathtt{sv} - 2) * (\mathtt{maxv} - 1) - 1③)$
required(VARIABLES, var)
required(FOUND, var)
where
$\mathtt{maxv} = $ maxval(VARIABLES.var)
$\mathtt{rv} = $ range(VARIABLES.var)
$\mathtt{sv} = |\mathtt{VARIABLES}|$
$\mathtt{minv} = $ minval(VARIABLES.var)

**Purpose**

The constraint MAX_SURF_GORGE(VALUE, VARIABLES) holds. In addition, FOUND is a collection of 0/1 variables where the value 1 indicates the position of the **found** letter in those occurrences of the pattern GORGE for which the feature value is VALUE.

The position of the **found** letter in an occurrence of a pattern is the first position where the occurrence of pattern is identified, even if the pattern is not complete.

An occurrence of the pattern GORGE is the *maximal* subsequence which matches the regular expression $(> | > (= | >)^*>)(< | < (= | <)^* <)$.

Assume that the occurrence of the pattern GORGE starts at position $i$ and ends at position $j$. The feature SURF computes the sum of the values from index $i + 1$ to index $j$.

**Example**

$$\left( \begin{array}{l} 11, \langle 1, 7, 3, 4, 4, 5, 5, 4, 2, 2, 6, 5, 4, 6, 5, 7 \rangle, \\ \langle 0, 0, 1, 0, 0, 0, 0, 0, 0, 0, 0, 0, 0, 0, 0, 0 \rangle \end{array} \right)$$

Figure 3.1010 provides an example where the POS_MAX_SURF_GORGE $(11, [1, 7, 3, 4, 4, 5, 5, 4, 2, 2, 6, 5, 4, 6, 5, 7], [0, 0, 1, 0, 0, 0, 0, 0, 0, 0, 0, 0, 0, 0, 0, 0])$ constraint holds.

**Typical**

$|\mathtt{VARIABLES}| > 2$
range(VARIABLES.var) $> 1$

Figure 3.1010: Illustrating the POS_MAX_SURF_GORGE constraint of the **Example** slot

**Arg. properties**
- Functional dependency: VALUE determined by VARIABLES.
- Functional dependency: FOUND determined by VARIABLES.

**Automaton**            Similar to the automaton of the MAX_SURF_GORGE constraint but use the decoration table 2.33.

# 3.446  POS_MAX_SURF_INCREASING

► ▷ ◁ ◀          **DESCRIPTION**                **AUTOMATON**

| | |
|---|---|
| **Origin** | Based on constraint MAX_SURF_INCREASING. |
| **Constraint** | POS_MAX_SURF_INCREASING(VALUE, VARIABLES, FOUND) |

| **Arguments** | | | |
|---|---|---|---|
| | VALUE | : | dvar |
| | VARIABLES | : | collection(var−dvar) |
| | FOUND | : | collection(var−dvar) |

**Restrictions**

$\mathtt{sv} \leq 1 \vee \mathtt{rv} \leq 1 \Rightarrow \mathtt{VALUE} = -\infty$

$\mathtt{VALUE} = -\infty \vee \mathtt{VALUE} \geq 2 * \mathtt{minv} + 1$

$\mathtt{VALUE} \leq 2 * \mathtt{maxv} - 1$ ①

required(VARIABLES, var)

required(FOUND, var)

where

$\mathtt{maxv} = \mathtt{maxval}(\mathtt{VARIABLES.var})$

$\mathtt{minv} = \mathtt{minval}(\mathtt{VARIABLES.var})$

$\mathtt{sv} = |\mathtt{VARIABLES}|$

$\mathtt{rv} = \mathtt{range}(\mathtt{VARIABLES.var})$

**Purpose**

The constraint MAX_SURF_INCREASING(VALUE, VARIABLES) holds. In addition, FOUND is a collection of 0/1 variables where the value 1 indicates the position of the **found** letter in those occurrences of the pattern INCREASING for which the feature value is VALUE.

The position of the **found** letter in an occurrence of a pattern is the first position where the occurrence of pattern is identified, even if the pattern is not complete.

An occurrence of the pattern INCREASING is the subsequence which matches the regular expression <.

Assume that the occurrence of the pattern INCREASING starts at position $i$ and ends at position $j$. The feature SURF computes the sum of the values from index $i$ to index $j+1$.

**Example**

$$\left( \begin{array}{l} 10, \langle 4, 3, 5, 5, 2, 1, 1, 3, 3, 4, 6, 6, 3, 1, 3, 3 \rangle, \\ \langle 0, 0, 0, 0, 0, 0, 0, 0, 0, 0, 1, 0, 0, 0, 0, 0 \rangle \end{array} \right)$$

Figure 3.1011 provides an example where the POS_MAX_SURF_INCREASING $(10, [4, 3, 5, 5, 2, 1, 1, 3, 3, 4, 6, 6, 3, 1, 3, 3], [0, 0, 0, 0, 0, 0, 0, 0, 0, 0, 1, 0, 0, 0, 0, 0])$ constraint holds.

**Typical**

$|\mathtt{VARIABLES}| > 1$

$\mathtt{range}(\mathtt{VARIABLES.var}) > 1$

Figure 3.1011: Illustrating the POS_MAX_SURF_INCREASING constraint of the **Example** slot

**Arg. properties**

- Functional dependency: VALUE determined by VARIABLES.
- Functional dependency: FOUND determined by VARIABLES.

**Automaton**   Similar to the automaton of the MAX_SURF_INCREASING constraint but use the decoration table 2.33.

# 3.447 POS_MAX_SURF_INCREASING_SEQUENCE

▶ ▷ ◁ ◀      **DESCRIPTION**      **AUTOMATON**

$< (< \mid =)^* < \mid$
$<$

| | |
|---|---|
| **Origin** | Based on constraint MAX_SURF_INCREASING_SEQUENCE. |
| **Constraint** | POS_MAX_SURF_INCREASING_SEQUENCE(VALUE, VARIABLES, FOUND) |

| | | |
|---|---|---|
| **Arguments** | VALUE | : dvar |
| | VARIABLES | : collection(var−dvar) |
| | FOUND | : collection(var−dvar) |

**Restrictions**

$\mathtt{sv} \leq 1 \vee \mathtt{rv} \leq 1 \Rightarrow \mathtt{VALUE} = -\infty$

$\mathtt{rv} = 2 \Rightarrow \mathtt{VALUE} = -\infty \vee \mathtt{VALUE} \geq 2 * \mathtt{minv} + 1$

$\mathtt{rv} \geq 3 \Rightarrow \mathtt{VALUE} = -\infty \vee \mathtt{VALUE} \geq \min(2 * \mathtt{minv} + 1, \mathtt{sv} * (\mathtt{minv} + 1))$

$\mathtt{rv} = 2 \Rightarrow \mathtt{VALUE} \leq 2 * \mathtt{maxv} - 1 \text{①}$

$\mathtt{rv} \geq 3 \Rightarrow \mathtt{VALUE} \leq \max(2 * \mathtt{maxv} - 1 \text{②}, \mathtt{sv} * (\mathtt{maxv} - 1) \text{③})$

required(VARIABLES, var)

required(FOUND, var)

where

maxv = maxval(VARIABLES.var)

rv = range(VARIABLES.var)

sv = |VARIABLES|

minv = minval(VARIABLES.var)

VALUE ≤ **5** = 2 ∗ 3 − 1 ①

VALUE ≤ **−5** = −2 ∗ 2 − 1 ②

VALUE ≤ **16** = 8 ∗ (3 − 1) ③

**Purpose**

The constraint MAX_SURF_INCREASING_SEQUENCE(VALUE, VARIABLES) holds. In addition, FOUND is a collection of 0/1 variables where the value 1 indicates the position of the **found** letter in those occurrences of the pattern INCREASING_SEQUENCE for which the feature value is VALUE.

The position of the **found** letter in an occurrence of a pattern is the first position where the occurrence of pattern is identified, even if the pattern is not complete.

An occurrence of the pattern INCREASING_SEQUENCE is the *maximal* subsequence which matches the regular expression $< (< \mid =)^* < \mid <$.

Assume that the occurrence of the pattern INCREASING_SEQUENCE starts at position $i$ and ends at position $j$. The feature SURF computes the sum of the values from index $i$ to index $j + 1$.

**Example**

$$\left( \begin{array}{l} 17, \langle 4, 3, 5, 5, 2, 1, 1, 3, 3, 4, 6, 6, 3, 1, 1, 3, 3 \rangle, \\ \langle 0, 0, 0, 0, 0, 0, 1, 0, 0, 0, 0, 0, 0, 0, 0, 0, 0 \rangle \end{array} \right)$$

Figure 3.1012 provides an example where the POS_MAX_SURF_INCREASING_SEQUENCE $(17, [4, 3, 5, 5, 2, 1, 1, 3, 3, 4, 6, 6, 3, 1, 1, 3, 3], [0, 0, 0, 0, 0, 0, 1, 0, 0, 0, 0, 0, 0, 0, 0, 0, 0])$ constraint holds.

Figure 3.1012: Illustrating the POS_MAX_SURF_INCREASING_SEQUENCE constraint of the **Example** slot

| **Typical** | $|\texttt{VARIABLES}| > 1$ |
| | $\text{range}(\texttt{VARIABLES.var}) > 1$ |
| | |
| **Arg. properties** | • Functional dependency: VALUE determined by VARIABLES. |
| | • Functional dependency: FOUND determined by VARIABLES. |

**Automaton**        Similar to the automaton of the MAX_SURF_INCREASING_SEQUENCE constraint but use
                     the decoration table 2.33.

## 3.448 POS_MAX_SURF_INCREASING_TERRACE

**Origin**  Based on constraint MAX_SURF_INCREASING_TERRACE.

**Constraint**  POS_MAX_SURF_INCREASING_TERRACE(VALUE, VARIABLES, FOUND)

**Arguments**
```
VALUE     : dvar
VARIABLES : collection(var−dvar)
FOUND     : collection(var−dvar)
```

**Restrictions**

$\mathtt{sv} \le 3 \lor \mathtt{rv} \le 2 \Rightarrow \mathtt{VALUE} = -\infty$

$\mathtt{VALUE} = -\infty \lor \mathtt{VALUE} \ge \min(2 * (\mathtt{minv} + 1), (\mathtt{sv} - 2) * (\mathtt{minv} + 1))$

$\mathtt{VALUE} \le \max(2 * (\mathtt{maxv} - 1)①, (\mathtt{sv} - 2) * (\mathtt{maxv} - 1)②)$

required(VARIABLES, var)

required(FOUND, var)

where

$\mathtt{maxv} = \mathtt{maxval}(\mathtt{VARIABLES.var})$
$\mathtt{sv} = |\mathtt{VARIABLES}|$
$\mathtt{minv} = \mathtt{minval}(\mathtt{VARIABLES.var})$
$\mathtt{rv} = \mathtt{range}(\mathtt{VARIABLES.var})$

**Purpose**

The constraint MAX_SURF_INCREASING_TERRACE(VALUE, VARIABLES) holds. In addition, FOUND is a collection of 0/1 variables where the value 1 indicates the position of the **found** letter in those occurrences of the pattern INCREASING_TERRACE for which the feature value is VALUE.

The position of the **found** letter in an occurrence of a pattern is the first position where the occurrence of pattern is identified, even if the pattern is not complete.

An occurrence of the pattern INCREASING_TERRACE is the *maximal* subsequence which matches the regular expression $<=^+<$.

Assume that the occurrence of the pattern INCREASING_TERRACE starts at position $i$ and ends at position $j$. The feature SURF computes the sum of the values from index $i + 1$ to index $j$.

**Example**

$$\left( \begin{array}{l} 10, \langle 1, 3, 3, 3, 2, 5, 5, 6, 4, 4, 2, 3, 3, 3, 4, 4 \rangle, \\ \langle 0, 0, 0, 0, 0, 0, 1, 0, 0, 0, 0, 0, 0, 0, 0, 0 \rangle \end{array} \right)$$

Figure 3.1013 provides an example where the POS_MAX_SURF_INCREASING_TERRACE $(10, [1, 3, 3, 3, 2, 5, 5, 6, 4, 4, 2, 3, 3, 3, 4, 4], [0, 0, 0, 0, 0, 0, 1, 0, 0, 0, 0, 0, 0, 0, 0, 0])$ constraint holds.

**Typical**

$|\mathtt{VARIABLES}| > 3$

range(VARIABLES.var) $> 2$

Figure 3.1013: Illustrating the POS_MAX_SURF_INCREASING_TERRACE constraint of the **Example** slot

**Arg. properties**
- Functional dependency: VALUE determined by VARIABLES.
- Functional dependency: FOUND determined by VARIABLES.

**Automaton**          Similar to the automaton of the MAX_SURF_INCREASING_TERRACE constraint but use the decoration table 2.33.

## 3.449  POS_MAX_SURF_INFLEXION

► ▷ ◁ ◀        **DESCRIPTION**        **AUTOMATON**

$< (< | =)^* > | > (> | =)^* <$

| | |
|---|---|
| **Origin** | Based on constraint MAX_SURF_INFLEXION. |
| **Constraint** | POS_MAX_SURF_INFLEXION(VALUE, VARIABLES, FOUND) |
| **Arguments** | VALUE : dvar<br>VARIABLES : collection(var−dvar)<br>FOUND : collection(var−dvar) |

**Restrictions**

$$\mathtt{sv} \leq 2 \vee \mathtt{rv} \leq 1 \Rightarrow \mathtt{VALUE} = -\infty$$
$$\mathtt{VALUE} = -\infty \vee \mathtt{VALUE} \geq \min(\mathtt{minv}, (\mathtt{sv}-2) * \mathtt{minv})$$
$$\mathtt{VALUE} \leq \max(\mathtt{maxv}①, (\mathtt{sv}-2) * \mathtt{maxv}②)$$
$$\mathtt{required}(\mathtt{VARIABLES}, \mathtt{var})$$
$$\mathtt{required}(\mathtt{FOUND}, \mathtt{var})$$
where
$$\mathtt{maxv} = \mathtt{maxval}(\mathtt{VARIABLES.var})$$
$$\mathtt{sv} = |\mathtt{VARIABLES}|$$
$$\mathtt{minv} = \mathtt{minval}(\mathtt{VARIABLES.var})$$
$$\mathtt{rv} = \mathtt{range}(\mathtt{VARIABLES.var})$$

VALUE ≤ −1

①

VALUE ≤ 12 = (8 − 2) * 2

②

**Purpose**

The constraint MAX_SURF_INFLEXION(VALUE, VARIABLES) holds. In addition, FOUND is a collection of 0/1 variables where the value 1 indicates the position of the **found** letter in those occurrences of the pattern INFLEXION for which the feature value is VALUE.

The position of the **found** letter in an occurrence of a pattern is the first position where the occurrence of pattern is identified, even if the pattern is not complete.

An occurrence of the pattern INFLEXION is the *maximal* subsequence which matches the regular expression $< (< | =)^* > | > (> | =)^* <$.

Assume that the occurrence of the pattern INFLEXION starts at position $i$ and ends at position $j$. The feature SURF computes the sum of the values from index $i + 1$ to index $j$.

**Example**

$$\left( \begin{array}{l} 14, \langle 1, 2, 6, 6, 4, 4, 3, 5, 2, 5, 1, 5, 3, 3, 4, 4 \rangle, \\ \langle 0, 0, 0, 1, 0, 0, 0, 0, 0, 0, 0, 0, 0, 0, 0, 0 \rangle \end{array} \right)$$

Figure 3.1014 provides an example where the POS_MAX_SURF_INFLEXION $(14, [1, 2, 6, 6, 4, 4, 3, 5, 2, 5, 1, 5, 3, 3, 4, 4], [0, 0, 0, 1, 0, 0, 0, 0, 0, 0, 0, 0, 0, 0, 0, 0])$ constraint holds.

**Typical**

$$|\mathtt{VARIABLES}| > 2$$
$$\mathtt{range}(\mathtt{VARIABLES.var}) > 1$$

Figure 3.1014: Illustrating the POS_MAX_SURF_INFLEXION constraint of the **Example** slot

**Arg. properties**
- Functional dependency: VALUE determined by VARIABLES.
- Functional dependency: FOUND determined by VARIABLES.

**Automaton**       Similar to the automaton of the MAX_SURF_INFLEXION constraint but use the decoration
                    table 2.33.

## 3.450 POS_MAX_SURF_PEAK

▶ ▷ ◁ ◀    **DESCRIPTION**    **AUTOMATON**

| | |
|---|---|
| **Origin** | Based on constraint MAX_SURF_PEAK. |
| **Constraint** | POS_MAX_SURF_PEAK(VALUE, VARIABLES, FOUND) |
| **Arguments** | VALUE : dvar<br>VARIABLES : collection(var−dvar)<br>FOUND : collection(var−dvar) |

**Restrictions**

$$\mathtt{sv} \leq 2 \lor \mathtt{rv} \leq 1 \Rightarrow \mathtt{VALUE} = -\infty$$
$$\mathtt{VALUE} = -\infty \lor \mathtt{VALUE} \geq \min(\mathtt{minv}+1, (\mathtt{sv}-2)*(\mathtt{minv}+1))$$
$$\mathtt{VALUE} \leq \max(\mathtt{maxv}①, (\mathtt{sv}-2)*\mathtt{maxv}②)$$
required(VARIABLES, var)
required(FOUND, var)
where
$$\mathtt{minv} = \mathtt{minval}(\mathtt{VARIABLES.var})$$
$$\mathtt{maxv} = \mathtt{maxval}(\mathtt{VARIABLES.var})$$
$$\mathtt{sv} = |\mathtt{VARIABLES}|$$
$$\mathtt{rv} = \mathtt{range}(\mathtt{VARIABLES.var})$$

**Purpose**

The constraint MAX_SURF_PEAK(VALUE, VARIABLES) holds. In addition, FOUND is a collection of 0/1 variables where the value 1 indicates the position of the **found** letter in those occurrences of the pattern PEAK for which the feature value is VALUE.

The position of the **found** letter in an occurrence of a pattern is the first position where the occurrence of pattern is identified, even if the pattern is not complete.

An occurrence of the pattern PEAK is the *maximal* subsequence which matches the regular expression $< (= | <)^* (> | =)^* >$.

Assume that the occurrence of the pattern PEAK starts at position $i$ and ends at position $j$. The feature SURF computes the sum of the values from index $i + 1$ to index $j$.

**Example**

$$\left( \begin{array}{l} 14, \langle 7, 5, 5, 1, 4, 5, 2, 2, 3, 5, 6, 2, 3, 3, 3, 1 \rangle, \\ \langle 0, 0, 0, 0, 0, 0, 0, 0, 0, 0, 0, 1, 0, 0, 0, 0, 0 \rangle \end{array} \right)$$

Figure 3.1015 provides an example where the POS_MAX_SURF_PEAK $(14, [7, 5, 5, 1, 4, 5, 2, 2, 3, 5, 6, 2, 3, 3, 3, 1], [0, 0, 0, 0, 0, 0, 0, 0, 0, 0, 0, 1, 0, 0, 0, 0, 0])$ constraint holds.

**Typical**

$$|\mathtt{VARIABLES}| > 2$$
$$\mathtt{range}(\mathtt{VARIABLES.var}) > 1$$

**Arg. properties**

- Functional dependency: VALUE determined by VARIABLES.
- Functional dependency: FOUND determined by VARIABLES.

Figure 3.1015: Illustrating the POS_MAX_SURF_PEAK constraint of the **Example** slot

**Automaton**        Similar to the automaton of the MAX_SURF_PEAK constraint but use the decoration table 2.33.

PL

# 3.451 POS_MAX_SURF_PLAIN

▶ ▷ ◁ ◀     DESCRIPTION     AUTOMATON

| | |
|---|---|
| **Origin** | Based on constraint MAX_SURF_PLAIN. |
| **Constraint** | POS_MAX_SURF_PLAIN(VALUE, VARIABLES, FOUND) |

**Arguments**

```
VALUE     : dvar
VARIABLES : collection(var-dvar)
FOUND     : collection(var-dvar)
```

**Restrictions**

$$\mathrm{sv} \leq 2 \vee \mathrm{rv} \leq 1 \Rightarrow \mathtt{VALUE} = -\infty$$
$$\mathtt{VALUE} = -\infty \vee \mathtt{VALUE} \geq \min(\mathrm{minv}, (\mathrm{sv} - 2) * \mathrm{minv})$$
$$\mathtt{VALUE} \leq \max(\mathrm{maxv} - 1 ①, (\mathrm{sv} - 2) * (\mathrm{maxv} - 1) ②)$$
required(VARIABLES, var)
required(FOUND, var)
where
$$\mathrm{maxv} = \mathtt{maxval}(\mathtt{VARIABLES.var})$$
$$\mathrm{sv} = |\mathtt{VARIABLES}|$$
$$\mathrm{minv} = \mathtt{minval}(\mathtt{VARIABLES.var})$$
$$\mathrm{rv} = \mathtt{range}(\mathtt{VARIABLES.var})$$

**Purpose**

The constraint MAX_SURF_PLAIN(VALUE, VARIABLES) holds. In addition, FOUND is a collection of 0/1 variables where the value 1 indicates the position of the **found** letter in those occurrences of the pattern PLAIN for which the feature value is VALUE.

The position of the **found** letter in an occurrence of a pattern is the first position where the occurrence of pattern is identified, even if the pattern is not complete.

An occurrence of the pattern PLAIN is the *maximal* subsequence which matches the regular expression >=*<.

Assume that the occurrence of the pattern PLAIN starts at position $i$ and ends at position $j$. The feature SURF computes the sum of the values from index $i + 1$ to index $j$.

**Example**

$$\left( \begin{array}{l} 6, \langle 2, 3, 6, 5, 7, 6, 6, 4, 5, 5, 4, 3, 3, 6, 6, 3 \rangle, \\ \langle 0, 0, 0, 0, 0, 0, 0, 0, 0, 0, 0, 0, 0, 1, 0, 0, 0 \rangle \end{array} \right)$$

Figure 3.1016 provides an example where the POS_MAX_SURF_PLAIN $(6, [2, 3, 6, 5, 7, 6, 6, 4, 5, 5, 4, 3, 3, 6, 6, 3], [0, 0, 0, 0, 0, 0, 0, 0, 0, 0, 0, 0, 0, 1, 0, 0, 0])$ constraint holds.

**Typical**

$$|\mathtt{VARIABLES}| > 2$$
$$\mathtt{range}(\mathtt{VARIABLES.var}) > 1$$

**Arg. properties**

- Functional dependency: VALUE determined by VARIABLES.
- Functional dependency: FOUND determined by VARIABLES.

Figure 3.1016: Illustrating the POS_MAX_SURF_PLAIN constraint of the **Example** slot

**Automaton**   Similar to the automaton of the MAX_SURF_PLAIN constraint but use the decoration table 2.33.

## 3.452 POS_MAX_SURF_PLATEAU

| | |
|---|---|
| **Origin** | Based on constraint MAX_SURF_PLATEAU. |
| **Constraint** | POS_MAX_SURF_PLATEAU(VALUE, VARIABLES, FOUND) |

| | | |
|---|---|---|
| **Arguments** | VALUE | : dvar |
| | VARIABLES | : collection(var−dvar) |
| | FOUND | : collection(var−dvar) |

**Restrictions**

$$\text{sv} \leq 2 \vee \text{rv} \leq 1 \Rightarrow \text{VALUE} = -\infty$$
$$\text{VALUE} = -\infty \vee \text{VALUE} \geq \min(\text{minv} + 1, (\text{sv} - 2) * (\text{minv} + 1))$$
$$\text{VALUE} \leq \max(\text{maxv}①, (\text{sv} - 2) * \text{maxv}②)$$
required(VARIABLES, var)
required(FOUND, var)
where
$$\text{maxv} = \text{maxval}(\text{VARIABLES.var})$$
$$\text{sv} = |\text{VARIABLES}|$$
$$\text{minv} = \text{minval}(\text{VARIABLES.var})$$
$$\text{rv} = \text{range}(\text{VARIABLES.var})$$

**Purpose**

The constraint MAX_SURF_PLATEAU(VALUE, VARIABLES) holds. In addition, FOUND is a collection of 0/1 variables where the value 1 indicates the position of the **found** letter in those occurrences of the pattern PLATEAU for which the feature value is VALUE.

The position of the **found** letter in an occurrence of a pattern is the first position where the occurrence of pattern is identified, even if the pattern is not complete.

An occurrence of the pattern PLATEAU is the *maximal* subsequence which matches the regular expression $< =^* >$.

Assume that the occurrence of the pattern PLATEAU starts at position $i$ and ends at position $j$. The feature SURF computes the sum of the values from index $i + 1$ to index $j$.

**Example**

$$\left( \begin{array}{l} 10, \langle 7, 5, 2, 3, 1, 2, 2, 4, 3, 3, 4, 5, 5, 2, 2, 5 \rangle, \\ \langle 0, 0, 0, 0, 0, 0, 0, 0, 0, 0, 0, 0, 0, 1, 0, 0, 0 \rangle \end{array} \right)$$

Figure 3.1017 provides an example where the POS_MAX_SURF_PLATEAU $(10, [7, 5, 2, 3, 1, 2, 2, 4, 3, 3, 4, 5, 5, 2, 2, 5], [0, 0, 0, 0, 0, 0, 0, 0, 0, 0, 0, 0, 0, 1, 0, 0, 0])$ constraint holds.

**Typical**

$$|\text{VARIABLES}| > 2$$
$$\text{range}(\text{VARIABLES.var}) > 1$$

Figure 3.1017: Illustrating the POS_MAX_SURF_PLATEAU constraint of the **Example** slot

**Arg. properties**

- Functional dependency: VALUE determined by VARIABLES.
- Functional dependency: FOUND determined by VARIABLES.

**Automaton**    Similar to the automaton of the MAX_SURF_PLATEAU constraint but use the decoration table 2.33.

# 3.453 POS_MAX_SURF_PROPER_PLAIN

▶ ▷ ◁ ◀     **DESCRIPTION**     **AUTOMATON**

| | |
|---|---|
| **Origin** | Based on constraint MAX_SURF_PROPER_PLAIN. |
| **Constraint** | POS_MAX_SURF_PROPER_PLAIN(VALUE, VARIABLES, FOUND) |

**Arguments**

```
VALUE     : dvar
VARIABLES : collection(var−dvar)
FOUND     : collection(var−dvar)
```

**Restrictions**

$sv \leq 3 \lor rv \leq 1 \Rightarrow VALUE = -\infty$
$VALUE = -\infty \lor VALUE \geq \min(2 * minv, (sv - 2) * minv)$
$VALUE \leq \max(2 * (maxv - 1) ①, (sv - 2) * (maxv - 1) ②)$
required(VARIABLES, var)
required(FOUND, var)
where
$maxv = maxval(VARIABLES.var)$
$sv = |VARIABLES|$
$minv = minval(VARIABLES.var)$
$rv = range(VARIABLES.var)$

**Purpose**

The constraint MAX_SURF_PROPER_PLAIN(VALUE, VARIABLES) holds. In addition, FOUND is a collection of 0/1 variables where the value 1 indicates the position of the **found** letter in those occurrences of the pattern PROPER_PLAIN for which the feature value is VALUE.

The position of the **found** letter in an occurrence of a pattern is the first position where the occurrence of pattern is identified, even if the pattern is not complete.

An occurrence of the pattern PROPER_PLAIN is the *maximal* subsequence which matches the regular expression $> =^+ <$.

Assume that the occurrence of the pattern PROPER_PLAIN starts at position $i$ and ends at position $j$. The feature SURF computes the sum of the values from index $i + 1$ to index $j$.

**Example**

$$\left( \begin{array}{l} 10, \langle 2, 7, 5, 5, 6, 3, 7, 4, 4, 5, 6, 5, 3, 3, 3, 5 \rangle, \\ \langle 0, 0, 0, 1, 0, 0, 0, 0, 0, 0, 0, 0, 0, 0, 0, 0 \rangle \end{array} \right)$$

Figure 3.1018 provides an example where the POS_MAX_SURF_PROPER_PLAIN $(10, [2, 7, 5, 5, 6, 3, 7, 4, 4, 5, 6, 5, 3, 3, 3, 5], [0, 0, 0, 1, 0, 0, 0, 0, 0, 0, 0, 0, 0, 0, 0, 0])$ constraint holds.

**Typical**

$|VARIABLES| > 3$
$range(VARIABLES.var) > 1$

Figure 3.1018: Illustrating the POS_MAX_SURF_PROPER_PLAIN constraint of the **Example** slot

**Arg. properties**
- Functional dependency: VALUE determined by VARIABLES.
- Functional dependency: FOUND determined by VARIABLES.

**Automaton**    Similar to the automaton of the MAX_SURF_PROPER_PLAIN constraint but use the decoration table 2.33.

# 3.454 POS_MAX_SURF_PROPER_PLATEAU

▶ ▷ ◁ ◀          DESCRIPTION          AUTOMATON

| | |
|---|---|
| **Origin** | Based on constraint MAX_SURF_PROPER_PLATEAU. |
| **Constraint** | POS_MAX_SURF_PROPER_PLATEAU(VALUE, VARIABLES, FOUND) |

| **Arguments** | | |
|---|---|---|
| | VALUE | : dvar |
| | VARIABLES | : collection(var−dvar) |
| | FOUND | : collection(var−dvar) |

**Restrictions**

$sv \leq 3 \vee rv \leq 1 \Rightarrow$ VALUE $= -\infty$

VALUE $= -\infty \vee$ VALUE $\geq \min(2 * (minv + 1), (sv - 2) * (minv + 1))$

VALUE $\leq \max(2 * maxv①, (sv - 2) * maxv②)$

required(VARIABLES, var)

required(FOUND, var)

where

maxv = maxval(VARIABLES.var)
sv = |VARIABLES|
minv = minval(VARIABLES.var)
rv = range(VARIABLES.var)

**Purpose**

The constraint MAX_SURF_PROPER_PLATEAU(VALUE, VARIABLES) holds. In addition, FOUND is a collection of 0/1 variables where the value 1 indicates the position of the **found** letter in those occurrences of the pattern PROPER_PLATEAU for which the feature value is VALUE.

The position of the **found** letter in an occurrence of a pattern is the first position where the occurrence of pattern is identified, even if the pattern is not complete.

An occurrence of the pattern PROPER_PLATEAU is the *maximal* subsequence which matches the regular expression $<=^+>$.

Assume that the occurrence of the pattern PROPER_PLATEAU starts at position $i$ and ends at position $j$. The feature SURF computes the sum of the values from index $i+1$ to index $j$.

**Example**

$$\left( \begin{array}{l} 15, \langle 7, 1, 3, 3, 2, 5, 1, 4, 4, 3, 2, 3, 5, 5, 5, 3 \rangle, \\ \langle 0, 0, 0, 0, 0, 0, 0, 0, 0, 0, 0, 0, 0, 0, 1, 0 \rangle \end{array} \right)$$

Figure 3.1019 provides an example where the POS_MAX_SURF_PROPER_PLATEAU $(15, [7, 1, 3, 3, 2, 5, 1, 4, 4, 3, 2, 3, 5, 5, 5, 3], [0, 0, 0, 0, 0, 0, 0, 0, 0, 0, 0, 0, 0, 0, 1, 0])$ constraint holds.

**Typical**

|VARIABLES| $> 3$
range(VARIABLES.var) $> 1$

Figure 3.1019: Illustrating the POS_MAX_SURF_PROPER_PLATEAU constraint of the **Example** slot

**Arg. properties**
- Functional dependency: VALUE determined by VARIABLES.
- Functional dependency: FOUND determined by VARIABLES.

**Automaton**    Similar to the automaton of the MAX_SURF_PROPER_PLATEAU constraint but use the decoration table 2.33.

# 3.455 POS_MAX_SURF_STEADY

▶ ▷ ◁ ◀        DESCRIPTION        AUTOMATON

**Origin**        Based on constraint MAX_SURF_STEADY.

**Constraint**        POS_MAX_SURF_STEADY(VALUE, VARIABLES, FOUND)

**Arguments**
```
VALUE     : dvar
VARIABLES : collection(var−dvar)
FOUND     : collection(var−dvar)
```

**Restrictions**
$\text{sv} \leq 1 \Rightarrow \text{VALUE} = -\infty$
$\text{VALUE} = -\infty \vee \text{VALUE} \geq 2 * \text{minv}$
$\text{VALUE} \leq 2 * \text{maxv}①$
required(VARIABLES, var)
required(FOUND, var)
where
$\text{maxv} = \text{maxval}(\text{VARIABLES.var})$
$\text{minv} = \text{minval}(\text{VARIABLES.var})$
$\text{sv} = |\text{VARIABLES}|$

**Purpose**

The constraint MAX_SURF_STEADY(VALUE, VARIABLES) holds. In addition, FOUND is a collection of 0/1 variables where the value 1 indicates the position of the **found** letter in those occurrences of the pattern STEADY for which the feature value is VALUE.

The position of the **found** letter in an occurrence of a pattern is the first position where the occurrence of pattern is identified, even if the pattern is not complete.

An occurrence of the pattern STEADY is the subsequence which matches the regular expression =.

Assume that the occurrence of the pattern STEADY starts at position $i$ and ends at position $j$. The feature SURF computes the sum of the values from index $i$ to index $j + 1$.

**Example**

$$\left( \begin{array}{l} 12, \langle 1, 1, 7, 3, 3, 5, 5, 5, 6, 5, 5, 5, 7, 2, 6, 6 \rangle , \\ \langle 0, 0, 0, 0, 0, 0, 0, 0, 0, 0, 0, 0, 0, 0, 1, 0 \rangle \end{array} \right)$$

Figure 3.1020 provides an example where the POS_MAX_SURF_STEADY $(12, [1, 1, 7, 3, 3, 5, 5, 5, 6, 5, 5, 5, 7, 2, 6, 6], [0, 0, 0, 0, 0, 0, 0, 0, 0, 0, 0, 0, 0, 0, 1, 0])$ constraint holds.

**Typical**        $|\text{VARIABLES}| > 1$

**Arg. properties**

- Functional dependency: VALUE determined by VARIABLES.
- Functional dependency: FOUND determined by VARIABLES.

Figure 3.1020: Illustrating the POS_MAX_SURF_STEADY constraint of the **Example** slot

**Automaton**          Similar to the automaton of the MAX_SURF_STEADY constraint but use the decoration table 2.33.

# 3.456   POS_MAX_SURF_STEADY_SEQUENCE

▶ ▷ ◁ ◀        **DESCRIPTION**        **AUTOMATON**

**Origin**          Based on constraint MAX_SURF_STEADY_SEQUENCE.

**Constraint**      POS_MAX_SURF_STEADY_SEQUENCE(VALUE, VARIABLES, FOUND)

**Arguments**       VALUE      : dvar
                    VARIABLES  : collection(var−dvar)
                    FOUND      : collection(var−dvar)

**Restrictions**    $sv \leq 1 \Rightarrow \text{VALUE} = -\infty$
                    $rv = 1 \Rightarrow \text{VALUE} = -\infty \vee \text{VALUE} \geq sv * \text{minv}$
                    $rv \geq 2 \Rightarrow \text{VALUE} = -\infty \vee \text{VALUE} \geq \min(2 * \text{minv}, sv * \text{minv})$
                    $rv = 1 \Rightarrow \text{VALUE} \leq sv * \text{maxv}$①
                    $rv \geq 2 \Rightarrow \text{VALUE} \leq \max(2 * \text{maxv}$②$, sv * \text{maxv}$③$)$
                    required(VARIABLES, var)
                    required(FOUND, var)
                    where
                    $\text{maxv} = \text{maxval}(\text{VARIABLES.var})$
                    $rv = \text{range}(\text{VARIABLES.var})$
                    $sv = |\text{VARIABLES}|$
                    $\text{minv} = \text{minval}(\text{VARIABLES.var})$

**Purpose**         The constraint MAX_SURF_STEADY_SEQUENCE(VALUE, VARIABLES) holds.  In addition, FOUND is a collection of 0/1 variables where the value 1 indicates the position of the **found** letter in those occurrences of the pattern STEADY_SEQUENCE for which the feature value is VALUE.

The position of the **found** letter in an occurrence of a pattern is the first position where the occurrence of pattern is identified, even if the pattern is not complete.

An occurrence of the pattern STEADY_SEQUENCE is the *maximal* subsequence which matches the regular expression $=^+$.

Assume that the occurrence of the pattern STEADY_SEQUENCE starts at position $i$ and ends at position $j$.  The feature SURF computes the sum of the values from index $i$ to index $j + 1$.

**Example**         $\left( \begin{array}{l} 15, \langle 3, 1, 1, 4, 5, 5, 5, 6, 2, 2, 4, 4, 3, 2, 1, 1 \rangle, \\ \langle 0, 0, 0, 0, 1, 0, 0, 0, 0, 0, 0, 0, 0, 0, 0, 0 \rangle \end{array} \right)$

Figure 3.1021 provides an example where the POS_MAX_SURF_STEADY_SEQUENCE $(15, [3, 1, 1, 4, 5, 5, 5, 6, 2, 2, 4, 4, 3, 2, 1, 1], [0, 0, 0, 0, 1, 0, 0, 0, 0, 0, 0, 0, 0, 0, 0, 0])$ constraint holds.

Figure 3.1021: Illustrating the POS_MAX_SURF_STEADY_SEQUENCE constraint of the **Example** slot

| **Typical** | $|\texttt{VARIABLES}| > 1$ |
|---|---|

| **Arg. properties** | • Functional dependency: VALUE determined by VARIABLES. |
|---|---|
| | • Functional dependency: FOUND determined by VARIABLES. |

**Automaton**    Similar to the automaton of the MAX_SURF_STEADY_SEQUENCE constraint but use the decoration table 2.33.

# 3.457    POS_MAX_SURF_STRICTLY_DECREASING_SEQUENCE

▶ ▷ ◁ ◀            **DESCRIPTION**          **AUTOMATON**

| | |
|---|---|
| **Origin** | Based on constraint MAX_SURF_STRICTLY_DECREASING_SEQUENCE. |
| **Constraint** | POS_MAX_SURF_STRICTLY_DECREASING_SEQUENCE(VALUE, VARIABLES, FOUND) |

| **Arguments** | | |
|---|---|---|
| | VALUE | : dvar |
| | VARIABLES | : collection(var−dvar) |
| | FOUND | : collection(var−dvar) |

**Restrictions**

$\text{sv} \leq 1 \vee \text{rv} \leq 1 \Rightarrow \text{VALUE} = -\infty$

$\text{minv} < 0 \Rightarrow \text{VALUE} = -\infty \vee \text{VALUE} \geq \ell_1 * \text{minv} + \lfloor \ell_1 * (\ell_1 - 1)/2 \rfloor$

$\text{minv} \geq 0 \Rightarrow \text{VALUE} = -\infty \vee \text{VALUE} \geq 2 * \text{minv} + 1$

$\text{maxv} > 0 \Rightarrow \text{VALUE} \leq \ell_2 * \text{maxv} - \lfloor \ell_2 * (\ell_2 - 1)/2 \rfloor$ ①

$\text{maxv} \leq 0 \Rightarrow \text{VALUE} \leq 2 * \text{maxv} - 1$②

required(VARIABLES, var)

required(FOUND, var)

where

$\text{maxv} = \text{maxval}(\text{VARIABLES.var})$

$\text{rv} = \text{range}(\text{VARIABLES.var})$

$\text{sv} = |\text{VARIABLES}|$

$\ell_1 = \min(\min(\text{sv}, \text{rv}), |\text{minv}|)$

$\ell_2 = \min(\min(\text{sv}, \text{rv}), |\text{maxv}|)$

$\text{minv} = \text{minval}(\text{VARIABLES.var})$

VALUE ≤ 3    VALUE ≤ 10    VALUE ≤ 9    VALUE ≤ −3

$\ell_1 = 1$    $\ell_1 = 1$    $\ell_1 = 1$
$\ell_2 = 2$    $\ell_2 = 4$    $\ell_2 = 3$

**Purpose**

The constraint MAX_SURF_STRICTLY_DECREASING_SEQUENCE(VALUE, VARIABLES) holds. In addition, FOUND is a collection of 0/1 variables where the value 1 indicates the position of the **found** letter in those occurrences of the pattern STRICTLY_DECREASING_SEQUENCE for which the feature value is VALUE.

The position of the **found** letter in an occurrence of a pattern is the first position where the occurrence of pattern is identified, even if the pattern is not complete.

An occurrence of the pattern STRICTLY_DECREASING_SEQUENCE is the *maximal* sub-sequence which matches the regular expression $>^+$.

Assume that the occurrence of the pattern STRICTLY_DECREASING_SEQUENCE starts at position $i$ and ends at position $j$. The feature SURF computes the sum of the values from index $i$ to index $j + 1$.

**Example**

$$\left( \begin{array}{l} 13, \langle 4, 4, 6, 4, 1, 1, 3, 4, 4, 6, 6, 5, 2, 2, 4, 3 \rangle, \\ \langle 0, 0, 0, 0, 0, 0, 0, 0, 0, 0, 0, 1, 0, 0, 0, 0, 0 \rangle \end{array} \right)$$

Figure 3.1022 provides an example where the POS_MAX_SURF_STRICTLY_DECREASING_SEQUENCE (13, [4, 4, 6, 4, 1, 1, 3, 4, 4, 6, 6, 5, 2, 2, 4, 3], [0, 0, 0, 0, 0, 0, 0, 0, 0, 0, 0, 1, 0, 0, 0, 0, 0]) constraint holds.

Figure 3.1022: Illustrating the POS_MAX_SURF_STRICTLY_DECREASING_SEQUENCE constraint of the **Example** slot

**Typical**

$$|\texttt{VARIABLES}| > 1$$
$$\texttt{range}(\texttt{VARIABLES.var}) > 1$$

**Arg. properties**

- Functional dependency: VALUE determined by VARIABLES.
- Functional dependency: FOUND determined by VARIABLES.

**Automaton**      Similar to the automaton of the MAX_SURF_STRICTLY_DECREASING_SEQUENCE con-
                   straint but use the decoration table 2.33.

## 3.458   POS_MAX_SURF_STRICTLY_INCREASING_SEQUENCE

► ▷ ◁ ◀        **DESCRIPTION**            **AUTOMATON**

| | |
|---|---|
| **Origin** | Based on constraint MAX_SURF_STRICTLY_INCREASING_SEQUENCE. |
| **Constraint** | POS_MAX_SURF_STRICTLY_INCREASING_SEQUENCE(VALUE, VARIABLES, FOUND) |
| **Arguments** | VALUE : dvar |
| | VARIABLES : collection(var−dvar) |
| | FOUND : collection(var−dvar) |

**Restrictions**

$\text{sv} \leq 1 \vee \text{rv} \leq 1 \Rightarrow \text{VALUE} = -\infty$

$\text{minv} < 0 \Rightarrow \text{VALUE} = -\infty \vee \text{VALUE} \geq \ell_1 * \text{minv} + \lfloor \ell_1 * (\ell_1 - 1)/2 \rfloor$

$\text{minv} \geq 0 \Rightarrow \text{VALUE} = -\infty \vee \text{VALUE} \geq 2 * \text{minv} + 1$

$\text{maxv} > 0 \Rightarrow \text{VALUE} \leq \ell_2 * \text{maxv} - \lfloor \ell_2 * (\ell_2 - 1)/2 \rfloor$ ①

$\text{maxv} \leq 0 \Rightarrow \text{VALUE} \leq 2 * \text{maxv} - 1$ ②

required(VARIABLES, var)

required(FOUND, var)

where

maxv = maxval(VARIABLES.var)

rv = range(VARIABLES.var)

sv = |VARIABLES|

$\ell_1 = \min(\min(\text{sv}, \text{rv}), |\text{minv}|)$

$\ell_2 = \min(\min(\text{sv}, \text{rv}), |\text{maxv}|)$

minv = minval(VARIABLES.var)

**Purpose**

The constraint MAX_SURF_STRICTLY_INCREASING_SEQUENCE(VALUE, VARIABLES) holds. In addition, FOUND is a collection of 0/1 variables where the value 1 indicates the position of the **found** letter in those occurrences of the pattern STRICTLY_INCREASING_SEQUENCE for which the feature value is VALUE.

The position of the **found** letter in an occurrence of a pattern is the first position where the occurrence of pattern is identified, even if the pattern is not complete.

An occurrence of the pattern STRICTLY_INCREASING_SEQUENCE is the *maximal* subsequence which matches the regular expression $<^+$.

Assume that the occurrence of the pattern STRICTLY_INCREASING_SEQUENCE starts at position $i$ and ends at position $j$. The feature SURF computes the sum of the values from index $i$ to index $j + 1$.

**Example**

$$\left( \begin{array}{l} 16, \langle 4, 3, 5, 5, 2, 1, 1, 2, 3, 4, 6, 6, 3, 1, 2, 3 \rangle, \\ \langle 0, 0, 0, 0, 0, 0, 0, 1, 0, 0, 0, 0, 0, 0, 0, 0 \rangle \end{array} \right)$$

Figure 3.1023 provides an example where the POS_MAX_SURF_STRICTLY_INCREASING_SEQUENCE $(16, [4, 3, 5, 5, 2, 1, 1, 2, 3, 4, 6, 6, 3, 1, 2, 3], [0, 0, 0, 0, 0, 0, 0, 1, 0, 0, 0, 0, 0, 0, 0, 0])$ constraint holds.

Figure 3.1023: Illustrating the POS_MAX_SURF_STRICTLY_INCREASING_SEQUENCE constraint of the **Example** slot

**Typical**
$|\text{VARIABLES}| > 1$
$\text{range}(\text{VARIABLES.var}) > 1$

**Arg. properties**
- Functional dependency: VALUE determined by VARIABLES.
- Functional dependency: FOUND determined by VARIABLES.

**Automaton**          Similar to the automaton of the MAX_SURF_STRICTLY_INCREASING_SEQUENCE constraint but use the decoration table 2.33.

# 3.459 POS_MAX_SURF_SUMMIT

▶ ▷ ◁ ◀          **DESCRIPTION**          **AUTOMATON**

$(< | < (= | <)^* <)( > | > (= | >)^* >)$

**Origin**           Based on constraint MAX_SURF_SUMMIT.

**Constraint**       POS_MAX_SURF_SUMMIT(VALUE, VARIABLES, FOUND)

**Arguments**
          VALUE      : dvar
          VARIABLES  : collection(var−dvar)
          FOUND      : collection(var−dvar)

**Restrictions**

$\mathtt{sv} \leq 2 \vee \mathtt{rv} \leq 1 \Rightarrow \mathtt{VALUE} = -\infty$

$\mathtt{rv} = 2 \Rightarrow \mathtt{VALUE} = -\infty \vee \mathtt{VALUE} \geq \mathtt{minv} + 1$

$\mathtt{rv} \geq 3 \Rightarrow$
  $\mathtt{VALUE} = -\infty \vee \mathtt{VALUE} \geq \min(\mathtt{minv} + 1, (\mathtt{sv} - 2) * (\mathtt{minv} + 1) + 1)$

$\mathtt{rv} = 2 \Rightarrow \mathtt{VALUE} \leq \mathtt{maxv}①$

$\mathtt{rv} \geq 3 \Rightarrow \mathtt{VALUE} \leq \max(\mathtt{maxv}②, (\mathtt{sv} - 2) * (\mathtt{maxv} - 1) + 1③)$

required(VARIABLES, var)

required(FOUND, var)

where

$\mathtt{maxv} = \mathtt{maxval}(\mathtt{VARIABLES.var})$
$\mathtt{rv} = \mathtt{range}(\mathtt{VARIABLES.var})$
$\mathtt{sv} = |\mathtt{VARIABLES}|$
$\mathtt{minv} = \mathtt{minval}(\mathtt{VARIABLES.var})$

**Purpose**

The constraint MAX_SURF_SUMMIT(VALUE, VARIABLES) holds. In addition, FOUND is a collection of 0/1 variables where the value 1 indicates the position of the **found** letter in those occurrences of the pattern SUMMIT for which the feature value is VALUE.

The position of the **found** letter in an occurrence of a pattern is the first position where the occurrence of pattern is identified, even if the pattern is not complete.

An occurrence of the pattern SUMMIT is the *maximal* subsequence which matches the regular expression $(< | < (= | <)^* <)( > | > (= | >)^* >)$.

Assume that the occurrence of the pattern SUMMIT starts at position $i$ and ends at position $j$. The feature SURF computes the sum of the values from index $i + 1$ to index $j$.

**Example**

$$\left( \begin{array}{l} 13, \langle 7, 1, 5, 4, 4, 3, 3, 4, 6, 6, 2, 3, 4, 2, 3, 1 \rangle, \\ \langle 0, 0, 1, 0, 0, 0, 0, 0, 0, 0, 0, 0, 0, 0, 0, 0 \rangle \end{array} \right)$$

Figure 3.1024 provides an example where the POS_MAX_SURF_SUMMIT $(13, [7, 1, 5, 4, 4, 3, 3, 4, 6, 6, 2, 3, 4, 2, 3, 1], [0, 0, 1, 0, 0, 0, 0, 0, 0, 0, 0, 0, 0, 0, 0, 0])$ constraint holds.

Figure 3.1024: Illustrating the POS_MAX_SURF_SUMMIT constraint of the **Example** slot

| **Typical** | $|\texttt{VARIABLES}| > 2$ |
| | range(VARIABLES.var) > 1 |

| **Arg. properties** | • Functional dependency: VALUE determined by VARIABLES. |
| | • Functional dependency: FOUND determined by VARIABLES. |

**Automaton**          Similar to the automaton of the MAX_SURF_SUMMIT constraint but use the decoration table 2.33.

## 3.460   POS_MAX_SURF_VALLEY

                **DESCRIPTION**           **AUTOMATON**

| | |
|---|---|
| **Origin** | Based on constraint MAX_SURF_VALLEY. |
| **Constraint** | POS_MAX_SURF_VALLEY(VALUE, VARIABLES, FOUND) |
| **Arguments** | VALUE      : dvar |
| | VARIABLES : collection(var−dvar) |
| | FOUND      : collection(var−dvar) |
| **Restrictions** | $\text{sv} \le 2 \lor \text{rv} \le 1 \Rightarrow \text{VALUE} = -\infty$ |
| | $\text{VALUE} = -\infty \lor \text{VALUE} \ge \min(\text{minv}, (\text{sv} - 2) * \text{minv})$ |
| | $\text{VALUE} \le \max(\text{maxv} - 1①, (\text{sv} - 2) * (\text{maxv} - 1)②)$ |
| | required(VARIABLES, var) |
| | required(FOUND, var) |
| | where |
| | maxv = maxval(VARIABLES.var) |
| | sv = \|VARIABLES\| |
| | minv = minval(VARIABLES.var) |
| | rv = range(VARIABLES.var) |
| **Purpose** | The constraint MAX_SURF_VALLEY(VALUE, VARIABLES) holds. In addition, FOUND is a collection of 0/1 variables where the value 1 indicates the position of the **found** letter in those occurrences of the pattern VALLEY for which the feature value is VALUE. |
| | The position of the **found** letter in an occurrence of a pattern is the first position where the occurrence of pattern is identified, even if the pattern is not complete. |
| | An occurrence of the pattern VALLEY is the *maximal* subsequence which matches the regular expression $> (= \| >)^* (< \| =)^* <$. |
| | Assume that the occurrence of the pattern VALLEY starts at position $i$ and ends at position $j$. The feature SURF computes the sum of the values from index $i + 1$ to index $j$. |
| **Example** | $\left( \begin{array}{l} 15, \langle 1, 3, 7, 4, 3, 6, 6, 5, 3, 3, 2, 6, 5, 5, 5, 7 \rangle, \\ \langle 0, 0, 0, 0, 0, 0, 0, 0, 0, 0, 0, 0, 0, 0, 1, 0 \rangle \end{array} \right)$ |

Figure 3.1025 provides an example where the POS_MAX_SURF_VALLEY $(15, [1, 3, 7, 4, 3, 6, 6, 5, 3, 3, 2, 6, 5, 5, 5, 7], [0, 0, 0, 0, 0, 0, 0, 0, 0, 0, 0, 0, 0, 0, 1, 0])$ constraint holds.

| | |
|---|---|
| **Typical** | $\|VARIABLES\| > 2$ |
| | range(VARIABLES.var) > 1 |
| **Arg. properties** | • Functional dependency: VALUE determined by VARIABLES. |
| | • Functional dependency: FOUND determined by VARIABLES. |

Figure 3.1025: Illustrating the POS_MAX_SURF_VALLEY constraint of the **Example** slot

**Automaton**        Similar to the automaton of the MAX_SURF_VALLEY constraint but use the decoration table 2.33.

# 3.461 POS_MAX_SURF_ZIGZAG

► ▷ ◁ ◀          DESCRIPTION          AUTOMATON

**Origin**          Based on constraint MAX_SURF_ZIGZAG.

$(<>)^+(<|<>)|(><)^+(>|><)$

**Constraint**          POS_MAX_SURF_ZIGZAG(VALUE, VARIABLES, FOUND)

**Arguments**

```
VALUE      : dvar
VARIABLES  : collection(var−dvar)
FOUND      : collection(var−dvar)
```

**Restrictions**

$\text{sv} \leq 3 \lor \text{rv} \leq 1 \Rightarrow \text{VALUE} = -\infty$

$\bigvee \left( \begin{array}{l} \text{VALUE} = -\infty, \\ \text{VALUE} \geq \min \left( \begin{array}{l} 2 * \text{minv} + 1, \\ \lfloor (\text{sv}-1)/2 \rfloor * \text{minv} + \lfloor (\text{sv}-2)/2 \rfloor * (\text{minv}+1) \end{array} \right) \end{array} \right)$

$\text{VALUE} \leq \max \left( \begin{array}{l} 2 * \text{maxv} - 1①, \\ \lfloor (\text{sv}-1)/2 \rfloor * \text{maxv} + \lfloor (\text{sv}-2)/2 \rfloor * (\text{maxv}-1)② \end{array} \right)$

required(VARIABLES, var)
required(FOUND, var)

where

minv = minval(VARIABLES.var)
maxv = maxval(VARIABLES.var)
sv = |VARIABLES|
rv = range(VARIABLES.var)

$\text{VALUE} \leq -3$ ①          $\text{VALUE} \leq 14 = 5*2+4*(2-1)$ ②

**Purpose**

The constraint MAX_SURF_ZIGZAG(VALUE, VARIABLES) holds. In addition, FOUND is a collection of 0/1 variables where the value 1 indicates the position of the **found** letter in those occurrences of the pattern ZIGZAG for which the feature value is VALUE.

The position of the **found** letter in an occurrence of a pattern is the first position where the occurrence of pattern is identified, even if the pattern is not complete.

An occurrence of the pattern ZIGZAG is the *maximal* subsequence which matches the regular expression $(<>)^+(<|<>)|(><)^+(>|><)$.

Assume that the occurrence of the pattern ZIGZAG starts at position $i$ and ends at position $j$. The feature SURF computes the sum of the values from index $i+1$ to index $j$.

**Example**

$\left( \begin{array}{l} 21, \langle 4, 1, 3, 1, 4, 6, 1, 5, 5, 2, 7, 2, 3, 1, 6, 1 \rangle, \\ \langle 0, 0, 0, 0, 0, 0, 0, 0, 0, 0, 1, 0, 0, 0, 0, 0 \rangle \end{array} \right)$

Figure 3.1026 provides an example where the POS_MAX_SURF_ZIGZAG $(21, [4, 1, 3, 1, 4, 6, 1, 5, 5, 2, 7, 2, 3, 1, 6, 1], [0, 0, 0, 0, 0, 0, 0, 0, 0, 0, 1, 0, 0, 0, 0, 0])$ constraint holds.

**Typical**

|VARIABLES| > 3
range(VARIABLES.var) > 1

Figure 3.1026: Illustrating the POS_MAX_SURF_ZIGZAG constraint of the **Example** slot

**Arg. properties**
- Functional dependency: VALUE determined by VARIABLES.
- Functional dependency: FOUND determined by VARIABLES.

**Automaton**        Similar to the automaton of the MAX_SURF_ZIGZAG constraint but use the decoration table 2.33.

# 3.462 POS_MAX_WIDTH_DECREASING_SEQUENCE

▶ ▷ ◁ ◀        DESCRIPTION        AUTOMATON

**Origin**            Based on constraint MAX_WIDTH_DECREASING_SEQUENCE.

**Constraint**        POS_MAX_WIDTH_DECREASING_SEQUENCE(VALUE, VARIABLES, FOUND)

**Arguments**         VALUE       : dvar
                      VARIABLES   : collection(var−dvar)
                      FOUND       : collection(var−dvar)

**Restrictions**      $\mathtt{sv} \leq 1 \vee \mathtt{rv} \leq 1 \Rightarrow \mathtt{VALUE} = 0$
                      $\mathtt{VALUE} = 0 \vee \mathtt{VALUE} \geq 2$
                      $\mathtt{rv} = 2 \Rightarrow \mathtt{VALUE} \leq 2①$
                      $\mathtt{rv} \geq 3 \Rightarrow \mathtt{VALUE} \leq \mathtt{sv}②$
                      required(VARIABLES, var)
                      required(FOUND, var)
                      where
                      $\mathtt{sv} = |\mathtt{VARIABLES}|$
                      $\mathtt{rv} = \mathrm{range}(\mathtt{VARIABLES.var})$

**Purpose**           The constraint MAX_WIDTH_DECREASING_SEQUENCE(VALUE, VARIABLES) holds. In
                      addition, FOUND is a collection of 0/1 variables where the value 1 indicates the position
                      of the **found** letter in those occurrences of the pattern DECREASING_SEQUENCE for
                      which the feature value is VALUE.

                      The position of the **found** letter in an occurrence of a pattern is the first position where
                      the occurrence of pattern is identified, even if the pattern is not complete.

                      An occurrence of the pattern DECREASING_SEQUENCE is the *maximal* subsequence
                      which matches the regular expression $> (> | =)^* | >$.

                      Assume that the occurrence of the pattern DECREASING_SEQUENCE starts at position $i$
                      and ends at position $j$. The feature WIDTH computes the value $j - i + 2$.

**Example**           $\left( \begin{array}{l} 5, \langle 3, 4, 2, 2, 5, 6, 6, 4, 4, 3, 1, 1, 4, 6, 4, 4 \rangle, \\ \langle 0, 0, 0, 0, 0, 0, 1, 0, 0, 0, 0, 0, 0, 0, 0, 0 \rangle \end{array} \right)$

                      Figure 3.1027 provides an example where the POS_MAX_WIDTH_DECREASING_SEQUENCE
                      $(5, [3, 4, 2, 2, 5, 6, 6, 4, 4, 3, 1, 1, 4, 6, 4, 4], [0, 0, 0, 0, 0, 0, 1, 0, 0, 0, 0, 0, 0, 0, 0, 0])$ con-
                      straint holds.

**Typical**           $|\mathtt{VARIABLES}| > 1$
                      $\mathrm{range}(\mathtt{VARIABLES.var}) > 1$

**Arg. properties**   • Functional dependency: VALUE determined by VARIABLES.
                      • Functional dependency: FOUND determined by VARIABLES.

Figure 3.1027: Illustrating the POS_MAX_WIDTH_DECREASING_SEQUENCE constraint of the **Example** slot

**Automaton**          Similar to the automaton of the MAX_WIDTH_DECREASING_SEQUENCE constraint but use the decoration table 2.33.

## 3.463  POS_MAX_WIDTH_DECREASING_TERRACE

▶ ▷ ◁ ◀          **DESCRIPTION**          **AUTOMATON**

| | |
|---|---|
| **Origin** | Based on constraint MAX_WIDTH_DECREASING_TERRACE. |
| **Constraint** | POS_MAX_WIDTH_DECREASING_TERRACE(VALUE, VARIABLES, FOUND) |
| **Arguments** | VALUE      : dvar<br>VARIABLES : collection(var−dvar)<br>FOUND     : collection(var−dvar) |
| **Restrictions** | $\mathtt{sv} \leq 3 \vee \mathtt{rv} \leq 2 \Rightarrow \mathtt{VALUE} = 0$<br>$\mathtt{VALUE} = 0 \vee \mathtt{VALUE} \geq 2$<br>$\mathtt{VALUE} \leq \max(0, \mathtt{sv} - 2①)$<br>required(VARIABLES, var)<br>required(FOUND, var)<br>where<br>$\mathtt{sv} = |\mathtt{VARIABLES}|$<br>$\mathtt{rv} = \mathtt{range}(\mathtt{VARIABLES.var})$ |

| | |
|---|---|
| **Purpose** | The constraint MAX_WIDTH_DECREASING_TERRACE(VALUE, VARIABLES) holds. In addition, FOUND is a collection of 0/1 variables where the value 1 indicates the position of the **found** letter in those occurrences of the pattern DECREASING_TERRACE for which the feature value is VALUE.<br>The position of the **found** letter in an occurrence of a pattern is the first position where the occurrence of pattern is identified, even if the pattern is not complete.<br>An occurrence of the pattern DECREASING_TERRACE is the *maximal* subsequence which matches the regular expression $> =^+ >$.<br>Assume that the occurrence of the pattern DECREASING_TERRACE starts at position $i$ and ends at position $j$. The feature WIDTH computes the value $j - i$. |
| **Example** | $\left( \begin{array}{l} 2, \langle 6, 4, 4, 4, 5, 2, 2, 1, 3, 3, 5, 4, 4, 3, 3, 3 \rangle, \\ \langle 0, 0, 0, 0, 0, 0, 1, 0, 0, 0, 0, 0, 1, 0, 0, 0 \rangle \end{array} \right)$ |

Figure 3.1028 provides an example where the POS_MAX_WIDTH_DECREASING_TERRACE $(2, [6, 4, 4, 4, 5, 2, 2, 1, 3, 3, 5, 4, 4, 3, 3, 3], [0, 0, 0, 0, 0, 0, 1, 0, 0, 0, 0, 0, 1, 0, 0, 0])$ constraint holds.

| | |
|---|---|
| **Typical** | $|\mathtt{VARIABLES}| > 3$<br>$\mathtt{range}(\mathtt{VARIABLES.var}) > 2$ |

| | |
|---|---|
| **Arg. properties** | • Functional dependency: VALUE determined by VARIABLES.<br>• Functional dependency: FOUND determined by VARIABLES. |

Figure 3.1028: Illustrating the POS_MAX_WIDTH_DECREASING_TERRACE constraint of the **Example** slot

**Automaton**      Similar to the automaton of the MAX_WIDTH_DECREASING_TERRACE constraint but use the decoration table 2.33.

## 3.464 POS_MAX_WIDTH_GORGE

▶ ▷ ◁ ◀          **DESCRIPTION**          **AUTOMATON**

$(> | > (= | >)^* >)(< | < (= | <)^* <)$

| | |
|---|---|
| **Origin** | Based on constraint MAX_WIDTH_GORGE. |
| **Constraint** | POS_MAX_WIDTH_GORGE(VALUE, VARIABLES, FOUND) |

**Arguments**

```
VALUE     : dvar
VARIABLES : collection(var−dvar)
FOUND     : collection(var−dvar)
```

**Restrictions**

$\mathtt{sv} \leq 2 \vee \mathtt{rv} \leq 1 \Rightarrow \mathtt{VALUE} = 0$

$\mathtt{VALUE} = 0 \vee \mathtt{VALUE} \geq 1$

$\mathtt{rv} = 2 \Rightarrow \mathtt{VALUE} \leq 1①$

$\mathtt{rv} \geq 3 \Rightarrow \mathtt{VALUE} \leq \max(0, \mathtt{sv} - 2②)$

required(VARIABLES, var)

required(FOUND, var)

where

$\mathtt{sv} = |\mathtt{VARIABLES}|$

$\mathtt{rv} = \mathrm{range}(\mathtt{VARIABLES.var})$

**Purpose**

The constraint MAX_WIDTH_GORGE(VALUE, VARIABLES) holds. In addition, FOUND is a collection of 0/1 variables where the value 1 indicates the position of the **found** letter in those occurrences of the pattern GORGE for which the feature value is VALUE.

The position of the **found** letter in an occurrence of a pattern is the first position where the occurrence of pattern is identified, even if the pattern is not complete.

An occurrence of the pattern GORGE is the *maximal* subsequence which matches the regular expression $(> | > (= | >)^* >)(< | < (= | <)^* <)$.

Assume that the occurrence of the pattern GORGE starts at position $i$ and ends at position $j$. The feature WIDTH computes the value $j - i$.

**Example**

$$\left( \begin{array}{l} 3, \langle 1, 7, 3, 4, 4, 5, 5, 4, 2, 2, 6, 5, 4, 6, 5, 7 \rangle, \\ \langle 0, 0, 1, 0, 0, 0, 0, 0, 0, 0, 0, 0, 0, 0, 0, 0 \rangle \end{array} \right)$$

Figure 3.1029 provides an example where the POS_MAX_WIDTH_GORGE $(3, [1, 7, 3, 4, 4, 5, 5, 4, 2, 2, 6, 5, 4, 6, 5, 7], [0, 0, 1, 0, 0, 0, 0, 0, 0, 0, 0, 0, 0, 0, 0, 0])$ constraint holds.

**Typical**

$|\mathtt{VARIABLES}| > 2$

$\mathrm{range}(\mathtt{VARIABLES.var}) > 1$

**Arg. properties**

- Functional dependency: VALUE determined by VARIABLES.
- Functional dependency: FOUND determined by VARIABLES.

Figure 3.1029: Illustrating the POS_MAX_WIDTH_GORGE constraint of the **Example** slot

**Automaton**     Similar to the automaton of the MAX_WIDTH_GORGE constraint but use the decoration table 2.33.

# 3.465  POS_MAX_WIDTH_INCREASING_SEQUENCE

▶ ▷ ◁ ◀        **DESCRIPTION**        **AUTOMATON**

| | |
|---|---|
| **Origin** | Based on constraint MAX_WIDTH_INCREASING_SEQUENCE. |
| **Constraint** | POS_MAX_WIDTH_INCREASING_SEQUENCE(VALUE, VARIABLES, FOUND) |
| **Arguments** | VALUE      : dvar <br> VARIABLES : collection(var−dvar) <br> FOUND     : collection(var−dvar) |

**Restrictions**

$\mathtt{sv} \leq 1 \lor \mathtt{rv} \leq 1 \Rightarrow \mathtt{VALUE} = 0$

$\mathtt{VALUE} = 0 \lor \mathtt{VALUE} \geq 2$

$\mathtt{rv} = 2 \Rightarrow \mathtt{VALUE} \leq 2$①

$\mathtt{rv} \geq 3 \Rightarrow \mathtt{VALUE} \leq \mathtt{sv}$②

required(VARIABLES, var)

required(FOUND, var)

where

$\mathtt{sv} = |\mathtt{VARIABLES}|$

$\mathtt{rv} = \mathtt{range}(\mathtt{VARIABLES.var})$

**Purpose**

The constraint MAX_WIDTH_INCREASING_SEQUENCE(VALUE, VARIABLES) holds. In addition, FOUND is a collection of 0/1 variables where the value 1 indicates the position of the **found** letter in those occurrences of the pattern INCREASING_SEQUENCE for which the feature value is VALUE.

The position of the **found** letter in an occurrence of a pattern is the first position where the occurrence of pattern is identified, even if the pattern is not complete.

An occurrence of the pattern INCREASING_SEQUENCE is the *maximal* subsequence which matches the regular expression $< (< | =)^* < | <$.

Assume that the occurrence of the pattern INCREASING_SEQUENCE starts at position $i$ and ends at position $j$. The feature WIDTH computes the value $j - i + 2$.

**Example**

$$\left( \begin{array}{l} 5, \langle 4, 3, 5, 5, 2, 1, 1, 3, 3, 4, 6, 6, 3, 1, 3, 3 \rangle , \\ \langle 0, 0, 0, 0, 0, 0, 1, 0, 0, 0, 0, 0, 0, 0, 0, 0 \rangle \end{array} \right)$$

Figure 3.1030 provides an example where the POS_MAX_WIDTH_INCREASING_SEQUENCE $(5, [4, 3, 5, 5, 2, 1, 1, 3, 3, 4, 6, 6, 3, 1, 3, 3], [0, 0, 0, 0, 0, 0, 1, 0, 0, 0, 0, 0, 0, 0, 0, 0])$ constraint holds.

**Typical**

$|\mathtt{VARIABLES}| > 1$

$\mathtt{range}(\mathtt{VARIABLES.var}) > 1$

**Arg. properties**

- Functional dependency: VALUE determined by VARIABLES.
- Functional dependency: FOUND determined by VARIABLES.

Figure 3.1030: Illustrating the POS_MAX_WIDTH_INCREASING_SEQUENCE constraint of the **Example** slot

**Automaton**   Similar to the automaton of the MAX_WIDTH_INCREASING_SEQUENCE constraint but use the decoration table 2.33.

# 3.466   POS_MAX_WIDTH_INCREASING_TERRACE

         **DESCRIPTION**                **AUTOMATON**

| | |
|---|---|
| **Origin** | Based on constraint MAX_WIDTH_INCREASING_TERRACE. |
| **Constraint** | POS_MAX_WIDTH_INCREASING_TERRACE(VALUE, VARIABLES, FOUND) |
| **Arguments** | VALUE      : dvar<br>VARIABLES : collection(var−dvar)<br>FOUND      : collection(var−dvar) |

**Restrictions**

$\mathtt{sv} \leq 3 \vee \mathtt{rv} \leq 2 \Rightarrow \mathtt{VALUE} = 0$

$\mathtt{VALUE} = 0 \vee \mathtt{VALUE} \geq 2$

$\mathtt{VALUE} \leq \max(0, \mathtt{sv} - 2①)$

required(VARIABLES, var)

required(FOUND, var)

where

$\mathtt{sv} = |\mathtt{VARIABLES}|$

$\mathtt{rv} = $range(VARIABLES.var)

**Purpose**

The constraint MAX_WIDTH_INCREASING_TERRACE(VALUE, VARIABLES) holds. In addition, FOUND is a collection of 0/1 variables where the value 1 indicates the position of the **found** letter in those occurrences of the pattern INCREASING_TERRACE for which the feature value is VALUE.

The position of the **found** letter in an occurrence of a pattern is the first position where the occurrence of pattern is identified, even if the pattern is not complete.

An occurrence of the pattern INCREASING_TERRACE is the *maximal* subsequence which matches the regular expression $<=^+<$.

Assume that the occurrence of the pattern INCREASING_TERRACE starts at position $i$ and ends at position $j$. The feature WIDTH computes the value $j - i$.

**Example**

$$\left( \begin{array}{l} 3, \langle 1, 3, 3, 3, 2, 5, 5, 6, 4, 4, 2, 3, 3, 3, 4, 4 \rangle \, , \\ \langle 0, 0, 0, 0, 0, 0, 0, 0, 0, 0, 0, 0, 0, 0, 1, 0, 0 \rangle \end{array} \right)$$

Figure 3.1031 provides an example where the POS_MAX_WIDTH_INCREASING_TERRACE $(3, [1, 3, 3, 3, 2, 5, 5, 6, 4, 4, 2, 3, 3, 3, 4, 4], [0, 0, 0, 0, 0, 0, 0, 0, 0, 0, 0, 0, 0, 0, 1, 0, 0])$ constraint holds.

**Typical**

$|\mathtt{VARIABLES}| > 3$

range(VARIABLES.var) $> 2$

**Arg. properties**

- Functional dependency: VALUE determined by VARIABLES.
- Functional dependency: FOUND determined by VARIABLES.

Figure 3.1031: Illustrating the POS_MAX_WIDTH_INCREASING_TERRACE constraint of the **Example** slot

**Automaton**  Similar to the automaton of the MAX_WIDTH_INCREASING_TERRACE constraint but use the decoration table 2.33.

# 3.467 POS_MAX_WIDTH_INFLEXION

▶ ▷ ◁ ◀    DESCRIPTION    AUTOMATON

$< (< | =)^* > | > (> | =)^* <$

| | |
|---|---|
| **Origin** | Based on constraint MAX_WIDTH_INFLEXION. |
| **Constraint** | POS_MAX_WIDTH_INFLEXION(VALUE, VARIABLES, FOUND) |

| **Arguments** | | |
|---|---|---|
| VALUE | : | dvar |
| VARIABLES | : | collection(var−dvar) |
| FOUND | : | collection(var−dvar) |

**Restrictions**

$sv \leq 2 \vee rv \leq 1 \Rightarrow VALUE = 0$
$VALUE \geq 0$
$VALUE \leq \max(0, sv - 2①)$
required(VARIABLES, var)
required(FOUND, var)
where
$sv = |VARIABLES|$
$rv = range(VARIABLES.var)$

VALUE ≤ 4 = 6 − 2

VALUE ≤ 4 = 6 − 2

**Purpose**

The constraint MAX_WIDTH_INFLEXION(VALUE, VARIABLES) holds. In addition, FOUND is a collection of 0/1 variables where the value 1 indicates the position of the **found** letter in those occurrences of the pattern INFLEXION for which the feature value is VALUE. The position of the **found** letter in an occurrence of a pattern is the first position where the occurrence of pattern is identified, even if the pattern is not complete.

An occurrence of the pattern INFLEXION is the *maximal* subsequence which matches the regular expression $< (< | =)^* > | > (> | =)^* <$.

Assume that the occurrence of the pattern INFLEXION starts at position $i$ and ends at position $j$. The feature WIDTH computes the value $j - i$.

**Example**

$$\left( \begin{array}{l} 3, \langle 1, 2, 6, 6, 4, 4, 3, 5, 2, 5, 1, 5, 3, 3, 4, 4 \rangle, \\ \langle 0, 0, 0, 1, 0, 0, 1, 0, 0, 0, 0, 0, 0, 0, 0, 0 \rangle \end{array} \right)$$

Figure 3.1032 provides an example where the POS_MAX_WIDTH_INFLEXION $(3, [1, 2, 6, 6, 4, 4, 3, 5, 2, 5, 1, 5, 3, 3, 4, 4], [0, 0, 0, 1, 0, 0, 1, 0, 0, 0, 0, 0, 0, 0, 0, 0])$ constraint holds.

| **Typical** | $|VARIABLES| > 2$ |
|---|---|
| | $range(VARIABLES.var) > 1$ |

**Arg. properties**

- Functional dependency: VALUE determined by VARIABLES.
- Functional dependency: FOUND determined by VARIABLES.

Figure 3.1032: Illustrating the POS_MAX_WIDTH_INFLEXION constraint of the **Example** slot

**Automaton**          Similar to the automaton of the MAX_WIDTH_INFLEXION constraint but use the decoration table 2.33.

### 3.468   POS_MAX_WIDTH_PEAK

▶ ▷ ◁ ◀        **DESCRIPTION**         **AUTOMATON**

$< (= | <)^* (> | =)^* >$

**Origin**          Based on constraint MAX_WIDTH_PEAK.

**Constraint**      POS_MAX_WIDTH_PEAK(VALUE, VARIABLES, FOUND)

**Arguments**       VALUE      : dvar
                    VARIABLES  : collection(var−dvar)
                    FOUND      : collection(var−dvar)

**Restrictions**    $\mathtt{sv} \leq 2 \vee \mathtt{rv} \leq 1 \Rightarrow \mathtt{VALUE} = 0$
                    $\mathtt{VALUE} \geq 0$
                    $\mathtt{VALUE} \leq \max(0, \mathtt{sv} - 2①)$
                    required(VARIABLES, var)
                    required(FOUND, var)
                    where
                    $\mathtt{sv} = |\mathtt{VARIABLES}|$
                    $\mathtt{rv} = $ range(VARIABLES.var)

VALUE $\leq$ **4** $= 6 - 2$

**Purpose**         The constraint MAX_WIDTH_PEAK(VALUE, VARIABLES) holds. In addition, FOUND is a
                    collection of 0/1 variables where the value 1 indicates the position of the **found** letter
                    in those occurrences of the pattern PEAK for which the feature value is VALUE.
                    The position of the **found** letter in an occurrence of a pattern is the first position where
                    the occurrence of pattern is identified, even if the pattern is not complete.
                    An occurrence of the pattern PEAK is the *maximal* subsequence which matches the reg-
                    ular expression $< (= | <)^* (> | =)^* >$.
                    Assume that the occurrence of the pattern PEAK starts at position $i$ and ends at position
                    $j$. The feature WIDTH computes the value $j - i$.

**Example**         $\left( \begin{array}{l} 3, \langle 7, 5, 5, 1, 4, 5, 2, 2, 3, 5, 6, 2, 3, 3, 3, 1 \rangle , \\ \langle 0, 0, 0, 0, 0, 0, 0, 0, 0, 0, 0, 1, 0, 0, 0, 1, 0 \rangle \end{array} \right)$

                    Figure   3.1033   provides   an   example   where   the   POS_MAX_WIDTH_PEAK
                    $(3, [7, 5, 5, 1, 4, 5, 2, 2, 3, 5, 6, 2, 3, 3, 3, 1], [0, 0, 0, 0, 0, 0, 0, 0, 0, 0, 0, 1, 0, 0, 0, 1, 0])$
                    constraint holds.

**Typical**         $|\mathtt{VARIABLES}| > 2$
                    range(VARIABLES.var) $> 1$

**Arg. properties** • Functional dependency: VALUE determined by VARIABLES.
                    • Functional dependency: FOUND determined by VARIABLES.

Figure 3.1033: Illustrating the POS_MAX_WIDTH_PEAK constraint of the **Example** slot

**Automaton**          Similar to the automaton of the MAX_WIDTH_PEAK constraint but use the decoration table 2.33.

# 3.469   POS_MAX_WIDTH_PLAIN

▶ ▷ ◁ ◀        **DESCRIPTION**            **AUTOMATON**

| | |
|---|---|
| **Origin** | Based on constraint MAX_WIDTH_PLAIN. |
| **Constraint** | POS_MAX_WIDTH_PLAIN(VALUE, VARIABLES, FOUND) |

**Arguments**

```
VALUE     : dvar
VARIABLES : collection(var−dvar)
FOUND     : collection(var−dvar)
```

**Restrictions**

$\text{sv} \leq 2 \vee \text{rv} \leq 1 \Rightarrow \text{VALUE} = 0$
$\text{VALUE} \geq 0$
$\text{VALUE} \leq \max(0, \text{sv} - 2①)$
required(VARIABLES, var)
required(FOUND, var)
where
$\text{sv} = |\text{VARIABLES}|$
$\text{rv} = \text{range}(\text{VARIABLES.var})$

**Purpose**

The constraint MAX_WIDTH_PLAIN(VALUE, VARIABLES) holds. In addition, FOUND is a collection of 0/1 variables where the value 1 indicates the position of the **found** letter in those occurrences of the pattern PLAIN for which the feature value is VALUE.

The position of the **found** letter in an occurrence of a pattern is the first position where the occurrence of pattern is identified, even if the pattern is not complete.

An occurrence of the pattern PLAIN is the *maximal* subsequence which matches the regular expression $> =^* <$.

Assume that the occurrence of the pattern PLAIN starts at position $i$ and ends at position $j$. The feature WIDTH computes the value $j - i$.

**Example**

$$\left( \begin{array}{l} 2, \langle 2, 3, 6, 5, 7, 6, 6, 4, 5, 5, 4, 3, 3, 6, 6, 3 \rangle, \\ \langle 0, 0, 0, 0, 0, 0, 0, 0, 0, 0, 0, 0, 0, 1, 0, 0, 0 \rangle \end{array} \right)$$

Figure 3.1034 provides an example where the POS_MAX_WIDTH_PLAIN $(2, [2, 3, 6, 5, 7, 6, 6, 4, 5, 5, 4, 3, 3, 6, 6, 3], [0, 0, 0, 0, 0, 0, 0, 0, 0, 0, 0, 0, 0, 1, 0, 0, 0])$ constraint holds.

**Typical**

$|\text{VARIABLES}| > 2$
$\text{range}(\text{VARIABLES.var}) > 1$

**Arg. properties**

- Functional dependency: VALUE determined by VARIABLES.
- Functional dependency: FOUND determined by VARIABLES.

Figure 3.1034: Illustrating the POS_MAX_WIDTH_PLAIN constraint of the **Example** slot

**Automaton**       Similar to the automaton of the MAX_WIDTH_PLAIN constraint but use the decoration table 2.33.

### 3.470 POS_MAX_WIDTH_PLATEAU

▶ ▷ ◁ ◀          **DESCRIPTION**              **AUTOMATON**

**Origin**          Based on constraint MAX_WIDTH_PLATEAU.

**Constraint**      POS_MAX_WIDTH_PLATEAU(VALUE, VARIABLES, FOUND)

**Arguments**       VALUE     : dvar
                    VARIABLES : collection(var−dvar)
                    FOUND     : collection(var−dvar)

**Restrictions**    $\mathtt{sv} \leq 2 \vee \mathtt{rv} \leq 1 \Rightarrow \mathtt{VALUE} = 0$
                    $\mathtt{VALUE} \geq 0$
                    $\mathtt{VALUE} \leq \max(0, \mathtt{sv} - 2①)$
                    required(VARIABLES, var)
                    required(FOUND, var)
                    where
                    $\mathtt{sv} = |\mathtt{VARIABLES}|$
                    $\mathtt{rv} = \mathtt{range}(\mathtt{VARIABLES.var})$

**Purpose**         The constraint MAX_WIDTH_PLATEAU(VALUE, VARIABLES) holds. In addition, FOUND
                    is a collection of 0/1 variables where the value 1 indicates the position of the **found**
                    letter in those occurrences of the pattern PLATEAU for which the feature value is VALUE.
                    The position of the **found** letter in an occurrence of a pattern is the first position where
                    the occurrence of pattern is identified, even if the pattern is not complete.

                    An occurrence of the pattern PLATEAU is the *maximal* subsequence which matches the
                    regular expression $<=^*>$.

                    Assume that the occurrence of the pattern PLATEAU starts at position $i$ and ends at position $j$. The feature WIDTH computes the value $j - i$.

**Example**         $\left( \begin{array}{l} 4, \langle 1, 3, 3, 5, 5, 5, 5, 2, 4, 4, 4, 3, 3, 1, 5, 5 \rangle, \\ \langle 0, 0, 0, 0, 0, 0, 1, 0, 0, 0, 0, 0, 0, 0, 0, 0 \rangle \end{array} \right)$

                    Figure 3.1035 provides an example where the POS_MAX_WIDTH_PLATEAU
                    $(4, [1, 3, 3, 5, 5, 5, 5, 2, 4, 4, 4, 3, 3, 1, 5, 5], [0, 0, 0, 0, 0, 0, 1, 0, 0, 0, 0, 0, 0, 0, 0, 0])$
                    constraint holds.

**Typical**         $|\mathtt{VARIABLES}| > 2$
                    $\mathtt{range}(\mathtt{VARIABLES.var}) > 1$

**Arg. properties** • Functional dependency: VALUE determined by VARIABLES.
                    • Functional dependency: FOUND determined by VARIABLES.

Figure 3.1035: Illustrating the POS_MAX_WIDTH_PLATEAU constraint of the **Example** slot

**Automaton**     Similar to the automaton of the MAX_WIDTH_PLATEAU constraint but use the decoration table 2.33.

## 3.471 POS_MAX_WIDTH_PROPER_PLAIN

► ▷ ◁ ◀    **DESCRIPTION**        **AUTOMATON**

| | |
|---|---|
| **Origin** | Based on constraint MAX_WIDTH_PROPER_PLAIN. |
| **Constraint** | POS_MAX_WIDTH_PROPER_PLAIN(VALUE, VARIABLES, FOUND) |

**Arguments**

```
VALUE     : dvar
VARIABLES : collection(var−dvar)
FOUND     : collection(var−dvar)
```

**Restrictions**

$\mathtt{sv} \leq 3 \vee \mathtt{rv} \leq 1 \Rightarrow \mathtt{VALUE} = 0$
$\mathtt{VALUE} = 0 \vee \mathtt{VALUE} \geq 2$
$\mathtt{VALUE} \leq \max(0, \mathtt{sv} - 2①)$
required(VARIABLES, var)
required(FOUND, var)
where
$\mathtt{sv} = |\mathtt{VARIABLES}|$
$\mathtt{rv} = \text{range}(\mathtt{VARIABLES.var})$

**Purpose**

The constraint MAX_WIDTH_PROPER_PLAIN(VALUE, VARIABLES) holds. In addition, FOUND is a collection of 0/1 variables where the value 1 indicates the position of the **found** letter in those occurrences of the pattern PROPER_PLAIN for which the feature value is VALUE.

The position of the **found** letter in an occurrence of a pattern is the first position where the occurrence of pattern is identified, even if the pattern is not complete.

An occurrence of the pattern PROPER_PLAIN is the *maximal* subsequence which matches the regular expression $> =^+ <$.

Assume that the occurrence of the pattern PROPER_PLAIN starts at position $i$ and ends at position $j$. The feature WIDTH computes the value $j - i$.

**Example**

$$\left( \begin{array}{l} 3, \langle 2, 7, 5, 5, 6, 3, 7, 4, 4, 5, 6, 5, 3, 3, 3, 5 \rangle, \\ \langle 0, 0, 0, 0, 0, 0, 0, 0, 0, 0, 0, 0, 0, 0, 1, 0 \rangle \end{array} \right)$$

Figure 3.1036 provides an example where the POS_MAX_WIDTH_PROPER_PLAIN $(3, [2, 7, 5, 5, 6, 3, 7, 4, 4, 5, 6, 5, 3, 3, 3, 5], [0, 0, 0, 0, 0, 0, 0, 0, 0, 0, 0, 0, 0, 0, 1, 0])$ constraint holds.

**Typical**

$|\mathtt{VARIABLES}| > 3$
$\text{range}(\mathtt{VARIABLES.var}) > 1$

**Arg. properties**

- Functional dependency: VALUE determined by VARIABLES.
- Functional dependency: FOUND determined by VARIABLES.

Figure 3.1036: Illustrating the POS_MAX_WIDTH_PROPER_PLAIN constraint of the **Example** slot

**Automaton**        Similar to the automaton of the MAX_WIDTH_PROPER_PLAIN constraint but use the decoration table 2.33.

## 3.472 POS_MAX_WIDTH_PROPER_PLATEAU

▶ ▷ ◁ ◀          **DESCRIPTION**          **AUTOMATON**

| | |
|---|---|
| **Origin** | Based on constraint MAX_WIDTH_PROPER_PLATEAU. |
| **Constraint** | POS_MAX_WIDTH_PROPER_PLATEAU(VALUE, VARIABLES, FOUND) |
| **Arguments** | VALUE     : dvar |
| | VARIABLES : collection(var−dvar) |
| | FOUND     : collection(var−dvar) |

**Restrictions**

$\mathtt{sv} \leq 3 \vee \mathtt{rv} \leq 1 \Rightarrow \mathtt{VALUE} = 0$
$\mathtt{VALUE} = 0 \vee \mathtt{VALUE} \geq 2$
$\mathtt{VALUE} \leq \max(0, \mathtt{sv} - 2①)$
required(VARIABLES, var)
required(FOUND, var)
where
$\mathtt{sv} = |\mathtt{VARIABLES}|$
$\mathtt{rv} = \mathrm{range}(\mathtt{VARIABLES.var})$

**Purpose**

The constraint MAX_WIDTH_PROPER_PLATEAU(VALUE, VARIABLES) holds. In addition, FOUND is a collection of 0/1 variables where the value 1 indicates the position of the **found** letter in those occurrences of the pattern PROPER_PLATEAU for which the feature value is VALUE.

The position of the **found** letter in an occurrence of a pattern is the first position where the occurrence of pattern is identified, even if the pattern is not complete.

An occurrence of the pattern PROPER_PLATEAU is the *maximal* subsequence which matches the regular expression $< =^+ >$.

Assume that the occurrence of the pattern PROPER_PLATEAU starts at position $i$ and ends at position $j$. The feature WIDTH computes the value $j - i$.

**Example**

$$\left( \begin{array}{l} 3, \langle 7, 1, 3, 3, 2, 5, 1, 4, 4, 3, 2, 3, 5, 5, 5, 3 \rangle, \\ \langle 0, 0, 0, 0, 0, 0, 0, 0, 0, 0, 0, 0, 0, 0, 1, 0 \rangle \end{array} \right)$$

Figure 3.1037 provides an example where the POS_MAX_WIDTH_PROPER_PLATEAU $(3, [7, 1, 3, 3, 2, 5, 1, 4, 4, 3, 2, 3, 5, 5, 5, 3], [0, 0, 0, 0, 0, 0, 0, 0, 0, 0, 0, 0, 0, 0, 1, 0])$ constraint holds.

**Typical**

$|\mathtt{VARIABLES}| > 3$
$\mathrm{range}(\mathtt{VARIABLES.var}) > 1$

**Arg. properties**

- Functional dependency: VALUE determined by VARIABLES.
- Functional dependency: FOUND determined by VARIABLES.

Figure 3.1037: Illustrating the POS_MAX_WIDTH_PROPER_PLATEAU constraint of the **Example** slot

**Automaton**      Similar to the automaton of the MAX_WIDTH_PROPER_PLATEAU constraint but use the
decoration table 2.33.

## 3.473   POS_MAX_WIDTH_STEADY_SEQUENCE

▶ ▷ ◁ ◀          **DESCRIPTION**                **AUTOMATON**

**Origin**          Based on constraint MAX_WIDTH_STEADY_SEQUENCE.

**Constraint**      POS_MAX_WIDTH_STEADY_SEQUENCE(VALUE, VARIABLES, FOUND)

**Arguments**
```
VALUE      :  dvar
VARIABLES  :  collection(var−dvar)
FOUND      :  collection(var−dvar)
```

**Restrictions**
$$\mathtt{sv} \leq 1 \Rightarrow \mathtt{VALUE} = 0$$
$$\mathtt{VALUE} = 0 \vee \mathtt{VALUE} \geq 2$$
$$\mathtt{rv} = 1 \wedge \mathtt{sv} \geq 2 \Rightarrow \mathtt{VALUE} \geq \mathtt{sv}$$
$$\mathtt{rv} \geq 2 \wedge \mathtt{sv} \geq 2 \Rightarrow \mathtt{VALUE} \geq 0$$
$$\mathtt{VALUE} \leq \mathtt{sv}①$$
required(VARIABLES, var)
required(FOUND, var)
where
$$\mathtt{sv} = |\mathtt{VARIABLES}|$$
$$\mathtt{rv} = \mathtt{range}(\mathtt{VARIABLES.var})$$

**Purpose**

The constraint MAX_WIDTH_STEADY_SEQUENCE(VALUE, VARIABLES) holds. In addition, FOUND is a collection of 0/1 variables where the value 1 indicates the position of the **found** letter in those occurrences of the pattern STEADY_SEQUENCE for which the feature value is VALUE.

The position of the **found** letter in an occurrence of a pattern is the first position where the occurrence of pattern is identified, even if the pattern is not complete.

An occurrence of the pattern STEADY_SEQUENCE is the *maximal* subsequence which matches the regular expression $=^+$.

Assume that the occurrence of the pattern STEADY_SEQUENCE starts at position $i$ and ends at position $j$. The feature WIDTH computes the value $j - i + 2$.

**Example**

$$\left( \begin{array}{l} 3, \langle 3, 1, 1, 4, 5, 5, 5, 6, 2, 2, 4, 4, 3, 2, 1, 1 \rangle, \\ \langle 0, 0, 0, 0, 1, 0, 0, 0, 0, 0, 0, 0, 0, 0, 0, 0 \rangle \end{array} \right)$$

Figure 3.1038 provides an example where the POS_MAX_WIDTH_STEADY_SEQUENCE $(3, [3, 1, 1, 4, 5, 5, 5, 6, 2, 2, 4, 4, 3, 2, 1, 1], [0, 0, 0, 0, 1, 0, 0, 0, 0, 0, 0, 0, 0, 0, 0, 0])$ constraint holds.

**Typical**        $|\mathtt{VARIABLES}| > 1$

**Arg. properties**
- Functional dependency: VALUE determined by VARIABLES.
- Functional dependency: FOUND determined by VARIABLES.

Figure 3.1038: Illustrating the POS_MAX_WIDTH_STEADY_SEQUENCE constraint of the **Example** slot

**Automaton**        Similar to the automaton of the MAX_WIDTH_STEADY_SEQUENCE constraint but use the
decoration table 2.33.

# 3.474    POS_MAX_WIDTH_STRICTLY_DECREASING_SEQUENCE

▶ ▷ ◁ ◀        **DESCRIPTION**        **AUTOMATON**

| | |
|---|---|
| **Origin** | Based on constraint MAX_WIDTH_STRICTLY_DECREASING_SEQUENCE. |
| **Constraint** | POS_MAX_WIDTH_STRICTLY_DECREASING_SEQUENCE(VALUE, VARIABLES, FOUND) |

**Arguments**

```
VALUE      :  dvar
VARIABLES  :  collection(var−dvar)
FOUND      :  collection(var−dvar)
```

**Restrictions**

$\mathtt{sv} \leq 1 \vee \mathtt{rv} \leq 1 \Rightarrow \mathtt{VALUE} = 0$
$\mathtt{VALUE} = 0 \vee \mathtt{VALUE} \geq 2$
$\mathtt{VALUE} \leq \min(\mathtt{sv}, \mathtt{rv}①)$
required(VARIABLES, var)
required(FOUND, var)
where
$\mathtt{sv} = |\mathtt{VARIABLES}|$
$\mathtt{rv} = $range(VARIABLES.var)

**Purpose**

The constraint MAX_WIDTH_STRICTLY_DECREASING_SEQUENCE(VALUE, VARIABLES) holds. In addition, FOUND is a collection of 0/1 variables where the value 1 indicates the position of the **found** letter in those occurrences of the pattern STRICTLY_DECREASING_SEQUENCE for which the feature value is VALUE.

The position of the **found** letter in an occurrence of a pattern is the first position where the occurrence of pattern is identified, even if the pattern is not complete.

An occurrence of the pattern STRICTLY_DECREASING_SEQUENCE is the *maximal* subsequence which matches the regular expression $>^+$.

Assume that the occurrence of the pattern STRICTLY_DECREASING_SEQUENCE starts at position $i$ and ends at position $j$. The feature WIDTH computes the value $j - i + 2$.

**Example**

$$\left( \begin{array}{l} 3, \langle 4, 4, 6, 4, 1, 1, 3, 4, 4, 6, 6, 5, 2, 2, 4, 3 \rangle , \\ \langle 0, 0, 1, 0, 0, 0, 0, 0, 0, 0, 1, 0, 0, 0, 0, 0 \rangle \end{array} \right)$$

Figure 3.1039 provides an example where the POS_MAX_WIDTH_STRICTLY_DECREASING_SEQUENCE $(3, [4, 4, 6, 4, 1, 1, 3, 4, 4, 6, 6, 5, 2, 2, 4, 3], [0, 0, 1, 0, 0, 0, 0, 0, 0, 0, 1, 0, 0, 0, 0, 0])$ constraint holds.

**Typical**

$|\mathtt{VARIABLES}| > 1$
range(VARIABLES.var) > 1

**Arg. properties**

- Functional dependency: VALUE determined by VARIABLES.
- Functional dependency: FOUND determined by VARIABLES.

Figure 3.1039: Illustrating the POS_MAX_WIDTH_STRICTLY_DECREASING_SEQUENCE constraint of the **Example** slot

**Automaton**          Similar to the automaton of the MAX_WIDTH_STRICTLY_DECREASING_SEQUENCE con-
                       straint but use the decoration table 2.33.

## 3.475　POS_MAX_WIDTH_STRICTLY_INCREASING_SEQUENCE

▶ ▷ ◁ ◀　　　DESCRIPTION　　　　AUTOMATON

| | |
|---|---|
| **Origin** | Based on constraint MAX_WIDTH_STRICTLY_INCREASING_SEQUENCE. |
| **Constraint** | POS_MAX_WIDTH_STRICTLY_INCREASING_SEQUENCE(VALUE, VARIABLES, FOUND) |
| **Arguments** | VALUE　　　: dvar<br>VARIABLES : collection(var−dvar)<br>FOUND　　　: collection(var−dvar) |

**Restrictions**

$\mathtt{sv} \leq 1 \vee \mathtt{rv} \leq 1 \Rightarrow \mathtt{VALUE} = 0$
$\mathtt{VALUE} = 0 \vee \mathtt{VALUE} \geq 2$
$\mathtt{VALUE} \leq \min(\mathtt{sv}, \mathtt{rv}①)$
required(VARIABLES, var)
required(FOUND, var)
where
$\mathtt{sv} = |\mathtt{VARIABLES}|$
$\mathtt{rv} = $ range(VARIABLES.var)

**Purpose**

The constraint MAX_WIDTH_STRICTLY_INCREASING_SEQUENCE(VALUE, VARIABLES) holds. In addition, FOUND is a collection of 0/1 variables where the value 1 indicates the position of the **found** letter in those occurrences of the pattern STRICTLY_INCREASING_SEQUENCE for which the feature value is VALUE.

The position of the **found** letter in an occurrence of a pattern is the first position where the occurrence of pattern is identified, even if the pattern is not complete.

An occurrence of the pattern STRICTLY_INCREASING_SEQUENCE is the *maximal* subsequence which matches the regular expression $<^+$.

Assume that the occurrence of the pattern STRICTLY_INCREASING_SEQUENCE starts at position $i$ and ends at position $j$. The feature WIDTH computes the value $j - i + 2$.

**Example**

$$\left( \begin{array}{l} 5, \langle 4, 3, 5, 5, 2, 1, 1, 2, 3, 4, 6, 6, 3, 1, 2, 3 \rangle, \\ \langle 0, 0, 0, 0, 0, 0, 1, 0, 0, 0, 0, 0, 0, 0, 0, 0 \rangle \end{array} \right)$$

Figure 3.1040 provides an example where the POS_MAX_WIDTH_STRICTLY_INCREASING_SEQUENCE $(5, [4, 3, 5, 5, 2, 1, 1, 2, 3, 4, 6, 6, 3, 1, 2, 3], [0, 0, 0, 0, 0, 0, 1, 0, 0, 0, 0, 0, 0, 0, 0, 0])$ constraint holds.

**Typical**

$|\mathtt{VARIABLES}| > 1$
range(VARIABLES.var) $> 1$

**Arg. properties**

- Functional dependency: VALUE determined by VARIABLES.
- Functional dependency: FOUND determined by VARIABLES.

Figure 3.1040: Illustrating the POS_MAX_WIDTH_STRICTLY_INCREASING_SEQUENCE constraint of the **Example** slot

**Automaton**    Similar to the automaton of the MAX_WIDTH_STRICTLY_INCREASING_SEQUENCE constraint but use the decoration table 2.33.

## 3.476　POS_MAX_WIDTH_SUMMIT

▶ ▷ ◁ ◀　　　**DESCRIPTION**　　　**AUTOMATON**

$(<|<(=|<)^*<)(>|>(=|>)^*>)$

| | |
|---|---|
| **Origin** | Based on constraint MAX_WIDTH_SUMMIT. |
| **Constraint** | POS_MAX_WIDTH_SUMMIT(VALUE, VARIABLES, FOUND) |

**Arguments**

```
VALUE     : dvar
VARIABLES : collection(var−dvar)
FOUND     : collection(var−dvar)
```

**Restrictions**

$\text{sv} \leq 2 \lor \text{rv} \leq 1 \Rightarrow \text{VALUE} = 0$

$\text{VALUE} \geq 0$

$\text{rv} = 2 \Rightarrow \text{VALUE} \leq 1$①

$\text{rv} \geq 3 \Rightarrow \text{VALUE} \leq \max(0, \text{sv} - 2$②$)$

required(VARIABLES, var)

required(FOUND, var)

where

$\text{sv} = |\text{VARIABLES}|$

$\text{rv} = \text{range}(\text{VARIABLES.var})$

**Purpose**

The constraint MAX_WIDTH_SUMMIT(VALUE, VARIABLES) holds. In addition, FOUND is a collection of 0/1 variables where the value 1 indicates the position of the **found** letter in those occurrences of the pattern SUMMIT for which the feature value is VALUE.

The position of the **found** letter in an occurrence of a pattern is the first position in the sequence where the occurrence of pattern is identified, even if the pattern is not complete.

An occurrence of the pattern SUMMIT is the *maximal* subsequence which matches the regular expression $(<|<(=|<)^*<)(>|>(=|>)^*>)$.

Assume that the occurrence of the pattern SUMMIT starts at position $i$ and ends at position $j$. The feature WIDTH computes the value $j - i$.

**Example**

$$\left( \begin{array}{l} 3, \langle 7, 1, 5, 4, 4, 3, 3, 4, 6, 6, 2, 3, 4, 2, 3, 1 \rangle , \\ \langle 0, 0, 1, 0, 0, 0, 0, 0, 0, 0, 0, 0, 0, 0, 0, 0 \rangle \end{array} \right)$$

Figure 3.1041 provides an example where the POS_MAX_WIDTH_SUMMIT $(3, [7, 1, 5, 4, 4, 3, 3, 4, 6, 6, 2, 3, 4, 2, 3, 1], [0, 0, 1, 0, 0, 0, 0, 0, 0, 0, 0, 0, 0, 0, 0, 0])$ constraint holds.

**Typical**

$|\text{VARIABLES}| > 2$

range(VARIABLES.var) > 1

**Arg. properties**

- Functional dependency: VALUE determined by VARIABLES.
- Functional dependency: FOUND determined by VARIABLES.

Figure 3.1041: Illustrating the POS_MAX_WIDTH_SUMMIT constraint of the **Example** slot

**Automaton**        Similar to the automaton of the MAX_WIDTH_SUMMIT constraint but use the decoration table 2.33.

## 3.477 POS_MAX_WIDTH_VALLEY

  DESCRIPTION    AUTOMATON

**Origin**  Based on constraint MAX_WIDTH_VALLEY.

**Constraint**  POS_MAX_WIDTH_VALLEY(VALUE, VARIABLES, FOUND)

**Arguments**
```
VALUE     : dvar
VARIABLES : collection(var−dvar)
FOUND     : collection(var−dvar)
```

**Restrictions**

$\mathtt{sv} \leq 2 \vee \mathtt{rv} \leq 1 \Rightarrow \mathtt{VALUE} = 0$

$\mathtt{VALUE} \geq 0$

$\mathtt{VALUE} \leq \max(0, \mathtt{sv} - 2\textcircled{1})$

required(VARIABLES, var)

required(FOUND, var)

where

$\mathtt{sv} = |\mathtt{VARIABLES}|$

$\mathtt{rv} = \mathrm{range}(\mathtt{VARIABLES.var})$

**Purpose**

The constraint MAX_WIDTH_VALLEY(VALUE, VARIABLES) holds. In addition, FOUND is a collection of 0/1 variables where the value 1 indicates the position of the **found** letter in those occurrences of the pattern VALLEY for which the feature value is VALUE.

The position of the **found** letter in an occurrence of a pattern is the first position where the occurrence of pattern is identified, even if the pattern is not complete.

An occurrence of the pattern VALLEY is the *maximal* subsequence which matches the regular expression $> (= | >)^* (< | =)^* <$.

Assume that the occurrence of the pattern VALLEY starts at position $i$ and ends at position $j$. The feature WIDTH computes the value $j - i$.

**Example**

$$\left( \begin{array}{l} 4, \langle 1, 3, 7, 4, 3, 6, 6, 5, 3, 3, 2, 6, 5, 5, 5, 7 \rangle, \\ \langle 0, 0, 0, 0, 0, 0, 0, 0, 0, 0, 0, 1, 0, 0, 0, 0 \rangle \end{array} \right)$$

Figure 3.1042 provides an example where the POS_MAX_WIDTH_VALLEY $(4, [1, 3, 7, 4, 3, 6, 6, 5, 3, 3, 2, 6, 5, 5, 5, 7], [0, 0, 0, 0, 0, 0, 0, 0, 0, 0, 0, 1, 0, 0, 0, 0])$ constraint holds.

**Typical**

$|\mathtt{VARIABLES}| > 2$

$\mathrm{range}(\mathtt{VARIABLES.var}) > 1$

**Arg. properties**

- Functional dependency: VALUE determined by VARIABLES.
- Functional dependency: FOUND determined by VARIABLES.

Figure 3.1042: Illustrating the POS_MAX_WIDTH_VALLEY constraint of the **Example** slot

**Automaton**    Similar to the automaton of the MAX_WIDTH_VALLEY constraint but use the decoration table 2.33.

## 3.478   POS_MAX_WIDTH_ZIGZAG

▶ ▷ ◁ ◀        **DESCRIPTION**              **AUTOMATON**

$(<>)^+(<\,|\,<>)\,|\,(><)^+(>\,|\,><)$

**Origin**            Based on constraint MAX_WIDTH_ZIGZAG.

**Constraint**        POS_MAX_WIDTH_ZIGZAG(VALUE, VARIABLES, FOUND)

**Arguments**         VALUE      : dvar
                      VARIABLES  : collection(var−dvar)
                      FOUND      : collection(var−dvar)

**Restrictions**      $\texttt{sv} \leq 3 \vee \texttt{rv} \leq 1 \Rightarrow \texttt{VALUE} = 0$
                      $\texttt{VALUE} = 0 \vee \texttt{VALUE} \geq 2$
                      $\texttt{VALUE} \leq \max(0, \texttt{sv} - 2\text{①})$
                      required(VARIABLES, var)
                      required(FOUND, var)
                      where
                      $\texttt{sv} = |\texttt{VARIABLES}|$
                      $\texttt{rv} = \text{range}(\texttt{VARIABLES.var})$

**Purpose**           The constraint MAX_WIDTH_ZIGZAG(VALUE, VARIABLES) holds. In addition, FOUND is
                      a collection of 0/1 variables where the value 1 indicates the position of the **found** letter
                      in those occurrences of the pattern ZIGZAG for which the feature value is VALUE.
                      The position of the **found** letter in an occurrence of a pattern is the first position where
                      the occurrence of pattern is identified, even if the pattern is not complete.
                      An occurrence of the pattern ZIGZAG is the *maximal* subsequence which matches the
                      regular expression $(<>)^+(<\,|\,<>)\,|\,(><)^+(>\,|\,><)$.
                      Assume that the occurrence of the pattern ZIGZAG starts at position $i$ and ends at position
                      $j$. The feature WIDTH computes the value $j - i$.

**Example**           $\left( \begin{array}{l} 6, \langle 4, 1, 3, 1, 4, 6, 1, 5, 5, 2, 7, 2, 3, 1, 6, 1 \rangle , \\ \langle 0, 0, 0, 0, 0, 0, 0, 0, 0, 0, 0, 1, 0, 0, 0, 0 \rangle \end{array} \right)$

                      Figure 3.1043 provides an example where the POS_MAX_WIDTH_ZIGZAG
                      $(6, [4, 1, 3, 1, 4, 6, 1, 5, 5, 2, 7, 2, 3, 1, 6, 1], [0, 0, 0, 0, 0, 0, 0, 0, 0, 0, 0, 1, 0, 0, 0, 0])$
                      constraint holds.

**Typical**           $|\texttt{VARIABLES}| > 3$
                      $\text{range}(\texttt{VARIABLES.var}) > 1$

**Arg. properties**   • Functional dependency: VALUE determined by VARIABLES.
                      • Functional dependency: FOUND determined by VARIABLES.

Figure 3.1043: Illustrating the POS_MAX_WIDTH_ZIGZAG constraint of the **Example** slot

**Automaton**      Similar to the automaton of the MAX_WIDTH_ZIGZAG constraint but use the decoration table 2.33.

## 3.479 POS_MIN_HEIGHT_DECREASING_TERRACE

   DESCRIPTION          AUTOMATON

| | |
|---|---|
| **Origin** | Based on constraint MIN_HEIGHT_DECREASING_TERRACE. |
| **Constraint** | POS_MIN_HEIGHT_DECREASING_TERRACE(VALUE, VARIABLES, FOUND) |

**Arguments**

```
VALUE     : dvar
VARIABLES : collection(var−dvar)
FOUND     : collection(var−dvar)
```

**Restrictions**

$sv \leq 3 \vee rv \leq 2 \Rightarrow \text{VALUE} = +\infty$
$\text{VALUE} \geq \text{minv} + 1 ①$
$\text{VALUE} = +\infty \vee \text{VALUE} \leq \text{maxv} - 1$
required(VARIABLES, var)
required(FOUND, var)
where
$\text{minv} = \text{minval}(\text{VARIABLES.var})$
$\text{maxv} = \text{maxval}(\text{VARIABLES.var})$
$sv = |\text{VARIABLES}|$
$rv = \text{range}(\text{VARIABLES.var})$

**Purpose**

The constraint MIN_HEIGHT_DECREASING_TERRACE(VALUE, VARIABLES) holds. In addition, FOUND is a collection of 0/1 variables where the value 1 indicates the position of the **found** letter in those occurrences of the pattern DECREASING_TERRACE for which the feature value is VALUE.

The position of the **found** letter in an occurrence of a pattern is the first position where the occurrence of pattern is identified, even if the pattern is not complete.

An occurrence of the pattern DECREASING_TERRACE is the *maximal* subsequence which matches the regular expression $> =^+ >$.

Assume that the occurrence of the pattern DECREASING_TERRACE starts at position $i$ and ends at position $j$. The feature MIN computes the minimum of the values from index $i + 1$ to index $j$.

**Example**

$$\left( \begin{array}{l} 2, \langle 6, 4, 4, 4, 5, 2, 2, 1, 3, 3, 5, 4, 4, 3, 3, 3 \rangle, \\ \langle 0, 0, 0, 0, 0, 0, 1, 0, 0, 0, 0, 0, 0, 0, 0, 0 \rangle \end{array} \right)$$

Figure 3.1044 provides an example where the POS_MIN_HEIGHT_DECREASING_TERRACE $(2, [6, 4, 4, 4, 5, 2, 2, 1, 3, 3, 5, 4, 4, 3, 3, 3], [0, 0, 0, 0, 0, 0, 1, 0, 0, 0, 0, 0, 0, 0, 0, 0])$ constraint holds.

**Typical**

$|\text{VARIABLES}| > 3$
range(VARIABLES.var) $> 2$

Figure 3.1044: Illustrating the POS_MIN_HEIGHT_DECREASING_TERRACE constraint of the **Example** slot

**Arg. properties**
- Functional dependency: VALUE determined by VARIABLES.
- Functional dependency: FOUND determined by VARIABLES.

**Automaton**        Similar to the automaton of the MIN_HEIGHT_DECREASING_TERRACE constraint but use
the decoration table 2.33.

## 3.480   POS_MIN_HEIGHT_INCREASING_TERRACE

        DESCRIPTION            AUTOMATON

| | |
|---|---|
| **Origin** | Based on constraint MIN_HEIGHT_INCREASING_TERRACE. |
| **Constraint** | POS_MIN_HEIGHT_INCREASING_TERRACE(VALUE, VARIABLES, FOUND) |
| **Arguments** | VALUE     : dvar<br>VARIABLES : collection(var−dvar)<br>FOUND     : collection(var−dvar) |

**Restrictions**

$sv \leq 3 \vee rv \leq 2 \Rightarrow VALUE = +\infty$

$VALUE \geq minv + 1①$

$VALUE = +\infty \vee VALUE \leq maxv − 1$

required(VARIABLES, var)

required(FOUND, var)

where

minv = minval(VARIABLES.var)

maxv = maxval(VARIABLES.var)

sv = |VARIABLES|

rv = range(VARIABLES.var)

**Purpose**

The constraint MIN_HEIGHT_INCREASING_TERRACE(VALUE, VARIABLES) holds. In addition, FOUND is a collection of 0/1 variables where the value 1 indicates the position of the **found** letter in those occurrences of the pattern INCREASING_TERRACE for which the feature value is VALUE.

The position of the **found** letter in an occurrence of a pattern is the first position where the occurrence of pattern is identified, even if the pattern is not complete.

An occurrence of the pattern INCREASING_TERRACE is the *maximal* subsequence which matches the regular expression $<=^+<$.

Assume that the occurrence of the pattern INCREASING_TERRACE starts at position $i$ and ends at position $j$. The feature MIN computes the minimum of the values from index $i + 1$ to index $j$.

**Example**

$$\left( \begin{array}{l} 3, \langle 1, 3, 3, 3, 2, 5, 5, 6, 4, 4, 2, 3, 3, 3, 4, 4 \rangle, \\ \langle 0, 0, 0, 0, 0, 0, 0, 0, 0, 0, 0, 0, 0, 0, 1, 0, 0 \rangle \end{array} \right)$$

Figure 3.1045 provides an example where the POS_MIN_HEIGHT_INCREASING_TERRACE $(3, [1, 3, 3, 3, 2, 5, 5, 6, 4, 4, 2, 3, 3, 3, 4, 4], [0, 0, 0, 0, 0, 0, 0, 0, 0, 0, 0, 0, 0, 0, 1, 0, 0])$ constraint holds.

**Typical**

|VARIABLES| > 3

range(VARIABLES.var) > 2

Figure 3.1045: Illustrating the POS_MIN_HEIGHT_INCREASING_TERRACE constraint of the **Example** slot

**Arg. properties**
- Functional dependency: VALUE determined by VARIABLES.
- Functional dependency: FOUND determined by VARIABLES.

**Automaton**    Similar to the automaton of the MIN_HEIGHT_INCREASING_TERRACE constraint but use
the decoration table 2.33.

# 3.481  POS_MIN_HEIGHT_PLAIN

        DESCRIPTION        AUTOMATON

| | |
|---|---|
| **Origin** | Based on constraint MIN_HEIGHT_PLAIN. |
| **Constraint** | POS_MIN_HEIGHT_PLAIN(VALUE, VARIABLES, FOUND) |
| **Arguments** | VALUE : dvar <br> VARIABLES : collection(var−dvar) <br> FOUND : collection(var−dvar) |

**Restrictions**

$\text{sv} \leq 2 \vee \text{rv} \leq 1 \Rightarrow \text{VALUE} = +\infty$

$\text{VALUE} \geq \text{minv}$①

$\text{VALUE} = +\infty \vee \text{VALUE} \leq \text{maxv} - 1$

required(VARIABLES, var)

required(FOUND, var)

where

minv = minval(VARIABLES.var)

maxv = maxval(VARIABLES.var)

sv = |VARIABLES|

rv = range(VARIABLES.var)

**Purpose**

The constraint MIN_HEIGHT_PLAIN(VALUE, VARIABLES) holds. In addition, FOUND is a collection of 0/1 variables where the value 1 indicates the position of the **found** letter in those occurrences of the pattern PLAIN for which the feature value is VALUE.

The position of the **found** letter in an occurrence of a pattern is the first position where the occurrence of pattern is identified, even if the pattern is not complete.

An occurrence of the pattern PLAIN is the *maximal* subsequence which matches the regular expression $> =^* <$.

Assume that the occurrence of the pattern PLAIN starts at position $i$ and ends at position $j$. The feature MIN computes the minimum of the values from index $i + 1$ to index $j$.

**Example**

$$\left( \begin{array}{l} 3, \langle 2, 3, 6, 5, 7, 6, 6, 4, 5, 5, 4, 3, 3, 6, 6, 3 \rangle, \\ \langle 0, 0, 0, 0, 0, 0, 0, 0, 0, 0, 0, 0, 1, 0, 0, 0 \rangle \end{array} \right)$$

Figure 3.1046 provides an example where the POS_MIN_HEIGHT_PLAIN $(3, [2, 3, 6, 5, 7, 6, 6, 4, 5, 5, 4, 3, 3, 6, 6, 3], [0, 0, 0, 0, 0, 0, 0, 0, 0, 0, 0, 0, 1, 0, 0, 0])$ constraint holds.

**Typical**

$|\text{VARIABLES}| > 2$

range(VARIABLES.var) > 1

**Arg. properties**

- Functional dependency: VALUE determined by VARIABLES.
- Functional dependency: FOUND determined by VARIABLES.

Figure 3.1046: Illustrating the POS_MIN_HEIGHT_PLAIN constraint of the **Example** slot

**Automaton**          Similar to the automaton of the MIN_HEIGHT_PLAIN constraint but use the decoration ta-
                       ble 2.33.

## 3.482 POS_MIN_HEIGHT_PLATEAU

**Origin**          Based on constraint MIN_HEIGHT_PLATEAU.

**Constraint**      POS_MIN_HEIGHT_PLATEAU(VALUE, VARIABLES, FOUND)

**Arguments**       
```
VALUE     :  dvar
VARIABLES :  collection(var−dvar)
FOUND     :  collection(var−dvar)
```

**Restrictions**    
$\mathtt{sv} \leq 2 \vee \mathtt{rv} \leq 1 \Rightarrow \mathtt{VALUE} = +\infty$

$\mathtt{VALUE} \geq \mathtt{minv} + 1$①

$\mathtt{VALUE} = +\infty \vee \mathtt{VALUE} \leq \mathtt{maxv}$

required(VARIABLES, var)

required(FOUND, var)

where

$\mathtt{minv} = \mathtt{minval}(\mathtt{VARIABLES.var})$

$\mathtt{maxv} = \mathtt{maxval}(\mathtt{VARIABLES.var})$

$\mathtt{sv} = |\mathtt{VARIABLES}|$

$\mathtt{rv} = \mathtt{range}(\mathtt{VARIABLES.var})$

**Purpose**         The constraint MIN_HEIGHT_PLATEAU(VALUE, VARIABLES) holds. In addition, FOUND is a collection of 0/1 variables where the value 1 indicates the position of the **found** letter in those occurrences of the pattern PLATEAU for which the feature value is VALUE. The position of the **found** letter in an occurrence of a pattern is the first position where the occurrence of pattern is identified, even if the pattern is not complete.

An occurrence of the pattern PLATEAU is the *maximal* subsequence which matches the regular expression $<=^*>$.

Assume that the occurrence of the pattern PLATEAU starts at position $i$ and ends at position $j$. The feature MIN computes the minimum of the values from index $i + 1$ to index $j$.

**Example**         
$$\left( \begin{array}{l} 3, \langle 7, 5, 2, 3, 1, 2, 2, 4, 3, 3, 4, 5, 5, 2, 2, 5 \rangle, \\ \langle 0, 0, 0, 1, 0, 0, 0, 0, 0, 0, 0, 0, 0, 0, 0, 0 \rangle \end{array} \right)$$

Figure 3.1047 provides an example where the POS_MIN_HEIGHT_PLATEAU $(3, [7, 5, 2, 3, 1, 2, 2, 4, 3, 3, 4, 5, 5, 2, 2, 5], [0, 0, 0, 1, 0, 0, 0, 0, 0, 0, 0, 0, 0, 0, 0, 0])$ constraint holds.

**Typical**         
$|\mathtt{VARIABLES}| > 2$

$\mathtt{range}(\mathtt{VARIABLES.var}) > 1$

Figure 3.1047: Illustrating the POS_MIN_HEIGHT_PLATEAU constraint of the **Example** slot

**Arg. properties**
- Functional dependency: VALUE determined by VARIABLES.
- Functional dependency: FOUND determined by VARIABLES.

**Automaton**          Similar to the automaton of the MIN_HEIGHT_PLATEAU constraint but use the decoration
                       table 2.33.

## 3.483   POS_MIN_HEIGHT_PROPER_PLAIN

        DESCRIPTION          AUTOMATON

| | |
|---|---|
| **Origin** | Based on constraint MIN_HEIGHT_PROPER_PLAIN. |
| **Constraint** | POS_MIN_HEIGHT_PROPER_PLAIN(VALUE, VARIABLES, FOUND) |

**Arguments**

```
VALUE     : dvar
VARIABLES : collection(var-dvar)
FOUND     : collection(var-dvar)
```

**Restrictions**

$\text{sv} \leq 3 \vee \text{rv} \leq 1 \Rightarrow \text{VALUE} = +\infty$

$\text{VALUE} \geq \text{minv}①$

$\text{VALUE} = +\infty \vee \text{VALUE} \leq \text{maxv} - 1$

required(VARIABLES, var)

required(FOUND, var)

where

minv = minval(VARIABLES.var)

maxv = maxval(VARIABLES.var)

sv = |VARIABLES|

rv = range(VARIABLES.var)

**Purpose**

The constraint MIN_HEIGHT_PROPER_PLAIN(VALUE, VARIABLES) holds. In addition, FOUND is a collection of 0/1 variables where the value 1 indicates the position of the **found** letter in those occurrences of the pattern PROPER_PLAIN for which the feature value is VALUE.

The position of the **found** letter in an occurrence of a pattern is the first position where the occurrence of pattern is identified, even if the pattern is not complete.

An occurrence of the pattern PROPER_PLAIN is the *maximal* subsequence which matches the regular expression $>=^+<$.

Assume that the occurrence of the pattern PROPER_PLAIN starts at position $i$ and ends at position $j$. The feature MIN computes the minimum of the values from index $i + 1$ to index $j$.

**Example**

$$\left( \begin{array}{l} 3, \langle 2, 7, 5, 5, 6, 3, 7, 4, 4, 5, 6, 5, 3, 3, 3, 5 \rangle, \\ \langle 0, 0, 0, 0, 0, 0, 0, 0, 0, 0, 0, 0, 0, 0, 1, 0 \rangle \end{array} \right)$$

Figure 3.1048 provides an example where the POS_MIN_HEIGHT_PROPER_PLAIN (3, [2, 7, 5, 5, 6, 3, 7, 4, 4, 5, 6, 5, 3, 3, 3, 5], [0, 0, 0, 0, 0, 0, 0, 0, 0, 0, 0, 0, 0, 0, 1, 0]) constraint holds.

**Typical**

|VARIABLES| > 3

range(VARIABLES.var) > 1

Figure 3.1048: Illustrating the POS_MIN_HEIGHT_PROPER_PLAIN constraint of the **Example** slot

**Arg. properties**
- Functional dependency: VALUE determined by VARIABLES.
- Functional dependency: FOUND determined by VARIABLES.

**Automaton**            Similar to the automaton of the MIN_HEIGHT_PROPER_PLAIN constraint but use the deco-
                         ration table 2.33.

## 3.484   POS_MIN_HEIGHT_PROPER_PLATEAU

▶ ▷ ◁ ◀    **DESCRIPTION**          **AUTOMATON**

| | |
|---|---|
| Origin | Based on constraint MIN_HEIGHT_PROPER_PLATEAU. |
| Constraint | POS_MIN_HEIGHT_PROPER_PLATEAU(VALUE, VARIABLES, FOUND) |
| Arguments | VALUE     : dvar |
| | VARIABLES : collection(var−dvar) |
| | FOUND     : collection(var−dvar) |
| Restrictions | $\mathtt{sv} \leq 3 \vee \mathtt{rv} \leq 1 \Rightarrow \mathtt{VALUE} = +\infty$ |
| | $\mathtt{VALUE} \geq \mathtt{minv} + 1$ ① |
| | $\mathtt{VALUE} = +\infty \vee \mathtt{VALUE} \leq \mathtt{maxv}$ |
| | required(VARIABLES, var) |
| | required(FOUND, var) |
| | where |
| | $\mathtt{minv} = \mathtt{minval}(\mathtt{VARIABLES.var})$ |
| | $\mathtt{maxv} = \mathtt{maxval}(\mathtt{VARIABLES.var})$ |
| | $\mathtt{sv} = |\mathtt{VARIABLES}|$ |
| | $\mathtt{rv} = \mathtt{range}(\mathtt{VARIABLES.var})$ |
| Purpose | The constraint MIN_HEIGHT_PROPER_PLATEAU(VALUE, VARIABLES) holds. In addition, FOUND is a collection of 0/1 variables where the value 1 indicates the position of the **found** letter in those occurrences of the pattern PROPER_PLATEAU for which the feature value is VALUE. |
| | The position of the **found** letter in an occurrence of a pattern is the first position where the occurrence of pattern is identified, even if the pattern is not complete. |
| | An occurrence of the pattern PROPER_PLATEAU is the *maximal* subsequence which matches the regular expression $<=^+>$. |
| | Assume that the occurrence of the pattern PROPER_PLATEAU starts at position $i$ and ends at position $j$. The feature MIN computes the minimum of the values from index $i + 1$ to index $j$. |
| Example | $\left( \begin{array}{l} 3, \langle 7, 1, 3, 3, 2, 5, 1, 4, 4, 3, 2, 3, 5, 5, 5, 3 \rangle, \\ \langle 0, 0, 0, 1, 0, 0, 0, 0, 0, 0, 0, 0, 0, 0, 0, 0 \rangle \end{array} \right)$ |
| | Figure 3.1049 provides an example where the POS_MIN_HEIGHT_PROPER_PLATEAU (3, [7, 1, 3, 3, 2, 5, 1, 4, 4, 3, 2, 3, 5, 5, 5, 3], [0, 0, 0, 1, 0, 0, 0, 0, 0, 0, 0, 0, 0, 0, 0, 0]) constraint holds. |
| Typical | $|\mathtt{VARIABLES}| > 3$ |
| | $\mathtt{range}(\mathtt{VARIABLES.var}) > 1$ |

Figure 3.1049: Illustrating the POS_MIN_HEIGHT_PROPER_PLATEAU constraint of the **Example** slot

**Arg. properties**
- Functional dependency: VALUE determined by VARIABLES.
- Functional dependency: FOUND determined by VARIABLES.

**Automaton**     Similar to the automaton of the MIN_HEIGHT_PROPER_PLATEAU constraint but use the decoration table 2.33.

## 3.485 POS_MIN_HEIGHT_STEADY

        DESCRIPTION          AUTOMATON

| | |
|---|---|
| **Origin** | Based on constraint MIN_HEIGHT_STEADY. |
| **Constraint** | POS_MIN_HEIGHT_STEADY(VALUE, VARIABLES, FOUND) |
| **Arguments** | VALUE : dvar |
| | VARIABLES : collection(var−dvar) |
| | FOUND : collection(var−dvar) |
| **Restrictions** | $\text{sv} \leq 1 \Rightarrow \text{VALUE} = +\infty$ |
| | $\text{VALUE} \geq \text{minv}①$ |
| | $\text{VALUE} = +\infty \lor \text{VALUE} \leq \text{maxv}$ |
| | required(VARIABLES, var) |
| | required(FOUND, var) |
| | where |
| | minv = minval(VARIABLES.var) |
| | maxv = maxval(VARIABLES.var) |
| | sv = \|VARIABLES\| |
| **Purpose** | The constraint MIN_HEIGHT_STEADY(VALUE, VARIABLES) holds. In addition, FOUND is a collection of 0/1 variables where the value 1 indicates the position of the **found** letter in those occurrences of the pattern STEADY for which the feature value is VALUE. |
| | The position of the **found** letter in an occurrence of a pattern is the first position where the occurrence of pattern is identified, even if the pattern is not complete. |
| | An occurrence of the pattern STEADY is the subsequence which matches the regular expression =. |
| | Assume that the occurrence of the pattern STEADY starts at position $i$ and ends at position $j$. The feature MIN computes the minimum of the values from index $i$ to index $j + 1$. |
| **Example** | $\left( \begin{array}{l} 1, \langle 1, 1, 7, 3, 3, 5, 5, 5, 6, 5, 5, 5, 7, 2, 6, 6 \rangle , \\ \langle 1, 0, 0, 0, 0, 0, 0, 0, 0, 0, 0, 0, 0, 0, 0, 0 \rangle \end{array} \right)$ |

Figure 3.1050 provides an example where the POS_MIN_HEIGHT_STEADY
$(1, [1, 1, 7, 3, 3, 5, 5, 5, 6, 5, 5, 5, 7, 2, 6, 6], [1, 0, 0, 0, 0, 0, 0, 0, 0, 0, 0, 0, 0, 0, 0, 0])$
constraint holds.

| | |
|---|---|
| **Typical** | \|VARIABLES\| > 1 |
| **Arg. properties** | • Functional dependency: VALUE determined by VARIABLES. |
| | • Functional dependency: FOUND determined by VARIABLES. |

Figure 3.1050: Illustrating the POS_MIN_HEIGHT_STEADY constraint of the **Example** slot

**Automaton**          Similar to the automaton of the MIN_HEIGHT_STEADY constraint but use the decoration
                       table 2.33.

## 3.486   POS_MIN_HEIGHT_STEADY_SEQUENCE

▶ ▷ ◁ ◀        **DESCRIPTION**          **AUTOMATON**

**Origin**          Based on constraint MIN_HEIGHT_STEADY_SEQUENCE.

**Constraint**      POS_MIN_HEIGHT_STEADY_SEQUENCE(VALUE, VARIABLES, FOUND)

**Arguments**       VALUE      : dvar
                    VARIABLES  : collection(var−dvar)
                    FOUND      : collection(var−dvar)

**Restrictions**    $\mathtt{sv} \leq 1 \Rightarrow \mathtt{VALUE} = +\infty$
                    $\mathtt{VALUE} \geq \mathtt{minv}①$
                    $\mathtt{VALUE} = +\infty \vee \mathtt{VALUE} \leq \mathtt{maxv}$
                    required(VARIABLES, var)
                    required(FOUND, var)
                    where
                    $\mathtt{minv} = \mathtt{minval}(\mathtt{VARIABLES.var})$
                    $\mathtt{maxv} = \mathtt{maxval}(\mathtt{VARIABLES.var})$
                    $\mathtt{sv} = |\mathtt{VARIABLES}|$

**Purpose**         The constraint MIN_HEIGHT_STEADY_SEQUENCE(VALUE, VARIABLES) holds. In addition, FOUND is a collection of 0/1 variables where the value 1 indicates the position of the **found** letter in those occurrences of the pattern STEADY_SEQUENCE for which the feature value is VALUE.

                    The position of the **found** letter in an occurrence of a pattern is the first position where the occurrence of pattern is identified, even if the pattern is not complete.

                    An occurrence of the pattern STEADY_SEQUENCE is the *maximal* subsequence which matches the regular expression $=^+$.

                    Assume that the occurrence of the pattern STEADY_SEQUENCE starts at position $i$ and ends at position $j$. The feature MIN computes the minimum of the values from index $i$ to index $j + 1$.

**Example**         $\left( \begin{array}{l} 1, \langle 3, 1, 1, 4, 5, 5, 5, 6, 2, 2, 4, 4, 3, 2, 1, 1 \rangle , \\ \langle 0, 1, 0, 0, 0, 0, 0, 0, 0, 0, 0, 0, 0, 0, 1, 0 \rangle \end{array} \right)$

                    Figure 3.1051 provides an example where the POS_MIN_HEIGHT_STEADY_SEQUENCE $(1, [3, 1, 1, 4, 5, 5, 5, 6, 2, 2, 4, 4, 3, 2, 1, 1], [0, 1, 0, 0, 0, 0, 0, 0, 0, 0, 0, 0, 0, 0, 1, 0])$ constraint holds.

**Typical**         $|\mathtt{VARIABLES}| > 1$

**Arg. properties**  • Functional dependency: VALUE determined by VARIABLES.
                    • Functional dependency: FOUND determined by VARIABLES.

Figure 3.1051: Illustrating the POS_MIN_HEIGHT_STEADY_SEQUENCE constraint of the **Example** slot

**Automaton**        Similar to the automaton of the MIN_HEIGHT_STEADY_SEQUENCE constraint but use the
                     decoration table 2.33.

## 3.487   POS_MIN_MAX_BUMP_ON_DECREASING_SEQUENCE

▶ ▷ ◁ ◀        DESCRIPTION          AUTOMATON

**Origin**          Based on constraint MIN_MAX_BUMP_ON_DECREASING_SEQUENCE.

**Constraint**      POS_MIN_MAX_BUMP_ON_DECREASING_SEQUENCE(VALUE, VARIABLES, FOUND)

**Arguments**       VALUE     : dvar
                    VARIABLES : collection(var−dvar)
                    FOUND     : collection(var−dvar)

**Restrictions**    $\text{sv} \leq 5 \vee \text{rv} \leq 2 \Rightarrow \text{VALUE} = +\infty$

                    $\text{VALUE} \geq \text{minv} + 2①$
                    $\text{VALUE} = +\infty \vee \text{VALUE} \leq \text{maxv}$
                    required(VARIABLES, var)
                    required(FOUND, var)
                    where
                    $\text{minv} = \text{minval}(\text{VARIABLES.var})$
                    $\text{maxv} = \text{maxval}(\text{VARIABLES.var})$
                    $\text{sv} = |\text{VARIABLES}|$
                    $\text{rv} = \text{range}(\text{VARIABLES.var})$

**Purpose**         The constraint MIN_MAX_BUMP_ON_DECREASING_SEQUENCE(VALUE, VARIABLES)
                    holds. In addition, FOUND is a collection of 0/1 variables where the value 1
                    indicates the position of the **found** letter in those occurrences of the pattern
                    BUMP_ON_DECREASING_SEQUENCE for which the feature value is VALUE.

                    The position of the **found** letter in an occurrence of a pattern is the first position where
                    the occurrence of pattern is identified, even if the pattern is not complete.

                    An occurrence of the pattern BUMP_ON_DECREASING_SEQUENCE is the subsequence
                    which matches the regular expression $>><>$.

                    Assume that the occurrence of the pattern BUMP_ON_DECREASING_SEQUENCE starts at
                    position $i$ and ends at position $j$. The feature MAX computes the maximum of the values
                    from index $i + 2$ to index $j$.

**Example**         $\left( \begin{array}{l} 5, \langle 7, 6, 5, 6, 5, 4, 1, 4, 7, 5, 4, 2, 5, 4, 3, 3 \rangle, \\ \langle 0, 0, 0, 0, 0, 0, 0, 0, 0, 0, 0, 0, 0, 0, 1, 0, 0 \rangle \end{array} \right)$

                    Figure 3.1052 provides an example where the POS_MIN_MAX_BUMP_ON_DECREASING_SEQUENCE
                    (5, [7, 6, 5, 6, 5, 4, 1, 4, 7, 5, 4, 2, 5, 4, 3, 3], [0, 0, 0, 0, 0, 0, 0, 0, 0, 0, 0, 0, 0, 0, 1, 0, 0])   con-
                    straint holds.

**Typical**         $|\text{VARIABLES}| > 5$
                    range(VARIABLES.var) > 2

Figure 3.1052: Illustrating the POS_MIN_MAX_BUMP_ON_DECREASING_SEQUENCE constraint of the **Example** slot

**Arg. properties**
- Functional dependency: VALUE determined by VARIABLES.
- Functional dependency: FOUND determined by VARIABLES.

**Automaton**    Similar to the automaton of the MIN_MAX_BUMP_ON_DECREASING_SEQUENCE constraint
but use the decoration table 2.33.

# 3.488 POS_MIN_MAX_DECREASING

◂ ▹ ◃ ◂    DESCRIPTION    AUTOMATON

| | |
|---|---|
| **Origin** | Based on constraint MIN_MAX_DECREASING. |
| **Constraint** | POS_MIN_MAX_DECREASING(VALUE, VARIABLES, FOUND) |
| **Arguments** | VALUE : dvar |
| | VARIABLES : collection(var−dvar) |
| | FOUND : collection(var−dvar) |

**Restrictions**

$sv \leq 1 \vee rv \leq 1 \Rightarrow$ VALUE $= +\infty$

VALUE $\geq$ minv $+ 1$①

VALUE $= +\infty \vee$ VALUE $\leq$ maxv

required(VARIABLES, var)

required(FOUND, var)

where

minv = minval(VARIABLES.var)

maxv = maxval(VARIABLES.var)

sv = |VARIABLES|

rv = range(VARIABLES.var)

**Purpose**

The constraint MIN_MAX_DECREASING(VALUE, VARIABLES) holds. In addition, FOUND is a collection of 0/1 variables where the value 1 indicates the position of the **found** letter in those occurrences of the pattern DECREASING for which the feature value is VALUE.

The position of the **found** letter in an occurrence of a pattern is the first position where the occurrence of pattern is identified, even if the pattern is not complete.

An occurrence of the pattern DECREASING is the subsequence which matches the regular expression >.

Assume that the occurrence of the pattern DECREASING starts at position $i$ and ends at position $j$. The feature MAX computes the maximum of the values from index $i$ to index $j + 1$.

**Example**

$$\left( \begin{array}{l} 3, \langle 3, 4, 2, 2, 5, 6, 6, 4, 4, 3, 1, 1, 4, 6, 4, 4 \rangle, \\ \langle 0, 0, 0, 0, 0, 0, 0, 0, 0, 0, 1, 0, 0, 0, 0, 0 \rangle \end{array} \right)$$

Figure 3.1053 provides an example where the POS_MIN_MAX_DECREASING $(3, [3, 4, 2, 2, 5, 6, 6, 4, 4, 3, 1, 1, 4, 6, 4, 4], [0, 0, 0, 0, 0, 0, 0, 0, 0, 0, 1, 0, 0, 0, 0, 0])$ constraint holds.

**Typical**

|VARIABLES| $> 1$

range(VARIABLES.var) $> 1$

Figure 3.1053: Illustrating the `POS_MIN_MAX_DECREASING` constraint of the **Example** slot

| Arg. properties | • Functional dependency: `VALUE` determined by `VARIABLES`. |
|---|---|
|  | • Functional dependency: `FOUND` determined by `VARIABLES`. |

**Automaton**          Similar to the automaton of the MIN_MAX_DECREASING constraint but use the decoration
                       table 2.33.

# 3.489 POS_MIN_MAX_DECREASING_SEQUENCE

    **DESCRIPTION**        **AUTOMATON**

**Origin**           Based on constraint MIN_MAX_DECREASING_SEQUENCE.

**Constraint**       POS_MIN_MAX_DECREASING_SEQUENCE(VALUE, VARIABLES, FOUND)

**Arguments**
```
VALUE       : dvar
VARIABLES   : collection(var−dvar)
FOUND       : collection(var−dvar)
```

**Restrictions**     $sv \leq 1 \lor rv \leq 1 \Rightarrow$ VALUE $= +\infty$
                     VALUE $\geq$ minv $+ 1$ ①
                     VALUE $= +\infty \lor$ VALUE $\leq$ maxv
                     required(VARIABLES, var)
                     required(FOUND, var)
                     where
                     minv $=$ minval(VARIABLES.var)
                     maxv $=$ maxval(VARIABLES.var)
                     sv $= |$VARIABLES$|$
                     rv $=$ range(VARIABLES.var)

**Purpose**
The constraint MIN_MAX_DECREASING_SEQUENCE(VALUE, VARIABLES) holds. In addition, FOUND is a collection of 0/1 variables where the value 1 indicates the position of the **found** letter in those occurrences of the pattern DECREASING_SEQUENCE for which the feature value is VALUE.

The position of the **found** letter in an occurrence of a pattern is the first position where the occurrence of pattern is identified, even if the pattern is not complete.

An occurrence of the pattern DECREASING_SEQUENCE is the *maximal* subsequence which matches the regular expression $> (> | =)^* | >$.

Assume that the occurrence of the pattern DECREASING_SEQUENCE starts at position $i$ and ends at position $j$. The feature MAX computes the maximum of the values from index $i$ to index $j + 1$.

**Example**
$$\left( \begin{array}{l} 4, \langle 3, 4, 2, 2, 5, 6, 6, 4, 4, 3, 1, 1, 4, 6, 4, 4 \rangle, \\ \langle 0, 1, 0, 0, 0, 0, 0, 0, 0, 0, 0, 0, 0, 0, 0, 0 \rangle \end{array} \right)$$

Figure 3.1054 provides an example where the POS_MIN_MAX_DECREASING_SEQUENCE $(4, [3, 4, 2, 2, 5, 6, 6, 4, 4, 3, 1, 1, 4, 6, 4, 4], [0, 1, 0, 0, 0, 0, 0, 0, 0, 0, 0, 0, 0, 0, 0, 0])$ constraint holds.

**Typical**
$|$VARIABLES$| > 1$
range(VARIABLES.var) $> 1$

Figure 3.1054: Illustrating the POS_MIN_MAX_DECREASING_SEQUENCE constraint of the **Example** slot

**Arg. properties**

- Functional dependency: VALUE determined by VARIABLES.
- Functional dependency: FOUND determined by VARIABLES.

**Automaton**        Similar to the automaton of the MIN_MAX_DECREASING_SEQUENCE constraint but use the
                     decoration table 2.33.

# 3.490 POS_MIN_MAX_DIP_ON_INCREASING_SEQUENCE

► ▷ ◁ ◀      **DESCRIPTION**      **AUTOMATON**

| | |
|---|---|
| **Origin** | Based on constraint MIN_MAX_DIP_ON_INCREASING_SEQUENCE. |
| **Constraint** | POS_MIN_MAX_DIP_ON_INCREASING_SEQUENCE(VALUE, VARIABLES, FOUND) |

| **Arguments** | |
|---|---|
| VALUE | : dvar |
| VARIABLES | : collection(var−dvar) |
| FOUND | : collection(var−dvar) |

**Restrictions**

$\mathtt{sv} \leq 5 \lor \mathtt{rv} \leq 2 \Rightarrow \mathtt{VALUE} = +\infty$

$\mathtt{VALUE} \geq \mathtt{minv} + 2$①

$\mathtt{VALUE} = +\infty \lor \mathtt{VALUE} \leq \mathtt{maxv}$

required(VARIABLES, var)

required(FOUND, var)

where

$\mathtt{minv} = $ minval(VARIABLES.var)

$\mathtt{maxv} = $ maxval(VARIABLES.var)

$\mathtt{sv} = |\mathtt{VARIABLES}|$

$\mathtt{rv} = $ range(VARIABLES.var)

**Purpose**

The constraint MIN_MAX_DIP_ON_INCREASING_SEQUENCE(VALUE, VARIABLES) holds. In addition, FOUND is a collection of 0/1 variables where the value 1 indicates the position of the **found** letter in those occurrences of the pattern DIP_ON_INCREASING_SEQUENCE for which the feature value is VALUE.

The position of the **found** letter in an occurrence of a pattern is the first position where the occurrence of pattern is identified, even if the pattern is not complete.

An occurrence of the pattern DIP_ON_INCREASING_SEQUENCE is the subsequence which matches the regular expression `<<><<`.

Assume that the occurrence of the pattern DIP_ON_INCREASING_SEQUENCE starts at position $i$ and ends at position $j$. The feature MAX computes the maximum of the values from index $i + 2$ to index $j$.

**Example**

$$\left( \begin{array}{l} 5, \langle 1, 2, 3, 2, 5, 6, 7, 4, 1, 3, 4, 6, 1, 2, 4, 4 \rangle, \\ \langle 0, 0, 0, 0, 1, 0, 0, 0, 0, 0, 0, 0, 0, 0, 0, 0 \rangle \end{array} \right)$$

Figure 3.1055 provides an example where the POS_MIN_MAX_DIP_ON_INCREASING_SEQUENCE $(5, [1, 2, 3, 2, 5, 6, 7, 4, 1, 3, 4, 6, 1, 2, 4, 4], [0, 0, 0, 0, 1, 0, 0, 0, 0, 0, 0, 0, 0, 0, 0, 0])$ constraint holds.

**Typical**

$|\mathtt{VARIABLES}| > 5$

range(VARIABLES.var) > 2

Figure 3.1055: Illustrating the POS_MIN_MAX_DIP_ON_INCREASING_SEQUENCE constraint of the **Example** slot

**Arg. properties**
- Functional dependency: VALUE determined by VARIABLES.
- Functional dependency: FOUND determined by VARIABLES.

**Automaton**     Similar to the automaton of the MIN_MAX_DIP_ON_INCREASING_SEQUENCE constraint but use the decoration table 2.33.

## 3.491   POS_MIN_MAX_INCREASING

▶ ▷ ◁ ◀        **DESCRIPTION**            **AUTOMATON**

| | |
|---|---|
| **Origin** | Based on constraint MIN_MAX_INCREASING. |
| **Constraint** | POS_MIN_MAX_INCREASING(VALUE, VARIABLES, FOUND) |
| **Arguments** | VALUE     : dvar |
| | VARIABLES : collection(var−dvar) |
| | FOUND     : collection(var−dvar) |

**Restrictions**

$sv \leq 1 \vee rv \leq 1 \Rightarrow$ VALUE $= +\infty$

VALUE $\geq$ minv $+ 1$①

VALUE $= +\infty \vee$ VALUE $\leq$ maxv

required(VARIABLES, var)

required(FOUND, var)

where

minv = minval(VARIABLES.var)

maxv = maxval(VARIABLES.var)

sv $= |$VARIABLES$|$

rv = range(VARIABLES.var)

**Purpose**

The constraint MIN_MAX_INCREASING(VALUE, VARIABLES) holds. In addition, FOUND is a collection of 0/1 variables where the value 1 indicates the position of the **found** letter in those occurrences of the pattern INCREASING for which the feature value is VALUE.

The position of the **found** letter in an occurrence of a pattern is the first position where the occurrence of pattern is identified, even if the pattern is not complete.

An occurrence of the pattern INCREASING is the subsequence which matches the regular expression $<$.

Assume that the occurrence of the pattern INCREASING starts at position $i$ and ends at position $j$. The feature MAX computes the maximum of the values from index $i$ to index $j + 1$.

**Example**

$$\left( \begin{array}{l} 3, \langle 4, 3, 5, 5, 2, 1, 1, 3, 3, 4, 6, 6, 3, 1, 1, 3, 3 \rangle, \\ \langle 0, 0, 0, 0, 0, 0, 1, 0, 0, 0, 0, 0, 0, 1, 0, 0 \rangle \end{array} \right)$$

Figure 3.1056 provides an example where the POS_MIN_MAX_INCREASING $(3, [4, 3, 5, 5, 2, 1, 1, 3, 3, 4, 6, 6, 3, 1, 1, 3, 3], [0, 0, 0, 0, 0, 0, 1, 0, 0, 0, 0, 0, 0, 1, 0, 0])$ constraint holds.

**Typical**

$|$VARIABLES$| > 1$

range(VARIABLES.var) $> 1$

Figure 3.1056: Illustrating the POS_MIN_MAX_INCREASING constraint of the **Example** slot

**Arg. properties**
- Functional dependency: VALUE determined by VARIABLES.
- Functional dependency: FOUND determined by VARIABLES.

**Automaton**       Similar to the automaton of the MIN_MAX_INCREASING constraint but use the decoration table 2.33.

## 3.492 POS_MIN_MAX_INCREASING_SEQUENCE

▶ ▷ ◁ ◀     **DESCRIPTION**     **AUTOMATON**

| | |
|---|---|
| **Origin** | Based on constraint MIN_MAX_INCREASING_SEQUENCE. |
| **Constraint** | POS_MIN_MAX_INCREASING_SEQUENCE(VALUE, VARIABLES, FOUND) |
| **Arguments** | VALUE : dvar<br>VARIABLES : collection(var−dvar)<br>FOUND : collection(var−dvar) |

**Restrictions**

$$\mathtt{sv} \leq 1 \lor \mathtt{rv} \leq 1 \Rightarrow \mathtt{VALUE} = +\infty$$
$$\mathtt{VALUE} \geq \mathtt{minv} + 1 \ ①$$
$$\mathtt{VALUE} = +\infty \lor \mathtt{VALUE} \leq \mathtt{maxv}$$
required(VARIABLES, var)
required(FOUND, var)
where
$$\mathtt{minv} = \mathtt{minval}(\mathtt{VARIABLES.var})$$
$$\mathtt{maxv} = \mathtt{maxval}(\mathtt{VARIABLES.var})$$
$$\mathtt{sv} = |\mathtt{VARIABLES}|$$
$$\mathtt{rv} = \mathtt{range}(\mathtt{VARIABLES.var})$$

**Purpose**

The constraint MIN_MAX_INCREASING_SEQUENCE(VALUE, VARIABLES) holds. In addition, FOUND is a collection of 0/1 variables where the value 1 indicates the position of the **found** letter in those occurrences of the pattern INCREASING_SEQUENCE for which the feature value is VALUE.

The position of the **found** letter in an occurrence of a pattern is the first position where the occurrence of pattern is identified, even if the pattern is not complete.

An occurrence of the pattern INCREASING_SEQUENCE is the *maximal* subsequence which matches the regular expression $< (< | =)^* < | <$.

Assume that the occurrence of the pattern INCREASING_SEQUENCE starts at position $i$ and ends at position $j$. The feature MAX computes the maximum of the values from index $i$ to index $j + 1$.

**Example**

$$\left( \begin{array}{l} 3, \langle 4, 3, 5, 5, 2, 1, 1, 3, 3, 4, 6, 6, 3, 1, 1, 3, 3 \rangle, \\ \langle 0, 0, 0, 0, 0, 0, 0, 0, 0, 0, 0, 0, 0, 0, 1, 0, 0 \rangle \end{array} \right)$$

Figure 3.1057 provides an example where the POS_MIN_MAX_INCREASING_SEQUENCE $(3, [4, 3, 5, 5, 2, 1, 1, 3, 3, 4, 6, 6, 3, 1, 1, 3, 3], [0, 0, 0, 0, 0, 0, 0, 0, 0, 0, 0, 0, 0, 0, 1, 0, 0])$ constraint holds.

**Typical**

$$|\mathtt{VARIABLES}| > 1$$
range(VARIABLES.var) > 1

Figure 3.1057: Illustrating the POS_MIN_MAX_INCREASING_SEQUENCE constraint of the **Example** slot

**Arg. properties**
- Functional dependency: VALUE determined by VARIABLES.
- Functional dependency: FOUND determined by VARIABLES.

**Automaton**          Similar to the automaton of the MIN_MAX_INCREASING_SEQUENCE constraint but use the
                       decoration table 2.33.

# 3.493 POS_MIN_MAX_INFLEXION

▶ ▷ ◁ ◀        **DESCRIPTION**        **AUTOMATON**

$< (< | =)^* > | > (> | =)^* <$

**Origin**            Based on constraint MIN_MAX_INFLEXION.

**Constraint**        POS_MIN_MAX_INFLEXION(VALUE, VARIABLES, FOUND)

**Arguments**         VALUE      : dvar
                      VARIABLES  : collection(var−dvar)
                      FOUND      : collection(var−dvar)

**Restrictions**      $\mathtt{sv} \leq 2 \lor \mathtt{rv} \leq 1 \Rightarrow \mathtt{VALUE} = +\infty$
                      $\mathtt{VALUE} \geq \mathtt{minv}$①
                      $\mathtt{VALUE} = +\infty \lor \mathtt{VALUE} \leq \mathtt{maxv}$
                      required(VARIABLES, var)
                      required(FOUND, var)
                      where
                      $\mathtt{minv} = \mathtt{minval}(\mathtt{VARIABLES.var})$
                      $\mathtt{maxv} = \mathtt{maxval}(\mathtt{VARIABLES.var})$
                      $\mathtt{sv} = |\mathtt{VARIABLES}|$
                      $\mathtt{rv} = \mathtt{range}(\mathtt{VARIABLES.var})$

**Purpose**           The constraint MIN_MAX_INFLEXION(VALUE, VARIABLES) holds. In addition, FOUND is
                      a collection of 0/1 variables where the value 1 indicates the position of the **found** letter
                      in those occurrences of the pattern INFLEXION for which the feature value is VALUE.
                      The position of the **found** letter in an occurrence of a pattern is the first position where
                      the occurrence of pattern is identified, even if the pattern is not complete.
                      An occurrence of the pattern INFLEXION is the *maximal* subsequence which matches the
                      regular expression $< (< | =)^* > | > (> | =)^* <$.
                      Assume that the occurrence of the pattern INFLEXION starts at position $i$ and ends at
                      position $j$. The feature MAX computes the maximum of the values from index $i + 1$ to
                      index $j$.

**Example**           $\left( \begin{array}{l} 1, \langle 1, 2, 6, 6, 4, 4, 3, 5, 2, 5, 1, 5, 3, 3, 4, 4 \rangle, \\ \langle 0, 0, 0, 0, 0, 0, 0, 0, 0, 0, 1, 0, 0, 0, 0, 0 \rangle \end{array} \right)$

                      Figure 3.1058 provides an example where the POS_MIN_MAX_INFLEXION
                      $(1, [1, 2, 6, 6, 4, 4, 3, 5, 2, 5, 1, 5, 3, 3, 4, 4], [0, 0, 0, 0, 0, 0, 0, 0, 0, 0, 1, 0, 0, 0, 0, 0])$
                      constraint holds.

**Typical**           $|\mathtt{VARIABLES}| > 2$
                      $\mathtt{range}(\mathtt{VARIABLES.var}) > 1$

Figure 3.1058: Illustrating the POS_MIN_MAX_INFLEXION constraint of the **Example** slot

**Arg. properties**
- Functional dependency: VALUE determined by VARIABLES.
- Functional dependency: FOUND determined by VARIABLES.

**Automaton**     Similar to the automaton of the MIN_MAX_INFLEXION constraint but use the decoration table 2.33.

# 3.494 POS_MIN_MAX_PEAK

**DESCRIPTION**          **AUTOMATON**

$< (= | <)^* (> | =)^* >$

| | |
|---|---|
| **Origin** | Based on constraint MIN_MAX_PEAK. |
| **Constraint** | POS_MIN_MAX_PEAK(VALUE, VARIABLES, FOUND) |
| **Arguments** | VALUE : dvar<br>VARIABLES : collection(var−dvar)<br>FOUND : collection(var−dvar) |

**Restrictions**

$\text{sv} \leq 2 \vee \text{rv} \leq 1 \Rightarrow \text{VALUE} = +\infty$

$\text{VALUE} \geq \text{minv} + 1①$

$\text{VALUE} = +\infty \vee \text{VALUE} \leq \text{maxv}$

required(VARIABLES, var)

required(FOUND, var)

where

$\text{minv} = \text{minval}(\text{VARIABLES.var})$

$\text{maxv} = \text{maxval}(\text{VARIABLES.var})$

$\text{sv} = |\text{VARIABLES}|$

$\text{rv} = \text{range}(\text{VARIABLES.var})$

**Purpose**

The constraint MIN_MAX_PEAK(VALUE, VARIABLES) holds. In addition, FOUND is a collection of 0/1 variables where the value 1 indicates the position of the **found** letter in those occurrences of the pattern PEAK for which the feature value is VALUE.

The position of the **found** letter in an occurrence of a pattern is the first position where the occurrence of pattern is identified, even if the pattern is not complete.

An occurrence of the pattern PEAK is the *maximal* subsequence which matches the regular expression $< (= | <)^* (> | =)^* >$.

Assume that the occurrence of the pattern PEAK starts at position $i$ and ends at position $j$. The feature MAX computes the maximum of the values from index $i + 1$ to index $j$.

**Example**

$$\left( \begin{array}{l} 3, \langle 7, 5, 5, 1, 4, 5, 2, 2, 3, 5, 6, 2, 3, 3, 3, 1 \rangle, \\ \langle 0, 0, 0, 0, 0, 0, 0, 0, 0, 0, 0, 0, 0, 0, 1, 0 \rangle \end{array} \right)$$

Figure 3.1059 provides an example where the POS_MIN_MAX_PEAK $(3, [7, 5, 5, 1, 4, 5, 2, 2, 3, 5, 6, 2, 3, 3, 3, 1], [0, 0, 0, 0, 0, 0, 0, 0, 0, 0, 0, 0, 0, 0, 1, 0])$ constraint holds.

| | |
|---|---|
| **Typical** | $|\text{VARIABLES}| > 2$<br>range(VARIABLES.var) $> 1$ |
| **Arg. properties** | • Functional dependency: VALUE determined by VARIABLES.<br>• Functional dependency: FOUND determined by VARIABLES. |

Figure 3.1059: Illustrating the POS_MIN_MAX_PEAK constraint of the **Example** slot

**Automaton**          Similar to the automaton of the MIN_MAX_PEAK constraint but use the decoration table 2.33.

# 3.495 POS_MIN_MAX_STRICTLY_DECREASING_SEQUENCE

    DESCRIPTION          AUTOMATON

| | |
|---|---|
| **Origin** | Based on constraint MIN_MAX_STRICTLY_DECREASING_SEQUENCE. |
| **Constraint** | POS_MIN_MAX_STRICTLY_DECREASING_SEQUENCE(VALUE, VARIABLES, FOUND) |
| **Arguments** | VALUE : dvar |
| | VARIABLES : collection(var−dvar) |
| | FOUND : collection(var−dvar) |

**Restrictions**

$sv \leq 1 \vee rv \leq 1 \Rightarrow VALUE = +\infty$

$VALUE \geq minv + 1①$

$VALUE = +\infty \vee VALUE \leq maxv$

required(VARIABLES, var)

required(FOUND, var)

where

$minv = minval(VARIABLES.var)$

$maxv = maxval(VARIABLES.var)$

$sv = |VARIABLES|$

$rv = range(VARIABLES.var)$

**Purpose**

The constraint MIN_MAX_STRICTLY_DECREASING_SEQUENCE(VALUE, VARIABLES) holds. In addition, FOUND is a collection of 0/1 variables where the value 1 indicates the position of the **found** letter in those occurrences of the pattern STRICTLY_DECREASING_SEQUENCE for which the feature value is VALUE.

The position of the **found** letter in an occurrence of a pattern is the first position where the occurrence of pattern is identified, even if the pattern is not complete.

An occurrence of the pattern STRICTLY_DECREASING_SEQUENCE is the *maximal* subsequence which matches the regular expression $>^+$.

Assume that the occurrence of the pattern STRICTLY_DECREASING_SEQUENCE starts at position $i$ and ends at position $j$. The feature MAX computes the maximum of the values from index $i$ to index $j + 1$.

**Example**

$$\left( \begin{array}{l} 4, \langle 4, 4, 6, 4, 1, 1, 3, 4, 4, 6, 6, 5, 2, 2, 4, 3 \rangle, \\ \langle 0, 0, 0, 0, 0, 0, 0, 0, 0, 0, 0, 0, 0, 0, 1, 0 \rangle \end{array} \right)$$

Figure 3.1060 provides an example where the POS_MIN_MAX_STRICTLY_DECREASING_SEQUENCE $(4, [4, 4, 6, 4, 1, 1, 3, 4, 4, 6, 6, 5, 2, 2, 4, 3], [0, 0, 0, 0, 0, 0, 0, 0, 0, 0, 0, 0, 0, 0, 1, 0])$ constraint holds.

**Typical**

$|VARIABLES| > 1$

$range(VARIABLES.var) > 1$

Figure 3.1060: Illustrating the POS_MIN_MAX_STRICTLY_DECREASING_SEQUENCE constraint of the **Example** slot

| Arg. properties | • Functional dependency: VALUE determined by VARIABLES. |
| | • Functional dependency: FOUND determined by VARIABLES. |

**Automaton**    Similar to the automaton of the MIN_MAX_STRICTLY_DECREASING_SEQUENCE constraint but use the decoration table 2.33.

## 3.496   POS_MIN_MAX_STRICTLY_INCREASING_SEQUENCE

▶ ▷ ◁ ◀        **DESCRIPTION**              **AUTOMATON**

| | |
|---|---|
| **Origin** | Based on constraint MIN_MAX_STRICTLY_INCREASING_SEQUENCE. |
| **Constraint** | POS_MIN_MAX_STRICTLY_INCREASING_SEQUENCE(VALUE, VARIABLES, FOUND) |

**Arguments**

    VALUE      : dvar
    VARIABLES  : collection(var−dvar)
    FOUND      : collection(var−dvar)

**Restrictions**

$sv \leq 1 \lor rv \leq 1 \Rightarrow$ VALUE $= +\infty$

VALUE $\geq$ minv $+ 1$ ①

VALUE $= +\infty \lor$ VALUE $\leq$ maxv

required(VARIABLES, var)

required(FOUND, var)

where

minv $=$ minval(VARIABLES.var)

maxv $=$ maxval(VARIABLES.var)

sv $= |$VARIABLES$|$

rv $=$ range(VARIABLES.var)

**Purpose**

The constraint MIN_MAX_STRICTLY_INCREASING_SEQUENCE(VALUE, VARIABLES) holds. In addition, FOUND is a collection of 0/1 variables where the value 1 indicates the position of the **found** letter in those occurrences of the pattern STRICTLY_INCREASING_SEQUENCE for which the feature value is VALUE.

The position of the **found** letter in an occurrence of a pattern is the first position where the occurrence of pattern is identified, even if the pattern is not complete.

An occurrence of the pattern STRICTLY_INCREASING_SEQUENCE is the *maximal* sub-sequence which matches the regular expression $<^+$.

Assume that the occurrence of the pattern STRICTLY_INCREASING_SEQUENCE starts at position $i$ and ends at position $j$. The feature MAX computes the maximum of the values from index $i$ to index $j + 1$.

**Example**

$$\left( \begin{array}{l} 3, \langle 4, 3, 5, 5, 2, 1, 1, 2, 3, 4, 6, 6, 3, 1, 2, 3 \rangle, \\ \langle 0, 0, 0, 0, 0, 0, 0, 0, 0, 0, 0, 0, 0, 0, 1, 0, 0 \rangle \end{array} \right)$$

Figure 3.1061 provides an example where the POS_MIN_MAX_STRICTLY_INCREASING_SEQUENCE $(3, [4, 3, 5, 5, 2, 1, 1, 2, 3, 4, 6, 6, 3, 1, 2, 3], [0, 0, 0, 0, 0, 0, 0, 0, 0, 0, 0, 0, 0, 0, 1, 0, 0])$ constraint holds.

**Typical**

$|$VARIABLES$| > 1$

range(VARIABLES.var) $> 1$

Figure 3.1061: Illustrating the POS_MIN_MAX_STRICTLY_INCREASING_SEQUENCE constraint of the **Example** slot

**Arg. properties**
- Functional dependency: VALUE determined by VARIABLES.
- Functional dependency: FOUND determined by VARIABLES.

**Automaton**    Similar to the automaton of the MIN_MAX_STRICTLY_INCREASING_SEQUENCE constraint but use the decoration table 2.33.

## 3.497 POS_MIN_MAX_SUMMIT

▶ ▷ ◁ ◀        **DESCRIPTION**        **AUTOMATON**

$(< | < (= | <)^* <)(> | > (= | >)^* >)$

| | |
|---|---|
| **Origin** | Based on constraint MIN_MAX_SUMMIT. |
| **Constraint** | POS_MIN_MAX_SUMMIT(VALUE, VARIABLES, FOUND) |
| **Arguments** | VALUE      :   dvar |
| | VARIABLES :   collection(var−dvar) |
| | FOUND       :   collection(var−dvar) |

**Restrictions**

$\text{sv} \leq 2 \vee \text{rv} \leq 1 \Rightarrow \text{VALUE} = +\infty$

$\text{VALUE} \geq \text{minv} + 1 ①$

$\text{VALUE} = +\infty \vee \text{VALUE} \leq \text{maxv}$

required(VARIABLES, var)

required(FOUND, var)

where

$\text{minv} = \text{minval}(\text{VARIABLES.var})$

$\text{maxv} = \text{maxval}(\text{VARIABLES.var})$

$\text{sv} = |\text{VARIABLES}|$

$\text{rv} = \text{range}(\text{VARIABLES.var})$

**Purpose**

The constraint MIN_MAX_SUMMIT(VALUE, VARIABLES) holds. In addition, FOUND is a collection of 0/1 variables where the value 1 indicates the position of the **found** letter in those occurrences of the pattern SUMMIT for which the feature value is VALUE.

The position of the **found** letter in an occurrence of a pattern is the first position where the occurrence of pattern is identified, even if the pattern is not complete.

An occurrence of the pattern SUMMIT is the *maximal* subsequence which matches the regular expression $(< | < (= | <)^* <)(> | > (= | >)^* >)$.

Assume that the occurrence of the pattern SUMMIT starts at position $i$ and ends at position $j$. The feature MAX computes the maximum of the values from index $i+1$ to index $j$.

**Example**

$$\left( \begin{array}{l} 3, \langle 7, 1, 5, 4, 4, 3, 3, 4, 6, 6, 2, 3, 4, 2, 3, 1 \rangle, \\ \langle 0, 0, 0, 0, 0, 0, 0, 0, 0, 0, 0, 0, 0, 0, 1, 0 \rangle \end{array} \right)$$

Figure 3.1062 provides an example where the POS_MIN_MAX_SUMMIT $(3, [7, 1, 5, 4, 4, 3, 3, 4, 6, 6, 2, 3, 4, 2, 3, 1], [0, 0, 0, 0, 0, 0, 0, 0, 0, 0, 0, 0, 0, 0, 1, 0])$ constraint holds.

**Typical**

$|\text{VARIABLES}| > 2$

$\text{range}(\text{VARIABLES.var}) > 1$

Figure 3.1062: Illustrating the POS_MIN_MAX_SUMMIT constraint of the **Example** slot

**Arg. properties**
- Functional dependency: VALUE determined by VARIABLES.
- Functional dependency: FOUND determined by VARIABLES.

**Automaton** Similar to the automaton of the MIN_MAX_SUMMIT constraint but use the decoration table 2.33.

## 3.498 POS_MIN_MAX_ZIGZAG

▶ ▷ ◁ ◀    DESCRIPTION    AUTOMATON

**Origin** Based on constraint MIN_MAX_ZIGZAG.

**Constraint** POS_MIN_MAX_ZIGZAG(VALUE, VARIABLES, FOUND)

**Arguments**

```
VALUE     : dvar
VARIABLES : collection(var−dvar)
FOUND     : collection(var−dvar)
```

**Restrictions**

$sv \leq 3 \lor rv \leq 1 \Rightarrow VALUE = +\infty$

$VALUE \geq minv + 1$①

$VALUE = +\infty \lor VALUE \leq maxv$

required(VARIABLES, var)

required(FOUND, var)

where

$minv = minval(VARIABLES.var)$

$maxv = maxval(VARIABLES.var)$

$sv = |VARIABLES|$

$rv = range(VARIABLES.var)$

**Purpose**

The constraint MIN_MAX_ZIGZAG(VALUE, VARIABLES) holds. In addition, FOUND is a collection of 0/1 variables where the value 1 indicates the position of the **found** letter in those occurrences of the pattern ZIGZAG for which the feature value is VALUE.

The position of the **found** letter in an occurrence of a pattern is the first position where the occurrence of pattern is identified, even if the pattern is not complete.

An occurrence of the pattern ZIGZAG is the *maximal* subsequence which matches the regular expression $(<>)^+(< | <>) | (><)^+(> | ><)$.

Assume that the occurrence of the pattern ZIGZAG starts at position $i$ and ends at position $j$. The feature MAX computes the maximum of the values from index $i + 1$ to index $j$.

**Example**

$$\left( \begin{array}{l} 3, \langle 4, 1, 3, 1, 4, 6, 1, 5, 5, 2, 7, 2, 3, 1, 6, 1 \rangle, \\ \langle 0, 0, 1, 0, 0, 0, 0, 0, 0, 0, 0, 0, 0, 0, 0, 0 \rangle \end{array} \right)$$

Figure 3.1063 provides an example where the POS_MIN_MAX_ZIGZAG $(3, [4, 1, 3, 1, 4, 6, 1, 5, 5, 2, 7, 2, 3, 1, 6, 1], [0, 0, 1, 0, 0, 0, 0, 0, 0, 0, 0, 0, 0, 0, 0, 0])$ constraint holds.

**Typical**

$|VARIABLES| > 3$

$range(VARIABLES.var) > 1$

**Arg. properties**

- Functional dependency: VALUE determined by VARIABLES.
- Functional dependency: FOUND determined by VARIABLES.

Figure 3.1063: Illustrating the POS_MIN_MAX_ZIGZAG constraint of the **Example** slot

**Automaton**        Similar to the automaton of the MIN_MAX_ZIGZAG constraint but use the decoration ta-
                     ble 2.33.

# 3.499   POS_MIN_MIN_BUMP_ON_DECREASING_SEQUENCE

▶ ▷ ◁ ◀    DESCRIPTION          AUTOMATON

**Origin**          Based on constraint MIN_MIN_BUMP_ON_DECREASING_SEQUENCE.

**Constraint**      POS_MIN_MIN_BUMP_ON_DECREASING_SEQUENCE(VALUE, VARIABLES, FOUND)

**Arguments**       VALUE      : dvar
                    VARIABLES  : collection(var−dvar)
                    FOUND      : collection(var−dvar)

**Restrictions**    $\mathtt{sv} \leq 5 \vee \mathtt{rv} \leq 2 \Rightarrow \mathtt{VALUE} = +\infty$

                    $\mathtt{VALUE} \geq \mathtt{minv}$①
                    $\mathtt{VALUE} = +\infty \vee \mathtt{VALUE} \leq \mathtt{maxv} - 2$
                    required(VARIABLES, var)
                    required(FOUND, var)
                    where
                    minv =minval(VARIABLES.var)
                    maxv =maxval(VARIABLES.var)
                    sv = |VARIABLES|
                    rv =range(VARIABLES.var)

**Purpose**         The constraint MIN_MIN_BUMP_ON_DECREASING_SEQUENCE(VALUE, VARIABLES) holds. In addition, FOUND is a collection of 0/1 variables where the value 1 indicates the position of the **found** letter in those occurrences of the pattern BUMP_ON_DECREASING_SEQUENCE for which the feature value is VALUE.
                    The position of the **found** letter in an occurrence of a pattern is the first position where the occurrence of pattern is identified, even if the pattern is not complete.
                    An occurrence of the pattern BUMP_ON_DECREASING_SEQUENCE is the subsequence which matches the regular expression >><>.
                    Assume that the occurrence of the pattern BUMP_ON_DECREASING_SEQUENCE starts at position $i$ and ends at position $j$. The feature MIN computes the minimum of the values from index $i + 2$ to index $j$.

**Example**         $\left( \begin{array}{l} 2, \langle 7, 6, 5, 6, 5, 4, 1, 4, 7, 5, 4, 2, 5, 4, 3, 3 \rangle, \\ \langle 0, 0, 0, 0, 0, 0, 0, 0, 0, 0, 0, 0, 0, 1, 0, 0 \rangle \end{array} \right)$

                    Figure 3.1064 provides an example where the POS_MIN_MIN_BUMP_ON_DECREASING_SEQUENCE $(2, [7, 6, 5, 6, 5, 4, 1, 4, 7, 5, 4, 2, 5, 4, 3, 3], [0, 0, 0, 0, 0, 0, 0, 0, 0, 0, 0, 0, 0, 1, 0, 0])$ constraint holds.

**Typical**         |VARIABLES| > 5
                    range(VARIABLES.var) > 2

Figure 3.1064: Illustrating the POS_MIN_MIN_BUMP_ON_DECREASING_SEQUENCE constraint of the **Example** slot

**Arg. properties**

- Functional dependency: VALUE determined by VARIABLES.
- Functional dependency: FOUND determined by VARIABLES.

**Automaton**        Similar to the automaton of the MIN_MIN_BUMP_ON_DECREASING_SEQUENCE constraint
                     but use the decoration table 2.33.

## 3.500 POS_MIN_MIN_DECREASING

    DESCRIPTION          AUTOMATON

| | |
|---|---|
| **Origin** | Based on constraint MIN_MIN_DECREASING. |
| **Constraint** | POS_MIN_MIN_DECREASING(VALUE, VARIABLES, FOUND) |

**Arguments**

```
VALUE      : dvar
VARIABLES  : collection(var−dvar)
FOUND      : collection(var−dvar)
```

**Restrictions**

$sv \leq 1 \lor rv \leq 1 \Rightarrow VALUE = +\infty$

$VALUE \geq minv①$

$VALUE = +\infty \lor VALUE \leq maxv − 1$

required(VARIABLES, var)

required(FOUND, var)

where

minv = minval(VARIABLES.var)

maxv = maxval(VARIABLES.var)

sv = |VARIABLES|

rv = range(VARIABLES.var)

**Purpose**

The constraint MIN_MIN_DECREASING(VALUE, VARIABLES) holds. In addition, FOUND is a collection of 0/1 variables where the value 1 indicates the position of the **found** letter in those occurrences of the pattern DECREASING for which the feature value is VALUE.

The position of the **found** letter in an occurrence of a pattern is the first position where the occurrence of pattern is identified, even if the pattern is not complete.

An occurrence of the pattern DECREASING is the subsequence which matches the regular expression >.

Assume that the occurrence of the pattern DECREASING starts at position $i$ and ends at position $j$. The feature MIN computes the minimum of the values from index $i$ to index $j + 1$.

**Example**

$$\left( \begin{array}{l} 1, \langle 3, 4, 2, 2, 5, 6, 6, 4, 4, 3, 1, 1, 4, 6, 4, 4 \rangle, \\ \langle 0, 0, 0, 0, 0, 0, 0, 0, 0, 0, 1, 0, 0, 0, 0, 0 \rangle \end{array} \right)$$

Figure 3.1065 provides an example where the POS_MIN_MIN_DECREASING $(1, [3, 4, 2, 2, 5, 6, 6, 4, 4, 3, 1, 1, 4, 6, 4, 4], [0, 0, 0, 0, 0, 0, 0, 0, 0, 0, 1, 0, 0, 0, 0, 0])$ constraint holds.

**Typical**

|VARIABLES| > 1

range(VARIABLES.var) > 1

Figure 3.1065: Illustrating the POS_MIN_MIN_DECREASING constraint of the **Example** slot

**Arg. properties**
- Functional dependency: VALUE determined by VARIABLES.
- Functional dependency: FOUND determined by VARIABLES.

**Automaton**  Similar to the automaton of the MIN_MIN_DECREASING constraint but use the decoration table 2.33.

# 3.501 POS_MIN_MIN_DECREASING_SEQUENCE

    DESCRIPTION    AUTOMATON

| | |
|---|---|
| **Origin** | Based on constraint MIN_MIN_DECREASING_SEQUENCE. |
| **Constraint** | POS_MIN_MIN_DECREASING_SEQUENCE(VALUE, VARIABLES, FOUND) |

| **Arguments** | | |
|---|---|---|
| | VALUE | : dvar |
| | VARIABLES | : collection(var−dvar) |
| | FOUND | : collection(var−dvar) |

**Restrictions**

$sv \leq 1 \vee rv \leq 1 \Rightarrow VALUE = +\infty$

$VALUE \geq minv①$

$VALUE = +\infty \vee VALUE \leq maxv − 1$

required(VARIABLES, var)

required(FOUND, var)

where

minv = minval(VARIABLES.var)

maxv = maxval(VARIABLES.var)

sv = |VARIABLES|

rv = range(VARIABLES.var)

**Purpose**

The constraint MIN_MIN_DECREASING_SEQUENCE(VALUE, VARIABLES) holds. In addition, FOUND is a collection of 0/1 variables where the value 1 indicates the position of the **found** letter in those occurrences of the pattern DECREASING_SEQUENCE for which the feature value is VALUE.

The position of the **found** letter in an occurrence of a pattern is the first position where the occurrence of pattern is identified, even if the pattern is not complete.

An occurrence of the pattern DECREASING_SEQUENCE is the *maximal* subsequence which matches the regular expression $> (> | =)^* > | >$.

Assume that the occurrence of the pattern DECREASING_SEQUENCE starts at position $i$ and ends at position $j$. The feature MIN computes the minimum of the values from index $i$ to index $j + 1$.

**Example**

$$\left( \begin{array}{l} 1, \langle 3, 4, 2, 2, 5, 6, 6, 4, 4, 3, 1, 1, 4, 6, 4, 4 \rangle, \\ \langle 0, 0, 0, 0, 0, 0, 1, 0, 0, 0, 0, 0, 0, 0, 0, 0 \rangle \end{array} \right)$$

Figure 3.1066 provides an example where the POS_MIN_MIN_DECREASING_SEQUENCE $(1, [3, 4, 2, 2, 5, 6, 6, 4, 4, 3, 1, 1, 4, 6, 4, 4], [0, 0, 0, 0, 0, 0, 1, 0, 0, 0, 0, 0, 0, 0, 0, 0])$ constraint holds.

**Typical**

|VARIABLES| > 1

range(VARIABLES.var) > 1

Figure 3.1066: Illustrating the POS_MIN_MIN_DECREASING_SEQUENCE constraint of the **Example** slot

**Arg. properties**

- Functional dependency: VALUE determined by VARIABLES.
- Functional dependency: FOUND determined by VARIABLES.

**Automaton**          Similar to the automaton of the MIN_MIN_DECREASING_SEQUENCE constraint but use the
                       decoration table 2.33.

# 3.502   POS_MIN_MIN_DIP_ON_INCREASING_SEQUENCE

► ▷ ◁ ◄      **DESCRIPTION**          **AUTOMATON**

| | |
|---|---|
| **Origin** | Based on constraint MIN_MIN_DIP_ON_INCREASING_SEQUENCE. |
| **Constraint** | POS_MIN_MIN_DIP_ON_INCREASING_SEQUENCE(VALUE, VARIABLES, FOUND) |
| **Arguments** | VALUE     : dvar<br>VARIABLES : collection(var−dvar)<br>FOUND     : collection(var−dvar) |
| **Restrictions** | $\text{sv} \leq 5 \lor \text{rv} \leq 2 \Rightarrow \text{VALUE} = +\infty$<br>$\text{VALUE} \geq \text{minv}①$<br>$\text{VALUE} = +\infty \lor \text{VALUE} \leq \text{maxv} - 2$<br>required(VARIABLES, var)<br>required(FOUND, var)<br>where<br>$\text{minv} = \text{minval}(\text{VARIABLES.var})$<br>$\text{maxv} = \text{maxval}(\text{VARIABLES.var})$<br>$\text{sv} = |\text{VARIABLES}|$<br>$\text{rv} = \text{range}(\text{VARIABLES.var})$ |

| | |
|---|---|
| **Purpose** | The constraint MIN_MIN_DIP_ON_INCREASING_SEQUENCE(VALUE, VARIABLES) holds. In addition, FOUND is a collection of 0/1 variables where the value 1 indicates the position of the **found** letter in those occurrences of the pattern DIP_ON_INCREASING_SEQUENCE for which the feature value is VALUE.<br>The position of the **found** letter in an occurrence of a pattern is the first position where the occurrence of pattern is identified, even if the pattern is not complete.<br>An occurrence of the pattern DIP_ON_INCREASING_SEQUENCE is the subsequence which matches the regular expression <<><<.<br>Assume that the occurrence of the pattern DIP_ON_INCREASING_SEQUENCE starts at position $i$ and ends at position $j$. The feature MIN computes the minimum of the values from index $i + 2$ to index $j$. |
| **Example** | $\left( \begin{array}{l} 1, \langle 1, 2, 3, 2, 5, 6, 7, 4, 1, 3, 4, 6, 1, 2, 4, 4 \rangle, \\ \langle 0, 0, 0, 0, 0, 0, 0, 0, 0, 0, 0, 0, 0, 1, 0, 0 \rangle \end{array} \right)$ |

Figure 3.1067 provides an example where the POS_MIN_MIN_DIP_ON_INCREASING_SEQUENCE $(1, [1, 2, 3, 2, 5, 6, 7, 4, 1, 3, 4, 6, 1, 2, 4, 4], [0, 0, 0, 0, 0, 0, 0, 0, 0, 0, 0, 0, 0, 1, 0, 0])$ constraint holds.

| | |
|---|---|
| **Typical** | $|\text{VARIABLES}| > 5$<br>range(VARIABLES.var) $> 2$ |

Figure 3.1067: Illustrating the `POS_MIN_MIN_DIP_ON_INCREASING_SEQUENCE` constraint of the **Example** slot

**Arg. properties**

- Functional dependency: `VALUE` determined by `VARIABLES`.
- Functional dependency: `FOUND` determined by `VARIABLES`.

**Automaton**     Similar to the automaton of the MIN_MIN_DIP_ON_INCREASING_SEQUENCE constraint but use the decoration table 2.33.

# 3.503 POS_MIN_MIN_GORGE

          **DESCRIPTION**          **AUTOMATON**

$(> | > (= | >)^*>)(< | < (= | <)^* <)$

| | |
|---|---|
| **Origin** | Based on constraint MIN_MIN_GORGE. |
| **Constraint** | POS_MIN_MIN_GORGE(VALUE, VARIABLES, FOUND) |
| **Arguments** | VALUE : dvar<br>VARIABLES : collection(var−dvar)<br>FOUND : collection(var−dvar) |
| **Restrictions** | $\mathtt{sv} \leq 2 \vee \mathtt{rv} \leq 1 \Rightarrow \mathtt{VALUE} = +\infty$<br>$\mathtt{VALUE} \geq \mathtt{minv}①$<br>$\mathtt{VALUE} = +\infty \vee \mathtt{VALUE} \leq \mathtt{maxv} - 1$<br>required(VARIABLES, var)<br>required(FOUND, var)<br>where<br>$\mathtt{minv} = \mathtt{minval}(\mathtt{VARIABLES.var})$<br>$\mathtt{maxv} = \mathtt{maxval}(\mathtt{VARIABLES.var})$<br>$\mathtt{sv} = |\mathtt{VARIABLES}|$<br>$\mathtt{rv} = \mathtt{range}(\mathtt{VARIABLES.var})$ |

| | |
|---|---|
| **Purpose** | The constraint MIN_MIN_GORGE(VALUE, VARIABLES) holds. In addition, FOUND is a collection of 0/1 variables where the value 1 indicates the position of the **found** letter in those occurrences of the pattern GORGE for which the feature value is VALUE.<br>The position of the **found** letter in an occurrence of a pattern is the first position where the occurrence of pattern is identified, even if the pattern is not complete.<br>An occurrence of the pattern GORGE is the *maximal* subsequence which matches the regular expression $(> | > (= | >)^*>)(< | < (= | <)^* <)$.<br>Assume that the occurrence of the pattern GORGE starts at position $i$ and ends at position $j$. The feature MIN computes the minimum of the values from index $i+1$ to index $j$. |

| | |
|---|---|
| **Example** | $\left(\begin{array}{l} 3, \langle 1, 7, 3, 4, 4, 5, 5, 4, 2, 2, 6, 5, 4, 6, 5, 7 \rangle, \\ \langle 0, 0, 1, 0, 0, 0, 0, 0, 0, 0, 0, 0, 0, 0, 0, 0 \rangle \end{array}\right)$ |

Figure 3.1068 provides an example where the POS_MIN_MIN_GORGE $(3, [1, 7, 3, 4, 4, 5, 5, 4, 2, 2, 6, 5, 4, 6, 5, 7], [0, 0, 1, 0, 0, 0, 0, 0, 0, 0, 0, 0, 0, 0, 0, 0])$ constraint holds.

| | |
|---|---|
| **Typical** | $|\mathtt{VARIABLES}| > 2$<br>$\mathtt{range}(\mathtt{VARIABLES.var}) > 1$ |

| | |
|---|---|
| **Arg. properties** | • Functional dependency: VALUE determined by VARIABLES.<br>• Functional dependency: FOUND determined by VARIABLES. |

Figure 3.1068: Illustrating the POS_MIN_MIN_GORGE constraint of the **Example** slot

**Automaton**          Similar to the automaton of the MIN_MIN_GORGE constraint but use the decoration table 2.33.

## 3.504 POS_MIN_MIN_INCREASING

▶ ▷ ◁ ◀    DESCRIPTION    AUTOMATON

**Origin**    Based on constraint MIN_MIN_INCREASING.

**Constraint**    POS_MIN_MIN_INCREASING(VALUE, VARIABLES, FOUND)

**Arguments**
```
VALUE     : dvar
VARIABLES : collection(var−dvar)
FOUND     : collection(var−dvar)
```

**Restrictions**
$\mathtt{sv} \leq 1 \vee \mathtt{rv} \leq 1 \Rightarrow \mathtt{VALUE} = +\infty$
$\mathtt{VALUE} \geq \mathtt{minv}①$
$\mathtt{VALUE} = +\infty \vee \mathtt{VALUE} \leq \mathtt{maxv} - 1$
required(VARIABLES, var)
required(FOUND, var)
where
$\mathtt{minv} = \mathtt{minval}(\mathtt{VARIABLES.var})$
$\mathtt{maxv} = \mathtt{maxval}(\mathtt{VARIABLES.var})$
$\mathtt{sv} = |\mathtt{VARIABLES}|$
$\mathtt{rv} = \mathtt{range}(\mathtt{VARIABLES.var})$

**Purpose**
The constraint MIN_MIN_INCREASING(VALUE, VARIABLES) holds. In addition, FOUND is a collection of 0/1 variables where the value 1 indicates the position of the **found** letter in those occurrences of the pattern INCREASING for which the feature value is VALUE.
The position of the **found** letter in an occurrence of a pattern is the first position where the occurrence of pattern is identified, even if the pattern is not complete.
An occurrence of the pattern INCREASING is the subsequence which matches the regular expression <.
Assume that the occurrence of the pattern INCREASING starts at position $i$ and ends at position $j$. The feature MIN computes the minimum of the values from index $i$ to index $j + 1$.

**Example**
$$\left( \begin{array}{l} 1, \langle 4, 3, 5, 5, 2, 1, 1, 3, 3, 4, 6, 6, 3, 1, 1, 3, 3 \rangle, \\ \langle 0, 0, 0, 0, 0, 0, 1, 0, 0, 0, 0, 0, 0, 0, 1, 0, 0 \rangle \end{array} \right)$$

Figure 3.1069 provides an example where the POS_MIN_MIN_INCREASING
$(1, [4, 3, 5, 5, 2, 1, 1, 3, 3, 4, 6, 6, 3, 1, 1, 3, 3], [0, 0, 0, 0, 0, 0, 1, 0, 0, 0, 0, 0, 0, 0, 1, 0, 0])$
constraint holds.

**Typical**
$|\mathtt{VARIABLES}| > 1$
$\mathtt{range}(\mathtt{VARIABLES.var}) > 1$

Figure 3.1069: Illustrating the POS_MIN_MIN_INCREASING constraint of the **Example** slot

| Arg. properties | • Functional dependency: VALUE determined by VARIABLES. |
|---|---|
| | • Functional dependency: FOUND determined by VARIABLES. |

**Automaton**        Similar to the automaton of the MIN_MIN_INCREASING constraint but use the decoration table 2.33.

## 3.505   POS_MIN_MIN_INCREASING_SEQUENCE

▶ ▷ ◁ ◀    **DESCRIPTION**    **AUTOMATON**

| | |
|---|---|
| **Origin** | Based on constraint MIN_MIN_INCREASING_SEQUENCE. |
| **Constraint** | POS_MIN_MIN_INCREASING_SEQUENCE(VALUE, VARIABLES, FOUND) |

| **Arguments** | |
|---|---|
| VALUE | : dvar |
| VARIABLES | : collection(var−dvar) |
| FOUND | : collection(var−dvar) |

**Restrictions**

$sv \leq 1 \lor rv \leq 1 \Rightarrow$ VALUE $= +\infty$
VALUE $\geq$ minv①
VALUE $= +\infty \lor$ VALUE $\leq$ maxv $- 1$
required(VARIABLES, var)
required(FOUND, var)
where
minv =minval(VARIABLES.var)
maxv =maxval(VARIABLES.var)
sv = |VARIABLES|
rv =range(VARIABLES.var)

**Purpose**

The constraint MIN_MIN_INCREASING_SEQUENCE(VALUE, VARIABLES) holds. In addition, FOUND is a collection of 0/1 variables where the value 1 indicates the position of the **found** letter in those occurrences of the pattern INCREASING_SEQUENCE for which the feature value is VALUE.

The position of the **found** letter in an occurrence of a pattern is the first position where the occurrence of pattern is identified, even if the pattern is not complete.

An occurrence of the pattern INCREASING_SEQUENCE is the *maximal* subsequence which matches the regular expression $< (< | =)^* < | <$.

Assume that the occurrence of the pattern INCREASING_SEQUENCE starts at position $i$ and ends at position $j$. The feature MIN computes the minimum of the values from index $i$ to index $j + 1$.

**Example**

$$\left( \begin{array}{l} 1, \langle 4, 3, 5, 5, 2, 1, 1, 3, 3, 4, 6, 6, 3, 1, 1, 3, 3 \rangle, \\ \langle 0, 0, 0, 0, 0, 0, 1, 0, 0, 0, 0, 0, 0, 0, 1, 0, 0 \rangle \end{array} \right)$$

Figure 3.1070 provides an example where the POS_MIN_MIN_INCREASING_SEQUENCE $(1, [4, 3, 5, 5, 2, 1, 1, 3, 3, 4, 6, 6, 3, 1, 1, 3, 3], [0, 0, 0, 0, 0, 0, 1, 0, 0, 0, 0, 0, 0, 0, 1, 0, 0])$ constraint holds.

**Typical**

|VARIABLES| > 1
range(VARIABLES.var) > 1

Figure 3.1070: Illustrating the POS_MIN_MIN_INCREASING_SEQUENCE constraint of the **Example** slot

**Arg. properties**
- Functional dependency: VALUE determined by VARIABLES.
- Functional dependency: FOUND determined by VARIABLES.

**Automaton**          Similar to the automaton of the MIN_MIN_INCREASING_SEQUENCE constraint but use the
                       decoration table 2.33.

## 3.506　POS_MIN_MIN_INFLEXION

$< (< | =)^* > | > (> | =)^* <$

| | |
|---|---|
| **Origin** | Based on constraint MIN_MIN_INFLEXION. |
| **Constraint** | POS_MIN_MIN_INFLEXION(VALUE, VARIABLES, FOUND) |
| **Arguments** | VALUE　　　　: dvar |
| | VARIABLES : collection(var−dvar) |
| | FOUND　　　　: collection(var−dvar) |
| **Restrictions** | $\mathtt{sv} \leq 2 \vee \mathtt{rv} \leq 1 \Rightarrow \mathtt{VALUE} = +\infty$ |
| | $\mathtt{VALUE} \geq \mathtt{minv}①$ |
| | $\mathtt{VALUE} = +\infty \vee \mathtt{VALUE} \leq \mathtt{maxv}$ |
| | required(VARIABLES, var) |
| | required(FOUND, var) |
| | where |
| | $\mathtt{minv} =$ minval(VARIABLES.var) |
| | $\mathtt{maxv} =$ maxval(VARIABLES.var) |
| | $\mathtt{sv} = |\mathtt{VARIABLES}|$ |
| | $\mathtt{rv} =$ range(VARIABLES.var) |

VALUE $\geq$ 0

**Purpose**

The constraint MIN_MIN_INFLEXION(VALUE, VARIABLES) holds. In addition, FOUND is a collection of 0/1 variables where the value 1 indicates the position of the **found** letter in those occurrences of the pattern INFLEXION for which the feature value is VALUE.

The position of the **found** letter in an occurrence of a pattern is the first position where the occurrence of pattern is identified, even if the pattern is not complete.

An occurrence of the pattern INFLEXION is the *maximal* subsequence which matches the regular expression $< (< | =)^* > | > (> | =)^* <$.

Assume that the occurrence of the pattern INFLEXION starts at position $i$ and ends at position $j$. The feature MIN computes the minimum of the values from index $i + 1$ to index $j$.

**Example**

$$\left( \begin{array}{l} 1, \langle 1, 2, 6, 6, 4, 4, 3, 5, 2, 5, 1, 5, 3, 3, 4, 4 \rangle, \\ \langle 0, 0, 0, 0, 0, 0, 0, 0, 0, 0, 1, 0, 0, 0, 0, 0 \rangle \end{array} \right)$$

Figure 3.1071 provides an example where the POS_MIN_MIN_INFLEXION $(1, [1, 2, 6, 6, 4, 4, 3, 5, 2, 5, 1, 5, 3, 3, 4, 4], [0, 0, 0, 0, 0, 0, 0, 0, 0, 0, 1, 0, 0, 0, 0, 0])$ constraint holds.

**Typical**

|VARIABLES| > 2
range(VARIABLES.var) > 1

Figure 3.1071: Illustrating the POS_MIN_MIN_INFLEXION constraint of the **Example** slot

**Arg. properties**
- Functional dependency: VALUE determined by VARIABLES.
- Functional dependency: FOUND determined by VARIABLES.

**Automaton**        Similar to the automaton of the MIN_MIN_INFLEXION constraint but use the decoration table 2.33.

# 3.507   POS_MIN_MIN_STRICTLY_DECREASING_SEQUENCE

▶ ▷ ◁ ◀      **DESCRIPTION**      **AUTOMATON**

| | |
|---|---|
| **Origin** | Based on constraint MIN_MIN_STRICTLY_DECREASING_SEQUENCE. |
| **Constraint** | POS_MIN_MIN_STRICTLY_DECREASING_SEQUENCE(VALUE, VARIABLES, FOUND) |
| **Arguments** | VALUE      :   dvar <br> VARIABLES :   collection(var−dvar) <br> FOUND      :   collection(var−dvar) |

**Restrictions**

$\mathtt{sv} \leq 1 \lor \mathtt{rv} \leq 1 \Rightarrow \mathtt{VALUE} = +\infty$

$\mathtt{VALUE} \geq \mathtt{minv}①$

$\mathtt{VALUE} = +\infty \lor \mathtt{VALUE} \leq \mathtt{maxv} - 1$

required(VARIABLES, var)

required(FOUND, var)

where

$\mathtt{minv} = \mathtt{minval}(\mathtt{VARIABLES.var})$

$\mathtt{maxv} = \mathtt{maxval}(\mathtt{VARIABLES.var})$

$\mathtt{sv} = |\mathtt{VARIABLES}|$

$\mathtt{rv} = \mathtt{range}(\mathtt{VARIABLES.var})$

**Purpose**

The constraint MIN_MIN_STRICTLY_DECREASING_SEQUENCE(VALUE, VARIABLES) holds. In addition, FOUND is a collection of 0/1 variables where the value 1 indicates the position of the **found** letter in those occurrences of the pattern STRICTLY_DECREASING_SEQUENCE for which the feature value is VALUE.

The position of the **found** letter in an occurrence of a pattern is the first position where the occurrence of pattern is identified, even if the pattern is not complete.

An occurrence of the pattern STRICTLY_DECREASING_SEQUENCE is the *maximal* subsequence which matches the regular expression $>^+$.

Assume that the occurrence of the pattern STRICTLY_DECREASING_SEQUENCE starts at position $i$ and ends at position $j$. The feature MIN computes the minimum of the values from index $i$ to index $j + 1$.

**Example**

$$\left( \begin{array}{l} 1, \langle 4, 4, 6, 4, 1, 1, 3, 4, 4, 6, 6, 5, 2, 2, 4, 3 \rangle, \\ \langle 0, 0, 1, 0, 0, 0, 0, 0, 0, 0, 0, 0, 0, 0, 0, 0 \rangle \end{array} \right)$$

Figure 3.1072 provides an example where the POS_MIN_MIN_STRICTLY_DECREASING_SEQUENCE $(1, [4, 4, 6, 4, 1, 1, 3, 4, 4, 6, 6, 5, 2, 2, 4, 3], [0, 0, 1, 0, 0, 0, 0, 0, 0, 0, 0, 0, 0, 0, 0, 0])$ constraint holds.

**Typical**

$|\mathtt{VARIABLES}| > 1$

range(VARIABLES.var) > 1

Figure 3.1072: Illustrating the POS_MIN_MIN_STRICTLY_DECREASING_SEQUENCE constraint of the **Example** slot

**Arg. properties**
- Functional dependency: VALUE determined by VARIABLES.
- Functional dependency: FOUND determined by VARIABLES.

**Automaton**    Similar to the automaton of the MIN_MIN_STRICTLY_DECREASING_SEQUENCE constraint
but use the decoration table 2.33.

## 3.508   POS_MIN_MIN_STRICTLY_INCREASING_SEQUENCE

▶ ▷ ◁ ◀        **DESCRIPTION**            **AUTOMATON**

**Origin**          Based on constraint MIN_MIN_STRICTLY_INCREASING_SEQUENCE.

**Constraint**      POS_MIN_MIN_STRICTLY_INCREASING_SEQUENCE(VALUE, VARIABLES, FOUND)

**Arguments**
```
VALUE       : dvar
VARIABLES   : collection(var−dvar)
FOUND       : collection(var−dvar)
```

**Restrictions**    $\mathtt{sv} \leq 1 \vee \mathtt{rv} \leq 1 \Rightarrow \mathtt{VALUE} = +\infty$

$\mathtt{VALUE} \geq \mathtt{minv}①$

$\mathtt{VALUE} = +\infty \vee \mathtt{VALUE} \leq \mathtt{maxv} - 1$

required(VARIABLES, var)

required(FOUND, var)

where

$\mathtt{minv} = \mathtt{minval}(\mathtt{VARIABLES.var})$

$\mathtt{maxv} = \mathtt{maxval}(\mathtt{VARIABLES.var})$

$\mathtt{sv} = |\mathtt{VARIABLES}|$

$\mathtt{rv} = \mathtt{range}(\mathtt{VARIABLES.var})$

**Purpose**
The constraint MIN_MIN_STRICTLY_INCREASING_SEQUENCE(VALUE, VARIABLES) holds. In addition, FOUND is a collection of 0/1 variables where the value 1 indicates the position of the **found** letter in those occurrences of the pattern STRICTLY_INCREASING_SEQUENCE for which the feature value is VALUE.

The position of the **found** letter in an occurrence of a pattern is the first position where the occurrence of pattern is identified, even if the pattern is not complete.

An occurrence of the pattern STRICTLY_INCREASING_SEQUENCE is the *maximal* sub-sequence which matches the regular expression $<^+$.

Assume that the occurrence of the pattern STRICTLY_INCREASING_SEQUENCE starts at position $i$ and ends at position $j$. The feature MIN computes the minimum of the values from index $i$ to index $j + 1$.

**Example**
$$\left( \begin{array}{l} 1, \langle 4, 3, 5, 5, 2, 1, 1, 2, 3, 4, 6, 6, 3, 1, 2, 3 \rangle, \\ \langle 0, 0, 0, 0, 0, 0, 1, 0, 0, 0, 0, 0, 0, 0, 1, 0, 0 \rangle \end{array} \right)$$

Figure 3.1073 provides an example where the POS_MIN_MIN_STRICTLY_INCREASING_SEQUENCE $(1, [4, 3, 5, 5, 2, 1, 1, 2, 3, 4, 6, 6, 3, 1, 2, 3], [0, 0, 0, 0, 0, 0, 1, 0, 0, 0, 0, 0, 0, 0, 1, 0, 0])$ constraint holds.

**Typical**
$|\mathtt{VARIABLES}| > 1$

$\mathtt{range}(\mathtt{VARIABLES.var}) > 1$

Figure 3.1073: Illustrating the POS_MIN_MIN_STRICTLY_INCREASING_SEQUENCE constraint of the **Example** slot

**Arg. properties**
- Functional dependency: VALUE determined by VARIABLES.
- Functional dependency: FOUND determined by VARIABLES.

**Automaton**      Similar to the automaton of the MIN_MIN_STRICTLY_INCREASING_SEQUENCE constraint but use the decoration table 2.33.

## 3.509 POS_MIN_MIN_VALLEY

    **DESCRIPTION**          **AUTOMATON**

$> (= | >)^* (< | =)^* <$

| | |
|---|---|
| **Origin** | Based on constraint MIN_MIN_VALLEY. |
| **Constraint** | POS_MIN_MIN_VALLEY(VALUE, VARIABLES, FOUND) |
| **Arguments** | VALUE     : dvar <br> VARIABLES : collection(var−dvar) <br> FOUND     : collection(var−dvar) |
| **Restrictions** | $\mathtt{sv} \leq 2 \vee \mathtt{rv} \leq 1 \Rightarrow \mathtt{VALUE} = +\infty$ <br> $\mathtt{VALUE} \geq \mathtt{minv}①$ <br> $\mathtt{VALUE} = +\infty \vee \mathtt{VALUE} \leq \mathtt{maxv} - 1$ <br> required(VARIABLES, var) <br> required(FOUND, var) <br> where <br> $\mathtt{minv} = \mathtt{minval}(\mathtt{VARIABLES.var})$ <br> $\mathtt{maxv} = \mathtt{maxval}(\mathtt{VARIABLES.var})$ <br> $\mathtt{sv} = |\mathtt{VARIABLES}|$ <br> $\mathtt{rv} = \mathtt{range}(\mathtt{VARIABLES.var})$ |

VALUE ≥ 0

| | |
|---|---|
| **Purpose** | The constraint MIN_MIN_VALLEY(VALUE, VARIABLES) holds. In addition, FOUND is a collection of 0/1 variables where the value 1 indicates the position of the **found** letter in those occurrences of the pattern VALLEY for which the feature value is VALUE. <br><br> The position of the **found** letter in an occurrence of a pattern is the first position where the occurrence of pattern is identified, even if the pattern is not complete. <br><br> An occurrence of the pattern VALLEY is the *maximal* subsequence which matches the regular expression $> (= | >)^* (< | =)^* <$. <br><br> Assume that the occurrence of the pattern VALLEY starts at position $i$ and ends at position $j$. The feature MIN computes the minimum of the values from index $i + 1$ to index $j$. |

| | |
|---|---|
| **Example** | $\left( \begin{array}{l} 2, \langle 1, 3, 7, 4, 3, 6, 6, 5, 3, 3, 2, 6, 5, 5, 5, 7 \rangle, \\ \langle 0, 0, 0, 0, 0, 0, 0, 0, 0, 0, 1, 0, 0, 0, 0, 0 \rangle \end{array} \right)$ |

Figure 3.1074 provides an example where the POS_MIN_MIN_VALLEY $(2, [1, 3, 7, 4, 3, 6, 6, 5, 3, 3, 2, 6, 5, 5, 5, 7], [0, 0, 0, 0, 0, 0, 0, 0, 0, 0, 1, 0, 0, 0, 0, 0])$ constraint holds.

| | |
|---|---|
| **Typical** | $|\mathtt{VARIABLES}| > 2$ <br> range(VARIABLES.var) $> 1$ |

| | |
|---|---|
| **Arg. properties** | • Functional dependency: VALUE determined by VARIABLES. <br> • Functional dependency: FOUND determined by VARIABLES. |

Figure 3.1074: Illustrating the POS_MIN_MIN_VALLEY constraint of the **Example** slot

**Automaton**          Similar to the automaton of the MIN_MIN_VALLEY constraint but use the decoration table 2.33.

## 3.510   POS_MIN_MIN_ZIGZAG

| | |
|---|---|
| **Origin** | Based on constraint MIN_MIN_ZIGZAG. |
| **Constraint** | POS_MIN_MIN_ZIGZAG(VALUE, VARIABLES, FOUND) |

**Arguments**

```
VALUE       : dvar
VARIABLES   : collection(var−dvar)
FOUND       : collection(var−dvar)
```

**Restrictions**

$\text{sv} \leq 3 \vee \text{rv} \leq 1 \Rightarrow \text{VALUE} = +\infty$

$\text{VALUE} \geq \text{minv}①$

$\text{VALUE} = +\infty \vee \text{VALUE} \leq \text{maxv} - 1$

required(VARIABLES, var)

required(FOUND, var)

where

$\text{minv} = \text{minval}(\text{VARIABLES.var})$

$\text{maxv} = \text{maxval}(\text{VARIABLES.var})$

$\text{sv} = |\text{VARIABLES}|$

$\text{rv} = \text{range}(\text{VARIABLES.var})$

**Purpose**

The constraint MIN_MIN_ZIGZAG(VALUE, VARIABLES) holds. In addition, FOUND is a collection of 0/1 variables where the value 1 indicates the position of the **found** letter in those occurrences of the pattern ZIGZAG for which the feature value is VALUE.

The position of the **found** letter in an occurrence of a pattern is the first position where the occurrence of pattern is identified, even if the pattern is not complete.

An occurrence of the pattern ZIGZAG is the *maximal* subsequence which matches the regular expression $(<>)^+(< | <>) | (><)^+(> | ><)$.

Assume that the occurrence of the pattern ZIGZAG starts at position $i$ and ends at position $j$. The feature MIN computes the minimum of the values from index $i + 1$ to index $j$.

**Example**

$$\left( \begin{array}{l} 1, \langle 4, 1, 3, 1, 4, 6, 1, 5, 5, 2, 7, 2, 3, 1, 6, 1 \rangle, \\ \langle 0, 0, 1, 0, 0, 0, 1, 0, 0, 0, 1, 0, 0, 0, 0, 0 \rangle \end{array} \right)$$

Figure 3.1075 provides an example where the POS_MIN_MIN_ZIGZAG $(1, [4, 1, 3, 1, 4, 6, 1, 5, 5, 2, 7, 2, 3, 1, 6, 1], [0, 0, 1, 0, 0, 0, 1, 0, 0, 0, 1, 0, 0, 0, 0, 0])$ constraint holds.

**Typical**

$|\text{VARIABLES}| > 3$

$\text{range}(\text{VARIABLES.var}) > 1$

**Arg. properties**

- Functional dependency: VALUE determined by VARIABLES.
- Functional dependency: FOUND determined by VARIABLES.

Figure 3.1075: Illustrating the POS_MIN_MIN_ZIGZAG constraint of the **Example** slot

**Automaton**  Similar to the automaton of the MIN_MIN_ZIGZAG constraint but use the decoration table 2.33.

# 3.511   POS_MIN_SURF_BUMP_ON_DECREASING_SEQUENCE

▶ ▷ ◁ ◀        **DESCRIPTION**              **AUTOMATON**

| | |
|---|---|
| **Origin** | Based on constraint MIN_SURF_BUMP_ON_DECREASING_SEQUENCE. |
| **Constraint** | POS_MIN_SURF_BUMP_ON_DECREASING_SEQUENCE(VALUE, VARIABLES, FOUND) |

**Arguments**

```
VALUE     : dvar
VARIABLES : collection(var−dvar)
FOUND     : collection(var−dvar)
```

**Restrictions**

$\mathtt{sv} \leq 5 \lor \mathtt{rv} \leq 2 \Rightarrow \mathtt{VALUE} = +\infty$

$\mathtt{VALUE} \geq 3 * \mathtt{minv} + 3 ①$

$\mathtt{VALUE} = +\infty \lor \mathtt{VALUE} \leq 3 * \mathtt{maxv} - 3$

required(VARIABLES, var)

required(FOUND, var)

where

minv = minval(VARIABLES.var)

rv = range(VARIABLES.var)

sv = |VARIABLES|

maxv = maxval(VARIABLES.var)

**Purpose**

The constraint MIN_SURF_BUMP_ON_DECREASING_SEQUENCE(VALUE, VARIABLES) holds. In addition, FOUND is a collection of 0/1 variables where the value 1 indicates the position of the **found** letter in those occurrences of the pattern BUMP_ON_DECREASING_SEQUENCE for which the feature value is VALUE.

The position of the **found** letter in an occurrence of a pattern is the first position where the occurrence of pattern is identified, even if the pattern is not complete.

An occurrence of the pattern BUMP_ON_DECREASING_SEQUENCE is the subsequence which matches the regular expression >><<>.

Assume that the occurrence of the pattern BUMP_ON_DECREASING_SEQUENCE starts at position $i$ and ends at position $j$. The feature SURF computes the sum of the values from index $i + 2$ to index $j$.

**Example**

$$\left( \begin{array}{l} 11, \langle 7, 6, 5, 6, 5, 4, 1, 4, 7, 5, 4, 2, 5, 4, 3, 3 \rangle, \\ \langle 0, 0, 0, 0, 0, 0, 0, 0, 0, 0, 0, 0, 0, 0, 1, 0, 0 \rangle \end{array} \right)$$

Figure 3.1076 provides an example where the POS_MIN_SURF_BUMP_ON_DECREASING_SEQUENCE $(11, [7, 6, 5, 6, 5, 4, 1, 4, 7, 5, 4, 2, 5, 4, 3, 3], [0, 0, 0, 0, 0, 0, 0, 0, 0, 0, 0, 0, 0, 0, 1, 0, 0])$ constraint holds.

**Typical**

|VARIABLES| > 5

range(VARIABLES.var) > 2

Figure 3.1076: Illustrating the POS_MIN_SURF_BUMP_ON_DECREASING_SEQUENCE constraint of the **Example** slot

| Arg. properties | |
|---|---|
| | • Functional dependency: VALUE determined by VARIABLES. |
| | • Functional dependency: FOUND determined by VARIABLES. |

**Automaton**     Similar to the automaton of the MIN_SURF_BUMP_ON_DECREASING_SEQUENCE constraint but use the decoration table 2.33.

## 3.512 POS_MIN_SURF_DECREASING

    DESCRIPTION    AUTOMATON

| | |
|---|---|
| **Origin** | Based on constraint MIN_SURF_DECREASING. |
| **Constraint** | POS_MIN_SURF_DECREASING(VALUE, VARIABLES, FOUND) |
| **Arguments** | VALUE : dvar <br> VARIABLES : collection(var−dvar) <br> FOUND : collection(var−dvar) |

**Restrictions**

$\text{sv} \leq 1 \vee \text{rv} \leq 1 \Rightarrow \text{VALUE} = +\infty$

$\text{VALUE} \geq 2 * \text{minv} + 1$ ①

$\text{VALUE} = +\infty \vee \text{VALUE} \leq 2 * \text{maxv} - 1$

required(VARIABLES, var)

required(FOUND, var)

where

$\text{minv} = \text{minval}(\text{VARIABLES.var})$

$\text{maxv} = \text{maxval}(\text{VARIABLES.var})$

$\text{sv} = |\text{VARIABLES}|$

$\text{rv} = \text{range}(\text{VARIABLES.var})$

VALUE ≥ **1** = 2 * 0 + 1

**Purpose**

The constraint MIN_SURF_DECREASING(VALUE, VARIABLES) holds. In addition, FOUND is a collection of 0/1 variables where the value 1 indicates the position of the **found** letter in those occurrences of the pattern DECREASING for which the feature value is VALUE.

The position of the **found** letter in an occurrence of a pattern is the first position where the occurrence of pattern is identified, even if the pattern is not complete.

An occurrence of the pattern DECREASING is the subsequence which matches the regular expression >.

Assume that the occurrence of the pattern DECREASING starts at position $i$ and ends at position $j$. The feature SURF computes the sum of the values from index $i$ to index $j+1$.

**Example**

$$\left( \begin{array}{l} 4, \langle 3, 4, 2, 2, 5, 6, 6, 4, 4, 3, 1, 1, 4, 6, 4, 4 \rangle, \\ \langle 0, 0, 0, 0, 0, 0, 0, 0, 0, 0, 1, 0, 0, 0, 0, 0 \rangle \end{array} \right)$$

Figure 3.1077 provides an example where the POS_MIN_SURF_DECREASING (4, [3, 4, 2, 2, 5, 6, 6, 4, 4, 3, 1, 1, 4, 6, 4, 4], [0, 0, 0, 0, 0, 0, 0, 0, 0, 0, 1, 0, 0, 0, 0, 0]) constraint holds.

**Typical**

$|\text{VARIABLES}| > 1$

$\text{range}(\text{VARIABLES.var}) > 1$

Figure 3.1077: Illustrating the POS_MIN_SURF_DECREASING constraint of the **Example** slot

| Arg. properties | |
|---|---|
| | • Functional dependency: VALUE determined by VARIABLES. |
| | • Functional dependency: FOUND determined by VARIABLES. |

**Automaton**       Similar to the automaton of the MIN_SURF_DECREASING constraint but use the decoration
                    table 2.33.

# 3.513　POS_MIN_SURF_DECREASING_SEQUENCE

　　**DESCRIPTION**　　　**AUTOMATON**

| | |
|---|---|
| **Origin** | Based on constraint MIN_SURF_DECREASING_SEQUENCE. |
| **Constraint** | POS_MIN_SURF_DECREASING_SEQUENCE(VALUE, VARIABLES, FOUND) |
| **Arguments** | VALUE　　:　dvar<br>VARIABLES　:　collection(var−dvar)<br>FOUND　　:　collection(var−dvar) |

**Restrictions**

$sv \leq 1 \lor rv \leq 1 \Rightarrow \text{VALUE} = +\infty$

$rv = 2 \Rightarrow \text{VALUE} \geq 2 * minv + 1$ ①

$rv \geq 3 \Rightarrow \text{VALUE} \geq \min(2 * minv + 1②, sv * (minv + 1)③)$

$rv = 2 \Rightarrow \text{VALUE} = +\infty \lor \text{VALUE} \leq 2 * maxv - 1$

$rv \geq 3 \Rightarrow \text{VALUE} = +\infty \lor \text{VALUE} \leq \max(2 * maxv - 1, sv * (maxv - 1))$

required(VARIABLES, var)

required(FOUND, var)

where

minv =minval(VARIABLES.var)

rv =range(VARIABLES.var)

sv =|VARIABLES|

maxv =maxval(VARIABLES.var)

**Purpose**

The constraint MIN_SURF_DECREASING_SEQUENCE(VALUE, VARIABLES) holds. In addition, FOUND is a collection of 0/1 variables where the value 1 indicates the position of the **found** letter in those occurrences of the pattern DECREASING_SEQUENCE for which the feature value is VALUE.

The position of the **found** letter in an occurrence of a pattern is the first position where the occurrence of pattern is identified, even if the pattern is not complete.

An occurrence of the pattern DECREASING_SEQUENCE is the *maximal* subsequence which matches the regular expression > (> | =)* > | >.

Assume that the occurrence of the pattern DECREASING_SEQUENCE starts at position $i$ and ends at position $j$. The feature SURF computes the sum of the values from index $i$ to index $j + 1$.

**Example**

$$\left( \begin{array}{l} 6, \langle 3, 4, 2, 2, 5, 6, 6, 4, 4, 3, 1, 1, 4, 6, 4, 4 \rangle, \\ \langle 0, 1, 0, 0, 0, 0, 0, 0, 0, 0, 0, 0, 0, 0, 0, 0 \rangle \end{array} \right)$$

Figure 3.1078 provides an example where the POS_MIN_SURF_DECREASING_SEQUENCE (6, [3, 4, 2, 2, 5, 6, 6, 4, 4, 3, 1, 1, 4, 6, 4, 4], [0, 1, 0, 0, 0, 0, 0, 0, 0, 0, 0, 0, 0, 0, 0, 0]) constraint holds.

Figure 3.1078: Illustrating the POS_MIN_SURF_DECREASING_SEQUENCE constraint of the **Example** slot

| **Typical** | $\mid\texttt{VARIABLES}\mid > 1$ |
| | $\texttt{range}(\texttt{VARIABLES.var}) > 1$ |

| **Arg. properties** | • Functional dependency: VALUE determined by VARIABLES. |
| | • Functional dependency: FOUND determined by VARIABLES. |

**Automaton**          Similar to the automaton of the MIN_SURF_DECREASING_SEQUENCE constraint but use
                       the decoration table 2.33.

## 3.514  POS_MIN_SURF_DECREASING_TERRACE

▶ ▷ ◁ ◀        **DESCRIPTION**          **AUTOMATON**

| | |
|---|---|
| **Origin** | Based on constraint MIN_SURF_DECREASING_TERRACE. |
| **Constraint** | POS_MIN_SURF_DECREASING_TERRACE(VALUE, VARIABLES, FOUND) |

**Arguments**

```
VALUE     : dvar
VARIABLES : collection(var-dvar)
FOUND     : collection(var-dvar)
```

**Restrictions**

$\text{sv} \leq 3 \lor \text{rv} \leq 2 \Rightarrow \text{VALUE} = +\infty$

$\text{VALUE} \geq \min(2 * (\text{minv} + 1)①, (\text{sv} - 2) * (\text{minv} + 1)②)$

$\text{VALUE} = +\infty \lor \text{VALUE} \leq \max(2 * (\text{maxv} - 1), (\text{sv} - 2) * (\text{maxv} - 1))$

required(VARIABLES, var)

required(FOUND, var)

where

$\text{minv} = \text{minval}(\text{VARIABLES.var})$
$\text{sv} = |\text{VARIABLES}|$
$\text{maxv} = \text{maxval}(\text{VARIABLES.var})$
$\text{rv} = \text{range}(\text{VARIABLES.var})$

**Purpose**

The constraint MIN_SURF_DECREASING_TERRACE(VALUE, VARIABLES) holds. In addition, FOUND is a collection of 0/1 variables where the value 1 indicates the position of the **found** letter in those occurrences of the pattern DECREASING_TERRACE for which the feature value is VALUE.

The position of the **found** letter in an occurrence of a pattern is the first position where the occurrence of pattern is identified, even if the pattern is not complete.

An occurrence of the pattern DECREASING_TERRACE is the *maximal* subsequence which matches the regular expression $> =^+ >$.

Assume that the occurrence of the pattern DECREASING_TERRACE starts at position $i$ and ends at position $j$. The feature SURF computes the sum of the values from index $i + 1$ to index $j$.

**Example**

$$\left( \begin{array}{l} 4, \langle 6, 4, 4, 4, 5, 2, 2, 1, 3, 3, 5, 4, 4, 3, 3, 3 \rangle, \\ \langle 0, 0, 0, 0, 0, 0, 1, 0, 0, 0, 0, 0, 0, 0, 0, 0 \rangle \end{array} \right)$$

Figure 3.1079 provides an example where the POS_MIN_SURF_DECREASING_TERRACE $(4, [6, 4, 4, 4, 5, 2, 2, 1, 3, 3, 5, 4, 4, 3, 3, 3], [0, 0, 0, 0, 0, 0, 1, 0, 0, 0, 0, 0, 0, 0, 0, 0])$ constraint holds.

**Typical**

$|\text{VARIABLES}| > 3$

$\text{range}(\text{VARIABLES.var}) > 2$

Figure 3.1079: Illustrating the POS_MIN_SURF_DECREASING_TERRACE constraint of the **Example** slot

**Arg. properties**

- Functional dependency: VALUE determined by VARIABLES.
- Functional dependency: FOUND determined by VARIABLES.

**Automaton**          Similar to the automaton of the MIN_SURF_DECREASING_TERRACE constraint but use the
                       decoration table 2.33.

# 3.515  POS_MIN_SURF_DIP_ON_INCREASING_SEQUENCE

▶ ▷ ◁ ◀     **DESCRIPTION**        **AUTOMATON**

| | |
|---|---|
| **Origin** | Based on constraint MIN_SURF_DIP_ON_INCREASING_SEQUENCE. |
| **Constraint** | POS_MIN_SURF_DIP_ON_INCREASING_SEQUENCE(VALUE, VARIABLES, FOUND) |

**Arguments**

```
VALUE      : dvar
VARIABLES  : collection(var−dvar)
FOUND      : collection(var−dvar)
```

**Restrictions**

$sv \leq 5 \vee rv \leq 2 \Rightarrow VALUE = +\infty$

$VALUE \geq 3 * minv + 3①$

$VALUE = +\infty \vee VALUE \leq 3 * maxv - 3$

required(VARIABLES, var)

required(FOUND, var)

where

$minv = minval(VARIABLES.var)$

$maxv = maxval(VARIABLES.var)$

$sv = |VARIABLES|$

$rv = range(VARIABLES.var)$

**Purpose**

The constraint MIN_SURF_DIP_ON_INCREASING_SEQUENCE(VALUE, VARIABLES) holds. In addition, FOUND is a collection of 0/1 variables where the value 1 indicates the position of the **found** letter in those occurrences of the pattern DIP_ON_INCREASING_SEQUENCE for which the feature value is VALUE.

The position of the **found** letter in an occurrence of a pattern is the first position where the occurrence of pattern is identified, even if the pattern is not complete.

An occurrence of the pattern DIP_ON_INCREASING_SEQUENCE is the subsequence which matches the regular expression `<<><<`.

Assume that the occurrence of the pattern DIP_ON_INCREASING_SEQUENCE starts at position $i$ and ends at position $j$. The feature SURF computes the sum of the values from index $i + 2$ to index $j$.

**Example**

$$\left( \begin{array}{l} 9, \langle 1, 2, 3, 2, 5, 6, 7, 4, 1, 3, 4, 6, 1, 2, 4, 4 \rangle, \\ \langle 0, 0, 0, 0, 0, 0, 0, 0, 0, 0, 0, 0, 0, 0, 1, 0, 0 \rangle \end{array} \right)$$

Figure 3.1080 provides an example where the POS_MIN_SURF_DIP_ON_INCREASING_SEQUENCE $(9, [1, 2, 3, 2, 5, 6, 7, 4, 1, 3, 4, 6, 1, 2, 4, 4], [0, 0, 0, 0, 0, 0, 0, 0, 0, 0, 0, 0, 0, 0, 1, 0, 0])$ constraint holds.

**Typical**

$|VARIABLES| > 5$

$range(VARIABLES.var) > 2$

Figure 3.1080: Illustrating the POS_MIN_SURF_DIP_ON_INCREASING_SEQUENCE constraint of the **Example** slot

**Arg. properties**
- Functional dependency: VALUE determined by VARIABLES.
- Functional dependency: FOUND determined by VARIABLES.

**Automaton**    Similar to the automaton of the MIN_SURF_DIP_ON_INCREASING_SEQUENCE constraint but use the decoration table 2.33.

# 3.516 POS_MIN_SURF_GORGE

► ▷ ◁ ◀        DESCRIPTION            AUTOMATON

$(> | > (= | >)^*)(< | < (= | <)^* <)$

| | |
|---|---|
| **Origin** | Based on constraint MIN_SURF_GORGE. |
| **Constraint** | POS_MIN_SURF_GORGE(VALUE, VARIABLES, FOUND) |
| **Arguments** | VALUE    : dvar<br>VARIABLES : collection(var−dvar)<br>FOUND    : collection(var−dvar) |

**Restrictions**

$$sv \leq 2 \vee rv \leq 1 \Rightarrow \text{VALUE} = +\infty$$
$$rv = 2 \Rightarrow \text{VALUE} \geq minv①$$
$$rv \geq 3 \Rightarrow \text{VALUE} \geq \min(minv②, (sv - 2) * (minv + 1) - 1③)$$
$$rv = 2 \Rightarrow \text{VALUE} = +\infty \vee \text{VALUE} \leq maxv - 1$$
$$rv \geq 3 \Rightarrow$$
$$\text{VALUE} = +\infty \vee \text{VALUE} \leq \max(maxv - 1, (sv - 2) * (maxv - 1) - 1)$$

required(VARIABLES, var)
required(FOUND, var)

where

minv = minval(VARIABLES.var)
rv = range(VARIABLES.var)
sv = |VARIABLES|
maxv = maxval(VARIABLES.var)

**Purpose**

The constraint MIN_SURF_GORGE(VALUE, VARIABLES) holds. In addition, FOUND is a collection of 0/1 variables where the value 1 indicates the position of the **found** letter in those occurrences of the pattern GORGE for which the feature value is VALUE.

The position of the **found** letter in an occurrence of a pattern is the first position where the occurrence of pattern is identified, even if the pattern is not complete.

An occurrence of the pattern GORGE is the *maximal* subsequence which matches the regular expression $(> | > (= | >)^* >)(< | < (= | <)^* <)$.

Assume that the occurrence of the pattern GORGE starts at position $i$ and ends at position $j$. The feature SURF computes the sum of the values from index $i + 1$ to index $j$.

**Example**

$$\left( \begin{array}{l} 5, \langle 1, 7, 3, 4, 4, 5, 5, 4, 2, 2, 6, 5, 4, 6, 5, 7 \rangle , \\ \langle 0, 0, 0, 0, 0, 0, 0, 0, 0, 0, 0, 0, 0, 0, 1, 0 \rangle \end{array} \right)$$

Figure 3.1081 provides an example where the POS_MIN_SURF_GORGE $(5, [1, 7, 3, 4, 4, 5, 5, 4, 2, 2, 6, 5, 4, 6, 5, 7], [0, 0, 0, 0, 0, 0, 0, 0, 0, 0, 0, 0, 0, 0, 1, 0])$ constraint holds.

**Typical**

|VARIABLES| > 2
range(VARIABLES.var) > 1

Figure 3.1081: Illustrating the POS_MIN_SURF_GORGE constraint of the **Example** slot

**Arg. properties**
- Functional dependency: VALUE determined by VARIABLES.
- Functional dependency: FOUND determined by VARIABLES.

**Automaton**      Similar to the automaton of the MIN_SURF_GORGE constraint but use the decoration table 2.33.

## 3.517   POS_MIN_SURF_INCREASING

▶ ▷ ◁ ◀         **DESCRIPTION**                    **AUTOMATON**

**Origin**          Based on constraint MIN_SURF_INCREASING.

**Constraint**      POS_MIN_SURF_INCREASING(VALUE, VARIABLES, FOUND)

**Arguments**       VALUE     :  dvar
                    VARIABLES :  collection(var−dvar)
                    FOUND     :  collection(var−dvar)

**Restrictions**    $\mathtt{sv} \leq 1 \lor \mathtt{rv} \leq 1 \Rightarrow \mathtt{VALUE} = +\infty$
                    $\mathtt{VALUE} \geq 2 * \mathtt{minv} + 1$①
                    $\mathtt{VALUE} = +\infty \lor \mathtt{VALUE} \leq 2 * \mathtt{maxv} - 1$
                    required(VARIABLES, var)
                    required(FOUND, var)
                    where
                    $\mathtt{minv} =$ minval(VARIABLES.var)
                    $\mathtt{maxv} =$ maxval(VARIABLES.var)
                    $\mathtt{sv} = |\mathtt{VARIABLES}|$
                    $\mathtt{rv} =$ range(VARIABLES.var)

VALUE $\geq$ **1** $= 2 * 0 + 1$

**Purpose**         The constraint MIN_SURF_INCREASING(VALUE, VARIABLES) holds. In addition, FOUND is a collection of 0/1 variables where the value 1 indicates the position of the **found** letter in those occurrences of the pattern INCREASING for which the feature value is VALUE.

                    The position of the **found** letter in an occurrence of a pattern is the first position where the occurrence of pattern is identified, even if the pattern is not complete.

                    An occurrence of the pattern INCREASING is the subsequence which matches the regular expression <.

                    Assume that the occurrence of the pattern INCREASING starts at position $i$ and ends at position $j$. The feature SURF computes the sum of the values from index $i$ to index $j+1$.

**Example**         $\left( \begin{array}{l} 4, \langle 4, 3, 5, 5, 2, 1, 1, 3, 3, 4, 6, 6, 3, 1, 3, 3 \rangle, \\ \langle 0, 0, 0, 0, 0, 0, 1, 0, 0, 0, 0, 0, 0, 1, 0, 0 \rangle \end{array} \right)$

                    Figure 3.1082 provides an example where the POS_MIN_SURF_INCREASING $(4, [4, 3, 5, 5, 2, 1, 1, 3, 3, 4, 6, 6, 3, 1, 3, 3], [0, 0, 0, 0, 0, 0, 1, 0, 0, 0, 0, 0, 0, 1, 0, 0])$ constraint holds.

**Typical**         $|\mathtt{VARIABLES}| > 1$
                    range(VARIABLES.var) $> 1$

Figure 3.1082: Illustrating the POS_MIN_SURF_INCREASING constraint of the **Example** slot

| Arg. properties | • Functional dependency: VALUE determined by VARIABLES. |
|---|---|
| | • Functional dependency: FOUND determined by VARIABLES. |

**Automaton**  Similar to the automaton of the MIN_SURF_INCREASING constraint but use the decoration table 2.33.

# 3.518 POS_MIN_SURF_INCREASING_SEQUENCE

▶ ▷ ◁ ◀        **DESCRIPTION**        **AUTOMATON**

$< (< | =)^* < |$
$<$

| | |
|---|---|
| **Origin** | Based on constraint MIN_SURF_INCREASING_SEQUENCE. |
| **Constraint** | POS_MIN_SURF_INCREASING_SEQUENCE(VALUE, VARIABLES, FOUND) |
| **Arguments** | VALUE : dvar |
| | VARIABLES : collection(var−dvar) |
| | FOUND : collection(var−dvar) |

**Restrictions**

$\mathtt{sv} \leq 1 \lor \mathtt{rv} \leq 1 \Rightarrow \mathtt{VALUE} = +\infty$

$\mathtt{rv} = 2 \Rightarrow \mathtt{VALUE} \geq 2 * \mathtt{minv} + 1$ ①

$\mathtt{rv} \geq 3 \Rightarrow \mathtt{VALUE} \geq \min(2 * \mathtt{minv} + 1$②$, \mathtt{sv} * (\mathtt{minv} + 1)$③$)$

$\mathtt{rv} = 2 \Rightarrow \mathtt{VALUE} = +\infty \lor \mathtt{VALUE} \leq 2 * \mathtt{maxv} - 1$

$\mathtt{rv} \geq 3 \Rightarrow \mathtt{VALUE} = +\infty \lor \mathtt{VALUE} \leq \max(2 * \mathtt{maxv} - 1, \mathtt{sv} * (\mathtt{maxv} - 1))$

required(VARIABLES, var)

required(FOUND, var)

where

minv =minval(VARIABLES.var)

rv =range(VARIABLES.var)

sv = |VARIABLES|

maxv =maxval(VARIABLES.var)

**Purpose**

The constraint MIN_SURF_INCREASING_SEQUENCE(VALUE, VARIABLES) holds. In addition, FOUND is a collection of 0/1 variables where the value 1 indicates the position of the **found** letter in those occurrences of the pattern INCREASING_SEQUENCE for which the feature value is VALUE.

The position of the **found** letter in an occurrence of a pattern is the first position where the occurrence of pattern is identified, even if the pattern is not complete.

An occurrence of the pattern INCREASING_SEQUENCE is the *maximal* subsequence which matches the regular expression $< (< | =)^* < | <$.

Assume that the occurrence of the pattern INCREASING_SEQUENCE starts at position $i$ and ends at position $j$. The feature SURF computes the sum of the values from index $i$ to index $j + 1$.

**Example**

$$\left( \begin{array}{l} 4, \langle 4, 3, 5, 5, 2, 1, 1, 3, 3, 4, 6, 6, 3, 1, 3, 3 \rangle, \\ \langle 0, 0, 0, 0, 0, 0, 0, 0, 0, 0, 0, 0, 0, 0, 1, 0, 0 \rangle \end{array} \right)$$

Figure 3.1083 provides an example where the POS_MIN_SURF_INCREASING_SEQUENCE $(4, [4, 3, 5, 5, 2, 1, 1, 3, 3, 4, 6, 6, 3, 1, 3, 3], [0, 0, 0, 0, 0, 0, 0, 0, 0, 0, 0, 0, 0, 0, 1, 0, 0])$ constraint holds.

Figure 3.1083: Illustrating the POS_MIN_SURF_INCREASING_SEQUENCE constraint of the **Example** slot

| **Typical** | $|\text{VARIABLES}| > 1$ |
| | $\text{range}(\text{VARIABLES.var}) > 1$ |

**Arg. properties**
- Functional dependency: VALUE determined by VARIABLES.
- Functional dependency: FOUND determined by VARIABLES.

**Automaton**          Similar to the automaton of the MIN_SURF_INCREASING_SEQUENCE constraint but use the
                       decoration table 2.33.

# 3.519   POS_MIN_SURF_INCREASING_TERRACE

        **DESCRIPTION**              **AUTOMATON**

| | |
|---|---|
| **Origin** | Based on constraint MIN_SURF_INCREASING_TERRACE. |
| **Constraint** | POS_MIN_SURF_INCREASING_TERRACE(VALUE, VARIABLES, FOUND) |
| **Arguments** | VALUE : dvar<br>VARIABLES : collection(var−dvar)<br>FOUND : collection(var−dvar) |

**Restrictions**

$sv \leq 3 \lor rv \leq 2 \Rightarrow VALUE = +\infty$

$VALUE \geq \min(2 * (minv + 1)①, (sv − 2) * (minv + 1)②)$

$VALUE = +\infty \lor VALUE \leq \max(2 * (maxv − 1), (sv − 2) * (maxv − 1))$

required(VARIABLES, var)

required(FOUND, var)

where

minv = minval(VARIABLES.var)
sv = |VARIABLES|
maxv = maxval(VARIABLES.var)
rv = range(VARIABLES.var)

**Purpose**

The constraint MIN_SURF_INCREASING_TERRACE(VALUE, VARIABLES) holds. In addition, FOUND is a collection of 0/1 variables where the value 1 indicates the position of the **found** letter in those occurrences of the pattern INCREASING_TERRACE for which the feature value is VALUE.

The position of the **found** letter in an occurrence of a pattern is the first position where the occurrence of pattern is identified, even if the pattern is not complete.

An occurrence of the pattern INCREASING_TERRACE is the *maximal* subsequence which matches the regular expression $<=^+<$.

Assume that the occurrence of the pattern INCREASING_TERRACE starts at position $i$ and ends at position $j$. The feature SURF computes the sum of the values from index $i + 1$ to index $j$.

**Example**

$$\left( \begin{array}{l} 9, \langle 1, 3, 3, 3, 2, 5, 5, 6, 4, 4, 2, 3, 3, 3, 4, 4 \rangle , \\ \langle 0, 0, 0, 0, 0, 0, 0, 0, 0, 0, 0, 0, 0, 0, 1, 0, 0 \rangle \end{array} \right)$$

Figure 3.1084 provides an example where the POS_MIN_SURF_INCREASING_TERRACE $(9, [1, 3, 3, 3, 2, 5, 5, 6, 4, 4, 2, 3, 3, 3, 4, 4], [0, 0, 0, 0, 0, 0, 0, 0, 0, 0, 0, 0, 0, 0, 1, 0, 0])$ constraint holds.

**Typical**

|VARIABLES| > 3

range(VARIABLES.var) > 2

Figure 3.1084: Illustrating the `POS_MIN_SURF_INCREASING_TERRACE` constraint of the **Example** slot

**Arg. properties**

- Functional dependency: `VALUE` determined by `VARIABLES`.
- Functional dependency: `FOUND` determined by `VARIABLES`.

**Automaton**      Similar to the automaton of the MIN_SURF_INCREASING_TERRACE constraint but use the decoration table 2.33.

## 3.520 POS_MIN_SURF_INFLEXION

        **DESCRIPTION**          **AUTOMATON**

$< (< | =)^* > | > (> | =)^* <$

| | |
|---|---|
| **Origin** | Based on constraint MIN_SURF_INFLEXION. |
| **Constraint** | POS_MIN_SURF_INFLEXION(VALUE, VARIABLES, FOUND) |
| **Arguments** | VALUE    : dvar |
| | VARIABLES : collection(var−dvar) |
| | FOUND    : collection(var−dvar) |

**Restrictions**

$\text{sv} \leq 2 \vee \text{rv} \leq 1 \Rightarrow \text{VALUE} = +\infty$

$\text{VALUE} \geq \min(\text{minv}①, (\text{sv} - 2) * \text{minv}②)$

$\text{VALUE} = +\infty \vee \text{VALUE} \leq \max(\text{maxv}, (\text{sv} - 2) * \text{maxv})$

required(VARIABLES, var)

required(FOUND, var)

where

minv = minval(VARIABLES.var)

sv = |VARIABLES|

maxv = maxval(VARIABLES.var)

rv = range(VARIABLES.var)

**Purpose**

The constraint MIN_SURF_INFLEXION(VALUE, VARIABLES) holds. In addition, FOUND is a collection of 0/1 variables where the value 1 indicates the position of the **found** letter in those occurrences of the pattern INFLEXION for which the feature value is VALUE.

The position of the **found** letter in an occurrence of a pattern is the first position where the occurrence of pattern is identified, even if the pattern is not complete.

An occurrence of the pattern INFLEXION is the *maximal* subsequence which matches the regular expression $< (< | =)^* > | > (> | =)^* <$.

Assume that the occurrence of the pattern INFLEXION starts at position $i$ and ends at position $j$. The feature SURF computes the sum of the values from index $i + 1$ to index $j$.

**Example**

$$\left( \begin{array}{l} 1, \langle 1, 2, 6, 6, 4, 4, 3, 5, 2, 5, 1, 5, 3, 3, 4, 4 \rangle, \\ \langle 0, 0, 0, 0, 0, 0, 0, 0, 0, 0, 1, 0, 0, 0, 0, 0 \rangle \end{array} \right)$$

Figure 3.1085 provides an example where the POS_MIN_SURF_INFLEXION $(1, [1, 2, 6, 6, 4, 4, 3, 5, 2, 5, 1, 5, 3, 3, 4, 4], [0, 0, 0, 0, 0, 0, 0, 0, 0, 0, 1, 0, 0, 0, 0, 0])$ constraint holds.

**Typical**

|VARIABLES| > 2

range(VARIABLES.var) > 1

Figure 3.1085: Illustrating the POS_MIN_SURF_INFLEXION constraint of the **Example** slot

| Arg. properties | • Functional dependency: VALUE determined by VARIABLES. |
|---|---|
|  | • Functional dependency: FOUND determined by VARIABLES. |

**Automaton**          Similar to the automaton of the MIN_SURF_INFLEXION constraint but use the decoration
                       table 2.33.

### 3.521 POS_MIN_SURF_PEAK

　　　　**DESCRIPTION**　　　　**AUTOMATON**

$< (= | <)^* (> | =)^* >$

| **Origin** | Based on constraint MIN_SURF_PEAK. |
|---|---|

| **Constraint** | POS_MIN_SURF_PEAK(VALUE, VARIABLES, FOUND) |
|---|---|

**Arguments**

```
VALUE     : dvar
VARIABLES : collection(var−dvar)
FOUND     : collection(var−dvar)
```

**Restrictions**

$\text{sv} \leq 2 \vee \text{rv} \leq 1 \Rightarrow \text{VALUE} = +\infty$

$\text{VALUE} \geq \min(\text{minv} + 1①, (\text{sv} - 2) * (\text{minv} + 1)②)$

$\text{VALUE} = +\infty \vee \text{VALUE} \leq \max(\text{maxv}, (\text{sv} - 2) * \text{maxv})$

required(VARIABLES, var)

required(FOUND, var)

where

$\text{minv} = \text{minval}(\text{VARIABLES.var})$

$\text{sv} = |\text{VARIABLES}|$

$\text{maxv} = \text{maxval}(\text{VARIABLES.var})$

$\text{rv} = \text{range}(\text{VARIABLES.var})$

**Purpose**

The constraint MIN_SURF_PEAK(VALUE, VARIABLES) holds. In addition, FOUND is a collection of 0/1 variables where the value 1 indicates the position of the **found** letter in those occurrences of the pattern PEAK for which the feature value is VALUE.

The position of the **found** letter in an occurrence of a pattern is the first position where the occurrence of pattern is identified, even if the pattern is not complete.

An occurrence of the pattern PEAK is the *maximal* subsequence which matches the regular expression $< (= | <)^* (> | =)^* >$.

Assume that the occurrence of the pattern PEAK starts at position $i$ and ends at position $j$. The feature SURF computes the sum of the values from index $i + 1$ to index $j$.

**Example**

$$\left( \begin{array}{l} 9, \langle 7, 5, 5, 1, 4, 5, 2, 2, 3, 5, 6, 2, 3, 3, 3, 1 \rangle, \\ \langle 0, 0, 0, 0, 0, 1, 0, 0, 0, 0, 0, 0, 0, 0, 1, 0 \rangle \end{array} \right)$$

Figure 3.1086 provides an example where the POS_MIN_SURF_PEAK $(9, [7, 5, 5, 1, 4, 5, 2, 2, 3, 5, 6, 2, 3, 3, 3, 1], [0, 0, 0, 0, 0, 1, 0, 0, 0, 0, 0, 0, 0, 0, 1, 0])$ constraint holds.

**Typical**

$|\text{VARIABLES}| > 2$

$\text{range}(\text{VARIABLES.var}) > 1$

**Arg. properties**

- Functional dependency: VALUE determined by VARIABLES.
- Functional dependency: FOUND determined by VARIABLES.

Figure 3.1086: Illustrating the POS_MIN_SURF_PEAK constraint of the **Example** slot

**Automaton**     Similar to the automaton of the MIN_SURF_PEAK constraint but use the decoration table 2.33.

## 3.522 POS_MIN_SURF_PLAIN

▶ ▷ ◁ ◀    DESCRIPTION    AUTOMATON

| | |
|---|---|
| **Origin** | Based on constraint MIN_SURF_PLAIN. |
| **Constraint** | POS_MIN_SURF_PLAIN(VALUE, VARIABLES, FOUND) |
| **Arguments** | VALUE : dvar |
| | VARIABLES : collection(var−dvar) |
| | FOUND : collection(var−dvar) |
| **Restrictions** | $sv \leq 2 \vee rv \leq 1 \Rightarrow VALUE = +\infty$ |
| | $VALUE \geq \min(minv①, (sv-2) * minv②)$ |
| | $VALUE = +\infty \vee VALUE \leq \max(maxv-1, (sv-2) * (maxv-1))$ |
| | required(VARIABLES, var) |
| | required(FOUND, var) |
| | where |
| | $minv = minval(VARIABLES.var)$ |
| | $sv = |VARIABLES|$ |
| | $maxv = maxval(VARIABLES.var)$ |
| | $rv = range(VARIABLES.var)$ |
| **Purpose** | The constraint MIN_SURF_PLAIN(VALUE, VARIABLES) holds. In addition, FOUND is a collection of 0/1 variables where the value 1 indicates the position of the **found** letter in those occurrences of the pattern PLAIN for which the feature value is VALUE. |
| | The position of the **found** letter in an occurrence of a pattern is the first position where the occurrence of pattern is identified, even if the pattern is not complete. |
| | An occurrence of the pattern PLAIN is the *maximal* subsequence which matches the regular expression $> =^* <$. |
| | Assume that the occurrence of the pattern PLAIN starts at position $i$ and ends at position $j$. The feature SURF computes the sum of the values from index $i + 1$ to index $j$. |
| **Example** | $\left( \begin{array}{l} 4, \langle 2, 3, 6, 5, 7, 6, 6, 4, 5, 5, 4, 3, 3, 6, 6, 3 \rangle, \\ \langle 0, 0, 0, 0, 0, 0, 0, 1, 0, 0, 0, 0, 0, 0, 0, 0 \rangle \end{array} \right)$ |

Figure 3.1087 provides an example where the POS_MIN_SURF_PLAIN $(4, [2, 3, 6, 5, 7, 6, 6, 4, 5, 5, 4, 3, 3, 6, 6, 3], [0, 0, 0, 0, 0, 0, 0, 1, 0, 0, 0, 0, 0, 0, 0, 0])$ constraint holds.

| | |
|---|---|
| **Typical** | $|VARIABLES| > 2$ |
| | range(VARIABLES.var) $> 1$ |
| **Arg. properties** | • Functional dependency: VALUE determined by VARIABLES. |
| | • Functional dependency: FOUND determined by VARIABLES. |

Figure 3.1087: Illustrating the POS_MIN_SURF_PLAIN constraint of the **Example** slot

**Automaton**          Similar to the automaton of the MIN_SURF_PLAIN constraint but use the decoration table 2.33.

# 3.523   POS_MIN_SURF_PLATEAU

▶ ▷ ◁ ◀        **DESCRIPTION**        **AUTOMATON**

| | |
|---|---|
| **Origin** | Based on constraint MIN_SURF_PLATEAU. |
| **Constraint** | POS_MIN_SURF_PLATEAU(VALUE, VARIABLES, FOUND) |
| **Arguments** | VALUE     : dvar |
| | VARIABLES : collection(var−dvar) |
| | FOUND     : collection(var−dvar) |
| **Restrictions** | $\text{sv} \leq 2 \lor \text{rv} \leq 1 \Rightarrow \text{VALUE} = +\infty$ |
| | $\text{VALUE} \geq \min(\text{minv} + 1 ①, (\text{sv} - 2) * (\text{minv} + 1) ②)$ |
| | $\text{VALUE} = +\infty \lor \text{VALUE} \leq \max(\text{maxv}, (\text{sv} - 2) * \text{maxv})$ |
| | required(VARIABLES, var) |
| | required(FOUND, var) |
| | where |
| | $\text{minv} = $ minval(VARIABLES.var) |
| | $\text{sv} = |\text{VARIABLES}|$ |
| | $\text{maxv} = $ maxval(VARIABLES.var) |
| | $\text{rv} = $ range(VARIABLES.var) |
| **Purpose** | The constraint MIN_SURF_PLATEAU(VALUE, VARIABLES) holds. In addition, FOUND is a collection of 0/1 variables where the value 1 indicates the position of the **found** letter in those occurrences of the pattern PLATEAU for which the feature value is VALUE. |
| | The position of the **found** letter in an occurrence of a pattern is the first position where the occurrence of pattern is identified, even if the pattern is not complete. |
| | An occurrence of the pattern PLATEAU is the *maximal* subsequence which matches the regular expression $< = ^* >$. |
| | Assume that the occurrence of the pattern PLATEAU starts at position $i$ and ends at position $j$. The feature SURF computes the sum of the values from index $i + 1$ to index $j$. |
| **Example** | $\left( \begin{array}{l} 3, \langle 7, 5, 2, 3, 1, 2, 2, 4, 3, 3, 4, 5, 5, 2, 2, 5 \rangle, \\ \langle 0, 0, 0, 1, 0, 0, 0, 0, 0, 0, 0, 0, 0, 0, 0, 0 \rangle \end{array} \right)$ |

Figure 3.1088 provides an example where the POS_MIN_SURF_PLATEAU (3, [7, 5, 2, 3, 1, 2, 2, 4, 3, 3, 4, 5, 5, 2, 2, 5], [0, 0, 0, 1, 0, 0, 0, 0, 0, 0, 0, 0, 0, 0, 0, 0]) constraint holds.

| | |
|---|---|
| **Typical** | $|\text{VARIABLES}| > 2$ |
| | range(VARIABLES.var) $> 1$ |

Figure 3.1088: Illustrating the POS_MIN_SURF_PLATEAU constraint of the **Example** slot

**Arg. properties**
- Functional dependency: VALUE determined by VARIABLES.
- Functional dependency: FOUND determined by VARIABLES.

**Automaton**          Similar to the automaton of the MIN_SURF_PLATEAU constraint but use the decoration table 2.33.

## 3.524  POS_MIN_SURF_PROPER_PLAIN

▶ ▷ ◁ ◀        DESCRIPTION          AUTOMATON

| | |
|---|---|
| **Origin** | Based on constraint MIN_SURF_PROPER_PLAIN. |
| **Constraint** | POS_MIN_SURF_PROPER_PLAIN(VALUE, VARIABLES, FOUND) |

**Arguments**

```
VALUE     : dvar
VARIABLES : collection(var−dvar)
FOUND     : collection(var−dvar)
```

**Restrictions**

$$sv \leq 3 \vee rv \leq 1 \Rightarrow \text{VALUE} = +\infty$$
$$\text{VALUE} \geq \min(2 * \text{minv}①, (sv - 2) * \text{minv}②)$$
$$\text{VALUE} = +\infty \vee \text{VALUE} \leq \max(2 * (\text{maxv} - 1), (sv - 2) * (\text{maxv} - 1))$$
required(VARIABLES, var)
required(FOUND, var)
where
$$\text{minv} = \text{minval}(\text{VARIABLES.var})$$
$$sv = |\text{VARIABLES}|$$
$$\text{maxv} = \text{maxval}(\text{VARIABLES.var})$$
$$rv = \text{range}(\text{VARIABLES.var})$$

**Purpose**

The constraint MIN_SURF_PROPER_PLAIN(VALUE, VARIABLES) holds. In addition, FOUND is a collection of 0/1 variables where the value 1 indicates the position of the **found** letter in those occurrences of the pattern PROPER_PLAIN for which the feature value is VALUE.

The position of the **found** letter in an occurrence of a pattern is the first position where the occurrence of pattern is identified, even if the pattern is not complete.

An occurrence of the pattern PROPER_PLAIN is the *maximal* subsequence which matches the regular expression $> =^+ <$.

Assume that the occurrence of the pattern PROPER_PLAIN starts at position $i$ and ends at position $j$. The feature SURF computes the sum of the values from index $i + 1$ to index $j$.

**Example**

$$\left( \begin{array}{l} 8, \langle 2, 7, 5, 5, 6, 3, 7, 4, 4, 5, 6, 5, 3, 3, 3, 5 \rangle, \\ \langle 0, 0, 0, 0, 0, 0, 0, 0, 1, 0, 0, 0, 0, 0, 0, 0 \rangle \end{array} \right)$$

Figure 3.1089 provides an example where the POS_MIN_SURF_PROPER_PLAIN (8, [2, 7, 5, 5, 6, 3, 7, 4, 4, 5, 6, 5, 3, 3, 3, 5], [0, 0, 0, 0, 0, 0, 0, 0, 1, 0, 0, 0, 0, 0, 0, 0]) constraint holds.

**Typical**

$$|\text{VARIABLES}| > 3$$
$$\text{range}(\text{VARIABLES.var}) > 1$$

Figure 3.1089: Illustrating the `POS_MIN_SURF_PROPER_PLAIN` constraint of the **Example** slot

**Arg. properties**
- Functional dependency: `VALUE` determined by `VARIABLES`.
- Functional dependency: `FOUND` determined by `VARIABLES`.

**Automaton**        Similar to the automaton of the MIN_SURF_PROPER_PLAIN constraint but use the decoration table 2.33.

## 3.525 POS_MIN_SURF_PROPER_PLATEAU

▶ ▷ ◁ ◀        DESCRIPTION        AUTOMATON

| | |
|---|---|
| **Origin** | Based on constraint MIN_SURF_PROPER_PLATEAU. |
| **Constraint** | POS_MIN_SURF_PROPER_PLATEAU(VALUE, VARIABLES, FOUND) |

**Arguments**

```
VALUE      : dvar
VARIABLES  : collection(var−dvar)
FOUND      : collection(var−dvar)
```

**Restrictions**

$sv \leq 3 \vee rv \leq 1 \Rightarrow VALUE = +\infty$

$VALUE \geq \min(2 * (minv + 1)①, (sv - 2) * (minv + 1)②)$

$VALUE = +\infty \vee VALUE \leq \max(2 * maxv, (sv - 2) * maxv)$

required(VARIABLES, var)

required(FOUND, var)

where

$minv = minval(VARIABLES.var)$

$sv = |VARIABLES|$

$maxv = maxval(VARIABLES.var)$

$rv = range(VARIABLES.var)$

**Purpose**

The constraint MIN_SURF_PROPER_PLATEAU(VALUE, VARIABLES) holds. In addition, FOUND is a collection of 0/1 variables where the value 1 indicates the position of the **found** letter in those occurrences of the pattern PROPER_PLATEAU for which the feature value is VALUE.

The position of the **found** letter in an occurrence of a pattern is the first position where the occurrence of pattern is identified, even if the pattern is not complete.

An occurrence of the pattern PROPER_PLATEAU is the *maximal* subsequence which matches the regular expression $<=^+>$.

Assume that the occurrence of the pattern PROPER_PLATEAU starts at position $i$ and ends at position $j$. The feature SURF computes the sum of the values from index $i+1$ to index $j$.

**Example**

$$\left( \begin{array}{l} 6, \langle 7, 1, 3, 3, 2, 5, 1, 4, 4, 3, 2, 3, 5, 5, 5, 3 \rangle, \\ \langle 0, 0, 0, 1, 0, 0, 0, 0, 0, 0, 0, 0, 0, 0, 0, 0 \rangle \end{array} \right)$$

Figure 3.1090 provides an example where the POS_MIN_SURF_PROPER_PLATEAU $(6, [7, 1, 3, 3, 2, 5, 1, 4, 4, 3, 2, 3, 5, 5, 5, 3], [0, 0, 0, 1, 0, 0, 0, 0, 0, 0, 0, 0, 0, 0, 0, 0])$ constraint holds.

**Typical**

$|VARIABLES| > 3$

$range(VARIABLES.var) > 1$

Figure 3.1090: Illustrating the POS_MIN_SURF_PROPER_PLATEAU constraint of the **Example** slot

**Arg. properties**
- *Functional dependency*: VALUE determined by VARIABLES.
- *Functional dependency*: FOUND determined by VARIABLES.

**Automaton**     Similar to the automaton of the MIN_SURF_PROPER_PLATEAU constraint but use the decoration table 2.33.

## 3.526   POS_MIN_SURF_STEADY

          **DESCRIPTION**          **AUTOMATON**

| | |
|---|---|
| **Origin** | Based on constraint MIN_SURF_STEADY. |
| **Constraint** | POS_MIN_SURF_STEADY(VALUE, VARIABLES, FOUND) |
| **Arguments** | VALUE     : dvar |
| | VARIABLES : collection(var−dvar) |
| | FOUND     : collection(var−dvar) |
| **Restrictions** | $\text{sv} \leq 1 \Rightarrow \text{VALUE} = +\infty$ |
| | $\text{VALUE} \geq 2 * \text{minv}①$ |
| | $\text{VALUE} = +\infty \vee \text{VALUE} \leq 2 * \text{maxv}$ |
| | required(VARIABLES, var) |
| | required(FOUND, var) |
| | where |
| | minv =minval(VARIABLES.var) |
| | maxv =maxval(VARIABLES.var) |
| | $\text{sv} = |\text{VARIABLES}|$ |

VALUE $\geq$ **2** $= 2 * 1$

| | |
|---|---|
| **Purpose** | The constraint MIN_SURF_STEADY(VALUE, VARIABLES) holds. In addition, FOUND is a collection of 0/1 variables where the value 1 indicates the position of the **found** letter in those occurrences of the pattern STEADY for which the feature value is VALUE. |
| | The position of the **found** letter in an occurrence of a pattern is the first position where the occurrence of pattern is identified, even if the pattern is not complete. |
| | An occurrence of the pattern STEADY is the subsequence which matches the regular expression =. |
| | Assume that the occurrence of the pattern STEADY starts at position $i$ and ends at position $j$. The feature SURF computes the sum of the values from index $i$ to index $j + 1$. |
| **Example** | $\left( \begin{array}{l} 2, \langle 1, 1, 7, 3, 3, 5, 5, 5, 6, 5, 5, 5, 7, 2, 6, 6 \rangle, \\ \langle 1, 0, 0, 0, 0, 0, 0, 0, 0, 0, 0, 0, 0, 0, 0, 0 \rangle \end{array} \right)$ |

Figure 3.1091 provides an example where the POS_MIN_SURF_STEADY $(2, [1, 1, 7, 3, 3, 5, 5, 5, 6, 5, 5, 5, 7, 2, 6, 6], [1, 0, 0, 0, 0, 0, 0, 0, 0, 0, 0, 0, 0, 0, 0, 0])$ constraint holds.

| | |
|---|---|
| **Typical** | $|\text{VARIABLES}| > 1$ |

| | |
|---|---|
| **Arg. properties** | • Functional dependency: VALUE determined by VARIABLES. |
| | • Functional dependency: FOUND determined by VARIABLES. |

Figure 3.1091: Illustrating the POS_MIN_SURF_STEADY constraint of the **Example** slot

**Automaton**          Similar to the automaton of the MIN_SURF_STEADY constraint but use the decoration table 2.33.

## 3.527   POS_MIN_SURF_STEADY_SEQUENCE

| | |
|---|---|
| **Origin** | Based on constraint MIN_SURF_STEADY_SEQUENCE. |
| **Constraint** | POS_MIN_SURF_STEADY_SEQUENCE(VALUE, VARIABLES, FOUND) |

**Arguments**

VALUE     : dvar
VARIABLES : collection(var−dvar)
FOUND     : collection(var−dvar)

**Restrictions**

$\mathtt{sv} \leq 1 \Rightarrow \mathtt{VALUE} = +\infty$
$\mathtt{rv} = 1 \Rightarrow \mathtt{VALUE} \geq \mathtt{sv} * \mathtt{minv}$①
$\mathtt{rv} \geq 2 \Rightarrow \mathtt{VALUE} \geq \min(2 * \mathtt{minv}②, \mathtt{sv} * \mathtt{minv}③)$
$\mathtt{rv} = 1 \Rightarrow \mathtt{VALUE} = +\infty \vee \mathtt{VALUE} \leq \mathtt{sv} * \mathtt{maxv}$
$\mathtt{rv} \geq 2 \Rightarrow \mathtt{VALUE} = +\infty \vee \mathtt{VALUE} \leq \max(2 * \mathtt{maxv}, \mathtt{sv} * \mathtt{maxv})$
required(VARIABLES, var)
required(FOUND, var)
where
minv =minval(VARIABLES.var)
rv =range(VARIABLES.var)
sv = |VARIABLES|
maxv =maxval(VARIABLES.var)

**Purpose**

The constraint MIN_SURF_STEADY_SEQUENCE(VALUE, VARIABLES) holds. In addition, FOUND is a collection of 0/1 variables where the value 1 indicates the position of the **found** letter in those occurrences of the pattern STEADY_SEQUENCE for which the feature value is VALUE.

The position of the **found** letter in an occurrence of a pattern is the first position where the occurrence of pattern is identified, even if the pattern is not complete.

An occurrence of the pattern STEADY_SEQUENCE is the *maximal* subsequence which matches the regular expression $=^+$.

Assume that the occurrence of the pattern STEADY_SEQUENCE starts at position $i$ and ends at position $j$. The feature SURF computes the sum of the values from index $i$ to index $j + 1$.

**Example**

$$\left( \begin{array}{l} 2, \langle 3, 1, 1, 4, 5, 5, 5, 6, 2, 2, 4, 4, 3, 2, 1, 1 \rangle, \\ \langle 0, 1, 0, 0, 0, 0, 0, 0, 0, 0, 0, 0, 0, 0, 1, 0 \rangle \end{array} \right)$$

Figure 3.1092 provides an example where the POS_MIN_SURF_STEADY_SEQUENCE $(2, [3, 1, 1, 4, 5, 5, 5, 6, 2, 2, 4, 4, 3, 2, 1, 1], [0, 1, 0, 0, 0, 0, 0, 0, 0, 0, 0, 0, 0, 0, 1, 0])$ constraint holds.

Figure 3.1092: Illustrating the `POS_MIN_SURF_STEADY_SEQUENCE` constraint of the **Example** slot

| Typical | $|\mathtt{VARIABLES}| > 1$ |
|---|---|
| **Arg. properties** | • Functional dependency: `VALUE` determined by `VARIABLES`. |
| | • Functional dependency: `FOUND` determined by `VARIABLES`. |

**Automaton**        Similar to the automaton of the MIN_SURF_STEADY_SEQUENCE constraint but use the dec-
                     oration table 2.33.

## 3.528   POS_MIN_SURF_STRICTLY_DECREASING_SEQUENCE

▶ ▷ ◁ ◀        **DESCRIPTION**            **AUTOMATON**

| | |
|---|---|
| **Origin** | Based on constraint MIN_SURF_STRICTLY_DECREASING_SEQUENCE. |
| **Constraint** | POS_MIN_SURF_STRICTLY_DECREASING_SEQUENCE(VALUE, VARIABLES, FOUND) |

| **Arguments** | | |
|---|---|---|
| | VALUE | : dvar |
| | VARIABLES | : collection(var−dvar) |
| | FOUND | : collection(var−dvar) |

**Restrictions**

$\texttt{sv} \leq 1 \vee \texttt{rv} \leq 1 \Rightarrow \texttt{VALUE} = +\infty$

$\texttt{minv} < 0 \Rightarrow \texttt{VALUE} \geq \ell_1 * \texttt{minv} + \lfloor \ell_1 * (\ell_1 - 1)/2 \rfloor$ ①

$\texttt{minv} \geq 0 \Rightarrow \texttt{VALUE} \geq 2 * \texttt{minv} + 1$ ②

$\texttt{maxv} > 0 \Rightarrow \texttt{VALUE} = +\infty \vee \texttt{VALUE} \leq \ell_2 * \texttt{maxv} - \lfloor \ell_2 * (\ell_2 - 1)/2 \rfloor$

$\texttt{maxv} \leq 0 \Rightarrow \texttt{VALUE} = +\infty \vee \texttt{VALUE} \leq 2 * \texttt{maxv} - 1$

required(VARIABLES, var)

required(FOUND, var)

where

$\texttt{minv} = \texttt{minval}(\texttt{VARIABLES.var})$

$\texttt{rv} = \texttt{range}(\texttt{VARIABLES.var})$

$\texttt{sv} = |\texttt{VARIABLES}|$

$\ell_1 = \min(\min(\texttt{sv}, \texttt{rv}), |\texttt{minv}|)$

$\ell_2 = \min(\min(\texttt{sv}, \texttt{rv}), |\texttt{maxv}|)$

$\texttt{maxv} = \texttt{maxval}(\texttt{VARIABLES.var})$

**Purpose**

The constraint MIN_SURF_STRICTLY_DECREASING_SEQUENCE(VALUE, VARIABLES) holds. In addition, FOUND is a collection of 0/1 variables where the value 1 indicates the position of the **found** letter in those occurrences of the pattern STRICTLY_DECREASING_SEQUENCE for which the feature value is VALUE.

The position of the **found** letter in an occurrence of a pattern is the first position where the occurrence of pattern is identified, even if the pattern is not complete.

An occurrence of the pattern STRICTLY_DECREASING_SEQUENCE is the *maximal* sub-sequence which matches the regular expression $>^+$.

Assume that the occurrence of the pattern STRICTLY_DECREASING_SEQUENCE starts at position $i$ and ends at position $j$. The feature SURF computes the sum of the values from index $i$ to index $j + 1$.

**Example**

$$\left( \begin{array}{l} 7, \langle 4, 4, 6, 4, 1, 1, 3, 4, 4, 6, 6, 5, 2, 2, 4, 3 \rangle, \\ \langle 0, 0, 0, 0, 0, 0, 0, 0, 0, 0, 0, 0, 0, 0, 1, 0 \rangle \end{array} \right)$$

Figure 3.1093 provides an example where the POS_MIN_SURF_STRICTLY_DECREASING_SEQUENCE $(7, [4, 4, 6, 4, 1, 1, 3, 4, 4, 6, 6, 5, 2, 2, 4, 3], [0, 0, 0, 0, 0, 0, 0, 0, 0, 0, 0, 0, 0, 0, 1, 0])$ constraint holds.

Figure 3.1093: Illustrating the POS_MIN_SURF_STRICTLY_DECREASING_SEQUENCE constraint of the **Example** slot

| **Typical** | $|\texttt{VARIABLES}| > 1$ |
| | $\texttt{range}(\texttt{VARIABLES.var}) > 1$ |

| **Arg. properties** | • Functional dependency: VALUE determined by VARIABLES. |
| | • Functional dependency: FOUND determined by VARIABLES. |

**Automaton**        Similar to the automaton of the MIN_SURF_STRICTLY_DECREASING_SEQUENCE constraint
                     but use the decoration table 2.33.

## 3.529   POS_MIN_SURF_STRICTLY_INCREASING_SEQUENCE

▶ ▷ ◁ ◀    **DESCRIPTION**        **AUTOMATON**

**Origin**        Based on constraint MIN_SURF_STRICTLY_INCREASING_SEQUENCE.

**Constraint**    POS_MIN_SURF_STRICTLY_INCREASING_SEQUENCE(VALUE, VARIABLES, FOUND)

**Arguments**
```
VALUE     : dvar
VARIABLES : collection(var−dvar)
FOUND     : collection(var−dvar)
```

**Restrictions**

$sv \leq 1 \vee rv \leq 1 \Rightarrow VALUE = +\infty$
$minv < 0 \Rightarrow VALUE \geq \ell_1 * minv + \lfloor \ell_1 * (\ell_1 - 1)/2 \rfloor$ ①
$minv \geq 0 \Rightarrow VALUE \geq 2 * minv + 1$ ②
$maxv > 0 \Rightarrow VALUE = +\infty \vee VALUE \leq \ell_2 * maxv - \lfloor \ell_2 * (\ell_2 - 1)/2 \rfloor$
$maxv \leq 0 \Rightarrow VALUE = +\infty \vee VALUE \leq 2 * maxv - 1$
required(VARIABLES, var)
required(FOUND, var)
where
$minv = $ minval(VARIABLES.var)
$rv = $ range(VARIABLES.var)
$sv = |$VARIABLES$|$
$\ell_1 = \min(\min(sv, rv), |minv|)$
$\ell_2 = \min(\min(sv, rv), |maxv|)$
$maxv = $ maxval(VARIABLES.var)

**Purpose**

The constraint MIN_SURF_STRICTLY_INCREASING_SEQUENCE(VALUE, VARIABLES) holds. In addition, FOUND is a collection of 0/1 variables where the value 1 indicates the position of the **found** letter in those occurrences of the pattern STRICTLY_INCREASING_SEQUENCE for which the feature value is VALUE.

The position of the **found** letter in an occurrence of a pattern is the first position where the occurrence of pattern is identified, even if the pattern is not complete.

An occurrence of the pattern STRICTLY_INCREASING_SEQUENCE is the *maximal* subsequence which matches the regular expression $<^+$.

Assume that the occurrence of the pattern STRICTLY_INCREASING_SEQUENCE starts at position $i$ and ends at position $j$. The feature SURF computes the sum of the values from index $i$ to index $j + 1$.

**Example**

$$\left( \begin{array}{l} 6, \langle 4, 3, 5, 5, 2, 1, 1, 2, 3, 4, 6, 6, 3, 1, 2, 3 \rangle, \\ \langle 0, 0, 0, 0, 0, 0, 0, 0, 0, 0, 0, 0, 0, 0, 1, 0, 0 \rangle \end{array} \right)$$

Figure 3.1094 provides an example where the POS_MIN_SURF_STRICTLY_INCREASING_SEQUENCE $(6, [4, 3, 5, 5, 2, 1, 1, 2, 3, 4, 6, 6, 3, 1, 2, 3], [0, 0, 0, 0, 0, 0, 0, 0, 0, 0, 0, 0, 0, 0, 1, 0, 0])$ constraint holds.

Figure 3.1094: Illustrating the POS_MIN_SURF_STRICTLY_INCREASING_SEQUENCE constraint of the **Example** slot

**Typical**

$|\mathtt{VARIABLES}| > 1$

$\mathtt{range}(\mathtt{VARIABLES.var}) > 1$

**Arg. properties**

- Functional dependency: VALUE determined by VARIABLES.
- Functional dependency: FOUND determined by VARIABLES.

**Automaton**  Similar to the automaton of the MIN_SURF_STRICTLY_INCREASING_SEQUENCE constraint but use the decoration table 2.33.

# 3.530 POS_MIN_SURF_SUMMIT

          **DESCRIPTION**          **AUTOMATON**

$(< | < (= | <)^* <)(> | > (= | >)^* >)$

| | |
|---|---|
| **Origin** | Based on constraint MIN_SURF_SUMMIT. |
| **Constraint** | POS_MIN_SURF_SUMMIT(VALUE, VARIABLES, FOUND) |

**Arguments**

```
VALUE     : dvar
VARIABLES : collection(var−dvar)
FOUND     : collection(var−dvar)
```

**Restrictions**

$\mathtt{sv} \leq 2 \lor \mathtt{rv} \leq 1 \Rightarrow \mathtt{VALUE} = +\infty$

$\mathtt{rv} = 2 \Rightarrow \mathtt{VALUE} \geq \mathtt{minv} + 1$①

$\mathtt{rv} \geq 3 \Rightarrow \mathtt{VALUE} \geq \min(\mathtt{minv} + 1$①$, (\mathtt{sv} - 2) * (\mathtt{minv} + 1) + 1$②$)$

$\mathtt{rv} = 2 \Rightarrow \mathtt{VALUE} = +\infty \lor \mathtt{VALUE} \leq \mathtt{maxv}$

$\mathtt{rv} \geq 3 \Rightarrow \mathtt{VALUE} = +\infty \lor \mathtt{VALUE} \leq \max(\mathtt{maxv}, (\mathtt{sv} - 2) * (\mathtt{maxv} - 1) + 1)$

required(VARIABLES, var)

required(FOUND, var)

where

minv = minval(VARIABLES.var)

rv = range(VARIABLES.var)

sv = |VARIABLES|

maxv = maxval(VARIABLES.var)

**Purpose**

The constraint MIN_SURF_SUMMIT(VALUE, VARIABLES) holds. In addition, FOUND is a collection of 0/1 variables where the value 1 indicates the position of the **found** letter in those occurrences of the pattern SUMMIT for which the feature value is VALUE.

The position of the **found** letter in an occurrence of a pattern is the first position where the occurrence of pattern is identified, even if the pattern is not complete.

An occurrence of the pattern SUMMIT is the *maximal* subsequence which matches the regular expression $(< | < (= | <)^* <)(> | > (= | >)^* >)$.

Assume that the occurrence of the pattern SUMMIT starts at position $i$ and ends at position $j$. The feature SURF computes the sum of the values from index $i + 1$ to index $j$.

**Example**

$$\left( \begin{array}{l} 3, \langle 7, 1, 5, 4, 4, 3, 3, 4, 6, 6, 2, 3, 4, 2, 3, 1 \rangle, \\ \langle 0, 0, 0, 0, 0, 0, 0, 0, 0, 0, 0, 0, 0, 0, 1, 0 \rangle \end{array} \right)$$

Figure 3.1095 provides an example where the POS_MIN_SURF_SUMMIT $(3, [7, 1, 5, 4, 4, 3, 3, 4, 6, 6, 2, 3, 4, 2, 3, 1], [0, 0, 0, 0, 0, 0, 0, 0, 0, 0, 0, 0, 0, 0, 1, 0])$ constraint holds.

**Typical**

$|\mathtt{VARIABLES}| > 2$

range(VARIABLES.var) $> 1$

Figure 3.1095: Illustrating the POS_MIN_SURF_SUMMIT constraint of the **Example** slot

**Arg. properties**
- Functional dependency: VALUE determined by VARIABLES.
- Functional dependency: FOUND determined by VARIABLES.

**Automaton**          Similar to the automaton of the MIN_SURF_SUMMIT constraint but use the decoration table 2.33.

## 3.531   POS_MIN_SURF_VALLEY

**Origin**         Based on constraint MIN_SURF_VALLEY.

**Constraint**     POS_MIN_SURF_VALLEY(VALUE, VARIABLES, FOUND)

**Arguments**      VALUE      : dvar
                   VARIABLES  : collection(var−dvar)
                   FOUND      : collection(var−dvar)

**Restrictions**   $\mathtt{sv} \leq 2 \vee \mathtt{rv} \leq 1 \Rightarrow \mathtt{VALUE} = +\infty$
                   $\mathtt{VALUE} \geq \min(\mathtt{minv}\text{①}, (\mathtt{sv} - 2) * \mathtt{minv}\text{②})$
                   $\mathtt{VALUE} = +\infty \vee \mathtt{VALUE} \leq \max(\mathtt{maxv} - 1, (\mathtt{sv} - 2) * (\mathtt{maxv} - 1))$
                   required(VARIABLES, var)
                   required(FOUND, var)
                   where
                   $\mathtt{minv} =$ minval(VARIABLES.var)
                   $\mathtt{sv} = |\mathtt{VARIABLES}|$
                   $\mathtt{maxv} =$ maxval(VARIABLES.var)
                   $\mathtt{rv} =$ range(VARIABLES.var)

**Purpose**        The constraint MIN_SURF_VALLEY(VALUE, VARIABLES) holds. In addition, FOUND is a
                   collection of 0/1 variables where the value 1 indicates the position of the **found** letter
                   in those occurrences of the pattern VALLEY for which the feature value is VALUE.
                   The position of the **found** letter in an occurrence of a pattern is the first position where
                   the occurrence of pattern is identified, even if the pattern is not complete.
                   An occurrence of the pattern VALLEY is the *maximal* subsequence which matches the
                   regular expression $> (= | >)^* (< | =)^* <$.
                   Assume that the occurrence of the pattern VALLEY starts at position $i$ and ends at position
                   $j$. The feature SURF computes the sum of the values from index $i + 1$ to index $j$.

**Example**        $\left(\begin{array}{l} 7, \langle 1, 3, 7, 4, 3, 6, 6, 5, 3, 3, 2, 6, 5, 5, 5, 7 \rangle, \\ \langle 0, 0, 0, 0, 1, 0, 0, 0, 0, 0, 0, 0, 0, 0, 0, 0 \rangle \end{array}\right)$

                   Figure 3.1096 provides an example where the POS_MIN_SURF_VALLEY
                   $(7, [1, 3, 7, 4, 3, 6, 6, 5, 3, 3, 2, 6, 5, 5, 5, 7], [0, 0, 0, 0, 1, 0, 0, 0, 0, 0, 0, 0, 0, 0, 0, 0])$
                   constraint holds.

**Typical**        $|\mathtt{VARIABLES}| > 2$
                   range(VARIABLES.var) $> 1$

**Arg. properties** • Functional dependency: VALUE determined by VARIABLES.
                    • Functional dependency: FOUND determined by VARIABLES.

Figure 3.1096: Illustrating the POS_MIN_SURF_VALLEY constraint of the **Example** slot

**Automaton**          Similar to the automaton of the MIN_SURF_VALLEY constraint but use the decoration table 2.33.

# 3.532 POS_MIN_SURF_ZIGZAG

► ▷ ◁ ◀        DESCRIPTION        AUTOMATON

| | |
|---|---|
| **Origin** | Based on constraint MIN_SURF_ZIGZAG. |
| **Constraint** | POS_MIN_SURF_ZIGZAG(VALUE, VARIABLES, FOUND) |

**Arguments**

```
VALUE     : dvar
VARIABLES : collection(var-dvar)
FOUND     : collection(var-dvar)
```

**Restrictions**

$\texttt{sv} \leq 3 \lor \texttt{rv} \leq 1 \Rightarrow \texttt{VALUE} = +\infty$

$\texttt{VALUE} \geq \min \left( \begin{array}{l} 2 * \texttt{minv} + 1 ①, \\ \lfloor (\texttt{sv} - 1)/2 \rfloor * \texttt{minv} + \lfloor (\texttt{sv} - 2)/2 \rfloor * (\texttt{minv} + 1) ② \end{array} \right)$

$\lor \left( \begin{array}{l} \texttt{VALUE} = +\infty, \\ \texttt{VALUE} \leq \max \left( \begin{array}{l} 2 * \texttt{maxv} - 1, \\ \lfloor (\texttt{sv} - 1)/2 \rfloor * \texttt{maxv} + \lfloor (\texttt{sv} - 2)/2 \rfloor * (\texttt{maxv} - 1) \end{array} \right) \end{array} \right)$

required(VARIABLES, var)
required(FOUND, var)

where

$\texttt{minv} = \texttt{minval}(\texttt{VARIABLES.var})$
$\texttt{sv} = |\texttt{VARIABLES}|$
$\texttt{maxv} = \texttt{maxval}(\texttt{VARIABLES.var})$
$\texttt{rv} = \texttt{range}(\texttt{VARIABLES.var})$

**Purpose**

The constraint MIN_SURF_ZIGZAG(VALUE, VARIABLES) holds. In addition, FOUND is a collection of 0/1 variables where the value 1 indicates the position of the **found** letter in those occurrences of the pattern ZIGZAG for which the feature value is VALUE.

The position of the **found** letter in an occurrence of a pattern is the first position where the occurrence of pattern is identified, even if the pattern is not complete.

An occurrence of the pattern ZIGZAG is the *maximal* subsequence which matches the regular expression $(< >)^+(< \mid < >) \mid (> <)^+(> \mid > <)$.

Assume that the occurrence of the pattern ZIGZAG starts at position $i$ and ends at position $j$. The feature SURF computes the sum of the values from index $i + 1$ to index $j$.

**Example**

$$\left( \begin{array}{l} 5, \langle 4, 1, 3, 1, 4, 6, 1, 5, 5, 2, 7, 2, 3, 1, 6, 1 \rangle, \\ \langle 0, 0, 1, 0, 0, 0, 0, 0, 0, 0, 0, 0, 0, 0, 0, 0 \rangle \end{array} \right)$$

Figure 3.1097 provides an example where the POS_MIN_SURF_ZIGZAG (5, [4, 1, 3, 1, 4, 6, 1, 5, 5, 2, 7, 2, 3, 1, 6, 1], [0, 0, 1, 0, 0, 0, 0, 0, 0, 0, 0, 0, 0, 0, 0, 0]) constraint holds.

**Typical**

$|\texttt{VARIABLES}| > 3$
$\texttt{range}(\texttt{VARIABLES.var}) > 1$

Figure 3.1097: Illustrating the POS_MIN_SURF_ZIGZAG constraint of the **Example** slot

**Arg. properties**
- Functional dependency: VALUE determined by VARIABLES.
- Functional dependency: FOUND determined by VARIABLES.

**Automaton**          Similar to the automaton of the MIN_SURF_ZIGZAG constraint but use the decoration ta-
                       ble 2.33.

# 3.533   POS_MIN_WIDTH_DECREASING_SEQUENCE

    **DESCRIPTION**          **AUTOMATON**

| | |
|---|---|
| **Origin** | Based on constraint MIN_WIDTH_DECREASING_SEQUENCE. |
| **Constraint** | POS_MIN_WIDTH_DECREASING_SEQUENCE(VALUE, VARIABLES, FOUND) |
| **Arguments** | VALUE : dvar <br> VARIABLES : collection(var−dvar) <br> FOUND : collection(var−dvar) |

**Restrictions**

$$\mathtt{sv} \leq 1 \vee \mathtt{rv} \leq 1 \Rightarrow \mathtt{VALUE} = \mathtt{sv} + 1$$
$$\mathtt{VALUE} \geq 2$$
$$\mathtt{rv} = 2 \Rightarrow \mathtt{VALUE} = \mathtt{sv} + 1 \vee \mathtt{VALUE} \leq 2 ①$$
$$\mathtt{rv} \geq 3 \Rightarrow \mathtt{VALUE} = \mathtt{sv} + 1 \vee \mathtt{VALUE} \leq \mathtt{sv} ②$$
required(VARIABLES, var)
required(FOUND, var)
where
$$\mathtt{sv} = |\mathtt{VARIABLES}|$$
$$\mathtt{rv} = \mathrm{range}(\mathtt{VARIABLES.var})$$

**Purpose**

The constraint MIN_WIDTH_DECREASING_SEQUENCE(VALUE, VARIABLES) holds. In addition, FOUND is a collection of 0/1 variables where the value 1 indicates the position of the **found** letter in those occurrences of the pattern DECREASING_SEQUENCE for which the feature value is VALUE.

The position of the **found** letter in an occurrence of a pattern is the first position where the occurrence of pattern is identified, even if the pattern is not complete.

An occurrence of the pattern DECREASING_SEQUENCE is the *maximal* subsequence which matches the regular expression $> (> | =)^* > | >$.

Assume that the occurrence of the pattern DECREASING_SEQUENCE starts at position $i$ and ends at position $j$. The feature WIDTH computes the value $j - i + 2$.

**Example**

$$\left( \begin{array}{l} 2, \langle 3, 4, 2, 2, 5, 6, 6, 4, 4, 3, 1, 1, 4, 6, 4, 4 \rangle, \\ \langle 0, 1, 0, 0, 0, 0, 0, 0, 0, 0, 0, 0, 0, 0, 1, 0, 0 \rangle \end{array} \right)$$

Figure 3.1098 provides an example where the POS_MIN_WIDTH_DECREASING_SEQUENCE $(2, [3, 4, 2, 2, 5, 6, 6, 4, 4, 3, 1, 1, 4, 6, 4, 4], [0, 1, 0, 0, 0, 0, 0, 0, 0, 0, 0, 0, 0, 0, 1, 0, 0])$ constraint holds.

**Typical**

$$|\mathtt{VARIABLES}| > 1$$
$$\mathrm{range}(\mathtt{VARIABLES.var}) > 1$$

**Arg. properties**

- Functional dependency: VALUE determined by VARIABLES.
- Functional dependency: FOUND determined by VARIABLES.

Figure 3.1098: Illustrating the POS_MIN_WIDTH_DECREASING_SEQUENCE constraint of the **Example** slot

**Automaton**          Similar to the automaton of the MIN_WIDTH_DECREASING_SEQUENCE constraint but use
                       the decoration table 2.33.

## 3.534   POS_MIN_WIDTH_DECREASING_TERRACE

▶ ▷ ◁ ◀        **DESCRIPTION**              **AUTOMATON**

| | |
|---|---|
| **Origin** | Based on constraint MIN_WIDTH_DECREASING_TERRACE. |
| **Constraint** | POS_MIN_WIDTH_DECREASING_TERRACE(VALUE, VARIABLES, FOUND) |

**Arguments**
```
VALUE     : dvar
VARIABLES : collection(var−dvar)
FOUND     : collection(var−dvar)
```

**Restrictions**

$sv \leq 3 \vee rv \leq 2 \Rightarrow$ VALUE $= sv + 1$

VALUE $\geq 2$

VALUE $= sv + 1 \vee$ VALUE $\leq sv - 2$ ①

required(VARIABLES, var)

required(FOUND, var)

where

$sv = |$VARIABLES$|$

$rv =$ range(VARIABLES.var)

**Purpose**

The constraint MIN_WIDTH_DECREASING_TERRACE(VALUE, VARIABLES) holds. In addition, FOUND is a collection of 0/1 variables where the value 1 indicates the position of the **found** letter in those occurrences of the pattern DECREASING_TERRACE for which the feature value is VALUE.

The position of the **found** letter in an occurrence of a pattern is the first position where the occurrence of pattern is identified, even if the pattern is not complete.

An occurrence of the pattern DECREASING_TERRACE is the *maximal* subsequence which matches the regular expression $> =^+ >$.

Assume that the occurrence of the pattern DECREASING_TERRACE starts at position $i$ and ends at position $j$. The feature WIDTH computes the value $j - i$.

**Example**

$$\left( \begin{array}{l} 2, \langle 6, 4, 4, 4, 5, 2, 2, 1, 3, 3, 5, 4, 4, 3, 3, 3 \rangle, \\ \langle 0, 0, 0, 0, 0, 0, 1, 0, 0, 0, 0, 0, 1, 0, 0, 0 \rangle \end{array} \right)$$

Figure 3.1099 provides an example where the POS_MIN_WIDTH_DECREASING_TERRACE $(2, [6, 4, 4, 4, 5, 2, 2, 1, 3, 3, 5, 4, 4, 3, 3, 3], [0, 0, 0, 0, 0, 0, 1, 0, 0, 0, 0, 0, 1, 0, 0, 0])$ constraint holds.

**Typical**

$|$VARIABLES$| > 3$

range(VARIABLES.var) $> 2$

**Arg. properties**

- Functional dependency: VALUE determined by VARIABLES.
- Functional dependency: FOUND determined by VARIABLES.

Figure 3.1099: Illustrating the POS_MIN_WIDTH_DECREASING_TERRACE constraint of the **Example** slot

**Automaton**          Similar to the automaton of the MIN_WIDTH_DECREASING_TERRACE constraint but use
the decoration table 2.33.

## 3.535   POS_MIN_WIDTH_GORGE

$(> \mid > (= \mid >)^* >)(< \mid < (= \mid <)^* <)$

| | |
|---|---|
| Origin | Based on constraint MIN_WIDTH_GORGE. |
| Constraint | POS_MIN_WIDTH_GORGE(VALUE, VARIABLES, FOUND) |
| Arguments | VALUE : dvar<br>VARIABLES : collection(var−dvar)<br>FOUND : collection(var−dvar) |
| Restrictions | $\mathtt{sv} \leq 2 \vee \mathtt{rv} \leq 1 \Rightarrow \mathtt{VALUE} = \mathtt{sv} + 1$<br>$\mathtt{VALUE} \geq 1$<br>$\mathtt{rv} = 2 \Rightarrow \mathtt{VALUE} = \mathtt{sv} + 1 \vee \mathtt{VALUE} \leq 1$①<br>$\mathtt{rv} \geq 3 \Rightarrow \mathtt{VALUE} = \mathtt{sv} + 1 \vee \mathtt{VALUE} \leq \mathtt{sv} - 2$②<br>required(VARIABLES, var)<br>required(FOUND, var)<br>where<br>$\mathtt{sv} = \lvert\mathtt{VARIABLES}\rvert$<br>$\mathtt{rv} = \text{range}(\mathtt{VARIABLES.var})$ |

The constraint MIN_WIDTH_GORGE(VALUE, VARIABLES) holds. In addition, FOUND is a collection of 0/1 variables where the value 1 indicates the position of the **found** letter in those occurrences of the pattern GORGE for which the feature value is VALUE.

The position of the **found** letter in an occurrence of a pattern is the first position in the sequence where the occurrence of pattern is identified, even if the pattern is not complete.

An occurrence of the pattern GORGE is the *maximal* subsequence which matches the regular expression $(> \mid > (= \mid >)^* >)(< \mid < (= \mid <)^* <)$.

Assume that the occurrence of the pattern GORGE starts at position $i$ and ends at position $j$. The feature WIDTH computes the value $j - i$.

| | |
|---|---|
| Example | $\left( \begin{array}{l} 1, \langle 1, 7, 3, 4, 4, 5, 5, 4, 2, 2, 6, 5, 4, 6, 5, 7 \rangle, \\ \langle 0, 0, 0, 0, 0, 0, 0, 0, 0, 0, 0, 0, 0, 0, 1, 0 \rangle \end{array} \right)$ |

Figure 3.1100 provides an example where the POS_MIN_WIDTH_GORGE $(1, [1, 7, 3, 4, 4, 5, 5, 4, 2, 2, 6, 5, 4, 6, 5, 7], [0, 0, 0, 0, 0, 0, 0, 0, 0, 0, 0, 0, 0, 0, 1, 0])$ constraint holds.

| | |
|---|---|
| Typical | $\lvert\mathtt{VARIABLES}\rvert > 2$<br>$\text{range}(\mathtt{VARIABLES.var}) > 1$ |
| Arg. properties | • Functional dependency: VALUE determined by VARIABLES.<br>• Functional dependency: FOUND determined by VARIABLES. |

Figure 3.1100: Illustrating the POS_MIN_WIDTH_GORGE constraint of the **Example** slot

**Automaton**          Similar to the automaton of the MIN_WIDTH_GORGE constraint but use the decoration table 2.33.

# 3.536  POS_MIN_WIDTH_INCREASING_SEQUENCE

**DESCRIPTION**          **AUTOMATON**

| | |
|---|---|
| **Origin** | Based on constraint MIN_WIDTH_INCREASING_SEQUENCE. |
| **Constraint** | POS_MIN_WIDTH_INCREASING_SEQUENCE(VALUE, VARIABLES, FOUND) |

**Arguments**

VALUE      : dvar
VARIABLES  : collection(var−dvar)
FOUND      : collection(var−dvar)

**Restrictions**

$sv \leq 1 \lor rv \leq 1 \Rightarrow$ VALUE $= sv + 1$
VALUE $\geq 2$
$rv = 2 \Rightarrow$ VALUE $= sv + 1 \lor$ VALUE $\leq 2$①
$rv \geq 3 \Rightarrow$ VALUE $= sv + 1 \lor$ VALUE $\leq sv$②
required(VARIABLES, var)
required(FOUND, var)

where

$sv = |$VARIABLES$|$
$rv =$ range(VARIABLES.var)

**Purpose**

The constraint MIN_WIDTH_INCREASING_SEQUENCE(VALUE, VARIABLES) holds. In addition, FOUND is a collection of 0/1 variables where the value 1 indicates the position of the **found** letter in those occurrences of the pattern INCREASING_SEQUENCE for which the feature value is VALUE.

The position of the **found** letter in an occurrence of a pattern is the first position where the occurrence of pattern is identified, even if the pattern is not complete.

An occurrence of the pattern INCREASING_SEQUENCE is the *maximal* subsequence which matches the regular expression $< (< | =)^* < | <$.

Assume that the occurrence of the pattern INCREASING_SEQUENCE starts at position $i$ and ends at position $j$. The feature WIDTH computes the value $j - i + 2$.

**Example**

$$\left( \begin{array}{l} 2, \langle 4, 3, 5, 5, 2, 1, 1, 3, 3, 4, 6, 6, 3, 1, 3, 3 \rangle , \\ \langle 0, 1, 0, 0, 0, 0, 0, 0, 0, 0, 0, 0, 0, 0, 1, 0, 0 \rangle \end{array} \right)$$

Figure 3.1101 provides an example where the POS_MIN_WIDTH_INCREASING_SEQUENCE $(2, [4, 3, 5, 5, 2, 1, 1, 3, 3, 4, 6, 6, 3, 1, 3, 3], [0, 1, 0, 0, 0, 0, 0, 0, 0, 0, 0, 0, 0, 0, 1, 0, 0])$ constraint holds.

**Typical**

$|$VARIABLES$| > 1$
range(VARIABLES.var) $> 1$

**Arg. properties**

- Functional dependency: VALUE determined by VARIABLES.
- Functional dependency: FOUND determined by VARIABLES.

Figure 3.1101: Illustrating the POS_MIN_WIDTH_INCREASING_SEQUENCE constraint of the **Example** slot

**Automaton**      Similar to the automaton of the MIN_WIDTH_INCREASING_SEQUENCE constraint but use
the decoration table 2.33.

# 3.537 POS_MIN_WIDTH_INCREASING_TERRACE

         DESCRIPTION            AUTOMATON

| | |
|---|---|
| **Origin** | Based on constraint MIN_WIDTH_INCREASING_TERRACE. |
| **Constraint** | POS_MIN_WIDTH_INCREASING_TERRACE(VALUE, VARIABLES, FOUND) |
| **Arguments** | VALUE     : dvar |
| | VARIABLES : collection(var−dvar) |
| | FOUND     : collection(var−dvar) |

**Restrictions**

$$sv \leq 3 \vee rv \leq 2 \Rightarrow VALUE = sv + 1$$
$$VALUE \geq 2$$
$$VALUE = sv + 1 \vee VALUE \leq sv - 2 \,①$$
required(VARIABLES, var)
required(FOUND, var)
where
$$sv = |VARIABLES|$$
$$rv = range(VARIABLES.var)$$

**Purpose**

The constraint MIN_WIDTH_INCREASING_TERRACE(VALUE, VARIABLES) holds. In addition, FOUND is a collection of 0/1 variables where the value 1 indicates the position of the **found** letter in those occurrences of the pattern INCREASING_TERRACE for which the feature value is VALUE.

The position of the **found** letter in an occurrence of a pattern is the first position where the occurrence of pattern is identified, even if the pattern is not complete.

An occurrence of the pattern INCREASING_TERRACE is the *maximal* subsequence which matches the regular expression $<=^+<$.

Assume that the occurrence of the pattern INCREASING_TERRACE starts at position $i$ and ends at position $j$. The feature WIDTH computes the value $j - i$.

**Example**

$$\left( \begin{array}{l} 2, \langle 1, 3, 3, 3, 2, 5, 5, 6, 4, 4, 2, 3, 3, 3, 4, 4 \rangle, \\ \langle 0, 0, 0, 0, 0, 0, 1, 0, 0, 0, 0, 0, 0, 0, 0, 0 \rangle \end{array} \right)$$

Figure 3.1102 provides an example where the POS_MIN_WIDTH_INCREASING_TERRACE $(2, [1, 3, 3, 3, 2, 5, 5, 6, 4, 4, 2, 3, 3, 3, 4, 4], [0, 0, 0, 0, 0, 0, 1, 0, 0, 0, 0, 0, 0, 0, 0, 0])$ constraint holds.

**Typical**

$$|VARIABLES| > 3$$
$$range(VARIABLES.var) > 2$$

**Arg. properties**

- Functional dependency: VALUE determined by VARIABLES.
- Functional dependency: FOUND determined by VARIABLES.

Figure 3.1102: Illustrating the POS_MIN_WIDTH_INCREASING_TERRACE constraint of the **Example** slot

**Automaton**          Similar to the automaton of the MIN_WIDTH_INCREASING_TERRACE constraint but use the
                       decoration table 2.33.

## 3.538   POS_MIN_WIDTH_INFLEXION

          **DESCRIPTION**          **AUTOMATON**

$< (< | =)^* > | > (>|=)^* <$

| | |
|---|---|
| **Origin** | Based on constraint MIN_WIDTH_INFLEXION. |
| **Constraint** | POS_MIN_WIDTH_INFLEXION(VALUE, VARIABLES, FOUND) |

**Arguments**

```
VALUE     : dvar
VARIABLES : collection(var−dvar)
FOUND     : collection(var−dvar)
```

**Restrictions**

$sv \leq 2 \vee rv \leq 1 \Rightarrow VALUE = sv + 1$

$VALUE \geq 1$

$VALUE = sv + 1 \vee VALUE \leq sv - 2 \textcircled{1}$

required(VARIABLES, var)

required(FOUND, var)

where

$sv = |VARIABLES|$

$rv = range(VARIABLES.var)$

VALUE $\leq 4 = 6 - 2$

VALUE $\leq 4 = 6 - 2$

**Purpose**

The constraint MIN_WIDTH_INFLEXION(VALUE, VARIABLES) holds. In addition, FOUND is a collection of 0/1 variables where the value 1 indicates the position of the **found** letter in those occurrences of the pattern INFLEXION for which the feature value is VALUE. The position of the **found** letter in an occurrence of a pattern is the first position where the occurrence of pattern is identified, even if the pattern is not complete.

An occurrence of the pattern INFLEXION is the *maximal* subsequence which matches the regular expression $< (< | =)^* > | > (>|=)^* <$.

Assume that the occurrence of the pattern INFLEXION starts at position $i$ and ends at position $j$. The feature WIDTH computes the value $j - i$.

**Example**

$$\left( \begin{array}{l} 1, \langle 1, 2, 6, 6, 4, 4, 3, 5, 2, 5, 1, 5, 3, 3, 4, 4 \rangle, \\ \langle 0, 0, 0, 0, 0, 0, 0, 1, 1, 1, 1, 1, 0, 0, 0, 0 \rangle \end{array} \right)$$

Figure 3.1103 provides an example where the POS_MIN_WIDTH_INFLEXION $(1, [1, 2, 6, 6, 4, 4, 3, 5, 2, 5, 1, 5, 3, 3, 4, 4], [0, 0, 0, 0, 0, 0, 0, 1, 1, 1, 1, 1, 0, 0, 0, 0])$ constraint holds.

**Typical**

$|VARIABLES| > 2$

$range(VARIABLES.var) > 1$

**Arg. properties**

- Functional dependency: VALUE determined by VARIABLES.
- Functional dependency: FOUND determined by VARIABLES.

Figure 3.1103: Illustrating the POS_MIN_WIDTH_INFLEXION constraint of the **Example** slot

**Automaton**  Similar to the automaton of the MIN_WIDTH_INFLEXION constraint but use the decoration table 2.33.

## 3.539  POS_MIN_WIDTH_PEAK

► ▷ ◁ ◄      **DESCRIPTION**          **AUTOMATON**

$< (= | <)^* (> | =)^* >$

**Origin**            Based on constraint MIN_WIDTH_PEAK.

**Constraint**        POS_MIN_WIDTH_PEAK(VALUE, VARIABLES, FOUND)

**Arguments**         VALUE     : dvar
                      VARIABLES : collection(var−dvar)
                      FOUND     : collection(var−dvar)

**Restrictions**      $\mathtt{sv} \leq 2 \lor \mathtt{rv} \leq 1 \Rightarrow \mathtt{VALUE} = \mathtt{sv} + 1$
                      $\mathtt{VALUE} \geq 1$
                      $\mathtt{VALUE} = \mathtt{sv} + 1 \lor \mathtt{VALUE} \leq \mathtt{sv} - 2$①
                      required(VARIABLES, var)
                      required(FOUND, var)
                      where
                      $\mathtt{sv} = |\mathtt{VARIABLES}|$
                      $\mathtt{rv} = \mathtt{range}(\mathtt{VARIABLES.var})$

VALUE $\leq$ **4** $= 6 - 2$

**Purpose**           The constraint MIN_WIDTH_PEAK(VALUE, VARIABLES) holds. In addition, FOUND is a
                      collection of 0/1 variables where the value 1 indicates the position of the **found** letter
                      in those occurrences of the pattern PEAK for which the feature value is VALUE.
                      The position of the **found** letter in an occurrence of a pattern is the first position where
                      the occurrence of pattern is identified, even if the pattern is not complete.
                      An occurrence of the pattern PEAK is the *maximal* subsequence which matches the reg-
                      ular expression $< (= | <)^* (> | =)^* >$.
                      Assume that the occurrence of the pattern PEAK starts at position $i$ and ends at position
                      $j$. The feature WIDTH computes the value $j - i$.

**Example**           $\left( \begin{array}{l} 2, \langle 7, 5, 5, 1, 4, 5, 2, 2, 3, 5, 6, 2, 3, 3, 3, 1 \rangle, \\ \langle 0, 0, 0, 0, 0, 1, 0, 0, 0, 0, 0, 0, 0, 0, 0, 0 \rangle \end{array} \right)$

                      Figure 3.1104 provides an example where the POS_MIN_WIDTH_PEAK
                      $(2, [7, 5, 5, 1, 4, 5, 2, 2, 3, 5, 6, 2, 3, 3, 3, 1], [0, 0, 0, 0, 0, 1, 0, 0, 0, 0, 0, 0, 0, 0, 0, 0])$
                      constraint holds.

**Typical**           $|\mathtt{VARIABLES}| > 2$
                      $\mathtt{range}(\mathtt{VARIABLES.var}) > 1$

**Arg. properties**   • Functional dependency: VALUE determined by VARIABLES.
                      • Functional dependency: FOUND determined by VARIABLES.

Figure 3.1104: Illustrating the POS_MIN_WIDTH_PEAK constraint of the **Example** slot

**Automaton**          Similar to the automaton of the MIN_WIDTH_PEAK constraint but use the decoration table 2.33.

## 3.540   POS_MIN_WIDTH_PLAIN

▶ ▷ ◁ ◀          **DESCRIPTION**          **AUTOMATON**

**Origin**          Based on constraint MIN_WIDTH_PLAIN.

**Constraint**      POS_MIN_WIDTH_PLAIN(VALUE, VARIABLES, FOUND)

**Arguments**
```
VALUE     : dvar
VARIABLES : collection(var−dvar)
FOUND     : collection(var−dvar)
```

**Restrictions**

$\mathtt{sv} \leq 2 \vee \mathtt{rv} \leq 1 \Rightarrow \mathtt{VALUE} = \mathtt{sv} + 1$

$\mathtt{VALUE} \geq 1$

$\mathtt{VALUE} = \mathtt{sv} + 1 \vee \mathtt{VALUE} \leq \mathtt{sv} - 2①$

required(VARIABLES, var)

required(FOUND, var)

where

$\mathtt{sv} = |\mathtt{VARIABLES}|$

$\mathtt{rv} = \mathtt{range}(\mathtt{VARIABLES.var})$

**Purpose**

The constraint MIN_WIDTH_PLAIN(VALUE, VARIABLES) holds. In addition, FOUND is a collection of 0/1 variables where the value 1 indicates the position of the **found** letter in those occurrences of the pattern PLAIN for which the feature value is VALUE.

The position of the **found** letter in an occurrence of a pattern is the first position where the occurrence of pattern is identified, even if the pattern is not complete.

An occurrence of the pattern PLAIN is the *maximal* subsequence which matches the regular expression $> =^* <$.

Assume that the occurrence of the pattern PLAIN starts at position $i$ and ends at position $j$. The feature WIDTH computes the value $j - i$.

**Example**

$$\left( \begin{array}{l} 1, \langle 2, 3, 6, 5, 7, 6, 6, 4, 5, 5, 4, 3, 3, 6, 6, 3 \rangle \,, \\ \langle 0, 0, 0, 1, 0, 0, 0, 1, 0, 0, 0, 0, 0, 0, 0, 0 \rangle \end{array} \right)$$

Figure 3.1105 provides an example where the POS_MIN_WIDTH_PLAIN $(1, [2, 3, 6, 5, 7, 6, 6, 4, 5, 5, 4, 3, 3, 6, 6, 3], [0, 0, 0, 1, 0, 0, 0, 1, 0, 0, 0, 0, 0, 0, 0, 0])$ constraint holds.

**Typical**

$|\mathtt{VARIABLES}| > 2$

$\mathtt{range}(\mathtt{VARIABLES.var}) > 1$

**Arg. properties**

- Functional dependency: VALUE determined by VARIABLES.
- Functional dependency: FOUND determined by VARIABLES.

Figure 3.1105: Illustrating the POS_MIN_WIDTH_PLAIN constraint of the **Example** slot

**Automaton**          Similar to the automaton of the MIN_WIDTH_PLAIN constraint but use the decoration table 2.33.

## 3.541  POS_MIN_WIDTH_PLATEAU

▶ ▷ ◁ ◀     **DESCRIPTION**          **AUTOMATON**

**Origin**          Based on constraint MIN_WIDTH_PLATEAU.

**Constraint**      POS_MIN_WIDTH_PLATEAU(VALUE, VARIABLES, FOUND)

**Arguments**
```
VALUE      : dvar
VARIABLES  : collection(var−dvar)
FOUND      : collection(var−dvar)
```

**Restrictions**    $\mathtt{sv} \leq 2 \vee \mathtt{rv} \leq 1 \Rightarrow \mathtt{VALUE} = \mathtt{sv} + 1$

$\mathtt{VALUE} \geq 1$

$\mathtt{VALUE} = \mathtt{sv} + 1 \vee \mathtt{VALUE} \leq \mathtt{sv} - 2$①

required(VARIABLES, var)

required(FOUND, var)

where

$\mathtt{sv} = |\mathtt{VARIABLES}|$

$\mathtt{rv} = \mathtt{range}(\mathtt{VARIABLES.var})$

**Purpose**

The constraint MIN_WIDTH_PLATEAU(VALUE, VARIABLES) holds. In addition, FOUND is a collection of 0/1 variables where the value 1 indicates the position of the **found** letter in those occurrences of the pattern PLATEAU for which the feature value is VALUE.

The position of the **found** letter in an occurrence of a pattern is the first position where the occurrence of pattern is identified, even if the pattern is not complete.

An occurrence of the pattern PLATEAU is the *maximal* subsequence which matches the regular expression $\mathtt{<=^*>}$.

Assume that the occurrence of the pattern PLATEAU starts at position $i$ and ends at position $j$. The feature WIDTH computes the value $j - i$.

**Example**

$$\left( \begin{array}{l} 3, \langle 1, 3, 3, 5, 5, 5, 5, 2, 4, 4, 4, 3, 3, 1, 5, 5 \rangle, \\ \langle 0, 0, 0, 0, 0, 0, 0, 0, 0, 0, 0, 1, 0, 0, 0, 0 \rangle \end{array} \right)$$

Figure 3.1106 provides an example where the POS_MIN_WIDTH_PLATEAU $(3, [1, 3, 3, 5, 5, 5, 5, 2, 4, 4, 4, 3, 3, 1, 5, 5], [0, 0, 0, 0, 0, 0, 0, 0, 0, 0, 0, 1, 0, 0, 0, 0])$ constraint holds.

**Typical**         $|\mathtt{VARIABLES}| > 2$

range(VARIABLES.var) > 1

**Arg. properties**
- Functional dependency: VALUE determined by VARIABLES.
- Functional dependency: FOUND determined by VARIABLES.

Figure 3.1106: Illustrating the POS_MIN_WIDTH_PLATEAU constraint of the **Example** slot

**Automaton**        Similar to the automaton of the MIN_WIDTH_PLATEAU constraint but use the decoration table 2.33.

## 3.542  POS_MIN_WIDTH_PROPER_PLAIN

▶ ▷ ◁ ◀        **DESCRIPTION**              **AUTOMATON**

| | |
|---|---|
| **Origin** | Based on constraint MIN_WIDTH_PROPER_PLAIN. |
| **Constraint** | POS_MIN_WIDTH_PROPER_PLAIN(VALUE, VARIABLES, FOUND) |
| **Arguments** | VALUE       :  dvar<br>VARIABLES  :  collection(var−dvar)<br>FOUND       :  collection(var−dvar) |
| **Restrictions** | $\mathtt{sv} \leq 3 \vee \mathtt{rv} \leq 1 \Rightarrow \mathtt{VALUE} = \mathtt{sv} + 1$<br>$\mathtt{VALUE} \geq 2$<br>$\mathtt{VALUE} = \mathtt{sv} + 1 \vee \mathtt{VALUE} \leq \mathtt{sv} - 2$ ①<br>required(VARIABLES, var)<br>required(FOUND, var)<br>where<br>$\mathtt{sv} = \lvert \mathtt{VARIABLES} \rvert$<br>$\mathtt{rv} = \mathtt{range}(\mathtt{VARIABLES.var})$ |
| **Purpose** | The constraint MIN_WIDTH_PROPER_PLAIN(VALUE, VARIABLES) holds. In addition, FOUND is a collection of 0/1 variables where the value 1 indicates the position of the **found** letter in those occurrences of the pattern PROPER_PLAIN for which the feature value is VALUE.<br><br>The position of the **found** letter in an occurrence of a pattern is the first position where the occurrence of pattern is identified, even if the pattern is not complete.<br><br>An occurrence of the pattern PROPER_PLAIN is the *maximal* subsequence which matches the regular expression $>=^+<$.<br><br>Assume that the occurrence of the pattern PROPER_PLAIN starts at position $i$ and ends at position $j$. The feature WIDTH computes the value $j - i$. |
| **Example** | $\left( \begin{array}{l} 2, \langle 2, 7, 5, 5, 6, 3, 7, 4, 4, 5, 6, 5, 3, 3, 3, 5 \rangle, \\ \langle 0, 0, 0, 1, 0, 0, 0, 0, 1, 0, 0, 0, 0, 0, 0, 0 \rangle \end{array} \right)$ |

Figure 3.1107 provides an example where the POS_MIN_WIDTH_PROPER_PLAIN $(2, [2, 7, 5, 5, 6, 3, 7, 4, 4, 5, 6, 5, 3, 3, 3, 5], [0, 0, 0, 1, 0, 0, 0, 0, 1, 0, 0, 0, 0, 0, 0, 0])$ constraint holds.

| | |
|---|---|
| **Typical** | $\lvert \mathtt{VARIABLES} \rvert > 3$<br>range(VARIABLES.var) > 1 |
| **Arg. properties** | • Functional dependency: VALUE determined by VARIABLES.<br>• Functional dependency: FOUND determined by VARIABLES. |

Figure 3.1107: Illustrating the POS_MIN_WIDTH_PROPER_PLAIN constraint of the **Example** slot

**Automaton**        Similar to the automaton of the MIN_WIDTH_PROPER_PLAIN constraint but use the decoration table 2.33.

# 3.543  POS_MIN_WIDTH_PROPER_PLATEAU

▶ ▷ ◁ ◀          DESCRIPTION          AUTOMATON

**Origin**          Based on constraint MIN_WIDTH_PROPER_PLATEAU.

**Constraint**      POS_MIN_WIDTH_PROPER_PLATEAU(VALUE, VARIABLES, FOUND)

**Arguments**
```
VALUE      :  dvar
VARIABLES  :  collection(var−dvar)
FOUND      :  collection(var−dvar)
```

**Restrictions**
$sv \leq 3 \vee rv \leq 1 \Rightarrow VALUE = sv + 1$

$VALUE \geq 2$

$VALUE = sv + 1 \vee VALUE \leq sv - 2$ ①

required(VARIABLES, var)

required(FOUND, var)

where

$sv = |VARIABLES|$

$rv = range(VARIABLES.var)$

**Purpose**
The constraint MIN_WIDTH_PROPER_PLATEAU(VALUE, VARIABLES) holds. In addition, FOUND is a collection of 0/1 variables where the value 1 indicates the position of the **found** letter in those occurrences of the pattern PROPER_PLATEAU for which the feature value is VALUE.

The position of the **found** letter in an occurrence of a pattern is the first position where the occurrence of pattern is identified, even if the pattern is not complete.

An occurrence of the pattern PROPER_PLATEAU is the *maximal* subsequence which matches the regular expression $< =^+ >$.

Assume that the occurrence of the pattern PROPER_PLATEAU starts at position $i$ and ends at position $j$. The feature WIDTH computes the value $j - i$.

**Example**
$$\left( \begin{array}{l} 2, \langle 7, 1, 3, 3, 2, 5, 1, 4, 4, 3, 2, 3, 5, 5, 5, 3 \rangle \,, \\ \langle 0, 0, 0, 1, 0, 0, 0, 0, 1, 0, 0, 0, 0, 0, 0, 0 \rangle \end{array} \right)$$

Figure 3.1108 provides an example where the POS_MIN_WIDTH_PROPER_PLATEAU $(2, [7, 1, 3, 3, 2, 5, 1, 4, 4, 3, 2, 3, 5, 5, 5, 3], [0, 0, 0, 1, 0, 0, 0, 0, 1, 0, 0, 0, 0, 0, 0, 0])$ constraint holds.

**Typical**
$|VARIABLES| > 3$

$range(VARIABLES.var) > 1$

**Arg. properties**
- Functional dependency: VALUE determined by VARIABLES.
- Functional dependency: FOUND determined by VARIABLES.

Figure 3.1108: Illustrating the POS_MIN_WIDTH_PROPER_PLATEAU constraint of the **Example** slot

**Automaton**          Similar to the automaton of the MIN_WIDTH_PROPER_PLATEAU constraint but use the decoration table 2.33.

# 3.544　POS_MIN_WIDTH_STEADY_SEQUENCE

▶ ▷ ◁ ◀　　　　**DESCRIPTION**　　　　**AUTOMATON**

| | |
|---|---|
| **Origin** | Based on constraint MIN_WIDTH_STEADY_SEQUENCE. |
| **Constraint** | POS_MIN_WIDTH_STEADY_SEQUENCE(VALUE, VARIABLES, FOUND) |
| **Arguments** | VALUE : dvar<br>VARIABLES : collection(var−dvar)<br>FOUND : collection(var−dvar) |

**Restrictions**

$\text{sv} \leq 1 \Rightarrow \text{VALUE} = \text{sv} + 1$
$\text{rv} = 1 \Rightarrow \text{VALUE} \geq \text{sv}$
$\text{rv} \geq 2 \Rightarrow \text{VALUE} \geq 2$
$\text{VALUE} = \text{sv} + 1 \lor \text{VALUE} \leq \text{sv}①$
required(VARIABLES, var)
required(FOUND, var)
where
$\text{rv} = \text{range}(\text{VARIABLES.var})$
$\text{sv} = |\text{VARIABLES}|$

**Purpose**

The constraint MIN_WIDTH_STEADY_SEQUENCE(VALUE, VARIABLES) holds. In addition, FOUND is a collection of 0/1 variables where the value 1 indicates the position of the **found** letter in those occurrences of the pattern STEADY_SEQUENCE for which the feature value is VALUE.

The position of the **found** letter in an occurrence of a pattern is the first position where the occurrence of pattern is identified, even if the pattern is not complete.

An occurrence of the pattern STEADY_SEQUENCE is the *maximal* subsequence which matches the regular expression $=^+$.

Assume that the occurrence of the pattern STEADY_SEQUENCE starts at position $i$ and ends at position $j$. The feature WIDTH computes the value $j - i + 2$.

**Example**

$$\left( \begin{array}{l} 2, \langle 3, 1, 1, 4, 5, 5, 5, 6, 2, 2, 4, 4, 3, 2, 1, 1 \rangle, \\ \langle 0, 1, 0, 0, 0, 0, 0, 0, 1, 0, 1, 0, 0, 0, 1, 0 \rangle \end{array} \right)$$

Figure 3.1109 provides an example where the POS_MIN_WIDTH_STEADY_SEQUENCE $(2, [3, 1, 1, 4, 5, 5, 5, 6, 2, 2, 4, 4, 3, 2, 1, 1], [0, 1, 0, 0, 0, 0, 0, 0, 1, 0, 1, 0, 0, 0, 1, 0])$ constraint holds.

| | |
|---|---|
| **Typical** | $|\text{VARIABLES}| > 1$ |
| **Arg. properties** | • Functional dependency: VALUE determined by VARIABLES.<br>• Functional dependency: FOUND determined by VARIABLES. |

Figure 3.1109: Illustrating the POS_MIN_WIDTH_STEADY_SEQUENCE constraint of the **Example** slot

**Automaton**        Similar to the automaton of the MIN_WIDTH_STEADY_SEQUENCE constraint but use the decoration table 2.33.

# 3.545   POS_MIN_WIDTH_STRICTLY_DECREASING_SEQUENCE

► ▷ ◁ ◀     **DESCRIPTION**          **AUTOMATON**

| | |
|---|---|
| **Origin** | Based on constraint MIN_WIDTH_STRICTLY_DECREASING_SEQUENCE. |
| **Constraint** | POS_MIN_WIDTH_STRICTLY_DECREASING_SEQUENCE(VALUE, VARIABLES, FOUND) |
| **Arguments** | VALUE      : dvar<br>VARIABLES : collection(var−dvar)<br>FOUND      : collection(var−dvar) |
| **Restrictions** | $\text{sv} \leq 1 \lor \text{rv} \leq 1 \Rightarrow \text{VALUE} = \text{sv} + 1$<br>$\text{VALUE} \geq 2$<br>$\text{VALUE} = \text{sv} + 1 \lor \text{VALUE} \leq \min(\text{sv}, \text{rv}①)$<br>required(VARIABLES, var)<br>required(FOUND, var)<br>where<br>$\text{sv} = |\text{VARIABLES}|$<br>$\text{rv} = \text{range}(\text{VARIABLES.var})$ |

| | |
|---|---|
| **Purpose** | The constraint MIN_WIDTH_STRICTLY_DECREASING_SEQUENCE(VALUE, VARIABLES) holds. In addition, FOUND is a collection of 0/1 variables where the value 1 indicates the position of the **found** letter in those occurrences of the pattern STRICTLY_DECREASING_SEQUENCE for which the feature value is VALUE.<br>The position of the **found** letter in an occurrence of a pattern is the first position where the occurrence of pattern is identified, even if the pattern is not complete.<br>An occurrence of the pattern STRICTLY_DECREASING_SEQUENCE is the *maximal* sub-sequence which matches the regular expression $>^{+}$.<br>Assume that the occurrence of the pattern STRICTLY_DECREASING_SEQUENCE starts at position $i$ and ends at position $j$. The feature WIDTH computes the value $j - i + 2$. |
| **Example** | $\left( \begin{array}{l} 2, \langle 4, 4, 6, 4, 1, 1, 3, 4, 4, 6, 6, 5, 2, 2, 4, 3 \rangle, \\ \langle 0, 0, 0, 0, 0, 0, 0, 0, 0, 0, 0, 0, 0, 0, 1, 0 \rangle \end{array} \right)$ |

Figure 3.1110 provides an example where the POS_MIN_WIDTH_STRICTLY_DECREASING_SEQUENCE $(2, [4, 4, 6, 4, 1, 1, 3, 4, 4, 6, 6, 5, 2, 2, 4, 3], [0, 0, 0, 0, 0, 0, 0, 0, 0, 0, 0, 0, 0, 0, 1, 0])$ constraint holds.

| | |
|---|---|
| **Typical** | $|\text{VARIABLES}| > 1$<br>$\text{range}(\text{VARIABLES.var}) > 1$ |

| | |
|---|---|
| **Arg. properties** | • Functional dependency: VALUE determined by VARIABLES.<br>• Functional dependency: FOUND determined by VARIABLES. |

Figure 3.1110: Illustrating the POS_MIN_WIDTH_STRICTLY_DECREASING_SEQUENCE constraint of the **Example** slot

**Automaton**     Similar to the automaton of the MIN_WIDTH_STRICTLY_DECREASING_SEQUENCE constraint but use the decoration table 2.33.

## 3.546　POS_MIN_WIDTH_STRICTLY_INCREASING_SEQUENCE

▶ ▷ ◁ ◀　　　DESCRIPTION　　　AUTOMATON

**Origin**　Based on constraint MIN_WIDTH_STRICTLY_INCREASING_SEQUENCE.

**Constraint**　POS_MIN_WIDTH_STRICTLY_INCREASING_SEQUENCE(VALUE, VARIABLES, FOUND)

**Arguments**
```
VALUE      : dvar
VARIABLES  : collection(var−dvar)
FOUND      : collection(var−dvar)
```

**Restrictions**

$\text{sv} \leq 1 \vee \text{rv} \leq 1 \Rightarrow \text{VALUE} = \text{sv} + 1$
$\text{VALUE} \geq 2$
$\text{VALUE} = \text{sv} + 1 \vee \text{VALUE} \leq \min(\text{sv}, \text{rv}①)$
required(VARIABLES, var)
required(FOUND, var)
where
$\text{sv} = |\text{VARIABLES}|$
$\text{rv} = \text{range}(\text{VARIABLES.var})$

**Purpose**

The constraint MIN_WIDTH_STRICTLY_INCREASING_SEQUENCE(VALUE, VARIABLES) holds. In addition, FOUND is a collection of 0/1 variables where the value 1 indicates the position of the **found** letter in those occurrences of the pattern STRICTLY_INCREASING_SEQUENCE for which the feature value is VALUE.

The position of the **found** letter in an occurrence of a pattern is the first position where the occurrence of pattern is identified, even if the pattern is not complete.

An occurrence of the pattern STRICTLY_INCREASING_SEQUENCE is the *maximal* subsequence which matches the regular expression $<^+$.

Assume that the occurrence of the pattern STRICTLY_INCREASING_SEQUENCE starts at position $i$ and ends at position $j$. The feature WIDTH computes the value $j - i + 2$.

**Example**

$$\left( \begin{array}{l} 2, \langle 4, 3, 5, 5, 2, 1, 1, 2, 3, 4, 6, 6, 3, 1, 2, 3 \rangle, \\ \langle 0, 1, 0, 0, 0, 0, 0, 0, 0, 0, 0, 0, 0, 0, 0, 0 \rangle \end{array} \right)$$

Figure 3.1111 provides an example where the POS_MIN_WIDTH_STRICTLY_INCREASING_SEQUENCE $(2, [4, 3, 5, 5, 2, 1, 1, 2, 3, 4, 6, 6, 3, 1, 2, 3], [0, 1, 0, 0, 0, 0, 0, 0, 0, 0, 0, 0, 0, 0, 0, 0])$ constraint holds.

**Typical**

$|\text{VARIABLES}| > 1$
$\text{range}(\text{VARIABLES.var}) > 1$

**Arg. properties**

- Functional dependency: VALUE determined by VARIABLES.
- Functional dependency: FOUND determined by VARIABLES.

Figure 3.1111: Illustrating the POS_MIN_WIDTH_STRICTLY_INCREASING_SEQUENCE constraint of the **Example** slot

**Automaton** Similar to the automaton of the MIN_WIDTH_STRICTLY_INCREASING_SEQUENCE constraint but use the decoration table 2.33.

## 3.547  POS_MIN_WIDTH_SUMMIT

$(< | < (= | <)^* <)(> | > (= | >)^* >)$

| | |
|---|---|
| **Origin** | Based on constraint MIN_WIDTH_SUMMIT. |
| **Constraint** | POS_MIN_WIDTH_SUMMIT(VALUE, VARIABLES, FOUND) |

| **Arguments** | | |
|---|---|---|
| | VALUE | : dvar |
| | VARIABLES | : collection(var−dvar) |
| | FOUND | : collection(var−dvar) |

**Restrictions**

$sv \leq 2 \lor rv \leq 1 \Rightarrow VALUE = sv + 1$

$VALUE \geq 1$

$rv = 2 \Rightarrow VALUE = sv + 1 \lor VALUE \leq 1$①

$rv \geq 3 \Rightarrow VALUE = sv + 1 \lor VALUE \leq sv - 2$②

required(VARIABLES, var)

required(FOUND, var)

where

$sv = |VARIABLES|$

$rv = range(VARIABLES.var)$

**Purpose**

The constraint MIN_WIDTH_SUMMIT(VALUE, VARIABLES) holds. In addition, FOUND is a collection of 0/1 variables where the value 1 indicates the position of the **found** letter in those occurrences of the pattern SUMMIT for which the feature value is VALUE.

The position of the **found** letter in an occurrence of a pattern is the first position where the occurrence of pattern is identified, even if the pattern is not complete.

An occurrence of the pattern SUMMIT is the *maximal* subsequence which matches the regular expression $(< | < (= | <)^* <)(> | > (= | >)^* >)$.

Assume that the occurrence of the pattern SUMMIT starts at position $i$ and ends at position $j$. The feature WIDTH computes the value $j − i$.

**Example**

$$\left( \begin{array}{l} 1, \langle 7, 1, 5, 4, 4, 3, 3, 4, 6, 6, 2, 3, 4, 2, 3, 1 \rangle, \\ \langle 0, 0, 0, 0, 0, 0, 0, 0, 0, 0, 0, 0, 0, 0, 1, 0 \rangle \end{array} \right)$$

Figure 3.1112 provides an example where the POS_MIN_WIDTH_SUMMIT $(1, [7, 1, 5, 4, 4, 3, 3, 4, 6, 6, 2, 3, 4, 2, 3, 1], [0, 0, 0, 0, 0, 0, 0, 0, 0, 0, 0, 0, 0, 0, 1, 0])$ constraint holds.

**Typical**

$|VARIABLES| > 2$

$range(VARIABLES.var) > 1$

**Arg. properties**

- Functional dependency: VALUE determined by VARIABLES.
- Functional dependency: FOUND determined by VARIABLES.

Figure 3.1112: Illustrating the POS_MIN_WIDTH_SUMMIT constraint of the **Example** slot

**Automaton**          Similar to the automaton of the MIN_WIDTH_SUMMIT constraint but use the decoration table 2.33.

## 3.548 POS_MIN_WIDTH_VALLEY

► ▷ ◁ ◀          **DESCRIPTION**          **AUTOMATON**

> $> (= | >)^* (< | =)^* <$

| | |
|---|---|
| **Origin** | Based on constraint MIN_WIDTH_VALLEY. |
| **Constraint** | POS_MIN_WIDTH_VALLEY(VALUE, VARIABLES, FOUND) |

**Arguments**

```
VALUE      : dvar
VARIABLES  : collection(var−dvar)
FOUND      : collection(var−dvar)
```

**Restrictions**

$$\mathtt{sv} \leq 2 \vee \mathtt{rv} \leq 1 \Rightarrow \mathtt{VALUE} = \mathtt{sv} + 1$$
$$\mathtt{VALUE} \geq 1$$
$$\mathtt{VALUE} = \mathtt{sv} + 1 \vee \mathtt{VALUE} \leq \mathtt{sv} - 2①$$
$$\mathtt{required}(\mathtt{VARIABLES}, \mathtt{var})$$
$$\mathtt{required}(\mathtt{FOUND}, \mathtt{var})$$

where

$$\mathtt{sv} = |\mathtt{VARIABLES}|$$
$$\mathtt{rv} = \mathtt{range}(\mathtt{VARIABLES.var})$$

VALUE $\leq$ **4** $= 6 - 2$

**Purpose**

The constraint MIN_WIDTH_VALLEY(VALUE, VARIABLES) holds. In addition, FOUND is a collection of 0/1 variables where the value 1 indicates the position of the **found** letter in those occurrences of the pattern VALLEY for which the feature value is VALUE.

The position of the **found** letter in an occurrence of a pattern is the first position where the occurrence of pattern is identified, even if the pattern is not complete.

An occurrence of the pattern VALLEY is the *maximal* subsequence which matches the regular expression $> (= | >)^* (< | =)^* <$.

Assume that the occurrence of the pattern VALLEY starts at position $i$ and ends at position $j$. The feature WIDTH computes the value $j - i$.

**Example**

$$\left( \begin{array}{l} 2, \langle 1, 3, 7, 4, 3, 6, 6, 5, 3, 3, 2, 6, 5, 5, 5, 7 \rangle, \\ \langle 0, 0, 0, 0, 1, 0, 0, 0, 0, 0, 0, 0, 0, 0, 0, 0 \rangle \end{array} \right)$$

Figure 3.1113 provides an example where the POS_MIN_WIDTH_VALLEY $(2, [1, 3, 7, 4, 3, 6, 6, 5, 3, 3, 2, 6, 5, 5, 5, 7], [0, 0, 0, 0, 1, 0, 0, 0, 0, 0, 0, 0, 0, 0, 0, 0])$ constraint holds.

**Typical**

$$|\mathtt{VARIABLES}| > 2$$
$$\mathtt{range}(\mathtt{VARIABLES.var}) > 1$$

**Arg. properties**

- Functional dependency: VALUE determined by VARIABLES.
- Functional dependency: FOUND determined by VARIABLES.

Figure 3.1113: Illustrating the POS_MIN_WIDTH_VALLEY constraint of the **Example** slot

**Automaton**  Similar to the automaton of the MIN_WIDTH_VALLEY constraint but use the decoration table 2.33.

## 3.549 POS_MIN_WIDTH_ZIGZAG

▶ ▷ ◁ ◀      **DESCRIPTION**          **AUTOMATON**

$$(<>)^+(<\mid<>)\mid(><)^+(>\mid><)$$

**Origin**          Based on constraint MIN_WIDTH_ZIGZAG.

**Constraint**      POS_MIN_WIDTH_ZIGZAG(VALUE, VARIABLES, FOUND)

**Arguments**       
```
VALUE     : dvar
VARIABLES : collection(var−dvar)
FOUND     : collection(var−dvar)
```

**Restrictions**    $\mathtt{sv} \leq 3 \vee \mathtt{rv} \leq 1 \Rightarrow \mathtt{VALUE} = \mathtt{sv} + 1$

$\mathtt{VALUE} \geq 2$

$\mathtt{VALUE} = \mathtt{sv} + 1 \vee \mathtt{VALUE} \leq \mathtt{sv} - 2①$

required(VARIABLES, var)

required(FOUND, var)

where

$\mathtt{sv} = |\mathtt{VARIABLES}|$

$\mathtt{rv} = \mathtt{range}(\mathtt{VARIABLES.var})$

**Purpose**         The constraint MIN_WIDTH_ZIGZAG(VALUE, VARIABLES) holds. In addition, FOUND is a collection of 0/1 variables where the value 1 indicates the position of the **found** letter in those occurrences of the pattern ZIGZAG for which the feature value is VALUE.

The position of the **found** letter in an occurrence of a pattern is the first position where the occurrence of pattern is identified, even if the pattern is not complete.

An occurrence of the pattern ZIGZAG is the *maximal* subsequence which matches the regular expression $(<>)^+(<\mid<>)\mid(><)^+(>\mid><)$.

Assume that the occurrence of the pattern ZIGZAG starts at position $i$ and ends at position $j$. The feature WIDTH computes the value $j - i$.

**Example**         
$$\left(\begin{array}{l} 2, \langle 4, 1, 3, 1, 4, 6, 1, 5, 5, 2, 7, 2, 3, 1, 6, 1\rangle\,, \\ \langle 0, 0, 0, 0, 0, 0, 1, 0, 0, 0, 0, 0, 0, 0, 0, 0\rangle \end{array}\right)$$

Figure 3.1114 provides an example where the POS_MIN_WIDTH_ZIGZAG $(2, [4, 1, 3, 1, 4, 6, 1, 5, 5, 2, 7, 2, 3, 1, 6, 1], [0, 0, 0, 0, 0, 0, 1, 0, 0, 0, 0, 0, 0, 0, 0, 0])$ constraint holds.

**Typical**         $|\mathtt{VARIABLES}| > 3$

range(VARIABLES.var) $> 1$

**Arg. properties** 
- Functional dependency: VALUE determined by VARIABLES.
- Functional dependency: FOUND determined by VARIABLES.

Figure 3.1114: Illustrating the POS_MIN_WIDTH_ZIGZAG constraint of the **Example** slot

**Automaton**          Similar to the automaton of the MIN_WIDTH_ZIGZAG constraint but use the decoration table 2.33.

## 3.550  SUM_HEIGHT_DECREASING_TERRACE

▶   ◁ ◀

**Origin**          Based on the DECREASING_TERRACE pattern.

**Constraint**      SUM_HEIGHT_DECREASING_TERRACE(VALUE, VARIABLES)

**Arguments**       VALUE     : dvar
                    VARIABLES : collection(var−dvar)

**Restrictions**

$$sv \leq 3 \vee rv \leq 2 \Rightarrow VALUE = 0$$

$$VALUE \geq \min_{q \in [\texttt{lb1}, \texttt{ub1}]} \min \left( 0, \sum \left( \begin{array}{c} \prod \left( \begin{array}{c} \min\left(\ 1, \max\left(\ 0, rv - 2\ \right)\ \right), \\ q, \\ \sum \left( \begin{array}{c} np1 * minv, \\ \lfloor (np1 + 1) * np1/2 \rfloor \end{array} \right) \end{array} \right), \\ \prod \left( \begin{array}{c} \min \left( 1, \max \left( 0, \left\lfloor \left( \begin{array}{c} \min\left( \begin{array}{c} rv, \\ |minv| + 1 \end{array} \right) - \\ \left( \max\left(\ 1, sv - sv \bmod 2\ \right) - \right) / \\ 2 * q \end{array} \right) \right\rfloor - 2 \right) \right), \\ \lfloor \left( \begin{array}{c} \max\left(\ 1, sv - sv \bmod 2\ \right) - \\ 2 * q \\ q * 2 * np1 \end{array} \right) /2 \rfloor, \\ minv + np1 + 1 \end{array} \right) \end{array} \right) \right),$$

$$VALUE \leq \max_{q \in [\texttt{lb2}, \texttt{ub2}]} \max \left( 0, \sum \left( \begin{array}{c} \prod \left( \begin{array}{c} \min\left(\ 1, \max\left(\ 0, rv - 2\ \right)\ \right), \\ q, \\ np2 * maxv - \\ \lfloor (np2 + 1) * np2/2 \rfloor \end{array} \right), \\ \prod \left( \begin{array}{c} \min \left( 1, \max \left( 0, \left\lfloor \left( \begin{array}{c} \min\left( \begin{array}{c} rv, \\ |maxv| + 1 \end{array} \right) - \\ \left( \max\left(\ 1, sv - sv \bmod 2\ \right) - \right) / \\ 2 * q \end{array} \right) \right\rfloor - 2 \right) \right), \\ \lfloor \left( \begin{array}{c} \max\left(\ 1, sv - sv \bmod 2\ \right) - \\ 2 * q \\ q * 2 * np2 \end{array} \right) /2 \rfloor, \\ maxv - np2 - 1 \end{array} \right) \end{array} \right) \right),$$

required(VARIABLES, var)

where

$sv = |VARIABLES|$

$rv = range(VARIABLES.var)$

$$np1 = \max \left( 0, \sum \left( \begin{array}{c} \lfloor (\max(1, sv - sv \bmod 2) - 2 * q)/2 * q \rfloor, \\ \min \left( 0, \begin{array}{c} \min(rv, |minv| + 1) - \\ \lfloor (\max(1, sv - sv \bmod 2) - 2 * q)/2 * q \rfloor \end{array} - 2 \right) \end{array} \right) \right)$$

$$np2 = \max \left( 0, \sum \left( \begin{array}{c} \lfloor (\max(1, sv - sv \bmod 2) - 2 * q)/2 * q \rfloor, \\ \min \left( 0, \begin{array}{c} \min(rv, |maxv| + 1) - \\ \lfloor (\max(1, sv - sv \bmod 2) - 2 * q)/2 * q \rfloor \end{array} - 2 \right) \end{array} \right) \right)$$

$maxv = maxval(VARIABLES.var)$

$minv = minval(VARIABLES.var)$

$$lb1 = \min \left( \sum \left( \begin{array}{c} \lfloor sv/4 \rfloor + 1, \\ \left\lfloor \min \left( \begin{array}{c} \max(1, sv - sv \bmod 2)/ \\ \max(1, sv - sv \bmod 2), \\ \max(1, (\min(|minv| + 1, rv) - 2) * 2 + 2) \end{array} \right) \right\rfloor, \\ \max \left( 0, \min \left( 1, \min \left( \begin{array}{c} \max(1, sv - sv \bmod 2) \bmod \\ \max(1, sv - sv \bmod 2), \\ \max(2, (\min(|minv| + 1, rv) - 2) * 2 + 2) \end{array} \right) - 3 \right) \right) \end{array} \right) \right)$$

$ub1 = \lfloor sv/4 \rfloor + 1$

$$lb2 = \min \left( \sum \left( \begin{array}{c} \lfloor sv/4 \rfloor + 1, \\ \left\lfloor \min \left( \begin{array}{c} \max(1, sv - sv \bmod 2)/ \\ \max(1, sv - sv \bmod 2), \\ \max(1, (\min(|maxv| + 1, rv) - 2) * 2 + 2) \end{array} \right) \right\rfloor, \\ \max \left( 0, \min \left( 1, \min \left( \begin{array}{c} \max(1, sv - sv \bmod 2) \bmod \\ \max(1, sv - sv \bmod 2), \\ \max(2, (\min(|maxv| + 1, rv) - 2) * 2 + 2) \end{array} \right) - 3 \right) \right) \end{array} \right) \right)$$

$ub2 = \lfloor sv/4 \rfloor + 1$

| Purpose | VALUE is the sum of all minimum values in each occurrence of the DECREAS-ING_TERRACE pattern in the time-series given by the VARIABLES collection. If the pattern does not occur, VALUE takes the default value 0.<br><br>An occurrence of the pattern DECREASING_TERRACE is the *maximal* subsequence which matches the regular expression $>=^+>$.<br><br>Assume that the occurrence of the pattern DECREASING_TERRACE starts at position $i$ and ends at position $j$. The feature MIN computes the minimum of the values from index $i+1$ to index $j$. |
|---|---|

| Example | $(6, \langle 6, 4, 4, 4, 5, 2, 2, 1, 3, 3, 5, 4, 4, 3, 3, 3 \rangle)$ |
|---|---|

Figure 3.1115 provides an example where the SUM_HEIGHT_DECREASING_TERRACE $(6, [6, 4, 4, 4, 5, 2, 2, 1, 3, 3, 5, 4, 4, 3, 3, 3])$ constraint holds.

Figure 3.1115: Illustrating the SUM_HEIGHT_DECREASING_TERRACE constraint of the **Example** slot

| Typical | $|\texttt{VARIABLES}| > 3$<br>range(VARIABLES.var) > 2 |
|---|---|

| Arg. properties | Functional dependency: VALUE determined by VARIABLES. |
|---|---|

**Automaton**    Figures 3.1116 and 3.1117 respectively depict the automaton associated with the constraint SUM_HEIGHT_DECREASING_TERRACE and its simplified form.

Figure 3.1116: Automaton for the SUM_HEIGHT_DECREASING_TERRACE constraint obtained by applying decoration Table 2.35 to the seed transducer of the DECREAS­ING_TERRACE pattern where default is 0

Figure 3.1117: Automaton for the SUM_HEIGHT_DECREASING_TERRACE constraint obtained by applying decoration Table 2.37 to the seed transducer of the DECREAS­ING_TERRACE pattern where default is 0

|   | $s$ | $r$ | $t$ |
|---|---|---|---|
| $s$ | $\overrightarrow{C} + \overleftarrow{C}$ | $\overrightarrow{C} + \overleftarrow{C}$ | $\overrightarrow{C} + \overleftarrow{C}$ |
| $r$ | $\overrightarrow{C} + \overleftarrow{C}$ | $\overrightarrow{C} + \overleftarrow{C}$ | $\min(\overrightarrow{D}, \overleftarrow{D}, \mathtt{VAR}_i)$ |
| $t$ | $\overrightarrow{C} + \overleftarrow{C}$ | $\min(\overrightarrow{D}, \overleftarrow{D}, \mathtt{VAR}_i)$ | $\min(\overrightarrow{D}, \overleftarrow{D}, \mathtt{VAR}_i)$ |

Table 3.142: Glue matrix for the SUM_HEIGHT_DECREASING_TERRACE constraint defined as the composition of the DECREASING_TERRACE pattern , the feature MIN , and the aggregator sum ; cells of the glue matrix are coloured with the colour of the constituent to which they are related.

## 3.551 SUM_HEIGHT_INCREASING_TERRACE

**DESCRIPTION**          **AUTOMATON**

**Origin**          Based on the INCREASING_TERRACE pattern.

**Constraint**          SUM_HEIGHT_INCREASING_TERRACE(VALUE, VARIABLES)

**Arguments**          VALUE     : dvar
                       VARIABLES : collection(var−dvar)

**Restrictions**

$\mathtt{sv} \leq 3 \vee \mathtt{rv} \leq 2 \Rightarrow \mathtt{VALUE} = 0$

$$\mathtt{VALUE} \geq \min_{\mathtt{q} \in [\mathtt{lb1}, \mathtt{ub1}]} \min \left( 0, \sum \left( \begin{array}{l} \prod \left( \begin{array}{l} \min\left(\ 1, \max\left(\ 0, \mathtt{rv} - 2\ \right)\ \right), \\ \mathtt{q}, \\ \sum \left( \begin{array}{l} \mathtt{np1} * \mathtt{minv}, \\ \lfloor (\mathtt{np1} + 1) * \mathtt{np1}/2 \rfloor \end{array} \right) \end{array} \right), \\ \prod \left( \begin{array}{l} \min\left( 1, \max\left( 0, \left\lfloor \left( \begin{array}{l} \min\left( \begin{array}{l} \mathtt{rv}, \\ |\mathtt{minv}| + 1 \end{array} \right) - \\ \left( \begin{array}{l} \max\left(\ 1, \mathtt{sv} - \mathtt{sv} \bmod 2\ \right) - \\ 2 * \mathtt{q} \end{array} \right)/ \end{array} \right\rfloor - 2 \right) \right), \\ \lfloor \left( \begin{array}{l} \max\left(\ 1, \mathtt{sv} - \mathtt{sv} \bmod 2\ \right) - \\ \mathtt{q} * 2 * \mathtt{np1} \end{array} \right)/2 \rfloor, \\ \mathtt{minv} + \mathtt{np1} + 1 \end{array} \right) \end{array} \right) \right)$$

$$\mathtt{VALUE} \leq \max_{\mathtt{q} \in [\mathtt{lb2}, \mathtt{ub2}]} \max \left( 0, \sum \left( \begin{array}{l} \prod \left( \begin{array}{l} \min\left(\ 1, \max\left(\ 0, \mathtt{rv} - 2\ \right)\ \right), \\ \mathtt{q}, \\ \mathtt{np2} * \mathtt{maxv} - \\ \lfloor (\mathtt{np2} + 1) * \mathtt{np2}/2 \rfloor \end{array} \right), \\ \prod \left( \begin{array}{l} \min\left( 1, \max\left( 0, \left\lfloor \left( \begin{array}{l} \min\left( \begin{array}{l} \mathtt{rv}, \\ |\mathtt{maxv}| + 1 \end{array} \right) - \\ \left( \begin{array}{l} \max\left(\ 1, \mathtt{sv} - \mathtt{sv} \bmod 2\ \right) - \\ 2 * \mathtt{q} \end{array} \right)/ \end{array} \right\rfloor - 2 \right) \right), \\ \lfloor \left( \begin{array}{l} \max\left(\ 1, \mathtt{sv} - \mathtt{sv} \bmod 2\ \right) - \\ \mathtt{q} * 2 * \mathtt{np2} \end{array} \right)/2 \rfloor, \\ \mathtt{maxv} - \mathtt{np2} - 1 \end{array} \right) \end{array} \right) \right)$$

${\color{teal}\texttt{required}}(\mathtt{VARIABLES}, \mathtt{var})$

where

$\mathtt{sv} = |\mathtt{VARIABLES}|$

$\mathtt{rv} = {\color{teal}\texttt{range}}(\mathtt{VARIABLES}.\mathtt{var})$

$\mathtt{np1} = \max \left( 0, \sum \left( \begin{array}{l} \lfloor (\max(1, \mathtt{sv} - \mathtt{sv} \bmod 2) - 2 * \mathtt{q})/2 * \mathtt{q} \rfloor, \\ \min \left( 0, \begin{array}{l} \min(\mathtt{rv}, |\mathtt{minv}| + 1) - \\ \lfloor (\max(1, \mathtt{sv} - \mathtt{sv} \bmod 2) - 2 * \mathtt{q})/2 * \mathtt{q} \rfloor \end{array} - 2 \right) \end{array} \right) \right)$

$\mathtt{np2} = \max \left( 0, \sum \left( \begin{array}{l} \lfloor (\max(1, \mathtt{sv} - \mathtt{sv} \bmod 2) - 2 * \mathtt{q})/2 * \mathtt{q} \rfloor, \\ \min \left( 0, \begin{array}{l} \min(\mathtt{rv}, |\mathtt{maxv}| + 1) - \\ \lfloor (\max(1, \mathtt{sv} - \mathtt{sv} \bmod 2) - 2 * \mathtt{q})/2 * \mathtt{q} \rfloor \end{array} - 2 \right) \end{array} \right) \right)$

$\mathtt{maxv} = {\color{teal}\texttt{maxval}}(\mathtt{VARIABLES}.\mathtt{var})$

$\mathtt{minv} = {\color{teal}\texttt{minval}}(\mathtt{VARIABLES}.\mathtt{var})$

$\mathtt{lb1} = \min \left( \sum \left( \begin{array}{l} \lfloor \mathtt{sv}/4 \rfloor + 1, \\ \left\lfloor \min\left( \begin{array}{l} \max(1, \mathtt{sv} - \mathtt{sv} \bmod 2)/ \\ \max(1, \mathtt{sv} - \mathtt{sv} \bmod 2), \\ \max(1, (\min(|\mathtt{minv}| + 1, \mathtt{rv}) - 2) * 2 + 2) \end{array} \right) \right\rfloor, \\ \max\left( 0, \min\left( 1, \min\left( \begin{array}{l} \max(1, \mathtt{sv} - \mathtt{sv} \bmod 2) \bmod \\ \max(1, \mathtt{sv} - \mathtt{sv} \bmod 2), \\ \max(2, (\min(|\mathtt{minv}| + 1, \mathtt{rv}) - 2) * 2 + 2) \end{array} \right) - 3 \right) \right) \end{array} \right) \right)$

$\mathtt{ub1} = \lfloor \mathtt{sv}/4 \rfloor + 1$

$\mathtt{lb2} = \min \left( \sum \left( \begin{array}{l} \lfloor \mathtt{sv}/4 \rfloor + 1, \\ \left\lfloor \min\left( \begin{array}{l} \max(1, \mathtt{sv} - \mathtt{sv} \bmod 2)/ \\ \max(1, \mathtt{sv} - \mathtt{sv} \bmod 2), \\ \max(1, (\min(|\mathtt{maxv}| + 1, \mathtt{rv}) - 2) * 2 + 2) \end{array} \right) \right\rfloor, \\ \max\left( 0, \min\left( 1, \min\left( \begin{array}{l} \max(1, \mathtt{sv} - \mathtt{sv} \bmod 2) \bmod \\ \max(1, \mathtt{sv} - \mathtt{sv} \bmod 2), \\ \max(2, (\min(|\mathtt{maxv}| + 1, \mathtt{rv}) - 2) * 2 + 2) \end{array} \right) - 3 \right) \right) \end{array} \right) \right)$

$\mathtt{ub2} = \lfloor \mathtt{sv}/4 \rfloor + 1$

**Purpose**

VALUE is the sum of all minimum values in each occurrence of the INCREAS­ING_TERRACE pattern in the time-series given by the VARIABLES collection. If the pattern does not occur, VALUE takes the default value 0.

An occurrence of the pattern INCREASING_TERRACE is the *maximal* subsequence which matches the regular expression $< =^+ <$.

Assume that the occurrence of the pattern INCREASING_TERRACE starts at position $i$ and ends at position $j$. The feature MIN computes the minimum of the values from index $i + 1$ to index $j$.

**Example**

$(8, \langle 1, 3, 3, 3, 2, 5, 5, 6, 4, 4, 2, 3, 3, 3, 4, 4 \rangle)$

Figure 3.1118 provides an example where the SUM_HEIGHT_INCREASING_TERRACE $(8, [1, 3, 3, 3, 2, 5, 5, 6, 4, 4, 2, 3, 3, 3, 4, 4])$ constraint holds.

Figure 3.1118: Illustrating the SUM_HEIGHT_INCREASING_TERRACE constraint of the **Example** slot

**Typical**

$|\text{VARIABLES}| > 3$

$\text{range}(\text{VARIABLES.var}) > 2$

**Arg. properties**

Functional dependency: VALUE determined by VARIABLES.

**Automaton** Figures 3.1119 and 3.1120 respectively depict the automaton associated with the constraint SUM_HEIGHT_INCREASING_TERRACE and its simplified form.

Figure 3.1119: Automaton for the SUM_HEIGHT_INCREASING_TERRACE constraint obtained by applying decoration Table 2.35 to the seed transducer of the INCREAS­ING_TERRACE pattern where default is 0

Figure 3.1120: Automaton for the SUM_HEIGHT_INCREASING_TERRACE constraint obtained by applying decoration Table 2.37 to the seed transducer of the INCREAS­ING_TERRACE pattern where default is 0

|   | $s$ | $r$ | $t$ |
|---|---|---|---|
| $s$ | $\overrightarrow{C} + \overleftarrow{C}$ | $\overrightarrow{C} + \overleftarrow{C}$ | $\overrightarrow{C} + \overleftarrow{C}$ |
| $r$ | $\overrightarrow{C} + \overleftarrow{C}$ | $\overrightarrow{C} + \overleftarrow{C}$ | $\min(\overrightarrow{D}, \overleftarrow{D}, \text{VAR}_i)$ |
| $t$ | $\overrightarrow{C} + \overleftarrow{C}$ | $\min(\overrightarrow{D}, \overleftarrow{D}, \text{VAR}_i)$ | $\min(\overrightarrow{D}, \overleftarrow{D}, \text{VAR}_i)$ |

Table 3.143: Glue matrix for the SUM_HEIGHT_INCREASING_TERRACE constraint defined as the composition of the INCREASING_TERRACE pattern, the feature MIN, and the aggregator sum; cells of the glue matrix are coloured with the colour of the constituent to which they are related.

# 3.552 SUM_HEIGHT_PLAIN

► ◁ ◀   DESCRIPTION   AUTOMATON

| | |
|---|---|
| **Origin** | Based on the PLAIN pattern. |
| **Constraint** | SUM_HEIGHT_PLAIN(VALUE, VARIABLES) |
| **Arguments** | VALUE       : dvar<br>VARIABLES  : collection(var−dvar) |

**Restrictions**

$\mathtt{sv} \leq 2 \vee \mathtt{rv} \leq 1 \Rightarrow \mathtt{VALUE} = 0$

$\mathtt{VALUE} = 0 \vee \mathtt{VALUE} \geq \min(\mathtt{minv}①, \mathtt{minv} * \mathtt{np}②)$

$\mathtt{VALUE} = 0 \vee \mathtt{VALUE} \leq \max(\mathtt{maxv} - 1, (\mathtt{maxv} - 1) * \mathtt{np})$

required(VARIABLES, var)

where

$\mathtt{sv} = |\mathtt{VARIABLES}|$

$\mathtt{np} = \max(0, \lfloor (\mathtt{sv} - 1)/2 \rfloor)$

$\mathtt{minv} = \mathtt{minval}(\mathtt{VARIABLES.var})$

$\mathtt{maxv} = \mathtt{maxval}(\mathtt{VARIABLES.var})$

$\mathtt{rv} = \mathtt{range}(\mathtt{VARIABLES.var})$

**Purpose**

VALUE is the sum of all minimum values in each occurrence of the PLAIN pattern in the time-series given by the VARIABLES collection. If the pattern does not occur, VALUE takes the default value 0.

An occurrence of the pattern PLAIN is the *maximal* subsequence which matches the regular expression > =* <.

Assume that the occurrence of the pattern PLAIN starts at position $i$ and ends at position $j$. The feature MIN computes the minimum of the values from index $i + 1$ to index $j$.

| | |
|---|---|
| **Example** | $(12, \langle 2, 3, 6, 5, 7, 6, 6, 4, 5, 5, 4, 3, 3, 6, 6, 3 \rangle)$ |

Figure 3.1121 provides an example where the SUM_HEIGHT_PLAIN $(12, [2, 3, 6, 5, 7, 6, 6, 4, 5, 5, 4, 3, 3, 6, 6, 3])$ constraint holds.

| | |
|---|---|
| **Typical** | $|\mathtt{VARIABLES}| > 2$<br>range(VARIABLES.var) > 1 |

| | |
|---|---|
| **Symmetry** | Items of VARIABLES can be reversed. |
| **Arg. properties** | Functional dependency: VALUE determined by VARIABLES. |

Figure 3.1121: Illustrating the SUM_HEIGHT_PLAIN constraint of the **Example** slot

**Automaton**     Figures 3.1122 and 3.1123 respectively depict the automaton associated with the constraint SUM_HEIGHT_PLAIN and its simplified form.

Figure 3.1122: Automaton for the SUM_HEIGHT_PLAIN constraint obtained by applying decoration Table 2.35 to the seed transducer of the PLAIN pattern where default is 0

Figure 3.1123: Automaton for the SUM_HEIGHT_PLAIN constraint obtained by applying decoration Table 2.37 to the seed transducer of the PLAIN pattern where default is 0

|   | $s$ | $r$ | $t$ |
|---|---|---|---|
| $s$ | $\overrightarrow{C} + \overleftarrow{C}$ | $\overrightarrow{C} + \overleftarrow{C}$ | $\overrightarrow{C} + \overleftarrow{C}$ |
| $r$ | $\overrightarrow{C} + \overleftarrow{C}$ | $\min(\overrightarrow{D}, \overleftarrow{D}, \text{VAR}_i)$ | $\min(\overrightarrow{D}, \overleftarrow{D}, \text{VAR}_i)$ |
| $t$ | $\overrightarrow{C} + \overleftarrow{C}$ | $\min(\overrightarrow{D}, \overleftarrow{D}, \text{VAR}_i)$ | $\min(\overrightarrow{D}, \overleftarrow{D}, \text{VAR}_i)$ |

Table 3.144: Glue matrix for the SUM_HEIGHT_PLAIN constraint defined as the composition of the PLAIN pattern, the feature MIN, and the aggregator sum; cells of the glue matrix are coloured with the colour of the constituent to which they are related.

## 3.553   SUM_HEIGHT_PLATEAU

    DESCRIPTION        AUTOMATON

| | |
|---|---|
| **Origin** | Based on the PLATEAU pattern. |
| **Constraint** | SUM_HEIGHT_PLATEAU(VALUE, VARIABLES) |
| **Arguments** | VALUE       : dvar<br>VARIABLES : collection(var−dvar) |

**Restrictions**

$\text{sv} \leq 2 \vee \text{rv} \leq 1 \Rightarrow \text{VALUE} = 0$

$\text{VALUE} = 0 \vee \text{VALUE} \geq \min(\text{minv} + 1①, (\text{minv} + 1) * \text{np}②)$

$\text{VALUE} = 0 \vee \text{VALUE} \leq \max(\text{maxv}, \text{maxv} * \text{np})$

required(VARIABLES, var)

where

$\text{sv} = |\text{VARIABLES}|$

$\text{np} = \max(0, \lfloor (\text{sv} - 1)/2 \rfloor)$

$\text{minv} = \text{minval}(\text{VARIABLES.var})$

$\text{maxv} = \text{maxval}(\text{VARIABLES.var})$

$\text{rv} = \text{range}(\text{VARIABLES.var})$

**Purpose**

VALUE is the sum of all minimum values in each occurrence of the PLATEAU pattern in the time-series given by the VARIABLES collection. If the pattern does not occur, VALUE takes the default value 0.

An occurrence of the pattern PLATEAU is the *maximal* subsequence which matches the regular expression $< =^* >$.

Assume that the occurrence of the pattern PLATEAU starts at position $i$ and ends at position $j$. The feature MIN computes the minimum of the values from index $i + 1$ to index $j$.

**Example**

$(12, \langle 7, 5, 2, 3, 1, 2, 2, 4, 3, 3, 4, 5, 5, 2, 2, 5 \rangle)$

Figure 3.1124 provides an example where the SUM_HEIGHT_PLATEAU $(12, [7, 5, 2, 3, 1, 2, 2, 4, 3, 3, 4, 5, 5, 2, 2, 5])$ constraint holds.

| | |
|---|---|
| **Typical** | $|\text{VARIABLES}| > 2$<br>range(VARIABLES.var) > 1 |
| **Symmetry** | Items of VARIABLES can be reversed. |
| **Arg. properties** | Functional dependency: VALUE determined by VARIABLES. |

Figure 3.1124: Illustrating the SUM_HEIGHT_PLATEAU constraint of the **Example** slot

**Automaton**     Figures 3.1125 and 3.1126 respectively depict the automaton associated with the constraint SUM_HEIGHT_PLATEAU and its simplified form.

Figure 3.1125: Automaton for the SUM_HEIGHT_PLATEAU constraint obtained by applying decoration Table 2.35 to the seed transducer of the PLATEAU pattern where default is 0

Figure 3.1126: Automaton for the SUM_HEIGHT_PLATEAU constraint obtained by applying decoration Table 2.37 to the seed transducer of the PLATEAU pattern where default is 0

|   | $s$ | $r$ | $t$ |
|---|-----|-----|-----|
| $s$ | $\overrightarrow{C} + \overleftarrow{C}$ | $\overrightarrow{C} + \overleftarrow{C}$ | $\overrightarrow{C} + \overleftarrow{C}$ |
| $r$ | $\overrightarrow{C} + \overleftarrow{C}$ | $\min(\overrightarrow{D}, \overleftarrow{D}, \mathtt{VAR}_i)$ | $\min(\overrightarrow{D}, \overleftarrow{D}, \mathtt{VAR}_i)$ |
| $t$ | $\overrightarrow{C} + \overleftarrow{C}$ | $\min(\overrightarrow{D}, \overleftarrow{D}, \mathtt{VAR}_i)$ | $\min(\overrightarrow{D}, \overleftarrow{D}, \mathtt{VAR}_i)$ |

Table 3.145: Glue matrix for the SUM_HEIGHT_PLATEAU constraint defined as the composition of the PLATEAU pattern , the feature MIN , and the aggregator sum ; cells of the glue matrix are coloured with the colour of the constituent to which they are related.

## 3.554  SUM_HEIGHT_PROPER_PLAIN

▶    ◁ ◀        DESCRIPTION        AUTOMATON

| | |
|---|---|
| **Origin** | Based on the PROPER_PLAIN pattern. |
| **Constraint** | SUM_HEIGHT_PROPER_PLAIN(VALUE, VARIABLES) |
| **Arguments** | VALUE     : dvar |
| | VARIABLES : collection(var−dvar) |

**Restrictions**

$\mathtt{sv} \le 3 \lor \mathtt{rv} \le 1 \Rightarrow \mathtt{VALUE} = 0$
$\mathtt{VALUE} = 0 \lor \mathtt{VALUE} \ge \min(\mathtt{minv}①, \mathtt{minv} * \mathtt{np}②)$
$\mathtt{VALUE} = 0 \lor \mathtt{VALUE} \le \max(\mathtt{maxv} - 1, (\mathtt{maxv} - 1) * \mathtt{np})$
required(VARIABLES, var)
where
$\mathtt{sv} = |\mathtt{VARIABLES}|$
$\mathtt{np} = \max(0, \lfloor (\mathtt{sv} - 1)/3 \rfloor)$
$\mathtt{minv} = \mathtt{minval}(\mathtt{VARIABLES.var})$
$\mathtt{maxv} = \mathtt{maxval}(\mathtt{VARIABLES.var})$
$\mathtt{rv} = \mathtt{range}(\mathtt{VARIABLES.var})$

**Purpose**

VALUE is the sum of all minimum values in each occurrence of the PROPER_PLAIN pattern in the time-series given by the VARIABLES collection. If the pattern does not occur, VALUE takes the default value 0.
An occurrence of the pattern PROPER_PLAIN is the *maximal* subsequence which matches the regular expression $>=^+<$.
Assume that the occurrence of the pattern PROPER_PLAIN starts at position $i$ and ends at position $j$. The feature MIN computes the minimum of the values from index $i+1$ to index $j$.

**Example**

$(12, \langle 2, 7, 5, 5, 6, 3, 7, 4, 4, 5, 6, 5, 3, 3, 3, 5 \rangle)$

Figure 3.1127 provides an example where the SUM_HEIGHT_PROPER_PLAIN $(12, [2, 7, 5, 5, 6, 3, 7, 4, 4, 5, 6, 5, 3, 3, 3, 5])$ constraint holds.

| | |
|---|---|
| **Typical** | $|\mathtt{VARIABLES}| > 3$ |
| | range(VARIABLES.var) $> 1$ |
| **Symmetry** | Items of VARIABLES can be reversed. |
| **Arg. properties** | Functional dependency: VALUE determined by VARIABLES. |

Figure 3.1127: Illustrating the SUM_HEIGHT_PROPER_PLAIN constraint of the **Example** slot

           Figures 3.1128 and 3.1129 respectively depict the automaton associated with the constraint
                SUM_HEIGHT_PROPER_PLAIN and its simplified form.

Figure 3.1128: Automaton for the SUM_HEIGHT_PROPER_PLAIN constraint obtained by applying decoration Table 2.35 to the seed transducer of the PROPER_PLAIN pattern where default is 0

Figure 3.1129: Automaton for the SUM_HEIGHT_PROPER_PLAIN constraint obtained by applying decoration Table 2.37 to the seed transducer of the PROPER_PLAIN pattern where default is 0

|   | $s$ | $r$ | $t$ |
|---|---|---|---|
| $s$ | $\overrightarrow{C} + \overleftarrow{C}$ | $\overrightarrow{C} + \overleftarrow{C}$ | $\overrightarrow{C} + \overleftarrow{C}$ |
| $r$ | $\overrightarrow{C} + \overleftarrow{C}$ | $\overrightarrow{C} + \overleftarrow{C}$ | $\min(\overrightarrow{D}, \overleftarrow{D}, \text{VAR}_i)$ |
| $t$ | $\overrightarrow{C} + \overleftarrow{C}$ | $\min(\overrightarrow{D}, \overleftarrow{D}, \text{VAR}_i)$ | $\min(\overrightarrow{D}, \overleftarrow{D}, \text{VAR}_i)$ |

Table 3.146: Glue matrix for the SUM_HEIGHT_PROPER_PLAIN constraint defined as the composition of the PROPER_PLAIN pattern , the feature MIN , and the aggregator sum ; cells of the glue matrix are coloured with the colour of the constituent to which they are related.

# 3.555   SUM_HEIGHT_PROPER_PLATEAU

    DESCRIPTION          AUTOMATON

| | |
|---|---|
| **Origin** | Based on the PROPER_PLATEAU pattern. |
| **Constraint** | SUM_HEIGHT_PROPER_PLATEAU(VALUE, VARIABLES) |
| **Arguments** | VALUE      : dvar |
| | VARIABLES  : collection(var−dvar) |
| **Restrictions** | $\mathtt{sv} \leq 3 \vee \mathtt{rv} \leq 1 \Rightarrow \mathtt{VALUE} = 0$ |
| | $\mathtt{VALUE} = 0 \vee \mathtt{VALUE} \geq \min(\mathtt{minv} + 1 ①, (\mathtt{minv} + 1) * \mathtt{np} ②)$ |
| | $\mathtt{VALUE} = 0 \vee \mathtt{VALUE} \leq \max(\mathtt{maxv}, \mathtt{maxv} * \mathtt{np})$ |
| | required(VARIABLES, var) |
| | where |
| | $\mathtt{sv} = |\mathtt{VARIABLES}|$ |
| | $\mathtt{np} = \max(0, \lfloor (\mathtt{sv} - 1)/3 \rfloor)$ |
| | $\mathtt{minv} = $ minval(VARIABLES.var) |
| | $\mathtt{maxv} = $ maxval(VARIABLES.var) |
| | $\mathtt{rv} = $ range(VARIABLES.var) |
| **Purpose** | VALUE is the sum of all minimum values in each occurrence of the PROPER_PLATEAU pattern in the time-series given by the VARIABLES collection. If the pattern does not occur, VALUE takes the default value 0. |
| | An occurrence of the pattern PROPER_PLATEAU is the *maximal* subsequence which matches the regular expression $<=^+>$. |
| | Assume that the occurrence of the pattern PROPER_PLATEAU starts at position $i$ and ends at position $j$. The feature MIN computes the minimum of the values from index $i + 1$ to index $j$. |
| **Example** | $(12, \langle 7, 1, 3, 3, 2, 5, 1, 4, 4, 3, 2, 3, 5, 5, 5, 3 \rangle)$ |

Figure 3.1130 provides an example where the SUM_HEIGHT_PROPER_PLATEAU $(12, [7, 1, 3, 3, 2, 5, 1, 4, 4, 3, 2, 3, 5, 5, 5, 3])$ constraint holds.

| | |
|---|---|
| **Typical** | $|\mathtt{VARIABLES}| > 3$ |
| | range(VARIABLES.var) $> 1$ |
| **Symmetry** | Items of VARIABLES can be reversed. |
| **Arg. properties** | Functional dependency: VALUE determined by VARIABLES. |

Figure 3.1130: Illustrating the SUM_HEIGHT_PROPER_PLATEAU constraint of the **Example** slot

                    Figures 3.1131 and 3.1132 respectively depict the automaton associated with the constraint
SUM_HEIGHT_PROPER_PLATEAU and its simplified form.

Figure 3.1131: Automaton for the SUM_HEIGHT_PROPER_PLATEAU constraint obtained by applying decoration Table 2.35 to the seed transducer of the PROPER_PLATEAU pattern where default is $0$

Figure 3.1132: Automaton for the SUM_HEIGHT_PROPER_PLATEAU constraint obtained by applying decoration Table 2.37 to the seed transducer of the PROPER_PLATEAU pattern where default is $0$

|   | $s$ | $r$ | $t$ |
|---|---|---|---|
| $s$ | $\overrightarrow{C} + \overleftarrow{C}$ | $\overrightarrow{C} + \overleftarrow{C}$ | $\overrightarrow{C} + \overleftarrow{C}$ |
| $r$ | $\overrightarrow{C} + \overleftarrow{C}$ | $\overrightarrow{C} + \overleftarrow{C}$ | $\min(\overrightarrow{D}, \overleftarrow{D}, \mathtt{VAR}_i)$ |
| $t$ | $\overrightarrow{C} + \overleftarrow{C}$ | $\min(\overrightarrow{D}, \overleftarrow{D}, \mathtt{VAR}_i)$ | $\min(\overrightarrow{D}, \overleftarrow{D}, \mathtt{VAR}_i)$ |

Table 3.147: Glue matrix for the SUM_HEIGHT_PROPER_PLATEAU constraint defined as the composition of the PROPER_PLATEAU pattern , the feature MIN , and the aggregator sum ; cells of the glue matrix are coloured with the colour of the constituent to which they are related.

## 3.556   SUM_HEIGHT_STEADY

▶   ◁ ◀        **DESCRIPTION**        **AUTOMATON**

| | |
|---|---|
| **Origin** | Based on the STEADY pattern. |
| **Constraint** | SUM_HEIGHT_STEADY(VALUE, VARIABLES) |
| **Arguments** | VALUE      : dvar |
| | VARIABLES  : collection(var−dvar) |

**Restrictions**

$sv \leq 1 \Rightarrow VALUE = 0$
$rv = 1 \Rightarrow VALUE \geq minv * np$①
$rv \geq 2 \Rightarrow VALUE = 0 \lor VALUE \geq \min(minv$②$, minv * np$③$)$
$rv = 1 \Rightarrow VALUE \leq maxv * np$
$rv \geq 2 \Rightarrow VALUE = 0 \lor VALUE \leq \max(maxv, maxv * np)$
required(VARIABLES, var)
where
$sv = |VARIABLES|$
$rv =$ range(VARIABLES.var)
$np = \max(0, sv - 1)$
$minv =$ minval(VARIABLES.var)
$maxv =$ maxval(VARIABLES.var)

**Purpose**

VALUE is the sum of all minimum values in each occurrence of the STEADY pattern in the time-series given by the VARIABLES collection. If the pattern does not occur, VALUE takes the default value 0.

An occurrence of the pattern STEADY is the subsequence which matches the regular expression =.

Assume that the occurrence of the pattern STEADY starts at position $i$ and ends at position $j$. The feature MIN computes the minimum of the values from index $i$ to index $j + 1$.

**Example**

$(30, \langle 1, 1, 7, 3, 3, 5, 5, 5, 6, 5, 5, 5, 7, 2, 6, 6 \rangle)$

Figure   3.1133   provides   an   example   where   the   SUM_HEIGHT_STEADY $(30, [1, 1, 7, 3, 3, 5, 5, 5, 6, 5, 5, 5, 7, 2, 6, 6])$ constraint holds.

| | |
|---|---|
| **Typical** | $|VARIABLES| > 1$ |

| | |
|---|---|
| **Symmetry** | Items of VARIABLES can be reversed. |
| **Arg. properties** | Functional dependency: VALUE determined by VARIABLES. |

Figure 3.1133: Illustrating the SUM_HEIGHT_STEADY constraint of the **Example** slot

**Automaton** Figures 3.1134 and 3.1135 respectively depict the automaton associated with the constraint SUM_HEIGHT_STEADY and its simplified form.

$$\left\{\begin{array}{l} C \leftarrow \texttt{default} \\ D \leftarrow +\infty \\ R \leftarrow \texttt{default} \end{array}\right\}$$

$$\overset{=}{\left\{\begin{array}{l} D \leftarrow +\infty \\ R \leftarrow R + \min(\min(D, \texttt{VAR}_i), \texttt{VAR}_{i+1}) \end{array}\right\}} \qquad \gneqq s \qquad \gneqq$$

$$\boxed{R + C}$$

Figure 3.1134: Automaton for the SUM_HEIGHT_STEADY constraint obtained by applying decoration Table 2.35 to the seed transducer of the STEADY pattern where default is 0

$$\{R \leftarrow \texttt{default}\}$$

$$\overset{=}{\{R \leftarrow R + \texttt{VAR}_i\}} \qquad \gneqq s \qquad \gneqq$$

$$\boxed{R}$$

Figure 3.1135: Automaton for the SUM_HEIGHT_STEADY constraint obtained by applying decoration Table 2.37 to the seed transducer of the STEADY pattern where default is 0

|  | $s$ |
|---|---|
| $s$ | $\overrightarrow{C} + \overleftarrow{C}$ |

Table 3.148: Glue matrix for the SUM_HEIGHT_STEADY constraint defined as the composition of the STEADY pattern , the feature MIN , and the aggregator sum ; cells of the glue matrix are coloured with the colour of the constituent to which they are related.

## 3.557 SUM_HEIGHT_STEADY_SEQUENCE

► ◁ ◄

**DESCRIPTION**                **AUTOMATON**

| Origin | Based on the STEADY_SEQUENCE pattern. |
|---|---|

| Constraint | SUM_HEIGHT_STEADY_SEQUENCE(VALUE, VARIABLES) |
|---|---|

| Arguments | VALUE       : dvar |
|---|---|
|  | VARIABLES : collection(var−dvar) |

**Restrictions**

$\mathtt{sv} \leq 1 \Rightarrow \mathtt{VALUE} = 0$

$\mathtt{rv} = 1 \Rightarrow \mathtt{VALUE} = 0 \vee \mathtt{VALUE} \geq \mathtt{minv}$

$\mathtt{rv} \geq 2 \wedge \mathtt{minv} = -1 \Rightarrow \mathtt{VALUE} = 0 \vee \mathtt{VALUE} \geq -1 * \lfloor (\mathtt{sv} - 2)/3 \rfloor - 1 ①$

$\mathtt{rv} \geq 2 \wedge \mathtt{minv} \neq -1 \wedge \mathtt{sv} \leq 4 \Rightarrow$
$$\bigvee \left( \begin{array}{l} \mathtt{VALUE} = 0, \\ \mathtt{VALUE} \geq \min(\mathtt{minv}②, \mathtt{minv} * \lfloor (\mathtt{np} + 1)/2 \rfloor + (\mathtt{minv} + 1) * \lfloor \mathtt{np}/2 \rfloor ③) \end{array} \right)$$

$\mathtt{rv} \geq 2 \wedge \mathtt{minv} \neq -1 \wedge \mathtt{sv} \geq 5 \Rightarrow$
$$\bigvee \left( \begin{array}{l} \mathtt{VALUE} = 0, \\ \mathtt{VALUE} \geq \min \left( \begin{array}{l} \mathtt{minv}④, \\ \mathtt{minv} * \lfloor (\mathtt{np} + 1)/2 \rfloor + (\mathtt{minv} + 1) * \lfloor \mathtt{np}/2 \rfloor - \\ (\mathtt{np} + 1) \bmod 2 * \mathtt{sv} \bmod 2 ⑤ \end{array} \right) \end{array} \right)$$

$\mathtt{rv} = 1 \Rightarrow \mathtt{VALUE} = 0 \vee \mathtt{VALUE} \leq \mathtt{maxv}$

$\mathtt{rv} \geq 2 \wedge \mathtt{maxv} = 1 \Rightarrow \mathtt{VALUE} = 0 \vee \mathtt{VALUE} \leq \lfloor (\mathtt{sv} - 2)/3 \rfloor + 1$

$\mathtt{rv} \geq 2 \wedge \mathtt{maxv} \neq 1 \wedge \mathtt{sv} \leq 4 \Rightarrow$
$$\bigvee \left( \begin{array}{l} \mathtt{VALUE} = 0, \\ \mathtt{VALUE} \leq \max(\mathtt{maxv}, \mathtt{maxv} * \lfloor (\mathtt{np} + 1)/2 \rfloor + (\mathtt{maxv} - 1) * \lfloor \mathtt{np}/2 \rfloor) \end{array} \right)$$

$\mathtt{rv} \geq 2 \wedge \mathtt{maxv} \neq 1 \wedge \mathtt{sv} \geq 5 \Rightarrow$
$$\bigvee \left( \begin{array}{l} \mathtt{VALUE} = 0, \\ \mathtt{VALUE} \leq \max \left( \begin{array}{l} \mathtt{maxv}, \\ \sum \left( \begin{array}{l} \mathtt{maxv} * \lfloor (\mathtt{np} + 1)/2 \rfloor, \\ (\mathtt{maxv} - 1) * \lfloor \mathtt{np}/2 \rfloor, \\ (\mathtt{np} + 1) \bmod 2 * \mathtt{sv} \bmod 2 \end{array} \right) \end{array} \right) \end{array} \right)$$

required(VARIABLES, var)

where

$\mathtt{sv} = |\mathtt{VARIABLES}|$

$\mathtt{rv} = \mathtt{range}(\mathtt{VARIABLES.var})$

$\mathtt{np} = \lfloor \mathtt{sv}/2 \rfloor$

$\mathtt{minv} = \mathtt{minval}(\mathtt{VARIABLES.var})$

$\mathtt{maxv} = \mathtt{maxval}(\mathtt{VARIABLES.var})$

**Purpose**

> VALUE is the sum of all minimum values in each occurrence of the STEADY_SEQUENCE pattern in the time-series given by the VARIABLES collection. If the pattern does not occur, VALUE takes the default value 0.
>
> An occurrence of the pattern STEADY_SEQUENCE is the *maximal* subsequence which matches the regular expression $=^+$.
>
> Assume that the occurrence of the pattern STEADY_SEQUENCE starts at position $i$ and ends at position $j$. The feature MIN computes the minimum of the values from index $i$ to index $j + 1$.

**Example**

$$(13, \langle 3, 1, 1, 4, 5, 5, 5, 6, 2, 2, 4, 4, 3, 2, 1, 1 \rangle)$$

Figure 3.1136 provides an example where the SUM_HEIGHT_STEADY_SEQUENCE $(13, [3, 1, 1, 4, 5, 5, 5, 6, 2, 2, 4, 4, 3, 2, 1, 1])$ constraint holds.

Figure 3.1136: Illustrating the SUM_HEIGHT_STEADY_SEQUENCE constraint of the **Example** slot

**Typical**

$|\text{VARIABLES}| > 1$

**Symmetry**

Items of VARIABLES can be reversed.

**Arg. properties**

Functional dependency: VALUE determined by VARIABLES.

**Automaton**       Figures 3.1137 and 3.1138 respectively depict the automaton associated with the constraint SUM_HEIGHT_STEADY_SEQUENCE and its simplified form.

Figure 3.1137: Automaton for the SUM_HEIGHT_STEADY_SEQUENCE constraint obtained by applying decoration Table 2.35 to the seed transducer of the STEADY_SEQUENCE pattern where default is $0$

Figure 3.1138: Automaton for the SUM_HEIGHT_STEADY_SEQUENCE constraint obtained by applying decoration Table 2.37 to the seed transducer of the STEADY_SEQUENCE pattern where default is $0$

|       | $s$ | $r$ |
|-------|-----|-----|
| $s$   | $\overrightarrow{C} + \overleftarrow{C}$ | $\overrightarrow{C} + \overleftarrow{C}$ |
| $r$   | $\overrightarrow{C} + \overleftarrow{C}$ | $\min(\overrightarrow{C}, \overleftarrow{C}, \overrightarrow{D}, \overleftarrow{D}, \text{VAR}_i)$ |

Table 3.149: Glue matrix for the SUM_HEIGHT_STEADY_SEQUENCE constraint defined as the composition of the STEADY_SEQUENCE pattern , the feature MIN , and the aggregator sum ; cells of the glue matrix are coloured with the colour of the constituent to which they are related.

# 3.558   SUM_MAX_BUMP_ON_DECREASING_SEQUENCE

▶    ◁ ◀     **DESCRIPTION**      **AUTOMATON**

| | |
|---|---|
| **Origin** | Based on the BUMP_ON_DECREASING_SEQUENCE pattern. |
| **Constraint** | SUM_MAX_BUMP_ON_DECREASING_SEQUENCE(VALUE, VARIABLES) |
| **Arguments** | VALUE      :   dvar <br> VARIABLES :   collection(var−dvar) |

**Restrictions**

$\mathtt{sv} \leq 5 \vee \mathtt{rv} \leq 2 \Rightarrow \mathtt{VALUE} = 0$

$\mathtt{VALUE} = 0 \vee \mathtt{VALUE} \geq \min(\mathtt{minv} + 2①, (\mathtt{minv} + 2) * \mathtt{np}②)$

$\mathtt{VALUE} = 0 \vee \mathtt{VALUE} \leq \max(\mathtt{maxv}, \mathtt{maxv} * \mathtt{np})$

required(VARIABLES, var)

where

$\mathtt{sv} = |\mathtt{VARIABLES}|$

$\mathtt{np} = \max(0, \lfloor (\mathtt{sv} - 3)/3 \rfloor)$

$\mathtt{minv} = \mathsf{minval}(\mathtt{VARIABLES.var})$

$\mathtt{maxv} = \mathsf{maxval}(\mathtt{VARIABLES.var})$

$\mathtt{rv} = \mathsf{range}(\mathtt{VARIABLES.var})$

**Purpose**

VALUE is the sum of all maximum values in each occurrence of the BUMP_ON_DECREASING_SEQUENCE pattern in the time-series given by the VARIABLES collection. If the pattern does not occur, VALUE takes the default value 0.

An occurrence of the pattern BUMP_ON_DECREASING_SEQUENCE is the subsequence which matches the regular expression `>><>`.

Assume that the occurrence of the pattern BUMP_ON_DECREASING_SEQUENCE starts at position $i$ and ends at position $j$. The feature MAX computes the maximum of the values from index $i + 2$ to index $j$.

**Example**

$(11, \langle 7, 6, 5, 6, 5, 4, 1, 4, 7, 5, 4, 2, 5, 4, 3, 3 \rangle)$

Figure 3.1139 provides an example where the SUM_MAX_BUMP_ON_DECREASING_SEQUENCE $(11, [7, 6, 5, 6, 5, 4, 1, 4, 7, 5, 4, 2, 5, 4, 3, 3])$ constraint holds.

**Typical**

$|\mathtt{VARIABLES}| > 5$

$\mathsf{range}(\mathtt{VARIABLES.var}) > 2$

**Arg. properties**

Functional dependency: VALUE determined by VARIABLES.

Figure 3.1139: Illustrating the SUM_MAX_BUMP_ON_DECREASING_SEQUENCE constraint of the **Example** slot

**Automaton**     Figures 3.1140 and 3.1141 respectively depict the automaton associated with the constraint
SUM_MAX_BUMP_ON_DECREASING_SEQUENCE and its simplified form.

Figure 3.1140: Automaton for the SUM_MAX_BUMP_ON_DECREASING_SEQUENCE
constraint obtained by applying decoration Table 2.35 to the seed transducer of the
BUMP_ON_DECREASING_SEQUENCE pattern where default is 0

Figure 3.1141: Automaton for the SUM_MAX_BUMP_ON_DECREASING_SEQUENCE constraint obtained by applying decoration Table 2.27 to the seed transducer of the BUMP_ON_DECREASING_SEQUENCE pattern where default is 0

## 3.559 SUM_MAX_DECREASING

► ◁ ◀    **DESCRIPTION**    **AUTOMATON**

**Origin**         Based on the DECREASING pattern.

**Constraint**     SUM_MAX_DECREASING(VALUE, VARIABLES)

**Arguments**      VALUE      : dvar
                   VARIABLES  : collection(var−dvar)

**Restrictions**   $\mathtt{sv} \leq 1 \vee \mathtt{rv} \leq 1 \Rightarrow \mathtt{VALUE} = 0$

$$\mathtt{VALUE} \geq \min_{\mathtt{q} \in [\mathtt{lb1}, \mathtt{ub1}]} \min \left( 0, \sum \left( \begin{array}{l} \mathtt{q} * \left( \sum \left( \begin{array}{l} \prod \left( \begin{array}{l} \mathtt{np} - 1, \\ \mathtt{minv} + 1 \end{array} \right), \\ \lfloor \prod \left( \begin{array}{l} \mathtt{np} - 1, \\ \mathtt{np} - 2 \end{array} \right) /2 \rfloor \end{array} \right) \right), \\ \prod \left( \begin{array}{l} \max \left( 0, \min \left( 1, \mathtt{np} - 1 \right) \right), \\ \mathtt{sv} \bmod \mathtt{np} * \mathtt{q}, \\ \mathtt{minv} + 1 + \mathtt{np} - 1 \end{array} \right), \\ \prod \left( \begin{array}{l} \max \left( 0, \min \left( 1, 2 - \mathtt{np} \right) \right), \\ \mathtt{minv} + 1, \\ \mathtt{q} - 1 \end{array} \right) \end{array} \right) \right) \quad ①$$

$$\mathtt{VALUE} \leq \max_{\mathtt{q} \in [\mathtt{lb2}, \mathtt{ub2}]} \max \left( 0, \sum \left( \begin{array}{l} \mathtt{q} * \left( \begin{array}{l} (\mathtt{np} - 1) * \mathtt{maxv} - \\ \lfloor \prod \left( \begin{array}{l} \mathtt{np} - 1, \\ \mathtt{np} - 2 \end{array} \right) /2 \rfloor \end{array} \right), \\ \prod \left( \begin{array}{l} \max \left( 0, \min \left( 1, \mathtt{np} - 1 \right) \right), \\ \mathtt{sv} \bmod \mathtt{np} * \mathtt{q}, \\ \mathtt{maxv} - \mathtt{np} + 1 \end{array} \right), \\ \prod \left( \begin{array}{l} \max \left( 0, \min \left( 1, 2 - \mathtt{np} \right) \right), \\ \mathtt{maxv}, \\ \mathtt{q} - 1 \end{array} \right) \end{array} \right) \right)$$

required(VARIABLES, var)
where
$\mathtt{sv} = |\mathtt{VARIABLES}|$
$\mathtt{np} = \lfloor \mathtt{sv}/\mathtt{q} \rfloor$
$\mathtt{maxv} = \mathtt{maxval}(\mathtt{VARIABLES.var})$
$\mathtt{minv} = \mathtt{minval}(\mathtt{VARIABLES.var})$
$\mathtt{rv} = \mathtt{range}(\mathtt{VARIABLES.var})$

$$\mathtt{lb1} = \min \left( \begin{array}{l} \lfloor \mathtt{sv}/2 \rfloor + 1, \\ \sum \left( \begin{array}{l} \lfloor \mathtt{sv}/\min(\min(\mathtt{sv}, \mathtt{rv}), |\mathtt{minv}| + 1) \rfloor, \\ \min(1, \mathtt{sv} \bmod \min(\min(\mathtt{sv}, \mathtt{rv}), |\mathtt{minv}| + 1)) \end{array} \right) \end{array} \right)$$

$\mathtt{ub1} = \lfloor \mathtt{sv}/2 \rfloor + 1$

$$\mathtt{lb2} = \min \left( \begin{array}{l} \lfloor \mathtt{sv}/2 \rfloor + 1, \\ \sum \left( \begin{array}{l} \lfloor \mathtt{sv}/\min(\min(\mathtt{sv}, \mathtt{rv}), |\mathtt{maxv}| + 1) \rfloor, \\ \min(1, \mathtt{sv} \bmod \min(\min(\mathtt{sv}, \mathtt{rv}), |\mathtt{maxv}| + 1)) \end{array} \right) \end{array} \right)$$

$\mathtt{ub2} = \lfloor \mathtt{sv}/2 \rfloor + 1$

**Purpose**

VALUE is the sum of all maximum values in each occurrence of the DECREASING pattern in the time-series given by the VARIABLES collection. If the pattern does not occur, VALUE takes the default value 0.

An occurrence of the pattern DECREASING is the subsequence which matches the regular expression >.

Assume that the occurrence of the pattern DECREASING starts at position $i$ and ends at position $j$. The feature MAX computes the maximum of the values from index $i$ to index $j + 1$.

**Example**

$(23, \langle 3, 4, 2, 2, 5, 6, 6, 4, 4, 3, 1, 1, 4, 6, 4, 4 \rangle)$

Figure 3.1142 provides an example where the SUM_MAX_DECREASING $(23, [3, 4, 2, 2, 5, 6, 6, 4, 4, 3, 1, 1, 4, 6, 4, 4])$ constraint holds.

Figure 3.1142: Illustrating the SUM_MAX_DECREASING constraint of the **Example** slot

**Typical**

$|\texttt{VARIABLES}| > 1$
$\texttt{range}(\texttt{VARIABLES.var}) > 1$

**Arg. properties**

Functional dependency: VALUE determined by VARIABLES.

**Automaton**   Figures 3.1143 and 3.1144 respectively depict the automaton associated with the constraint SUM_MAX_DECREASING and its simplified form.

Figure 3.1143: Automaton for the SUM_MAX_DECREASING constraint obtained by applying decoration Table 2.35 to the seed transducer of the DECREASING pattern where default is 0

Figure 3.1144: Automaton for the SUM_MAX_DECREASING constraint obtained by applying decoration Table 2.37 to the seed transducer of the DECREASING pattern where default is 0

Table 3.150: Glue matrix for the SUM_MAX_DECREASING constraint defined as the composition of the DECREASING pattern , the feature MAX , and the aggregator sum ; cells of the glue matrix are coloured with the colour of the constituent to which they are related.

## 3.560   SUM_MAX_DECREASING_SEQUENCE

► ◁ ◀        **DESCRIPTION**        **AUTOMATON**

$> (> | =)^* > |$
$>$

**Origin**            Based on the DECREASING_SEQUENCE pattern.

**Constraint**        SUM_MAX_DECREASING_SEQUENCE(VALUE, VARIABLES)

**Arguments**         VALUE      : dvar
                      VARIABLES  : collection(var−dvar)

**Restrictions**      $sv \leq 1 \vee rv \leq 1 \Rightarrow \text{VALUE} = 0$
                      $\text{VALUE} = 0 \vee \text{VALUE} \geq \min(minv + 1①, (minv + 1) * np②)$
                      $\text{VALUE} = 0 \vee \text{VALUE} \leq \max(maxv③, maxv * np④)$
                      required(VARIABLES, var)
                      where
                      $sv = |\text{VARIABLES}|$
                      $np = \lfloor sv/2 \rfloor$
                      $minv = minval(\text{VARIABLES.var})$
                      $maxv = maxval(\text{VARIABLES.var})$
                      $rv = range(\text{VARIABLES.var})$

**Purpose**           VALUE is the sum of all maximum values in each occurrence of the DECREAS-
                      ING_SEQUENCE pattern in the time-series given by the VARIABLES collection. If the
                      pattern does not occur, VALUE takes the default value 0.
                      An occurrence of the pattern DECREASING_SEQUENCE is the *maximal* subsequence
                      which matches the regular expression $> (> | =)^* > | >$.
                      Assume that the occurrence of the pattern DECREASING_SEQUENCE starts at position
                      $i$ and ends at position $j$. The feature MAX computes the maximum of the values from
                      index $i$ to index $j + 1$.

**Example**           $(16, \langle 3, 4, 2, 2, 5, 6, 6, 4, 4, 3, 1, 1, 4, 6, 4, 4 \rangle)$

                      Figure 3.1145 provides an example where the SUM_MAX_DECREASING_SEQUENCE
                      $(16, [3, 4, 2, 2, 5, 6, 6, 4, 4, 3, 1, 1, 4, 6, 4, 4])$ constraint holds.

**Typical**           $|\text{VARIABLES}| > 1$
                      $range(\text{VARIABLES.var}) > 1$

**Arg. properties**   Functional dependency: VALUE determined by VARIABLES.

Figure 3.1145: Illustrating the SUM_MAX_DECREASING_SEQUENCE constraint of the **Example** slot

    Figures 3.1146 and 3.1147 respectively depict the automaton associated with the constraint
SUM_MAX_DECREASING_SEQUENCE and its simplified form.

Figure 3.1146: Automaton for the SUM_MAX_DECREASING_SEQUENCE constraint obtained by applying decoration Table 2.35 to the seed transducer of the DECREASING_SEQUENCE pattern where default is 0

Figure 3.1147: Automaton for the SUM_MAX_DECREASING_SEQUENCE constraint obtained by applying decoration Table 2.37 to the seed transducer of the DECREASING_SEQUENCE pattern where default is 0

|   | $s$ | $t$ |
|---|---|---|
| $s$ | $\overrightarrow{C} + \overleftarrow{C}$ | $\overrightarrow{C} + \overleftarrow{C}$ |
| $t$ | $\overrightarrow{C} + \overleftarrow{C}$ | $\max(\overrightarrow{C}, \overleftarrow{C}, \overrightarrow{D}, \overleftarrow{D}, \mathtt{VAR}_i)$ |

Table 3.151: Glue matrix for the SUM_MAX_DECREASING_SEQUENCE constraint defined as the composition of the DECREASING_SEQUENCE pattern, the feature MAX, and the aggregator sum; cells of the glue matrix are coloured with the colour of the constituent to which they are related.

# 3.561 SUM_MAX_DIP_ON_INCREASING_SEQUENCE

▶    ◁ ◀        **DESCRIPTION**        **AUTOMATON**

| | |
|---|---|
| **Origin** | Based on the DIP_ON_INCREASING_SEQUENCE pattern. |
| **Constraint** | SUM_MAX_DIP_ON_INCREASING_SEQUENCE(VALUE, VARIABLES) |
| **Arguments** | VALUE      : dvar<br>VARIABLES : collection(var−dvar) |

**Restrictions**

$\text{sv} \leq 5 \vee \text{rv} \leq 2 \Rightarrow \text{VALUE} = 0$
$\text{VALUE} = 0 \vee \text{VALUE} \geq \min(\text{minv} + 2①, (\text{minv} + 2) * \text{np}②)$
$\text{VALUE} = 0 \vee \text{VALUE} \leq \max(\text{maxv}, \text{maxv} * \text{np})$
required(VARIABLES, var)
where
$\text{sv} = |\text{VARIABLES}|$
$\text{np} = \max(0, \lfloor (\text{sv} - 3)/3 \rfloor)$
$\text{minv} = \text{minval}(\text{VARIABLES.var})$
$\text{maxv} = \text{maxval}(\text{VARIABLES.var})$
$\text{rv} = \text{range}(\text{VARIABLES.var})$

**Purpose**

VALUE is the sum of all maximum values in each occurrence of the DIP_ON_INCREASING_SEQUENCE pattern in the time-series given by the VARIABLES collection. If the pattern does not occur, VALUE takes the default value 0.

An occurrence of the pattern DIP_ON_INCREASING_SEQUENCE is the subsequence which matches the regular expression $<<><<$.

Assume that the occurrence of the pattern DIP_ON_INCREASING_SEQUENCE starts at position $i$ and ends at position $j$. The feature MAX computes the maximum of the values from index $i + 2$ to index $j$.

**Example**

$(11, \langle 1, 2, 3, 2, 5, 6, 7, 4, 1, 3, 4, 6, 1, 2, 4, 4 \rangle)$

Figure 3.1148 provides an example where the SUM_MAX_DIP_ON_INCREASING_SEQUENCE $(11, [1, 2, 3, 2, 5, 6, 7, 4, 1, 3, 4, 6, 1, 2, 4, 4])$ constraint holds.

**Typical**

$|\text{VARIABLES}| > 5$
$\text{range}(\text{VARIABLES.var}) > 2$

**Arg. properties**

Functional dependency: VALUE determined by VARIABLES.

Figure 3.1148: Illustrating the SUM_MAX_DIP_ON_INCREASING_SEQUENCE constraint of the **Example** slot

**Automaton**    Figures 3.1149 and 3.1150 respectively depict the automaton associated with the constraint SUM_MAX_DIP_ON_INCREASING_SEQUENCE and its simplified form.

Figure 3.1149: Automaton for the SUM_MAX_DIP_ON_INCREASING_SEQUENCE constraint obtained by applying decoration Table 2.35 to the seed transducer of the DIP_ON_INCREASING_SEQUENCE pattern where default is 0

Figure 3.1150: Automaton for the SUM_MAX_DIP_ON_INCREASING_SEQUENCE constraint obtained by applying decoration Table 2.25 to the seed transducer of the DIP_ON_INCREASING_SEQUENCE pattern where default is $0$

## 3.562 SUM_MAX_INCREASING

► ◁ ◀   **DESCRIPTION**   **AUTOMATON**

**Origin**       Based on the INCREASING pattern.

**Constraint**   SUM_MAX_INCREASING(VALUE, VARIABLES)

**Arguments**    VALUE      : dvar
                 VARIABLES  : collection(var−dvar)

**Restrictions**   $\text{sv} \leq 1 \vee \text{rv} \leq 1 \Rightarrow \text{VALUE} = 0$

$$\text{VALUE} \geq \min_{\text{q} \in [\text{lb1}, \text{ub1}]} \min \left( 0, \sum \left( \begin{array}{l} \text{q} * \left( \sum \left( \begin{array}{l} \prod \left( \begin{array}{l} \text{np}-1, \\ \text{minv}+1 \end{array} \right), \\ \lfloor \prod \left( \begin{array}{l} \text{np}-1, \\ \text{np}-2 \end{array} \right)/2 \rfloor \end{array} \right) \right), \\ \prod \left( \begin{array}{l} \max \left( 0, \min \left( 1, \text{np}-1 \right) \right), \\ \text{sv} \bmod \text{np} * \text{q}, \\ \text{minv}+1+\text{np}-1 \end{array} \right), \\ \prod \left( \begin{array}{l} \max \left( 0, \min \left( 1, 2-\text{np} \right) \right), \\ \text{minv}+1, \\ \text{q}-1 \end{array} \right) \end{array} \right) \right)$$ ①

$$\text{VALUE} \leq \max_{\text{q} \in [\text{lb2}, \text{ub2}]} \max \left( 0, \sum \left( \begin{array}{l} \text{q} * \left( \begin{array}{l} (\text{np}-1) * \text{maxv}- \\ \lfloor \prod \left( \begin{array}{l} \text{np}-1, \\ \text{np}-2 \end{array} \right)/2 \rfloor \end{array} \right), \\ \prod \left( \begin{array}{l} \max \left( 0, \min \left( 1, \text{np}-1 \right) \right), \\ \text{sv} \bmod \text{np} * \text{q}, \\ \text{maxv}-\text{np}+1 \end{array} \right), \\ \prod \left( \begin{array}{l} \max \left( 0, \min \left( 1, 2-\text{np} \right) \right), \\ \text{maxv}, \\ \text{q}-1 \end{array} \right) \end{array} \right) \right)$$

required(VARIABLES, var)
where
$\text{sv} = |\text{VARIABLES}|$
$\text{np} = \lfloor \text{sv}/\text{q} \rfloor$
$\text{maxv} = \text{maxval}(\text{VARIABLES.var})$
$\text{minv} = \text{minval}(\text{VARIABLES.var})$
$\text{rv} = \text{range}(\text{VARIABLES.var})$
$\text{lb1} = \min \left( \begin{array}{l} \lfloor \text{sv}/2 \rfloor + 1, \\ \sum \left( \begin{array}{l} \lfloor \text{sv}/\min(\min(\text{sv}, \text{rv}), |\text{minv}| + 1) \rfloor, \\ \min(1, \text{sv} \bmod \min(\min(\text{sv}, \text{rv}), |\text{minv}| + 1)) \end{array} \right) \end{array} \right)$
$\text{ub1} = \lfloor \text{sv}/2 \rfloor + 1$
$\text{lb2} = \min \left( \begin{array}{l} \lfloor \text{sv}/2 \rfloor + 1, \\ \sum \left( \begin{array}{l} \lfloor \text{sv}/\min(\min(\text{sv}, \text{rv}), |\text{maxv}| + 1) \rfloor, \\ \min(1, \text{sv} \bmod \min(\min(\text{sv}, \text{rv}), |\text{maxv}| + 1)) \end{array} \right) \end{array} \right)$
$\text{ub2} = \lfloor \text{sv}/2 \rfloor + 1$

VALUE ≥ −33   VALUE ≥ −57

**Purpose**

> VALUE is the sum of all maximum values in each occurrence of the INCREASING pattern in the time-series given by the VARIABLES collection. If the pattern does not occur, VALUE takes the default value 0.
>
> An occurrence of the pattern INCREASING is the subsequence which matches the regular expression <.
>
> Assume that the occurrence of the pattern INCREASING starts at position $i$ and ends at position $j$. The feature MAX computes the maximum of the values from index $i$ to index $j + 1$.

**Example**

> $(21, \langle 4, 3, 5, 5, 2, 1, 1, 3, 3, 4, 6, 6, 3, 1, 3, 3 \rangle)$

Figure 3.1151 provides an example where the SUM_MAX_INCREASING $(21, [4, 3, 5, 5, 2, 1, 1, 3, 3, 4, 6, 6, 3, 1, 3, 3])$ constraint holds.

Figure 3.1151: Illustrating the SUM_MAX_INCREASING constraint of the **Example** slot

**Typical**

> $|\text{VARIABLES}| > 1$
> $\text{range}(\text{VARIABLES.var}) > 1$

**Arg. properties**

> Functional dependency: VALUE determined by VARIABLES.

**Automaton**    Figures 3.1152 and 3.1153 respectively depict the automaton associated with the constraint SUM_MAX_INCREASING and its simplified form.

Figure 3.1152: Automaton for the SUM_MAX_INCREASING constraint obtained by applying decoration Table 2.35 to the seed transducer of the INCREASING pattern where default is 0

Figure 3.1153: Automaton for the SUM_MAX_INCREASING constraint obtained by applying decoration Table 2.38 to the seed transducer of the INCREASING pattern where default is 0

Table 3.152: Glue matrix for the SUM_MAX_INCREASING constraint defined as the composition of the INCREASING pattern , the feature MAX , and the aggregator sum ; cells of the glue matrix are coloured with the colour of the constituent to which they are related.

# 3.563   SUM_MAX_INCREASING_SEQUENCE

▶   ◁ ◀      **DESCRIPTION**      **AUTOMATON**

| **Origin** | Based on the INCREASING_SEQUENCE pattern. |
|---|---|

| **Constraint** | SUM_MAX_INCREASING_SEQUENCE(VALUE, VARIABLES) |
|---|---|

| **Arguments** | VALUE      : dvar |
|---|---|
| | VARIABLES  : collection(var−dvar) |

**Restrictions**

$sv \leq 1 \vee rv \leq 1 \Rightarrow VALUE = 0$
$VALUE = 0 \vee VALUE \geq \min(minv + 1①, (minv + 1) * np②)$
$VALUE = 0 \vee VALUE \leq \max(maxv③, maxv * np④)$
required(VARIABLES, var)
where
$sv = |VARIABLES|$
$np = \lfloor sv/2 \rfloor$
$minv = minval(VARIABLES.var)$
$maxv = maxval(VARIABLES.var)$
$rv = range(VARIABLES.var)$

**Purpose**

VALUE is the sum of all maximum values in each occurrence of the INCREAS-ING_SEQUENCE pattern in the time-series given by the VARIABLES collection. If the pattern does not occur, VALUE takes the default value 0.

An occurrence of the pattern INCREASING_SEQUENCE is the *maximal* subsequence which matches the regular expression $< (< | =)^* < | <$.

Assume that the occurrence of the pattern INCREASING_SEQUENCE starts at position $i$ and ends at position $j$. The feature MAX computes the maximum of the values from index $i$ to index $j + 1$.

**Example**

$(14, \langle 4, 3, 5, 5, 2, 1, 1, 3, 3, 4, 6, 6, 3, 1, 3, 3 \rangle)$

Figure 3.1154 provides an example where the SUM_MAX_INCREASING_SEQUENCE $(14, [4, 3, 5, 5, 2, 1, 1, 3, 3, 4, 6, 6, 3, 1, 3, 3])$ constraint holds.

| **Typical** | $|VARIABLES| > 1$ |
|---|---|
| | $range(VARIABLES.var) > 1$ |

| **Arg. properties** | Functional dependency: VALUE determined by VARIABLES. |
|---|---|

Figure 3.1154: Illustrating the SUM_MAX_INCREASING_SEQUENCE constraint of the **Example** slot

**Automaton**     Figures 3.1155 and 3.1156 respectively depict the automaton associated with the constraint
SUM_MAX_INCREASING_SEQUENCE and its simplified form.

Figure 3.1155: Automaton for the SUM_MAX_INCREASING_SEQUENCE constraint obtained by applying decoration Table 2.35 to the seed transducer of the INCREASING_SEQUENCE pattern where default is $0$

Figure 3.1156: Automaton for the SUM_MAX_INCREASING_SEQUENCE constraint obtained by applying decoration Table 2.24 to the seed transducer of the INCREASING_SEQUENCE pattern where default is $0$

|   | $s$ | $t$ |
|---|-----|-----|
| $s$ | $\overrightarrow{C} + \overleftarrow{C}$ | $\overrightarrow{C} + \overleftarrow{C}$ |
| $t$ | $\overrightarrow{C} + \overleftarrow{C}$ | $\max(\overrightarrow{C}, \overleftarrow{C}, \overrightarrow{D}, \overleftarrow{D}, \mathtt{VAR}_i)$ |

Table 3.153: Glue matrix for the SUM_MAX_INCREASING_SEQUENCE constraint defined as the composition of the INCREASING_SEQUENCE pattern, the feature MAX, and the aggregator sum; cells of the glue matrix are coloured with the colour of the constituent to which they are related.

# 3.564　SUM_MAX_INFLEXION

▶　◁ ◀　　DESCRIPTION　　AUTOMATON

$< (< | =)^* > | > (> | =)^* <$

| | |
|---|---|
| **Origin** | Based on the INFLEXION pattern. |
| **Constraint** | SUM_MAX_INFLEXION(VALUE, VARIABLES) |
| **Arguments** | VALUE　　　: dvar <br> VARIABLES : collection(var−dvar) |

**Restrictions**

$$\mathtt{sv} \leq 2 \vee \mathtt{rv} \leq 1 \Rightarrow \mathtt{VALUE} = 0$$

$$\bigvee \left( \begin{array}{l} \mathtt{VALUE} = 0, \\ \mathtt{VALUE} \geq \min(\mathtt{minv}①, \mathtt{minv} * \lfloor (\mathtt{np}+1)/2 \rfloor + (\mathtt{minv}+1) * \lfloor \mathtt{np}/2 \rfloor ②) \end{array} \right)$$

$$\bigvee \left( \begin{array}{l} \mathtt{VALUE} = 0, \\ \mathtt{VALUE} \leq \max(\mathtt{maxv}③, \mathtt{maxv} * \lfloor (\mathtt{np}+1)/2 \rfloor + (\mathtt{maxv}-1) * \lfloor \mathtt{np}/2 \rfloor ④) \end{array} \right)$$

required(VARIABLES, var)

where

$$\mathtt{sv} = |\mathtt{VARIABLES}|$$
$$\mathtt{np} = \max(0, \mathtt{sv} - 2)$$
$$\mathtt{minv} = \mathtt{minval}(\mathtt{VARIABLES.var})$$
$$\mathtt{maxv} = \mathtt{maxval}(\mathtt{VARIABLES.var})$$
$$\mathtt{rv} = \mathtt{range}(\mathtt{VARIABLES.var})$$

**Purpose**

VALUE is the sum of all maximum values in each occurrence of the INFLEXION pattern in the time-series given by the VARIABLES collection. If the pattern does not occur, VALUE takes the default value 0.

An occurrence of the pattern INFLEXION is the *maximal* subsequence which matches the regular expression $< (< | =)^* > | > (> | =)^* <$.

Assume that the occurrence of the pattern INFLEXION starts at position $i$ and ends at position $j$. The feature MAX computes the maximum of the values from index $i + 1$ to index $j$.

**Example**

$$(31, \langle 1, 2, 6, 6, 4, 4, 3, 5, 2, 5, 1, 5, 3, 3, 4, 4 \rangle)$$

Figure 3.1157 provides an example where the SUM_MAX_INFLEXION $(31, [1, 2, 6, 6, 4, 4, 3, 5, 2, 5, 1, 5, 3, 3, 4, 4])$ constraint holds.

| | |
|---|---|
| **Typical** | $\|\mathtt{VARIABLES}\| > 2$ <br> range(VARIABLES.var) > 1 |
| **Symmetry** | Items of VARIABLES can be reversed. |
| **Arg. properties** | Functional dependency: VALUE determined by VARIABLES. |

Figure 3.1157: Illustrating the SUM_MAX_INFLEXION constraint of the **Example** slot

**Automaton**     Figures 3.1158 and 3.1159 respectively depict the automaton associated with the constraint SUM_MAX_INFLEXION and its simplified form.

Figure 3.1158: Automaton for the SUM_MAX_INFLEXION constraint obtained by applying decoration Table 2.35 to the seed transducer of the INFLEXION pattern where default is 0 (transition $r \to t$ has the same accumulators updates as transition $t \to r$)

Figure 3.1159: Automaton for the SUM_MAX_INFLEXION constraint obtained by applying decoration Table 2.25 to the seed transducer of the INFLEXION pattern where default is 0 (transition $r \to t$ has the same accumulators updates as transition $t \to r$)

# 3.565   SUM_MAX_PEAK

► ◁ ◄     DESCRIPTION          AUTOMATON

$< (= | <)^* (> | =)^* >$

| **Origin** | Based on the PEAK pattern. |
|---|---|

| **Constraint** | SUM_MAX_PEAK(VALUE, VARIABLES) |
|---|---|

| **Arguments** | VALUE      : dvar |
|---|---|
| | VARIABLES : collection(var−dvar) |

**Restrictions**

$sv \leq 2 \vee rv \leq 1 \Rightarrow \text{VALUE} = 0$

$\text{VALUE} = 0 \vee \text{VALUE} \geq \min(minv + 1①, (minv + 1) * np②)$

$\text{VALUE} = 0 \vee \text{VALUE} \leq \max(maxv③, maxv * np④)$

required(VARIABLES, var)

where

$sv = |\text{VARIABLES}|$

$np = \max(0, \lfloor (sv - 1)/2 \rfloor)$

$minv = \text{minval}(\text{VARIABLES.var})$

$maxv = \text{maxval}(\text{VARIABLES.var})$

$rv = \text{range}(\text{VARIABLES.var})$

**Purpose**

VALUE is the sum of all maximum values in each occurrence of the PEAK pattern in the time-series given by the VARIABLES collection. If the pattern does not occur, VALUE takes the default value 0.

An occurrence of the pattern PEAK is the *maximal* subsequence which matches the regular expression $< (= | <)^* (> | =)^* >$.

Assume that the occurrence of the pattern PEAK starts at position $i$ and ends at position $j$. The feature MAX computes the maximum of the values from index $i + 1$ to index $j$.

| **Example** | $(14, \langle 7, 5, 5, 1, 4, 5, 2, 2, 3, 5, 6, 2, 3, 3, 3, 1 \rangle)$ |
|---|---|

Figure 3.1160 provides an example where the SUM_MAX_PEAK $(14, [7, 5, 5, 1, 4, 5, 2, 2, 3, 5, 6, 2, 3, 3, 3, 1])$ constraint holds.

| **Typical** | $|\text{VARIABLES}| > 2$ |
|---|---|
| | range(VARIABLES.var) > 1 |

| **Symmetry** | Items of VARIABLES can be reversed. |
|---|---|

| **Arg. properties** | Functional dependency: VALUE determined by VARIABLES. |
|---|---|

Figure 3.1160: Illustrating the SUM_MAX_PEAK constraint of the **Example** slot

**Automaton**  Figures 3.1161 and 3.1162 respectively depict the automaton associated with the constraint SUM_MAX_PEAK and its simplified form.

Figure 3.1161: Automaton for the SUM_MAX_PEAK constraint obtained by applying decoration Table 2.35 to the seed transducer of the PEAK pattern where default is 0

Figure 3.1162: Automaton for the SUM_MAX_PEAK constraint obtained by applying decoration Table 2.37 to the seed transducer of the PEAK pattern where default is 0

|   | $s$ | $r$ | $t$ |
|---|---|---|---|
| $s$ | $\overrightarrow{C} + \overleftarrow{C}$ | $\overrightarrow{C} + \overleftarrow{C}$ | $\overrightarrow{C} + \overleftarrow{C}$ |
| $r$ | $\overrightarrow{C} + \overleftarrow{C}$ | $\max(\overrightarrow{D}, \overleftarrow{D}, \mathtt{VAR}_i)$ | $\max(\overleftarrow{C}, \overrightarrow{D}, \overleftarrow{D}, \mathtt{VAR}_i)$ |
| $t$ | $\overrightarrow{C} + \overleftarrow{C}$ | $\max(\overrightarrow{C}, \overrightarrow{D}, \overleftarrow{D}, \mathtt{VAR}_i)$ | $\overrightarrow{C} + \overleftarrow{C}$ |

Table 3.154: Glue matrix for the SUM_MAX_PEAK constraint defined as the composition of the PEAK pattern , the feature MAX , and the aggregator sum ; cells of the glue matrix are coloured with the colour of the constituent to which they are related.

# 3.566   SUM_MAX_STRICTLY_DECREASING_SEQUENCE

▶   ◁ ◀        DESCRIPTION        AUTOMATON

**Origin**          Based on the STRICTLY_DECREASING_SEQUENCE pattern.

**Constraint**      SUM_MAX_STRICTLY_DECREASING_SEQUENCE(VALUE, VARIABLES)

**Arguments**       VALUE      : dvar
                    VARIABLES  : collection(var−dvar)

**Restrictions**    $sv \leq 1 \lor rv \leq 1 \Rightarrow$ VALUE $= 0$
                    VALUE $= 0 \lor$ VALUE $\geq \min(minv + 1①, (minv + 1) * np②)$
                    VALUE $= 0 \lor$ VALUE $\leq \max(maxv③, maxv * np④)$
                    required(VARIABLES, var)
                    where
                    $sv = |$VARIABLES$|$
                    $np = \lfloor sv/2 \rfloor$
                    $minv =$ minval(VARIABLES.var)
                    $maxv =$ maxval(VARIABLES.var)
                    $rv =$ range(VARIABLES.var)

**Purpose**         VALUE is the sum of all maximum values in each occurrence of the
                    STRICTLY_DECREASING_SEQUENCE pattern in the time-series given by the VARIABLES
                    collection. If the pattern does not occur, VALUE takes the default value 0.
                    An occurrence of the pattern STRICTLY_DECREASING_SEQUENCE is the *maximal* sub-
                    sequence which matches the regular expression $>^+$.
                    Assume that the occurrence of the pattern STRICTLY_DECREASING_SEQUENCE starts at
                    position $i$ and ends at position $j$. The feature MAX computes the maximum of the values
                    from index $i$ to index $j + 1$.

**Example**         $(16, \langle 4, 4, 6, 4, 1, 1, 3, 4, 4, 6, 6, 5, 2, 2, 4, 3 \rangle)$

                    Figure 3.1163 provides an example where the SUM_MAX_STRICTLY_DECREASING_SEQUENCE
                    $(16, [4, 4, 6, 4, 1, 1, 3, 4, 4, 6, 6, 5, 2, 2, 4, 3])$ constraint holds.

**Typical**         $|$VARIABLES$| > 1$
                    range(VARIABLES.var) $> 1$

**Arg. properties** Functional dependency: VALUE determined by VARIABLES.

Figure 3.1163: Illustrating the SUM_MAX_STRICTLY_DECREASING_SEQUENCE constraint of the **Example** slot

**Automaton**          Figures 3.1164 and 3.1165 respectively depict the automaton associated with the constraint
SUM_MAX_STRICTLY_DECREASING_SEQUENCE and its simplified form.

Figure 3.1164: Automaton for the SUM_MAX_STRICTLY_DECREASING_SEQUENCE
constraint obtained by applying decoration Table 2.35 to the seed transducer of the
STRICTLY_DECREASING_SEQUENCE pattern where default is $0$

Figure 3.1165: Automaton for the SUM_MAX_STRICTLY_DECREASING_SEQUENCE
constraint obtained by applying decoration Table 2.37 to the seed transducer of the
STRICTLY_DECREASING_SEQUENCE pattern where default is $0$

|  | $s$ | $r$ |
|---|---|---|
| $s$ | $\overrightarrow{C} + \overleftarrow{C}$ | $\overrightarrow{C} + \overleftarrow{C}$ |
| $r$ | $\overrightarrow{C} + \overleftarrow{C}$ | $\max(\overrightarrow{C}, \overleftarrow{C}, \overrightarrow{D}, \overleftarrow{D}, \text{VAR}_i)$ |

Table 3.155: Glue matrix for the SUM\_MAX\_STRICTLY\_DECREASING\_SEQUENCE constraint defined as the composition of the STRICTLY\_DECREASING\_SEQUENCE pattern, the feature MAX, and the aggregator sum; cells of the glue matrix are coloured with the colour of the constituent to which they are related.

## 3.567  SUM_MAX_STRICTLY_INCREASING_SEQUENCE

► ◁ ◄    **DESCRIPTION**          **AUTOMATON**

| | |
|---|---|
| **Origin** | Based on the STRICTLY_INCREASING_SEQUENCE pattern. |
| **Constraint** | SUM_MAX_STRICTLY_INCREASING_SEQUENCE(VALUE, VARIABLES) |
| **Arguments** | VALUE : dvar |
| | VARIABLES : collection(var−dvar) |

**Restrictions**

$\mathtt{sv} \leq 1 \vee \mathtt{rv} \leq 1 \Rightarrow \mathtt{VALUE} = 0$

$\mathtt{VALUE} = 0 \vee \mathtt{VALUE} \geq \min(\mathtt{minv} + 1①, (\mathtt{minv} + 1) * \mathtt{np}②)$

$\mathtt{VALUE} = 0 \vee \mathtt{VALUE} \leq \max(\mathtt{maxv}③, \mathtt{maxv} * \mathtt{np}④)$

required(VARIABLES, var)

where

$\mathtt{sv} = |\mathtt{VARIABLES}|$

$\mathtt{np} = \lfloor \mathtt{sv}/2 \rfloor$

$\mathtt{minv} = \mathtt{minval}(\mathtt{VARIABLES.var})$

$\mathtt{maxv} = \mathtt{maxval}(\mathtt{VARIABLES.var})$

$\mathtt{rv} = \mathtt{range}(\mathtt{VARIABLES.var})$

**Purpose**

VALUE is the sum of all maximum values in each occurrence of the STRICTLY_INCREASING_SEQUENCE pattern in the time-series given by the VARIABLES collection. If the pattern does not occur, VALUE takes the default value 0.

An occurrence of the pattern STRICTLY_INCREASING_SEQUENCE is the *maximal* sub-sequence which matches the regular expression $<^+$.

Assume that the occurrence of the pattern STRICTLY_INCREASING_SEQUENCE starts at position $i$ and ends at position $j$. The feature MAX computes the maximum of the values from index $i$ to index $j + 1$.

**Example**

$(14, \langle 4, 3, 5, 5, 2, 1, 1, 2, 3, 4, 6, 6, 3, 1, 2, 3 \rangle)$

Figure 3.1166 provides an example where the SUM_MAX_STRICTLY_INCREASING_SEQUENCE $(14, [4, 3, 5, 5, 2, 1, 1, 2, 3, 4, 6, 6, 3, 1, 2, 3])$ constraint holds.

**Typical**

$|\mathtt{VARIABLES}| > 1$

$\mathtt{range}(\mathtt{VARIABLES.var}) > 1$

**Arg. properties**

Functional dependency: VALUE determined by VARIABLES.

Figure 3.1166: Illustrating the SUM_MAX_STRICTLY_INCREASING_SEQUENCE constraint of the **Example** slot

**Automaton**     Figures 3.1167 and 3.1168 respectively depict the automaton associated with the constraint
SUM_MAX_STRICTLY_INCREASING_SEQUENCE and its simplified form.

Figure 3.1167: Automaton for the SUM_MAX_STRICTLY_INCREASING_SEQUENCE
constraint obtained by applying decoration Table 2.35 to the seed transducer of the
STRICTLY_INCREASING_SEQUENCE pattern where default is 0

Figure 3.1168: Automaton for the SUM_MAX_STRICTLY_INCREASING_SEQUENCE
constraint obtained by applying decoration Table 2.24 to the seed transducer of the
STRICTLY_INCREASING_SEQUENCE pattern where default is 0

|  | $s$ | $r$ |
|---|---|---|
| $s$ | $\overrightarrow{C} + \overleftarrow{C}$ | $\overrightarrow{C} + \overleftarrow{C}$ |
| $r$ | $\overrightarrow{C} + \overleftarrow{C}$ | $\max(\overrightarrow{C}, \overleftarrow{C}, \overrightarrow{D}, \overleftarrow{D}, \mathtt{VAR}_i)$ |

Table 3.156: Glue matrix for the SUM_MAX_STRICTLY_INCREASING_SEQUENCE constraint defined as the composition of the STRICTLY_INCREASING_SEQUENCE pattern , the feature MAX , and the aggregator sum ; cells of the glue matrix are coloured with the colour of the constituent to which they are related.

# 3.568   SUM_MAX_SUMMIT

     DESCRIPTION        AUTOMATON

$(< | < (= | <)^* <) (> | > (= | >)^* >)$

| | |
|---|---|
| **Origin** | Based on the SUMMIT pattern. |
| **Constraint** | SUM_MAX_SUMMIT(VALUE, VARIABLES) |
| **Arguments** | VALUE : dvar |
| | VARIABLES : collection(var−dvar) |

**Restrictions**

$sv \leq 2 \vee rv \leq 1 \Rightarrow \text{VALUE} = 0$

$\text{VALUE} = 0 \vee \text{VALUE} \geq \min(\text{minv} + 1①, (\text{minv} + 1) * \text{np}②)$

$\text{VALUE} = 0 \vee \text{VALUE} \leq \max(\text{maxv}③, \text{maxv} * \text{np}④)$

required(VARIABLES, var)

where

$sv = |\text{VARIABLES}|$

$np = \max(0, \lfloor (sv - 1)/2 \rfloor)$

$minv = \text{minval}(\text{VARIABLES.var})$

$maxv = \text{maxval}(\text{VARIABLES.var})$

$rv = \text{range}(\text{VARIABLES.var})$

**Purpose**

VALUE is the sum of all maximum values in each occurrence of the SUMMIT pattern in the time-series given by the VARIABLES collection. If the pattern does not occur, VALUE takes the default value 0.

An occurrence of the pattern SUMMIT is the *maximal* subsequence which matches the regular expression $(< | < (= | <)^* <) (> | > (= | >)^* >)$.

Assume that the occurrence of the pattern SUMMIT starts at position $i$ and ends at position $j$. The feature MAX computes the maximum of the values from index $i + 1$ to index $j$.

**Example**

$(12, \langle 7, 1, 5, 4, 4, 3, 3, 4, 6, 6, 2, 3, 4, 2, 3, 1 \rangle)$

Figure 3.1169 provides an example where the SUM_MAX_SUMMIT $(12, [7, 1, 5, 4, 4, 3, 3, 4, 6, 6, 2, 3, 4, 2, 3, 1])$ constraint holds.

| | |
|---|---|
| **Typical** | $|\text{VARIABLES}| > 2$ |
| | range(VARIABLES.var) > 1 |
| **Symmetry** | Items of VARIABLES can be reversed. |
| **Arg. properties** | Functional dependency: VALUE determined by VARIABLES. |

Figure 3.1169: Illustrating the SUM_MAX_SUMMIT constraint of the **Example** slot

**Automaton**     Figures 3.1170 and 3.1171 respectively depict the automaton associated with the constraint SUM_MAX_SUMMIT and its simplified form.

Figure 3.1170: Automaton for the SUM_MAX_SUMMIT constraint obtained by applying decoration Table 2.35 to the seed transducer of the SUMMIT pattern where default is 0 (transition $u \rightarrow r$ has the same accumulator update as transition $r \rightarrow u$)

| | $s$ | $r$ | $t$ | $u$ |
|---|---|---|---|---|
| $s$ | $\overrightarrow{C} + \overleftarrow{C}$ | $\overrightarrow{C} + \overleftarrow{C}$ | $\overrightarrow{C} + \overleftarrow{C}$ | $\overrightarrow{C} + \overleftarrow{C}$ |
| $r$ | $\overrightarrow{C} + \overleftarrow{C}$ | $\max(\overrightarrow{D}, \overleftarrow{D}, \mathtt{VAR}_i)$ | $\max(\overleftarrow{C}, \overrightarrow{D}, \overleftarrow{D}, \mathtt{VAR}_i)$ | $\overrightarrow{C} + \overleftarrow{C}$ |
| $t$ | $\overrightarrow{C} + \overleftarrow{C}$ | $\max(\overrightarrow{C}, \overrightarrow{D}, \overleftarrow{D}, \mathtt{VAR}_i)$ | $\overrightarrow{C} + \overleftarrow{C}$ | $\max(\overrightarrow{C}, \overrightarrow{D}, \overleftarrow{D}, \mathtt{VAR}_i)$ |
| $u$ | $\overrightarrow{C} + \overleftarrow{C}$ | $\overrightarrow{C} + \overleftarrow{C}$ | $\max(\overleftarrow{C}, \overrightarrow{D}, \overleftarrow{D}, \mathtt{VAR}_i)$ | $\overrightarrow{C} + \overleftarrow{C}$ |

Table 3.157: Glue matrix for the SUM_MAX_SUMMIT constraint defined as the composition of the SUMMIT pattern , the feature MAX , and the aggregator sum ; cells of the glue matrix are coloured with the colour of the constituent to which they are related.

Figure 3.1171: Automaton for the SUM_MAX_SUMMIT constraint obtained by applying decoration Table 2.37 to the seed transducer of the SUMMIT pattern where default is 0 (transition $u \to r$ has the same accumulator update as transition $r \to u$)

# 3.569 SUM_MAX_ZIGZAG

► ◁ ◀    DESCRIPTION    AUTOMATON

$(<>)^+(<|<>)|(><)^+(>|><)$

**Origin**    Based on the ZIGZAG pattern.

**Constraint**    SUM_MAX_ZIGZAG(VALUE, VARIABLES)

**Arguments**
VALUE      : dvar
VARIABLES  : collection(var−dvar)

**Restrictions**

$\mathtt{sv} \leq 3 \vee \mathtt{rv} \leq 1 \Rightarrow \mathtt{VALUE} = 0$

$$\mathtt{VALUE} \geq \min_{\mathtt{q}\in[0,\mathtt{qub}]} \min \left( 0, \sum \left( \begin{array}{l} \prod \left( \begin{array}{l} \lfloor(\mathtt{sv}-3*\mathtt{q})/4\rfloor, \\ \mathtt{minv}+1 \end{array} \right), \\ \mathtt{q}*(\mathtt{minv}+2), \\ \prod \left( \begin{array}{l} \min\left( 1, \max\left( \mathtt{maxv}-\mathtt{minv}-1, 0 \right) \right), \\ \max\left( -1, \min\left( 0, \mathtt{q}*(\mathtt{minv}+2) \right) \right), \\ \min \left( \begin{array}{l} \mathtt{q}, \\ \left( \begin{array}{l} (\mathtt{sv}-\lfloor(\mathtt{sv}-3*\mathtt{q})/4\rfloor)*4- \\ \mathtt{q}*3 \end{array} \right) \bmod 3 \end{array} \right) \end{array} \right) \end{array} \right) \right)$$

$\mathtt{rv} = 2 \Rightarrow \mathtt{VALUE} = 0 \vee \mathtt{VALUE} \leq \max(\mathtt{maxv}, \mathtt{maxv}*\mathtt{np1})$
$\mathtt{rv} \geq 3 \Rightarrow \mathtt{VALUE} = 0 \vee \mathtt{VALUE} \leq \max(\mathtt{maxv}, \max(\mathtt{maxv}*\mathtt{np1}, \mathtt{maxv}*\mathtt{np2}))$
required(VARIABLES, var)
where
$\mathtt{sv} = |\mathtt{VARIABLES}|$
$\mathtt{rv} = \mathtt{range}(\mathtt{VARIABLES.var})$
$\mathtt{np1} = \lfloor \mathtt{sv}/4 \rfloor$
$\mathtt{np2} = \max(0, \lfloor(\mathtt{sv}-1)/3\rfloor)$
$\mathtt{minv} = \mathtt{minval}(\mathtt{VARIABLES.var})$
$\mathtt{maxv} = \mathtt{maxval}(\mathtt{VARIABLES.var})$
$\mathtt{qub} = \min(1, \max(\mathtt{maxv}-\mathtt{minv}-1, 0)) * \mathtt{np2}$

**Purpose**

VALUE is the sum of all maximum values in each occurrence of the ZIGZAG pattern in the time-series given by the VARIABLES collection. If the pattern does not occur, VALUE takes the default value 0.

An occurrence of the pattern ZIGZAG is the *maximal* subsequence which matches the regular expression $(<>)^+(<|<>)|(><)^+(>|><)$.

Assume that the occurrence of the pattern ZIGZAG starts at position $i$ and ends at position $j$. The feature MAX computes the maximum of the values from index $i + 1$ to index $j$.

**Example**    $(16, \langle 4, 1, 3, 1, 4, 6, 1, 5, 5, 2, 7, 2, 3, 1, 6, 1 \rangle)$

Figure 3.1172 provides an example where the SUM_MAX_ZIGZAG $(16, [4, 1, 3, 1, 4, 6, 1, 5, 5, 2, 7, 2, 3, 1, 6, 1])$ constraint holds.

Figure 3.1172: Illustrating the SUM_MAX_ZIGZAG constraint of the **Example** slot

| **Typical** | $|\texttt{VARIABLES}| > 3$ |
| | $\texttt{range}(\texttt{VARIABLES.var}) > 1$ |
| | |
| **Symmetry** | Items of VARIABLES can be reversed. |
| | |
| **Arg. properties** | Functional dependency: VALUE determined by VARIABLES. |

**Automaton**        Figures 3.1173 and 3.1174 respectively depict the automaton associated with the constraint
                     SUM_MAX_ZIGZAG and its simplified form.

Figure 3.1173: Automaton for the SUM_MAX_ZIGZAG constraint obtained by applying decoration Table 2.35 to the seed transducer of the ZIGZAG pattern where default is 0; (1) missing transitions from $a, b, c, d, e, f$ to $s$ are labelled by $=$; (2) on transitions from $b, c, e, f$ to $s$ the accumulator $D$ is reset to its initial value; (3) on transitions from $c, f$ to $s$ the accumulator $R$ is updated wrt $C$ and the accumulator $C$ is reset to its initial value

Figure 3.1174: Automaton for the SUM_MAX_ZIGZAG constraint obtained by applying decoration Table 2.23 to the seed transducer of the ZIGZAG pattern where default is 0; missing transitions from $a, b, c, d, e, f$ to $s$ are labelled by $=$; (2) on transitions from $b, c, e, f$ to $s$ the accumulator $D$ is reset to its initial value; (3) on transitions from $c, f$ to $s$ the accumulator $R$ is updated wrt $C$ and the accumulator $C$ is reset to its initial value.

| | $s$ | $a$ | $b$ | $c$ | $d$ | $e$ | $f$ |
|---|---|---|---|---|---|---|---|
| $s$ | $\overleftarrow{C}+\overrightarrow{C}$ | $\overleftarrow{C}+\overrightarrow{C}$ | $\overleftarrow{C}+\overrightarrow{C}$ | $\overleftarrow{C}+\overrightarrow{C}$ | $\overleftarrow{C}+\overrightarrow{C}$ | $\overleftarrow{C}+\overrightarrow{C}$ | $\overleftarrow{C}+\overrightarrow{C}$ |
| $a$ | $\overleftarrow{C}+\overrightarrow{C}$ | $\overleftarrow{C}+\overrightarrow{C}$ | $\max(\overrightarrow{D},\overleftarrow{D},\text{VAR}_i)$ | $\max(\overleftarrow{C},\overrightarrow{D},\overleftarrow{D},\text{VAR}_i)$ | $\overleftarrow{C}+\overrightarrow{C}$ | $\max(\overrightarrow{D},\overleftarrow{D},\text{VAR}_i)$ | $\overleftarrow{C}+\overrightarrow{C}$ |
| $b$ | $\overleftarrow{C}+\overrightarrow{C}$ | $\overleftarrow{C}+\overrightarrow{C}$ | $\overleftarrow{C}+\overrightarrow{C}$ | $\overleftarrow{C}+\overrightarrow{C}$ | $\max(\overrightarrow{D},\overleftarrow{D},\text{VAR}_i)$ | $\overleftarrow{C}+\overrightarrow{C}$ | $\max(\overleftarrow{C},\overrightarrow{D},\overleftarrow{D},\text{VAR}_i)$ |
| $c$ | $\overleftarrow{C}+\overrightarrow{C}$ | $\max(\overrightarrow{C},\overrightarrow{D},\overleftarrow{D},\text{VAR}_i)$ | $\max(\overrightarrow{D},\overleftarrow{D},\text{VAR}_i)$ | $\max(\overrightarrow{C},\overleftarrow{C},\overrightarrow{D},\overleftarrow{D},\text{VAR}_i)$ | $\overleftarrow{C}+\overrightarrow{C}$ | $\max(\overleftarrow{C},\overrightarrow{D},\overleftarrow{D},\text{VAR}_i)$ | $\overleftarrow{C}+\overrightarrow{C}$ |
| $d$ | $\overleftarrow{C}+\overrightarrow{C}$ | $\overleftarrow{C}+\overrightarrow{C}$ | $\overleftarrow{C}+\overrightarrow{C}$ | $\overleftarrow{C}+\overrightarrow{C}$ | $\overleftarrow{C}+\overrightarrow{C}$ | $\overleftarrow{C}+\overrightarrow{C}$ | $\max(\overleftarrow{C},\overrightarrow{D},\overleftarrow{D},\text{VAR}_i)$ |
| $e$ | $\overleftarrow{C}+\overrightarrow{C}$ | $\max(\overrightarrow{D},\overleftarrow{D},\text{VAR}_i)$ | $\overleftarrow{C}+\overrightarrow{C}$ | $\max(\overleftarrow{C},\overrightarrow{D},\overleftarrow{D},\text{VAR}_i)$ | $\overleftarrow{C}+\overrightarrow{C}$ | $\max(\overrightarrow{D},\overleftarrow{D},\text{VAR}_i)$ | $\overleftarrow{C}+\overrightarrow{C}$ |
| $f$ | $\overleftarrow{C}+\overrightarrow{C}$ | $\overleftarrow{C}+\overrightarrow{C}$ | $\max(\overrightarrow{C},\overrightarrow{D},\overleftarrow{D},\text{VAR}_i)$ | $\overleftarrow{C}+\overrightarrow{C}$ | $\max(\overrightarrow{C},\overrightarrow{D},\overleftarrow{D},\text{VAR}_i)$ | $\overleftarrow{C}+\overrightarrow{C}$ | $\max(\overrightarrow{C},\overleftarrow{C},\overrightarrow{D},\overleftarrow{D},\text{VAR}_i)$ |

Table 3.158: Glue matrix for the SUM_MAX_ZIGZAG constraint defined as the composition of the ZIGZAG pattern , the feature MAX , and the aggregator sum ; cells of the glue matrix are coloured with the colour of the constituent to which they are related.

# 3.570   SUM_MIN_BUMP_ON_DECREASING_SEQUENCE

► ◁ ◀        **DESCRIPTION**        **AUTOMATON**

| **Origin** | Based on the BUMP_ON_DECREASING_SEQUENCE pattern. |
|---|---|
| **Constraint** | SUM_MIN_BUMP_ON_DECREASING_SEQUENCE(VALUE, VARIABLES) |

**Arguments**

```
VALUE     : dvar
VARIABLES : collection(var−dvar)
```

**Restrictions**

$\text{sv} \leq 5 \vee \text{rv} \leq 2 \Rightarrow \text{VALUE} = 0$

$\text{VALUE} = 0 \vee \text{VALUE} \geq \min(\text{minv}, \text{minv} * \text{np})$

$\text{VALUE} = 0 \vee \text{VALUE} \leq \max(\text{maxv} - 2①, (\text{maxv} - 2) * \text{np}②)$

required(VARIABLES, var)

where

$\text{sv} = |\text{VARIABLES}|$

$\text{np} = \max(0, \lfloor(\text{sv} - 3)/3\rfloor)$

minv = minval(VARIABLES.var)

maxv = maxval(VARIABLES.var)

rv = range(VARIABLES.var)

**Purpose**

VALUE is the sum of all minimum values in each occurrence of the BUMP_ON_DECREASING_SEQUENCE pattern in the time-series given by the VARIABLES collection. If the pattern does not occur, VALUE takes the default value 0.

An occurrence of the pattern BUMP_ON_DECREASING_SEQUENCE is the subsequence which matches the regular expression `>><<>>`.

Assume that the occurrence of the pattern BUMP_ON_DECREASING_SEQUENCE starts at position $i$ and ends at position $j$. The feature MIN computes the minimum of the values from index $i + 2$ to index $j$.

**Example**

$(7, \langle 7, 6, 5, 6, 5, 4, 1, 4, 7, 5, 4, 2, 5, 4, 3, 3 \rangle)$

Figure 3.1175 provides an example where the SUM_MIN_BUMP_ON_DECREASING_SEQUENCE $(7, [7, 6, 5, 6, 5, 4, 1, 4, 7, 5, 4, 2, 5, 4, 3, 3])$ constraint holds.

**Typical**

$|\text{VARIABLES}| > 5$

range(VARIABLES.var) > 2

**Arg. properties**

Functional dependency: VALUE determined by VARIABLES.

Figure 3.1175: Illustrating the SUM_MIN_BUMP_ON_DECREASING_SEQUENCE constraint of the **Example** slot

**Automaton**     Figures 3.1176 and 3.1177 respectively depict the automaton associated with the constraint SUM_MIN_BUMP_ON_DECREASING_SEQUENCE and its simplified form.

Figure 3.1176: Automaton for the SUM_MIN_BUMP_ON_DECREASING_SEQUENCE constraint obtained by applying decoration Table 2.35 to the seed transducer of the BUMP_ON_DECREASING_SEQUENCE pattern where default is 0

Figure 3.1177: Automaton for the SUM_MIN_BUMP_ON_DECREASING_SEQUENCE constraint obtained by applying decoration Table 2.25 to the seed transducer of the BUMP_ON_DECREASING_SEQUENCE pattern where default is 0

## 3.571   SUM_MIN_DECREASING

      **DESCRIPTION**          **AUTOMATON**

| | |
|---|---|
| **Origin** | Based on the DECREASING pattern. |
| **Constraint** | SUM_MIN_DECREASING(VALUE, VARIABLES) |
| **Arguments** | VALUE : dvar |
| | VARIABLES : collection(var−dvar) |

**Restrictions**

$$\mathtt{sv} \leq 1 \vee \mathtt{rv} \leq 1 \Rightarrow \mathtt{VALUE} = 0$$

$$\mathtt{VALUE} \geq \min_{\mathtt{q} \in [\mathtt{lb1}, \mathtt{ub1}]} \min \left( 0, \sum \left( \prod \begin{pmatrix} \mathtt{q} * \left( \sum \begin{pmatrix} (\mathtt{np}-1)*\mathtt{minv}, \\ \lfloor \prod \begin{pmatrix} \mathtt{np}-1, \\ \mathtt{np}-2 \end{pmatrix} /2 \rfloor \end{pmatrix} \right), \\ \max\left(0, \min\left(1, \mathtt{np}-1\right)\right), \\ \mathtt{sv} \bmod \mathtt{np} * \mathtt{q}, \\ \mathtt{minv} + \mathtt{np} - 1 \\ \max\left(0, \min\left(1, 2-\mathtt{np}\right)\right), \\ \mathtt{minv}, \\ \mathtt{q}-1 \end{pmatrix} \right) \right)$$ ①

$$\mathtt{VALUE} \leq \max_{\mathtt{q} \in [\mathtt{lb2}, \mathtt{ub2}]} \max \left( 0, \sum \left( \prod \begin{pmatrix} \mathtt{q} * \left( \prod \begin{pmatrix} \mathtt{np}-1, \\ \mathtt{maxv}-1 \end{pmatrix} - \\ \lfloor \prod \begin{pmatrix} \mathtt{np}-1, \\ \mathtt{np}-2 \end{pmatrix} /2 \rfloor \right), \\ \max\left(0, \min\left(1, \mathtt{np}-1\right)\right), \\ \mathtt{sv} \bmod \mathtt{np} * \mathtt{q}, \\ \mathtt{maxv} - 1 - \mathtt{np} + 1 \\ \max\left(0, \min\left(1, 2-\mathtt{np}\right)\right), \\ \mathtt{maxv}-1, \\ \mathtt{q}-1 \end{pmatrix} \right) \right)$$

required(VARIABLES, var)

where

$$\mathtt{sv} = |\mathtt{VARIABLES}|$$
$$\mathtt{np} = \lfloor \mathtt{sv}/\mathtt{q} \rfloor$$
$$\mathtt{maxv} = \mathtt{maxval}(\mathtt{VARIABLES.var})$$
$$\mathtt{minv} = \mathtt{minval}(\mathtt{VARIABLES.var})$$
$$\mathtt{rv} = \mathtt{range}(\mathtt{VARIABLES.var})$$

$$\mathtt{lb1} = \min \left( \sum \begin{pmatrix} \lfloor \mathtt{sv}/2 \rfloor + 1, \\ \lfloor \mathtt{sv}/\min(\min(\mathtt{sv},\mathtt{rv}), |\mathtt{minv}| + 1) \rfloor, \\ \min(1, \mathtt{sv} \bmod \min(\min(\mathtt{sv},\mathtt{rv}), |\mathtt{minv}| + 1)) \end{pmatrix} \right)$$
$$\mathtt{ub1} = \lfloor \mathtt{sv}/2 \rfloor + 1$$
$$\mathtt{lb2} = \min \left( \sum \begin{pmatrix} \lfloor \mathtt{sv}/2 \rfloor + 1, \\ \lfloor \mathtt{sv}/\min(\min(\mathtt{sv},\mathtt{rv}), |\mathtt{maxv}| + 1) \rfloor, \\ \min(1, \mathtt{sv} \bmod \min(\min(\mathtt{sv},\mathtt{rv}), |\mathtt{maxv}| + 1)) \end{pmatrix} \right)$$
$$\mathtt{ub2} = \lfloor \mathtt{sv}/2 \rfloor + 1$$

| | |
|---|---|
| **Purpose** | VALUE is the sum of all minimum values in each occurrence of the DECREASING pattern in the time-series given by the VARIABLES collection. If the pattern does not occur, VALUE takes the default value 0.<br><br>An occurrence of the pattern DECREASING is the subsequence which matches the regular expression $>$.<br><br>Assume that the occurrence of the pattern DECREASING starts at position $i$ and ends at position $j$. The feature MIN computes the minimum of the values from index $i$ to index $j + 1$. |
| **Example** | $(14, \langle 3, 4, 2, 2, 5, 6, 6, 4, 4, 3, 1, 1, 4, 6, 4, 4 \rangle)$ |

Figure 3.1178 provides an example where the SUM_MIN_DECREASING $(14, [3, 4, 2, 2, 5, 6, 6, 4, 4, 3, 1, 1, 4, 6, 4, 4])$ constraint holds.

Figure 3.1178: Illustrating the SUM_MIN_DECREASING constraint of the **Example** slot

| | |
|---|---|
| **Typical** | $|\text{VARIABLES}| > 1$<br>$\text{range}(\text{VARIABLES.var}) > 1$ |

| | |
|---|---|
| **Arg. properties** | Functional dependency: VALUE determined by VARIABLES. |

**Automaton**     Figures 3.1179 and 3.1180 respectively depict the automaton associated with the constraint SUM_MIN_DECREASING and its simplified form.

Figure 3.1179: Automaton for the SUM_MIN_DECREASING constraint obtained by applying decoration Table 2.35 to the seed transducer of the DECREASING pattern where `default` is $0$

Figure 3.1180: Automaton for the SUM_MIN_DECREASING constraint obtained by applying decoration Table 2.38 to the seed transducer of the DECREASING pattern where `default` is $0$

Table 3.159: Glue matrix for the SUM_MIN_DECREASING constraint defined as the composition of the DECREASING pattern , the feature MIN , and the aggregator sum ; cells of the glue matrix are coloured with the colour of the constituent to which they are related.

## 3.572   SUM_MIN_DECREASING_SEQUENCE

▶    ◁ ◀     **DESCRIPTION**     **AUTOMATON**

| | |
|---|---|
| **Origin** | Based on the DECREASING_SEQUENCE pattern. |
| **Constraint** | SUM_MIN_DECREASING_SEQUENCE(VALUE, VARIABLES) |
| **Arguments** | VALUE      :   dvar <br> VARIABLES   :   collection(var−dvar) |

**Restrictions**

$$\text{sv} \leq 1 \lor \text{rv} \leq 1 \Rightarrow \text{VALUE} = 0$$
$$\text{VALUE} = 0 \lor \text{VALUE} \geq \min(\text{minv}①, \text{minv} * \text{np}②)$$
$$\text{VALUE} = 0 \lor \text{VALUE} \leq \max(\text{maxv} - 1③, (\text{maxv} - 1) * \text{np}④)$$
required(VARIABLES, var)
where
$$\text{sv} = |\text{VARIABLES}|$$
$$\text{np} = \lfloor \text{sv}/2 \rfloor$$
$$\text{minv} = \text{minval}(\text{VARIABLES.var})$$
$$\text{maxv} = \text{maxval}(\text{VARIABLES.var})$$
$$\text{rv} = \text{range}(\text{VARIABLES.var})$$

**Purpose**

VALUE is the sum of all minimum values in each occurrence of the DECREAS-ING_SEQUENCE pattern in the time-series given by the VARIABLES collection. If the pattern does not occur, VALUE takes the default value 0.

An occurrence of the pattern DECREASING_SEQUENCE is the *maximal* subsequence which matches the regular expression $> (> | =)^* > | >$.

Assume that the occurrence of the pattern DECREASING_SEQUENCE starts at position $i$ and ends at position $j$. The feature MIN computes the minimum of the values from index $i$ to index $j + 1$.

**Example**

$$(7, \langle 3, 4, 2, 2, 5, 6, 6, 4, 4, 3, 1, 1, 4, 6, 4, 4 \rangle)$$

Figure 3.1181 provides an example where the SUM_MIN_DECREASING_SEQUENCE $(7, [3, 4, 2, 2, 5, 6, 6, 4, 4, 3, 1, 1, 4, 6, 4, 4])$ constraint holds.

**Typical**

$$|\text{VARIABLES}| > 1$$
$$\text{range}(\text{VARIABLES.var}) > 1$$

**Arg. properties**

Functional dependency: VALUE determined by VARIABLES.

Figure 3.1181: Illustrating the SUM_MIN_DECREASING_SEQUENCE constraint of the **Example** slot

**Automaton** Figures 3.1182 and 3.1183 respectively depict the automaton associated with the constraint SUM_MIN_DECREASING_SEQUENCE and its simplified form.

Figure 3.1182: Automaton for the SUM_MIN_DECREASING_SEQUENCE constraint obtained by applying decoration Table 2.35 to the seed transducer of the DECREAS-ING_SEQUENCE pattern where default is 0

Figure 3.1183: Automaton for the SUM_MIN_DECREASING_SEQUENCE constraint obtained by applying decoration Table 2.24 to the seed transducer of the DECREAS-ING_SEQUENCE pattern where default is 0

| | $s$ | $t$ |
|---|---|---|
| $s$ | $\overrightarrow{C} + \overleftarrow{C}$ | $\overrightarrow{C} + \overleftarrow{C}$ |
| $t$ | $\overrightarrow{C} + \overleftarrow{C}$ | $\min(\overrightarrow{C}, \overleftarrow{C}, \overrightarrow{D}, \overleftarrow{D}, \text{VAR}_i)$ |

Table 3.160: Glue matrix for the SUM\_MIN\_DECREASING\_SEQUENCE constraint defined as the composition of the DECREASING\_SEQUENCE pattern , the feature MIN , and the aggregator sum ; cells of the glue matrix are coloured with the colour of the constituent to which they are related.

# 3.573   SUM_MIN_DIP_ON_INCREASING_SEQUENCE

▶   ◁ ◀        **DESCRIPTION**          **AUTOMATON**

| | |
|---|---|
| **Origin** | Based on the DIP_ON_INCREASING_SEQUENCE pattern. |
| **Constraint** | SUM_MIN_DIP_ON_INCREASING_SEQUENCE(VALUE, VARIABLES) |
| **Arguments** | VALUE       : dvar <br> VARIABLES : collection(var−dvar) |

**Restrictions**

$sv \leq 5 \vee rv \leq 2 \Rightarrow VALUE = 0$

$VALUE = 0 \vee VALUE \geq \min(minv, minv * np)$

$VALUE = 0 \vee VALUE \leq \max(maxv - 2①, (maxv - 2) * np②)$

required(VARIABLES, var)

where

$sv = |VARIABLES|$

$np = \max(0, \lfloor (sv - 3)/3 \rfloor)$

$minv = minval(VARIABLES.var)$

$maxv = maxval(VARIABLES.var)$

$rv = range(VARIABLES.var)$

**Purpose**

VALUE is the sum of all minimum values in each occurrence of the DIP_ON_INCREASING_SEQUENCE pattern in the time-series given by the VARIABLES collection. If the pattern does not occur, VALUE takes the default value 0.

An occurrence of the pattern DIP_ON_INCREASING_SEQUENCE is the subsequence which matches the regular expression $<<><<$.

Assume that the occurrence of the pattern DIP_ON_INCREASING_SEQUENCE starts at position $i$ and ends at position $j$. The feature MIN computes the minimum of the values from index $i + 2$ to index $j$.

**Example**

$(3, \langle 1, 2, 3, 2, 5, 6, 7, 4, 1, 3, 4, 6, 1, 2, 4, 4 \rangle)$

Figure 3.1184 provides an example where the SUM_MIN_DIP_ON_INCREASING_SEQUENCE $(3, [1, 2, 3, 2, 5, 6, 7, 4, 1, 3, 4, 6, 1, 2, 4, 4])$ constraint holds.

| | |
|---|---|
| **Typical** | $|VARIABLES| > 5$ <br> range(VARIABLES.var) > 2 |
| **Arg. properties** | Functional dependency: VALUE determined by VARIABLES. |

Figure 3.1184: Illustrating the SUM_MIN_DIP_ON_INCREASING_SEQUENCE constraint of the **Example** slot

**Automaton**     Figures 3.1185 and 3.1186 respectively depict the automaton associated with the constraint SUM_MIN_DIP_ON_INCREASING_SEQUENCE and its simplified form.

Figure 3.1185: Automaton for the SUM_MIN_DIP_ON_INCREASING_SEQUENCE constraint obtained by applying decoration Table 2.35 to the seed transducer of the DIP_ON_INCREASING_SEQUENCE pattern where default is $0$

Figure 3.1186: Automaton for the SUM_MIN_DIP_ON_INCREASING_SEQUENCE constraint obtained by applying decoration Table 2.27 to the seed transducer of the DIP_ON_INCREASING_SEQUENCE pattern where default is 0

# 3.574 SUM_MIN_GORGE

$(> | > (= | >)^*>)(< | < (= | <)^* <)$

| | |
|---|---|
| **Origin** | Based on the GORGE pattern. |
| **Constraint** | SUM_MIN_GORGE(VALUE, VARIABLES) |
| **Arguments** | VALUE       : dvar<br>VARIABLES : collection(var−dvar) |

**Restrictions**

$\text{sv} \leq 2 \vee \text{rv} \leq 1 \Rightarrow \text{VALUE} = 0$

$\text{VALUE} = 0 \vee \text{VALUE} \geq \min(\text{minv}①, \text{minv} * \text{np}②)$

$\text{VALUE} = 0 \vee \text{VALUE} \leq \max(\text{maxv} - 1③, (\text{maxv} - 1) * \text{np}④)$

required(VARIABLES, var)

where

$\text{sv} = |\text{VARIABLES}|$

$\text{np} = \max(0, \lfloor (\text{sv} - 1)/2 \rfloor)$

$\text{minv} = \text{minval}(\text{VARIABLES.var})$

$\text{maxv} = \text{maxval}(\text{VARIABLES.var})$

$\text{rv} = \text{range}(\text{VARIABLES.var})$

VALUE ≥ 1

VALUE ≥ −9 = −3 * 3

VALUE ≤ −2

VALUE ≤ 9 = (4 − 1) * 3

**Purpose**

VALUE is the sum of all minimum values in each occurrence of the GORGE pattern in the time-series given by the VARIABLES collection. If the pattern does not occur, VALUE takes the default value 0.

An occurrence of the pattern GORGE is the *maximal* subsequence which matches the regular expression $(> | > (= | >)^*>)(< | < (= | <)^* <)$.

Assume that the occurrence of the pattern GORGE starts at position $i$ and ends at position $j$. The feature MIN computes the minimum of the values from index $i + 1$ to index $j$.

**Example**

$(12, \langle 1, 7, 3, 4, 4, 5, 5, 4, 2, 2, 6, 5, 4, 6, 5, 7 \rangle)$

Figure 3.1187 provides an example where the SUM_MIN_GORGE $(12, [1, 7, 3, 4, 4, 5, 5, 4, 2, 2, 6, 5, 4, 6, 5, 7])$ constraint holds.

| | |
|---|---|
| **Typical** | $|\text{VARIABLES}| > 2$<br>range(VARIABLES.var) > 1 |
| **Symmetry** | Items of VARIABLES can be reversed. |
| **Arg. properties** | Functional dependency: VALUE determined by VARIABLES. |

Figure 3.1187: Illustrating the SUM_MIN_GORGE constraint of the **Example** slot

**Automaton**      Figures 3.1188 and 3.1189 respectively depict the automaton associated with the constraint
SUM_MIN_GORGE and its simplified form.

Figure 3.1188: Automaton for the SUM_MIN_GORGE constraint obtained by applying
decoration Table 2.35 to the seed transducer of the GORGE pattern where default is 0
(transition $u \rightarrow r$ has the same accumulator update as transition $r \rightarrow u$)

|   | $s$ | $r$ | $t$ | $u$ |
|---|---|---|---|---|
| $s$ | $\overrightarrow{C} + \overleftarrow{C}$ | $\overrightarrow{C} + \overleftarrow{C}$ | $\overrightarrow{C} + \overleftarrow{C}$ | $\overrightarrow{C} + \overleftarrow{C}$ |
| $r$ | $\overrightarrow{C} + \overleftarrow{C}$ | $\min(\overrightarrow{D}, \overleftarrow{D}, \text{VAR}_i)$ | $\min(\overleftarrow{C}, \overrightarrow{D}, \overleftarrow{D}, \text{VAR}_i)$ | $\overrightarrow{C} + \overleftarrow{C}$ |
| $t$ | $\overrightarrow{C} + \overleftarrow{C}$ | $\min(\overrightarrow{C}, \overrightarrow{D}, \overleftarrow{D}, \text{VAR}_i)$ | $\overrightarrow{C} + \overleftarrow{C}$ | $\min(\overrightarrow{C}, \overrightarrow{D}, \overleftarrow{D}, \text{VAR}_i)$ |
| $u$ | $\overrightarrow{C} + \overleftarrow{C}$ | $\overrightarrow{C} + \overleftarrow{C}$ | $\min(\overleftarrow{C}, \overrightarrow{D}, \overleftarrow{D}, \text{VAR}_i)$ | $\overrightarrow{C} + \overleftarrow{C}$ |

Table 3.161: Glue matrix for the SUM_MIN_GORGE constraint defined as the compo-
sition of the GORGE pattern , the feature MIN , and the aggregator sum ; cells of the
glue matrix are coloured with the colour of the constituent to which they are related.

Figure 3.1189: Automaton for the SUM_MIN_GORGE constraint obtained by applying decoration Table 2.37 to the seed transducer of the GORGE pattern where default is 0

## 3.575 SUM_MIN_INCREASING

► ◁ ◀        **DESCRIPTION**        **AUTOMATON**

**Origin**         Based on the INCREASING pattern.

**Constraint**     SUM_MIN_INCREASING(VALUE, VARIABLES)

**Arguments**      VALUE       : dvar
                   VARIABLES   : collection(var−dvar)

**Restrictions**   $sv \leq 1 \vee rv \leq 1 \Rightarrow$ VALUE $= 0$

$$\text{VALUE} \geq \min_{q \in [lb1, ub1]} \min\left(0, \sum \left(\begin{array}{l} q * \left(\sum \left(\begin{array}{l} (np-1) * minv, \\ \lfloor \prod \left(\begin{array}{l} np-1, \\ np-2 \end{array}\right) /2 \rfloor \end{array}\right)\right), \\ \prod \left(\begin{array}{l} \max\left(0, \min\left(1, np-1\right)\right), \\ sv \bmod np * q, \\ minv + np - 1 \end{array}\right), \\ \prod \left(\begin{array}{l} \max\left(0, \min\left(1, 2-np\right)\right), \\ minv, \\ q - 1 \end{array}\right) \end{array}\right)\right)$$ ①

$$\text{VALUE} \leq \max_{q \in [lb2, ub2]} \max\left(0, \sum \left(\begin{array}{l} q * \left(\prod \left(\begin{array}{l} np-1, \\ maxv-1 \end{array}\right) - \\ \lfloor \prod \left(\begin{array}{l} np-1, \\ np-2 \end{array}\right) /2 \rfloor \right), \\ \prod \left(\begin{array}{l} \max\left(0, \min\left(1, np-1\right)\right), \\ sv \bmod np * q, \\ maxv - 1 - np + 1 \end{array}\right), \\ \prod \left(\begin{array}{l} \max\left(0, \min\left(1, 2-np\right)\right), \\ maxv - 1, \\ q - 1 \end{array}\right) \end{array}\right)\right)$$

required(VARIABLES, var)
where
$sv = |\text{VARIABLES}|$
$np = \lfloor sv/q \rfloor$
$maxv = \text{maxval}(\text{VARIABLES.var})$
$minv = \text{minval}(\text{VARIABLES.var})$
$rv = \text{range}(\text{VARIABLES.var})$
$lb1 = \min\left(\begin{array}{l} \lfloor sv/2 \rfloor + 1, \\ \sum \left(\begin{array}{l} \lfloor sv/\min(\min(sv, rv), |minv| + 1) \rfloor, \\ \min(1, sv \bmod \min(\min(sv, rv), |minv| + 1)) \end{array}\right) \end{array}\right)$
$ub1 = \lfloor sv/2 \rfloor + 1$
$lb2 = \min\left(\begin{array}{l} \lfloor sv/2 \rfloor + 1, \\ \sum \left(\begin{array}{l} \lfloor sv/\min(\min(sv, rv), |maxv| + 1) \rfloor, \\ \min(1, sv \bmod \min(\min(sv, rv), |maxv| + 1)) \end{array}\right) \end{array}\right)$
$ub2 = \lfloor sv/2 \rfloor + 1$

VALUE ≥ **−41**

VALUE ≥ **−65**

**Purpose**

VALUE is the sum of all minimum values in each occurrence of the INCREASING pattern in the time-series given by the VARIABLES collection. If the pattern does not occur, VALUE takes the default value 0.

An occurrence of the pattern INCREASING is the subsequence which matches the regular expression <.

Assume that the occurrence of the pattern INCREASING starts at position $i$ and ends at position $j$. The feature MIN computes the minimum of the values from index $i$ to index $j + 1$.

**Example**

$$(12, \langle 4, 3, 5, 5, 2, 1, 1, 3, 3, 4, 6, 6, 3, 1, 3, 3 \rangle)$$

Figure 3.1190 provides an example where the SUM_MIN_INCREASING $(12, [4, 3, 5, 5, 2, 1, 1, 3, 3, 4, 6, 6, 3, 1, 3, 3])$ constraint holds.

Figure 3.1190: Illustrating the SUM_MIN_INCREASING constraint of the **Example** slot

**Typical**

$|\texttt{VARIABLES}| > 1$

$\texttt{range}(\texttt{VARIABLES.var}) > 1$

**Arg. properties**

Functional dependency: VALUE determined by VARIABLES.

**Automaton**     Figures 3.1191 and 3.1192 respectively depict the automaton associated with the constraint SUM_MIN_INCREASING and its simplified form.

Figure 3.1191: Automaton for the SUM_MIN_INCREASING constraint obtained by applying decoration Table 2.35 to the seed transducer of the INCREASING pattern where `default` is $0$

Figure 3.1192: Automaton for the SUM_MIN_INCREASING constraint obtained by applying decoration Table 2.37 to the seed transducer of the INCREASING pattern where `default` is $0$

Table 3.162: Glue matrix for the SUM_MIN_INCREASING constraint defined as the composition of the INCREASING pattern, the feature MIN, and the aggregator sum; cells of the glue matrix are coloured with the colour of the constituent to which they are related.

# 3.576   SUM_MIN_INCREASING_SEQUENCE

▶   ◁ ◀      **DESCRIPTION**      **AUTOMATON**

| | |
|---|---|
| **Origin** | Based on the INCREASING_SEQUENCE pattern. |
| **Constraint** | SUM_MIN_INCREASING_SEQUENCE(VALUE, VARIABLES) |
| **Arguments** | VALUE      : dvar<br>VARIABLES : collection(var−dvar) |

**Restrictions**

$\text{sv} \leq 1 \lor \text{rv} \leq 1 \Rightarrow \text{VALUE} = 0$

$\text{VALUE} = 0 \lor \text{VALUE} \geq \min(\text{minv}①, \text{minv} * \text{np}②)$

$\text{VALUE} = 0 \lor \text{VALUE} \leq \max(\text{maxv} - 1③, (\text{maxv} - 1) * \text{np}④)$

required(VARIABLES, var)

where

$\text{sv} = |\text{VARIABLES}|$

$\text{np} = \lfloor \text{sv}/2 \rfloor$

$\text{minv} = \text{minval}(\text{VARIABLES.var})$

$\text{maxv} = \text{maxval}(\text{VARIABLES.var})$

$\text{rv} = \text{range}(\text{VARIABLES.var})$

**Purpose**

VALUE is the sum of all minimum values in each occurrence of the INCREAS-ING_SEQUENCE pattern in the time-series given by the VARIABLES collection. If the pattern does not occur, VALUE takes the default value 0.

An occurrence of the pattern INCREASING_SEQUENCE is the *maximal* subsequence which matches the regular expression $< (< | =)^* < | <$.

Assume that the occurrence of the pattern INCREASING_SEQUENCE starts at position $i$ and ends at position $j$. The feature MIN computes the minimum of the values from index $i$ to index $j + 1$.

**Example**

$(5, \langle 4, 3, 5, 5, 2, 1, 1, 3, 3, 4, 6, 6, 3, 1, 3, 3 \rangle)$

Figure 3.1193 provides an example where the SUM_MIN_INCREASING_SEQUENCE $(5, [4, 3, 5, 5, 2, 1, 1, 3, 3, 4, 6, 6, 3, 1, 3, 3])$ constraint holds.

**Typical**

$|\text{VARIABLES}| > 1$

range(VARIABLES.var) > 1

**Arg. properties**

Functional dependency: VALUE determined by VARIABLES.

Figure 3.1193: Illustrating the SUM_MIN_INCREASING_SEQUENCE constraint of the **Example** slot

Figure 3.1194: Automaton for the SUM_MIN_INCREASING_SEQUENCE constraint obtained by applying decoration Table 2.35 to the seed transducer of the INCREASING_SEQUENCE pattern where default is $0$

Figure 3.1195: Automaton for the SUM_MIN_INCREASING_SEQUENCE constraint obtained by applying decoration Table 2.37 to the seed transducer of the INCREASING_SEQUENCE pattern where default is $0$

|   | $s$ | $t$ |
|---|---|---|
| $s$ | $\overrightarrow{C} + \overleftarrow{C}$ | $\overrightarrow{C} + \overleftarrow{C}$ |
| $t$ | $\overrightarrow{C} + \overleftarrow{C}$ | $\min(\overrightarrow{C}, \overleftarrow{C}, \overrightarrow{D}, \overleftarrow{D}, \text{VAR}_i)$ |

Table 3.163: Glue matrix for the SUM_MIN_INCREASING_SEQUENCE constraint defined as the composition of the INCREASING_SEQUENCE pattern , the feature MIN , and the aggregator sum ; cells of the glue matrix are coloured with the colour of the constituent to which they are related.

## 3.577  SUM_MIN_INFLEXION

| | |
|---|---|
| **Origin** | Based on the INFLEXION pattern. |
| **Constraint** | SUM_MIN_INFLEXION(VALUE, VARIABLES) |
| **Arguments** | VALUE      : dvar<br>VARIABLES  : collection(var−dvar) |

**Restrictions**

$$\mathtt{sv} \leq 2 \vee \mathtt{rv} \leq 1 \Rightarrow \mathtt{VALUE} = 0$$

$$\bigvee \left( \begin{array}{l} \mathtt{VALUE} = 0, \\ \mathtt{VALUE} \geq \min(\mathtt{minv}①, \mathtt{minv} * \lfloor (\mathtt{np}+1)/2 \rfloor + (\mathtt{minv}+1) * \lfloor \mathtt{np}/2 \rfloor ②) \end{array} \right)$$

$$\bigvee \left( \begin{array}{l} \mathtt{VALUE} = 0, \\ \mathtt{VALUE} \leq \max(\mathtt{maxv}③, \mathtt{maxv} * \lfloor (\mathtt{np}+1)/2 \rfloor + (\mathtt{maxv}-1) * \lfloor \mathtt{np}/2 \rfloor ④) \end{array} \right)$$

required(VARIABLES, var)

where

$$\mathtt{sv} = |\mathtt{VARIABLES}|$$
$$\mathtt{np} = \max(0, \mathtt{sv} - 2)$$
$$\mathtt{minv} = \mathtt{minval}(\mathtt{VARIABLES.var})$$
$$\mathtt{maxv} = \mathtt{maxval}(\mathtt{VARIABLES.var})$$
$$\mathtt{rv} = \mathtt{range}(\mathtt{VARIABLES.var})$$

**Purpose**

VALUE is the sum of all minimum values in each occurrence of the INFLEXION pattern in the time-series given by the VARIABLES collection. If the pattern does not occur, VALUE takes the default value 0.

An occurrence of the pattern INFLEXION is the *maximal* subsequence which matches the regular expression $< (< | =)^* > | > (> | =)^* <$.

Assume that the occurrence of the pattern INFLEXION starts at position $i$ and ends at position $j$. The feature MIN computes the minimum of the values from index $i + 1$ to index $j$.

| | |
|---|---|
| **Example** | $(26, \langle 1, 2, 6, 6, 4, 4, 3, 5, 2, 5, 1, 5, 3, 3, 4, 4 \rangle)$ |

Figure 3.1196 provides an example where the SUM_MIN_INFLEXION $(26, [1, 2, 6, 6, 4, 4, 3, 5, 2, 5, 1, 5, 3, 3, 4, 4])$ constraint holds.

| | |
|---|---|
| **Typical** | $\|\mathtt{VARIABLES}\| > 2$<br>range(VARIABLES.var) > 1 |
| **Symmetry** | Items of VARIABLES can be reversed. |
| **Arg. properties** | Functional dependency: VALUE determined by VARIABLES. |

Figure 3.1196: Illustrating the SUM_MIN_INFLEXION constraint of the **Example** slot

**Automaton**         Figures 3.1197 and 3.1198 respectively depict the automaton associated with the constraint
                      SUM_MIN_INFLEXION and its simplified form.

Figure 3.1197: Automaton for the SUM_MIN_INFLEXION constraint obtained by ap-
plying decoration Table 2.35 to the seed transducer of the INFLEXION pattern where
default is 0 (transition $r \rightarrow t$ has the same accumulators updates as transition $t \rightarrow r$)

Figure 3.1198: Automaton for the SUM_MIN_INFLEXION constraint obtained by ap-
plying decoration Table 2.25 to the seed transducer of the INFLEXION pattern where
default is 0 (transition $r \rightarrow t$ has the same accumulators updates as transition $t \rightarrow r$)

## 3.578  SUM_MIN_STRICTLY_DECREASING_SEQUENCE

▶   ◁ ◀        DESCRIPTION            AUTOMATON

**Origin**            Based on the STRICTLY_DECREASING_SEQUENCE pattern.

**Constraint**        SUM_MIN_STRICTLY_DECREASING_SEQUENCE(VALUE, VARIABLES)

**Arguments**         VALUE      : dvar
                      VARIABLES  : collection(var−dvar)

**Restrictions**      $sv \leq 1 \lor rv \leq 1 \Rightarrow VALUE = 0$
                      $VALUE = 0 \lor VALUE \geq \min(minv①, minv * np②)$
                      $VALUE = 0 \lor VALUE \leq \max(maxv - 1③, (maxv - 1) * np④)$
                      required(VARIABLES, var)
                      where
                      $sv = |VARIABLES|$
                      $np = \lfloor sv/2 \rfloor$
                      $minv = \text{minval}(VARIABLES.var)$
                      $maxv = \text{maxval}(VARIABLES.var)$
                      $rv = \text{range}(VARIABLES.var)$

**Purpose**           VALUE is the sum of all minimum values in each occurrence of the STRICTLY_DECREASING_SEQUENCE pattern in the time-series given by the VARIABLES collection. If the pattern does not occur, VALUE takes the default value 0.

                      An occurrence of the pattern STRICTLY_DECREASING_SEQUENCE is the *maximal* sub-sequence which matches the regular expression $>^+$.

                      Assume that the occurrence of the pattern STRICTLY_DECREASING_SEQUENCE starts at position $i$ and ends at position $j$. The feature MIN computes the minimum of the values from index $i$ to index $j + 1$.

**Example**           $(6, \langle 4, 4, 6, 4, 1, 1, 3, 4, 4, 6, 6, 5, 2, 2, 4, 3 \rangle)$

                      Figure 3.1199 provides an example where the SUM_MIN_STRICTLY_DECREASING_SEQUENCE $(6, [4, 4, 6, 4, 1, 1, 3, 4, 4, 6, 6, 5, 2, 2, 4, 3])$ constraint holds.

**Typical**           $|VARIABLES| > 1$
                      $\text{range}(VARIABLES.var) > 1$

**Arg. properties**   Functional dependency: VALUE determined by VARIABLES.

Figure 3.1199: Illustrating the SUM_MIN_STRICTLY_DECREASING_SEQUENCE constraint of the **Example** slot

**Automaton**     Figures 3.1200 and 3.1201 respectively depict the automaton associated with the constraint
SUM_MIN_STRICTLY_DECREASING_SEQUENCE and its simplified form.

Figure 3.1200:  Automaton for the SUM_MIN_STRICTLY_DECREASING_SEQUENCE
constraint obtained by applying decoration Table 2.35 to the seed transducer of the
STRICTLY_DECREASING_SEQUENCE pattern where default is 0

Figure 3.1201:  Automaton for the SUM_MIN_STRICTLY_DECREASING_SEQUENCE
constraint obtained by applying decoration Table 2.24 to the seed transducer of the
STRICTLY_DECREASING_SEQUENCE pattern where default is 0

|     | $s$                                         | $r$                                                                                       |
| --- | ------------------------------------------- | ----------------------------------------------------------------------------------------- |
| $s$ | $\overrightarrow{C} + \overleftarrow{C}$    | $\overrightarrow{C} + \overleftarrow{C}$                                                  |
| $r$ | $\overrightarrow{C} + \overleftarrow{C}$    | $\min(\overrightarrow{C}, \overleftarrow{C}, \overrightarrow{D}, \overleftarrow{D}, \texttt{VAR}_i)$ |

Table 3.164: Glue matrix for the SUM_MIN_STRICTLY_DECREASING_SEQUENCE constraint defined as the composition of the STRICTLY_DECREASING_SEQUENCE pattern, the feature MIN, and the aggregator sum; cells of the glue matrix are coloured with the colour of the constituent to which they are related.

## 3.579   SUM_MIN_STRICTLY_INCREASING_SEQUENCE

▶     ◁ ◀        DESCRIPTION          AUTOMATON

| | |
|---|---|
| **Origin** | Based on the STRICTLY_INCREASING_SEQUENCE pattern. |
| **Constraint** | SUM_MIN_STRICTLY_INCREASING_SEQUENCE(VALUE, VARIABLES) |
| **Arguments** | VALUE       : dvar<br>VARIABLES  : collection(var−dvar) |

**Restrictions**

$\text{sv} \leq 1 \lor \text{rv} \leq 1 \Rightarrow \text{VALUE} = 0$

$\text{VALUE} = 0 \lor \text{VALUE} \geq \min(\text{minv}①, \text{minv} * \text{np}②)$

$\text{VALUE} = 0 \lor \text{VALUE} \leq \max(\text{maxv} - 1③, (\text{maxv} - 1) * \text{np}④)$

required(VARIABLES, var)

where

$\text{sv} = |\text{VARIABLES}|$

$\text{np} = \lfloor \text{sv}/2 \rfloor$

$\text{minv} = \text{minval}(\text{VARIABLES.var})$

$\text{maxv} = \text{maxval}(\text{VARIABLES.var})$

$\text{rv} = \text{range}(\text{VARIABLES.var})$

**Purpose**

VALUE is the sum of all minimum values in each occurrence of the STRICTLY_INCREASING_SEQUENCE pattern in the time-series given by the VARIABLES collection. If the pattern does not occur, VALUE takes the default value 0.

An occurrence of the pattern STRICTLY_INCREASING_SEQUENCE is the *maximal* subsequence which matches the regular expression $<^+$.

Assume that the occurrence of the pattern STRICTLY_INCREASING_SEQUENCE starts at position $i$ and ends at position $j$. The feature MIN computes the minimum of the values from index $i$ to index $j + 1$.

**Example**

$(5, \langle 4, 3, 5, 5, 2, 1, 1, 2, 3, 4, 6, 6, 3, 1, 2, 3 \rangle)$

Figure 3.1202 provides an example where the SUM_MIN_STRICTLY_INCREASING_SEQUENCE $(5, [4, 3, 5, 5, 2, 1, 1, 2, 3, 4, 6, 6, 3, 1, 2, 3])$ constraint holds.

**Typical**

$|\text{VARIABLES}| > 1$

$\text{range}(\text{VARIABLES.var}) > 1$

**Arg. properties**

Functional dependency: VALUE determined by VARIABLES.

Figure 3.1202: Illustrating the SUM_MIN_STRICTLY_INCREASING_SEQUENCE constraint of the **Example** slot

**Automaton**  Figures 3.1203 and 3.1204 respectively depict the automaton associated with the constraint SUM_MIN_STRICTLY_INCREASING_SEQUENCE and its simplified form.

Figure 3.1203: Automaton for the SUM_MIN_STRICTLY_INCREASING_SEQUENCE constraint obtained by applying decoration Table 2.35 to the seed transducer of the STRICTLY_INCREASING_SEQUENCE pattern where default is 0

Figure 3.1204: Automaton for the SUM_MIN_STRICTLY_INCREASING_SEQUENCE constraint obtained by applying decoration Table 2.37 to the seed transducer of the STRICTLY_INCREASING_SEQUENCE pattern where default is 0

| | $s$ | $r$ |
|---|---|---|
| $s$ | $\overrightarrow{C} + \overleftarrow{C}$ | $\overrightarrow{C} + \overleftarrow{C}$ |
| $r$ | $\overrightarrow{C} + \overleftarrow{C}$ | $\min(\overrightarrow{C}, \overleftarrow{C}, \overrightarrow{D}, \overleftarrow{D}, \texttt{VAR}_i)$ |

Table 3.165: Glue matrix for the SUM_MIN_STRICTLY_INCREASING_SEQUENCE constraint defined as the composition of the STRICTLY_INCREASING_SEQUENCE pattern, the feature MIN, and the aggregator sum; cells of the glue matrix are coloured with the colour of the constituent to which they are related.

## 3.580   SUM_MIN_VALLEY

▶   ◁ ◀      DESCRIPTION          AUTOMATON

$> (= | >)^* (< | =)^* <$

| | |
|---|---|
| **Origin** | Based on the VALLEY pattern. |
| **Constraint** | SUM_MIN_VALLEY(VALUE, VARIABLES) |
| **Arguments** | VALUE      : dvar<br>VARIABLES : collection(var−dvar) |

**Restrictions**

$\text{sv} \leq 2 \vee \text{rv} \leq 1 \Rightarrow \text{VALUE} = 0$

$\text{VALUE} = 0 \vee \text{VALUE} \geq \min(\text{minv}①, \text{minv} * \text{np}②)$

$\text{VALUE} = 0 \vee \text{VALUE} \leq \max(\text{maxv} - 1③, (\text{maxv} - 1) * \text{np}④)$

required(VARIABLES, var)

where

$\text{sv} = |\text{VARIABLES}|$

$\text{np} = \max(0, \lfloor (\text{sv} - 1)/2 \rfloor)$

$\text{minv} = \text{minval}(\text{VARIABLES.var})$

$\text{maxv} = \text{maxval}(\text{VARIABLES.var})$

$\text{rv} = \text{range}(\text{VARIABLES.var})$

**Purpose**

VALUE is the sum of all minimum values in each occurrence of the VALLEY pattern in the time-series given by the VARIABLES collection. If the pattern does not occur, VALUE takes the default value 0.

An occurrence of the pattern VALLEY is the *maximal* subsequence which matches the regular expression $> (= | >)^* (< | =)^* <$.

Assume that the occurrence of the pattern VALLEY starts at position $i$ and ends at position $j$. The feature MIN computes the minimum of the values from index $i + 1$ to index $j$.

**Example**

$(10, \langle 1, 3, 7, 4, 3, 6, 6, 5, 3, 3, 2, 6, 5, 5, 5, 7 \rangle)$

Figure 3.1205 provides an example where the SUM_MIN_VALLEY $(10, [1, 3, 7, 4, 3, 6, 6, 5, 3, 3, 2, 6, 5, 5, 5, 7])$ constraint holds.

**Typical**

$|\text{VARIABLES}| > 2$

range(VARIABLES.var) $> 1$

**Symmetry**

Items of VARIABLES can be reversed.

**Arg. properties**

Functional dependency: VALUE determined by VARIABLES.

Figure 3.1205: Illustrating the SUM_MIN_VALLEY constraint of the **Example** slot

**Automaton**    Figures 3.1206 and 3.1207 respectively depict the automaton associated with the constraint SUM_MIN_VALLEY and its simplified form.

Figure 3.1206: Automaton for the SUM_MIN_VALLEY constraint obtained by applying decoration Table 2.35 to the seed transducer of the VALLEY pattern where default is 0

Figure 3.1207: Automaton for the SUM_MIN_VALLEY constraint obtained by applying decoration Table 2.37 to the seed transducer of the VALLEY pattern where default is 0

|   | $s$ | $r$ | $t$ |
|---|---|---|---|
| $s$ | $\overrightarrow{C}+\overleftarrow{C}$ | $\overrightarrow{C}+\overleftarrow{C}$ | $\overrightarrow{C}+\overleftarrow{C}$ |
| $r$ | $\overrightarrow{C}+\overleftarrow{C}$ | $\min(\overrightarrow{D},\overleftarrow{D},\mathtt{VAR}_i)$ | $\min(\overleftarrow{C},\overrightarrow{D},\overleftarrow{D},\mathtt{VAR}_i)$ |
| $t$ | $\overrightarrow{C}+\overleftarrow{C}$ | $\min(\overrightarrow{C},\overrightarrow{D},\overleftarrow{D},\mathtt{VAR}_i)$ | $\overrightarrow{C}+\overleftarrow{C}$ |

Table 3.166: Glue matrix for the SUM_MIN_VALLEY constraint defined as the composition of the VALLEY pattern , the feature MIN , and the aggregator sum ; cells of the glue matrix are coloured with the colour of the constituent to which they are related.

## 3.581 SUM_MIN_ZIGZAG

**DESCRIPTION**        **AUTOMATON**

**Origin**        Based on the ZIGZAG pattern.

**Constraint**        SUM_MIN_ZIGZAG(VALUE, VARIABLES)

**Arguments**        VALUE : dvar
                     VARIABLES : collection(var−dvar)

**Restrictions**

$$sv \leq 3 \vee rv \leq 1 \Rightarrow VALUE = 0$$

$$rv = 2 \Rightarrow VALUE = 0 \vee VALUE \geq \min(minv, minv * np1)$$

$$rv \geq 3 \Rightarrow VALUE = 0 \vee VALUE \geq \min(minv, \min(minv * np1, minv * np2))$$

$$VALUE \leq \max_{q \in [0, qub]} \max \left( 0, \sum \left( \prod \left( \begin{array}{l} \left\lfloor (sv - 3 * q)/4 \right\rfloor, \\ maxv - 1 \end{array} \right), \\ q * (maxv - 2), \\ \min \left( 1, \max \left( maxv - minv - 1, 0 \right) \right), \\ \min \left( 1, \max \left( 0, q * (maxv - 2) \right) \right), \\ \min \left( \begin{array}{l} q, \\ \left( (sv - \lfloor (sv - 3 * q)/4 \rfloor) * 4 - \\ q * 3 \end{array} \right) \mod 3 \right) \end{array} \right) \right) \right)$$

required(VARIABLES, var)
where
$$sv = |VARIABLES|$$
$$rv = range(VARIABLES.var)$$
$$np1 = \lfloor sv/4 \rfloor$$
$$np2 = \max(0, \lfloor (sv - 1)/3 \rfloor)$$
$$minv = minval(VARIABLES.var)$$
$$maxv = maxval(VARIABLES.var)$$
$$qub = \min(1, \max(maxv - minv - 1, 0)) * np2$$

**Purpose**        VALUE is the sum of all minimum values in each occurrence of the ZIGZAG pattern in the time-series given by the VARIABLES collection. If the pattern does not occur, VALUE takes the default value 0.

An occurrence of the pattern ZIGZAG is the *maximal* subsequence which matches the regular expression $(< >)^+(< | < >) | (> <)^+(> | > <)$.

Assume that the occurrence of the pattern ZIGZAG starts at position $i$ and ends at position $j$. The feature MIN computes the minimum of the values from index $i + 1$ to index $j$.

**Example**        $(3, \langle 4, 1, 3, 1, 4, 6, 1, 5, 5, 2, 7, 2, 3, 1, 6, 1 \rangle)$

Figure 3.1208 provides an example where the SUM_MIN_ZIGZAG
$(3, [4, 1, 3, 1, 4, 6, 1, 5, 5, 2, 7, 2, 3, 1, 6, 1])$ constraint holds.

Figure 3.1208: Illustrating the SUM_MIN_ZIGZAG constraint of the **Example** slot

| Typical | $|\texttt{VARIABLES}| > 3$ |
|---|---|
| | $\texttt{range}(\texttt{VARIABLES.var}) > 1$ |

| Symmetry | Items of VARIABLES can be reversed. |
|---|---|

| Arg. properties | Functional dependency: VALUE determined by VARIABLES. |
|---|---|

**Automaton**        Figures 3.1209 and 3.1210 respectively depict the automaton associated with the constraint
                     SUM_MIN_ZIGZAG and its simplified form.

Figure 3.1209: Automaton for the SUM_MIN_ZIGZAG constraint obtained by applying decoration Table 2.35 to the seed transducer of the ZIGZAG pattern where default is 0; (1) missing transitions from $a, b, c, d, e, f$ to $s$ are labelled by $=$; (2) on transitions from $b, c, e, f$ to $s$ the accumulator $D$ is reset to its initial value; (3) on transitions from $c, f$ to $s$ the accumulator $R$ is updated wrt $C$ and the accumulator $C$ is reset to its initial value

Figure 3.1210: Automaton for the SUM_MIN_ZIGZAG constraint obtained by applying decoration Table 2.23 to the seed transducer of the ZIGZAG pattern where default is 0; missing transitions from $a, b, c, d, e, f$ to $s$ are labelled by $=$; (2) on transitions from $b, c, e, f$ to $s$ the accumulator $D$ is reset to its initial value; (3) on transitions from $c, f$ to $s$ the accumulator $R$ is updated wrt $C$ and the accumulator $C$ is reset to its initial value.

| | s | a | b | c | d | e | f |
|---|---|---|---|---|---|---|---|
| s | $\overleftarrow{C}+\overrightarrow{C}$ | $\overrightarrow{C}+\overleftarrow{C}$ | $\overrightarrow{C}+\overleftarrow{C}$ | $\overrightarrow{C}+\overleftarrow{C}$ | $\overrightarrow{C}+\overleftarrow{C}$ | $\overrightarrow{C}+\overleftarrow{C}$ | $\overrightarrow{C}+\overleftarrow{C}$ |
| a | $\overrightarrow{C}+\overleftarrow{C}$ | $\overrightarrow{C}+\overleftarrow{C}$ | $\overrightarrow{C}+\overleftarrow{C}$ | $\min(\overrightarrow{C},\overrightarrow{D},\overleftarrow{D},\text{VAR}_i)$ | $\overrightarrow{C}+\overleftarrow{C}$ | $\min(\overrightarrow{D},\overleftarrow{D},\text{VAR}_i)$ | $\overrightarrow{C}+\overleftarrow{C}$ |
| b | $\overrightarrow{C}+\overleftarrow{C}$ | $\overrightarrow{C}+\overleftarrow{C}$ | $\min(\overrightarrow{D},\overleftarrow{D},\text{VAR}_i)$ | $\overrightarrow{C}+\overleftarrow{C}$ | $\min(\overrightarrow{D},\overleftarrow{D},\text{VAR}_i)$ | $\overrightarrow{C}+\overleftarrow{C}$ | $\min(\overleftarrow{C},\overrightarrow{D},\overleftarrow{D},\text{VAR}_i)$ |
| c | $\overrightarrow{C}+\overleftarrow{C}$ | $\min(\overrightarrow{C},\overrightarrow{D},\overleftarrow{D},\text{VAR}_i)$ | $\overrightarrow{C}+\overleftarrow{C}$ | $\min(\overrightarrow{C},\overrightarrow{C},\overrightarrow{D},\overleftarrow{D},\text{VAR}_i)$ | $\overrightarrow{C}+\overleftarrow{C}$ | $\min(\overrightarrow{C},\overrightarrow{D},\overleftarrow{D},\text{VAR}_i)$ | $\overrightarrow{C}+\overleftarrow{C}$ |
| d | $\overrightarrow{C}+\overleftarrow{C}$ | $\overrightarrow{C}+\overleftarrow{C}$ | $\min(\overrightarrow{D},\overleftarrow{D},\text{VAR}_i)$ | $\overrightarrow{C}+\overleftarrow{C}$ | $\overrightarrow{C}+\overleftarrow{C}$ | $\overrightarrow{C}+\overleftarrow{C}$ | $\min(\overleftarrow{C},\overrightarrow{D},\overleftarrow{D},\text{VAR}_i)$ |
| e | $\overrightarrow{C}+\overleftarrow{C}$ | $\min(\overrightarrow{D},\overleftarrow{D},\text{VAR}_i)$ | $\overrightarrow{C}+\overleftarrow{C}$ | $\min(\overleftarrow{C},\overrightarrow{D},\overleftarrow{D},\text{VAR}_i)$ | $\overrightarrow{C}+\overleftarrow{C}$ | $\min(\overrightarrow{D},\overleftarrow{D},\text{VAR}_i)$ | $\overrightarrow{C}+\overleftarrow{C}$ |
| f | $\overrightarrow{C}+\overleftarrow{C}$ | $\overrightarrow{C}+\overleftarrow{C}$ | $\min(\overleftarrow{C},\overrightarrow{D},\overleftarrow{D},\text{VAR}_i)$ | $\overrightarrow{C}+\overleftarrow{C}$ | $\min(\overleftarrow{C},\overrightarrow{D},\overleftarrow{D},\text{VAR}_i)$ | $\overrightarrow{C}+\overleftarrow{C}$ | $\min(\overleftarrow{C},\overleftarrow{C},\overrightarrow{D},\overleftarrow{D},\text{VAR}_i)$ |

Table 3.167: Glue matrix for the SUM_MIN_ZIGZAG constraint defined as the composition of the ZIGZAG pattern, the feature MIN, and the aggregator sum; cells of the glue matrix are coloured with the colour of the constituent to which they are related.

# 3.582 SUM_RANGE_DECREASING

| | |
|---|---|
| **Origin** | Based on the DECREASING pattern. |

**Constraint**     SUM_RANGE_DECREASING(VALUE, VARIABLES)

**Arguments**
```
VALUE      : dvar
VARIABLES  : collection(var−dvar)
```

**Restrictions**

$$\mathtt{sv} \leq 1 \vee \mathtt{rv} \leq 1 \Rightarrow \mathtt{VALUE} = 0$$
$$\mathtt{VALUE} \geq 0$$
$$\mathtt{VALUE} \leq \lfloor \mathtt{sv}/2 \rfloor * (\mathtt{rv} - 1) \textcircled{1}$$
$$\mathtt{required}(\mathtt{VARIABLES}, \mathtt{var})$$
where
$$\mathtt{rv} = \mathtt{range}(\mathtt{VARIABLES.var})$$
$$\mathtt{sv} = |\mathtt{VARIABLES}|$$

**Purpose**

VALUE is the sum of the differences between the largest and smallest value in each occurrence of the DECREASING pattern in the time-series given by the VARIABLES collection. If the pattern does not occur, VALUE takes the default value 0.

An occurrence of the pattern DECREASING is the subsequence which matches the regular expression >.

Assume that the occurrence of the pattern DECREASING starts at position $i$ and ends at position $j$. The feature RANGE computes the range of the values from index $i$ to index $j + 1$.

**Example**

$$(9, \langle 3, 4, 2, 2, 5, 6, 6, 4, 4, 3, 1, 1, 4, 6, 4, 4 \rangle)$$

Figure 3.1211 provides an example where the SUM_RANGE_DECREASING $(9, [3, 4, 2, 2, 5, 6, 6, 4, 4, 3, 1, 1, 4, 6, 4, 4])$ constraint holds.

**Typical**

$$|\mathtt{VARIABLES}| > 1$$
$$\mathtt{range}(\mathtt{VARIABLES.var}) > 1$$

**Symmetry**     One and the same constant can be added to the var attribute of all items of VARIABLES.

**Arg. properties**     Functional dependency: VALUE determined by VARIABLES.

Figure 3.1211: Illustrating the SUM_RANGE_DECREASING constraint of the **Example** slot

**Automaton**     Figures 3.1212 and 3.1213 respectively depict the automaton associated with the constraint SUM_RANGE_DECREASING and its simplified form.

Figure 3.1212: Automaton for the SUM_RANGE_DECREASING constraint obtained by applying decoration Table 2.46 to the seed transducer of the DECREASING pattern where default is 0

Figure 3.1213: Automaton for the SUM_RANGE_DECREASING constraint obtained by applying decoration Table 2.44 to the seed transducer of the DECREASING pattern where default is 0

# 3.583　SUM_RANGE_DECREASING_SEQUENCE

　　　**DESCRIPTION**　　　　　**AUTOMATON**

**Origin**　　　Based on the DECREASING_SEQUENCE pattern.

**Constraint**　　　SUM_RANGE_DECREASING_SEQUENCE(VALUE, VARIABLES)

**Arguments**
```
VALUE      : dvar
VARIABLES  : collection(var−dvar)
```

**Restrictions**

$\mathtt{sv} \leq 1 \lor \mathtt{rv} \leq 1 \Rightarrow \mathtt{VALUE} = 0$
$\mathtt{VALUE} \geq 0$
$\mathtt{VALUE} \leq \lfloor \mathtt{sv}/2 \rfloor * (\mathtt{rv} - 1)$①
required(VARIABLES, var)
where
$\mathtt{rv} = $range(VARIABLES.var)
$\mathtt{sv} = |\mathtt{VARIABLES}|$

**Purpose**

VALUE is the sum of the differences between the largest and smallest value in each occurrence of the DECREASING_SEQUENCE pattern in the time-series given by the VARIABLES collection. If the pattern does not occur, VALUE takes the default value 0.

An occurrence of the pattern DECREASING_SEQUENCE is the *maximal* subsequence which matches the regular expression $> (> | =)^* > | >$.

Assume that the occurrence of the pattern DECREASING_SEQUENCE starts at position $i$ and ends at position $j$. The feature RANGE computes the range of the values from index $i$ to index $j + 1$.

**Example**

$(9, \langle 3, 4, 2, 2, 5, 6, 6, 4, 4, 3, 1, 1, 4, 6, 4, 4 \rangle)$

Figure 3.1214 provides an example where the SUM_RANGE_DECREASING_SEQUENCE $(9, [3, 4, 2, 2, 5, 6, 6, 4, 4, 3, 1, 1, 4, 6, 4, 4])$ constraint holds.

**Typical**

$|\mathtt{VARIABLES}| > 1$
range(VARIABLES.var) $> 1$

**Symmetry**　　　One and the same constant can be added to the var attribute of all items of VARIABLES.

**Arg. properties**　　　Functional dependency: VALUE determined by VARIABLES.

Figure 3.1214: Illustrating the SUM_RANGE_DECREASING_SEQUENCE constraint of the **Example** slot

**Automaton**     Figures 3.1215 and 3.1216 respectively depict the automaton associated with the constraint SUM_RANGE_DECREASING_SEQUENCE and its simplified form.

Figure 3.1215: Automaton for the SUM_RANGE_DECREASING_SEQUENCE constraint obtained by applying decoration Table 2.46 to the seed transducer of the DECREAS-ING_SEQUENCE pattern where default is $0$

Figure 3.1216: Automaton for the SUM_RANGE_DECREASING_SEQUENCE constraint obtained by applying decoration Table 2.44 to the seed transducer of the DECREAS-ING_SEQUENCE pattern where default is $0$

# 3.584   SUM_RANGE_INCREASING

▶    ◁ ◀        **DESCRIPTION**          **AUTOMATON**

**Origin**          Based on the INCREASING pattern.

**Constraint**      SUM_RANGE_INCREASING(VALUE, VARIABLES)

**Arguments**       VALUE      : dvar
                    VARIABLES  : collection(var−dvar)

**Restrictions**    $\mathrm{sv} \leq 1 \vee \mathrm{rv} \leq 1 \Rightarrow \mathtt{VALUE} = 0$
                    $\mathtt{VALUE} \geq 0$
                    $\mathtt{VALUE} \leq \lfloor \mathrm{sv}/2 \rfloor * (\mathrm{rv} - 1)$①
                    required(VARIABLES, var)
                    where
                    $\mathrm{rv} = \mathrm{range}(\mathtt{VARIABLES.var})$
                    $\mathrm{sv} = |\mathtt{VARIABLES}|$

**Purpose**         VALUE is the sum of the differences between the largest and smallest value in each occur-
                    rence of the INCREASING pattern in the time-series given by the VARIABLES collection.
                    If the pattern does not occur, VALUE takes the default value 0.
                    An occurrence of the pattern INCREASING is the subsequence which matches the regular
                    expression <.
                    Assume that the occurrence of the pattern INCREASING starts at position $i$ and ends at
                    position $j$. The feature RANGE computes the range of the values from index $i$ to index
                    $j + 1$.

**Example**         $(9, \langle 4, 3, 5, 5, 2, 1, 1, 3, 3, 4, 6, 6, 3, 1, 3, 3 \rangle)$

                    Figure 3.1217 provides an example where the SUM_RANGE_INCREASING
                    $(9, [4, 3, 5, 5, 2, 1, 1, 3, 3, 4, 6, 6, 3, 1, 3, 3])$ constraint holds.

**Typical**         $|\mathtt{VARIABLES}| > 1$
                    $\mathrm{range}(\mathtt{VARIABLES.var}) > 1$

**Symmetry**        One and the same constant can be added to the var attribute of all items of VARIABLES.

**Arg. properties** Functional dependency: VALUE determined by VARIABLES.

Figure 3.1217: Illustrating the SUM_RANGE_INCREASING constraint of the **Example** slot

**Automaton**    Figures 3.1218 and 3.1219 respectively depict the automaton associated with the constraint SUM_RANGE_INCREASING and its simplified form.

$$\left\{ \begin{array}{l} C \leftarrow \texttt{default} \\ H \leftarrow \texttt{VAR}_1 \\ R \leftarrow \texttt{default} \end{array} \right\}$$

$$\left\{ \begin{array}{l} H \leftarrow \texttt{VAR}_{i+1} \\ R \leftarrow R + |H - \texttt{VAR}_{i+1}| \end{array} \right\} \overset{<}{\underset{\geq}{\circlearrowleft}} s \quad \{ H \leftarrow \overset{\geq}{\texttt{VAR}_{i+1}} \}$$

$$\boxed{R + C}$$

Figure 3.1218: Automaton for the SUM_RANGE_INCREASING constraint obtained by applying decoration Table 2.46 to the seed transducer of the INCREASING pattern where default is 0

$$\{ R \leftarrow \texttt{default} \}$$

$$\{ R \leftarrow R + \max(0, \texttt{VAR}_{i+1} - \texttt{VAR}_i) \} \overset{<}{\underset{\geq}{\circlearrowleft}} s \quad \geq$$

$$\boxed{R}$$

Figure 3.1219: Automaton for the SUM_RANGE_INCREASING constraint obtained by applying decoration Table 2.45 to the seed transducer of the INCREASING pattern where default is 0; $R_i - R_{i-2} - X_i + X_{i-2} \geq 0$ and $R_i - R_{i-1} - X_i + X_{i-1} \geq 0$ are linear invariants.

# 3.585   SUM_RANGE_INCREASING_SEQUENCE

▶    ◁ ◀        **DESCRIPTION**            **AUTOMATON**

**Origin**            Based on the INCREASING_SEQUENCE pattern.

**Constraint**        SUM_RANGE_INCREASING_SEQUENCE(VALUE, VARIABLES)

**Arguments**         VALUE      : dvar
                      VARIABLES  : collection(var−dvar)

**Restrictions**      $\mathrm{sv} \leq 1 \vee \mathrm{rv} \leq 1 \Rightarrow \mathtt{VALUE} = 0$
                      $\mathtt{VALUE} \geq 0$
                      $\mathtt{VALUE} \leq \lfloor \mathrm{sv}/2 \rfloor * (\mathrm{rv} - 1)$ ①
                      required(VARIABLES, var)
                      where
                      $\mathrm{rv} =$ range(VARIABLES.var)
                      $\mathrm{sv} = |\mathtt{VARIABLES}|$

**Purpose**           VALUE is the sum of the differences between the largest and smallest value in each occur-
                      rence of the INCREASING_SEQUENCE pattern in the time-series given by the VARIABLES
                      collection. If the pattern does not occur, VALUE takes the default value 0.
                      An occurrence of the pattern INCREASING_SEQUENCE is the *maximal* subsequence
                      which matches the regular expression $< (< | =)^* < | <$.
                      Assume that the occurrence of the pattern INCREASING_SEQUENCE starts at position $i$
                      and ends at position $j$. The feature RANGE computes the range of the values from index
                      $i$ to index $j + 1$.

**Example**           $(9, \langle 4, 3, 5, 5, 2, 1, 1, 3, 3, 4, 6, 6, 3, 1, 3, 3 \rangle)$

                      Figure 3.1220 provides an example where the SUM_RANGE_INCREASING_SEQUENCE
                      $(9, [4, 3, 5, 5, 2, 1, 1, 3, 3, 4, 6, 6, 3, 1, 3, 3])$ constraint holds.

**Typical**           $|\mathtt{VARIABLES}| > 1$
                      range(VARIABLES.var) > 1

**Symmetry**          One and the same constant can be added to the var attribute of all items of VARIABLES.

**Arg. properties**   Functional dependency: VALUE determined by VARIABLES.

Figure 3.1220: Illustrating the SUM_RANGE_INCREASING_SEQUENCE constraint of the **Example** slot

**Automaton**    Figures 3.1221 and 3.1222 respectively depict the automaton associated with the constraint SUM_RANGE_INCREASING_SEQUENCE and its simplified form.

Figure 3.1221: Automaton for the SUM_RANGE_INCREASING_SEQUENCE constraint obtained by applying decoration Table 2.46 to the seed transducer of the INCREAS-ING_SEQUENCE pattern where default is 0

Figure 3.1222: Automaton for the SUM_RANGE_INCREASING_SEQUENCE constraint obtained by applying decoration Table 2.45 to the seed transducer of the INCREAS-ING_SEQUENCE pattern where default is 0

# 3.586  SUM_RANGE_STRICTLY_DECREASING_SEQUENCE

▶    ◁ ◀        DESCRIPTION              AUTOMATON

| | |
|---|---|
| **Origin** | Based on the STRICTLY_DECREASING_SEQUENCE pattern. |
| **Constraint** | SUM_RANGE_STRICTLY_DECREASING_SEQUENCE(VALUE, VARIABLES) |
| **Arguments** | VALUE      : dvar<br>VARIABLES  : collection(var−dvar) |

**Restrictions**

$$\mathtt{sv} \leq 1 \vee \mathtt{rv} \leq 1 \Rightarrow \mathtt{VALUE} = 0$$
$$\mathtt{VALUE} \geq 0$$
$$\mathtt{VALUE} \leq \lfloor \mathtt{sv}/2 \rfloor * (\mathtt{rv} - 1) \text{①}$$
required(VARIABLES, var)
where
$$\mathtt{rv} = \mathtt{range}(\mathtt{VARIABLES.var})$$
$$\mathtt{sv} = |\mathtt{VARIABLES}|$$

**Purpose**

VALUE is the sum of the differences between the largest and smallest value in each occurrence of the STRICTLY_DECREASING_SEQUENCE pattern in the time-series given by the VARIABLES collection. If the pattern does not occur, VALUE takes the default value 0.

An occurrence of the pattern STRICTLY_DECREASING_SEQUENCE is the *maximal* subsequence which matches the regular expression $>^+$.

Assume that the occurrence of the pattern STRICTLY_DECREASING_SEQUENCE starts at position $i$ and ends at position $j$. The feature RANGE computes the range of the values from index $i$ to index $j + 1$.

**Example**

$$(10, \langle 4, 4, 6, 4, 1, 1, 3, 4, 4, 6, 6, 5, 2, 2, 4, 3 \rangle)$$

Figure 3.1223 provides an example where the SUM_RANGE_STRICTLY_DECREASING_SEQUENCE $(10, [4, 4, 6, 4, 1, 1, 3, 4, 4, 6, 6, 5, 2, 2, 4, 3])$ constraint holds.

| | |
|---|---|
| **Typical** | $\lvert \mathtt{VARIABLES} \rvert > 1$<br>range(VARIABLES.var) $> 1$ |
| **Symmetry** | One and the same constant can be added to the var attribute of all items of VARIABLES. |
| **Arg. properties** | Functional dependency: VALUE determined by VARIABLES. |

Figure 3.1223: Illustrating the SUM_RANGE_STRICTLY_DECREASING_SEQUENCE constraint of the **Example** slot

**Automaton**　　　　Figures 3.1224 and 3.1225 respectively depict the automaton associated with the constraint
SUM_RANGE_STRICTLY_DECREASING_SEQUENCE and its simplified form.

Figure 3.1224: Automaton for the SUM_RANGE_STRICTLY_DECREASING_SEQUENCE constraint obtained by applying decoration Table 2.46 to the seed transducer of the STRICTLY_DECREASING_SEQUENCE pattern where default is 0

Figure 3.1225: Automaton for the SUM_RANGE_STRICTLY_DECREASING_SEQUENCE constraint obtained by applying decoration Table 2.44 to the seed transducer of the STRICTLY_DECREASING_SEQUENCE pattern where default is 0

# 3.587   SUM_RANGE_STRICTLY_INCREASING_SEQUENCE

▶   ◁ ◀    DESCRIPTION    AUTOMATON

**Origin**   Based on the STRICTLY_INCREASING_SEQUENCE pattern.

**Constraint**   SUM_RANGE_STRICTLY_INCREASING_SEQUENCE(VALUE, VARIABLES)

**Arguments**
```
VALUE     : dvar
VARIABLES : collection(var−dvar)
```

**Restrictions**
$\mathtt{sv} \leq 1 \lor \mathtt{rv} \leq 1 \Rightarrow \mathtt{VALUE} = 0$
$\mathtt{VALUE} \geq 0$
$\mathtt{VALUE} \leq \lfloor \mathtt{sv}/2 \rfloor * (\mathtt{rv} - 1)$①
required(VARIABLES, var)
where
$\mathtt{rv} =$ range(VARIABLES.var)
$\mathtt{sv} = |\mathtt{VARIABLES}|$

**Purpose**
VALUE is the sum of the differences between the largest and smallest value in each occurrence of the STRICTLY_INCREASING_SEQUENCE pattern in the time-series given by the VARIABLES collection. If the pattern does not occur, VALUE takes the default value 0.
An occurrence of the pattern STRICTLY_INCREASING_SEQUENCE is the *maximal* subsequence which matches the regular expression $<^+$.
Assume that the occurrence of the pattern STRICTLY_INCREASING_SEQUENCE starts at position $i$ and ends at position $j$. The feature RANGE computes the range of the values from index $i$ to index $j + 1$.

**Example**   $(9, \langle 4, 3, 5, 5, 2, 1, 1, 2, 3, 4, 6, 6, 3, 1, 2, 3 \rangle)$

Figure 3.1226 provides an example where the SUM_RANGE_STRICTLY_INCREASING_SEQUENCE $(9, [4, 3, 5, 5, 2, 1, 1, 2, 3, 4, 6, 6, 3, 1, 2, 3])$ constraint holds.

**Typical**
$|\mathtt{VARIABLES}| > 1$
range(VARIABLES.var) $> 1$

**Symmetry**   One and the same constant can be added to the var attribute of all items of VARIABLES.

**Arg. properties**   Functional dependency: VALUE determined by VARIABLES.

Figure 3.1226: Illustrating the SUM_RANGE_STRICTLY_INCREASING_SEQUENCE constraint of the **Example** slot

**Automaton** Figures 3.1227 and 3.1228 respectively depict the automaton associated with the constraint SUM_RANGE_STRICTLY_INCREASING_SEQUENCE and its simplified form.

Figure 3.1227: Automaton for the SUM_RANGE_STRICTLY_INCREASING_SEQUENCE constraint obtained by applying decoration Table 2.46 to the seed transducer of the STRICTLY_INCREASING_SEQUENCE pattern where default is $0$

Figure 3.1228: Automaton for the SUM_RANGE_STRICTLY_INCREASING_SEQUENCE constraint obtained by applying decoration Table 2.45 to the seed transducer of the STRICTLY_INCREASING_SEQUENCE pattern where default is $0$

## 3.588   SUM_SURF_BUMP_ON_DECREASING_SEQUENCE

    **DESCRIPTION**          **AUTOMATON**

| | |
|---|---|
| **Origin** | Based on the BUMP_ON_DECREASING_SEQUENCE pattern. |
| **Constraint** | SUM_SURF_BUMP_ON_DECREASING_SEQUENCE(VALUE, VARIABLES) |
| **Arguments** | VALUE      : dvar<br>VARIABLES : collection(var−dvar) |

**Restrictions**

$$sv \leq 5 \vee rv \leq 2 \Rightarrow \text{VALUE} = 0$$
$$\text{VALUE} = 0 \vee \text{VALUE} \geq \min(3 * \text{minv} + 3, (3 * \text{minv} + 3) * \text{np})$$
$$\text{VALUE} = 0 \vee \text{VALUE} \leq \max(3 * \text{maxv} - 3①, (3 * \text{maxv} - 3) * \text{np}②)$$
required(VARIABLES, var)
where
$$sv = |\text{VARIABLES}|$$
$$np = \max(0, \lfloor(sv - 3)/3\rfloor)$$
$$minv = \text{minval}(\text{VARIABLES.var})$$
$$maxv = \text{maxval}(\text{VARIABLES.var})$$
$$rv = \text{range}(\text{VARIABLES.var})$$

VALUE ≤ −**6** = −1 ∗ 3 − 3 ①          VALUE ≤ **18** = (3 ∗ 4 − 3) ∗ 2 ②

**Purpose**

VALUE is the sum of the surface of occurrences of the BUMP_ON_DECREASING_SEQUENCE pattern in the time-series given by the VARIABLES collection. If the pattern does not occur, VALUE takes the default value 0.

An occurrence of the pattern BUMP_ON_DECREASING_SEQUENCE is the subsequence which matches the regular expression >><>>.

Assume that the occurrence of the pattern BUMP_ON_DECREASING_SEQUENCE starts at position $i$ and ends at position $j$. The feature SURF computes the sum of the values from index $i + 2$ to index $j$.

**Example**

$$(27, \langle 7, 6, 5, 6, 5, 4, 1, 4, 7, 5, 4, 2, 5, 4, 3, 3 \rangle)$$

Figure 3.1229 provides an example where the SUM_SURF_BUMP_ON_DECREASING_SEQUENCE $(27, [7, 6, 5, 6, 5, 4, 1, 4, 7, 5, 4, 2, 5, 4, 3, 3])$ constraint holds.

**Typical**

$$|\text{VARIABLES}| > 5$$
$$\text{range}(\text{VARIABLES.var}) > 2$$

**Arg. properties**

Functional dependency: VALUE determined by VARIABLES.

Figure 3.1229: Illustrating the SUM_SURF_BUMP_ON_DECREASING_SEQUENCE constraint of the **Example** slot

**Automaton**  Figures 3.1230 and 3.1231 respectively depict the automaton associated with the constraint SUM_SURF_BUMP_ON_DECREASING_SEQUENCE and its simplified form.

Figure 3.1230: Automaton for the SUM_SURF_BUMP_ON_DECREASING_SEQUENCE constraint obtained by applying decoration Table 2.35 to the seed transducer of the BUMP_ON_DECREASING_SEQUENCE pattern where default is 0

Figure 3.1231: Automaton for the SUM_SURF_BUMP_ON_DECREASING_SEQUENCE constraint obtained by applying decoration Table 2.25 to the seed transducer of the BUMP_ON_DECREASING_SEQUENCE pattern where default is 0

### 3.589   SUM_SURF_DECREASING

▶    ◁ ◀        **DESCRIPTION**        **AUTOMATON**

**Origin**        Based on the DECREASING pattern.

**Constraint**    SUM_SURF_DECREASING(VALUE, VARIABLES)

**Arguments**     VALUE      : dvar
                  VARIABLES  : collection(var−dvar)

**Restrictions**

$$\text{sv} \leq 1 \lor \text{rv} \leq 1 \Rightarrow \text{VALUE} = 0$$

$$\text{VALUE} \geq \min_{\text{q} \in [\text{lb1}, \text{ub1}]} \min \left( 0, \sum \left( \begin{array}{l} \text{q} * (\sum \left( \begin{array}{l} 2*(\text{np}-1)*\text{minv}, \\ \text{np}*\text{np} \end{array} \right) - \\ 2*\text{np} \end{array} +1), \\ \prod \left( \begin{array}{l} \max\left(\ 0, \min\left(\ 1, \text{np}-1\ \right)\ \right), \\ \text{sv} \bmod \text{np} * \text{q}, \\ \sum \left( \begin{array}{l} 2*\text{minv}-1, \\ 2*\text{np} \end{array} \right) \end{array} \right), \\ \prod \left( \begin{array}{l} \max\left(\ 0, \min\left(\ 1, 2-\text{np}\ \right)\ \right), \\ 2*\text{minv}+1, \\ \text{q}-1 \end{array} \right) \end{array} \right) \right) \quad ①$$

$$\text{VALUE} \leq \max_{\text{q} \in [\text{lb2}, \text{ub2}]} \max \left( 0, \sum \left( \begin{array}{l} \text{q} * (\sum \left( \begin{array}{l} 2*(\text{np}-1)*\text{maxv}- \\ \text{np}*\text{np} \\ 2*\text{np} \end{array} \right) -1), \\ \prod \left( \begin{array}{l} \max\left(\ 0, \min\left(\ 1, \text{np}-1\ \right)\ \right), \\ \text{sv} \bmod \text{np} * \text{q}, \\ 2*\text{maxv}+1- \\ 2*\text{np} \end{array} \right), \\ \prod \left( \begin{array}{l} \max\left(\ 0, \min\left(\ 1, 2-\text{np}\ \right)\ \right), \\ 2*\text{maxv}-1, \\ \text{q}-1 \end{array} \right) \end{array} \right) \right)$$

required(VARIABLES, var)

where

$$\text{sv} = |\text{VARIABLES}|$$
$$\text{np} = \lfloor \text{sv}/\text{q} \rfloor$$
$$\text{maxv} = \text{maxval}(\text{VARIABLES.var})$$
$$\text{minv} = \text{minval}(\text{VARIABLES.var})$$
$$\text{rv} = \text{range}(\text{VARIABLES.var})$$

$$\text{lb1} = \min \left( \begin{array}{l} \lfloor \text{sv}/2 \rfloor + 1, \\ \sum \left( \begin{array}{l} \lfloor \text{sv}/\min(\min(\text{sv}, \text{rv}), |\text{minv}|+1) \rfloor, \\ \min(1, \text{sv} \bmod \min(\min(\text{sv}, \text{rv}), |\text{minv}|+1)) \end{array} \right) \end{array} \right)$$
$$\text{ub1} = \lfloor \text{sv}/2 \rfloor + 1$$
$$\text{lb2} = \min \left( \begin{array}{l} \lfloor \text{sv}/2 \rfloor + 1, \\ \sum \left( \begin{array}{l} \lfloor \text{sv}/\min(\min(\text{sv}, \text{rv}), |\text{maxv}|+1) \rfloor, \\ \min(1, \text{sv} \bmod \min(\min(\text{sv}, \text{rv}), |\text{maxv}|+1)) \end{array} \right) \end{array} \right)$$
$$\text{ub2} = \lfloor \text{sv}/2 \rfloor + 1$$

**Purpose**

VALUE is the sum of the surface of occurrences of the DECREASING pattern in the time-series given by the VARIABLES collection. If the pattern does not occur, VALUE takes the default value 0.

An occurrence of the pattern DECREASING is the subsequence which matches the regular expression >.

Assume that the occurrence of the pattern DECREASING starts at position $i$ and ends at position $j$. The feature SURF computes the sum of the values from index $i$ to index $j+1$.

**Example**

$(37, \langle 3, 4, 2, 2, 5, 6, 6, 4, 4, 3, 1, 1, 4, 6, 4, 4 \rangle)$

Figure 3.1232 provides an example where the SUM_SURF_DECREASING $(37, [3, 4, 2, 2, 5, 6, 6, 4, 4, 3, 1, 1, 4, 6, 4, 4])$ constraint holds.

Figure 3.1232: Illustrating the SUM_SURF_DECREASING constraint of the **Example** slot

**Typical**

$|\text{VARIABLES}| > 1$
$\text{range}(\text{VARIABLES.var}) > 1$

**Arg. properties**

Functional dependency: VALUE determined by VARIABLES.

**Automaton**    Figures 3.1233 and 3.1234 respectively depict the automaton associated with the constraint SUM_SURF_DECREASING and its simplified form.

Figure 3.1233: Automaton for the SUM_SURF_DECREASING constraint obtained by applying decoration Table 2.35 to the seed transducer of the DECREASING pattern where default is 0

Figure 3.1234: Automaton for the SUM_SURF_DECREASING constraint obtained by applying decoration Table 2.38 to the seed transducer of the DECREASING pattern where default is 0

Table 3.168: Glue matrix for the SUM_SURF_DECREASING constraint defined as the composition of the DECREASING pattern , the feature SURF , and the aggregator sum ; cells of the glue matrix are coloured with the colour of the constituent to which they are related.

# 3.590  SUM_SURF_DECREASING_SEQUENCE

▶     ◁ ◀     DESCRIPTION          AUTOMATON

| **Origin** | Based on the DECREASING_SEQUENCE pattern. |
|---|---|

| **Constraint** | SUM_SURF_DECREASING_SEQUENCE(VALUE, VARIABLES) |
|---|---|

| **Arguments** | VALUE     : dvar |
|---|---|
|  | VARIABLES : collection(var−dvar) |

**Restrictions**

$sv \le 1 \vee rv \le 1 \Rightarrow \texttt{VALUE} = 0$

$rv = 2 \Rightarrow \texttt{VALUE} = 0 \vee \texttt{VALUE} \ge \min(2 * minv + 1, (2 * minv + 1) * np)$

$rv \ge 3 \Rightarrow$
$$\bigvee \left( \begin{array}{l} \texttt{VALUE} = 0, \\ \texttt{VALUE} \ge \min \left( \begin{array}{l} 2 * minv + 1 ①, \\ (2 * minv + 1) * np + \min(0, sv \bmod 2 * (minv + 2)) ② \end{array} \right) \end{array} \right)$$

$rv = 2 \Rightarrow \texttt{VALUE} = 0 \vee \texttt{VALUE} \le \max(2 * maxv - 1, (2 * maxv - 1) * np)$

$rv \ge 3 \Rightarrow$
$$\bigvee \left( \begin{array}{l} \texttt{VALUE} = 0, \\ \texttt{VALUE} \le \max \left( \begin{array}{l} 2 * maxv - 1 ③, \\ (2 * maxv - 1) * np + \max(0, sv \bmod 2 * (maxv - 2)) ④ \end{array} \right) \end{array} \right)$$

required(VARIABLES, var)
where

$sv = |\texttt{VARIABLES}|$
$np = \lfloor sv/2 \rfloor$
$minv = minval(\texttt{VARIABLES}.var)$
$maxv = maxval(\texttt{VARIABLES}.var)$
$rv = range(\texttt{VARIABLES}.var)$

**Purpose**

VALUE is the sum of the surface of occurrences of the DECREASING_SEQUENCE pattern in the time-series given by the VARIABLES collection. If the pattern does not occur, VALUE takes the default value 0.

An occurrence of the pattern DECREASING_SEQUENCE is the *maximal* subsequence which matches the regular expression $> (> | =)^* > | >$.

Assume that the occurrence of the pattern DECREASING_SEQUENCE starts at position $i$ and ends at position $j$. The feature SURF computes the sum of the values from index $i$ to index $j + 1$.

**Example**

$$(34, \langle 3, 4, 2, 2, 5, 6, 6, 4, 4, 3, 1, 1, 4, 6, 4, 4 \rangle)$$

Figure 3.1235 provides an example where the SUM_SURF_DECREASING_SEQUENCE $(34, [3, 4, 2, 2, 5, 6, 6, 4, 4, 3, 1, 1, 4, 6, 4, 4])$ constraint holds.

Figure 3.1235: Illustrating the SUM_SURF_DECREASING_SEQUENCE constraint of the **Example** slot

**Typical**            $|\texttt{VARIABLES}| > 1$

                       $\texttt{range}(\texttt{VARIABLES.var}) > 1$

**Arg. properties**    Functional dependency: VALUE determined by VARIABLES.

**Automaton**   Figures 3.1236 and 3.1237 respectively depict the automaton associated with the constraint SUM_SURF_DECREASING_SEQUENCE and its simplified form.

Figure 3.1236: Automaton for the SUM_SURF_DECREASING_SEQUENCE constraint obtained by applying decoration Table 2.35 to the seed transducer of the DECREAS-ING_SEQUENCE pattern where default is 0

Figure 3.1237: Automaton for the SUM_SURF_DECREASING_SEQUENCE constraint obtained by applying decoration Table 2.29 to the seed transducer of the DECREAS-ING_SEQUENCE pattern where default is 0

| | $s$ | $t$ |
|---|---|---|
| $s$ | $\overrightarrow{C} + \overleftarrow{C}$ | $\overrightarrow{C} + \overleftarrow{C}$ |
| $t$ | $\overrightarrow{C} + \overleftarrow{C}$ | $\overrightarrow{C} + \overleftarrow{C} + \overrightarrow{D} + \overleftarrow{D} + \text{VAR}_i$ |

Table 3.169: Glue matrix for the SUM_SURF_DECREASING_SEQUENCE constraint defined as the composition of the DECREASING_SEQUENCE pattern, the feature SURF, and the aggregator sum; cells of the glue matrix are coloured with the colour of the constituent to which they are related.

# 3.591   SUM_SURF_DECREASING_TERRACE

      DESCRIPTION          AUTOMATON

| **Origin** | Based on the DECREASING_TERRACE pattern. |
| **Constraint** | SUM_SURF_DECREASING_TERRACE(VALUE, VARIABLES) |
| **Arguments** | VALUE      : dvar<br>VARIABLES  : collection(var−dvar) |

**Restrictions**

$\text{sv} \leq 3 \lor \text{rv} \leq 2 \Rightarrow \text{VALUE} = 0$

$\text{VALUE} = 0 \lor \text{VALUE} \geq \min(2 * (\text{minv} + 1)①, (\text{sv} - 2) * (\text{minv} + 1)②)$

$\text{VALUE} = 0 \lor \text{VALUE} \leq \max(2 * (\text{maxv} - 1)③, (\text{sv} - 2) * (\text{maxv} - 1)④)$

required(VARIABLES, var)

where

minv = minval(VARIABLES.var)

maxv = maxval(VARIABLES.var)

sv = |VARIABLES|

rv = range(VARIABLES.var)

**Purpose**

VALUE is the sum of the surface of occurrences of the DECREASING_TERRACE pattern in the time-series given by the VARIABLES collection. If the pattern does not occur, VALUE takes the default value 0.

An occurrence of the pattern DECREASING_TERRACE is the *maximal* subsequence which matches the regular expression $> =^+ >$.

Assume that the occurrence of the pattern DECREASING_TERRACE starts at position $i$ and ends at position $j$. The feature SURF computes the sum of the values from index $i + 1$ to index $j$.

**Example**

$(12, \langle 6, 4, 4, 4, 5, 2, 2, 1, 3, 3, 5, 4, 4, 3, 3, 3 \rangle)$

Figure 3.1238 provides an example where the SUM_SURF_DECREASING_TERRACE $(12, [6, 4, 4, 4, 5, 2, 2, 1, 3, 3, 5, 4, 4, 3, 3, 3])$ constraint holds.

**Typical**

$|\text{VARIABLES}| > 3$

range(VARIABLES.var) > 2

**Arg. properties**

Functional dependency: VALUE determined by VARIABLES.

Figure 3.1238: Illustrating the SUM_SURF_DECREASING_TERRACE constraint of the **Example** slot

**Automaton**　　　Figures 3.1239 and 3.1240 respectively depict the automaton associated with the constraint SUM_SURF_DECREASING_TERRACE and its simplified form.

Figure 3.1239: Automaton for the SUM_SURF_DECREASING_TERRACE constraint obtained by applying decoration Table 2.35 to the seed transducer of the DECREAS­ING_TERRACE pattern where default is 0

Figure 3.1240: Automaton for the SUM_SURF_DECREASING_TERRACE constraint obtained by applying decoration Table 2.28 to the seed transducer of the DECREAS­ING_TERRACE pattern where default is 0

|   | $s$ | $r$ | $t$ |
|---|---|---|---|
| $s$ | $\overrightarrow{C} + \overleftarrow{C}$ | $\overrightarrow{C} + \overleftarrow{C}$ | $\overrightarrow{C} + \overleftarrow{C}$ |
| $r$ | $\overrightarrow{C} + \overleftarrow{C}$ | $\overrightarrow{C} + \overleftarrow{C}$ | $\overrightarrow{D} + \overleftarrow{D} + \mathtt{VAR}_i$ |
| $t$ | $\overrightarrow{C} + \overleftarrow{C}$ | $\overrightarrow{D} + \overleftarrow{D} + \mathtt{VAR}_i$ | $\overrightarrow{D} + \overleftarrow{D} + \mathtt{VAR}_i$ |

Table 3.170: Glue matrix for the SUM_SURF_DECREASING_TERRACE constraint defined as the composition of the DECREASING_TERRACE pattern , the feature SURF , and the aggregator sum ; cells of the glue matrix are coloured with the colour of the constituent to which they are related.

## 3.592 SUM_SURF_DIP_ON_INCREASING_SEQUENCE

▶ ◁ ◀    **DESCRIPTION**    **AUTOMATON**

| **Origin** | Based on the DIP_ON_INCREASING_SEQUENCE pattern. |
|---|---|

| **Constraint** | SUM_SURF_DIP_ON_INCREASING_SEQUENCE(VALUE, VARIABLES) |
|---|---|

| **Arguments** | VALUE : dvar |
|---|---|
| | VARIABLES : collection(var−dvar) |

**Restrictions**

$sv \leq 5 \lor rv \leq 2 \Rightarrow VALUE = 0$
$VALUE = 0 \lor VALUE \geq \min(3 * minv + 3, (3 * minv + 3) * np)$
$VALUE = 0 \lor VALUE \leq \max(3 * maxv - 3①, (3 * maxv - 3) * np②)$
required(VARIABLES, var)
where
$sv = |VARIABLES|$
$np = \max(0, \lfloor (sv - 3)/3 \rfloor)$
$minv = minval(VARIABLES.var)$
$maxv = maxval(VARIABLES.var)$
$rv = range(VARIABLES.var)$

**Purpose**

VALUE is the sum of the surface of occurrences of the DIP_ON_INCREASING_SEQUENCE pattern in the time-series given by the VARIABLES collection. If the pattern does not occur, VALUE takes the default value 0.

An occurrence of the pattern DIP_ON_INCREASING_SEQUENCE is the subsequence which matches the regular expression $<<><<$.

Assume that the occurrence of the pattern DIP_ON_INCREASING_SEQUENCE starts at position $i$ and ends at position $j$. The feature SURF computes the sum of the values from index $i + 2$ to index $j$.

**Example**

$(19, \langle 1, 2, 3, 2, 5, 6, 7, 4, 1, 3, 4, 6, 1, 2, 4, 4 \rangle)$

Figure 3.1241 provides an example where the SUM_SURF_DIP_ON_INCREASING_SEQUENCE $(19, [1, 2, 3, 2, 5, 6, 7, 4, 1, 3, 4, 6, 1, 2, 4, 4])$ constraint holds.

| **Typical** | $|VARIABLES| > 5$ |
|---|---|
| | $range(VARIABLES.var) > 2$ |

| **Arg. properties** | Functional dependency: VALUE determined by VARIABLES. |
|---|---|

Figure 3.1241: Illustrating the SUM_SURF_DIP_ON_INCREASING_SEQUENCE constraint of the **Example** slot

**Automaton**     Figures 3.1242 and 3.1243 respectively depict the automaton associated with the constraint
SUM_SURF_DIP_ON_INCREASING_SEQUENCE and its simplified form.

Figure 3.1242: Automaton for the SUM_SURF_DIP_ON_INCREASING_SEQUENCE constraint obtained by applying decoration Table 2.35 to the seed transducer of the DIP_ON_INCREASING_SEQUENCE pattern where default is 0

Figure 3.1243: Automaton for the SUM_SURF_DIP_ON_INCREASING_SEQUENCE constraint obtained by applying decoration Table 2.25 to the seed transducer of the DIP_ON_INCREASING_SEQUENCE pattern where default is 0

# 3.593 SUM_SURF_GORGE

   DESCRIPTION   AUTOMATON

$$(> | > (= | >)^* >)(< | < (= | <)^* <)$$

| **Origin** | Based on the GORGE pattern. |
| --- | --- |

**Constraint** SUM_SURF_GORGE(VALUE, VARIABLES)

**Arguments**

VALUE : dvar
VARIABLES : collection(var−dvar)

**Restrictions**

$\mathtt{sv} \leq 2 \vee \mathtt{rv} \leq 1 \Rightarrow \mathtt{VALUE} = 0$

$\mathtt{rv} = 2 \Rightarrow \mathtt{VALUE} = 0 \vee \mathtt{VALUE} \geq \min(\mathtt{minv}①, \mathtt{minv} * \mathtt{np}②)$

$\mathtt{rv} \geq 3 \Rightarrow$
$\vee \left( \begin{array}{l} \mathtt{VALUE} = 0, \\ \mathtt{VALUE} \geq \min(\mathtt{minv}③, \min((\mathtt{sv} - 2) * (\mathtt{minv} + 1) - 1④, \mathtt{minv} * \mathtt{np}⑤)) \end{array} \right)$

$\mathtt{rv} = 2 \Rightarrow \mathtt{VALUE} = 0 \vee \mathtt{VALUE} \leq \max(\mathtt{maxv} - 1, (\mathtt{maxv} - 1) * \mathtt{np})$

$\mathtt{rv} \geq 3 \Rightarrow$
$\vee \left( \begin{array}{l} \mathtt{VALUE} = 0, \\ \mathtt{VALUE} \leq \max(\mathtt{maxv} - 1, \max((\mathtt{sv} - 2) * (\mathtt{maxv} - 1) - 1, (\mathtt{maxv} - 1) * \mathtt{np})) \end{array} \right)$

required(VARIABLES, var)

where

$\mathtt{sv} = |\mathtt{VARIABLES}|$
$\mathtt{np} = \max(0, \lfloor (\mathtt{sv} - 1)/2 \rfloor)$
$\mathtt{minv} = \mathtt{minval}(\mathtt{VARIABLES.var})$
$\mathtt{maxv} = \mathtt{maxval}(\mathtt{VARIABLES.var})$
$\mathtt{rv} = \mathtt{range}(\mathtt{VARIABLES.var})$

**Purpose**

VALUE is the sum of the surface of occurrences of the GORGE pattern in the time-series given by the VARIABLES collection. If the pattern does not occur, VALUE takes the default value 0.

An occurrence of the pattern GORGE is the *maximal* subsequence which matches the regular expression $(> | > (= | >)^* >)(< | < (= | <)^* <)$.

Assume that the occurrence of the pattern GORGE starts at position $i$ and ends at position $j$. The feature SURF computes the sum of the values from index $i + 1$ to index $j$.

**Example** $(25, \langle 1, 7, 3, 4, 4, 5, 5, 4, 2, 2, 6, 5, 4, 6, 5, 7 \rangle)$

Figure 3.1244 provides an example where the SUM_SURF_GORGE $(25, [1, 7, 3, 4, 4, 5, 5, 4, 2, 2, 6, 5, 4, 6, 5, 7])$ constraint holds.

**Typical**

$|\mathtt{VARIABLES}| > 2$
range(VARIABLES.var) > 1

Figure 3.1244: Illustrating the SUM_SURF_GORGE constraint of the **Example** slot

**Symmetry**        Items of VARIABLES can be reversed.

**Arg. properties**   Functional dependency: VALUE determined by VARIABLES.

**Automaton**      Figures 3.1245 and 3.1246 respectively depict the automaton associated with the constraint
SUM_SURF_GORGE and its simplified form.

Figure 3.1245: Automaton for the SUM_SURF_GORGE constraint obtained by applying
decoration Table 2.35 to the seed transducer of the GORGE pattern where default is 0
(transition $u \rightarrow r$ has the same accumulator update as transition $r \rightarrow u$)

| | $s$ | $r$ | $t$ | $u$ |
|---|---|---|---|---|
| $s$ | $\overrightarrow{C} + \overleftarrow{C}$ | $\overrightarrow{C} + \overleftarrow{C}$ | $\overrightarrow{C} + \overleftarrow{C}$ | $\overrightarrow{C} + \overleftarrow{C}$ |
| $r$ | $\overrightarrow{C} + \overleftarrow{C}$ | $\overrightarrow{D} + \overleftarrow{D} + \text{VAR}_i$ | $\overleftarrow{C} + \overrightarrow{D} + \overleftarrow{D} + \text{VAR}_i$ | $\overrightarrow{C} + \overleftarrow{C}$ |
| $t$ | $\overrightarrow{C} + \overleftarrow{C}$ | $\overrightarrow{C} + \overrightarrow{D} + \overleftarrow{D} + \text{VAR}_i$ | $\overrightarrow{C} + \overleftarrow{C}$ | $\overrightarrow{C} + \overrightarrow{D} + \overleftarrow{D} + \text{VAR}_i$ |
| $u$ | $\overrightarrow{C} + \overleftarrow{C}$ | $\overrightarrow{C} + \overleftarrow{C}$ | $\overleftarrow{C} + \overrightarrow{D} + \overleftarrow{D} + \text{VAR}_i$ | $\overrightarrow{C} + \overleftarrow{C}$ |

Table 3.171: Glue matrix for the SUM_SURF_GORGE constraint defined as the compo-
sition of the GORGE pattern , the feature SURF , and the aggregator sum ; cells of the
glue matrix are coloured with the colour of the constituent to which they are related.

Figure 3.1246: Automaton for the SUM_SURF_GORGE constraint obtained by applying decoration Table 2.25 to the seed transducer of the GORGE pattern where default is 0 (transition $u \rightarrow r$ has the same accumulator update as transition $r \rightarrow u$)

## 3.594   SUM_SURF_INCREASING

      **DESCRIPTION**          **AUTOMATON**

| | |
|---|---|
| **Origin** | Based on the INCREASING pattern. |
| **Constraint** | SUM_SURF_INCREASING(VALUE, VARIABLES) |
| **Arguments** | VALUE      : dvar |
| | VARIABLES : collection(var−dvar) |

**Restrictions**

$$\mathtt{sv} \le 1 \lor \mathtt{rv} \le 1 \Rightarrow \mathtt{VALUE} = 0$$

$$\mathtt{VALUE} \ge \min_{\mathtt{q} \in [\mathtt{lb1},\mathtt{ub1}]} \min \left( 0, \sum \left( \begin{array}{l} \mathtt{q} * ( \sum \left( \begin{array}{l} 2*(\mathtt{np}-1)*\mathtt{minv}, \\ \mathtt{np}*\mathtt{np} \end{array} \right) - \\ 2*\mathtt{np} \\[2pt] \prod \left( \begin{array}{l} \max \left( 0, \min \left( 1, \mathtt{np}-1 \right) \right), \\ \mathtt{sv} \bmod \mathtt{np} * \mathtt{q}, \\ \sum \left( \begin{array}{l} 2*\mathtt{minv}-1, \\ 2*\mathtt{np} \end{array} \right) \end{array} \right), \\ \prod \left( \begin{array}{l} \max \left( 0, \min \left( 1, 2-\mathtt{np} \right) \right), \\ 2*\mathtt{minv}+1, \\ \mathtt{q}-1 \end{array} \right) \end{array} \right) \right) \qquad ①$$

$$\mathtt{VALUE} \le \max_{\mathtt{q} \in [\mathtt{lb2},\mathtt{ub2}]} \max \left( 0, \sum \left( \begin{array}{l} \mathtt{q} * ( \sum \left( \begin{array}{l} 2*(\mathtt{np}-1)*\mathtt{maxv}- \\ \mathtt{np}*\mathtt{np} \\ 2*\mathtt{np} \end{array} , \right) -1), \\[2pt] \prod \left( \begin{array}{l} \max \left( 0, \min \left( 1, \mathtt{np}-1 \right) \right), \\ \mathtt{sv} \bmod \mathtt{np} * \mathtt{q}, \\ 2*\mathtt{maxv}+1- \\ 2*\mathtt{np} \end{array} \right), \\ \prod \left( \begin{array}{l} \max \left( 0, \min \left( 1, 2-\mathtt{np} \right) \right), \\ 2*\mathtt{maxv}-1, \\ \mathtt{q}-1 \end{array} \right) \end{array} \right) \right)$$

required(VARIABLES, var)
where

$$\mathtt{sv} = |\mathtt{VARIABLES}|$$
$$\mathtt{np} = \lfloor \mathtt{sv}/\mathtt{q} \rfloor$$
$$\mathtt{maxv} = \mathtt{maxval}(\mathtt{VARIABLES.var})$$
$$\mathtt{minv} = \mathtt{minval}(\mathtt{VARIABLES.var})$$
$$\mathtt{rv} = \mathtt{range}(\mathtt{VARIABLES.var})$$

$$\mathtt{lb1} = \min \left( \begin{array}{l} \lfloor \mathtt{sv}/2 \rfloor + 1, \\ \sum \left( \begin{array}{l} \lfloor \mathtt{sv}/\min(\min(\mathtt{sv},\mathtt{rv}), |\mathtt{minv}| + 1) \rfloor, \\ \min(1, \mathtt{sv} \bmod \min(\min(\mathtt{sv},\mathtt{rv}), |\mathtt{minv}| + 1)) \end{array} \right) \end{array} \right)$$
$$\mathtt{ub1} = \lfloor \mathtt{sv}/2 \rfloor + 1$$
$$\mathtt{lb2} = \min \left( \begin{array}{l} \lfloor \mathtt{sv}/2 \rfloor + 1, \\ \sum \left( \begin{array}{l} \lfloor \mathtt{sv}/\min(\min(\mathtt{sv},\mathtt{rv}), |\mathtt{maxv}| + 1) \rfloor, \\ \min(1, \mathtt{sv} \bmod \min(\min(\mathtt{sv},\mathtt{rv}), |\mathtt{maxv}| + 1)) \end{array} \right) \end{array} \right)$$
$$\mathtt{ub2} = \lfloor \mathtt{sv}/2 \rfloor + 1$$

**Purpose**

VALUE is the sum of the surface of occurrences of the INCREASING pattern in the time-series given by the VARIABLES collection. If the pattern does not occur, VALUE takes the default value 0.

An occurrence of the pattern INCREASING is the subsequence which matches the regular expression <.

Assume that the occurrence of the pattern INCREASING starts at position $i$ and ends at position $j$. The feature SURF computes the sum of the values from index $i$ to index $j+1$.

**Example**

$$(33, \langle 4, 3, 5, 5, 2, 1, 1, 3, 3, 4, 6, 6, 3, 1, 3, 3 \rangle)$$

Figure 3.1247 provides an example where the SUM_SURF_INCREASING $(33, [4, 3, 5, 5, 2, 1, 1, 3, 3, 4, 6, 6, 3, 1, 3, 3])$ constraint holds.

Figure 3.1247: Illustrating the SUM_SURF_INCREASING constraint of the **Example** slot

**Typical**

$|\texttt{VARIABLES}| > 1$
$\texttt{range}(\texttt{VARIABLES.var}) > 1$

**Arg. properties**

Functional dependency: VALUE determined by VARIABLES.

**Automaton**  Figures 3.1248 and 3.1249 respectively depict the automaton associated with the constraint SUM_SURF_INCREASING and its simplified form.

Figure 3.1248: Automaton for the SUM_SURF_INCREASING constraint obtained by applying decoration Table 2.35 to the seed transducer of the INCREASING pattern where default is $0$

Figure 3.1249: Automaton for the SUM_SURF_INCREASING constraint obtained by applying decoration Table 2.38 to the seed transducer of the INCREASING pattern where default is $0$

Table 3.172: Glue matrix for the SUM_SURF_INCREASING constraint defined as the composition of the INCREASING pattern , the feature SURF , and the aggregator sum ; cells of the glue matrix are coloured with the colour of the constituent to which they are related.

# 3.595   SUM_SURF_INCREASING_SEQUENCE

▶   ◁ ◀      DESCRIPTION          AUTOMATON

$< (< | =)^* < |$
$<$

| **Origin** | Based on the INCREASING_SEQUENCE pattern. |
| **Constraint** | SUM_SURF_INCREASING_SEQUENCE(VALUE, VARIABLES) |
| **Arguments** | VALUE     : dvar<br>VARIABLES : collection(var−dvar) |

**Restrictions**

$sv \leq 1 \vee rv \leq 1 \Rightarrow VALUE = 0$

$rv = 2 \Rightarrow VALUE = 0 \vee VALUE \geq \min(2 * minv + 1, (2 * minv + 1) * np)$

$rv \geq 3 \Rightarrow$
$$\bigvee \left( \begin{array}{l} VALUE = 0, \\ VALUE \geq \min \left( \begin{array}{l} 2 * minv + 1 ①, \\ (2 * minv + 1) * np + \min(0, sv \bmod 2 * (minv + 2)) ② \end{array} \right) \end{array} \right)$$

$rv = 2 \Rightarrow VALUE = 0 \vee VALUE \leq \max(2 * maxv - 1, (2 * maxv - 1) * np)$

$rv \geq 3 \Rightarrow$
$$\bigvee \left( \begin{array}{l} VALUE = 0, \\ VALUE \leq \max \left( \begin{array}{l} 2 * maxv - 1 ③, \\ (2 * maxv - 1) * np + \max(0, sv \bmod 2 * (maxv - 2)) ④ \end{array} \right) \end{array} \right)$$

required(VARIABLES, var)
where
$sv = |VARIABLES|$
$np = \lfloor sv/2 \rfloor$
$minv = minval(VARIABLES.var)$
$maxv = maxval(VARIABLES.var)$
$rv = range(VARIABLES.var)$

**Purpose**

VALUE is the sum of the surface of occurrences of the INCREASING_SEQUENCE pattern in the time-series given by the VARIABLES collection. If the pattern does not occur, VALUE takes the default value 0.

An occurrence of the pattern INCREASING_SEQUENCE is the *maximal* subsequence which matches the regular expression $< (< | =)^* < | <$.

Assume that the occurrence of the pattern INCREASING_SEQUENCE starts at position $i$ and ends at position $j$. The feature SURF computes the sum of the values from index $i$ to index $j + 1$.

**Example**

$(29, \langle 4, 3, 5, 5, 2, 1, 1, 3, 3, 4, 6, 6, 3, 1, 3, 3 \rangle)$

Figure 3.1250 provides an example where the SUM_SURF_INCREASING_SEQUENCE $(29, [4, 3, 5, 5, 2, 1, 1, 3, 3, 4, 6, 6, 3, 1, 3, 3])$ constraint holds.

Figure 3.1250: Illustrating the SUM_SURF_INCREASING_SEQUENCE constraint of the **Example** slot

| **Typical** | $|\texttt{VARIABLES}| > 1$ |
| | $\texttt{range}(\texttt{VARIABLES.var}) > 1$ |

| **Arg. properties** | Functional dependency: VALUE determined by VARIABLES. |

**Automaton**          Figures 3.1251 and 3.1252 respectively depict the automaton associated with the constraint SUM_SURF_INCREASING_SEQUENCE and its simplified form.

Figure 3.1251: Automaton for the SUM_SURF_INCREASING_SEQUENCE constraint obtained by applying decoration Table 2.35 to the seed transducer of the INCREASING_SEQUENCE pattern where default is 0

Figure 3.1252: Automaton for the SUM_SURF_INCREASING_SEQUENCE constraint obtained by applying decoration Table 2.29 to the seed transducer of the INCREASING_SEQUENCE pattern where default is 0

| | $s$ | $t$ |
|---|---|---|
| $s$ | $\overrightarrow{C} + \overleftarrow{C}$ | $\overrightarrow{C} + \overleftarrow{C}$ |
| $t$ | $\overrightarrow{C} + \overleftarrow{C}$ | $\overrightarrow{C} + \overleftarrow{C} + \overrightarrow{D} + \overleftarrow{D} + \mathtt{VAR}_i$ |

Table 3.173: Glue matrix for the SUM_SURF_INCREASING_SEQUENCE constraint defined as the composition of the INCREASING_SEQUENCE pattern , the feature SURF , and the aggregator sum ; cells of the glue matrix are coloured with the colour of the constituent to which they are related.

# 3.596   SUM_SURF_INCREASING_TERRACE

▶    ◁ ◀      DESCRIPTION        AUTOMATON

| **Origin** | Based on the INCREASING_TERRACE pattern. |

| **Constraint** | SUM_SURF_INCREASING_TERRACE(VALUE, VARIABLES) |

| **Arguments** | VALUE     : dvar |
|  | VARIABLES : collection(var−dvar) |

**Restrictions**

$$\text{sv} \leq 3 \vee \text{rv} \leq 2 \Rightarrow \text{VALUE} = 0$$
$$\text{VALUE} = 0 \vee \text{VALUE} \geq \min(2*(\text{minv}+1)①, (\text{sv}-2)*(\text{minv}+1)②)$$
$$\text{VALUE} = 0 \vee \text{VALUE} \leq \max(2*(\text{maxv}-1)③, (\text{sv}-2)*(\text{maxv}-1)④)$$
$$\text{required}(\text{VARIABLES}, \text{var})$$
where
$$\text{minv} = \text{minval}(\text{VARIABLES.var})$$
$$\text{maxv} = \text{maxval}(\text{VARIABLES.var})$$
$$\text{sv} = |\text{VARIABLES}|$$
$$\text{rv} = \text{range}(\text{VARIABLES.var})$$

**Purpose**

VALUE is the sum of the surface of occurrences of the INCREASING_TERRACE pattern in the time-series given by the VARIABLES collection. If the pattern does not occur, VALUE takes the default value 0.

An occurrence of the pattern INCREASING_TERRACE is the *maximal* subsequence which matches the regular expression $<=^+<$.

Assume that the occurrence of the pattern INCREASING_TERRACE starts at position $i$ and ends at position $j$. The feature SURF computes the sum of the values from index $i+1$ to index $j$.

**Example**

$$(19, \langle 1, 3, 3, 3, 2, 5, 5, 6, 4, 4, 2, 3, 3, 3, 4, 4 \rangle)$$

Figure 3.1253 provides an example where the SUM_SURF_INCREASING_TERRACE $(19, [1, 3, 3, 3, 2, 5, 5, 6, 4, 4, 2, 3, 3, 3, 4, 4])$ constraint holds.

**Typical**

$$|\text{VARIABLES}| > 3$$
$$\text{range}(\text{VARIABLES.var}) > 2$$

**Arg. properties**

Functional dependency: VALUE determined by VARIABLES.

Figure 3.1253: Illustrating the SUM_SURF_INCREASING_TERRACE constraint of the **Example** slot

**Automaton**     Figures 3.1254 and 3.1255 respectively depict the automaton associated with the constraint SUM_SURF_INCREASING_TERRACE and its simplified form.

Figure 3.1254: Automaton for the SUM_SURF_INCREASING_TERRACE constraint obtained by applying decoration Table 2.35 to the seed transducer of the INCREASING_TERRACE pattern where default is 0

Figure 3.1255: Automaton for the SUM_SURF_INCREASING_TERRACE constraint obtained by applying decoration Table 2.28 to the seed transducer of the INCREASING_TERRACE pattern where default is 0

|   | $s$ | $r$ | $t$ |
|---|---|---|---|
| $s$ | $\overrightarrow{C} + \overleftarrow{C}$ | $\overrightarrow{C} + \overleftarrow{C}$ | $\overrightarrow{C} + \overleftarrow{C}$ |
| $r$ | $\overrightarrow{C} + \overleftarrow{C}$ | $\overrightarrow{C} + \overleftarrow{C}$ | $\overrightarrow{D} + \overleftarrow{D} + \text{VAR}_i$ |
| $t$ | $\overrightarrow{C} + \overleftarrow{C}$ | $\overrightarrow{D} + \overleftarrow{D} + \text{VAR}_i$ | $\overrightarrow{D} + \overleftarrow{D} + \text{VAR}_i$ |

Table 3.174: Glue matrix for the SUM_SURF_INCREASING_TERRACE constraint defined as the composition of the INCREASING_TERRACE pattern , the feature SURF , and the aggregator sum ; cells of the glue matrix are coloured with the colour of the constituent to which they are related.

## 3.597 SUM_SURF_INFLEXION

▶    ◁ ◀        DESCRIPTION            AUTOMATON

$< (< | =)^* > | > (> | =)^* <$

| Origin | Based on the INFLEXION pattern. |
|---|---|

| Constraint | SUM_SURF_INFLEXION(VALUE, VARIABLES) |
|---|---|

| Arguments | VALUE     : dvar |
|---|---|
|  | VARIABLES : collection(var−dvar) |

**Restrictions**

$\text{sv} \leq 2 \vee \text{rv} \leq 1 \Rightarrow \text{VALUE} = 0$

$\text{VALUE} = 0 \vee \text{VALUE} \geq \min(\text{minv}①, (\text{sv} − 2) * \text{minv}②)$

$\text{VALUE} = 0 \vee \text{VALUE} \leq \max(\text{maxv}③, (\text{sv} − 2) * \text{maxv}④)$

required(VARIABLES, var)

where

minv = minval(VARIABLES.var)

maxv = maxval(VARIABLES.var)

sv = |VARIABLES|

rv = range(VARIABLES.var)

VALUE $\geq$ **0** ①     VALUE $\geq$ **−24** $= (10 − 2) * −3$ ②

VALUE $\leq$ **−1** ③     VALUE $\leq$ **16** $= (10 − 2) * 2$ ④

**Purpose**

VALUE is the sum of the surface of occurrences of the INFLEXION pattern in the time-series given by the VARIABLES collection. If the pattern does not occur, VALUE takes the default value 0.

An occurrence of the pattern INFLEXION is the *maximal* subsequence which matches the regular expression $< (< | =)^* > | > (> | =)^* <$.

Assume that the occurrence of the pattern INFLEXION starts at position $i$ and ends at position $j$. The feature SURF computes the sum of the values from index $i + 1$ to index $j$.

| Example | $(49, \langle 1, 2, 6, 6, 4, 4, 3, 5, 2, 5, 1, 5, 3, 3, 4, 4 \rangle)$ |
|---|---|

Figure 3.1256 provides an example where the SUM_SURF_INFLEXION $(49, [1, 2, 6, 6, 4, 4, 3, 5, 2, 5, 1, 5, 3, 3, 4, 4])$ constraint holds.

| Typical | $|\text{VARIABLES}| > 2$ |
|---|---|
|  | range(VARIABLES.var) $> 1$ |

| Symmetry | Items of VARIABLES can be reversed. |
|---|---|

| Arg. properties | Functional dependency: VALUE determined by VARIABLES. |
|---|---|

Figure 3.1256: Illustrating the SUM_SURF_INFLEXION constraint of the **Example** slot

**Automaton**     Figures 3.1257 and 3.1258 respectively depict the automaton associated with the constraint SUM_SURF_INFLEXION and its simplified form.

Figure 3.1257: Automaton for the SUM_SURF_INFLEXION constraint obtained by applying decoration Table 2.35 to the seed transducer of the INFLEXION pattern where default is 0 (transition $r \rightarrow t$ has the same accumulators updates as transition $t \rightarrow r$)

Figure 3.1258: Automaton for the SUM_SURF_INFLEXION constraint obtained by applying decoration Table 2.25 to the seed transducer of the INFLEXION pattern where default is 0 (transition $r \rightarrow t$ has the same accumulators updates as transition $t \rightarrow r$)

# 3.598   SUM_SURF_PEAK

        **DESCRIPTION**              **AUTOMATON**

$< (= | <)^* (> | =)^* >$

| | |
|---|---|
| **Origin** | Based on the PEAK pattern. |
| **Constraint** | SUM_SURF_PEAK(VALUE, VARIABLES) |
| **Arguments** | VALUE : dvar<br>VARIABLES : collection(var−dvar) |

**Restrictions**

$\text{sv} \leq 2 \vee \text{rv} \leq 1 \Rightarrow \text{VALUE} = 0$

$\text{VALUE} = 0 \vee \text{VALUE} \geq \min(\text{minv} + 1①, (\text{sv} - 2) * (\text{minv} + 1)②)$

$\text{VALUE} = 0 \vee \text{VALUE} \leq \max(\text{maxv}③, (\text{sv} - 2) * \text{maxv}④)$

required(VARIABLES, var)

where

minv = minval(VARIABLES.var)

maxv = maxval(VARIABLES.var)

sv = |VARIABLES|

rv = range(VARIABLES.var)

VALUE ≥ **1**

VALUE ≥ **−18** = (11 − 2) ∗ (−3 + 1)

VALUE ≤ **−1**

VALUE ≤ **18** = (11 − 2) ∗ 2

**Purpose**

VALUE is the sum of the surface of occurrences of the PEAK pattern in the time-series given by the VARIABLES collection. If the pattern does not occur, VALUE takes the default value 0.

An occurrence of the pattern PEAK is the *maximal* subsequence which matches the regular expression $< (= | <)^* (> | =)^* >$.

Assume that the occurrence of the pattern PEAK starts at position $i$ and ends at position $j$. The feature SURF computes the sum of the values from index $i + 1$ to index $j$.

**Example**

$(32, \langle 7, 5, 5, 1, 4, 5, 2, 2, 3, 5, 6, 2, 3, 3, 3, 1 \rangle)$

Figure 3.1259 provides an example where the SUM_SURF_PEAK $(32, [7, 5, 5, 1, 4, 5, 2, 2, 3, 5, 6, 2, 3, 3, 3, 1])$ constraint holds.

**Typical**

$|\text{VARIABLES}| > 2$

range(VARIABLES.var) > 1

**Symmetry**     Items of VARIABLES can be reversed.

**Arg. properties**     Functional dependency: VALUE determined by VARIABLES.

Figure 3.1259: Illustrating the SUM_SURF_PEAK constraint of the **Example** slot

**Automaton**  Figures 3.1260 and 3.1261 respectively depict the automaton associated with the constraint SUM_SURF_PEAK and its simplified form.

Figure 3.1260: Automaton for the SUM_SURF_PEAK constraint obtained by applying decoration Table 2.35 to the seed transducer of the PEAK pattern where default is 0

Figure 3.1261: Automaton for the SUM_SURF_PEAK constraint obtained by applying decoration Table 2.25 to the seed transducer of the PEAK pattern where default is 0

| | $s$ | $r$ | $t$ |
|---|---|---|---|
| $s$ | $\overrightarrow{C} + \overleftarrow{C}$ | $\overrightarrow{C} + \overleftarrow{C}$ | $\overrightarrow{C} + \overleftarrow{C}$ |
| $r$ | $\overrightarrow{C} + \overleftarrow{C}$ | $\overrightarrow{D} + \overleftarrow{D} + \mathtt{VAR}_i$ | $\overleftarrow{C} + \overrightarrow{D} + \overleftarrow{D} + \mathtt{VAR}_i$ |
| $t$ | $\overrightarrow{C} + \overleftarrow{C}$ | $\overrightarrow{C} + \overrightarrow{D} + \overleftarrow{D} + \mathtt{VAR}_i$ | $\overrightarrow{C} + \overleftarrow{C}$ |

Table 3.175: Glue matrix for the SUM_SURF_PEAK constraint defined as the composition of the PEAK pattern , the feature SURF , and the aggregator sum ; cells of the glue matrix are coloured with the colour of the constituent to which they are related.

## 3.599  SUM_SURF_PLAIN

▶    ◁ ◀        DESCRIPTION          AUTOMATON

| | |
|---|---|
| **Origin** | Based on the PLAIN pattern. |
| **Constraint** | SUM_SURF_PLAIN(VALUE, VARIABLES) |
| **Arguments** | VALUE      :  dvar<br>VARIABLES  :  collection(var−dvar) |
| **Restrictions** | $\mathtt{sv} \leq 2 \vee \mathtt{rv} \leq 1 \Rightarrow \mathtt{VALUE} = 0$<br>$\mathtt{VALUE} = 0 \vee \mathtt{VALUE} \geq \min(\mathtt{minv}①, \mathtt{minv} * (\mathtt{sv}-2)②)$<br>$\mathtt{VALUE} = 0 \vee \mathtt{VALUE} \leq \max(\mathtt{maxv}-1③, (\mathtt{maxv}-1) * (\mathtt{sv}-2)④)$<br>required(VARIABLES, var)<br>where<br>minv =minval(VARIABLES.var)<br>maxv =maxval(VARIABLES.var)<br>sv =\|VARIABLES\|<br>rv =range(VARIABLES.var) |
| **Purpose** | VALUE is the sum of the surface of occurrences of the PLAIN pattern in the time-series given by the VARIABLES collection. If the pattern does not occur, VALUE takes the default value 0.<br>An occurrence of the pattern PLAIN is the *maximal* subsequence which matches the regular expression $> =^* <$.<br>Assume that the occurrence of the pattern PLAIN starts at position $i$ and ends at position $j$. The feature SURF computes the sum of the values from index $i+1$ to index $j$. |
| **Example** | $(15, \langle 2, 3, 6, 5, 7, 6, 6, 4, 5, 5, 4, 3, 3, 6, 6, 3 \rangle)$ |

Figure    3.1262    provides    an    example    where    the    SUM_SURF_PLAIN
$(15, [2, 3, 6, 5, 7, 6, 6, 4, 5, 5, 4, 3, 3, 6, 6, 3])$ constraint holds.

| | |
|---|---|
| **Typical** | $\|\mathtt{VARIABLES}\| > 2$<br>range(VARIABLES.var) $> 1$ |
| **Symmetry** | Items of VARIABLES can be reversed. |
| **Arg. properties** | Functional dependency: VALUE determined by VARIABLES. |

Figure 3.1262: Illustrating the SUM_SURF_PLAIN constraint of the **Example** slot

**Automaton**    Figures 3.1263 and 3.1264 respectively depict the automaton associated with the constraint SUM_SURF_PLAIN and its simplified form.

Figure 3.1263: Automaton for the SUM_SURF_PLAIN constraint obtained by applying decoration Table 2.35 to the seed transducer of the PLAIN pattern where default is 0

Figure 3.1264: Automaton for the SUM_SURF_PLAIN constraint obtained by applying decoration Table 2.28 to the seed transducer of the PLAIN pattern where default is 0

|   | $s$ | $r$ | $t$ |
|---|-----|-----|-----|
| $s$ | $\overrightarrow{C} + \overleftarrow{C}$ | $\overrightarrow{C} + \overleftarrow{C}$ | $\overrightarrow{C} + \overleftarrow{C}$ |
| $r$ | $\overrightarrow{C} + \overleftarrow{C}$ | $\overrightarrow{D} + \overleftarrow{D} + \mathtt{VAR}_i$ | $\overrightarrow{D} + \overleftarrow{D} + \mathtt{VAR}_i$ |
| $t$ | $\overrightarrow{C} + \overleftarrow{C}$ | $\overrightarrow{D} + \overleftarrow{D} + \mathtt{VAR}_i$ | $\overrightarrow{D} + \overleftarrow{D} + \mathtt{VAR}_i$ |

Table 3.176: Glue matrix for the SUM_SURF_PLAIN constraint defined as the composition of the PLAIN pattern , the feature SURF , and the aggregator sum ; cells of the glue matrix are coloured with the colour of the constituent to which they are related.

# 3.600 SUM_SURF_PLATEAU

▶    ◁ ◀    DESCRIPTION    AUTOMATON

| | |
|---|---|
| **Origin** | Based on the PLATEAU pattern. |
| **Constraint** | SUM_SURF_PLATEAU(VALUE, VARIABLES) |
| **Arguments** | VALUE : dvar<br>VARIABLES : collection(var−dvar) |

**Restrictions**

$\text{sv} \leq 2 \vee \text{rv} \leq 1 \Rightarrow \text{VALUE} = 0$

$\text{VALUE} = 0 \vee \text{VALUE} \geq \min(\text{minv} + 1①, (\text{sv} - 2) * (\text{minv} + 1)②)$

$\text{VALUE} = 0 \vee \text{VALUE} \leq \max(\text{maxv}③, (\text{sv} - 2) * \text{maxv}④)$

required(VARIABLES, var)

where

minv = minval(VARIABLES.var)
maxv = maxval(VARIABLES.var)
sv = |VARIABLES|
rv = range(VARIABLES.var)

**Purpose**

VALUE is the sum of the surface of occurrences of the PLATEAU pattern in the time-series given by the VARIABLES collection. If the pattern does not occur, VALUE takes the default value 0.

An occurrence of the pattern PLATEAU is the *maximal* subsequence which matches the regular expression $<=^* >$.

Assume that the occurrence of the pattern PLATEAU starts at position $i$ and ends at position $j$. The feature SURF computes the sum of the values from index $i + 1$ to index $j$.

**Example**

$(17, \langle 7, 5, 2, 3, 1, 2, 2, 4, 3, 3, 4, 5, 5, 2, 2, 5 \rangle)$

Figure 3.1265 provides an example where the SUM_SURF_PLATEAU $(17, [7, 5, 2, 3, 1, 2, 2, 4, 3, 3, 4, 5, 5, 2, 2, 5])$ constraint holds.

| | |
|---|---|
| **Typical** | $\|\text{VARIABLES}\| > 2$<br>range(VARIABLES.var) $> 1$ |
| **Symmetry** | Items of VARIABLES can be reversed. |
| **Arg. properties** | Functional dependency: VALUE determined by VARIABLES. |

Figure 3.1265: Illustrating the SUM_SURF_PLATEAU constraint of the **Example** slot

**Automaton**     Figures 3.1266 and 3.1267 respectively depict the automaton associated with the constraint SUM_SURF_PLATEAU and its simplified form.

Figure 3.1266: Automaton for the SUM_SURF_PLATEAU constraint obtained by applying decoration Table 2.35 to the seed transducer of the PLATEAU pattern where default is 0

Figure 3.1267: Automaton for the SUM_SURF_PLATEAU constraint obtained by applying decoration Table 2.28 to the seed transducer of the PLATEAU pattern where default is 0

| | $s$ | $r$ | $t$ |
|---|---|---|---|
| $s$ | $\overrightarrow{C} + \overleftarrow{C}$ | $\overrightarrow{C} + \overleftarrow{C}$ | $\overrightarrow{C} + \overleftarrow{C}$ |
| $r$ | $\overrightarrow{C} + \overleftarrow{C}$ | $\overrightarrow{D} + \overleftarrow{D} + \text{VAR}_i$ | $\overrightarrow{D} + \overleftarrow{D} + \text{VAR}_i$ |
| $t$ | $\overrightarrow{C} + \overleftarrow{C}$ | $\overrightarrow{D} + \overleftarrow{D} + \text{VAR}_i$ | $\overrightarrow{D} + \overleftarrow{D} + \text{VAR}_i$ |

Table 3.177: Glue matrix for the SUM_SURF_PLATEAU constraint defined as the composition of the PLATEAU pattern , the feature SURF , and the aggregator sum ; cells of the glue matrix are coloured with the colour of the constituent to which they are related.

## 3.601   SUM_SURF_PROPER_PLAIN

▶   ◁ ◀        DESCRIPTION           AUTOMATON

**Origin**         Based on the PROPER_PLAIN pattern.

**Constraint**     SUM_SURF_PROPER_PLAIN(VALUE, VARIABLES)

**Arguments**      VALUE      : dvar
                   VARIABLES  : collection(var−dvar)

**Restrictions**   $sv \leq 3 \vee rv \leq 1 \Rightarrow \text{VALUE} = 0$
                   $\text{VALUE} = 0 \vee \text{VALUE} \geq \min(2 * \text{minv}①, (sv − 2) * \text{minv}②)$
                   $\text{VALUE} = 0 \vee \text{VALUE} \leq \max(2 * (\text{maxv} − 1)③, (sv − 2) * (\text{maxv} − 1)④)$
                   required(VARIABLES, var)
                   where
                   minv = minval(VARIABLES.var)
                   maxv = maxval(VARIABLES.var)
                   $sv = |\text{VARIABLES}|$
                   $rv = \text{range}(\text{VARIABLES.var})$

**Purpose**        VALUE is the sum of the surface of occurrences of the PROPER_PLAIN pattern in the time-
                   series given by the VARIABLES collection. If the pattern does not occur, VALUE takes the
                   default value 0.
                   An occurrence of the pattern PROPER_PLAIN is the *maximal* subsequence which matches
                   the regular expression $>=^+<$.
                   Assume that the occurrence of the pattern PROPER_PLAIN starts at position $i$ and ends at
                   position $j$. The feature SURF computes the sum of the values from index $i + 1$ to index
                   $j$.

**Example**        $(27, \langle 2, 7, 5, 5, 6, 3, 7, 4, 4, 5, 6, 5, 3, 3, 3, 5\rangle)$

                   Figure 3.1268 provides an example where the SUM_SURF_PROPER_PLAIN
                   $(27, [2, 7, 5, 5, 6, 3, 7, 4, 4, 5, 6, 5, 3, 3, 3, 5])$ constraint holds.

**Typical**        $|\text{VARIABLES}| > 3$
                   range(VARIABLES.var) > 1

**Symmetry**       Items of VARIABLES can be reversed.

**Arg. properties** Functional dependency: VALUE determined by VARIABLES.

Figure 3.1268: Illustrating the SUM_SURF_PROPER_PLAIN constraint of the **Example** slot

**Automaton**     Figures 3.1269 and 3.1270 respectively depict the automaton associated with the constraint
SUM_SURF_PROPER_PLAIN and its simplified form.

Figure 3.1269: Automaton for the SUM_SURF_PROPER_PLAIN constraint obtained by
applying decoration Table 2.35 to the seed transducer of the PROPER_PLAIN pattern
where default is 0

Figure 3.1270: Automaton for the SUM_SURF_PROPER_PLAIN constraint obtained by
applying decoration Table 2.28 to the seed transducer of the PROPER_PLAIN pattern
where default is 0

|   | $s$ | $r$ | $t$ |
|---|---|---|---|
| $s$ | $\overrightarrow{C} + \overleftarrow{C}$ | $\overrightarrow{C} + \overleftarrow{C}$ | $\overrightarrow{C} + \overleftarrow{C}$ |
| $r$ | $\overrightarrow{C} + \overleftarrow{C}$ | $\overrightarrow{C} + \overleftarrow{C}$ | $\overrightarrow{D} + \overleftarrow{D} + \text{VAR}_i$ |
| $t$ | $\overrightarrow{C} + \overleftarrow{C}$ | $\overrightarrow{D} + \overleftarrow{D} + \text{VAR}_i$ | $\overrightarrow{D} + \overleftarrow{D} + \text{VAR}_i$ |

Table 3.178: Glue matrix for the SUM_SURF_PROPER_PLAIN constraint defined as the composition of the PROPER_PLAIN pattern , the feature SURF , and the aggregator sum ; cells of the glue matrix are coloured with the colour of the constituent to which they are related.

## 3.602   SUM_SURF_PROPER_PLATEAU

| | |
|---|---|
| **Origin** | Based on the PROPER_PLATEAU pattern. |
| **Constraint** | SUM_SURF_PROPER_PLATEAU(VALUE, VARIABLES) |
| **Arguments** | VALUE     : dvar |
| | VARIABLES : collection(var−dvar) |

**Restrictions**

$\mathtt{sv} \leq 3 \vee \mathtt{rv} \leq 1 \Rightarrow \mathtt{VALUE} = 0$

$\mathtt{VALUE} = 0 \vee \mathtt{VALUE} \geq \min(2 * (\mathtt{minv} + 1)①, (\mathtt{sv} - 2) * (\mathtt{minv} + 1)②)$

$\mathtt{VALUE} = 0 \vee \mathtt{VALUE} \leq \max(2 * \mathtt{maxv}③, (\mathtt{sv} - 2) * \mathtt{maxv}④)$

required(VARIABLES, var)

where

minv =minval(VARIABLES.var)

maxv =maxval(VARIABLES.var)

sv =|VARIABLES|

rv =range(VARIABLES.var)

**Purpose**

VALUE is the sum of the surface of occurrences of the PROPER_PLATEAU pattern in the time-series given by the VARIABLES collection. If the pattern does not occur, VALUE takes the default value 0.

An occurrence of the pattern PROPER_PLATEAU is the *maximal* subsequence which matches the regular expression $<=^+>$.

Assume that the occurrence of the pattern PROPER_PLATEAU starts at position $i$ and ends at position $j$. The feature SURF computes the sum of the values from index $i + 1$ to index $j$.

| | |
|---|---|
| **Example** | $(29, \langle 7, 1, 3, 3, 2, 5, 1, 4, 4, 3, 2, 3, 5, 5, 5, 3 \rangle)$ |

Figure 3.1271 provides an example where the SUM_SURF_PROPER_PLATEAU $(29, [7, 1, 3, 3, 2, 5, 1, 4, 4, 3, 2, 3, 5, 5, 5, 3])$ constraint holds.

| | |
|---|---|
| **Typical** | $|\mathtt{VARIABLES}| > 3$ |
| | range(VARIABLES.var) > 1 |
| **Symmetry** | Items of VARIABLES can be reversed. |
| **Arg. properties** | Functional dependency: VALUE determined by VARIABLES. |

Figure 3.1271: Illustrating the SUM_SURF_PROPER_PLATEAU constraint of the **Example** slot

**Automaton**     Figures 3.1272 and 3.1273 respectively depict the automaton associated with the constraint
SUM_SURF_PROPER_PLATEAU and its simplified form.

Figure 3.1272: Automaton for the SUM_SURF_PROPER_PLATEAU constraint obtained by applying decoration Table 2.35 to the seed transducer of the PROPER_PLATEAU pattern where default is 0

Figure 3.1273: Automaton for the SUM_SURF_PROPER_PLATEAU constraint obtained by applying decoration Table 2.28 to the seed transducer of the PROPER_PLATEAU pattern where default is 0

|   | $s$ | $r$ | $t$ |
|---|---|---|---|
| $s$ | $\overrightarrow{C} + \overleftarrow{C}$ | $\overrightarrow{C} + \overleftarrow{C}$ | $\overrightarrow{C} + \overleftarrow{C}$ |
| $r$ | $\overrightarrow{C} + \overleftarrow{C}$ | $\overrightarrow{C} + \overleftarrow{C}$ | $\overrightarrow{D} + \overleftarrow{D} + \mathtt{VAR}_i$ |
| $t$ | $\overrightarrow{C} + \overleftarrow{C}$ | $\overrightarrow{D} + \overleftarrow{D} + \mathtt{VAR}_i$ | $\overrightarrow{D} + \overleftarrow{D} + \mathtt{VAR}_i$ |

Table 3.179: Glue matrix for the SUM_SURF_PROPER_PLATEAU constraint defined as the composition of the PROPER_PLATEAU pattern , the feature SURF , and the aggregator sum ; cells of the glue matrix are coloured with the colour of the constituent to which they are related.

# 3.603   SUM_SURF_STEADY

          **DESCRIPTION**          **AUTOMATON**

| | |
|---|---|
| **Origin** | Based on the STEADY pattern. |
| **Constraint** | SUM_SURF_STEADY(VALUE, VARIABLES) |
| **Arguments** | VALUE        : dvar<br>VARIABLES  : collection(var−dvar) |

**Restrictions**

$\text{sv} \leq 1 \Rightarrow \text{VALUE} = 0$
$\text{rv} = 1 \Rightarrow \text{VALUE} \geq 2 * \text{minv} * \text{np} ①$
$\text{rv} \geq 2 \Rightarrow \text{VALUE} = 0 \lor \text{VALUE} \geq \min(2 * \text{minv} ②, 2 * \text{minv} * \text{np} ③)$
$\text{rv} = 1 \Rightarrow \text{VALUE} \leq 2 * \text{maxv} * \text{np} ④$
$\text{rv} \geq 2 \Rightarrow \text{VALUE} = 0 \lor \text{VALUE} \leq \max(2 * \text{maxv} ⑤, 2 * \text{maxv} * \text{np} ⑥)$
required(VARIABLES, var)
where
$\text{sv} = |\text{VARIABLES}|$
$\text{np} = \max(0, \text{sv} - 1)$
$\text{minv} = \text{minval}(\text{VARIABLES.var})$
$\text{maxv} = \text{maxval}(\text{VARIABLES.var})$
$\text{rv} = \text{range}(\text{VARIABLES.var})$

**Purpose**

VALUE is the sum of the surface of occurrences of the STEADY pattern in the time-series given by the VARIABLES collection. If the pattern does not occur, VALUE takes the default value 0.
An occurrence of the pattern STEADY is the subsequence which matches the regular expression =.
Assume that the occurrence of the pattern STEADY starts at position $i$ and ends at position $j$. The feature SURF computes the sum of the values from index $i$ to index $j + 1$.

**Example**

$(60, \langle 1, 1, 7, 3, 3, 5, 5, 6, 5, 5, 5, 7, 2, 6, 6 \rangle)$

Figure   3.1274   provides   an   example   where   the   SUM_SURF_STEADY $(60, [1, 1, 7, 3, 3, 5, 5, 6, 5, 5, 5, 7, 2, 6, 6])$ constraint holds.

| | |
|---|---|
| **Typical** | $|\text{VARIABLES}| > 1$ |
| **Symmetry** | Items of VARIABLES can be reversed. |
| **Arg. properties** | Functional dependency: VALUE determined by VARIABLES. |

Figure 3.1274: Illustrating the SUM_SURF_STEADY constraint of the **Example** slot

**Automaton**     Figures 3.1275 and 3.1276 respectively depict the automaton associated with the constraint
SUM_SURF_STEADY and its simplified form.

Figure 3.1275: Automaton for the SUM_SURF_STEADY constraint obtained by applying
decoration Table 2.35 to the seed transducer of the STEADY pattern where default is 0

Figure 3.1276: Automaton for the SUM_SURF_STEADY constraint obtained by applying
decoration Table 2.38 to the seed transducer of the STEADY pattern where default is 0

Table 3.180: Glue matrix for the SUM_SURF_STEADY constraint defined as the compo-
sition of the STEADY pattern , the feature SURF , and the aggregator sum ; cells of
the glue matrix are coloured with the colour of the constituent to which they are related.

## 3.604  SUM_SURF_STEADY_SEQUENCE

► ◁ ◀        **DESCRIPTION**              **AUTOMATON**

| **Origin** | Based on the STEADY_SEQUENCE pattern. |
|---|---|

| **Constraint** | SUM_SURF_STEADY_SEQUENCE(VALUE, VARIABLES) |
|---|---|

| **Arguments** | VALUE     : dvar |
|---|---|
| | VARIABLES : collection(var−dvar) |

**Restrictions**

$\mathtt{sv} \leq 1 \Rightarrow \mathtt{VALUE} = 0$

$\mathtt{rv} = 1 \wedge \mathtt{sv} \geq 2 \Rightarrow \mathtt{VALUE} \geq \mathtt{minv} * \mathtt{sv}$①

$\mathtt{rv} \geq 2 \Rightarrow \mathtt{VALUE} = 0 \vee \mathtt{VALUE} \geq \min(2 * \mathtt{minv}②, \mathtt{minv} * \mathtt{sv}③)$

$\mathtt{rv} = 1 \wedge \mathtt{sv} \geq 2 \Rightarrow \mathtt{VALUE} \leq \mathtt{maxv} * \mathtt{sv}$④

$\mathtt{rv} \geq 2 \Rightarrow \mathtt{VALUE} = 0 \vee \mathtt{VALUE} \leq \max(2 * \mathtt{maxv}⑤, \mathtt{maxv} * \mathtt{sv}⑥)$

required(VARIABLES, var)

where

minv = minval(VARIABLES.var)

maxv = maxval(VARIABLES.var)

sv = |VARIABLES|

rv = range(VARIABLES.var)

**Purpose**

VALUE is the sum of the surface of occurrences of the STEADY_SEQUENCE pattern in the time-series given by the VARIABLES collection. If the pattern does not occur, VALUE takes the default value 0.

An occurrence of the pattern STEADY_SEQUENCE is the *maximal* subsequence which matches the regular expression $=^{+}$.

Assume that the occurrence of the pattern STEADY_SEQUENCE starts at position $i$ and ends at position $j$. The feature SURF computes the sum of the values from index $i$ to index $j + 1$.

| **Example** | $(31, \langle 3, 1, 1, 4, 5, 5, 5, 6, 2, 2, 4, 4, 3, 2, 1, 1 \rangle)$ |
|---|---|

Figure 3.1277 provides an example where the SUM_SURF_STEADY_SEQUENCE $(31, [3, 1, 1, 4, 5, 5, 5, 6, 2, 2, 4, 4, 3, 2, 1, 1])$ constraint holds.

| **Typical** | $|\mathtt{VARIABLES}| > 1$ |
|---|---|

| **Symmetry** | Items of VARIABLES can be reversed. |
|---|---|

| **Arg. properties** | Functional dependency: VALUE determined by VARIABLES. |
|---|---|

Figure 3.1277: Illustrating the SUM_SURF_STEADY_SEQUENCE constraint of the **Example** slot

**Automaton**  Figures 3.1278 and 3.1279 respectively depict the automaton associated with the constraint SUM_SURF_STEADY_SEQUENCE and its simplified form.

Figure 3.1278: Automaton for the SUM_SURF_STEADY_SEQUENCE constraint obtained by applying decoration Table 2.35 to the seed transducer of the STEADY_SEQUENCE pattern where default is 0

Figure 3.1279: Automaton for the SUM_SURF_STEADY_SEQUENCE constraint obtained by applying decoration Table 2.38 to the seed transducer of the STEADY_SEQUENCE pattern where default is 0

| | $s$ | $r$ |
|---|---|---|
| $s$ | $\overrightarrow{C} + \overleftarrow{C}$ | $\overrightarrow{C} + \overleftarrow{C}$ |
| $r$ | $\overrightarrow{C} + \overleftarrow{C}$ | $\overrightarrow{C} + \overleftarrow{C} + \overrightarrow{D} + \overleftarrow{D} + \texttt{VAR}_i$ |

Table 3.181: Glue matrix for the SUM_SURF_STEADY_SEQUENCE constraint defined as the composition of the STEADY_SEQUENCE pattern, the feature SURF, and the aggregator sum; cells of the glue matrix are coloured with the colour of the constituent to which they are related.

# 3.605 SUM_SURF_STRICTLY_DECREASING_SEQUENCE

▶    ◁ ◀    **DESCRIPTION**        **AUTOMATON**

| | |
|---|---|
| **Origin** | Based on the STRICTLY_DECREASING_SEQUENCE pattern. |
| **Constraint** | SUM_SURF_STRICTLY_DECREASING_SEQUENCE(VALUE, VARIABLES) |
| **Arguments** | VALUE      :  dvar<br>VARIABLES  :  collection(var−dvar) |

**Restrictions**

$\text{sv} \leq 1 \vee \text{rv} \leq 1 \Rightarrow \text{VALUE} = 0$

$\text{rv} = 2 \Rightarrow \text{VALUE} = 0 \vee \text{VALUE} \geq \min(2 * \text{minv} + 1, (2 * \text{minv} + 1) * \text{np})$

$\text{rv} \geq 3 \Rightarrow$
$$\bigvee \begin{pmatrix} \text{VALUE} = 0, \\ \text{VALUE} \geq \min \begin{pmatrix} 2 * \text{minv} + 1, \\ (2 * \text{minv} + 1) * \text{np} + \min(0, \text{sv} \bmod 2 * (\text{minv} + 2)) \end{pmatrix} \end{pmatrix}$$

$\text{rv} = 2 \Rightarrow \text{VALUE} = 0 \vee \text{VALUE} \leq \max(2 * \text{maxv} - 1①, (2 * \text{maxv} - 1) * \text{np}②)$

$\text{rv} \geq 3 \Rightarrow$
$$\bigvee \begin{pmatrix} \text{VALUE} = 0, \\ \text{VALUE} \leq \max \begin{pmatrix} 2 * \text{maxv} - 1③, \\ (2 * \text{maxv} - 1) * \text{np} + \max(0, \text{sv} \bmod 2 * (\text{maxv} - 2))④ \end{pmatrix} \end{pmatrix}$$

required(VARIABLES, var)

where

$\text{sv} = |\text{VARIABLES}|$
$\text{np} = \lfloor \text{sv}/2 \rfloor$
$\text{minv} =$ minval(VARIABLES.var)
$\text{maxv} =$ maxval(VARIABLES.var)
$\text{rv} =$ range(VARIABLES.var)

**Purpose**

VALUE is the sum of the surface of occurrences of the STRICTLY_DECREASING_SEQUENCE pattern in the time-series given by the VARIABLES collection. If the pattern does not occur, VALUE takes the default value 0.

An occurrence of the pattern STRICTLY_DECREASING_SEQUENCE is the *maximal* subsequence which matches the regular expression $>^+$.

Assume that the occurrence of the pattern STRICTLY_DECREASING_SEQUENCE starts at position $i$ and ends at position $j$. The feature SURF computes the sum of the values from index $i$ to index $j + 1$.

**Example**

$(31, \langle 4, 4, 6, 4, 1, 1, 3, 4, 4, 6, 6, 5, 2, 2, 4, 3 \rangle)$

Figure 3.1280 provides an example where the SUM_SURF_STRICTLY_DECREASING_SEQUENCE $(31, [4, 4, 6, 4, 1, 1, 3, 4, 4, 6, 6, 5, 2, 2, 4, 3])$ constraint holds.

Figure 3.1280: Illustrating the SUM_SURF_STRICTLY_DECREASING_SEQUENCE constraint of the **Example** slot

| **Typical** | $|\text{VARIABLES}| > 1$ |
| | $\text{range}(\text{VARIABLES.var}) > 1$ |

| **Arg. properties** | Functional dependency: VALUE determined by VARIABLES. |

**Automaton**      Figures 3.1281 and 3.1282 respectively depict the automaton associated with the constraint
SUM_SURF_STRICTLY_DECREASING_SEQUENCE and its simplified form.

Figure 3.1281: Automaton for the SUM_SURF_STRICTLY_DECREASING_SEQUENCE
constraint obtained by applying decoration Table 2.35 to the seed transducer of the
STRICTLY_DECREASING_SEQUENCE pattern where default is $0$

Figure 3.1282: Automaton for the SUM_SURF_STRICTLY_DECREASING_SEQUENCE
constraint obtained by applying decoration Table 2.38 to the seed transducer of the
STRICTLY_DECREASING_SEQUENCE pattern where default is $0$

|   | $s$ | $r$ |
|---|---|---|
| $s$ | $\overrightarrow{C} + \overleftarrow{C}$ | $\overrightarrow{C} + \overleftarrow{C}$ |
| $r$ | $\overrightarrow{C} + \overleftarrow{C}$ | $\overrightarrow{C} + \overleftarrow{C} + \overrightarrow{D} + \overleftarrow{D} + \text{VAR}_i$ |

Table 3.182: Glue matrix for the SUM_SURF_STRICTLY_DECREASING_SEQUENCE constraint defined as the composition of the STRICTLY_DECREASING_SEQUENCE pattern , the feature SURF , and the aggregator sum ; cells of the glue matrix are coloured with the colour of the constituent to which they are related.

# 3.606 SUM_SURF_STRICTLY_INCREASING_SEQUENCE

▶    ◁  ◀        DESCRIPTION            AUTOMATON

**Origin**          Based on the STRICTLY_INCREASING_SEQUENCE pattern.

**Constraint**      SUM_SURF_STRICTLY_INCREASING_SEQUENCE(VALUE, VARIABLES)

**Arguments**       VALUE      : dvar
                    VARIABLES  : collection(var−dvar)

**Restrictions**    $\mathtt{sv} \leq 1 \vee \mathtt{rv} \leq 1 \Rightarrow \mathtt{VALUE} = 0$

$\mathtt{rv} = 2 \Rightarrow \mathtt{VALUE} = 0 \vee \mathtt{VALUE} \geq \min(2 * \mathtt{minv} + 1, (2 * \mathtt{minv} + 1) * \mathtt{np})$

$\mathtt{rv} \geq 3 \Rightarrow$
$$\bigvee \left( \begin{array}{l} \mathtt{VALUE} = 0, \\ \mathtt{VALUE} \geq \min \left( \begin{array}{l} 2 * \mathtt{minv} + 1, \\ (2 * \mathtt{minv} + 1) * \mathtt{np} + \min(0, \mathtt{sv} \bmod 2 * (\mathtt{minv} + 2)) \end{array} \right) \end{array} \right)$$

$\mathtt{rv} = 2 \Rightarrow \mathtt{VALUE} = 0 \vee \mathtt{VALUE} \leq \max(2 * \mathtt{maxv} - 1①, (2 * \mathtt{maxv} - 1) * \mathtt{np}②)$

$\mathtt{rv} \geq 3 \Rightarrow$
$$\bigvee \left( \begin{array}{l} \mathtt{VALUE} = 0, \\ \mathtt{VALUE} \leq \max \left( \begin{array}{l} 2 * \mathtt{maxv} - 1③, \\ (2 * \mathtt{maxv} - 1) * \mathtt{np} + \max(0, \mathtt{sv} \bmod 2 * (\mathtt{maxv} - 2))④ \end{array} \right) \end{array} \right)$$

required(VARIABLES, var)

where
$\mathtt{sv} = |\mathtt{VARIABLES}|$
$\mathtt{np} = \lfloor \mathtt{sv}/2 \rfloor$
$\mathtt{minv} =$ minval(VARIABLES.var)
$\mathtt{maxv} =$ maxval(VARIABLES.var)
$\mathtt{rv} =$ range(VARIABLES.var)

**Purpose**

VALUE is the sum of the surface of occurrences of the STRICTLY_INCREASING_SEQUENCE pattern in the time-series given by the VARIABLES collection. If the pattern does not occur, VALUE takes the default value 0.

An occurrence of the pattern STRICTLY_INCREASING_SEQUENCE is the *maximal* sub-sequence which matches the regular expression $<^+$.

Assume that the occurrence of the pattern STRICTLY_INCREASING_SEQUENCE starts at position $i$ and ends at position $j$. The feature SURF computes the sum of the values from index $i$ to index $j + 1$.

**Example**         $(30, \langle 4, 3, 5, 5, 2, 1, 1, 2, 3, 4, 6, 6, 3, 1, 2, 3 \rangle)$

Figure 3.1283 provides an example where the SUM_SURF_STRICTLY_INCREASING_SEQUENCE $(30, [4, 3, 5, 5, 2, 1, 1, 2, 3, 4, 6, 6, 3, 1, 2, 3])$ constraint holds.

Figure 3.1283: Illustrating the SUM_SURF_STRICTLY_INCREASING_SEQUENCE constraint of the **Example** slot

| Typical | $|\texttt{VARIABLES}| > 1$ |
| --- | --- |
| | $\texttt{range}(\texttt{VARIABLES.var}) > 1$ |

| Arg. properties | Functional dependency: VALUE determined by VARIABLES. |
| --- | --- |

**Automaton**　　Figures 3.1284 and 3.1285 respectively depict the automaton associated with the constraint SUM_SURF_STRICTLY_INCREASING_SEQUENCE and its simplified form.

Figure 3.1284: Automaton for the SUM_SURF_STRICTLY_INCREASING_SEQUENCE constraint obtained by applying decoration Table 2.35 to the seed transducer of the STRICTLY_INCREASING_SEQUENCE pattern where default is $0$

Figure 3.1285: Automaton for the SUM_SURF_STRICTLY_INCREASING_SEQUENCE constraint obtained by applying decoration Table 2.38 to the seed transducer of the STRICTLY_INCREASING_SEQUENCE pattern where default is $0$

|   | $s$ | $r$ |
|---|---|---|
| $s$ | $\overrightarrow{C} + \overleftarrow{C}$ | $\overrightarrow{C} + \overleftarrow{C}$ |
| $r$ | $\overrightarrow{C} + \overleftarrow{C}$ | $\overrightarrow{C} + \overleftarrow{C} + \overrightarrow{D} + \overleftarrow{D} + \text{VAR}_i$ |

Table 3.183: Glue matrix for the SUM_SURF_STRICTLY_INCREASING_SEQUENCE constraint defined as the composition of the STRICTLY_INCREASING_SEQUENCE pattern, the feature SURF, and the aggregator sum; cells of the glue matrix are coloured with the colour of the constituent to which they are related.

# 3.607  SUM_SURF_SUMMIT

▶   ◁ ◀   DESCRIPTION        AUTOMATON

$$(< | < (= | <)^* <)(> | > (= | >)^* >)$$

| **Origin** | Based on the SUMMIT pattern. |
|---|---|

| **Constraint** | SUM_SURF_SUMMIT(VALUE, VARIABLES) |
|---|---|

**Arguments**

    VALUE     : dvar
    VARIABLES : collection(var−dvar)

**Restrictions**

$sv \leq 2 \vee rv \leq 1 \Rightarrow VALUE = 0$

$rv = 2 \Rightarrow VALUE = 0 \vee VALUE \geq \min(\text{minv} + 1①, (\text{minv} + 1) * \text{np}②)$

$rv \geq 3 \Rightarrow$
$$\bigvee \left( \begin{array}{l} VALUE = 0, \\ VALUE \geq \min((\text{minv} + 1) * \text{np}③, \min(\text{minv} + 1④, (sv - 2) * (\text{minv} + 1) + 1⑤)) \end{array} \right)$$

$rv = 2 \Rightarrow VALUE = 0 \vee VALUE \leq \max(\text{maxv}, \text{maxv} * \text{np})$

$rv \geq 3 \Rightarrow$
$$\bigvee \left( \begin{array}{l} VALUE = 0, \\ VALUE \leq \max(\text{maxv} * \text{np}, \max(\text{maxv}, (sv - 2) * (\text{maxv} - 1) + 1)) \end{array} \right)$$

required(VARIABLES, var)

where
$sv = |VARIABLES|$
$np = \max(0, \lfloor (sv - 1)/2 \rfloor)$
$\text{minv} = \text{minval}(VARIABLES.var)$
$\text{maxv} = \text{maxval}(VARIABLES.var)$
$rv = \text{range}(VARIABLES.var)$

**Purpose**

VALUE is the sum of the surface of occurrences of the SUMMIT pattern in the time-series given by the VARIABLES collection. If the pattern does not occur, VALUE takes the default value 0.

An occurrence of the pattern SUMMIT is the *maximal* subsequence which matches the regular expression $(< | < (= | <)^* <)(> | > (= | >)^* >)$.

Assume that the occurrence of the pattern SUMMIT starts at position $i$ and ends at position $j$. The feature SURF computes the sum of the values from index $i + 1$ to index $j$.

| **Example** | $(23, \langle 7, 1, 5, 4, 4, 3, 3, 4, 6, 6, 2, 3, 4, 2, 3, 1 \rangle)$ |
|---|---|

Figure 3.1286 provides an example where the SUM_SURF_SUMMIT $(23, [7, 1, 5, 4, 4, 3, 3, 4, 6, 6, 2, 3, 4, 2, 3, 1])$ constraint holds.

**Typical**

$|VARIABLES| > 2$
$\text{range}(VARIABLES.var) > 1$

Figure 3.1286: Illustrating the SUM_SURF_SUMMIT constraint of the **Example** slot

| **Symmetry** | Items of VARIABLES can be reversed. |
|---|---|
| **Arg. properties** | Functional dependency: VALUE determined by VARIABLES. |

**Automaton**     Figures 3.1287 and 3.1288 respectively depict the automaton associated with the constraint SUM_SURF_SUMMIT and its simplified form.

Figure 3.1287: Automaton for the SUM_SURF_SUMMIT constraint obtained by applying decoration Table 2.35 to the seed transducer of the SUMMIT pattern where default is 0 (transition $u \to r$ has the same accumulator update as transition $r \to u$)

Figure 3.1288: Automaton for the SUM_SURF_SUMMIT constraint obtained by applying decoration Table 2.25 to the seed transducer of the SUMMIT pattern where default is 0 (transition $u \rightarrow r$ has the same accumulator update as transition $r \rightarrow u$)

| | $s$ | $r$ | $t$ | $u$ |
|---|---|---|---|---|
| $s$ | $\overrightarrow{C} + \overleftarrow{C}$ | $\overrightarrow{C} + \overleftarrow{C}$ | $\overrightarrow{C} + \overleftarrow{C}$ | $\overrightarrow{C} + \overleftarrow{C}$ |
| $r$ | $\overrightarrow{C} + \overleftarrow{C}$ | $\overrightarrow{D} + \overleftarrow{D} + \text{VAR}_i$ | $\overleftarrow{C} + \overrightarrow{D} + \overleftarrow{D} + \text{VAR}_i$ | $\overrightarrow{C} + \overleftarrow{C}$ |
| $t$ | $\overrightarrow{C} + \overleftarrow{C}$ | $\overrightarrow{C} + \overrightarrow{D} + \overleftarrow{D} + \text{VAR}_i$ | $\overrightarrow{C} + \overleftarrow{C}$ | $\overrightarrow{C} + \overrightarrow{D} + \overleftarrow{D} + \text{VAR}_i$ |
| $u$ | $\overrightarrow{C} + \overleftarrow{C}$ | $\overrightarrow{C} + \overleftarrow{C}$ | $\overleftarrow{C} + \overrightarrow{D} + \overleftarrow{D} + \text{VAR}_i$ | $\overrightarrow{C} + \overleftarrow{C}$ |

Table 3.184: Glue matrix for the SUM_SURF_SUMMIT constraint defined as the composition of the SUMMIT pattern, the feature SURF, and the aggregator sum; cells of the glue matrix are coloured with the colour of the constituent to which they are related.

## 3.608 SUM_SURF_VALLEY

        DESCRIPTION            AUTOMATON

$> (= | >)^* (< | =)^* <$

| **Origin** | Based on the VALLEY pattern. |
|---|---|

| **Constraint** | SUM_SURF_VALLEY(VALUE, VARIABLES) |
|---|---|

| **Arguments** | VALUE      : dvar |
|---|---|
|  | VARIABLES  : collection(var−dvar) |

**Restrictions**

$\text{sv} \leq 2 \vee \text{rv} \leq 1 \Rightarrow \text{VALUE} = 0$
$\text{VALUE} = 0 \vee \text{VALUE} \geq \min(\text{minv}①, \text{minv} * (\text{sv} - 2)②)$
$\text{VALUE} = 0 \vee \text{VALUE} \leq \max(\text{maxv} - 1③, (\text{maxv} - 1) * (\text{sv} - 2)④)$
required(VARIABLES, var)
where
minv =minval(VARIABLES.var)
maxv =maxval(VARIABLES.var)
$\text{sv} = |\text{VARIABLES}|$
rv =range(VARIABLES.var)

**Purpose**

VALUE is the sum of the surface of occurrences of the VALLEY pattern in the time-series given by the VARIABLES collection. If the pattern does not occur, VALUE takes the default value 0.

An occurrence of the pattern VALLEY is the *maximal* subsequence which matches the regular expression $> (= | >)^* (< | =)^* <$.

Assume that the occurrence of the pattern VALLEY starts at position $i$ and ends at position $j$. The feature SURF computes the sum of the values from index $i + 1$ to index $j$.

| **Example** | $(35, \langle 1, 3, 7, 4, 3, 6, 6, 5, 3, 3, 2, 6, 5, 5, 5, 7 \rangle)$ |
|---|---|

Figure 3.1289 provides an example where the SUM_SURF_VALLEY $(35, [1, 3, 7, 4, 3, 6, 6, 5, 3, 3, 2, 6, 5, 5, 5, 7])$ constraint holds.

| **Typical** | $|\text{VARIABLES}| > 2$ |
|---|---|
|  | range(VARIABLES.var) $> 1$ |

| **Symmetry** | Items of VARIABLES can be reversed. |
|---|---|

| **Arg. properties** | Functional dependency: VALUE determined by VARIABLES. |
|---|---|

Figure 3.1289: Illustrating the SUM_SURF_VALLEY constraint of the **Example** slot

**Automaton**   Figures 3.1290 and 3.1291 respectively depict the automaton associated with the constraint SUM_SURF_VALLEY and its simplified form.

Figure 3.1290: Automaton for the SUM_SURF_VALLEY constraint obtained by applying decoration Table 2.35 to the seed transducer of the VALLEY pattern where default is 0

Figure 3.1291: Automaton for the SUM_SURF_VALLEY constraint obtained by applying decoration Table 2.25 to the seed transducer of the VALLEY pattern where default is 0

| | $s$ | $r$ | $t$ |
|---|---|---|---|
| $s$ | $\overrightarrow{C} + \overleftarrow{C}$ | $\overrightarrow{C} + \overleftarrow{C}$ | $\overrightarrow{C} + \overleftarrow{C}$ |
| $r$ | $\overrightarrow{C} + \overleftarrow{C}$ | $\overrightarrow{D} + \overleftarrow{D} + \texttt{VAR}_i$ | $\overleftarrow{C} + \overrightarrow{D} + \overleftarrow{D} + \texttt{VAR}_i$ |
| $t$ | $\overrightarrow{C} + \overleftarrow{C}$ | $\overrightarrow{C} + \overrightarrow{D} + \overleftarrow{D} + \texttt{VAR}_i$ | $\overrightarrow{C} + \overleftarrow{C}$ |

Table 3.185: Glue matrix for the SUM_SURF_VALLEY constraint defined as the composition of the VALLEY pattern, the feature SURF, and the aggregator sum; cells of the glue matrix are coloured with the colour of the constituent to which they are related.

## 3.609  SUM_SURF_ZIGZAG

▶   ◁ ◀      DESCRIPTION          AUTOMATON

| **Origin** | Based on the ZIGZAG pattern. |
|---|---|

| **Constraint** | SUM_SURF_ZIGZAG(VALUE, VARIABLES) |
|---|---|

| **Arguments** | VALUE      : dvar |
|---|---|
| | VARIABLES : collection(var−dvar) |

**Restrictions**

$sv \leq 3 \vee rv \leq 1 \Rightarrow \text{VALUE} = 0$

$$\bigvee \left( \begin{array}{l} \text{VALUE} = 0, \\ \text{VALUE} \geq \min \left( \begin{array}{l} 2 * \text{minv} + 1 ①, \\ \lfloor (sv-1)/2 \rfloor * \text{minv} + \lfloor (sv-2)/2 \rfloor * (\text{minv}+1) ② \end{array} \right) \end{array} \right)$$

$$\bigvee \left( \begin{array}{l} \text{VALUE} = 0, \\ \text{VALUE} \leq \max \left( \begin{array}{l} 2 * \text{maxv} - 1 ③, \\ \lfloor (sv-1)/2 \rfloor * \text{maxv} + \lfloor (sv-2)/2 \rfloor * (\text{maxv}-1) ④ \end{array} \right) \end{array} \right)$$

required(VARIABLES, var)

where

minv = minval(VARIABLES.var)
maxv = maxval(VARIABLES.var)
sv = |VARIABLES|
rv = range(VARIABLES.var)

**Purpose**

VALUE is the sum of the surface of occurrences of the ZIGZAG pattern in the time-series given by the VARIABLES collection. If the pattern does not occur, VALUE takes the default value 0.

An occurrence of the pattern ZIGZAG is the *maximal* subsequence which matches the regular expression $(<>)^+(<\,|\,<>)\,|\,(><)^+(>\,|\,><)$.

Assume that the occurrence of the pattern ZIGZAG starts at position $i$ and ends at position $j$. The feature SURF computes the sum of the values from index $i+1$ to index $j$.

**Example**

$(33, \langle 4, 1, 3, 1, 4, 6, 1, 5, 5, 2, 7, 2, 3, 1, 6, 1 \rangle)$

Figure 3.1292 provides an example where the SUM_SURF_ZIGZAG $(33, [4, 1, 3, 1, 4, 6, 1, 5, 5, 2, 7, 2, 3, 1, 6, 1])$ constraint holds.

| **Typical** | $|\text{VARIABLES}| > 3$ |
|---|---|
| | range(VARIABLES.var) > 1 |

| **Symmetry** | Items of VARIABLES can be reversed. |
|---|---|

| **Arg. properties** | Functional dependency: VALUE determined by VARIABLES. |
|---|---|

Figure 3.1292: Illustrating the SUM_SURF_ZIGZAG constraint of the **Example** slot

**Automaton**        Figures 3.1293 and 3.1294 respectively depict the automaton associated with the constraint SUM_SURF_ZIGZAG and its simplified form.

Figure 3.1293: Automaton for the SUM_SURF_ZIGZAG constraint obtained by applying decoration Table 2.35 to the seed transducer of the ZIGZAG pattern where default is 0; (1) missing transitions from $a, b, c, d, e, f$ to $s$ are labelled by $=$; (2) on transitions from $b, c, e, f$ to $s$ the accumulator $D$ is reset to its initial value; (3) on transitions from $c, f$ to $s$ the accumulator $R$ is updated wrt $C$ and the accumulator $C$ is reset to its initial value

Figure 3.1294: Automaton for the SUM_SURF_ZIGZAG constraint obtained by applying decoration Table 2.28 to the seed transducer of the ZIGZAG pattern where default is 0; (1) missing transitions from $a, b, c, d, e, f$ to $s$ are labelled by $=$; (2) on transitions from $b, c, e, f$ to $s$ the accumulator $D$ is reset to its initial value

| | s | a | b | c | d | e | f |
|---|---|---|---|---|---|---|---|
| **s** | $\overleftarrow{C}+\overrightarrow{C}$ | $\overleftarrow{C}+\overrightarrow{C}$ | $\overleftarrow{C}+\overrightarrow{C}$ | $\overleftarrow{C}+\overrightarrow{C}$ | $\overleftarrow{C}+\overrightarrow{C}$ | $\overleftarrow{C}+\overrightarrow{C}$ | $\overrightarrow{C}+\overleftarrow{C}$ |
| **a** | $\overleftarrow{C}+\overrightarrow{C}$ | $\overleftarrow{C}+\overrightarrow{C}$ | $\overleftarrow{C}+\overrightarrow{C}$ | $\overleftarrow{C}+\overrightarrow{C}$ | $\overleftarrow{C}+\overrightarrow{C}$ | $\overleftarrow{C}+\overrightarrow{C}$ | $\overrightarrow{C}+\overleftarrow{C}$ |
| **b** | $\overleftarrow{C}+\overrightarrow{C}$ | $\overrightarrow{C}+\overrightarrow{D}+\overleftarrow{D}+VAR_i$ | $\overrightarrow{D}+\overleftarrow{D}+VAR_i$ | $\overleftarrow{C}+\overrightarrow{D}+\overleftarrow{D}+VAR_i$ | $\overrightarrow{D}+\overleftarrow{D}+VAR_i$ | $\overrightarrow{D}+\overleftarrow{D}+VAR_i$ | $\overleftarrow{C}+\overrightarrow{D}+\overleftarrow{D}+VAR_i$ |
| **c** | $\overleftarrow{C}+\overrightarrow{C}$ | $\overrightarrow{C}+\overleftarrow{C}$ | $\overrightarrow{C}+\overleftarrow{C}$ | $\overrightarrow{C}+\overleftarrow{C}+\overrightarrow{D}+\overleftarrow{D}+VAR_i$ | $\overrightarrow{C}+\overleftarrow{C}$ | $\overrightarrow{C}+\overrightarrow{D}+\overleftarrow{D}+VAR_i$ | $\overrightarrow{C}+\overleftarrow{C}$ |
| **d** | $\overleftarrow{C}+\overrightarrow{C}$ | $\overrightarrow{D}+\overleftarrow{D}+VAR_i$ | $\overrightarrow{D}+\overleftarrow{D}+VAR_i$ | $\overrightarrow{C}+\overleftarrow{D}+\overleftarrow{D}+VAR_i$ | $\overrightarrow{D}+\overleftarrow{D}+VAR_i$ | $\overrightarrow{D}+\overleftarrow{D}+VAR_i$ | $\overleftarrow{C}+\overrightarrow{D}+\overleftarrow{D}+VAR_i$ |
| **e** | $\overleftarrow{C}+\overrightarrow{C}$ | $\overrightarrow{C}+\overleftarrow{C}$ | $\overrightarrow{C}+\overleftarrow{C}$ | $\overleftarrow{C}+\overrightarrow{D}+\overleftarrow{D}+VAR_i$ | $\overrightarrow{C}+\overleftarrow{C}$ | $\overrightarrow{C}+\overleftarrow{C}$ | $\overrightarrow{C}+\overleftarrow{C}$ |
| **f** | $\overleftarrow{C}+\overrightarrow{C}$ | $\overrightarrow{C}+\overleftarrow{C}$ | $\overrightarrow{C}+\overrightarrow{D}+\overleftarrow{D}+VAR_i$ | $\overrightarrow{C}+\overleftarrow{C}$ | $\overrightarrow{C}+\overrightarrow{D}+\overleftarrow{D}+VAR_i$ | $\overrightarrow{C}+\overleftarrow{C}$ | $\overrightarrow{C}+\overleftarrow{C}+\overrightarrow{D}+\overleftarrow{D}+VAR_i$ |

Table 3.186: Glue matrix for the SUM_SURF_ZIGZAG constraint defined as the composition of the ZIGZAG pattern, the feature SURF, and the aggregator sum; cells of the glue matrix are coloured with the colour of the constituent to which they are related.

# 3.610　SUM_WIDTH_DECREASING_SEQUENCE

▶　　◁ ◀　　**DESCRIPTION**　　　**AUTOMATON**

| | |
|---|---|
| **Origin** | Based on the DECREASING_SEQUENCE pattern. |
| **Constraint** | SUM_WIDTH_DECREASING_SEQUENCE(VALUE, VARIABLES) |
| **Arguments** | VALUE　　　: dvar <br> VARIABLES : collection(var−dvar) |

**Restrictions**

$$sv \leq 1 \vee rv \leq 1 \Rightarrow \text{VALUE} = 0$$
$$\text{VALUE} = 0 \vee \text{VALUE} \geq 2$$
$$rv = 2 \Rightarrow \text{VALUE} \leq 2 * \text{np}①$$
$$rv \geq 3 \Rightarrow \text{VALUE} \leq sv②$$
$$\text{required}(\text{VARIABLES}, \text{var})$$

where

$$rv = \text{range}(\text{VARIABLES.var})$$
$$sv = |\text{VARIABLES}|$$
$$np = \lfloor sv/2 \rfloor$$

**Purpose**

VALUE is the sum of the width of occurrences of the DECREASING_SEQUENCE pattern in the time-series given by the VARIABLES collection. If the pattern does not occur, VALUE takes the default value 0.

An occurrence of the pattern DECREASING_SEQUENCE is the *maximal* subsequence which matches the regular expression $> (> | =)^* | >$.

Assume that the occurrence of the pattern DECREASING_SEQUENCE starts at position $i$ and ends at position $j$. The feature WIDTH computes the value $j - i + 2$.

**Example**

$$(9, \langle 3, 4, 2, 2, 5, 6, 6, 4, 4, 3, 1, 1, 4, 6, 4, 4 \rangle)$$

Figure 3.1295 provides an example where the SUM_WIDTH_DECREASING_SEQUENCE $(9, [3, 4, 2, 2, 5, 6, 6, 4, 4, 3, 1, 1, 4, 6, 4, 4])$ constraint holds.

| | |
|---|---|
| **Typical** | $|\text{VARIABLES}| > 1$ <br> $\text{range}(\text{VARIABLES.var}) > 1$ |
| **Symmetry** | One and the same constant can be added to the var attribute of all items of VARIABLES. |
| **Arg. properties** | Functional dependency: VALUE determined by VARIABLES. |

Figure 3.1295: Illustrating the SUM_WIDTH_DECREASING_SEQUENCE constraint of the **Example** slot

**Automaton**  Figures 3.1296 and 3.1297 respectively depict the automaton associated with the constraint SUM_WIDTH_DECREASING_SEQUENCE and its simplified form.

Figure 3.1296: Automaton for the SUM_WIDTH_DECREASING_SEQUENCE constraint obtained by applying decoration Table 2.35 to the seed transducer of the DECREASING_SEQUENCE pattern where default is 0

Table 3.187: Glue matrix for the SUM_WIDTH_DECREASING_SEQUENCE constraint defined as the composition of the DECREASING_SEQUENCE pattern , the feature WIDTH , and the aggregator sum ; cells of the glue matrix are coloured with the colour of the constituent to which they are related.

Figure 3.1297: Automaton for the SUM_WIDTH_DECREASING_SEQUENCE constraint obtained by applying decoration Table 2.29 to the seed transducer of the DECREAS-ING_SEQUENCE pattern where default is 0; $R_i - R_{i-1} \geq 0$ and $-R_i + R_{i-1} + D_{i-1} + 1 \geq 0$ are linear invariants.

# 3.611  SUM_WIDTH_DECREASING_TERRACE

► ◁ ◀      **DESCRIPTION**          **AUTOMATON**

**Origin**            Based on the DECREASING_TERRACE pattern.

**Constraint**        SUM_WIDTH_DECREASING_TERRACE(VALUE, VARIABLES)

**Arguments**         VALUE      : dvar
                      VARIABLES  : collection(var−dvar)

**Restrictions**      $\mathtt{sv} \leq 3 \vee \mathtt{rv} \leq 2 \Rightarrow \mathtt{VALUE} = 0$
                      $\mathtt{VALUE} = 0 \vee \mathtt{VALUE} \geq 2$
                      $\mathtt{VALUE} \leq \max(0, \mathtt{sv} - 2\textcircled{1})$
                      required(VARIABLES, var)
                      where
                      $\mathtt{sv} = |\mathtt{VARIABLES}|$
                      $\mathtt{rv} = \mathtt{range}(\mathtt{VARIABLES.var})$

**Purpose**           VALUE is the sum of the width of occurrences of the DECREASING_TERRACE pattern in
                      the time-series given by the VARIABLES collection. If the pattern does not occur, VALUE
                      takes the default value 0.
                      An occurrence of the pattern DECREASING_TERRACE is the *maximal* subsequence which
                      matches the regular expression $> =^+ >$.
                      Assume that the occurrence of the pattern DECREASING_TERRACE starts at position $i$
                      and ends at position $j$. The feature WIDTH computes the value $j - i$.

**Example**           $(4, \langle 6, 4, 4, 4, 5, 2, 2, 1, 3, 3, 5, 4, 4, 3, 3, 3 \rangle)$

                      Figure 3.1298 provides an example where the SUM_WIDTH_DECREASING_TERRACE
                      $(4, [6, 4, 4, 4, 5, 2, 2, 1, 3, 3, 5, 4, 4, 3, 3, 3])$ constraint holds.

**Typical**           $|\mathtt{VARIABLES}| > 3$
                      $\mathtt{range}(\mathtt{VARIABLES.var}) > 2$

**Symmetry**          One and the same constant can be added to the var attribute of all items of VARIABLES.

**Arg. properties**   Functional dependency: VALUE determined by VARIABLES.

Figure 3.1298: Illustrating the SUM_WIDTH_DECREASING_TERRACE constraint of the **Example** slot

**Automaton**  Figures 3.1299 and 3.1300 respectively depict the automaton associated with the constraint SUM_WIDTH_DECREASING_TERRACE and its simplified form.

Figure 3.1299: Automaton for the SUM_WIDTH_DECREASING_TERRACE constraint obtained by applying decoration Table 2.35 to the seed transducer of the DECREAS-ING_TERRACE pattern where default is 0

|   | $s$ | $r$ | $t$ |
|---|---|---|---|
| $s$ | $\overrightarrow{C} + \overleftarrow{C}$ | $\overrightarrow{C} + \overleftarrow{C}$ | $\overrightarrow{C} + \overleftarrow{C}$ |
| $r$ | $\overrightarrow{C} + \overleftarrow{C}$ | $\overrightarrow{C} + \overleftarrow{C}$ | $\overrightarrow{D} + \overleftarrow{D} + 1$ |
| $t$ | $\overrightarrow{C} + \overleftarrow{C}$ | $\overrightarrow{D} + \overleftarrow{D} + 1$ | $\overrightarrow{D} + \overleftarrow{D} + 1$ |

Table 3.188: Glue matrix for the SUM_WIDTH_DECREASING_TERRACE constraint defined as the composition of the DECREASING_TERRACE pattern , the feature WIDTH , and the aggregator sum ; cells of the glue matrix are coloured with the colour of the constituent to which they are related.

Figure 3.1300: Automaton for the SUM_WIDTH_DECREASING_TERRACE constraint obtained by applying decoration Table 2.28 to the seed transducer of the DECREAS-ING_TERRACE pattern where default is $0$; $R_i - R_{i-1} \geq 0$ and $-R_i + R_{i-1} + D_{i-1} + 1 \geq 0$ are linear invariants.

# 3.612  SUM_WIDTH_GORGE

► ◁ ◀        **DESCRIPTION**        **AUTOMATON**

$(> \mid > (= \mid >)^{*} >)(< \mid < (= \mid <)^{*} <)$

| | |
|---|---|
| **Origin** | Based on the GORGE pattern. |
| **Constraint** | SUM_WIDTH_GORGE(VALUE, VARIABLES) |
| **Arguments** | VALUE     : dvar<br>VARIABLES : collection(var−dvar) |
| **Restrictions** | $\mathtt{sv} \leq 2 \vee \mathtt{rv} \leq 1 \Rightarrow \mathtt{VALUE} = 0$<br>$\mathtt{VALUE} = 0 \vee \mathtt{VALUE} \geq 1$<br>$\mathtt{rv} = 2 \Rightarrow \mathtt{VALUE} \leq \mathtt{np}①$<br>$\mathtt{rv} \geq 3 \Rightarrow \mathtt{VALUE} \leq \max(0, \mathtt{sv} - 2②)$<br>required(VARIABLES, var)<br>where<br>$\mathtt{rv} =$ range(VARIABLES.var)<br>$\mathtt{sv} = \mid$VARIABLES$\mid$<br>$\mathtt{np} = \max(0, \lfloor (\mathtt{sv} - 1)/2 \rfloor)$ |
| **Purpose** | VALUE is the sum of the width of occurrences of the GORGE pattern in the time-series given by the VARIABLES collection. If the pattern does not occur, VALUE takes the default value 0.<br>An occurrence of the pattern GORGE is the *maximal* subsequence which matches the regular expression $(> \mid > (= \mid >)^{*} >)(< \mid < (= \mid <)^{*} <)$.<br>Assume that the occurrence of the pattern GORGE starts at position $i$ and ends at position $j$. The feature WIDTH computes the value $j - i$. |
| **Example** | $(6, \langle 1, 7, 3, 4, 4, 5, 5, 4, 2, 2, 6, 5, 4, 6, 5, 7 \rangle)$ |

Figure 3.1301 provides an example where the SUM_WIDTH_GORGE $(6, [1, 7, 3, 4, 4, 5, 5, 4, 2, 2, 6, 5, 4, 6, 5, 7])$ constraint holds.

| | |
|---|---|
| **Typical** | $\mid$VARIABLES$\mid > 2$<br>range(VARIABLES.var) $> 1$ |
| **Symmetries** | • Items of VARIABLES can be reversed.<br>• One and the same constant can be added to the var attribute of all items of VARIABLES. |
| **Arg. properties** | Functional dependency: VALUE determined by VARIABLES. |

Figure 3.1301: Illustrating the SUM_WIDTH_GORGE constraint of the **Example** slot

**Automaton**    Figures 3.1302 and 3.1303 respectively depict the automaton associated with the constraint SUM_WIDTH_GORGE and its simplified form.

Figure 3.1302: Automaton for the SUM_WIDTH_GORGE constraint obtained by applying decoration Table 2.35 to the seed transducer of the GORGE pattern where default is 0 (transition $u \rightarrow r$ has the same accumulator update as transition $r \rightarrow u$)

Figure 3.1303: Automaton for the SUM_WIDTH_GORGE constraint obtained by applying decoration Table 2.25 to the seed transducer of the GORGE pattern where default is 0 (transition $u \rightarrow r$ has the same accumulator update as transition $r \rightarrow u$); $R_i - R_{i-1} \geq 0$ and $-R_i + R_{i-1} + D_{i-1} + 1 \geq 0$ are linear invariants.

|   | $s$ | $r$ | $t$ | $u$ |
|---|---|---|---|---|
| $s$ | $\overrightarrow{C}+\overleftarrow{C}$ | $\overrightarrow{C}+\overleftarrow{C}$ | $\overrightarrow{C}+\overleftarrow{C}$ | $\overrightarrow{C}+\overleftarrow{C}$ |
| $r$ | $\overrightarrow{C}+\overleftarrow{C}$ | $\overrightarrow{D}+\overleftarrow{D}+1$ | $\overleftarrow{C}+\overrightarrow{D}+\overleftarrow{D}+1$ | $\overrightarrow{C}+\overleftarrow{C}$ |
| $t$ | $\overrightarrow{C}+\overleftarrow{C}$ | $\overrightarrow{C}+\overrightarrow{D}+\overleftarrow{D}+1$ | $\overrightarrow{C}+\overleftarrow{C}$ | $\overrightarrow{C}+\overrightarrow{D}+\overleftarrow{D}+1$ |
| $u$ | $\overrightarrow{C}+\overleftarrow{C}$ | $\overrightarrow{C}+\overleftarrow{C}$ | $\overleftarrow{C}+\overrightarrow{D}+\overleftarrow{D}+1$ | $\overrightarrow{C}+\overleftarrow{C}$ |

Table 3.189: Glue matrix for the SUM_WIDTH_GORGE constraint defined as the composition of the GORGE pattern , the feature WIDTH , and the aggregator sum ; cells of the glue matrix are coloured with the colour of the constituent to which they are related.

# 3.613   SUM_WIDTH_INCREASING_SEQUENCE

► ◁ ◄    **DESCRIPTION**        **AUTOMATON**

$< (< | =)^* < |$
$<$

| | |
|---|---|
| **Origin** | Based on the INCREASING_SEQUENCE pattern. |
| **Constraint** | SUM_WIDTH_INCREASING_SEQUENCE(VALUE, VARIABLES) |
| **Arguments** | VALUE       : dvar<br>VARIABLES : collection(var−dvar) |
| **Restrictions** | $\mathtt{sv} \leq 1 \vee \mathtt{rv} \leq 1 \Rightarrow \mathtt{VALUE} = 0$<br>$\mathtt{VALUE} = 0 \vee \mathtt{VALUE} \geq 2$<br>$\mathtt{rv} = 2 \Rightarrow \mathtt{VALUE} \leq 2 * \mathtt{np}①$<br>$\mathtt{rv} \geq 3 \Rightarrow \mathtt{VALUE} \leq \mathtt{sv}②$<br>required(VARIABLES, var)<br>where<br>$\mathtt{rv} =$ range(VARIABLES.var)<br>$\mathtt{sv} = |\mathtt{VARIABLES}|$<br>$\mathtt{np} = \lfloor \mathtt{sv}/2 \rfloor$ |

VALUE ≤ **6** = 2 * 3

VALUE ≤ **7**

| | |
|---|---|
| **Purpose** | VALUE is the sum of the width of occurrences of the INCREASING_SEQUENCE pattern in the time-series given by the VARIABLES collection. If the pattern does not occur, VALUE takes the default value 0.<br>An occurrence of the pattern INCREASING_SEQUENCE is the *maximal* subsequence which matches the regular expression $< (< | =)^* < | <$.<br>Assume that the occurrence of the pattern INCREASING_SEQUENCE starts at position $i$ and ends at position $j$. The feature WIDTH computes the value $j - i + 2$. |

| | |
|---|---|
| **Example** | $(9, \langle 4, 3, 5, 5, 2, 1, 1, 3, 3, 4, 6, 6, 3, 1, 3, 3 \rangle)$ |

Figure 3.1304 provides an example where the SUM_WIDTH_INCREASING_SEQUENCE $(9, [4, 3, 5, 5, 2, 1, 1, 3, 3, 4, 6, 6, 3, 1, 3, 3])$ constraint holds.

| | |
|---|---|
| **Typical** | $|\mathtt{VARIABLES}| > 1$<br>range(VARIABLES.var) > 1 |
| **Symmetry** | One and the same constant can be added to the var attribute of all items of VARIABLES. |
| **Arg. properties** | Functional dependency: VALUE determined by VARIABLES. |

Figure 3.1304: Illustrating the SUM_WIDTH_INCREASING_SEQUENCE constraint of the **Example** slot

**Automaton**     Figures 3.1305 and 3.1306 respectively depict the automaton associated with the constraint
                  SUM_WIDTH_INCREASING_SEQUENCE and its simplified form.

Figure 3.1305: Automaton for the SUM_WIDTH_INCREASING_SEQUENCE constraint
obtained by applying decoration Table 2.35 to the seed transducer of the INCREAS-
ING_SEQUENCE pattern where default is 0

Table 3.190: Glue matrix for the SUM_WIDTH_INCREASING_SEQUENCE con-
straint defined as the composition of the INCREASING_SEQUENCE pattern , the
feature WIDTH , and the aggregator sum ; cells of the glue matrix are coloured with
the colour of the constituent to which they are related.

Figure 3.1306: Automaton for the SUM_WIDTH_INCREASING_SEQUENCE constraint obtained by applying decoration Table 2.29 to the seed transducer of the INCREAS-ING_SEQUENCE pattern where default is 0; $R_i - R_{i-1} \geq 0$ and $-R_i + R_{i-1} + D_{i-1} + 1 \geq 0$ are linear invariants.

# 3.614 SUM_WIDTH_INCREASING_TERRACE

   **DESCRIPTION**      **AUTOMATON**

| | |
|---|---|
| **Origin** | Based on the INCREASING_TERRACE pattern. |
| **Constraint** | SUM_WIDTH_INCREASING_TERRACE(VALUE, VARIABLES) |
| **Arguments** | VALUE : dvar<br>VARIABLES : collection(var−dvar) |

**Restrictions**

$\mathtt{sv} \leq 3 \vee \mathtt{rv} \leq 2 \Rightarrow \mathtt{VALUE} = 0$
$\mathtt{VALUE} = 0 \vee \mathtt{VALUE} \geq 2$
$\mathtt{VALUE} \leq \max(0, \mathtt{sv} - 2①)$
required(VARIABLES, var)
where
$\mathtt{sv} = |\mathtt{VARIABLES}|$
$\mathtt{rv} = $ range(VARIABLES.var)

**Purpose**

VALUE is the sum of the width of occurrences of the INCREASING_TERRACE pattern in the time-series given by the VARIABLES collection. If the pattern does not occur, VALUE takes the default value 0.

An occurrence of the pattern INCREASING_TERRACE is the *maximal* subsequence which matches the regular expression $<=^+<$.

Assume that the occurrence of the pattern INCREASING_TERRACE starts at position $i$ and ends at position $j$. The feature WIDTH computes the value $j - i$.

**Example**

$(5, \langle 1, 3, 3, 3, 2, 5, 5, 6, 4, 4, 2, 3, 3, 3, 4, 4 \rangle)$

Figure 3.1307 provides an example where the SUM_WIDTH_INCREASING_TERRACE $(5, [1, 3, 3, 3, 2, 5, 5, 6, 4, 4, 2, 3, 3, 3, 4, 4])$ constraint holds.

**Typical**

$|\mathtt{VARIABLES}| > 3$
range(VARIABLES.var) $> 2$

**Symmetry**

One and the same constant can be added to the var attribute of all items of VARIABLES.

**Arg. properties**

Functional dependency: VALUE determined by VARIABLES.

Figure 3.1307: Illustrating the SUM_WIDTH_INCREASING_TERRACE constraint of the **Example** slot

**Automaton**    Figures 3.1308 and 3.1309 respectively depict the automaton associated with the constraint
SUM_WIDTH_INCREASING_TERRACE and its simplified form.

Figure 3.1308: Automaton for the SUM_WIDTH_INCREASING_TERRACE constraint
obtained by applying decoration Table 2.35 to the seed transducer of the INCREAS-
ING_TERRACE pattern where default is 0

|     | $s$ | $r$ | $t$ |
|-----|-----|-----|-----|
| $s$ | $\overrightarrow{C} + \overleftarrow{C}$ | $\overrightarrow{C} + \overleftarrow{C}$ | $\overrightarrow{C} + \overleftarrow{C}$ |
| $r$ | $\overrightarrow{C} + \overleftarrow{C}$ | $\overrightarrow{C} + \overleftarrow{C}$ | $\overrightarrow{D} + \overleftarrow{D} + 1$ |
| $t$ | $\overrightarrow{C} + \overleftarrow{C}$ | $\overrightarrow{D} + \overleftarrow{D} + 1$ | $\overrightarrow{D} + \overleftarrow{D} + 1$ |

Table 3.191: Glue matrix for the SUM_WIDTH_INCREASING_TERRACE constraint de-
fined as the composition of the INCREASING_TERRACE pattern , the feature WIDTH ,
and the aggregator sum ; cells of the glue matrix are coloured with the colour of the
constituent to which they are related.

Figure 3.1309: Automaton for the SUM_WIDTH_INCREASING_TERRACE constraint obtained by applying decoration Table 2.28 to the seed transducer of the INCREAS-ING_TERRACE pattern where default is 0; $R_i - R_{i-1} \geq 0$ and $-R_i + R_{i-1} + D_{i-1} + 1 \geq 0$ are linear invariants.

# 3.615　SUM_WIDTH_INFLEXION

　　**DESCRIPTION**　　　　**AUTOMATON**

$< (< | =)^* > | > (> | =)^* <$

| | |
|---|---|
| **Origin** | Based on the INFLEXION pattern. |
| **Constraint** | SUM_WIDTH_INFLEXION(VALUE, VARIABLES) |
| **Arguments** | VALUE　　　: dvar<br>VARIABLES : collection(var−dvar) |
| **Restrictions** | $\mathtt{sv} \leq 2 \vee \mathtt{rv} \leq 1 \Rightarrow \mathtt{VALUE} = 0$<br>$\mathtt{VALUE} = 0 \vee \mathtt{VALUE} \geq 1$<br>$\mathtt{VALUE} \leq \max(0, \mathtt{sv} - 2①)$<br>required(VARIABLES, var)<br>where<br>$\mathtt{sv} = \|\mathtt{VARIABLES}\|$<br>$\mathtt{rv} = \mathtt{range}(\mathtt{VARIABLES.var})$ |

VALUE ≤ **5** = 7 − 2

| | |
|---|---|
| **Purpose** | VALUE is the sum of the width of occurrences of the INFLEXION pattern in the time-series given by the VARIABLES collection. If the pattern does not occur, VALUE takes the default value 0.<br>An occurrence of the pattern INFLEXION is the *maximal* subsequence which matches the regular expression $< (< | =)^* > | > (> | =)^* <$.<br>Assume that the occurrence of the pattern INFLEXION starts at position $i$ and ends at position $j$. The feature WIDTH computes the value $j - i$. |

| | |
|---|---|
| **Example** | $(13, \langle 1, 2, 6, 6, 4, 4, 3, 5, 2, 5, 1, 5, 3, 3, 4, 4 \rangle)$ |

Figure 3.1310 provides an example where the SUM_WIDTH_INFLEXION $(13, [1, 2, 6, 6, 4, 4, 3, 5, 2, 5, 1, 5, 3, 3, 4, 4])$ constraint holds.

| | |
|---|---|
| **Typical** | $\|\mathtt{VARIABLES}\| > 2$<br>$\mathtt{range}(\mathtt{VARIABLES.var}) > 1$ |
| **Symmetries** | • Items of VARIABLES can be reversed.<br>• One and the same constant can be added to the var attribute of all items of VARIABLES. |
| **Arg. properties** | Functional dependency: VALUE determined by VARIABLES. |

Figure 3.1310: Illustrating the SUM_WIDTH_INFLEXION constraint of the **Example** slot

**Automaton**    Figures 3.1311 and 3.1312 respectively depict the automaton associated with the constraint SUM_WIDTH_INFLEXION and its simplified form.

Figure 3.1311: Automaton for the SUM_WIDTH_INFLEXION constraint obtained by applying decoration Table 2.35 to the seed transducer of the INFLEXION pattern where default is 0 (transition $r \rightarrow t$ has the same accumulators updates as transition $t \rightarrow r$)

Figure 3.1312: Automaton for the SUM_WIDTH_INFLEXION constraint obtained by applying decoration Table 2.25 to the seed transducer of the INFLEXION pattern where default is 0 (transition $r \rightarrow t$ has the same accumulators updates as transition $t \rightarrow r$); $R_i - R_{i-1} \geq 0$ and $-R_i + R_{i-1} + D_{i-1} + 1 \geq 0$ are linear invariants.

## 3.616  SUM_WIDTH_PEAK

▶    ◁ ◀    **DESCRIPTION**        **AUTOMATON**

$< (= | <)^* (> | =)^* >$

| **Origin** | Based on the PEAK pattern. |

| **Constraint** | SUM_WIDTH_PEAK(VALUE, VARIABLES) |

| **Arguments** | VALUE      : dvar |
| | VARIABLES : collection(var−dvar) |

**Restrictions**

$\mathtt{sv} \leq 2 \vee \mathtt{rv} \leq 1 \Rightarrow \mathtt{VALUE} = 0$
$\mathtt{VALUE} = 0 \vee \mathtt{VALUE} \geq 1$
$\mathtt{VALUE} \leq \max(0, \mathtt{sv} - 2①)$
required(VARIABLES, var)
where
$\mathtt{sv} = |\mathtt{VARIABLES}|$
$\mathtt{rv} = \text{range}(\mathtt{VARIABLES.var})$

**Purpose**

VALUE is the sum of the width of occurrences of the PEAK pattern in the time-series given by the VARIABLES collection. If the pattern does not occur, VALUE takes the default value 0.

An occurrence of the pattern PEAK is the *maximal* subsequence which matches the regular expression $< (= | <)^* (> | =)^* >$.

Assume that the occurrence of the pattern PEAK starts at position $i$ and ends at position $j$. The feature WIDTH computes the value $j - i$.

**Example**

$(8, \langle 7, 5, 5, 1, 4, 5, 2, 2, 3, 5, 6, 2, 3, 3, 3, 1 \rangle)$

Figure 3.1313 provides an example where the SUM_WIDTH_PEAK $(8, [7, 5, 5, 1, 4, 5, 2, 2, 3, 5, 6, 2, 3, 3, 3, 1])$ constraint holds.

**Typical**

$|\mathtt{VARIABLES}| > 2$
$\text{range}(\mathtt{VARIABLES.var}) > 1$

**Symmetries**

- Items of VARIABLES can be reversed.
- One and the same constant can be added to the var attribute of all items of VARIABLES.

**Arg. properties**

Functional dependency: VALUE determined by VARIABLES.

Figure 3.1313: Illustrating the SUM_WIDTH_PEAK constraint of the **Example** slot

**Automaton**     Figures 3.1314 and 3.1315 respectively depict the automaton associated with the constraint SUM_WIDTH_PEAK and its simplified form.

Figure 3.1314: Automaton for the SUM_WIDTH_PEAK constraint obtained by applying decoration Table 2.35 to the seed transducer of the PEAK pattern where default is 0

Figure 3.1315: Automaton for the SUM_WIDTH_PEAK constraint obtained by applying decoration Table 2.25 to the seed transducer of the PEAK pattern where default is 0; $R_i - R_{i-1} \geq 0$ and $-R_i + R_{i-1} + D_{i-1} + 1 \geq 0$ are linear invariants.

|   | $s$ | $r$ | $t$ |
|---|---|---|---|
| $s$ | $\overrightarrow{C} + \overleftarrow{C}$ | $\overrightarrow{C} + \overleftarrow{C}$ | $\overrightarrow{C} + \overleftarrow{C}$ |
| $r$ | $\overrightarrow{C} + \overleftarrow{C}$ | $\overrightarrow{D} + \overleftarrow{D} + 1$ | $\overleftarrow{C} + \overrightarrow{D} + \overleftarrow{D} + 1$ |
| $t$ | $\overrightarrow{C} + \overleftarrow{C}$ | $\overrightarrow{C} + \overrightarrow{D} + \overleftarrow{D} + 1$ | $\overrightarrow{C} + \overleftarrow{C}$ |

Table 3.192: Glue matrix for the SUM_WIDTH_PEAK constraint defined as the composition of the PEAK pattern , the feature WIDTH , and the aggregator sum ; cells of the glue matrix are coloured with the colour of the constituent to which they are related.

# 3.617 SUM_WIDTH_PLAIN

▶    ◁ ◀        DESCRIPTION              AUTOMATON

| | |
|---|---|
| **Origin** | Based on the PLAIN pattern. |
| **Constraint** | SUM_WIDTH_PLAIN(VALUE, VARIABLES) |
| **Arguments** | VALUE     : dvar |
| | VARIABLES : collection(var−dvar) |

**Restrictions**

$\text{sv} \leq 2 \vee \text{rv} \leq 1 \Rightarrow \text{VALUE} = 0$
$\text{VALUE} = 0 \vee \text{VALUE} \geq 1$
$\text{VALUE} \leq \max(0, \text{sv} - 2①)$
required(VARIABLES, var)
where
$\text{sv} = |\text{VARIABLES}|$
$\text{rv} = \text{range}(\text{VARIABLES.var})$

**Purpose**

VALUE is the sum of the width of occurrences of the PLAIN pattern in the time-series given by the VARIABLES collection. If the pattern does not occur, VALUE takes the default value 0.

An occurrence of the pattern PLAIN is the *maximal* subsequence which matches the regular expression $> =^* <$.

Assume that the occurrence of the pattern PLAIN starts at position $i$ and ends at position $j$. The feature WIDTH computes the value $j - i$.

**Example**

$(4, \langle 2, 3, 6, 5, 7, 6, 6, 4, 5, 5, 4, 3, 3, 6, 6, 3 \rangle)$

Figure 3.1316 provides an example where the SUM_WIDTH_PLAIN $(4, [2, 3, 6, 5, 7, 6, 6, 4, 5, 5, 4, 3, 3, 6, 6, 3])$ constraint holds.

**Typical**

$|\text{VARIABLES}| > 2$
$\text{range}(\text{VARIABLES.var}) > 1$

**Symmetries**

- Items of VARIABLES can be reversed.
- One and the same constant can be added to the var attribute of all items of VARIABLES.

**Arg. properties**

Functional dependency: VALUE determined by VARIABLES.

Figure 3.1316: Illustrating the SUM_WIDTH_PLAIN constraint of the **Example** slot

**Automaton**   Figures 3.1317 and 3.1318 respectively depict the automaton associated with the constraint SUM_WIDTH_PLAIN and its simplified form.

Figure 3.1317: Automaton for the SUM_WIDTH_PLAIN constraint obtained by applying decoration Table 2.35 to the seed transducer of the PLAIN pattern where default is 0

Figure 3.1318: Automaton for the SUM_WIDTH_PLAIN constraint obtained by applying decoration Table 2.28 to the seed transducer of the PLAIN pattern where default is 0; $R_i - R_{i-1} \geq 0$ and $-R_i + R_{i-1} + D_{i-1} + 1 \geq 0$ are linear invariants.

|   | $s$ | $r$ | $t$ |
|---|---|---|---|
| $s$ | $\overrightarrow{C} + \overleftarrow{C}$ | $\overrightarrow{C} + \overleftarrow{C}$ | $\overrightarrow{C} + \overleftarrow{C}$ |
| $r$ | $\overrightarrow{C} + \overleftarrow{C}$ | $\overrightarrow{D} + \overleftarrow{D} + 1$ | $\overrightarrow{D} + \overleftarrow{D} + 1$ |
| $t$ | $\overrightarrow{C} + \overleftarrow{C}$ | $\overrightarrow{D} + \overleftarrow{D} + 1$ | $\overrightarrow{D} + \overleftarrow{D} + 1$ |

Table 3.193: Glue matrix for the SUM_WIDTH_PLAIN constraint defined as the composition of the PLAIN pattern , the feature WIDTH , and the aggregator sum ; cells of the glue matrix are coloured with the colour of the constituent to which they are related.

## 3.618 SUM_WIDTH_PLATEAU

    DESCRIPTION          AUTOMATON

| | |
|---|---|
| **Origin** | Based on the PLATEAU pattern. |
| **Constraint** | SUM_WIDTH_PLATEAU(VALUE, VARIABLES) |
| **Arguments** | VALUE      : dvar<br>VARIABLES  : collection(var−dvar) |

**Restrictions**

$\mathtt{sv} \leq 2 \vee \mathtt{rv} \leq 1 \Rightarrow \mathtt{VALUE} = 0$
$\mathtt{VALUE} = 0 \vee \mathtt{VALUE} \geq 1$
$\mathtt{VALUE} \leq \max(0, \mathtt{sv} - 2①)$
required(VARIABLES, var)
where
$\mathtt{sv} = |\mathtt{VARIABLES}|$
$\mathtt{rv} = \text{range}(\mathtt{VARIABLES.var})$

**Purpose**

VALUE is the sum of the width of occurrences of the PLATEAU pattern in the time-series given by the VARIABLES collection. If the pattern does not occur, VALUE takes the default value 0.

An occurrence of the pattern PLATEAU is the *maximal* subsequence which matches the regular expression $< =^* >$.

Assume that the occurrence of the pattern PLATEAU starts at position $i$ and ends at position $j$. The feature WIDTH computes the value $j - i$.

**Example**

$(7, \langle 1, 3, 3, 5, 5, 5, 5, 2, 4, 4, 4, 3, 3, 1, 5, 5 \rangle)$

Figure 3.1319 provides an example where the SUM_WIDTH_PLATEAU $(7, [1, 3, 3, 5, 5, 5, 5, 2, 4, 4, 4, 3, 3, 1, 5, 5])$ constraint holds.

**Typical**

$|\mathtt{VARIABLES}| > 2$
$\text{range}(\mathtt{VARIABLES.var}) > 1$

**Symmetries**

- Items of VARIABLES can be reversed.
- One and the same constant can be added to the var attribute of all items of VARIABLES.

**Arg. properties**

Functional dependency: VALUE determined by VARIABLES.

Figure 3.1319: Illustrating the SUM_WIDTH_PLATEAU constraint of the **Example** slot

 Figures 3.1320 and 3.1321 respectively depict the automaton associated with the constraint SUM_WIDTH_PLATEAU and its simplified form.

Figure 3.1320: Automaton for the SUM_WIDTH_PLATEAU constraint obtained by applying decoration Table 2.35 to the seed transducer of the PLATEAU pattern where default is 0

Figure 3.1321: Automaton for the SUM_WIDTH_PLATEAU constraint obtained by applying decoration Table 2.28 to the seed transducer of the PLATEAU pattern where default is 0; $R_i - R_{i-1} \geq 0$ and $-R_i + R_{i-1} + D_{i-1} + 1 \geq 0$ are linear invariants.

| | $s$ | $r$ | $t$ |
|---|---|---|---|
| $s$ | $\overrightarrow{C} + \overleftarrow{C}$ | $\overrightarrow{C} + \overleftarrow{C}$ | $\overrightarrow{C} + \overleftarrow{C}$ |
| $r$ | $\overrightarrow{C} + \overleftarrow{C}$ | $\overrightarrow{D} + \overleftarrow{D} + 1$ | $\overrightarrow{D} + \overleftarrow{D} + 1$ |
| $t$ | $\overrightarrow{C} + \overleftarrow{C}$ | $\overrightarrow{D} + \overleftarrow{D} + 1$ | $\overrightarrow{D} + \overleftarrow{D} + 1$ |

Table 3.194: Glue matrix for the SUM_WIDTH_PLATEAU constraint defined as the composition of the PLATEAU pattern , the feature WIDTH , and the aggregator sum ; cells of the glue matrix are coloured with the colour of the constituent to which they are related.

## 3.619　SUM_WIDTH_PROPER_PLAIN

| | |
|---|---|
| **Origin** | Based on the PROPER_PLAIN pattern. |
| **Constraint** | SUM_WIDTH_PROPER_PLAIN(VALUE, VARIABLES) |
| **Arguments** | VALUE : dvar<br>VARIABLES : collection(var−dvar) |

**Restrictions**

$\mathtt{sv} \leq 3 \lor \mathtt{rv} \leq 1 \Rightarrow \mathtt{VALUE} = 0$
$\mathtt{VALUE} = 0 \lor \mathtt{VALUE} \geq 2$
$\mathtt{VALUE} \leq \max(0, \mathtt{sv} - 2①)$
required(VARIABLES, var)
where
$\mathtt{sv} = |\mathtt{VARIABLES}|$
$\mathtt{rv} = \mathtt{range}(\mathtt{VARIABLES.var})$

**Purpose**

VALUE is the sum of the width of occurrences of the PROPER_PLAIN pattern in the time-series given by the VARIABLES collection. If the pattern does not occur, VALUE takes the default value 0.

An occurrence of the pattern PROPER_PLAIN is the *maximal* subsequence which matches the regular expression $> =^+ <$.

Assume that the occurrence of the pattern PROPER_PLAIN starts at position $i$ and ends at position $j$. The feature WIDTH computes the value $j - i$.

**Example**

$(7, \langle 2, 7, 5, 5, 6, 3, 7, 4, 4, 5, 6, 5, 3, 3, 3, 5 \rangle)$

Figure 3.1322 provides an example where the SUM_WIDTH_PROPER_PLAIN $(7, [2, 7, 5, 5, 6, 3, 7, 4, 4, 5, 6, 5, 3, 3, 3, 5])$ constraint holds.

**Typical**

$|\mathtt{VARIABLES}| > 3$
$\mathtt{range}(\mathtt{VARIABLES.var}) > 1$

**Symmetries**

- Items of VARIABLES can be reversed.
- One and the same constant can be added to the var attribute of all items of VARIABLES.

**Arg. properties**

Functional dependency: VALUE determined by VARIABLES.

Figure 3.1322: Illustrating the SUM_WIDTH_PROPER_PLAIN constraint of the **Example** slot

 Figures 3.1323 and 3.1324 respectively depict the automaton associated with the constraint SUM_WIDTH_PROPER_PLAIN and its simplified form.

Figure 3.1323: Automaton for the SUM_WIDTH_PROPER_PLAIN constraint obtained by applying decoration Table 2.35 to the seed transducer of the PROPER_PLAIN pattern where default is 0

Figure 3.1324: Automaton for the SUM_WIDTH_PROPER_PLAIN constraint obtained by applying decoration Table 2.28 to the seed transducer of the PROPER_PLAIN pattern where default is 0; $R_i - R_{i-1} \geq 0$ and $-R_i + R_{i-1} + D_{i-1} + 1 \geq 0$ are linear invariants.

|   | $s$ | $r$ | $t$ |
|---|---|---|---|
| $s$ | $\overrightarrow{C} + \overleftarrow{C}$ | $\overrightarrow{C} + \overleftarrow{C}$ | $\overrightarrow{C} + \overleftarrow{C}$ |
| $r$ | $\overrightarrow{C} + \overleftarrow{C}$ | $\overrightarrow{C} + \overleftarrow{C}$ | $\overrightarrow{D} + \overleftarrow{D} + 1$ |
| $t$ | $\overrightarrow{C} + \overleftarrow{C}$ | $\overrightarrow{D} + \overleftarrow{D} + 1$ | $\overrightarrow{D} + \overleftarrow{D} + 1$ |

Table 3.195: Glue matrix for the SUM_WIDTH_PROPER_PLAIN constraint defined as the composition of the PROPER_PLAIN pattern , the feature WIDTH , and the aggregator sum ; cells of the glue matrix are coloured with the colour of the constituent to which they are related.

## 3.620 SUM_WIDTH_PROPER_PLATEAU

     DESCRIPTION          AUTOMATON

| | |
|---|---|
| **Origin** | Based on the PROPER_PLATEAU pattern. |
| **Constraint** | SUM_WIDTH_PROPER_PLATEAU(VALUE, VARIABLES) |
| **Arguments** | VALUE      : dvar<br>VARIABLES  : collection(var−dvar) |

**Restrictions**

$$\mathtt{sv} \leq 3 \vee \mathtt{rv} \leq 1 \Rightarrow \mathtt{VALUE} = 0$$
$$\mathtt{VALUE} = 0 \vee \mathtt{VALUE} \geq 2$$
$$\mathtt{VALUE} \leq \max(0, \mathtt{sv} - 2①)$$
required(VARIABLES, var)
where
$$\mathtt{sv} = |\mathtt{VARIABLES}|$$
$$\mathtt{rv} = \mathtt{range}(\mathtt{VARIABLES.var})$$

**Purpose**

VALUE is the sum of the width of occurrences of the PROPER_PLATEAU pattern in the time-series given by the VARIABLES collection. If the pattern does not occur, VALUE takes the default value 0.

An occurrence of the pattern PROPER_PLATEAU is the *maximal* subsequence which matches the regular expression $<=^+>$.

Assume that the occurrence of the pattern PROPER_PLATEAU starts at position $i$ and ends at position $j$. The feature WIDTH computes the value $j - i$.

**Example**

$$(7, \langle 7, 1, 3, 3, 2, 5, 1, 4, 4, 3, 2, 3, 5, 5, 5, 3 \rangle)$$

Figure 3.1325 provides an example where the SUM_WIDTH_PROPER_PLATEAU $(7, [7, 1, 3, 3, 2, 5, 1, 4, 4, 3, 2, 3, 5, 5, 5, 3])$ constraint holds.

**Typical**

$$|\mathtt{VARIABLES}| > 3$$
$$\mathtt{range}(\mathtt{VARIABLES.var}) > 1$$

**Symmetries**

- Items of VARIABLES can be reversed.
- One and the same constant can be added to the var attribute of all items of VARIABLES.

**Arg. properties**

Functional dependency: VALUE determined by VARIABLES.

Figure 3.1325: Illustrating the SUM_WIDTH_PROPER_PLATEAU constraint of the **Example** slot

**Automaton**     Figures 3.1326 and 3.1327 respectively depict the automaton associated with the constraint
                  SUM_WIDTH_PROPER_PLATEAU and its simplified form.

Figure 3.1326: Automaton for the SUM_WIDTH_PROPER_PLATEAU constraint obtained
by applying decoration Table 2.35 to the seed transducer of the PROPER_PLATEAU
pattern where default is 0

Figure 3.1327: Automaton for the SUM_WIDTH_PROPER_PLATEAU constraint obtained
by applying decoration Table 2.28 to the seed transducer of the PROPER_PLATEAU
pattern where default is 0; $R_i - R_{i-1} \geq 0$ and $-R_i + R_{i-1} + D_{i-1} + 1 \geq 0$ are linear
invariants.

|   | $s$ | $r$ | $t$ |
|---|---|---|---|
| $s$ | $\overrightarrow{C} + \overleftarrow{C}$ | $\overrightarrow{C} + \overleftarrow{C}$ | $\overrightarrow{C} + \overleftarrow{C}$ |
| $r$ | $\overrightarrow{C} + \overleftarrow{C}$ | $\overrightarrow{C} + \overleftarrow{C}$ | $\overrightarrow{D} + \overleftarrow{D} + 1$ |
| $t$ | $\overrightarrow{C} + \overleftarrow{C}$ | $\overrightarrow{D} + \overleftarrow{D} + 1$ | $\overrightarrow{D} + \overleftarrow{D} + 1$ |

Table 3.196: Glue matrix for the SUM_WIDTH_PROPER_PLATEAU constraint defined as the composition of the PROPER_PLATEAU pattern , the feature WIDTH , and the aggregator sum ; cells of the glue matrix are coloured with the colour of the constituent to which they are related.

# 3.621   SUM_WIDTH_STEADY_SEQUENCE

▶   ◁ ◀       DESCRIPTION            AUTOMATON

| | |
|---|---|
| **Origin** | Based on the STEADY_SEQUENCE pattern. |
| **Constraint** | SUM_WIDTH_STEADY_SEQUENCE(VALUE, VARIABLES) |
| **Arguments** | VALUE      : dvar |
| | VARIABLES  : collection(var−dvar) |

**Restrictions**

$\text{sv} \leq 1 \Rightarrow \text{VALUE} = 0$
$\text{rv} = 1 \Rightarrow \text{VALUE} \geq \text{sv}$
$\text{rv} \geq 2 \Rightarrow \text{VALUE} = 0 \lor \text{VALUE} \geq 2$
$\text{VALUE} \leq \text{sv}①$
required(VARIABLES, var)
where
$\text{sv} = |\text{VARIABLES}|$
$\text{rv} = \text{range}(\text{VARIABLES.var})$

**Purpose**

VALUE is the sum of the width of occurrences of the STEADY_SEQUENCE pattern in the time-series given by the VARIABLES collection. If the pattern does not occur, VALUE takes the default value 0.

An occurrence of the pattern STEADY_SEQUENCE is the *maximal* subsequence which matches the regular expression $=^+$.

Assume that the occurrence of the pattern STEADY_SEQUENCE starts at position $i$ and ends at position $j$. The feature WIDTH computes the value $j - i + 2$.

**Example**

$(11, \langle 3, 1, 1, 4, 5, 5, 5, 6, 2, 2, 4, 4, 3, 2, 1, 1 \rangle)$

Figure 3.1328 provides an example where the SUM_WIDTH_STEADY_SEQUENCE $(11, [3, 1, 1, 4, 5, 5, 5, 6, 2, 2, 4, 4, 3, 2, 1, 1])$ constraint holds.

| | |
|---|---|
| **Typical** | $|\text{VARIABLES}| > 1$ |

**Symmetries**

- Items of VARIABLES can be reversed.
- One and the same constant can be added to the var attribute of all items of VARIABLES.

| | |
|---|---|
| **Arg. properties** | Functional dependency: VALUE determined by VARIABLES. |

Figure 3.1328: Illustrating the SUM_WIDTH_STEADY_SEQUENCE constraint of the **Example** slot

**Automaton**      Figures 3.1329 and 3.1330 respectively depict the automaton associated with the constraint SUM_WIDTH_STEADY_SEQUENCE and its simplified form.

Figure 3.1329: Automaton for the SUM_WIDTH_STEADY_SEQUENCE constraint obtained by applying decoration Table 2.35 to the seed transducer of the STEADY_SEQUENCE pattern where default is 0

Figure 3.1330: Automaton for the SUM_WIDTH_STEADY_SEQUENCE constraint obtained by applying decoration Table 2.38 to the seed transducer of the STEADY_SEQUENCE pattern where default is 0; $R_i - R_{i-1} \geq 0$ and $-R_i + R_{i-1} + 2 \geq 0$ are linear invariants.

|     | $s$ | $r$ |
| --- | --- | --- |
| $s$ | $\overrightarrow{C} + \overleftarrow{C}$ | $\overrightarrow{C} + \overleftarrow{C}$ |
| $r$ | $\overrightarrow{C} + \overleftarrow{C}$ | $\overrightarrow{C} + \overleftarrow{C} + \overrightarrow{D} + \overleftarrow{D} + 1$ |

Table 3.197: Glue matrix for the SUM_WIDTH_STEADY_SEQUENCE constraint defined as the composition of the STEADY_SEQUENCE pattern , the feature WIDTH , and the aggregator sum ; cells of the glue matrix are coloured with the colour of the constituent to which they are related.

# 3.622   SUM_WIDTH_STRICTLY_DECREASING_SEQUENCE

► ◁ ◄         DESCRIPTION          AUTOMATON

**Origin**         Based on the STRICTLY_DECREASING_SEQUENCE pattern.

**Constraint**     SUM_WIDTH_STRICTLY_DECREASING_SEQUENCE(VALUE, VARIABLES)

**Arguments**
```
VALUE      :  dvar
VARIABLES  :  collection(var−dvar)
```

**Restrictions**   $\mathtt{sv} \leq 1 \vee \mathtt{rv} \leq 1 \Rightarrow \mathtt{VALUE} = 0$
$\mathtt{VALUE} = 0 \vee \mathtt{VALUE} \geq 2$
$\mathtt{rv} = 2 \Rightarrow \mathtt{VALUE} \leq \mathtt{sv} - \mathtt{sv} \bmod 2①$
$\mathtt{rv} \geq 3 \Rightarrow \mathtt{VALUE} \leq \mathtt{sv}②$
required(VARIABLES, var)
where
$\mathtt{sv} = |\mathtt{VARIABLES}|$
$\mathtt{rv} =$ range(VARIABLES.var)

**Purpose**
VALUE is the sum of the width of occurrences of the STRICTLY_DECREASING_SEQUENCE pattern in the time-series given by the VARIABLES collection. If the pattern does not occur, VALUE takes the default value 0.
An occurrence of the pattern STRICTLY_DECREASING_SEQUENCE is the *maximal* subsequence which matches the regular expression $>^+$.
Assume that the occurrence of the pattern STRICTLY_DECREASING_SEQUENCE starts at position $i$ and ends at position $j$. The feature WIDTH computes the value $j - i + 2$.

**Example**       $(8, \langle 4, 4, 6, 4, 1, 1, 3, 4, 4, 6, 6, 5, 2, 2, 4, 3 \rangle)$

Figure 3.1331 provides an example where the SUM_WIDTH_STRICTLY_DECREASING_SEQUENCE $(8, [4, 4, 6, 4, 1, 1, 3, 4, 4, 6, 6, 5, 2, 2, 4, 3])$ constraint holds.

**Typical**       $|\mathtt{VARIABLES}| > 1$
range(VARIABLES.var) $> 1$

**Symmetry**      One and the same constant can be added to the var attribute of all items of VARIABLES.

**Arg. properties**   Functional dependency: VALUE determined by VARIABLES.

Figure 3.1331: Illustrating the SUM_WIDTH_STRICTLY_DECREASING_SEQUENCE constraint of the **Example** slot

 Figures 3.1332 and 3.1333 respectively depict the automaton associated with the constraint SUM_WIDTH_STRICTLY_DECREASING_SEQUENCE and its simplified form.

Figure 3.1332: Automaton for the SUM_WIDTH_STRICTLY_DECREASING_SEQUENCE constraint obtained by applying decoration Table 2.35 to the seed transducer of the STRICTLY_DECREASING_SEQUENCE pattern where default is 0

Figure 3.1333: Automaton for the SUM_WIDTH_STRICTLY_DECREASING_SEQUENCE constraint obtained by applying decoration Table 2.38 to the seed transducer of the STRICTLY_DECREASING_SEQUENCE pattern where default is 0; $R_i - R_{i-1} \geq 0$ and $-R_i + R_{i-1} + 2 \geq 0$ are linear invariants.

|   | s | r |
|---|---|---|
| s | $\overrightarrow{C} + \overleftarrow{C}$ | $\overrightarrow{C} + \overleftarrow{C}$ |
| r | $\overrightarrow{C} + \overleftarrow{C}$ | $\overrightarrow{C} + \overleftarrow{C} + \overrightarrow{D} + \overleftarrow{D} + 1$ |

Table 3.198: Glue matrix for the SUM_WIDTH_STRICTLY_DECREASING_SEQUENCE constraint defined as the composition of the STRICTLY_DECREASING_SEQUENCE pattern, the feature WIDTH, and the aggregator sum; cells of the glue matrix are coloured with the colour of the constituent to which they are related.

# 3.623   SUM_WIDTH_STRICTLY_INCREASING_SEQUENCE

▶    ◁  ◀        **DESCRIPTION**            **AUTOMATON**

**Origin**          Based on the STRICTLY_INCREASING_SEQUENCE pattern.

**Constraint**      SUM_WIDTH_STRICTLY_INCREASING_SEQUENCE(VALUE, VARIABLES)

**Arguments**       VALUE      :  dvar
                    VARIABLES  :  collection(var−dvar)

**Restrictions**    $\mathtt{sv} \leq 1 \vee \mathtt{rv} \leq 1 \Rightarrow \mathtt{VALUE} = 0$
                    $\mathtt{VALUE} = 0 \vee \mathtt{VALUE} \geq 2$
                    $\mathtt{rv} = 2 \Rightarrow \mathtt{VALUE} \leq \mathtt{sv} - \mathtt{sv} \bmod 2 ①$
                    $\mathtt{rv} \geq 3 \Rightarrow \mathtt{VALUE} \leq \mathtt{sv} ②$
                    required(VARIABLES, var)
                    where
                    $\mathtt{sv} = |\mathtt{VARIABLES}|$
                    $\mathtt{rv} =$ range(VARIABLES.var)

**Purpose**         VALUE is the sum of the width of occurrences of the STRICTLY_INCREASING_SEQUENCE
                    pattern in the time-series given by the VARIABLES collection. If the pattern does not
                    occur, VALUE takes the default value 0.
                    An occurrence of the pattern STRICTLY_INCREASING_SEQUENCE is the *maximal* sub-
                    sequence which matches the regular expression $<^+$.
                    Assume that the occurrence of the pattern STRICTLY_INCREASING_SEQUENCE starts at
                    position $i$ and ends at position $j$. The feature WIDTH computes the value $j - i + 2$.

**Example**         $(10, \langle 4, 3, 5, 5, 2, 1, 1, 2, 3, 4, 6, 6, 3, 1, 2, 3 \rangle)$

                    Figure 3.1334 provides an example where the SUM_WIDTH_STRICTLY_INCREASING_SEQUENCE
                    $(10, [4, 3, 5, 5, 2, 1, 1, 2, 3, 4, 6, 6, 3, 1, 2, 3])$ constraint holds.

**Typical**         $|\mathtt{VARIABLES}| > 1$
                    range(VARIABLES.var) $> 1$

**Symmetry**        One and the same constant can be added to the var attribute of all items of VARIABLES.

**Arg. properties** Functional dependency: VALUE determined by VARIABLES.

Figure 3.1334: Illustrating the SUM_WIDTH_STRICTLY_INCREASING_SEQUENCE constraint of the **Example** slot

**Automaton**       Figures 3.1335 and 3.1336 respectively depict the automaton associated with the constraint
                    SUM_WIDTH_STRICTLY_INCREASING_SEQUENCE and its simplified form.

Figure 3.1335: Automaton for the SUM_WIDTH_STRICTLY_INCREASING_SEQUENCE
constraint obtained by applying decoration Table 2.35 to the seed transducer of the
STRICTLY_INCREASING_SEQUENCE pattern where default is 0

Figure 3.1336: Automaton for the SUM_WIDTH_STRICTLY_INCREASING_SEQUENCE
constraint obtained by applying decoration Table 2.38 to the seed transducer of the
STRICTLY_INCREASING_SEQUENCE pattern where default is 0; $R_i - R_{i-1} \geq 0$ and
$-R_i + R_{i-1} + 2 \geq 0$ are linear invariants.

|     | $s$ | $r$ |
| --- | --- | --- |
| $s$ | $\overrightarrow{C} + \overleftarrow{C}$ | $\overrightarrow{C} + \overleftarrow{C}$ |
| $r$ | $\overrightarrow{C} + \overleftarrow{C}$ | $\overrightarrow{C} + \overleftarrow{C} + \overrightarrow{D} + \overleftarrow{D} + 1$ |

Table 3.199: Glue matrix for the SUM_WIDTH_STRICTLY_INCREASING_SEQUENCE constraint defined as the composition of the STRICTLY_INCREASING_SEQUENCE pattern , the feature WIDTH , and the aggregator sum ; cells of the glue matrix are coloured with the colour of the constituent to which they are related.

# 3.624  SUM_WIDTH_SUMMIT

$(< | < (= | <)^* <)(>| > (= | >)^* >)$

| | |
|---|---|
| **Origin** | Based on the SUMMIT pattern. |
| **Constraint** | SUM_WIDTH_SUMMIT(VALUE, VARIABLES) |
| **Arguments** | VALUE      : dvar |
| | VARIABLES  : collection(var−dvar) |

**Restrictions**

$sv \leq 2 \vee rv \leq 1 \Rightarrow \text{VALUE} = 0$
$\text{VALUE} = 0 \vee \text{VALUE} \geq 1$
$rv = 2 \Rightarrow \text{VALUE} \leq np①$
$rv \geq 3 \Rightarrow \text{VALUE} \leq \max(0, sv - 2②)$
required(VARIABLES, var)
where
$sv = |\text{VARIABLES}|$
$rv = \text{range}(\text{VARIABLES.var})$
$np = \max(0, \lfloor (sv - 1)/2 \rfloor)$

**Purpose**

VALUE is the sum of the width of occurrences of the SUMMIT pattern in the time-series given by the VARIABLES collection. If the pattern does not occur, VALUE takes the default value 0.

An occurrence of the pattern SUMMIT is the *maximal* subsequence which matches the regular expression $(< | < (= | <)^* <)(>| > (= | >)^* >)$.

Assume that the occurrence of the pattern SUMMIT starts at position $i$ and ends at position $j$. The feature WIDTH computes the value $j - i$.

**Example**

$(6, \langle 7, 1, 5, 4, 4, 3, 3, 4, 6, 6, 2, 3, 4, 2, 3, 1 \rangle)$

Figure 3.1337 provides an example where the SUM_WIDTH_SUMMIT $(6, [7, 1, 5, 4, 4, 3, 3, 4, 6, 6, 2, 3, 4, 2, 3, 1])$ constraint holds.

**Typical**

$|\text{VARIABLES}| > 2$
range(VARIABLES.var) > 1

**Symmetries**

- Items of VARIABLES can be reversed.
- One and the same constant can be added to the var attribute of all items of VARIABLES.

**Arg. properties**

Functional dependency: VALUE determined by VARIABLES.

Figure 3.1337: Illustrating the SUM_WIDTH_SUMMIT constraint of the **Example** slot

**Automaton**    Figures 3.1338 and 3.1339 respectively depict the automaton associated with the constraint
SUM_WIDTH_SUMMIT and its simplified form.

Figure 3.1338: Automaton for the SUM_WIDTH_SUMMIT constraint obtained by apply-
ing decoration Table 2.35 to the seed transducer of the SUMMIT pattern where default
is 0 (transition $u \to r$ has the same accumulator update as transition $r \to u$)

Figure 3.1339: Automaton for the SUM_WIDTH_SUMMIT constraint obtained by applying decoration Table 2.25 to the seed transducer of the SUMMIT pattern where default is 0 (transition $u \to r$ has the same accumulator update as transition $r \to u$); $R_i - R_{i-1} \geq 0$ and $-R_i + R_{i-1} + D_{i-1} + 1 \geq 0$ are linear invariants.

| | $s$ | $r$ | $t$ | $u$ |
|---|---|---|---|---|
| $s$ | $\overrightarrow{C} + \overleftarrow{C}$ | $\overrightarrow{C} + \overleftarrow{C}$ | $\overrightarrow{C} + \overleftarrow{C}$ | $\overrightarrow{C} + \overleftarrow{C}$ |
| $r$ | $\overrightarrow{C} + \overleftarrow{C}$ | $\overrightarrow{D} + \overleftarrow{D} + 1$ | $\overleftarrow{C} + \overrightarrow{D} + \overleftarrow{D} + 1$ | $\overrightarrow{C} + \overleftarrow{C}$ |
| $t$ | $\overrightarrow{C} + \overleftarrow{C}$ | $\overrightarrow{C} + \overrightarrow{D} + \overleftarrow{D} + 1$ | $\overrightarrow{C} + \overleftarrow{C}$ | $\overrightarrow{C} + \overrightarrow{D} + \overleftarrow{D} + 1$ |
| $u$ | $\overrightarrow{C} + \overleftarrow{C}$ | $\overrightarrow{C} + \overleftarrow{C}$ | $\overleftarrow{C} + \overrightarrow{D} + \overleftarrow{D} + 1$ | $\overrightarrow{C} + \overleftarrow{C}$ |

Table 3.200: Glue matrix for the SUM_WIDTH_SUMMIT constraint defined as the composition of the SUMMIT pattern , the feature WIDTH , and the aggregator sum ; cells of the glue matrix are coloured with the colour of the constituent to which they are related.

## 3.625  SUM_WIDTH_VALLEY

▶    ◁ ◀        DESCRIPTION            AUTOMATON

| | |
|---|---|
| **Origin** | Based on the VALLEY pattern. |
| **Constraint** | SUM_WIDTH_VALLEY(VALUE, VARIABLES) |
| **Arguments** | VALUE      :  dvar<br>VARIABLES  :  collection(var−dvar) |

**Restrictions**

$$sv \leq 2 \vee rv \leq 1 \Rightarrow \text{VALUE} = 0$$
$$\text{VALUE} = 0 \vee \text{VALUE} \geq 1$$
$$\text{VALUE} \leq \max(0, sv - 2①)$$
required(VARIABLES, var)
where
$$sv = |\text{VARIABLES}|$$
$$rv = \text{range}(\text{VARIABLES.var})$$

**Purpose**

VALUE is the sum of the width of occurrences of the VALLEY pattern in the time-series given by the VARIABLES collection. If the pattern does not occur, VALUE takes the default value 0.

An occurrence of the pattern VALLEY is the *maximal* subsequence which matches the regular expression $> (= | >)^* (< | =)^* <$.

Assume that the occurrence of the pattern VALLEY starts at position $i$ and ends at position $j$. The feature WIDTH computes the value $j - i$.

**Example**

$$(9, \langle 1, 3, 7, 4, 3, 6, 6, 5, 3, 3, 2, 6, 5, 5, 5, 7 \rangle)$$

Figure 3.1340 provides an example where the SUM_WIDTH_VALLEY $(9, [1, 3, 7, 4, 3, 6, 6, 5, 3, 3, 2, 6, 5, 5, 5, 7])$ constraint holds.

**Typical**

$$|\text{VARIABLES}| > 2$$
$$\text{range}(\text{VARIABLES.var}) > 1$$

**Symmetries**

- Items of VARIABLES can be reversed.
- One and the same constant can be added to the var attribute of all items of VARIABLES.

**Arg. properties**

Functional dependency: VALUE determined by VARIABLES.

Figure 3.1340: Illustrating the SUM_WIDTH_VALLEY constraint of the **Example** slot

**Automaton** Figures 3.1341 and 3.1342 respectively depict the automaton associated with the constraint SUM_WIDTH_VALLEY and its simplified form.

Figure 3.1341: Automaton for the SUM_WIDTH_VALLEY constraint obtained by applying decoration Table 2.35 to the seed transducer of the VALLEY pattern where default is 0

Figure 3.1342: Automaton for the SUM_WIDTH_VALLEY constraint obtained by applying decoration Table 2.25 to the seed transducer of the VALLEY pattern where default is 0; $R_i - R_{i-1} \geq 0$ and $-R_i + R_{i-1} + D_{i-1} + 1 \geq 0$ are linear invariants.

|   | $s$ | $r$ | $t$ |
|---|---|---|---|
| $s$ | $\overrightarrow{C} + \overleftarrow{C}$ | $\overrightarrow{C} + \overleftarrow{C}$ | $\overrightarrow{C} + \overleftarrow{C}$ |
| $r$ | $\overrightarrow{C} + \overleftarrow{C}$ | $\overrightarrow{D} + \overleftarrow{D} + 1$ | $\overleftarrow{C} + \overrightarrow{D} + \overleftarrow{D} + 1$ |
| $t$ | $\overrightarrow{C} + \overleftarrow{C}$ | $\overrightarrow{C} + \overrightarrow{D} + \overleftarrow{D} + 1$ | $\overrightarrow{C} + \overleftarrow{C}$ |

Table 3.201: Glue matrix for the SUM_WIDTH_VALLEY constraint defined as the composition of the VALLEY pattern , the feature WIDTH , and the aggregator sum ; cells of the glue matrix are coloured with the colour of the constituent to which they are related.

# 3.626   SUM_WIDTH_ZIGZAG

◁ ◀     **DESCRIPTION**      **AUTOMATON**

| | |
|---|---|
| **Origin** | Based on the ZIGZAG pattern. |
| **Constraint** | SUM_WIDTH_ZIGZAG(VALUE, VARIABLES) |
| **Arguments** | VALUE      : dvar <br> VARIABLES : collection(var−dvar) |

**Restrictions**

$\mathtt{sv} \leq 3 \vee \mathtt{rv} \leq 1 \Rightarrow \mathtt{VALUE} = 0$
$\mathtt{VALUE} = 0 \vee \mathtt{VALUE} \geq 2$
$\mathtt{VALUE} \leq \max(0, \mathtt{sv} - 2①)$
required(VARIABLES, var)
where
$\mathtt{sv} = |\mathtt{VARIABLES}|$
$\mathtt{rv} = \mathtt{range}(\mathtt{VARIABLES.var})$

**Purpose**

VALUE is the sum of the width of occurrences of the ZIGZAG pattern in the time-series given by the VARIABLES collection. If the pattern does not occur, VALUE takes the default value 0.

An occurrence of the pattern ZIGZAG is the *maximal* subsequence which matches the regular expression $(<>)^+(< | <>) | (><)^+(> | ><)$.

Assume that the occurrence of the pattern ZIGZAG starts at position $i$ and ends at position $j$. The feature WIDTH computes the value $j - i$.

**Example**

$(11, \langle 4, 1, 3, 1, 4, 6, 1, 5, 5, 2, 7, 2, 3, 1, 6, 1 \rangle)$

Figure 3.1343 provides an example where the SUM_WIDTH_ZIGZAG $(11, [4, 1, 3, 1, 4, 6, 1, 5, 5, 2, 7, 2, 3, 1, 6, 1])$ constraint holds.

**Typical**

$|\mathtt{VARIABLES}| > 3$
$\mathtt{range}(\mathtt{VARIABLES.var}) > 1$

**Symmetries**

- Items of VARIABLES can be reversed.
- One and the same constant can be added to the var attribute of all items of VARIABLES.

**Arg. properties**

Functional dependency: VALUE determined by VARIABLES.

Figure 3.1343: Illustrating the SUM_WIDTH_ZIGZAG constraint of the **Example** slot

**Automaton**        Figures 3.1344 and 3.1345 respectively depict the automaton associated with the constraint SUM_WIDTH_ZIGZAG and its simplified form.

Figure 3.1344: Automaton for the SUM_WIDTH_ZIGZAG constraint obtained by applying decoration Table 2.35 to the seed transducer of the ZIGZAG pattern where default is 0; (1) missing transitions from $a, b, c, d, e, f$ to $s$ are labelled by $=$; (2) on transitions from $b, c, e, f$ to $s$ the accumulator $D$ is reset to its initial value; (3) on transitions from $c, f$ to $s$ the accumulator $R$ is updated wrt $C$ and the accumulator $C$ is reset to its initial value

Figure 3.1345: Automaton for the SUM_WIDTH_ZIGZAG constraint obtained by applying decoration Table 2.39 to the seed transducer of the ZIGZAG pattern where default is 0; missing transitions from $a, b, c, d, e, f$ to $s$ are labelled by $=$; $R_i - R_{i-1} \geq 0$ and $-R_i + R_{i-1} + 2 \geq 0$ are linear invariants.

| | s | a | b | c | d | e | f |
|---|---|---|---|---|---|---|---|
| **s** | $\overleftarrow{C}+\overleftarrow{C}$ | $\overleftarrow{C}+\overleftarrow{C}$ | $\overleftarrow{C}+\overleftarrow{C}$ | $\overleftarrow{C}+\overleftarrow{C}$ | $\overleftarrow{C}+\overleftarrow{C}$ | $\overleftarrow{C}+\overleftarrow{C}$ | $\overleftarrow{C}+\overleftarrow{C}$ |
| **a** | $\overrightarrow{C}+\overleftarrow{C}$ | $\overrightarrow{C}+\overleftarrow{C}$ | $\overrightarrow{C}+\overleftarrow{C}$ | $\overrightarrow{C}+\overrightarrow{D}+\overleftarrow{D}+1$ | $\overrightarrow{D}+\overleftarrow{D}+1$ | $\overrightarrow{C}+\overleftarrow{C}$ | $\overrightarrow{C}+\overleftarrow{C}$ |
| **b** | $\overrightarrow{C}+\overleftarrow{C}$ | $\overrightarrow{C}+\overleftarrow{C}$ | $\overrightarrow{D}+\overleftarrow{D}+1$ | $\overrightarrow{C}+\overleftarrow{C}$ | $\overrightarrow{D}+\overleftarrow{D}+1$ | $\overrightarrow{C}+\overleftarrow{C}$ | $\overrightarrow{C}+\overrightarrow{D}+\overleftarrow{D}+1$ |
| **c** | $\overrightarrow{C}+\overleftarrow{C}$ | $\overrightarrow{C}+\overrightarrow{D}+\overleftarrow{D}+1$ | $\overrightarrow{C}+\overleftarrow{C}$ | $\overrightarrow{C}+\overrightarrow{D}+\overleftarrow{D}+1$ | $\overleftarrow{C}+\overrightarrow{C}+\overrightarrow{D}+\overleftarrow{D}+1$ | $\overrightarrow{C}+\overrightarrow{D}+\overleftarrow{D}+1$ | $\overrightarrow{C}+\overleftarrow{C}$ |
| **d** | $\overrightarrow{C}+\overleftarrow{C}$ | $\overrightarrow{D}+\overleftarrow{D}+1$ | $\overrightarrow{C}+\overleftarrow{C}$ | $\overrightarrow{C}+\overleftarrow{C}$ | $\overrightarrow{C}+\overleftarrow{C}$ | $\overrightarrow{C}+\overleftarrow{C}$ | $\overrightarrow{C}+\overrightarrow{D}+\overleftarrow{D}+1$ |
| **e** | $\overrightarrow{C}+\overleftarrow{C}$ | $\overrightarrow{C}+\overleftarrow{C}$ | $\overrightarrow{D}+\overleftarrow{D}+1$ | $\overrightarrow{C}+\overleftarrow{C}$ | $\overrightarrow{C}+\overrightarrow{D}+\overleftarrow{D}+1$ | $\overrightarrow{C}+\overleftarrow{C}$ | $\overrightarrow{C}+\overleftarrow{C}$ |
| **f** | $\overrightarrow{C}+\overleftarrow{C}$ | $\overrightarrow{C}+\overleftarrow{C}$ | $\overleftarrow{C}+\overrightarrow{C}+\overrightarrow{D}+\overleftarrow{D}+1$ | $\overrightarrow{C}+\overleftarrow{C}$ | $\overleftarrow{C}+\overrightarrow{C}+\overrightarrow{D}+\overleftarrow{D}+1$ | $\overrightarrow{C}+\overleftarrow{C}$ | $\overleftarrow{C}+\overrightarrow{C}+\overrightarrow{D}+\overleftarrow{D}+1$ |

Table 3.202: Glue matrix for the SUM_WIDTH_ZIGZAG constraint defined as the composition of the ZIGZAG pattern, the feature WIDTH, and the aggregator sum; cells of the glue matrix are coloured with the colour of the constituent to which they are related.

# Appendix A

# Electronic Constraint Catalogue

The file attached to each constraint is available from a link located on the leftmost upper corner of the first page of each constraint provided you are using Adobe Reader.

- The file `eval.pl` containing all utilities is available from this link .

- The file `src.pl` allowing to download all time series constraints in SICStus is available from the link .

- Finally the MiniZinc implementation of the time-series constraints is available from this link . It corresponds to a reformulation of the corresponding automata constraints.

# Index

# A

# D

# F

# I

# L

# M

# N